%% file: neurips_2023.tex
\documentclass{article}

\PassOptionsToPackage{numbers, compress}{natbib}

     \usepackage[final]{neurips_2023}

\usepackage[utf8]{inputenc} 
\usepackage[T1]{fontenc}    
\usepackage{hyperref}       
\usepackage{url}            
\usepackage{booktabs}       
\usepackage{amsfonts}       
\usepackage{nicefrac}       
\usepackage{microtype}      
\usepackage{xcolor}         

\usepackage{amsmath}
\usepackage{amssymb}
\usepackage[capitalise]{cleveref}
\usepackage{todonotes}
\usepackage{float}
\usepackage{graphicx}
\usepackage{bbm}
\usepackage{subcaption}

\usepackage{wasysym}
\usepackage{siunitx}
\usepackage{chemmacros}

\usepackage{pgfplots} 

\usepackage{algorithm}
\usepackage[noend]{algpseudocode}

\usepackage{makecell}

\usepackage{amsthm}
\usepackage{thmtools,thm-restate}

\declaretheorem{proposition}
\declaretheorem{definition}

\newcommand*\act{\bullet}
\newcommand*\cat{\circ}
\newcommand*\compose{\cdot}

\newcommand*\din{d_\mathrm{in}}
\newcommand*\dout{d_\mathrm{out}}
\newcommand*\dlift{\hat{d}_\mathrm{in}}

\newcommand*\dlocal{d_\mathrm{loc}}

\newtheorem{innercustomproposition}{Proposition}
\newenvironment{customproposition}[1]
  {\renewcommand\theinnercustomproposition{#1}\innercustomproposition}
  {\endinnercustomproposition}

\input{math_commands}

\title{
Provable Adversarial Robustness\\for Group Equivariant Tasks:\\Graphs, Point Clouds, Molecules, and More
}

\author{%
  Jan Schuchardt, Yan Scholten, Stephan G\"unnemann
    \\
    \texttt{\{j.schuchardt, y.scholten, s.guennemann\}@tum.de} \\
  Department of Computer Science \& Munich Data Science Institute\\
  Technical University of Munich\\
}

\begin{document}

\include{includeonly_chapters/main}

\clearpage

\setcounter{page}{16}

\include{includeonly_chapters/appendix}


\end{document}

%% file: math_commands.tex
\usepackage{amsmath,amsfonts,bm}

\def\eqref#1{equation~\ref{#1}}

\def\1{\bm{1}}

\newcommand{\indicator}{\mathbbm{1}}

\def\vone{{\bm{1}}}

\def\vrho{{\bm{\rho}}}
\def\vkappa{{\bm{\kappa}}}
\def\va{{\bm{a}}}
\def\vb{{\bm{b}}}

\def\vq{{\bm{q}}}
\def\vr{{\bm{r}}}
\def\vs{{\bm{s}}}
\def\vt{{\bm{t}}}

\def\vv{{\bm{v}}}

\def\vx{{\bm{x}}}
\def\vy{{\bm{y}}}
\def\vz{{\bm{z}}}

\def\evq{{q}}
\def\evr{{r}}
\def\evs{{s}}

\def\evv{{v}}

\def\evx{{x}}
\def\evy{{y}}
\def\evz{{z}}

\def\mZero{{\bm{0}}}
\def\mA{{\bm{A}}}

\def\mC{{\bm{C}}}
\def\mD{{\bm{D}}}

\def\mH{{\bm{H}}}

\def\mL{{\bm{L}}}

\def\mP{{\bm{P}}}
\def\mQ{{\bm{Q}}}
\def\mR{{\bm{R}}}
\def\mS{{\bm{S}}}
\def\mT{{\bm{T}}}
\def\mU{{\bm{U}}}
\def\mV{{\bm{V}}}
\def\mW{{\bm{W}}}
\def\mX{{\bm{X}}}
\def\mY{{\bm{Y}}}
\def\mZ{{\bm{Z}}}

\def\mPhi{{\bm{\Phi}}}
\def\mLambda{{\bm{\Lambda}}}

\def\mOmega{{\bm{\Omega}}}
\def\mPsi{{\bm{\Psi}}}
\def\mDelta{{\bm{\Delta}}}
\def\eye{{\bm{I}}}

\DeclareMathAlphabet{\mathsfit}{\encodingdefault}{\sfdefault}{m}{sl}
\SetMathAlphabet{\mathsfit}{bold}{\encodingdefault}{\sfdefault}{bx}{n}

\def\sA{{\mathbb{A}}}
\def\sB{{\mathbb{B}}}

\def\sD{{\mathbb{D}}}
\def\sF{{\mathbb{F}}}
\def\sG{{\mathbb{G}}}
\def\sH{{\mathbb{H}}}
\def\sI{{\mathbb{I}}}

\def\sL{{\mathbb{L}}}

\def\sN{{\mathbb{N}}}

\def\sR{{\mathbb{R}}}
\def\sS{{\mathbb{S}}}
\def\sT{{\mathbb{T}}}

\def\sV{{\mathbb{V}}}

\def\sX{{\mathbb{X}}}
\def\sY{{\mathbb{Y}}}

\def\emA{{A}}

\def\emC{{C}}
\def\emD{{D}}

\def\emH{{H}}

\def\emL{{L}}

\def\emQ{{Q}}
\def\emR{{R}}
\def\emS{{S}}
\def\emT{{T}}
\def\emU{{U}}
\def\emV{{V}}
\def\emW{{W}}
\def\emX{{X}}

\def\emZ{{Z}}

\def\emPhi{{\Phi}}
\def\emPsi{{\Psi}}
\def\emDelta{{\Delta}}
\def\emOmega{{\Omega}}

\DeclareMathOperator*{\argmax}{arg\,max}
\DeclareMathOperator*{\argmin}{arg\,min}

%% file: includeonly_chapters/main.tex
\maketitle

\input{sections/abstract}

\input{sections/introduction}

\input{sections/related_work}

\input{sections/background}

\input{sections/defining_robustness}

\input{sections/certification}

\input{sections/graph_edit_certificates}

\input{sections/future_work}

\input{sections/experiments}

\input{sections/conclusion}

\input{sections/acknowledgements}

\bibliographystyle{unsrtnat} 
\bibliography{references/references}

%% file: sections/abstract.tex
\begin{abstract}
A machine learning model is traditionally considered robust if its prediction remains (almost) constant under input perturbations with small norm.
However, real-world tasks like molecular property prediction or point cloud segmentation have inherent equivariances,
such as rotation or permutation equivariance.
In such tasks, even perturbations with large norm do not necessarily change an input's semantic content. Furthermore, there are perturbations for which a model's prediction explicitly needs to change.
For the first time, we propose a sound notion of adversarial robustness that accounts for task equivariance.
We then demonstrate that provable robustness can be achieved by (1) choosing a model that matches the task's equivariances (2) certifying traditional adversarial robustness.
Certification methods are, however, unavailable for many models, such as those with continuous equivariances.
We close this gap by developing the framework of equivariance-preserving randomized smoothing, which enables architecture-agnostic certification.
We additionally derive the first architecture-specific graph edit distance certificates, i.e.\ sound robustness guarantees for isomorphism equivariant tasks like node classification.
Overall, a sound notion of robustness is an important prerequisite for future work at the intersection of robust  and geometric machine learning.
\end{abstract}

%% file: sections/introduction.tex
\section{Introduction}\label{section:introduction}

Group equivariance and adversarial robustness are 
two important model properties when applying machine learning to real-world tasks
involving images, graphs, point clouds and other data types:

Group equivariance is an ubiquitous form of task symmetry~\citep{Bronstein2021}.
For instance, we do not know how to optimally classify point clouds, but know that the label is not affected by permutation.
We cannot calculate molecular forces in closed form, but know that the force vectors rotate as the molecule rotates. 
Directly enforcing such equivariances in models is an effective inductive bias, 
as demonstrated by the success of convolutional layers~\cite{Fukushima1979}, transformers~\cite{Vaswani2017} and graph neural networks~\cite{Gori2005}.

Adversarial robustness~\cite{Szegedy2014,Nguyen2015,Goodfellow2015} is a generalized notion of model Lipschitzness:
\textit{A small change to a model's input $\smash{x}$ should only cause a small change to its prediction $\smash{f(x)}$.}
If a model is adversarially robust, then its accuracy will not be negatively affected by sensor noise, measurement errors or other small perturbations that are unavoidable when working with real-world data.

For the first time, we address the following question:
\textit{What is adversarial robustness in tasks that are group equivariant?}
This is necessary because the notions of input and output similarity used in prior work on adversarial robustness (e.g.~$\smash{\ell_p}$ distances) are not suitable for group equivariant tasks.

\begin{figure}
	\centering
	\begin{minipage}{0.48\textwidth}
		\centering
       \includegraphics[width=\linewidth]{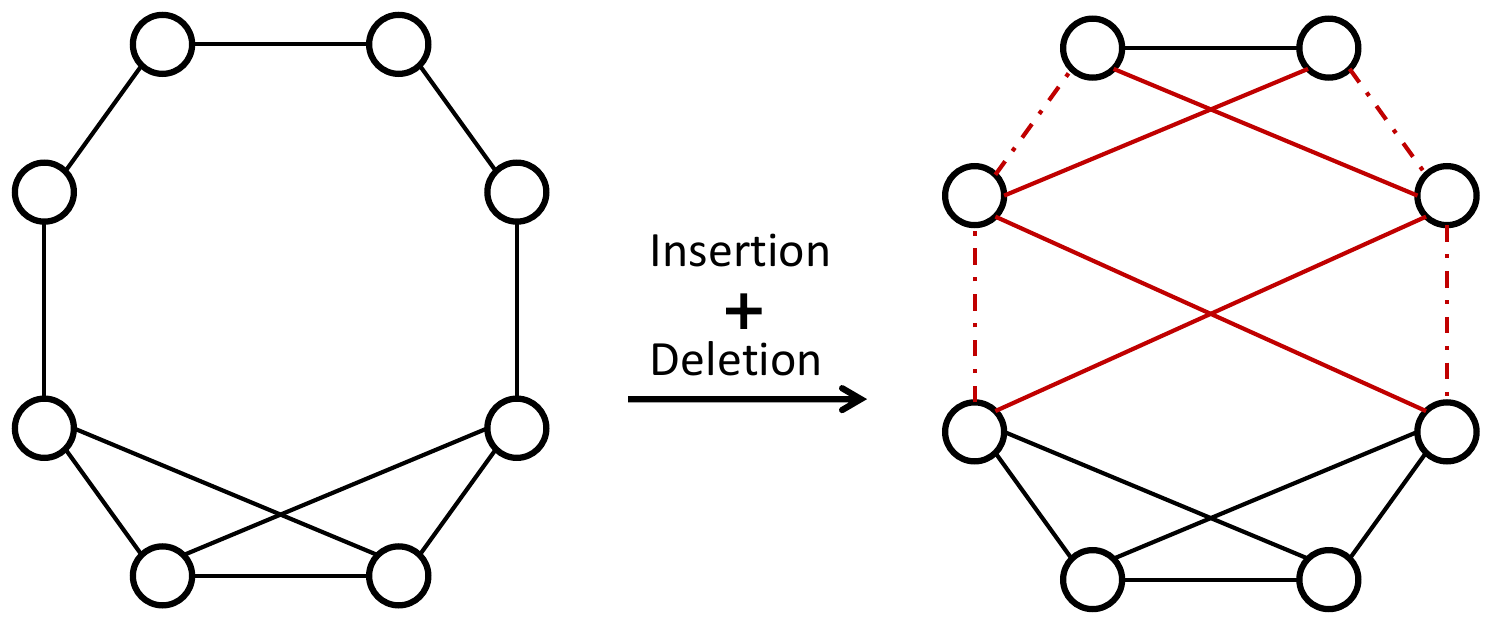}
         \vspace{-0.2cm}
  		\caption{The right graph is constructed by inserting and deleting four edges.
         While their $\ell_0$ distance is large, the graphs are isomorphic and should thus have the same set of node labels.
        \label{fig:intro_example_graph}
		}
	\end{minipage}
	\hfill
	\begin{minipage}{0.48\textwidth}
		\centering
		\includegraphics[width=\linewidth]{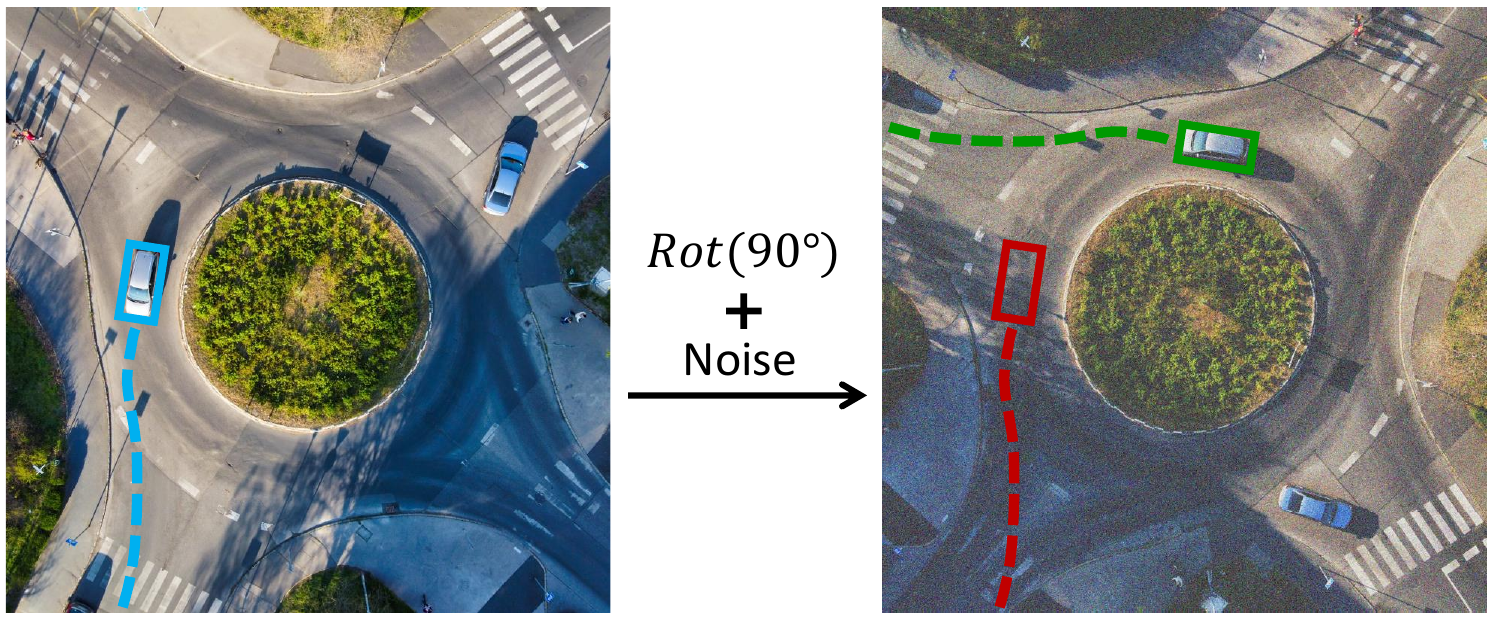}
        \vspace{-0.2cm}
		\caption{The predicted trajectory (blue) should rotate as the image rotates (green), even in the presence of camera noise and other perturbations. It should not remain constant (red).
        \label{fig:intro_example_trajectory}
		}
	\end{minipage}
     \vspace{-0.5cm}
\end{figure}

\cref{fig:intro_example_graph} illustrates why group equivariant tasks require rethinking input similarity.
The right graph is constructed by modifying a large fraction of edges, meaning the perturbation of the adjacency matrix has a large $\smash{\ell_0}$ norm.
Prior work~\cite{Zuegner2018,Zuegner2019meta,Zuegner2019cert,Bojchevski2019,Liu2020graphcert,Zuegner2020,Bojchevski2020,Jin2020,Schuchardt2021,Sun2022} may deem it too large for a node classifiers' prediction to be robust.
However, we know the graphs are isomorphic, meaning they are the same geometric object and should have the same set of labels.
\cref{fig:intro_example_trajectory} illustrates why group equivariant tasks also require rethinking output similarity. 
Prior work considers a prediction robust if it remains (almost) constant.
But when predicting trajectories from drone footage, we know that they should rotate as the drone rotates -- even in the presence of camera noise.
The predictions should explicitly not remain constant.

To address these two issues we propose a sound notion of adversarial robustness for group equivariant tasks that
a) measures input similarity by lifting distance functions to group invariant distance functions and 
b) jointly considers transformations of the input and output space.
This notion of adversarial robustness applies to arbitrary tasks, groups and distance functions.

A natural goal after introducing a novel notion of robustness is developing (provably) robust models.
We show that robustness can be guaranteed by being robust under the traditional notion of adversarial robustness -- if the model's equivariances match the equivariances of the task it is used for. 
Importantly, this implies that existing robustness guarantees may actually hold for significantly larger sets of perturbed inputs.
For instance, proving the robustness of a graph neural network w.r.t.\  $\smash{\ell_0}$ distance is in fact proving robustness w.r.t\ graph edit distance with uniform cost for insertion and deletion.

Although equivariant models make provable robustness more attainable, there are no certification procedures for many architectures.
For instance, there is no prior work on proving robustness for rotation equivariant models.
To close this gap, we develop the framework of equivariance-preserving randomized smoothing. It specifies sufficient conditions under which models retains their  equivariances when undergoing randomized smoothing -- a state-of-the-art approach for simultaneously increasing and proving the robustness of arbitrary models~\cite{Li2019,Lecuyer2019,Cohen2019}.
In addition to that, we generalize the aforementioned graph edit guarantees to arbitrary, user specified costs.
Varying these costs allows for a fine-grained analysis of a model's robustness to graph perturbations.


To summarize, our core contributions are that we
\vspace{-0.15cm}
\begin{itemize}
    \itemsep0pt
    \item propose a sound notion of adversarial robustness for group equivariant tasks,
    \item show that using equivariant models facilitates achieving (provable) robustness,
    \item develop the framework of equivariance-preserving randomized smoothing,
    \item and generalize each existing graph and node classification robustness certificate for $\smash{\ell_0}$ perturbations to graph edit distance perturbations with user-specified costs.\end{itemize}
\vspace{-0.15cm}
Overall, reconsidering what adversarial robustness means in  equivariant tasks is an important prerequisite for future work at the intersection of robust  and geometric machine learning.

%% file: sections/related_work.tex
\section{Related work}\label{section:related_work}

There is little prior work that studies equivariance and adversarial robustness jointly.
The work that exists~\cite{Kamath2021,Tramer2020,Sun2021,Singla2021,Dumont2018,Schuchardt2022,Jin2022,Chuang2022,Chen2023,Xiang2019,Yang2019} only considers group invariant classification (a special case of equivariance) and does not consider how a task's equivariances should influence what we consider robust.

\textbf{Model invariance and robustness.}
Prior work mostly focuses on a trade-off between invariance and robustness in image classification, i.e., 
whether increasing robustness to rotation or translation decreases robustness to $\smash{\ell_p}$ perturbations and vice-versa~\cite{Kamath2021,Tramer2020,Sun2021,Singla2021,Dumont2018}.
\citet{Schuchardt2022} used knowledge about the invariances of point cloud classifiers to prove that they are constant within larger regions than could be shown using previous approaches.

\textbf{Group invariant distances.}
Recently, stability results for graph classifiers under isomorphism invariant optimal transport distances have been derived~\cite{Jin2022,Chuang2022,Chen2023}.
For point cloud classifiers, using  the permutation invariant Chamfer or Hausdorff distance to craft attacks has been proposed~\cite{Xiang2019,Yang2019}.
These works only focus on invariance and specific domains and do not consider that distances should be task-dependent:
A rotation invariant distance for images may be desirable when segmenting cell nuclei, but not when classifying hand-written digits, since it would fail to distinguish $\smash{6}$ and $\smash{9}$.

\textbf{String edit distance.}
In concurrent work, \citet{Huang2023} use randomized smoothing to prove robustness of classifiers w.r.t.\ string edit distance, i.e., the number of substitutions that are needed to convert one string from alphabet $\Sigma \cup \{\perp\}$ into another, up to insertion of alignment tokens $\perp$.
Their work further emphasizes the need for invariant distance functions in domains with symmetries, and the usefulness of randomized smoothing for proving robustness w.r.t.\ such distances.

\textbf{Robustness of models with equivariances.} 
Aside from work that studies invariance and adversarial robustness jointly, 
there is a rich literature investigating the robustness of models that happen to have equivariances.
This includes convolutions~\cite{Szegedy2014,Nguyen2015,Goodfellow2015,Akhtar2018}, transformers\cite{Hsieh2019,Shi2020,Bonaert2021,Mahmood2021}, point cloud models~\cite{Xiang2019,Yang2019,Zhou2020,Wen2020,Huang2020,Liu2020,Lorenz2021,Liu2021,Denipitiyage2021,Kim2021,Chu2022} and graph neural networks~\cite{Zuegner2018,Zuegner2019meta,Zuegner2019cert,Bojchevski2019,Liu2020graphcert,Zuegner2020,Bojchevski2020,Jin2020,Schuchardt2021,Sun2022,Geisler2020,Geisler2021,Scholten2022,Geisler2022defenses,Schuchardt2023,Gosch2023,Campi2023,Gosch2023training,Scholten2023}.
The models are however treated as a series of matrix multiplications and nonlinearities,
without accounting for their equivariances or the equivariances of the tasks they are used for.
Nevertheless, many methods can actually be reused for proving (non-)robustness under our proposed notion of adversarial robustness (see~\cref{section:certification}).

\textbf{Transformation-specific robustness.} 
A subfield of robust machine learning focuses on robustness to unnoticeable parametric transformations (e.g.\ small rotations)~\cite{Lorenz2021,Chu2022,Kanbak2018,Engstrom2019,Balunovic2019,Zhao2020iso,Fischer2020,Ruoss2021,Mohapatra2021,Li2021,Alfarra2022,Murarev2022}.
These works implicitly assume that large transformations lead to easily identifiable out-of-distribution samples. This is not the case with equivariant tasks:
For instance, a molecule rotated by $\smash{180^\circ}$ is still the same geometric object.
Furthermore, they do not consider unstructured perturbations.
Nevertheless, transformation-specific robustness can be framed as a special case of our proposed notion (see~\cref{appendix:transformation_specific}).

\textbf{Semantics-aware robustness.}
Our work is closely related to different proposals to include ground truth labels in the definition of adversarial robustness
~\cite{Tramer2020,Gosch2023,Jacobsen2018,Jacobsen2019,Gilmer2018,Suggala2019,Chowdhury2022,Kollovieh2023}.
A problem is that the ground truth is usually unknown, which limits experimental evaluation to simple data generating distributions~\cite{Gosch2023,Chowdhury2022} or using human study participants~\cite{Tramer2020,Jacobsen2019,Kollovieh2023}.
\citet{Geisler2022} overcome this problem in the context of neural combinatorial optimization by using adversarial perturbations that are known to change the ground truth of a decision problem.
Group equivariant tasks admit a similar approach, since we know how the ground truth changes for specific input transformations.

%% file: sections/background.tex
\section{Background}\label{section:background}

\textbf{Group theory and equivariance.}
Discussing group equivariance requires  a few algebraic concepts.
A group is a set $\smash{\sG}$ with  identity element $\smash{e}$ and associative operator $\smash{\cdot : \sG \times \sG \rightarrow \sG}$ such that $\smash{\forall g \in \sG: e\compose g = g\compose e = g}$ and each $\smash{g \in \sG}$ has an inverse element $\smash{g^{-1}}$ with $\smash{g^{-1}\compose g = g\compose g^{-1} = e}$.
We are interested in transformations with a group structure, such as rotations.
Given set $\smash{\sX}$, a (left) group action is a function $\smash{\act_\sX : \sG \times \sX \rightarrow \sX}$ that transforms elements of $\smash{\sX}$ and preserves the structure of $\smash{\sG}$, i.e.\ $\smash{g \act_\sX h \act_\sX x = (g\compose h) \act_\sX x}$. For instance, rotation by $\smash{\phi^{\circ}}$ and then by $\smash{\psi^{\circ}}$  is the same as rotation by $\smash{(\psi + \phi)^{\circ}}$.
A group may act differently on different sets.
For example, rotation group $\mathit{SO}(3)$ may act on point clouds via matrix multipliciation while acting on class labels via the identity function.
When clear from context, we drop the subscripts.
A function $\smash{f : \sX \rightarrow \sY}$ is equivariant if each action on its input is equivalent to an action on its output, i.e.\ $\smash{\forall x \in \sX,  g \in \sG : f(g \act_\sX x) = g \act_\sY f (x)}$. 

\textbf{Adversarial robustness}\label{section:background_robustness}
means that any small change to clean input $x$ only causes a small change to prediction $f(x)$, i.e.\ 
$\max_{x' \in \sB_x} \dout(f(x), f(x')) \leq \delta$
with
$\sB_x = \{x' \mid \din(x,x') \leq \epsilon\}$, 
 and $\smash{\din}$ and $\smash{\dout}$ quantifying input and output distance.
We refer to the set of admissible perturbed inputs $\sB_x$ as the \textit{perturbation model.}
A common special case is $\dout(y, y') = \indicator[y \neq y']$ and $\delta=0$~\cite{Szegedy2014,Nguyen2015,Goodfellow2015}.
Other forms of robustness involve training data~\cite{Wang2022,Tian2022}, but we focus on test-time perturbations.

\label{section:background_randomized_smoothing}
\textbf{Randomized smoothing}~\cite{Li2019,Lecuyer2019,Cohen2019}  
is a paradigm for increasing and proving the robustness of models in an architecture-agnostic manner.
It works by randomizing the inputs of a \textit{base model} $\smash{h}$ to construct a more robust \textit{smoothed model} $\smash{f}$.
While originally proposed for provably robust image classification, it has evolved into a much more general framework that can be applied to various domains and tasks~\cite{Bojchevski2020,Cohen2019,Schuchardt2023,Kumar2020,Pautov2022,Chiang2020,Wu2021,Fischer2021,Sukenik2022,Wollschlaeger2023}.
Consider a measurable input space $\smash{(\sX, \sD)}$ and a measurable base model $\smash{h : \sX \rightarrow \sV}$ that maps to a measurable intermediate space $\smash{(\sV, \sF)}$ (e.g.\ logits).
Given the base model $h$ and a family of probability measures $\smash{(\mu_x)_{x \in \sX}}$ on $\smash{(\sX, \sD)}$ indexed by $\smash{\sX}$ (e.g.\ Gaussians with mean $x$), we can define the input-dependent pushforward measure $\mu_x \cat h^{-1}$ for any input $x \in \sX$.\footnote{Note that $h^{-1}$ is not the inverse but the pre-image. The base model need not be invertible.}
We can further define a \textit{smoothing scheme} $\smash{\xi : \Delta(\sV, \sF) \rightarrow \sY}$ that maps from probability measures $\smash{\Delta(\sV, \sF)}$ on the intermediate space to an output space $\smash{\sY}$ (e.g.\ logit distributions to labels).\footnote{$\sV$ and $\sY$ can also be the same set, as is the case with majority voting for classifiers.}
This lets us construct the smoothed model $\smash{f(x) = \xi( \mu_x \cat h^{-1})}$, which makes a prediction for input $\smash{x \in \sX}$ based on some quantity of the input-dependent pushforward measure $\smash{\mu_x \cat h^{-1}}$ (e.g.\ the expected value).
Intuitively, if two inputs $x$ and $x'$ are sufficiently similar, then the smoothing measures $\mu_x$ and $\mu_x'$ will have a large overlap, thus leading to similar smoothed predictions.
The combination of measures $(\mu_x)_{x \in \sX}$ and smoothing scheme $\xi$ determines for which $\din$ and $\dout$ robustness is guaranteed.
We discuss these combinations in more detail in~\cref{appendix:combing_measures_schemes}.

%% file: sections/defining_robustness.tex
\section{Redefining robustness for group equivariant tasks}\label{section:defining_robustness}

For the first time, we seek to define adversarial robustness for group equivariant tasks.
By task we mean an unknown function $\smash{y : \sX \rightarrow \sY}$ that is equivariant with respect to the action of a group $\sG$.
We assume that it is approximated by a (learned) model $\smash{f : \sX \rightarrow \sY}$ whose robustness we want to determine.
The model does not have to be equivariant.
Like traditional adversarial robustness, we assume there are functions $\smash{\din : \sX \times \sX \rightarrow \sR_+}$ and $\smash{\dout : \sY \times \sY \rightarrow \sR_+}$, 
which define what constitutes a small change in domain $\sX$ and co-domain $\sY$, if we are not concerned with group symmetries. For instance, $\ell_2$ distance is a natural notion of similarity between Euclidean coordinates.

For ease of exposition, we further assume that all considered optimization domains are compact (so that minima and maxima exist) and that group $\sG$ acts isometrically on $\sX$, i.e.\ $\forall x, x' \in \sX, \forall g \in \sG: \din(g \act x, g \act x') = \din(x, x')$. This covers most practically relevant use cases.
For completeness, we discuss non-compact sets and non-isometric actions in~\cref{appendix:non_compact,appendix:non_isometric}.

\subsection{Perturbation model for group equivariant tasks}\label{section:input_distance}
Our first goal is to resolve the shortcomings of using typical input distance functions $\din$ to define what constitutes small input perturbations in group equivariant tasks.
We seek some improved function $\dlift$ that accounts for the equivariance of $y$ and simultaneously captures the similarity of objects in domain $\sX$, as defined by original distance function $\din$.
To this end, we define three desiderata.

\textbf{Desideratum 1. }
We know that any perturbed $\smash{x' \in \sX}$ and $\smash{g \act x'}$ with $\smash{g \in \sG}$ are up to symmetry the same geometric object with the same semantic content, i.e.\ $\smash{y(g \act x') = g \act y(x')}$.
But as illustrated in~\cref{fig:intro_example_graph}, group actions may cause a drastic change w.r.t.\ distance $\din$.
Thus, even if $f(x)$ is ``robust'' within $\smash{\sB_x = \{x' \mid \din(x,x') \leq \epsilon\}}$ for some $\epsilon$, there may still be a $\smash{x' \in \sB_x}$ and $\smash{g \act x' \notin \sB_x}$ that lead to two completely different predictions.
If a prediction can be altered without changing the semantic content of an input, it can hardly be considered  robust.
This problem cannot be resolved by requiring robustness for very large $\epsilon$ so that $\sB_x$ covers all symmetric objects. 
Doing so would also include objects with substantially different semantic content from clean input $x$, which should actually lead to different predictions.
Instead, we need a $\smash{\dlift}$ that is constant for all symmetric forms of $x'$, so that robustness to one implies robustness to all: 
$\smash{\forall x,x' \in \sX, g \in \sG : \dlift(x, g \act x') = \dlift(x, x')}$.

\textbf{Desideratum 2. }
While the first desideratum accounts for equivariance, a model should also be robust to sensor noise, measurement errors and other small perturbations that are not necessarily group actions.
Therefore, elements that are close with respect to the original distance function $\din$ should remain close under our new distance function $\dlift$, i.e.\ $\smash{\forall x, x': \dlift(x,x') \leq \din(x,x')}$.

\textbf{Desideratum 3. }
The first two desiderata could be fulfilled by $\smash{\dlift(x,x') = 0}$ or a function that arbitrarily changes the  ordering of inputs w.r.t.\ distance.
To appropriately capture similarity, $\dlift$ should not only underapproximate $\din$, but preserve it as best as possible. Let $\sD$ be the set of functions from $\smash{\sX \times \sX}$ to $\sR_+$ that fulfill the first two desiderata.
We 
require that $\smash{\dlift(x, x') = \max_{\gamma \in \sD} \gamma(x, x')}$.
\begin{restatable}[]{proposition}{induceddistance}
A function $\smash{\dlift : \sX \times \sX \rightarrow \sR_+}$ that fulfills all three desiderata for any original distance function  $\smash{\din : \sX \times \sX \rightarrow \sR_+}$ exists and is uniquely defined: $\smash{\dlift(x,x') = \min_{g \in \sG} \din(x, g \act x')}$.
\end{restatable}
We prove this result in~\cref{appendix:action_induced_distance}. 
We refer to $\dlift$ as the \textit{action-induced distance}.
It is the distance after optimally aligning the perturbed input $x'$ with the clean input $x$ via a group action.

\textbf{Example: Point cloud registration distance. }
Let $\smash{\sX = \sR^{N \times D}}$. 
Consider $\smash{\sG = \mathit{S}_N \times \mathit{SE}(D)}$, the set of permutation matrices, rotation matrices and translation vectors. Let $\sG$ act via
$\smash(\mP, \mR, \vt) \act \mX = \mP \left( \mX \mR^T + \1_N \vt^T \right)$
and let $\din$ be the Frobenius distance.
Then $\dlift$ is the \textit{registration distance}~\cite{Huang2021}, i.e.\ the distance after finding an optimal correspondence and applying an optimal rigid transformation.

\textbf{Example: Graph edit distance. }
Let $\sX$ be the set of all adjacency matrices $\{0,1\}^{N \times N}$ with $\smash{\din(\mA, \mA') = ||\mA - \mA'||_0}$.
Consider $\smash{\sG =\mathit{S}_N}$, the set of permutation matrices.
Let $\sG$ act via $\smash{\mP \act \mA = \mP \mA \mP^T}$. Then $\smash{\dlift(\mA, \mA')}$ is the \textit{graph edit distance} with uniform cost~\cite{Almohamad1993,Justice2006,Lerouge2017}. That is, the number edges that have to be inserted or deleted to transform $\mA$ into a graph isomorphic to $\mA'$.

As demonstrated by these examples, we do not claim to have invented this notion of distance for equivariant domains.
Our contribution is justifying, in a principled manner, why this notion of distance should be used to define adversarial robustness for group equivariant tasks.

\textbf{Perturbation model.} 
Now that we have an appropriate input distance for group equivariant tasks, we can use it to define the set of admissible perturbed inputs $\sB_x$ as $\smash{\{x' \mid \min_{g \in \sG} \din(x, g \act x')
 \leq \epsilon\}}$ with some small $\epsilon \in \sR_+$.
Because every $g \in \sG$ has an inverse element $g^{-1}$, this set is identical to $\smash{\{g \act x' \mid g \in \sG,  \din(x,x') \leq \epsilon\}}$.
We use this equivalent representation because it lets us disentangle group actions from other perturbations. 
That is, $f(x)$ should be robust to inputs that can be generated via a small perturbation w.r.t.\ the original distance $\din$ followed by a group action.


\subsection{Output distance for group equivariant tasks}\label{section:output_distance}
\cref{fig:intro_example_trajectory} illustrates that we also need to reconsider what a small change to a model's prediction is.
Letting a group element act on a prediction may cause a large change w.r.t.\ $\dout$.
In particular, we may have $\smash{\dout(y(x), y(g \act x)) = \dout(y(x), g \act y(x)) \gg 0}$, even though $x$ and $g \act x$ have a distance of zero w.r.t.\ the action-induced distance.
Thus, we would consider even the ground truth $y$ itself to be non-robust.
Using action-induced distances, i.e.\ $\smash{\min_{g \in \sG} \dout(y, g \act y')}$, is not appropriate either. Action-induced distances cannot distinguish between a model that transforms its predictions in compliance with the ground truth and one that applies  arbitrary group actions.

To appropriately measure output distance, we need to account for differences between clean prediction $f(x)$ and perturbed prediction $\smash{f(g \act x')}$ that are caused by the specific group element $g \in \sG$ acting on the model's input.
To do so, we need to first revert the effect of the group action before comparing the predictions.
That is, we need to measure output distance using $\smash{\dout(f(x), g^{-1} \act f(g \act x'))}$.

\subsection{Proposed definition of robustness for group equivariant tasks}
Combining the perturbation model from~\cref{section:input_distance} with the output distance from~\cref{section:output_distance} leads to the following definition of adversarial robustness for group equivariant tasks:
\begin{definition}\label{definition:equivariant_robustness}
Assume that ground truth function $y : \sX \rightarrow \sY$ is equivariant with respect to the action of group $\sG$.
Then, a prediction $f(x)$ for clean input $x \in \sX$ is $(\sG, \din, \dout, \epsilon, \delta)$-equivariant-robust if
\begin{equation}\label{eq:equivariant_robustness}
(
    \max_{x' \in \sX} \max_{g \in \sG} \dout(f(x), g^{-1} \act f(g \act x')) \text{ s.t. }  \din(x,  x') \leq \epsilon
)
\leq \delta.
\end{equation}\end{definition}
Simply speaking, prediction $f(x)$ should be considered robust if it is robust to unnoticeable perturbations and is approximately equivariant around $x$.
For perturbations of size $\epsilon$ w.r.t.\ the original $\din$, the prediction should not change by more than $\delta$ and no group action should cause this error to increase beyond $\delta$.
Note that~\cref{definition:equivariant_robustness} depends on the equivariances of the task $y$, not those of the model $f$.
Further note that the constraint involves original distance $\din$, not action-induced distance $\dlift$.

\textbf{Special cases.}
If the task is not equivariant, i.e.\ $\sG = \{e\}$, we recover the traditional notion of  adversarial robustness because $e$ acts via identity.
For $\epsilon = 0$, constant $\delta$ is a bound on the equivariance error, a common evaluation metric in geometric machine learning (see, e.g.~\cite{Zhang2019,Fuchs2020,Sosnovik2020,Bouchacourt2021,Gruver2023}).

\textbf{Other invariant distances.} 
As discussed in~\cref{section:related_work}, alternative invariant distances 
have been proposed for certain group invariant classification tasks.
This includes Hausdorff and Chamfer distance~\cite{Xiang2019,Yang2019}, as well as optimal transport for graphs~\cite{Jin2022,Chuang2022,Chen2023}.
Any group invariant function $\din$ is preserved by the action-induced distance:
$\min_{g \in \sG} \din( x,  g \act x') = \min_{g \in \sG} \din(x,  x') = \din(x,  x')$.
Thus, all previous results for invariant classification are compatible with our notion of robustness.

\textbf{Local budgets and local robustness.}
For data that is composed of $N$ distinct elements, such as point clouds, one may want local budgets $\epsilon_1,\dots, \epsilon_N$.
For tasks that involve $M$ distinct predictions, such as segmentation, one may only be interested in the robustness of some subset of predictions.
\cref{definition:equivariant_robustness} can be extended to accommodate local budgets and robustness, see~\cref{appendix:local_robustness}.

%% file: sections/certification.tex
\section{Achieving provable robustness for group equivariant tasks}\label{section:certification}

Now we have a sound notion of robustness that overcomes the limitations discussed in~\cref{section:introduction}.
But it is not clear how to achieve provable robustness.
Given a task $y : \sX \rightarrow \sY$ that is equivariant with respect to the action of a group $\sG$, we 
 want a model $f : \sX \rightarrow \sY$ and a corresponding algorithm that can verify the $\smash{(\sG, \din, \dout, \epsilon, \delta)}$-equivariant-robustness of a prediction $f(x)$.

A challenge is that the action-induced distance $\dlift$, which defines our perturbation model and thus underlies the optimization domain in~\cref{eq:equivariant_robustness}, is generally not tractable. For example, deciding whether the graph edit distance is smaller than some $\epsilon \in \sR$ is NP-hard~\cite{Zeng2009}.
A solution would be  to relax the optimization domain in order to pessimistically bound the model's actual robustness.
This is, in essence, the approach taken by works on robustness to graph optimal transport perturbation models (e.g.~\cite{Jin2022,Chuang2022}).
Instead of optimizing over discrete correspondences between objects, they optimize over couplings that define soft correspondences.
There is however a more straight-forward solution that lets us take advantage of years of research in geometric and robust machine learning: Applying the principles of geometric machine learning and using a model with matching equivariances.
\begin{proposition}\label{proposition:certification_reduction}
    Consider a model $f : \sX \rightarrow \sY$ that is equivariant with respect to the action of a group $\sG$.
    Any prediction $f(x)$ is 
    $\smash{(\sG, \din, \dout, \epsilon, \delta)}$-equivariant-robust if and only if it is $\smash{(\{e\}, \din, \dout, \epsilon, \delta)}$-equivariant-robust, i.e.\ fulfills traditional adversarial robustness.
\end{proposition}
\vspace{-0.4cm}
\begin{proof}
    Because $f$ is equivariant, we have $\forall g :  g^{-1} \act f(g \act x') = g^{-1} \act g \act f( x') = f( x') = e^{-1} \act f ( e \act x')$ and thus ${\dout(f(x), g^{-1} \act f(g \act x')) = \dout(f(x),  e^{-1} \act f(e \act x'))}$.
\end{proof}
\vspace{-0.2cm}
By using model $f$ with the same equivariances as ground truth $y$, we reduce the problem of proving robustness to that of proving traditional adversarial robustness i.e.\ $\smash{\sG = \{e\}}$.
Again, note that robustness does not mean that a prediction remains constant, but that it transforms in compliance with semantics-preserving transformations of the input, even under small perturbations (see~\cref{eq:equivariant_robustness}).

\textbf{Discussion.}
\cref{proposition:certification_reduction} provides another strong argument for the use of geometric machine learning models.
These models facilitate the problem of achieving provable robustness for group equivariant tasks --- 
without relaxing the action-induced distance and thus weakening our guarantees.
We can instead build upon existing robustness certification procedures for equivariant models, such as transformers, PointNet and graph convolutional networks,
under traditional perturbation models, like $\ell_2$ or $\ell_0$ perturbations~\cite{Zuegner2018,Zuegner2019meta,Zuegner2019cert,Bojchevski2019,Liu2020graphcert,Zuegner2020,Bojchevski2020,Jin2020,Schuchardt2021,Sun2022,Hsieh2019,Shi2020,Bonaert2021,Mahmood2021,Zhou2020,Wen2020,Huang2020,Liu2020,Lorenz2021,Liu2021,Denipitiyage2021,Kim2021,Chu2022}.
We just need to use our knowledge about the equivariances of tasks and models to reinterpret what is actually certified by these procedures.

\textbf{Relation to orbit-based certificates.}
A related result was discussed for group \textit{invariant} point cloud classifiers in~\cite{Schuchardt2022}:
If $\sX = \sR^{N \times D}$ and classifier $f$ is constant within Frobenius norm ball $\sB$ and invariant w.r.t.\ the action of group $\sG$, then it is also constant within $\smash{\{g \act \mX'  \mid  \mX' \in \sB, g \in \sG\}}$.
However, this work did not discuss  whether being constant within this set is desirable, how it relates to task invariance and what this result tells us about the adversarial robustness of the classifier.

\textbf{Adversarial attacks.}
While we focus on provable robustness, \cref{proposition:certification_reduction} also applies to adversarial attacks, i.e.\ proving non-robustness via counterexample.
Even when the equivariances of $y$ and $f$ do not match,  traditional  attacks are feasible solutions to~\cref{eq:equivariant_robustness} since they amount to constraining $g$ to set $\{e\}$.
For completeness, we perform experiments with adversarial attacks in~\cref{appendix:extra_experiments_attacks}.

\subsection{Equivariance-preserving randomized smoothing}\label{section:equivariance_smoothing}
Equivariant models reduce the problem of proving robustness for group equivariant tasks to that of proving traditional robustness.
However, specialized procedures to make these proofs are only available for a limited range of architectures.
For example, specialized certification procedures for point cloud models are limited to the PointNet architecture~\cite{Lorenz2021,Mu2022}, and 
there are no procedures to prove the robustness of models with continuous equivariances, such as rotation equivariance.

For such models, one could try to apply randomized smoothing (recall~\cref{section:background_randomized_smoothing}).
By choosing a suitable smoothing scheme $\xi$ and measures $(\mu_x)_{x \in \sX}$ for distances $\din$ and $\dout$, one can transform any base model $h$ into smoothed model $f$ and prove its $\smash{(\{e\}, \din, \dout, \epsilon, \delta)}$-equivariant-robustness.
However, base model $h$ having the same equivariances as task $y$ does not guarantee that smoothed model $f$ has the same equivariances. Thus, one can generally not use~\cref{proposition:certification_reduction} to prove $\smash{(\sG, \din, \dout, \epsilon, \delta)}$-equivariant-robustness.
We propose the following sufficient condition for verifying that a specific smoothing scheme and family of measures are equivariance-preserving (proof in~\cref{appendix:proof_equivariance_smoothing}).
\begin{proposition}\label{proposition:equivariance_preservation}
    Assume two measurable spaces $\smash{(\sX, \sD)}$, $\smash{(\sV, \sF)}$, an output space $\sY$ and a measurable base model $\smash{h : \sX \rightarrow \sV}$ that is equivariant with respect to the action of group $\sG$.
    Further assume that $\sG$ acts on $\sX$ and $\sV$ via measurable functions.
    Let $\smash{\xi : \Delta(\sV, \sF) \rightarrow \sY}$ be a smoothing scheme that maps from the set of probability measures $\Delta(\sV, \sF)$ on intermediate space $\smash{(\sV, \sF)}$ to the output space.
    Define $T_{\sX, g}(\cdot)$ to be the group action on set $\sX$ for a fixed $g$, i.e.\ $\smash{T_{\sX, g}(x) = g \act_\sX x}$. 
    Then, the smoothed model $f(x) = \xi(\mu_x \cat h^{-1})$ is equivariant with respect to the action of group $\sG$ if both
    \vspace{-0.2cm}
    \begin{itemize}
        \itemsep0pt
        \item the family of measures $\smash{(\mu_x)_{x \in \sX}}$ is equivariant, i.e.\ $\smash{\forall x \in \sX, g \in \sG: \mu_{g \act x} = \mu_x \cat T_{\sX,g}^{-1}}$, 
        \item and smoothing scheme $\xi$ is equivariant, i.e.\ $\smash{\forall \nu \in \Delta(\sV, \sF), g \in \sG : \xi (\nu \cat T_{\sV, g}^{-1} ) = g \act \xi(\nu)}$.
    \end{itemize}
    \vspace{-0.2cm}
\end{proposition}
Note that $\cat$ is a composition of functions, whereas $\act$ is a group action. The intuition behind~\cref{proposition:equivariance_preservation} is that, if family of measures $\smash{(\mu_x)_{x \in \sX}}$, and base model $h$, and  smoothing scheme $\xi$ are equivariant, then we have a chain of equivariant functions which is overall equivariant.
We provide a visual example of such an equivariant chain of functions in~\cref{fig:explainy_smoothing}.

In the following, we show that various schemes and measures preserve practically relevant equivariances and can thus be used in conjunction with~\cref{proposition:certification_reduction} to prove $(\sG, \din, \dout, \epsilon, \delta)$-equivariant-robustness.
For the sake of readability, we provide a high-level discussion, leaving the formal propositions  in~\cref{appendix:proof_scheme_measure_equivariances}.
We summarize our results in~\cref{table:measures,table:schemes}. In~\cref{appendix:combing_measures_schemes}, we discuss how to derive robustness guarantees for arbitrary combinations of these schemes and measures.

\textbf{Componentwise smoothing schemes.} 
The most common type of smoothing scheme can be applied whenever the intermediate and output space have $M$ distinct components,
i.e.\ $\sV = \sA^M$ and $\sY = \sB^M$ for some $\sA, \sB$. It smooths each of the $M$ base model outputs independently.
This includes majority voting~\cite{Cohen2019}, expected value smoothing~\cite{Kumar2020,Pautov2022} and median smoothing~\cite{Chiang2020},
which can be used for tasks like classification, segmentation, node classification, regression, uncertainty estimation and object detection~\cite{Bojchevski2020,Cohen2019,Schuchardt2023,Kumar2020,Pautov2022,Chiang2020,Wu2021,Fischer2021,Sukenik2022,Wollschlaeger2023}.
Such schemes preserve equivariance to groups acting on $\sV$ and $\sY$ via permutation.
Note that $\sG$ need not be the symmetric group $\mathit{S}_N$ to act via permutation.
For instance, the identity function is a permutation, meaning componentwise smoothing schemes can be used 
to prove robustness for arbitrary group invariant tasks. See~\cref{proposition:componentwise_schemes}.

\textbf{Expected value smoothing scheme.}
When the intermediate and output space are real-valued, i.e.\ $\sV = \sY = \sR^M$, one can make predictions via the expected value~\cite{Kumar2020,Pautov2022}. Due to linearity of expectation, this scheme does not only preserve permutation equivariance but also equivariance to affine transformations. 
However, certifying the smoothed predictions requires that the support of the output distribution is bounded by a hyperrectangle, i.e.
$\mathrm{supp}(\mu_x \cat h^{-1}) \subseteq \{\vy \in \sR^M \mid \forall m: a_m \leq \evy_m \leq b_m\}$ for some $\va,\vb \in \sR^M$~\cite{Kumar2020,Pautov2022}.
See~\cref{proposition:expected_value_scheme}.

\textbf{Median smoothing scheme.}
When the support of the model's output distribution is real-valued and potentially unbounded, one can make smoothed predictions via the elementwise median~\cite{Chiang2020}.
This scheme does not only preserve permutation equivariance but also equivariance to elementwise linear transformations, such as scaling.
See~\cref{proposition:median_scheme}.

\textbf{Center smoothing scheme.}
Center smoothing~\cite{Kumar2021} is a flexible scheme that can be used whenever $\dout$ fulfills a relaxed triangle inequality.
It predicts the center of the smallest $\dout$ ball with measure of at least $\frac{1}{2}$.
Center smoothing has been applied to challenging tasks like image reconstruction, dimensionality reduction, object detection and image segmentation~\cite{Kumar2021,Schuchardt2023}.
We prove that center smoothing is equivariant to any group acting isometrically w.r.t.\ $\dout$. 
For example, when $\sV = \sY = \sR^{M \times D}$ and $\dout$ is the Frobenius norm, center smoothing guarantees 
$\smash{(\sG, \din, \dout, \epsilon, \delta)}$-equivariant-robustness for any group acting on $\smash{\sR^{N \times D}}$ via permutation, rotation, translation or reflection. See~\cref{proposition:center_scheme}.

\textbf{Product measures.}
Many randomized smoothing methods use product measures.
That is, they use independent noise to randomize elements of an $N$-dimensional input space $\sX = \sA^N$.
The most popular example are exponential family distributions, such as Gaussian, uniform and Laplacian noise, which can be used to prove robustness for various $\ell_p$ distances~\cite{Lecuyer2019,Cohen2019,Lee2019,Yang2020}, Mahalanobis distance~\cite{Fischer2020,Eiras2021} and Wasserstein distance~\cite{Levine2020}.
Product measures preserve equivariance to groups acting via permutation.
Again, group $\sG$ need not be symmetric group $\mathit{S}_n$ to act via permutation.
For instance, rotating a square image by $90^\circ$ (see~\cref{fig:intro_example_trajectory}) is a permutation of its pixels.
See~\cref{proposition:independent_noise}.

\textbf{Isotropic Gaussian measures. }
Isotropic Gaussian measures are particularly useful for $\smash{\sX = \sR^{N \times D}}$. They guarantee robustness when $\din$ is the Frobenius distance~\cite{Liu2021,Chu2022} and preserve equivariance to isometries, i.e.\ permutation, rotation, translation and reflection. Combined with~\cref{proposition:certification_reduction}, this guarantees robustness w.r.t.\ the aforementioned point cloud registration distance. 
See~\cref{proposition:gaussian_noise}.

\textbf{Transformation-specific measures.}
A standard tool for proving transformation-specific  robustness (see~\cref{section:related_work}) is transformation-specific smoothing~\cite{Chu2022,Fischer2020,Li2021,Alfarra2022,Murarev2022}, i.e.\ applying randomly sampled transformations from a parametric family $\smash{(\psi_\theta)_{\theta \in \Theta}}$ with $\psi_\theta : \sX \rightarrow \sX$ to the inputs of a model.
If all $\psi_\theta$ with $\theta \in \Theta$ are equivariant to the actions of group $\sG$, then transformation-specific smoothing preserves this equivariance. 
For example, random scaling~\cite{Murarev2022} preserves rotation equivariance and additive noise (e.g.\ Gaussian) preserves translation equivariance for $\sX = \sR^{N \times D}$.
See~\cref{proposition:transformation_noise}.

\textbf{Sparsity-aware measures.}
Sparsity-aware noise~\cite{Bojchevski2020} can be applied to discrete graph-structured data to guarantee robustness to edge and attribute insertions and deletions.
We prove that sparsity-aware measures preserve equivariance to groups acting via graph isomorphisms.
As we discuss in the next section, this guarantees robustness w.r.t.\  the graph edit distance. See~\cref{proposition:sparsity_noise}.

%% file: sections/graph_edit_certificates.tex
\subsection{Deterministic edit distance certificates for graph and node classification}\label{section:graph_edit_certificates}
Besides sparsity-aware smoothing, there are also deterministic procedures for proving robustness for specific graph neural networks, namely graph convolutional networks~\cite{Kipf2017} and APPNP~\cite{Klicpera2019}.
They are however limited to uniform costs.
To be more specific, let $\sX$ be the set of all graphs $\{0,1\}^{N \times D} \times \{0,1\}^{N \times N}$ with $N$ nodes and $D$ binary  attributes and let $(\mA)_{+} = \max(\mA, 0)$ and $(\mA)_- = \min(\mA, 0)$ with elementwise maximum and minimum. Define $\din((\mX,\mA), (\mX', \mA'))$ as
\begin{equation}\label{eq:ged_base_distance}
    c_\mX^+ \cdot ||(\mX' - \mX)_{+}||_0 +  c_\mX^- \cdot ||(\mX' - \mX)_{-}||_0
    + c_\mA^+ \cdot ||(\mA' - \mA)_{+}||_0 +  c_\mA^- \cdot ||(\mA' - \mA)_{-}||_0, 
\end{equation}
with costs $c_\mX^+, c_\mX^-, c_\mA^+, c_\mA^-$ for insertion and deletion of attributes and edges.
Prior work can only prove robustness for costs in $\{\infty, 1\}$, i.e.\ disallow certain types of perturbations and use uniform cost for the remaining ones.
In~\cref{appendix:graph_edit_distance} we generalize each existing deterministic graph and node classification guarantee to non-uniform costs. This includes procedures based on 
convex outer adversarial polytopes~\cite{Zuegner2019cert}, policy iteration~\cite{Bojchevski2019}, interval bound propagation~\cite{Liu2020graphcert}, bilinear programming~\cite{Zuegner2020}, and simultaneous linearization and dualization~\cite{Jin2020}.
Our proofs mostly require solving different knapsack problems with local constraints and only two distinct costs -- which can be done efficiently via dynamic programming or linear relaxations with analytic solutions (see~\cref{appendix:knapsack}).
As per~\cref{proposition:certification_reduction}, proving $\smash{(\{e\}, \din, \dout, \epsilon, \delta)}$-equivariant-robustness 
of isomorphism equivariant models in this way proves $(\smash{\mathit{S}_N, \din, \dout, \epsilon, \delta)}$-equivariant-robustness.
Thus, these procedures guarantee robustness w.r.t\ graph edit distance 
$\smash{\dlift((\mX, \mA), (\mX', \mA')) = \min_{\mP \in \mathit{S}_N} \din((\mX, \mA), (\mP \mX', \mP \mA' \mP^T))}$.

%% file: sections/future_work.tex
\subsection{Limitations }\label{section:future_work}
Group equivariance covers various common task symmetries. However, there are symmetries that do not fit into this framework, such as local gauge equivariance~\cite{Cohen2019gauge,Haan2021,He2021} or wave function symmetries~\cite{Han2019,Luo2019,Pfau2020,Hermann2020,Wilson2021,Gao2022a,Gao2023a,Wilson2023,Gao2023b,Scherbela2023}. 
A limitation of relying on group equivariant models for provable robustness is that there are  tasks where non-equivariant models have better empirical performance, for example vision transformers~\cite{Khan2022}.
Models that are in principle equivariant may also lose their equivariance due to domain-specific artifacts like image interpolation.
Finally, it should be noted that providing guarantees of the form $(\sG, \din, \dout, \epsilon, \delta)$ requires a-priori knowledge that the task is equivariant to group  $\sG$.
These limitations are however not relevant for many important domains in which equivariant models are the de-facto standard (e.g.\ graphs, point clouds and molecules).

%% file: sections/experiments.tex
\section{Experimental evaluation}
In the following, we demonstrate how sound robustness guarantees for tasks with discrete and continuous domains and equivariances can be obtained via equivariance-preserving randomized smoothing. We additionally evaluate our graph edit distance certificates and how the cost of edit operations affects  provable robustness. 
All experimental details are specified in~\cref{appendix:experimental_setup}.
Certificates are evaluated on a separate test set.
Randomized smoothing methods are evaluated using sampling and hold with high probability.
Each experiment is repeated $5$ times. We visualize the standard deviation using shaded areas.
When the shaded area is too small, we report its maximum value.
An implementaton will be made available at \href{https://www.cs.cit.tum.de/daml/equivariance-robustness}{https://cs.cit.tum.de/daml/equivariance-robustness}.

\begin{figure}[t]
	\centering
	\begin{minipage}[b]{0.49\textwidth}
		\centering
        \input{figures/experiments/pointclouds/0.2.pgf}
        \vspace{-0.4cm}
		\caption{Provable robustness of smoothed ($\sigma=0.2$) PointNet and DGCNN point cloud classifiers  on ModelNet40. Correspondence distance $\epsilon$ is the Frobenius distance between point clouds after finding an  optimal matching via permutation.}
        \label{fig:exp_main_pointcloud}
	\end{minipage}
	\hfill
	\begin{minipage}[b]{0.49\textwidth}
		\centering
        \input{figures/experiments/force_fields/0.00001.pgf}
        \vspace{-0.4cm}
  		\caption{Provable robustness of smoothed ($\sigma = \SI{1}{\femto\meter}$) DimeNet++ force predictions on MD17.
         The average provable bounds $\delta$ on prediction changes 
         are $2$ to $13$ times smaller than the average test errors
         ($0.19$ to $\SI[per-mode=repeated-symbol]{0.74}{\kilo \cal \per \mol \per \angstrom}$).
         }
        \label{fig:exp_main_molecules}
	\end{minipage}
    \vspace{-0.7cm}
\end{figure}

\textbf{Point cloud classification.}
The first task we consider is point cloud classification, i.e. $\sX = \sR^{N \times 3}$ and $\sY = \{1,\dots, K\}$.
Natural distances $\din$ and $\dout$ are the Frobenius distance and 0-1-loss.
The task is permutation invariant, i.e. symmetric group $\sG = \mathit{S}_N$ acts on $\sX$ via permutation and on $\sY$ via identity.
Thus, we need to prove robustness w.r.t.\ to the correspondence distance $\dlift$, i.e.\ the Frobenius distance after finding an optimal matching between rows via permutation.
To preserve invariance and use~\cref{proposition:certification_reduction} while randomly smoothing, we use  Gaussian measures and majority voting, i.e.\ we predict the most likely label under Gaussian input perturbations.
As invariant base models $h$ we choose PointNet~\cite{Qi2017} and DGCNN~\cite{Wang2019}, which are used in prior work on robust point cloud classification~\cite{Xiang2019,Yang2019}.
We evaluate the robustness guarantees for smoothing standard deviation $\sigma=0.2$ on ModelNet40~\cite{Wu2015}. This dataset consists of point cloud representations of CAD models from $40$ different classes.
\cref{fig:exp_main_pointcloud} shows the certified accuracy, i.e.\ the percentage of predictions that are correct and provably robust.
Both smoothed models have a high accuracy above $80\%$, but DGCNN has significantly higher provable robustness.
In~\cref{appendix:extra_experiments_pointclouds} we repeat the experiment for different values of $\sigma$ and compare our guarantees to those of 3DVerifier~\cite{Mu2022}.

\textbf{Molecular force prediction.}
Next, we consider a  task with continuous equivariances: 
Predicting the forces acting on each atom in a molecule, which can for instance be used to simulate molecular dynamics.
We have $\sX=\sR^{N \times 3}$ (atomic numbers are treated as constant) and $\sY=\sR^{N \times 3}$.
Suitable distances $\din$ and $\dout$ are the Frobenius distance and the average $\ell_2$  error $\sum_{n=1}^N ||\mY_n - \mY'_n||_2 \mathbin{/} N$.
The task is permutation, rotation and translation equivariant, i.e. $\sG = \mathit{S}_N \times \mathit{SE}(3)$.
Thus, $\dlift$ is the point cloud registration distance.
To preserve equivariance, we use Gaussian measures and center smoothing.
We choose DimeNet++~\cite{Gasteiger2020} as equivariant base model $h$.
We evaluate the provable robustness for smoothing standard deviation $\sigma = \SI{1}{\femto\meter}$ on MD17~\cite{Chmiela2017}, a collection of $8$ datasets, each consisting of a large number of configurations of a specific molecule.
With just $1000$ training samples, the models achieve low average test errors between $0.19$ and $\SI[per-mode=repeated-symbol]{0.74}{\kilo \cal \per \mol \per \angstrom}$.
We use  $1000$ samples per test set to evaluate the robustness guarantees.
\cref{fig:exp_main_molecules} shows the average upper bounds $\delta$ on the change of the predicted force vectors for small perturbations between $0$ and $\SI{2}{\femto\meter}$. These average $\delta$ are smaller than the average test errors by factors between $2$ and $13$.
The standard deviation across seeds is below $\SI[per-mode=repeated-symbol]{3e-4}{\kilo \cal \per \mol \per \angstrom}$ for all $\epsilon$.
In~\cref{appendix:extra_experiments_molecules} we show that the maximum $\epsilon$ and $\delta$ grow approximately linearly with $\sigma$ and repeat our experiments with SchNet~\cite{Schutt2018} and SphereNet~\cite{Liu2022} base models.
Note that we are not (primarily) proving robustness to malicious perturbations, but robustness to measurement errors or errors in previous simulation steps.

\textbf{Node classification.}
Finally, we consider a task with discrete domain and co-domain: Node classification, i.e.\ $\sX = \{0,1\}^{N \times D} \times \{0,1\}^{N \times N}$ and $\sY = \{1,\dots, K\}^N$.
Distance $\din$ is naturally defined via a weighted sum of inserted and deleted bits (see~\cref{eq:ged_base_distance}).
Output distance $\dout$ is the $\ell_0$ distance, i.e.\ the number of changed predictions.
The task is equivariant to symmetric group $\mathit{S}_n$ acting on $\sX$ and $\sY$ via isomorphisms and permutations, respectively.
To preserve equivariance, we use sparsity-aware measures and majority voting.
The flip probabilites (see~\cref{appendix:ged_randomized_smoothing}) are set to
$p_\mX^+=0,p_\mX^-=0,p_\mA^+=0.001,p_\mA^-=0.8$.
We use a $2$-layer graph convolutional network~\cite{Kipf2017} as our isomorphism equivariant base model $h$.
\cref{fig:exp_main_graph_smoothing} shows the resulting guarantees on Cora ML~\cite{McCallum2000,Bojchevski2018} for graph edit perturbations of the adjacency, i.e.\ $c_\mX^+ = c_\mX^- = \infty$, for varying costs  $c_\mA^+$ and $c_\mA^-$.
We observe that increasing the cost for edge insertions significantly increases the provable robustness for the same budgets $\epsilon$, whereas the cost for edge deletions has virtually no effect.
This suggests that insertions are much more effective at changing a model's predictions, which was empirically observed in prior work~\cite{Zuegner2020patterns}.
We repeat the experiment with other flip probabilities in~\cref{appendix:extra_experiments_ged}.
The standard deviation of certified accuracies across all seeds was below
$\SI{2.1}{p{.}p{.}}$ everywhere.
Note that here, certified accuracy does not refer to the percentage of predictions that are correct and constant, but those that are correct and permute in compliance with isomorphisms of the input graph.

\textbf{Deterministic edit distance certificates.} 
In~\cref{fig:exp_main_graph_deterministic} we repeat the node classification experiment with our generalization of the convex polytope method from~\cite{Zuegner2019cert}. We consider feature perturbations, i.e.\ $c_\mA^+ = c_\mA^- = \infty$, for varying  $c_\mX^+$ and $c_\mX^-$.
Again, the cost of insertions has a larger effect, suggesting that the model is less robust to them.
In~\cref{appendix:extra_experiments_ged} we repeat the experiments with the four other graph edit distance certificates and also evaluate them on Citeseer~\cite{Sen2008} and the TUDataset~\cite{Morris2020}.
Overall, our generalizations of existing graph robustness guarantees let us prove robustness to more complex threat models, even though evaluating the edit distance is computationally hard.

\begin{figure}[t]
	\centering
	\begin{minipage}{0.49\textwidth}
    \begin{subfigure}[t]{0.49\linewidth}
		\centering
       \input{figures/experiments/graphs/sparse_smoothing/nodes_structure/node_classification-Cora-GCN-hidden=32-p_adj_plus=0.001-p_adj_minus=0.8-p_att_plus=0.0-p_att_minus=0.0-multi_class_cert-B.pgf}
    \end{subfigure}
    \begin{subfigure}[t]{0.49\linewidth}
		\centering
       \input{figures/experiments/graphs/sparse_smoothing/nodes_structure/node_classification-Cora-GCN-hidden=32-p_adj_plus=0.001-p_adj_minus=0.8-p_att_plus=0.0-p_att_minus=0.0-multi_class_cert-A.pgf}
    \end{subfigure}
    \vspace{-0.2cm}
    \caption{Randomized smoothing guarantees for GCNs on Cora-ML.
             Increasing the cost of adversarial edge insertions increases the provable robustness for the same perturbation budgets $\epsilon$.}
    \label{fig:exp_main_graph_smoothing}
	\end{minipage}
	\hfill
	\begin{minipage}{0.49\textwidth}
    \begin{subfigure}[t]{0.49\linewidth}
		\centering
       \input{figures/experiments/graphs/attributes_zuegner/cora_cost_add.pgf}
    \end{subfigure}
    \begin{subfigure}[t]{0.49\linewidth}
		\centering
       \input{figures/experiments/graphs/attributes_zuegner/cora_cost_del.pgf}
    \end{subfigure}
    \vspace{-0.2cm}
    \caption{Convex adversarial polytope guarantees for GCNs on Cora-ML.
            Increasing the cost of attribute insertions increases the provable robustness for the same perturbation budgets $\epsilon$.}
    \label{fig:exp_main_graph_deterministic}
	\end{minipage}
    \vspace{-0.6cm}
\end{figure}

%% file: figures/experiments/pointclouds/0.2.pgf
\begingroup%
\makeatletter%
\begin{pgfpicture}%
\pgfpathrectangle{\pgfpointorigin}{\pgfqpoint{2.750000in}{1.699593in}}%
\pgfusepath{use as bounding box, clip}%
\begin{pgfscope}%
\pgfsetbuttcap%
\pgfsetmiterjoin%
\definecolor{currentfill}{rgb}{1.000000,1.000000,1.000000}%
\pgfsetfillcolor{currentfill}%
\pgfsetlinewidth{0.000000pt}%
\definecolor{currentstroke}{rgb}{1.000000,1.000000,1.000000}%
\pgfsetstrokecolor{currentstroke}%
\pgfsetdash{}{0pt}%
\pgfpathmoveto{\pgfqpoint{0.000000in}{0.000000in}}%
\pgfpathlineto{\pgfqpoint{2.750000in}{0.000000in}}%
\pgfpathlineto{\pgfqpoint{2.750000in}{1.699593in}}%
\pgfpathlineto{\pgfqpoint{0.000000in}{1.699593in}}%
\pgfpathlineto{\pgfqpoint{0.000000in}{0.000000in}}%
\pgfpathclose%
\pgfusepath{fill}%
\end{pgfscope}%
\begin{pgfscope}%
\pgfsetbuttcap%
\pgfsetmiterjoin%
\definecolor{currentfill}{rgb}{1.000000,1.000000,1.000000}%
\pgfsetfillcolor{currentfill}%
\pgfsetlinewidth{0.000000pt}%
\definecolor{currentstroke}{rgb}{0.000000,0.000000,0.000000}%
\pgfsetstrokecolor{currentstroke}%
\pgfsetstrokeopacity{0.000000}%
\pgfsetdash{}{0pt}%
\pgfpathmoveto{\pgfqpoint{0.520798in}{0.442177in}}%
\pgfpathlineto{\pgfqpoint{2.673611in}{0.442177in}}%
\pgfpathlineto{\pgfqpoint{2.673611in}{1.585085in}}%
\pgfpathlineto{\pgfqpoint{0.520798in}{1.585085in}}%
\pgfpathlineto{\pgfqpoint{0.520798in}{0.442177in}}%
\pgfpathclose%
\pgfusepath{fill}%
\end{pgfscope}%
\begin{pgfscope}%
\pgfpathrectangle{\pgfqpoint{0.520798in}{0.442177in}}{\pgfqpoint{2.152813in}{1.142908in}}%
\pgfusepath{clip}%
\pgfsetroundcap%
\pgfsetroundjoin%
\pgfsetlinewidth{0.501875pt}%
\definecolor{currentstroke}{rgb}{0.800000,0.800000,0.800000}%
\pgfsetstrokecolor{currentstroke}%
\pgfsetdash{}{0pt}%
\pgfpathmoveto{\pgfqpoint{0.520798in}{0.442177in}}%
\pgfpathlineto{\pgfqpoint{0.520798in}{1.585085in}}%
\pgfusepath{stroke}%
\end{pgfscope}%
\begin{pgfscope}%
\definecolor{textcolor}{rgb}{0.150000,0.150000,0.150000}%
\pgfsetstrokecolor{textcolor}%
\pgfsetfillcolor{textcolor}%
\pgftext[x=0.520798in,y=0.351899in,,top]{\color{textcolor}\rmfamily\fontsize{8.000000}{9.600000}\selectfont \(\displaystyle {0.0}\)}%
\end{pgfscope}%
\begin{pgfscope}%
\pgfpathrectangle{\pgfqpoint{0.520798in}{0.442177in}}{\pgfqpoint{2.152813in}{1.142908in}}%
\pgfusepath{clip}%
\pgfsetroundcap%
\pgfsetroundjoin%
\pgfsetlinewidth{0.501875pt}%
\definecolor{currentstroke}{rgb}{0.800000,0.800000,0.800000}%
\pgfsetstrokecolor{currentstroke}%
\pgfsetdash{}{0pt}%
\pgfpathmoveto{\pgfqpoint{1.135887in}{0.442177in}}%
\pgfpathlineto{\pgfqpoint{1.135887in}{1.585085in}}%
\pgfusepath{stroke}%
\end{pgfscope}%
\begin{pgfscope}%
\definecolor{textcolor}{rgb}{0.150000,0.150000,0.150000}%
\pgfsetstrokecolor{textcolor}%
\pgfsetfillcolor{textcolor}%
\pgftext[x=1.135887in,y=0.351899in,,top]{\color{textcolor}\rmfamily\fontsize{8.000000}{9.600000}\selectfont \(\displaystyle {0.2}\)}%
\end{pgfscope}%
\begin{pgfscope}%
\pgfpathrectangle{\pgfqpoint{0.520798in}{0.442177in}}{\pgfqpoint{2.152813in}{1.142908in}}%
\pgfusepath{clip}%
\pgfsetroundcap%
\pgfsetroundjoin%
\pgfsetlinewidth{0.501875pt}%
\definecolor{currentstroke}{rgb}{0.800000,0.800000,0.800000}%
\pgfsetstrokecolor{currentstroke}%
\pgfsetdash{}{0pt}%
\pgfpathmoveto{\pgfqpoint{1.750977in}{0.442177in}}%
\pgfpathlineto{\pgfqpoint{1.750977in}{1.585085in}}%
\pgfusepath{stroke}%
\end{pgfscope}%
\begin{pgfscope}%
\definecolor{textcolor}{rgb}{0.150000,0.150000,0.150000}%
\pgfsetstrokecolor{textcolor}%
\pgfsetfillcolor{textcolor}%
\pgftext[x=1.750977in,y=0.351899in,,top]{\color{textcolor}\rmfamily\fontsize{8.000000}{9.600000}\selectfont \(\displaystyle {0.4}\)}%
\end{pgfscope}%
\begin{pgfscope}%
\pgfpathrectangle{\pgfqpoint{0.520798in}{0.442177in}}{\pgfqpoint{2.152813in}{1.142908in}}%
\pgfusepath{clip}%
\pgfsetroundcap%
\pgfsetroundjoin%
\pgfsetlinewidth{0.501875pt}%
\definecolor{currentstroke}{rgb}{0.800000,0.800000,0.800000}%
\pgfsetstrokecolor{currentstroke}%
\pgfsetdash{}{0pt}%
\pgfpathmoveto{\pgfqpoint{2.366066in}{0.442177in}}%
\pgfpathlineto{\pgfqpoint{2.366066in}{1.585085in}}%
\pgfusepath{stroke}%
\end{pgfscope}%
\begin{pgfscope}%
\definecolor{textcolor}{rgb}{0.150000,0.150000,0.150000}%
\pgfsetstrokecolor{textcolor}%
\pgfsetfillcolor{textcolor}%
\pgftext[x=2.366066in,y=0.351899in,,top]{\color{textcolor}\rmfamily\fontsize{8.000000}{9.600000}\selectfont \(\displaystyle {0.6}\)}%
\end{pgfscope}%
\begin{pgfscope}%
\definecolor{textcolor}{rgb}{0.150000,0.150000,0.150000}%
\pgfsetstrokecolor{textcolor}%
\pgfsetfillcolor{textcolor}%
\pgftext[x=1.597204in,y=0.198219in,,top]{\color{textcolor}\rmfamily\fontsize{10.000000}{12.000000}\selectfont Correspondence distance \(\displaystyle \epsilon\)}%
\end{pgfscope}%
\begin{pgfscope}%
\pgfpathrectangle{\pgfqpoint{0.520798in}{0.442177in}}{\pgfqpoint{2.152813in}{1.142908in}}%
\pgfusepath{clip}%
\pgfsetroundcap%
\pgfsetroundjoin%
\pgfsetlinewidth{0.501875pt}%
\definecolor{currentstroke}{rgb}{0.800000,0.800000,0.800000}%
\pgfsetstrokecolor{currentstroke}%
\pgfsetdash{}{0pt}%
\pgfpathmoveto{\pgfqpoint{0.520798in}{0.442177in}}%
\pgfpathlineto{\pgfqpoint{2.673611in}{0.442177in}}%
\pgfusepath{stroke}%
\end{pgfscope}%
\begin{pgfscope}%
\definecolor{textcolor}{rgb}{0.150000,0.150000,0.150000}%
\pgfsetstrokecolor{textcolor}%
\pgfsetfillcolor{textcolor}%
\pgftext[x=0.273151in, y=0.403915in, left, base]{\color{textcolor}\rmfamily\fontsize{8.000000}{9.600000}\selectfont 0\%}%
\end{pgfscope}%
\begin{pgfscope}%
\pgfpathrectangle{\pgfqpoint{0.520798in}{0.442177in}}{\pgfqpoint{2.152813in}{1.142908in}}%
\pgfusepath{clip}%
\pgfsetroundcap%
\pgfsetroundjoin%
\pgfsetlinewidth{0.501875pt}%
\definecolor{currentstroke}{rgb}{0.800000,0.800000,0.800000}%
\pgfsetstrokecolor{currentstroke}%
\pgfsetdash{}{0pt}%
\pgfpathmoveto{\pgfqpoint{0.520798in}{0.670758in}}%
\pgfpathlineto{\pgfqpoint{2.673611in}{0.670758in}}%
\pgfusepath{stroke}%
\end{pgfscope}%
\begin{pgfscope}%
\definecolor{textcolor}{rgb}{0.150000,0.150000,0.150000}%
\pgfsetstrokecolor{textcolor}%
\pgfsetfillcolor{textcolor}%
\pgftext[x=0.214138in, y=0.632496in, left, base]{\color{textcolor}\rmfamily\fontsize{8.000000}{9.600000}\selectfont 20\%}%
\end{pgfscope}%
\begin{pgfscope}%
\pgfpathrectangle{\pgfqpoint{0.520798in}{0.442177in}}{\pgfqpoint{2.152813in}{1.142908in}}%
\pgfusepath{clip}%
\pgfsetroundcap%
\pgfsetroundjoin%
\pgfsetlinewidth{0.501875pt}%
\definecolor{currentstroke}{rgb}{0.800000,0.800000,0.800000}%
\pgfsetstrokecolor{currentstroke}%
\pgfsetdash{}{0pt}%
\pgfpathmoveto{\pgfqpoint{0.520798in}{0.899340in}}%
\pgfpathlineto{\pgfqpoint{2.673611in}{0.899340in}}%
\pgfusepath{stroke}%
\end{pgfscope}%
\begin{pgfscope}%
\definecolor{textcolor}{rgb}{0.150000,0.150000,0.150000}%
\pgfsetstrokecolor{textcolor}%
\pgfsetfillcolor{textcolor}%
\pgftext[x=0.214138in, y=0.861078in, left, base]{\color{textcolor}\rmfamily\fontsize{8.000000}{9.600000}\selectfont 40\%}%
\end{pgfscope}%
\begin{pgfscope}%
\pgfpathrectangle{\pgfqpoint{0.520798in}{0.442177in}}{\pgfqpoint{2.152813in}{1.142908in}}%
\pgfusepath{clip}%
\pgfsetroundcap%
\pgfsetroundjoin%
\pgfsetlinewidth{0.501875pt}%
\definecolor{currentstroke}{rgb}{0.800000,0.800000,0.800000}%
\pgfsetstrokecolor{currentstroke}%
\pgfsetdash{}{0pt}%
\pgfpathmoveto{\pgfqpoint{0.520798in}{1.127922in}}%
\pgfpathlineto{\pgfqpoint{2.673611in}{1.127922in}}%
\pgfusepath{stroke}%
\end{pgfscope}%
\begin{pgfscope}%
\definecolor{textcolor}{rgb}{0.150000,0.150000,0.150000}%
\pgfsetstrokecolor{textcolor}%
\pgfsetfillcolor{textcolor}%
\pgftext[x=0.214138in, y=1.089660in, left, base]{\color{textcolor}\rmfamily\fontsize{8.000000}{9.600000}\selectfont 60\%}%
\end{pgfscope}%
\begin{pgfscope}%
\pgfpathrectangle{\pgfqpoint{0.520798in}{0.442177in}}{\pgfqpoint{2.152813in}{1.142908in}}%
\pgfusepath{clip}%
\pgfsetroundcap%
\pgfsetroundjoin%
\pgfsetlinewidth{0.501875pt}%
\definecolor{currentstroke}{rgb}{0.800000,0.800000,0.800000}%
\pgfsetstrokecolor{currentstroke}%
\pgfsetdash{}{0pt}%
\pgfpathmoveto{\pgfqpoint{0.520798in}{1.356504in}}%
\pgfpathlineto{\pgfqpoint{2.673611in}{1.356504in}}%
\pgfusepath{stroke}%
\end{pgfscope}%
\begin{pgfscope}%
\definecolor{textcolor}{rgb}{0.150000,0.150000,0.150000}%
\pgfsetstrokecolor{textcolor}%
\pgfsetfillcolor{textcolor}%
\pgftext[x=0.214138in, y=1.318241in, left, base]{\color{textcolor}\rmfamily\fontsize{8.000000}{9.600000}\selectfont 80\%}%
\end{pgfscope}%
\begin{pgfscope}%
\pgfpathrectangle{\pgfqpoint{0.520798in}{0.442177in}}{\pgfqpoint{2.152813in}{1.142908in}}%
\pgfusepath{clip}%
\pgfsetroundcap%
\pgfsetroundjoin%
\pgfsetlinewidth{0.501875pt}%
\definecolor{currentstroke}{rgb}{0.800000,0.800000,0.800000}%
\pgfsetstrokecolor{currentstroke}%
\pgfsetdash{}{0pt}%
\pgfpathmoveto{\pgfqpoint{0.520798in}{1.585085in}}%
\pgfpathlineto{\pgfqpoint{2.673611in}{1.585085in}}%
\pgfusepath{stroke}%
\end{pgfscope}%
\begin{pgfscope}%
\definecolor{textcolor}{rgb}{0.150000,0.150000,0.150000}%
\pgfsetstrokecolor{textcolor}%
\pgfsetfillcolor{textcolor}%
\pgftext[x=0.155124in, y=1.546823in, left, base]{\color{textcolor}\rmfamily\fontsize{8.000000}{9.600000}\selectfont 100\%}%
\end{pgfscope}%
\begin{pgfscope}%
\definecolor{textcolor}{rgb}{0.150000,0.150000,0.150000}%
\pgfsetstrokecolor{textcolor}%
\pgfsetfillcolor{textcolor}%
\pgftext[x=0.099569in,y=1.013631in,,bottom,rotate=90.000000]{\color{textcolor}\rmfamily\fontsize{10.000000}{12.000000}\selectfont Cert. Acc.}%
\end{pgfscope}%
\begin{pgfscope}%
\pgfsetrectcap%
\pgfsetmiterjoin%
\pgfsetlinewidth{0.752812pt}%
\definecolor{currentstroke}{rgb}{0.700000,0.700000,0.700000}%
\pgfsetstrokecolor{currentstroke}%
\pgfsetdash{}{0pt}%
\pgfpathmoveto{\pgfqpoint{0.520798in}{0.442177in}}%
\pgfpathlineto{\pgfqpoint{0.520798in}{1.585085in}}%
\pgfusepath{stroke}%
\end{pgfscope}%
\begin{pgfscope}%
\pgfsetrectcap%
\pgfsetmiterjoin%
\pgfsetlinewidth{0.752812pt}%
\definecolor{currentstroke}{rgb}{0.700000,0.700000,0.700000}%
\pgfsetstrokecolor{currentstroke}%
\pgfsetdash{}{0pt}%
\pgfpathmoveto{\pgfqpoint{2.673611in}{0.442177in}}%
\pgfpathlineto{\pgfqpoint{2.673611in}{1.585085in}}%
\pgfusepath{stroke}%
\end{pgfscope}%
\begin{pgfscope}%
\pgfsetrectcap%
\pgfsetmiterjoin%
\pgfsetlinewidth{0.752812pt}%
\definecolor{currentstroke}{rgb}{0.700000,0.700000,0.700000}%
\pgfsetstrokecolor{currentstroke}%
\pgfsetdash{}{0pt}%
\pgfpathmoveto{\pgfqpoint{0.520798in}{0.442177in}}%
\pgfpathlineto{\pgfqpoint{2.673611in}{0.442177in}}%
\pgfusepath{stroke}%
\end{pgfscope}%
\begin{pgfscope}%
\pgfsetrectcap%
\pgfsetmiterjoin%
\pgfsetlinewidth{0.752812pt}%
\definecolor{currentstroke}{rgb}{0.700000,0.700000,0.700000}%
\pgfsetstrokecolor{currentstroke}%
\pgfsetdash{}{0pt}%
\pgfpathmoveto{\pgfqpoint{0.520798in}{1.585085in}}%
\pgfpathlineto{\pgfqpoint{2.673611in}{1.585085in}}%
\pgfusepath{stroke}%
\end{pgfscope}%
\begin{pgfscope}%
\pgfpathrectangle{\pgfqpoint{0.520798in}{0.442177in}}{\pgfqpoint{2.152813in}{1.142908in}}%
\pgfusepath{clip}%
\pgfsetroundcap%
\pgfsetroundjoin%
\pgfsetlinewidth{1.003750pt}%
\definecolor{currentstroke}{rgb}{0.003922,0.450980,0.698039}%
\pgfsetstrokecolor{currentstroke}%
\pgfsetdash{}{0pt}%
\pgfpathmoveto{\pgfqpoint{0.520798in}{1.360301in}}%
\pgfpathlineto{\pgfqpoint{0.551863in}{1.353169in}}%
\pgfpathlineto{\pgfqpoint{0.582928in}{1.345389in}}%
\pgfpathlineto{\pgfqpoint{0.613993in}{1.338073in}}%
\pgfpathlineto{\pgfqpoint{0.645058in}{1.328996in}}%
\pgfpathlineto{\pgfqpoint{0.676124in}{1.321864in}}%
\pgfpathlineto{\pgfqpoint{0.707189in}{1.313621in}}%
\pgfpathlineto{\pgfqpoint{0.738254in}{1.304823in}}%
\pgfpathlineto{\pgfqpoint{0.769319in}{1.295839in}}%
\pgfpathlineto{\pgfqpoint{0.800384in}{1.286021in}}%
\pgfpathlineto{\pgfqpoint{0.831449in}{1.276111in}}%
\pgfpathlineto{\pgfqpoint{0.862514in}{1.265552in}}%
\pgfpathlineto{\pgfqpoint{0.893579in}{1.255457in}}%
\pgfpathlineto{\pgfqpoint{0.924645in}{1.244899in}}%
\pgfpathlineto{\pgfqpoint{0.955710in}{1.234340in}}%
\pgfpathlineto{\pgfqpoint{0.986775in}{1.223411in}}%
\pgfpathlineto{\pgfqpoint{1.017840in}{1.211834in}}%
\pgfpathlineto{\pgfqpoint{1.048905in}{1.201553in}}%
\pgfpathlineto{\pgfqpoint{1.079970in}{1.190069in}}%
\pgfpathlineto{\pgfqpoint{1.111035in}{1.178584in}}%
\pgfpathlineto{\pgfqpoint{1.142100in}{1.164877in}}%
\pgfpathlineto{\pgfqpoint{1.173166in}{1.152095in}}%
\pgfpathlineto{\pgfqpoint{1.204231in}{1.140333in}}%
\pgfpathlineto{\pgfqpoint{1.235296in}{1.126903in}}%
\pgfpathlineto{\pgfqpoint{1.266361in}{1.113196in}}%
\pgfpathlineto{\pgfqpoint{1.297426in}{1.102822in}}%
\pgfpathlineto{\pgfqpoint{1.328491in}{1.088281in}}%
\pgfpathlineto{\pgfqpoint{1.359556in}{1.077260in}}%
\pgfpathlineto{\pgfqpoint{1.390621in}{1.063645in}}%
\pgfpathlineto{\pgfqpoint{1.421687in}{1.050771in}}%
\pgfpathlineto{\pgfqpoint{1.452752in}{1.036415in}}%
\pgfpathlineto{\pgfqpoint{1.483817in}{1.022800in}}%
\pgfpathlineto{\pgfqpoint{1.514882in}{1.008074in}}%
\pgfpathlineto{\pgfqpoint{1.545947in}{0.994459in}}%
\pgfpathlineto{\pgfqpoint{1.577012in}{0.978158in}}%
\pgfpathlineto{\pgfqpoint{1.608077in}{0.963247in}}%
\pgfpathlineto{\pgfqpoint{1.639142in}{0.947131in}}%
\pgfpathlineto{\pgfqpoint{1.670208in}{0.930738in}}%
\pgfpathlineto{\pgfqpoint{1.701273in}{0.915271in}}%
\pgfpathlineto{\pgfqpoint{1.732338in}{0.899155in}}%
\pgfpathlineto{\pgfqpoint{1.763403in}{0.883132in}}%
\pgfpathlineto{\pgfqpoint{1.794468in}{0.866646in}}%
\pgfpathlineto{\pgfqpoint{1.825533in}{0.849141in}}%
\pgfpathlineto{\pgfqpoint{1.856598in}{0.833952in}}%
\pgfpathlineto{\pgfqpoint{1.887663in}{0.817836in}}%
\pgfpathlineto{\pgfqpoint{1.918729in}{0.801906in}}%
\pgfpathlineto{\pgfqpoint{1.949794in}{0.783012in}}%
\pgfpathlineto{\pgfqpoint{1.980859in}{0.769582in}}%
\pgfpathlineto{\pgfqpoint{2.011924in}{0.755134in}}%
\pgfpathlineto{\pgfqpoint{2.042989in}{0.742074in}}%
\pgfpathlineto{\pgfqpoint{2.074054in}{0.729201in}}%
\pgfpathlineto{\pgfqpoint{2.105119in}{0.716419in}}%
\pgfpathlineto{\pgfqpoint{2.136184in}{0.704101in}}%
\pgfpathlineto{\pgfqpoint{2.167250in}{0.692431in}}%
\pgfpathlineto{\pgfqpoint{2.198315in}{0.680946in}}%
\pgfpathlineto{\pgfqpoint{2.229380in}{0.670018in}}%
\pgfpathlineto{\pgfqpoint{2.260445in}{0.654643in}}%
\pgfpathlineto{\pgfqpoint{2.291510in}{0.642788in}}%
\pgfpathlineto{\pgfqpoint{2.322575in}{0.627969in}}%
\pgfpathlineto{\pgfqpoint{2.353640in}{0.605000in}}%
\pgfpathlineto{\pgfqpoint{2.384705in}{0.591477in}}%
\pgfpathlineto{\pgfqpoint{2.415771in}{0.569990in}}%
\pgfpathlineto{\pgfqpoint{2.446836in}{0.539333in}}%
\pgfpathlineto{\pgfqpoint{2.477901in}{0.539333in}}%
\pgfpathlineto{\pgfqpoint{2.508966in}{0.442177in}}%
\pgfpathlineto{\pgfqpoint{2.540031in}{0.442177in}}%
\pgfpathlineto{\pgfqpoint{2.571096in}{0.442177in}}%
\pgfpathlineto{\pgfqpoint{2.602161in}{0.442177in}}%
\pgfpathlineto{\pgfqpoint{2.633226in}{0.442177in}}%
\pgfpathlineto{\pgfqpoint{2.664292in}{0.442177in}}%
\pgfpathlineto{\pgfqpoint{2.675278in}{0.442177in}}%
\pgfusepath{stroke}%
\end{pgfscope}%
\begin{pgfscope}%
\pgfpathrectangle{\pgfqpoint{0.520798in}{0.442177in}}{\pgfqpoint{2.152813in}{1.142908in}}%
\pgfusepath{clip}%
\pgfsetbuttcap%
\pgfsetroundjoin%
\definecolor{currentfill}{rgb}{0.003922,0.450980,0.698039}%
\pgfsetfillcolor{currentfill}%
\pgfsetfillopacity{0.500000}%
\pgfsetlinewidth{0.000000pt}%
\definecolor{currentstroke}{rgb}{0.003922,0.450980,0.698039}%
\pgfsetstrokecolor{currentstroke}%
\pgfsetstrokeopacity{0.500000}%
\pgfsetdash{}{0pt}%
\pgfpathmoveto{\pgfqpoint{0.520798in}{1.370802in}}%
\pgfpathlineto{\pgfqpoint{0.520798in}{1.349799in}}%
\pgfpathlineto{\pgfqpoint{0.551863in}{1.343167in}}%
\pgfpathlineto{\pgfqpoint{0.582928in}{1.333927in}}%
\pgfpathlineto{\pgfqpoint{0.613993in}{1.327510in}}%
\pgfpathlineto{\pgfqpoint{0.645058in}{1.318468in}}%
\pgfpathlineto{\pgfqpoint{0.676124in}{1.310845in}}%
\pgfpathlineto{\pgfqpoint{0.707189in}{1.301298in}}%
\pgfpathlineto{\pgfqpoint{0.738254in}{1.291178in}}%
\pgfpathlineto{\pgfqpoint{0.769319in}{1.280868in}}%
\pgfpathlineto{\pgfqpoint{0.800384in}{1.270267in}}%
\pgfpathlineto{\pgfqpoint{0.831449in}{1.259313in}}%
\pgfpathlineto{\pgfqpoint{0.862514in}{1.248565in}}%
\pgfpathlineto{\pgfqpoint{0.893579in}{1.238206in}}%
\pgfpathlineto{\pgfqpoint{0.924645in}{1.227266in}}%
\pgfpathlineto{\pgfqpoint{0.955710in}{1.218029in}}%
\pgfpathlineto{\pgfqpoint{0.986775in}{1.205768in}}%
\pgfpathlineto{\pgfqpoint{1.017840in}{1.195493in}}%
\pgfpathlineto{\pgfqpoint{1.048905in}{1.184651in}}%
\pgfpathlineto{\pgfqpoint{1.079970in}{1.174735in}}%
\pgfpathlineto{\pgfqpoint{1.111035in}{1.162759in}}%
\pgfpathlineto{\pgfqpoint{1.142100in}{1.149256in}}%
\pgfpathlineto{\pgfqpoint{1.173166in}{1.136512in}}%
\pgfpathlineto{\pgfqpoint{1.204231in}{1.125146in}}%
\pgfpathlineto{\pgfqpoint{1.235296in}{1.111697in}}%
\pgfpathlineto{\pgfqpoint{1.266361in}{1.099061in}}%
\pgfpathlineto{\pgfqpoint{1.297426in}{1.087872in}}%
\pgfpathlineto{\pgfqpoint{1.328491in}{1.072571in}}%
\pgfpathlineto{\pgfqpoint{1.359556in}{1.062895in}}%
\pgfpathlineto{\pgfqpoint{1.390621in}{1.048261in}}%
\pgfpathlineto{\pgfqpoint{1.421687in}{1.035159in}}%
\pgfpathlineto{\pgfqpoint{1.452752in}{1.019968in}}%
\pgfpathlineto{\pgfqpoint{1.483817in}{1.006218in}}%
\pgfpathlineto{\pgfqpoint{1.514882in}{0.992142in}}%
\pgfpathlineto{\pgfqpoint{1.545947in}{0.978477in}}%
\pgfpathlineto{\pgfqpoint{1.577012in}{0.963165in}}%
\pgfpathlineto{\pgfqpoint{1.608077in}{0.946907in}}%
\pgfpathlineto{\pgfqpoint{1.639142in}{0.932316in}}%
\pgfpathlineto{\pgfqpoint{1.670208in}{0.913453in}}%
\pgfpathlineto{\pgfqpoint{1.701273in}{0.898053in}}%
\pgfpathlineto{\pgfqpoint{1.732338in}{0.881735in}}%
\pgfpathlineto{\pgfqpoint{1.763403in}{0.865148in}}%
\pgfpathlineto{\pgfqpoint{1.794468in}{0.849675in}}%
\pgfpathlineto{\pgfqpoint{1.825533in}{0.830457in}}%
\pgfpathlineto{\pgfqpoint{1.856598in}{0.814786in}}%
\pgfpathlineto{\pgfqpoint{1.887663in}{0.799969in}}%
\pgfpathlineto{\pgfqpoint{1.918729in}{0.785451in}}%
\pgfpathlineto{\pgfqpoint{1.949794in}{0.767611in}}%
\pgfpathlineto{\pgfqpoint{1.980859in}{0.752913in}}%
\pgfpathlineto{\pgfqpoint{2.011924in}{0.736471in}}%
\pgfpathlineto{\pgfqpoint{2.042989in}{0.723045in}}%
\pgfpathlineto{\pgfqpoint{2.074054in}{0.710697in}}%
\pgfpathlineto{\pgfqpoint{2.105119in}{0.697239in}}%
\pgfpathlineto{\pgfqpoint{2.136184in}{0.685298in}}%
\pgfpathlineto{\pgfqpoint{2.167250in}{0.673837in}}%
\pgfpathlineto{\pgfqpoint{2.198315in}{0.663175in}}%
\pgfpathlineto{\pgfqpoint{2.229380in}{0.650458in}}%
\pgfpathlineto{\pgfqpoint{2.260445in}{0.634193in}}%
\pgfpathlineto{\pgfqpoint{2.291510in}{0.621306in}}%
\pgfpathlineto{\pgfqpoint{2.322575in}{0.606909in}}%
\pgfpathlineto{\pgfqpoint{2.353640in}{0.583341in}}%
\pgfpathlineto{\pgfqpoint{2.384705in}{0.570409in}}%
\pgfpathlineto{\pgfqpoint{2.415771in}{0.551756in}}%
\pgfpathlineto{\pgfqpoint{2.446836in}{0.524888in}}%
\pgfpathlineto{\pgfqpoint{2.477901in}{0.524888in}}%
\pgfpathlineto{\pgfqpoint{2.508966in}{0.442177in}}%
\pgfpathlineto{\pgfqpoint{2.540031in}{0.442177in}}%
\pgfpathlineto{\pgfqpoint{2.571096in}{0.442177in}}%
\pgfpathlineto{\pgfqpoint{2.602161in}{0.442177in}}%
\pgfpathlineto{\pgfqpoint{2.633226in}{0.442177in}}%
\pgfpathlineto{\pgfqpoint{2.664292in}{0.442177in}}%
\pgfpathlineto{\pgfqpoint{2.695357in}{0.442177in}}%
\pgfpathlineto{\pgfqpoint{2.726422in}{0.442177in}}%
\pgfpathlineto{\pgfqpoint{2.757487in}{0.442177in}}%
\pgfpathlineto{\pgfqpoint{2.788552in}{0.442177in}}%
\pgfpathlineto{\pgfqpoint{2.819617in}{0.442177in}}%
\pgfpathlineto{\pgfqpoint{2.850682in}{0.442177in}}%
\pgfpathlineto{\pgfqpoint{2.881747in}{0.442177in}}%
\pgfpathlineto{\pgfqpoint{2.912813in}{0.442177in}}%
\pgfpathlineto{\pgfqpoint{2.943878in}{0.442177in}}%
\pgfpathlineto{\pgfqpoint{2.974943in}{0.442177in}}%
\pgfpathlineto{\pgfqpoint{3.006008in}{0.442177in}}%
\pgfpathlineto{\pgfqpoint{3.037073in}{0.442177in}}%
\pgfpathlineto{\pgfqpoint{3.068138in}{0.442177in}}%
\pgfpathlineto{\pgfqpoint{3.099203in}{0.442177in}}%
\pgfpathlineto{\pgfqpoint{3.130268in}{0.442177in}}%
\pgfpathlineto{\pgfqpoint{3.161334in}{0.442177in}}%
\pgfpathlineto{\pgfqpoint{3.192399in}{0.442177in}}%
\pgfpathlineto{\pgfqpoint{3.223464in}{0.442177in}}%
\pgfpathlineto{\pgfqpoint{3.254529in}{0.442177in}}%
\pgfpathlineto{\pgfqpoint{3.285594in}{0.442177in}}%
\pgfpathlineto{\pgfqpoint{3.316659in}{0.442177in}}%
\pgfpathlineto{\pgfqpoint{3.347724in}{0.442177in}}%
\pgfpathlineto{\pgfqpoint{3.378789in}{0.442177in}}%
\pgfpathlineto{\pgfqpoint{3.409854in}{0.442177in}}%
\pgfpathlineto{\pgfqpoint{3.440920in}{0.442177in}}%
\pgfpathlineto{\pgfqpoint{3.471985in}{0.442177in}}%
\pgfpathlineto{\pgfqpoint{3.503050in}{0.442177in}}%
\pgfpathlineto{\pgfqpoint{3.534115in}{0.442177in}}%
\pgfpathlineto{\pgfqpoint{3.565180in}{0.442177in}}%
\pgfpathlineto{\pgfqpoint{3.596245in}{0.442177in}}%
\pgfpathlineto{\pgfqpoint{3.596245in}{0.442177in}}%
\pgfpathlineto{\pgfqpoint{3.596245in}{0.442177in}}%
\pgfpathlineto{\pgfqpoint{3.565180in}{0.442177in}}%
\pgfpathlineto{\pgfqpoint{3.534115in}{0.442177in}}%
\pgfpathlineto{\pgfqpoint{3.503050in}{0.442177in}}%
\pgfpathlineto{\pgfqpoint{3.471985in}{0.442177in}}%
\pgfpathlineto{\pgfqpoint{3.440920in}{0.442177in}}%
\pgfpathlineto{\pgfqpoint{3.409854in}{0.442177in}}%
\pgfpathlineto{\pgfqpoint{3.378789in}{0.442177in}}%
\pgfpathlineto{\pgfqpoint{3.347724in}{0.442177in}}%
\pgfpathlineto{\pgfqpoint{3.316659in}{0.442177in}}%
\pgfpathlineto{\pgfqpoint{3.285594in}{0.442177in}}%
\pgfpathlineto{\pgfqpoint{3.254529in}{0.442177in}}%
\pgfpathlineto{\pgfqpoint{3.223464in}{0.442177in}}%
\pgfpathlineto{\pgfqpoint{3.192399in}{0.442177in}}%
\pgfpathlineto{\pgfqpoint{3.161334in}{0.442177in}}%
\pgfpathlineto{\pgfqpoint{3.130268in}{0.442177in}}%
\pgfpathlineto{\pgfqpoint{3.099203in}{0.442177in}}%
\pgfpathlineto{\pgfqpoint{3.068138in}{0.442177in}}%
\pgfpathlineto{\pgfqpoint{3.037073in}{0.442177in}}%
\pgfpathlineto{\pgfqpoint{3.006008in}{0.442177in}}%
\pgfpathlineto{\pgfqpoint{2.974943in}{0.442177in}}%
\pgfpathlineto{\pgfqpoint{2.943878in}{0.442177in}}%
\pgfpathlineto{\pgfqpoint{2.912813in}{0.442177in}}%
\pgfpathlineto{\pgfqpoint{2.881747in}{0.442177in}}%
\pgfpathlineto{\pgfqpoint{2.850682in}{0.442177in}}%
\pgfpathlineto{\pgfqpoint{2.819617in}{0.442177in}}%
\pgfpathlineto{\pgfqpoint{2.788552in}{0.442177in}}%
\pgfpathlineto{\pgfqpoint{2.757487in}{0.442177in}}%
\pgfpathlineto{\pgfqpoint{2.726422in}{0.442177in}}%
\pgfpathlineto{\pgfqpoint{2.695357in}{0.442177in}}%
\pgfpathlineto{\pgfqpoint{2.664292in}{0.442177in}}%
\pgfpathlineto{\pgfqpoint{2.633226in}{0.442177in}}%
\pgfpathlineto{\pgfqpoint{2.602161in}{0.442177in}}%
\pgfpathlineto{\pgfqpoint{2.571096in}{0.442177in}}%
\pgfpathlineto{\pgfqpoint{2.540031in}{0.442177in}}%
\pgfpathlineto{\pgfqpoint{2.508966in}{0.442177in}}%
\pgfpathlineto{\pgfqpoint{2.477901in}{0.553778in}}%
\pgfpathlineto{\pgfqpoint{2.446836in}{0.553778in}}%
\pgfpathlineto{\pgfqpoint{2.415771in}{0.588224in}}%
\pgfpathlineto{\pgfqpoint{2.384705in}{0.612546in}}%
\pgfpathlineto{\pgfqpoint{2.353640in}{0.626658in}}%
\pgfpathlineto{\pgfqpoint{2.322575in}{0.649029in}}%
\pgfpathlineto{\pgfqpoint{2.291510in}{0.664269in}}%
\pgfpathlineto{\pgfqpoint{2.260445in}{0.675093in}}%
\pgfpathlineto{\pgfqpoint{2.229380in}{0.689577in}}%
\pgfpathlineto{\pgfqpoint{2.198315in}{0.698718in}}%
\pgfpathlineto{\pgfqpoint{2.167250in}{0.711026in}}%
\pgfpathlineto{\pgfqpoint{2.136184in}{0.722904in}}%
\pgfpathlineto{\pgfqpoint{2.105119in}{0.735599in}}%
\pgfpathlineto{\pgfqpoint{2.074054in}{0.747704in}}%
\pgfpathlineto{\pgfqpoint{2.042989in}{0.761104in}}%
\pgfpathlineto{\pgfqpoint{2.011924in}{0.773797in}}%
\pgfpathlineto{\pgfqpoint{1.980859in}{0.786251in}}%
\pgfpathlineto{\pgfqpoint{1.949794in}{0.798413in}}%
\pgfpathlineto{\pgfqpoint{1.918729in}{0.818361in}}%
\pgfpathlineto{\pgfqpoint{1.887663in}{0.835703in}}%
\pgfpathlineto{\pgfqpoint{1.856598in}{0.853117in}}%
\pgfpathlineto{\pgfqpoint{1.825533in}{0.867825in}}%
\pgfpathlineto{\pgfqpoint{1.794468in}{0.883617in}}%
\pgfpathlineto{\pgfqpoint{1.763403in}{0.901116in}}%
\pgfpathlineto{\pgfqpoint{1.732338in}{0.916575in}}%
\pgfpathlineto{\pgfqpoint{1.701273in}{0.932488in}}%
\pgfpathlineto{\pgfqpoint{1.670208in}{0.948023in}}%
\pgfpathlineto{\pgfqpoint{1.639142in}{0.961946in}}%
\pgfpathlineto{\pgfqpoint{1.608077in}{0.979586in}}%
\pgfpathlineto{\pgfqpoint{1.577012in}{0.993152in}}%
\pgfpathlineto{\pgfqpoint{1.545947in}{1.010441in}}%
\pgfpathlineto{\pgfqpoint{1.514882in}{1.024006in}}%
\pgfpathlineto{\pgfqpoint{1.483817in}{1.039382in}}%
\pgfpathlineto{\pgfqpoint{1.452752in}{1.052862in}}%
\pgfpathlineto{\pgfqpoint{1.421687in}{1.066383in}}%
\pgfpathlineto{\pgfqpoint{1.390621in}{1.079029in}}%
\pgfpathlineto{\pgfqpoint{1.359556in}{1.091625in}}%
\pgfpathlineto{\pgfqpoint{1.328491in}{1.103992in}}%
\pgfpathlineto{\pgfqpoint{1.297426in}{1.117773in}}%
\pgfpathlineto{\pgfqpoint{1.266361in}{1.127330in}}%
\pgfpathlineto{\pgfqpoint{1.235296in}{1.142109in}}%
\pgfpathlineto{\pgfqpoint{1.204231in}{1.155519in}}%
\pgfpathlineto{\pgfqpoint{1.173166in}{1.167679in}}%
\pgfpathlineto{\pgfqpoint{1.142100in}{1.180497in}}%
\pgfpathlineto{\pgfqpoint{1.111035in}{1.194409in}}%
\pgfpathlineto{\pgfqpoint{1.079970in}{1.205402in}}%
\pgfpathlineto{\pgfqpoint{1.048905in}{1.218456in}}%
\pgfpathlineto{\pgfqpoint{1.017840in}{1.228175in}}%
\pgfpathlineto{\pgfqpoint{0.986775in}{1.241055in}}%
\pgfpathlineto{\pgfqpoint{0.955710in}{1.250651in}}%
\pgfpathlineto{\pgfqpoint{0.924645in}{1.262532in}}%
\pgfpathlineto{\pgfqpoint{0.893579in}{1.272708in}}%
\pgfpathlineto{\pgfqpoint{0.862514in}{1.282540in}}%
\pgfpathlineto{\pgfqpoint{0.831449in}{1.292909in}}%
\pgfpathlineto{\pgfqpoint{0.800384in}{1.301775in}}%
\pgfpathlineto{\pgfqpoint{0.769319in}{1.310809in}}%
\pgfpathlineto{\pgfqpoint{0.738254in}{1.318467in}}%
\pgfpathlineto{\pgfqpoint{0.707189in}{1.325945in}}%
\pgfpathlineto{\pgfqpoint{0.676124in}{1.332884in}}%
\pgfpathlineto{\pgfqpoint{0.645058in}{1.339523in}}%
\pgfpathlineto{\pgfqpoint{0.613993in}{1.348635in}}%
\pgfpathlineto{\pgfqpoint{0.582928in}{1.356852in}}%
\pgfpathlineto{\pgfqpoint{0.551863in}{1.363172in}}%
\pgfpathlineto{\pgfqpoint{0.520798in}{1.370802in}}%
\pgfpathlineto{\pgfqpoint{0.520798in}{1.370802in}}%
\pgfpathclose%
\pgfusepath{fill}%
\end{pgfscope}%
\begin{pgfscope}%
\pgfpathrectangle{\pgfqpoint{0.520798in}{0.442177in}}{\pgfqpoint{2.152813in}{1.142908in}}%
\pgfusepath{clip}%
\pgfsetroundcap%
\pgfsetroundjoin%
\pgfsetlinewidth{1.003750pt}%
\definecolor{currentstroke}{rgb}{0.870588,0.560784,0.019608}%
\pgfsetstrokecolor{currentstroke}%
\pgfsetdash{}{0pt}%
\pgfpathmoveto{\pgfqpoint{0.520798in}{1.394199in}}%
\pgfpathlineto{\pgfqpoint{0.551863in}{1.388642in}}%
\pgfpathlineto{\pgfqpoint{0.582928in}{1.383363in}}%
\pgfpathlineto{\pgfqpoint{0.613993in}{1.377991in}}%
\pgfpathlineto{\pgfqpoint{0.645058in}{1.370767in}}%
\pgfpathlineto{\pgfqpoint{0.676124in}{1.364191in}}%
\pgfpathlineto{\pgfqpoint{0.707189in}{1.357522in}}%
\pgfpathlineto{\pgfqpoint{0.738254in}{1.350298in}}%
\pgfpathlineto{\pgfqpoint{0.769319in}{1.344556in}}%
\pgfpathlineto{\pgfqpoint{0.800384in}{1.338073in}}%
\pgfpathlineto{\pgfqpoint{0.831449in}{1.331404in}}%
\pgfpathlineto{\pgfqpoint{0.862514in}{1.324272in}}%
\pgfpathlineto{\pgfqpoint{0.893579in}{1.315474in}}%
\pgfpathlineto{\pgfqpoint{0.924645in}{1.305564in}}%
\pgfpathlineto{\pgfqpoint{0.955710in}{1.296857in}}%
\pgfpathlineto{\pgfqpoint{0.986775in}{1.286669in}}%
\pgfpathlineto{\pgfqpoint{1.017840in}{1.275926in}}%
\pgfpathlineto{\pgfqpoint{1.048905in}{1.266386in}}%
\pgfpathlineto{\pgfqpoint{1.079970in}{1.255179in}}%
\pgfpathlineto{\pgfqpoint{1.111035in}{1.244806in}}%
\pgfpathlineto{\pgfqpoint{1.142100in}{1.234062in}}%
\pgfpathlineto{\pgfqpoint{1.173166in}{1.226097in}}%
\pgfpathlineto{\pgfqpoint{1.204231in}{1.217576in}}%
\pgfpathlineto{\pgfqpoint{1.235296in}{1.207666in}}%
\pgfpathlineto{\pgfqpoint{1.266361in}{1.197015in}}%
\pgfpathlineto{\pgfqpoint{1.297426in}{1.185438in}}%
\pgfpathlineto{\pgfqpoint{1.328491in}{1.176176in}}%
\pgfpathlineto{\pgfqpoint{1.359556in}{1.165154in}}%
\pgfpathlineto{\pgfqpoint{1.390621in}{1.153670in}}%
\pgfpathlineto{\pgfqpoint{1.421687in}{1.144778in}}%
\pgfpathlineto{\pgfqpoint{1.452752in}{1.134683in}}%
\pgfpathlineto{\pgfqpoint{1.483817in}{1.123847in}}%
\pgfpathlineto{\pgfqpoint{1.514882in}{1.110510in}}%
\pgfpathlineto{\pgfqpoint{1.545947in}{1.098747in}}%
\pgfpathlineto{\pgfqpoint{1.577012in}{1.088744in}}%
\pgfpathlineto{\pgfqpoint{1.608077in}{1.077260in}}%
\pgfpathlineto{\pgfqpoint{1.639142in}{1.065960in}}%
\pgfpathlineto{\pgfqpoint{1.670208in}{1.052253in}}%
\pgfpathlineto{\pgfqpoint{1.701273in}{1.039935in}}%
\pgfpathlineto{\pgfqpoint{1.732338in}{1.025764in}}%
\pgfpathlineto{\pgfqpoint{1.763403in}{1.012612in}}%
\pgfpathlineto{\pgfqpoint{1.794468in}{0.999090in}}%
\pgfpathlineto{\pgfqpoint{1.825533in}{0.985938in}}%
\pgfpathlineto{\pgfqpoint{1.856598in}{0.972972in}}%
\pgfpathlineto{\pgfqpoint{1.887663in}{0.958986in}}%
\pgfpathlineto{\pgfqpoint{1.918729in}{0.943426in}}%
\pgfpathlineto{\pgfqpoint{1.949794in}{0.928422in}}%
\pgfpathlineto{\pgfqpoint{1.980859in}{0.914900in}}%
\pgfpathlineto{\pgfqpoint{2.011924in}{0.897673in}}%
\pgfpathlineto{\pgfqpoint{2.042989in}{0.881928in}}%
\pgfpathlineto{\pgfqpoint{2.074054in}{0.866275in}}%
\pgfpathlineto{\pgfqpoint{2.105119in}{0.849512in}}%
\pgfpathlineto{\pgfqpoint{2.136184in}{0.832470in}}%
\pgfpathlineto{\pgfqpoint{2.167250in}{0.818762in}}%
\pgfpathlineto{\pgfqpoint{2.198315in}{0.803758in}}%
\pgfpathlineto{\pgfqpoint{2.229380in}{0.791070in}}%
\pgfpathlineto{\pgfqpoint{2.260445in}{0.775510in}}%
\pgfpathlineto{\pgfqpoint{2.291510in}{0.763099in}}%
\pgfpathlineto{\pgfqpoint{2.322575in}{0.749669in}}%
\pgfpathlineto{\pgfqpoint{2.353640in}{0.728552in}}%
\pgfpathlineto{\pgfqpoint{2.384705in}{0.716697in}}%
\pgfpathlineto{\pgfqpoint{2.415771in}{0.697988in}}%
\pgfpathlineto{\pgfqpoint{2.446836in}{0.660663in}}%
\pgfpathlineto{\pgfqpoint{2.477901in}{0.660663in}}%
\pgfpathlineto{\pgfqpoint{2.508966in}{0.442177in}}%
\pgfpathlineto{\pgfqpoint{2.540031in}{0.442177in}}%
\pgfpathlineto{\pgfqpoint{2.571096in}{0.442177in}}%
\pgfpathlineto{\pgfqpoint{2.602161in}{0.442177in}}%
\pgfpathlineto{\pgfqpoint{2.633226in}{0.442177in}}%
\pgfpathlineto{\pgfqpoint{2.664292in}{0.442177in}}%
\pgfpathlineto{\pgfqpoint{2.675278in}{0.442177in}}%
\pgfusepath{stroke}%
\end{pgfscope}%
\begin{pgfscope}%
\pgfpathrectangle{\pgfqpoint{0.520798in}{0.442177in}}{\pgfqpoint{2.152813in}{1.142908in}}%
\pgfusepath{clip}%
\pgfsetbuttcap%
\pgfsetroundjoin%
\definecolor{currentfill}{rgb}{0.870588,0.560784,0.019608}%
\pgfsetfillcolor{currentfill}%
\pgfsetfillopacity{0.500000}%
\pgfsetlinewidth{0.000000pt}%
\definecolor{currentstroke}{rgb}{0.870588,0.560784,0.019608}%
\pgfsetstrokecolor{currentstroke}%
\pgfsetstrokeopacity{0.500000}%
\pgfsetdash{}{0pt}%
\pgfpathmoveto{\pgfqpoint{0.520798in}{1.397767in}}%
\pgfpathlineto{\pgfqpoint{0.520798in}{1.390631in}}%
\pgfpathlineto{\pgfqpoint{0.551863in}{1.385298in}}%
\pgfpathlineto{\pgfqpoint{0.582928in}{1.379940in}}%
\pgfpathlineto{\pgfqpoint{0.613993in}{1.374984in}}%
\pgfpathlineto{\pgfqpoint{0.645058in}{1.367309in}}%
\pgfpathlineto{\pgfqpoint{0.676124in}{1.359866in}}%
\pgfpathlineto{\pgfqpoint{0.707189in}{1.352812in}}%
\pgfpathlineto{\pgfqpoint{0.738254in}{1.344274in}}%
\pgfpathlineto{\pgfqpoint{0.769319in}{1.339056in}}%
\pgfpathlineto{\pgfqpoint{0.800384in}{1.332565in}}%
\pgfpathlineto{\pgfqpoint{0.831449in}{1.325953in}}%
\pgfpathlineto{\pgfqpoint{0.862514in}{1.318877in}}%
\pgfpathlineto{\pgfqpoint{0.893579in}{1.308831in}}%
\pgfpathlineto{\pgfqpoint{0.924645in}{1.298197in}}%
\pgfpathlineto{\pgfqpoint{0.955710in}{1.288406in}}%
\pgfpathlineto{\pgfqpoint{0.986775in}{1.276830in}}%
\pgfpathlineto{\pgfqpoint{1.017840in}{1.264767in}}%
\pgfpathlineto{\pgfqpoint{1.048905in}{1.257081in}}%
\pgfpathlineto{\pgfqpoint{1.079970in}{1.246322in}}%
\pgfpathlineto{\pgfqpoint{1.111035in}{1.235612in}}%
\pgfpathlineto{\pgfqpoint{1.142100in}{1.224940in}}%
\pgfpathlineto{\pgfqpoint{1.173166in}{1.217506in}}%
\pgfpathlineto{\pgfqpoint{1.204231in}{1.207641in}}%
\pgfpathlineto{\pgfqpoint{1.235296in}{1.200176in}}%
\pgfpathlineto{\pgfqpoint{1.266361in}{1.189151in}}%
\pgfpathlineto{\pgfqpoint{1.297426in}{1.177180in}}%
\pgfpathlineto{\pgfqpoint{1.328491in}{1.168567in}}%
\pgfpathlineto{\pgfqpoint{1.359556in}{1.155703in}}%
\pgfpathlineto{\pgfqpoint{1.390621in}{1.142573in}}%
\pgfpathlineto{\pgfqpoint{1.421687in}{1.134589in}}%
\pgfpathlineto{\pgfqpoint{1.452752in}{1.125345in}}%
\pgfpathlineto{\pgfqpoint{1.483817in}{1.112682in}}%
\pgfpathlineto{\pgfqpoint{1.514882in}{1.097489in}}%
\pgfpathlineto{\pgfqpoint{1.545947in}{1.084875in}}%
\pgfpathlineto{\pgfqpoint{1.577012in}{1.072574in}}%
\pgfpathlineto{\pgfqpoint{1.608077in}{1.060742in}}%
\pgfpathlineto{\pgfqpoint{1.639142in}{1.050727in}}%
\pgfpathlineto{\pgfqpoint{1.670208in}{1.036463in}}%
\pgfpathlineto{\pgfqpoint{1.701273in}{1.022400in}}%
\pgfpathlineto{\pgfqpoint{1.732338in}{1.007149in}}%
\pgfpathlineto{\pgfqpoint{1.763403in}{0.993296in}}%
\pgfpathlineto{\pgfqpoint{1.794468in}{0.978140in}}%
\pgfpathlineto{\pgfqpoint{1.825533in}{0.964292in}}%
\pgfpathlineto{\pgfqpoint{1.856598in}{0.950055in}}%
\pgfpathlineto{\pgfqpoint{1.887663in}{0.938667in}}%
\pgfpathlineto{\pgfqpoint{1.918729in}{0.924950in}}%
\pgfpathlineto{\pgfqpoint{1.949794in}{0.910866in}}%
\pgfpathlineto{\pgfqpoint{1.980859in}{0.898321in}}%
\pgfpathlineto{\pgfqpoint{2.011924in}{0.879880in}}%
\pgfpathlineto{\pgfqpoint{2.042989in}{0.865132in}}%
\pgfpathlineto{\pgfqpoint{2.074054in}{0.851683in}}%
\pgfpathlineto{\pgfqpoint{2.105119in}{0.835913in}}%
\pgfpathlineto{\pgfqpoint{2.136184in}{0.821155in}}%
\pgfpathlineto{\pgfqpoint{2.167250in}{0.807712in}}%
\pgfpathlineto{\pgfqpoint{2.198315in}{0.793794in}}%
\pgfpathlineto{\pgfqpoint{2.229380in}{0.779604in}}%
\pgfpathlineto{\pgfqpoint{2.260445in}{0.767172in}}%
\pgfpathlineto{\pgfqpoint{2.291510in}{0.754048in}}%
\pgfpathlineto{\pgfqpoint{2.322575in}{0.740594in}}%
\pgfpathlineto{\pgfqpoint{2.353640in}{0.717112in}}%
\pgfpathlineto{\pgfqpoint{2.384705in}{0.704304in}}%
\pgfpathlineto{\pgfqpoint{2.415771in}{0.685967in}}%
\pgfpathlineto{\pgfqpoint{2.446836in}{0.646479in}}%
\pgfpathlineto{\pgfqpoint{2.477901in}{0.646479in}}%
\pgfpathlineto{\pgfqpoint{2.508966in}{0.442177in}}%
\pgfpathlineto{\pgfqpoint{2.540031in}{0.442177in}}%
\pgfpathlineto{\pgfqpoint{2.571096in}{0.442177in}}%
\pgfpathlineto{\pgfqpoint{2.602161in}{0.442177in}}%
\pgfpathlineto{\pgfqpoint{2.633226in}{0.442177in}}%
\pgfpathlineto{\pgfqpoint{2.664292in}{0.442177in}}%
\pgfpathlineto{\pgfqpoint{2.695357in}{0.442177in}}%
\pgfpathlineto{\pgfqpoint{2.726422in}{0.442177in}}%
\pgfpathlineto{\pgfqpoint{2.757487in}{0.442177in}}%
\pgfpathlineto{\pgfqpoint{2.788552in}{0.442177in}}%
\pgfpathlineto{\pgfqpoint{2.819617in}{0.442177in}}%
\pgfpathlineto{\pgfqpoint{2.850682in}{0.442177in}}%
\pgfpathlineto{\pgfqpoint{2.881747in}{0.442177in}}%
\pgfpathlineto{\pgfqpoint{2.912813in}{0.442177in}}%
\pgfpathlineto{\pgfqpoint{2.943878in}{0.442177in}}%
\pgfpathlineto{\pgfqpoint{2.974943in}{0.442177in}}%
\pgfpathlineto{\pgfqpoint{3.006008in}{0.442177in}}%
\pgfpathlineto{\pgfqpoint{3.037073in}{0.442177in}}%
\pgfpathlineto{\pgfqpoint{3.068138in}{0.442177in}}%
\pgfpathlineto{\pgfqpoint{3.099203in}{0.442177in}}%
\pgfpathlineto{\pgfqpoint{3.130268in}{0.442177in}}%
\pgfpathlineto{\pgfqpoint{3.161334in}{0.442177in}}%
\pgfpathlineto{\pgfqpoint{3.192399in}{0.442177in}}%
\pgfpathlineto{\pgfqpoint{3.223464in}{0.442177in}}%
\pgfpathlineto{\pgfqpoint{3.254529in}{0.442177in}}%
\pgfpathlineto{\pgfqpoint{3.285594in}{0.442177in}}%
\pgfpathlineto{\pgfqpoint{3.316659in}{0.442177in}}%
\pgfpathlineto{\pgfqpoint{3.347724in}{0.442177in}}%
\pgfpathlineto{\pgfqpoint{3.378789in}{0.442177in}}%
\pgfpathlineto{\pgfqpoint{3.409854in}{0.442177in}}%
\pgfpathlineto{\pgfqpoint{3.440920in}{0.442177in}}%
\pgfpathlineto{\pgfqpoint{3.471985in}{0.442177in}}%
\pgfpathlineto{\pgfqpoint{3.503050in}{0.442177in}}%
\pgfpathlineto{\pgfqpoint{3.534115in}{0.442177in}}%
\pgfpathlineto{\pgfqpoint{3.565180in}{0.442177in}}%
\pgfpathlineto{\pgfqpoint{3.596245in}{0.442177in}}%
\pgfpathlineto{\pgfqpoint{3.596245in}{0.442177in}}%
\pgfpathlineto{\pgfqpoint{3.596245in}{0.442177in}}%
\pgfpathlineto{\pgfqpoint{3.565180in}{0.442177in}}%
\pgfpathlineto{\pgfqpoint{3.534115in}{0.442177in}}%
\pgfpathlineto{\pgfqpoint{3.503050in}{0.442177in}}%
\pgfpathlineto{\pgfqpoint{3.471985in}{0.442177in}}%
\pgfpathlineto{\pgfqpoint{3.440920in}{0.442177in}}%
\pgfpathlineto{\pgfqpoint{3.409854in}{0.442177in}}%
\pgfpathlineto{\pgfqpoint{3.378789in}{0.442177in}}%
\pgfpathlineto{\pgfqpoint{3.347724in}{0.442177in}}%
\pgfpathlineto{\pgfqpoint{3.316659in}{0.442177in}}%
\pgfpathlineto{\pgfqpoint{3.285594in}{0.442177in}}%
\pgfpathlineto{\pgfqpoint{3.254529in}{0.442177in}}%
\pgfpathlineto{\pgfqpoint{3.223464in}{0.442177in}}%
\pgfpathlineto{\pgfqpoint{3.192399in}{0.442177in}}%
\pgfpathlineto{\pgfqpoint{3.161334in}{0.442177in}}%
\pgfpathlineto{\pgfqpoint{3.130268in}{0.442177in}}%
\pgfpathlineto{\pgfqpoint{3.099203in}{0.442177in}}%
\pgfpathlineto{\pgfqpoint{3.068138in}{0.442177in}}%
\pgfpathlineto{\pgfqpoint{3.037073in}{0.442177in}}%
\pgfpathlineto{\pgfqpoint{3.006008in}{0.442177in}}%
\pgfpathlineto{\pgfqpoint{2.974943in}{0.442177in}}%
\pgfpathlineto{\pgfqpoint{2.943878in}{0.442177in}}%
\pgfpathlineto{\pgfqpoint{2.912813in}{0.442177in}}%
\pgfpathlineto{\pgfqpoint{2.881747in}{0.442177in}}%
\pgfpathlineto{\pgfqpoint{2.850682in}{0.442177in}}%
\pgfpathlineto{\pgfqpoint{2.819617in}{0.442177in}}%
\pgfpathlineto{\pgfqpoint{2.788552in}{0.442177in}}%
\pgfpathlineto{\pgfqpoint{2.757487in}{0.442177in}}%
\pgfpathlineto{\pgfqpoint{2.726422in}{0.442177in}}%
\pgfpathlineto{\pgfqpoint{2.695357in}{0.442177in}}%
\pgfpathlineto{\pgfqpoint{2.664292in}{0.442177in}}%
\pgfpathlineto{\pgfqpoint{2.633226in}{0.442177in}}%
\pgfpathlineto{\pgfqpoint{2.602161in}{0.442177in}}%
\pgfpathlineto{\pgfqpoint{2.571096in}{0.442177in}}%
\pgfpathlineto{\pgfqpoint{2.540031in}{0.442177in}}%
\pgfpathlineto{\pgfqpoint{2.508966in}{0.442177in}}%
\pgfpathlineto{\pgfqpoint{2.477901in}{0.674847in}}%
\pgfpathlineto{\pgfqpoint{2.446836in}{0.674847in}}%
\pgfpathlineto{\pgfqpoint{2.415771in}{0.710009in}}%
\pgfpathlineto{\pgfqpoint{2.384705in}{0.729090in}}%
\pgfpathlineto{\pgfqpoint{2.353640in}{0.739992in}}%
\pgfpathlineto{\pgfqpoint{2.322575in}{0.758744in}}%
\pgfpathlineto{\pgfqpoint{2.291510in}{0.772150in}}%
\pgfpathlineto{\pgfqpoint{2.260445in}{0.783847in}}%
\pgfpathlineto{\pgfqpoint{2.229380in}{0.802535in}}%
\pgfpathlineto{\pgfqpoint{2.198315in}{0.813722in}}%
\pgfpathlineto{\pgfqpoint{2.167250in}{0.829812in}}%
\pgfpathlineto{\pgfqpoint{2.136184in}{0.843784in}}%
\pgfpathlineto{\pgfqpoint{2.105119in}{0.863110in}}%
\pgfpathlineto{\pgfqpoint{2.074054in}{0.880868in}}%
\pgfpathlineto{\pgfqpoint{2.042989in}{0.898724in}}%
\pgfpathlineto{\pgfqpoint{2.011924in}{0.915466in}}%
\pgfpathlineto{\pgfqpoint{1.980859in}{0.931479in}}%
\pgfpathlineto{\pgfqpoint{1.949794in}{0.945978in}}%
\pgfpathlineto{\pgfqpoint{1.918729in}{0.961903in}}%
\pgfpathlineto{\pgfqpoint{1.887663in}{0.979305in}}%
\pgfpathlineto{\pgfqpoint{1.856598in}{0.995889in}}%
\pgfpathlineto{\pgfqpoint{1.825533in}{1.007585in}}%
\pgfpathlineto{\pgfqpoint{1.794468in}{1.020040in}}%
\pgfpathlineto{\pgfqpoint{1.763403in}{1.031928in}}%
\pgfpathlineto{\pgfqpoint{1.732338in}{1.044379in}}%
\pgfpathlineto{\pgfqpoint{1.701273in}{1.057470in}}%
\pgfpathlineto{\pgfqpoint{1.670208in}{1.068043in}}%
\pgfpathlineto{\pgfqpoint{1.639142in}{1.081193in}}%
\pgfpathlineto{\pgfqpoint{1.608077in}{1.093777in}}%
\pgfpathlineto{\pgfqpoint{1.577012in}{1.104915in}}%
\pgfpathlineto{\pgfqpoint{1.545947in}{1.112619in}}%
\pgfpathlineto{\pgfqpoint{1.514882in}{1.123530in}}%
\pgfpathlineto{\pgfqpoint{1.483817in}{1.135011in}}%
\pgfpathlineto{\pgfqpoint{1.452752in}{1.144021in}}%
\pgfpathlineto{\pgfqpoint{1.421687in}{1.154968in}}%
\pgfpathlineto{\pgfqpoint{1.390621in}{1.164767in}}%
\pgfpathlineto{\pgfqpoint{1.359556in}{1.174606in}}%
\pgfpathlineto{\pgfqpoint{1.328491in}{1.183785in}}%
\pgfpathlineto{\pgfqpoint{1.297426in}{1.193696in}}%
\pgfpathlineto{\pgfqpoint{1.266361in}{1.204879in}}%
\pgfpathlineto{\pgfqpoint{1.235296in}{1.215156in}}%
\pgfpathlineto{\pgfqpoint{1.204231in}{1.227511in}}%
\pgfpathlineto{\pgfqpoint{1.173166in}{1.234688in}}%
\pgfpathlineto{\pgfqpoint{1.142100in}{1.243184in}}%
\pgfpathlineto{\pgfqpoint{1.111035in}{1.254000in}}%
\pgfpathlineto{\pgfqpoint{1.079970in}{1.264037in}}%
\pgfpathlineto{\pgfqpoint{1.048905in}{1.275691in}}%
\pgfpathlineto{\pgfqpoint{1.017840in}{1.287085in}}%
\pgfpathlineto{\pgfqpoint{0.986775in}{1.296509in}}%
\pgfpathlineto{\pgfqpoint{0.955710in}{1.305309in}}%
\pgfpathlineto{\pgfqpoint{0.924645in}{1.312930in}}%
\pgfpathlineto{\pgfqpoint{0.893579in}{1.322116in}}%
\pgfpathlineto{\pgfqpoint{0.862514in}{1.329668in}}%
\pgfpathlineto{\pgfqpoint{0.831449in}{1.336855in}}%
\pgfpathlineto{\pgfqpoint{0.800384in}{1.343580in}}%
\pgfpathlineto{\pgfqpoint{0.769319in}{1.350055in}}%
\pgfpathlineto{\pgfqpoint{0.738254in}{1.356322in}}%
\pgfpathlineto{\pgfqpoint{0.707189in}{1.362232in}}%
\pgfpathlineto{\pgfqpoint{0.676124in}{1.368515in}}%
\pgfpathlineto{\pgfqpoint{0.645058in}{1.374225in}}%
\pgfpathlineto{\pgfqpoint{0.613993in}{1.380998in}}%
\pgfpathlineto{\pgfqpoint{0.582928in}{1.386786in}}%
\pgfpathlineto{\pgfqpoint{0.551863in}{1.391987in}}%
\pgfpathlineto{\pgfqpoint{0.520798in}{1.397767in}}%
\pgfpathlineto{\pgfqpoint{0.520798in}{1.397767in}}%
\pgfpathclose%
\pgfusepath{fill}%
\end{pgfscope}%
\begin{pgfscope}%
\pgfsetbuttcap%
\pgfsetmiterjoin%
\definecolor{currentfill}{rgb}{1.000000,1.000000,1.000000}%
\pgfsetfillcolor{currentfill}%
\pgfsetfillopacity{0.800000}%
\pgfsetlinewidth{1.003750pt}%
\definecolor{currentstroke}{rgb}{0.800000,0.800000,0.800000}%
\pgfsetstrokecolor{currentstroke}%
\pgfsetstrokeopacity{0.800000}%
\pgfsetdash{}{0pt}%
\pgfpathmoveto{\pgfqpoint{1.754318in}{1.186330in}}%
\pgfpathlineto{\pgfqpoint{2.618055in}{1.186330in}}%
\pgfpathlineto{\pgfqpoint{2.618055in}{1.529530in}}%
\pgfpathlineto{\pgfqpoint{1.754318in}{1.529530in}}%
\pgfpathlineto{\pgfqpoint{1.754318in}{1.186330in}}%
\pgfpathclose%
\pgfusepath{stroke,fill}%
\end{pgfscope}%
\begin{pgfscope}%
\pgfsetroundcap%
\pgfsetroundjoin%
\pgfsetlinewidth{1.003750pt}%
\definecolor{currentstroke}{rgb}{0.003922,0.450980,0.698039}%
\pgfsetstrokecolor{currentstroke}%
\pgfsetdash{}{0pt}%
\pgfpathmoveto{\pgfqpoint{1.798763in}{1.446196in}}%
\pgfpathlineto{\pgfqpoint{1.909874in}{1.446196in}}%
\pgfpathlineto{\pgfqpoint{2.020985in}{1.446196in}}%
\pgfusepath{stroke}%
\end{pgfscope}%
\begin{pgfscope}%
\definecolor{textcolor}{rgb}{0.150000,0.150000,0.150000}%
\pgfsetstrokecolor{textcolor}%
\pgfsetfillcolor{textcolor}%
\pgftext[x=2.109874in,y=1.407307in,left,base]{\color{textcolor}\rmfamily\fontsize{8.000000}{9.600000}\selectfont PointNet}%
\end{pgfscope}%
\begin{pgfscope}%
\pgfsetroundcap%
\pgfsetroundjoin%
\pgfsetlinewidth{1.003750pt}%
\definecolor{currentstroke}{rgb}{0.870588,0.560784,0.019608}%
\pgfsetstrokecolor{currentstroke}%
\pgfsetdash{}{0pt}%
\pgfpathmoveto{\pgfqpoint{1.798763in}{1.291263in}}%
\pgfpathlineto{\pgfqpoint{1.909874in}{1.291263in}}%
\pgfpathlineto{\pgfqpoint{2.020985in}{1.291263in}}%
\pgfusepath{stroke}%
\end{pgfscope}%
\begin{pgfscope}%
\definecolor{textcolor}{rgb}{0.150000,0.150000,0.150000}%
\pgfsetstrokecolor{textcolor}%
\pgfsetfillcolor{textcolor}%
\pgftext[x=2.109874in,y=1.252374in,left,base]{\color{textcolor}\rmfamily\fontsize{8.000000}{9.600000}\selectfont DGCNN}%
\end{pgfscope}%
\end{pgfpicture}%
\makeatother%
\endgroup%

%% file: figures/experiments/graphs/sparse_smoothing/nodes_structure/node_classification-Cora-GCN-hidden=32-p_adj_plus=0.001-p_adj_minus=0.8-p_att_plus=0.0-p_att_minus=0.0-multi_class_cert-B.pgf
\begingroup%
\makeatletter%
\begin{pgfpicture}%
\pgfpathrectangle{\pgfpointorigin}{\pgfqpoint{1.375000in}{1.581250in}}%
\pgfusepath{use as bounding box, clip}%
\begin{pgfscope}%
\pgfsetbuttcap%
\pgfsetmiterjoin%
\definecolor{currentfill}{rgb}{1.000000,1.000000,1.000000}%
\pgfsetfillcolor{currentfill}%
\pgfsetlinewidth{0.000000pt}%
\definecolor{currentstroke}{rgb}{1.000000,1.000000,1.000000}%
\pgfsetstrokecolor{currentstroke}%
\pgfsetdash{}{0pt}%
\pgfpathmoveto{\pgfqpoint{0.000000in}{0.000000in}}%
\pgfpathlineto{\pgfqpoint{1.375000in}{0.000000in}}%
\pgfpathlineto{\pgfqpoint{1.375000in}{1.581250in}}%
\pgfpathlineto{\pgfqpoint{0.000000in}{1.581250in}}%
\pgfpathlineto{\pgfqpoint{0.000000in}{0.000000in}}%
\pgfpathclose%
\pgfusepath{fill}%
\end{pgfscope}%
\begin{pgfscope}%
\pgfsetbuttcap%
\pgfsetmiterjoin%
\definecolor{currentfill}{rgb}{1.000000,1.000000,1.000000}%
\pgfsetfillcolor{currentfill}%
\pgfsetlinewidth{0.000000pt}%
\definecolor{currentstroke}{rgb}{0.000000,0.000000,0.000000}%
\pgfsetstrokecolor{currentstroke}%
\pgfsetstrokeopacity{0.000000}%
\pgfsetdash{}{0pt}%
\pgfpathmoveto{\pgfqpoint{0.520798in}{0.442177in}}%
\pgfpathlineto{\pgfqpoint{1.239803in}{0.442177in}}%
\pgfpathlineto{\pgfqpoint{1.239803in}{1.314061in}}%
\pgfpathlineto{\pgfqpoint{0.520798in}{1.314061in}}%
\pgfpathlineto{\pgfqpoint{0.520798in}{0.442177in}}%
\pgfpathclose%
\pgfusepath{fill}%
\end{pgfscope}%
\begin{pgfscope}%
\pgfpathrectangle{\pgfqpoint{0.520798in}{0.442177in}}{\pgfqpoint{0.719005in}{0.871884in}}%
\pgfusepath{clip}%
\pgfsetroundcap%
\pgfsetroundjoin%
\pgfsetlinewidth{0.501875pt}%
\definecolor{currentstroke}{rgb}{0.800000,0.800000,0.800000}%
\pgfsetstrokecolor{currentstroke}%
\pgfsetdash{}{0pt}%
\pgfpathmoveto{\pgfqpoint{0.520798in}{0.442177in}}%
\pgfpathlineto{\pgfqpoint{0.520798in}{1.314061in}}%
\pgfusepath{stroke}%
\end{pgfscope}%
\begin{pgfscope}%
\definecolor{textcolor}{rgb}{0.150000,0.150000,0.150000}%
\pgfsetstrokecolor{textcolor}%
\pgfsetfillcolor{textcolor}%
\pgftext[x=0.520798in,y=0.351899in,,top]{\color{textcolor}\rmfamily\fontsize{8.000000}{9.600000}\selectfont \(\displaystyle {0}\)}%
\end{pgfscope}%
\begin{pgfscope}%
\pgfpathrectangle{\pgfqpoint{0.520798in}{0.442177in}}{\pgfqpoint{0.719005in}{0.871884in}}%
\pgfusepath{clip}%
\pgfsetroundcap%
\pgfsetroundjoin%
\pgfsetlinewidth{0.501875pt}%
\definecolor{currentstroke}{rgb}{0.800000,0.800000,0.800000}%
\pgfsetstrokecolor{currentstroke}%
\pgfsetdash{}{0pt}%
\pgfpathmoveto{\pgfqpoint{0.880300in}{0.442177in}}%
\pgfpathlineto{\pgfqpoint{0.880300in}{1.314061in}}%
\pgfusepath{stroke}%
\end{pgfscope}%
\begin{pgfscope}%
\definecolor{textcolor}{rgb}{0.150000,0.150000,0.150000}%
\pgfsetstrokecolor{textcolor}%
\pgfsetfillcolor{textcolor}%
\pgftext[x=0.880300in,y=0.351899in,,top]{\color{textcolor}\rmfamily\fontsize{8.000000}{9.600000}\selectfont \(\displaystyle {5}\)}%
\end{pgfscope}%
\begin{pgfscope}%
\pgfpathrectangle{\pgfqpoint{0.520798in}{0.442177in}}{\pgfqpoint{0.719005in}{0.871884in}}%
\pgfusepath{clip}%
\pgfsetroundcap%
\pgfsetroundjoin%
\pgfsetlinewidth{0.501875pt}%
\definecolor{currentstroke}{rgb}{0.800000,0.800000,0.800000}%
\pgfsetstrokecolor{currentstroke}%
\pgfsetdash{}{0pt}%
\pgfpathmoveto{\pgfqpoint{1.239803in}{0.442177in}}%
\pgfpathlineto{\pgfqpoint{1.239803in}{1.314061in}}%
\pgfusepath{stroke}%
\end{pgfscope}%
\begin{pgfscope}%
\definecolor{textcolor}{rgb}{0.150000,0.150000,0.150000}%
\pgfsetstrokecolor{textcolor}%
\pgfsetfillcolor{textcolor}%
\pgftext[x=1.239803in,y=0.351899in,,top]{\color{textcolor}\rmfamily\fontsize{8.000000}{9.600000}\selectfont \(\displaystyle {10}\)}%
\end{pgfscope}%
\begin{pgfscope}%
\definecolor{textcolor}{rgb}{0.150000,0.150000,0.150000}%
\pgfsetstrokecolor{textcolor}%
\pgfsetfillcolor{textcolor}%
\pgftext[x=0.880300in,y=0.198219in,,top]{\color{textcolor}\rmfamily\fontsize{10.000000}{12.000000}\selectfont Edit distance \(\displaystyle \epsilon\)}%
\end{pgfscope}%
\begin{pgfscope}%
\pgfpathrectangle{\pgfqpoint{0.520798in}{0.442177in}}{\pgfqpoint{0.719005in}{0.871884in}}%
\pgfusepath{clip}%
\pgfsetroundcap%
\pgfsetroundjoin%
\pgfsetlinewidth{0.501875pt}%
\definecolor{currentstroke}{rgb}{0.800000,0.800000,0.800000}%
\pgfsetstrokecolor{currentstroke}%
\pgfsetdash{}{0pt}%
\pgfpathmoveto{\pgfqpoint{0.520798in}{0.442177in}}%
\pgfpathlineto{\pgfqpoint{1.239803in}{0.442177in}}%
\pgfusepath{stroke}%
\end{pgfscope}%
\begin{pgfscope}%
\definecolor{textcolor}{rgb}{0.150000,0.150000,0.150000}%
\pgfsetstrokecolor{textcolor}%
\pgfsetfillcolor{textcolor}%
\pgftext[x=0.273151in, y=0.403915in, left, base]{\color{textcolor}\rmfamily\fontsize{8.000000}{9.600000}\selectfont 0\%}%
\end{pgfscope}%
\begin{pgfscope}%
\pgfpathrectangle{\pgfqpoint{0.520798in}{0.442177in}}{\pgfqpoint{0.719005in}{0.871884in}}%
\pgfusepath{clip}%
\pgfsetroundcap%
\pgfsetroundjoin%
\pgfsetlinewidth{0.501875pt}%
\definecolor{currentstroke}{rgb}{0.800000,0.800000,0.800000}%
\pgfsetstrokecolor{currentstroke}%
\pgfsetdash{}{0pt}%
\pgfpathmoveto{\pgfqpoint{0.520798in}{0.660148in}}%
\pgfpathlineto{\pgfqpoint{1.239803in}{0.660148in}}%
\pgfusepath{stroke}%
\end{pgfscope}%
\begin{pgfscope}%
\definecolor{textcolor}{rgb}{0.150000,0.150000,0.150000}%
\pgfsetstrokecolor{textcolor}%
\pgfsetfillcolor{textcolor}%
\pgftext[x=0.214138in, y=0.621886in, left, base]{\color{textcolor}\rmfamily\fontsize{8.000000}{9.600000}\selectfont 25\%}%
\end{pgfscope}%
\begin{pgfscope}%
\pgfpathrectangle{\pgfqpoint{0.520798in}{0.442177in}}{\pgfqpoint{0.719005in}{0.871884in}}%
\pgfusepath{clip}%
\pgfsetroundcap%
\pgfsetroundjoin%
\pgfsetlinewidth{0.501875pt}%
\definecolor{currentstroke}{rgb}{0.800000,0.800000,0.800000}%
\pgfsetstrokecolor{currentstroke}%
\pgfsetdash{}{0pt}%
\pgfpathmoveto{\pgfqpoint{0.520798in}{0.878119in}}%
\pgfpathlineto{\pgfqpoint{1.239803in}{0.878119in}}%
\pgfusepath{stroke}%
\end{pgfscope}%
\begin{pgfscope}%
\definecolor{textcolor}{rgb}{0.150000,0.150000,0.150000}%
\pgfsetstrokecolor{textcolor}%
\pgfsetfillcolor{textcolor}%
\pgftext[x=0.214138in, y=0.839857in, left, base]{\color{textcolor}\rmfamily\fontsize{8.000000}{9.600000}\selectfont 50\%}%
\end{pgfscope}%
\begin{pgfscope}%
\pgfpathrectangle{\pgfqpoint{0.520798in}{0.442177in}}{\pgfqpoint{0.719005in}{0.871884in}}%
\pgfusepath{clip}%
\pgfsetroundcap%
\pgfsetroundjoin%
\pgfsetlinewidth{0.501875pt}%
\definecolor{currentstroke}{rgb}{0.800000,0.800000,0.800000}%
\pgfsetstrokecolor{currentstroke}%
\pgfsetdash{}{0pt}%
\pgfpathmoveto{\pgfqpoint{0.520798in}{1.096090in}}%
\pgfpathlineto{\pgfqpoint{1.239803in}{1.096090in}}%
\pgfusepath{stroke}%
\end{pgfscope}%
\begin{pgfscope}%
\definecolor{textcolor}{rgb}{0.150000,0.150000,0.150000}%
\pgfsetstrokecolor{textcolor}%
\pgfsetfillcolor{textcolor}%
\pgftext[x=0.214138in, y=1.057828in, left, base]{\color{textcolor}\rmfamily\fontsize{8.000000}{9.600000}\selectfont 75\%}%
\end{pgfscope}%
\begin{pgfscope}%
\pgfpathrectangle{\pgfqpoint{0.520798in}{0.442177in}}{\pgfqpoint{0.719005in}{0.871884in}}%
\pgfusepath{clip}%
\pgfsetroundcap%
\pgfsetroundjoin%
\pgfsetlinewidth{0.501875pt}%
\definecolor{currentstroke}{rgb}{0.800000,0.800000,0.800000}%
\pgfsetstrokecolor{currentstroke}%
\pgfsetdash{}{0pt}%
\pgfpathmoveto{\pgfqpoint{0.520798in}{1.314061in}}%
\pgfpathlineto{\pgfqpoint{1.239803in}{1.314061in}}%
\pgfusepath{stroke}%
\end{pgfscope}%
\begin{pgfscope}%
\definecolor{textcolor}{rgb}{0.150000,0.150000,0.150000}%
\pgfsetstrokecolor{textcolor}%
\pgfsetfillcolor{textcolor}%
\pgftext[x=0.155124in, y=1.275799in, left, base]{\color{textcolor}\rmfamily\fontsize{8.000000}{9.600000}\selectfont 100\%}%
\end{pgfscope}%
\begin{pgfscope}%
\definecolor{textcolor}{rgb}{0.150000,0.150000,0.150000}%
\pgfsetstrokecolor{textcolor}%
\pgfsetfillcolor{textcolor}%
\pgftext[x=0.099569in,y=0.878119in,,bottom,rotate=90.000000]{\color{textcolor}\rmfamily\fontsize{10.000000}{12.000000}\selectfont Cert. Acc.}%
\end{pgfscope}%
\begin{pgfscope}%
\pgfsetrectcap%
\pgfsetmiterjoin%
\pgfsetlinewidth{0.752812pt}%
\definecolor{currentstroke}{rgb}{0.700000,0.700000,0.700000}%
\pgfsetstrokecolor{currentstroke}%
\pgfsetdash{}{0pt}%
\pgfpathmoveto{\pgfqpoint{0.520798in}{0.442177in}}%
\pgfpathlineto{\pgfqpoint{0.520798in}{1.314061in}}%
\pgfusepath{stroke}%
\end{pgfscope}%
\begin{pgfscope}%
\pgfsetrectcap%
\pgfsetmiterjoin%
\pgfsetlinewidth{0.752812pt}%
\definecolor{currentstroke}{rgb}{0.700000,0.700000,0.700000}%
\pgfsetstrokecolor{currentstroke}%
\pgfsetdash{}{0pt}%
\pgfpathmoveto{\pgfqpoint{1.239803in}{0.442177in}}%
\pgfpathlineto{\pgfqpoint{1.239803in}{1.314061in}}%
\pgfusepath{stroke}%
\end{pgfscope}%
\begin{pgfscope}%
\pgfsetrectcap%
\pgfsetmiterjoin%
\pgfsetlinewidth{0.752812pt}%
\definecolor{currentstroke}{rgb}{0.700000,0.700000,0.700000}%
\pgfsetstrokecolor{currentstroke}%
\pgfsetdash{}{0pt}%
\pgfpathmoveto{\pgfqpoint{0.520798in}{0.442177in}}%
\pgfpathlineto{\pgfqpoint{1.239803in}{0.442177in}}%
\pgfusepath{stroke}%
\end{pgfscope}%
\begin{pgfscope}%
\pgfsetrectcap%
\pgfsetmiterjoin%
\pgfsetlinewidth{0.752812pt}%
\definecolor{currentstroke}{rgb}{0.700000,0.700000,0.700000}%
\pgfsetstrokecolor{currentstroke}%
\pgfsetdash{}{0pt}%
\pgfpathmoveto{\pgfqpoint{0.520798in}{1.314061in}}%
\pgfpathlineto{\pgfqpoint{1.239803in}{1.314061in}}%
\pgfusepath{stroke}%
\end{pgfscope}%
\begin{pgfscope}%
\pgfpathrectangle{\pgfqpoint{0.520798in}{0.442177in}}{\pgfqpoint{0.719005in}{0.871884in}}%
\pgfusepath{clip}%
\pgfsetroundcap%
\pgfsetroundjoin%
\pgfsetlinewidth{1.003750pt}%
\definecolor{currentstroke}{rgb}{0.003922,0.450980,0.698039}%
\pgfsetstrokecolor{currentstroke}%
\pgfsetdash{}{0pt}%
\pgfpathmoveto{\pgfqpoint{0.520798in}{1.044627in}}%
\pgfpathlineto{\pgfqpoint{0.556748in}{1.044627in}}%
\pgfpathlineto{\pgfqpoint{0.556748in}{0.912559in}}%
\pgfpathlineto{\pgfqpoint{0.628649in}{0.912559in}}%
\pgfpathlineto{\pgfqpoint{0.628649in}{0.712085in}}%
\pgfpathlineto{\pgfqpoint{0.700549in}{0.712085in}}%
\pgfpathlineto{\pgfqpoint{0.700549in}{0.466692in}}%
\pgfpathlineto{\pgfqpoint{0.772450in}{0.466692in}}%
\pgfpathlineto{\pgfqpoint{0.772450in}{0.446526in}}%
\pgfpathlineto{\pgfqpoint{0.844350in}{0.446526in}}%
\pgfpathlineto{\pgfqpoint{0.844350in}{0.444470in}}%
\pgfpathlineto{\pgfqpoint{0.916251in}{0.444470in}}%
\pgfpathlineto{\pgfqpoint{0.916251in}{0.443917in}}%
\pgfpathlineto{\pgfqpoint{0.988151in}{0.443917in}}%
\pgfpathlineto{\pgfqpoint{0.988151in}{0.443521in}}%
\pgfpathlineto{\pgfqpoint{1.060052in}{0.443521in}}%
\pgfpathlineto{\pgfqpoint{1.060052in}{0.442572in}}%
\pgfpathlineto{\pgfqpoint{1.239803in}{0.442572in}}%
\pgfpathlineto{\pgfqpoint{1.239803in}{0.442493in}}%
\pgfpathlineto{\pgfqpoint{1.241470in}{0.442493in}}%
\pgfusepath{stroke}%
\end{pgfscope}%
\begin{pgfscope}%
\pgfpathrectangle{\pgfqpoint{0.520798in}{0.442177in}}{\pgfqpoint{0.719005in}{0.871884in}}%
\pgfusepath{clip}%
\pgfsetbuttcap%
\pgfsetroundjoin%
\definecolor{currentfill}{rgb}{0.003922,0.450980,0.698039}%
\pgfsetfillcolor{currentfill}%
\pgfsetfillopacity{0.500000}%
\pgfsetlinewidth{0.000000pt}%
\definecolor{currentstroke}{rgb}{0.003922,0.450980,0.698039}%
\pgfsetstrokecolor{currentstroke}%
\pgfsetstrokeopacity{0.500000}%
\pgfsetdash{}{0pt}%
\pgfpathmoveto{\pgfqpoint{0.520798in}{1.063547in}}%
\pgfpathlineto{\pgfqpoint{0.520798in}{1.025707in}}%
\pgfpathlineto{\pgfqpoint{0.556748in}{1.025707in}}%
\pgfpathlineto{\pgfqpoint{0.556748in}{0.901723in}}%
\pgfpathlineto{\pgfqpoint{0.628649in}{0.901723in}}%
\pgfpathlineto{\pgfqpoint{0.628649in}{0.705865in}}%
\pgfpathlineto{\pgfqpoint{0.700549in}{0.705865in}}%
\pgfpathlineto{\pgfqpoint{0.700549in}{0.463919in}}%
\pgfpathlineto{\pgfqpoint{0.772450in}{0.463919in}}%
\pgfpathlineto{\pgfqpoint{0.772450in}{0.445495in}}%
\pgfpathlineto{\pgfqpoint{0.844350in}{0.445495in}}%
\pgfpathlineto{\pgfqpoint{0.844350in}{0.443590in}}%
\pgfpathlineto{\pgfqpoint{0.916251in}{0.443590in}}%
\pgfpathlineto{\pgfqpoint{0.916251in}{0.443380in}}%
\pgfpathlineto{\pgfqpoint{0.988151in}{0.443380in}}%
\pgfpathlineto{\pgfqpoint{0.988151in}{0.443118in}}%
\pgfpathlineto{\pgfqpoint{1.060052in}{0.443118in}}%
\pgfpathlineto{\pgfqpoint{1.060052in}{0.442572in}}%
\pgfpathlineto{\pgfqpoint{1.239803in}{0.442572in}}%
\pgfpathlineto{\pgfqpoint{1.239803in}{0.442335in}}%
\pgfpathlineto{\pgfqpoint{1.419554in}{0.442335in}}%
\pgfpathlineto{\pgfqpoint{1.419554in}{0.442177in}}%
\pgfpathlineto{\pgfqpoint{2.066659in}{0.442177in}}%
\pgfpathlineto{\pgfqpoint{2.066659in}{0.442177in}}%
\pgfpathlineto{\pgfqpoint{2.677813in}{0.442177in}}%
\pgfpathlineto{\pgfqpoint{2.677813in}{0.442177in}}%
\pgfpathlineto{\pgfqpoint{2.677813in}{0.442177in}}%
\pgfpathlineto{\pgfqpoint{2.066659in}{0.442177in}}%
\pgfpathlineto{\pgfqpoint{2.066659in}{0.442177in}}%
\pgfpathlineto{\pgfqpoint{1.419554in}{0.442177in}}%
\pgfpathlineto{\pgfqpoint{1.419554in}{0.442651in}}%
\pgfpathlineto{\pgfqpoint{1.239803in}{0.442651in}}%
\pgfpathlineto{\pgfqpoint{1.239803in}{0.442572in}}%
\pgfpathlineto{\pgfqpoint{1.060052in}{0.442572in}}%
\pgfpathlineto{\pgfqpoint{1.060052in}{0.443924in}}%
\pgfpathlineto{\pgfqpoint{0.988151in}{0.443924in}}%
\pgfpathlineto{\pgfqpoint{0.988151in}{0.444453in}}%
\pgfpathlineto{\pgfqpoint{0.916251in}{0.444453in}}%
\pgfpathlineto{\pgfqpoint{0.916251in}{0.445351in}}%
\pgfpathlineto{\pgfqpoint{0.844350in}{0.445351in}}%
\pgfpathlineto{\pgfqpoint{0.844350in}{0.447557in}}%
\pgfpathlineto{\pgfqpoint{0.772450in}{0.447557in}}%
\pgfpathlineto{\pgfqpoint{0.772450in}{0.469466in}}%
\pgfpathlineto{\pgfqpoint{0.700549in}{0.469466in}}%
\pgfpathlineto{\pgfqpoint{0.700549in}{0.718305in}}%
\pgfpathlineto{\pgfqpoint{0.628649in}{0.718305in}}%
\pgfpathlineto{\pgfqpoint{0.628649in}{0.923396in}}%
\pgfpathlineto{\pgfqpoint{0.556748in}{0.923396in}}%
\pgfpathlineto{\pgfqpoint{0.556748in}{1.063547in}}%
\pgfpathlineto{\pgfqpoint{0.520798in}{1.063547in}}%
\pgfpathlineto{\pgfqpoint{0.520798in}{1.063547in}}%
\pgfpathclose%
\pgfusepath{fill}%
\end{pgfscope}%
\begin{pgfscope}%
\pgfpathrectangle{\pgfqpoint{0.520798in}{0.442177in}}{\pgfqpoint{0.719005in}{0.871884in}}%
\pgfusepath{clip}%
\pgfsetroundcap%
\pgfsetroundjoin%
\pgfsetlinewidth{1.003750pt}%
\definecolor{currentstroke}{rgb}{0.870588,0.560784,0.019608}%
\pgfsetstrokecolor{currentstroke}%
\pgfsetdash{}{0pt}%
\pgfpathmoveto{\pgfqpoint{0.520798in}{1.044627in}}%
\pgfpathlineto{\pgfqpoint{0.556748in}{1.044627in}}%
\pgfpathlineto{\pgfqpoint{0.556748in}{0.937866in}}%
\pgfpathlineto{\pgfqpoint{0.628649in}{0.937866in}}%
\pgfpathlineto{\pgfqpoint{0.628649in}{0.846921in}}%
\pgfpathlineto{\pgfqpoint{0.700549in}{0.846921in}}%
\pgfpathlineto{\pgfqpoint{0.700549in}{0.761749in}}%
\pgfpathlineto{\pgfqpoint{0.772450in}{0.761749in}}%
\pgfpathlineto{\pgfqpoint{0.772450in}{0.691998in}}%
\pgfpathlineto{\pgfqpoint{0.844350in}{0.691998in}}%
\pgfpathlineto{\pgfqpoint{0.844350in}{0.638539in}}%
\pgfpathlineto{\pgfqpoint{0.916251in}{0.638539in}}%
\pgfpathlineto{\pgfqpoint{0.916251in}{0.466692in}}%
\pgfpathlineto{\pgfqpoint{0.988151in}{0.466692in}}%
\pgfpathlineto{\pgfqpoint{0.988151in}{0.461868in}}%
\pgfpathlineto{\pgfqpoint{1.060052in}{0.461868in}}%
\pgfpathlineto{\pgfqpoint{1.060052in}{0.446526in}}%
\pgfpathlineto{\pgfqpoint{1.131952in}{0.446526in}}%
\pgfpathlineto{\pgfqpoint{1.131952in}{0.445261in}}%
\pgfpathlineto{\pgfqpoint{1.203853in}{0.445261in}}%
\pgfpathlineto{\pgfqpoint{1.203853in}{0.444391in}}%
\pgfpathlineto{\pgfqpoint{1.241470in}{0.444391in}}%
\pgfusepath{stroke}%
\end{pgfscope}%
\begin{pgfscope}%
\pgfpathrectangle{\pgfqpoint{0.520798in}{0.442177in}}{\pgfqpoint{0.719005in}{0.871884in}}%
\pgfusepath{clip}%
\pgfsetbuttcap%
\pgfsetroundjoin%
\definecolor{currentfill}{rgb}{0.870588,0.560784,0.019608}%
\pgfsetfillcolor{currentfill}%
\pgfsetfillopacity{0.500000}%
\pgfsetlinewidth{0.000000pt}%
\definecolor{currentstroke}{rgb}{0.870588,0.560784,0.019608}%
\pgfsetstrokecolor{currentstroke}%
\pgfsetstrokeopacity{0.500000}%
\pgfsetdash{}{0pt}%
\pgfpathmoveto{\pgfqpoint{0.520798in}{1.063547in}}%
\pgfpathlineto{\pgfqpoint{0.520798in}{1.025707in}}%
\pgfpathlineto{\pgfqpoint{0.556748in}{1.025707in}}%
\pgfpathlineto{\pgfqpoint{0.556748in}{0.925638in}}%
\pgfpathlineto{\pgfqpoint{0.628649in}{0.925638in}}%
\pgfpathlineto{\pgfqpoint{0.628649in}{0.836703in}}%
\pgfpathlineto{\pgfqpoint{0.700549in}{0.836703in}}%
\pgfpathlineto{\pgfqpoint{0.700549in}{0.752714in}}%
\pgfpathlineto{\pgfqpoint{0.772450in}{0.752714in}}%
\pgfpathlineto{\pgfqpoint{0.772450in}{0.686105in}}%
\pgfpathlineto{\pgfqpoint{0.844350in}{0.686105in}}%
\pgfpathlineto{\pgfqpoint{0.844350in}{0.632893in}}%
\pgfpathlineto{\pgfqpoint{0.916251in}{0.632893in}}%
\pgfpathlineto{\pgfqpoint{0.916251in}{0.463919in}}%
\pgfpathlineto{\pgfqpoint{0.988151in}{0.463919in}}%
\pgfpathlineto{\pgfqpoint{0.988151in}{0.460041in}}%
\pgfpathlineto{\pgfqpoint{1.060052in}{0.460041in}}%
\pgfpathlineto{\pgfqpoint{1.060052in}{0.445495in}}%
\pgfpathlineto{\pgfqpoint{1.131952in}{0.445495in}}%
\pgfpathlineto{\pgfqpoint{1.131952in}{0.444188in}}%
\pgfpathlineto{\pgfqpoint{1.203853in}{0.444188in}}%
\pgfpathlineto{\pgfqpoint{1.203853in}{0.443577in}}%
\pgfpathlineto{\pgfqpoint{1.275753in}{0.443577in}}%
\pgfpathlineto{\pgfqpoint{1.275753in}{0.443404in}}%
\pgfpathlineto{\pgfqpoint{1.347654in}{0.443404in}}%
\pgfpathlineto{\pgfqpoint{1.347654in}{0.443126in}}%
\pgfpathlineto{\pgfqpoint{1.419554in}{0.443126in}}%
\pgfpathlineto{\pgfqpoint{1.419554in}{0.443009in}}%
\pgfpathlineto{\pgfqpoint{1.491455in}{0.443009in}}%
\pgfpathlineto{\pgfqpoint{1.491455in}{0.442889in}}%
\pgfpathlineto{\pgfqpoint{1.563355in}{0.442889in}}%
\pgfpathlineto{\pgfqpoint{1.563355in}{0.442751in}}%
\pgfpathlineto{\pgfqpoint{1.635256in}{0.442751in}}%
\pgfpathlineto{\pgfqpoint{1.635256in}{0.442572in}}%
\pgfpathlineto{\pgfqpoint{1.922858in}{0.442572in}}%
\pgfpathlineto{\pgfqpoint{1.922858in}{0.442335in}}%
\pgfpathlineto{\pgfqpoint{2.246410in}{0.442335in}}%
\pgfpathlineto{\pgfqpoint{2.246410in}{0.442177in}}%
\pgfpathlineto{\pgfqpoint{2.498062in}{0.442177in}}%
\pgfpathlineto{\pgfqpoint{2.498062in}{0.442177in}}%
\pgfpathlineto{\pgfqpoint{2.677813in}{0.442177in}}%
\pgfpathlineto{\pgfqpoint{2.677813in}{0.442177in}}%
\pgfpathlineto{\pgfqpoint{2.677813in}{0.442177in}}%
\pgfpathlineto{\pgfqpoint{2.498062in}{0.442177in}}%
\pgfpathlineto{\pgfqpoint{2.498062in}{0.442177in}}%
\pgfpathlineto{\pgfqpoint{2.246410in}{0.442177in}}%
\pgfpathlineto{\pgfqpoint{2.246410in}{0.442651in}}%
\pgfpathlineto{\pgfqpoint{1.922858in}{0.442651in}}%
\pgfpathlineto{\pgfqpoint{1.922858in}{0.442572in}}%
\pgfpathlineto{\pgfqpoint{1.635256in}{0.442572in}}%
\pgfpathlineto{\pgfqpoint{1.635256in}{0.443343in}}%
\pgfpathlineto{\pgfqpoint{1.563355in}{0.443343in}}%
\pgfpathlineto{\pgfqpoint{1.563355in}{0.443521in}}%
\pgfpathlineto{\pgfqpoint{1.491455in}{0.443521in}}%
\pgfpathlineto{\pgfqpoint{1.491455in}{0.443717in}}%
\pgfpathlineto{\pgfqpoint{1.419554in}{0.443717in}}%
\pgfpathlineto{\pgfqpoint{1.419554in}{0.444075in}}%
\pgfpathlineto{\pgfqpoint{1.347654in}{0.444075in}}%
\pgfpathlineto{\pgfqpoint{1.347654in}{0.444588in}}%
\pgfpathlineto{\pgfqpoint{1.275753in}{0.444588in}}%
\pgfpathlineto{\pgfqpoint{1.275753in}{0.445205in}}%
\pgfpathlineto{\pgfqpoint{1.203853in}{0.445205in}}%
\pgfpathlineto{\pgfqpoint{1.203853in}{0.446334in}}%
\pgfpathlineto{\pgfqpoint{1.131952in}{0.446334in}}%
\pgfpathlineto{\pgfqpoint{1.131952in}{0.447557in}}%
\pgfpathlineto{\pgfqpoint{1.060052in}{0.447557in}}%
\pgfpathlineto{\pgfqpoint{1.060052in}{0.463696in}}%
\pgfpathlineto{\pgfqpoint{0.988151in}{0.463696in}}%
\pgfpathlineto{\pgfqpoint{0.988151in}{0.469466in}}%
\pgfpathlineto{\pgfqpoint{0.916251in}{0.469466in}}%
\pgfpathlineto{\pgfqpoint{0.916251in}{0.644184in}}%
\pgfpathlineto{\pgfqpoint{0.844350in}{0.644184in}}%
\pgfpathlineto{\pgfqpoint{0.844350in}{0.697892in}}%
\pgfpathlineto{\pgfqpoint{0.772450in}{0.697892in}}%
\pgfpathlineto{\pgfqpoint{0.772450in}{0.770785in}}%
\pgfpathlineto{\pgfqpoint{0.700549in}{0.770785in}}%
\pgfpathlineto{\pgfqpoint{0.700549in}{0.857139in}}%
\pgfpathlineto{\pgfqpoint{0.628649in}{0.857139in}}%
\pgfpathlineto{\pgfqpoint{0.628649in}{0.950093in}}%
\pgfpathlineto{\pgfqpoint{0.556748in}{0.950093in}}%
\pgfpathlineto{\pgfqpoint{0.556748in}{1.063547in}}%
\pgfpathlineto{\pgfqpoint{0.520798in}{1.063547in}}%
\pgfpathlineto{\pgfqpoint{0.520798in}{1.063547in}}%
\pgfpathclose%
\pgfusepath{fill}%
\end{pgfscope}%
\begin{pgfscope}%
\pgfpathrectangle{\pgfqpoint{0.520798in}{0.442177in}}{\pgfqpoint{0.719005in}{0.871884in}}%
\pgfusepath{clip}%
\pgfsetroundcap%
\pgfsetroundjoin%
\pgfsetlinewidth{1.003750pt}%
\definecolor{currentstroke}{rgb}{0.007843,0.619608,0.450980}%
\pgfsetstrokecolor{currentstroke}%
\pgfsetdash{}{0pt}%
\pgfpathmoveto{\pgfqpoint{0.520798in}{1.044627in}}%
\pgfpathlineto{\pgfqpoint{0.556748in}{1.044627in}}%
\pgfpathlineto{\pgfqpoint{0.556748in}{0.937866in}}%
\pgfpathlineto{\pgfqpoint{0.628649in}{0.937866in}}%
\pgfpathlineto{\pgfqpoint{0.628649in}{0.846921in}}%
\pgfpathlineto{\pgfqpoint{0.700549in}{0.846921in}}%
\pgfpathlineto{\pgfqpoint{0.700549in}{0.761749in}}%
\pgfpathlineto{\pgfqpoint{0.772450in}{0.761749in}}%
\pgfpathlineto{\pgfqpoint{0.772450in}{0.691998in}}%
\pgfpathlineto{\pgfqpoint{0.844350in}{0.691998in}}%
\pgfpathlineto{\pgfqpoint{0.844350in}{0.638539in}}%
\pgfpathlineto{\pgfqpoint{0.916251in}{0.638539in}}%
\pgfpathlineto{\pgfqpoint{0.916251in}{0.597969in}}%
\pgfpathlineto{\pgfqpoint{0.988151in}{0.597969in}}%
\pgfpathlineto{\pgfqpoint{0.988151in}{0.566178in}}%
\pgfpathlineto{\pgfqpoint{1.060052in}{0.566178in}}%
\pgfpathlineto{\pgfqpoint{1.060052in}{0.538262in}}%
\pgfpathlineto{\pgfqpoint{1.131952in}{0.538262in}}%
\pgfpathlineto{\pgfqpoint{1.131952in}{0.517384in}}%
\pgfpathlineto{\pgfqpoint{1.203853in}{0.517384in}}%
\pgfpathlineto{\pgfqpoint{1.203853in}{0.500777in}}%
\pgfpathlineto{\pgfqpoint{1.241470in}{0.500777in}}%
\pgfusepath{stroke}%
\end{pgfscope}%
\begin{pgfscope}%
\pgfpathrectangle{\pgfqpoint{0.520798in}{0.442177in}}{\pgfqpoint{0.719005in}{0.871884in}}%
\pgfusepath{clip}%
\pgfsetbuttcap%
\pgfsetroundjoin%
\definecolor{currentfill}{rgb}{0.007843,0.619608,0.450980}%
\pgfsetfillcolor{currentfill}%
\pgfsetfillopacity{0.500000}%
\pgfsetlinewidth{0.000000pt}%
\definecolor{currentstroke}{rgb}{0.007843,0.619608,0.450980}%
\pgfsetstrokecolor{currentstroke}%
\pgfsetstrokeopacity{0.500000}%
\pgfsetdash{}{0pt}%
\pgfpathmoveto{\pgfqpoint{0.520798in}{1.063547in}}%
\pgfpathlineto{\pgfqpoint{0.520798in}{1.025707in}}%
\pgfpathlineto{\pgfqpoint{0.556748in}{1.025707in}}%
\pgfpathlineto{\pgfqpoint{0.556748in}{0.925638in}}%
\pgfpathlineto{\pgfqpoint{0.628649in}{0.925638in}}%
\pgfpathlineto{\pgfqpoint{0.628649in}{0.836703in}}%
\pgfpathlineto{\pgfqpoint{0.700549in}{0.836703in}}%
\pgfpathlineto{\pgfqpoint{0.700549in}{0.752714in}}%
\pgfpathlineto{\pgfqpoint{0.772450in}{0.752714in}}%
\pgfpathlineto{\pgfqpoint{0.772450in}{0.686105in}}%
\pgfpathlineto{\pgfqpoint{0.844350in}{0.686105in}}%
\pgfpathlineto{\pgfqpoint{0.844350in}{0.632893in}}%
\pgfpathlineto{\pgfqpoint{0.916251in}{0.632893in}}%
\pgfpathlineto{\pgfqpoint{0.916251in}{0.592765in}}%
\pgfpathlineto{\pgfqpoint{0.988151in}{0.592765in}}%
\pgfpathlineto{\pgfqpoint{0.988151in}{0.560236in}}%
\pgfpathlineto{\pgfqpoint{1.060052in}{0.560236in}}%
\pgfpathlineto{\pgfqpoint{1.060052in}{0.532040in}}%
\pgfpathlineto{\pgfqpoint{1.131952in}{0.532040in}}%
\pgfpathlineto{\pgfqpoint{1.131952in}{0.512449in}}%
\pgfpathlineto{\pgfqpoint{1.203853in}{0.512449in}}%
\pgfpathlineto{\pgfqpoint{1.203853in}{0.495792in}}%
\pgfpathlineto{\pgfqpoint{1.275753in}{0.495792in}}%
\pgfpathlineto{\pgfqpoint{1.275753in}{0.482366in}}%
\pgfpathlineto{\pgfqpoint{1.347654in}{0.482366in}}%
\pgfpathlineto{\pgfqpoint{1.347654in}{0.463919in}}%
\pgfpathlineto{\pgfqpoint{1.419554in}{0.463919in}}%
\pgfpathlineto{\pgfqpoint{1.419554in}{0.460041in}}%
\pgfpathlineto{\pgfqpoint{1.491455in}{0.460041in}}%
\pgfpathlineto{\pgfqpoint{1.491455in}{0.457229in}}%
\pgfpathlineto{\pgfqpoint{1.563355in}{0.457229in}}%
\pgfpathlineto{\pgfqpoint{1.563355in}{0.454448in}}%
\pgfpathlineto{\pgfqpoint{1.635256in}{0.454448in}}%
\pgfpathlineto{\pgfqpoint{1.635256in}{0.445495in}}%
\pgfpathlineto{\pgfqpoint{1.707156in}{0.445495in}}%
\pgfpathlineto{\pgfqpoint{1.707156in}{0.444188in}}%
\pgfpathlineto{\pgfqpoint{1.779057in}{0.444188in}}%
\pgfpathlineto{\pgfqpoint{1.779057in}{0.444042in}}%
\pgfpathlineto{\pgfqpoint{1.850957in}{0.444042in}}%
\pgfpathlineto{\pgfqpoint{1.850957in}{0.443904in}}%
\pgfpathlineto{\pgfqpoint{1.922858in}{0.443904in}}%
\pgfpathlineto{\pgfqpoint{1.922858in}{0.443308in}}%
\pgfpathlineto{\pgfqpoint{1.994758in}{0.443308in}}%
\pgfpathlineto{\pgfqpoint{1.994758in}{0.443118in}}%
\pgfpathlineto{\pgfqpoint{2.066659in}{0.443118in}}%
\pgfpathlineto{\pgfqpoint{2.066659in}{0.443113in}}%
\pgfpathlineto{\pgfqpoint{2.138559in}{0.443113in}}%
\pgfpathlineto{\pgfqpoint{2.138559in}{0.442988in}}%
\pgfpathlineto{\pgfqpoint{2.210460in}{0.442988in}}%
\pgfpathlineto{\pgfqpoint{2.210460in}{0.442889in}}%
\pgfpathlineto{\pgfqpoint{2.282360in}{0.442889in}}%
\pgfpathlineto{\pgfqpoint{2.282360in}{0.442718in}}%
\pgfpathlineto{\pgfqpoint{2.354261in}{0.442718in}}%
\pgfpathlineto{\pgfqpoint{2.354261in}{0.442493in}}%
\pgfpathlineto{\pgfqpoint{2.426161in}{0.442493in}}%
\pgfpathlineto{\pgfqpoint{2.426161in}{0.442537in}}%
\pgfpathlineto{\pgfqpoint{2.498062in}{0.442537in}}%
\pgfpathlineto{\pgfqpoint{2.498062in}{0.442493in}}%
\pgfpathlineto{\pgfqpoint{2.605913in}{0.442493in}}%
\pgfpathlineto{\pgfqpoint{2.605913in}{0.442572in}}%
\pgfpathlineto{\pgfqpoint{2.677813in}{0.442572in}}%
\pgfpathlineto{\pgfqpoint{2.677813in}{0.442572in}}%
\pgfpathlineto{\pgfqpoint{2.677813in}{0.442572in}}%
\pgfpathlineto{\pgfqpoint{2.605913in}{0.442572in}}%
\pgfpathlineto{\pgfqpoint{2.605913in}{0.442809in}}%
\pgfpathlineto{\pgfqpoint{2.498062in}{0.442809in}}%
\pgfpathlineto{\pgfqpoint{2.498062in}{0.442924in}}%
\pgfpathlineto{\pgfqpoint{2.426161in}{0.442924in}}%
\pgfpathlineto{\pgfqpoint{2.426161in}{0.443126in}}%
\pgfpathlineto{\pgfqpoint{2.354261in}{0.443126in}}%
\pgfpathlineto{\pgfqpoint{2.354261in}{0.443218in}}%
\pgfpathlineto{\pgfqpoint{2.282360in}{0.443218in}}%
\pgfpathlineto{\pgfqpoint{2.282360in}{0.443521in}}%
\pgfpathlineto{\pgfqpoint{2.210460in}{0.443521in}}%
\pgfpathlineto{\pgfqpoint{2.210460in}{0.443580in}}%
\pgfpathlineto{\pgfqpoint{2.138559in}{0.443580in}}%
\pgfpathlineto{\pgfqpoint{2.138559in}{0.443613in}}%
\pgfpathlineto{\pgfqpoint{2.066659in}{0.443613in}}%
\pgfpathlineto{\pgfqpoint{2.066659in}{0.443924in}}%
\pgfpathlineto{\pgfqpoint{1.994758in}{0.443924in}}%
\pgfpathlineto{\pgfqpoint{1.994758in}{0.444842in}}%
\pgfpathlineto{\pgfqpoint{1.922858in}{0.444842in}}%
\pgfpathlineto{\pgfqpoint{1.922858in}{0.445353in}}%
\pgfpathlineto{\pgfqpoint{1.850957in}{0.445353in}}%
\pgfpathlineto{\pgfqpoint{1.850957in}{0.446005in}}%
\pgfpathlineto{\pgfqpoint{1.779057in}{0.446005in}}%
\pgfpathlineto{\pgfqpoint{1.779057in}{0.446334in}}%
\pgfpathlineto{\pgfqpoint{1.707156in}{0.446334in}}%
\pgfpathlineto{\pgfqpoint{1.707156in}{0.447557in}}%
\pgfpathlineto{\pgfqpoint{1.635256in}{0.447557in}}%
\pgfpathlineto{\pgfqpoint{1.635256in}{0.456794in}}%
\pgfpathlineto{\pgfqpoint{1.563355in}{0.456794in}}%
\pgfpathlineto{\pgfqpoint{1.563355in}{0.459074in}}%
\pgfpathlineto{\pgfqpoint{1.491455in}{0.459074in}}%
\pgfpathlineto{\pgfqpoint{1.491455in}{0.463696in}}%
\pgfpathlineto{\pgfqpoint{1.419554in}{0.463696in}}%
\pgfpathlineto{\pgfqpoint{1.419554in}{0.469466in}}%
\pgfpathlineto{\pgfqpoint{1.347654in}{0.469466in}}%
\pgfpathlineto{\pgfqpoint{1.347654in}{0.492142in}}%
\pgfpathlineto{\pgfqpoint{1.275753in}{0.492142in}}%
\pgfpathlineto{\pgfqpoint{1.275753in}{0.505762in}}%
\pgfpathlineto{\pgfqpoint{1.203853in}{0.505762in}}%
\pgfpathlineto{\pgfqpoint{1.203853in}{0.522319in}}%
\pgfpathlineto{\pgfqpoint{1.131952in}{0.522319in}}%
\pgfpathlineto{\pgfqpoint{1.131952in}{0.544484in}}%
\pgfpathlineto{\pgfqpoint{1.060052in}{0.544484in}}%
\pgfpathlineto{\pgfqpoint{1.060052in}{0.572120in}}%
\pgfpathlineto{\pgfqpoint{0.988151in}{0.572120in}}%
\pgfpathlineto{\pgfqpoint{0.988151in}{0.603173in}}%
\pgfpathlineto{\pgfqpoint{0.916251in}{0.603173in}}%
\pgfpathlineto{\pgfqpoint{0.916251in}{0.644184in}}%
\pgfpathlineto{\pgfqpoint{0.844350in}{0.644184in}}%
\pgfpathlineto{\pgfqpoint{0.844350in}{0.697892in}}%
\pgfpathlineto{\pgfqpoint{0.772450in}{0.697892in}}%
\pgfpathlineto{\pgfqpoint{0.772450in}{0.770785in}}%
\pgfpathlineto{\pgfqpoint{0.700549in}{0.770785in}}%
\pgfpathlineto{\pgfqpoint{0.700549in}{0.857139in}}%
\pgfpathlineto{\pgfqpoint{0.628649in}{0.857139in}}%
\pgfpathlineto{\pgfqpoint{0.628649in}{0.950093in}}%
\pgfpathlineto{\pgfqpoint{0.556748in}{0.950093in}}%
\pgfpathlineto{\pgfqpoint{0.556748in}{1.063547in}}%
\pgfpathlineto{\pgfqpoint{0.520798in}{1.063547in}}%
\pgfpathlineto{\pgfqpoint{0.520798in}{1.063547in}}%
\pgfpathclose%
\pgfusepath{fill}%
\end{pgfscope}%
\begin{pgfscope}%
\definecolor{textcolor}{rgb}{0.150000,0.150000,0.150000}%
\pgfsetstrokecolor{textcolor}%
\pgfsetfillcolor{textcolor}%
\pgftext[x=0.880300in,y=1.397395in,,base]{\color{textcolor}\rmfamily\fontsize{9.000000}{10.800000}\selectfont \(\displaystyle c_A^-=1\)}%
\end{pgfscope}%
\begin{pgfscope}%
\pgfsetbuttcap%
\pgfsetmiterjoin%
\definecolor{currentfill}{rgb}{1.000000,1.000000,1.000000}%
\pgfsetfillcolor{currentfill}%
\pgfsetfillopacity{0.800000}%
\pgfsetlinewidth{1.003750pt}%
\definecolor{currentstroke}{rgb}{0.800000,0.800000,0.800000}%
\pgfsetstrokecolor{currentstroke}%
\pgfsetstrokeopacity{0.800000}%
\pgfsetdash{}{0pt}%
\pgfpathmoveto{\pgfqpoint{0.785843in}{0.638103in}}%
\pgfpathlineto{\pgfqpoint{1.191192in}{0.638103in}}%
\pgfpathlineto{\pgfqpoint{1.191192in}{1.265450in}}%
\pgfpathlineto{\pgfqpoint{0.785843in}{1.265450in}}%
\pgfpathlineto{\pgfqpoint{0.785843in}{0.638103in}}%
\pgfpathclose%
\pgfusepath{stroke,fill}%
\end{pgfscope}%
\begin{pgfscope}%
\definecolor{textcolor}{rgb}{0.150000,0.150000,0.150000}%
\pgfsetstrokecolor{textcolor}%
\pgfsetfillcolor{textcolor}%
\pgftext[x=0.918271in,y=1.113275in,left,base]{\color{textcolor}\rmfamily\fontsize{9.000000}{10.800000}\selectfont \(\displaystyle c_A^+\)}%
\end{pgfscope}%
\begin{pgfscope}%
\pgfsetroundcap%
\pgfsetroundjoin%
\pgfsetlinewidth{1.003750pt}%
\definecolor{currentstroke}{rgb}{0.003922,0.450980,0.698039}%
\pgfsetstrokecolor{currentstroke}%
\pgfsetdash{}{0pt}%
\pgfpathmoveto{\pgfqpoint{0.824732in}{1.001052in}}%
\pgfpathlineto{\pgfqpoint{0.873343in}{1.001052in}}%
\pgfpathlineto{\pgfqpoint{0.873343in}{1.001052in}}%
\pgfpathlineto{\pgfqpoint{0.970565in}{1.001052in}}%
\pgfpathlineto{\pgfqpoint{0.970565in}{1.001052in}}%
\pgfpathlineto{\pgfqpoint{1.019176in}{1.001052in}}%
\pgfusepath{stroke}%
\end{pgfscope}%
\begin{pgfscope}%
\definecolor{textcolor}{rgb}{0.150000,0.150000,0.150000}%
\pgfsetstrokecolor{textcolor}%
\pgfsetfillcolor{textcolor}%
\pgftext[x=1.096954in,y=0.967024in,left,base]{\color{textcolor}\rmfamily\fontsize{7.000000}{8.400000}\selectfont 1}%
\end{pgfscope}%
\begin{pgfscope}%
\pgfsetroundcap%
\pgfsetroundjoin%
\pgfsetlinewidth{1.003750pt}%
\definecolor{currentstroke}{rgb}{0.870588,0.560784,0.019608}%
\pgfsetstrokecolor{currentstroke}%
\pgfsetdash{}{0pt}%
\pgfpathmoveto{\pgfqpoint{0.824732in}{0.865486in}}%
\pgfpathlineto{\pgfqpoint{0.873343in}{0.865486in}}%
\pgfpathlineto{\pgfqpoint{0.873343in}{0.865486in}}%
\pgfpathlineto{\pgfqpoint{0.970565in}{0.865486in}}%
\pgfpathlineto{\pgfqpoint{0.970565in}{0.865486in}}%
\pgfpathlineto{\pgfqpoint{1.019176in}{0.865486in}}%
\pgfusepath{stroke}%
\end{pgfscope}%
\begin{pgfscope}%
\definecolor{textcolor}{rgb}{0.150000,0.150000,0.150000}%
\pgfsetstrokecolor{textcolor}%
\pgfsetfillcolor{textcolor}%
\pgftext[x=1.096954in,y=0.831458in,left,base]{\color{textcolor}\rmfamily\fontsize{7.000000}{8.400000}\selectfont 2}%
\end{pgfscope}%
\begin{pgfscope}%
\pgfsetroundcap%
\pgfsetroundjoin%
\pgfsetlinewidth{1.003750pt}%
\definecolor{currentstroke}{rgb}{0.007843,0.619608,0.450980}%
\pgfsetstrokecolor{currentstroke}%
\pgfsetdash{}{0pt}%
\pgfpathmoveto{\pgfqpoint{0.824732in}{0.729919in}}%
\pgfpathlineto{\pgfqpoint{0.873343in}{0.729919in}}%
\pgfpathlineto{\pgfqpoint{0.873343in}{0.729919in}}%
\pgfpathlineto{\pgfqpoint{0.970565in}{0.729919in}}%
\pgfpathlineto{\pgfqpoint{0.970565in}{0.729919in}}%
\pgfpathlineto{\pgfqpoint{1.019176in}{0.729919in}}%
\pgfusepath{stroke}%
\end{pgfscope}%
\begin{pgfscope}%
\definecolor{textcolor}{rgb}{0.150000,0.150000,0.150000}%
\pgfsetstrokecolor{textcolor}%
\pgfsetfillcolor{textcolor}%
\pgftext[x=1.096954in,y=0.695892in,left,base]{\color{textcolor}\rmfamily\fontsize{7.000000}{8.400000}\selectfont 4}%
\end{pgfscope}%
\end{pgfpicture}%
\makeatother%
\endgroup%

%% file: figures/experiments/graphs/sparse_smoothing/nodes_structure/node_classification-Cora-GCN-hidden=32-p_adj_plus=0.001-p_adj_minus=0.8-p_att_plus=0.0-p_att_minus=0.0-multi_class_cert-A.pgf
\begingroup%
\makeatletter%
\begin{pgfpicture}%
\pgfpathrectangle{\pgfpointorigin}{\pgfqpoint{1.375000in}{1.581250in}}%
\pgfusepath{use as bounding box, clip}%
\begin{pgfscope}%
\pgfsetbuttcap%
\pgfsetmiterjoin%
\definecolor{currentfill}{rgb}{1.000000,1.000000,1.000000}%
\pgfsetfillcolor{currentfill}%
\pgfsetlinewidth{0.000000pt}%
\definecolor{currentstroke}{rgb}{1.000000,1.000000,1.000000}%
\pgfsetstrokecolor{currentstroke}%
\pgfsetdash{}{0pt}%
\pgfpathmoveto{\pgfqpoint{0.000000in}{0.000000in}}%
\pgfpathlineto{\pgfqpoint{1.375000in}{0.000000in}}%
\pgfpathlineto{\pgfqpoint{1.375000in}{1.581250in}}%
\pgfpathlineto{\pgfqpoint{0.000000in}{1.581250in}}%
\pgfpathlineto{\pgfqpoint{0.000000in}{0.000000in}}%
\pgfpathclose%
\pgfusepath{fill}%
\end{pgfscope}%
\begin{pgfscope}%
\pgfsetbuttcap%
\pgfsetmiterjoin%
\definecolor{currentfill}{rgb}{1.000000,1.000000,1.000000}%
\pgfsetfillcolor{currentfill}%
\pgfsetlinewidth{0.000000pt}%
\definecolor{currentstroke}{rgb}{0.000000,0.000000,0.000000}%
\pgfsetstrokecolor{currentstroke}%
\pgfsetstrokeopacity{0.000000}%
\pgfsetdash{}{0pt}%
\pgfpathmoveto{\pgfqpoint{0.520798in}{0.442177in}}%
\pgfpathlineto{\pgfqpoint{1.239803in}{0.442177in}}%
\pgfpathlineto{\pgfqpoint{1.239803in}{1.314061in}}%
\pgfpathlineto{\pgfqpoint{0.520798in}{1.314061in}}%
\pgfpathlineto{\pgfqpoint{0.520798in}{0.442177in}}%
\pgfpathclose%
\pgfusepath{fill}%
\end{pgfscope}%
\begin{pgfscope}%
\pgfpathrectangle{\pgfqpoint{0.520798in}{0.442177in}}{\pgfqpoint{0.719005in}{0.871884in}}%
\pgfusepath{clip}%
\pgfsetroundcap%
\pgfsetroundjoin%
\pgfsetlinewidth{0.501875pt}%
\definecolor{currentstroke}{rgb}{0.800000,0.800000,0.800000}%
\pgfsetstrokecolor{currentstroke}%
\pgfsetdash{}{0pt}%
\pgfpathmoveto{\pgfqpoint{0.520798in}{0.442177in}}%
\pgfpathlineto{\pgfqpoint{0.520798in}{1.314061in}}%
\pgfusepath{stroke}%
\end{pgfscope}%
\begin{pgfscope}%
\definecolor{textcolor}{rgb}{0.150000,0.150000,0.150000}%
\pgfsetstrokecolor{textcolor}%
\pgfsetfillcolor{textcolor}%
\pgftext[x=0.520798in,y=0.351899in,,top]{\color{textcolor}\rmfamily\fontsize{8.000000}{9.600000}\selectfont \(\displaystyle {0}\)}%
\end{pgfscope}%
\begin{pgfscope}%
\pgfpathrectangle{\pgfqpoint{0.520798in}{0.442177in}}{\pgfqpoint{0.719005in}{0.871884in}}%
\pgfusepath{clip}%
\pgfsetroundcap%
\pgfsetroundjoin%
\pgfsetlinewidth{0.501875pt}%
\definecolor{currentstroke}{rgb}{0.800000,0.800000,0.800000}%
\pgfsetstrokecolor{currentstroke}%
\pgfsetdash{}{0pt}%
\pgfpathmoveto{\pgfqpoint{0.880300in}{0.442177in}}%
\pgfpathlineto{\pgfqpoint{0.880300in}{1.314061in}}%
\pgfusepath{stroke}%
\end{pgfscope}%
\begin{pgfscope}%
\definecolor{textcolor}{rgb}{0.150000,0.150000,0.150000}%
\pgfsetstrokecolor{textcolor}%
\pgfsetfillcolor{textcolor}%
\pgftext[x=0.880300in,y=0.351899in,,top]{\color{textcolor}\rmfamily\fontsize{8.000000}{9.600000}\selectfont \(\displaystyle {5}\)}%
\end{pgfscope}%
\begin{pgfscope}%
\pgfpathrectangle{\pgfqpoint{0.520798in}{0.442177in}}{\pgfqpoint{0.719005in}{0.871884in}}%
\pgfusepath{clip}%
\pgfsetroundcap%
\pgfsetroundjoin%
\pgfsetlinewidth{0.501875pt}%
\definecolor{currentstroke}{rgb}{0.800000,0.800000,0.800000}%
\pgfsetstrokecolor{currentstroke}%
\pgfsetdash{}{0pt}%
\pgfpathmoveto{\pgfqpoint{1.239803in}{0.442177in}}%
\pgfpathlineto{\pgfqpoint{1.239803in}{1.314061in}}%
\pgfusepath{stroke}%
\end{pgfscope}%
\begin{pgfscope}%
\definecolor{textcolor}{rgb}{0.150000,0.150000,0.150000}%
\pgfsetstrokecolor{textcolor}%
\pgfsetfillcolor{textcolor}%
\pgftext[x=1.239803in,y=0.351899in,,top]{\color{textcolor}\rmfamily\fontsize{8.000000}{9.600000}\selectfont \(\displaystyle {10}\)}%
\end{pgfscope}%
\begin{pgfscope}%
\definecolor{textcolor}{rgb}{0.150000,0.150000,0.150000}%
\pgfsetstrokecolor{textcolor}%
\pgfsetfillcolor{textcolor}%
\pgftext[x=0.880300in,y=0.198219in,,top]{\color{textcolor}\rmfamily\fontsize{10.000000}{12.000000}\selectfont Edit distance \(\displaystyle \epsilon\)}%
\end{pgfscope}%
\begin{pgfscope}%
\pgfpathrectangle{\pgfqpoint{0.520798in}{0.442177in}}{\pgfqpoint{0.719005in}{0.871884in}}%
\pgfusepath{clip}%
\pgfsetroundcap%
\pgfsetroundjoin%
\pgfsetlinewidth{0.501875pt}%
\definecolor{currentstroke}{rgb}{0.800000,0.800000,0.800000}%
\pgfsetstrokecolor{currentstroke}%
\pgfsetdash{}{0pt}%
\pgfpathmoveto{\pgfqpoint{0.520798in}{0.442177in}}%
\pgfpathlineto{\pgfqpoint{1.239803in}{0.442177in}}%
\pgfusepath{stroke}%
\end{pgfscope}%
\begin{pgfscope}%
\definecolor{textcolor}{rgb}{0.150000,0.150000,0.150000}%
\pgfsetstrokecolor{textcolor}%
\pgfsetfillcolor{textcolor}%
\pgftext[x=0.273151in, y=0.403915in, left, base]{\color{textcolor}\rmfamily\fontsize{8.000000}{9.600000}\selectfont 0\%}%
\end{pgfscope}%
\begin{pgfscope}%
\pgfpathrectangle{\pgfqpoint{0.520798in}{0.442177in}}{\pgfqpoint{0.719005in}{0.871884in}}%
\pgfusepath{clip}%
\pgfsetroundcap%
\pgfsetroundjoin%
\pgfsetlinewidth{0.501875pt}%
\definecolor{currentstroke}{rgb}{0.800000,0.800000,0.800000}%
\pgfsetstrokecolor{currentstroke}%
\pgfsetdash{}{0pt}%
\pgfpathmoveto{\pgfqpoint{0.520798in}{0.660148in}}%
\pgfpathlineto{\pgfqpoint{1.239803in}{0.660148in}}%
\pgfusepath{stroke}%
\end{pgfscope}%
\begin{pgfscope}%
\definecolor{textcolor}{rgb}{0.150000,0.150000,0.150000}%
\pgfsetstrokecolor{textcolor}%
\pgfsetfillcolor{textcolor}%
\pgftext[x=0.214138in, y=0.621886in, left, base]{\color{textcolor}\rmfamily\fontsize{8.000000}{9.600000}\selectfont 25\%}%
\end{pgfscope}%
\begin{pgfscope}%
\pgfpathrectangle{\pgfqpoint{0.520798in}{0.442177in}}{\pgfqpoint{0.719005in}{0.871884in}}%
\pgfusepath{clip}%
\pgfsetroundcap%
\pgfsetroundjoin%
\pgfsetlinewidth{0.501875pt}%
\definecolor{currentstroke}{rgb}{0.800000,0.800000,0.800000}%
\pgfsetstrokecolor{currentstroke}%
\pgfsetdash{}{0pt}%
\pgfpathmoveto{\pgfqpoint{0.520798in}{0.878119in}}%
\pgfpathlineto{\pgfqpoint{1.239803in}{0.878119in}}%
\pgfusepath{stroke}%
\end{pgfscope}%
\begin{pgfscope}%
\definecolor{textcolor}{rgb}{0.150000,0.150000,0.150000}%
\pgfsetstrokecolor{textcolor}%
\pgfsetfillcolor{textcolor}%
\pgftext[x=0.214138in, y=0.839857in, left, base]{\color{textcolor}\rmfamily\fontsize{8.000000}{9.600000}\selectfont 50\%}%
\end{pgfscope}%
\begin{pgfscope}%
\pgfpathrectangle{\pgfqpoint{0.520798in}{0.442177in}}{\pgfqpoint{0.719005in}{0.871884in}}%
\pgfusepath{clip}%
\pgfsetroundcap%
\pgfsetroundjoin%
\pgfsetlinewidth{0.501875pt}%
\definecolor{currentstroke}{rgb}{0.800000,0.800000,0.800000}%
\pgfsetstrokecolor{currentstroke}%
\pgfsetdash{}{0pt}%
\pgfpathmoveto{\pgfqpoint{0.520798in}{1.096090in}}%
\pgfpathlineto{\pgfqpoint{1.239803in}{1.096090in}}%
\pgfusepath{stroke}%
\end{pgfscope}%
\begin{pgfscope}%
\definecolor{textcolor}{rgb}{0.150000,0.150000,0.150000}%
\pgfsetstrokecolor{textcolor}%
\pgfsetfillcolor{textcolor}%
\pgftext[x=0.214138in, y=1.057828in, left, base]{\color{textcolor}\rmfamily\fontsize{8.000000}{9.600000}\selectfont 75\%}%
\end{pgfscope}%
\begin{pgfscope}%
\pgfpathrectangle{\pgfqpoint{0.520798in}{0.442177in}}{\pgfqpoint{0.719005in}{0.871884in}}%
\pgfusepath{clip}%
\pgfsetroundcap%
\pgfsetroundjoin%
\pgfsetlinewidth{0.501875pt}%
\definecolor{currentstroke}{rgb}{0.800000,0.800000,0.800000}%
\pgfsetstrokecolor{currentstroke}%
\pgfsetdash{}{0pt}%
\pgfpathmoveto{\pgfqpoint{0.520798in}{1.314061in}}%
\pgfpathlineto{\pgfqpoint{1.239803in}{1.314061in}}%
\pgfusepath{stroke}%
\end{pgfscope}%
\begin{pgfscope}%
\definecolor{textcolor}{rgb}{0.150000,0.150000,0.150000}%
\pgfsetstrokecolor{textcolor}%
\pgfsetfillcolor{textcolor}%
\pgftext[x=0.155124in, y=1.275799in, left, base]{\color{textcolor}\rmfamily\fontsize{8.000000}{9.600000}\selectfont 100\%}%
\end{pgfscope}%
\begin{pgfscope}%
\definecolor{textcolor}{rgb}{0.150000,0.150000,0.150000}%
\pgfsetstrokecolor{textcolor}%
\pgfsetfillcolor{textcolor}%
\pgftext[x=0.099569in,y=0.878119in,,bottom,rotate=90.000000]{\color{textcolor}\rmfamily\fontsize{10.000000}{12.000000}\selectfont Cert. Acc.}%
\end{pgfscope}%
\begin{pgfscope}%
\pgfsetrectcap%
\pgfsetmiterjoin%
\pgfsetlinewidth{0.752812pt}%
\definecolor{currentstroke}{rgb}{0.700000,0.700000,0.700000}%
\pgfsetstrokecolor{currentstroke}%
\pgfsetdash{}{0pt}%
\pgfpathmoveto{\pgfqpoint{0.520798in}{0.442177in}}%
\pgfpathlineto{\pgfqpoint{0.520798in}{1.314061in}}%
\pgfusepath{stroke}%
\end{pgfscope}%
\begin{pgfscope}%
\pgfsetrectcap%
\pgfsetmiterjoin%
\pgfsetlinewidth{0.752812pt}%
\definecolor{currentstroke}{rgb}{0.700000,0.700000,0.700000}%
\pgfsetstrokecolor{currentstroke}%
\pgfsetdash{}{0pt}%
\pgfpathmoveto{\pgfqpoint{1.239803in}{0.442177in}}%
\pgfpathlineto{\pgfqpoint{1.239803in}{1.314061in}}%
\pgfusepath{stroke}%
\end{pgfscope}%
\begin{pgfscope}%
\pgfsetrectcap%
\pgfsetmiterjoin%
\pgfsetlinewidth{0.752812pt}%
\definecolor{currentstroke}{rgb}{0.700000,0.700000,0.700000}%
\pgfsetstrokecolor{currentstroke}%
\pgfsetdash{}{0pt}%
\pgfpathmoveto{\pgfqpoint{0.520798in}{0.442177in}}%
\pgfpathlineto{\pgfqpoint{1.239803in}{0.442177in}}%
\pgfusepath{stroke}%
\end{pgfscope}%
\begin{pgfscope}%
\pgfsetrectcap%
\pgfsetmiterjoin%
\pgfsetlinewidth{0.752812pt}%
\definecolor{currentstroke}{rgb}{0.700000,0.700000,0.700000}%
\pgfsetstrokecolor{currentstroke}%
\pgfsetdash{}{0pt}%
\pgfpathmoveto{\pgfqpoint{0.520798in}{1.314061in}}%
\pgfpathlineto{\pgfqpoint{1.239803in}{1.314061in}}%
\pgfusepath{stroke}%
\end{pgfscope}%
\begin{pgfscope}%
\pgfpathrectangle{\pgfqpoint{0.520798in}{0.442177in}}{\pgfqpoint{0.719005in}{0.871884in}}%
\pgfusepath{clip}%
\pgfsetroundcap%
\pgfsetroundjoin%
\pgfsetlinewidth{1.003750pt}%
\definecolor{currentstroke}{rgb}{0.003922,0.450980,0.698039}%
\pgfsetstrokecolor{currentstroke}%
\pgfsetdash{}{0pt}%
\pgfpathmoveto{\pgfqpoint{0.520798in}{1.044627in}}%
\pgfpathlineto{\pgfqpoint{0.556748in}{1.044627in}}%
\pgfpathlineto{\pgfqpoint{0.556748in}{0.912559in}}%
\pgfpathlineto{\pgfqpoint{0.628649in}{0.912559in}}%
\pgfpathlineto{\pgfqpoint{0.628649in}{0.712085in}}%
\pgfpathlineto{\pgfqpoint{0.700549in}{0.712085in}}%
\pgfpathlineto{\pgfqpoint{0.700549in}{0.466692in}}%
\pgfpathlineto{\pgfqpoint{0.772450in}{0.466692in}}%
\pgfpathlineto{\pgfqpoint{0.772450in}{0.446526in}}%
\pgfpathlineto{\pgfqpoint{0.844350in}{0.446526in}}%
\pgfpathlineto{\pgfqpoint{0.844350in}{0.444470in}}%
\pgfpathlineto{\pgfqpoint{0.916251in}{0.444470in}}%
\pgfpathlineto{\pgfqpoint{0.916251in}{0.443917in}}%
\pgfpathlineto{\pgfqpoint{0.988151in}{0.443917in}}%
\pgfpathlineto{\pgfqpoint{0.988151in}{0.443521in}}%
\pgfpathlineto{\pgfqpoint{1.060052in}{0.443521in}}%
\pgfpathlineto{\pgfqpoint{1.060052in}{0.442572in}}%
\pgfpathlineto{\pgfqpoint{1.239803in}{0.442572in}}%
\pgfpathlineto{\pgfqpoint{1.239803in}{0.442493in}}%
\pgfpathlineto{\pgfqpoint{1.241470in}{0.442493in}}%
\pgfusepath{stroke}%
\end{pgfscope}%
\begin{pgfscope}%
\pgfpathrectangle{\pgfqpoint{0.520798in}{0.442177in}}{\pgfqpoint{0.719005in}{0.871884in}}%
\pgfusepath{clip}%
\pgfsetbuttcap%
\pgfsetroundjoin%
\definecolor{currentfill}{rgb}{0.003922,0.450980,0.698039}%
\pgfsetfillcolor{currentfill}%
\pgfsetfillopacity{0.500000}%
\pgfsetlinewidth{0.000000pt}%
\definecolor{currentstroke}{rgb}{0.003922,0.450980,0.698039}%
\pgfsetstrokecolor{currentstroke}%
\pgfsetstrokeopacity{0.500000}%
\pgfsetdash{}{0pt}%
\pgfpathmoveto{\pgfqpoint{0.520798in}{1.063547in}}%
\pgfpathlineto{\pgfqpoint{0.520798in}{1.025707in}}%
\pgfpathlineto{\pgfqpoint{0.556748in}{1.025707in}}%
\pgfpathlineto{\pgfqpoint{0.556748in}{0.901723in}}%
\pgfpathlineto{\pgfqpoint{0.628649in}{0.901723in}}%
\pgfpathlineto{\pgfqpoint{0.628649in}{0.705865in}}%
\pgfpathlineto{\pgfqpoint{0.700549in}{0.705865in}}%
\pgfpathlineto{\pgfqpoint{0.700549in}{0.463919in}}%
\pgfpathlineto{\pgfqpoint{0.772450in}{0.463919in}}%
\pgfpathlineto{\pgfqpoint{0.772450in}{0.445495in}}%
\pgfpathlineto{\pgfqpoint{0.844350in}{0.445495in}}%
\pgfpathlineto{\pgfqpoint{0.844350in}{0.443590in}}%
\pgfpathlineto{\pgfqpoint{0.916251in}{0.443590in}}%
\pgfpathlineto{\pgfqpoint{0.916251in}{0.443380in}}%
\pgfpathlineto{\pgfqpoint{0.988151in}{0.443380in}}%
\pgfpathlineto{\pgfqpoint{0.988151in}{0.443118in}}%
\pgfpathlineto{\pgfqpoint{1.060052in}{0.443118in}}%
\pgfpathlineto{\pgfqpoint{1.060052in}{0.442572in}}%
\pgfpathlineto{\pgfqpoint{1.239803in}{0.442572in}}%
\pgfpathlineto{\pgfqpoint{1.239803in}{0.442335in}}%
\pgfpathlineto{\pgfqpoint{1.419554in}{0.442335in}}%
\pgfpathlineto{\pgfqpoint{1.419554in}{0.442177in}}%
\pgfpathlineto{\pgfqpoint{2.066659in}{0.442177in}}%
\pgfpathlineto{\pgfqpoint{2.066659in}{0.442177in}}%
\pgfpathlineto{\pgfqpoint{2.677813in}{0.442177in}}%
\pgfpathlineto{\pgfqpoint{2.677813in}{0.442177in}}%
\pgfpathlineto{\pgfqpoint{2.677813in}{0.442177in}}%
\pgfpathlineto{\pgfqpoint{2.066659in}{0.442177in}}%
\pgfpathlineto{\pgfqpoint{2.066659in}{0.442177in}}%
\pgfpathlineto{\pgfqpoint{1.419554in}{0.442177in}}%
\pgfpathlineto{\pgfqpoint{1.419554in}{0.442651in}}%
\pgfpathlineto{\pgfqpoint{1.239803in}{0.442651in}}%
\pgfpathlineto{\pgfqpoint{1.239803in}{0.442572in}}%
\pgfpathlineto{\pgfqpoint{1.060052in}{0.442572in}}%
\pgfpathlineto{\pgfqpoint{1.060052in}{0.443924in}}%
\pgfpathlineto{\pgfqpoint{0.988151in}{0.443924in}}%
\pgfpathlineto{\pgfqpoint{0.988151in}{0.444453in}}%
\pgfpathlineto{\pgfqpoint{0.916251in}{0.444453in}}%
\pgfpathlineto{\pgfqpoint{0.916251in}{0.445351in}}%
\pgfpathlineto{\pgfqpoint{0.844350in}{0.445351in}}%
\pgfpathlineto{\pgfqpoint{0.844350in}{0.447557in}}%
\pgfpathlineto{\pgfqpoint{0.772450in}{0.447557in}}%
\pgfpathlineto{\pgfqpoint{0.772450in}{0.469466in}}%
\pgfpathlineto{\pgfqpoint{0.700549in}{0.469466in}}%
\pgfpathlineto{\pgfqpoint{0.700549in}{0.718305in}}%
\pgfpathlineto{\pgfqpoint{0.628649in}{0.718305in}}%
\pgfpathlineto{\pgfqpoint{0.628649in}{0.923396in}}%
\pgfpathlineto{\pgfqpoint{0.556748in}{0.923396in}}%
\pgfpathlineto{\pgfqpoint{0.556748in}{1.063547in}}%
\pgfpathlineto{\pgfqpoint{0.520798in}{1.063547in}}%
\pgfpathlineto{\pgfqpoint{0.520798in}{1.063547in}}%
\pgfpathclose%
\pgfusepath{fill}%
\end{pgfscope}%
\begin{pgfscope}%
\pgfpathrectangle{\pgfqpoint{0.520798in}{0.442177in}}{\pgfqpoint{0.719005in}{0.871884in}}%
\pgfusepath{clip}%
\pgfsetroundcap%
\pgfsetroundjoin%
\pgfsetlinewidth{1.003750pt}%
\definecolor{currentstroke}{rgb}{0.870588,0.560784,0.019608}%
\pgfsetstrokecolor{currentstroke}%
\pgfsetdash{}{0pt}%
\pgfpathmoveto{\pgfqpoint{0.520798in}{1.044627in}}%
\pgfpathlineto{\pgfqpoint{0.556748in}{1.044627in}}%
\pgfpathlineto{\pgfqpoint{0.556748in}{0.912559in}}%
\pgfpathlineto{\pgfqpoint{0.628649in}{0.912559in}}%
\pgfpathlineto{\pgfqpoint{0.628649in}{0.712085in}}%
\pgfpathlineto{\pgfqpoint{0.700549in}{0.712085in}}%
\pgfpathlineto{\pgfqpoint{0.700549in}{0.466692in}}%
\pgfpathlineto{\pgfqpoint{0.772450in}{0.466692in}}%
\pgfpathlineto{\pgfqpoint{0.772450in}{0.446526in}}%
\pgfpathlineto{\pgfqpoint{0.844350in}{0.446526in}}%
\pgfpathlineto{\pgfqpoint{0.844350in}{0.444549in}}%
\pgfpathlineto{\pgfqpoint{0.916251in}{0.444549in}}%
\pgfpathlineto{\pgfqpoint{0.916251in}{0.443917in}}%
\pgfpathlineto{\pgfqpoint{0.988151in}{0.443917in}}%
\pgfpathlineto{\pgfqpoint{0.988151in}{0.443521in}}%
\pgfpathlineto{\pgfqpoint{1.060052in}{0.443521in}}%
\pgfpathlineto{\pgfqpoint{1.060052in}{0.442572in}}%
\pgfpathlineto{\pgfqpoint{1.239803in}{0.442572in}}%
\pgfpathlineto{\pgfqpoint{1.239803in}{0.442493in}}%
\pgfpathlineto{\pgfqpoint{1.241470in}{0.442493in}}%
\pgfusepath{stroke}%
\end{pgfscope}%
\begin{pgfscope}%
\pgfpathrectangle{\pgfqpoint{0.520798in}{0.442177in}}{\pgfqpoint{0.719005in}{0.871884in}}%
\pgfusepath{clip}%
\pgfsetbuttcap%
\pgfsetroundjoin%
\definecolor{currentfill}{rgb}{0.870588,0.560784,0.019608}%
\pgfsetfillcolor{currentfill}%
\pgfsetfillopacity{0.500000}%
\pgfsetlinewidth{0.000000pt}%
\definecolor{currentstroke}{rgb}{0.870588,0.560784,0.019608}%
\pgfsetstrokecolor{currentstroke}%
\pgfsetstrokeopacity{0.500000}%
\pgfsetdash{}{0pt}%
\pgfpathmoveto{\pgfqpoint{0.520798in}{1.063547in}}%
\pgfpathlineto{\pgfqpoint{0.520798in}{1.025707in}}%
\pgfpathlineto{\pgfqpoint{0.556748in}{1.025707in}}%
\pgfpathlineto{\pgfqpoint{0.556748in}{0.901723in}}%
\pgfpathlineto{\pgfqpoint{0.628649in}{0.901723in}}%
\pgfpathlineto{\pgfqpoint{0.628649in}{0.705865in}}%
\pgfpathlineto{\pgfqpoint{0.700549in}{0.705865in}}%
\pgfpathlineto{\pgfqpoint{0.700549in}{0.463919in}}%
\pgfpathlineto{\pgfqpoint{0.772450in}{0.463919in}}%
\pgfpathlineto{\pgfqpoint{0.772450in}{0.445495in}}%
\pgfpathlineto{\pgfqpoint{0.844350in}{0.445495in}}%
\pgfpathlineto{\pgfqpoint{0.844350in}{0.443581in}}%
\pgfpathlineto{\pgfqpoint{0.916251in}{0.443581in}}%
\pgfpathlineto{\pgfqpoint{0.916251in}{0.443380in}}%
\pgfpathlineto{\pgfqpoint{0.988151in}{0.443380in}}%
\pgfpathlineto{\pgfqpoint{0.988151in}{0.443118in}}%
\pgfpathlineto{\pgfqpoint{1.060052in}{0.443118in}}%
\pgfpathlineto{\pgfqpoint{1.060052in}{0.442572in}}%
\pgfpathlineto{\pgfqpoint{1.239803in}{0.442572in}}%
\pgfpathlineto{\pgfqpoint{1.239803in}{0.442335in}}%
\pgfpathlineto{\pgfqpoint{1.419554in}{0.442335in}}%
\pgfpathlineto{\pgfqpoint{1.419554in}{0.442177in}}%
\pgfpathlineto{\pgfqpoint{2.066659in}{0.442177in}}%
\pgfpathlineto{\pgfqpoint{2.066659in}{0.442177in}}%
\pgfpathlineto{\pgfqpoint{2.677813in}{0.442177in}}%
\pgfpathlineto{\pgfqpoint{2.677813in}{0.442177in}}%
\pgfpathlineto{\pgfqpoint{2.677813in}{0.442177in}}%
\pgfpathlineto{\pgfqpoint{2.066659in}{0.442177in}}%
\pgfpathlineto{\pgfqpoint{2.066659in}{0.442177in}}%
\pgfpathlineto{\pgfqpoint{1.419554in}{0.442177in}}%
\pgfpathlineto{\pgfqpoint{1.419554in}{0.442651in}}%
\pgfpathlineto{\pgfqpoint{1.239803in}{0.442651in}}%
\pgfpathlineto{\pgfqpoint{1.239803in}{0.442572in}}%
\pgfpathlineto{\pgfqpoint{1.060052in}{0.442572in}}%
\pgfpathlineto{\pgfqpoint{1.060052in}{0.443924in}}%
\pgfpathlineto{\pgfqpoint{0.988151in}{0.443924in}}%
\pgfpathlineto{\pgfqpoint{0.988151in}{0.444453in}}%
\pgfpathlineto{\pgfqpoint{0.916251in}{0.444453in}}%
\pgfpathlineto{\pgfqpoint{0.916251in}{0.445518in}}%
\pgfpathlineto{\pgfqpoint{0.844350in}{0.445518in}}%
\pgfpathlineto{\pgfqpoint{0.844350in}{0.447557in}}%
\pgfpathlineto{\pgfqpoint{0.772450in}{0.447557in}}%
\pgfpathlineto{\pgfqpoint{0.772450in}{0.469466in}}%
\pgfpathlineto{\pgfqpoint{0.700549in}{0.469466in}}%
\pgfpathlineto{\pgfqpoint{0.700549in}{0.718305in}}%
\pgfpathlineto{\pgfqpoint{0.628649in}{0.718305in}}%
\pgfpathlineto{\pgfqpoint{0.628649in}{0.923396in}}%
\pgfpathlineto{\pgfqpoint{0.556748in}{0.923396in}}%
\pgfpathlineto{\pgfqpoint{0.556748in}{1.063547in}}%
\pgfpathlineto{\pgfqpoint{0.520798in}{1.063547in}}%
\pgfpathlineto{\pgfqpoint{0.520798in}{1.063547in}}%
\pgfpathclose%
\pgfusepath{fill}%
\end{pgfscope}%
\begin{pgfscope}%
\pgfpathrectangle{\pgfqpoint{0.520798in}{0.442177in}}{\pgfqpoint{0.719005in}{0.871884in}}%
\pgfusepath{clip}%
\pgfsetroundcap%
\pgfsetroundjoin%
\pgfsetlinewidth{1.003750pt}%
\definecolor{currentstroke}{rgb}{0.007843,0.619608,0.450980}%
\pgfsetstrokecolor{currentstroke}%
\pgfsetdash{}{0pt}%
\pgfpathmoveto{\pgfqpoint{0.520798in}{1.044627in}}%
\pgfpathlineto{\pgfqpoint{0.556748in}{1.044627in}}%
\pgfpathlineto{\pgfqpoint{0.556748in}{0.912559in}}%
\pgfpathlineto{\pgfqpoint{0.628649in}{0.912559in}}%
\pgfpathlineto{\pgfqpoint{0.628649in}{0.712085in}}%
\pgfpathlineto{\pgfqpoint{0.700549in}{0.712085in}}%
\pgfpathlineto{\pgfqpoint{0.700549in}{0.466692in}}%
\pgfpathlineto{\pgfqpoint{0.772450in}{0.466692in}}%
\pgfpathlineto{\pgfqpoint{0.772450in}{0.446526in}}%
\pgfpathlineto{\pgfqpoint{0.844350in}{0.446526in}}%
\pgfpathlineto{\pgfqpoint{0.844350in}{0.444549in}}%
\pgfpathlineto{\pgfqpoint{0.916251in}{0.444549in}}%
\pgfpathlineto{\pgfqpoint{0.916251in}{0.443917in}}%
\pgfpathlineto{\pgfqpoint{0.988151in}{0.443917in}}%
\pgfpathlineto{\pgfqpoint{0.988151in}{0.443521in}}%
\pgfpathlineto{\pgfqpoint{1.060052in}{0.443521in}}%
\pgfpathlineto{\pgfqpoint{1.060052in}{0.442572in}}%
\pgfpathlineto{\pgfqpoint{1.239803in}{0.442572in}}%
\pgfpathlineto{\pgfqpoint{1.239803in}{0.442493in}}%
\pgfpathlineto{\pgfqpoint{1.241470in}{0.442493in}}%
\pgfusepath{stroke}%
\end{pgfscope}%
\begin{pgfscope}%
\pgfpathrectangle{\pgfqpoint{0.520798in}{0.442177in}}{\pgfqpoint{0.719005in}{0.871884in}}%
\pgfusepath{clip}%
\pgfsetbuttcap%
\pgfsetroundjoin%
\definecolor{currentfill}{rgb}{0.007843,0.619608,0.450980}%
\pgfsetfillcolor{currentfill}%
\pgfsetfillopacity{0.500000}%
\pgfsetlinewidth{0.000000pt}%
\definecolor{currentstroke}{rgb}{0.007843,0.619608,0.450980}%
\pgfsetstrokecolor{currentstroke}%
\pgfsetstrokeopacity{0.500000}%
\pgfsetdash{}{0pt}%
\pgfpathmoveto{\pgfqpoint{0.520798in}{1.063547in}}%
\pgfpathlineto{\pgfqpoint{0.520798in}{1.025707in}}%
\pgfpathlineto{\pgfqpoint{0.556748in}{1.025707in}}%
\pgfpathlineto{\pgfqpoint{0.556748in}{0.901723in}}%
\pgfpathlineto{\pgfqpoint{0.628649in}{0.901723in}}%
\pgfpathlineto{\pgfqpoint{0.628649in}{0.705865in}}%
\pgfpathlineto{\pgfqpoint{0.700549in}{0.705865in}}%
\pgfpathlineto{\pgfqpoint{0.700549in}{0.463919in}}%
\pgfpathlineto{\pgfqpoint{0.772450in}{0.463919in}}%
\pgfpathlineto{\pgfqpoint{0.772450in}{0.445495in}}%
\pgfpathlineto{\pgfqpoint{0.844350in}{0.445495in}}%
\pgfpathlineto{\pgfqpoint{0.844350in}{0.443581in}}%
\pgfpathlineto{\pgfqpoint{0.916251in}{0.443581in}}%
\pgfpathlineto{\pgfqpoint{0.916251in}{0.443380in}}%
\pgfpathlineto{\pgfqpoint{0.988151in}{0.443380in}}%
\pgfpathlineto{\pgfqpoint{0.988151in}{0.443118in}}%
\pgfpathlineto{\pgfqpoint{1.060052in}{0.443118in}}%
\pgfpathlineto{\pgfqpoint{1.060052in}{0.442572in}}%
\pgfpathlineto{\pgfqpoint{1.239803in}{0.442572in}}%
\pgfpathlineto{\pgfqpoint{1.239803in}{0.442335in}}%
\pgfpathlineto{\pgfqpoint{1.419554in}{0.442335in}}%
\pgfpathlineto{\pgfqpoint{1.419554in}{0.442177in}}%
\pgfpathlineto{\pgfqpoint{2.066659in}{0.442177in}}%
\pgfpathlineto{\pgfqpoint{2.066659in}{0.442177in}}%
\pgfpathlineto{\pgfqpoint{2.677813in}{0.442177in}}%
\pgfpathlineto{\pgfqpoint{2.677813in}{0.442177in}}%
\pgfpathlineto{\pgfqpoint{2.677813in}{0.442177in}}%
\pgfpathlineto{\pgfqpoint{2.066659in}{0.442177in}}%
\pgfpathlineto{\pgfqpoint{2.066659in}{0.442177in}}%
\pgfpathlineto{\pgfqpoint{1.419554in}{0.442177in}}%
\pgfpathlineto{\pgfqpoint{1.419554in}{0.442651in}}%
\pgfpathlineto{\pgfqpoint{1.239803in}{0.442651in}}%
\pgfpathlineto{\pgfqpoint{1.239803in}{0.442572in}}%
\pgfpathlineto{\pgfqpoint{1.060052in}{0.442572in}}%
\pgfpathlineto{\pgfqpoint{1.060052in}{0.443924in}}%
\pgfpathlineto{\pgfqpoint{0.988151in}{0.443924in}}%
\pgfpathlineto{\pgfqpoint{0.988151in}{0.444453in}}%
\pgfpathlineto{\pgfqpoint{0.916251in}{0.444453in}}%
\pgfpathlineto{\pgfqpoint{0.916251in}{0.445518in}}%
\pgfpathlineto{\pgfqpoint{0.844350in}{0.445518in}}%
\pgfpathlineto{\pgfqpoint{0.844350in}{0.447557in}}%
\pgfpathlineto{\pgfqpoint{0.772450in}{0.447557in}}%
\pgfpathlineto{\pgfqpoint{0.772450in}{0.469466in}}%
\pgfpathlineto{\pgfqpoint{0.700549in}{0.469466in}}%
\pgfpathlineto{\pgfqpoint{0.700549in}{0.718305in}}%
\pgfpathlineto{\pgfqpoint{0.628649in}{0.718305in}}%
\pgfpathlineto{\pgfqpoint{0.628649in}{0.923396in}}%
\pgfpathlineto{\pgfqpoint{0.556748in}{0.923396in}}%
\pgfpathlineto{\pgfqpoint{0.556748in}{1.063547in}}%
\pgfpathlineto{\pgfqpoint{0.520798in}{1.063547in}}%
\pgfpathlineto{\pgfqpoint{0.520798in}{1.063547in}}%
\pgfpathclose%
\pgfusepath{fill}%
\end{pgfscope}%
\begin{pgfscope}%
\definecolor{textcolor}{rgb}{0.150000,0.150000,0.150000}%
\pgfsetstrokecolor{textcolor}%
\pgfsetfillcolor{textcolor}%
\pgftext[x=0.880300in,y=1.397395in,,base]{\color{textcolor}\rmfamily\fontsize{9.000000}{10.800000}\selectfont \(\displaystyle c_A^+=1\)}%
\end{pgfscope}%
\begin{pgfscope}%
\pgfsetbuttcap%
\pgfsetmiterjoin%
\definecolor{currentfill}{rgb}{1.000000,1.000000,1.000000}%
\pgfsetfillcolor{currentfill}%
\pgfsetfillopacity{0.800000}%
\pgfsetlinewidth{1.003750pt}%
\definecolor{currentstroke}{rgb}{0.800000,0.800000,0.800000}%
\pgfsetstrokecolor{currentstroke}%
\pgfsetstrokeopacity{0.800000}%
\pgfsetdash{}{0pt}%
\pgfpathmoveto{\pgfqpoint{0.785843in}{0.638103in}}%
\pgfpathlineto{\pgfqpoint{1.191192in}{0.638103in}}%
\pgfpathlineto{\pgfqpoint{1.191192in}{1.265450in}}%
\pgfpathlineto{\pgfqpoint{0.785843in}{1.265450in}}%
\pgfpathlineto{\pgfqpoint{0.785843in}{0.638103in}}%
\pgfpathclose%
\pgfusepath{stroke,fill}%
\end{pgfscope}%
\begin{pgfscope}%
\definecolor{textcolor}{rgb}{0.150000,0.150000,0.150000}%
\pgfsetstrokecolor{textcolor}%
\pgfsetfillcolor{textcolor}%
\pgftext[x=0.917113in,y=1.113275in,left,base]{\color{textcolor}\rmfamily\fontsize{9.000000}{10.800000}\selectfont \(\displaystyle c_A^-\)}%
\end{pgfscope}%
\begin{pgfscope}%
\pgfsetroundcap%
\pgfsetroundjoin%
\pgfsetlinewidth{1.003750pt}%
\definecolor{currentstroke}{rgb}{0.003922,0.450980,0.698039}%
\pgfsetstrokecolor{currentstroke}%
\pgfsetdash{}{0pt}%
\pgfpathmoveto{\pgfqpoint{0.824732in}{1.001052in}}%
\pgfpathlineto{\pgfqpoint{0.873343in}{1.001052in}}%
\pgfpathlineto{\pgfqpoint{0.873343in}{1.001052in}}%
\pgfpathlineto{\pgfqpoint{0.970565in}{1.001052in}}%
\pgfpathlineto{\pgfqpoint{0.970565in}{1.001052in}}%
\pgfpathlineto{\pgfqpoint{1.019176in}{1.001052in}}%
\pgfusepath{stroke}%
\end{pgfscope}%
\begin{pgfscope}%
\definecolor{textcolor}{rgb}{0.150000,0.150000,0.150000}%
\pgfsetstrokecolor{textcolor}%
\pgfsetfillcolor{textcolor}%
\pgftext[x=1.096954in,y=0.967024in,left,base]{\color{textcolor}\rmfamily\fontsize{7.000000}{8.400000}\selectfont 1}%
\end{pgfscope}%
\begin{pgfscope}%
\pgfsetroundcap%
\pgfsetroundjoin%
\pgfsetlinewidth{1.003750pt}%
\definecolor{currentstroke}{rgb}{0.870588,0.560784,0.019608}%
\pgfsetstrokecolor{currentstroke}%
\pgfsetdash{}{0pt}%
\pgfpathmoveto{\pgfqpoint{0.824732in}{0.865486in}}%
\pgfpathlineto{\pgfqpoint{0.873343in}{0.865486in}}%
\pgfpathlineto{\pgfqpoint{0.873343in}{0.865486in}}%
\pgfpathlineto{\pgfqpoint{0.970565in}{0.865486in}}%
\pgfpathlineto{\pgfqpoint{0.970565in}{0.865486in}}%
\pgfpathlineto{\pgfqpoint{1.019176in}{0.865486in}}%
\pgfusepath{stroke}%
\end{pgfscope}%
\begin{pgfscope}%
\definecolor{textcolor}{rgb}{0.150000,0.150000,0.150000}%
\pgfsetstrokecolor{textcolor}%
\pgfsetfillcolor{textcolor}%
\pgftext[x=1.096954in,y=0.831458in,left,base]{\color{textcolor}\rmfamily\fontsize{7.000000}{8.400000}\selectfont 2}%
\end{pgfscope}%
\begin{pgfscope}%
\pgfsetroundcap%
\pgfsetroundjoin%
\pgfsetlinewidth{1.003750pt}%
\definecolor{currentstroke}{rgb}{0.007843,0.619608,0.450980}%
\pgfsetstrokecolor{currentstroke}%
\pgfsetdash{}{0pt}%
\pgfpathmoveto{\pgfqpoint{0.824732in}{0.729919in}}%
\pgfpathlineto{\pgfqpoint{0.873343in}{0.729919in}}%
\pgfpathlineto{\pgfqpoint{0.873343in}{0.729919in}}%
\pgfpathlineto{\pgfqpoint{0.970565in}{0.729919in}}%
\pgfpathlineto{\pgfqpoint{0.970565in}{0.729919in}}%
\pgfpathlineto{\pgfqpoint{1.019176in}{0.729919in}}%
\pgfusepath{stroke}%
\end{pgfscope}%
\begin{pgfscope}%
\definecolor{textcolor}{rgb}{0.150000,0.150000,0.150000}%
\pgfsetstrokecolor{textcolor}%
\pgfsetfillcolor{textcolor}%
\pgftext[x=1.096954in,y=0.695892in,left,base]{\color{textcolor}\rmfamily\fontsize{7.000000}{8.400000}\selectfont 4}%
\end{pgfscope}%
\end{pgfpicture}%
\makeatother%
\endgroup%

%% file: sections/conclusion.tex
\section{Conclusion}
The main goal of this paper is to define adversarial robustness for group equivariant tasks.
To this end, we introduce action-induced distances as the appropriate notion of input distance and consider how predictions should transform for semantics-preserving transformations of inputs. 
If the equivariances of a task and model match, the proposed notion of  robustness can be guaranteed by proving traditional adversarial robustness.
This has two consequences:
Firstly, specialized certification procedures for equivariant architectures can be reused. One just has to reinterpret what is actually guaranteed by these procedures, e.g.\ robustness to perturbations bounded by graph edit distance.
Secondly, randomized smoothing can be used to provide architecture-agnostic guarantees --- but only if the smoothing scheme and measures preserve the model's equivariances.
We experimentally demonstrated the generality of this equivariance-preserving randomized smoothing approach by certifying robustness for graph, point cloud and molecule  models.
Overall, our work provides a sound foundation for future work at the intersection of robust and geometric machine learning.

\textbf{Future work.} Based on~\cref{proposition:certification_reduction}, a direction for future work is to continue making equivariant models more robust to classic threat models, without caring about equivariance.
A more interesting direction is to make attacks, defenses and certificates equivariance-aware. For instance, knowledge about model equivariances could be used to reduce the search space for attacks, disrupt gradient-based attacks (similar to~\cite{Zhang2023}), or derive stronger randomized smoothing guarantees (as proposed for invariant models in~\cite{Schuchardt2022}).
Finally, developing procedures to certify non-equivariant models (e.g.\ vision transformers) or models that are only ``almost equivariant'' (e.g.\ due to interpolation artifacts) under the proposed notion of robustness would be desirable for equivariant computer vision tasks.


%% file: sections/acknowledgements.tex
\section{Acknowledgments and disclosure of funding}
The authors would like to thank Hongwei Jin for assistance with the implementation of their topology attack certificates,
Tom Wollschl{\"a}ger for providing access to code for training molecular force models, 
and Aman Saxena for valuable discussions concerning non-compact sets and non-isometric actions. 
This work has been funded by the Munich Center for Machine Learning, by the DAAD program
Konrad Zuse Schools of Excellence in Artificial Intelligence (sponsored by the Federal Ministry of
Education and Research), and by the German Research Foundation, grant GU 1409/4-1. The authors of this work take full responsibility for its content.

%% file: includeonly_chapters/appendix.tex
\appendix

\input{appendix/extra_experiments}

\clearpage

\input{appendix/experimental_setup}

\clearpage

\input{appendix/action_induced_distance}

\clearpage

\input{appendix/combining_measures_schemes}

\clearpage

\input{appendix/equivariance_smoothing}

\clearpage

\input{appendix/graph_edit_distance}

\clearpage

\input{appendix/local_robustness}

\clearpage

\input{appendix/non_compact_sets}

\input{appendix/non_isometric_actions}

\clearpage

\input{appendix/transformation_specific}

\clearpage

%% file: appendix/extra_experiments.tex
\section{Additional experiments}
\label{appendix:extra_experiments}

\subsection{Point cloud classification}
\label{appendix:extra_experiments_pointclouds}

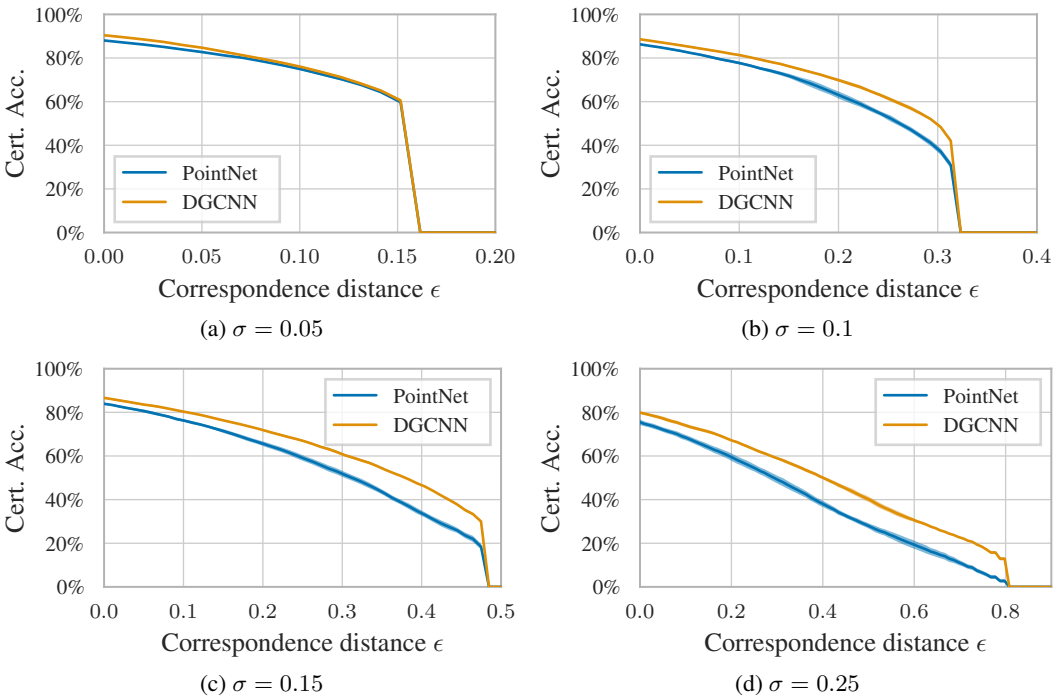
\begin{figure}[hb]
     \centering
     \begin{subfigure}[b]{0.49\textwidth}
         \centering
         \input{figures/experiments/pointclouds/0.05.pgf}
         \vskip-0.2cm
         \caption{$\sigma=0.05$}
     \end{subfigure}
     \hfill
     \begin{subfigure}[b]{0.49\textwidth}
         \centering
         \input{figures/experiments/pointclouds/0.1.pgf}
         \vskip-0.2cm
         \caption{$\sigma=0.1$}
     \end{subfigure}
     \begin{subfigure}[b]{0.49\textwidth}
         \centering
         \input{figures/experiments/pointclouds/0.15.pgf}
         \vskip-0.2cm
         \caption{$\sigma=0.15$}
     \end{subfigure}
     \hfill
     \begin{subfigure}[b]{0.49\textwidth}
         \centering
         \input{figures/experiments/pointclouds/0.25.pgf}
         \vskip-0.2cm
         \caption{$\sigma=0.25$}
     \end{subfigure}
     \caption{Provable robustness of smoothed PointNet and DGCNN point cloud classifiers  on ModelNet40 for different smoothing standard deivations $\sigma$. Correspondence distance $\epsilon$ is the Frobenius distance between point clouds after finding an  optimal matching via permutation.
     For larger $\sigma$, DGCNN offers better accuracy and provable robustness than PointNet.}
     \label{fig:exp_appendix_pointclouds}
\end{figure}

\textbf{Different smoothing standard deviations.} 
In~\cref{fig:exp_appendix_pointclouds} we repeat our experiments on permutation invariant point cloud classification for different smoothing standard deviations $\sigma \in \{0.05, 0.1, 0.15, 0.25\}$.
The maximum certifiable correspondence distance $\epsilon$ increases linearly with $\sigma$.
For $\sigma=0.05$, the average natural accuracy (i.e.\ certified accuracy at $\epsilon=0$) of the smoothed PointNet and DGCNN models is $88\%$ and $90.4\%$, respectively. Their certified accuracies are almost identical for most $\epsilon$.
For the larger $\sigma=0.25$, their accuracy decreases to $75.5\%$ and $79.9\%$, respectively, i.e.\ the difference in accuracy grows by $\SI{2}{p{.}p{.}}$.
DGCNN also offers stronger robustness guarantees with certified accuracies that are up to $\SI{6.3}{p{.}p{.}}$larger.
The most important takeaway should however be that equivariance-preserving randomized smoothing provides sound robustness guarantees for equivariant tasks where discrete groups act on continuous data.

\textbf{Comparison to 3DVerifier.}
As shown in~\cref{fig:exp_appendix_pointclouds}, Gaussian randomized smoothing with $\sigma \in [0.05, 0.25]$ lets us retain relatively high accuracy while certifying robustness to correspondence distance perturbations of sizes up to $\epsilon=0.8$.
However, as discussed in~\cref{section:certification}, \cref{proposition:certification_reduction} does not make any assumptions about how we prove traditional adversarial robustness.
For example, 3DVerifier~\cite{Mu2022}, which is a specialized certification procedure for PointNet, can on average prove robustness robustness to $\ell_2$ norm perturbations of size $\epsilon=0.959$ (see Table 2 in~\cite{Mu2022}).
Because both the point cloud classification task and the PointNet architecture are permutation invariant,
this directly translates to correspondence distance perturbations with an average size of $0.959$.
This experiment demonstrates  that developing specialized certification procedures for a wider range of equivariant models is a promising direction for future work.
Nevertheless, equivariance-preserving randomized smoothing remains valuable as a general-purpose certification procedure.
For instance, 3DVerifier cannot prove robustness for DGCNN or other point cloud classification architectures.

\subsection{Molecular force prediction}
\label{appendix:extra_experiments_molecules}

\begin{figure}[H]
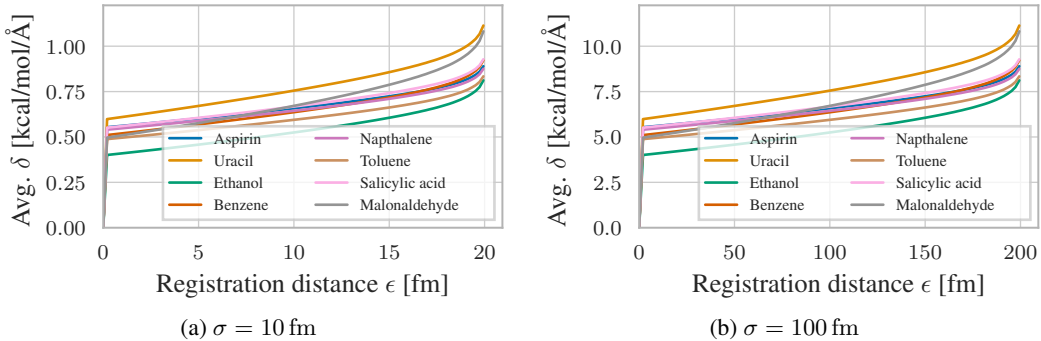

     \centering
     \begin{subfigure}[b]{0.49\textwidth}
         \centering
         \input{figures/experiments/force_fields/0.0001.pgf}
         \vskip-0.2cm
         \caption{$\sigma=\SI{10}{\femto\meter}$}
     \end{subfigure}
     \hfill
     \begin{subfigure}[b]{0.49\textwidth}
         \centering
         \input{figures/experiments/force_fields/0.001.pgf}
         \vskip-0.2cm
         \caption{$\sigma=\SI{100}{\femto\meter}$}
     \end{subfigure}
     \caption{Provable robustness of smoothed DimeNet++ force predictions on MD17
     with smoothing standard deviations $\sigma \in \{\SI{10}{\femto\meter},\SI{100}{\femto\meter}\}$.
     The average certified output distance $\delta$ and the maximum certifiable registration distance $\epsilon$
     grow (approximately) linearly with $\sigma$.
         }
     \label{fig:exp_appendix_molecules}
\end{figure}

\textbf{Different smoothing standard deviations.} 
In~\cref{fig:exp_appendix_molecules} we repeat our experiments on permutation and isometry equivariant  force prediction on MD17 with different smoothing standard deviations $\sigma \in \{\SI{10}{\femto\meter},\SI{100}{\femto\meter}\}$.
The maximum certifiable registration distance grows linearly with $\sigma$. This is inherent to Gaussian smoothing (see~\cite{Cohen2019}).
The average certified output distance $\delta$ also grows approximately linearly.
This can potentially be explained as follows:
In a sufficiently small region around clean input $\mX$, the model may have approximately linear behavior. 
If $\sigma$ is small, the output distribution will be similar to a normal distribution whose covariances grow linearly with $\sigma^2$.
The certified $\delta$ of center smoothing are always between the median and maximum sampled distance from the smoothed prediction (see~\cite{Kumar2021}), which will in this case also increase approximately linearly.

\begin{figure}[H]
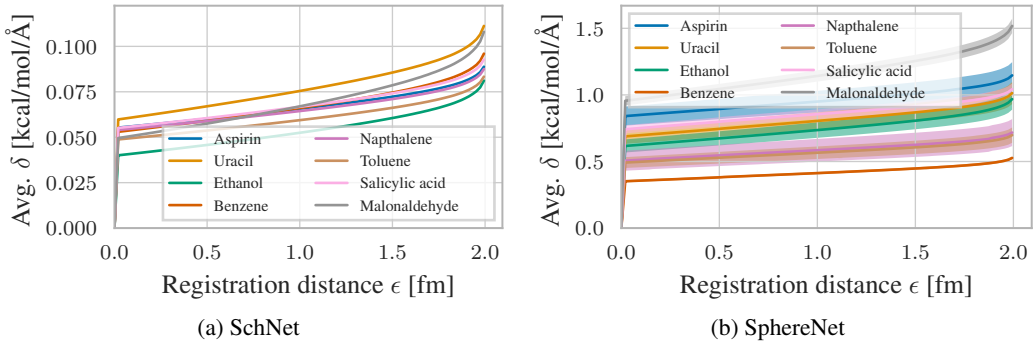

     \centering
     \begin{subfigure}[b]{0.49\textwidth}
         \centering
         \input{figures/experiments/force_fields/schnet/0.00001.pgf}
         \vskip-0.2cm
         \caption{SchNet}
     \end{subfigure}
     \hfill
     \begin{subfigure}[b]{0.49\textwidth}
         \centering
         \input{figures/experiments/force_fields/spherenet/0.00001.pgf}
         \vskip-0.2cm
         \caption{SphereNet}
     \end{subfigure}
     \label{fig:exp_appendix_molecules_models}
     \caption{Provable robustness of smoothed SchNet and SphereNet force predictions on MD17
     with a smoothing standard deviation $\sigma = \SI{1}{\femto\meter}$.
     While SphereNet is more accurate, its provable robustness guarantees are weaker and have higher variance across different random weight initializations.
     }
\end{figure}

\textbf{Different base models.}
In~\cref{fig:exp_appendix_molecules_models} we repeat our experiment with $\sigma = \SI{1}{\femto\meter}$ and different base models, namely SchNet~\cite{Schutt2018} and SphereNet~\cite{Liu2022}.
SphereNet achieves lower average test errors between $0.45$ and $\SI[per-mode=repeated-symbol]{1.4}{\kilo \cal \per \mol \per \angstrom}$
than SchNet, which has test errors between between $0.51$ and $\SI[per-mode=repeated-symbol]{1.89}{\kilo \cal \per \mol \per \angstrom}$.
However, the provable robustness guarantees obtained for SphereNet are worse than those for SchNet by more than an order of magnitude.
For adversarial budgets of up to $\SI{2}{\femto\meter}$, the average certified output distance $\delta$ for SphereNet varies between
$\SI[per-mode=repeated-symbol]{0.35}{\kilo \cal \per \mol \per \angstrom}$ (Benzene) and $\SI[per-mode=repeated-symbol]{1.61}{\kilo \cal \per \mol \per \angstrom}$ (Malonaldehyde).
For comparison, the average certified output $\delta$ for SchNet vary between
$\SI[per-mode=repeated-symbol]{0.04}{\kilo \cal \per \mol \per \angstrom}$ (Ethanol) and $\SI[per-mode=repeated-symbol]{0.12}{\kilo \cal \per \mol \per \angstrom}$ (Uracil) and are quite similar (but not identical to) those for DimeNet++.
We also observe a much higher variance across random seeds.

\textbf{Conclusion.} 
While these observations may be of interest to some readers that are specifically interested in machine learning for molecules, the most important takeaway should be that  equivariance-preserving randomized smoothing can be used to obtain sound robustness guarantees for equivariant tasks in which continuous groups act on continuous data.

\vspace{-0.2cm}
\vspace{-0.1cm}

\clearpage

\subsection{Node and graph classification}
\label{appendix:extra_experiments_ged}
Next, we evaluate the different robustness guarantees for graph and node classification that we generalized to graph edit distance perturbations with user-specified costs (see~\cref{appendix:graph_edit_distance}).

\textbf{Convex outer adversarial polytopes for attribute perturbations.}
In~\cref{fig:exp_appendix_attributes_zuegner_citeseer} we repeat our experiment on proving the robustness of $2$-layer graph convolutional networks to graph edit distance perturbations. We use our generalization of the convex relaxation approach (see~\cref{appendix:graph_edit_distance_convex}), which was originally proposed in~\cite{Zuegner2019cert}, 
on the Citeseer~\cite{Sen2008} node classification dataset.
As before, we set $c_\mA^+ = c_\mA^- = \infty$ and vary the cost of attribute insertion $c_\mX^+$ and deletion $c_\mX^-$.
The result is qualitatively similar to our results on Cora-ML.
Increasing the cost of attribute insertions leads to significantly higher certified accuracy. 
With $c_\mX^- = 1$, the average certified accuracy for  $c_\mX^+ = 4$ at edit distance $\epsilon=25$ is $23.4\%$, which is much larger than the $6.6\%$ for $c_\mX^+ = 1$.
Increasing the cost of deletions does however not have a noticeable effect, which matches the empirical observation that insertions are  more effective when attacking graph neural networks~\cite{Zuegner2020patterns}.
The main difference to Cora-ML is that we have a larger standard deviation across seeds.
This is not necessarily surprising, since many graph benchmarks are known to be sensitive to the choice of data split and model initialization~\cite{Shchur2018}.
Note that the split and initialization for each seed are the same in all experiments.

\begin{figure}[h]
	\centering
	\begin{minipage}{0.49\textwidth}
    \begin{subfigure}[t]{0.49\linewidth}
		\centering
       \input{figures/experiments/graphs/attributes_zuegner/cora_cost_add.pgf}
    \end{subfigure}
    \begin{subfigure}[t]{0.49\linewidth}
		\centering
       \input{figures/experiments/graphs/attributes_zuegner/cora_cost_del.pgf}
    \end{subfigure}
    \vspace{-0.2cm}
    \caption{Convex outer adversarial polytope robustness guarantees for GCNs under attribute perturbations  on Cora-ML.}
    \label{fig:exp_appendix_attributes_zuegner_cora}
	\end{minipage}
	\hfill
	\begin{minipage}{0.49\textwidth}
    \begin{subfigure}[t]{0.49\linewidth}
		\centering
         \input{figures/experiments/graphs/attributes_zuegner/citeseer_cost_add.pgf}
    \end{subfigure}
    \begin{subfigure}[t]{0.49\linewidth}
		\centering
         \input{figures/experiments/graphs/attributes_zuegner/citeseer_cost_del.pgf}
    \end{subfigure}
    \vspace{-0.2cm}
    \caption{Convex  outer adversarial polytope robustness guarantees for GCNs under attribute perturbations on Citeseer.}
    \label{fig:exp_appendix_attributes_zuegner_citeseer}
	\end{minipage}
\end{figure}

\textbf{Interval bound propagation for attribute perturbations.}
In~\cref{fig:exp_appendix_attributes_ibp_cora,fig:exp_appendix_attributes_ibp_citeseer} we repeat the same experiment using our generalization of the interval bound propagation approach (see~\cref{appendix:certs_ibp}) originally proposed in~\cite{Liu2020graphcert}.
Unlike before, increasing a single cost parameter while keeping the other one at $1$ has no noticeable effect on the certified accuracy.
A potential explanation is that interval bound propagation only offers very loose guarantees.
For example, the maximum certifiable edit distance on Cora (see~\cref{fig:exp_appendix_attributes_ibp_cora}) is $10$, compared to over $50$ with the previously discussed convex relaxation approach.
The relaxation might simply be too loose to accurately capture the effect of different perturbation types on the models.

\begin{figure}[h]
	\centering
	\begin{minipage}{0.49\textwidth}
    \begin{subfigure}[t]{0.49\linewidth}
		\centering
       \input{figures/experiments/graphs/attributes_ibp/cora_cost_add.pgf}
    \end{subfigure}
    \begin{subfigure}[t]{0.49\linewidth}
		\centering
       \input{figures/experiments/graphs/attributes_ibp/cora_cost_del.pgf}
    \end{subfigure}
    \vspace{-0.2cm}
    \caption{Interval bound propagation robustness guarantees for GCNs under attribute perturbations on Cora-ML.}
    \label{fig:exp_appendix_attributes_ibp_cora}
	\end{minipage}
	\hfill
	\begin{minipage}{0.49\textwidth}
    \begin{subfigure}[t]{0.49\linewidth}
		\centering
         \input{figures/experiments/graphs/attributes_ibp/citeseer_cost_add.pgf}
    \end{subfigure}
    \begin{subfigure}[t]{0.49\linewidth}
		\centering
         \input{figures/experiments/graphs/attributes_ibp/citeseer_cost_del.pgf}
    \end{subfigure}
    \vspace{-0.2cm}
    \caption{Interval bound propagation robustness guarantees for GCNs under attribute perturbations on Citeseer.}
    \label{fig:exp_appendix_attributes_ibp_citeseer}
	\end{minipage}
\end{figure}

\textbf{Policy iteration for adjacency perturbations.} 
Next, we consider perturbations of the adjacency matrix.
We begin with our  generalization of the policy iteration approach (see~\cref{appendix:graph_edit_distance_policy_iteration}) originally proposed in~\cite{Bojchevski2019}.
This robustness guarantee is specifically designed for \(\pi\)-PPNP, which is APPNP~\cite{Klicpera2019} where the degree normalized adjacency matrix $\mD^{-1 \mathbin{/} 2} \mA \mD^{-1 \mathbin{/} 2}$ is replaced with $\mD^{-1} \mA$.
Different from the other guarantees, we do not have a single global budget $\epsilon$. Instead, we only have nodewise budgets $\rho_1,\dots,\rho_N$ with $\rho_n = (\emD_{n,n} - c + s)$ with an arbitrarily chosen constant $c$ and attack strength $s$. 
Like in~\cite{Bojchevski2019}, we set $c=11$ and vary attack strength $s$.
\cref{fig:exp_appendix_adjacency_policy_iteration_cora,fig:exp_appendix_adjacency_policy_iteration_citeseer} shows the resulting certified accuracies on Cora-ML and Citeseer for $c_\mX^+ = c_\mX^- = \infty$ and varying $c_\mA^+, c_\mA^-$.
Again, increasing the cost of adversarial perturbations to $c_\mA^+ = 4$ has a larger effect, in some cases almost tripling the certified accuracy. 
But there is also a small increase of up to  $\SI{2.9}{p{.}p{.}}$ certified accuracy when increasing the deletion cost $c_\mA^-$ from $1$ to $4$ on Citeseer.

\begin{figure}[h]
	\centering
	\begin{minipage}{0.49\textwidth}
    \begin{subfigure}[t]{0.49\linewidth}
		\centering
       \input{figures/experiments/graphs/piPPNP/piPPNP-Cora-B.pgf}
    \end{subfigure}
    \begin{subfigure}[t]{0.49\linewidth}
		\centering
       \input{figures/experiments/graphs/piPPNP/piPPNP-Cora-A.pgf}
    \end{subfigure}
    \vspace{-0.2cm}
    \caption{Robustness guarantees for \(\pi\)-PPNP under adjacency perturbations on Cora-ML. Note that we do not have a global budget $\epsilon$ and instead vary per-node budgets.}
    \label{fig:exp_appendix_adjacency_policy_iteration_cora}
	\end{minipage}
	\hfill
	\begin{minipage}{0.49\textwidth}
    \begin{subfigure}[t]{0.49\linewidth}
		\centering
       \input{figures/experiments/graphs/piPPNP/piPPNP-Citeseer-B.pgf}
    \end{subfigure}
    \begin{subfigure}[t]{0.49\linewidth}
		\centering
        \input{figures/experiments/graphs/piPPNP/piPPNP-Citeseer-A.pgf}
    \end{subfigure}
    \vspace{-0.2cm}
    \caption{Robustness guarantees for \(\pi\)-PPNP under adjacency perturbations on Citeseer. Note that we do not have a global budget $\epsilon$ and instead vary per-node budgets.}\label{fig:pippnpCiteseer}
    \label{fig:exp_appendix_adjacency_policy_iteration_citeseer}
	\end{minipage}
\end{figure}

\begin{figure}[H]
	\centering
	\begin{minipage}{0.49\textwidth}
    \begin{subfigure}[t]{0.49\linewidth}
		\centering
       \input{figures/experiments/graphs/structure_jin/proteins_cost_add.pgf}
    \end{subfigure}
    \begin{subfigure}[t]{0.49\linewidth}
		\centering
       \input{figures/experiments/graphs/structure_jin/proteins_cost_del.pgf}
    \end{subfigure}
    \vspace{-0.2cm}
    \caption{Linearization and dualization robustness guarantees for GCN graph classifiers under adjacency perturbations on PROTEINS.}
    \label{fig:exp_appendix_adjacency_linearization_proteins}
	\end{minipage}
	\hfill
	\begin{minipage}{0.49\textwidth}
    \begin{subfigure}[t]{0.49\linewidth}
		\centering
         \input{figures/experiments/graphs/structure_jin/mutag_cost_add.pgf}
    \end{subfigure}
    \begin{subfigure}[t]{0.49\linewidth}
		\centering
         \input{figures/experiments/graphs/structure_jin/mutag_cost_del.pgf}
    \end{subfigure}
    \vspace{-0.2cm}
    \caption{Linearization and dualization robustness guarantees for GCN graph classifiers under adjacency perturbations on MUTAG.}
    \label{fig:exp_appendix_adjacency_linearization_mutag}
	\end{minipage}
\end{figure}

\textbf{Bilinear programming for adjacency perturbations.} 
Generalizing the bilinear programming method for proving robustness of GCNs to adjacency perturbations proposed in~\cite{Zuegner2020}
from $c_\mX^+, c_\mX^-, c_\mA^+ = \infty, c_\mX^-=1$ to $c_\mX^- \in \sR_+$ does not require any modifications, since scaling the cost is equivalent to scaling the adversarial budgets (see~\cref{appendix:bilinear_programming}). We thus refer to~\cite{Zuegner2020} for experimental results.

\textbf{Linearization and dualization for adjacency perturbations.}\label{appendix:linearization_and_dualization}
Next, we consider GCN-based node classifiers and adjacency perturbations. We prove robustness via our generalization of the procedure proposed in~\cite{Jin2020}.
This guarantee is specifically designed for single layer GCNs followed by a linear layer and max-pooling.
Since Cora-ML and Citeseer are node classification datasets, we instead use the PROTEINS and MUTAG datasets from the TUDataset~\cite{Morris2020} collection, which were also used in~\cite{Jin2020}.
\cref{fig:exp_appendix_adjacency_linearization_proteins,fig:exp_appendix_adjacency_linearization_mutag} show the resulting certified accuracies for fixed $c_\mX^+ = c_\mX^- = \infty$ and varying adjacency perturbation costs $c_\mA^+$ and $c_\mA^-$.
Unlike in all previous experiments, increasing the cost of deletions also leads to a significant increase for the same edit distance budget $\epsilon$.
On PROTEINS, choosing $c_\mA^+ = 1$ and $c_\mA^-=4$ even yields stronger robustness guarantees than $c_\mA^+=4$ and $c_\mA^-=1$. For example, the average certified accuracies for budget $\epsilon$ are $61.2\%$ and $55.6\%$, respectively (see~\cref{fig:exp_appendix_adjacency_linearization_proteins}).
This may be caused by one of multiple factory.
Firstly, we have a different task -- graph instead of node classificaton.
Secondly, we have a different architecture that aggregates information from the entire graph via max-pooling instead of just using local information.
Thirdly, the considered graphs have a significantly different distribution than Cora-ML and Citeseer.
For instance, Cora-ML and Citeser consist of $2708$ and $3327$ nodes, respectively, whereas the average number of nodes in PROTEINS and MUTAG is $17.9$ and $39.1$. Cora-ML and Citeseer also have much sparser adjacency matrices, with just $0.14\%$ and $0.08\%$ non-zero entries.

\textbf{Sparsity-aware randomized smoothing for attribute and adjacency perturbations.}
Finally, we perform additional experiments with sparsity-aware randomized smoothing~\cite{Bojchevski2020}.
As we discussed in~\cref{section:equivariance_smoothing}, this randomized smoothing method preserves isomorphism equivariance and can thus be used to prove adversarial robustness for arbitrary isomorphism equivariant models and tasks w.r.t.\ graph edit distance and both attribute and adjacency perturbations.
\cref{fig:exp_appendix_attributes_smoothing_cora_gcn,fig:exp_appendix_adjacency_smoothing_cora_gcn,fig:exp_appendix_attributes_smoothing_citeseer_gcn,fig:exp_appendix_adjacency_smoothing_citeseer_gcn,fig:exp_appendix_attributes_smoothing_cora_gat,fig:exp_appendix_adjacency_smoothing_cora_gat,fig:exp_appendix_attributes_smoothing_citeseer_gat,fig:exp_appendix_adjacency_smoothing_citeseer_gat,fig:exp_appendix_attributes_smoothing_cora_appnp,fig:exp_appendix_adjacency_smoothing_cora_appnp,fig:exp_appendix_attributes_smoothing_citeseer_appnp,fig:exp_appendix_adjacency_smoothing_citeseer_appnp} show the certified accuracy of graph convolutional networks, graph attention networks~\cite{Velickovic2018} (GAT), and APPNP on Cora-ML and Citeseer for varying costs of attribute and adjacency perturbations
with smoothing parameters $p_\mX^+=0.001,p_\mX^-=0.8,p_\mA^+=0,p_\mA^-=0$ and $p_\mX^+=0,p_\mX^-=0,p_\mA^+=0.001,p_\mA^-=0.8$
While the different models and datasets differ slightly with respect to (certified) accuracy, we observe the same effect as in our previous node classification experiments:
Only increasing the cost of insertions has a large effect on provable robustness for any graph edit distance $\epsilon$,
which suggests that these perturbations are significantly more effective at changing the prediction of the models.

\textbf{Conclusion.}
Overall, there are three main takeaways from evaluating the different graph edit distance robustness guarantees.
Firstly, even though evaluating the graph edit distance is not tractable, it is possible to prove robust under it to provide sound robustness guarantees for isomorphism equivariant tasks.
Secondly, unless they use very loose relaxations (like interval bound propagation), 
generalizing existing guarantees to non-uniform costs provides more fine-grained insights into the adversarial robustness of different graph and node classification architectures.
Thirdly, equivariance-preserving randomized smoothing is  an effective way of obtaining sound robustness guarantees for equivariant tasks where discrete groups act on discrete data.

\begin{figure}[H]
	\centering
	\begin{minipage}{0.49\textwidth}
    \begin{subfigure}[t]{0.49\linewidth}
		\centering
       \input{figures/experiments/graphs/sparse_smoothing/nodes_attributes/node_classification-Cora-GCN-hidden=32-p_adj_plus=0.0-p_adj_minus=0.0-p_att_plus=0.001-p_att_minus=0.8-multi_class_cert-B.pgf}
    \end{subfigure}
    \begin{subfigure}[t]{0.49\linewidth}
		\centering
       \input{figures/experiments/graphs/sparse_smoothing/nodes_attributes/node_classification-Cora-GCN-hidden=32-p_adj_plus=0.0-p_adj_minus=0.0-p_att_plus=0.001-p_att_minus=0.8-multi_class_cert-A.pgf}
    \end{subfigure}
    \vspace{-0.2cm}
    \caption{Randomized smoothing robustness guarantees for GCNs under attribute perturbations on Cora-ML. The Smoothing parameters are set to $p_\mX^+=0.001,p_\mX^-=0.8,p_\mA^+=0,p_\mA^-=0$.}
    \label{fig:exp_appendix_attributes_smoothing_cora_gcn}
	\end{minipage}
	\hfill
	\begin{minipage}{0.49\textwidth}
    \begin{subfigure}[t]{0.49\linewidth}
		\centering
       \input{figures/experiments/graphs/sparse_smoothing/nodes_structure/node_classification-Cora-GCN-hidden=32-p_adj_plus=0.001-p_adj_minus=0.8-p_att_plus=0.0-p_att_minus=0.0-multi_class_cert-B.pgf}
    \end{subfigure}
    \begin{subfigure}[t]{0.49\linewidth}
		\centering
       \input{figures/experiments/graphs/sparse_smoothing/nodes_structure/node_classification-Cora-GCN-hidden=32-p_adj_plus=0.001-p_adj_minus=0.8-p_att_plus=0.0-p_att_minus=0.0-multi_class_cert-A.pgf}
    \end{subfigure}
    \vspace{-0.2cm}
    \caption{Randomized smoothing robustness guarantees for GCNs under adjacency perturbations on Cora-ML. The moothing parameters are set to $p_\mX^+=0,p_\mX^-=0,p_\mA^+=0.001,p_\mA^-=0.8$.}
    \label{fig:exp_appendix_adjacency_smoothing_cora_gcn}
	\end{minipage}
\end{figure}

\begin{figure}[H]
	\centering
	\begin{minipage}{0.49\textwidth}
    \begin{subfigure}[t]{0.49\linewidth}
		\centering
       \input{figures/experiments/graphs/sparse_smoothing/nodes_attributes/node_classification-Citeseer-GCN-hidden=32-p_adj_plus=0.0-p_adj_minus=0.0-p_att_plus=0.001-p_att_minus=0.8-multi_class_cert-B.pgf}
    \end{subfigure}
    \begin{subfigure}[t]{0.49\linewidth}
		\centering
       \input{figures/experiments/graphs/sparse_smoothing/nodes_attributes/node_classification-Citeseer-GCN-hidden=32-p_adj_plus=0.0-p_adj_minus=0.0-p_att_plus=0.001-p_att_minus=0.8-multi_class_cert-A.pgf}
    \end{subfigure}
    \vspace{-0.2cm}
    \caption{Randomized smoothing robustness guarantees for GCNs under attribute perturbations on Citeseer. The Smoothing parameters are set to $p_\mX^+=0.001,p_\mX^-=0.8,p_\mA^+=0,p_\mA^-=0$.}
    \label{fig:exp_appendix_attributes_smoothing_citeseer_gcn}
	\end{minipage}
	\hfill
	\begin{minipage}{0.49\textwidth}
    \begin{subfigure}[t]{0.49\linewidth}
		\centering
       \input{figures/experiments/graphs/sparse_smoothing/nodes_structure/node_classification-Citeseer-GCN-hidden=32-p_adj_plus=0.001-p_adj_minus=0.8-p_att_plus=0.0-p_att_minus=0.0-multi_class_cert-B.pgf}
    \end{subfigure}
    \begin{subfigure}[t]{0.49\linewidth}
		\centering
       \input{figures/experiments/graphs/sparse_smoothing/nodes_structure/node_classification-Citeseer-GCN-hidden=32-p_adj_plus=0.001-p_adj_minus=0.8-p_att_plus=0.0-p_att_minus=0.0-multi_class_cert-A.pgf}
    \end{subfigure}
    \vspace{-0.2cm}
    \caption{Randomized smoothing robustness guarantees for GCNs under adjacency perturbations on Citeseer. The moothing parameters are set to $p_\mX^+=0,p_\mX^-=0,p_\mA^+=0.001,p_\mA^-=0.8$.}
    \label{fig:exp_appendix_adjacency_smoothing_citeseer_gcn}
	\end{minipage}
\end{figure}

\clearpage

\begin{figure}[H]
	\centering
	\begin{minipage}{0.49\textwidth}
    \begin{subfigure}[t]{0.49\linewidth}
		\centering
       \input{figures/experiments/graphs/sparse_smoothing/nodes_attributes/node_classification-Cora-GAT-hidden=8-p_adj_plus=0.0-p_adj_minus=0.0-p_att_plus=0.001-p_att_minus=0.8-multi_class_cert-B.pgf}
    \end{subfigure}
    \begin{subfigure}[t]{0.49\linewidth}
		\centering
       \input{figures/experiments/graphs/sparse_smoothing/nodes_attributes/node_classification-Cora-GAT-hidden=8-p_adj_plus=0.0-p_adj_minus=0.0-p_att_plus=0.001-p_att_minus=0.8-multi_class_cert-A.pgf}
    \end{subfigure}
    \vspace{-0.2cm}
    \caption{Randomized smoothing robustness guarantees for GATs under attribute perturbations on Cora-ML. The Smoothing parameters are set to $p_\mX^+=0.001,p_\mX^-=0.8,p_\mA^+=0,p_\mA^-=0$.}
    \label{fig:exp_appendix_attributes_smoothing_cora_gat}
	\end{minipage}
	\hfill
	\begin{minipage}{0.49\textwidth}
    \begin{subfigure}[t]{0.49\linewidth}
		\centering
       \input{figures/experiments/graphs/sparse_smoothing/nodes_structure/node_classification-Cora-GAT-hidden=8-p_adj_plus=0.001-p_adj_minus=0.8-p_att_plus=0.0-p_att_minus=0.0-multi_class_cert-B.pgf}
    \end{subfigure}
    \begin{subfigure}[t]{0.49\linewidth}
		\centering
       \input{figures/experiments/graphs/sparse_smoothing/nodes_structure/node_classification-Cora-GAT-hidden=8-p_adj_plus=0.001-p_adj_minus=0.8-p_att_plus=0.0-p_att_minus=0.0-multi_class_cert-A.pgf}
    \end{subfigure}
    \vspace{-0.2cm}
    \caption{Randomized smoothing robustness guarantees for GATs under adjacency perturbations on Cora-ML. The moothing parameters are set to $p_\mX^+=0,p_\mX^-=0,p_\mA^+=0.001,p_\mA^-=0.8$.}
    \label{fig:exp_appendix_adjacency_smoothing_cora_gat}
	\end{minipage}
\end{figure}

\vspace{-0.60cm}
\begin{figure}[H]
	\centering
	\begin{minipage}{0.49\textwidth}
    \begin{subfigure}[t]{0.49\linewidth}
		\centering
       \input{figures/experiments/graphs/sparse_smoothing/nodes_attributes/node_classification-Citeseer-GAT-hidden=8-p_adj_plus=0.0-p_adj_minus=0.0-p_att_plus=0.001-p_att_minus=0.8-multi_class_cert-B.pgf}
    \end{subfigure}
    \begin{subfigure}[t]{0.49\linewidth}
		\centering
       \input{figures/experiments/graphs/sparse_smoothing/nodes_attributes/node_classification-Citeseer-GAT-hidden=8-p_adj_plus=0.0-p_adj_minus=0.0-p_att_plus=0.001-p_att_minus=0.8-multi_class_cert-A.pgf}
    \end{subfigure}
    \vspace{-0.2cm}
    \caption{Randomized smoothing robustness guarantees for GATs under attribute perturbations on Citeseer. The Smoothing parameters are set to $p_\mX^+=0.001,p_\mX^-=0.8,p_\mA^+=0,p_\mA^-=0$.}
    \label{fig:exp_appendix_attributes_smoothing_citeseer_gat}
	\end{minipage}
	\hfill
	\begin{minipage}{0.49\textwidth}
    \begin{subfigure}[t]{0.49\linewidth}
		\centering
       \input{figures/experiments/graphs/sparse_smoothing/nodes_structure/node_classification-Citeseer-GAT-hidden=8-p_adj_plus=0.001-p_adj_minus=0.8-p_att_plus=0.0-p_att_minus=0.0-multi_class_cert-B.pgf}
    \end{subfigure}
    \begin{subfigure}[t]{0.49\linewidth}
		\centering
       \input{figures/experiments/graphs/sparse_smoothing/nodes_structure/node_classification-Citeseer-GAT-hidden=8-p_adj_plus=0.001-p_adj_minus=0.8-p_att_plus=0.0-p_att_minus=0.0-multi_class_cert-A.pgf}
    \end{subfigure}
    \vspace{-0.2cm}
    \caption{Randomized smoothing robustness guarantees for GATs under adjacency perturbations on Citeseer. The moothing parameters are set to $p_\mX^+=0,p_\mX^-=0,p_\mA^+=0.001,p_\mA^-=0.8$.}
    \label{fig:exp_appendix_adjacency_smoothing_citeseer_gat}
	\end{minipage}
\end{figure}
\vspace{-0.60cm}

\begin{figure}[H]
	\centering
	\begin{minipage}{0.49\textwidth}
    \begin{subfigure}[t]{0.49\linewidth}
		\centering
       \input{figures/experiments/graphs/sparse_smoothing/nodes_attributes/node_classification-Cora-APPNP-hidden=32-p_adj_plus=0.0-p_adj_minus=0.0-p_att_plus=0.001-p_att_minus=0.8-multi_class_cert-B.pgf}
    \end{subfigure}
    \begin{subfigure}[t]{0.49\linewidth}
		\centering
       \input{figures/experiments/graphs/sparse_smoothing/nodes_attributes/node_classification-Cora-APPNP-hidden=32-p_adj_plus=0.0-p_adj_minus=0.0-p_att_plus=0.001-p_att_minus=0.8-multi_class_cert-A.pgf}
    \end{subfigure}
    \vspace{-0.2cm}
    \caption{Randomized smoothing robustness guarantees for APPNP under attribute perturbations on Cora-ML. The Smoothing parameters are set to $p_\mX^+=0.001,p_\mX^-=0.8,p_\mA^+=0,p_\mA^-=0$.}
    \label{fig:exp_appendix_attributes_smoothing_cora_appnp}
	\end{minipage}
	\hfill
	\begin{minipage}{0.49\textwidth}
    \begin{subfigure}[t]{0.49\linewidth}
		\centering
       \input{figures/experiments/graphs/sparse_smoothing/nodes_structure/node_classification-Cora-APPNP-hidden=32-p_adj_plus=0.001-p_adj_minus=0.8-p_att_plus=0.0-p_att_minus=0.0-multi_class_cert-B.pgf}
    \end{subfigure}
    \begin{subfigure}[t]{0.49\linewidth}
		\centering
       \input{figures/experiments/graphs/sparse_smoothing/nodes_structure/node_classification-Cora-APPNP-hidden=32-p_adj_plus=0.001-p_adj_minus=0.8-p_att_plus=0.0-p_att_minus=0.0-multi_class_cert-A.pgf}
    \end{subfigure}
    \vspace{-0.2cm}
    \caption{Randomized smoothing robustness guarantees for APPNP under adjacency perturbations on Cora-ML. The moothing parameters are set to $p_\mX^+=0,p_\mX^-=0,p_\mA^+=0.001,p_\mA^-=0.8$.}
    \label{fig:exp_appendix_adjacency_smoothing_cora_appnp}
	\end{minipage}
\end{figure}
\vspace{-0.60cm}

\begin{figure}[H]
	\centering
	\begin{minipage}{0.49\textwidth}
    \begin{subfigure}[t]{0.49\linewidth}
		\centering
       \input{figures/experiments/graphs/sparse_smoothing/nodes_attributes/node_classification-Citeseer-APPNP-hidden=32-p_adj_plus=0.0-p_adj_minus=0.0-p_att_plus=0.001-p_att_minus=0.8-multi_class_cert-B.pgf}
    \end{subfigure}
    \begin{subfigure}[t]{0.49\linewidth}
		\centering
       \input{figures/experiments/graphs/sparse_smoothing/nodes_attributes/node_classification-Citeseer-APPNP-hidden=32-p_adj_plus=0.0-p_adj_minus=0.0-p_att_plus=0.001-p_att_minus=0.8-multi_class_cert-A.pgf}
    \end{subfigure}
    \vspace{-0.2cm}
    \caption{Randomized smoothing robustness guarantees for APPNP under attribute perturbations on Citeseer. The Smoothing parameters are set to $p_\mX^+=0.001,p_\mX^-=0.8,p_\mA^+=0,p_\mA^-=0$.}
    \label{fig:exp_appendix_attributes_smoothing_citeseer_appnp}
	\end{minipage}
	\hfill
	\begin{minipage}{0.49\textwidth}
    \begin{subfigure}[t]{0.49\linewidth}
		\centering
       \input{figures/experiments/graphs/sparse_smoothing/nodes_structure/node_classification-Citeseer-APPNP-hidden=32-p_adj_plus=0.001-p_adj_minus=0.8-p_att_plus=0.0-p_att_minus=0.0-multi_class_cert-B.pgf}
    \end{subfigure}
    \begin{subfigure}[t]{0.49\linewidth}
		\centering
       \input{figures/experiments/graphs/sparse_smoothing/nodes_structure/node_classification-Citeseer-APPNP-hidden=32-p_adj_plus=0.001-p_adj_minus=0.8-p_att_plus=0.0-p_att_minus=0.0-multi_class_cert-A.pgf}
    \end{subfigure}
    \vspace{-0.2cm}
    \caption{Randomized smoothing robustness guarantees for APPNP under adjacency perturbations on Citeseer. The moothing parameters are set to $p_\mX^+=0,p_\mX^-=0,p_\mA^+=0.001,p_\mA^-=0.8$.}
    \label{fig:exp_appendix_adjacency_smoothing_citeseer_appnp}
	\end{minipage}
\end{figure}

\clearpage

\subsection{Adversarial attacks}\label{appendix:extra_experiments_attacks}
As discussed in~\cref{section:certification}, any traditional adversarial attack is a valid adversarial attack under our proposed notion of adversarial robustness for equivariant tasks. This holds no matter if the equivariances of task $y$ and model $f$ match or not.
As an example we conduct adversarial attacks on a PointNet classifier and a graph convolutional network.
We use the same datasets, models and hyperparameters as for the experiments shown in~\cref{fig:exp_main_pointcloud} (without randomized smoothing)  and~\cref{fig:exp_appendix_attributes_zuegner_cora}.
To attack the PointNet classifier, we use a single gradient step w.r.t.\ cross-entropy loss, which we scale to have an $\ell_2$ norm of $\epsilon$.
To attack the graph convolutional network, we use the method from Section 4.4 of~\cite{Zuegner2019cert}, which is directly derived from their certification procedure.
As can be seen, future work is needed to further improve the robustness of equivariant models to our proposed perturbation model.

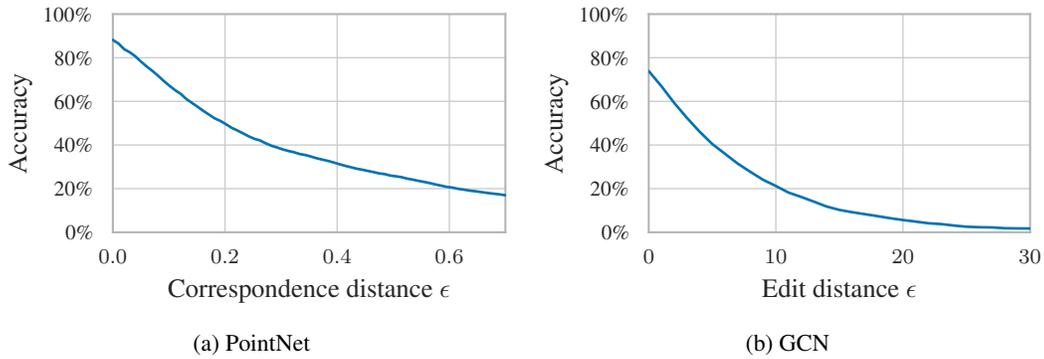
\begin{figure}[H]
    \centering
    \begin{subfigure}{0.49\textwidth}
        \input{figures/experiments/attacks/pointnet_attack.pgf}
        \caption{PointNet}
    \end{subfigure}
    \hfill
    \begin{subfigure}{0.49\textwidth}
        \input{figures/experiments/attacks/gcn_attack.pgf}
        \caption{GCN}
    \end{subfigure}
    \caption{Adversarial attacks on (a) PointNet using a single gradient step with $\ell_2$ norm $\epsilon$ and (b) GCN (with perturbation costs $c_X^+ = c_X^- = 1$) using the method from Section 4.4 of~\cite{Zuegner2019cert}.}
    \vspace{-0.5cm}
\end{figure}

%% file: figures/experiments/pointclouds/0.05.pgf
\begingroup%
\makeatletter%
\begin{pgfpicture}%
\pgfpathrectangle{\pgfpointorigin}{\pgfqpoint{2.750000in}{1.699593in}}%
\pgfusepath{use as bounding box, clip}%
\begin{pgfscope}%
\pgfsetbuttcap%
\pgfsetmiterjoin%
\definecolor{currentfill}{rgb}{1.000000,1.000000,1.000000}%
\pgfsetfillcolor{currentfill}%
\pgfsetlinewidth{0.000000pt}%
\definecolor{currentstroke}{rgb}{1.000000,1.000000,1.000000}%
\pgfsetstrokecolor{currentstroke}%
\pgfsetdash{}{0pt}%
\pgfpathmoveto{\pgfqpoint{0.000000in}{0.000000in}}%
\pgfpathlineto{\pgfqpoint{2.750000in}{0.000000in}}%
\pgfpathlineto{\pgfqpoint{2.750000in}{1.699593in}}%
\pgfpathlineto{\pgfqpoint{0.000000in}{1.699593in}}%
\pgfpathlineto{\pgfqpoint{0.000000in}{0.000000in}}%
\pgfpathclose%
\pgfusepath{fill}%
\end{pgfscope}%
\begin{pgfscope}%
\pgfsetbuttcap%
\pgfsetmiterjoin%
\definecolor{currentfill}{rgb}{1.000000,1.000000,1.000000}%
\pgfsetfillcolor{currentfill}%
\pgfsetlinewidth{0.000000pt}%
\definecolor{currentstroke}{rgb}{0.000000,0.000000,0.000000}%
\pgfsetstrokecolor{currentstroke}%
\pgfsetstrokeopacity{0.000000}%
\pgfsetdash{}{0pt}%
\pgfpathmoveto{\pgfqpoint{0.520798in}{0.442177in}}%
\pgfpathlineto{\pgfqpoint{2.569063in}{0.442177in}}%
\pgfpathlineto{\pgfqpoint{2.569063in}{1.585085in}}%
\pgfpathlineto{\pgfqpoint{0.520798in}{1.585085in}}%
\pgfpathlineto{\pgfqpoint{0.520798in}{0.442177in}}%
\pgfpathclose%
\pgfusepath{fill}%
\end{pgfscope}%
\begin{pgfscope}%
\pgfpathrectangle{\pgfqpoint{0.520798in}{0.442177in}}{\pgfqpoint{2.048265in}{1.142908in}}%
\pgfusepath{clip}%
\pgfsetroundcap%
\pgfsetroundjoin%
\pgfsetlinewidth{0.501875pt}%
\definecolor{currentstroke}{rgb}{0.800000,0.800000,0.800000}%
\pgfsetstrokecolor{currentstroke}%
\pgfsetdash{}{0pt}%
\pgfpathmoveto{\pgfqpoint{0.520798in}{0.442177in}}%
\pgfpathlineto{\pgfqpoint{0.520798in}{1.585085in}}%
\pgfusepath{stroke}%
\end{pgfscope}%
\begin{pgfscope}%
\definecolor{textcolor}{rgb}{0.150000,0.150000,0.150000}%
\pgfsetstrokecolor{textcolor}%
\pgfsetfillcolor{textcolor}%
\pgftext[x=0.520798in,y=0.351899in,,top]{\color{textcolor}\rmfamily\fontsize{8.000000}{9.600000}\selectfont \(\displaystyle {0.00}\)}%
\end{pgfscope}%
\begin{pgfscope}%
\pgfpathrectangle{\pgfqpoint{0.520798in}{0.442177in}}{\pgfqpoint{2.048265in}{1.142908in}}%
\pgfusepath{clip}%
\pgfsetroundcap%
\pgfsetroundjoin%
\pgfsetlinewidth{0.501875pt}%
\definecolor{currentstroke}{rgb}{0.800000,0.800000,0.800000}%
\pgfsetstrokecolor{currentstroke}%
\pgfsetdash{}{0pt}%
\pgfpathmoveto{\pgfqpoint{1.032864in}{0.442177in}}%
\pgfpathlineto{\pgfqpoint{1.032864in}{1.585085in}}%
\pgfusepath{stroke}%
\end{pgfscope}%
\begin{pgfscope}%
\definecolor{textcolor}{rgb}{0.150000,0.150000,0.150000}%
\pgfsetstrokecolor{textcolor}%
\pgfsetfillcolor{textcolor}%
\pgftext[x=1.032864in,y=0.351899in,,top]{\color{textcolor}\rmfamily\fontsize{8.000000}{9.600000}\selectfont \(\displaystyle {0.05}\)}%
\end{pgfscope}%
\begin{pgfscope}%
\pgfpathrectangle{\pgfqpoint{0.520798in}{0.442177in}}{\pgfqpoint{2.048265in}{1.142908in}}%
\pgfusepath{clip}%
\pgfsetroundcap%
\pgfsetroundjoin%
\pgfsetlinewidth{0.501875pt}%
\definecolor{currentstroke}{rgb}{0.800000,0.800000,0.800000}%
\pgfsetstrokecolor{currentstroke}%
\pgfsetdash{}{0pt}%
\pgfpathmoveto{\pgfqpoint{1.544931in}{0.442177in}}%
\pgfpathlineto{\pgfqpoint{1.544931in}{1.585085in}}%
\pgfusepath{stroke}%
\end{pgfscope}%
\begin{pgfscope}%
\definecolor{textcolor}{rgb}{0.150000,0.150000,0.150000}%
\pgfsetstrokecolor{textcolor}%
\pgfsetfillcolor{textcolor}%
\pgftext[x=1.544931in,y=0.351899in,,top]{\color{textcolor}\rmfamily\fontsize{8.000000}{9.600000}\selectfont \(\displaystyle {0.10}\)}%
\end{pgfscope}%
\begin{pgfscope}%
\pgfpathrectangle{\pgfqpoint{0.520798in}{0.442177in}}{\pgfqpoint{2.048265in}{1.142908in}}%
\pgfusepath{clip}%
\pgfsetroundcap%
\pgfsetroundjoin%
\pgfsetlinewidth{0.501875pt}%
\definecolor{currentstroke}{rgb}{0.800000,0.800000,0.800000}%
\pgfsetstrokecolor{currentstroke}%
\pgfsetdash{}{0pt}%
\pgfpathmoveto{\pgfqpoint{2.056997in}{0.442177in}}%
\pgfpathlineto{\pgfqpoint{2.056997in}{1.585085in}}%
\pgfusepath{stroke}%
\end{pgfscope}%
\begin{pgfscope}%
\definecolor{textcolor}{rgb}{0.150000,0.150000,0.150000}%
\pgfsetstrokecolor{textcolor}%
\pgfsetfillcolor{textcolor}%
\pgftext[x=2.056997in,y=0.351899in,,top]{\color{textcolor}\rmfamily\fontsize{8.000000}{9.600000}\selectfont \(\displaystyle {0.15}\)}%
\end{pgfscope}%
\begin{pgfscope}%
\pgfpathrectangle{\pgfqpoint{0.520798in}{0.442177in}}{\pgfqpoint{2.048265in}{1.142908in}}%
\pgfusepath{clip}%
\pgfsetroundcap%
\pgfsetroundjoin%
\pgfsetlinewidth{0.501875pt}%
\definecolor{currentstroke}{rgb}{0.800000,0.800000,0.800000}%
\pgfsetstrokecolor{currentstroke}%
\pgfsetdash{}{0pt}%
\pgfpathmoveto{\pgfqpoint{2.569063in}{0.442177in}}%
\pgfpathlineto{\pgfqpoint{2.569063in}{1.585085in}}%
\pgfusepath{stroke}%
\end{pgfscope}%
\begin{pgfscope}%
\definecolor{textcolor}{rgb}{0.150000,0.150000,0.150000}%
\pgfsetstrokecolor{textcolor}%
\pgfsetfillcolor{textcolor}%
\pgftext[x=2.569063in,y=0.351899in,,top]{\color{textcolor}\rmfamily\fontsize{8.000000}{9.600000}\selectfont \(\displaystyle {0.20}\)}%
\end{pgfscope}%
\begin{pgfscope}%
\definecolor{textcolor}{rgb}{0.150000,0.150000,0.150000}%
\pgfsetstrokecolor{textcolor}%
\pgfsetfillcolor{textcolor}%
\pgftext[x=1.544931in,y=0.198219in,,top]{\color{textcolor}\rmfamily\fontsize{10.000000}{12.000000}\selectfont Correspondence distance \(\displaystyle \epsilon\)}%
\end{pgfscope}%
\begin{pgfscope}%
\pgfpathrectangle{\pgfqpoint{0.520798in}{0.442177in}}{\pgfqpoint{2.048265in}{1.142908in}}%
\pgfusepath{clip}%
\pgfsetroundcap%
\pgfsetroundjoin%
\pgfsetlinewidth{0.501875pt}%
\definecolor{currentstroke}{rgb}{0.800000,0.800000,0.800000}%
\pgfsetstrokecolor{currentstroke}%
\pgfsetdash{}{0pt}%
\pgfpathmoveto{\pgfqpoint{0.520798in}{0.442177in}}%
\pgfpathlineto{\pgfqpoint{2.569063in}{0.442177in}}%
\pgfusepath{stroke}%
\end{pgfscope}%
\begin{pgfscope}%
\definecolor{textcolor}{rgb}{0.150000,0.150000,0.150000}%
\pgfsetstrokecolor{textcolor}%
\pgfsetfillcolor{textcolor}%
\pgftext[x=0.273151in, y=0.403915in, left, base]{\color{textcolor}\rmfamily\fontsize{8.000000}{9.600000}\selectfont 0\%}%
\end{pgfscope}%
\begin{pgfscope}%
\pgfpathrectangle{\pgfqpoint{0.520798in}{0.442177in}}{\pgfqpoint{2.048265in}{1.142908in}}%
\pgfusepath{clip}%
\pgfsetroundcap%
\pgfsetroundjoin%
\pgfsetlinewidth{0.501875pt}%
\definecolor{currentstroke}{rgb}{0.800000,0.800000,0.800000}%
\pgfsetstrokecolor{currentstroke}%
\pgfsetdash{}{0pt}%
\pgfpathmoveto{\pgfqpoint{0.520798in}{0.670758in}}%
\pgfpathlineto{\pgfqpoint{2.569063in}{0.670758in}}%
\pgfusepath{stroke}%
\end{pgfscope}%
\begin{pgfscope}%
\definecolor{textcolor}{rgb}{0.150000,0.150000,0.150000}%
\pgfsetstrokecolor{textcolor}%
\pgfsetfillcolor{textcolor}%
\pgftext[x=0.214138in, y=0.632496in, left, base]{\color{textcolor}\rmfamily\fontsize{8.000000}{9.600000}\selectfont 20\%}%
\end{pgfscope}%
\begin{pgfscope}%
\pgfpathrectangle{\pgfqpoint{0.520798in}{0.442177in}}{\pgfqpoint{2.048265in}{1.142908in}}%
\pgfusepath{clip}%
\pgfsetroundcap%
\pgfsetroundjoin%
\pgfsetlinewidth{0.501875pt}%
\definecolor{currentstroke}{rgb}{0.800000,0.800000,0.800000}%
\pgfsetstrokecolor{currentstroke}%
\pgfsetdash{}{0pt}%
\pgfpathmoveto{\pgfqpoint{0.520798in}{0.899340in}}%
\pgfpathlineto{\pgfqpoint{2.569063in}{0.899340in}}%
\pgfusepath{stroke}%
\end{pgfscope}%
\begin{pgfscope}%
\definecolor{textcolor}{rgb}{0.150000,0.150000,0.150000}%
\pgfsetstrokecolor{textcolor}%
\pgfsetfillcolor{textcolor}%
\pgftext[x=0.214138in, y=0.861078in, left, base]{\color{textcolor}\rmfamily\fontsize{8.000000}{9.600000}\selectfont 40\%}%
\end{pgfscope}%
\begin{pgfscope}%
\pgfpathrectangle{\pgfqpoint{0.520798in}{0.442177in}}{\pgfqpoint{2.048265in}{1.142908in}}%
\pgfusepath{clip}%
\pgfsetroundcap%
\pgfsetroundjoin%
\pgfsetlinewidth{0.501875pt}%
\definecolor{currentstroke}{rgb}{0.800000,0.800000,0.800000}%
\pgfsetstrokecolor{currentstroke}%
\pgfsetdash{}{0pt}%
\pgfpathmoveto{\pgfqpoint{0.520798in}{1.127922in}}%
\pgfpathlineto{\pgfqpoint{2.569063in}{1.127922in}}%
\pgfusepath{stroke}%
\end{pgfscope}%
\begin{pgfscope}%
\definecolor{textcolor}{rgb}{0.150000,0.150000,0.150000}%
\pgfsetstrokecolor{textcolor}%
\pgfsetfillcolor{textcolor}%
\pgftext[x=0.214138in, y=1.089660in, left, base]{\color{textcolor}\rmfamily\fontsize{8.000000}{9.600000}\selectfont 60\%}%
\end{pgfscope}%
\begin{pgfscope}%
\pgfpathrectangle{\pgfqpoint{0.520798in}{0.442177in}}{\pgfqpoint{2.048265in}{1.142908in}}%
\pgfusepath{clip}%
\pgfsetroundcap%
\pgfsetroundjoin%
\pgfsetlinewidth{0.501875pt}%
\definecolor{currentstroke}{rgb}{0.800000,0.800000,0.800000}%
\pgfsetstrokecolor{currentstroke}%
\pgfsetdash{}{0pt}%
\pgfpathmoveto{\pgfqpoint{0.520798in}{1.356504in}}%
\pgfpathlineto{\pgfqpoint{2.569063in}{1.356504in}}%
\pgfusepath{stroke}%
\end{pgfscope}%
\begin{pgfscope}%
\definecolor{textcolor}{rgb}{0.150000,0.150000,0.150000}%
\pgfsetstrokecolor{textcolor}%
\pgfsetfillcolor{textcolor}%
\pgftext[x=0.214138in, y=1.318241in, left, base]{\color{textcolor}\rmfamily\fontsize{8.000000}{9.600000}\selectfont 80\%}%
\end{pgfscope}%
\begin{pgfscope}%
\pgfpathrectangle{\pgfqpoint{0.520798in}{0.442177in}}{\pgfqpoint{2.048265in}{1.142908in}}%
\pgfusepath{clip}%
\pgfsetroundcap%
\pgfsetroundjoin%
\pgfsetlinewidth{0.501875pt}%
\definecolor{currentstroke}{rgb}{0.800000,0.800000,0.800000}%
\pgfsetstrokecolor{currentstroke}%
\pgfsetdash{}{0pt}%
\pgfpathmoveto{\pgfqpoint{0.520798in}{1.585085in}}%
\pgfpathlineto{\pgfqpoint{2.569063in}{1.585085in}}%
\pgfusepath{stroke}%
\end{pgfscope}%
\begin{pgfscope}%
\definecolor{textcolor}{rgb}{0.150000,0.150000,0.150000}%
\pgfsetstrokecolor{textcolor}%
\pgfsetfillcolor{textcolor}%
\pgftext[x=0.155124in, y=1.546823in, left, base]{\color{textcolor}\rmfamily\fontsize{8.000000}{9.600000}\selectfont 100\%}%
\end{pgfscope}%
\begin{pgfscope}%
\definecolor{textcolor}{rgb}{0.150000,0.150000,0.150000}%
\pgfsetstrokecolor{textcolor}%
\pgfsetfillcolor{textcolor}%
\pgftext[x=0.099569in,y=1.013631in,,bottom,rotate=90.000000]{\color{textcolor}\rmfamily\fontsize{10.000000}{12.000000}\selectfont Cert. Acc.}%
\end{pgfscope}%
\begin{pgfscope}%
\pgfsetrectcap%
\pgfsetmiterjoin%
\pgfsetlinewidth{0.752812pt}%
\definecolor{currentstroke}{rgb}{0.700000,0.700000,0.700000}%
\pgfsetstrokecolor{currentstroke}%
\pgfsetdash{}{0pt}%
\pgfpathmoveto{\pgfqpoint{0.520798in}{0.442177in}}%
\pgfpathlineto{\pgfqpoint{0.520798in}{1.585085in}}%
\pgfusepath{stroke}%
\end{pgfscope}%
\begin{pgfscope}%
\pgfsetrectcap%
\pgfsetmiterjoin%
\pgfsetlinewidth{0.752812pt}%
\definecolor{currentstroke}{rgb}{0.700000,0.700000,0.700000}%
\pgfsetstrokecolor{currentstroke}%
\pgfsetdash{}{0pt}%
\pgfpathmoveto{\pgfqpoint{2.569063in}{0.442177in}}%
\pgfpathlineto{\pgfqpoint{2.569063in}{1.585085in}}%
\pgfusepath{stroke}%
\end{pgfscope}%
\begin{pgfscope}%
\pgfsetrectcap%
\pgfsetmiterjoin%
\pgfsetlinewidth{0.752812pt}%
\definecolor{currentstroke}{rgb}{0.700000,0.700000,0.700000}%
\pgfsetstrokecolor{currentstroke}%
\pgfsetdash{}{0pt}%
\pgfpathmoveto{\pgfqpoint{0.520798in}{0.442177in}}%
\pgfpathlineto{\pgfqpoint{2.569063in}{0.442177in}}%
\pgfusepath{stroke}%
\end{pgfscope}%
\begin{pgfscope}%
\pgfsetrectcap%
\pgfsetmiterjoin%
\pgfsetlinewidth{0.752812pt}%
\definecolor{currentstroke}{rgb}{0.700000,0.700000,0.700000}%
\pgfsetstrokecolor{currentstroke}%
\pgfsetdash{}{0pt}%
\pgfpathmoveto{\pgfqpoint{0.520798in}{1.585085in}}%
\pgfpathlineto{\pgfqpoint{2.569063in}{1.585085in}}%
\pgfusepath{stroke}%
\end{pgfscope}%
\begin{pgfscope}%
\pgfpathrectangle{\pgfqpoint{0.520798in}{0.442177in}}{\pgfqpoint{2.048265in}{1.142908in}}%
\pgfusepath{clip}%
\pgfsetroundcap%
\pgfsetroundjoin%
\pgfsetlinewidth{1.003750pt}%
\definecolor{currentstroke}{rgb}{0.003922,0.450980,0.698039}%
\pgfsetstrokecolor{currentstroke}%
\pgfsetdash{}{0pt}%
\pgfpathmoveto{\pgfqpoint{0.520798in}{1.448288in}}%
\pgfpathlineto{\pgfqpoint{0.624246in}{1.437452in}}%
\pgfpathlineto{\pgfqpoint{0.727693in}{1.426986in}}%
\pgfpathlineto{\pgfqpoint{0.831141in}{1.414760in}}%
\pgfpathlineto{\pgfqpoint{0.934589in}{1.400682in}}%
\pgfpathlineto{\pgfqpoint{1.038037in}{1.386975in}}%
\pgfpathlineto{\pgfqpoint{1.141484in}{1.371230in}}%
\pgfpathlineto{\pgfqpoint{1.244932in}{1.357800in}}%
\pgfpathlineto{\pgfqpoint{1.348380in}{1.339091in}}%
\pgfpathlineto{\pgfqpoint{1.451828in}{1.319086in}}%
\pgfpathlineto{\pgfqpoint{1.555275in}{1.296672in}}%
\pgfpathlineto{\pgfqpoint{1.658723in}{1.271480in}}%
\pgfpathlineto{\pgfqpoint{1.762171in}{1.245732in}}%
\pgfpathlineto{\pgfqpoint{1.865619in}{1.215446in}}%
\pgfpathlineto{\pgfqpoint{1.969066in}{1.178028in}}%
\pgfpathlineto{\pgfqpoint{2.072514in}{1.125236in}}%
\pgfpathlineto{\pgfqpoint{2.175962in}{0.442177in}}%
\pgfpathlineto{\pgfqpoint{2.279410in}{0.442177in}}%
\pgfpathlineto{\pgfqpoint{2.382857in}{0.442177in}}%
\pgfpathlineto{\pgfqpoint{2.486305in}{0.442177in}}%
\pgfpathlineto{\pgfqpoint{2.570730in}{0.442177in}}%
\pgfusepath{stroke}%
\end{pgfscope}%
\begin{pgfscope}%
\pgfpathrectangle{\pgfqpoint{0.520798in}{0.442177in}}{\pgfqpoint{2.048265in}{1.142908in}}%
\pgfusepath{clip}%
\pgfsetbuttcap%
\pgfsetroundjoin%
\definecolor{currentfill}{rgb}{0.003922,0.450980,0.698039}%
\pgfsetfillcolor{currentfill}%
\pgfsetfillopacity{0.500000}%
\pgfsetlinewidth{0.000000pt}%
\definecolor{currentstroke}{rgb}{0.003922,0.450980,0.698039}%
\pgfsetstrokecolor{currentstroke}%
\pgfsetstrokeopacity{0.500000}%
\pgfsetdash{}{0pt}%
\pgfpathmoveto{\pgfqpoint{0.520798in}{1.450568in}}%
\pgfpathlineto{\pgfqpoint{0.520798in}{1.446008in}}%
\pgfpathlineto{\pgfqpoint{0.624246in}{1.434943in}}%
\pgfpathlineto{\pgfqpoint{0.727693in}{1.422953in}}%
\pgfpathlineto{\pgfqpoint{0.831141in}{1.411241in}}%
\pgfpathlineto{\pgfqpoint{0.934589in}{1.396588in}}%
\pgfpathlineto{\pgfqpoint{1.038037in}{1.382453in}}%
\pgfpathlineto{\pgfqpoint{1.141484in}{1.368139in}}%
\pgfpathlineto{\pgfqpoint{1.244932in}{1.354534in}}%
\pgfpathlineto{\pgfqpoint{1.348380in}{1.335511in}}%
\pgfpathlineto{\pgfqpoint{1.451828in}{1.313999in}}%
\pgfpathlineto{\pgfqpoint{1.555275in}{1.291955in}}%
\pgfpathlineto{\pgfqpoint{1.658723in}{1.264191in}}%
\pgfpathlineto{\pgfqpoint{1.762171in}{1.238257in}}%
\pgfpathlineto{\pgfqpoint{1.865619in}{1.211331in}}%
\pgfpathlineto{\pgfqpoint{1.969066in}{1.172237in}}%
\pgfpathlineto{\pgfqpoint{2.072514in}{1.118680in}}%
\pgfpathlineto{\pgfqpoint{2.175962in}{0.442177in}}%
\pgfpathlineto{\pgfqpoint{2.279410in}{0.442177in}}%
\pgfpathlineto{\pgfqpoint{2.382857in}{0.442177in}}%
\pgfpathlineto{\pgfqpoint{2.486305in}{0.442177in}}%
\pgfpathlineto{\pgfqpoint{2.589753in}{0.442177in}}%
\pgfpathlineto{\pgfqpoint{2.693201in}{0.442177in}}%
\pgfpathlineto{\pgfqpoint{2.796648in}{0.442177in}}%
\pgfpathlineto{\pgfqpoint{2.900096in}{0.442177in}}%
\pgfpathlineto{\pgfqpoint{3.003544in}{0.442177in}}%
\pgfpathlineto{\pgfqpoint{3.106992in}{0.442177in}}%
\pgfpathlineto{\pgfqpoint{3.210439in}{0.442177in}}%
\pgfpathlineto{\pgfqpoint{3.313887in}{0.442177in}}%
\pgfpathlineto{\pgfqpoint{3.417335in}{0.442177in}}%
\pgfpathlineto{\pgfqpoint{3.520783in}{0.442177in}}%
\pgfpathlineto{\pgfqpoint{3.624230in}{0.442177in}}%
\pgfpathlineto{\pgfqpoint{3.727678in}{0.442177in}}%
\pgfpathlineto{\pgfqpoint{3.831126in}{0.442177in}}%
\pgfpathlineto{\pgfqpoint{3.934574in}{0.442177in}}%
\pgfpathlineto{\pgfqpoint{4.038021in}{0.442177in}}%
\pgfpathlineto{\pgfqpoint{4.141469in}{0.442177in}}%
\pgfpathlineto{\pgfqpoint{4.244917in}{0.442177in}}%
\pgfpathlineto{\pgfqpoint{4.348365in}{0.442177in}}%
\pgfpathlineto{\pgfqpoint{4.451812in}{0.442177in}}%
\pgfpathlineto{\pgfqpoint{4.555260in}{0.442177in}}%
\pgfpathlineto{\pgfqpoint{4.658708in}{0.442177in}}%
\pgfpathlineto{\pgfqpoint{4.762156in}{0.442177in}}%
\pgfpathlineto{\pgfqpoint{4.865603in}{0.442177in}}%
\pgfpathlineto{\pgfqpoint{4.969051in}{0.442177in}}%
\pgfpathlineto{\pgfqpoint{5.072499in}{0.442177in}}%
\pgfpathlineto{\pgfqpoint{5.175947in}{0.442177in}}%
\pgfpathlineto{\pgfqpoint{5.279394in}{0.442177in}}%
\pgfpathlineto{\pgfqpoint{5.382842in}{0.442177in}}%
\pgfpathlineto{\pgfqpoint{5.486290in}{0.442177in}}%
\pgfpathlineto{\pgfqpoint{5.589738in}{0.442177in}}%
\pgfpathlineto{\pgfqpoint{5.693185in}{0.442177in}}%
\pgfpathlineto{\pgfqpoint{5.796633in}{0.442177in}}%
\pgfpathlineto{\pgfqpoint{5.900081in}{0.442177in}}%
\pgfpathlineto{\pgfqpoint{6.003529in}{0.442177in}}%
\pgfpathlineto{\pgfqpoint{6.106976in}{0.442177in}}%
\pgfpathlineto{\pgfqpoint{6.210424in}{0.442177in}}%
\pgfpathlineto{\pgfqpoint{6.313872in}{0.442177in}}%
\pgfpathlineto{\pgfqpoint{6.417320in}{0.442177in}}%
\pgfpathlineto{\pgfqpoint{6.520767in}{0.442177in}}%
\pgfpathlineto{\pgfqpoint{6.624215in}{0.442177in}}%
\pgfpathlineto{\pgfqpoint{6.727663in}{0.442177in}}%
\pgfpathlineto{\pgfqpoint{6.831111in}{0.442177in}}%
\pgfpathlineto{\pgfqpoint{6.934558in}{0.442177in}}%
\pgfpathlineto{\pgfqpoint{7.038006in}{0.442177in}}%
\pgfpathlineto{\pgfqpoint{7.141454in}{0.442177in}}%
\pgfpathlineto{\pgfqpoint{7.244902in}{0.442177in}}%
\pgfpathlineto{\pgfqpoint{7.348349in}{0.442177in}}%
\pgfpathlineto{\pgfqpoint{7.451797in}{0.442177in}}%
\pgfpathlineto{\pgfqpoint{7.555245in}{0.442177in}}%
\pgfpathlineto{\pgfqpoint{7.658693in}{0.442177in}}%
\pgfpathlineto{\pgfqpoint{7.762140in}{0.442177in}}%
\pgfpathlineto{\pgfqpoint{7.865588in}{0.442177in}}%
\pgfpathlineto{\pgfqpoint{7.969036in}{0.442177in}}%
\pgfpathlineto{\pgfqpoint{8.072484in}{0.442177in}}%
\pgfpathlineto{\pgfqpoint{8.175931in}{0.442177in}}%
\pgfpathlineto{\pgfqpoint{8.279379in}{0.442177in}}%
\pgfpathlineto{\pgfqpoint{8.382827in}{0.442177in}}%
\pgfpathlineto{\pgfqpoint{8.486275in}{0.442177in}}%
\pgfpathlineto{\pgfqpoint{8.589722in}{0.442177in}}%
\pgfpathlineto{\pgfqpoint{8.693170in}{0.442177in}}%
\pgfpathlineto{\pgfqpoint{8.796618in}{0.442177in}}%
\pgfpathlineto{\pgfqpoint{8.900066in}{0.442177in}}%
\pgfpathlineto{\pgfqpoint{9.003513in}{0.442177in}}%
\pgfpathlineto{\pgfqpoint{9.106961in}{0.442177in}}%
\pgfpathlineto{\pgfqpoint{9.210409in}{0.442177in}}%
\pgfpathlineto{\pgfqpoint{9.313857in}{0.442177in}}%
\pgfpathlineto{\pgfqpoint{9.417304in}{0.442177in}}%
\pgfpathlineto{\pgfqpoint{9.520752in}{0.442177in}}%
\pgfpathlineto{\pgfqpoint{9.624200in}{0.442177in}}%
\pgfpathlineto{\pgfqpoint{9.727648in}{0.442177in}}%
\pgfpathlineto{\pgfqpoint{9.831095in}{0.442177in}}%
\pgfpathlineto{\pgfqpoint{9.934543in}{0.442177in}}%
\pgfpathlineto{\pgfqpoint{10.037991in}{0.442177in}}%
\pgfpathlineto{\pgfqpoint{10.141439in}{0.442177in}}%
\pgfpathlineto{\pgfqpoint{10.244886in}{0.442177in}}%
\pgfpathlineto{\pgfqpoint{10.348334in}{0.442177in}}%
\pgfpathlineto{\pgfqpoint{10.451782in}{0.442177in}}%
\pgfpathlineto{\pgfqpoint{10.555230in}{0.442177in}}%
\pgfpathlineto{\pgfqpoint{10.658677in}{0.442177in}}%
\pgfpathlineto{\pgfqpoint{10.762125in}{0.442177in}}%
\pgfpathlineto{\pgfqpoint{10.762125in}{0.442177in}}%
\pgfpathlineto{\pgfqpoint{10.762125in}{0.442177in}}%
\pgfpathlineto{\pgfqpoint{10.658677in}{0.442177in}}%
\pgfpathlineto{\pgfqpoint{10.555230in}{0.442177in}}%
\pgfpathlineto{\pgfqpoint{10.451782in}{0.442177in}}%
\pgfpathlineto{\pgfqpoint{10.348334in}{0.442177in}}%
\pgfpathlineto{\pgfqpoint{10.244886in}{0.442177in}}%
\pgfpathlineto{\pgfqpoint{10.141439in}{0.442177in}}%
\pgfpathlineto{\pgfqpoint{10.037991in}{0.442177in}}%
\pgfpathlineto{\pgfqpoint{9.934543in}{0.442177in}}%
\pgfpathlineto{\pgfqpoint{9.831095in}{0.442177in}}%
\pgfpathlineto{\pgfqpoint{9.727648in}{0.442177in}}%
\pgfpathlineto{\pgfqpoint{9.624200in}{0.442177in}}%
\pgfpathlineto{\pgfqpoint{9.520752in}{0.442177in}}%
\pgfpathlineto{\pgfqpoint{9.417304in}{0.442177in}}%
\pgfpathlineto{\pgfqpoint{9.313857in}{0.442177in}}%
\pgfpathlineto{\pgfqpoint{9.210409in}{0.442177in}}%
\pgfpathlineto{\pgfqpoint{9.106961in}{0.442177in}}%
\pgfpathlineto{\pgfqpoint{9.003513in}{0.442177in}}%
\pgfpathlineto{\pgfqpoint{8.900066in}{0.442177in}}%
\pgfpathlineto{\pgfqpoint{8.796618in}{0.442177in}}%
\pgfpathlineto{\pgfqpoint{8.693170in}{0.442177in}}%
\pgfpathlineto{\pgfqpoint{8.589722in}{0.442177in}}%
\pgfpathlineto{\pgfqpoint{8.486275in}{0.442177in}}%
\pgfpathlineto{\pgfqpoint{8.382827in}{0.442177in}}%
\pgfpathlineto{\pgfqpoint{8.279379in}{0.442177in}}%
\pgfpathlineto{\pgfqpoint{8.175931in}{0.442177in}}%
\pgfpathlineto{\pgfqpoint{8.072484in}{0.442177in}}%
\pgfpathlineto{\pgfqpoint{7.969036in}{0.442177in}}%
\pgfpathlineto{\pgfqpoint{7.865588in}{0.442177in}}%
\pgfpathlineto{\pgfqpoint{7.762140in}{0.442177in}}%
\pgfpathlineto{\pgfqpoint{7.658693in}{0.442177in}}%
\pgfpathlineto{\pgfqpoint{7.555245in}{0.442177in}}%
\pgfpathlineto{\pgfqpoint{7.451797in}{0.442177in}}%
\pgfpathlineto{\pgfqpoint{7.348349in}{0.442177in}}%
\pgfpathlineto{\pgfqpoint{7.244902in}{0.442177in}}%
\pgfpathlineto{\pgfqpoint{7.141454in}{0.442177in}}%
\pgfpathlineto{\pgfqpoint{7.038006in}{0.442177in}}%
\pgfpathlineto{\pgfqpoint{6.934558in}{0.442177in}}%
\pgfpathlineto{\pgfqpoint{6.831111in}{0.442177in}}%
\pgfpathlineto{\pgfqpoint{6.727663in}{0.442177in}}%
\pgfpathlineto{\pgfqpoint{6.624215in}{0.442177in}}%
\pgfpathlineto{\pgfqpoint{6.520767in}{0.442177in}}%
\pgfpathlineto{\pgfqpoint{6.417320in}{0.442177in}}%
\pgfpathlineto{\pgfqpoint{6.313872in}{0.442177in}}%
\pgfpathlineto{\pgfqpoint{6.210424in}{0.442177in}}%
\pgfpathlineto{\pgfqpoint{6.106976in}{0.442177in}}%
\pgfpathlineto{\pgfqpoint{6.003529in}{0.442177in}}%
\pgfpathlineto{\pgfqpoint{5.900081in}{0.442177in}}%
\pgfpathlineto{\pgfqpoint{5.796633in}{0.442177in}}%
\pgfpathlineto{\pgfqpoint{5.693185in}{0.442177in}}%
\pgfpathlineto{\pgfqpoint{5.589738in}{0.442177in}}%
\pgfpathlineto{\pgfqpoint{5.486290in}{0.442177in}}%
\pgfpathlineto{\pgfqpoint{5.382842in}{0.442177in}}%
\pgfpathlineto{\pgfqpoint{5.279394in}{0.442177in}}%
\pgfpathlineto{\pgfqpoint{5.175947in}{0.442177in}}%
\pgfpathlineto{\pgfqpoint{5.072499in}{0.442177in}}%
\pgfpathlineto{\pgfqpoint{4.969051in}{0.442177in}}%
\pgfpathlineto{\pgfqpoint{4.865603in}{0.442177in}}%
\pgfpathlineto{\pgfqpoint{4.762156in}{0.442177in}}%
\pgfpathlineto{\pgfqpoint{4.658708in}{0.442177in}}%
\pgfpathlineto{\pgfqpoint{4.555260in}{0.442177in}}%
\pgfpathlineto{\pgfqpoint{4.451812in}{0.442177in}}%
\pgfpathlineto{\pgfqpoint{4.348365in}{0.442177in}}%
\pgfpathlineto{\pgfqpoint{4.244917in}{0.442177in}}%
\pgfpathlineto{\pgfqpoint{4.141469in}{0.442177in}}%
\pgfpathlineto{\pgfqpoint{4.038021in}{0.442177in}}%
\pgfpathlineto{\pgfqpoint{3.934574in}{0.442177in}}%
\pgfpathlineto{\pgfqpoint{3.831126in}{0.442177in}}%
\pgfpathlineto{\pgfqpoint{3.727678in}{0.442177in}}%
\pgfpathlineto{\pgfqpoint{3.624230in}{0.442177in}}%
\pgfpathlineto{\pgfqpoint{3.520783in}{0.442177in}}%
\pgfpathlineto{\pgfqpoint{3.417335in}{0.442177in}}%
\pgfpathlineto{\pgfqpoint{3.313887in}{0.442177in}}%
\pgfpathlineto{\pgfqpoint{3.210439in}{0.442177in}}%
\pgfpathlineto{\pgfqpoint{3.106992in}{0.442177in}}%
\pgfpathlineto{\pgfqpoint{3.003544in}{0.442177in}}%
\pgfpathlineto{\pgfqpoint{2.900096in}{0.442177in}}%
\pgfpathlineto{\pgfqpoint{2.796648in}{0.442177in}}%
\pgfpathlineto{\pgfqpoint{2.693201in}{0.442177in}}%
\pgfpathlineto{\pgfqpoint{2.589753in}{0.442177in}}%
\pgfpathlineto{\pgfqpoint{2.486305in}{0.442177in}}%
\pgfpathlineto{\pgfqpoint{2.382857in}{0.442177in}}%
\pgfpathlineto{\pgfqpoint{2.279410in}{0.442177in}}%
\pgfpathlineto{\pgfqpoint{2.175962in}{0.442177in}}%
\pgfpathlineto{\pgfqpoint{2.072514in}{1.131792in}}%
\pgfpathlineto{\pgfqpoint{1.969066in}{1.183820in}}%
\pgfpathlineto{\pgfqpoint{1.865619in}{1.219561in}}%
\pgfpathlineto{\pgfqpoint{1.762171in}{1.253207in}}%
\pgfpathlineto{\pgfqpoint{1.658723in}{1.278769in}}%
\pgfpathlineto{\pgfqpoint{1.555275in}{1.301389in}}%
\pgfpathlineto{\pgfqpoint{1.451828in}{1.324172in}}%
\pgfpathlineto{\pgfqpoint{1.348380in}{1.342671in}}%
\pgfpathlineto{\pgfqpoint{1.244932in}{1.361067in}}%
\pgfpathlineto{\pgfqpoint{1.141484in}{1.374321in}}%
\pgfpathlineto{\pgfqpoint{1.038037in}{1.391497in}}%
\pgfpathlineto{\pgfqpoint{0.934589in}{1.404777in}}%
\pgfpathlineto{\pgfqpoint{0.831141in}{1.418280in}}%
\pgfpathlineto{\pgfqpoint{0.727693in}{1.431019in}}%
\pgfpathlineto{\pgfqpoint{0.624246in}{1.439961in}}%
\pgfpathlineto{\pgfqpoint{0.520798in}{1.450568in}}%
\pgfpathlineto{\pgfqpoint{0.520798in}{1.450568in}}%
\pgfpathclose%
\pgfusepath{fill}%
\end{pgfscope}%
\begin{pgfscope}%
\pgfpathrectangle{\pgfqpoint{0.520798in}{0.442177in}}{\pgfqpoint{2.048265in}{1.142908in}}%
\pgfusepath{clip}%
\pgfsetroundcap%
\pgfsetroundjoin%
\pgfsetlinewidth{1.003750pt}%
\definecolor{currentstroke}{rgb}{0.870588,0.560784,0.019608}%
\pgfsetstrokecolor{currentstroke}%
\pgfsetdash{}{0pt}%
\pgfpathmoveto{\pgfqpoint{0.520798in}{1.475703in}}%
\pgfpathlineto{\pgfqpoint{0.624246in}{1.465052in}}%
\pgfpathlineto{\pgfqpoint{0.727693in}{1.453567in}}%
\pgfpathlineto{\pgfqpoint{0.831141in}{1.440879in}}%
\pgfpathlineto{\pgfqpoint{0.934589in}{1.424207in}}%
\pgfpathlineto{\pgfqpoint{1.038037in}{1.409759in}}%
\pgfpathlineto{\pgfqpoint{1.141484in}{1.390680in}}%
\pgfpathlineto{\pgfqpoint{1.244932in}{1.371137in}}%
\pgfpathlineto{\pgfqpoint{1.348380in}{1.351965in}}%
\pgfpathlineto{\pgfqpoint{1.451828in}{1.331774in}}%
\pgfpathlineto{\pgfqpoint{1.555275in}{1.309176in}}%
\pgfpathlineto{\pgfqpoint{1.658723in}{1.283150in}}%
\pgfpathlineto{\pgfqpoint{1.762171in}{1.255642in}}%
\pgfpathlineto{\pgfqpoint{1.865619in}{1.222392in}}%
\pgfpathlineto{\pgfqpoint{1.969066in}{1.186086in}}%
\pgfpathlineto{\pgfqpoint{2.072514in}{1.134220in}}%
\pgfpathlineto{\pgfqpoint{2.175962in}{0.442177in}}%
\pgfpathlineto{\pgfqpoint{2.279410in}{0.442177in}}%
\pgfpathlineto{\pgfqpoint{2.382857in}{0.442177in}}%
\pgfpathlineto{\pgfqpoint{2.486305in}{0.442177in}}%
\pgfpathlineto{\pgfqpoint{2.570730in}{0.442177in}}%
\pgfusepath{stroke}%
\end{pgfscope}%
\begin{pgfscope}%
\pgfpathrectangle{\pgfqpoint{0.520798in}{0.442177in}}{\pgfqpoint{2.048265in}{1.142908in}}%
\pgfusepath{clip}%
\pgfsetbuttcap%
\pgfsetroundjoin%
\definecolor{currentfill}{rgb}{0.870588,0.560784,0.019608}%
\pgfsetfillcolor{currentfill}%
\pgfsetfillopacity{0.500000}%
\pgfsetlinewidth{0.000000pt}%
\definecolor{currentstroke}{rgb}{0.870588,0.560784,0.019608}%
\pgfsetstrokecolor{currentstroke}%
\pgfsetstrokeopacity{0.500000}%
\pgfsetdash{}{0pt}%
\pgfpathmoveto{\pgfqpoint{0.520798in}{1.477843in}}%
\pgfpathlineto{\pgfqpoint{0.520798in}{1.473563in}}%
\pgfpathlineto{\pgfqpoint{0.624246in}{1.463385in}}%
\pgfpathlineto{\pgfqpoint{0.727693in}{1.449828in}}%
\pgfpathlineto{\pgfqpoint{0.831141in}{1.436804in}}%
\pgfpathlineto{\pgfqpoint{0.934589in}{1.420759in}}%
\pgfpathlineto{\pgfqpoint{1.038037in}{1.406941in}}%
\pgfpathlineto{\pgfqpoint{1.141484in}{1.385962in}}%
\pgfpathlineto{\pgfqpoint{1.244932in}{1.367684in}}%
\pgfpathlineto{\pgfqpoint{1.348380in}{1.346357in}}%
\pgfpathlineto{\pgfqpoint{1.451828in}{1.325459in}}%
\pgfpathlineto{\pgfqpoint{1.555275in}{1.304663in}}%
\pgfpathlineto{\pgfqpoint{1.658723in}{1.277919in}}%
\pgfpathlineto{\pgfqpoint{1.762171in}{1.249593in}}%
\pgfpathlineto{\pgfqpoint{1.865619in}{1.216809in}}%
\pgfpathlineto{\pgfqpoint{1.969066in}{1.181581in}}%
\pgfpathlineto{\pgfqpoint{2.072514in}{1.126330in}}%
\pgfpathlineto{\pgfqpoint{2.175962in}{0.442177in}}%
\pgfpathlineto{\pgfqpoint{2.279410in}{0.442177in}}%
\pgfpathlineto{\pgfqpoint{2.382857in}{0.442177in}}%
\pgfpathlineto{\pgfqpoint{2.486305in}{0.442177in}}%
\pgfpathlineto{\pgfqpoint{2.589753in}{0.442177in}}%
\pgfpathlineto{\pgfqpoint{2.693201in}{0.442177in}}%
\pgfpathlineto{\pgfqpoint{2.796648in}{0.442177in}}%
\pgfpathlineto{\pgfqpoint{2.900096in}{0.442177in}}%
\pgfpathlineto{\pgfqpoint{3.003544in}{0.442177in}}%
\pgfpathlineto{\pgfqpoint{3.106992in}{0.442177in}}%
\pgfpathlineto{\pgfqpoint{3.210439in}{0.442177in}}%
\pgfpathlineto{\pgfqpoint{3.313887in}{0.442177in}}%
\pgfpathlineto{\pgfqpoint{3.417335in}{0.442177in}}%
\pgfpathlineto{\pgfqpoint{3.520783in}{0.442177in}}%
\pgfpathlineto{\pgfqpoint{3.624230in}{0.442177in}}%
\pgfpathlineto{\pgfqpoint{3.727678in}{0.442177in}}%
\pgfpathlineto{\pgfqpoint{3.831126in}{0.442177in}}%
\pgfpathlineto{\pgfqpoint{3.934574in}{0.442177in}}%
\pgfpathlineto{\pgfqpoint{4.038021in}{0.442177in}}%
\pgfpathlineto{\pgfqpoint{4.141469in}{0.442177in}}%
\pgfpathlineto{\pgfqpoint{4.244917in}{0.442177in}}%
\pgfpathlineto{\pgfqpoint{4.348365in}{0.442177in}}%
\pgfpathlineto{\pgfqpoint{4.451812in}{0.442177in}}%
\pgfpathlineto{\pgfqpoint{4.555260in}{0.442177in}}%
\pgfpathlineto{\pgfqpoint{4.658708in}{0.442177in}}%
\pgfpathlineto{\pgfqpoint{4.762156in}{0.442177in}}%
\pgfpathlineto{\pgfqpoint{4.865603in}{0.442177in}}%
\pgfpathlineto{\pgfqpoint{4.969051in}{0.442177in}}%
\pgfpathlineto{\pgfqpoint{5.072499in}{0.442177in}}%
\pgfpathlineto{\pgfqpoint{5.175947in}{0.442177in}}%
\pgfpathlineto{\pgfqpoint{5.279394in}{0.442177in}}%
\pgfpathlineto{\pgfqpoint{5.382842in}{0.442177in}}%
\pgfpathlineto{\pgfqpoint{5.486290in}{0.442177in}}%
\pgfpathlineto{\pgfqpoint{5.589738in}{0.442177in}}%
\pgfpathlineto{\pgfqpoint{5.693185in}{0.442177in}}%
\pgfpathlineto{\pgfqpoint{5.796633in}{0.442177in}}%
\pgfpathlineto{\pgfqpoint{5.900081in}{0.442177in}}%
\pgfpathlineto{\pgfqpoint{6.003529in}{0.442177in}}%
\pgfpathlineto{\pgfqpoint{6.106976in}{0.442177in}}%
\pgfpathlineto{\pgfqpoint{6.210424in}{0.442177in}}%
\pgfpathlineto{\pgfqpoint{6.313872in}{0.442177in}}%
\pgfpathlineto{\pgfqpoint{6.417320in}{0.442177in}}%
\pgfpathlineto{\pgfqpoint{6.520767in}{0.442177in}}%
\pgfpathlineto{\pgfqpoint{6.624215in}{0.442177in}}%
\pgfpathlineto{\pgfqpoint{6.727663in}{0.442177in}}%
\pgfpathlineto{\pgfqpoint{6.831111in}{0.442177in}}%
\pgfpathlineto{\pgfqpoint{6.934558in}{0.442177in}}%
\pgfpathlineto{\pgfqpoint{7.038006in}{0.442177in}}%
\pgfpathlineto{\pgfqpoint{7.141454in}{0.442177in}}%
\pgfpathlineto{\pgfqpoint{7.244902in}{0.442177in}}%
\pgfpathlineto{\pgfqpoint{7.348349in}{0.442177in}}%
\pgfpathlineto{\pgfqpoint{7.451797in}{0.442177in}}%
\pgfpathlineto{\pgfqpoint{7.555245in}{0.442177in}}%
\pgfpathlineto{\pgfqpoint{7.658693in}{0.442177in}}%
\pgfpathlineto{\pgfqpoint{7.762140in}{0.442177in}}%
\pgfpathlineto{\pgfqpoint{7.865588in}{0.442177in}}%
\pgfpathlineto{\pgfqpoint{7.969036in}{0.442177in}}%
\pgfpathlineto{\pgfqpoint{8.072484in}{0.442177in}}%
\pgfpathlineto{\pgfqpoint{8.175931in}{0.442177in}}%
\pgfpathlineto{\pgfqpoint{8.279379in}{0.442177in}}%
\pgfpathlineto{\pgfqpoint{8.382827in}{0.442177in}}%
\pgfpathlineto{\pgfqpoint{8.486275in}{0.442177in}}%
\pgfpathlineto{\pgfqpoint{8.589722in}{0.442177in}}%
\pgfpathlineto{\pgfqpoint{8.693170in}{0.442177in}}%
\pgfpathlineto{\pgfqpoint{8.796618in}{0.442177in}}%
\pgfpathlineto{\pgfqpoint{8.900066in}{0.442177in}}%
\pgfpathlineto{\pgfqpoint{9.003513in}{0.442177in}}%
\pgfpathlineto{\pgfqpoint{9.106961in}{0.442177in}}%
\pgfpathlineto{\pgfqpoint{9.210409in}{0.442177in}}%
\pgfpathlineto{\pgfqpoint{9.313857in}{0.442177in}}%
\pgfpathlineto{\pgfqpoint{9.417304in}{0.442177in}}%
\pgfpathlineto{\pgfqpoint{9.520752in}{0.442177in}}%
\pgfpathlineto{\pgfqpoint{9.624200in}{0.442177in}}%
\pgfpathlineto{\pgfqpoint{9.727648in}{0.442177in}}%
\pgfpathlineto{\pgfqpoint{9.831095in}{0.442177in}}%
\pgfpathlineto{\pgfqpoint{9.934543in}{0.442177in}}%
\pgfpathlineto{\pgfqpoint{10.037991in}{0.442177in}}%
\pgfpathlineto{\pgfqpoint{10.141439in}{0.442177in}}%
\pgfpathlineto{\pgfqpoint{10.244886in}{0.442177in}}%
\pgfpathlineto{\pgfqpoint{10.348334in}{0.442177in}}%
\pgfpathlineto{\pgfqpoint{10.451782in}{0.442177in}}%
\pgfpathlineto{\pgfqpoint{10.555230in}{0.442177in}}%
\pgfpathlineto{\pgfqpoint{10.658677in}{0.442177in}}%
\pgfpathlineto{\pgfqpoint{10.762125in}{0.442177in}}%
\pgfpathlineto{\pgfqpoint{10.762125in}{0.442177in}}%
\pgfpathlineto{\pgfqpoint{10.762125in}{0.442177in}}%
\pgfpathlineto{\pgfqpoint{10.658677in}{0.442177in}}%
\pgfpathlineto{\pgfqpoint{10.555230in}{0.442177in}}%
\pgfpathlineto{\pgfqpoint{10.451782in}{0.442177in}}%
\pgfpathlineto{\pgfqpoint{10.348334in}{0.442177in}}%
\pgfpathlineto{\pgfqpoint{10.244886in}{0.442177in}}%
\pgfpathlineto{\pgfqpoint{10.141439in}{0.442177in}}%
\pgfpathlineto{\pgfqpoint{10.037991in}{0.442177in}}%
\pgfpathlineto{\pgfqpoint{9.934543in}{0.442177in}}%
\pgfpathlineto{\pgfqpoint{9.831095in}{0.442177in}}%
\pgfpathlineto{\pgfqpoint{9.727648in}{0.442177in}}%
\pgfpathlineto{\pgfqpoint{9.624200in}{0.442177in}}%
\pgfpathlineto{\pgfqpoint{9.520752in}{0.442177in}}%
\pgfpathlineto{\pgfqpoint{9.417304in}{0.442177in}}%
\pgfpathlineto{\pgfqpoint{9.313857in}{0.442177in}}%
\pgfpathlineto{\pgfqpoint{9.210409in}{0.442177in}}%
\pgfpathlineto{\pgfqpoint{9.106961in}{0.442177in}}%
\pgfpathlineto{\pgfqpoint{9.003513in}{0.442177in}}%
\pgfpathlineto{\pgfqpoint{8.900066in}{0.442177in}}%
\pgfpathlineto{\pgfqpoint{8.796618in}{0.442177in}}%
\pgfpathlineto{\pgfqpoint{8.693170in}{0.442177in}}%
\pgfpathlineto{\pgfqpoint{8.589722in}{0.442177in}}%
\pgfpathlineto{\pgfqpoint{8.486275in}{0.442177in}}%
\pgfpathlineto{\pgfqpoint{8.382827in}{0.442177in}}%
\pgfpathlineto{\pgfqpoint{8.279379in}{0.442177in}}%
\pgfpathlineto{\pgfqpoint{8.175931in}{0.442177in}}%
\pgfpathlineto{\pgfqpoint{8.072484in}{0.442177in}}%
\pgfpathlineto{\pgfqpoint{7.969036in}{0.442177in}}%
\pgfpathlineto{\pgfqpoint{7.865588in}{0.442177in}}%
\pgfpathlineto{\pgfqpoint{7.762140in}{0.442177in}}%
\pgfpathlineto{\pgfqpoint{7.658693in}{0.442177in}}%
\pgfpathlineto{\pgfqpoint{7.555245in}{0.442177in}}%
\pgfpathlineto{\pgfqpoint{7.451797in}{0.442177in}}%
\pgfpathlineto{\pgfqpoint{7.348349in}{0.442177in}}%
\pgfpathlineto{\pgfqpoint{7.244902in}{0.442177in}}%
\pgfpathlineto{\pgfqpoint{7.141454in}{0.442177in}}%
\pgfpathlineto{\pgfqpoint{7.038006in}{0.442177in}}%
\pgfpathlineto{\pgfqpoint{6.934558in}{0.442177in}}%
\pgfpathlineto{\pgfqpoint{6.831111in}{0.442177in}}%
\pgfpathlineto{\pgfqpoint{6.727663in}{0.442177in}}%
\pgfpathlineto{\pgfqpoint{6.624215in}{0.442177in}}%
\pgfpathlineto{\pgfqpoint{6.520767in}{0.442177in}}%
\pgfpathlineto{\pgfqpoint{6.417320in}{0.442177in}}%
\pgfpathlineto{\pgfqpoint{6.313872in}{0.442177in}}%
\pgfpathlineto{\pgfqpoint{6.210424in}{0.442177in}}%
\pgfpathlineto{\pgfqpoint{6.106976in}{0.442177in}}%
\pgfpathlineto{\pgfqpoint{6.003529in}{0.442177in}}%
\pgfpathlineto{\pgfqpoint{5.900081in}{0.442177in}}%
\pgfpathlineto{\pgfqpoint{5.796633in}{0.442177in}}%
\pgfpathlineto{\pgfqpoint{5.693185in}{0.442177in}}%
\pgfpathlineto{\pgfqpoint{5.589738in}{0.442177in}}%
\pgfpathlineto{\pgfqpoint{5.486290in}{0.442177in}}%
\pgfpathlineto{\pgfqpoint{5.382842in}{0.442177in}}%
\pgfpathlineto{\pgfqpoint{5.279394in}{0.442177in}}%
\pgfpathlineto{\pgfqpoint{5.175947in}{0.442177in}}%
\pgfpathlineto{\pgfqpoint{5.072499in}{0.442177in}}%
\pgfpathlineto{\pgfqpoint{4.969051in}{0.442177in}}%
\pgfpathlineto{\pgfqpoint{4.865603in}{0.442177in}}%
\pgfpathlineto{\pgfqpoint{4.762156in}{0.442177in}}%
\pgfpathlineto{\pgfqpoint{4.658708in}{0.442177in}}%
\pgfpathlineto{\pgfqpoint{4.555260in}{0.442177in}}%
\pgfpathlineto{\pgfqpoint{4.451812in}{0.442177in}}%
\pgfpathlineto{\pgfqpoint{4.348365in}{0.442177in}}%
\pgfpathlineto{\pgfqpoint{4.244917in}{0.442177in}}%
\pgfpathlineto{\pgfqpoint{4.141469in}{0.442177in}}%
\pgfpathlineto{\pgfqpoint{4.038021in}{0.442177in}}%
\pgfpathlineto{\pgfqpoint{3.934574in}{0.442177in}}%
\pgfpathlineto{\pgfqpoint{3.831126in}{0.442177in}}%
\pgfpathlineto{\pgfqpoint{3.727678in}{0.442177in}}%
\pgfpathlineto{\pgfqpoint{3.624230in}{0.442177in}}%
\pgfpathlineto{\pgfqpoint{3.520783in}{0.442177in}}%
\pgfpathlineto{\pgfqpoint{3.417335in}{0.442177in}}%
\pgfpathlineto{\pgfqpoint{3.313887in}{0.442177in}}%
\pgfpathlineto{\pgfqpoint{3.210439in}{0.442177in}}%
\pgfpathlineto{\pgfqpoint{3.106992in}{0.442177in}}%
\pgfpathlineto{\pgfqpoint{3.003544in}{0.442177in}}%
\pgfpathlineto{\pgfqpoint{2.900096in}{0.442177in}}%
\pgfpathlineto{\pgfqpoint{2.796648in}{0.442177in}}%
\pgfpathlineto{\pgfqpoint{2.693201in}{0.442177in}}%
\pgfpathlineto{\pgfqpoint{2.589753in}{0.442177in}}%
\pgfpathlineto{\pgfqpoint{2.486305in}{0.442177in}}%
\pgfpathlineto{\pgfqpoint{2.382857in}{0.442177in}}%
\pgfpathlineto{\pgfqpoint{2.279410in}{0.442177in}}%
\pgfpathlineto{\pgfqpoint{2.175962in}{0.442177in}}%
\pgfpathlineto{\pgfqpoint{2.072514in}{1.142109in}}%
\pgfpathlineto{\pgfqpoint{1.969066in}{1.190591in}}%
\pgfpathlineto{\pgfqpoint{1.865619in}{1.227976in}}%
\pgfpathlineto{\pgfqpoint{1.762171in}{1.261692in}}%
\pgfpathlineto{\pgfqpoint{1.658723in}{1.288381in}}%
\pgfpathlineto{\pgfqpoint{1.555275in}{1.313688in}}%
\pgfpathlineto{\pgfqpoint{1.451828in}{1.338090in}}%
\pgfpathlineto{\pgfqpoint{1.348380in}{1.357573in}}%
\pgfpathlineto{\pgfqpoint{1.244932in}{1.374590in}}%
\pgfpathlineto{\pgfqpoint{1.141484in}{1.395397in}}%
\pgfpathlineto{\pgfqpoint{1.038037in}{1.412577in}}%
\pgfpathlineto{\pgfqpoint{0.934589in}{1.427656in}}%
\pgfpathlineto{\pgfqpoint{0.831141in}{1.444954in}}%
\pgfpathlineto{\pgfqpoint{0.727693in}{1.457307in}}%
\pgfpathlineto{\pgfqpoint{0.624246in}{1.466719in}}%
\pgfpathlineto{\pgfqpoint{0.520798in}{1.477843in}}%
\pgfpathlineto{\pgfqpoint{0.520798in}{1.477843in}}%
\pgfpathclose%
\pgfusepath{fill}%
\end{pgfscope}%
\begin{pgfscope}%
\pgfsetbuttcap%
\pgfsetmiterjoin%
\definecolor{currentfill}{rgb}{1.000000,1.000000,1.000000}%
\pgfsetfillcolor{currentfill}%
\pgfsetfillopacity{0.800000}%
\pgfsetlinewidth{1.003750pt}%
\definecolor{currentstroke}{rgb}{0.800000,0.800000,0.800000}%
\pgfsetstrokecolor{currentstroke}%
\pgfsetstrokeopacity{0.800000}%
\pgfsetdash{}{0pt}%
\pgfpathmoveto{\pgfqpoint{0.576353in}{0.497732in}}%
\pgfpathlineto{\pgfqpoint{1.440091in}{0.497732in}}%
\pgfpathlineto{\pgfqpoint{1.440091in}{0.840932in}}%
\pgfpathlineto{\pgfqpoint{0.576353in}{0.840932in}}%
\pgfpathlineto{\pgfqpoint{0.576353in}{0.497732in}}%
\pgfpathclose%
\pgfusepath{stroke,fill}%
\end{pgfscope}%
\begin{pgfscope}%
\pgfsetroundcap%
\pgfsetroundjoin%
\pgfsetlinewidth{1.003750pt}%
\definecolor{currentstroke}{rgb}{0.003922,0.450980,0.698039}%
\pgfsetstrokecolor{currentstroke}%
\pgfsetdash{}{0pt}%
\pgfpathmoveto{\pgfqpoint{0.620798in}{0.757598in}}%
\pgfpathlineto{\pgfqpoint{0.731909in}{0.757598in}}%
\pgfpathlineto{\pgfqpoint{0.843020in}{0.757598in}}%
\pgfusepath{stroke}%
\end{pgfscope}%
\begin{pgfscope}%
\definecolor{textcolor}{rgb}{0.150000,0.150000,0.150000}%
\pgfsetstrokecolor{textcolor}%
\pgfsetfillcolor{textcolor}%
\pgftext[x=0.931909in,y=0.718710in,left,base]{\color{textcolor}\rmfamily\fontsize{8.000000}{9.600000}\selectfont PointNet}%
\end{pgfscope}%
\begin{pgfscope}%
\pgfsetroundcap%
\pgfsetroundjoin%
\pgfsetlinewidth{1.003750pt}%
\definecolor{currentstroke}{rgb}{0.870588,0.560784,0.019608}%
\pgfsetstrokecolor{currentstroke}%
\pgfsetdash{}{0pt}%
\pgfpathmoveto{\pgfqpoint{0.620798in}{0.602665in}}%
\pgfpathlineto{\pgfqpoint{0.731909in}{0.602665in}}%
\pgfpathlineto{\pgfqpoint{0.843020in}{0.602665in}}%
\pgfusepath{stroke}%
\end{pgfscope}%
\begin{pgfscope}%
\definecolor{textcolor}{rgb}{0.150000,0.150000,0.150000}%
\pgfsetstrokecolor{textcolor}%
\pgfsetfillcolor{textcolor}%
\pgftext[x=0.931909in,y=0.563777in,left,base]{\color{textcolor}\rmfamily\fontsize{8.000000}{9.600000}\selectfont DGCNN}%
\end{pgfscope}%
\end{pgfpicture}%
\makeatother%
\endgroup%

%% file: figures/experiments/pointclouds/0.1.pgf
\begingroup%
\makeatletter%
\begin{pgfpicture}%
\pgfpathrectangle{\pgfpointorigin}{\pgfqpoint{2.750000in}{1.699593in}}%
\pgfusepath{use as bounding box, clip}%
\begin{pgfscope}%
\pgfsetbuttcap%
\pgfsetmiterjoin%
\definecolor{currentfill}{rgb}{1.000000,1.000000,1.000000}%
\pgfsetfillcolor{currentfill}%
\pgfsetlinewidth{0.000000pt}%
\definecolor{currentstroke}{rgb}{1.000000,1.000000,1.000000}%
\pgfsetstrokecolor{currentstroke}%
\pgfsetdash{}{0pt}%
\pgfpathmoveto{\pgfqpoint{0.000000in}{0.000000in}}%
\pgfpathlineto{\pgfqpoint{2.750000in}{0.000000in}}%
\pgfpathlineto{\pgfqpoint{2.750000in}{1.699593in}}%
\pgfpathlineto{\pgfqpoint{0.000000in}{1.699593in}}%
\pgfpathlineto{\pgfqpoint{0.000000in}{0.000000in}}%
\pgfpathclose%
\pgfusepath{fill}%
\end{pgfscope}%
\begin{pgfscope}%
\pgfsetbuttcap%
\pgfsetmiterjoin%
\definecolor{currentfill}{rgb}{1.000000,1.000000,1.000000}%
\pgfsetfillcolor{currentfill}%
\pgfsetlinewidth{0.000000pt}%
\definecolor{currentstroke}{rgb}{0.000000,0.000000,0.000000}%
\pgfsetstrokecolor{currentstroke}%
\pgfsetstrokeopacity{0.000000}%
\pgfsetdash{}{0pt}%
\pgfpathmoveto{\pgfqpoint{0.520798in}{0.442177in}}%
\pgfpathlineto{\pgfqpoint{2.598467in}{0.442177in}}%
\pgfpathlineto{\pgfqpoint{2.598467in}{1.585085in}}%
\pgfpathlineto{\pgfqpoint{0.520798in}{1.585085in}}%
\pgfpathlineto{\pgfqpoint{0.520798in}{0.442177in}}%
\pgfpathclose%
\pgfusepath{fill}%
\end{pgfscope}%
\begin{pgfscope}%
\pgfpathrectangle{\pgfqpoint{0.520798in}{0.442177in}}{\pgfqpoint{2.077669in}{1.142908in}}%
\pgfusepath{clip}%
\pgfsetroundcap%
\pgfsetroundjoin%
\pgfsetlinewidth{0.501875pt}%
\definecolor{currentstroke}{rgb}{0.800000,0.800000,0.800000}%
\pgfsetstrokecolor{currentstroke}%
\pgfsetdash{}{0pt}%
\pgfpathmoveto{\pgfqpoint{0.520798in}{0.442177in}}%
\pgfpathlineto{\pgfqpoint{0.520798in}{1.585085in}}%
\pgfusepath{stroke}%
\end{pgfscope}%
\begin{pgfscope}%
\definecolor{textcolor}{rgb}{0.150000,0.150000,0.150000}%
\pgfsetstrokecolor{textcolor}%
\pgfsetfillcolor{textcolor}%
\pgftext[x=0.520798in,y=0.351899in,,top]{\color{textcolor}\rmfamily\fontsize{8.000000}{9.600000}\selectfont \(\displaystyle {0.0}\)}%
\end{pgfscope}%
\begin{pgfscope}%
\pgfpathrectangle{\pgfqpoint{0.520798in}{0.442177in}}{\pgfqpoint{2.077669in}{1.142908in}}%
\pgfusepath{clip}%
\pgfsetroundcap%
\pgfsetroundjoin%
\pgfsetlinewidth{0.501875pt}%
\definecolor{currentstroke}{rgb}{0.800000,0.800000,0.800000}%
\pgfsetstrokecolor{currentstroke}%
\pgfsetdash{}{0pt}%
\pgfpathmoveto{\pgfqpoint{1.040215in}{0.442177in}}%
\pgfpathlineto{\pgfqpoint{1.040215in}{1.585085in}}%
\pgfusepath{stroke}%
\end{pgfscope}%
\begin{pgfscope}%
\definecolor{textcolor}{rgb}{0.150000,0.150000,0.150000}%
\pgfsetstrokecolor{textcolor}%
\pgfsetfillcolor{textcolor}%
\pgftext[x=1.040215in,y=0.351899in,,top]{\color{textcolor}\rmfamily\fontsize{8.000000}{9.600000}\selectfont \(\displaystyle {0.1}\)}%
\end{pgfscope}%
\begin{pgfscope}%
\pgfpathrectangle{\pgfqpoint{0.520798in}{0.442177in}}{\pgfqpoint{2.077669in}{1.142908in}}%
\pgfusepath{clip}%
\pgfsetroundcap%
\pgfsetroundjoin%
\pgfsetlinewidth{0.501875pt}%
\definecolor{currentstroke}{rgb}{0.800000,0.800000,0.800000}%
\pgfsetstrokecolor{currentstroke}%
\pgfsetdash{}{0pt}%
\pgfpathmoveto{\pgfqpoint{1.559633in}{0.442177in}}%
\pgfpathlineto{\pgfqpoint{1.559633in}{1.585085in}}%
\pgfusepath{stroke}%
\end{pgfscope}%
\begin{pgfscope}%
\definecolor{textcolor}{rgb}{0.150000,0.150000,0.150000}%
\pgfsetstrokecolor{textcolor}%
\pgfsetfillcolor{textcolor}%
\pgftext[x=1.559633in,y=0.351899in,,top]{\color{textcolor}\rmfamily\fontsize{8.000000}{9.600000}\selectfont \(\displaystyle {0.2}\)}%
\end{pgfscope}%
\begin{pgfscope}%
\pgfpathrectangle{\pgfqpoint{0.520798in}{0.442177in}}{\pgfqpoint{2.077669in}{1.142908in}}%
\pgfusepath{clip}%
\pgfsetroundcap%
\pgfsetroundjoin%
\pgfsetlinewidth{0.501875pt}%
\definecolor{currentstroke}{rgb}{0.800000,0.800000,0.800000}%
\pgfsetstrokecolor{currentstroke}%
\pgfsetdash{}{0pt}%
\pgfpathmoveto{\pgfqpoint{2.079050in}{0.442177in}}%
\pgfpathlineto{\pgfqpoint{2.079050in}{1.585085in}}%
\pgfusepath{stroke}%
\end{pgfscope}%
\begin{pgfscope}%
\definecolor{textcolor}{rgb}{0.150000,0.150000,0.150000}%
\pgfsetstrokecolor{textcolor}%
\pgfsetfillcolor{textcolor}%
\pgftext[x=2.079050in,y=0.351899in,,top]{\color{textcolor}\rmfamily\fontsize{8.000000}{9.600000}\selectfont \(\displaystyle {0.3}\)}%
\end{pgfscope}%
\begin{pgfscope}%
\pgfpathrectangle{\pgfqpoint{0.520798in}{0.442177in}}{\pgfqpoint{2.077669in}{1.142908in}}%
\pgfusepath{clip}%
\pgfsetroundcap%
\pgfsetroundjoin%
\pgfsetlinewidth{0.501875pt}%
\definecolor{currentstroke}{rgb}{0.800000,0.800000,0.800000}%
\pgfsetstrokecolor{currentstroke}%
\pgfsetdash{}{0pt}%
\pgfpathmoveto{\pgfqpoint{2.598467in}{0.442177in}}%
\pgfpathlineto{\pgfqpoint{2.598467in}{1.585085in}}%
\pgfusepath{stroke}%
\end{pgfscope}%
\begin{pgfscope}%
\definecolor{textcolor}{rgb}{0.150000,0.150000,0.150000}%
\pgfsetstrokecolor{textcolor}%
\pgfsetfillcolor{textcolor}%
\pgftext[x=2.598467in,y=0.351899in,,top]{\color{textcolor}\rmfamily\fontsize{8.000000}{9.600000}\selectfont \(\displaystyle {0.4}\)}%
\end{pgfscope}%
\begin{pgfscope}%
\definecolor{textcolor}{rgb}{0.150000,0.150000,0.150000}%
\pgfsetstrokecolor{textcolor}%
\pgfsetfillcolor{textcolor}%
\pgftext[x=1.559633in,y=0.198219in,,top]{\color{textcolor}\rmfamily\fontsize{10.000000}{12.000000}\selectfont Correspondence distance \(\displaystyle \epsilon\)}%
\end{pgfscope}%
\begin{pgfscope}%
\pgfpathrectangle{\pgfqpoint{0.520798in}{0.442177in}}{\pgfqpoint{2.077669in}{1.142908in}}%
\pgfusepath{clip}%
\pgfsetroundcap%
\pgfsetroundjoin%
\pgfsetlinewidth{0.501875pt}%
\definecolor{currentstroke}{rgb}{0.800000,0.800000,0.800000}%
\pgfsetstrokecolor{currentstroke}%
\pgfsetdash{}{0pt}%
\pgfpathmoveto{\pgfqpoint{0.520798in}{0.442177in}}%
\pgfpathlineto{\pgfqpoint{2.598467in}{0.442177in}}%
\pgfusepath{stroke}%
\end{pgfscope}%
\begin{pgfscope}%
\definecolor{textcolor}{rgb}{0.150000,0.150000,0.150000}%
\pgfsetstrokecolor{textcolor}%
\pgfsetfillcolor{textcolor}%
\pgftext[x=0.273151in, y=0.403915in, left, base]{\color{textcolor}\rmfamily\fontsize{8.000000}{9.600000}\selectfont 0\%}%
\end{pgfscope}%
\begin{pgfscope}%
\pgfpathrectangle{\pgfqpoint{0.520798in}{0.442177in}}{\pgfqpoint{2.077669in}{1.142908in}}%
\pgfusepath{clip}%
\pgfsetroundcap%
\pgfsetroundjoin%
\pgfsetlinewidth{0.501875pt}%
\definecolor{currentstroke}{rgb}{0.800000,0.800000,0.800000}%
\pgfsetstrokecolor{currentstroke}%
\pgfsetdash{}{0pt}%
\pgfpathmoveto{\pgfqpoint{0.520798in}{0.670758in}}%
\pgfpathlineto{\pgfqpoint{2.598467in}{0.670758in}}%
\pgfusepath{stroke}%
\end{pgfscope}%
\begin{pgfscope}%
\definecolor{textcolor}{rgb}{0.150000,0.150000,0.150000}%
\pgfsetstrokecolor{textcolor}%
\pgfsetfillcolor{textcolor}%
\pgftext[x=0.214138in, y=0.632496in, left, base]{\color{textcolor}\rmfamily\fontsize{8.000000}{9.600000}\selectfont 20\%}%
\end{pgfscope}%
\begin{pgfscope}%
\pgfpathrectangle{\pgfqpoint{0.520798in}{0.442177in}}{\pgfqpoint{2.077669in}{1.142908in}}%
\pgfusepath{clip}%
\pgfsetroundcap%
\pgfsetroundjoin%
\pgfsetlinewidth{0.501875pt}%
\definecolor{currentstroke}{rgb}{0.800000,0.800000,0.800000}%
\pgfsetstrokecolor{currentstroke}%
\pgfsetdash{}{0pt}%
\pgfpathmoveto{\pgfqpoint{0.520798in}{0.899340in}}%
\pgfpathlineto{\pgfqpoint{2.598467in}{0.899340in}}%
\pgfusepath{stroke}%
\end{pgfscope}%
\begin{pgfscope}%
\definecolor{textcolor}{rgb}{0.150000,0.150000,0.150000}%
\pgfsetstrokecolor{textcolor}%
\pgfsetfillcolor{textcolor}%
\pgftext[x=0.214138in, y=0.861078in, left, base]{\color{textcolor}\rmfamily\fontsize{8.000000}{9.600000}\selectfont 40\%}%
\end{pgfscope}%
\begin{pgfscope}%
\pgfpathrectangle{\pgfqpoint{0.520798in}{0.442177in}}{\pgfqpoint{2.077669in}{1.142908in}}%
\pgfusepath{clip}%
\pgfsetroundcap%
\pgfsetroundjoin%
\pgfsetlinewidth{0.501875pt}%
\definecolor{currentstroke}{rgb}{0.800000,0.800000,0.800000}%
\pgfsetstrokecolor{currentstroke}%
\pgfsetdash{}{0pt}%
\pgfpathmoveto{\pgfqpoint{0.520798in}{1.127922in}}%
\pgfpathlineto{\pgfqpoint{2.598467in}{1.127922in}}%
\pgfusepath{stroke}%
\end{pgfscope}%
\begin{pgfscope}%
\definecolor{textcolor}{rgb}{0.150000,0.150000,0.150000}%
\pgfsetstrokecolor{textcolor}%
\pgfsetfillcolor{textcolor}%
\pgftext[x=0.214138in, y=1.089660in, left, base]{\color{textcolor}\rmfamily\fontsize{8.000000}{9.600000}\selectfont 60\%}%
\end{pgfscope}%
\begin{pgfscope}%
\pgfpathrectangle{\pgfqpoint{0.520798in}{0.442177in}}{\pgfqpoint{2.077669in}{1.142908in}}%
\pgfusepath{clip}%
\pgfsetroundcap%
\pgfsetroundjoin%
\pgfsetlinewidth{0.501875pt}%
\definecolor{currentstroke}{rgb}{0.800000,0.800000,0.800000}%
\pgfsetstrokecolor{currentstroke}%
\pgfsetdash{}{0pt}%
\pgfpathmoveto{\pgfqpoint{0.520798in}{1.356504in}}%
\pgfpathlineto{\pgfqpoint{2.598467in}{1.356504in}}%
\pgfusepath{stroke}%
\end{pgfscope}%
\begin{pgfscope}%
\definecolor{textcolor}{rgb}{0.150000,0.150000,0.150000}%
\pgfsetstrokecolor{textcolor}%
\pgfsetfillcolor{textcolor}%
\pgftext[x=0.214138in, y=1.318241in, left, base]{\color{textcolor}\rmfamily\fontsize{8.000000}{9.600000}\selectfont 80\%}%
\end{pgfscope}%
\begin{pgfscope}%
\pgfpathrectangle{\pgfqpoint{0.520798in}{0.442177in}}{\pgfqpoint{2.077669in}{1.142908in}}%
\pgfusepath{clip}%
\pgfsetroundcap%
\pgfsetroundjoin%
\pgfsetlinewidth{0.501875pt}%
\definecolor{currentstroke}{rgb}{0.800000,0.800000,0.800000}%
\pgfsetstrokecolor{currentstroke}%
\pgfsetdash{}{0pt}%
\pgfpathmoveto{\pgfqpoint{0.520798in}{1.585085in}}%
\pgfpathlineto{\pgfqpoint{2.598467in}{1.585085in}}%
\pgfusepath{stroke}%
\end{pgfscope}%
\begin{pgfscope}%
\definecolor{textcolor}{rgb}{0.150000,0.150000,0.150000}%
\pgfsetstrokecolor{textcolor}%
\pgfsetfillcolor{textcolor}%
\pgftext[x=0.155124in, y=1.546823in, left, base]{\color{textcolor}\rmfamily\fontsize{8.000000}{9.600000}\selectfont 100\%}%
\end{pgfscope}%
\begin{pgfscope}%
\definecolor{textcolor}{rgb}{0.150000,0.150000,0.150000}%
\pgfsetstrokecolor{textcolor}%
\pgfsetfillcolor{textcolor}%
\pgftext[x=0.099569in,y=1.013631in,,bottom,rotate=90.000000]{\color{textcolor}\rmfamily\fontsize{10.000000}{12.000000}\selectfont Cert. Acc.}%
\end{pgfscope}%
\begin{pgfscope}%
\pgfsetrectcap%
\pgfsetmiterjoin%
\pgfsetlinewidth{0.752812pt}%
\definecolor{currentstroke}{rgb}{0.700000,0.700000,0.700000}%
\pgfsetstrokecolor{currentstroke}%
\pgfsetdash{}{0pt}%
\pgfpathmoveto{\pgfqpoint{0.520798in}{0.442177in}}%
\pgfpathlineto{\pgfqpoint{0.520798in}{1.585085in}}%
\pgfusepath{stroke}%
\end{pgfscope}%
\begin{pgfscope}%
\pgfsetrectcap%
\pgfsetmiterjoin%
\pgfsetlinewidth{0.752812pt}%
\definecolor{currentstroke}{rgb}{0.700000,0.700000,0.700000}%
\pgfsetstrokecolor{currentstroke}%
\pgfsetdash{}{0pt}%
\pgfpathmoveto{\pgfqpoint{2.598467in}{0.442177in}}%
\pgfpathlineto{\pgfqpoint{2.598467in}{1.585085in}}%
\pgfusepath{stroke}%
\end{pgfscope}%
\begin{pgfscope}%
\pgfsetrectcap%
\pgfsetmiterjoin%
\pgfsetlinewidth{0.752812pt}%
\definecolor{currentstroke}{rgb}{0.700000,0.700000,0.700000}%
\pgfsetstrokecolor{currentstroke}%
\pgfsetdash{}{0pt}%
\pgfpathmoveto{\pgfqpoint{0.520798in}{0.442177in}}%
\pgfpathlineto{\pgfqpoint{2.598467in}{0.442177in}}%
\pgfusepath{stroke}%
\end{pgfscope}%
\begin{pgfscope}%
\pgfsetrectcap%
\pgfsetmiterjoin%
\pgfsetlinewidth{0.752812pt}%
\definecolor{currentstroke}{rgb}{0.700000,0.700000,0.700000}%
\pgfsetstrokecolor{currentstroke}%
\pgfsetdash{}{0pt}%
\pgfpathmoveto{\pgfqpoint{0.520798in}{1.585085in}}%
\pgfpathlineto{\pgfqpoint{2.598467in}{1.585085in}}%
\pgfusepath{stroke}%
\end{pgfscope}%
\begin{pgfscope}%
\pgfpathrectangle{\pgfqpoint{0.520798in}{0.442177in}}{\pgfqpoint{2.077669in}{1.142908in}}%
\pgfusepath{clip}%
\pgfsetroundcap%
\pgfsetroundjoin%
\pgfsetlinewidth{1.003750pt}%
\definecolor{currentstroke}{rgb}{0.003922,0.450980,0.698039}%
\pgfsetstrokecolor{currentstroke}%
\pgfsetdash{}{0pt}%
\pgfpathmoveto{\pgfqpoint{0.520798in}{1.428190in}}%
\pgfpathlineto{\pgfqpoint{0.573264in}{1.419947in}}%
\pgfpathlineto{\pgfqpoint{0.625731in}{1.412352in}}%
\pgfpathlineto{\pgfqpoint{0.678197in}{1.403368in}}%
\pgfpathlineto{\pgfqpoint{0.730663in}{1.394199in}}%
\pgfpathlineto{\pgfqpoint{0.783130in}{1.383455in}}%
\pgfpathlineto{\pgfqpoint{0.835596in}{1.373823in}}%
\pgfpathlineto{\pgfqpoint{0.888063in}{1.362153in}}%
\pgfpathlineto{\pgfqpoint{0.940529in}{1.350020in}}%
\pgfpathlineto{\pgfqpoint{0.992996in}{1.340110in}}%
\pgfpathlineto{\pgfqpoint{1.045462in}{1.329089in}}%
\pgfpathlineto{\pgfqpoint{1.097928in}{1.316585in}}%
\pgfpathlineto{\pgfqpoint{1.150395in}{1.301859in}}%
\pgfpathlineto{\pgfqpoint{1.202861in}{1.289818in}}%
\pgfpathlineto{\pgfqpoint{1.255328in}{1.275926in}}%
\pgfpathlineto{\pgfqpoint{1.307794in}{1.260366in}}%
\pgfpathlineto{\pgfqpoint{1.360260in}{1.241009in}}%
\pgfpathlineto{\pgfqpoint{1.412727in}{1.223133in}}%
\pgfpathlineto{\pgfqpoint{1.465193in}{1.202294in}}%
\pgfpathlineto{\pgfqpoint{1.517660in}{1.180807in}}%
\pgfpathlineto{\pgfqpoint{1.570126in}{1.158393in}}%
\pgfpathlineto{\pgfqpoint{1.622592in}{1.136628in}}%
\pgfpathlineto{\pgfqpoint{1.675059in}{1.114585in}}%
\pgfpathlineto{\pgfqpoint{1.727525in}{1.089763in}}%
\pgfpathlineto{\pgfqpoint{1.779992in}{1.065405in}}%
\pgfpathlineto{\pgfqpoint{1.832458in}{1.039657in}}%
\pgfpathlineto{\pgfqpoint{1.884924in}{1.010575in}}%
\pgfpathlineto{\pgfqpoint{1.937391in}{0.981585in}}%
\pgfpathlineto{\pgfqpoint{1.989857in}{0.947131in}}%
\pgfpathlineto{\pgfqpoint{2.042324in}{0.911473in}}%
\pgfpathlineto{\pgfqpoint{2.094790in}{0.866183in}}%
\pgfpathlineto{\pgfqpoint{2.147256in}{0.795886in}}%
\pgfpathlineto{\pgfqpoint{2.199723in}{0.442177in}}%
\pgfpathlineto{\pgfqpoint{2.252189in}{0.442177in}}%
\pgfpathlineto{\pgfqpoint{2.304656in}{0.442177in}}%
\pgfpathlineto{\pgfqpoint{2.357122in}{0.442177in}}%
\pgfpathlineto{\pgfqpoint{2.409588in}{0.442177in}}%
\pgfpathlineto{\pgfqpoint{2.462055in}{0.442177in}}%
\pgfpathlineto{\pgfqpoint{2.514521in}{0.442177in}}%
\pgfpathlineto{\pgfqpoint{2.566988in}{0.442177in}}%
\pgfpathlineto{\pgfqpoint{2.600134in}{0.442177in}}%
\pgfusepath{stroke}%
\end{pgfscope}%
\begin{pgfscope}%
\pgfpathrectangle{\pgfqpoint{0.520798in}{0.442177in}}{\pgfqpoint{2.077669in}{1.142908in}}%
\pgfusepath{clip}%
\pgfsetbuttcap%
\pgfsetroundjoin%
\definecolor{currentfill}{rgb}{0.003922,0.450980,0.698039}%
\pgfsetfillcolor{currentfill}%
\pgfsetfillopacity{0.500000}%
\pgfsetlinewidth{0.000000pt}%
\definecolor{currentstroke}{rgb}{0.003922,0.450980,0.698039}%
\pgfsetstrokecolor{currentstroke}%
\pgfsetstrokeopacity{0.500000}%
\pgfsetdash{}{0pt}%
\pgfpathmoveto{\pgfqpoint{0.520798in}{1.430944in}}%
\pgfpathlineto{\pgfqpoint{0.520798in}{1.425436in}}%
\pgfpathlineto{\pgfqpoint{0.573264in}{1.417417in}}%
\pgfpathlineto{\pgfqpoint{0.625731in}{1.408545in}}%
\pgfpathlineto{\pgfqpoint{0.678197in}{1.397947in}}%
\pgfpathlineto{\pgfqpoint{0.730663in}{1.388502in}}%
\pgfpathlineto{\pgfqpoint{0.783130in}{1.379391in}}%
\pgfpathlineto{\pgfqpoint{0.835596in}{1.368491in}}%
\pgfpathlineto{\pgfqpoint{0.888063in}{1.355413in}}%
\pgfpathlineto{\pgfqpoint{0.940529in}{1.343620in}}%
\pgfpathlineto{\pgfqpoint{0.992996in}{1.333111in}}%
\pgfpathlineto{\pgfqpoint{1.045462in}{1.322124in}}%
\pgfpathlineto{\pgfqpoint{1.097928in}{1.309745in}}%
\pgfpathlineto{\pgfqpoint{1.150395in}{1.294729in}}%
\pgfpathlineto{\pgfqpoint{1.202861in}{1.280656in}}%
\pgfpathlineto{\pgfqpoint{1.255328in}{1.265965in}}%
\pgfpathlineto{\pgfqpoint{1.307794in}{1.249093in}}%
\pgfpathlineto{\pgfqpoint{1.360260in}{1.227091in}}%
\pgfpathlineto{\pgfqpoint{1.412727in}{1.205276in}}%
\pgfpathlineto{\pgfqpoint{1.465193in}{1.183310in}}%
\pgfpathlineto{\pgfqpoint{1.517660in}{1.163654in}}%
\pgfpathlineto{\pgfqpoint{1.570126in}{1.139529in}}%
\pgfpathlineto{\pgfqpoint{1.622592in}{1.119186in}}%
\pgfpathlineto{\pgfqpoint{1.675059in}{1.099448in}}%
\pgfpathlineto{\pgfqpoint{1.727525in}{1.076429in}}%
\pgfpathlineto{\pgfqpoint{1.779992in}{1.053791in}}%
\pgfpathlineto{\pgfqpoint{1.832458in}{1.023630in}}%
\pgfpathlineto{\pgfqpoint{1.884924in}{0.997244in}}%
\pgfpathlineto{\pgfqpoint{1.937391in}{0.967608in}}%
\pgfpathlineto{\pgfqpoint{1.989857in}{0.932992in}}%
\pgfpathlineto{\pgfqpoint{2.042324in}{0.895331in}}%
\pgfpathlineto{\pgfqpoint{2.094790in}{0.849245in}}%
\pgfpathlineto{\pgfqpoint{2.147256in}{0.778692in}}%
\pgfpathlineto{\pgfqpoint{2.199723in}{0.442177in}}%
\pgfpathlineto{\pgfqpoint{2.252189in}{0.442177in}}%
\pgfpathlineto{\pgfqpoint{2.304656in}{0.442177in}}%
\pgfpathlineto{\pgfqpoint{2.357122in}{0.442177in}}%
\pgfpathlineto{\pgfqpoint{2.409588in}{0.442177in}}%
\pgfpathlineto{\pgfqpoint{2.462055in}{0.442177in}}%
\pgfpathlineto{\pgfqpoint{2.514521in}{0.442177in}}%
\pgfpathlineto{\pgfqpoint{2.566988in}{0.442177in}}%
\pgfpathlineto{\pgfqpoint{2.619454in}{0.442177in}}%
\pgfpathlineto{\pgfqpoint{2.671920in}{0.442177in}}%
\pgfpathlineto{\pgfqpoint{2.724387in}{0.442177in}}%
\pgfpathlineto{\pgfqpoint{2.776853in}{0.442177in}}%
\pgfpathlineto{\pgfqpoint{2.829320in}{0.442177in}}%
\pgfpathlineto{\pgfqpoint{2.881786in}{0.442177in}}%
\pgfpathlineto{\pgfqpoint{2.934252in}{0.442177in}}%
\pgfpathlineto{\pgfqpoint{2.986719in}{0.442177in}}%
\pgfpathlineto{\pgfqpoint{3.039185in}{0.442177in}}%
\pgfpathlineto{\pgfqpoint{3.091652in}{0.442177in}}%
\pgfpathlineto{\pgfqpoint{3.144118in}{0.442177in}}%
\pgfpathlineto{\pgfqpoint{3.196584in}{0.442177in}}%
\pgfpathlineto{\pgfqpoint{3.249051in}{0.442177in}}%
\pgfpathlineto{\pgfqpoint{3.301517in}{0.442177in}}%
\pgfpathlineto{\pgfqpoint{3.353984in}{0.442177in}}%
\pgfpathlineto{\pgfqpoint{3.406450in}{0.442177in}}%
\pgfpathlineto{\pgfqpoint{3.458916in}{0.442177in}}%
\pgfpathlineto{\pgfqpoint{3.511383in}{0.442177in}}%
\pgfpathlineto{\pgfqpoint{3.563849in}{0.442177in}}%
\pgfpathlineto{\pgfqpoint{3.616316in}{0.442177in}}%
\pgfpathlineto{\pgfqpoint{3.668782in}{0.442177in}}%
\pgfpathlineto{\pgfqpoint{3.721248in}{0.442177in}}%
\pgfpathlineto{\pgfqpoint{3.773715in}{0.442177in}}%
\pgfpathlineto{\pgfqpoint{3.826181in}{0.442177in}}%
\pgfpathlineto{\pgfqpoint{3.878648in}{0.442177in}}%
\pgfpathlineto{\pgfqpoint{3.931114in}{0.442177in}}%
\pgfpathlineto{\pgfqpoint{3.983580in}{0.442177in}}%
\pgfpathlineto{\pgfqpoint{4.036047in}{0.442177in}}%
\pgfpathlineto{\pgfqpoint{4.088513in}{0.442177in}}%
\pgfpathlineto{\pgfqpoint{4.140980in}{0.442177in}}%
\pgfpathlineto{\pgfqpoint{4.193446in}{0.442177in}}%
\pgfpathlineto{\pgfqpoint{4.245912in}{0.442177in}}%
\pgfpathlineto{\pgfqpoint{4.298379in}{0.442177in}}%
\pgfpathlineto{\pgfqpoint{4.350845in}{0.442177in}}%
\pgfpathlineto{\pgfqpoint{4.403312in}{0.442177in}}%
\pgfpathlineto{\pgfqpoint{4.455778in}{0.442177in}}%
\pgfpathlineto{\pgfqpoint{4.508244in}{0.442177in}}%
\pgfpathlineto{\pgfqpoint{4.560711in}{0.442177in}}%
\pgfpathlineto{\pgfqpoint{4.613177in}{0.442177in}}%
\pgfpathlineto{\pgfqpoint{4.665644in}{0.442177in}}%
\pgfpathlineto{\pgfqpoint{4.718110in}{0.442177in}}%
\pgfpathlineto{\pgfqpoint{4.770576in}{0.442177in}}%
\pgfpathlineto{\pgfqpoint{4.823043in}{0.442177in}}%
\pgfpathlineto{\pgfqpoint{4.875509in}{0.442177in}}%
\pgfpathlineto{\pgfqpoint{4.927976in}{0.442177in}}%
\pgfpathlineto{\pgfqpoint{4.980442in}{0.442177in}}%
\pgfpathlineto{\pgfqpoint{5.032908in}{0.442177in}}%
\pgfpathlineto{\pgfqpoint{5.085375in}{0.442177in}}%
\pgfpathlineto{\pgfqpoint{5.137841in}{0.442177in}}%
\pgfpathlineto{\pgfqpoint{5.190308in}{0.442177in}}%
\pgfpathlineto{\pgfqpoint{5.242774in}{0.442177in}}%
\pgfpathlineto{\pgfqpoint{5.295240in}{0.442177in}}%
\pgfpathlineto{\pgfqpoint{5.347707in}{0.442177in}}%
\pgfpathlineto{\pgfqpoint{5.400173in}{0.442177in}}%
\pgfpathlineto{\pgfqpoint{5.452640in}{0.442177in}}%
\pgfpathlineto{\pgfqpoint{5.505106in}{0.442177in}}%
\pgfpathlineto{\pgfqpoint{5.557572in}{0.442177in}}%
\pgfpathlineto{\pgfqpoint{5.610039in}{0.442177in}}%
\pgfpathlineto{\pgfqpoint{5.662505in}{0.442177in}}%
\pgfpathlineto{\pgfqpoint{5.714972in}{0.442177in}}%
\pgfpathlineto{\pgfqpoint{5.714972in}{0.442177in}}%
\pgfpathlineto{\pgfqpoint{5.714972in}{0.442177in}}%
\pgfpathlineto{\pgfqpoint{5.662505in}{0.442177in}}%
\pgfpathlineto{\pgfqpoint{5.610039in}{0.442177in}}%
\pgfpathlineto{\pgfqpoint{5.557572in}{0.442177in}}%
\pgfpathlineto{\pgfqpoint{5.505106in}{0.442177in}}%
\pgfpathlineto{\pgfqpoint{5.452640in}{0.442177in}}%
\pgfpathlineto{\pgfqpoint{5.400173in}{0.442177in}}%
\pgfpathlineto{\pgfqpoint{5.347707in}{0.442177in}}%
\pgfpathlineto{\pgfqpoint{5.295240in}{0.442177in}}%
\pgfpathlineto{\pgfqpoint{5.242774in}{0.442177in}}%
\pgfpathlineto{\pgfqpoint{5.190308in}{0.442177in}}%
\pgfpathlineto{\pgfqpoint{5.137841in}{0.442177in}}%
\pgfpathlineto{\pgfqpoint{5.085375in}{0.442177in}}%
\pgfpathlineto{\pgfqpoint{5.032908in}{0.442177in}}%
\pgfpathlineto{\pgfqpoint{4.980442in}{0.442177in}}%
\pgfpathlineto{\pgfqpoint{4.927976in}{0.442177in}}%
\pgfpathlineto{\pgfqpoint{4.875509in}{0.442177in}}%
\pgfpathlineto{\pgfqpoint{4.823043in}{0.442177in}}%
\pgfpathlineto{\pgfqpoint{4.770576in}{0.442177in}}%
\pgfpathlineto{\pgfqpoint{4.718110in}{0.442177in}}%
\pgfpathlineto{\pgfqpoint{4.665644in}{0.442177in}}%
\pgfpathlineto{\pgfqpoint{4.613177in}{0.442177in}}%
\pgfpathlineto{\pgfqpoint{4.560711in}{0.442177in}}%
\pgfpathlineto{\pgfqpoint{4.508244in}{0.442177in}}%
\pgfpathlineto{\pgfqpoint{4.455778in}{0.442177in}}%
\pgfpathlineto{\pgfqpoint{4.403312in}{0.442177in}}%
\pgfpathlineto{\pgfqpoint{4.350845in}{0.442177in}}%
\pgfpathlineto{\pgfqpoint{4.298379in}{0.442177in}}%
\pgfpathlineto{\pgfqpoint{4.245912in}{0.442177in}}%
\pgfpathlineto{\pgfqpoint{4.193446in}{0.442177in}}%
\pgfpathlineto{\pgfqpoint{4.140980in}{0.442177in}}%
\pgfpathlineto{\pgfqpoint{4.088513in}{0.442177in}}%
\pgfpathlineto{\pgfqpoint{4.036047in}{0.442177in}}%
\pgfpathlineto{\pgfqpoint{3.983580in}{0.442177in}}%
\pgfpathlineto{\pgfqpoint{3.931114in}{0.442177in}}%
\pgfpathlineto{\pgfqpoint{3.878648in}{0.442177in}}%
\pgfpathlineto{\pgfqpoint{3.826181in}{0.442177in}}%
\pgfpathlineto{\pgfqpoint{3.773715in}{0.442177in}}%
\pgfpathlineto{\pgfqpoint{3.721248in}{0.442177in}}%
\pgfpathlineto{\pgfqpoint{3.668782in}{0.442177in}}%
\pgfpathlineto{\pgfqpoint{3.616316in}{0.442177in}}%
\pgfpathlineto{\pgfqpoint{3.563849in}{0.442177in}}%
\pgfpathlineto{\pgfqpoint{3.511383in}{0.442177in}}%
\pgfpathlineto{\pgfqpoint{3.458916in}{0.442177in}}%
\pgfpathlineto{\pgfqpoint{3.406450in}{0.442177in}}%
\pgfpathlineto{\pgfqpoint{3.353984in}{0.442177in}}%
\pgfpathlineto{\pgfqpoint{3.301517in}{0.442177in}}%
\pgfpathlineto{\pgfqpoint{3.249051in}{0.442177in}}%
\pgfpathlineto{\pgfqpoint{3.196584in}{0.442177in}}%
\pgfpathlineto{\pgfqpoint{3.144118in}{0.442177in}}%
\pgfpathlineto{\pgfqpoint{3.091652in}{0.442177in}}%
\pgfpathlineto{\pgfqpoint{3.039185in}{0.442177in}}%
\pgfpathlineto{\pgfqpoint{2.986719in}{0.442177in}}%
\pgfpathlineto{\pgfqpoint{2.934252in}{0.442177in}}%
\pgfpathlineto{\pgfqpoint{2.881786in}{0.442177in}}%
\pgfpathlineto{\pgfqpoint{2.829320in}{0.442177in}}%
\pgfpathlineto{\pgfqpoint{2.776853in}{0.442177in}}%
\pgfpathlineto{\pgfqpoint{2.724387in}{0.442177in}}%
\pgfpathlineto{\pgfqpoint{2.671920in}{0.442177in}}%
\pgfpathlineto{\pgfqpoint{2.619454in}{0.442177in}}%
\pgfpathlineto{\pgfqpoint{2.566988in}{0.442177in}}%
\pgfpathlineto{\pgfqpoint{2.514521in}{0.442177in}}%
\pgfpathlineto{\pgfqpoint{2.462055in}{0.442177in}}%
\pgfpathlineto{\pgfqpoint{2.409588in}{0.442177in}}%
\pgfpathlineto{\pgfqpoint{2.357122in}{0.442177in}}%
\pgfpathlineto{\pgfqpoint{2.304656in}{0.442177in}}%
\pgfpathlineto{\pgfqpoint{2.252189in}{0.442177in}}%
\pgfpathlineto{\pgfqpoint{2.199723in}{0.442177in}}%
\pgfpathlineto{\pgfqpoint{2.147256in}{0.813080in}}%
\pgfpathlineto{\pgfqpoint{2.094790in}{0.883121in}}%
\pgfpathlineto{\pgfqpoint{2.042324in}{0.927615in}}%
\pgfpathlineto{\pgfqpoint{1.989857in}{0.961271in}}%
\pgfpathlineto{\pgfqpoint{1.937391in}{0.995562in}}%
\pgfpathlineto{\pgfqpoint{1.884924in}{1.023905in}}%
\pgfpathlineto{\pgfqpoint{1.832458in}{1.055684in}}%
\pgfpathlineto{\pgfqpoint{1.779992in}{1.077018in}}%
\pgfpathlineto{\pgfqpoint{1.727525in}{1.103097in}}%
\pgfpathlineto{\pgfqpoint{1.675059in}{1.129722in}}%
\pgfpathlineto{\pgfqpoint{1.622592in}{1.154070in}}%
\pgfpathlineto{\pgfqpoint{1.570126in}{1.177258in}}%
\pgfpathlineto{\pgfqpoint{1.517660in}{1.197960in}}%
\pgfpathlineto{\pgfqpoint{1.465193in}{1.221279in}}%
\pgfpathlineto{\pgfqpoint{1.412727in}{1.240991in}}%
\pgfpathlineto{\pgfqpoint{1.360260in}{1.254926in}}%
\pgfpathlineto{\pgfqpoint{1.307794in}{1.271639in}}%
\pgfpathlineto{\pgfqpoint{1.255328in}{1.285886in}}%
\pgfpathlineto{\pgfqpoint{1.202861in}{1.298981in}}%
\pgfpathlineto{\pgfqpoint{1.150395in}{1.308989in}}%
\pgfpathlineto{\pgfqpoint{1.097928in}{1.323425in}}%
\pgfpathlineto{\pgfqpoint{1.045462in}{1.336053in}}%
\pgfpathlineto{\pgfqpoint{0.992996in}{1.347109in}}%
\pgfpathlineto{\pgfqpoint{0.940529in}{1.356421in}}%
\pgfpathlineto{\pgfqpoint{0.888063in}{1.368893in}}%
\pgfpathlineto{\pgfqpoint{0.835596in}{1.379155in}}%
\pgfpathlineto{\pgfqpoint{0.783130in}{1.387520in}}%
\pgfpathlineto{\pgfqpoint{0.730663in}{1.399896in}}%
\pgfpathlineto{\pgfqpoint{0.678197in}{1.408789in}}%
\pgfpathlineto{\pgfqpoint{0.625731in}{1.416160in}}%
\pgfpathlineto{\pgfqpoint{0.573264in}{1.422477in}}%
\pgfpathlineto{\pgfqpoint{0.520798in}{1.430944in}}%
\pgfpathlineto{\pgfqpoint{0.520798in}{1.430944in}}%
\pgfpathclose%
\pgfusepath{fill}%
\end{pgfscope}%
\begin{pgfscope}%
\pgfpathrectangle{\pgfqpoint{0.520798in}{0.442177in}}{\pgfqpoint{2.077669in}{1.142908in}}%
\pgfusepath{clip}%
\pgfsetroundcap%
\pgfsetroundjoin%
\pgfsetlinewidth{1.003750pt}%
\definecolor{currentstroke}{rgb}{0.870588,0.560784,0.019608}%
\pgfsetstrokecolor{currentstroke}%
\pgfsetdash{}{0pt}%
\pgfpathmoveto{\pgfqpoint{0.520798in}{1.454586in}}%
\pgfpathlineto{\pgfqpoint{0.573264in}{1.446992in}}%
\pgfpathlineto{\pgfqpoint{0.625731in}{1.439119in}}%
\pgfpathlineto{\pgfqpoint{0.678197in}{1.431339in}}%
\pgfpathlineto{\pgfqpoint{0.730663in}{1.423281in}}%
\pgfpathlineto{\pgfqpoint{0.783130in}{1.414853in}}%
\pgfpathlineto{\pgfqpoint{0.835596in}{1.406240in}}%
\pgfpathlineto{\pgfqpoint{0.888063in}{1.397626in}}%
\pgfpathlineto{\pgfqpoint{0.940529in}{1.389290in}}%
\pgfpathlineto{\pgfqpoint{0.992996in}{1.379473in}}%
\pgfpathlineto{\pgfqpoint{1.045462in}{1.370952in}}%
\pgfpathlineto{\pgfqpoint{1.097928in}{1.359930in}}%
\pgfpathlineto{\pgfqpoint{1.150395in}{1.348168in}}%
\pgfpathlineto{\pgfqpoint{1.202861in}{1.336220in}}%
\pgfpathlineto{\pgfqpoint{1.255328in}{1.325476in}}%
\pgfpathlineto{\pgfqpoint{1.307794in}{1.311121in}}%
\pgfpathlineto{\pgfqpoint{1.360260in}{1.297321in}}%
\pgfpathlineto{\pgfqpoint{1.412727in}{1.283243in}}%
\pgfpathlineto{\pgfqpoint{1.465193in}{1.269257in}}%
\pgfpathlineto{\pgfqpoint{1.517660in}{1.253049in}}%
\pgfpathlineto{\pgfqpoint{1.570126in}{1.237397in}}%
\pgfpathlineto{\pgfqpoint{1.622592in}{1.220262in}}%
\pgfpathlineto{\pgfqpoint{1.675059in}{1.203684in}}%
\pgfpathlineto{\pgfqpoint{1.727525in}{1.182103in}}%
\pgfpathlineto{\pgfqpoint{1.779992in}{1.161820in}}%
\pgfpathlineto{\pgfqpoint{1.832458in}{1.139036in}}%
\pgfpathlineto{\pgfqpoint{1.884924in}{1.116345in}}%
\pgfpathlineto{\pgfqpoint{1.937391in}{1.093468in}}%
\pgfpathlineto{\pgfqpoint{1.989857in}{1.066053in}}%
\pgfpathlineto{\pgfqpoint{2.042324in}{1.036322in}}%
\pgfpathlineto{\pgfqpoint{2.094790in}{0.994459in}}%
\pgfpathlineto{\pgfqpoint{2.147256in}{0.922680in}}%
\pgfpathlineto{\pgfqpoint{2.199723in}{0.442177in}}%
\pgfpathlineto{\pgfqpoint{2.252189in}{0.442177in}}%
\pgfpathlineto{\pgfqpoint{2.304656in}{0.442177in}}%
\pgfpathlineto{\pgfqpoint{2.357122in}{0.442177in}}%
\pgfpathlineto{\pgfqpoint{2.409588in}{0.442177in}}%
\pgfpathlineto{\pgfqpoint{2.462055in}{0.442177in}}%
\pgfpathlineto{\pgfqpoint{2.514521in}{0.442177in}}%
\pgfpathlineto{\pgfqpoint{2.566988in}{0.442177in}}%
\pgfpathlineto{\pgfqpoint{2.600134in}{0.442177in}}%
\pgfusepath{stroke}%
\end{pgfscope}%
\begin{pgfscope}%
\pgfpathrectangle{\pgfqpoint{0.520798in}{0.442177in}}{\pgfqpoint{2.077669in}{1.142908in}}%
\pgfusepath{clip}%
\pgfsetbuttcap%
\pgfsetroundjoin%
\definecolor{currentfill}{rgb}{0.870588,0.560784,0.019608}%
\pgfsetfillcolor{currentfill}%
\pgfsetfillopacity{0.500000}%
\pgfsetlinewidth{0.000000pt}%
\definecolor{currentstroke}{rgb}{0.870588,0.560784,0.019608}%
\pgfsetstrokecolor{currentstroke}%
\pgfsetstrokeopacity{0.500000}%
\pgfsetdash{}{0pt}%
\pgfpathmoveto{\pgfqpoint{0.520798in}{1.455531in}}%
\pgfpathlineto{\pgfqpoint{0.520798in}{1.453642in}}%
\pgfpathlineto{\pgfqpoint{0.573264in}{1.445274in}}%
\pgfpathlineto{\pgfqpoint{0.625731in}{1.437426in}}%
\pgfpathlineto{\pgfqpoint{0.678197in}{1.428545in}}%
\pgfpathlineto{\pgfqpoint{0.730663in}{1.419011in}}%
\pgfpathlineto{\pgfqpoint{0.783130in}{1.409767in}}%
\pgfpathlineto{\pgfqpoint{0.835596in}{1.401708in}}%
\pgfpathlineto{\pgfqpoint{0.888063in}{1.393248in}}%
\pgfpathlineto{\pgfqpoint{0.940529in}{1.385546in}}%
\pgfpathlineto{\pgfqpoint{0.992996in}{1.375632in}}%
\pgfpathlineto{\pgfqpoint{1.045462in}{1.366437in}}%
\pgfpathlineto{\pgfqpoint{1.097928in}{1.355964in}}%
\pgfpathlineto{\pgfqpoint{1.150395in}{1.341930in}}%
\pgfpathlineto{\pgfqpoint{1.202861in}{1.331075in}}%
\pgfpathlineto{\pgfqpoint{1.255328in}{1.320915in}}%
\pgfpathlineto{\pgfqpoint{1.307794in}{1.306034in}}%
\pgfpathlineto{\pgfqpoint{1.360260in}{1.291379in}}%
\pgfpathlineto{\pgfqpoint{1.412727in}{1.278645in}}%
\pgfpathlineto{\pgfqpoint{1.465193in}{1.263555in}}%
\pgfpathlineto{\pgfqpoint{1.517660in}{1.246050in}}%
\pgfpathlineto{\pgfqpoint{1.570126in}{1.229443in}}%
\pgfpathlineto{\pgfqpoint{1.622592in}{1.210595in}}%
\pgfpathlineto{\pgfqpoint{1.675059in}{1.194341in}}%
\pgfpathlineto{\pgfqpoint{1.727525in}{1.174651in}}%
\pgfpathlineto{\pgfqpoint{1.779992in}{1.153500in}}%
\pgfpathlineto{\pgfqpoint{1.832458in}{1.129557in}}%
\pgfpathlineto{\pgfqpoint{1.884924in}{1.106843in}}%
\pgfpathlineto{\pgfqpoint{1.937391in}{1.083210in}}%
\pgfpathlineto{\pgfqpoint{1.989857in}{1.057567in}}%
\pgfpathlineto{\pgfqpoint{2.042324in}{1.030270in}}%
\pgfpathlineto{\pgfqpoint{2.094790in}{0.987770in}}%
\pgfpathlineto{\pgfqpoint{2.147256in}{0.913827in}}%
\pgfpathlineto{\pgfqpoint{2.199723in}{0.442177in}}%
\pgfpathlineto{\pgfqpoint{2.252189in}{0.442177in}}%
\pgfpathlineto{\pgfqpoint{2.304656in}{0.442177in}}%
\pgfpathlineto{\pgfqpoint{2.357122in}{0.442177in}}%
\pgfpathlineto{\pgfqpoint{2.409588in}{0.442177in}}%
\pgfpathlineto{\pgfqpoint{2.462055in}{0.442177in}}%
\pgfpathlineto{\pgfqpoint{2.514521in}{0.442177in}}%
\pgfpathlineto{\pgfqpoint{2.566988in}{0.442177in}}%
\pgfpathlineto{\pgfqpoint{2.619454in}{0.442177in}}%
\pgfpathlineto{\pgfqpoint{2.671920in}{0.442177in}}%
\pgfpathlineto{\pgfqpoint{2.724387in}{0.442177in}}%
\pgfpathlineto{\pgfqpoint{2.776853in}{0.442177in}}%
\pgfpathlineto{\pgfqpoint{2.829320in}{0.442177in}}%
\pgfpathlineto{\pgfqpoint{2.881786in}{0.442177in}}%
\pgfpathlineto{\pgfqpoint{2.934252in}{0.442177in}}%
\pgfpathlineto{\pgfqpoint{2.986719in}{0.442177in}}%
\pgfpathlineto{\pgfqpoint{3.039185in}{0.442177in}}%
\pgfpathlineto{\pgfqpoint{3.091652in}{0.442177in}}%
\pgfpathlineto{\pgfqpoint{3.144118in}{0.442177in}}%
\pgfpathlineto{\pgfqpoint{3.196584in}{0.442177in}}%
\pgfpathlineto{\pgfqpoint{3.249051in}{0.442177in}}%
\pgfpathlineto{\pgfqpoint{3.301517in}{0.442177in}}%
\pgfpathlineto{\pgfqpoint{3.353984in}{0.442177in}}%
\pgfpathlineto{\pgfqpoint{3.406450in}{0.442177in}}%
\pgfpathlineto{\pgfqpoint{3.458916in}{0.442177in}}%
\pgfpathlineto{\pgfqpoint{3.511383in}{0.442177in}}%
\pgfpathlineto{\pgfqpoint{3.563849in}{0.442177in}}%
\pgfpathlineto{\pgfqpoint{3.616316in}{0.442177in}}%
\pgfpathlineto{\pgfqpoint{3.668782in}{0.442177in}}%
\pgfpathlineto{\pgfqpoint{3.721248in}{0.442177in}}%
\pgfpathlineto{\pgfqpoint{3.773715in}{0.442177in}}%
\pgfpathlineto{\pgfqpoint{3.826181in}{0.442177in}}%
\pgfpathlineto{\pgfqpoint{3.878648in}{0.442177in}}%
\pgfpathlineto{\pgfqpoint{3.931114in}{0.442177in}}%
\pgfpathlineto{\pgfqpoint{3.983580in}{0.442177in}}%
\pgfpathlineto{\pgfqpoint{4.036047in}{0.442177in}}%
\pgfpathlineto{\pgfqpoint{4.088513in}{0.442177in}}%
\pgfpathlineto{\pgfqpoint{4.140980in}{0.442177in}}%
\pgfpathlineto{\pgfqpoint{4.193446in}{0.442177in}}%
\pgfpathlineto{\pgfqpoint{4.245912in}{0.442177in}}%
\pgfpathlineto{\pgfqpoint{4.298379in}{0.442177in}}%
\pgfpathlineto{\pgfqpoint{4.350845in}{0.442177in}}%
\pgfpathlineto{\pgfqpoint{4.403312in}{0.442177in}}%
\pgfpathlineto{\pgfqpoint{4.455778in}{0.442177in}}%
\pgfpathlineto{\pgfqpoint{4.508244in}{0.442177in}}%
\pgfpathlineto{\pgfqpoint{4.560711in}{0.442177in}}%
\pgfpathlineto{\pgfqpoint{4.613177in}{0.442177in}}%
\pgfpathlineto{\pgfqpoint{4.665644in}{0.442177in}}%
\pgfpathlineto{\pgfqpoint{4.718110in}{0.442177in}}%
\pgfpathlineto{\pgfqpoint{4.770576in}{0.442177in}}%
\pgfpathlineto{\pgfqpoint{4.823043in}{0.442177in}}%
\pgfpathlineto{\pgfqpoint{4.875509in}{0.442177in}}%
\pgfpathlineto{\pgfqpoint{4.927976in}{0.442177in}}%
\pgfpathlineto{\pgfqpoint{4.980442in}{0.442177in}}%
\pgfpathlineto{\pgfqpoint{5.032908in}{0.442177in}}%
\pgfpathlineto{\pgfqpoint{5.085375in}{0.442177in}}%
\pgfpathlineto{\pgfqpoint{5.137841in}{0.442177in}}%
\pgfpathlineto{\pgfqpoint{5.190308in}{0.442177in}}%
\pgfpathlineto{\pgfqpoint{5.242774in}{0.442177in}}%
\pgfpathlineto{\pgfqpoint{5.295240in}{0.442177in}}%
\pgfpathlineto{\pgfqpoint{5.347707in}{0.442177in}}%
\pgfpathlineto{\pgfqpoint{5.400173in}{0.442177in}}%
\pgfpathlineto{\pgfqpoint{5.452640in}{0.442177in}}%
\pgfpathlineto{\pgfqpoint{5.505106in}{0.442177in}}%
\pgfpathlineto{\pgfqpoint{5.557572in}{0.442177in}}%
\pgfpathlineto{\pgfqpoint{5.610039in}{0.442177in}}%
\pgfpathlineto{\pgfqpoint{5.662505in}{0.442177in}}%
\pgfpathlineto{\pgfqpoint{5.714972in}{0.442177in}}%
\pgfpathlineto{\pgfqpoint{5.714972in}{0.442177in}}%
\pgfpathlineto{\pgfqpoint{5.714972in}{0.442177in}}%
\pgfpathlineto{\pgfqpoint{5.662505in}{0.442177in}}%
\pgfpathlineto{\pgfqpoint{5.610039in}{0.442177in}}%
\pgfpathlineto{\pgfqpoint{5.557572in}{0.442177in}}%
\pgfpathlineto{\pgfqpoint{5.505106in}{0.442177in}}%
\pgfpathlineto{\pgfqpoint{5.452640in}{0.442177in}}%
\pgfpathlineto{\pgfqpoint{5.400173in}{0.442177in}}%
\pgfpathlineto{\pgfqpoint{5.347707in}{0.442177in}}%
\pgfpathlineto{\pgfqpoint{5.295240in}{0.442177in}}%
\pgfpathlineto{\pgfqpoint{5.242774in}{0.442177in}}%
\pgfpathlineto{\pgfqpoint{5.190308in}{0.442177in}}%
\pgfpathlineto{\pgfqpoint{5.137841in}{0.442177in}}%
\pgfpathlineto{\pgfqpoint{5.085375in}{0.442177in}}%
\pgfpathlineto{\pgfqpoint{5.032908in}{0.442177in}}%
\pgfpathlineto{\pgfqpoint{4.980442in}{0.442177in}}%
\pgfpathlineto{\pgfqpoint{4.927976in}{0.442177in}}%
\pgfpathlineto{\pgfqpoint{4.875509in}{0.442177in}}%
\pgfpathlineto{\pgfqpoint{4.823043in}{0.442177in}}%
\pgfpathlineto{\pgfqpoint{4.770576in}{0.442177in}}%
\pgfpathlineto{\pgfqpoint{4.718110in}{0.442177in}}%
\pgfpathlineto{\pgfqpoint{4.665644in}{0.442177in}}%
\pgfpathlineto{\pgfqpoint{4.613177in}{0.442177in}}%
\pgfpathlineto{\pgfqpoint{4.560711in}{0.442177in}}%
\pgfpathlineto{\pgfqpoint{4.508244in}{0.442177in}}%
\pgfpathlineto{\pgfqpoint{4.455778in}{0.442177in}}%
\pgfpathlineto{\pgfqpoint{4.403312in}{0.442177in}}%
\pgfpathlineto{\pgfqpoint{4.350845in}{0.442177in}}%
\pgfpathlineto{\pgfqpoint{4.298379in}{0.442177in}}%
\pgfpathlineto{\pgfqpoint{4.245912in}{0.442177in}}%
\pgfpathlineto{\pgfqpoint{4.193446in}{0.442177in}}%
\pgfpathlineto{\pgfqpoint{4.140980in}{0.442177in}}%
\pgfpathlineto{\pgfqpoint{4.088513in}{0.442177in}}%
\pgfpathlineto{\pgfqpoint{4.036047in}{0.442177in}}%
\pgfpathlineto{\pgfqpoint{3.983580in}{0.442177in}}%
\pgfpathlineto{\pgfqpoint{3.931114in}{0.442177in}}%
\pgfpathlineto{\pgfqpoint{3.878648in}{0.442177in}}%
\pgfpathlineto{\pgfqpoint{3.826181in}{0.442177in}}%
\pgfpathlineto{\pgfqpoint{3.773715in}{0.442177in}}%
\pgfpathlineto{\pgfqpoint{3.721248in}{0.442177in}}%
\pgfpathlineto{\pgfqpoint{3.668782in}{0.442177in}}%
\pgfpathlineto{\pgfqpoint{3.616316in}{0.442177in}}%
\pgfpathlineto{\pgfqpoint{3.563849in}{0.442177in}}%
\pgfpathlineto{\pgfqpoint{3.511383in}{0.442177in}}%
\pgfpathlineto{\pgfqpoint{3.458916in}{0.442177in}}%
\pgfpathlineto{\pgfqpoint{3.406450in}{0.442177in}}%
\pgfpathlineto{\pgfqpoint{3.353984in}{0.442177in}}%
\pgfpathlineto{\pgfqpoint{3.301517in}{0.442177in}}%
\pgfpathlineto{\pgfqpoint{3.249051in}{0.442177in}}%
\pgfpathlineto{\pgfqpoint{3.196584in}{0.442177in}}%
\pgfpathlineto{\pgfqpoint{3.144118in}{0.442177in}}%
\pgfpathlineto{\pgfqpoint{3.091652in}{0.442177in}}%
\pgfpathlineto{\pgfqpoint{3.039185in}{0.442177in}}%
\pgfpathlineto{\pgfqpoint{2.986719in}{0.442177in}}%
\pgfpathlineto{\pgfqpoint{2.934252in}{0.442177in}}%
\pgfpathlineto{\pgfqpoint{2.881786in}{0.442177in}}%
\pgfpathlineto{\pgfqpoint{2.829320in}{0.442177in}}%
\pgfpathlineto{\pgfqpoint{2.776853in}{0.442177in}}%
\pgfpathlineto{\pgfqpoint{2.724387in}{0.442177in}}%
\pgfpathlineto{\pgfqpoint{2.671920in}{0.442177in}}%
\pgfpathlineto{\pgfqpoint{2.619454in}{0.442177in}}%
\pgfpathlineto{\pgfqpoint{2.566988in}{0.442177in}}%
\pgfpathlineto{\pgfqpoint{2.514521in}{0.442177in}}%
\pgfpathlineto{\pgfqpoint{2.462055in}{0.442177in}}%
\pgfpathlineto{\pgfqpoint{2.409588in}{0.442177in}}%
\pgfpathlineto{\pgfqpoint{2.357122in}{0.442177in}}%
\pgfpathlineto{\pgfqpoint{2.304656in}{0.442177in}}%
\pgfpathlineto{\pgfqpoint{2.252189in}{0.442177in}}%
\pgfpathlineto{\pgfqpoint{2.199723in}{0.442177in}}%
\pgfpathlineto{\pgfqpoint{2.147256in}{0.931533in}}%
\pgfpathlineto{\pgfqpoint{2.094790in}{1.001148in}}%
\pgfpathlineto{\pgfqpoint{2.042324in}{1.042375in}}%
\pgfpathlineto{\pgfqpoint{1.989857in}{1.074538in}}%
\pgfpathlineto{\pgfqpoint{1.937391in}{1.103726in}}%
\pgfpathlineto{\pgfqpoint{1.884924in}{1.125846in}}%
\pgfpathlineto{\pgfqpoint{1.832458in}{1.148515in}}%
\pgfpathlineto{\pgfqpoint{1.779992in}{1.170140in}}%
\pgfpathlineto{\pgfqpoint{1.727525in}{1.189556in}}%
\pgfpathlineto{\pgfqpoint{1.675059in}{1.213027in}}%
\pgfpathlineto{\pgfqpoint{1.622592in}{1.229929in}}%
\pgfpathlineto{\pgfqpoint{1.570126in}{1.245350in}}%
\pgfpathlineto{\pgfqpoint{1.517660in}{1.260048in}}%
\pgfpathlineto{\pgfqpoint{1.465193in}{1.274959in}}%
\pgfpathlineto{\pgfqpoint{1.412727in}{1.287840in}}%
\pgfpathlineto{\pgfqpoint{1.360260in}{1.303263in}}%
\pgfpathlineto{\pgfqpoint{1.307794in}{1.316207in}}%
\pgfpathlineto{\pgfqpoint{1.255328in}{1.330038in}}%
\pgfpathlineto{\pgfqpoint{1.202861in}{1.341365in}}%
\pgfpathlineto{\pgfqpoint{1.150395in}{1.354406in}}%
\pgfpathlineto{\pgfqpoint{1.097928in}{1.363897in}}%
\pgfpathlineto{\pgfqpoint{1.045462in}{1.375467in}}%
\pgfpathlineto{\pgfqpoint{0.992996in}{1.383314in}}%
\pgfpathlineto{\pgfqpoint{0.940529in}{1.393034in}}%
\pgfpathlineto{\pgfqpoint{0.888063in}{1.402004in}}%
\pgfpathlineto{\pgfqpoint{0.835596in}{1.410771in}}%
\pgfpathlineto{\pgfqpoint{0.783130in}{1.419939in}}%
\pgfpathlineto{\pgfqpoint{0.730663in}{1.427552in}}%
\pgfpathlineto{\pgfqpoint{0.678197in}{1.434133in}}%
\pgfpathlineto{\pgfqpoint{0.625731in}{1.440812in}}%
\pgfpathlineto{\pgfqpoint{0.573264in}{1.448709in}}%
\pgfpathlineto{\pgfqpoint{0.520798in}{1.455531in}}%
\pgfpathlineto{\pgfqpoint{0.520798in}{1.455531in}}%
\pgfpathclose%
\pgfusepath{fill}%
\end{pgfscope}%
\begin{pgfscope}%
\pgfsetbuttcap%
\pgfsetmiterjoin%
\definecolor{currentfill}{rgb}{1.000000,1.000000,1.000000}%
\pgfsetfillcolor{currentfill}%
\pgfsetfillopacity{0.800000}%
\pgfsetlinewidth{1.003750pt}%
\definecolor{currentstroke}{rgb}{0.800000,0.800000,0.800000}%
\pgfsetstrokecolor{currentstroke}%
\pgfsetstrokeopacity{0.800000}%
\pgfsetdash{}{0pt}%
\pgfpathmoveto{\pgfqpoint{0.576353in}{0.497732in}}%
\pgfpathlineto{\pgfqpoint{1.440091in}{0.497732in}}%
\pgfpathlineto{\pgfqpoint{1.440091in}{0.840932in}}%
\pgfpathlineto{\pgfqpoint{0.576353in}{0.840932in}}%
\pgfpathlineto{\pgfqpoint{0.576353in}{0.497732in}}%
\pgfpathclose%
\pgfusepath{stroke,fill}%
\end{pgfscope}%
\begin{pgfscope}%
\pgfsetroundcap%
\pgfsetroundjoin%
\pgfsetlinewidth{1.003750pt}%
\definecolor{currentstroke}{rgb}{0.003922,0.450980,0.698039}%
\pgfsetstrokecolor{currentstroke}%
\pgfsetdash{}{0pt}%
\pgfpathmoveto{\pgfqpoint{0.620798in}{0.757598in}}%
\pgfpathlineto{\pgfqpoint{0.731909in}{0.757598in}}%
\pgfpathlineto{\pgfqpoint{0.843020in}{0.757598in}}%
\pgfusepath{stroke}%
\end{pgfscope}%
\begin{pgfscope}%
\definecolor{textcolor}{rgb}{0.150000,0.150000,0.150000}%
\pgfsetstrokecolor{textcolor}%
\pgfsetfillcolor{textcolor}%
\pgftext[x=0.931909in,y=0.718710in,left,base]{\color{textcolor}\rmfamily\fontsize{8.000000}{9.600000}\selectfont PointNet}%
\end{pgfscope}%
\begin{pgfscope}%
\pgfsetroundcap%
\pgfsetroundjoin%
\pgfsetlinewidth{1.003750pt}%
\definecolor{currentstroke}{rgb}{0.870588,0.560784,0.019608}%
\pgfsetstrokecolor{currentstroke}%
\pgfsetdash{}{0pt}%
\pgfpathmoveto{\pgfqpoint{0.620798in}{0.602665in}}%
\pgfpathlineto{\pgfqpoint{0.731909in}{0.602665in}}%
\pgfpathlineto{\pgfqpoint{0.843020in}{0.602665in}}%
\pgfusepath{stroke}%
\end{pgfscope}%
\begin{pgfscope}%
\definecolor{textcolor}{rgb}{0.150000,0.150000,0.150000}%
\pgfsetstrokecolor{textcolor}%
\pgfsetfillcolor{textcolor}%
\pgftext[x=0.931909in,y=0.563777in,left,base]{\color{textcolor}\rmfamily\fontsize{8.000000}{9.600000}\selectfont DGCNN}%
\end{pgfscope}%
\end{pgfpicture}%
\makeatother%
\endgroup%

%% file: figures/experiments/pointclouds/0.15.pgf
\begingroup%
\makeatletter%
\begin{pgfpicture}%
\pgfpathrectangle{\pgfpointorigin}{\pgfqpoint{2.750000in}{1.699593in}}%
\pgfusepath{use as bounding box, clip}%
\begin{pgfscope}%
\pgfsetbuttcap%
\pgfsetmiterjoin%
\definecolor{currentfill}{rgb}{1.000000,1.000000,1.000000}%
\pgfsetfillcolor{currentfill}%
\pgfsetlinewidth{0.000000pt}%
\definecolor{currentstroke}{rgb}{1.000000,1.000000,1.000000}%
\pgfsetstrokecolor{currentstroke}%
\pgfsetdash{}{0pt}%
\pgfpathmoveto{\pgfqpoint{0.000000in}{0.000000in}}%
\pgfpathlineto{\pgfqpoint{2.750000in}{0.000000in}}%
\pgfpathlineto{\pgfqpoint{2.750000in}{1.699593in}}%
\pgfpathlineto{\pgfqpoint{0.000000in}{1.699593in}}%
\pgfpathlineto{\pgfqpoint{0.000000in}{0.000000in}}%
\pgfpathclose%
\pgfusepath{fill}%
\end{pgfscope}%
\begin{pgfscope}%
\pgfsetbuttcap%
\pgfsetmiterjoin%
\definecolor{currentfill}{rgb}{1.000000,1.000000,1.000000}%
\pgfsetfillcolor{currentfill}%
\pgfsetlinewidth{0.000000pt}%
\definecolor{currentstroke}{rgb}{0.000000,0.000000,0.000000}%
\pgfsetstrokecolor{currentstroke}%
\pgfsetstrokeopacity{0.000000}%
\pgfsetdash{}{0pt}%
\pgfpathmoveto{\pgfqpoint{0.520798in}{0.442177in}}%
\pgfpathlineto{\pgfqpoint{2.598467in}{0.442177in}}%
\pgfpathlineto{\pgfqpoint{2.598467in}{1.585085in}}%
\pgfpathlineto{\pgfqpoint{0.520798in}{1.585085in}}%
\pgfpathlineto{\pgfqpoint{0.520798in}{0.442177in}}%
\pgfpathclose%
\pgfusepath{fill}%
\end{pgfscope}%
\begin{pgfscope}%
\pgfpathrectangle{\pgfqpoint{0.520798in}{0.442177in}}{\pgfqpoint{2.077669in}{1.142908in}}%
\pgfusepath{clip}%
\pgfsetroundcap%
\pgfsetroundjoin%
\pgfsetlinewidth{0.501875pt}%
\definecolor{currentstroke}{rgb}{0.800000,0.800000,0.800000}%
\pgfsetstrokecolor{currentstroke}%
\pgfsetdash{}{0pt}%
\pgfpathmoveto{\pgfqpoint{0.520798in}{0.442177in}}%
\pgfpathlineto{\pgfqpoint{0.520798in}{1.585085in}}%
\pgfusepath{stroke}%
\end{pgfscope}%
\begin{pgfscope}%
\definecolor{textcolor}{rgb}{0.150000,0.150000,0.150000}%
\pgfsetstrokecolor{textcolor}%
\pgfsetfillcolor{textcolor}%
\pgftext[x=0.520798in,y=0.351899in,,top]{\color{textcolor}\rmfamily\fontsize{8.000000}{9.600000}\selectfont \(\displaystyle {0.0}\)}%
\end{pgfscope}%
\begin{pgfscope}%
\pgfpathrectangle{\pgfqpoint{0.520798in}{0.442177in}}{\pgfqpoint{2.077669in}{1.142908in}}%
\pgfusepath{clip}%
\pgfsetroundcap%
\pgfsetroundjoin%
\pgfsetlinewidth{0.501875pt}%
\definecolor{currentstroke}{rgb}{0.800000,0.800000,0.800000}%
\pgfsetstrokecolor{currentstroke}%
\pgfsetdash{}{0pt}%
\pgfpathmoveto{\pgfqpoint{0.936332in}{0.442177in}}%
\pgfpathlineto{\pgfqpoint{0.936332in}{1.585085in}}%
\pgfusepath{stroke}%
\end{pgfscope}%
\begin{pgfscope}%
\definecolor{textcolor}{rgb}{0.150000,0.150000,0.150000}%
\pgfsetstrokecolor{textcolor}%
\pgfsetfillcolor{textcolor}%
\pgftext[x=0.936332in,y=0.351899in,,top]{\color{textcolor}\rmfamily\fontsize{8.000000}{9.600000}\selectfont \(\displaystyle {0.1}\)}%
\end{pgfscope}%
\begin{pgfscope}%
\pgfpathrectangle{\pgfqpoint{0.520798in}{0.442177in}}{\pgfqpoint{2.077669in}{1.142908in}}%
\pgfusepath{clip}%
\pgfsetroundcap%
\pgfsetroundjoin%
\pgfsetlinewidth{0.501875pt}%
\definecolor{currentstroke}{rgb}{0.800000,0.800000,0.800000}%
\pgfsetstrokecolor{currentstroke}%
\pgfsetdash{}{0pt}%
\pgfpathmoveto{\pgfqpoint{1.351866in}{0.442177in}}%
\pgfpathlineto{\pgfqpoint{1.351866in}{1.585085in}}%
\pgfusepath{stroke}%
\end{pgfscope}%
\begin{pgfscope}%
\definecolor{textcolor}{rgb}{0.150000,0.150000,0.150000}%
\pgfsetstrokecolor{textcolor}%
\pgfsetfillcolor{textcolor}%
\pgftext[x=1.351866in,y=0.351899in,,top]{\color{textcolor}\rmfamily\fontsize{8.000000}{9.600000}\selectfont \(\displaystyle {0.2}\)}%
\end{pgfscope}%
\begin{pgfscope}%
\pgfpathrectangle{\pgfqpoint{0.520798in}{0.442177in}}{\pgfqpoint{2.077669in}{1.142908in}}%
\pgfusepath{clip}%
\pgfsetroundcap%
\pgfsetroundjoin%
\pgfsetlinewidth{0.501875pt}%
\definecolor{currentstroke}{rgb}{0.800000,0.800000,0.800000}%
\pgfsetstrokecolor{currentstroke}%
\pgfsetdash{}{0pt}%
\pgfpathmoveto{\pgfqpoint{1.767400in}{0.442177in}}%
\pgfpathlineto{\pgfqpoint{1.767400in}{1.585085in}}%
\pgfusepath{stroke}%
\end{pgfscope}%
\begin{pgfscope}%
\definecolor{textcolor}{rgb}{0.150000,0.150000,0.150000}%
\pgfsetstrokecolor{textcolor}%
\pgfsetfillcolor{textcolor}%
\pgftext[x=1.767400in,y=0.351899in,,top]{\color{textcolor}\rmfamily\fontsize{8.000000}{9.600000}\selectfont \(\displaystyle {0.3}\)}%
\end{pgfscope}%
\begin{pgfscope}%
\pgfpathrectangle{\pgfqpoint{0.520798in}{0.442177in}}{\pgfqpoint{2.077669in}{1.142908in}}%
\pgfusepath{clip}%
\pgfsetroundcap%
\pgfsetroundjoin%
\pgfsetlinewidth{0.501875pt}%
\definecolor{currentstroke}{rgb}{0.800000,0.800000,0.800000}%
\pgfsetstrokecolor{currentstroke}%
\pgfsetdash{}{0pt}%
\pgfpathmoveto{\pgfqpoint{2.182933in}{0.442177in}}%
\pgfpathlineto{\pgfqpoint{2.182933in}{1.585085in}}%
\pgfusepath{stroke}%
\end{pgfscope}%
\begin{pgfscope}%
\definecolor{textcolor}{rgb}{0.150000,0.150000,0.150000}%
\pgfsetstrokecolor{textcolor}%
\pgfsetfillcolor{textcolor}%
\pgftext[x=2.182933in,y=0.351899in,,top]{\color{textcolor}\rmfamily\fontsize{8.000000}{9.600000}\selectfont \(\displaystyle {0.4}\)}%
\end{pgfscope}%
\begin{pgfscope}%
\pgfpathrectangle{\pgfqpoint{0.520798in}{0.442177in}}{\pgfqpoint{2.077669in}{1.142908in}}%
\pgfusepath{clip}%
\pgfsetroundcap%
\pgfsetroundjoin%
\pgfsetlinewidth{0.501875pt}%
\definecolor{currentstroke}{rgb}{0.800000,0.800000,0.800000}%
\pgfsetstrokecolor{currentstroke}%
\pgfsetdash{}{0pt}%
\pgfpathmoveto{\pgfqpoint{2.598467in}{0.442177in}}%
\pgfpathlineto{\pgfqpoint{2.598467in}{1.585085in}}%
\pgfusepath{stroke}%
\end{pgfscope}%
\begin{pgfscope}%
\definecolor{textcolor}{rgb}{0.150000,0.150000,0.150000}%
\pgfsetstrokecolor{textcolor}%
\pgfsetfillcolor{textcolor}%
\pgftext[x=2.598467in,y=0.351899in,,top]{\color{textcolor}\rmfamily\fontsize{8.000000}{9.600000}\selectfont \(\displaystyle {0.5}\)}%
\end{pgfscope}%
\begin{pgfscope}%
\definecolor{textcolor}{rgb}{0.150000,0.150000,0.150000}%
\pgfsetstrokecolor{textcolor}%
\pgfsetfillcolor{textcolor}%
\pgftext[x=1.559633in,y=0.198219in,,top]{\color{textcolor}\rmfamily\fontsize{10.000000}{12.000000}\selectfont Correspondence distance \(\displaystyle \epsilon\)}%
\end{pgfscope}%
\begin{pgfscope}%
\pgfpathrectangle{\pgfqpoint{0.520798in}{0.442177in}}{\pgfqpoint{2.077669in}{1.142908in}}%
\pgfusepath{clip}%
\pgfsetroundcap%
\pgfsetroundjoin%
\pgfsetlinewidth{0.501875pt}%
\definecolor{currentstroke}{rgb}{0.800000,0.800000,0.800000}%
\pgfsetstrokecolor{currentstroke}%
\pgfsetdash{}{0pt}%
\pgfpathmoveto{\pgfqpoint{0.520798in}{0.442177in}}%
\pgfpathlineto{\pgfqpoint{2.598467in}{0.442177in}}%
\pgfusepath{stroke}%
\end{pgfscope}%
\begin{pgfscope}%
\definecolor{textcolor}{rgb}{0.150000,0.150000,0.150000}%
\pgfsetstrokecolor{textcolor}%
\pgfsetfillcolor{textcolor}%
\pgftext[x=0.273151in, y=0.403915in, left, base]{\color{textcolor}\rmfamily\fontsize{8.000000}{9.600000}\selectfont 0\%}%
\end{pgfscope}%
\begin{pgfscope}%
\pgfpathrectangle{\pgfqpoint{0.520798in}{0.442177in}}{\pgfqpoint{2.077669in}{1.142908in}}%
\pgfusepath{clip}%
\pgfsetroundcap%
\pgfsetroundjoin%
\pgfsetlinewidth{0.501875pt}%
\definecolor{currentstroke}{rgb}{0.800000,0.800000,0.800000}%
\pgfsetstrokecolor{currentstroke}%
\pgfsetdash{}{0pt}%
\pgfpathmoveto{\pgfqpoint{0.520798in}{0.670758in}}%
\pgfpathlineto{\pgfqpoint{2.598467in}{0.670758in}}%
\pgfusepath{stroke}%
\end{pgfscope}%
\begin{pgfscope}%
\definecolor{textcolor}{rgb}{0.150000,0.150000,0.150000}%
\pgfsetstrokecolor{textcolor}%
\pgfsetfillcolor{textcolor}%
\pgftext[x=0.214138in, y=0.632496in, left, base]{\color{textcolor}\rmfamily\fontsize{8.000000}{9.600000}\selectfont 20\%}%
\end{pgfscope}%
\begin{pgfscope}%
\pgfpathrectangle{\pgfqpoint{0.520798in}{0.442177in}}{\pgfqpoint{2.077669in}{1.142908in}}%
\pgfusepath{clip}%
\pgfsetroundcap%
\pgfsetroundjoin%
\pgfsetlinewidth{0.501875pt}%
\definecolor{currentstroke}{rgb}{0.800000,0.800000,0.800000}%
\pgfsetstrokecolor{currentstroke}%
\pgfsetdash{}{0pt}%
\pgfpathmoveto{\pgfqpoint{0.520798in}{0.899340in}}%
\pgfpathlineto{\pgfqpoint{2.598467in}{0.899340in}}%
\pgfusepath{stroke}%
\end{pgfscope}%
\begin{pgfscope}%
\definecolor{textcolor}{rgb}{0.150000,0.150000,0.150000}%
\pgfsetstrokecolor{textcolor}%
\pgfsetfillcolor{textcolor}%
\pgftext[x=0.214138in, y=0.861078in, left, base]{\color{textcolor}\rmfamily\fontsize{8.000000}{9.600000}\selectfont 40\%}%
\end{pgfscope}%
\begin{pgfscope}%
\pgfpathrectangle{\pgfqpoint{0.520798in}{0.442177in}}{\pgfqpoint{2.077669in}{1.142908in}}%
\pgfusepath{clip}%
\pgfsetroundcap%
\pgfsetroundjoin%
\pgfsetlinewidth{0.501875pt}%
\definecolor{currentstroke}{rgb}{0.800000,0.800000,0.800000}%
\pgfsetstrokecolor{currentstroke}%
\pgfsetdash{}{0pt}%
\pgfpathmoveto{\pgfqpoint{0.520798in}{1.127922in}}%
\pgfpathlineto{\pgfqpoint{2.598467in}{1.127922in}}%
\pgfusepath{stroke}%
\end{pgfscope}%
\begin{pgfscope}%
\definecolor{textcolor}{rgb}{0.150000,0.150000,0.150000}%
\pgfsetstrokecolor{textcolor}%
\pgfsetfillcolor{textcolor}%
\pgftext[x=0.214138in, y=1.089660in, left, base]{\color{textcolor}\rmfamily\fontsize{8.000000}{9.600000}\selectfont 60\%}%
\end{pgfscope}%
\begin{pgfscope}%
\pgfpathrectangle{\pgfqpoint{0.520798in}{0.442177in}}{\pgfqpoint{2.077669in}{1.142908in}}%
\pgfusepath{clip}%
\pgfsetroundcap%
\pgfsetroundjoin%
\pgfsetlinewidth{0.501875pt}%
\definecolor{currentstroke}{rgb}{0.800000,0.800000,0.800000}%
\pgfsetstrokecolor{currentstroke}%
\pgfsetdash{}{0pt}%
\pgfpathmoveto{\pgfqpoint{0.520798in}{1.356504in}}%
\pgfpathlineto{\pgfqpoint{2.598467in}{1.356504in}}%
\pgfusepath{stroke}%
\end{pgfscope}%
\begin{pgfscope}%
\definecolor{textcolor}{rgb}{0.150000,0.150000,0.150000}%
\pgfsetstrokecolor{textcolor}%
\pgfsetfillcolor{textcolor}%
\pgftext[x=0.214138in, y=1.318241in, left, base]{\color{textcolor}\rmfamily\fontsize{8.000000}{9.600000}\selectfont 80\%}%
\end{pgfscope}%
\begin{pgfscope}%
\pgfpathrectangle{\pgfqpoint{0.520798in}{0.442177in}}{\pgfqpoint{2.077669in}{1.142908in}}%
\pgfusepath{clip}%
\pgfsetroundcap%
\pgfsetroundjoin%
\pgfsetlinewidth{0.501875pt}%
\definecolor{currentstroke}{rgb}{0.800000,0.800000,0.800000}%
\pgfsetstrokecolor{currentstroke}%
\pgfsetdash{}{0pt}%
\pgfpathmoveto{\pgfqpoint{0.520798in}{1.585085in}}%
\pgfpathlineto{\pgfqpoint{2.598467in}{1.585085in}}%
\pgfusepath{stroke}%
\end{pgfscope}%
\begin{pgfscope}%
\definecolor{textcolor}{rgb}{0.150000,0.150000,0.150000}%
\pgfsetstrokecolor{textcolor}%
\pgfsetfillcolor{textcolor}%
\pgftext[x=0.155124in, y=1.546823in, left, base]{\color{textcolor}\rmfamily\fontsize{8.000000}{9.600000}\selectfont 100\%}%
\end{pgfscope}%
\begin{pgfscope}%
\definecolor{textcolor}{rgb}{0.150000,0.150000,0.150000}%
\pgfsetstrokecolor{textcolor}%
\pgfsetfillcolor{textcolor}%
\pgftext[x=0.099569in,y=1.013631in,,bottom,rotate=90.000000]{\color{textcolor}\rmfamily\fontsize{10.000000}{12.000000}\selectfont Cert. Acc.}%
\end{pgfscope}%
\begin{pgfscope}%
\pgfsetrectcap%
\pgfsetmiterjoin%
\pgfsetlinewidth{0.752812pt}%
\definecolor{currentstroke}{rgb}{0.700000,0.700000,0.700000}%
\pgfsetstrokecolor{currentstroke}%
\pgfsetdash{}{0pt}%
\pgfpathmoveto{\pgfqpoint{0.520798in}{0.442177in}}%
\pgfpathlineto{\pgfqpoint{0.520798in}{1.585085in}}%
\pgfusepath{stroke}%
\end{pgfscope}%
\begin{pgfscope}%
\pgfsetrectcap%
\pgfsetmiterjoin%
\pgfsetlinewidth{0.752812pt}%
\definecolor{currentstroke}{rgb}{0.700000,0.700000,0.700000}%
\pgfsetstrokecolor{currentstroke}%
\pgfsetdash{}{0pt}%
\pgfpathmoveto{\pgfqpoint{2.598467in}{0.442177in}}%
\pgfpathlineto{\pgfqpoint{2.598467in}{1.585085in}}%
\pgfusepath{stroke}%
\end{pgfscope}%
\begin{pgfscope}%
\pgfsetrectcap%
\pgfsetmiterjoin%
\pgfsetlinewidth{0.752812pt}%
\definecolor{currentstroke}{rgb}{0.700000,0.700000,0.700000}%
\pgfsetstrokecolor{currentstroke}%
\pgfsetdash{}{0pt}%
\pgfpathmoveto{\pgfqpoint{0.520798in}{0.442177in}}%
\pgfpathlineto{\pgfqpoint{2.598467in}{0.442177in}}%
\pgfusepath{stroke}%
\end{pgfscope}%
\begin{pgfscope}%
\pgfsetrectcap%
\pgfsetmiterjoin%
\pgfsetlinewidth{0.752812pt}%
\definecolor{currentstroke}{rgb}{0.700000,0.700000,0.700000}%
\pgfsetstrokecolor{currentstroke}%
\pgfsetdash{}{0pt}%
\pgfpathmoveto{\pgfqpoint{0.520798in}{1.585085in}}%
\pgfpathlineto{\pgfqpoint{2.598467in}{1.585085in}}%
\pgfusepath{stroke}%
\end{pgfscope}%
\begin{pgfscope}%
\pgfpathrectangle{\pgfqpoint{0.520798in}{0.442177in}}{\pgfqpoint{2.077669in}{1.142908in}}%
\pgfusepath{clip}%
\pgfsetroundcap%
\pgfsetroundjoin%
\pgfsetlinewidth{1.003750pt}%
\definecolor{currentstroke}{rgb}{0.003922,0.450980,0.698039}%
\pgfsetstrokecolor{currentstroke}%
\pgfsetdash{}{0pt}%
\pgfpathmoveto{\pgfqpoint{0.520798in}{1.401238in}}%
\pgfpathlineto{\pgfqpoint{0.562771in}{1.395125in}}%
\pgfpathlineto{\pgfqpoint{0.604744in}{1.386234in}}%
\pgfpathlineto{\pgfqpoint{0.646717in}{1.378269in}}%
\pgfpathlineto{\pgfqpoint{0.688690in}{1.370767in}}%
\pgfpathlineto{\pgfqpoint{0.730663in}{1.362987in}}%
\pgfpathlineto{\pgfqpoint{0.772637in}{1.353632in}}%
\pgfpathlineto{\pgfqpoint{0.814610in}{1.343815in}}%
\pgfpathlineto{\pgfqpoint{0.856583in}{1.334460in}}%
\pgfpathlineto{\pgfqpoint{0.898556in}{1.321772in}}%
\pgfpathlineto{\pgfqpoint{0.940529in}{1.312973in}}%
\pgfpathlineto{\pgfqpoint{0.982502in}{1.302507in}}%
\pgfpathlineto{\pgfqpoint{1.024475in}{1.292875in}}%
\pgfpathlineto{\pgfqpoint{1.066448in}{1.281575in}}%
\pgfpathlineto{\pgfqpoint{1.108422in}{1.269257in}}%
\pgfpathlineto{\pgfqpoint{1.150395in}{1.256846in}}%
\pgfpathlineto{\pgfqpoint{1.192368in}{1.243880in}}%
\pgfpathlineto{\pgfqpoint{1.234341in}{1.230080in}}%
\pgfpathlineto{\pgfqpoint{1.276314in}{1.215817in}}%
\pgfpathlineto{\pgfqpoint{1.318287in}{1.203035in}}%
\pgfpathlineto{\pgfqpoint{1.360260in}{1.189513in}}%
\pgfpathlineto{\pgfqpoint{1.402233in}{1.175713in}}%
\pgfpathlineto{\pgfqpoint{1.444207in}{1.162376in}}%
\pgfpathlineto{\pgfqpoint{1.486180in}{1.147001in}}%
\pgfpathlineto{\pgfqpoint{1.528153in}{1.130423in}}%
\pgfpathlineto{\pgfqpoint{1.570126in}{1.113566in}}%
\pgfpathlineto{\pgfqpoint{1.612099in}{1.098006in}}%
\pgfpathlineto{\pgfqpoint{1.654072in}{1.080501in}}%
\pgfpathlineto{\pgfqpoint{1.696045in}{1.062441in}}%
\pgfpathlineto{\pgfqpoint{1.738018in}{1.047529in}}%
\pgfpathlineto{\pgfqpoint{1.779992in}{1.029098in}}%
\pgfpathlineto{\pgfqpoint{1.821965in}{1.012057in}}%
\pgfpathlineto{\pgfqpoint{1.863938in}{0.994737in}}%
\pgfpathlineto{\pgfqpoint{1.905911in}{0.975009in}}%
\pgfpathlineto{\pgfqpoint{1.947884in}{0.954263in}}%
\pgfpathlineto{\pgfqpoint{1.989857in}{0.931757in}}%
\pgfpathlineto{\pgfqpoint{2.031830in}{0.906750in}}%
\pgfpathlineto{\pgfqpoint{2.073803in}{0.886096in}}%
\pgfpathlineto{\pgfqpoint{2.115776in}{0.864516in}}%
\pgfpathlineto{\pgfqpoint{2.157750in}{0.840620in}}%
\pgfpathlineto{\pgfqpoint{2.199723in}{0.819689in}}%
\pgfpathlineto{\pgfqpoint{2.241696in}{0.796719in}}%
\pgfpathlineto{\pgfqpoint{2.283669in}{0.775232in}}%
\pgfpathlineto{\pgfqpoint{2.325642in}{0.756616in}}%
\pgfpathlineto{\pgfqpoint{2.367615in}{0.739018in}}%
\pgfpathlineto{\pgfqpoint{2.409588in}{0.712437in}}%
\pgfpathlineto{\pgfqpoint{2.451561in}{0.693357in}}%
\pgfpathlineto{\pgfqpoint{2.493535in}{0.652976in}}%
\pgfpathlineto{\pgfqpoint{2.535508in}{0.442177in}}%
\pgfpathlineto{\pgfqpoint{2.577481in}{0.442177in}}%
\pgfpathlineto{\pgfqpoint{2.600134in}{0.442177in}}%
\pgfusepath{stroke}%
\end{pgfscope}%
\begin{pgfscope}%
\pgfpathrectangle{\pgfqpoint{0.520798in}{0.442177in}}{\pgfqpoint{2.077669in}{1.142908in}}%
\pgfusepath{clip}%
\pgfsetbuttcap%
\pgfsetroundjoin%
\definecolor{currentfill}{rgb}{0.003922,0.450980,0.698039}%
\pgfsetfillcolor{currentfill}%
\pgfsetfillopacity{0.500000}%
\pgfsetlinewidth{0.000000pt}%
\definecolor{currentstroke}{rgb}{0.003922,0.450980,0.698039}%
\pgfsetstrokecolor{currentstroke}%
\pgfsetstrokeopacity{0.500000}%
\pgfsetdash{}{0pt}%
\pgfpathmoveto{\pgfqpoint{0.520798in}{1.407074in}}%
\pgfpathlineto{\pgfqpoint{0.520798in}{1.395402in}}%
\pgfpathlineto{\pgfqpoint{0.562771in}{1.390024in}}%
\pgfpathlineto{\pgfqpoint{0.604744in}{1.380423in}}%
\pgfpathlineto{\pgfqpoint{0.646717in}{1.372661in}}%
\pgfpathlineto{\pgfqpoint{0.688690in}{1.364558in}}%
\pgfpathlineto{\pgfqpoint{0.730663in}{1.355201in}}%
\pgfpathlineto{\pgfqpoint{0.772637in}{1.345435in}}%
\pgfpathlineto{\pgfqpoint{0.814610in}{1.336903in}}%
\pgfpathlineto{\pgfqpoint{0.856583in}{1.327773in}}%
\pgfpathlineto{\pgfqpoint{0.898556in}{1.314204in}}%
\pgfpathlineto{\pgfqpoint{0.940529in}{1.306284in}}%
\pgfpathlineto{\pgfqpoint{0.982502in}{1.294404in}}%
\pgfpathlineto{\pgfqpoint{1.024475in}{1.286019in}}%
\pgfpathlineto{\pgfqpoint{1.066448in}{1.274530in}}%
\pgfpathlineto{\pgfqpoint{1.108422in}{1.261247in}}%
\pgfpathlineto{\pgfqpoint{1.150395in}{1.247004in}}%
\pgfpathlineto{\pgfqpoint{1.192368in}{1.233968in}}%
\pgfpathlineto{\pgfqpoint{1.234341in}{1.219320in}}%
\pgfpathlineto{\pgfqpoint{1.276314in}{1.204433in}}%
\pgfpathlineto{\pgfqpoint{1.318287in}{1.192167in}}%
\pgfpathlineto{\pgfqpoint{1.360260in}{1.176972in}}%
\pgfpathlineto{\pgfqpoint{1.402233in}{1.163249in}}%
\pgfpathlineto{\pgfqpoint{1.444207in}{1.149240in}}%
\pgfpathlineto{\pgfqpoint{1.486180in}{1.133634in}}%
\pgfpathlineto{\pgfqpoint{1.528153in}{1.115296in}}%
\pgfpathlineto{\pgfqpoint{1.570126in}{1.099000in}}%
\pgfpathlineto{\pgfqpoint{1.612099in}{1.083137in}}%
\pgfpathlineto{\pgfqpoint{1.654072in}{1.065157in}}%
\pgfpathlineto{\pgfqpoint{1.696045in}{1.046336in}}%
\pgfpathlineto{\pgfqpoint{1.738018in}{1.031444in}}%
\pgfpathlineto{\pgfqpoint{1.779992in}{1.013521in}}%
\pgfpathlineto{\pgfqpoint{1.821965in}{0.997213in}}%
\pgfpathlineto{\pgfqpoint{1.863938in}{0.979475in}}%
\pgfpathlineto{\pgfqpoint{1.905911in}{0.959976in}}%
\pgfpathlineto{\pgfqpoint{1.947884in}{0.940872in}}%
\pgfpathlineto{\pgfqpoint{1.989857in}{0.919154in}}%
\pgfpathlineto{\pgfqpoint{2.031830in}{0.894151in}}%
\pgfpathlineto{\pgfqpoint{2.073803in}{0.873395in}}%
\pgfpathlineto{\pgfqpoint{2.115776in}{0.849607in}}%
\pgfpathlineto{\pgfqpoint{2.157750in}{0.825470in}}%
\pgfpathlineto{\pgfqpoint{2.199723in}{0.806003in}}%
\pgfpathlineto{\pgfqpoint{2.241696in}{0.782825in}}%
\pgfpathlineto{\pgfqpoint{2.283669in}{0.759267in}}%
\pgfpathlineto{\pgfqpoint{2.325642in}{0.740380in}}%
\pgfpathlineto{\pgfqpoint{2.367615in}{0.725485in}}%
\pgfpathlineto{\pgfqpoint{2.409588in}{0.697779in}}%
\pgfpathlineto{\pgfqpoint{2.451561in}{0.677167in}}%
\pgfpathlineto{\pgfqpoint{2.493535in}{0.638164in}}%
\pgfpathlineto{\pgfqpoint{2.535508in}{0.442177in}}%
\pgfpathlineto{\pgfqpoint{2.577481in}{0.442177in}}%
\pgfpathlineto{\pgfqpoint{2.619454in}{0.442177in}}%
\pgfpathlineto{\pgfqpoint{2.661427in}{0.442177in}}%
\pgfpathlineto{\pgfqpoint{2.703400in}{0.442177in}}%
\pgfpathlineto{\pgfqpoint{2.745373in}{0.442177in}}%
\pgfpathlineto{\pgfqpoint{2.787346in}{0.442177in}}%
\pgfpathlineto{\pgfqpoint{2.829320in}{0.442177in}}%
\pgfpathlineto{\pgfqpoint{2.871293in}{0.442177in}}%
\pgfpathlineto{\pgfqpoint{2.913266in}{0.442177in}}%
\pgfpathlineto{\pgfqpoint{2.955239in}{0.442177in}}%
\pgfpathlineto{\pgfqpoint{2.997212in}{0.442177in}}%
\pgfpathlineto{\pgfqpoint{3.039185in}{0.442177in}}%
\pgfpathlineto{\pgfqpoint{3.081158in}{0.442177in}}%
\pgfpathlineto{\pgfqpoint{3.123131in}{0.442177in}}%
\pgfpathlineto{\pgfqpoint{3.165104in}{0.442177in}}%
\pgfpathlineto{\pgfqpoint{3.207078in}{0.442177in}}%
\pgfpathlineto{\pgfqpoint{3.249051in}{0.442177in}}%
\pgfpathlineto{\pgfqpoint{3.291024in}{0.442177in}}%
\pgfpathlineto{\pgfqpoint{3.332997in}{0.442177in}}%
\pgfpathlineto{\pgfqpoint{3.374970in}{0.442177in}}%
\pgfpathlineto{\pgfqpoint{3.416943in}{0.442177in}}%
\pgfpathlineto{\pgfqpoint{3.458916in}{0.442177in}}%
\pgfpathlineto{\pgfqpoint{3.500889in}{0.442177in}}%
\pgfpathlineto{\pgfqpoint{3.542863in}{0.442177in}}%
\pgfpathlineto{\pgfqpoint{3.584836in}{0.442177in}}%
\pgfpathlineto{\pgfqpoint{3.626809in}{0.442177in}}%
\pgfpathlineto{\pgfqpoint{3.668782in}{0.442177in}}%
\pgfpathlineto{\pgfqpoint{3.710755in}{0.442177in}}%
\pgfpathlineto{\pgfqpoint{3.752728in}{0.442177in}}%
\pgfpathlineto{\pgfqpoint{3.794701in}{0.442177in}}%
\pgfpathlineto{\pgfqpoint{3.836674in}{0.442177in}}%
\pgfpathlineto{\pgfqpoint{3.878648in}{0.442177in}}%
\pgfpathlineto{\pgfqpoint{3.920621in}{0.442177in}}%
\pgfpathlineto{\pgfqpoint{3.962594in}{0.442177in}}%
\pgfpathlineto{\pgfqpoint{4.004567in}{0.442177in}}%
\pgfpathlineto{\pgfqpoint{4.046540in}{0.442177in}}%
\pgfpathlineto{\pgfqpoint{4.088513in}{0.442177in}}%
\pgfpathlineto{\pgfqpoint{4.130486in}{0.442177in}}%
\pgfpathlineto{\pgfqpoint{4.172459in}{0.442177in}}%
\pgfpathlineto{\pgfqpoint{4.214433in}{0.442177in}}%
\pgfpathlineto{\pgfqpoint{4.256406in}{0.442177in}}%
\pgfpathlineto{\pgfqpoint{4.298379in}{0.442177in}}%
\pgfpathlineto{\pgfqpoint{4.340352in}{0.442177in}}%
\pgfpathlineto{\pgfqpoint{4.382325in}{0.442177in}}%
\pgfpathlineto{\pgfqpoint{4.424298in}{0.442177in}}%
\pgfpathlineto{\pgfqpoint{4.466271in}{0.442177in}}%
\pgfpathlineto{\pgfqpoint{4.508244in}{0.442177in}}%
\pgfpathlineto{\pgfqpoint{4.550217in}{0.442177in}}%
\pgfpathlineto{\pgfqpoint{4.592191in}{0.442177in}}%
\pgfpathlineto{\pgfqpoint{4.634164in}{0.442177in}}%
\pgfpathlineto{\pgfqpoint{4.676137in}{0.442177in}}%
\pgfpathlineto{\pgfqpoint{4.676137in}{0.442177in}}%
\pgfpathlineto{\pgfqpoint{4.676137in}{0.442177in}}%
\pgfpathlineto{\pgfqpoint{4.634164in}{0.442177in}}%
\pgfpathlineto{\pgfqpoint{4.592191in}{0.442177in}}%
\pgfpathlineto{\pgfqpoint{4.550217in}{0.442177in}}%
\pgfpathlineto{\pgfqpoint{4.508244in}{0.442177in}}%
\pgfpathlineto{\pgfqpoint{4.466271in}{0.442177in}}%
\pgfpathlineto{\pgfqpoint{4.424298in}{0.442177in}}%
\pgfpathlineto{\pgfqpoint{4.382325in}{0.442177in}}%
\pgfpathlineto{\pgfqpoint{4.340352in}{0.442177in}}%
\pgfpathlineto{\pgfqpoint{4.298379in}{0.442177in}}%
\pgfpathlineto{\pgfqpoint{4.256406in}{0.442177in}}%
\pgfpathlineto{\pgfqpoint{4.214433in}{0.442177in}}%
\pgfpathlineto{\pgfqpoint{4.172459in}{0.442177in}}%
\pgfpathlineto{\pgfqpoint{4.130486in}{0.442177in}}%
\pgfpathlineto{\pgfqpoint{4.088513in}{0.442177in}}%
\pgfpathlineto{\pgfqpoint{4.046540in}{0.442177in}}%
\pgfpathlineto{\pgfqpoint{4.004567in}{0.442177in}}%
\pgfpathlineto{\pgfqpoint{3.962594in}{0.442177in}}%
\pgfpathlineto{\pgfqpoint{3.920621in}{0.442177in}}%
\pgfpathlineto{\pgfqpoint{3.878648in}{0.442177in}}%
\pgfpathlineto{\pgfqpoint{3.836674in}{0.442177in}}%
\pgfpathlineto{\pgfqpoint{3.794701in}{0.442177in}}%
\pgfpathlineto{\pgfqpoint{3.752728in}{0.442177in}}%
\pgfpathlineto{\pgfqpoint{3.710755in}{0.442177in}}%
\pgfpathlineto{\pgfqpoint{3.668782in}{0.442177in}}%
\pgfpathlineto{\pgfqpoint{3.626809in}{0.442177in}}%
\pgfpathlineto{\pgfqpoint{3.584836in}{0.442177in}}%
\pgfpathlineto{\pgfqpoint{3.542863in}{0.442177in}}%
\pgfpathlineto{\pgfqpoint{3.500889in}{0.442177in}}%
\pgfpathlineto{\pgfqpoint{3.458916in}{0.442177in}}%
\pgfpathlineto{\pgfqpoint{3.416943in}{0.442177in}}%
\pgfpathlineto{\pgfqpoint{3.374970in}{0.442177in}}%
\pgfpathlineto{\pgfqpoint{3.332997in}{0.442177in}}%
\pgfpathlineto{\pgfqpoint{3.291024in}{0.442177in}}%
\pgfpathlineto{\pgfqpoint{3.249051in}{0.442177in}}%
\pgfpathlineto{\pgfqpoint{3.207078in}{0.442177in}}%
\pgfpathlineto{\pgfqpoint{3.165104in}{0.442177in}}%
\pgfpathlineto{\pgfqpoint{3.123131in}{0.442177in}}%
\pgfpathlineto{\pgfqpoint{3.081158in}{0.442177in}}%
\pgfpathlineto{\pgfqpoint{3.039185in}{0.442177in}}%
\pgfpathlineto{\pgfqpoint{2.997212in}{0.442177in}}%
\pgfpathlineto{\pgfqpoint{2.955239in}{0.442177in}}%
\pgfpathlineto{\pgfqpoint{2.913266in}{0.442177in}}%
\pgfpathlineto{\pgfqpoint{2.871293in}{0.442177in}}%
\pgfpathlineto{\pgfqpoint{2.829320in}{0.442177in}}%
\pgfpathlineto{\pgfqpoint{2.787346in}{0.442177in}}%
\pgfpathlineto{\pgfqpoint{2.745373in}{0.442177in}}%
\pgfpathlineto{\pgfqpoint{2.703400in}{0.442177in}}%
\pgfpathlineto{\pgfqpoint{2.661427in}{0.442177in}}%
\pgfpathlineto{\pgfqpoint{2.619454in}{0.442177in}}%
\pgfpathlineto{\pgfqpoint{2.577481in}{0.442177in}}%
\pgfpathlineto{\pgfqpoint{2.535508in}{0.442177in}}%
\pgfpathlineto{\pgfqpoint{2.493535in}{0.667787in}}%
\pgfpathlineto{\pgfqpoint{2.451561in}{0.709547in}}%
\pgfpathlineto{\pgfqpoint{2.409588in}{0.727094in}}%
\pgfpathlineto{\pgfqpoint{2.367615in}{0.752551in}}%
\pgfpathlineto{\pgfqpoint{2.325642in}{0.772851in}}%
\pgfpathlineto{\pgfqpoint{2.283669in}{0.791197in}}%
\pgfpathlineto{\pgfqpoint{2.241696in}{0.810614in}}%
\pgfpathlineto{\pgfqpoint{2.199723in}{0.833374in}}%
\pgfpathlineto{\pgfqpoint{2.157750in}{0.855770in}}%
\pgfpathlineto{\pgfqpoint{2.115776in}{0.879424in}}%
\pgfpathlineto{\pgfqpoint{2.073803in}{0.898797in}}%
\pgfpathlineto{\pgfqpoint{2.031830in}{0.919348in}}%
\pgfpathlineto{\pgfqpoint{1.989857in}{0.944359in}}%
\pgfpathlineto{\pgfqpoint{1.947884in}{0.967654in}}%
\pgfpathlineto{\pgfqpoint{1.905911in}{0.990043in}}%
\pgfpathlineto{\pgfqpoint{1.863938in}{1.009999in}}%
\pgfpathlineto{\pgfqpoint{1.821965in}{1.026900in}}%
\pgfpathlineto{\pgfqpoint{1.779992in}{1.044675in}}%
\pgfpathlineto{\pgfqpoint{1.738018in}{1.063615in}}%
\pgfpathlineto{\pgfqpoint{1.696045in}{1.078546in}}%
\pgfpathlineto{\pgfqpoint{1.654072in}{1.095845in}}%
\pgfpathlineto{\pgfqpoint{1.612099in}{1.112875in}}%
\pgfpathlineto{\pgfqpoint{1.570126in}{1.128132in}}%
\pgfpathlineto{\pgfqpoint{1.528153in}{1.145549in}}%
\pgfpathlineto{\pgfqpoint{1.486180in}{1.160368in}}%
\pgfpathlineto{\pgfqpoint{1.444207in}{1.175511in}}%
\pgfpathlineto{\pgfqpoint{1.402233in}{1.188177in}}%
\pgfpathlineto{\pgfqpoint{1.360260in}{1.202054in}}%
\pgfpathlineto{\pgfqpoint{1.318287in}{1.213904in}}%
\pgfpathlineto{\pgfqpoint{1.276314in}{1.227200in}}%
\pgfpathlineto{\pgfqpoint{1.234341in}{1.240839in}}%
\pgfpathlineto{\pgfqpoint{1.192368in}{1.253792in}}%
\pgfpathlineto{\pgfqpoint{1.150395in}{1.266689in}}%
\pgfpathlineto{\pgfqpoint{1.108422in}{1.277267in}}%
\pgfpathlineto{\pgfqpoint{1.066448in}{1.288621in}}%
\pgfpathlineto{\pgfqpoint{1.024475in}{1.299731in}}%
\pgfpathlineto{\pgfqpoint{0.982502in}{1.310610in}}%
\pgfpathlineto{\pgfqpoint{0.940529in}{1.319662in}}%
\pgfpathlineto{\pgfqpoint{0.898556in}{1.329339in}}%
\pgfpathlineto{\pgfqpoint{0.856583in}{1.341148in}}%
\pgfpathlineto{\pgfqpoint{0.814610in}{1.350727in}}%
\pgfpathlineto{\pgfqpoint{0.772637in}{1.361830in}}%
\pgfpathlineto{\pgfqpoint{0.730663in}{1.370772in}}%
\pgfpathlineto{\pgfqpoint{0.688690in}{1.376976in}}%
\pgfpathlineto{\pgfqpoint{0.646717in}{1.383877in}}%
\pgfpathlineto{\pgfqpoint{0.604744in}{1.392045in}}%
\pgfpathlineto{\pgfqpoint{0.562771in}{1.400227in}}%
\pgfpathlineto{\pgfqpoint{0.520798in}{1.407074in}}%
\pgfpathlineto{\pgfqpoint{0.520798in}{1.407074in}}%
\pgfpathclose%
\pgfusepath{fill}%
\end{pgfscope}%
\begin{pgfscope}%
\pgfpathrectangle{\pgfqpoint{0.520798in}{0.442177in}}{\pgfqpoint{2.077669in}{1.142908in}}%
\pgfusepath{clip}%
\pgfsetroundcap%
\pgfsetroundjoin%
\pgfsetlinewidth{1.003750pt}%
\definecolor{currentstroke}{rgb}{0.870588,0.560784,0.019608}%
\pgfsetstrokecolor{currentstroke}%
\pgfsetdash{}{0pt}%
\pgfpathmoveto{\pgfqpoint{0.520798in}{1.432265in}}%
\pgfpathlineto{\pgfqpoint{0.562771in}{1.425597in}}%
\pgfpathlineto{\pgfqpoint{0.604744in}{1.418280in}}%
\pgfpathlineto{\pgfqpoint{0.646717in}{1.411519in}}%
\pgfpathlineto{\pgfqpoint{0.688690in}{1.404017in}}%
\pgfpathlineto{\pgfqpoint{0.730663in}{1.396792in}}%
\pgfpathlineto{\pgfqpoint{0.772637in}{1.391328in}}%
\pgfpathlineto{\pgfqpoint{0.814610in}{1.384382in}}%
\pgfpathlineto{\pgfqpoint{0.856583in}{1.375953in}}%
\pgfpathlineto{\pgfqpoint{0.898556in}{1.367803in}}%
\pgfpathlineto{\pgfqpoint{0.940529in}{1.359375in}}%
\pgfpathlineto{\pgfqpoint{0.982502in}{1.351595in}}%
\pgfpathlineto{\pgfqpoint{1.024475in}{1.343167in}}%
\pgfpathlineto{\pgfqpoint{1.066448in}{1.333442in}}%
\pgfpathlineto{\pgfqpoint{1.108422in}{1.323068in}}%
\pgfpathlineto{\pgfqpoint{1.150395in}{1.313158in}}%
\pgfpathlineto{\pgfqpoint{1.192368in}{1.303619in}}%
\pgfpathlineto{\pgfqpoint{1.234341in}{1.293986in}}%
\pgfpathlineto{\pgfqpoint{1.276314in}{1.283520in}}%
\pgfpathlineto{\pgfqpoint{1.318287in}{1.272962in}}%
\pgfpathlineto{\pgfqpoint{1.360260in}{1.261385in}}%
\pgfpathlineto{\pgfqpoint{1.402233in}{1.249344in}}%
\pgfpathlineto{\pgfqpoint{1.444207in}{1.238323in}}%
\pgfpathlineto{\pgfqpoint{1.486180in}{1.227023in}}%
\pgfpathlineto{\pgfqpoint{1.528153in}{1.216465in}}%
\pgfpathlineto{\pgfqpoint{1.570126in}{1.204795in}}%
\pgfpathlineto{\pgfqpoint{1.612099in}{1.190995in}}%
\pgfpathlineto{\pgfqpoint{1.654072in}{1.175991in}}%
\pgfpathlineto{\pgfqpoint{1.696045in}{1.163858in}}%
\pgfpathlineto{\pgfqpoint{1.738018in}{1.149039in}}%
\pgfpathlineto{\pgfqpoint{1.779992in}{1.132275in}}%
\pgfpathlineto{\pgfqpoint{1.821965in}{1.119586in}}%
\pgfpathlineto{\pgfqpoint{1.863938in}{1.105786in}}%
\pgfpathlineto{\pgfqpoint{1.905911in}{1.092912in}}%
\pgfpathlineto{\pgfqpoint{1.947884in}{1.075870in}}%
\pgfpathlineto{\pgfqpoint{1.989857in}{1.056884in}}%
\pgfpathlineto{\pgfqpoint{2.031830in}{1.039657in}}%
\pgfpathlineto{\pgfqpoint{2.073803in}{1.023541in}}%
\pgfpathlineto{\pgfqpoint{2.115776in}{1.005573in}}%
\pgfpathlineto{\pgfqpoint{2.157750in}{0.985197in}}%
\pgfpathlineto{\pgfqpoint{2.199723in}{0.968155in}}%
\pgfpathlineto{\pgfqpoint{2.241696in}{0.945927in}}%
\pgfpathlineto{\pgfqpoint{2.283669in}{0.921754in}}%
\pgfpathlineto{\pgfqpoint{2.325642in}{0.898692in}}%
\pgfpathlineto{\pgfqpoint{2.367615in}{0.874796in}}%
\pgfpathlineto{\pgfqpoint{2.409588in}{0.845344in}}%
\pgfpathlineto{\pgfqpoint{2.451561in}{0.823486in}}%
\pgfpathlineto{\pgfqpoint{2.493535in}{0.785142in}}%
\pgfpathlineto{\pgfqpoint{2.535508in}{0.442177in}}%
\pgfpathlineto{\pgfqpoint{2.577481in}{0.442177in}}%
\pgfpathlineto{\pgfqpoint{2.600134in}{0.442177in}}%
\pgfusepath{stroke}%
\end{pgfscope}%
\begin{pgfscope}%
\pgfpathrectangle{\pgfqpoint{0.520798in}{0.442177in}}{\pgfqpoint{2.077669in}{1.142908in}}%
\pgfusepath{clip}%
\pgfsetbuttcap%
\pgfsetroundjoin%
\definecolor{currentfill}{rgb}{0.870588,0.560784,0.019608}%
\pgfsetfillcolor{currentfill}%
\pgfsetfillopacity{0.500000}%
\pgfsetlinewidth{0.000000pt}%
\definecolor{currentstroke}{rgb}{0.870588,0.560784,0.019608}%
\pgfsetstrokecolor{currentstroke}%
\pgfsetstrokeopacity{0.500000}%
\pgfsetdash{}{0pt}%
\pgfpathmoveto{\pgfqpoint{0.520798in}{1.436570in}}%
\pgfpathlineto{\pgfqpoint{0.520798in}{1.427961in}}%
\pgfpathlineto{\pgfqpoint{0.562771in}{1.420638in}}%
\pgfpathlineto{\pgfqpoint{0.604744in}{1.412908in}}%
\pgfpathlineto{\pgfqpoint{0.646717in}{1.406005in}}%
\pgfpathlineto{\pgfqpoint{0.688690in}{1.398680in}}%
\pgfpathlineto{\pgfqpoint{0.730663in}{1.391966in}}%
\pgfpathlineto{\pgfqpoint{0.772637in}{1.387037in}}%
\pgfpathlineto{\pgfqpoint{0.814610in}{1.379321in}}%
\pgfpathlineto{\pgfqpoint{0.856583in}{1.371931in}}%
\pgfpathlineto{\pgfqpoint{0.898556in}{1.364670in}}%
\pgfpathlineto{\pgfqpoint{0.940529in}{1.356323in}}%
\pgfpathlineto{\pgfqpoint{0.982502in}{1.349319in}}%
\pgfpathlineto{\pgfqpoint{1.024475in}{1.342538in}}%
\pgfpathlineto{\pgfqpoint{1.066448in}{1.331821in}}%
\pgfpathlineto{\pgfqpoint{1.108422in}{1.321847in}}%
\pgfpathlineto{\pgfqpoint{1.150395in}{1.309663in}}%
\pgfpathlineto{\pgfqpoint{1.192368in}{1.299125in}}%
\pgfpathlineto{\pgfqpoint{1.234341in}{1.291292in}}%
\pgfpathlineto{\pgfqpoint{1.276314in}{1.281591in}}%
\pgfpathlineto{\pgfqpoint{1.318287in}{1.270637in}}%
\pgfpathlineto{\pgfqpoint{1.360260in}{1.256717in}}%
\pgfpathlineto{\pgfqpoint{1.402233in}{1.243399in}}%
\pgfpathlineto{\pgfqpoint{1.444207in}{1.231998in}}%
\pgfpathlineto{\pgfqpoint{1.486180in}{1.220933in}}%
\pgfpathlineto{\pgfqpoint{1.528153in}{1.210915in}}%
\pgfpathlineto{\pgfqpoint{1.570126in}{1.200060in}}%
\pgfpathlineto{\pgfqpoint{1.612099in}{1.185888in}}%
\pgfpathlineto{\pgfqpoint{1.654072in}{1.169739in}}%
\pgfpathlineto{\pgfqpoint{1.696045in}{1.156285in}}%
\pgfpathlineto{\pgfqpoint{1.738018in}{1.141731in}}%
\pgfpathlineto{\pgfqpoint{1.779992in}{1.125613in}}%
\pgfpathlineto{\pgfqpoint{1.821965in}{1.112969in}}%
\pgfpathlineto{\pgfqpoint{1.863938in}{1.098672in}}%
\pgfpathlineto{\pgfqpoint{1.905911in}{1.086004in}}%
\pgfpathlineto{\pgfqpoint{1.947884in}{1.069363in}}%
\pgfpathlineto{\pgfqpoint{1.989857in}{1.051574in}}%
\pgfpathlineto{\pgfqpoint{2.031830in}{1.035221in}}%
\pgfpathlineto{\pgfqpoint{2.073803in}{1.018961in}}%
\pgfpathlineto{\pgfqpoint{2.115776in}{1.000782in}}%
\pgfpathlineto{\pgfqpoint{2.157750in}{0.980987in}}%
\pgfpathlineto{\pgfqpoint{2.199723in}{0.965280in}}%
\pgfpathlineto{\pgfqpoint{2.241696in}{0.944705in}}%
\pgfpathlineto{\pgfqpoint{2.283669in}{0.917554in}}%
\pgfpathlineto{\pgfqpoint{2.325642in}{0.892559in}}%
\pgfpathlineto{\pgfqpoint{2.367615in}{0.868271in}}%
\pgfpathlineto{\pgfqpoint{2.409588in}{0.837636in}}%
\pgfpathlineto{\pgfqpoint{2.451561in}{0.817180in}}%
\pgfpathlineto{\pgfqpoint{2.493535in}{0.780201in}}%
\pgfpathlineto{\pgfqpoint{2.535508in}{0.442177in}}%
\pgfpathlineto{\pgfqpoint{2.577481in}{0.442177in}}%
\pgfpathlineto{\pgfqpoint{2.619454in}{0.442177in}}%
\pgfpathlineto{\pgfqpoint{2.661427in}{0.442177in}}%
\pgfpathlineto{\pgfqpoint{2.703400in}{0.442177in}}%
\pgfpathlineto{\pgfqpoint{2.745373in}{0.442177in}}%
\pgfpathlineto{\pgfqpoint{2.787346in}{0.442177in}}%
\pgfpathlineto{\pgfqpoint{2.829320in}{0.442177in}}%
\pgfpathlineto{\pgfqpoint{2.871293in}{0.442177in}}%
\pgfpathlineto{\pgfqpoint{2.913266in}{0.442177in}}%
\pgfpathlineto{\pgfqpoint{2.955239in}{0.442177in}}%
\pgfpathlineto{\pgfqpoint{2.997212in}{0.442177in}}%
\pgfpathlineto{\pgfqpoint{3.039185in}{0.442177in}}%
\pgfpathlineto{\pgfqpoint{3.081158in}{0.442177in}}%
\pgfpathlineto{\pgfqpoint{3.123131in}{0.442177in}}%
\pgfpathlineto{\pgfqpoint{3.165104in}{0.442177in}}%
\pgfpathlineto{\pgfqpoint{3.207078in}{0.442177in}}%
\pgfpathlineto{\pgfqpoint{3.249051in}{0.442177in}}%
\pgfpathlineto{\pgfqpoint{3.291024in}{0.442177in}}%
\pgfpathlineto{\pgfqpoint{3.332997in}{0.442177in}}%
\pgfpathlineto{\pgfqpoint{3.374970in}{0.442177in}}%
\pgfpathlineto{\pgfqpoint{3.416943in}{0.442177in}}%
\pgfpathlineto{\pgfqpoint{3.458916in}{0.442177in}}%
\pgfpathlineto{\pgfqpoint{3.500889in}{0.442177in}}%
\pgfpathlineto{\pgfqpoint{3.542863in}{0.442177in}}%
\pgfpathlineto{\pgfqpoint{3.584836in}{0.442177in}}%
\pgfpathlineto{\pgfqpoint{3.626809in}{0.442177in}}%
\pgfpathlineto{\pgfqpoint{3.668782in}{0.442177in}}%
\pgfpathlineto{\pgfqpoint{3.710755in}{0.442177in}}%
\pgfpathlineto{\pgfqpoint{3.752728in}{0.442177in}}%
\pgfpathlineto{\pgfqpoint{3.794701in}{0.442177in}}%
\pgfpathlineto{\pgfqpoint{3.836674in}{0.442177in}}%
\pgfpathlineto{\pgfqpoint{3.878648in}{0.442177in}}%
\pgfpathlineto{\pgfqpoint{3.920621in}{0.442177in}}%
\pgfpathlineto{\pgfqpoint{3.962594in}{0.442177in}}%
\pgfpathlineto{\pgfqpoint{4.004567in}{0.442177in}}%
\pgfpathlineto{\pgfqpoint{4.046540in}{0.442177in}}%
\pgfpathlineto{\pgfqpoint{4.088513in}{0.442177in}}%
\pgfpathlineto{\pgfqpoint{4.130486in}{0.442177in}}%
\pgfpathlineto{\pgfqpoint{4.172459in}{0.442177in}}%
\pgfpathlineto{\pgfqpoint{4.214433in}{0.442177in}}%
\pgfpathlineto{\pgfqpoint{4.256406in}{0.442177in}}%
\pgfpathlineto{\pgfqpoint{4.298379in}{0.442177in}}%
\pgfpathlineto{\pgfqpoint{4.340352in}{0.442177in}}%
\pgfpathlineto{\pgfqpoint{4.382325in}{0.442177in}}%
\pgfpathlineto{\pgfqpoint{4.424298in}{0.442177in}}%
\pgfpathlineto{\pgfqpoint{4.466271in}{0.442177in}}%
\pgfpathlineto{\pgfqpoint{4.508244in}{0.442177in}}%
\pgfpathlineto{\pgfqpoint{4.550217in}{0.442177in}}%
\pgfpathlineto{\pgfqpoint{4.592191in}{0.442177in}}%
\pgfpathlineto{\pgfqpoint{4.634164in}{0.442177in}}%
\pgfpathlineto{\pgfqpoint{4.676137in}{0.442177in}}%
\pgfpathlineto{\pgfqpoint{4.676137in}{0.442177in}}%
\pgfpathlineto{\pgfqpoint{4.676137in}{0.442177in}}%
\pgfpathlineto{\pgfqpoint{4.634164in}{0.442177in}}%
\pgfpathlineto{\pgfqpoint{4.592191in}{0.442177in}}%
\pgfpathlineto{\pgfqpoint{4.550217in}{0.442177in}}%
\pgfpathlineto{\pgfqpoint{4.508244in}{0.442177in}}%
\pgfpathlineto{\pgfqpoint{4.466271in}{0.442177in}}%
\pgfpathlineto{\pgfqpoint{4.424298in}{0.442177in}}%
\pgfpathlineto{\pgfqpoint{4.382325in}{0.442177in}}%
\pgfpathlineto{\pgfqpoint{4.340352in}{0.442177in}}%
\pgfpathlineto{\pgfqpoint{4.298379in}{0.442177in}}%
\pgfpathlineto{\pgfqpoint{4.256406in}{0.442177in}}%
\pgfpathlineto{\pgfqpoint{4.214433in}{0.442177in}}%
\pgfpathlineto{\pgfqpoint{4.172459in}{0.442177in}}%
\pgfpathlineto{\pgfqpoint{4.130486in}{0.442177in}}%
\pgfpathlineto{\pgfqpoint{4.088513in}{0.442177in}}%
\pgfpathlineto{\pgfqpoint{4.046540in}{0.442177in}}%
\pgfpathlineto{\pgfqpoint{4.004567in}{0.442177in}}%
\pgfpathlineto{\pgfqpoint{3.962594in}{0.442177in}}%
\pgfpathlineto{\pgfqpoint{3.920621in}{0.442177in}}%
\pgfpathlineto{\pgfqpoint{3.878648in}{0.442177in}}%
\pgfpathlineto{\pgfqpoint{3.836674in}{0.442177in}}%
\pgfpathlineto{\pgfqpoint{3.794701in}{0.442177in}}%
\pgfpathlineto{\pgfqpoint{3.752728in}{0.442177in}}%
\pgfpathlineto{\pgfqpoint{3.710755in}{0.442177in}}%
\pgfpathlineto{\pgfqpoint{3.668782in}{0.442177in}}%
\pgfpathlineto{\pgfqpoint{3.626809in}{0.442177in}}%
\pgfpathlineto{\pgfqpoint{3.584836in}{0.442177in}}%
\pgfpathlineto{\pgfqpoint{3.542863in}{0.442177in}}%
\pgfpathlineto{\pgfqpoint{3.500889in}{0.442177in}}%
\pgfpathlineto{\pgfqpoint{3.458916in}{0.442177in}}%
\pgfpathlineto{\pgfqpoint{3.416943in}{0.442177in}}%
\pgfpathlineto{\pgfqpoint{3.374970in}{0.442177in}}%
\pgfpathlineto{\pgfqpoint{3.332997in}{0.442177in}}%
\pgfpathlineto{\pgfqpoint{3.291024in}{0.442177in}}%
\pgfpathlineto{\pgfqpoint{3.249051in}{0.442177in}}%
\pgfpathlineto{\pgfqpoint{3.207078in}{0.442177in}}%
\pgfpathlineto{\pgfqpoint{3.165104in}{0.442177in}}%
\pgfpathlineto{\pgfqpoint{3.123131in}{0.442177in}}%
\pgfpathlineto{\pgfqpoint{3.081158in}{0.442177in}}%
\pgfpathlineto{\pgfqpoint{3.039185in}{0.442177in}}%
\pgfpathlineto{\pgfqpoint{2.997212in}{0.442177in}}%
\pgfpathlineto{\pgfqpoint{2.955239in}{0.442177in}}%
\pgfpathlineto{\pgfqpoint{2.913266in}{0.442177in}}%
\pgfpathlineto{\pgfqpoint{2.871293in}{0.442177in}}%
\pgfpathlineto{\pgfqpoint{2.829320in}{0.442177in}}%
\pgfpathlineto{\pgfqpoint{2.787346in}{0.442177in}}%
\pgfpathlineto{\pgfqpoint{2.745373in}{0.442177in}}%
\pgfpathlineto{\pgfqpoint{2.703400in}{0.442177in}}%
\pgfpathlineto{\pgfqpoint{2.661427in}{0.442177in}}%
\pgfpathlineto{\pgfqpoint{2.619454in}{0.442177in}}%
\pgfpathlineto{\pgfqpoint{2.577481in}{0.442177in}}%
\pgfpathlineto{\pgfqpoint{2.535508in}{0.442177in}}%
\pgfpathlineto{\pgfqpoint{2.493535in}{0.790083in}}%
\pgfpathlineto{\pgfqpoint{2.451561in}{0.829792in}}%
\pgfpathlineto{\pgfqpoint{2.409588in}{0.853052in}}%
\pgfpathlineto{\pgfqpoint{2.367615in}{0.881322in}}%
\pgfpathlineto{\pgfqpoint{2.325642in}{0.904824in}}%
\pgfpathlineto{\pgfqpoint{2.283669in}{0.925953in}}%
\pgfpathlineto{\pgfqpoint{2.241696in}{0.947149in}}%
\pgfpathlineto{\pgfqpoint{2.199723in}{0.971031in}}%
\pgfpathlineto{\pgfqpoint{2.157750in}{0.989407in}}%
\pgfpathlineto{\pgfqpoint{2.115776in}{1.010364in}}%
\pgfpathlineto{\pgfqpoint{2.073803in}{1.028122in}}%
\pgfpathlineto{\pgfqpoint{2.031830in}{1.044093in}}%
\pgfpathlineto{\pgfqpoint{1.989857in}{1.062193in}}%
\pgfpathlineto{\pgfqpoint{1.947884in}{1.082377in}}%
\pgfpathlineto{\pgfqpoint{1.905911in}{1.099821in}}%
\pgfpathlineto{\pgfqpoint{1.863938in}{1.112900in}}%
\pgfpathlineto{\pgfqpoint{1.821965in}{1.126203in}}%
\pgfpathlineto{\pgfqpoint{1.779992in}{1.138937in}}%
\pgfpathlineto{\pgfqpoint{1.738018in}{1.156347in}}%
\pgfpathlineto{\pgfqpoint{1.696045in}{1.171431in}}%
\pgfpathlineto{\pgfqpoint{1.654072in}{1.182242in}}%
\pgfpathlineto{\pgfqpoint{1.612099in}{1.196101in}}%
\pgfpathlineto{\pgfqpoint{1.570126in}{1.209530in}}%
\pgfpathlineto{\pgfqpoint{1.528153in}{1.222014in}}%
\pgfpathlineto{\pgfqpoint{1.486180in}{1.233114in}}%
\pgfpathlineto{\pgfqpoint{1.444207in}{1.244648in}}%
\pgfpathlineto{\pgfqpoint{1.402233in}{1.255289in}}%
\pgfpathlineto{\pgfqpoint{1.360260in}{1.266052in}}%
\pgfpathlineto{\pgfqpoint{1.318287in}{1.275287in}}%
\pgfpathlineto{\pgfqpoint{1.276314in}{1.285450in}}%
\pgfpathlineto{\pgfqpoint{1.234341in}{1.296680in}}%
\pgfpathlineto{\pgfqpoint{1.192368in}{1.308112in}}%
\pgfpathlineto{\pgfqpoint{1.150395in}{1.316653in}}%
\pgfpathlineto{\pgfqpoint{1.108422in}{1.324290in}}%
\pgfpathlineto{\pgfqpoint{1.066448in}{1.335062in}}%
\pgfpathlineto{\pgfqpoint{1.024475in}{1.343795in}}%
\pgfpathlineto{\pgfqpoint{0.982502in}{1.353871in}}%
\pgfpathlineto{\pgfqpoint{0.940529in}{1.362427in}}%
\pgfpathlineto{\pgfqpoint{0.898556in}{1.370936in}}%
\pgfpathlineto{\pgfqpoint{0.856583in}{1.379976in}}%
\pgfpathlineto{\pgfqpoint{0.814610in}{1.389443in}}%
\pgfpathlineto{\pgfqpoint{0.772637in}{1.395619in}}%
\pgfpathlineto{\pgfqpoint{0.730663in}{1.401619in}}%
\pgfpathlineto{\pgfqpoint{0.688690in}{1.409353in}}%
\pgfpathlineto{\pgfqpoint{0.646717in}{1.417032in}}%
\pgfpathlineto{\pgfqpoint{0.604744in}{1.423652in}}%
\pgfpathlineto{\pgfqpoint{0.562771in}{1.430555in}}%
\pgfpathlineto{\pgfqpoint{0.520798in}{1.436570in}}%
\pgfpathlineto{\pgfqpoint{0.520798in}{1.436570in}}%
\pgfpathclose%
\pgfusepath{fill}%
\end{pgfscope}%
\begin{pgfscope}%
\pgfsetbuttcap%
\pgfsetmiterjoin%
\definecolor{currentfill}{rgb}{1.000000,1.000000,1.000000}%
\pgfsetfillcolor{currentfill}%
\pgfsetfillopacity{0.800000}%
\pgfsetlinewidth{1.003750pt}%
\definecolor{currentstroke}{rgb}{0.800000,0.800000,0.800000}%
\pgfsetstrokecolor{currentstroke}%
\pgfsetstrokeopacity{0.800000}%
\pgfsetdash{}{0pt}%
\pgfpathmoveto{\pgfqpoint{1.679175in}{1.186330in}}%
\pgfpathlineto{\pgfqpoint{2.542912in}{1.186330in}}%
\pgfpathlineto{\pgfqpoint{2.542912in}{1.529530in}}%
\pgfpathlineto{\pgfqpoint{1.679175in}{1.529530in}}%
\pgfpathlineto{\pgfqpoint{1.679175in}{1.186330in}}%
\pgfpathclose%
\pgfusepath{stroke,fill}%
\end{pgfscope}%
\begin{pgfscope}%
\pgfsetroundcap%
\pgfsetroundjoin%
\pgfsetlinewidth{1.003750pt}%
\definecolor{currentstroke}{rgb}{0.003922,0.450980,0.698039}%
\pgfsetstrokecolor{currentstroke}%
\pgfsetdash{}{0pt}%
\pgfpathmoveto{\pgfqpoint{1.723619in}{1.446196in}}%
\pgfpathlineto{\pgfqpoint{1.834730in}{1.446196in}}%
\pgfpathlineto{\pgfqpoint{1.945841in}{1.446196in}}%
\pgfusepath{stroke}%
\end{pgfscope}%
\begin{pgfscope}%
\definecolor{textcolor}{rgb}{0.150000,0.150000,0.150000}%
\pgfsetstrokecolor{textcolor}%
\pgfsetfillcolor{textcolor}%
\pgftext[x=2.034730in,y=1.407307in,left,base]{\color{textcolor}\rmfamily\fontsize{8.000000}{9.600000}\selectfont PointNet}%
\end{pgfscope}%
\begin{pgfscope}%
\pgfsetroundcap%
\pgfsetroundjoin%
\pgfsetlinewidth{1.003750pt}%
\definecolor{currentstroke}{rgb}{0.870588,0.560784,0.019608}%
\pgfsetstrokecolor{currentstroke}%
\pgfsetdash{}{0pt}%
\pgfpathmoveto{\pgfqpoint{1.723619in}{1.291263in}}%
\pgfpathlineto{\pgfqpoint{1.834730in}{1.291263in}}%
\pgfpathlineto{\pgfqpoint{1.945841in}{1.291263in}}%
\pgfusepath{stroke}%
\end{pgfscope}%
\begin{pgfscope}%
\definecolor{textcolor}{rgb}{0.150000,0.150000,0.150000}%
\pgfsetstrokecolor{textcolor}%
\pgfsetfillcolor{textcolor}%
\pgftext[x=2.034730in,y=1.252374in,left,base]{\color{textcolor}\rmfamily\fontsize{8.000000}{9.600000}\selectfont DGCNN}%
\end{pgfscope}%
\end{pgfpicture}%
\makeatother%
\endgroup%

%% file: figures/experiments/pointclouds/0.25.pgf
\begingroup%
\makeatletter%
\begin{pgfpicture}%
\pgfpathrectangle{\pgfpointorigin}{\pgfqpoint{2.750000in}{1.699593in}}%
\pgfusepath{use as bounding box, clip}%
\begin{pgfscope}%
\pgfsetbuttcap%
\pgfsetmiterjoin%
\definecolor{currentfill}{rgb}{1.000000,1.000000,1.000000}%
\pgfsetfillcolor{currentfill}%
\pgfsetlinewidth{0.000000pt}%
\definecolor{currentstroke}{rgb}{1.000000,1.000000,1.000000}%
\pgfsetstrokecolor{currentstroke}%
\pgfsetdash{}{0pt}%
\pgfpathmoveto{\pgfqpoint{0.000000in}{0.000000in}}%
\pgfpathlineto{\pgfqpoint{2.750000in}{0.000000in}}%
\pgfpathlineto{\pgfqpoint{2.750000in}{1.699593in}}%
\pgfpathlineto{\pgfqpoint{0.000000in}{1.699593in}}%
\pgfpathlineto{\pgfqpoint{0.000000in}{0.000000in}}%
\pgfpathclose%
\pgfusepath{fill}%
\end{pgfscope}%
\begin{pgfscope}%
\pgfsetbuttcap%
\pgfsetmiterjoin%
\definecolor{currentfill}{rgb}{1.000000,1.000000,1.000000}%
\pgfsetfillcolor{currentfill}%
\pgfsetlinewidth{0.000000pt}%
\definecolor{currentstroke}{rgb}{0.000000,0.000000,0.000000}%
\pgfsetstrokecolor{currentstroke}%
\pgfsetstrokeopacity{0.000000}%
\pgfsetdash{}{0pt}%
\pgfpathmoveto{\pgfqpoint{0.520798in}{0.442177in}}%
\pgfpathlineto{\pgfqpoint{2.673611in}{0.442177in}}%
\pgfpathlineto{\pgfqpoint{2.673611in}{1.585085in}}%
\pgfpathlineto{\pgfqpoint{0.520798in}{1.585085in}}%
\pgfpathlineto{\pgfqpoint{0.520798in}{0.442177in}}%
\pgfpathclose%
\pgfusepath{fill}%
\end{pgfscope}%
\begin{pgfscope}%
\pgfpathrectangle{\pgfqpoint{0.520798in}{0.442177in}}{\pgfqpoint{2.152813in}{1.142908in}}%
\pgfusepath{clip}%
\pgfsetroundcap%
\pgfsetroundjoin%
\pgfsetlinewidth{0.501875pt}%
\definecolor{currentstroke}{rgb}{0.800000,0.800000,0.800000}%
\pgfsetstrokecolor{currentstroke}%
\pgfsetdash{}{0pt}%
\pgfpathmoveto{\pgfqpoint{0.520798in}{0.442177in}}%
\pgfpathlineto{\pgfqpoint{0.520798in}{1.585085in}}%
\pgfusepath{stroke}%
\end{pgfscope}%
\begin{pgfscope}%
\definecolor{textcolor}{rgb}{0.150000,0.150000,0.150000}%
\pgfsetstrokecolor{textcolor}%
\pgfsetfillcolor{textcolor}%
\pgftext[x=0.520798in,y=0.351899in,,top]{\color{textcolor}\rmfamily\fontsize{8.000000}{9.600000}\selectfont \(\displaystyle {0.0}\)}%
\end{pgfscope}%
\begin{pgfscope}%
\pgfpathrectangle{\pgfqpoint{0.520798in}{0.442177in}}{\pgfqpoint{2.152813in}{1.142908in}}%
\pgfusepath{clip}%
\pgfsetroundcap%
\pgfsetroundjoin%
\pgfsetlinewidth{0.501875pt}%
\definecolor{currentstroke}{rgb}{0.800000,0.800000,0.800000}%
\pgfsetstrokecolor{currentstroke}%
\pgfsetdash{}{0pt}%
\pgfpathmoveto{\pgfqpoint{0.999201in}{0.442177in}}%
\pgfpathlineto{\pgfqpoint{0.999201in}{1.585085in}}%
\pgfusepath{stroke}%
\end{pgfscope}%
\begin{pgfscope}%
\definecolor{textcolor}{rgb}{0.150000,0.150000,0.150000}%
\pgfsetstrokecolor{textcolor}%
\pgfsetfillcolor{textcolor}%
\pgftext[x=0.999201in,y=0.351899in,,top]{\color{textcolor}\rmfamily\fontsize{8.000000}{9.600000}\selectfont \(\displaystyle {0.2}\)}%
\end{pgfscope}%
\begin{pgfscope}%
\pgfpathrectangle{\pgfqpoint{0.520798in}{0.442177in}}{\pgfqpoint{2.152813in}{1.142908in}}%
\pgfusepath{clip}%
\pgfsetroundcap%
\pgfsetroundjoin%
\pgfsetlinewidth{0.501875pt}%
\definecolor{currentstroke}{rgb}{0.800000,0.800000,0.800000}%
\pgfsetstrokecolor{currentstroke}%
\pgfsetdash{}{0pt}%
\pgfpathmoveto{\pgfqpoint{1.477604in}{0.442177in}}%
\pgfpathlineto{\pgfqpoint{1.477604in}{1.585085in}}%
\pgfusepath{stroke}%
\end{pgfscope}%
\begin{pgfscope}%
\definecolor{textcolor}{rgb}{0.150000,0.150000,0.150000}%
\pgfsetstrokecolor{textcolor}%
\pgfsetfillcolor{textcolor}%
\pgftext[x=1.477604in,y=0.351899in,,top]{\color{textcolor}\rmfamily\fontsize{8.000000}{9.600000}\selectfont \(\displaystyle {0.4}\)}%
\end{pgfscope}%
\begin{pgfscope}%
\pgfpathrectangle{\pgfqpoint{0.520798in}{0.442177in}}{\pgfqpoint{2.152813in}{1.142908in}}%
\pgfusepath{clip}%
\pgfsetroundcap%
\pgfsetroundjoin%
\pgfsetlinewidth{0.501875pt}%
\definecolor{currentstroke}{rgb}{0.800000,0.800000,0.800000}%
\pgfsetstrokecolor{currentstroke}%
\pgfsetdash{}{0pt}%
\pgfpathmoveto{\pgfqpoint{1.956007in}{0.442177in}}%
\pgfpathlineto{\pgfqpoint{1.956007in}{1.585085in}}%
\pgfusepath{stroke}%
\end{pgfscope}%
\begin{pgfscope}%
\definecolor{textcolor}{rgb}{0.150000,0.150000,0.150000}%
\pgfsetstrokecolor{textcolor}%
\pgfsetfillcolor{textcolor}%
\pgftext[x=1.956007in,y=0.351899in,,top]{\color{textcolor}\rmfamily\fontsize{8.000000}{9.600000}\selectfont \(\displaystyle {0.6}\)}%
\end{pgfscope}%
\begin{pgfscope}%
\pgfpathrectangle{\pgfqpoint{0.520798in}{0.442177in}}{\pgfqpoint{2.152813in}{1.142908in}}%
\pgfusepath{clip}%
\pgfsetroundcap%
\pgfsetroundjoin%
\pgfsetlinewidth{0.501875pt}%
\definecolor{currentstroke}{rgb}{0.800000,0.800000,0.800000}%
\pgfsetstrokecolor{currentstroke}%
\pgfsetdash{}{0pt}%
\pgfpathmoveto{\pgfqpoint{2.434410in}{0.442177in}}%
\pgfpathlineto{\pgfqpoint{2.434410in}{1.585085in}}%
\pgfusepath{stroke}%
\end{pgfscope}%
\begin{pgfscope}%
\definecolor{textcolor}{rgb}{0.150000,0.150000,0.150000}%
\pgfsetstrokecolor{textcolor}%
\pgfsetfillcolor{textcolor}%
\pgftext[x=2.434410in,y=0.351899in,,top]{\color{textcolor}\rmfamily\fontsize{8.000000}{9.600000}\selectfont \(\displaystyle {0.8}\)}%
\end{pgfscope}%
\begin{pgfscope}%
\definecolor{textcolor}{rgb}{0.150000,0.150000,0.150000}%
\pgfsetstrokecolor{textcolor}%
\pgfsetfillcolor{textcolor}%
\pgftext[x=1.597204in,y=0.198219in,,top]{\color{textcolor}\rmfamily\fontsize{10.000000}{12.000000}\selectfont Correspondence distance \(\displaystyle \epsilon\)}%
\end{pgfscope}%
\begin{pgfscope}%
\pgfpathrectangle{\pgfqpoint{0.520798in}{0.442177in}}{\pgfqpoint{2.152813in}{1.142908in}}%
\pgfusepath{clip}%
\pgfsetroundcap%
\pgfsetroundjoin%
\pgfsetlinewidth{0.501875pt}%
\definecolor{currentstroke}{rgb}{0.800000,0.800000,0.800000}%
\pgfsetstrokecolor{currentstroke}%
\pgfsetdash{}{0pt}%
\pgfpathmoveto{\pgfqpoint{0.520798in}{0.442177in}}%
\pgfpathlineto{\pgfqpoint{2.673611in}{0.442177in}}%
\pgfusepath{stroke}%
\end{pgfscope}%
\begin{pgfscope}%
\definecolor{textcolor}{rgb}{0.150000,0.150000,0.150000}%
\pgfsetstrokecolor{textcolor}%
\pgfsetfillcolor{textcolor}%
\pgftext[x=0.273151in, y=0.403915in, left, base]{\color{textcolor}\rmfamily\fontsize{8.000000}{9.600000}\selectfont 0\%}%
\end{pgfscope}%
\begin{pgfscope}%
\pgfpathrectangle{\pgfqpoint{0.520798in}{0.442177in}}{\pgfqpoint{2.152813in}{1.142908in}}%
\pgfusepath{clip}%
\pgfsetroundcap%
\pgfsetroundjoin%
\pgfsetlinewidth{0.501875pt}%
\definecolor{currentstroke}{rgb}{0.800000,0.800000,0.800000}%
\pgfsetstrokecolor{currentstroke}%
\pgfsetdash{}{0pt}%
\pgfpathmoveto{\pgfqpoint{0.520798in}{0.670758in}}%
\pgfpathlineto{\pgfqpoint{2.673611in}{0.670758in}}%
\pgfusepath{stroke}%
\end{pgfscope}%
\begin{pgfscope}%
\definecolor{textcolor}{rgb}{0.150000,0.150000,0.150000}%
\pgfsetstrokecolor{textcolor}%
\pgfsetfillcolor{textcolor}%
\pgftext[x=0.214138in, y=0.632496in, left, base]{\color{textcolor}\rmfamily\fontsize{8.000000}{9.600000}\selectfont 20\%}%
\end{pgfscope}%
\begin{pgfscope}%
\pgfpathrectangle{\pgfqpoint{0.520798in}{0.442177in}}{\pgfqpoint{2.152813in}{1.142908in}}%
\pgfusepath{clip}%
\pgfsetroundcap%
\pgfsetroundjoin%
\pgfsetlinewidth{0.501875pt}%
\definecolor{currentstroke}{rgb}{0.800000,0.800000,0.800000}%
\pgfsetstrokecolor{currentstroke}%
\pgfsetdash{}{0pt}%
\pgfpathmoveto{\pgfqpoint{0.520798in}{0.899340in}}%
\pgfpathlineto{\pgfqpoint{2.673611in}{0.899340in}}%
\pgfusepath{stroke}%
\end{pgfscope}%
\begin{pgfscope}%
\definecolor{textcolor}{rgb}{0.150000,0.150000,0.150000}%
\pgfsetstrokecolor{textcolor}%
\pgfsetfillcolor{textcolor}%
\pgftext[x=0.214138in, y=0.861078in, left, base]{\color{textcolor}\rmfamily\fontsize{8.000000}{9.600000}\selectfont 40\%}%
\end{pgfscope}%
\begin{pgfscope}%
\pgfpathrectangle{\pgfqpoint{0.520798in}{0.442177in}}{\pgfqpoint{2.152813in}{1.142908in}}%
\pgfusepath{clip}%
\pgfsetroundcap%
\pgfsetroundjoin%
\pgfsetlinewidth{0.501875pt}%
\definecolor{currentstroke}{rgb}{0.800000,0.800000,0.800000}%
\pgfsetstrokecolor{currentstroke}%
\pgfsetdash{}{0pt}%
\pgfpathmoveto{\pgfqpoint{0.520798in}{1.127922in}}%
\pgfpathlineto{\pgfqpoint{2.673611in}{1.127922in}}%
\pgfusepath{stroke}%
\end{pgfscope}%
\begin{pgfscope}%
\definecolor{textcolor}{rgb}{0.150000,0.150000,0.150000}%
\pgfsetstrokecolor{textcolor}%
\pgfsetfillcolor{textcolor}%
\pgftext[x=0.214138in, y=1.089660in, left, base]{\color{textcolor}\rmfamily\fontsize{8.000000}{9.600000}\selectfont 60\%}%
\end{pgfscope}%
\begin{pgfscope}%
\pgfpathrectangle{\pgfqpoint{0.520798in}{0.442177in}}{\pgfqpoint{2.152813in}{1.142908in}}%
\pgfusepath{clip}%
\pgfsetroundcap%
\pgfsetroundjoin%
\pgfsetlinewidth{0.501875pt}%
\definecolor{currentstroke}{rgb}{0.800000,0.800000,0.800000}%
\pgfsetstrokecolor{currentstroke}%
\pgfsetdash{}{0pt}%
\pgfpathmoveto{\pgfqpoint{0.520798in}{1.356504in}}%
\pgfpathlineto{\pgfqpoint{2.673611in}{1.356504in}}%
\pgfusepath{stroke}%
\end{pgfscope}%
\begin{pgfscope}%
\definecolor{textcolor}{rgb}{0.150000,0.150000,0.150000}%
\pgfsetstrokecolor{textcolor}%
\pgfsetfillcolor{textcolor}%
\pgftext[x=0.214138in, y=1.318241in, left, base]{\color{textcolor}\rmfamily\fontsize{8.000000}{9.600000}\selectfont 80\%}%
\end{pgfscope}%
\begin{pgfscope}%
\pgfpathrectangle{\pgfqpoint{0.520798in}{0.442177in}}{\pgfqpoint{2.152813in}{1.142908in}}%
\pgfusepath{clip}%
\pgfsetroundcap%
\pgfsetroundjoin%
\pgfsetlinewidth{0.501875pt}%
\definecolor{currentstroke}{rgb}{0.800000,0.800000,0.800000}%
\pgfsetstrokecolor{currentstroke}%
\pgfsetdash{}{0pt}%
\pgfpathmoveto{\pgfqpoint{0.520798in}{1.585085in}}%
\pgfpathlineto{\pgfqpoint{2.673611in}{1.585085in}}%
\pgfusepath{stroke}%
\end{pgfscope}%
\begin{pgfscope}%
\definecolor{textcolor}{rgb}{0.150000,0.150000,0.150000}%
\pgfsetstrokecolor{textcolor}%
\pgfsetfillcolor{textcolor}%
\pgftext[x=0.155124in, y=1.546823in, left, base]{\color{textcolor}\rmfamily\fontsize{8.000000}{9.600000}\selectfont 100\%}%
\end{pgfscope}%
\begin{pgfscope}%
\definecolor{textcolor}{rgb}{0.150000,0.150000,0.150000}%
\pgfsetstrokecolor{textcolor}%
\pgfsetfillcolor{textcolor}%
\pgftext[x=0.099569in,y=1.013631in,,bottom,rotate=90.000000]{\color{textcolor}\rmfamily\fontsize{10.000000}{12.000000}\selectfont Cert. Acc.}%
\end{pgfscope}%
\begin{pgfscope}%
\pgfsetrectcap%
\pgfsetmiterjoin%
\pgfsetlinewidth{0.752812pt}%
\definecolor{currentstroke}{rgb}{0.700000,0.700000,0.700000}%
\pgfsetstrokecolor{currentstroke}%
\pgfsetdash{}{0pt}%
\pgfpathmoveto{\pgfqpoint{0.520798in}{0.442177in}}%
\pgfpathlineto{\pgfqpoint{0.520798in}{1.585085in}}%
\pgfusepath{stroke}%
\end{pgfscope}%
\begin{pgfscope}%
\pgfsetrectcap%
\pgfsetmiterjoin%
\pgfsetlinewidth{0.752812pt}%
\definecolor{currentstroke}{rgb}{0.700000,0.700000,0.700000}%
\pgfsetstrokecolor{currentstroke}%
\pgfsetdash{}{0pt}%
\pgfpathmoveto{\pgfqpoint{2.673611in}{0.442177in}}%
\pgfpathlineto{\pgfqpoint{2.673611in}{1.585085in}}%
\pgfusepath{stroke}%
\end{pgfscope}%
\begin{pgfscope}%
\pgfsetrectcap%
\pgfsetmiterjoin%
\pgfsetlinewidth{0.752812pt}%
\definecolor{currentstroke}{rgb}{0.700000,0.700000,0.700000}%
\pgfsetstrokecolor{currentstroke}%
\pgfsetdash{}{0pt}%
\pgfpathmoveto{\pgfqpoint{0.520798in}{0.442177in}}%
\pgfpathlineto{\pgfqpoint{2.673611in}{0.442177in}}%
\pgfusepath{stroke}%
\end{pgfscope}%
\begin{pgfscope}%
\pgfsetrectcap%
\pgfsetmiterjoin%
\pgfsetlinewidth{0.752812pt}%
\definecolor{currentstroke}{rgb}{0.700000,0.700000,0.700000}%
\pgfsetstrokecolor{currentstroke}%
\pgfsetdash{}{0pt}%
\pgfpathmoveto{\pgfqpoint{0.520798in}{1.585085in}}%
\pgfpathlineto{\pgfqpoint{2.673611in}{1.585085in}}%
\pgfusepath{stroke}%
\end{pgfscope}%
\begin{pgfscope}%
\pgfpathrectangle{\pgfqpoint{0.520798in}{0.442177in}}{\pgfqpoint{2.152813in}{1.142908in}}%
\pgfusepath{clip}%
\pgfsetroundcap%
\pgfsetroundjoin%
\pgfsetlinewidth{1.003750pt}%
\definecolor{currentstroke}{rgb}{0.003922,0.450980,0.698039}%
\pgfsetstrokecolor{currentstroke}%
\pgfsetdash{}{0pt}%
\pgfpathmoveto{\pgfqpoint{0.520798in}{1.304545in}}%
\pgfpathlineto{\pgfqpoint{0.544960in}{1.295931in}}%
\pgfpathlineto{\pgfqpoint{0.569121in}{1.289726in}}%
\pgfpathlineto{\pgfqpoint{0.593283in}{1.282872in}}%
\pgfpathlineto{\pgfqpoint{0.617445in}{1.274444in}}%
\pgfpathlineto{\pgfqpoint{0.641607in}{1.265645in}}%
\pgfpathlineto{\pgfqpoint{0.665768in}{1.257680in}}%
\pgfpathlineto{\pgfqpoint{0.689930in}{1.250826in}}%
\pgfpathlineto{\pgfqpoint{0.714092in}{1.242954in}}%
\pgfpathlineto{\pgfqpoint{0.738254in}{1.231747in}}%
\pgfpathlineto{\pgfqpoint{0.762416in}{1.223411in}}%
\pgfpathlineto{\pgfqpoint{0.786577in}{1.213408in}}%
\pgfpathlineto{\pgfqpoint{0.810739in}{1.202850in}}%
\pgfpathlineto{\pgfqpoint{0.834901in}{1.192662in}}%
\pgfpathlineto{\pgfqpoint{0.859063in}{1.182474in}}%
\pgfpathlineto{\pgfqpoint{0.883224in}{1.172008in}}%
\pgfpathlineto{\pgfqpoint{0.907386in}{1.161172in}}%
\pgfpathlineto{\pgfqpoint{0.931548in}{1.152280in}}%
\pgfpathlineto{\pgfqpoint{0.955710in}{1.142092in}}%
\pgfpathlineto{\pgfqpoint{0.979871in}{1.131534in}}%
\pgfpathlineto{\pgfqpoint{1.004033in}{1.119401in}}%
\pgfpathlineto{\pgfqpoint{1.028195in}{1.107268in}}%
\pgfpathlineto{\pgfqpoint{1.052357in}{1.096617in}}%
\pgfpathlineto{\pgfqpoint{1.076518in}{1.085410in}}%
\pgfpathlineto{\pgfqpoint{1.100680in}{1.073925in}}%
\pgfpathlineto{\pgfqpoint{1.124842in}{1.062626in}}%
\pgfpathlineto{\pgfqpoint{1.149004in}{1.049659in}}%
\pgfpathlineto{\pgfqpoint{1.173166in}{1.039564in}}%
\pgfpathlineto{\pgfqpoint{1.197327in}{1.026042in}}%
\pgfpathlineto{\pgfqpoint{1.221489in}{1.013724in}}%
\pgfpathlineto{\pgfqpoint{1.245651in}{1.001128in}}%
\pgfpathlineto{\pgfqpoint{1.269813in}{0.990013in}}%
\pgfpathlineto{\pgfqpoint{1.293974in}{0.976584in}}%
\pgfpathlineto{\pgfqpoint{1.318136in}{0.963432in}}%
\pgfpathlineto{\pgfqpoint{1.342298in}{0.951299in}}%
\pgfpathlineto{\pgfqpoint{1.366460in}{0.938981in}}%
\pgfpathlineto{\pgfqpoint{1.390621in}{0.926385in}}%
\pgfpathlineto{\pgfqpoint{1.414783in}{0.910917in}}%
\pgfpathlineto{\pgfqpoint{1.438945in}{0.896932in}}%
\pgfpathlineto{\pgfqpoint{1.463107in}{0.883780in}}%
\pgfpathlineto{\pgfqpoint{1.487268in}{0.872203in}}%
\pgfpathlineto{\pgfqpoint{1.511430in}{0.859607in}}%
\pgfpathlineto{\pgfqpoint{1.535592in}{0.847937in}}%
\pgfpathlineto{\pgfqpoint{1.559754in}{0.832562in}}%
\pgfpathlineto{\pgfqpoint{1.583915in}{0.821634in}}%
\pgfpathlineto{\pgfqpoint{1.608077in}{0.809964in}}%
\pgfpathlineto{\pgfqpoint{1.632239in}{0.799220in}}%
\pgfpathlineto{\pgfqpoint{1.656401in}{0.788847in}}%
\pgfpathlineto{\pgfqpoint{1.680563in}{0.777362in}}%
\pgfpathlineto{\pgfqpoint{1.704724in}{0.767452in}}%
\pgfpathlineto{\pgfqpoint{1.728886in}{0.756152in}}%
\pgfpathlineto{\pgfqpoint{1.753048in}{0.744853in}}%
\pgfpathlineto{\pgfqpoint{1.777210in}{0.734943in}}%
\pgfpathlineto{\pgfqpoint{1.801371in}{0.727441in}}%
\pgfpathlineto{\pgfqpoint{1.825533in}{0.716234in}}%
\pgfpathlineto{\pgfqpoint{1.849695in}{0.707065in}}%
\pgfpathlineto{\pgfqpoint{1.873857in}{0.696043in}}%
\pgfpathlineto{\pgfqpoint{1.898018in}{0.685948in}}%
\pgfpathlineto{\pgfqpoint{1.922180in}{0.676038in}}%
\pgfpathlineto{\pgfqpoint{1.946342in}{0.667054in}}%
\pgfpathlineto{\pgfqpoint{1.970504in}{0.657051in}}%
\pgfpathlineto{\pgfqpoint{1.994665in}{0.647604in}}%
\pgfpathlineto{\pgfqpoint{2.018827in}{0.638064in}}%
\pgfpathlineto{\pgfqpoint{2.042989in}{0.627135in}}%
\pgfpathlineto{\pgfqpoint{2.067151in}{0.619818in}}%
\pgfpathlineto{\pgfqpoint{2.091313in}{0.610094in}}%
\pgfpathlineto{\pgfqpoint{2.115474in}{0.603147in}}%
\pgfpathlineto{\pgfqpoint{2.139636in}{0.590181in}}%
\pgfpathlineto{\pgfqpoint{2.163798in}{0.582586in}}%
\pgfpathlineto{\pgfqpoint{2.187960in}{0.570731in}}%
\pgfpathlineto{\pgfqpoint{2.212121in}{0.560635in}}%
\pgfpathlineto{\pgfqpoint{2.236283in}{0.548595in}}%
\pgfpathlineto{\pgfqpoint{2.260445in}{0.543223in}}%
\pgfpathlineto{\pgfqpoint{2.284607in}{0.527386in}}%
\pgfpathlineto{\pgfqpoint{2.308768in}{0.517846in}}%
\pgfpathlineto{\pgfqpoint{2.332930in}{0.507287in}}%
\pgfpathlineto{\pgfqpoint{2.357092in}{0.493858in}}%
\pgfpathlineto{\pgfqpoint{2.381254in}{0.493858in}}%
\pgfpathlineto{\pgfqpoint{2.405415in}{0.473389in}}%
\pgfpathlineto{\pgfqpoint{2.429577in}{0.473389in}}%
\pgfpathlineto{\pgfqpoint{2.453739in}{0.442177in}}%
\pgfpathlineto{\pgfqpoint{2.477901in}{0.442177in}}%
\pgfpathlineto{\pgfqpoint{2.502063in}{0.442177in}}%
\pgfpathlineto{\pgfqpoint{2.526224in}{0.442177in}}%
\pgfpathlineto{\pgfqpoint{2.550386in}{0.442177in}}%
\pgfpathlineto{\pgfqpoint{2.574548in}{0.442177in}}%
\pgfpathlineto{\pgfqpoint{2.598710in}{0.442177in}}%
\pgfpathlineto{\pgfqpoint{2.622871in}{0.442177in}}%
\pgfpathlineto{\pgfqpoint{2.647033in}{0.442177in}}%
\pgfpathlineto{\pgfqpoint{2.671195in}{0.442177in}}%
\pgfpathlineto{\pgfqpoint{2.675278in}{0.442177in}}%
\pgfusepath{stroke}%
\end{pgfscope}%
\begin{pgfscope}%
\pgfpathrectangle{\pgfqpoint{0.520798in}{0.442177in}}{\pgfqpoint{2.152813in}{1.142908in}}%
\pgfusepath{clip}%
\pgfsetbuttcap%
\pgfsetroundjoin%
\definecolor{currentfill}{rgb}{0.003922,0.450980,0.698039}%
\pgfsetfillcolor{currentfill}%
\pgfsetfillopacity{0.500000}%
\pgfsetlinewidth{0.000000pt}%
\definecolor{currentstroke}{rgb}{0.003922,0.450980,0.698039}%
\pgfsetstrokecolor{currentstroke}%
\pgfsetstrokeopacity{0.500000}%
\pgfsetdash{}{0pt}%
\pgfsys@defobject{currentmarker}{\pgfqpoint{0.520798in}{0.442177in}}{\pgfqpoint{2.912813in}{1.319425in}}{%
\pgfpathmoveto{\pgfqpoint{0.520798in}{1.319425in}}%
\pgfpathlineto{\pgfqpoint{0.520798in}{1.289664in}}%
\pgfpathlineto{\pgfqpoint{0.544960in}{1.283311in}}%
\pgfpathlineto{\pgfqpoint{0.569121in}{1.277305in}}%
\pgfpathlineto{\pgfqpoint{0.593283in}{1.269729in}}%
\pgfpathlineto{\pgfqpoint{0.617445in}{1.262041in}}%
\pgfpathlineto{\pgfqpoint{0.641607in}{1.253367in}}%
\pgfpathlineto{\pgfqpoint{0.665768in}{1.244664in}}%
\pgfpathlineto{\pgfqpoint{0.689930in}{1.238068in}}%
\pgfpathlineto{\pgfqpoint{0.714092in}{1.228791in}}%
\pgfpathlineto{\pgfqpoint{0.738254in}{1.216879in}}%
\pgfpathlineto{\pgfqpoint{0.762416in}{1.209481in}}%
\pgfpathlineto{\pgfqpoint{0.786577in}{1.199124in}}%
\pgfpathlineto{\pgfqpoint{0.810739in}{1.188440in}}%
\pgfpathlineto{\pgfqpoint{0.834901in}{1.178157in}}%
\pgfpathlineto{\pgfqpoint{0.859063in}{1.165998in}}%
\pgfpathlineto{\pgfqpoint{0.883224in}{1.156082in}}%
\pgfpathlineto{\pgfqpoint{0.907386in}{1.144112in}}%
\pgfpathlineto{\pgfqpoint{0.931548in}{1.134787in}}%
\pgfpathlineto{\pgfqpoint{0.955710in}{1.124280in}}%
\pgfpathlineto{\pgfqpoint{0.979871in}{1.112857in}}%
\pgfpathlineto{\pgfqpoint{1.004033in}{1.099794in}}%
\pgfpathlineto{\pgfqpoint{1.028195in}{1.087895in}}%
\pgfpathlineto{\pgfqpoint{1.052357in}{1.077495in}}%
\pgfpathlineto{\pgfqpoint{1.076518in}{1.065344in}}%
\pgfpathlineto{\pgfqpoint{1.100680in}{1.054369in}}%
\pgfpathlineto{\pgfqpoint{1.124842in}{1.044295in}}%
\pgfpathlineto{\pgfqpoint{1.149004in}{1.031330in}}%
\pgfpathlineto{\pgfqpoint{1.173166in}{1.020669in}}%
\pgfpathlineto{\pgfqpoint{1.197327in}{1.006383in}}%
\pgfpathlineto{\pgfqpoint{1.221489in}{0.993609in}}%
\pgfpathlineto{\pgfqpoint{1.245651in}{0.980644in}}%
\pgfpathlineto{\pgfqpoint{1.269813in}{0.968537in}}%
\pgfpathlineto{\pgfqpoint{1.293974in}{0.956107in}}%
\pgfpathlineto{\pgfqpoint{1.318136in}{0.943688in}}%
\pgfpathlineto{\pgfqpoint{1.342298in}{0.932262in}}%
\pgfpathlineto{\pgfqpoint{1.366460in}{0.920528in}}%
\pgfpathlineto{\pgfqpoint{1.390621in}{0.908771in}}%
\pgfpathlineto{\pgfqpoint{1.414783in}{0.893118in}}%
\pgfpathlineto{\pgfqpoint{1.438945in}{0.880871in}}%
\pgfpathlineto{\pgfqpoint{1.463107in}{0.866944in}}%
\pgfpathlineto{\pgfqpoint{1.487268in}{0.856080in}}%
\pgfpathlineto{\pgfqpoint{1.511430in}{0.844241in}}%
\pgfpathlineto{\pgfqpoint{1.535592in}{0.834365in}}%
\pgfpathlineto{\pgfqpoint{1.559754in}{0.818694in}}%
\pgfpathlineto{\pgfqpoint{1.583915in}{0.809411in}}%
\pgfpathlineto{\pgfqpoint{1.608077in}{0.797865in}}%
\pgfpathlineto{\pgfqpoint{1.632239in}{0.787494in}}%
\pgfpathlineto{\pgfqpoint{1.656401in}{0.775318in}}%
\pgfpathlineto{\pgfqpoint{1.680563in}{0.764204in}}%
\pgfpathlineto{\pgfqpoint{1.704724in}{0.753794in}}%
\pgfpathlineto{\pgfqpoint{1.728886in}{0.741505in}}%
\pgfpathlineto{\pgfqpoint{1.753048in}{0.727925in}}%
\pgfpathlineto{\pgfqpoint{1.777210in}{0.718564in}}%
\pgfpathlineto{\pgfqpoint{1.801371in}{0.710917in}}%
\pgfpathlineto{\pgfqpoint{1.825533in}{0.698162in}}%
\pgfpathlineto{\pgfqpoint{1.849695in}{0.687843in}}%
\pgfpathlineto{\pgfqpoint{1.873857in}{0.675541in}}%
\pgfpathlineto{\pgfqpoint{1.898018in}{0.664961in}}%
\pgfpathlineto{\pgfqpoint{1.922180in}{0.655243in}}%
\pgfpathlineto{\pgfqpoint{1.946342in}{0.645112in}}%
\pgfpathlineto{\pgfqpoint{1.970504in}{0.636374in}}%
\pgfpathlineto{\pgfqpoint{1.994665in}{0.627051in}}%
\pgfpathlineto{\pgfqpoint{2.018827in}{0.616666in}}%
\pgfpathlineto{\pgfqpoint{2.042989in}{0.608553in}}%
\pgfpathlineto{\pgfqpoint{2.067151in}{0.600710in}}%
\pgfpathlineto{\pgfqpoint{2.091313in}{0.592410in}}%
\pgfpathlineto{\pgfqpoint{2.115474in}{0.585144in}}%
\pgfpathlineto{\pgfqpoint{2.139636in}{0.573021in}}%
\pgfpathlineto{\pgfqpoint{2.163798in}{0.566813in}}%
\pgfpathlineto{\pgfqpoint{2.187960in}{0.556911in}}%
\pgfpathlineto{\pgfqpoint{2.212121in}{0.547945in}}%
\pgfpathlineto{\pgfqpoint{2.236283in}{0.536798in}}%
\pgfpathlineto{\pgfqpoint{2.260445in}{0.531408in}}%
\pgfpathlineto{\pgfqpoint{2.284607in}{0.516027in}}%
\pgfpathlineto{\pgfqpoint{2.308768in}{0.507198in}}%
\pgfpathlineto{\pgfqpoint{2.332930in}{0.496922in}}%
\pgfpathlineto{\pgfqpoint{2.357092in}{0.483592in}}%
\pgfpathlineto{\pgfqpoint{2.381254in}{0.483592in}}%
\pgfpathlineto{\pgfqpoint{2.405415in}{0.464658in}}%
\pgfpathlineto{\pgfqpoint{2.429577in}{0.464658in}}%
\pgfpathlineto{\pgfqpoint{2.453739in}{0.442177in}}%
\pgfpathlineto{\pgfqpoint{2.477901in}{0.442177in}}%
\pgfpathlineto{\pgfqpoint{2.502063in}{0.442177in}}%
\pgfpathlineto{\pgfqpoint{2.526224in}{0.442177in}}%
\pgfpathlineto{\pgfqpoint{2.550386in}{0.442177in}}%
\pgfpathlineto{\pgfqpoint{2.574548in}{0.442177in}}%
\pgfpathlineto{\pgfqpoint{2.598710in}{0.442177in}}%
\pgfpathlineto{\pgfqpoint{2.622871in}{0.442177in}}%
\pgfpathlineto{\pgfqpoint{2.647033in}{0.442177in}}%
\pgfpathlineto{\pgfqpoint{2.671195in}{0.442177in}}%
\pgfpathlineto{\pgfqpoint{2.695357in}{0.442177in}}%
\pgfpathlineto{\pgfqpoint{2.719518in}{0.442177in}}%
\pgfpathlineto{\pgfqpoint{2.743680in}{0.442177in}}%
\pgfpathlineto{\pgfqpoint{2.767842in}{0.442177in}}%
\pgfpathlineto{\pgfqpoint{2.792004in}{0.442177in}}%
\pgfpathlineto{\pgfqpoint{2.816165in}{0.442177in}}%
\pgfpathlineto{\pgfqpoint{2.840327in}{0.442177in}}%
\pgfpathlineto{\pgfqpoint{2.864489in}{0.442177in}}%
\pgfpathlineto{\pgfqpoint{2.888651in}{0.442177in}}%
\pgfpathlineto{\pgfqpoint{2.912813in}{0.442177in}}%
\pgfpathlineto{\pgfqpoint{2.912813in}{0.442177in}}%
\pgfpathlineto{\pgfqpoint{2.912813in}{0.442177in}}%
\pgfpathlineto{\pgfqpoint{2.888651in}{0.442177in}}%
\pgfpathlineto{\pgfqpoint{2.864489in}{0.442177in}}%
\pgfpathlineto{\pgfqpoint{2.840327in}{0.442177in}}%
\pgfpathlineto{\pgfqpoint{2.816165in}{0.442177in}}%
\pgfpathlineto{\pgfqpoint{2.792004in}{0.442177in}}%
\pgfpathlineto{\pgfqpoint{2.767842in}{0.442177in}}%
\pgfpathlineto{\pgfqpoint{2.743680in}{0.442177in}}%
\pgfpathlineto{\pgfqpoint{2.719518in}{0.442177in}}%
\pgfpathlineto{\pgfqpoint{2.695357in}{0.442177in}}%
\pgfpathlineto{\pgfqpoint{2.671195in}{0.442177in}}%
\pgfpathlineto{\pgfqpoint{2.647033in}{0.442177in}}%
\pgfpathlineto{\pgfqpoint{2.622871in}{0.442177in}}%
\pgfpathlineto{\pgfqpoint{2.598710in}{0.442177in}}%
\pgfpathlineto{\pgfqpoint{2.574548in}{0.442177in}}%
\pgfpathlineto{\pgfqpoint{2.550386in}{0.442177in}}%
\pgfpathlineto{\pgfqpoint{2.526224in}{0.442177in}}%
\pgfpathlineto{\pgfqpoint{2.502063in}{0.442177in}}%
\pgfpathlineto{\pgfqpoint{2.477901in}{0.442177in}}%
\pgfpathlineto{\pgfqpoint{2.453739in}{0.442177in}}%
\pgfpathlineto{\pgfqpoint{2.429577in}{0.482120in}}%
\pgfpathlineto{\pgfqpoint{2.405415in}{0.482120in}}%
\pgfpathlineto{\pgfqpoint{2.381254in}{0.504124in}}%
\pgfpathlineto{\pgfqpoint{2.357092in}{0.504124in}}%
\pgfpathlineto{\pgfqpoint{2.332930in}{0.517653in}}%
\pgfpathlineto{\pgfqpoint{2.308768in}{0.528493in}}%
\pgfpathlineto{\pgfqpoint{2.284607in}{0.538744in}}%
\pgfpathlineto{\pgfqpoint{2.260445in}{0.555039in}}%
\pgfpathlineto{\pgfqpoint{2.236283in}{0.560392in}}%
\pgfpathlineto{\pgfqpoint{2.212121in}{0.573326in}}%
\pgfpathlineto{\pgfqpoint{2.187960in}{0.584551in}}%
\pgfpathlineto{\pgfqpoint{2.163798in}{0.598359in}}%
\pgfpathlineto{\pgfqpoint{2.139636in}{0.607340in}}%
\pgfpathlineto{\pgfqpoint{2.115474in}{0.621151in}}%
\pgfpathlineto{\pgfqpoint{2.091313in}{0.627778in}}%
\pgfpathlineto{\pgfqpoint{2.067151in}{0.638927in}}%
\pgfpathlineto{\pgfqpoint{2.042989in}{0.645718in}}%
\pgfpathlineto{\pgfqpoint{2.018827in}{0.659463in}}%
\pgfpathlineto{\pgfqpoint{1.994665in}{0.668157in}}%
\pgfpathlineto{\pgfqpoint{1.970504in}{0.677728in}}%
\pgfpathlineto{\pgfqpoint{1.946342in}{0.688996in}}%
\pgfpathlineto{\pgfqpoint{1.922180in}{0.696833in}}%
\pgfpathlineto{\pgfqpoint{1.898018in}{0.706935in}}%
\pgfpathlineto{\pgfqpoint{1.873857in}{0.716546in}}%
\pgfpathlineto{\pgfqpoint{1.849695in}{0.726286in}}%
\pgfpathlineto{\pgfqpoint{1.825533in}{0.734306in}}%
\pgfpathlineto{\pgfqpoint{1.801371in}{0.743965in}}%
\pgfpathlineto{\pgfqpoint{1.777210in}{0.751322in}}%
\pgfpathlineto{\pgfqpoint{1.753048in}{0.761781in}}%
\pgfpathlineto{\pgfqpoint{1.728886in}{0.770800in}}%
\pgfpathlineto{\pgfqpoint{1.704724in}{0.781110in}}%
\pgfpathlineto{\pgfqpoint{1.680563in}{0.790520in}}%
\pgfpathlineto{\pgfqpoint{1.656401in}{0.802375in}}%
\pgfpathlineto{\pgfqpoint{1.632239in}{0.810946in}}%
\pgfpathlineto{\pgfqpoint{1.608077in}{0.822062in}}%
\pgfpathlineto{\pgfqpoint{1.583915in}{0.833856in}}%
\pgfpathlineto{\pgfqpoint{1.559754in}{0.846430in}}%
\pgfpathlineto{\pgfqpoint{1.535592in}{0.861509in}}%
\pgfpathlineto{\pgfqpoint{1.511430in}{0.874973in}}%
\pgfpathlineto{\pgfqpoint{1.487268in}{0.888327in}}%
\pgfpathlineto{\pgfqpoint{1.463107in}{0.900617in}}%
\pgfpathlineto{\pgfqpoint{1.438945in}{0.912993in}}%
\pgfpathlineto{\pgfqpoint{1.414783in}{0.928717in}}%
\pgfpathlineto{\pgfqpoint{1.390621in}{0.943998in}}%
\pgfpathlineto{\pgfqpoint{1.366460in}{0.957433in}}%
\pgfpathlineto{\pgfqpoint{1.342298in}{0.970336in}}%
\pgfpathlineto{\pgfqpoint{1.318136in}{0.983176in}}%
\pgfpathlineto{\pgfqpoint{1.293974in}{0.997061in}}%
\pgfpathlineto{\pgfqpoint{1.269813in}{1.011490in}}%
\pgfpathlineto{\pgfqpoint{1.245651in}{1.021611in}}%
\pgfpathlineto{\pgfqpoint{1.221489in}{1.033838in}}%
\pgfpathlineto{\pgfqpoint{1.197327in}{1.045701in}}%
\pgfpathlineto{\pgfqpoint{1.173166in}{1.058459in}}%
\pgfpathlineto{\pgfqpoint{1.149004in}{1.067989in}}%
\pgfpathlineto{\pgfqpoint{1.124842in}{1.080957in}}%
\pgfpathlineto{\pgfqpoint{1.100680in}{1.093482in}}%
\pgfpathlineto{\pgfqpoint{1.076518in}{1.105476in}}%
\pgfpathlineto{\pgfqpoint{1.052357in}{1.115738in}}%
\pgfpathlineto{\pgfqpoint{1.028195in}{1.126641in}}%
\pgfpathlineto{\pgfqpoint{1.004033in}{1.139008in}}%
\pgfpathlineto{\pgfqpoint{0.979871in}{1.150211in}}%
\pgfpathlineto{\pgfqpoint{0.955710in}{1.159905in}}%
\pgfpathlineto{\pgfqpoint{0.931548in}{1.169774in}}%
\pgfpathlineto{\pgfqpoint{0.907386in}{1.178231in}}%
\pgfpathlineto{\pgfqpoint{0.883224in}{1.187935in}}%
\pgfpathlineto{\pgfqpoint{0.859063in}{1.198950in}}%
\pgfpathlineto{\pgfqpoint{0.834901in}{1.207167in}}%
\pgfpathlineto{\pgfqpoint{0.810739in}{1.217260in}}%
\pgfpathlineto{\pgfqpoint{0.786577in}{1.227693in}}%
\pgfpathlineto{\pgfqpoint{0.762416in}{1.237341in}}%
\pgfpathlineto{\pgfqpoint{0.738254in}{1.246615in}}%
\pgfpathlineto{\pgfqpoint{0.714092in}{1.257117in}}%
\pgfpathlineto{\pgfqpoint{0.689930in}{1.263584in}}%
\pgfpathlineto{\pgfqpoint{0.665768in}{1.270696in}}%
\pgfpathlineto{\pgfqpoint{0.641607in}{1.277923in}}%
\pgfpathlineto{\pgfqpoint{0.617445in}{1.286847in}}%
\pgfpathlineto{\pgfqpoint{0.593283in}{1.296015in}}%
\pgfpathlineto{\pgfqpoint{0.569121in}{1.302146in}}%
\pgfpathlineto{\pgfqpoint{0.544960in}{1.308551in}}%
\pgfpathlineto{\pgfqpoint{0.520798in}{1.319425in}}%
\pgfpathlineto{\pgfqpoint{0.520798in}{1.319425in}}%
\pgfpathclose%
\pgfusepath{fill}%
}%
\begin{pgfscope}%
\pgfsys@transformshift{0.000000in}{0.000000in}%
\pgfsys@useobject{currentmarker}{}%
\end{pgfscope}%
\end{pgfscope}%
\begin{pgfscope}%
\pgfpathrectangle{\pgfqpoint{0.520798in}{0.442177in}}{\pgfqpoint{2.152813in}{1.142908in}}%
\pgfusepath{clip}%
\pgfsetroundcap%
\pgfsetroundjoin%
\pgfsetlinewidth{1.003750pt}%
\definecolor{currentstroke}{rgb}{0.870588,0.560784,0.019608}%
\pgfsetstrokecolor{currentstroke}%
\pgfsetdash{}{0pt}%
\pgfpathmoveto{\pgfqpoint{0.520798in}{1.355114in}}%
\pgfpathlineto{\pgfqpoint{0.544960in}{1.349465in}}%
\pgfpathlineto{\pgfqpoint{0.569121in}{1.343537in}}%
\pgfpathlineto{\pgfqpoint{0.593283in}{1.337517in}}%
\pgfpathlineto{\pgfqpoint{0.617445in}{1.331404in}}%
\pgfpathlineto{\pgfqpoint{0.641607in}{1.324550in}}%
\pgfpathlineto{\pgfqpoint{0.665768in}{1.316215in}}%
\pgfpathlineto{\pgfqpoint{0.689930in}{1.309268in}}%
\pgfpathlineto{\pgfqpoint{0.714092in}{1.303341in}}%
\pgfpathlineto{\pgfqpoint{0.738254in}{1.294542in}}%
\pgfpathlineto{\pgfqpoint{0.762416in}{1.287133in}}%
\pgfpathlineto{\pgfqpoint{0.786577in}{1.278982in}}%
\pgfpathlineto{\pgfqpoint{0.810739in}{1.273055in}}%
\pgfpathlineto{\pgfqpoint{0.834901in}{1.266479in}}%
\pgfpathlineto{\pgfqpoint{0.859063in}{1.260551in}}%
\pgfpathlineto{\pgfqpoint{0.883224in}{1.253790in}}%
\pgfpathlineto{\pgfqpoint{0.907386in}{1.246381in}}%
\pgfpathlineto{\pgfqpoint{0.931548in}{1.237211in}}%
\pgfpathlineto{\pgfqpoint{0.955710in}{1.227949in}}%
\pgfpathlineto{\pgfqpoint{0.979871in}{1.217669in}}%
\pgfpathlineto{\pgfqpoint{1.004033in}{1.208037in}}%
\pgfpathlineto{\pgfqpoint{1.028195in}{1.200720in}}%
\pgfpathlineto{\pgfqpoint{1.052357in}{1.189976in}}%
\pgfpathlineto{\pgfqpoint{1.076518in}{1.180529in}}%
\pgfpathlineto{\pgfqpoint{1.100680in}{1.170156in}}%
\pgfpathlineto{\pgfqpoint{1.124842in}{1.159690in}}%
\pgfpathlineto{\pgfqpoint{1.149004in}{1.149595in}}%
\pgfpathlineto{\pgfqpoint{1.173166in}{1.141444in}}%
\pgfpathlineto{\pgfqpoint{1.197327in}{1.131627in}}%
\pgfpathlineto{\pgfqpoint{1.221489in}{1.122272in}}%
\pgfpathlineto{\pgfqpoint{1.245651in}{1.113473in}}%
\pgfpathlineto{\pgfqpoint{1.269813in}{1.104675in}}%
\pgfpathlineto{\pgfqpoint{1.293974in}{1.094301in}}%
\pgfpathlineto{\pgfqpoint{1.318136in}{1.084391in}}%
\pgfpathlineto{\pgfqpoint{1.342298in}{1.074481in}}%
\pgfpathlineto{\pgfqpoint{1.366460in}{1.064201in}}%
\pgfpathlineto{\pgfqpoint{1.390621in}{1.053179in}}%
\pgfpathlineto{\pgfqpoint{1.414783in}{1.042991in}}%
\pgfpathlineto{\pgfqpoint{1.438945in}{1.033173in}}%
\pgfpathlineto{\pgfqpoint{1.463107in}{1.020485in}}%
\pgfpathlineto{\pgfqpoint{1.487268in}{1.010019in}}%
\pgfpathlineto{\pgfqpoint{1.511430in}{0.998442in}}%
\pgfpathlineto{\pgfqpoint{1.535592in}{0.987698in}}%
\pgfpathlineto{\pgfqpoint{1.559754in}{0.975472in}}%
\pgfpathlineto{\pgfqpoint{1.583915in}{0.964729in}}%
\pgfpathlineto{\pgfqpoint{1.608077in}{0.953429in}}%
\pgfpathlineto{\pgfqpoint{1.632239in}{0.942222in}}%
\pgfpathlineto{\pgfqpoint{1.656401in}{0.929997in}}%
\pgfpathlineto{\pgfqpoint{1.680563in}{0.917679in}}%
\pgfpathlineto{\pgfqpoint{1.704724in}{0.907027in}}%
\pgfpathlineto{\pgfqpoint{1.728886in}{0.895080in}}%
\pgfpathlineto{\pgfqpoint{1.753048in}{0.881002in}}%
\pgfpathlineto{\pgfqpoint{1.777210in}{0.867757in}}%
\pgfpathlineto{\pgfqpoint{1.801371in}{0.856087in}}%
\pgfpathlineto{\pgfqpoint{1.825533in}{0.845714in}}%
\pgfpathlineto{\pgfqpoint{1.849695in}{0.834507in}}%
\pgfpathlineto{\pgfqpoint{1.873857in}{0.823486in}}%
\pgfpathlineto{\pgfqpoint{1.898018in}{0.813946in}}%
\pgfpathlineto{\pgfqpoint{1.922180in}{0.803943in}}%
\pgfpathlineto{\pgfqpoint{1.946342in}{0.795515in}}%
\pgfpathlineto{\pgfqpoint{1.970504in}{0.785698in}}%
\pgfpathlineto{\pgfqpoint{1.994665in}{0.777825in}}%
\pgfpathlineto{\pgfqpoint{2.018827in}{0.768934in}}%
\pgfpathlineto{\pgfqpoint{2.042989in}{0.757634in}}%
\pgfpathlineto{\pgfqpoint{2.067151in}{0.749947in}}%
\pgfpathlineto{\pgfqpoint{2.091313in}{0.738462in}}%
\pgfpathlineto{\pgfqpoint{2.115474in}{0.731238in}}%
\pgfpathlineto{\pgfqpoint{2.139636in}{0.720772in}}%
\pgfpathlineto{\pgfqpoint{2.163798in}{0.713733in}}%
\pgfpathlineto{\pgfqpoint{2.187960in}{0.703175in}}%
\pgfpathlineto{\pgfqpoint{2.212121in}{0.695858in}}%
\pgfpathlineto{\pgfqpoint{2.236283in}{0.684651in}}%
\pgfpathlineto{\pgfqpoint{2.260445in}{0.678261in}}%
\pgfpathlineto{\pgfqpoint{2.284607in}{0.663905in}}%
\pgfpathlineto{\pgfqpoint{2.308768in}{0.653346in}}%
\pgfpathlineto{\pgfqpoint{2.332930in}{0.638990in}}%
\pgfpathlineto{\pgfqpoint{2.357092in}{0.622227in}}%
\pgfpathlineto{\pgfqpoint{2.381254in}{0.622227in}}%
\pgfpathlineto{\pgfqpoint{2.405415in}{0.589532in}}%
\pgfpathlineto{\pgfqpoint{2.429577in}{0.589532in}}%
\pgfpathlineto{\pgfqpoint{2.453739in}{0.442177in}}%
\pgfpathlineto{\pgfqpoint{2.477901in}{0.442177in}}%
\pgfpathlineto{\pgfqpoint{2.502063in}{0.442177in}}%
\pgfpathlineto{\pgfqpoint{2.526224in}{0.442177in}}%
\pgfpathlineto{\pgfqpoint{2.550386in}{0.442177in}}%
\pgfpathlineto{\pgfqpoint{2.574548in}{0.442177in}}%
\pgfpathlineto{\pgfqpoint{2.598710in}{0.442177in}}%
\pgfpathlineto{\pgfqpoint{2.622871in}{0.442177in}}%
\pgfpathlineto{\pgfqpoint{2.647033in}{0.442177in}}%
\pgfpathlineto{\pgfqpoint{2.671195in}{0.442177in}}%
\pgfpathlineto{\pgfqpoint{2.675278in}{0.442177in}}%
\pgfusepath{stroke}%
\end{pgfscope}%
\begin{pgfscope}%
\pgfpathrectangle{\pgfqpoint{0.520798in}{0.442177in}}{\pgfqpoint{2.152813in}{1.142908in}}%
\pgfusepath{clip}%
\pgfsetbuttcap%
\pgfsetroundjoin%
\definecolor{currentfill}{rgb}{0.870588,0.560784,0.019608}%
\pgfsetfillcolor{currentfill}%
\pgfsetfillopacity{0.500000}%
\pgfsetlinewidth{0.000000pt}%
\definecolor{currentstroke}{rgb}{0.870588,0.560784,0.019608}%
\pgfsetstrokecolor{currentstroke}%
\pgfsetstrokeopacity{0.500000}%
\pgfsetdash{}{0pt}%
\pgfsys@defobject{currentmarker}{\pgfqpoint{0.520798in}{0.442177in}}{\pgfqpoint{2.912813in}{1.362024in}}{%
\pgfpathmoveto{\pgfqpoint{0.520798in}{1.362024in}}%
\pgfpathlineto{\pgfqpoint{0.520798in}{1.348204in}}%
\pgfpathlineto{\pgfqpoint{0.544960in}{1.342300in}}%
\pgfpathlineto{\pgfqpoint{0.569121in}{1.336419in}}%
\pgfpathlineto{\pgfqpoint{0.593283in}{1.329461in}}%
\pgfpathlineto{\pgfqpoint{0.617445in}{1.323175in}}%
\pgfpathlineto{\pgfqpoint{0.641607in}{1.315157in}}%
\pgfpathlineto{\pgfqpoint{0.665768in}{1.306494in}}%
\pgfpathlineto{\pgfqpoint{0.689930in}{1.300092in}}%
\pgfpathlineto{\pgfqpoint{0.714092in}{1.294058in}}%
\pgfpathlineto{\pgfqpoint{0.738254in}{1.285743in}}%
\pgfpathlineto{\pgfqpoint{0.762416in}{1.278491in}}%
\pgfpathlineto{\pgfqpoint{0.786577in}{1.270403in}}%
\pgfpathlineto{\pgfqpoint{0.810739in}{1.263603in}}%
\pgfpathlineto{\pgfqpoint{0.834901in}{1.258241in}}%
\pgfpathlineto{\pgfqpoint{0.859063in}{1.252030in}}%
\pgfpathlineto{\pgfqpoint{0.883224in}{1.245173in}}%
\pgfpathlineto{\pgfqpoint{0.907386in}{1.238887in}}%
\pgfpathlineto{\pgfqpoint{0.931548in}{1.230422in}}%
\pgfpathlineto{\pgfqpoint{0.955710in}{1.221255in}}%
\pgfpathlineto{\pgfqpoint{0.979871in}{1.212977in}}%
\pgfpathlineto{\pgfqpoint{1.004033in}{1.203220in}}%
\pgfpathlineto{\pgfqpoint{1.028195in}{1.194472in}}%
\pgfpathlineto{\pgfqpoint{1.052357in}{1.183858in}}%
\pgfpathlineto{\pgfqpoint{1.076518in}{1.173564in}}%
\pgfpathlineto{\pgfqpoint{1.100680in}{1.163719in}}%
\pgfpathlineto{\pgfqpoint{1.124842in}{1.152572in}}%
\pgfpathlineto{\pgfqpoint{1.149004in}{1.142423in}}%
\pgfpathlineto{\pgfqpoint{1.173166in}{1.134087in}}%
\pgfpathlineto{\pgfqpoint{1.197327in}{1.125370in}}%
\pgfpathlineto{\pgfqpoint{1.221489in}{1.115952in}}%
\pgfpathlineto{\pgfqpoint{1.245651in}{1.106973in}}%
\pgfpathlineto{\pgfqpoint{1.269813in}{1.095706in}}%
\pgfpathlineto{\pgfqpoint{1.293974in}{1.085867in}}%
\pgfpathlineto{\pgfqpoint{1.318136in}{1.075865in}}%
\pgfpathlineto{\pgfqpoint{1.342298in}{1.065990in}}%
\pgfpathlineto{\pgfqpoint{1.366460in}{1.054909in}}%
\pgfpathlineto{\pgfqpoint{1.390621in}{1.044130in}}%
\pgfpathlineto{\pgfqpoint{1.414783in}{1.034454in}}%
\pgfpathlineto{\pgfqpoint{1.438945in}{1.024831in}}%
\pgfpathlineto{\pgfqpoint{1.463107in}{1.012398in}}%
\pgfpathlineto{\pgfqpoint{1.487268in}{1.001821in}}%
\pgfpathlineto{\pgfqpoint{1.511430in}{0.988664in}}%
\pgfpathlineto{\pgfqpoint{1.535592in}{0.978131in}}%
\pgfpathlineto{\pgfqpoint{1.559754in}{0.963843in}}%
\pgfpathlineto{\pgfqpoint{1.583915in}{0.954117in}}%
\pgfpathlineto{\pgfqpoint{1.608077in}{0.941170in}}%
\pgfpathlineto{\pgfqpoint{1.632239in}{0.928218in}}%
\pgfpathlineto{\pgfqpoint{1.656401in}{0.917042in}}%
\pgfpathlineto{\pgfqpoint{1.680563in}{0.904954in}}%
\pgfpathlineto{\pgfqpoint{1.704724in}{0.892886in}}%
\pgfpathlineto{\pgfqpoint{1.728886in}{0.881295in}}%
\pgfpathlineto{\pgfqpoint{1.753048in}{0.866553in}}%
\pgfpathlineto{\pgfqpoint{1.777210in}{0.853653in}}%
\pgfpathlineto{\pgfqpoint{1.801371in}{0.843021in}}%
\pgfpathlineto{\pgfqpoint{1.825533in}{0.832562in}}%
\pgfpathlineto{\pgfqpoint{1.849695in}{0.820957in}}%
\pgfpathlineto{\pgfqpoint{1.873857in}{0.810101in}}%
\pgfpathlineto{\pgfqpoint{1.898018in}{0.801553in}}%
\pgfpathlineto{\pgfqpoint{1.922180in}{0.792385in}}%
\pgfpathlineto{\pgfqpoint{1.946342in}{0.786184in}}%
\pgfpathlineto{\pgfqpoint{1.970504in}{0.777574in}}%
\pgfpathlineto{\pgfqpoint{1.994665in}{0.770559in}}%
\pgfpathlineto{\pgfqpoint{2.018827in}{0.762662in}}%
\pgfpathlineto{\pgfqpoint{2.042989in}{0.752990in}}%
\pgfpathlineto{\pgfqpoint{2.067151in}{0.745041in}}%
\pgfpathlineto{\pgfqpoint{2.091313in}{0.735678in}}%
\pgfpathlineto{\pgfqpoint{2.115474in}{0.728348in}}%
\pgfpathlineto{\pgfqpoint{2.139636in}{0.717374in}}%
\pgfpathlineto{\pgfqpoint{2.163798in}{0.710653in}}%
\pgfpathlineto{\pgfqpoint{2.187960in}{0.700326in}}%
\pgfpathlineto{\pgfqpoint{2.212121in}{0.693601in}}%
\pgfpathlineto{\pgfqpoint{2.236283in}{0.682156in}}%
\pgfpathlineto{\pgfqpoint{2.260445in}{0.675326in}}%
\pgfpathlineto{\pgfqpoint{2.284607in}{0.659748in}}%
\pgfpathlineto{\pgfqpoint{2.308768in}{0.647890in}}%
\pgfpathlineto{\pgfqpoint{2.332930in}{0.632448in}}%
\pgfpathlineto{\pgfqpoint{2.357092in}{0.614867in}}%
\pgfpathlineto{\pgfqpoint{2.381254in}{0.614867in}}%
\pgfpathlineto{\pgfqpoint{2.405415in}{0.581467in}}%
\pgfpathlineto{\pgfqpoint{2.429577in}{0.581467in}}%
\pgfpathlineto{\pgfqpoint{2.453739in}{0.442177in}}%
\pgfpathlineto{\pgfqpoint{2.477901in}{0.442177in}}%
\pgfpathlineto{\pgfqpoint{2.502063in}{0.442177in}}%
\pgfpathlineto{\pgfqpoint{2.526224in}{0.442177in}}%
\pgfpathlineto{\pgfqpoint{2.550386in}{0.442177in}}%
\pgfpathlineto{\pgfqpoint{2.574548in}{0.442177in}}%
\pgfpathlineto{\pgfqpoint{2.598710in}{0.442177in}}%
\pgfpathlineto{\pgfqpoint{2.622871in}{0.442177in}}%
\pgfpathlineto{\pgfqpoint{2.647033in}{0.442177in}}%
\pgfpathlineto{\pgfqpoint{2.671195in}{0.442177in}}%
\pgfpathlineto{\pgfqpoint{2.695357in}{0.442177in}}%
\pgfpathlineto{\pgfqpoint{2.719518in}{0.442177in}}%
\pgfpathlineto{\pgfqpoint{2.743680in}{0.442177in}}%
\pgfpathlineto{\pgfqpoint{2.767842in}{0.442177in}}%
\pgfpathlineto{\pgfqpoint{2.792004in}{0.442177in}}%
\pgfpathlineto{\pgfqpoint{2.816165in}{0.442177in}}%
\pgfpathlineto{\pgfqpoint{2.840327in}{0.442177in}}%
\pgfpathlineto{\pgfqpoint{2.864489in}{0.442177in}}%
\pgfpathlineto{\pgfqpoint{2.888651in}{0.442177in}}%
\pgfpathlineto{\pgfqpoint{2.912813in}{0.442177in}}%
\pgfpathlineto{\pgfqpoint{2.912813in}{0.442177in}}%
\pgfpathlineto{\pgfqpoint{2.912813in}{0.442177in}}%
\pgfpathlineto{\pgfqpoint{2.888651in}{0.442177in}}%
\pgfpathlineto{\pgfqpoint{2.864489in}{0.442177in}}%
\pgfpathlineto{\pgfqpoint{2.840327in}{0.442177in}}%
\pgfpathlineto{\pgfqpoint{2.816165in}{0.442177in}}%
\pgfpathlineto{\pgfqpoint{2.792004in}{0.442177in}}%
\pgfpathlineto{\pgfqpoint{2.767842in}{0.442177in}}%
\pgfpathlineto{\pgfqpoint{2.743680in}{0.442177in}}%
\pgfpathlineto{\pgfqpoint{2.719518in}{0.442177in}}%
\pgfpathlineto{\pgfqpoint{2.695357in}{0.442177in}}%
\pgfpathlineto{\pgfqpoint{2.671195in}{0.442177in}}%
\pgfpathlineto{\pgfqpoint{2.647033in}{0.442177in}}%
\pgfpathlineto{\pgfqpoint{2.622871in}{0.442177in}}%
\pgfpathlineto{\pgfqpoint{2.598710in}{0.442177in}}%
\pgfpathlineto{\pgfqpoint{2.574548in}{0.442177in}}%
\pgfpathlineto{\pgfqpoint{2.550386in}{0.442177in}}%
\pgfpathlineto{\pgfqpoint{2.526224in}{0.442177in}}%
\pgfpathlineto{\pgfqpoint{2.502063in}{0.442177in}}%
\pgfpathlineto{\pgfqpoint{2.477901in}{0.442177in}}%
\pgfpathlineto{\pgfqpoint{2.453739in}{0.442177in}}%
\pgfpathlineto{\pgfqpoint{2.429577in}{0.597598in}}%
\pgfpathlineto{\pgfqpoint{2.405415in}{0.597598in}}%
\pgfpathlineto{\pgfqpoint{2.381254in}{0.629586in}}%
\pgfpathlineto{\pgfqpoint{2.357092in}{0.629586in}}%
\pgfpathlineto{\pgfqpoint{2.332930in}{0.645533in}}%
\pgfpathlineto{\pgfqpoint{2.308768in}{0.658802in}}%
\pgfpathlineto{\pgfqpoint{2.284607in}{0.668061in}}%
\pgfpathlineto{\pgfqpoint{2.260445in}{0.681195in}}%
\pgfpathlineto{\pgfqpoint{2.236283in}{0.687147in}}%
\pgfpathlineto{\pgfqpoint{2.212121in}{0.698115in}}%
\pgfpathlineto{\pgfqpoint{2.187960in}{0.706024in}}%
\pgfpathlineto{\pgfqpoint{2.163798in}{0.716813in}}%
\pgfpathlineto{\pgfqpoint{2.139636in}{0.724170in}}%
\pgfpathlineto{\pgfqpoint{2.115474in}{0.734129in}}%
\pgfpathlineto{\pgfqpoint{2.091313in}{0.741247in}}%
\pgfpathlineto{\pgfqpoint{2.067151in}{0.754853in}}%
\pgfpathlineto{\pgfqpoint{2.042989in}{0.762278in}}%
\pgfpathlineto{\pgfqpoint{2.018827in}{0.775206in}}%
\pgfpathlineto{\pgfqpoint{1.994665in}{0.785091in}}%
\pgfpathlineto{\pgfqpoint{1.970504in}{0.793822in}}%
\pgfpathlineto{\pgfqpoint{1.946342in}{0.804846in}}%
\pgfpathlineto{\pgfqpoint{1.922180in}{0.815502in}}%
\pgfpathlineto{\pgfqpoint{1.898018in}{0.826339in}}%
\pgfpathlineto{\pgfqpoint{1.873857in}{0.836871in}}%
\pgfpathlineto{\pgfqpoint{1.849695in}{0.848058in}}%
\pgfpathlineto{\pgfqpoint{1.825533in}{0.858867in}}%
\pgfpathlineto{\pgfqpoint{1.801371in}{0.869154in}}%
\pgfpathlineto{\pgfqpoint{1.777210in}{0.881862in}}%
\pgfpathlineto{\pgfqpoint{1.753048in}{0.895450in}}%
\pgfpathlineto{\pgfqpoint{1.728886in}{0.908864in}}%
\pgfpathlineto{\pgfqpoint{1.704724in}{0.921169in}}%
\pgfpathlineto{\pgfqpoint{1.680563in}{0.930403in}}%
\pgfpathlineto{\pgfqpoint{1.656401in}{0.942952in}}%
\pgfpathlineto{\pgfqpoint{1.632239in}{0.956227in}}%
\pgfpathlineto{\pgfqpoint{1.608077in}{0.965688in}}%
\pgfpathlineto{\pgfqpoint{1.583915in}{0.975340in}}%
\pgfpathlineto{\pgfqpoint{1.559754in}{0.987102in}}%
\pgfpathlineto{\pgfqpoint{1.535592in}{0.997265in}}%
\pgfpathlineto{\pgfqpoint{1.511430in}{1.008219in}}%
\pgfpathlineto{\pgfqpoint{1.487268in}{1.018217in}}%
\pgfpathlineto{\pgfqpoint{1.463107in}{1.028572in}}%
\pgfpathlineto{\pgfqpoint{1.438945in}{1.041516in}}%
\pgfpathlineto{\pgfqpoint{1.414783in}{1.051528in}}%
\pgfpathlineto{\pgfqpoint{1.390621in}{1.062228in}}%
\pgfpathlineto{\pgfqpoint{1.366460in}{1.073492in}}%
\pgfpathlineto{\pgfqpoint{1.342298in}{1.082973in}}%
\pgfpathlineto{\pgfqpoint{1.318136in}{1.092917in}}%
\pgfpathlineto{\pgfqpoint{1.293974in}{1.102736in}}%
\pgfpathlineto{\pgfqpoint{1.269813in}{1.113643in}}%
\pgfpathlineto{\pgfqpoint{1.245651in}{1.119974in}}%
\pgfpathlineto{\pgfqpoint{1.221489in}{1.128592in}}%
\pgfpathlineto{\pgfqpoint{1.197327in}{1.137884in}}%
\pgfpathlineto{\pgfqpoint{1.173166in}{1.148801in}}%
\pgfpathlineto{\pgfqpoint{1.149004in}{1.156766in}}%
\pgfpathlineto{\pgfqpoint{1.124842in}{1.166808in}}%
\pgfpathlineto{\pgfqpoint{1.100680in}{1.176593in}}%
\pgfpathlineto{\pgfqpoint{1.076518in}{1.187494in}}%
\pgfpathlineto{\pgfqpoint{1.052357in}{1.196094in}}%
\pgfpathlineto{\pgfqpoint{1.028195in}{1.206967in}}%
\pgfpathlineto{\pgfqpoint{1.004033in}{1.212853in}}%
\pgfpathlineto{\pgfqpoint{0.979871in}{1.222361in}}%
\pgfpathlineto{\pgfqpoint{0.955710in}{1.234644in}}%
\pgfpathlineto{\pgfqpoint{0.931548in}{1.244001in}}%
\pgfpathlineto{\pgfqpoint{0.907386in}{1.253874in}}%
\pgfpathlineto{\pgfqpoint{0.883224in}{1.262407in}}%
\pgfpathlineto{\pgfqpoint{0.859063in}{1.269072in}}%
\pgfpathlineto{\pgfqpoint{0.834901in}{1.274716in}}%
\pgfpathlineto{\pgfqpoint{0.810739in}{1.282506in}}%
\pgfpathlineto{\pgfqpoint{0.786577in}{1.287561in}}%
\pgfpathlineto{\pgfqpoint{0.762416in}{1.295774in}}%
\pgfpathlineto{\pgfqpoint{0.738254in}{1.303341in}}%
\pgfpathlineto{\pgfqpoint{0.714092in}{1.312624in}}%
\pgfpathlineto{\pgfqpoint{0.689930in}{1.318444in}}%
\pgfpathlineto{\pgfqpoint{0.665768in}{1.325936in}}%
\pgfpathlineto{\pgfqpoint{0.641607in}{1.333944in}}%
\pgfpathlineto{\pgfqpoint{0.617445in}{1.339633in}}%
\pgfpathlineto{\pgfqpoint{0.593283in}{1.345573in}}%
\pgfpathlineto{\pgfqpoint{0.569121in}{1.350655in}}%
\pgfpathlineto{\pgfqpoint{0.544960in}{1.356629in}}%
\pgfpathlineto{\pgfqpoint{0.520798in}{1.362024in}}%
\pgfpathlineto{\pgfqpoint{0.520798in}{1.362024in}}%
\pgfpathclose%
\pgfusepath{fill}%
}%
\begin{pgfscope}%
\pgfsys@transformshift{0.000000in}{0.000000in}%
\pgfsys@useobject{currentmarker}{}%
\end{pgfscope}%
\end{pgfscope}%
\begin{pgfscope}%
\pgfsetbuttcap%
\pgfsetmiterjoin%
\definecolor{currentfill}{rgb}{1.000000,1.000000,1.000000}%
\pgfsetfillcolor{currentfill}%
\pgfsetfillopacity{0.800000}%
\pgfsetlinewidth{1.003750pt}%
\definecolor{currentstroke}{rgb}{0.800000,0.800000,0.800000}%
\pgfsetstrokecolor{currentstroke}%
\pgfsetstrokeopacity{0.800000}%
\pgfsetdash{}{0pt}%
\pgfpathmoveto{\pgfqpoint{1.754318in}{1.186330in}}%
\pgfpathlineto{\pgfqpoint{2.618055in}{1.186330in}}%
\pgfpathlineto{\pgfqpoint{2.618055in}{1.529530in}}%
\pgfpathlineto{\pgfqpoint{1.754318in}{1.529530in}}%
\pgfpathlineto{\pgfqpoint{1.754318in}{1.186330in}}%
\pgfpathclose%
\pgfusepath{stroke,fill}%
\end{pgfscope}%
\begin{pgfscope}%
\pgfsetroundcap%
\pgfsetroundjoin%
\pgfsetlinewidth{1.003750pt}%
\definecolor{currentstroke}{rgb}{0.003922,0.450980,0.698039}%
\pgfsetstrokecolor{currentstroke}%
\pgfsetdash{}{0pt}%
\pgfpathmoveto{\pgfqpoint{1.798763in}{1.446196in}}%
\pgfpathlineto{\pgfqpoint{1.909874in}{1.446196in}}%
\pgfpathlineto{\pgfqpoint{2.020985in}{1.446196in}}%
\pgfusepath{stroke}%
\end{pgfscope}%
\begin{pgfscope}%
\definecolor{textcolor}{rgb}{0.150000,0.150000,0.150000}%
\pgfsetstrokecolor{textcolor}%
\pgfsetfillcolor{textcolor}%
\pgftext[x=2.109874in,y=1.407307in,left,base]{\color{textcolor}\rmfamily\fontsize{8.000000}{9.600000}\selectfont PointNet}%
\end{pgfscope}%
\begin{pgfscope}%
\pgfsetroundcap%
\pgfsetroundjoin%
\pgfsetlinewidth{1.003750pt}%
\definecolor{currentstroke}{rgb}{0.870588,0.560784,0.019608}%
\pgfsetstrokecolor{currentstroke}%
\pgfsetdash{}{0pt}%
\pgfpathmoveto{\pgfqpoint{1.798763in}{1.291263in}}%
\pgfpathlineto{\pgfqpoint{1.909874in}{1.291263in}}%
\pgfpathlineto{\pgfqpoint{2.020985in}{1.291263in}}%
\pgfusepath{stroke}%
\end{pgfscope}%
\begin{pgfscope}%
\definecolor{textcolor}{rgb}{0.150000,0.150000,0.150000}%
\pgfsetstrokecolor{textcolor}%
\pgfsetfillcolor{textcolor}%
\pgftext[x=2.109874in,y=1.252374in,left,base]{\color{textcolor}\rmfamily\fontsize{8.000000}{9.600000}\selectfont DGCNN}%
\end{pgfscope}%
\end{pgfpicture}%
\makeatother%
\endgroup%

%% file: figures/experiments/graphs/piPPNP/piPPNP-Cora-B.pgf
\begingroup%
\makeatletter%
\begin{pgfpicture}%
\pgfpathrectangle{\pgfpointorigin}{\pgfqpoint{1.375000in}{1.581250in}}%
\pgfusepath{use as bounding box, clip}%
\begin{pgfscope}%
\pgfsetbuttcap%
\pgfsetmiterjoin%
\definecolor{currentfill}{rgb}{1.000000,1.000000,1.000000}%
\pgfsetfillcolor{currentfill}%
\pgfsetlinewidth{0.000000pt}%
\definecolor{currentstroke}{rgb}{1.000000,1.000000,1.000000}%
\pgfsetstrokecolor{currentstroke}%
\pgfsetdash{}{0pt}%
\pgfpathmoveto{\pgfqpoint{0.000000in}{0.000000in}}%
\pgfpathlineto{\pgfqpoint{1.375000in}{0.000000in}}%
\pgfpathlineto{\pgfqpoint{1.375000in}{1.581250in}}%
\pgfpathlineto{\pgfqpoint{0.000000in}{1.581250in}}%
\pgfpathlineto{\pgfqpoint{0.000000in}{0.000000in}}%
\pgfpathclose%
\pgfusepath{fill}%
\end{pgfscope}%
\begin{pgfscope}%
\pgfsetbuttcap%
\pgfsetmiterjoin%
\definecolor{currentfill}{rgb}{1.000000,1.000000,1.000000}%
\pgfsetfillcolor{currentfill}%
\pgfsetlinewidth{0.000000pt}%
\definecolor{currentstroke}{rgb}{0.000000,0.000000,0.000000}%
\pgfsetstrokecolor{currentstroke}%
\pgfsetstrokeopacity{0.000000}%
\pgfsetdash{}{0pt}%
\pgfpathmoveto{\pgfqpoint{0.520798in}{0.442177in}}%
\pgfpathlineto{\pgfqpoint{1.269207in}{0.442177in}}%
\pgfpathlineto{\pgfqpoint{1.269207in}{1.314061in}}%
\pgfpathlineto{\pgfqpoint{0.520798in}{1.314061in}}%
\pgfpathlineto{\pgfqpoint{0.520798in}{0.442177in}}%
\pgfpathclose%
\pgfusepath{fill}%
\end{pgfscope}%
\begin{pgfscope}%
\pgfpathrectangle{\pgfqpoint{0.520798in}{0.442177in}}{\pgfqpoint{0.748409in}{0.871884in}}%
\pgfusepath{clip}%
\pgfsetroundcap%
\pgfsetroundjoin%
\pgfsetlinewidth{0.501875pt}%
\definecolor{currentstroke}{rgb}{0.800000,0.800000,0.800000}%
\pgfsetstrokecolor{currentstroke}%
\pgfsetdash{}{0pt}%
\pgfpathmoveto{\pgfqpoint{0.520798in}{0.442177in}}%
\pgfpathlineto{\pgfqpoint{0.520798in}{1.314061in}}%
\pgfusepath{stroke}%
\end{pgfscope}%
\begin{pgfscope}%
\definecolor{textcolor}{rgb}{0.150000,0.150000,0.150000}%
\pgfsetstrokecolor{textcolor}%
\pgfsetfillcolor{textcolor}%
\pgftext[x=0.520798in,y=0.351899in,,top]{\color{textcolor}\rmfamily\fontsize{8.000000}{9.600000}\selectfont \(\displaystyle {0}\)}%
\end{pgfscope}%
\begin{pgfscope}%
\pgfpathrectangle{\pgfqpoint{0.520798in}{0.442177in}}{\pgfqpoint{0.748409in}{0.871884in}}%
\pgfusepath{clip}%
\pgfsetroundcap%
\pgfsetroundjoin%
\pgfsetlinewidth{0.501875pt}%
\definecolor{currentstroke}{rgb}{0.800000,0.800000,0.800000}%
\pgfsetstrokecolor{currentstroke}%
\pgfsetdash{}{0pt}%
\pgfpathmoveto{\pgfqpoint{0.853424in}{0.442177in}}%
\pgfpathlineto{\pgfqpoint{0.853424in}{1.314061in}}%
\pgfusepath{stroke}%
\end{pgfscope}%
\begin{pgfscope}%
\definecolor{textcolor}{rgb}{0.150000,0.150000,0.150000}%
\pgfsetstrokecolor{textcolor}%
\pgfsetfillcolor{textcolor}%
\pgftext[x=0.853424in,y=0.351899in,,top]{\color{textcolor}\rmfamily\fontsize{8.000000}{9.600000}\selectfont \(\displaystyle {4}\)}%
\end{pgfscope}%
\begin{pgfscope}%
\pgfpathrectangle{\pgfqpoint{0.520798in}{0.442177in}}{\pgfqpoint{0.748409in}{0.871884in}}%
\pgfusepath{clip}%
\pgfsetroundcap%
\pgfsetroundjoin%
\pgfsetlinewidth{0.501875pt}%
\definecolor{currentstroke}{rgb}{0.800000,0.800000,0.800000}%
\pgfsetstrokecolor{currentstroke}%
\pgfsetdash{}{0pt}%
\pgfpathmoveto{\pgfqpoint{1.269207in}{0.442177in}}%
\pgfpathlineto{\pgfqpoint{1.269207in}{1.314061in}}%
\pgfusepath{stroke}%
\end{pgfscope}%
\begin{pgfscope}%
\definecolor{textcolor}{rgb}{0.150000,0.150000,0.150000}%
\pgfsetstrokecolor{textcolor}%
\pgfsetfillcolor{textcolor}%
\pgftext[x=1.269207in,y=0.351899in,,top]{\color{textcolor}\rmfamily\fontsize{8.000000}{9.600000}\selectfont \(\displaystyle {9}\)}%
\end{pgfscope}%
\begin{pgfscope}%
\definecolor{textcolor}{rgb}{0.150000,0.150000,0.150000}%
\pgfsetstrokecolor{textcolor}%
\pgfsetfillcolor{textcolor}%
\pgftext[x=0.895002in,y=0.198219in,,top]{\color{textcolor}\rmfamily\fontsize{10.000000}{12.000000}\selectfont attack strength}%
\end{pgfscope}%
\begin{pgfscope}%
\pgfpathrectangle{\pgfqpoint{0.520798in}{0.442177in}}{\pgfqpoint{0.748409in}{0.871884in}}%
\pgfusepath{clip}%
\pgfsetroundcap%
\pgfsetroundjoin%
\pgfsetlinewidth{0.501875pt}%
\definecolor{currentstroke}{rgb}{0.800000,0.800000,0.800000}%
\pgfsetstrokecolor{currentstroke}%
\pgfsetdash{}{0pt}%
\pgfpathmoveto{\pgfqpoint{0.520798in}{0.442177in}}%
\pgfpathlineto{\pgfqpoint{1.269207in}{0.442177in}}%
\pgfusepath{stroke}%
\end{pgfscope}%
\begin{pgfscope}%
\definecolor{textcolor}{rgb}{0.150000,0.150000,0.150000}%
\pgfsetstrokecolor{textcolor}%
\pgfsetfillcolor{textcolor}%
\pgftext[x=0.273151in, y=0.403915in, left, base]{\color{textcolor}\rmfamily\fontsize{8.000000}{9.600000}\selectfont 0\%}%
\end{pgfscope}%
\begin{pgfscope}%
\pgfpathrectangle{\pgfqpoint{0.520798in}{0.442177in}}{\pgfqpoint{0.748409in}{0.871884in}}%
\pgfusepath{clip}%
\pgfsetroundcap%
\pgfsetroundjoin%
\pgfsetlinewidth{0.501875pt}%
\definecolor{currentstroke}{rgb}{0.800000,0.800000,0.800000}%
\pgfsetstrokecolor{currentstroke}%
\pgfsetdash{}{0pt}%
\pgfpathmoveto{\pgfqpoint{0.520798in}{0.660148in}}%
\pgfpathlineto{\pgfqpoint{1.269207in}{0.660148in}}%
\pgfusepath{stroke}%
\end{pgfscope}%
\begin{pgfscope}%
\definecolor{textcolor}{rgb}{0.150000,0.150000,0.150000}%
\pgfsetstrokecolor{textcolor}%
\pgfsetfillcolor{textcolor}%
\pgftext[x=0.214138in, y=0.621886in, left, base]{\color{textcolor}\rmfamily\fontsize{8.000000}{9.600000}\selectfont 25\%}%
\end{pgfscope}%
\begin{pgfscope}%
\pgfpathrectangle{\pgfqpoint{0.520798in}{0.442177in}}{\pgfqpoint{0.748409in}{0.871884in}}%
\pgfusepath{clip}%
\pgfsetroundcap%
\pgfsetroundjoin%
\pgfsetlinewidth{0.501875pt}%
\definecolor{currentstroke}{rgb}{0.800000,0.800000,0.800000}%
\pgfsetstrokecolor{currentstroke}%
\pgfsetdash{}{0pt}%
\pgfpathmoveto{\pgfqpoint{0.520798in}{0.878119in}}%
\pgfpathlineto{\pgfqpoint{1.269207in}{0.878119in}}%
\pgfusepath{stroke}%
\end{pgfscope}%
\begin{pgfscope}%
\definecolor{textcolor}{rgb}{0.150000,0.150000,0.150000}%
\pgfsetstrokecolor{textcolor}%
\pgfsetfillcolor{textcolor}%
\pgftext[x=0.214138in, y=0.839857in, left, base]{\color{textcolor}\rmfamily\fontsize{8.000000}{9.600000}\selectfont 50\%}%
\end{pgfscope}%
\begin{pgfscope}%
\pgfpathrectangle{\pgfqpoint{0.520798in}{0.442177in}}{\pgfqpoint{0.748409in}{0.871884in}}%
\pgfusepath{clip}%
\pgfsetroundcap%
\pgfsetroundjoin%
\pgfsetlinewidth{0.501875pt}%
\definecolor{currentstroke}{rgb}{0.800000,0.800000,0.800000}%
\pgfsetstrokecolor{currentstroke}%
\pgfsetdash{}{0pt}%
\pgfpathmoveto{\pgfqpoint{0.520798in}{1.096090in}}%
\pgfpathlineto{\pgfqpoint{1.269207in}{1.096090in}}%
\pgfusepath{stroke}%
\end{pgfscope}%
\begin{pgfscope}%
\definecolor{textcolor}{rgb}{0.150000,0.150000,0.150000}%
\pgfsetstrokecolor{textcolor}%
\pgfsetfillcolor{textcolor}%
\pgftext[x=0.214138in, y=1.057828in, left, base]{\color{textcolor}\rmfamily\fontsize{8.000000}{9.600000}\selectfont 75\%}%
\end{pgfscope}%
\begin{pgfscope}%
\pgfpathrectangle{\pgfqpoint{0.520798in}{0.442177in}}{\pgfqpoint{0.748409in}{0.871884in}}%
\pgfusepath{clip}%
\pgfsetroundcap%
\pgfsetroundjoin%
\pgfsetlinewidth{0.501875pt}%
\definecolor{currentstroke}{rgb}{0.800000,0.800000,0.800000}%
\pgfsetstrokecolor{currentstroke}%
\pgfsetdash{}{0pt}%
\pgfpathmoveto{\pgfqpoint{0.520798in}{1.314061in}}%
\pgfpathlineto{\pgfqpoint{1.269207in}{1.314061in}}%
\pgfusepath{stroke}%
\end{pgfscope}%
\begin{pgfscope}%
\definecolor{textcolor}{rgb}{0.150000,0.150000,0.150000}%
\pgfsetstrokecolor{textcolor}%
\pgfsetfillcolor{textcolor}%
\pgftext[x=0.155124in, y=1.275799in, left, base]{\color{textcolor}\rmfamily\fontsize{8.000000}{9.600000}\selectfont 100\%}%
\end{pgfscope}%
\begin{pgfscope}%
\definecolor{textcolor}{rgb}{0.150000,0.150000,0.150000}%
\pgfsetstrokecolor{textcolor}%
\pgfsetfillcolor{textcolor}%
\pgftext[x=0.099569in,y=0.878119in,,bottom,rotate=90.000000]{\color{textcolor}\rmfamily\fontsize{10.000000}{12.000000}\selectfont Cert. Acc.}%
\end{pgfscope}%
\begin{pgfscope}%
\pgfsetrectcap%
\pgfsetmiterjoin%
\pgfsetlinewidth{0.752812pt}%
\definecolor{currentstroke}{rgb}{0.700000,0.700000,0.700000}%
\pgfsetstrokecolor{currentstroke}%
\pgfsetdash{}{0pt}%
\pgfpathmoveto{\pgfqpoint{0.520798in}{0.442177in}}%
\pgfpathlineto{\pgfqpoint{0.520798in}{1.314061in}}%
\pgfusepath{stroke}%
\end{pgfscope}%
\begin{pgfscope}%
\pgfsetrectcap%
\pgfsetmiterjoin%
\pgfsetlinewidth{0.752812pt}%
\definecolor{currentstroke}{rgb}{0.700000,0.700000,0.700000}%
\pgfsetstrokecolor{currentstroke}%
\pgfsetdash{}{0pt}%
\pgfpathmoveto{\pgfqpoint{1.269207in}{0.442177in}}%
\pgfpathlineto{\pgfqpoint{1.269207in}{1.314061in}}%
\pgfusepath{stroke}%
\end{pgfscope}%
\begin{pgfscope}%
\pgfsetrectcap%
\pgfsetmiterjoin%
\pgfsetlinewidth{0.752812pt}%
\definecolor{currentstroke}{rgb}{0.700000,0.700000,0.700000}%
\pgfsetstrokecolor{currentstroke}%
\pgfsetdash{}{0pt}%
\pgfpathmoveto{\pgfqpoint{0.520798in}{0.442177in}}%
\pgfpathlineto{\pgfqpoint{1.269207in}{0.442177in}}%
\pgfusepath{stroke}%
\end{pgfscope}%
\begin{pgfscope}%
\pgfsetrectcap%
\pgfsetmiterjoin%
\pgfsetlinewidth{0.752812pt}%
\definecolor{currentstroke}{rgb}{0.700000,0.700000,0.700000}%
\pgfsetstrokecolor{currentstroke}%
\pgfsetdash{}{0pt}%
\pgfpathmoveto{\pgfqpoint{0.520798in}{1.314061in}}%
\pgfpathlineto{\pgfqpoint{1.269207in}{1.314061in}}%
\pgfusepath{stroke}%
\end{pgfscope}%
\begin{pgfscope}%
\pgfpathrectangle{\pgfqpoint{0.520798in}{0.442177in}}{\pgfqpoint{0.748409in}{0.871884in}}%
\pgfusepath{clip}%
\pgfsetroundcap%
\pgfsetroundjoin%
\pgfsetlinewidth{1.003750pt}%
\definecolor{currentstroke}{rgb}{0.003922,0.450980,0.698039}%
\pgfsetstrokecolor{currentstroke}%
\pgfsetdash{}{0pt}%
\pgfpathmoveto{\pgfqpoint{0.520798in}{0.935019in}}%
\pgfpathlineto{\pgfqpoint{0.562376in}{0.935019in}}%
\pgfpathlineto{\pgfqpoint{0.562376in}{0.897929in}}%
\pgfpathlineto{\pgfqpoint{0.645533in}{0.897929in}}%
\pgfpathlineto{\pgfqpoint{0.645533in}{0.853564in}}%
\pgfpathlineto{\pgfqpoint{0.728689in}{0.853564in}}%
\pgfpathlineto{\pgfqpoint{0.728689in}{0.799551in}}%
\pgfpathlineto{\pgfqpoint{0.811846in}{0.799551in}}%
\pgfpathlineto{\pgfqpoint{0.811846in}{0.723790in}}%
\pgfpathlineto{\pgfqpoint{0.895002in}{0.723790in}}%
\pgfpathlineto{\pgfqpoint{0.895002in}{0.627546in}}%
\pgfpathlineto{\pgfqpoint{0.978159in}{0.627546in}}%
\pgfpathlineto{\pgfqpoint{0.978159in}{0.525609in}}%
\pgfpathlineto{\pgfqpoint{1.061316in}{0.525609in}}%
\pgfpathlineto{\pgfqpoint{1.061316in}{0.464241in}}%
\pgfpathlineto{\pgfqpoint{1.144472in}{0.464241in}}%
\pgfpathlineto{\pgfqpoint{1.144472in}{0.447238in}}%
\pgfpathlineto{\pgfqpoint{1.227629in}{0.447238in}}%
\pgfpathlineto{\pgfqpoint{1.227629in}{0.442177in}}%
\pgfpathlineto{\pgfqpoint{1.270874in}{0.442177in}}%
\pgfusepath{stroke}%
\end{pgfscope}%
\begin{pgfscope}%
\pgfpathrectangle{\pgfqpoint{0.520798in}{0.442177in}}{\pgfqpoint{0.748409in}{0.871884in}}%
\pgfusepath{clip}%
\pgfsetbuttcap%
\pgfsetroundjoin%
\definecolor{currentfill}{rgb}{0.003922,0.450980,0.698039}%
\pgfsetfillcolor{currentfill}%
\pgfsetfillopacity{0.500000}%
\pgfsetlinewidth{0.000000pt}%
\definecolor{currentstroke}{rgb}{0.003922,0.450980,0.698039}%
\pgfsetstrokecolor{currentstroke}%
\pgfsetstrokeopacity{0.500000}%
\pgfsetdash{}{0pt}%
\pgfpathmoveto{\pgfqpoint{0.520798in}{0.948913in}}%
\pgfpathlineto{\pgfqpoint{0.520798in}{0.921125in}}%
\pgfpathlineto{\pgfqpoint{0.562376in}{0.921125in}}%
\pgfpathlineto{\pgfqpoint{0.562376in}{0.883499in}}%
\pgfpathlineto{\pgfqpoint{0.645533in}{0.883499in}}%
\pgfpathlineto{\pgfqpoint{0.645533in}{0.839560in}}%
\pgfpathlineto{\pgfqpoint{0.728689in}{0.839560in}}%
\pgfpathlineto{\pgfqpoint{0.728689in}{0.788147in}}%
\pgfpathlineto{\pgfqpoint{0.811846in}{0.788147in}}%
\pgfpathlineto{\pgfqpoint{0.811846in}{0.712298in}}%
\pgfpathlineto{\pgfqpoint{0.895002in}{0.712298in}}%
\pgfpathlineto{\pgfqpoint{0.895002in}{0.618685in}}%
\pgfpathlineto{\pgfqpoint{0.978159in}{0.618685in}}%
\pgfpathlineto{\pgfqpoint{0.978159in}{0.519540in}}%
\pgfpathlineto{\pgfqpoint{1.061316in}{0.519540in}}%
\pgfpathlineto{\pgfqpoint{1.061316in}{0.463139in}}%
\pgfpathlineto{\pgfqpoint{1.144472in}{0.463139in}}%
\pgfpathlineto{\pgfqpoint{1.144472in}{0.445024in}}%
\pgfpathlineto{\pgfqpoint{1.227629in}{0.445024in}}%
\pgfpathlineto{\pgfqpoint{1.227629in}{0.442177in}}%
\pgfpathlineto{\pgfqpoint{1.310785in}{0.442177in}}%
\pgfpathlineto{\pgfqpoint{1.310785in}{0.442177in}}%
\pgfpathlineto{\pgfqpoint{1.393942in}{0.442177in}}%
\pgfpathlineto{\pgfqpoint{1.393942in}{0.442177in}}%
\pgfpathlineto{\pgfqpoint{1.477098in}{0.442177in}}%
\pgfpathlineto{\pgfqpoint{1.477098in}{0.442177in}}%
\pgfpathlineto{\pgfqpoint{1.560255in}{0.442177in}}%
\pgfpathlineto{\pgfqpoint{1.560255in}{0.442177in}}%
\pgfpathlineto{\pgfqpoint{1.643412in}{0.442177in}}%
\pgfpathlineto{\pgfqpoint{1.643412in}{0.442177in}}%
\pgfpathlineto{\pgfqpoint{1.726568in}{0.442177in}}%
\pgfpathlineto{\pgfqpoint{1.726568in}{0.442177in}}%
\pgfpathlineto{\pgfqpoint{1.809725in}{0.442177in}}%
\pgfpathlineto{\pgfqpoint{1.809725in}{0.442177in}}%
\pgfpathlineto{\pgfqpoint{1.892881in}{0.442177in}}%
\pgfpathlineto{\pgfqpoint{1.892881in}{0.442177in}}%
\pgfpathlineto{\pgfqpoint{1.976038in}{0.442177in}}%
\pgfpathlineto{\pgfqpoint{1.976038in}{0.442177in}}%
\pgfpathlineto{\pgfqpoint{2.059194in}{0.442177in}}%
\pgfpathlineto{\pgfqpoint{2.059194in}{0.442177in}}%
\pgfpathlineto{\pgfqpoint{2.142351in}{0.442177in}}%
\pgfpathlineto{\pgfqpoint{2.142351in}{0.442177in}}%
\pgfpathlineto{\pgfqpoint{2.225508in}{0.442177in}}%
\pgfpathlineto{\pgfqpoint{2.225508in}{0.442177in}}%
\pgfpathlineto{\pgfqpoint{2.308664in}{0.442177in}}%
\pgfpathlineto{\pgfqpoint{2.308664in}{0.442177in}}%
\pgfpathlineto{\pgfqpoint{2.391821in}{0.442177in}}%
\pgfpathlineto{\pgfqpoint{2.391821in}{0.442177in}}%
\pgfpathlineto{\pgfqpoint{2.474977in}{0.442177in}}%
\pgfpathlineto{\pgfqpoint{2.474977in}{0.442177in}}%
\pgfpathlineto{\pgfqpoint{2.558134in}{0.442177in}}%
\pgfpathlineto{\pgfqpoint{2.558134in}{0.442177in}}%
\pgfpathlineto{\pgfqpoint{2.641290in}{0.442177in}}%
\pgfpathlineto{\pgfqpoint{2.641290in}{0.442177in}}%
\pgfpathlineto{\pgfqpoint{2.724447in}{0.442177in}}%
\pgfpathlineto{\pgfqpoint{2.724447in}{0.442177in}}%
\pgfpathlineto{\pgfqpoint{2.807604in}{0.442177in}}%
\pgfpathlineto{\pgfqpoint{2.807604in}{0.442177in}}%
\pgfpathlineto{\pgfqpoint{2.890760in}{0.442177in}}%
\pgfpathlineto{\pgfqpoint{2.890760in}{0.442177in}}%
\pgfpathlineto{\pgfqpoint{2.932338in}{0.442177in}}%
\pgfpathlineto{\pgfqpoint{2.932338in}{0.442177in}}%
\pgfpathlineto{\pgfqpoint{2.932338in}{0.442177in}}%
\pgfpathlineto{\pgfqpoint{2.890760in}{0.442177in}}%
\pgfpathlineto{\pgfqpoint{2.890760in}{0.442177in}}%
\pgfpathlineto{\pgfqpoint{2.807604in}{0.442177in}}%
\pgfpathlineto{\pgfqpoint{2.807604in}{0.442177in}}%
\pgfpathlineto{\pgfqpoint{2.724447in}{0.442177in}}%
\pgfpathlineto{\pgfqpoint{2.724447in}{0.442177in}}%
\pgfpathlineto{\pgfqpoint{2.641290in}{0.442177in}}%
\pgfpathlineto{\pgfqpoint{2.641290in}{0.442177in}}%
\pgfpathlineto{\pgfqpoint{2.558134in}{0.442177in}}%
\pgfpathlineto{\pgfqpoint{2.558134in}{0.442177in}}%
\pgfpathlineto{\pgfqpoint{2.474977in}{0.442177in}}%
\pgfpathlineto{\pgfqpoint{2.474977in}{0.442177in}}%
\pgfpathlineto{\pgfqpoint{2.391821in}{0.442177in}}%
\pgfpathlineto{\pgfqpoint{2.391821in}{0.442177in}}%
\pgfpathlineto{\pgfqpoint{2.308664in}{0.442177in}}%
\pgfpathlineto{\pgfqpoint{2.308664in}{0.442177in}}%
\pgfpathlineto{\pgfqpoint{2.225508in}{0.442177in}}%
\pgfpathlineto{\pgfqpoint{2.225508in}{0.442177in}}%
\pgfpathlineto{\pgfqpoint{2.142351in}{0.442177in}}%
\pgfpathlineto{\pgfqpoint{2.142351in}{0.442177in}}%
\pgfpathlineto{\pgfqpoint{2.059194in}{0.442177in}}%
\pgfpathlineto{\pgfqpoint{2.059194in}{0.442177in}}%
\pgfpathlineto{\pgfqpoint{1.976038in}{0.442177in}}%
\pgfpathlineto{\pgfqpoint{1.976038in}{0.442177in}}%
\pgfpathlineto{\pgfqpoint{1.892881in}{0.442177in}}%
\pgfpathlineto{\pgfqpoint{1.892881in}{0.442177in}}%
\pgfpathlineto{\pgfqpoint{1.809725in}{0.442177in}}%
\pgfpathlineto{\pgfqpoint{1.809725in}{0.442177in}}%
\pgfpathlineto{\pgfqpoint{1.726568in}{0.442177in}}%
\pgfpathlineto{\pgfqpoint{1.726568in}{0.442177in}}%
\pgfpathlineto{\pgfqpoint{1.643412in}{0.442177in}}%
\pgfpathlineto{\pgfqpoint{1.643412in}{0.442177in}}%
\pgfpathlineto{\pgfqpoint{1.560255in}{0.442177in}}%
\pgfpathlineto{\pgfqpoint{1.560255in}{0.442177in}}%
\pgfpathlineto{\pgfqpoint{1.477098in}{0.442177in}}%
\pgfpathlineto{\pgfqpoint{1.477098in}{0.442177in}}%
\pgfpathlineto{\pgfqpoint{1.393942in}{0.442177in}}%
\pgfpathlineto{\pgfqpoint{1.393942in}{0.442177in}}%
\pgfpathlineto{\pgfqpoint{1.310785in}{0.442177in}}%
\pgfpathlineto{\pgfqpoint{1.310785in}{0.442177in}}%
\pgfpathlineto{\pgfqpoint{1.227629in}{0.442177in}}%
\pgfpathlineto{\pgfqpoint{1.227629in}{0.449452in}}%
\pgfpathlineto{\pgfqpoint{1.144472in}{0.449452in}}%
\pgfpathlineto{\pgfqpoint{1.144472in}{0.465342in}}%
\pgfpathlineto{\pgfqpoint{1.061316in}{0.465342in}}%
\pgfpathlineto{\pgfqpoint{1.061316in}{0.531678in}}%
\pgfpathlineto{\pgfqpoint{0.978159in}{0.531678in}}%
\pgfpathlineto{\pgfqpoint{0.978159in}{0.636407in}}%
\pgfpathlineto{\pgfqpoint{0.895002in}{0.636407in}}%
\pgfpathlineto{\pgfqpoint{0.895002in}{0.735281in}}%
\pgfpathlineto{\pgfqpoint{0.811846in}{0.735281in}}%
\pgfpathlineto{\pgfqpoint{0.811846in}{0.810954in}}%
\pgfpathlineto{\pgfqpoint{0.728689in}{0.810954in}}%
\pgfpathlineto{\pgfqpoint{0.728689in}{0.867568in}}%
\pgfpathlineto{\pgfqpoint{0.645533in}{0.867568in}}%
\pgfpathlineto{\pgfqpoint{0.645533in}{0.912360in}}%
\pgfpathlineto{\pgfqpoint{0.562376in}{0.912360in}}%
\pgfpathlineto{\pgfqpoint{0.562376in}{0.948913in}}%
\pgfpathlineto{\pgfqpoint{0.520798in}{0.948913in}}%
\pgfpathlineto{\pgfqpoint{0.520798in}{0.948913in}}%
\pgfpathclose%
\pgfusepath{fill}%
\end{pgfscope}%
\begin{pgfscope}%
\pgfpathrectangle{\pgfqpoint{0.520798in}{0.442177in}}{\pgfqpoint{0.748409in}{0.871884in}}%
\pgfusepath{clip}%
\pgfsetroundcap%
\pgfsetroundjoin%
\pgfsetlinewidth{1.003750pt}%
\definecolor{currentstroke}{rgb}{0.870588,0.560784,0.019608}%
\pgfsetstrokecolor{currentstroke}%
\pgfsetdash{}{0pt}%
\pgfpathmoveto{\pgfqpoint{0.520798in}{1.012757in}}%
\pgfpathlineto{\pgfqpoint{0.562376in}{1.012757in}}%
\pgfpathlineto{\pgfqpoint{0.562376in}{0.989823in}}%
\pgfpathlineto{\pgfqpoint{0.645533in}{0.989823in}}%
\pgfpathlineto{\pgfqpoint{0.645533in}{0.957795in}}%
\pgfpathlineto{\pgfqpoint{0.728689in}{0.957795in}}%
\pgfpathlineto{\pgfqpoint{0.728689in}{0.920626in}}%
\pgfpathlineto{\pgfqpoint{0.811846in}{0.920626in}}%
\pgfpathlineto{\pgfqpoint{0.811846in}{0.868827in}}%
\pgfpathlineto{\pgfqpoint{0.895002in}{0.868827in}}%
\pgfpathlineto{\pgfqpoint{0.895002in}{0.797653in}}%
\pgfpathlineto{\pgfqpoint{0.978159in}{0.797653in}}%
\pgfpathlineto{\pgfqpoint{0.978159in}{0.671516in}}%
\pgfpathlineto{\pgfqpoint{1.061316in}{0.671516in}}%
\pgfpathlineto{\pgfqpoint{1.061316in}{0.539053in}}%
\pgfpathlineto{\pgfqpoint{1.144472in}{0.539053in}}%
\pgfpathlineto{\pgfqpoint{1.144472in}{0.471042in}}%
\pgfpathlineto{\pgfqpoint{1.227629in}{0.471042in}}%
\pgfpathlineto{\pgfqpoint{1.227629in}{0.449294in}}%
\pgfpathlineto{\pgfqpoint{1.269207in}{0.449294in}}%
\pgfusepath{stroke}%
\end{pgfscope}%
\begin{pgfscope}%
\pgfpathrectangle{\pgfqpoint{0.520798in}{0.442177in}}{\pgfqpoint{0.748409in}{0.871884in}}%
\pgfusepath{clip}%
\pgfsetbuttcap%
\pgfsetroundjoin%
\definecolor{currentfill}{rgb}{0.870588,0.560784,0.019608}%
\pgfsetfillcolor{currentfill}%
\pgfsetfillopacity{0.500000}%
\pgfsetlinewidth{0.000000pt}%
\definecolor{currentstroke}{rgb}{0.870588,0.560784,0.019608}%
\pgfsetstrokecolor{currentstroke}%
\pgfsetstrokeopacity{0.500000}%
\pgfsetdash{}{0pt}%
\pgfsys@defobject{currentmarker}{\pgfqpoint{0.520798in}{0.449294in}}{\pgfqpoint{1.269207in}{1.012757in}}{%
\pgfpathmoveto{\pgfqpoint{0.520798in}{1.012757in}}%
\pgfpathlineto{\pgfqpoint{0.520798in}{1.012757in}}%
\pgfpathlineto{\pgfqpoint{0.562376in}{1.012757in}}%
\pgfpathlineto{\pgfqpoint{0.562376in}{0.989823in}}%
\pgfpathlineto{\pgfqpoint{0.645533in}{0.989823in}}%
\pgfpathlineto{\pgfqpoint{0.645533in}{0.957795in}}%
\pgfpathlineto{\pgfqpoint{0.728689in}{0.957795in}}%
\pgfpathlineto{\pgfqpoint{0.728689in}{0.920626in}}%
\pgfpathlineto{\pgfqpoint{0.811846in}{0.920626in}}%
\pgfpathlineto{\pgfqpoint{0.811846in}{0.868827in}}%
\pgfpathlineto{\pgfqpoint{0.895002in}{0.868827in}}%
\pgfpathlineto{\pgfqpoint{0.895002in}{0.797653in}}%
\pgfpathlineto{\pgfqpoint{0.978159in}{0.797653in}}%
\pgfpathlineto{\pgfqpoint{0.978159in}{0.671516in}}%
\pgfpathlineto{\pgfqpoint{1.061316in}{0.671516in}}%
\pgfpathlineto{\pgfqpoint{1.061316in}{0.539053in}}%
\pgfpathlineto{\pgfqpoint{1.144472in}{0.539053in}}%
\pgfpathlineto{\pgfqpoint{1.144472in}{0.471042in}}%
\pgfpathlineto{\pgfqpoint{1.227629in}{0.471042in}}%
\pgfpathlineto{\pgfqpoint{1.227629in}{0.449294in}}%
\pgfpathlineto{\pgfqpoint{1.269207in}{0.449294in}}%
\pgfpathlineto{\pgfqpoint{1.269207in}{0.449294in}}%
\pgfpathlineto{\pgfqpoint{1.269207in}{0.449294in}}%
\pgfpathlineto{\pgfqpoint{1.227629in}{0.449294in}}%
\pgfpathlineto{\pgfqpoint{1.227629in}{0.471042in}}%
\pgfpathlineto{\pgfqpoint{1.144472in}{0.471042in}}%
\pgfpathlineto{\pgfqpoint{1.144472in}{0.539053in}}%
\pgfpathlineto{\pgfqpoint{1.061316in}{0.539053in}}%
\pgfpathlineto{\pgfqpoint{1.061316in}{0.671516in}}%
\pgfpathlineto{\pgfqpoint{0.978159in}{0.671516in}}%
\pgfpathlineto{\pgfqpoint{0.978159in}{0.797653in}}%
\pgfpathlineto{\pgfqpoint{0.895002in}{0.797653in}}%
\pgfpathlineto{\pgfqpoint{0.895002in}{0.868827in}}%
\pgfpathlineto{\pgfqpoint{0.811846in}{0.868827in}}%
\pgfpathlineto{\pgfqpoint{0.811846in}{0.920626in}}%
\pgfpathlineto{\pgfqpoint{0.728689in}{0.920626in}}%
\pgfpathlineto{\pgfqpoint{0.728689in}{0.957795in}}%
\pgfpathlineto{\pgfqpoint{0.645533in}{0.957795in}}%
\pgfpathlineto{\pgfqpoint{0.645533in}{0.989823in}}%
\pgfpathlineto{\pgfqpoint{0.562376in}{0.989823in}}%
\pgfpathlineto{\pgfqpoint{0.562376in}{1.012757in}}%
\pgfpathlineto{\pgfqpoint{0.520798in}{1.012757in}}%
\pgfpathlineto{\pgfqpoint{0.520798in}{1.012757in}}%
\pgfpathclose%
\pgfusepath{fill}%
}%
\begin{pgfscope}%
\pgfsys@transformshift{0.000000in}{0.000000in}%
\pgfsys@useobject{currentmarker}{}%
\end{pgfscope}%
\end{pgfscope}%
\begin{pgfscope}%
\pgfpathrectangle{\pgfqpoint{0.520798in}{0.442177in}}{\pgfqpoint{0.748409in}{0.871884in}}%
\pgfusepath{clip}%
\pgfsetroundcap%
\pgfsetroundjoin%
\pgfsetlinewidth{1.003750pt}%
\definecolor{currentstroke}{rgb}{0.007843,0.619608,0.450980}%
\pgfsetstrokecolor{currentstroke}%
\pgfsetdash{}{0pt}%
\pgfpathmoveto{\pgfqpoint{0.520798in}{1.068115in}}%
\pgfpathlineto{\pgfqpoint{0.562376in}{1.068115in}}%
\pgfpathlineto{\pgfqpoint{0.562376in}{1.059811in}}%
\pgfpathlineto{\pgfqpoint{0.645533in}{1.059811in}}%
\pgfpathlineto{\pgfqpoint{0.645533in}{1.047949in}}%
\pgfpathlineto{\pgfqpoint{0.728689in}{1.047949in}}%
\pgfpathlineto{\pgfqpoint{0.728689in}{1.029760in}}%
\pgfpathlineto{\pgfqpoint{0.811846in}{1.029760in}}%
\pgfpathlineto{\pgfqpoint{0.811846in}{0.997336in}}%
\pgfpathlineto{\pgfqpoint{0.895002in}{0.997336in}}%
\pgfpathlineto{\pgfqpoint{0.895002in}{0.962935in}}%
\pgfpathlineto{\pgfqpoint{0.978159in}{0.962935in}}%
\pgfpathlineto{\pgfqpoint{0.978159in}{0.898878in}}%
\pgfpathlineto{\pgfqpoint{1.061316in}{0.898878in}}%
\pgfpathlineto{\pgfqpoint{1.061316in}{0.785790in}}%
\pgfpathlineto{\pgfqpoint{1.144472in}{0.785790in}}%
\pgfpathlineto{\pgfqpoint{1.144472in}{0.650955in}}%
\pgfpathlineto{\pgfqpoint{1.227629in}{0.650955in}}%
\pgfpathlineto{\pgfqpoint{1.227629in}{0.533913in}}%
\pgfpathlineto{\pgfqpoint{1.269207in}{0.533913in}}%
\pgfusepath{stroke}%
\end{pgfscope}%
\begin{pgfscope}%
\pgfpathrectangle{\pgfqpoint{0.520798in}{0.442177in}}{\pgfqpoint{0.748409in}{0.871884in}}%
\pgfusepath{clip}%
\pgfsetbuttcap%
\pgfsetroundjoin%
\definecolor{currentfill}{rgb}{0.007843,0.619608,0.450980}%
\pgfsetfillcolor{currentfill}%
\pgfsetfillopacity{0.500000}%
\pgfsetlinewidth{0.000000pt}%
\definecolor{currentstroke}{rgb}{0.007843,0.619608,0.450980}%
\pgfsetstrokecolor{currentstroke}%
\pgfsetstrokeopacity{0.500000}%
\pgfsetdash{}{0pt}%
\pgfsys@defobject{currentmarker}{\pgfqpoint{0.520798in}{0.533913in}}{\pgfqpoint{1.269207in}{1.068115in}}{%
\pgfpathmoveto{\pgfqpoint{0.520798in}{1.068115in}}%
\pgfpathlineto{\pgfqpoint{0.520798in}{1.068115in}}%
\pgfpathlineto{\pgfqpoint{0.562376in}{1.068115in}}%
\pgfpathlineto{\pgfqpoint{0.562376in}{1.059811in}}%
\pgfpathlineto{\pgfqpoint{0.645533in}{1.059811in}}%
\pgfpathlineto{\pgfqpoint{0.645533in}{1.047949in}}%
\pgfpathlineto{\pgfqpoint{0.728689in}{1.047949in}}%
\pgfpathlineto{\pgfqpoint{0.728689in}{1.029760in}}%
\pgfpathlineto{\pgfqpoint{0.811846in}{1.029760in}}%
\pgfpathlineto{\pgfqpoint{0.811846in}{0.997336in}}%
\pgfpathlineto{\pgfqpoint{0.895002in}{0.997336in}}%
\pgfpathlineto{\pgfqpoint{0.895002in}{0.962935in}}%
\pgfpathlineto{\pgfqpoint{0.978159in}{0.962935in}}%
\pgfpathlineto{\pgfqpoint{0.978159in}{0.898878in}}%
\pgfpathlineto{\pgfqpoint{1.061316in}{0.898878in}}%
\pgfpathlineto{\pgfqpoint{1.061316in}{0.785790in}}%
\pgfpathlineto{\pgfqpoint{1.144472in}{0.785790in}}%
\pgfpathlineto{\pgfqpoint{1.144472in}{0.650955in}}%
\pgfpathlineto{\pgfqpoint{1.227629in}{0.650955in}}%
\pgfpathlineto{\pgfqpoint{1.227629in}{0.533913in}}%
\pgfpathlineto{\pgfqpoint{1.269207in}{0.533913in}}%
\pgfpathlineto{\pgfqpoint{1.269207in}{0.533913in}}%
\pgfpathlineto{\pgfqpoint{1.269207in}{0.533913in}}%
\pgfpathlineto{\pgfqpoint{1.227629in}{0.533913in}}%
\pgfpathlineto{\pgfqpoint{1.227629in}{0.650955in}}%
\pgfpathlineto{\pgfqpoint{1.144472in}{0.650955in}}%
\pgfpathlineto{\pgfqpoint{1.144472in}{0.785790in}}%
\pgfpathlineto{\pgfqpoint{1.061316in}{0.785790in}}%
\pgfpathlineto{\pgfqpoint{1.061316in}{0.898878in}}%
\pgfpathlineto{\pgfqpoint{0.978159in}{0.898878in}}%
\pgfpathlineto{\pgfqpoint{0.978159in}{0.962935in}}%
\pgfpathlineto{\pgfqpoint{0.895002in}{0.962935in}}%
\pgfpathlineto{\pgfqpoint{0.895002in}{0.997336in}}%
\pgfpathlineto{\pgfqpoint{0.811846in}{0.997336in}}%
\pgfpathlineto{\pgfqpoint{0.811846in}{1.029760in}}%
\pgfpathlineto{\pgfqpoint{0.728689in}{1.029760in}}%
\pgfpathlineto{\pgfqpoint{0.728689in}{1.047949in}}%
\pgfpathlineto{\pgfqpoint{0.645533in}{1.047949in}}%
\pgfpathlineto{\pgfqpoint{0.645533in}{1.059811in}}%
\pgfpathlineto{\pgfqpoint{0.562376in}{1.059811in}}%
\pgfpathlineto{\pgfqpoint{0.562376in}{1.068115in}}%
\pgfpathlineto{\pgfqpoint{0.520798in}{1.068115in}}%
\pgfpathlineto{\pgfqpoint{0.520798in}{1.068115in}}%
\pgfpathclose%
\pgfusepath{fill}%
}%
\begin{pgfscope}%
\pgfsys@transformshift{0.000000in}{0.000000in}%
\pgfsys@useobject{currentmarker}{}%
\end{pgfscope}%
\end{pgfscope}%
\begin{pgfscope}%
\definecolor{textcolor}{rgb}{0.150000,0.150000,0.150000}%
\pgfsetstrokecolor{textcolor}%
\pgfsetfillcolor{textcolor}%
\pgftext[x=0.895002in,y=1.397395in,,base]{\color{textcolor}\rmfamily\fontsize{9.000000}{10.800000}\selectfont \(\displaystyle c_A^-=1\)}%
\end{pgfscope}%
\begin{pgfscope}%
\pgfsetbuttcap%
\pgfsetmiterjoin%
\definecolor{currentfill}{rgb}{1.000000,1.000000,1.000000}%
\pgfsetfillcolor{currentfill}%
\pgfsetfillopacity{0.800000}%
\pgfsetlinewidth{1.003750pt}%
\definecolor{currentstroke}{rgb}{0.800000,0.800000,0.800000}%
\pgfsetstrokecolor{currentstroke}%
\pgfsetstrokeopacity{0.800000}%
\pgfsetdash{}{0pt}%
\pgfpathmoveto{\pgfqpoint{0.815247in}{0.638103in}}%
\pgfpathlineto{\pgfqpoint{1.220596in}{0.638103in}}%
\pgfpathlineto{\pgfqpoint{1.220596in}{1.265450in}}%
\pgfpathlineto{\pgfqpoint{0.815247in}{1.265450in}}%
\pgfpathlineto{\pgfqpoint{0.815247in}{0.638103in}}%
\pgfpathclose%
\pgfusepath{stroke,fill}%
\end{pgfscope}%
\begin{pgfscope}%
\definecolor{textcolor}{rgb}{0.150000,0.150000,0.150000}%
\pgfsetstrokecolor{textcolor}%
\pgfsetfillcolor{textcolor}%
\pgftext[x=0.947675in,y=1.113275in,left,base]{\color{textcolor}\rmfamily\fontsize{9.000000}{10.800000}\selectfont \(\displaystyle c_A^+\)}%
\end{pgfscope}%
\begin{pgfscope}%
\pgfsetroundcap%
\pgfsetroundjoin%
\pgfsetlinewidth{1.003750pt}%
\definecolor{currentstroke}{rgb}{0.003922,0.450980,0.698039}%
\pgfsetstrokecolor{currentstroke}%
\pgfsetdash{}{0pt}%
\pgfpathmoveto{\pgfqpoint{0.854136in}{1.001052in}}%
\pgfpathlineto{\pgfqpoint{0.902747in}{1.001052in}}%
\pgfpathlineto{\pgfqpoint{0.902747in}{1.001052in}}%
\pgfpathlineto{\pgfqpoint{0.999969in}{1.001052in}}%
\pgfpathlineto{\pgfqpoint{0.999969in}{1.001052in}}%
\pgfpathlineto{\pgfqpoint{1.048580in}{1.001052in}}%
\pgfusepath{stroke}%
\end{pgfscope}%
\begin{pgfscope}%
\definecolor{textcolor}{rgb}{0.150000,0.150000,0.150000}%
\pgfsetstrokecolor{textcolor}%
\pgfsetfillcolor{textcolor}%
\pgftext[x=1.126358in,y=0.967024in,left,base]{\color{textcolor}\rmfamily\fontsize{7.000000}{8.400000}\selectfont 1}%
\end{pgfscope}%
\begin{pgfscope}%
\pgfsetroundcap%
\pgfsetroundjoin%
\pgfsetlinewidth{1.003750pt}%
\definecolor{currentstroke}{rgb}{0.870588,0.560784,0.019608}%
\pgfsetstrokecolor{currentstroke}%
\pgfsetdash{}{0pt}%
\pgfpathmoveto{\pgfqpoint{0.854136in}{0.865486in}}%
\pgfpathlineto{\pgfqpoint{0.902747in}{0.865486in}}%
\pgfpathlineto{\pgfqpoint{0.902747in}{0.865486in}}%
\pgfpathlineto{\pgfqpoint{0.999969in}{0.865486in}}%
\pgfpathlineto{\pgfqpoint{0.999969in}{0.865486in}}%
\pgfpathlineto{\pgfqpoint{1.048580in}{0.865486in}}%
\pgfusepath{stroke}%
\end{pgfscope}%
\begin{pgfscope}%
\definecolor{textcolor}{rgb}{0.150000,0.150000,0.150000}%
\pgfsetstrokecolor{textcolor}%
\pgfsetfillcolor{textcolor}%
\pgftext[x=1.126358in,y=0.831458in,left,base]{\color{textcolor}\rmfamily\fontsize{7.000000}{8.400000}\selectfont 2}%
\end{pgfscope}%
\begin{pgfscope}%
\pgfsetroundcap%
\pgfsetroundjoin%
\pgfsetlinewidth{1.003750pt}%
\definecolor{currentstroke}{rgb}{0.007843,0.619608,0.450980}%
\pgfsetstrokecolor{currentstroke}%
\pgfsetdash{}{0pt}%
\pgfpathmoveto{\pgfqpoint{0.854136in}{0.729919in}}%
\pgfpathlineto{\pgfqpoint{0.902747in}{0.729919in}}%
\pgfpathlineto{\pgfqpoint{0.902747in}{0.729919in}}%
\pgfpathlineto{\pgfqpoint{0.999969in}{0.729919in}}%
\pgfpathlineto{\pgfqpoint{0.999969in}{0.729919in}}%
\pgfpathlineto{\pgfqpoint{1.048580in}{0.729919in}}%
\pgfusepath{stroke}%
\end{pgfscope}%
\begin{pgfscope}%
\definecolor{textcolor}{rgb}{0.150000,0.150000,0.150000}%
\pgfsetstrokecolor{textcolor}%
\pgfsetfillcolor{textcolor}%
\pgftext[x=1.126358in,y=0.695892in,left,base]{\color{textcolor}\rmfamily\fontsize{7.000000}{8.400000}\selectfont 4}%
\end{pgfscope}%
\end{pgfpicture}%
\makeatother%
\endgroup%

%% file: figures/experiments/graphs/piPPNP/piPPNP-Cora-A.pgf
\begingroup%
\makeatletter%
\begin{pgfpicture}%
\pgfpathrectangle{\pgfpointorigin}{\pgfqpoint{1.375000in}{1.581250in}}%
\pgfusepath{use as bounding box, clip}%
\begin{pgfscope}%
\pgfsetbuttcap%
\pgfsetmiterjoin%
\definecolor{currentfill}{rgb}{1.000000,1.000000,1.000000}%
\pgfsetfillcolor{currentfill}%
\pgfsetlinewidth{0.000000pt}%
\definecolor{currentstroke}{rgb}{1.000000,1.000000,1.000000}%
\pgfsetstrokecolor{currentstroke}%
\pgfsetdash{}{0pt}%
\pgfpathmoveto{\pgfqpoint{0.000000in}{0.000000in}}%
\pgfpathlineto{\pgfqpoint{1.375000in}{0.000000in}}%
\pgfpathlineto{\pgfqpoint{1.375000in}{1.581250in}}%
\pgfpathlineto{\pgfqpoint{0.000000in}{1.581250in}}%
\pgfpathlineto{\pgfqpoint{0.000000in}{0.000000in}}%
\pgfpathclose%
\pgfusepath{fill}%
\end{pgfscope}%
\begin{pgfscope}%
\pgfsetbuttcap%
\pgfsetmiterjoin%
\definecolor{currentfill}{rgb}{1.000000,1.000000,1.000000}%
\pgfsetfillcolor{currentfill}%
\pgfsetlinewidth{0.000000pt}%
\definecolor{currentstroke}{rgb}{0.000000,0.000000,0.000000}%
\pgfsetstrokecolor{currentstroke}%
\pgfsetstrokeopacity{0.000000}%
\pgfsetdash{}{0pt}%
\pgfpathmoveto{\pgfqpoint{0.520798in}{0.442177in}}%
\pgfpathlineto{\pgfqpoint{1.269207in}{0.442177in}}%
\pgfpathlineto{\pgfqpoint{1.269207in}{1.314061in}}%
\pgfpathlineto{\pgfqpoint{0.520798in}{1.314061in}}%
\pgfpathlineto{\pgfqpoint{0.520798in}{0.442177in}}%
\pgfpathclose%
\pgfusepath{fill}%
\end{pgfscope}%
\begin{pgfscope}%
\pgfpathrectangle{\pgfqpoint{0.520798in}{0.442177in}}{\pgfqpoint{0.748409in}{0.871884in}}%
\pgfusepath{clip}%
\pgfsetroundcap%
\pgfsetroundjoin%
\pgfsetlinewidth{0.501875pt}%
\definecolor{currentstroke}{rgb}{0.800000,0.800000,0.800000}%
\pgfsetstrokecolor{currentstroke}%
\pgfsetdash{}{0pt}%
\pgfpathmoveto{\pgfqpoint{0.520798in}{0.442177in}}%
\pgfpathlineto{\pgfqpoint{0.520798in}{1.314061in}}%
\pgfusepath{stroke}%
\end{pgfscope}%
\begin{pgfscope}%
\definecolor{textcolor}{rgb}{0.150000,0.150000,0.150000}%
\pgfsetstrokecolor{textcolor}%
\pgfsetfillcolor{textcolor}%
\pgftext[x=0.520798in,y=0.351899in,,top]{\color{textcolor}\rmfamily\fontsize{8.000000}{9.600000}\selectfont \(\displaystyle {0}\)}%
\end{pgfscope}%
\begin{pgfscope}%
\pgfpathrectangle{\pgfqpoint{0.520798in}{0.442177in}}{\pgfqpoint{0.748409in}{0.871884in}}%
\pgfusepath{clip}%
\pgfsetroundcap%
\pgfsetroundjoin%
\pgfsetlinewidth{0.501875pt}%
\definecolor{currentstroke}{rgb}{0.800000,0.800000,0.800000}%
\pgfsetstrokecolor{currentstroke}%
\pgfsetdash{}{0pt}%
\pgfpathmoveto{\pgfqpoint{0.853424in}{0.442177in}}%
\pgfpathlineto{\pgfqpoint{0.853424in}{1.314061in}}%
\pgfusepath{stroke}%
\end{pgfscope}%
\begin{pgfscope}%
\definecolor{textcolor}{rgb}{0.150000,0.150000,0.150000}%
\pgfsetstrokecolor{textcolor}%
\pgfsetfillcolor{textcolor}%
\pgftext[x=0.853424in,y=0.351899in,,top]{\color{textcolor}\rmfamily\fontsize{8.000000}{9.600000}\selectfont \(\displaystyle {4}\)}%
\end{pgfscope}%
\begin{pgfscope}%
\pgfpathrectangle{\pgfqpoint{0.520798in}{0.442177in}}{\pgfqpoint{0.748409in}{0.871884in}}%
\pgfusepath{clip}%
\pgfsetroundcap%
\pgfsetroundjoin%
\pgfsetlinewidth{0.501875pt}%
\definecolor{currentstroke}{rgb}{0.800000,0.800000,0.800000}%
\pgfsetstrokecolor{currentstroke}%
\pgfsetdash{}{0pt}%
\pgfpathmoveto{\pgfqpoint{1.269207in}{0.442177in}}%
\pgfpathlineto{\pgfqpoint{1.269207in}{1.314061in}}%
\pgfusepath{stroke}%
\end{pgfscope}%
\begin{pgfscope}%
\definecolor{textcolor}{rgb}{0.150000,0.150000,0.150000}%
\pgfsetstrokecolor{textcolor}%
\pgfsetfillcolor{textcolor}%
\pgftext[x=1.269207in,y=0.351899in,,top]{\color{textcolor}\rmfamily\fontsize{8.000000}{9.600000}\selectfont \(\displaystyle {9}\)}%
\end{pgfscope}%
\begin{pgfscope}%
\definecolor{textcolor}{rgb}{0.150000,0.150000,0.150000}%
\pgfsetstrokecolor{textcolor}%
\pgfsetfillcolor{textcolor}%
\pgftext[x=0.895002in,y=0.198219in,,top]{\color{textcolor}\rmfamily\fontsize{10.000000}{12.000000}\selectfont attack strength}%
\end{pgfscope}%
\begin{pgfscope}%
\pgfpathrectangle{\pgfqpoint{0.520798in}{0.442177in}}{\pgfqpoint{0.748409in}{0.871884in}}%
\pgfusepath{clip}%
\pgfsetroundcap%
\pgfsetroundjoin%
\pgfsetlinewidth{0.501875pt}%
\definecolor{currentstroke}{rgb}{0.800000,0.800000,0.800000}%
\pgfsetstrokecolor{currentstroke}%
\pgfsetdash{}{0pt}%
\pgfpathmoveto{\pgfqpoint{0.520798in}{0.442177in}}%
\pgfpathlineto{\pgfqpoint{1.269207in}{0.442177in}}%
\pgfusepath{stroke}%
\end{pgfscope}%
\begin{pgfscope}%
\definecolor{textcolor}{rgb}{0.150000,0.150000,0.150000}%
\pgfsetstrokecolor{textcolor}%
\pgfsetfillcolor{textcolor}%
\pgftext[x=0.273151in, y=0.403915in, left, base]{\color{textcolor}\rmfamily\fontsize{8.000000}{9.600000}\selectfont 0\%}%
\end{pgfscope}%
\begin{pgfscope}%
\pgfpathrectangle{\pgfqpoint{0.520798in}{0.442177in}}{\pgfqpoint{0.748409in}{0.871884in}}%
\pgfusepath{clip}%
\pgfsetroundcap%
\pgfsetroundjoin%
\pgfsetlinewidth{0.501875pt}%
\definecolor{currentstroke}{rgb}{0.800000,0.800000,0.800000}%
\pgfsetstrokecolor{currentstroke}%
\pgfsetdash{}{0pt}%
\pgfpathmoveto{\pgfqpoint{0.520798in}{0.660148in}}%
\pgfpathlineto{\pgfqpoint{1.269207in}{0.660148in}}%
\pgfusepath{stroke}%
\end{pgfscope}%
\begin{pgfscope}%
\definecolor{textcolor}{rgb}{0.150000,0.150000,0.150000}%
\pgfsetstrokecolor{textcolor}%
\pgfsetfillcolor{textcolor}%
\pgftext[x=0.214138in, y=0.621886in, left, base]{\color{textcolor}\rmfamily\fontsize{8.000000}{9.600000}\selectfont 25\%}%
\end{pgfscope}%
\begin{pgfscope}%
\pgfpathrectangle{\pgfqpoint{0.520798in}{0.442177in}}{\pgfqpoint{0.748409in}{0.871884in}}%
\pgfusepath{clip}%
\pgfsetroundcap%
\pgfsetroundjoin%
\pgfsetlinewidth{0.501875pt}%
\definecolor{currentstroke}{rgb}{0.800000,0.800000,0.800000}%
\pgfsetstrokecolor{currentstroke}%
\pgfsetdash{}{0pt}%
\pgfpathmoveto{\pgfqpoint{0.520798in}{0.878119in}}%
\pgfpathlineto{\pgfqpoint{1.269207in}{0.878119in}}%
\pgfusepath{stroke}%
\end{pgfscope}%
\begin{pgfscope}%
\definecolor{textcolor}{rgb}{0.150000,0.150000,0.150000}%
\pgfsetstrokecolor{textcolor}%
\pgfsetfillcolor{textcolor}%
\pgftext[x=0.214138in, y=0.839857in, left, base]{\color{textcolor}\rmfamily\fontsize{8.000000}{9.600000}\selectfont 50\%}%
\end{pgfscope}%
\begin{pgfscope}%
\pgfpathrectangle{\pgfqpoint{0.520798in}{0.442177in}}{\pgfqpoint{0.748409in}{0.871884in}}%
\pgfusepath{clip}%
\pgfsetroundcap%
\pgfsetroundjoin%
\pgfsetlinewidth{0.501875pt}%
\definecolor{currentstroke}{rgb}{0.800000,0.800000,0.800000}%
\pgfsetstrokecolor{currentstroke}%
\pgfsetdash{}{0pt}%
\pgfpathmoveto{\pgfqpoint{0.520798in}{1.096090in}}%
\pgfpathlineto{\pgfqpoint{1.269207in}{1.096090in}}%
\pgfusepath{stroke}%
\end{pgfscope}%
\begin{pgfscope}%
\definecolor{textcolor}{rgb}{0.150000,0.150000,0.150000}%
\pgfsetstrokecolor{textcolor}%
\pgfsetfillcolor{textcolor}%
\pgftext[x=0.214138in, y=1.057828in, left, base]{\color{textcolor}\rmfamily\fontsize{8.000000}{9.600000}\selectfont 75\%}%
\end{pgfscope}%
\begin{pgfscope}%
\pgfpathrectangle{\pgfqpoint{0.520798in}{0.442177in}}{\pgfqpoint{0.748409in}{0.871884in}}%
\pgfusepath{clip}%
\pgfsetroundcap%
\pgfsetroundjoin%
\pgfsetlinewidth{0.501875pt}%
\definecolor{currentstroke}{rgb}{0.800000,0.800000,0.800000}%
\pgfsetstrokecolor{currentstroke}%
\pgfsetdash{}{0pt}%
\pgfpathmoveto{\pgfqpoint{0.520798in}{1.314061in}}%
\pgfpathlineto{\pgfqpoint{1.269207in}{1.314061in}}%
\pgfusepath{stroke}%
\end{pgfscope}%
\begin{pgfscope}%
\definecolor{textcolor}{rgb}{0.150000,0.150000,0.150000}%
\pgfsetstrokecolor{textcolor}%
\pgfsetfillcolor{textcolor}%
\pgftext[x=0.155124in, y=1.275799in, left, base]{\color{textcolor}\rmfamily\fontsize{8.000000}{9.600000}\selectfont 100\%}%
\end{pgfscope}%
\begin{pgfscope}%
\definecolor{textcolor}{rgb}{0.150000,0.150000,0.150000}%
\pgfsetstrokecolor{textcolor}%
\pgfsetfillcolor{textcolor}%
\pgftext[x=0.099569in,y=0.878119in,,bottom,rotate=90.000000]{\color{textcolor}\rmfamily\fontsize{10.000000}{12.000000}\selectfont Cert. Acc.}%
\end{pgfscope}%
\begin{pgfscope}%
\pgfsetrectcap%
\pgfsetmiterjoin%
\pgfsetlinewidth{0.752812pt}%
\definecolor{currentstroke}{rgb}{0.700000,0.700000,0.700000}%
\pgfsetstrokecolor{currentstroke}%
\pgfsetdash{}{0pt}%
\pgfpathmoveto{\pgfqpoint{0.520798in}{0.442177in}}%
\pgfpathlineto{\pgfqpoint{0.520798in}{1.314061in}}%
\pgfusepath{stroke}%
\end{pgfscope}%
\begin{pgfscope}%
\pgfsetrectcap%
\pgfsetmiterjoin%
\pgfsetlinewidth{0.752812pt}%
\definecolor{currentstroke}{rgb}{0.700000,0.700000,0.700000}%
\pgfsetstrokecolor{currentstroke}%
\pgfsetdash{}{0pt}%
\pgfpathmoveto{\pgfqpoint{1.269207in}{0.442177in}}%
\pgfpathlineto{\pgfqpoint{1.269207in}{1.314061in}}%
\pgfusepath{stroke}%
\end{pgfscope}%
\begin{pgfscope}%
\pgfsetrectcap%
\pgfsetmiterjoin%
\pgfsetlinewidth{0.752812pt}%
\definecolor{currentstroke}{rgb}{0.700000,0.700000,0.700000}%
\pgfsetstrokecolor{currentstroke}%
\pgfsetdash{}{0pt}%
\pgfpathmoveto{\pgfqpoint{0.520798in}{0.442177in}}%
\pgfpathlineto{\pgfqpoint{1.269207in}{0.442177in}}%
\pgfusepath{stroke}%
\end{pgfscope}%
\begin{pgfscope}%
\pgfsetrectcap%
\pgfsetmiterjoin%
\pgfsetlinewidth{0.752812pt}%
\definecolor{currentstroke}{rgb}{0.700000,0.700000,0.700000}%
\pgfsetstrokecolor{currentstroke}%
\pgfsetdash{}{0pt}%
\pgfpathmoveto{\pgfqpoint{0.520798in}{1.314061in}}%
\pgfpathlineto{\pgfqpoint{1.269207in}{1.314061in}}%
\pgfusepath{stroke}%
\end{pgfscope}%
\begin{pgfscope}%
\pgfpathrectangle{\pgfqpoint{0.520798in}{0.442177in}}{\pgfqpoint{0.748409in}{0.871884in}}%
\pgfusepath{clip}%
\pgfsetroundcap%
\pgfsetroundjoin%
\pgfsetlinewidth{1.003750pt}%
\definecolor{currentstroke}{rgb}{0.003922,0.450980,0.698039}%
\pgfsetstrokecolor{currentstroke}%
\pgfsetdash{}{0pt}%
\pgfpathmoveto{\pgfqpoint{0.520798in}{0.935019in}}%
\pgfpathlineto{\pgfqpoint{0.562376in}{0.935019in}}%
\pgfpathlineto{\pgfqpoint{0.562376in}{0.897929in}}%
\pgfpathlineto{\pgfqpoint{0.645533in}{0.897929in}}%
\pgfpathlineto{\pgfqpoint{0.645533in}{0.853564in}}%
\pgfpathlineto{\pgfqpoint{0.728689in}{0.853564in}}%
\pgfpathlineto{\pgfqpoint{0.728689in}{0.799551in}}%
\pgfpathlineto{\pgfqpoint{0.811846in}{0.799551in}}%
\pgfpathlineto{\pgfqpoint{0.811846in}{0.723790in}}%
\pgfpathlineto{\pgfqpoint{0.895002in}{0.723790in}}%
\pgfpathlineto{\pgfqpoint{0.895002in}{0.627546in}}%
\pgfpathlineto{\pgfqpoint{0.978159in}{0.627546in}}%
\pgfpathlineto{\pgfqpoint{0.978159in}{0.525609in}}%
\pgfpathlineto{\pgfqpoint{1.061316in}{0.525609in}}%
\pgfpathlineto{\pgfqpoint{1.061316in}{0.464241in}}%
\pgfpathlineto{\pgfqpoint{1.144472in}{0.464241in}}%
\pgfpathlineto{\pgfqpoint{1.144472in}{0.447238in}}%
\pgfpathlineto{\pgfqpoint{1.227629in}{0.447238in}}%
\pgfpathlineto{\pgfqpoint{1.227629in}{0.442177in}}%
\pgfpathlineto{\pgfqpoint{1.270874in}{0.442177in}}%
\pgfusepath{stroke}%
\end{pgfscope}%
\begin{pgfscope}%
\pgfpathrectangle{\pgfqpoint{0.520798in}{0.442177in}}{\pgfqpoint{0.748409in}{0.871884in}}%
\pgfusepath{clip}%
\pgfsetbuttcap%
\pgfsetroundjoin%
\definecolor{currentfill}{rgb}{0.003922,0.450980,0.698039}%
\pgfsetfillcolor{currentfill}%
\pgfsetfillopacity{0.500000}%
\pgfsetlinewidth{0.000000pt}%
\definecolor{currentstroke}{rgb}{0.003922,0.450980,0.698039}%
\pgfsetstrokecolor{currentstroke}%
\pgfsetstrokeopacity{0.500000}%
\pgfsetdash{}{0pt}%
\pgfpathmoveto{\pgfqpoint{0.520798in}{0.948913in}}%
\pgfpathlineto{\pgfqpoint{0.520798in}{0.921125in}}%
\pgfpathlineto{\pgfqpoint{0.562376in}{0.921125in}}%
\pgfpathlineto{\pgfqpoint{0.562376in}{0.883499in}}%
\pgfpathlineto{\pgfqpoint{0.645533in}{0.883499in}}%
\pgfpathlineto{\pgfqpoint{0.645533in}{0.839560in}}%
\pgfpathlineto{\pgfqpoint{0.728689in}{0.839560in}}%
\pgfpathlineto{\pgfqpoint{0.728689in}{0.788147in}}%
\pgfpathlineto{\pgfqpoint{0.811846in}{0.788147in}}%
\pgfpathlineto{\pgfqpoint{0.811846in}{0.712298in}}%
\pgfpathlineto{\pgfqpoint{0.895002in}{0.712298in}}%
\pgfpathlineto{\pgfqpoint{0.895002in}{0.618685in}}%
\pgfpathlineto{\pgfqpoint{0.978159in}{0.618685in}}%
\pgfpathlineto{\pgfqpoint{0.978159in}{0.519540in}}%
\pgfpathlineto{\pgfqpoint{1.061316in}{0.519540in}}%
\pgfpathlineto{\pgfqpoint{1.061316in}{0.463139in}}%
\pgfpathlineto{\pgfqpoint{1.144472in}{0.463139in}}%
\pgfpathlineto{\pgfqpoint{1.144472in}{0.445024in}}%
\pgfpathlineto{\pgfqpoint{1.227629in}{0.445024in}}%
\pgfpathlineto{\pgfqpoint{1.227629in}{0.442177in}}%
\pgfpathlineto{\pgfqpoint{1.310785in}{0.442177in}}%
\pgfpathlineto{\pgfqpoint{1.310785in}{0.442177in}}%
\pgfpathlineto{\pgfqpoint{1.393942in}{0.442177in}}%
\pgfpathlineto{\pgfqpoint{1.393942in}{0.442177in}}%
\pgfpathlineto{\pgfqpoint{1.477098in}{0.442177in}}%
\pgfpathlineto{\pgfqpoint{1.477098in}{0.442177in}}%
\pgfpathlineto{\pgfqpoint{1.560255in}{0.442177in}}%
\pgfpathlineto{\pgfqpoint{1.560255in}{0.442177in}}%
\pgfpathlineto{\pgfqpoint{1.643412in}{0.442177in}}%
\pgfpathlineto{\pgfqpoint{1.643412in}{0.442177in}}%
\pgfpathlineto{\pgfqpoint{1.726568in}{0.442177in}}%
\pgfpathlineto{\pgfqpoint{1.726568in}{0.442177in}}%
\pgfpathlineto{\pgfqpoint{1.809725in}{0.442177in}}%
\pgfpathlineto{\pgfqpoint{1.809725in}{0.442177in}}%
\pgfpathlineto{\pgfqpoint{1.892881in}{0.442177in}}%
\pgfpathlineto{\pgfqpoint{1.892881in}{0.442177in}}%
\pgfpathlineto{\pgfqpoint{1.976038in}{0.442177in}}%
\pgfpathlineto{\pgfqpoint{1.976038in}{0.442177in}}%
\pgfpathlineto{\pgfqpoint{2.059194in}{0.442177in}}%
\pgfpathlineto{\pgfqpoint{2.059194in}{0.442177in}}%
\pgfpathlineto{\pgfqpoint{2.142351in}{0.442177in}}%
\pgfpathlineto{\pgfqpoint{2.142351in}{0.442177in}}%
\pgfpathlineto{\pgfqpoint{2.225508in}{0.442177in}}%
\pgfpathlineto{\pgfqpoint{2.225508in}{0.442177in}}%
\pgfpathlineto{\pgfqpoint{2.308664in}{0.442177in}}%
\pgfpathlineto{\pgfqpoint{2.308664in}{0.442177in}}%
\pgfpathlineto{\pgfqpoint{2.391821in}{0.442177in}}%
\pgfpathlineto{\pgfqpoint{2.391821in}{0.442177in}}%
\pgfpathlineto{\pgfqpoint{2.474977in}{0.442177in}}%
\pgfpathlineto{\pgfqpoint{2.474977in}{0.442177in}}%
\pgfpathlineto{\pgfqpoint{2.558134in}{0.442177in}}%
\pgfpathlineto{\pgfqpoint{2.558134in}{0.442177in}}%
\pgfpathlineto{\pgfqpoint{2.641290in}{0.442177in}}%
\pgfpathlineto{\pgfqpoint{2.641290in}{0.442177in}}%
\pgfpathlineto{\pgfqpoint{2.724447in}{0.442177in}}%
\pgfpathlineto{\pgfqpoint{2.724447in}{0.442177in}}%
\pgfpathlineto{\pgfqpoint{2.807604in}{0.442177in}}%
\pgfpathlineto{\pgfqpoint{2.807604in}{0.442177in}}%
\pgfpathlineto{\pgfqpoint{2.890760in}{0.442177in}}%
\pgfpathlineto{\pgfqpoint{2.890760in}{0.442177in}}%
\pgfpathlineto{\pgfqpoint{2.932338in}{0.442177in}}%
\pgfpathlineto{\pgfqpoint{2.932338in}{0.442177in}}%
\pgfpathlineto{\pgfqpoint{2.932338in}{0.442177in}}%
\pgfpathlineto{\pgfqpoint{2.890760in}{0.442177in}}%
\pgfpathlineto{\pgfqpoint{2.890760in}{0.442177in}}%
\pgfpathlineto{\pgfqpoint{2.807604in}{0.442177in}}%
\pgfpathlineto{\pgfqpoint{2.807604in}{0.442177in}}%
\pgfpathlineto{\pgfqpoint{2.724447in}{0.442177in}}%
\pgfpathlineto{\pgfqpoint{2.724447in}{0.442177in}}%
\pgfpathlineto{\pgfqpoint{2.641290in}{0.442177in}}%
\pgfpathlineto{\pgfqpoint{2.641290in}{0.442177in}}%
\pgfpathlineto{\pgfqpoint{2.558134in}{0.442177in}}%
\pgfpathlineto{\pgfqpoint{2.558134in}{0.442177in}}%
\pgfpathlineto{\pgfqpoint{2.474977in}{0.442177in}}%
\pgfpathlineto{\pgfqpoint{2.474977in}{0.442177in}}%
\pgfpathlineto{\pgfqpoint{2.391821in}{0.442177in}}%
\pgfpathlineto{\pgfqpoint{2.391821in}{0.442177in}}%
\pgfpathlineto{\pgfqpoint{2.308664in}{0.442177in}}%
\pgfpathlineto{\pgfqpoint{2.308664in}{0.442177in}}%
\pgfpathlineto{\pgfqpoint{2.225508in}{0.442177in}}%
\pgfpathlineto{\pgfqpoint{2.225508in}{0.442177in}}%
\pgfpathlineto{\pgfqpoint{2.142351in}{0.442177in}}%
\pgfpathlineto{\pgfqpoint{2.142351in}{0.442177in}}%
\pgfpathlineto{\pgfqpoint{2.059194in}{0.442177in}}%
\pgfpathlineto{\pgfqpoint{2.059194in}{0.442177in}}%
\pgfpathlineto{\pgfqpoint{1.976038in}{0.442177in}}%
\pgfpathlineto{\pgfqpoint{1.976038in}{0.442177in}}%
\pgfpathlineto{\pgfqpoint{1.892881in}{0.442177in}}%
\pgfpathlineto{\pgfqpoint{1.892881in}{0.442177in}}%
\pgfpathlineto{\pgfqpoint{1.809725in}{0.442177in}}%
\pgfpathlineto{\pgfqpoint{1.809725in}{0.442177in}}%
\pgfpathlineto{\pgfqpoint{1.726568in}{0.442177in}}%
\pgfpathlineto{\pgfqpoint{1.726568in}{0.442177in}}%
\pgfpathlineto{\pgfqpoint{1.643412in}{0.442177in}}%
\pgfpathlineto{\pgfqpoint{1.643412in}{0.442177in}}%
\pgfpathlineto{\pgfqpoint{1.560255in}{0.442177in}}%
\pgfpathlineto{\pgfqpoint{1.560255in}{0.442177in}}%
\pgfpathlineto{\pgfqpoint{1.477098in}{0.442177in}}%
\pgfpathlineto{\pgfqpoint{1.477098in}{0.442177in}}%
\pgfpathlineto{\pgfqpoint{1.393942in}{0.442177in}}%
\pgfpathlineto{\pgfqpoint{1.393942in}{0.442177in}}%
\pgfpathlineto{\pgfqpoint{1.310785in}{0.442177in}}%
\pgfpathlineto{\pgfqpoint{1.310785in}{0.442177in}}%
\pgfpathlineto{\pgfqpoint{1.227629in}{0.442177in}}%
\pgfpathlineto{\pgfqpoint{1.227629in}{0.449452in}}%
\pgfpathlineto{\pgfqpoint{1.144472in}{0.449452in}}%
\pgfpathlineto{\pgfqpoint{1.144472in}{0.465342in}}%
\pgfpathlineto{\pgfqpoint{1.061316in}{0.465342in}}%
\pgfpathlineto{\pgfqpoint{1.061316in}{0.531678in}}%
\pgfpathlineto{\pgfqpoint{0.978159in}{0.531678in}}%
\pgfpathlineto{\pgfqpoint{0.978159in}{0.636407in}}%
\pgfpathlineto{\pgfqpoint{0.895002in}{0.636407in}}%
\pgfpathlineto{\pgfqpoint{0.895002in}{0.735281in}}%
\pgfpathlineto{\pgfqpoint{0.811846in}{0.735281in}}%
\pgfpathlineto{\pgfqpoint{0.811846in}{0.810954in}}%
\pgfpathlineto{\pgfqpoint{0.728689in}{0.810954in}}%
\pgfpathlineto{\pgfqpoint{0.728689in}{0.867568in}}%
\pgfpathlineto{\pgfqpoint{0.645533in}{0.867568in}}%
\pgfpathlineto{\pgfqpoint{0.645533in}{0.912360in}}%
\pgfpathlineto{\pgfqpoint{0.562376in}{0.912360in}}%
\pgfpathlineto{\pgfqpoint{0.562376in}{0.948913in}}%
\pgfpathlineto{\pgfqpoint{0.520798in}{0.948913in}}%
\pgfpathlineto{\pgfqpoint{0.520798in}{0.948913in}}%
\pgfpathclose%
\pgfusepath{fill}%
\end{pgfscope}%
\begin{pgfscope}%
\pgfpathrectangle{\pgfqpoint{0.520798in}{0.442177in}}{\pgfqpoint{0.748409in}{0.871884in}}%
\pgfusepath{clip}%
\pgfsetroundcap%
\pgfsetroundjoin%
\pgfsetlinewidth{1.003750pt}%
\definecolor{currentstroke}{rgb}{0.870588,0.560784,0.019608}%
\pgfsetstrokecolor{currentstroke}%
\pgfsetdash{}{0pt}%
\pgfpathmoveto{\pgfqpoint{0.520798in}{0.923789in}}%
\pgfpathlineto{\pgfqpoint{0.562376in}{0.923789in}}%
\pgfpathlineto{\pgfqpoint{0.562376in}{0.887016in}}%
\pgfpathlineto{\pgfqpoint{0.645533in}{0.887016in}}%
\pgfpathlineto{\pgfqpoint{0.645533in}{0.849056in}}%
\pgfpathlineto{\pgfqpoint{0.728689in}{0.849056in}}%
\pgfpathlineto{\pgfqpoint{0.728689in}{0.799630in}}%
\pgfpathlineto{\pgfqpoint{0.811846in}{0.799630in}}%
\pgfpathlineto{\pgfqpoint{0.811846in}{0.721733in}}%
\pgfpathlineto{\pgfqpoint{0.895002in}{0.721733in}}%
\pgfpathlineto{\pgfqpoint{0.895002in}{0.622880in}}%
\pgfpathlineto{\pgfqpoint{0.978159in}{0.622880in}}%
\pgfpathlineto{\pgfqpoint{0.978159in}{0.524818in}}%
\pgfpathlineto{\pgfqpoint{1.061316in}{0.524818in}}%
\pgfpathlineto{\pgfqpoint{1.061316in}{0.466297in}}%
\pgfpathlineto{\pgfqpoint{1.144472in}{0.466297in}}%
\pgfpathlineto{\pgfqpoint{1.144472in}{0.449690in}}%
\pgfpathlineto{\pgfqpoint{1.227629in}{0.449690in}}%
\pgfpathlineto{\pgfqpoint{1.227629in}{0.442177in}}%
\pgfpathlineto{\pgfqpoint{1.269207in}{0.442177in}}%
\pgfusepath{stroke}%
\end{pgfscope}%
\begin{pgfscope}%
\pgfpathrectangle{\pgfqpoint{0.520798in}{0.442177in}}{\pgfqpoint{0.748409in}{0.871884in}}%
\pgfusepath{clip}%
\pgfsetbuttcap%
\pgfsetroundjoin%
\definecolor{currentfill}{rgb}{0.870588,0.560784,0.019608}%
\pgfsetfillcolor{currentfill}%
\pgfsetfillopacity{0.500000}%
\pgfsetlinewidth{0.000000pt}%
\definecolor{currentstroke}{rgb}{0.870588,0.560784,0.019608}%
\pgfsetstrokecolor{currentstroke}%
\pgfsetstrokeopacity{0.500000}%
\pgfsetdash{}{0pt}%
\pgfsys@defobject{currentmarker}{\pgfqpoint{0.520798in}{0.442177in}}{\pgfqpoint{1.269207in}{0.923789in}}{%
\pgfpathmoveto{\pgfqpoint{0.520798in}{0.923789in}}%
\pgfpathlineto{\pgfqpoint{0.520798in}{0.923789in}}%
\pgfpathlineto{\pgfqpoint{0.562376in}{0.923789in}}%
\pgfpathlineto{\pgfqpoint{0.562376in}{0.887016in}}%
\pgfpathlineto{\pgfqpoint{0.645533in}{0.887016in}}%
\pgfpathlineto{\pgfqpoint{0.645533in}{0.849056in}}%
\pgfpathlineto{\pgfqpoint{0.728689in}{0.849056in}}%
\pgfpathlineto{\pgfqpoint{0.728689in}{0.799630in}}%
\pgfpathlineto{\pgfqpoint{0.811846in}{0.799630in}}%
\pgfpathlineto{\pgfqpoint{0.811846in}{0.721733in}}%
\pgfpathlineto{\pgfqpoint{0.895002in}{0.721733in}}%
\pgfpathlineto{\pgfqpoint{0.895002in}{0.622880in}}%
\pgfpathlineto{\pgfqpoint{0.978159in}{0.622880in}}%
\pgfpathlineto{\pgfqpoint{0.978159in}{0.524818in}}%
\pgfpathlineto{\pgfqpoint{1.061316in}{0.524818in}}%
\pgfpathlineto{\pgfqpoint{1.061316in}{0.466297in}}%
\pgfpathlineto{\pgfqpoint{1.144472in}{0.466297in}}%
\pgfpathlineto{\pgfqpoint{1.144472in}{0.449690in}}%
\pgfpathlineto{\pgfqpoint{1.227629in}{0.449690in}}%
\pgfpathlineto{\pgfqpoint{1.227629in}{0.442177in}}%
\pgfpathlineto{\pgfqpoint{1.269207in}{0.442177in}}%
\pgfpathlineto{\pgfqpoint{1.269207in}{0.442177in}}%
\pgfpathlineto{\pgfqpoint{1.269207in}{0.442177in}}%
\pgfpathlineto{\pgfqpoint{1.227629in}{0.442177in}}%
\pgfpathlineto{\pgfqpoint{1.227629in}{0.449690in}}%
\pgfpathlineto{\pgfqpoint{1.144472in}{0.449690in}}%
\pgfpathlineto{\pgfqpoint{1.144472in}{0.466297in}}%
\pgfpathlineto{\pgfqpoint{1.061316in}{0.466297in}}%
\pgfpathlineto{\pgfqpoint{1.061316in}{0.524818in}}%
\pgfpathlineto{\pgfqpoint{0.978159in}{0.524818in}}%
\pgfpathlineto{\pgfqpoint{0.978159in}{0.622880in}}%
\pgfpathlineto{\pgfqpoint{0.895002in}{0.622880in}}%
\pgfpathlineto{\pgfqpoint{0.895002in}{0.721733in}}%
\pgfpathlineto{\pgfqpoint{0.811846in}{0.721733in}}%
\pgfpathlineto{\pgfqpoint{0.811846in}{0.799630in}}%
\pgfpathlineto{\pgfqpoint{0.728689in}{0.799630in}}%
\pgfpathlineto{\pgfqpoint{0.728689in}{0.849056in}}%
\pgfpathlineto{\pgfqpoint{0.645533in}{0.849056in}}%
\pgfpathlineto{\pgfqpoint{0.645533in}{0.887016in}}%
\pgfpathlineto{\pgfqpoint{0.562376in}{0.887016in}}%
\pgfpathlineto{\pgfqpoint{0.562376in}{0.923789in}}%
\pgfpathlineto{\pgfqpoint{0.520798in}{0.923789in}}%
\pgfpathlineto{\pgfqpoint{0.520798in}{0.923789in}}%
\pgfpathclose%
\pgfusepath{fill}%
}%
\begin{pgfscope}%
\pgfsys@transformshift{0.000000in}{0.000000in}%
\pgfsys@useobject{currentmarker}{}%
\end{pgfscope}%
\end{pgfscope}%
\begin{pgfscope}%
\pgfpathrectangle{\pgfqpoint{0.520798in}{0.442177in}}{\pgfqpoint{0.748409in}{0.871884in}}%
\pgfusepath{clip}%
\pgfsetroundcap%
\pgfsetroundjoin%
\pgfsetlinewidth{1.003750pt}%
\definecolor{currentstroke}{rgb}{0.007843,0.619608,0.450980}%
\pgfsetstrokecolor{currentstroke}%
\pgfsetdash{}{0pt}%
\pgfpathmoveto{\pgfqpoint{0.520798in}{0.926557in}}%
\pgfpathlineto{\pgfqpoint{0.562376in}{0.926557in}}%
\pgfpathlineto{\pgfqpoint{0.562376in}{0.888993in}}%
\pgfpathlineto{\pgfqpoint{0.645533in}{0.888993in}}%
\pgfpathlineto{\pgfqpoint{0.645533in}{0.851033in}}%
\pgfpathlineto{\pgfqpoint{0.728689in}{0.851033in}}%
\pgfpathlineto{\pgfqpoint{0.728689in}{0.801607in}}%
\pgfpathlineto{\pgfqpoint{0.811846in}{0.801607in}}%
\pgfpathlineto{\pgfqpoint{0.811846in}{0.725688in}}%
\pgfpathlineto{\pgfqpoint{0.895002in}{0.725688in}}%
\pgfpathlineto{\pgfqpoint{0.895002in}{0.625648in}}%
\pgfpathlineto{\pgfqpoint{0.978159in}{0.625648in}}%
\pgfpathlineto{\pgfqpoint{0.978159in}{0.525609in}}%
\pgfpathlineto{\pgfqpoint{1.061316in}{0.525609in}}%
\pgfpathlineto{\pgfqpoint{1.061316in}{0.467088in}}%
\pgfpathlineto{\pgfqpoint{1.144472in}{0.467088in}}%
\pgfpathlineto{\pgfqpoint{1.144472in}{0.449690in}}%
\pgfpathlineto{\pgfqpoint{1.227629in}{0.449690in}}%
\pgfpathlineto{\pgfqpoint{1.227629in}{0.442177in}}%
\pgfpathlineto{\pgfqpoint{1.269207in}{0.442177in}}%
\pgfusepath{stroke}%
\end{pgfscope}%
\begin{pgfscope}%
\pgfpathrectangle{\pgfqpoint{0.520798in}{0.442177in}}{\pgfqpoint{0.748409in}{0.871884in}}%
\pgfusepath{clip}%
\pgfsetbuttcap%
\pgfsetroundjoin%
\definecolor{currentfill}{rgb}{0.007843,0.619608,0.450980}%
\pgfsetfillcolor{currentfill}%
\pgfsetfillopacity{0.500000}%
\pgfsetlinewidth{0.000000pt}%
\definecolor{currentstroke}{rgb}{0.007843,0.619608,0.450980}%
\pgfsetstrokecolor{currentstroke}%
\pgfsetstrokeopacity{0.500000}%
\pgfsetdash{}{0pt}%
\pgfsys@defobject{currentmarker}{\pgfqpoint{0.520798in}{0.442177in}}{\pgfqpoint{1.269207in}{0.926557in}}{%
\pgfpathmoveto{\pgfqpoint{0.520798in}{0.926557in}}%
\pgfpathlineto{\pgfqpoint{0.520798in}{0.926557in}}%
\pgfpathlineto{\pgfqpoint{0.562376in}{0.926557in}}%
\pgfpathlineto{\pgfqpoint{0.562376in}{0.888993in}}%
\pgfpathlineto{\pgfqpoint{0.645533in}{0.888993in}}%
\pgfpathlineto{\pgfqpoint{0.645533in}{0.851033in}}%
\pgfpathlineto{\pgfqpoint{0.728689in}{0.851033in}}%
\pgfpathlineto{\pgfqpoint{0.728689in}{0.801607in}}%
\pgfpathlineto{\pgfqpoint{0.811846in}{0.801607in}}%
\pgfpathlineto{\pgfqpoint{0.811846in}{0.725688in}}%
\pgfpathlineto{\pgfqpoint{0.895002in}{0.725688in}}%
\pgfpathlineto{\pgfqpoint{0.895002in}{0.625648in}}%
\pgfpathlineto{\pgfqpoint{0.978159in}{0.625648in}}%
\pgfpathlineto{\pgfqpoint{0.978159in}{0.525609in}}%
\pgfpathlineto{\pgfqpoint{1.061316in}{0.525609in}}%
\pgfpathlineto{\pgfqpoint{1.061316in}{0.467088in}}%
\pgfpathlineto{\pgfqpoint{1.144472in}{0.467088in}}%
\pgfpathlineto{\pgfqpoint{1.144472in}{0.449690in}}%
\pgfpathlineto{\pgfqpoint{1.227629in}{0.449690in}}%
\pgfpathlineto{\pgfqpoint{1.227629in}{0.442177in}}%
\pgfpathlineto{\pgfqpoint{1.269207in}{0.442177in}}%
\pgfpathlineto{\pgfqpoint{1.269207in}{0.442177in}}%
\pgfpathlineto{\pgfqpoint{1.269207in}{0.442177in}}%
\pgfpathlineto{\pgfqpoint{1.227629in}{0.442177in}}%
\pgfpathlineto{\pgfqpoint{1.227629in}{0.449690in}}%
\pgfpathlineto{\pgfqpoint{1.144472in}{0.449690in}}%
\pgfpathlineto{\pgfqpoint{1.144472in}{0.467088in}}%
\pgfpathlineto{\pgfqpoint{1.061316in}{0.467088in}}%
\pgfpathlineto{\pgfqpoint{1.061316in}{0.525609in}}%
\pgfpathlineto{\pgfqpoint{0.978159in}{0.525609in}}%
\pgfpathlineto{\pgfqpoint{0.978159in}{0.625648in}}%
\pgfpathlineto{\pgfqpoint{0.895002in}{0.625648in}}%
\pgfpathlineto{\pgfqpoint{0.895002in}{0.725688in}}%
\pgfpathlineto{\pgfqpoint{0.811846in}{0.725688in}}%
\pgfpathlineto{\pgfqpoint{0.811846in}{0.801607in}}%
\pgfpathlineto{\pgfqpoint{0.728689in}{0.801607in}}%
\pgfpathlineto{\pgfqpoint{0.728689in}{0.851033in}}%
\pgfpathlineto{\pgfqpoint{0.645533in}{0.851033in}}%
\pgfpathlineto{\pgfqpoint{0.645533in}{0.888993in}}%
\pgfpathlineto{\pgfqpoint{0.562376in}{0.888993in}}%
\pgfpathlineto{\pgfqpoint{0.562376in}{0.926557in}}%
\pgfpathlineto{\pgfqpoint{0.520798in}{0.926557in}}%
\pgfpathlineto{\pgfqpoint{0.520798in}{0.926557in}}%
\pgfpathclose%
\pgfusepath{fill}%
}%
\begin{pgfscope}%
\pgfsys@transformshift{0.000000in}{0.000000in}%
\pgfsys@useobject{currentmarker}{}%
\end{pgfscope}%
\end{pgfscope}%
\begin{pgfscope}%
\definecolor{textcolor}{rgb}{0.150000,0.150000,0.150000}%
\pgfsetstrokecolor{textcolor}%
\pgfsetfillcolor{textcolor}%
\pgftext[x=0.895002in,y=1.397395in,,base]{\color{textcolor}\rmfamily\fontsize{9.000000}{10.800000}\selectfont \(\displaystyle c_A^+=1\)}%
\end{pgfscope}%
\begin{pgfscope}%
\pgfsetbuttcap%
\pgfsetmiterjoin%
\definecolor{currentfill}{rgb}{1.000000,1.000000,1.000000}%
\pgfsetfillcolor{currentfill}%
\pgfsetfillopacity{0.800000}%
\pgfsetlinewidth{1.003750pt}%
\definecolor{currentstroke}{rgb}{0.800000,0.800000,0.800000}%
\pgfsetstrokecolor{currentstroke}%
\pgfsetstrokeopacity{0.800000}%
\pgfsetdash{}{0pt}%
\pgfpathmoveto{\pgfqpoint{0.815247in}{0.638103in}}%
\pgfpathlineto{\pgfqpoint{1.220596in}{0.638103in}}%
\pgfpathlineto{\pgfqpoint{1.220596in}{1.265450in}}%
\pgfpathlineto{\pgfqpoint{0.815247in}{1.265450in}}%
\pgfpathlineto{\pgfqpoint{0.815247in}{0.638103in}}%
\pgfpathclose%
\pgfusepath{stroke,fill}%
\end{pgfscope}%
\begin{pgfscope}%
\definecolor{textcolor}{rgb}{0.150000,0.150000,0.150000}%
\pgfsetstrokecolor{textcolor}%
\pgfsetfillcolor{textcolor}%
\pgftext[x=0.946517in,y=1.113275in,left,base]{\color{textcolor}\rmfamily\fontsize{9.000000}{10.800000}\selectfont \(\displaystyle c_A^-\)}%
\end{pgfscope}%
\begin{pgfscope}%
\pgfsetroundcap%
\pgfsetroundjoin%
\pgfsetlinewidth{1.003750pt}%
\definecolor{currentstroke}{rgb}{0.003922,0.450980,0.698039}%
\pgfsetstrokecolor{currentstroke}%
\pgfsetdash{}{0pt}%
\pgfpathmoveto{\pgfqpoint{0.854136in}{1.001052in}}%
\pgfpathlineto{\pgfqpoint{0.902747in}{1.001052in}}%
\pgfpathlineto{\pgfqpoint{0.902747in}{1.001052in}}%
\pgfpathlineto{\pgfqpoint{0.999969in}{1.001052in}}%
\pgfpathlineto{\pgfqpoint{0.999969in}{1.001052in}}%
\pgfpathlineto{\pgfqpoint{1.048580in}{1.001052in}}%
\pgfusepath{stroke}%
\end{pgfscope}%
\begin{pgfscope}%
\definecolor{textcolor}{rgb}{0.150000,0.150000,0.150000}%
\pgfsetstrokecolor{textcolor}%
\pgfsetfillcolor{textcolor}%
\pgftext[x=1.126358in,y=0.967024in,left,base]{\color{textcolor}\rmfamily\fontsize{7.000000}{8.400000}\selectfont 1}%
\end{pgfscope}%
\begin{pgfscope}%
\pgfsetroundcap%
\pgfsetroundjoin%
\pgfsetlinewidth{1.003750pt}%
\definecolor{currentstroke}{rgb}{0.870588,0.560784,0.019608}%
\pgfsetstrokecolor{currentstroke}%
\pgfsetdash{}{0pt}%
\pgfpathmoveto{\pgfqpoint{0.854136in}{0.865486in}}%
\pgfpathlineto{\pgfqpoint{0.902747in}{0.865486in}}%
\pgfpathlineto{\pgfqpoint{0.902747in}{0.865486in}}%
\pgfpathlineto{\pgfqpoint{0.999969in}{0.865486in}}%
\pgfpathlineto{\pgfqpoint{0.999969in}{0.865486in}}%
\pgfpathlineto{\pgfqpoint{1.048580in}{0.865486in}}%
\pgfusepath{stroke}%
\end{pgfscope}%
\begin{pgfscope}%
\definecolor{textcolor}{rgb}{0.150000,0.150000,0.150000}%
\pgfsetstrokecolor{textcolor}%
\pgfsetfillcolor{textcolor}%
\pgftext[x=1.126358in,y=0.831458in,left,base]{\color{textcolor}\rmfamily\fontsize{7.000000}{8.400000}\selectfont 2}%
\end{pgfscope}%
\begin{pgfscope}%
\pgfsetroundcap%
\pgfsetroundjoin%
\pgfsetlinewidth{1.003750pt}%
\definecolor{currentstroke}{rgb}{0.007843,0.619608,0.450980}%
\pgfsetstrokecolor{currentstroke}%
\pgfsetdash{}{0pt}%
\pgfpathmoveto{\pgfqpoint{0.854136in}{0.729919in}}%
\pgfpathlineto{\pgfqpoint{0.902747in}{0.729919in}}%
\pgfpathlineto{\pgfqpoint{0.902747in}{0.729919in}}%
\pgfpathlineto{\pgfqpoint{0.999969in}{0.729919in}}%
\pgfpathlineto{\pgfqpoint{0.999969in}{0.729919in}}%
\pgfpathlineto{\pgfqpoint{1.048580in}{0.729919in}}%
\pgfusepath{stroke}%
\end{pgfscope}%
\begin{pgfscope}%
\definecolor{textcolor}{rgb}{0.150000,0.150000,0.150000}%
\pgfsetstrokecolor{textcolor}%
\pgfsetfillcolor{textcolor}%
\pgftext[x=1.126358in,y=0.695892in,left,base]{\color{textcolor}\rmfamily\fontsize{7.000000}{8.400000}\selectfont 4}%
\end{pgfscope}%
\end{pgfpicture}%
\makeatother%
\endgroup%

%% file: figures/experiments/graphs/piPPNP/piPPNP-Citeseer-B.pgf
\begingroup%
\makeatletter%
\begin{pgfpicture}%
\pgfpathrectangle{\pgfpointorigin}{\pgfqpoint{1.375000in}{1.581250in}}%
\pgfusepath{use as bounding box, clip}%
\begin{pgfscope}%
\pgfsetbuttcap%
\pgfsetmiterjoin%
\definecolor{currentfill}{rgb}{1.000000,1.000000,1.000000}%
\pgfsetfillcolor{currentfill}%
\pgfsetlinewidth{0.000000pt}%
\definecolor{currentstroke}{rgb}{1.000000,1.000000,1.000000}%
\pgfsetstrokecolor{currentstroke}%
\pgfsetdash{}{0pt}%
\pgfpathmoveto{\pgfqpoint{0.000000in}{0.000000in}}%
\pgfpathlineto{\pgfqpoint{1.375000in}{0.000000in}}%
\pgfpathlineto{\pgfqpoint{1.375000in}{1.581250in}}%
\pgfpathlineto{\pgfqpoint{0.000000in}{1.581250in}}%
\pgfpathlineto{\pgfqpoint{0.000000in}{0.000000in}}%
\pgfpathclose%
\pgfusepath{fill}%
\end{pgfscope}%
\begin{pgfscope}%
\pgfsetbuttcap%
\pgfsetmiterjoin%
\definecolor{currentfill}{rgb}{1.000000,1.000000,1.000000}%
\pgfsetfillcolor{currentfill}%
\pgfsetlinewidth{0.000000pt}%
\definecolor{currentstroke}{rgb}{0.000000,0.000000,0.000000}%
\pgfsetstrokecolor{currentstroke}%
\pgfsetstrokeopacity{0.000000}%
\pgfsetdash{}{0pt}%
\pgfpathmoveto{\pgfqpoint{0.520798in}{0.442177in}}%
\pgfpathlineto{\pgfqpoint{1.269207in}{0.442177in}}%
\pgfpathlineto{\pgfqpoint{1.269207in}{1.314061in}}%
\pgfpathlineto{\pgfqpoint{0.520798in}{1.314061in}}%
\pgfpathlineto{\pgfqpoint{0.520798in}{0.442177in}}%
\pgfpathclose%
\pgfusepath{fill}%
\end{pgfscope}%
\begin{pgfscope}%
\pgfpathrectangle{\pgfqpoint{0.520798in}{0.442177in}}{\pgfqpoint{0.748409in}{0.871884in}}%
\pgfusepath{clip}%
\pgfsetroundcap%
\pgfsetroundjoin%
\pgfsetlinewidth{0.501875pt}%
\definecolor{currentstroke}{rgb}{0.800000,0.800000,0.800000}%
\pgfsetstrokecolor{currentstroke}%
\pgfsetdash{}{0pt}%
\pgfpathmoveto{\pgfqpoint{0.520798in}{0.442177in}}%
\pgfpathlineto{\pgfqpoint{0.520798in}{1.314061in}}%
\pgfusepath{stroke}%
\end{pgfscope}%
\begin{pgfscope}%
\definecolor{textcolor}{rgb}{0.150000,0.150000,0.150000}%
\pgfsetstrokecolor{textcolor}%
\pgfsetfillcolor{textcolor}%
\pgftext[x=0.520798in,y=0.351899in,,top]{\color{textcolor}\rmfamily\fontsize{8.000000}{9.600000}\selectfont \(\displaystyle {0}\)}%
\end{pgfscope}%
\begin{pgfscope}%
\pgfpathrectangle{\pgfqpoint{0.520798in}{0.442177in}}{\pgfqpoint{0.748409in}{0.871884in}}%
\pgfusepath{clip}%
\pgfsetroundcap%
\pgfsetroundjoin%
\pgfsetlinewidth{0.501875pt}%
\definecolor{currentstroke}{rgb}{0.800000,0.800000,0.800000}%
\pgfsetstrokecolor{currentstroke}%
\pgfsetdash{}{0pt}%
\pgfpathmoveto{\pgfqpoint{0.853424in}{0.442177in}}%
\pgfpathlineto{\pgfqpoint{0.853424in}{1.314061in}}%
\pgfusepath{stroke}%
\end{pgfscope}%
\begin{pgfscope}%
\definecolor{textcolor}{rgb}{0.150000,0.150000,0.150000}%
\pgfsetstrokecolor{textcolor}%
\pgfsetfillcolor{textcolor}%
\pgftext[x=0.853424in,y=0.351899in,,top]{\color{textcolor}\rmfamily\fontsize{8.000000}{9.600000}\selectfont \(\displaystyle {4}\)}%
\end{pgfscope}%
\begin{pgfscope}%
\pgfpathrectangle{\pgfqpoint{0.520798in}{0.442177in}}{\pgfqpoint{0.748409in}{0.871884in}}%
\pgfusepath{clip}%
\pgfsetroundcap%
\pgfsetroundjoin%
\pgfsetlinewidth{0.501875pt}%
\definecolor{currentstroke}{rgb}{0.800000,0.800000,0.800000}%
\pgfsetstrokecolor{currentstroke}%
\pgfsetdash{}{0pt}%
\pgfpathmoveto{\pgfqpoint{1.269207in}{0.442177in}}%
\pgfpathlineto{\pgfqpoint{1.269207in}{1.314061in}}%
\pgfusepath{stroke}%
\end{pgfscope}%
\begin{pgfscope}%
\definecolor{textcolor}{rgb}{0.150000,0.150000,0.150000}%
\pgfsetstrokecolor{textcolor}%
\pgfsetfillcolor{textcolor}%
\pgftext[x=1.269207in,y=0.351899in,,top]{\color{textcolor}\rmfamily\fontsize{8.000000}{9.600000}\selectfont \(\displaystyle {9}\)}%
\end{pgfscope}%
\begin{pgfscope}%
\definecolor{textcolor}{rgb}{0.150000,0.150000,0.150000}%
\pgfsetstrokecolor{textcolor}%
\pgfsetfillcolor{textcolor}%
\pgftext[x=0.895002in,y=0.198219in,,top]{\color{textcolor}\rmfamily\fontsize{10.000000}{12.000000}\selectfont attack strength}%
\end{pgfscope}%
\begin{pgfscope}%
\pgfpathrectangle{\pgfqpoint{0.520798in}{0.442177in}}{\pgfqpoint{0.748409in}{0.871884in}}%
\pgfusepath{clip}%
\pgfsetroundcap%
\pgfsetroundjoin%
\pgfsetlinewidth{0.501875pt}%
\definecolor{currentstroke}{rgb}{0.800000,0.800000,0.800000}%
\pgfsetstrokecolor{currentstroke}%
\pgfsetdash{}{0pt}%
\pgfpathmoveto{\pgfqpoint{0.520798in}{0.442177in}}%
\pgfpathlineto{\pgfqpoint{1.269207in}{0.442177in}}%
\pgfusepath{stroke}%
\end{pgfscope}%
\begin{pgfscope}%
\definecolor{textcolor}{rgb}{0.150000,0.150000,0.150000}%
\pgfsetstrokecolor{textcolor}%
\pgfsetfillcolor{textcolor}%
\pgftext[x=0.273151in, y=0.403915in, left, base]{\color{textcolor}\rmfamily\fontsize{8.000000}{9.600000}\selectfont 0\%}%
\end{pgfscope}%
\begin{pgfscope}%
\pgfpathrectangle{\pgfqpoint{0.520798in}{0.442177in}}{\pgfqpoint{0.748409in}{0.871884in}}%
\pgfusepath{clip}%
\pgfsetroundcap%
\pgfsetroundjoin%
\pgfsetlinewidth{0.501875pt}%
\definecolor{currentstroke}{rgb}{0.800000,0.800000,0.800000}%
\pgfsetstrokecolor{currentstroke}%
\pgfsetdash{}{0pt}%
\pgfpathmoveto{\pgfqpoint{0.520798in}{0.660148in}}%
\pgfpathlineto{\pgfqpoint{1.269207in}{0.660148in}}%
\pgfusepath{stroke}%
\end{pgfscope}%
\begin{pgfscope}%
\definecolor{textcolor}{rgb}{0.150000,0.150000,0.150000}%
\pgfsetstrokecolor{textcolor}%
\pgfsetfillcolor{textcolor}%
\pgftext[x=0.214138in, y=0.621886in, left, base]{\color{textcolor}\rmfamily\fontsize{8.000000}{9.600000}\selectfont 25\%}%
\end{pgfscope}%
\begin{pgfscope}%
\pgfpathrectangle{\pgfqpoint{0.520798in}{0.442177in}}{\pgfqpoint{0.748409in}{0.871884in}}%
\pgfusepath{clip}%
\pgfsetroundcap%
\pgfsetroundjoin%
\pgfsetlinewidth{0.501875pt}%
\definecolor{currentstroke}{rgb}{0.800000,0.800000,0.800000}%
\pgfsetstrokecolor{currentstroke}%
\pgfsetdash{}{0pt}%
\pgfpathmoveto{\pgfqpoint{0.520798in}{0.878119in}}%
\pgfpathlineto{\pgfqpoint{1.269207in}{0.878119in}}%
\pgfusepath{stroke}%
\end{pgfscope}%
\begin{pgfscope}%
\definecolor{textcolor}{rgb}{0.150000,0.150000,0.150000}%
\pgfsetstrokecolor{textcolor}%
\pgfsetfillcolor{textcolor}%
\pgftext[x=0.214138in, y=0.839857in, left, base]{\color{textcolor}\rmfamily\fontsize{8.000000}{9.600000}\selectfont 50\%}%
\end{pgfscope}%
\begin{pgfscope}%
\pgfpathrectangle{\pgfqpoint{0.520798in}{0.442177in}}{\pgfqpoint{0.748409in}{0.871884in}}%
\pgfusepath{clip}%
\pgfsetroundcap%
\pgfsetroundjoin%
\pgfsetlinewidth{0.501875pt}%
\definecolor{currentstroke}{rgb}{0.800000,0.800000,0.800000}%
\pgfsetstrokecolor{currentstroke}%
\pgfsetdash{}{0pt}%
\pgfpathmoveto{\pgfqpoint{0.520798in}{1.096090in}}%
\pgfpathlineto{\pgfqpoint{1.269207in}{1.096090in}}%
\pgfusepath{stroke}%
\end{pgfscope}%
\begin{pgfscope}%
\definecolor{textcolor}{rgb}{0.150000,0.150000,0.150000}%
\pgfsetstrokecolor{textcolor}%
\pgfsetfillcolor{textcolor}%
\pgftext[x=0.214138in, y=1.057828in, left, base]{\color{textcolor}\rmfamily\fontsize{8.000000}{9.600000}\selectfont 75\%}%
\end{pgfscope}%
\begin{pgfscope}%
\pgfpathrectangle{\pgfqpoint{0.520798in}{0.442177in}}{\pgfqpoint{0.748409in}{0.871884in}}%
\pgfusepath{clip}%
\pgfsetroundcap%
\pgfsetroundjoin%
\pgfsetlinewidth{0.501875pt}%
\definecolor{currentstroke}{rgb}{0.800000,0.800000,0.800000}%
\pgfsetstrokecolor{currentstroke}%
\pgfsetdash{}{0pt}%
\pgfpathmoveto{\pgfqpoint{0.520798in}{1.314061in}}%
\pgfpathlineto{\pgfqpoint{1.269207in}{1.314061in}}%
\pgfusepath{stroke}%
\end{pgfscope}%
\begin{pgfscope}%
\definecolor{textcolor}{rgb}{0.150000,0.150000,0.150000}%
\pgfsetstrokecolor{textcolor}%
\pgfsetfillcolor{textcolor}%
\pgftext[x=0.155124in, y=1.275799in, left, base]{\color{textcolor}\rmfamily\fontsize{8.000000}{9.600000}\selectfont 100\%}%
\end{pgfscope}%
\begin{pgfscope}%
\definecolor{textcolor}{rgb}{0.150000,0.150000,0.150000}%
\pgfsetstrokecolor{textcolor}%
\pgfsetfillcolor{textcolor}%
\pgftext[x=0.099569in,y=0.878119in,,bottom,rotate=90.000000]{\color{textcolor}\rmfamily\fontsize{10.000000}{12.000000}\selectfont Cert. Acc.}%
\end{pgfscope}%
\begin{pgfscope}%
\pgfsetrectcap%
\pgfsetmiterjoin%
\pgfsetlinewidth{0.752812pt}%
\definecolor{currentstroke}{rgb}{0.700000,0.700000,0.700000}%
\pgfsetstrokecolor{currentstroke}%
\pgfsetdash{}{0pt}%
\pgfpathmoveto{\pgfqpoint{0.520798in}{0.442177in}}%
\pgfpathlineto{\pgfqpoint{0.520798in}{1.314061in}}%
\pgfusepath{stroke}%
\end{pgfscope}%
\begin{pgfscope}%
\pgfsetrectcap%
\pgfsetmiterjoin%
\pgfsetlinewidth{0.752812pt}%
\definecolor{currentstroke}{rgb}{0.700000,0.700000,0.700000}%
\pgfsetstrokecolor{currentstroke}%
\pgfsetdash{}{0pt}%
\pgfpathmoveto{\pgfqpoint{1.269207in}{0.442177in}}%
\pgfpathlineto{\pgfqpoint{1.269207in}{1.314061in}}%
\pgfusepath{stroke}%
\end{pgfscope}%
\begin{pgfscope}%
\pgfsetrectcap%
\pgfsetmiterjoin%
\pgfsetlinewidth{0.752812pt}%
\definecolor{currentstroke}{rgb}{0.700000,0.700000,0.700000}%
\pgfsetstrokecolor{currentstroke}%
\pgfsetdash{}{0pt}%
\pgfpathmoveto{\pgfqpoint{0.520798in}{0.442177in}}%
\pgfpathlineto{\pgfqpoint{1.269207in}{0.442177in}}%
\pgfusepath{stroke}%
\end{pgfscope}%
\begin{pgfscope}%
\pgfsetrectcap%
\pgfsetmiterjoin%
\pgfsetlinewidth{0.752812pt}%
\definecolor{currentstroke}{rgb}{0.700000,0.700000,0.700000}%
\pgfsetstrokecolor{currentstroke}%
\pgfsetdash{}{0pt}%
\pgfpathmoveto{\pgfqpoint{0.520798in}{1.314061in}}%
\pgfpathlineto{\pgfqpoint{1.269207in}{1.314061in}}%
\pgfusepath{stroke}%
\end{pgfscope}%
\begin{pgfscope}%
\pgfpathrectangle{\pgfqpoint{0.520798in}{0.442177in}}{\pgfqpoint{0.748409in}{0.871884in}}%
\pgfusepath{clip}%
\pgfsetroundcap%
\pgfsetroundjoin%
\pgfsetlinewidth{1.003750pt}%
\definecolor{currentstroke}{rgb}{0.003922,0.450980,0.698039}%
\pgfsetstrokecolor{currentstroke}%
\pgfsetdash{}{0pt}%
\pgfpathmoveto{\pgfqpoint{0.520798in}{0.885447in}}%
\pgfpathlineto{\pgfqpoint{0.562376in}{0.885447in}}%
\pgfpathlineto{\pgfqpoint{0.562376in}{0.854931in}}%
\pgfpathlineto{\pgfqpoint{0.645533in}{0.854931in}}%
\pgfpathlineto{\pgfqpoint{0.645533in}{0.823209in}}%
\pgfpathlineto{\pgfqpoint{0.728689in}{0.823209in}}%
\pgfpathlineto{\pgfqpoint{0.728689in}{0.790745in}}%
\pgfpathlineto{\pgfqpoint{0.811846in}{0.790745in}}%
\pgfpathlineto{\pgfqpoint{0.811846in}{0.741307in}}%
\pgfpathlineto{\pgfqpoint{0.895002in}{0.741307in}}%
\pgfpathlineto{\pgfqpoint{0.895002in}{0.673783in}}%
\pgfpathlineto{\pgfqpoint{0.978159in}{0.673783in}}%
\pgfpathlineto{\pgfqpoint{0.978159in}{0.594107in}}%
\pgfpathlineto{\pgfqpoint{1.061316in}{0.594107in}}%
\pgfpathlineto{\pgfqpoint{1.061316in}{0.507661in}}%
\pgfpathlineto{\pgfqpoint{1.144472in}{0.507661in}}%
\pgfpathlineto{\pgfqpoint{1.144472in}{0.455255in}}%
\pgfpathlineto{\pgfqpoint{1.227629in}{0.455255in}}%
\pgfpathlineto{\pgfqpoint{1.227629in}{0.442177in}}%
\pgfpathlineto{\pgfqpoint{1.270874in}{0.442177in}}%
\pgfusepath{stroke}%
\end{pgfscope}%
\begin{pgfscope}%
\pgfpathrectangle{\pgfqpoint{0.520798in}{0.442177in}}{\pgfqpoint{0.748409in}{0.871884in}}%
\pgfusepath{clip}%
\pgfsetbuttcap%
\pgfsetroundjoin%
\definecolor{currentfill}{rgb}{0.003922,0.450980,0.698039}%
\pgfsetfillcolor{currentfill}%
\pgfsetfillopacity{0.500000}%
\pgfsetlinewidth{0.000000pt}%
\definecolor{currentstroke}{rgb}{0.003922,0.450980,0.698039}%
\pgfsetstrokecolor{currentstroke}%
\pgfsetstrokeopacity{0.500000}%
\pgfsetdash{}{0pt}%
\pgfpathmoveto{\pgfqpoint{0.520798in}{0.898744in}}%
\pgfpathlineto{\pgfqpoint{0.520798in}{0.872149in}}%
\pgfpathlineto{\pgfqpoint{0.562376in}{0.872149in}}%
\pgfpathlineto{\pgfqpoint{0.562376in}{0.842812in}}%
\pgfpathlineto{\pgfqpoint{0.645533in}{0.842812in}}%
\pgfpathlineto{\pgfqpoint{0.645533in}{0.807228in}}%
\pgfpathlineto{\pgfqpoint{0.728689in}{0.807228in}}%
\pgfpathlineto{\pgfqpoint{0.728689in}{0.776979in}}%
\pgfpathlineto{\pgfqpoint{0.811846in}{0.776979in}}%
\pgfpathlineto{\pgfqpoint{0.811846in}{0.730042in}}%
\pgfpathlineto{\pgfqpoint{0.895002in}{0.730042in}}%
\pgfpathlineto{\pgfqpoint{0.895002in}{0.667647in}}%
\pgfpathlineto{\pgfqpoint{0.978159in}{0.667647in}}%
\pgfpathlineto{\pgfqpoint{0.978159in}{0.589634in}}%
\pgfpathlineto{\pgfqpoint{1.061316in}{0.589634in}}%
\pgfpathlineto{\pgfqpoint{1.061316in}{0.499578in}}%
\pgfpathlineto{\pgfqpoint{1.144472in}{0.499578in}}%
\pgfpathlineto{\pgfqpoint{1.144472in}{0.453692in}}%
\pgfpathlineto{\pgfqpoint{1.227629in}{0.453692in}}%
\pgfpathlineto{\pgfqpoint{1.227629in}{0.442177in}}%
\pgfpathlineto{\pgfqpoint{1.310785in}{0.442177in}}%
\pgfpathlineto{\pgfqpoint{1.310785in}{0.442177in}}%
\pgfpathlineto{\pgfqpoint{1.393942in}{0.442177in}}%
\pgfpathlineto{\pgfqpoint{1.393942in}{0.442177in}}%
\pgfpathlineto{\pgfqpoint{1.477098in}{0.442177in}}%
\pgfpathlineto{\pgfqpoint{1.477098in}{0.442177in}}%
\pgfpathlineto{\pgfqpoint{1.560255in}{0.442177in}}%
\pgfpathlineto{\pgfqpoint{1.560255in}{0.442177in}}%
\pgfpathlineto{\pgfqpoint{1.643412in}{0.442177in}}%
\pgfpathlineto{\pgfqpoint{1.643412in}{0.442177in}}%
\pgfpathlineto{\pgfqpoint{1.726568in}{0.442177in}}%
\pgfpathlineto{\pgfqpoint{1.726568in}{0.442177in}}%
\pgfpathlineto{\pgfqpoint{1.809725in}{0.442177in}}%
\pgfpathlineto{\pgfqpoint{1.809725in}{0.442177in}}%
\pgfpathlineto{\pgfqpoint{1.892881in}{0.442177in}}%
\pgfpathlineto{\pgfqpoint{1.892881in}{0.442177in}}%
\pgfpathlineto{\pgfqpoint{1.976038in}{0.442177in}}%
\pgfpathlineto{\pgfqpoint{1.976038in}{0.442177in}}%
\pgfpathlineto{\pgfqpoint{2.059194in}{0.442177in}}%
\pgfpathlineto{\pgfqpoint{2.059194in}{0.442177in}}%
\pgfpathlineto{\pgfqpoint{2.142351in}{0.442177in}}%
\pgfpathlineto{\pgfqpoint{2.142351in}{0.442177in}}%
\pgfpathlineto{\pgfqpoint{2.225508in}{0.442177in}}%
\pgfpathlineto{\pgfqpoint{2.225508in}{0.442177in}}%
\pgfpathlineto{\pgfqpoint{2.308664in}{0.442177in}}%
\pgfpathlineto{\pgfqpoint{2.308664in}{0.442177in}}%
\pgfpathlineto{\pgfqpoint{2.391821in}{0.442177in}}%
\pgfpathlineto{\pgfqpoint{2.391821in}{0.442177in}}%
\pgfpathlineto{\pgfqpoint{2.474977in}{0.442177in}}%
\pgfpathlineto{\pgfqpoint{2.474977in}{0.442177in}}%
\pgfpathlineto{\pgfqpoint{2.558134in}{0.442177in}}%
\pgfpathlineto{\pgfqpoint{2.558134in}{0.442177in}}%
\pgfpathlineto{\pgfqpoint{2.641290in}{0.442177in}}%
\pgfpathlineto{\pgfqpoint{2.641290in}{0.442177in}}%
\pgfpathlineto{\pgfqpoint{2.724447in}{0.442177in}}%
\pgfpathlineto{\pgfqpoint{2.724447in}{0.442177in}}%
\pgfpathlineto{\pgfqpoint{2.807604in}{0.442177in}}%
\pgfpathlineto{\pgfqpoint{2.807604in}{0.442177in}}%
\pgfpathlineto{\pgfqpoint{2.890760in}{0.442177in}}%
\pgfpathlineto{\pgfqpoint{2.890760in}{0.442177in}}%
\pgfpathlineto{\pgfqpoint{2.932338in}{0.442177in}}%
\pgfpathlineto{\pgfqpoint{2.932338in}{0.442177in}}%
\pgfpathlineto{\pgfqpoint{2.932338in}{0.442177in}}%
\pgfpathlineto{\pgfqpoint{2.890760in}{0.442177in}}%
\pgfpathlineto{\pgfqpoint{2.890760in}{0.442177in}}%
\pgfpathlineto{\pgfqpoint{2.807604in}{0.442177in}}%
\pgfpathlineto{\pgfqpoint{2.807604in}{0.442177in}}%
\pgfpathlineto{\pgfqpoint{2.724447in}{0.442177in}}%
\pgfpathlineto{\pgfqpoint{2.724447in}{0.442177in}}%
\pgfpathlineto{\pgfqpoint{2.641290in}{0.442177in}}%
\pgfpathlineto{\pgfqpoint{2.641290in}{0.442177in}}%
\pgfpathlineto{\pgfqpoint{2.558134in}{0.442177in}}%
\pgfpathlineto{\pgfqpoint{2.558134in}{0.442177in}}%
\pgfpathlineto{\pgfqpoint{2.474977in}{0.442177in}}%
\pgfpathlineto{\pgfqpoint{2.474977in}{0.442177in}}%
\pgfpathlineto{\pgfqpoint{2.391821in}{0.442177in}}%
\pgfpathlineto{\pgfqpoint{2.391821in}{0.442177in}}%
\pgfpathlineto{\pgfqpoint{2.308664in}{0.442177in}}%
\pgfpathlineto{\pgfqpoint{2.308664in}{0.442177in}}%
\pgfpathlineto{\pgfqpoint{2.225508in}{0.442177in}}%
\pgfpathlineto{\pgfqpoint{2.225508in}{0.442177in}}%
\pgfpathlineto{\pgfqpoint{2.142351in}{0.442177in}}%
\pgfpathlineto{\pgfqpoint{2.142351in}{0.442177in}}%
\pgfpathlineto{\pgfqpoint{2.059194in}{0.442177in}}%
\pgfpathlineto{\pgfqpoint{2.059194in}{0.442177in}}%
\pgfpathlineto{\pgfqpoint{1.976038in}{0.442177in}}%
\pgfpathlineto{\pgfqpoint{1.976038in}{0.442177in}}%
\pgfpathlineto{\pgfqpoint{1.892881in}{0.442177in}}%
\pgfpathlineto{\pgfqpoint{1.892881in}{0.442177in}}%
\pgfpathlineto{\pgfqpoint{1.809725in}{0.442177in}}%
\pgfpathlineto{\pgfqpoint{1.809725in}{0.442177in}}%
\pgfpathlineto{\pgfqpoint{1.726568in}{0.442177in}}%
\pgfpathlineto{\pgfqpoint{1.726568in}{0.442177in}}%
\pgfpathlineto{\pgfqpoint{1.643412in}{0.442177in}}%
\pgfpathlineto{\pgfqpoint{1.643412in}{0.442177in}}%
\pgfpathlineto{\pgfqpoint{1.560255in}{0.442177in}}%
\pgfpathlineto{\pgfqpoint{1.560255in}{0.442177in}}%
\pgfpathlineto{\pgfqpoint{1.477098in}{0.442177in}}%
\pgfpathlineto{\pgfqpoint{1.477098in}{0.442177in}}%
\pgfpathlineto{\pgfqpoint{1.393942in}{0.442177in}}%
\pgfpathlineto{\pgfqpoint{1.393942in}{0.442177in}}%
\pgfpathlineto{\pgfqpoint{1.310785in}{0.442177in}}%
\pgfpathlineto{\pgfqpoint{1.310785in}{0.442177in}}%
\pgfpathlineto{\pgfqpoint{1.227629in}{0.442177in}}%
\pgfpathlineto{\pgfqpoint{1.227629in}{0.456818in}}%
\pgfpathlineto{\pgfqpoint{1.144472in}{0.456818in}}%
\pgfpathlineto{\pgfqpoint{1.144472in}{0.515744in}}%
\pgfpathlineto{\pgfqpoint{1.061316in}{0.515744in}}%
\pgfpathlineto{\pgfqpoint{1.061316in}{0.598581in}}%
\pgfpathlineto{\pgfqpoint{0.978159in}{0.598581in}}%
\pgfpathlineto{\pgfqpoint{0.978159in}{0.679918in}}%
\pgfpathlineto{\pgfqpoint{0.895002in}{0.679918in}}%
\pgfpathlineto{\pgfqpoint{0.895002in}{0.752572in}}%
\pgfpathlineto{\pgfqpoint{0.811846in}{0.752572in}}%
\pgfpathlineto{\pgfqpoint{0.811846in}{0.804511in}}%
\pgfpathlineto{\pgfqpoint{0.728689in}{0.804511in}}%
\pgfpathlineto{\pgfqpoint{0.728689in}{0.839190in}}%
\pgfpathlineto{\pgfqpoint{0.645533in}{0.839190in}}%
\pgfpathlineto{\pgfqpoint{0.645533in}{0.867049in}}%
\pgfpathlineto{\pgfqpoint{0.562376in}{0.867049in}}%
\pgfpathlineto{\pgfqpoint{0.562376in}{0.898744in}}%
\pgfpathlineto{\pgfqpoint{0.520798in}{0.898744in}}%
\pgfpathlineto{\pgfqpoint{0.520798in}{0.898744in}}%
\pgfpathclose%
\pgfusepath{fill}%
\end{pgfscope}%
\begin{pgfscope}%
\pgfpathrectangle{\pgfqpoint{0.520798in}{0.442177in}}{\pgfqpoint{0.748409in}{0.871884in}}%
\pgfusepath{clip}%
\pgfsetroundcap%
\pgfsetroundjoin%
\pgfsetlinewidth{1.003750pt}%
\definecolor{currentstroke}{rgb}{0.870588,0.560784,0.019608}%
\pgfsetstrokecolor{currentstroke}%
\pgfsetdash{}{0pt}%
\pgfpathmoveto{\pgfqpoint{0.520798in}{0.966699in}}%
\pgfpathlineto{\pgfqpoint{0.562376in}{0.966699in}}%
\pgfpathlineto{\pgfqpoint{0.562376in}{0.945829in}}%
\pgfpathlineto{\pgfqpoint{0.645533in}{0.945829in}}%
\pgfpathlineto{\pgfqpoint{0.645533in}{0.921250in}}%
\pgfpathlineto{\pgfqpoint{0.728689in}{0.921250in}}%
\pgfpathlineto{\pgfqpoint{0.728689in}{0.893887in}}%
\pgfpathlineto{\pgfqpoint{0.811846in}{0.893887in}}%
\pgfpathlineto{\pgfqpoint{0.811846in}{0.859105in}}%
\pgfpathlineto{\pgfqpoint{0.895002in}{0.859105in}}%
\pgfpathlineto{\pgfqpoint{0.895002in}{0.799278in}}%
\pgfpathlineto{\pgfqpoint{0.978159in}{0.799278in}}%
\pgfpathlineto{\pgfqpoint{0.978159in}{0.722757in}}%
\pgfpathlineto{\pgfqpoint{1.061316in}{0.722757in}}%
\pgfpathlineto{\pgfqpoint{1.061316in}{0.628612in}}%
\pgfpathlineto{\pgfqpoint{1.144472in}{0.628612in}}%
\pgfpathlineto{\pgfqpoint{1.144472in}{0.534467in}}%
\pgfpathlineto{\pgfqpoint{1.227629in}{0.534467in}}%
\pgfpathlineto{\pgfqpoint{1.227629in}{0.467220in}}%
\pgfpathlineto{\pgfqpoint{1.269207in}{0.467220in}}%
\pgfusepath{stroke}%
\end{pgfscope}%
\begin{pgfscope}%
\pgfpathrectangle{\pgfqpoint{0.520798in}{0.442177in}}{\pgfqpoint{0.748409in}{0.871884in}}%
\pgfusepath{clip}%
\pgfsetbuttcap%
\pgfsetroundjoin%
\definecolor{currentfill}{rgb}{0.870588,0.560784,0.019608}%
\pgfsetfillcolor{currentfill}%
\pgfsetfillopacity{0.500000}%
\pgfsetlinewidth{0.000000pt}%
\definecolor{currentstroke}{rgb}{0.870588,0.560784,0.019608}%
\pgfsetstrokecolor{currentstroke}%
\pgfsetstrokeopacity{0.500000}%
\pgfsetdash{}{0pt}%
\pgfsys@defobject{currentmarker}{\pgfqpoint{0.520798in}{0.467220in}}{\pgfqpoint{1.269207in}{0.966699in}}{%
\pgfpathmoveto{\pgfqpoint{0.520798in}{0.966699in}}%
\pgfpathlineto{\pgfqpoint{0.520798in}{0.966699in}}%
\pgfpathlineto{\pgfqpoint{0.562376in}{0.966699in}}%
\pgfpathlineto{\pgfqpoint{0.562376in}{0.945829in}}%
\pgfpathlineto{\pgfqpoint{0.645533in}{0.945829in}}%
\pgfpathlineto{\pgfqpoint{0.645533in}{0.921250in}}%
\pgfpathlineto{\pgfqpoint{0.728689in}{0.921250in}}%
\pgfpathlineto{\pgfqpoint{0.728689in}{0.893887in}}%
\pgfpathlineto{\pgfqpoint{0.811846in}{0.893887in}}%
\pgfpathlineto{\pgfqpoint{0.811846in}{0.859105in}}%
\pgfpathlineto{\pgfqpoint{0.895002in}{0.859105in}}%
\pgfpathlineto{\pgfqpoint{0.895002in}{0.799278in}}%
\pgfpathlineto{\pgfqpoint{0.978159in}{0.799278in}}%
\pgfpathlineto{\pgfqpoint{0.978159in}{0.722757in}}%
\pgfpathlineto{\pgfqpoint{1.061316in}{0.722757in}}%
\pgfpathlineto{\pgfqpoint{1.061316in}{0.628612in}}%
\pgfpathlineto{\pgfqpoint{1.144472in}{0.628612in}}%
\pgfpathlineto{\pgfqpoint{1.144472in}{0.534467in}}%
\pgfpathlineto{\pgfqpoint{1.227629in}{0.534467in}}%
\pgfpathlineto{\pgfqpoint{1.227629in}{0.467220in}}%
\pgfpathlineto{\pgfqpoint{1.269207in}{0.467220in}}%
\pgfpathlineto{\pgfqpoint{1.269207in}{0.467220in}}%
\pgfpathlineto{\pgfqpoint{1.269207in}{0.467220in}}%
\pgfpathlineto{\pgfqpoint{1.227629in}{0.467220in}}%
\pgfpathlineto{\pgfqpoint{1.227629in}{0.534467in}}%
\pgfpathlineto{\pgfqpoint{1.144472in}{0.534467in}}%
\pgfpathlineto{\pgfqpoint{1.144472in}{0.628612in}}%
\pgfpathlineto{\pgfqpoint{1.061316in}{0.628612in}}%
\pgfpathlineto{\pgfqpoint{1.061316in}{0.722757in}}%
\pgfpathlineto{\pgfqpoint{0.978159in}{0.722757in}}%
\pgfpathlineto{\pgfqpoint{0.978159in}{0.799278in}}%
\pgfpathlineto{\pgfqpoint{0.895002in}{0.799278in}}%
\pgfpathlineto{\pgfqpoint{0.895002in}{0.859105in}}%
\pgfpathlineto{\pgfqpoint{0.811846in}{0.859105in}}%
\pgfpathlineto{\pgfqpoint{0.811846in}{0.893887in}}%
\pgfpathlineto{\pgfqpoint{0.728689in}{0.893887in}}%
\pgfpathlineto{\pgfqpoint{0.728689in}{0.921250in}}%
\pgfpathlineto{\pgfqpoint{0.645533in}{0.921250in}}%
\pgfpathlineto{\pgfqpoint{0.645533in}{0.945829in}}%
\pgfpathlineto{\pgfqpoint{0.562376in}{0.945829in}}%
\pgfpathlineto{\pgfqpoint{0.562376in}{0.966699in}}%
\pgfpathlineto{\pgfqpoint{0.520798in}{0.966699in}}%
\pgfpathlineto{\pgfqpoint{0.520798in}{0.966699in}}%
\pgfpathclose%
\pgfusepath{fill}%
}%
\begin{pgfscope}%
\pgfsys@transformshift{0.000000in}{0.000000in}%
\pgfsys@useobject{currentmarker}{}%
\end{pgfscope}%
\end{pgfscope}%
\begin{pgfscope}%
\pgfpathrectangle{\pgfqpoint{0.520798in}{0.442177in}}{\pgfqpoint{0.748409in}{0.871884in}}%
\pgfusepath{clip}%
\pgfsetroundcap%
\pgfsetroundjoin%
\pgfsetlinewidth{1.003750pt}%
\definecolor{currentstroke}{rgb}{0.007843,0.619608,0.450980}%
\pgfsetstrokecolor{currentstroke}%
\pgfsetdash{}{0pt}%
\pgfpathmoveto{\pgfqpoint{0.520798in}{0.998699in}}%
\pgfpathlineto{\pgfqpoint{0.562376in}{0.998699in}}%
\pgfpathlineto{\pgfqpoint{0.562376in}{0.988032in}}%
\pgfpathlineto{\pgfqpoint{0.645533in}{0.988032in}}%
\pgfpathlineto{\pgfqpoint{0.645533in}{0.981076in}}%
\pgfpathlineto{\pgfqpoint{0.728689in}{0.981076in}}%
\pgfpathlineto{\pgfqpoint{0.728689in}{0.967626in}}%
\pgfpathlineto{\pgfqpoint{0.811846in}{0.967626in}}%
\pgfpathlineto{\pgfqpoint{0.811846in}{0.950003in}}%
\pgfpathlineto{\pgfqpoint{0.895002in}{0.950003in}}%
\pgfpathlineto{\pgfqpoint{0.895002in}{0.919858in}}%
\pgfpathlineto{\pgfqpoint{0.978159in}{0.919858in}}%
\pgfpathlineto{\pgfqpoint{0.978159in}{0.880438in}}%
\pgfpathlineto{\pgfqpoint{1.061316in}{0.880438in}}%
\pgfpathlineto{\pgfqpoint{1.061316in}{0.811800in}}%
\pgfpathlineto{\pgfqpoint{1.144472in}{0.811800in}}%
\pgfpathlineto{\pgfqpoint{1.144472in}{0.727394in}}%
\pgfpathlineto{\pgfqpoint{1.227629in}{0.727394in}}%
\pgfpathlineto{\pgfqpoint{1.227629in}{0.634641in}}%
\pgfpathlineto{\pgfqpoint{1.269207in}{0.634641in}}%
\pgfusepath{stroke}%
\end{pgfscope}%
\begin{pgfscope}%
\pgfpathrectangle{\pgfqpoint{0.520798in}{0.442177in}}{\pgfqpoint{0.748409in}{0.871884in}}%
\pgfusepath{clip}%
\pgfsetbuttcap%
\pgfsetroundjoin%
\definecolor{currentfill}{rgb}{0.007843,0.619608,0.450980}%
\pgfsetfillcolor{currentfill}%
\pgfsetfillopacity{0.500000}%
\pgfsetlinewidth{0.000000pt}%
\definecolor{currentstroke}{rgb}{0.007843,0.619608,0.450980}%
\pgfsetstrokecolor{currentstroke}%
\pgfsetstrokeopacity{0.500000}%
\pgfsetdash{}{0pt}%
\pgfsys@defobject{currentmarker}{\pgfqpoint{0.520798in}{0.634641in}}{\pgfqpoint{1.269207in}{0.998699in}}{%
\pgfpathmoveto{\pgfqpoint{0.520798in}{0.998699in}}%
\pgfpathlineto{\pgfqpoint{0.520798in}{0.998699in}}%
\pgfpathlineto{\pgfqpoint{0.562376in}{0.998699in}}%
\pgfpathlineto{\pgfqpoint{0.562376in}{0.988032in}}%
\pgfpathlineto{\pgfqpoint{0.645533in}{0.988032in}}%
\pgfpathlineto{\pgfqpoint{0.645533in}{0.981076in}}%
\pgfpathlineto{\pgfqpoint{0.728689in}{0.981076in}}%
\pgfpathlineto{\pgfqpoint{0.728689in}{0.967626in}}%
\pgfpathlineto{\pgfqpoint{0.811846in}{0.967626in}}%
\pgfpathlineto{\pgfqpoint{0.811846in}{0.950003in}}%
\pgfpathlineto{\pgfqpoint{0.895002in}{0.950003in}}%
\pgfpathlineto{\pgfqpoint{0.895002in}{0.919858in}}%
\pgfpathlineto{\pgfqpoint{0.978159in}{0.919858in}}%
\pgfpathlineto{\pgfqpoint{0.978159in}{0.880438in}}%
\pgfpathlineto{\pgfqpoint{1.061316in}{0.880438in}}%
\pgfpathlineto{\pgfqpoint{1.061316in}{0.811800in}}%
\pgfpathlineto{\pgfqpoint{1.144472in}{0.811800in}}%
\pgfpathlineto{\pgfqpoint{1.144472in}{0.727394in}}%
\pgfpathlineto{\pgfqpoint{1.227629in}{0.727394in}}%
\pgfpathlineto{\pgfqpoint{1.227629in}{0.634641in}}%
\pgfpathlineto{\pgfqpoint{1.269207in}{0.634641in}}%
\pgfpathlineto{\pgfqpoint{1.269207in}{0.634641in}}%
\pgfpathlineto{\pgfqpoint{1.269207in}{0.634641in}}%
\pgfpathlineto{\pgfqpoint{1.227629in}{0.634641in}}%
\pgfpathlineto{\pgfqpoint{1.227629in}{0.727394in}}%
\pgfpathlineto{\pgfqpoint{1.144472in}{0.727394in}}%
\pgfpathlineto{\pgfqpoint{1.144472in}{0.811800in}}%
\pgfpathlineto{\pgfqpoint{1.061316in}{0.811800in}}%
\pgfpathlineto{\pgfqpoint{1.061316in}{0.880438in}}%
\pgfpathlineto{\pgfqpoint{0.978159in}{0.880438in}}%
\pgfpathlineto{\pgfqpoint{0.978159in}{0.919858in}}%
\pgfpathlineto{\pgfqpoint{0.895002in}{0.919858in}}%
\pgfpathlineto{\pgfqpoint{0.895002in}{0.950003in}}%
\pgfpathlineto{\pgfqpoint{0.811846in}{0.950003in}}%
\pgfpathlineto{\pgfqpoint{0.811846in}{0.967626in}}%
\pgfpathlineto{\pgfqpoint{0.728689in}{0.967626in}}%
\pgfpathlineto{\pgfqpoint{0.728689in}{0.981076in}}%
\pgfpathlineto{\pgfqpoint{0.645533in}{0.981076in}}%
\pgfpathlineto{\pgfqpoint{0.645533in}{0.988032in}}%
\pgfpathlineto{\pgfqpoint{0.562376in}{0.988032in}}%
\pgfpathlineto{\pgfqpoint{0.562376in}{0.998699in}}%
\pgfpathlineto{\pgfqpoint{0.520798in}{0.998699in}}%
\pgfpathlineto{\pgfqpoint{0.520798in}{0.998699in}}%
\pgfpathclose%
\pgfusepath{fill}%
}%
\begin{pgfscope}%
\pgfsys@transformshift{0.000000in}{0.000000in}%
\pgfsys@useobject{currentmarker}{}%
\end{pgfscope}%
\end{pgfscope}%
\begin{pgfscope}%
\definecolor{textcolor}{rgb}{0.150000,0.150000,0.150000}%
\pgfsetstrokecolor{textcolor}%
\pgfsetfillcolor{textcolor}%
\pgftext[x=0.895002in,y=1.397395in,,base]{\color{textcolor}\rmfamily\fontsize{9.000000}{10.800000}\selectfont \(\displaystyle c_A^-=1\)}%
\end{pgfscope}%
\begin{pgfscope}%
\pgfsetbuttcap%
\pgfsetmiterjoin%
\definecolor{currentfill}{rgb}{1.000000,1.000000,1.000000}%
\pgfsetfillcolor{currentfill}%
\pgfsetfillopacity{0.800000}%
\pgfsetlinewidth{1.003750pt}%
\definecolor{currentstroke}{rgb}{0.800000,0.800000,0.800000}%
\pgfsetstrokecolor{currentstroke}%
\pgfsetstrokeopacity{0.800000}%
\pgfsetdash{}{0pt}%
\pgfpathmoveto{\pgfqpoint{0.815247in}{0.638103in}}%
\pgfpathlineto{\pgfqpoint{1.220596in}{0.638103in}}%
\pgfpathlineto{\pgfqpoint{1.220596in}{1.265450in}}%
\pgfpathlineto{\pgfqpoint{0.815247in}{1.265450in}}%
\pgfpathlineto{\pgfqpoint{0.815247in}{0.638103in}}%
\pgfpathclose%
\pgfusepath{stroke,fill}%
\end{pgfscope}%
\begin{pgfscope}%
\definecolor{textcolor}{rgb}{0.150000,0.150000,0.150000}%
\pgfsetstrokecolor{textcolor}%
\pgfsetfillcolor{textcolor}%
\pgftext[x=0.947675in,y=1.113275in,left,base]{\color{textcolor}\rmfamily\fontsize{9.000000}{10.800000}\selectfont \(\displaystyle c_A^+\)}%
\end{pgfscope}%
\begin{pgfscope}%
\pgfsetroundcap%
\pgfsetroundjoin%
\pgfsetlinewidth{1.003750pt}%
\definecolor{currentstroke}{rgb}{0.003922,0.450980,0.698039}%
\pgfsetstrokecolor{currentstroke}%
\pgfsetdash{}{0pt}%
\pgfpathmoveto{\pgfqpoint{0.854136in}{1.001052in}}%
\pgfpathlineto{\pgfqpoint{0.902747in}{1.001052in}}%
\pgfpathlineto{\pgfqpoint{0.902747in}{1.001052in}}%
\pgfpathlineto{\pgfqpoint{0.999969in}{1.001052in}}%
\pgfpathlineto{\pgfqpoint{0.999969in}{1.001052in}}%
\pgfpathlineto{\pgfqpoint{1.048580in}{1.001052in}}%
\pgfusepath{stroke}%
\end{pgfscope}%
\begin{pgfscope}%
\definecolor{textcolor}{rgb}{0.150000,0.150000,0.150000}%
\pgfsetstrokecolor{textcolor}%
\pgfsetfillcolor{textcolor}%
\pgftext[x=1.126358in,y=0.967024in,left,base]{\color{textcolor}\rmfamily\fontsize{7.000000}{8.400000}\selectfont 1}%
\end{pgfscope}%
\begin{pgfscope}%
\pgfsetroundcap%
\pgfsetroundjoin%
\pgfsetlinewidth{1.003750pt}%
\definecolor{currentstroke}{rgb}{0.870588,0.560784,0.019608}%
\pgfsetstrokecolor{currentstroke}%
\pgfsetdash{}{0pt}%
\pgfpathmoveto{\pgfqpoint{0.854136in}{0.865486in}}%
\pgfpathlineto{\pgfqpoint{0.902747in}{0.865486in}}%
\pgfpathlineto{\pgfqpoint{0.902747in}{0.865486in}}%
\pgfpathlineto{\pgfqpoint{0.999969in}{0.865486in}}%
\pgfpathlineto{\pgfqpoint{0.999969in}{0.865486in}}%
\pgfpathlineto{\pgfqpoint{1.048580in}{0.865486in}}%
\pgfusepath{stroke}%
\end{pgfscope}%
\begin{pgfscope}%
\definecolor{textcolor}{rgb}{0.150000,0.150000,0.150000}%
\pgfsetstrokecolor{textcolor}%
\pgfsetfillcolor{textcolor}%
\pgftext[x=1.126358in,y=0.831458in,left,base]{\color{textcolor}\rmfamily\fontsize{7.000000}{8.400000}\selectfont 2}%
\end{pgfscope}%
\begin{pgfscope}%
\pgfsetroundcap%
\pgfsetroundjoin%
\pgfsetlinewidth{1.003750pt}%
\definecolor{currentstroke}{rgb}{0.007843,0.619608,0.450980}%
\pgfsetstrokecolor{currentstroke}%
\pgfsetdash{}{0pt}%
\pgfpathmoveto{\pgfqpoint{0.854136in}{0.729919in}}%
\pgfpathlineto{\pgfqpoint{0.902747in}{0.729919in}}%
\pgfpathlineto{\pgfqpoint{0.902747in}{0.729919in}}%
\pgfpathlineto{\pgfqpoint{0.999969in}{0.729919in}}%
\pgfpathlineto{\pgfqpoint{0.999969in}{0.729919in}}%
\pgfpathlineto{\pgfqpoint{1.048580in}{0.729919in}}%
\pgfusepath{stroke}%
\end{pgfscope}%
\begin{pgfscope}%
\definecolor{textcolor}{rgb}{0.150000,0.150000,0.150000}%
\pgfsetstrokecolor{textcolor}%
\pgfsetfillcolor{textcolor}%
\pgftext[x=1.126358in,y=0.695892in,left,base]{\color{textcolor}\rmfamily\fontsize{7.000000}{8.400000}\selectfont 4}%
\end{pgfscope}%
\end{pgfpicture}%
\makeatother%
\endgroup%

%% file: figures/experiments/graphs/piPPNP/piPPNP-Citeseer-A.pgf
\begingroup%
\makeatletter%
\begin{pgfpicture}%
\pgfpathrectangle{\pgfpointorigin}{\pgfqpoint{1.375000in}{1.581250in}}%
\pgfusepath{use as bounding box, clip}%
\begin{pgfscope}%
\pgfsetbuttcap%
\pgfsetmiterjoin%
\definecolor{currentfill}{rgb}{1.000000,1.000000,1.000000}%
\pgfsetfillcolor{currentfill}%
\pgfsetlinewidth{0.000000pt}%
\definecolor{currentstroke}{rgb}{1.000000,1.000000,1.000000}%
\pgfsetstrokecolor{currentstroke}%
\pgfsetdash{}{0pt}%
\pgfpathmoveto{\pgfqpoint{0.000000in}{0.000000in}}%
\pgfpathlineto{\pgfqpoint{1.375000in}{0.000000in}}%
\pgfpathlineto{\pgfqpoint{1.375000in}{1.581250in}}%
\pgfpathlineto{\pgfqpoint{0.000000in}{1.581250in}}%
\pgfpathlineto{\pgfqpoint{0.000000in}{0.000000in}}%
\pgfpathclose%
\pgfusepath{fill}%
\end{pgfscope}%
\begin{pgfscope}%
\pgfsetbuttcap%
\pgfsetmiterjoin%
\definecolor{currentfill}{rgb}{1.000000,1.000000,1.000000}%
\pgfsetfillcolor{currentfill}%
\pgfsetlinewidth{0.000000pt}%
\definecolor{currentstroke}{rgb}{0.000000,0.000000,0.000000}%
\pgfsetstrokecolor{currentstroke}%
\pgfsetstrokeopacity{0.000000}%
\pgfsetdash{}{0pt}%
\pgfpathmoveto{\pgfqpoint{0.520798in}{0.442177in}}%
\pgfpathlineto{\pgfqpoint{1.269207in}{0.442177in}}%
\pgfpathlineto{\pgfqpoint{1.269207in}{1.314061in}}%
\pgfpathlineto{\pgfqpoint{0.520798in}{1.314061in}}%
\pgfpathlineto{\pgfqpoint{0.520798in}{0.442177in}}%
\pgfpathclose%
\pgfusepath{fill}%
\end{pgfscope}%
\begin{pgfscope}%
\pgfpathrectangle{\pgfqpoint{0.520798in}{0.442177in}}{\pgfqpoint{0.748409in}{0.871884in}}%
\pgfusepath{clip}%
\pgfsetroundcap%
\pgfsetroundjoin%
\pgfsetlinewidth{0.501875pt}%
\definecolor{currentstroke}{rgb}{0.800000,0.800000,0.800000}%
\pgfsetstrokecolor{currentstroke}%
\pgfsetdash{}{0pt}%
\pgfpathmoveto{\pgfqpoint{0.520798in}{0.442177in}}%
\pgfpathlineto{\pgfqpoint{0.520798in}{1.314061in}}%
\pgfusepath{stroke}%
\end{pgfscope}%
\begin{pgfscope}%
\definecolor{textcolor}{rgb}{0.150000,0.150000,0.150000}%
\pgfsetstrokecolor{textcolor}%
\pgfsetfillcolor{textcolor}%
\pgftext[x=0.520798in,y=0.351899in,,top]{\color{textcolor}\rmfamily\fontsize{8.000000}{9.600000}\selectfont \(\displaystyle {0}\)}%
\end{pgfscope}%
\begin{pgfscope}%
\pgfpathrectangle{\pgfqpoint{0.520798in}{0.442177in}}{\pgfqpoint{0.748409in}{0.871884in}}%
\pgfusepath{clip}%
\pgfsetroundcap%
\pgfsetroundjoin%
\pgfsetlinewidth{0.501875pt}%
\definecolor{currentstroke}{rgb}{0.800000,0.800000,0.800000}%
\pgfsetstrokecolor{currentstroke}%
\pgfsetdash{}{0pt}%
\pgfpathmoveto{\pgfqpoint{0.853424in}{0.442177in}}%
\pgfpathlineto{\pgfqpoint{0.853424in}{1.314061in}}%
\pgfusepath{stroke}%
\end{pgfscope}%
\begin{pgfscope}%
\definecolor{textcolor}{rgb}{0.150000,0.150000,0.150000}%
\pgfsetstrokecolor{textcolor}%
\pgfsetfillcolor{textcolor}%
\pgftext[x=0.853424in,y=0.351899in,,top]{\color{textcolor}\rmfamily\fontsize{8.000000}{9.600000}\selectfont \(\displaystyle {4}\)}%
\end{pgfscope}%
\begin{pgfscope}%
\pgfpathrectangle{\pgfqpoint{0.520798in}{0.442177in}}{\pgfqpoint{0.748409in}{0.871884in}}%
\pgfusepath{clip}%
\pgfsetroundcap%
\pgfsetroundjoin%
\pgfsetlinewidth{0.501875pt}%
\definecolor{currentstroke}{rgb}{0.800000,0.800000,0.800000}%
\pgfsetstrokecolor{currentstroke}%
\pgfsetdash{}{0pt}%
\pgfpathmoveto{\pgfqpoint{1.269207in}{0.442177in}}%
\pgfpathlineto{\pgfqpoint{1.269207in}{1.314061in}}%
\pgfusepath{stroke}%
\end{pgfscope}%
\begin{pgfscope}%
\definecolor{textcolor}{rgb}{0.150000,0.150000,0.150000}%
\pgfsetstrokecolor{textcolor}%
\pgfsetfillcolor{textcolor}%
\pgftext[x=1.269207in,y=0.351899in,,top]{\color{textcolor}\rmfamily\fontsize{8.000000}{9.600000}\selectfont \(\displaystyle {9}\)}%
\end{pgfscope}%
\begin{pgfscope}%
\definecolor{textcolor}{rgb}{0.150000,0.150000,0.150000}%
\pgfsetstrokecolor{textcolor}%
\pgfsetfillcolor{textcolor}%
\pgftext[x=0.895002in,y=0.198219in,,top]{\color{textcolor}\rmfamily\fontsize{10.000000}{12.000000}\selectfont attack strength}%
\end{pgfscope}%
\begin{pgfscope}%
\pgfpathrectangle{\pgfqpoint{0.520798in}{0.442177in}}{\pgfqpoint{0.748409in}{0.871884in}}%
\pgfusepath{clip}%
\pgfsetroundcap%
\pgfsetroundjoin%
\pgfsetlinewidth{0.501875pt}%
\definecolor{currentstroke}{rgb}{0.800000,0.800000,0.800000}%
\pgfsetstrokecolor{currentstroke}%
\pgfsetdash{}{0pt}%
\pgfpathmoveto{\pgfqpoint{0.520798in}{0.442177in}}%
\pgfpathlineto{\pgfqpoint{1.269207in}{0.442177in}}%
\pgfusepath{stroke}%
\end{pgfscope}%
\begin{pgfscope}%
\definecolor{textcolor}{rgb}{0.150000,0.150000,0.150000}%
\pgfsetstrokecolor{textcolor}%
\pgfsetfillcolor{textcolor}%
\pgftext[x=0.273151in, y=0.403915in, left, base]{\color{textcolor}\rmfamily\fontsize{8.000000}{9.600000}\selectfont 0\%}%
\end{pgfscope}%
\begin{pgfscope}%
\pgfpathrectangle{\pgfqpoint{0.520798in}{0.442177in}}{\pgfqpoint{0.748409in}{0.871884in}}%
\pgfusepath{clip}%
\pgfsetroundcap%
\pgfsetroundjoin%
\pgfsetlinewidth{0.501875pt}%
\definecolor{currentstroke}{rgb}{0.800000,0.800000,0.800000}%
\pgfsetstrokecolor{currentstroke}%
\pgfsetdash{}{0pt}%
\pgfpathmoveto{\pgfqpoint{0.520798in}{0.660148in}}%
\pgfpathlineto{\pgfqpoint{1.269207in}{0.660148in}}%
\pgfusepath{stroke}%
\end{pgfscope}%
\begin{pgfscope}%
\definecolor{textcolor}{rgb}{0.150000,0.150000,0.150000}%
\pgfsetstrokecolor{textcolor}%
\pgfsetfillcolor{textcolor}%
\pgftext[x=0.214138in, y=0.621886in, left, base]{\color{textcolor}\rmfamily\fontsize{8.000000}{9.600000}\selectfont 25\%}%
\end{pgfscope}%
\begin{pgfscope}%
\pgfpathrectangle{\pgfqpoint{0.520798in}{0.442177in}}{\pgfqpoint{0.748409in}{0.871884in}}%
\pgfusepath{clip}%
\pgfsetroundcap%
\pgfsetroundjoin%
\pgfsetlinewidth{0.501875pt}%
\definecolor{currentstroke}{rgb}{0.800000,0.800000,0.800000}%
\pgfsetstrokecolor{currentstroke}%
\pgfsetdash{}{0pt}%
\pgfpathmoveto{\pgfqpoint{0.520798in}{0.878119in}}%
\pgfpathlineto{\pgfqpoint{1.269207in}{0.878119in}}%
\pgfusepath{stroke}%
\end{pgfscope}%
\begin{pgfscope}%
\definecolor{textcolor}{rgb}{0.150000,0.150000,0.150000}%
\pgfsetstrokecolor{textcolor}%
\pgfsetfillcolor{textcolor}%
\pgftext[x=0.214138in, y=0.839857in, left, base]{\color{textcolor}\rmfamily\fontsize{8.000000}{9.600000}\selectfont 50\%}%
\end{pgfscope}%
\begin{pgfscope}%
\pgfpathrectangle{\pgfqpoint{0.520798in}{0.442177in}}{\pgfqpoint{0.748409in}{0.871884in}}%
\pgfusepath{clip}%
\pgfsetroundcap%
\pgfsetroundjoin%
\pgfsetlinewidth{0.501875pt}%
\definecolor{currentstroke}{rgb}{0.800000,0.800000,0.800000}%
\pgfsetstrokecolor{currentstroke}%
\pgfsetdash{}{0pt}%
\pgfpathmoveto{\pgfqpoint{0.520798in}{1.096090in}}%
\pgfpathlineto{\pgfqpoint{1.269207in}{1.096090in}}%
\pgfusepath{stroke}%
\end{pgfscope}%
\begin{pgfscope}%
\definecolor{textcolor}{rgb}{0.150000,0.150000,0.150000}%
\pgfsetstrokecolor{textcolor}%
\pgfsetfillcolor{textcolor}%
\pgftext[x=0.214138in, y=1.057828in, left, base]{\color{textcolor}\rmfamily\fontsize{8.000000}{9.600000}\selectfont 75\%}%
\end{pgfscope}%
\begin{pgfscope}%
\pgfpathrectangle{\pgfqpoint{0.520798in}{0.442177in}}{\pgfqpoint{0.748409in}{0.871884in}}%
\pgfusepath{clip}%
\pgfsetroundcap%
\pgfsetroundjoin%
\pgfsetlinewidth{0.501875pt}%
\definecolor{currentstroke}{rgb}{0.800000,0.800000,0.800000}%
\pgfsetstrokecolor{currentstroke}%
\pgfsetdash{}{0pt}%
\pgfpathmoveto{\pgfqpoint{0.520798in}{1.314061in}}%
\pgfpathlineto{\pgfqpoint{1.269207in}{1.314061in}}%
\pgfusepath{stroke}%
\end{pgfscope}%
\begin{pgfscope}%
\definecolor{textcolor}{rgb}{0.150000,0.150000,0.150000}%
\pgfsetstrokecolor{textcolor}%
\pgfsetfillcolor{textcolor}%
\pgftext[x=0.155124in, y=1.275799in, left, base]{\color{textcolor}\rmfamily\fontsize{8.000000}{9.600000}\selectfont 100\%}%
\end{pgfscope}%
\begin{pgfscope}%
\definecolor{textcolor}{rgb}{0.150000,0.150000,0.150000}%
\pgfsetstrokecolor{textcolor}%
\pgfsetfillcolor{textcolor}%
\pgftext[x=0.099569in,y=0.878119in,,bottom,rotate=90.000000]{\color{textcolor}\rmfamily\fontsize{10.000000}{12.000000}\selectfont Cert. Acc.}%
\end{pgfscope}%
\begin{pgfscope}%
\pgfsetrectcap%
\pgfsetmiterjoin%
\pgfsetlinewidth{0.752812pt}%
\definecolor{currentstroke}{rgb}{0.700000,0.700000,0.700000}%
\pgfsetstrokecolor{currentstroke}%
\pgfsetdash{}{0pt}%
\pgfpathmoveto{\pgfqpoint{0.520798in}{0.442177in}}%
\pgfpathlineto{\pgfqpoint{0.520798in}{1.314061in}}%
\pgfusepath{stroke}%
\end{pgfscope}%
\begin{pgfscope}%
\pgfsetrectcap%
\pgfsetmiterjoin%
\pgfsetlinewidth{0.752812pt}%
\definecolor{currentstroke}{rgb}{0.700000,0.700000,0.700000}%
\pgfsetstrokecolor{currentstroke}%
\pgfsetdash{}{0pt}%
\pgfpathmoveto{\pgfqpoint{1.269207in}{0.442177in}}%
\pgfpathlineto{\pgfqpoint{1.269207in}{1.314061in}}%
\pgfusepath{stroke}%
\end{pgfscope}%
\begin{pgfscope}%
\pgfsetrectcap%
\pgfsetmiterjoin%
\pgfsetlinewidth{0.752812pt}%
\definecolor{currentstroke}{rgb}{0.700000,0.700000,0.700000}%
\pgfsetstrokecolor{currentstroke}%
\pgfsetdash{}{0pt}%
\pgfpathmoveto{\pgfqpoint{0.520798in}{0.442177in}}%
\pgfpathlineto{\pgfqpoint{1.269207in}{0.442177in}}%
\pgfusepath{stroke}%
\end{pgfscope}%
\begin{pgfscope}%
\pgfsetrectcap%
\pgfsetmiterjoin%
\pgfsetlinewidth{0.752812pt}%
\definecolor{currentstroke}{rgb}{0.700000,0.700000,0.700000}%
\pgfsetstrokecolor{currentstroke}%
\pgfsetdash{}{0pt}%
\pgfpathmoveto{\pgfqpoint{0.520798in}{1.314061in}}%
\pgfpathlineto{\pgfqpoint{1.269207in}{1.314061in}}%
\pgfusepath{stroke}%
\end{pgfscope}%
\begin{pgfscope}%
\pgfpathrectangle{\pgfqpoint{0.520798in}{0.442177in}}{\pgfqpoint{0.748409in}{0.871884in}}%
\pgfusepath{clip}%
\pgfsetroundcap%
\pgfsetroundjoin%
\pgfsetlinewidth{1.003750pt}%
\definecolor{currentstroke}{rgb}{0.003922,0.450980,0.698039}%
\pgfsetstrokecolor{currentstroke}%
\pgfsetdash{}{0pt}%
\pgfpathmoveto{\pgfqpoint{0.520798in}{0.885447in}}%
\pgfpathlineto{\pgfqpoint{0.562376in}{0.885447in}}%
\pgfpathlineto{\pgfqpoint{0.562376in}{0.854931in}}%
\pgfpathlineto{\pgfqpoint{0.645533in}{0.854931in}}%
\pgfpathlineto{\pgfqpoint{0.645533in}{0.823209in}}%
\pgfpathlineto{\pgfqpoint{0.728689in}{0.823209in}}%
\pgfpathlineto{\pgfqpoint{0.728689in}{0.790745in}}%
\pgfpathlineto{\pgfqpoint{0.811846in}{0.790745in}}%
\pgfpathlineto{\pgfqpoint{0.811846in}{0.741307in}}%
\pgfpathlineto{\pgfqpoint{0.895002in}{0.741307in}}%
\pgfpathlineto{\pgfqpoint{0.895002in}{0.673783in}}%
\pgfpathlineto{\pgfqpoint{0.978159in}{0.673783in}}%
\pgfpathlineto{\pgfqpoint{0.978159in}{0.594107in}}%
\pgfpathlineto{\pgfqpoint{1.061316in}{0.594107in}}%
\pgfpathlineto{\pgfqpoint{1.061316in}{0.507661in}}%
\pgfpathlineto{\pgfqpoint{1.144472in}{0.507661in}}%
\pgfpathlineto{\pgfqpoint{1.144472in}{0.455255in}}%
\pgfpathlineto{\pgfqpoint{1.227629in}{0.455255in}}%
\pgfpathlineto{\pgfqpoint{1.227629in}{0.442177in}}%
\pgfpathlineto{\pgfqpoint{1.270874in}{0.442177in}}%
\pgfusepath{stroke}%
\end{pgfscope}%
\begin{pgfscope}%
\pgfpathrectangle{\pgfqpoint{0.520798in}{0.442177in}}{\pgfqpoint{0.748409in}{0.871884in}}%
\pgfusepath{clip}%
\pgfsetbuttcap%
\pgfsetroundjoin%
\definecolor{currentfill}{rgb}{0.003922,0.450980,0.698039}%
\pgfsetfillcolor{currentfill}%
\pgfsetfillopacity{0.500000}%
\pgfsetlinewidth{0.000000pt}%
\definecolor{currentstroke}{rgb}{0.003922,0.450980,0.698039}%
\pgfsetstrokecolor{currentstroke}%
\pgfsetstrokeopacity{0.500000}%
\pgfsetdash{}{0pt}%
\pgfpathmoveto{\pgfqpoint{0.520798in}{0.898744in}}%
\pgfpathlineto{\pgfqpoint{0.520798in}{0.872149in}}%
\pgfpathlineto{\pgfqpoint{0.562376in}{0.872149in}}%
\pgfpathlineto{\pgfqpoint{0.562376in}{0.842812in}}%
\pgfpathlineto{\pgfqpoint{0.645533in}{0.842812in}}%
\pgfpathlineto{\pgfqpoint{0.645533in}{0.807228in}}%
\pgfpathlineto{\pgfqpoint{0.728689in}{0.807228in}}%
\pgfpathlineto{\pgfqpoint{0.728689in}{0.776979in}}%
\pgfpathlineto{\pgfqpoint{0.811846in}{0.776979in}}%
\pgfpathlineto{\pgfqpoint{0.811846in}{0.730042in}}%
\pgfpathlineto{\pgfqpoint{0.895002in}{0.730042in}}%
\pgfpathlineto{\pgfqpoint{0.895002in}{0.667647in}}%
\pgfpathlineto{\pgfqpoint{0.978159in}{0.667647in}}%
\pgfpathlineto{\pgfqpoint{0.978159in}{0.589634in}}%
\pgfpathlineto{\pgfqpoint{1.061316in}{0.589634in}}%
\pgfpathlineto{\pgfqpoint{1.061316in}{0.499578in}}%
\pgfpathlineto{\pgfqpoint{1.144472in}{0.499578in}}%
\pgfpathlineto{\pgfqpoint{1.144472in}{0.453692in}}%
\pgfpathlineto{\pgfqpoint{1.227629in}{0.453692in}}%
\pgfpathlineto{\pgfqpoint{1.227629in}{0.442177in}}%
\pgfpathlineto{\pgfqpoint{1.310785in}{0.442177in}}%
\pgfpathlineto{\pgfqpoint{1.310785in}{0.442177in}}%
\pgfpathlineto{\pgfqpoint{1.393942in}{0.442177in}}%
\pgfpathlineto{\pgfqpoint{1.393942in}{0.442177in}}%
\pgfpathlineto{\pgfqpoint{1.477098in}{0.442177in}}%
\pgfpathlineto{\pgfqpoint{1.477098in}{0.442177in}}%
\pgfpathlineto{\pgfqpoint{1.560255in}{0.442177in}}%
\pgfpathlineto{\pgfqpoint{1.560255in}{0.442177in}}%
\pgfpathlineto{\pgfqpoint{1.643412in}{0.442177in}}%
\pgfpathlineto{\pgfqpoint{1.643412in}{0.442177in}}%
\pgfpathlineto{\pgfqpoint{1.726568in}{0.442177in}}%
\pgfpathlineto{\pgfqpoint{1.726568in}{0.442177in}}%
\pgfpathlineto{\pgfqpoint{1.809725in}{0.442177in}}%
\pgfpathlineto{\pgfqpoint{1.809725in}{0.442177in}}%
\pgfpathlineto{\pgfqpoint{1.892881in}{0.442177in}}%
\pgfpathlineto{\pgfqpoint{1.892881in}{0.442177in}}%
\pgfpathlineto{\pgfqpoint{1.976038in}{0.442177in}}%
\pgfpathlineto{\pgfqpoint{1.976038in}{0.442177in}}%
\pgfpathlineto{\pgfqpoint{2.059194in}{0.442177in}}%
\pgfpathlineto{\pgfqpoint{2.059194in}{0.442177in}}%
\pgfpathlineto{\pgfqpoint{2.142351in}{0.442177in}}%
\pgfpathlineto{\pgfqpoint{2.142351in}{0.442177in}}%
\pgfpathlineto{\pgfqpoint{2.225508in}{0.442177in}}%
\pgfpathlineto{\pgfqpoint{2.225508in}{0.442177in}}%
\pgfpathlineto{\pgfqpoint{2.308664in}{0.442177in}}%
\pgfpathlineto{\pgfqpoint{2.308664in}{0.442177in}}%
\pgfpathlineto{\pgfqpoint{2.391821in}{0.442177in}}%
\pgfpathlineto{\pgfqpoint{2.391821in}{0.442177in}}%
\pgfpathlineto{\pgfqpoint{2.474977in}{0.442177in}}%
\pgfpathlineto{\pgfqpoint{2.474977in}{0.442177in}}%
\pgfpathlineto{\pgfqpoint{2.558134in}{0.442177in}}%
\pgfpathlineto{\pgfqpoint{2.558134in}{0.442177in}}%
\pgfpathlineto{\pgfqpoint{2.641290in}{0.442177in}}%
\pgfpathlineto{\pgfqpoint{2.641290in}{0.442177in}}%
\pgfpathlineto{\pgfqpoint{2.724447in}{0.442177in}}%
\pgfpathlineto{\pgfqpoint{2.724447in}{0.442177in}}%
\pgfpathlineto{\pgfqpoint{2.807604in}{0.442177in}}%
\pgfpathlineto{\pgfqpoint{2.807604in}{0.442177in}}%
\pgfpathlineto{\pgfqpoint{2.890760in}{0.442177in}}%
\pgfpathlineto{\pgfqpoint{2.890760in}{0.442177in}}%
\pgfpathlineto{\pgfqpoint{2.932338in}{0.442177in}}%
\pgfpathlineto{\pgfqpoint{2.932338in}{0.442177in}}%
\pgfpathlineto{\pgfqpoint{2.932338in}{0.442177in}}%
\pgfpathlineto{\pgfqpoint{2.890760in}{0.442177in}}%
\pgfpathlineto{\pgfqpoint{2.890760in}{0.442177in}}%
\pgfpathlineto{\pgfqpoint{2.807604in}{0.442177in}}%
\pgfpathlineto{\pgfqpoint{2.807604in}{0.442177in}}%
\pgfpathlineto{\pgfqpoint{2.724447in}{0.442177in}}%
\pgfpathlineto{\pgfqpoint{2.724447in}{0.442177in}}%
\pgfpathlineto{\pgfqpoint{2.641290in}{0.442177in}}%
\pgfpathlineto{\pgfqpoint{2.641290in}{0.442177in}}%
\pgfpathlineto{\pgfqpoint{2.558134in}{0.442177in}}%
\pgfpathlineto{\pgfqpoint{2.558134in}{0.442177in}}%
\pgfpathlineto{\pgfqpoint{2.474977in}{0.442177in}}%
\pgfpathlineto{\pgfqpoint{2.474977in}{0.442177in}}%
\pgfpathlineto{\pgfqpoint{2.391821in}{0.442177in}}%
\pgfpathlineto{\pgfqpoint{2.391821in}{0.442177in}}%
\pgfpathlineto{\pgfqpoint{2.308664in}{0.442177in}}%
\pgfpathlineto{\pgfqpoint{2.308664in}{0.442177in}}%
\pgfpathlineto{\pgfqpoint{2.225508in}{0.442177in}}%
\pgfpathlineto{\pgfqpoint{2.225508in}{0.442177in}}%
\pgfpathlineto{\pgfqpoint{2.142351in}{0.442177in}}%
\pgfpathlineto{\pgfqpoint{2.142351in}{0.442177in}}%
\pgfpathlineto{\pgfqpoint{2.059194in}{0.442177in}}%
\pgfpathlineto{\pgfqpoint{2.059194in}{0.442177in}}%
\pgfpathlineto{\pgfqpoint{1.976038in}{0.442177in}}%
\pgfpathlineto{\pgfqpoint{1.976038in}{0.442177in}}%
\pgfpathlineto{\pgfqpoint{1.892881in}{0.442177in}}%
\pgfpathlineto{\pgfqpoint{1.892881in}{0.442177in}}%
\pgfpathlineto{\pgfqpoint{1.809725in}{0.442177in}}%
\pgfpathlineto{\pgfqpoint{1.809725in}{0.442177in}}%
\pgfpathlineto{\pgfqpoint{1.726568in}{0.442177in}}%
\pgfpathlineto{\pgfqpoint{1.726568in}{0.442177in}}%
\pgfpathlineto{\pgfqpoint{1.643412in}{0.442177in}}%
\pgfpathlineto{\pgfqpoint{1.643412in}{0.442177in}}%
\pgfpathlineto{\pgfqpoint{1.560255in}{0.442177in}}%
\pgfpathlineto{\pgfqpoint{1.560255in}{0.442177in}}%
\pgfpathlineto{\pgfqpoint{1.477098in}{0.442177in}}%
\pgfpathlineto{\pgfqpoint{1.477098in}{0.442177in}}%
\pgfpathlineto{\pgfqpoint{1.393942in}{0.442177in}}%
\pgfpathlineto{\pgfqpoint{1.393942in}{0.442177in}}%
\pgfpathlineto{\pgfqpoint{1.310785in}{0.442177in}}%
\pgfpathlineto{\pgfqpoint{1.310785in}{0.442177in}}%
\pgfpathlineto{\pgfqpoint{1.227629in}{0.442177in}}%
\pgfpathlineto{\pgfqpoint{1.227629in}{0.456818in}}%
\pgfpathlineto{\pgfqpoint{1.144472in}{0.456818in}}%
\pgfpathlineto{\pgfqpoint{1.144472in}{0.515744in}}%
\pgfpathlineto{\pgfqpoint{1.061316in}{0.515744in}}%
\pgfpathlineto{\pgfqpoint{1.061316in}{0.598581in}}%
\pgfpathlineto{\pgfqpoint{0.978159in}{0.598581in}}%
\pgfpathlineto{\pgfqpoint{0.978159in}{0.679918in}}%
\pgfpathlineto{\pgfqpoint{0.895002in}{0.679918in}}%
\pgfpathlineto{\pgfqpoint{0.895002in}{0.752572in}}%
\pgfpathlineto{\pgfqpoint{0.811846in}{0.752572in}}%
\pgfpathlineto{\pgfqpoint{0.811846in}{0.804511in}}%
\pgfpathlineto{\pgfqpoint{0.728689in}{0.804511in}}%
\pgfpathlineto{\pgfqpoint{0.728689in}{0.839190in}}%
\pgfpathlineto{\pgfqpoint{0.645533in}{0.839190in}}%
\pgfpathlineto{\pgfqpoint{0.645533in}{0.867049in}}%
\pgfpathlineto{\pgfqpoint{0.562376in}{0.867049in}}%
\pgfpathlineto{\pgfqpoint{0.562376in}{0.898744in}}%
\pgfpathlineto{\pgfqpoint{0.520798in}{0.898744in}}%
\pgfpathlineto{\pgfqpoint{0.520798in}{0.898744in}}%
\pgfpathclose%
\pgfusepath{fill}%
\end{pgfscope}%
\begin{pgfscope}%
\pgfpathrectangle{\pgfqpoint{0.520798in}{0.442177in}}{\pgfqpoint{0.748409in}{0.871884in}}%
\pgfusepath{clip}%
\pgfsetroundcap%
\pgfsetroundjoin%
\pgfsetlinewidth{1.003750pt}%
\definecolor{currentstroke}{rgb}{0.870588,0.560784,0.019608}%
\pgfsetstrokecolor{currentstroke}%
\pgfsetdash{}{0pt}%
\pgfpathmoveto{\pgfqpoint{0.520798in}{0.910119in}}%
\pgfpathlineto{\pgfqpoint{0.562376in}{0.910119in}}%
\pgfpathlineto{\pgfqpoint{0.562376in}{0.875336in}}%
\pgfpathlineto{\pgfqpoint{0.645533in}{0.875336in}}%
\pgfpathlineto{\pgfqpoint{0.645533in}{0.845655in}}%
\pgfpathlineto{\pgfqpoint{0.728689in}{0.845655in}}%
\pgfpathlineto{\pgfqpoint{0.728689in}{0.810409in}}%
\pgfpathlineto{\pgfqpoint{0.811846in}{0.810409in}}%
\pgfpathlineto{\pgfqpoint{0.811846in}{0.755684in}}%
\pgfpathlineto{\pgfqpoint{0.895002in}{0.755684in}}%
\pgfpathlineto{\pgfqpoint{0.895002in}{0.686583in}}%
\pgfpathlineto{\pgfqpoint{0.978159in}{0.686583in}}%
\pgfpathlineto{\pgfqpoint{0.978159in}{0.598467in}}%
\pgfpathlineto{\pgfqpoint{1.061316in}{0.598467in}}%
\pgfpathlineto{\pgfqpoint{1.061316in}{0.524264in}}%
\pgfpathlineto{\pgfqpoint{1.144472in}{0.524264in}}%
\pgfpathlineto{\pgfqpoint{1.144472in}{0.459336in}}%
\pgfpathlineto{\pgfqpoint{1.227629in}{0.459336in}}%
\pgfpathlineto{\pgfqpoint{1.227629in}{0.442177in}}%
\pgfpathlineto{\pgfqpoint{1.269207in}{0.442177in}}%
\pgfusepath{stroke}%
\end{pgfscope}%
\begin{pgfscope}%
\pgfpathrectangle{\pgfqpoint{0.520798in}{0.442177in}}{\pgfqpoint{0.748409in}{0.871884in}}%
\pgfusepath{clip}%
\pgfsetbuttcap%
\pgfsetroundjoin%
\definecolor{currentfill}{rgb}{0.870588,0.560784,0.019608}%
\pgfsetfillcolor{currentfill}%
\pgfsetfillopacity{0.500000}%
\pgfsetlinewidth{0.000000pt}%
\definecolor{currentstroke}{rgb}{0.870588,0.560784,0.019608}%
\pgfsetstrokecolor{currentstroke}%
\pgfsetstrokeopacity{0.500000}%
\pgfsetdash{}{0pt}%
\pgfsys@defobject{currentmarker}{\pgfqpoint{0.520798in}{0.442177in}}{\pgfqpoint{1.269207in}{0.910119in}}{%
\pgfpathmoveto{\pgfqpoint{0.520798in}{0.910119in}}%
\pgfpathlineto{\pgfqpoint{0.520798in}{0.910119in}}%
\pgfpathlineto{\pgfqpoint{0.562376in}{0.910119in}}%
\pgfpathlineto{\pgfqpoint{0.562376in}{0.875336in}}%
\pgfpathlineto{\pgfqpoint{0.645533in}{0.875336in}}%
\pgfpathlineto{\pgfqpoint{0.645533in}{0.845655in}}%
\pgfpathlineto{\pgfqpoint{0.728689in}{0.845655in}}%
\pgfpathlineto{\pgfqpoint{0.728689in}{0.810409in}}%
\pgfpathlineto{\pgfqpoint{0.811846in}{0.810409in}}%
\pgfpathlineto{\pgfqpoint{0.811846in}{0.755684in}}%
\pgfpathlineto{\pgfqpoint{0.895002in}{0.755684in}}%
\pgfpathlineto{\pgfqpoint{0.895002in}{0.686583in}}%
\pgfpathlineto{\pgfqpoint{0.978159in}{0.686583in}}%
\pgfpathlineto{\pgfqpoint{0.978159in}{0.598467in}}%
\pgfpathlineto{\pgfqpoint{1.061316in}{0.598467in}}%
\pgfpathlineto{\pgfqpoint{1.061316in}{0.524264in}}%
\pgfpathlineto{\pgfqpoint{1.144472in}{0.524264in}}%
\pgfpathlineto{\pgfqpoint{1.144472in}{0.459336in}}%
\pgfpathlineto{\pgfqpoint{1.227629in}{0.459336in}}%
\pgfpathlineto{\pgfqpoint{1.227629in}{0.442177in}}%
\pgfpathlineto{\pgfqpoint{1.269207in}{0.442177in}}%
\pgfpathlineto{\pgfqpoint{1.269207in}{0.442177in}}%
\pgfpathlineto{\pgfqpoint{1.269207in}{0.442177in}}%
\pgfpathlineto{\pgfqpoint{1.227629in}{0.442177in}}%
\pgfpathlineto{\pgfqpoint{1.227629in}{0.459336in}}%
\pgfpathlineto{\pgfqpoint{1.144472in}{0.459336in}}%
\pgfpathlineto{\pgfqpoint{1.144472in}{0.524264in}}%
\pgfpathlineto{\pgfqpoint{1.061316in}{0.524264in}}%
\pgfpathlineto{\pgfqpoint{1.061316in}{0.598467in}}%
\pgfpathlineto{\pgfqpoint{0.978159in}{0.598467in}}%
\pgfpathlineto{\pgfqpoint{0.978159in}{0.686583in}}%
\pgfpathlineto{\pgfqpoint{0.895002in}{0.686583in}}%
\pgfpathlineto{\pgfqpoint{0.895002in}{0.755684in}}%
\pgfpathlineto{\pgfqpoint{0.811846in}{0.755684in}}%
\pgfpathlineto{\pgfqpoint{0.811846in}{0.810409in}}%
\pgfpathlineto{\pgfqpoint{0.728689in}{0.810409in}}%
\pgfpathlineto{\pgfqpoint{0.728689in}{0.845655in}}%
\pgfpathlineto{\pgfqpoint{0.645533in}{0.845655in}}%
\pgfpathlineto{\pgfqpoint{0.645533in}{0.875336in}}%
\pgfpathlineto{\pgfqpoint{0.562376in}{0.875336in}}%
\pgfpathlineto{\pgfqpoint{0.562376in}{0.910119in}}%
\pgfpathlineto{\pgfqpoint{0.520798in}{0.910119in}}%
\pgfpathlineto{\pgfqpoint{0.520798in}{0.910119in}}%
\pgfpathclose%
\pgfusepath{fill}%
}%
\begin{pgfscope}%
\pgfsys@transformshift{0.000000in}{0.000000in}%
\pgfsys@useobject{currentmarker}{}%
\end{pgfscope}%
\end{pgfscope}%
\begin{pgfscope}%
\pgfpathrectangle{\pgfqpoint{0.520798in}{0.442177in}}{\pgfqpoint{0.748409in}{0.871884in}}%
\pgfusepath{clip}%
\pgfsetroundcap%
\pgfsetroundjoin%
\pgfsetlinewidth{1.003750pt}%
\definecolor{currentstroke}{rgb}{0.007843,0.619608,0.450980}%
\pgfsetstrokecolor{currentstroke}%
\pgfsetdash{}{0pt}%
\pgfpathmoveto{\pgfqpoint{0.520798in}{0.910583in}}%
\pgfpathlineto{\pgfqpoint{0.562376in}{0.910583in}}%
\pgfpathlineto{\pgfqpoint{0.562376in}{0.876728in}}%
\pgfpathlineto{\pgfqpoint{0.645533in}{0.876728in}}%
\pgfpathlineto{\pgfqpoint{0.645533in}{0.846119in}}%
\pgfpathlineto{\pgfqpoint{0.728689in}{0.846119in}}%
\pgfpathlineto{\pgfqpoint{0.728689in}{0.810873in}}%
\pgfpathlineto{\pgfqpoint{0.811846in}{0.810873in}}%
\pgfpathlineto{\pgfqpoint{0.811846in}{0.757539in}}%
\pgfpathlineto{\pgfqpoint{0.895002in}{0.757539in}}%
\pgfpathlineto{\pgfqpoint{0.895002in}{0.688902in}}%
\pgfpathlineto{\pgfqpoint{0.978159in}{0.688902in}}%
\pgfpathlineto{\pgfqpoint{0.978159in}{0.599858in}}%
\pgfpathlineto{\pgfqpoint{1.061316in}{0.599858in}}%
\pgfpathlineto{\pgfqpoint{1.061316in}{0.525655in}}%
\pgfpathlineto{\pgfqpoint{1.144472in}{0.525655in}}%
\pgfpathlineto{\pgfqpoint{1.144472in}{0.460728in}}%
\pgfpathlineto{\pgfqpoint{1.227629in}{0.460728in}}%
\pgfpathlineto{\pgfqpoint{1.227629in}{0.442177in}}%
\pgfpathlineto{\pgfqpoint{1.269207in}{0.442177in}}%
\pgfusepath{stroke}%
\end{pgfscope}%
\begin{pgfscope}%
\pgfpathrectangle{\pgfqpoint{0.520798in}{0.442177in}}{\pgfqpoint{0.748409in}{0.871884in}}%
\pgfusepath{clip}%
\pgfsetbuttcap%
\pgfsetroundjoin%
\definecolor{currentfill}{rgb}{0.007843,0.619608,0.450980}%
\pgfsetfillcolor{currentfill}%
\pgfsetfillopacity{0.500000}%
\pgfsetlinewidth{0.000000pt}%
\definecolor{currentstroke}{rgb}{0.007843,0.619608,0.450980}%
\pgfsetstrokecolor{currentstroke}%
\pgfsetstrokeopacity{0.500000}%
\pgfsetdash{}{0pt}%
\pgfsys@defobject{currentmarker}{\pgfqpoint{0.520798in}{0.442177in}}{\pgfqpoint{1.269207in}{0.910583in}}{%
\pgfpathmoveto{\pgfqpoint{0.520798in}{0.910583in}}%
\pgfpathlineto{\pgfqpoint{0.520798in}{0.910583in}}%
\pgfpathlineto{\pgfqpoint{0.562376in}{0.910583in}}%
\pgfpathlineto{\pgfqpoint{0.562376in}{0.876728in}}%
\pgfpathlineto{\pgfqpoint{0.645533in}{0.876728in}}%
\pgfpathlineto{\pgfqpoint{0.645533in}{0.846119in}}%
\pgfpathlineto{\pgfqpoint{0.728689in}{0.846119in}}%
\pgfpathlineto{\pgfqpoint{0.728689in}{0.810873in}}%
\pgfpathlineto{\pgfqpoint{0.811846in}{0.810873in}}%
\pgfpathlineto{\pgfqpoint{0.811846in}{0.757539in}}%
\pgfpathlineto{\pgfqpoint{0.895002in}{0.757539in}}%
\pgfpathlineto{\pgfqpoint{0.895002in}{0.688902in}}%
\pgfpathlineto{\pgfqpoint{0.978159in}{0.688902in}}%
\pgfpathlineto{\pgfqpoint{0.978159in}{0.599858in}}%
\pgfpathlineto{\pgfqpoint{1.061316in}{0.599858in}}%
\pgfpathlineto{\pgfqpoint{1.061316in}{0.525655in}}%
\pgfpathlineto{\pgfqpoint{1.144472in}{0.525655in}}%
\pgfpathlineto{\pgfqpoint{1.144472in}{0.460728in}}%
\pgfpathlineto{\pgfqpoint{1.227629in}{0.460728in}}%
\pgfpathlineto{\pgfqpoint{1.227629in}{0.442177in}}%
\pgfpathlineto{\pgfqpoint{1.269207in}{0.442177in}}%
\pgfpathlineto{\pgfqpoint{1.269207in}{0.442177in}}%
\pgfpathlineto{\pgfqpoint{1.269207in}{0.442177in}}%
\pgfpathlineto{\pgfqpoint{1.227629in}{0.442177in}}%
\pgfpathlineto{\pgfqpoint{1.227629in}{0.460728in}}%
\pgfpathlineto{\pgfqpoint{1.144472in}{0.460728in}}%
\pgfpathlineto{\pgfqpoint{1.144472in}{0.525655in}}%
\pgfpathlineto{\pgfqpoint{1.061316in}{0.525655in}}%
\pgfpathlineto{\pgfqpoint{1.061316in}{0.599858in}}%
\pgfpathlineto{\pgfqpoint{0.978159in}{0.599858in}}%
\pgfpathlineto{\pgfqpoint{0.978159in}{0.688902in}}%
\pgfpathlineto{\pgfqpoint{0.895002in}{0.688902in}}%
\pgfpathlineto{\pgfqpoint{0.895002in}{0.757539in}}%
\pgfpathlineto{\pgfqpoint{0.811846in}{0.757539in}}%
\pgfpathlineto{\pgfqpoint{0.811846in}{0.810873in}}%
\pgfpathlineto{\pgfqpoint{0.728689in}{0.810873in}}%
\pgfpathlineto{\pgfqpoint{0.728689in}{0.846119in}}%
\pgfpathlineto{\pgfqpoint{0.645533in}{0.846119in}}%
\pgfpathlineto{\pgfqpoint{0.645533in}{0.876728in}}%
\pgfpathlineto{\pgfqpoint{0.562376in}{0.876728in}}%
\pgfpathlineto{\pgfqpoint{0.562376in}{0.910583in}}%
\pgfpathlineto{\pgfqpoint{0.520798in}{0.910583in}}%
\pgfpathlineto{\pgfqpoint{0.520798in}{0.910583in}}%
\pgfpathclose%
\pgfusepath{fill}%
}%
\begin{pgfscope}%
\pgfsys@transformshift{0.000000in}{0.000000in}%
\pgfsys@useobject{currentmarker}{}%
\end{pgfscope}%
\end{pgfscope}%
\begin{pgfscope}%
\definecolor{textcolor}{rgb}{0.150000,0.150000,0.150000}%
\pgfsetstrokecolor{textcolor}%
\pgfsetfillcolor{textcolor}%
\pgftext[x=0.895002in,y=1.397395in,,base]{\color{textcolor}\rmfamily\fontsize{9.000000}{10.800000}\selectfont \(\displaystyle c_A^+=1\)}%
\end{pgfscope}%
\begin{pgfscope}%
\pgfsetbuttcap%
\pgfsetmiterjoin%
\definecolor{currentfill}{rgb}{1.000000,1.000000,1.000000}%
\pgfsetfillcolor{currentfill}%
\pgfsetfillopacity{0.800000}%
\pgfsetlinewidth{1.003750pt}%
\definecolor{currentstroke}{rgb}{0.800000,0.800000,0.800000}%
\pgfsetstrokecolor{currentstroke}%
\pgfsetstrokeopacity{0.800000}%
\pgfsetdash{}{0pt}%
\pgfpathmoveto{\pgfqpoint{0.815247in}{0.638103in}}%
\pgfpathlineto{\pgfqpoint{1.220596in}{0.638103in}}%
\pgfpathlineto{\pgfqpoint{1.220596in}{1.265450in}}%
\pgfpathlineto{\pgfqpoint{0.815247in}{1.265450in}}%
\pgfpathlineto{\pgfqpoint{0.815247in}{0.638103in}}%
\pgfpathclose%
\pgfusepath{stroke,fill}%
\end{pgfscope}%
\begin{pgfscope}%
\definecolor{textcolor}{rgb}{0.150000,0.150000,0.150000}%
\pgfsetstrokecolor{textcolor}%
\pgfsetfillcolor{textcolor}%
\pgftext[x=0.946517in,y=1.113275in,left,base]{\color{textcolor}\rmfamily\fontsize{9.000000}{10.800000}\selectfont \(\displaystyle c_A^-\)}%
\end{pgfscope}%
\begin{pgfscope}%
\pgfsetroundcap%
\pgfsetroundjoin%
\pgfsetlinewidth{1.003750pt}%
\definecolor{currentstroke}{rgb}{0.003922,0.450980,0.698039}%
\pgfsetstrokecolor{currentstroke}%
\pgfsetdash{}{0pt}%
\pgfpathmoveto{\pgfqpoint{0.854136in}{1.001052in}}%
\pgfpathlineto{\pgfqpoint{0.902747in}{1.001052in}}%
\pgfpathlineto{\pgfqpoint{0.902747in}{1.001052in}}%
\pgfpathlineto{\pgfqpoint{0.999969in}{1.001052in}}%
\pgfpathlineto{\pgfqpoint{0.999969in}{1.001052in}}%
\pgfpathlineto{\pgfqpoint{1.048580in}{1.001052in}}%
\pgfusepath{stroke}%
\end{pgfscope}%
\begin{pgfscope}%
\definecolor{textcolor}{rgb}{0.150000,0.150000,0.150000}%
\pgfsetstrokecolor{textcolor}%
\pgfsetfillcolor{textcolor}%
\pgftext[x=1.126358in,y=0.967024in,left,base]{\color{textcolor}\rmfamily\fontsize{7.000000}{8.400000}\selectfont 1}%
\end{pgfscope}%
\begin{pgfscope}%
\pgfsetroundcap%
\pgfsetroundjoin%
\pgfsetlinewidth{1.003750pt}%
\definecolor{currentstroke}{rgb}{0.870588,0.560784,0.019608}%
\pgfsetstrokecolor{currentstroke}%
\pgfsetdash{}{0pt}%
\pgfpathmoveto{\pgfqpoint{0.854136in}{0.865486in}}%
\pgfpathlineto{\pgfqpoint{0.902747in}{0.865486in}}%
\pgfpathlineto{\pgfqpoint{0.902747in}{0.865486in}}%
\pgfpathlineto{\pgfqpoint{0.999969in}{0.865486in}}%
\pgfpathlineto{\pgfqpoint{0.999969in}{0.865486in}}%
\pgfpathlineto{\pgfqpoint{1.048580in}{0.865486in}}%
\pgfusepath{stroke}%
\end{pgfscope}%
\begin{pgfscope}%
\definecolor{textcolor}{rgb}{0.150000,0.150000,0.150000}%
\pgfsetstrokecolor{textcolor}%
\pgfsetfillcolor{textcolor}%
\pgftext[x=1.126358in,y=0.831458in,left,base]{\color{textcolor}\rmfamily\fontsize{7.000000}{8.400000}\selectfont 2}%
\end{pgfscope}%
\begin{pgfscope}%
\pgfsetroundcap%
\pgfsetroundjoin%
\pgfsetlinewidth{1.003750pt}%
\definecolor{currentstroke}{rgb}{0.007843,0.619608,0.450980}%
\pgfsetstrokecolor{currentstroke}%
\pgfsetdash{}{0pt}%
\pgfpathmoveto{\pgfqpoint{0.854136in}{0.729919in}}%
\pgfpathlineto{\pgfqpoint{0.902747in}{0.729919in}}%
\pgfpathlineto{\pgfqpoint{0.902747in}{0.729919in}}%
\pgfpathlineto{\pgfqpoint{0.999969in}{0.729919in}}%
\pgfpathlineto{\pgfqpoint{0.999969in}{0.729919in}}%
\pgfpathlineto{\pgfqpoint{1.048580in}{0.729919in}}%
\pgfusepath{stroke}%
\end{pgfscope}%
\begin{pgfscope}%
\definecolor{textcolor}{rgb}{0.150000,0.150000,0.150000}%
\pgfsetstrokecolor{textcolor}%
\pgfsetfillcolor{textcolor}%
\pgftext[x=1.126358in,y=0.695892in,left,base]{\color{textcolor}\rmfamily\fontsize{7.000000}{8.400000}\selectfont 4}%
\end{pgfscope}%
\end{pgfpicture}%
\makeatother%
\endgroup%

%% file: figures/experiments/graphs/sparse_smoothing/nodes_attributes/node_classification-Cora-GCN-hidden=32-p_adj_plus=0.0-p_adj_minus=0.0-p_att_plus=0.001-p_att_minus=0.8-multi_class_cert-B.pgf
\begingroup%
\makeatletter%
\begin{pgfpicture}%
\pgfpathrectangle{\pgfpointorigin}{\pgfqpoint{1.375000in}{1.581250in}}%
\pgfusepath{use as bounding box, clip}%
\begin{pgfscope}%
\pgfsetbuttcap%
\pgfsetmiterjoin%
\definecolor{currentfill}{rgb}{1.000000,1.000000,1.000000}%
\pgfsetfillcolor{currentfill}%
\pgfsetlinewidth{0.000000pt}%
\definecolor{currentstroke}{rgb}{1.000000,1.000000,1.000000}%
\pgfsetstrokecolor{currentstroke}%
\pgfsetdash{}{0pt}%
\pgfpathmoveto{\pgfqpoint{0.000000in}{0.000000in}}%
\pgfpathlineto{\pgfqpoint{1.375000in}{0.000000in}}%
\pgfpathlineto{\pgfqpoint{1.375000in}{1.581250in}}%
\pgfpathlineto{\pgfqpoint{0.000000in}{1.581250in}}%
\pgfpathlineto{\pgfqpoint{0.000000in}{0.000000in}}%
\pgfpathclose%
\pgfusepath{fill}%
\end{pgfscope}%
\begin{pgfscope}%
\pgfsetbuttcap%
\pgfsetmiterjoin%
\definecolor{currentfill}{rgb}{1.000000,1.000000,1.000000}%
\pgfsetfillcolor{currentfill}%
\pgfsetlinewidth{0.000000pt}%
\definecolor{currentstroke}{rgb}{0.000000,0.000000,0.000000}%
\pgfsetstrokecolor{currentstroke}%
\pgfsetstrokeopacity{0.000000}%
\pgfsetdash{}{0pt}%
\pgfpathmoveto{\pgfqpoint{0.520798in}{0.442177in}}%
\pgfpathlineto{\pgfqpoint{1.239803in}{0.442177in}}%
\pgfpathlineto{\pgfqpoint{1.239803in}{1.314061in}}%
\pgfpathlineto{\pgfqpoint{0.520798in}{1.314061in}}%
\pgfpathlineto{\pgfqpoint{0.520798in}{0.442177in}}%
\pgfpathclose%
\pgfusepath{fill}%
\end{pgfscope}%
\begin{pgfscope}%
\pgfpathrectangle{\pgfqpoint{0.520798in}{0.442177in}}{\pgfqpoint{0.719005in}{0.871884in}}%
\pgfusepath{clip}%
\pgfsetroundcap%
\pgfsetroundjoin%
\pgfsetlinewidth{0.501875pt}%
\definecolor{currentstroke}{rgb}{0.800000,0.800000,0.800000}%
\pgfsetstrokecolor{currentstroke}%
\pgfsetdash{}{0pt}%
\pgfpathmoveto{\pgfqpoint{0.520798in}{0.442177in}}%
\pgfpathlineto{\pgfqpoint{0.520798in}{1.314061in}}%
\pgfusepath{stroke}%
\end{pgfscope}%
\begin{pgfscope}%
\definecolor{textcolor}{rgb}{0.150000,0.150000,0.150000}%
\pgfsetstrokecolor{textcolor}%
\pgfsetfillcolor{textcolor}%
\pgftext[x=0.520798in,y=0.351899in,,top]{\color{textcolor}\rmfamily\fontsize{8.000000}{9.600000}\selectfont \(\displaystyle {0}\)}%
\end{pgfscope}%
\begin{pgfscope}%
\pgfpathrectangle{\pgfqpoint{0.520798in}{0.442177in}}{\pgfqpoint{0.719005in}{0.871884in}}%
\pgfusepath{clip}%
\pgfsetroundcap%
\pgfsetroundjoin%
\pgfsetlinewidth{0.501875pt}%
\definecolor{currentstroke}{rgb}{0.800000,0.800000,0.800000}%
\pgfsetstrokecolor{currentstroke}%
\pgfsetdash{}{0pt}%
\pgfpathmoveto{\pgfqpoint{0.880300in}{0.442177in}}%
\pgfpathlineto{\pgfqpoint{0.880300in}{1.314061in}}%
\pgfusepath{stroke}%
\end{pgfscope}%
\begin{pgfscope}%
\definecolor{textcolor}{rgb}{0.150000,0.150000,0.150000}%
\pgfsetstrokecolor{textcolor}%
\pgfsetfillcolor{textcolor}%
\pgftext[x=0.880300in,y=0.351899in,,top]{\color{textcolor}\rmfamily\fontsize{8.000000}{9.600000}\selectfont \(\displaystyle {5}\)}%
\end{pgfscope}%
\begin{pgfscope}%
\pgfpathrectangle{\pgfqpoint{0.520798in}{0.442177in}}{\pgfqpoint{0.719005in}{0.871884in}}%
\pgfusepath{clip}%
\pgfsetroundcap%
\pgfsetroundjoin%
\pgfsetlinewidth{0.501875pt}%
\definecolor{currentstroke}{rgb}{0.800000,0.800000,0.800000}%
\pgfsetstrokecolor{currentstroke}%
\pgfsetdash{}{0pt}%
\pgfpathmoveto{\pgfqpoint{1.239803in}{0.442177in}}%
\pgfpathlineto{\pgfqpoint{1.239803in}{1.314061in}}%
\pgfusepath{stroke}%
\end{pgfscope}%
\begin{pgfscope}%
\definecolor{textcolor}{rgb}{0.150000,0.150000,0.150000}%
\pgfsetstrokecolor{textcolor}%
\pgfsetfillcolor{textcolor}%
\pgftext[x=1.239803in,y=0.351899in,,top]{\color{textcolor}\rmfamily\fontsize{8.000000}{9.600000}\selectfont \(\displaystyle {10}\)}%
\end{pgfscope}%
\begin{pgfscope}%
\definecolor{textcolor}{rgb}{0.150000,0.150000,0.150000}%
\pgfsetstrokecolor{textcolor}%
\pgfsetfillcolor{textcolor}%
\pgftext[x=0.880300in,y=0.198219in,,top]{\color{textcolor}\rmfamily\fontsize{10.000000}{12.000000}\selectfont Edit distance \(\displaystyle \epsilon\)}%
\end{pgfscope}%
\begin{pgfscope}%
\pgfpathrectangle{\pgfqpoint{0.520798in}{0.442177in}}{\pgfqpoint{0.719005in}{0.871884in}}%
\pgfusepath{clip}%
\pgfsetroundcap%
\pgfsetroundjoin%
\pgfsetlinewidth{0.501875pt}%
\definecolor{currentstroke}{rgb}{0.800000,0.800000,0.800000}%
\pgfsetstrokecolor{currentstroke}%
\pgfsetdash{}{0pt}%
\pgfpathmoveto{\pgfqpoint{0.520798in}{0.442177in}}%
\pgfpathlineto{\pgfqpoint{1.239803in}{0.442177in}}%
\pgfusepath{stroke}%
\end{pgfscope}%
\begin{pgfscope}%
\definecolor{textcolor}{rgb}{0.150000,0.150000,0.150000}%
\pgfsetstrokecolor{textcolor}%
\pgfsetfillcolor{textcolor}%
\pgftext[x=0.273151in, y=0.403915in, left, base]{\color{textcolor}\rmfamily\fontsize{8.000000}{9.600000}\selectfont 0\%}%
\end{pgfscope}%
\begin{pgfscope}%
\pgfpathrectangle{\pgfqpoint{0.520798in}{0.442177in}}{\pgfqpoint{0.719005in}{0.871884in}}%
\pgfusepath{clip}%
\pgfsetroundcap%
\pgfsetroundjoin%
\pgfsetlinewidth{0.501875pt}%
\definecolor{currentstroke}{rgb}{0.800000,0.800000,0.800000}%
\pgfsetstrokecolor{currentstroke}%
\pgfsetdash{}{0pt}%
\pgfpathmoveto{\pgfqpoint{0.520798in}{0.660148in}}%
\pgfpathlineto{\pgfqpoint{1.239803in}{0.660148in}}%
\pgfusepath{stroke}%
\end{pgfscope}%
\begin{pgfscope}%
\definecolor{textcolor}{rgb}{0.150000,0.150000,0.150000}%
\pgfsetstrokecolor{textcolor}%
\pgfsetfillcolor{textcolor}%
\pgftext[x=0.214138in, y=0.621886in, left, base]{\color{textcolor}\rmfamily\fontsize{8.000000}{9.600000}\selectfont 25\%}%
\end{pgfscope}%
\begin{pgfscope}%
\pgfpathrectangle{\pgfqpoint{0.520798in}{0.442177in}}{\pgfqpoint{0.719005in}{0.871884in}}%
\pgfusepath{clip}%
\pgfsetroundcap%
\pgfsetroundjoin%
\pgfsetlinewidth{0.501875pt}%
\definecolor{currentstroke}{rgb}{0.800000,0.800000,0.800000}%
\pgfsetstrokecolor{currentstroke}%
\pgfsetdash{}{0pt}%
\pgfpathmoveto{\pgfqpoint{0.520798in}{0.878119in}}%
\pgfpathlineto{\pgfqpoint{1.239803in}{0.878119in}}%
\pgfusepath{stroke}%
\end{pgfscope}%
\begin{pgfscope}%
\definecolor{textcolor}{rgb}{0.150000,0.150000,0.150000}%
\pgfsetstrokecolor{textcolor}%
\pgfsetfillcolor{textcolor}%
\pgftext[x=0.214138in, y=0.839857in, left, base]{\color{textcolor}\rmfamily\fontsize{8.000000}{9.600000}\selectfont 50\%}%
\end{pgfscope}%
\begin{pgfscope}%
\pgfpathrectangle{\pgfqpoint{0.520798in}{0.442177in}}{\pgfqpoint{0.719005in}{0.871884in}}%
\pgfusepath{clip}%
\pgfsetroundcap%
\pgfsetroundjoin%
\pgfsetlinewidth{0.501875pt}%
\definecolor{currentstroke}{rgb}{0.800000,0.800000,0.800000}%
\pgfsetstrokecolor{currentstroke}%
\pgfsetdash{}{0pt}%
\pgfpathmoveto{\pgfqpoint{0.520798in}{1.096090in}}%
\pgfpathlineto{\pgfqpoint{1.239803in}{1.096090in}}%
\pgfusepath{stroke}%
\end{pgfscope}%
\begin{pgfscope}%
\definecolor{textcolor}{rgb}{0.150000,0.150000,0.150000}%
\pgfsetstrokecolor{textcolor}%
\pgfsetfillcolor{textcolor}%
\pgftext[x=0.214138in, y=1.057828in, left, base]{\color{textcolor}\rmfamily\fontsize{8.000000}{9.600000}\selectfont 75\%}%
\end{pgfscope}%
\begin{pgfscope}%
\pgfpathrectangle{\pgfqpoint{0.520798in}{0.442177in}}{\pgfqpoint{0.719005in}{0.871884in}}%
\pgfusepath{clip}%
\pgfsetroundcap%
\pgfsetroundjoin%
\pgfsetlinewidth{0.501875pt}%
\definecolor{currentstroke}{rgb}{0.800000,0.800000,0.800000}%
\pgfsetstrokecolor{currentstroke}%
\pgfsetdash{}{0pt}%
\pgfpathmoveto{\pgfqpoint{0.520798in}{1.314061in}}%
\pgfpathlineto{\pgfqpoint{1.239803in}{1.314061in}}%
\pgfusepath{stroke}%
\end{pgfscope}%
\begin{pgfscope}%
\definecolor{textcolor}{rgb}{0.150000,0.150000,0.150000}%
\pgfsetstrokecolor{textcolor}%
\pgfsetfillcolor{textcolor}%
\pgftext[x=0.155124in, y=1.275799in, left, base]{\color{textcolor}\rmfamily\fontsize{8.000000}{9.600000}\selectfont 100\%}%
\end{pgfscope}%
\begin{pgfscope}%
\definecolor{textcolor}{rgb}{0.150000,0.150000,0.150000}%
\pgfsetstrokecolor{textcolor}%
\pgfsetfillcolor{textcolor}%
\pgftext[x=0.099569in,y=0.878119in,,bottom,rotate=90.000000]{\color{textcolor}\rmfamily\fontsize{10.000000}{12.000000}\selectfont Cert. Acc.}%
\end{pgfscope}%
\begin{pgfscope}%
\pgfsetrectcap%
\pgfsetmiterjoin%
\pgfsetlinewidth{0.752812pt}%
\definecolor{currentstroke}{rgb}{0.700000,0.700000,0.700000}%
\pgfsetstrokecolor{currentstroke}%
\pgfsetdash{}{0pt}%
\pgfpathmoveto{\pgfqpoint{0.520798in}{0.442177in}}%
\pgfpathlineto{\pgfqpoint{0.520798in}{1.314061in}}%
\pgfusepath{stroke}%
\end{pgfscope}%
\begin{pgfscope}%
\pgfsetrectcap%
\pgfsetmiterjoin%
\pgfsetlinewidth{0.752812pt}%
\definecolor{currentstroke}{rgb}{0.700000,0.700000,0.700000}%
\pgfsetstrokecolor{currentstroke}%
\pgfsetdash{}{0pt}%
\pgfpathmoveto{\pgfqpoint{1.239803in}{0.442177in}}%
\pgfpathlineto{\pgfqpoint{1.239803in}{1.314061in}}%
\pgfusepath{stroke}%
\end{pgfscope}%
\begin{pgfscope}%
\pgfsetrectcap%
\pgfsetmiterjoin%
\pgfsetlinewidth{0.752812pt}%
\definecolor{currentstroke}{rgb}{0.700000,0.700000,0.700000}%
\pgfsetstrokecolor{currentstroke}%
\pgfsetdash{}{0pt}%
\pgfpathmoveto{\pgfqpoint{0.520798in}{0.442177in}}%
\pgfpathlineto{\pgfqpoint{1.239803in}{0.442177in}}%
\pgfusepath{stroke}%
\end{pgfscope}%
\begin{pgfscope}%
\pgfsetrectcap%
\pgfsetmiterjoin%
\pgfsetlinewidth{0.752812pt}%
\definecolor{currentstroke}{rgb}{0.700000,0.700000,0.700000}%
\pgfsetstrokecolor{currentstroke}%
\pgfsetdash{}{0pt}%
\pgfpathmoveto{\pgfqpoint{0.520798in}{1.314061in}}%
\pgfpathlineto{\pgfqpoint{1.239803in}{1.314061in}}%
\pgfusepath{stroke}%
\end{pgfscope}%
\begin{pgfscope}%
\pgfpathrectangle{\pgfqpoint{0.520798in}{0.442177in}}{\pgfqpoint{0.719005in}{0.871884in}}%
\pgfusepath{clip}%
\pgfsetroundcap%
\pgfsetroundjoin%
\pgfsetlinewidth{1.003750pt}%
\definecolor{currentstroke}{rgb}{0.003922,0.450980,0.698039}%
\pgfsetstrokecolor{currentstroke}%
\pgfsetdash{}{0pt}%
\pgfpathmoveto{\pgfqpoint{0.520798in}{1.158190in}}%
\pgfpathlineto{\pgfqpoint{0.556748in}{1.158190in}}%
\pgfpathlineto{\pgfqpoint{0.556748in}{1.090495in}}%
\pgfpathlineto{\pgfqpoint{0.628649in}{1.090495in}}%
\pgfpathlineto{\pgfqpoint{0.628649in}{0.978040in}}%
\pgfpathlineto{\pgfqpoint{0.700549in}{0.978040in}}%
\pgfpathlineto{\pgfqpoint{0.700549in}{0.703624in}}%
\pgfpathlineto{\pgfqpoint{0.772450in}{0.703624in}}%
\pgfpathlineto{\pgfqpoint{0.772450in}{0.585949in}}%
\pgfpathlineto{\pgfqpoint{0.844350in}{0.585949in}}%
\pgfpathlineto{\pgfqpoint{0.844350in}{0.569816in}}%
\pgfpathlineto{\pgfqpoint{0.916251in}{0.569816in}}%
\pgfpathlineto{\pgfqpoint{0.916251in}{0.561275in}}%
\pgfpathlineto{\pgfqpoint{0.988151in}{0.561275in}}%
\pgfpathlineto{\pgfqpoint{0.988151in}{0.547673in}}%
\pgfpathlineto{\pgfqpoint{1.060052in}{0.547673in}}%
\pgfpathlineto{\pgfqpoint{1.060052in}{0.507657in}}%
\pgfpathlineto{\pgfqpoint{1.131952in}{0.507657in}}%
\pgfpathlineto{\pgfqpoint{1.131952in}{0.500461in}}%
\pgfpathlineto{\pgfqpoint{1.203853in}{0.500461in}}%
\pgfpathlineto{\pgfqpoint{1.203853in}{0.481244in}}%
\pgfpathlineto{\pgfqpoint{1.241470in}{0.481244in}}%
\pgfusepath{stroke}%
\end{pgfscope}%
\begin{pgfscope}%
\pgfpathrectangle{\pgfqpoint{0.520798in}{0.442177in}}{\pgfqpoint{0.719005in}{0.871884in}}%
\pgfusepath{clip}%
\pgfsetbuttcap%
\pgfsetroundjoin%
\definecolor{currentfill}{rgb}{0.003922,0.450980,0.698039}%
\pgfsetfillcolor{currentfill}%
\pgfsetfillopacity{0.500000}%
\pgfsetlinewidth{0.000000pt}%
\definecolor{currentstroke}{rgb}{0.003922,0.450980,0.698039}%
\pgfsetstrokecolor{currentstroke}%
\pgfsetstrokeopacity{0.500000}%
\pgfsetdash{}{0pt}%
\pgfpathmoveto{\pgfqpoint{0.520798in}{1.173204in}}%
\pgfpathlineto{\pgfqpoint{0.520798in}{1.143176in}}%
\pgfpathlineto{\pgfqpoint{0.556748in}{1.143176in}}%
\pgfpathlineto{\pgfqpoint{0.556748in}{1.071028in}}%
\pgfpathlineto{\pgfqpoint{0.628649in}{1.071028in}}%
\pgfpathlineto{\pgfqpoint{0.628649in}{0.960594in}}%
\pgfpathlineto{\pgfqpoint{0.700549in}{0.960594in}}%
\pgfpathlineto{\pgfqpoint{0.700549in}{0.696977in}}%
\pgfpathlineto{\pgfqpoint{0.772450in}{0.696977in}}%
\pgfpathlineto{\pgfqpoint{0.772450in}{0.575272in}}%
\pgfpathlineto{\pgfqpoint{0.844350in}{0.575272in}}%
\pgfpathlineto{\pgfqpoint{0.844350in}{0.558900in}}%
\pgfpathlineto{\pgfqpoint{0.916251in}{0.558900in}}%
\pgfpathlineto{\pgfqpoint{0.916251in}{0.550809in}}%
\pgfpathlineto{\pgfqpoint{0.988151in}{0.550809in}}%
\pgfpathlineto{\pgfqpoint{0.988151in}{0.538459in}}%
\pgfpathlineto{\pgfqpoint{1.060052in}{0.538459in}}%
\pgfpathlineto{\pgfqpoint{1.060052in}{0.499955in}}%
\pgfpathlineto{\pgfqpoint{1.131952in}{0.499955in}}%
\pgfpathlineto{\pgfqpoint{1.131952in}{0.492637in}}%
\pgfpathlineto{\pgfqpoint{1.203853in}{0.492637in}}%
\pgfpathlineto{\pgfqpoint{1.203853in}{0.476098in}}%
\pgfpathlineto{\pgfqpoint{1.275753in}{0.476098in}}%
\pgfpathlineto{\pgfqpoint{1.275753in}{0.460224in}}%
\pgfpathlineto{\pgfqpoint{1.347654in}{0.460224in}}%
\pgfpathlineto{\pgfqpoint{1.347654in}{0.455551in}}%
\pgfpathlineto{\pgfqpoint{1.419554in}{0.455551in}}%
\pgfpathlineto{\pgfqpoint{1.419554in}{0.442177in}}%
\pgfpathlineto{\pgfqpoint{2.066659in}{0.442177in}}%
\pgfpathlineto{\pgfqpoint{2.066659in}{0.442177in}}%
\pgfpathlineto{\pgfqpoint{2.677813in}{0.442177in}}%
\pgfpathlineto{\pgfqpoint{2.677813in}{0.442177in}}%
\pgfpathlineto{\pgfqpoint{2.677813in}{0.442177in}}%
\pgfpathlineto{\pgfqpoint{2.066659in}{0.442177in}}%
\pgfpathlineto{\pgfqpoint{2.066659in}{0.442177in}}%
\pgfpathlineto{\pgfqpoint{1.419554in}{0.442177in}}%
\pgfpathlineto{\pgfqpoint{1.419554in}{0.462492in}}%
\pgfpathlineto{\pgfqpoint{1.347654in}{0.462492in}}%
\pgfpathlineto{\pgfqpoint{1.347654in}{0.466676in}}%
\pgfpathlineto{\pgfqpoint{1.275753in}{0.466676in}}%
\pgfpathlineto{\pgfqpoint{1.275753in}{0.486389in}}%
\pgfpathlineto{\pgfqpoint{1.203853in}{0.486389in}}%
\pgfpathlineto{\pgfqpoint{1.203853in}{0.508284in}}%
\pgfpathlineto{\pgfqpoint{1.131952in}{0.508284in}}%
\pgfpathlineto{\pgfqpoint{1.131952in}{0.515359in}}%
\pgfpathlineto{\pgfqpoint{1.060052in}{0.515359in}}%
\pgfpathlineto{\pgfqpoint{1.060052in}{0.556887in}}%
\pgfpathlineto{\pgfqpoint{0.988151in}{0.556887in}}%
\pgfpathlineto{\pgfqpoint{0.988151in}{0.571741in}}%
\pgfpathlineto{\pgfqpoint{0.916251in}{0.571741in}}%
\pgfpathlineto{\pgfqpoint{0.916251in}{0.580732in}}%
\pgfpathlineto{\pgfqpoint{0.844350in}{0.580732in}}%
\pgfpathlineto{\pgfqpoint{0.844350in}{0.596625in}}%
\pgfpathlineto{\pgfqpoint{0.772450in}{0.596625in}}%
\pgfpathlineto{\pgfqpoint{0.772450in}{0.710270in}}%
\pgfpathlineto{\pgfqpoint{0.700549in}{0.710270in}}%
\pgfpathlineto{\pgfqpoint{0.700549in}{0.995485in}}%
\pgfpathlineto{\pgfqpoint{0.628649in}{0.995485in}}%
\pgfpathlineto{\pgfqpoint{0.628649in}{1.109962in}}%
\pgfpathlineto{\pgfqpoint{0.556748in}{1.109962in}}%
\pgfpathlineto{\pgfqpoint{0.556748in}{1.173204in}}%
\pgfpathlineto{\pgfqpoint{0.520798in}{1.173204in}}%
\pgfpathlineto{\pgfqpoint{0.520798in}{1.173204in}}%
\pgfpathclose%
\pgfusepath{fill}%
\end{pgfscope}%
\begin{pgfscope}%
\pgfpathrectangle{\pgfqpoint{0.520798in}{0.442177in}}{\pgfqpoint{0.719005in}{0.871884in}}%
\pgfusepath{clip}%
\pgfsetroundcap%
\pgfsetroundjoin%
\pgfsetlinewidth{1.003750pt}%
\definecolor{currentstroke}{rgb}{0.870588,0.560784,0.019608}%
\pgfsetstrokecolor{currentstroke}%
\pgfsetdash{}{0pt}%
\pgfpathmoveto{\pgfqpoint{0.520798in}{1.158190in}}%
\pgfpathlineto{\pgfqpoint{0.556748in}{1.158190in}}%
\pgfpathlineto{\pgfqpoint{0.556748in}{1.107261in}}%
\pgfpathlineto{\pgfqpoint{0.628649in}{1.107261in}}%
\pgfpathlineto{\pgfqpoint{0.628649in}{1.054433in}}%
\pgfpathlineto{\pgfqpoint{0.700549in}{1.054433in}}%
\pgfpathlineto{\pgfqpoint{0.700549in}{1.006589in}}%
\pgfpathlineto{\pgfqpoint{0.772450in}{1.006589in}}%
\pgfpathlineto{\pgfqpoint{0.772450in}{0.966652in}}%
\pgfpathlineto{\pgfqpoint{0.844350in}{0.966652in}}%
\pgfpathlineto{\pgfqpoint{0.844350in}{0.929404in}}%
\pgfpathlineto{\pgfqpoint{0.916251in}{0.929404in}}%
\pgfpathlineto{\pgfqpoint{0.916251in}{0.703624in}}%
\pgfpathlineto{\pgfqpoint{0.988151in}{0.703624in}}%
\pgfpathlineto{\pgfqpoint{0.988151in}{0.687412in}}%
\pgfpathlineto{\pgfqpoint{1.060052in}{0.687412in}}%
\pgfpathlineto{\pgfqpoint{1.060052in}{0.585949in}}%
\pgfpathlineto{\pgfqpoint{1.131952in}{0.585949in}}%
\pgfpathlineto{\pgfqpoint{1.131952in}{0.575668in}}%
\pgfpathlineto{\pgfqpoint{1.203853in}{0.575668in}}%
\pgfpathlineto{\pgfqpoint{1.203853in}{0.565704in}}%
\pgfpathlineto{\pgfqpoint{1.241470in}{0.565704in}}%
\pgfusepath{stroke}%
\end{pgfscope}%
\begin{pgfscope}%
\pgfpathrectangle{\pgfqpoint{0.520798in}{0.442177in}}{\pgfqpoint{0.719005in}{0.871884in}}%
\pgfusepath{clip}%
\pgfsetbuttcap%
\pgfsetroundjoin%
\definecolor{currentfill}{rgb}{0.870588,0.560784,0.019608}%
\pgfsetfillcolor{currentfill}%
\pgfsetfillopacity{0.500000}%
\pgfsetlinewidth{0.000000pt}%
\definecolor{currentstroke}{rgb}{0.870588,0.560784,0.019608}%
\pgfsetstrokecolor{currentstroke}%
\pgfsetstrokeopacity{0.500000}%
\pgfsetdash{}{0pt}%
\pgfpathmoveto{\pgfqpoint{0.520798in}{1.173204in}}%
\pgfpathlineto{\pgfqpoint{0.520798in}{1.143176in}}%
\pgfpathlineto{\pgfqpoint{0.556748in}{1.143176in}}%
\pgfpathlineto{\pgfqpoint{0.556748in}{1.089244in}}%
\pgfpathlineto{\pgfqpoint{0.628649in}{1.089244in}}%
\pgfpathlineto{\pgfqpoint{0.628649in}{1.037364in}}%
\pgfpathlineto{\pgfqpoint{0.700549in}{1.037364in}}%
\pgfpathlineto{\pgfqpoint{0.700549in}{0.989410in}}%
\pgfpathlineto{\pgfqpoint{0.772450in}{0.989410in}}%
\pgfpathlineto{\pgfqpoint{0.772450in}{0.949253in}}%
\pgfpathlineto{\pgfqpoint{0.844350in}{0.949253in}}%
\pgfpathlineto{\pgfqpoint{0.844350in}{0.914783in}}%
\pgfpathlineto{\pgfqpoint{0.916251in}{0.914783in}}%
\pgfpathlineto{\pgfqpoint{0.916251in}{0.696977in}}%
\pgfpathlineto{\pgfqpoint{0.988151in}{0.696977in}}%
\pgfpathlineto{\pgfqpoint{0.988151in}{0.679970in}}%
\pgfpathlineto{\pgfqpoint{1.060052in}{0.679970in}}%
\pgfpathlineto{\pgfqpoint{1.060052in}{0.575272in}}%
\pgfpathlineto{\pgfqpoint{1.131952in}{0.575272in}}%
\pgfpathlineto{\pgfqpoint{1.131952in}{0.564416in}}%
\pgfpathlineto{\pgfqpoint{1.203853in}{0.564416in}}%
\pgfpathlineto{\pgfqpoint{1.203853in}{0.554940in}}%
\pgfpathlineto{\pgfqpoint{1.275753in}{0.554940in}}%
\pgfpathlineto{\pgfqpoint{1.275753in}{0.547843in}}%
\pgfpathlineto{\pgfqpoint{1.347654in}{0.547843in}}%
\pgfpathlineto{\pgfqpoint{1.347654in}{0.541183in}}%
\pgfpathlineto{\pgfqpoint{1.419554in}{0.541183in}}%
\pgfpathlineto{\pgfqpoint{1.419554in}{0.536037in}}%
\pgfpathlineto{\pgfqpoint{1.491455in}{0.536037in}}%
\pgfpathlineto{\pgfqpoint{1.491455in}{0.530083in}}%
\pgfpathlineto{\pgfqpoint{1.563355in}{0.530083in}}%
\pgfpathlineto{\pgfqpoint{1.563355in}{0.525144in}}%
\pgfpathlineto{\pgfqpoint{1.635256in}{0.525144in}}%
\pgfpathlineto{\pgfqpoint{1.635256in}{0.498952in}}%
\pgfpathlineto{\pgfqpoint{1.707156in}{0.498952in}}%
\pgfpathlineto{\pgfqpoint{1.707156in}{0.496284in}}%
\pgfpathlineto{\pgfqpoint{1.779057in}{0.496284in}}%
\pgfpathlineto{\pgfqpoint{1.779057in}{0.490162in}}%
\pgfpathlineto{\pgfqpoint{1.850957in}{0.490162in}}%
\pgfpathlineto{\pgfqpoint{1.850957in}{0.486836in}}%
\pgfpathlineto{\pgfqpoint{1.922858in}{0.486836in}}%
\pgfpathlineto{\pgfqpoint{1.922858in}{0.475976in}}%
\pgfpathlineto{\pgfqpoint{1.994758in}{0.475976in}}%
\pgfpathlineto{\pgfqpoint{1.994758in}{0.460083in}}%
\pgfpathlineto{\pgfqpoint{2.066659in}{0.460083in}}%
\pgfpathlineto{\pgfqpoint{2.066659in}{0.456510in}}%
\pgfpathlineto{\pgfqpoint{2.138559in}{0.456510in}}%
\pgfpathlineto{\pgfqpoint{2.138559in}{0.454840in}}%
\pgfpathlineto{\pgfqpoint{2.210460in}{0.454840in}}%
\pgfpathlineto{\pgfqpoint{2.210460in}{0.448173in}}%
\pgfpathlineto{\pgfqpoint{2.282360in}{0.448173in}}%
\pgfpathlineto{\pgfqpoint{2.282360in}{0.442177in}}%
\pgfpathlineto{\pgfqpoint{2.498062in}{0.442177in}}%
\pgfpathlineto{\pgfqpoint{2.498062in}{0.442177in}}%
\pgfpathlineto{\pgfqpoint{2.677813in}{0.442177in}}%
\pgfpathlineto{\pgfqpoint{2.677813in}{0.442177in}}%
\pgfpathlineto{\pgfqpoint{2.677813in}{0.442177in}}%
\pgfpathlineto{\pgfqpoint{2.498062in}{0.442177in}}%
\pgfpathlineto{\pgfqpoint{2.498062in}{0.442177in}}%
\pgfpathlineto{\pgfqpoint{2.282360in}{0.442177in}}%
\pgfpathlineto{\pgfqpoint{2.282360in}{0.452630in}}%
\pgfpathlineto{\pgfqpoint{2.210460in}{0.452630in}}%
\pgfpathlineto{\pgfqpoint{2.210460in}{0.462095in}}%
\pgfpathlineto{\pgfqpoint{2.138559in}{0.462095in}}%
\pgfpathlineto{\pgfqpoint{2.138559in}{0.464538in}}%
\pgfpathlineto{\pgfqpoint{2.066659in}{0.464538in}}%
\pgfpathlineto{\pgfqpoint{2.066659in}{0.466500in}}%
\pgfpathlineto{\pgfqpoint{1.994758in}{0.466500in}}%
\pgfpathlineto{\pgfqpoint{1.994758in}{0.485879in}}%
\pgfpathlineto{\pgfqpoint{1.922858in}{0.485879in}}%
\pgfpathlineto{\pgfqpoint{1.922858in}{0.501432in}}%
\pgfpathlineto{\pgfqpoint{1.850957in}{0.501432in}}%
\pgfpathlineto{\pgfqpoint{1.850957in}{0.506172in}}%
\pgfpathlineto{\pgfqpoint{1.779057in}{0.506172in}}%
\pgfpathlineto{\pgfqpoint{1.779057in}{0.511597in}}%
\pgfpathlineto{\pgfqpoint{1.707156in}{0.511597in}}%
\pgfpathlineto{\pgfqpoint{1.707156in}{0.515255in}}%
\pgfpathlineto{\pgfqpoint{1.635256in}{0.515255in}}%
\pgfpathlineto{\pgfqpoint{1.635256in}{0.542523in}}%
\pgfpathlineto{\pgfqpoint{1.563355in}{0.542523in}}%
\pgfpathlineto{\pgfqpoint{1.563355in}{0.546124in}}%
\pgfpathlineto{\pgfqpoint{1.491455in}{0.546124in}}%
\pgfpathlineto{\pgfqpoint{1.491455in}{0.551875in}}%
\pgfpathlineto{\pgfqpoint{1.419554in}{0.551875in}}%
\pgfpathlineto{\pgfqpoint{1.419554in}{0.560331in}}%
\pgfpathlineto{\pgfqpoint{1.347654in}{0.560331in}}%
\pgfpathlineto{\pgfqpoint{1.347654in}{0.568064in}}%
\pgfpathlineto{\pgfqpoint{1.275753in}{0.568064in}}%
\pgfpathlineto{\pgfqpoint{1.275753in}{0.576468in}}%
\pgfpathlineto{\pgfqpoint{1.203853in}{0.576468in}}%
\pgfpathlineto{\pgfqpoint{1.203853in}{0.586921in}}%
\pgfpathlineto{\pgfqpoint{1.131952in}{0.586921in}}%
\pgfpathlineto{\pgfqpoint{1.131952in}{0.596625in}}%
\pgfpathlineto{\pgfqpoint{1.060052in}{0.596625in}}%
\pgfpathlineto{\pgfqpoint{1.060052in}{0.694853in}}%
\pgfpathlineto{\pgfqpoint{0.988151in}{0.694853in}}%
\pgfpathlineto{\pgfqpoint{0.988151in}{0.710270in}}%
\pgfpathlineto{\pgfqpoint{0.916251in}{0.710270in}}%
\pgfpathlineto{\pgfqpoint{0.916251in}{0.944026in}}%
\pgfpathlineto{\pgfqpoint{0.844350in}{0.944026in}}%
\pgfpathlineto{\pgfqpoint{0.844350in}{0.984051in}}%
\pgfpathlineto{\pgfqpoint{0.772450in}{0.984051in}}%
\pgfpathlineto{\pgfqpoint{0.772450in}{1.023767in}}%
\pgfpathlineto{\pgfqpoint{0.700549in}{1.023767in}}%
\pgfpathlineto{\pgfqpoint{0.700549in}{1.071502in}}%
\pgfpathlineto{\pgfqpoint{0.628649in}{1.071502in}}%
\pgfpathlineto{\pgfqpoint{0.628649in}{1.125277in}}%
\pgfpathlineto{\pgfqpoint{0.556748in}{1.125277in}}%
\pgfpathlineto{\pgfqpoint{0.556748in}{1.173204in}}%
\pgfpathlineto{\pgfqpoint{0.520798in}{1.173204in}}%
\pgfpathlineto{\pgfqpoint{0.520798in}{1.173204in}}%
\pgfpathclose%
\pgfusepath{fill}%
\end{pgfscope}%
\begin{pgfscope}%
\pgfpathrectangle{\pgfqpoint{0.520798in}{0.442177in}}{\pgfqpoint{0.719005in}{0.871884in}}%
\pgfusepath{clip}%
\pgfsetroundcap%
\pgfsetroundjoin%
\pgfsetlinewidth{1.003750pt}%
\definecolor{currentstroke}{rgb}{0.007843,0.619608,0.450980}%
\pgfsetstrokecolor{currentstroke}%
\pgfsetdash{}{0pt}%
\pgfpathmoveto{\pgfqpoint{0.520798in}{1.158190in}}%
\pgfpathlineto{\pgfqpoint{0.556748in}{1.158190in}}%
\pgfpathlineto{\pgfqpoint{0.556748in}{1.107261in}}%
\pgfpathlineto{\pgfqpoint{0.628649in}{1.107261in}}%
\pgfpathlineto{\pgfqpoint{0.628649in}{1.054433in}}%
\pgfpathlineto{\pgfqpoint{0.700549in}{1.054433in}}%
\pgfpathlineto{\pgfqpoint{0.700549in}{1.006589in}}%
\pgfpathlineto{\pgfqpoint{0.772450in}{1.006589in}}%
\pgfpathlineto{\pgfqpoint{0.772450in}{0.966652in}}%
\pgfpathlineto{\pgfqpoint{0.844350in}{0.966652in}}%
\pgfpathlineto{\pgfqpoint{0.844350in}{0.929404in}}%
\pgfpathlineto{\pgfqpoint{0.916251in}{0.929404in}}%
\pgfpathlineto{\pgfqpoint{0.916251in}{0.892552in}}%
\pgfpathlineto{\pgfqpoint{0.988151in}{0.892552in}}%
\pgfpathlineto{\pgfqpoint{0.988151in}{0.859811in}}%
\pgfpathlineto{\pgfqpoint{1.060052in}{0.859811in}}%
\pgfpathlineto{\pgfqpoint{1.060052in}{0.827388in}}%
\pgfpathlineto{\pgfqpoint{1.131952in}{0.827388in}}%
\pgfpathlineto{\pgfqpoint{1.131952in}{0.803347in}}%
\pgfpathlineto{\pgfqpoint{1.203853in}{0.803347in}}%
\pgfpathlineto{\pgfqpoint{1.203853in}{0.775351in}}%
\pgfpathlineto{\pgfqpoint{1.241470in}{0.775351in}}%
\pgfusepath{stroke}%
\end{pgfscope}%
\begin{pgfscope}%
\pgfpathrectangle{\pgfqpoint{0.520798in}{0.442177in}}{\pgfqpoint{0.719005in}{0.871884in}}%
\pgfusepath{clip}%
\pgfsetbuttcap%
\pgfsetroundjoin%
\definecolor{currentfill}{rgb}{0.007843,0.619608,0.450980}%
\pgfsetfillcolor{currentfill}%
\pgfsetfillopacity{0.500000}%
\pgfsetlinewidth{0.000000pt}%
\definecolor{currentstroke}{rgb}{0.007843,0.619608,0.450980}%
\pgfsetstrokecolor{currentstroke}%
\pgfsetstrokeopacity{0.500000}%
\pgfsetdash{}{0pt}%
\pgfpathmoveto{\pgfqpoint{0.520798in}{1.173204in}}%
\pgfpathlineto{\pgfqpoint{0.520798in}{1.143176in}}%
\pgfpathlineto{\pgfqpoint{0.556748in}{1.143176in}}%
\pgfpathlineto{\pgfqpoint{0.556748in}{1.089244in}}%
\pgfpathlineto{\pgfqpoint{0.628649in}{1.089244in}}%
\pgfpathlineto{\pgfqpoint{0.628649in}{1.037364in}}%
\pgfpathlineto{\pgfqpoint{0.700549in}{1.037364in}}%
\pgfpathlineto{\pgfqpoint{0.700549in}{0.989410in}}%
\pgfpathlineto{\pgfqpoint{0.772450in}{0.989410in}}%
\pgfpathlineto{\pgfqpoint{0.772450in}{0.949253in}}%
\pgfpathlineto{\pgfqpoint{0.844350in}{0.949253in}}%
\pgfpathlineto{\pgfqpoint{0.844350in}{0.914783in}}%
\pgfpathlineto{\pgfqpoint{0.916251in}{0.914783in}}%
\pgfpathlineto{\pgfqpoint{0.916251in}{0.877582in}}%
\pgfpathlineto{\pgfqpoint{0.988151in}{0.877582in}}%
\pgfpathlineto{\pgfqpoint{0.988151in}{0.846099in}}%
\pgfpathlineto{\pgfqpoint{1.060052in}{0.846099in}}%
\pgfpathlineto{\pgfqpoint{1.060052in}{0.818032in}}%
\pgfpathlineto{\pgfqpoint{1.131952in}{0.818032in}}%
\pgfpathlineto{\pgfqpoint{1.131952in}{0.795334in}}%
\pgfpathlineto{\pgfqpoint{1.203853in}{0.795334in}}%
\pgfpathlineto{\pgfqpoint{1.203853in}{0.769596in}}%
\pgfpathlineto{\pgfqpoint{1.275753in}{0.769596in}}%
\pgfpathlineto{\pgfqpoint{1.275753in}{0.746234in}}%
\pgfpathlineto{\pgfqpoint{1.347654in}{0.746234in}}%
\pgfpathlineto{\pgfqpoint{1.347654in}{0.696977in}}%
\pgfpathlineto{\pgfqpoint{1.419554in}{0.696977in}}%
\pgfpathlineto{\pgfqpoint{1.419554in}{0.679970in}}%
\pgfpathlineto{\pgfqpoint{1.491455in}{0.679970in}}%
\pgfpathlineto{\pgfqpoint{1.491455in}{0.665650in}}%
\pgfpathlineto{\pgfqpoint{1.563355in}{0.665650in}}%
\pgfpathlineto{\pgfqpoint{1.563355in}{0.651775in}}%
\pgfpathlineto{\pgfqpoint{1.635256in}{0.651775in}}%
\pgfpathlineto{\pgfqpoint{1.635256in}{0.575272in}}%
\pgfpathlineto{\pgfqpoint{1.707156in}{0.575272in}}%
\pgfpathlineto{\pgfqpoint{1.707156in}{0.564416in}}%
\pgfpathlineto{\pgfqpoint{1.779057in}{0.564416in}}%
\pgfpathlineto{\pgfqpoint{1.779057in}{0.557907in}}%
\pgfpathlineto{\pgfqpoint{1.850957in}{0.557907in}}%
\pgfpathlineto{\pgfqpoint{1.850957in}{0.550981in}}%
\pgfpathlineto{\pgfqpoint{1.922858in}{0.550981in}}%
\pgfpathlineto{\pgfqpoint{1.922858in}{0.545081in}}%
\pgfpathlineto{\pgfqpoint{1.994758in}{0.545081in}}%
\pgfpathlineto{\pgfqpoint{1.994758in}{0.540599in}}%
\pgfpathlineto{\pgfqpoint{2.066659in}{0.540599in}}%
\pgfpathlineto{\pgfqpoint{2.066659in}{0.534831in}}%
\pgfpathlineto{\pgfqpoint{2.138559in}{0.534831in}}%
\pgfpathlineto{\pgfqpoint{2.138559in}{0.530664in}}%
\pgfpathlineto{\pgfqpoint{2.210460in}{0.530664in}}%
\pgfpathlineto{\pgfqpoint{2.210460in}{0.527644in}}%
\pgfpathlineto{\pgfqpoint{2.282360in}{0.527644in}}%
\pgfpathlineto{\pgfqpoint{2.282360in}{0.523570in}}%
\pgfpathlineto{\pgfqpoint{2.354261in}{0.523570in}}%
\pgfpathlineto{\pgfqpoint{2.354261in}{0.520179in}}%
\pgfpathlineto{\pgfqpoint{2.426161in}{0.520179in}}%
\pgfpathlineto{\pgfqpoint{2.426161in}{0.516850in}}%
\pgfpathlineto{\pgfqpoint{2.498062in}{0.516850in}}%
\pgfpathlineto{\pgfqpoint{2.498062in}{0.514107in}}%
\pgfpathlineto{\pgfqpoint{2.569962in}{0.514107in}}%
\pgfpathlineto{\pgfqpoint{2.569962in}{0.510799in}}%
\pgfpathlineto{\pgfqpoint{2.641863in}{0.510799in}}%
\pgfpathlineto{\pgfqpoint{2.641863in}{0.505947in}}%
\pgfpathlineto{\pgfqpoint{2.677813in}{0.505947in}}%
\pgfpathlineto{\pgfqpoint{2.677813in}{0.520913in}}%
\pgfpathlineto{\pgfqpoint{2.677813in}{0.520913in}}%
\pgfpathlineto{\pgfqpoint{2.641863in}{0.520913in}}%
\pgfpathlineto{\pgfqpoint{2.641863in}{0.522704in}}%
\pgfpathlineto{\pgfqpoint{2.569962in}{0.522704in}}%
\pgfpathlineto{\pgfqpoint{2.569962in}{0.526830in}}%
\pgfpathlineto{\pgfqpoint{2.498062in}{0.526830in}}%
\pgfpathlineto{\pgfqpoint{2.498062in}{0.529148in}}%
\pgfpathlineto{\pgfqpoint{2.426161in}{0.529148in}}%
\pgfpathlineto{\pgfqpoint{2.426161in}{0.532937in}}%
\pgfpathlineto{\pgfqpoint{2.354261in}{0.532937in}}%
\pgfpathlineto{\pgfqpoint{2.354261in}{0.537454in}}%
\pgfpathlineto{\pgfqpoint{2.282360in}{0.537454in}}%
\pgfpathlineto{\pgfqpoint{2.282360in}{0.540814in}}%
\pgfpathlineto{\pgfqpoint{2.210460in}{0.540814in}}%
\pgfpathlineto{\pgfqpoint{2.210460in}{0.546018in}}%
\pgfpathlineto{\pgfqpoint{2.138559in}{0.546018in}}%
\pgfpathlineto{\pgfqpoint{2.138559in}{0.551025in}}%
\pgfpathlineto{\pgfqpoint{2.066659in}{0.551025in}}%
\pgfpathlineto{\pgfqpoint{2.066659in}{0.556012in}}%
\pgfpathlineto{\pgfqpoint{1.994758in}{0.556012in}}%
\pgfpathlineto{\pgfqpoint{1.994758in}{0.565924in}}%
\pgfpathlineto{\pgfqpoint{1.922858in}{0.565924in}}%
\pgfpathlineto{\pgfqpoint{1.922858in}{0.571728in}}%
\pgfpathlineto{\pgfqpoint{1.850957in}{0.571728in}}%
\pgfpathlineto{\pgfqpoint{1.850957in}{0.579985in}}%
\pgfpathlineto{\pgfqpoint{1.779057in}{0.579985in}}%
\pgfpathlineto{\pgfqpoint{1.779057in}{0.586921in}}%
\pgfpathlineto{\pgfqpoint{1.707156in}{0.586921in}}%
\pgfpathlineto{\pgfqpoint{1.707156in}{0.596625in}}%
\pgfpathlineto{\pgfqpoint{1.635256in}{0.596625in}}%
\pgfpathlineto{\pgfqpoint{1.635256in}{0.666900in}}%
\pgfpathlineto{\pgfqpoint{1.563355in}{0.666900in}}%
\pgfpathlineto{\pgfqpoint{1.563355in}{0.680229in}}%
\pgfpathlineto{\pgfqpoint{1.491455in}{0.680229in}}%
\pgfpathlineto{\pgfqpoint{1.491455in}{0.694853in}}%
\pgfpathlineto{\pgfqpoint{1.419554in}{0.694853in}}%
\pgfpathlineto{\pgfqpoint{1.419554in}{0.710270in}}%
\pgfpathlineto{\pgfqpoint{1.347654in}{0.710270in}}%
\pgfpathlineto{\pgfqpoint{1.347654in}{0.753540in}}%
\pgfpathlineto{\pgfqpoint{1.275753in}{0.753540in}}%
\pgfpathlineto{\pgfqpoint{1.275753in}{0.781106in}}%
\pgfpathlineto{\pgfqpoint{1.203853in}{0.781106in}}%
\pgfpathlineto{\pgfqpoint{1.203853in}{0.811359in}}%
\pgfpathlineto{\pgfqpoint{1.131952in}{0.811359in}}%
\pgfpathlineto{\pgfqpoint{1.131952in}{0.836743in}}%
\pgfpathlineto{\pgfqpoint{1.060052in}{0.836743in}}%
\pgfpathlineto{\pgfqpoint{1.060052in}{0.873524in}}%
\pgfpathlineto{\pgfqpoint{0.988151in}{0.873524in}}%
\pgfpathlineto{\pgfqpoint{0.988151in}{0.907521in}}%
\pgfpathlineto{\pgfqpoint{0.916251in}{0.907521in}}%
\pgfpathlineto{\pgfqpoint{0.916251in}{0.944026in}}%
\pgfpathlineto{\pgfqpoint{0.844350in}{0.944026in}}%
\pgfpathlineto{\pgfqpoint{0.844350in}{0.984051in}}%
\pgfpathlineto{\pgfqpoint{0.772450in}{0.984051in}}%
\pgfpathlineto{\pgfqpoint{0.772450in}{1.023767in}}%
\pgfpathlineto{\pgfqpoint{0.700549in}{1.023767in}}%
\pgfpathlineto{\pgfqpoint{0.700549in}{1.071502in}}%
\pgfpathlineto{\pgfqpoint{0.628649in}{1.071502in}}%
\pgfpathlineto{\pgfqpoint{0.628649in}{1.125277in}}%
\pgfpathlineto{\pgfqpoint{0.556748in}{1.125277in}}%
\pgfpathlineto{\pgfqpoint{0.556748in}{1.173204in}}%
\pgfpathlineto{\pgfqpoint{0.520798in}{1.173204in}}%
\pgfpathlineto{\pgfqpoint{0.520798in}{1.173204in}}%
\pgfpathclose%
\pgfusepath{fill}%
\end{pgfscope}%
\begin{pgfscope}%
\definecolor{textcolor}{rgb}{0.150000,0.150000,0.150000}%
\pgfsetstrokecolor{textcolor}%
\pgfsetfillcolor{textcolor}%
\pgftext[x=0.880300in,y=1.397395in,,base]{\color{textcolor}\rmfamily\fontsize{9.000000}{10.800000}\selectfont \(\displaystyle c_X^-=1\)}%
\end{pgfscope}%
\begin{pgfscope}%
\pgfsetbuttcap%
\pgfsetmiterjoin%
\definecolor{currentfill}{rgb}{1.000000,1.000000,1.000000}%
\pgfsetfillcolor{currentfill}%
\pgfsetfillopacity{0.800000}%
\pgfsetlinewidth{1.003750pt}%
\definecolor{currentstroke}{rgb}{0.800000,0.800000,0.800000}%
\pgfsetstrokecolor{currentstroke}%
\pgfsetstrokeopacity{0.800000}%
\pgfsetdash{}{0pt}%
\pgfpathmoveto{\pgfqpoint{0.785843in}{0.638103in}}%
\pgfpathlineto{\pgfqpoint{1.191192in}{0.638103in}}%
\pgfpathlineto{\pgfqpoint{1.191192in}{1.265450in}}%
\pgfpathlineto{\pgfqpoint{0.785843in}{1.265450in}}%
\pgfpathlineto{\pgfqpoint{0.785843in}{0.638103in}}%
\pgfpathclose%
\pgfusepath{stroke,fill}%
\end{pgfscope}%
\begin{pgfscope}%
\definecolor{textcolor}{rgb}{0.150000,0.150000,0.150000}%
\pgfsetstrokecolor{textcolor}%
\pgfsetfillcolor{textcolor}%
\pgftext[x=0.912291in,y=1.113275in,left,base]{\color{textcolor}\rmfamily\fontsize{9.000000}{10.800000}\selectfont \(\displaystyle c_X^+\)}%
\end{pgfscope}%
\begin{pgfscope}%
\pgfsetroundcap%
\pgfsetroundjoin%
\pgfsetlinewidth{1.003750pt}%
\definecolor{currentstroke}{rgb}{0.003922,0.450980,0.698039}%
\pgfsetstrokecolor{currentstroke}%
\pgfsetdash{}{0pt}%
\pgfpathmoveto{\pgfqpoint{0.824732in}{1.001052in}}%
\pgfpathlineto{\pgfqpoint{0.873343in}{1.001052in}}%
\pgfpathlineto{\pgfqpoint{0.873343in}{1.001052in}}%
\pgfpathlineto{\pgfqpoint{0.970565in}{1.001052in}}%
\pgfpathlineto{\pgfqpoint{0.970565in}{1.001052in}}%
\pgfpathlineto{\pgfqpoint{1.019176in}{1.001052in}}%
\pgfusepath{stroke}%
\end{pgfscope}%
\begin{pgfscope}%
\definecolor{textcolor}{rgb}{0.150000,0.150000,0.150000}%
\pgfsetstrokecolor{textcolor}%
\pgfsetfillcolor{textcolor}%
\pgftext[x=1.096954in,y=0.967024in,left,base]{\color{textcolor}\rmfamily\fontsize{7.000000}{8.400000}\selectfont 1}%
\end{pgfscope}%
\begin{pgfscope}%
\pgfsetroundcap%
\pgfsetroundjoin%
\pgfsetlinewidth{1.003750pt}%
\definecolor{currentstroke}{rgb}{0.870588,0.560784,0.019608}%
\pgfsetstrokecolor{currentstroke}%
\pgfsetdash{}{0pt}%
\pgfpathmoveto{\pgfqpoint{0.824732in}{0.865486in}}%
\pgfpathlineto{\pgfqpoint{0.873343in}{0.865486in}}%
\pgfpathlineto{\pgfqpoint{0.873343in}{0.865486in}}%
\pgfpathlineto{\pgfqpoint{0.970565in}{0.865486in}}%
\pgfpathlineto{\pgfqpoint{0.970565in}{0.865486in}}%
\pgfpathlineto{\pgfqpoint{1.019176in}{0.865486in}}%
\pgfusepath{stroke}%
\end{pgfscope}%
\begin{pgfscope}%
\definecolor{textcolor}{rgb}{0.150000,0.150000,0.150000}%
\pgfsetstrokecolor{textcolor}%
\pgfsetfillcolor{textcolor}%
\pgftext[x=1.096954in,y=0.831458in,left,base]{\color{textcolor}\rmfamily\fontsize{7.000000}{8.400000}\selectfont 2}%
\end{pgfscope}%
\begin{pgfscope}%
\pgfsetroundcap%
\pgfsetroundjoin%
\pgfsetlinewidth{1.003750pt}%
\definecolor{currentstroke}{rgb}{0.007843,0.619608,0.450980}%
\pgfsetstrokecolor{currentstroke}%
\pgfsetdash{}{0pt}%
\pgfpathmoveto{\pgfqpoint{0.824732in}{0.729919in}}%
\pgfpathlineto{\pgfqpoint{0.873343in}{0.729919in}}%
\pgfpathlineto{\pgfqpoint{0.873343in}{0.729919in}}%
\pgfpathlineto{\pgfqpoint{0.970565in}{0.729919in}}%
\pgfpathlineto{\pgfqpoint{0.970565in}{0.729919in}}%
\pgfpathlineto{\pgfqpoint{1.019176in}{0.729919in}}%
\pgfusepath{stroke}%
\end{pgfscope}%
\begin{pgfscope}%
\definecolor{textcolor}{rgb}{0.150000,0.150000,0.150000}%
\pgfsetstrokecolor{textcolor}%
\pgfsetfillcolor{textcolor}%
\pgftext[x=1.096954in,y=0.695892in,left,base]{\color{textcolor}\rmfamily\fontsize{7.000000}{8.400000}\selectfont 4}%
\end{pgfscope}%
\end{pgfpicture}%
\makeatother%
\endgroup%

%% file: figures/experiments/graphs/sparse_smoothing/nodes_attributes/node_classification-Cora-GCN-hidden=32-p_adj_plus=0.0-p_adj_minus=0.0-p_att_plus=0.001-p_att_minus=0.8-multi_class_cert-A.pgf
\begingroup%
\makeatletter%
\begin{pgfpicture}%
\pgfpathrectangle{\pgfpointorigin}{\pgfqpoint{1.375000in}{1.581250in}}%
\pgfusepath{use as bounding box, clip}%
\begin{pgfscope}%
\pgfsetbuttcap%
\pgfsetmiterjoin%
\definecolor{currentfill}{rgb}{1.000000,1.000000,1.000000}%
\pgfsetfillcolor{currentfill}%
\pgfsetlinewidth{0.000000pt}%
\definecolor{currentstroke}{rgb}{1.000000,1.000000,1.000000}%
\pgfsetstrokecolor{currentstroke}%
\pgfsetdash{}{0pt}%
\pgfpathmoveto{\pgfqpoint{0.000000in}{0.000000in}}%
\pgfpathlineto{\pgfqpoint{1.375000in}{0.000000in}}%
\pgfpathlineto{\pgfqpoint{1.375000in}{1.581250in}}%
\pgfpathlineto{\pgfqpoint{0.000000in}{1.581250in}}%
\pgfpathlineto{\pgfqpoint{0.000000in}{0.000000in}}%
\pgfpathclose%
\pgfusepath{fill}%
\end{pgfscope}%
\begin{pgfscope}%
\pgfsetbuttcap%
\pgfsetmiterjoin%
\definecolor{currentfill}{rgb}{1.000000,1.000000,1.000000}%
\pgfsetfillcolor{currentfill}%
\pgfsetlinewidth{0.000000pt}%
\definecolor{currentstroke}{rgb}{0.000000,0.000000,0.000000}%
\pgfsetstrokecolor{currentstroke}%
\pgfsetstrokeopacity{0.000000}%
\pgfsetdash{}{0pt}%
\pgfpathmoveto{\pgfqpoint{0.520798in}{0.442177in}}%
\pgfpathlineto{\pgfqpoint{1.239803in}{0.442177in}}%
\pgfpathlineto{\pgfqpoint{1.239803in}{1.314061in}}%
\pgfpathlineto{\pgfqpoint{0.520798in}{1.314061in}}%
\pgfpathlineto{\pgfqpoint{0.520798in}{0.442177in}}%
\pgfpathclose%
\pgfusepath{fill}%
\end{pgfscope}%
\begin{pgfscope}%
\pgfpathrectangle{\pgfqpoint{0.520798in}{0.442177in}}{\pgfqpoint{0.719005in}{0.871884in}}%
\pgfusepath{clip}%
\pgfsetroundcap%
\pgfsetroundjoin%
\pgfsetlinewidth{0.501875pt}%
\definecolor{currentstroke}{rgb}{0.800000,0.800000,0.800000}%
\pgfsetstrokecolor{currentstroke}%
\pgfsetdash{}{0pt}%
\pgfpathmoveto{\pgfqpoint{0.520798in}{0.442177in}}%
\pgfpathlineto{\pgfqpoint{0.520798in}{1.314061in}}%
\pgfusepath{stroke}%
\end{pgfscope}%
\begin{pgfscope}%
\definecolor{textcolor}{rgb}{0.150000,0.150000,0.150000}%
\pgfsetstrokecolor{textcolor}%
\pgfsetfillcolor{textcolor}%
\pgftext[x=0.520798in,y=0.351899in,,top]{\color{textcolor}\rmfamily\fontsize{8.000000}{9.600000}\selectfont \(\displaystyle {0}\)}%
\end{pgfscope}%
\begin{pgfscope}%
\pgfpathrectangle{\pgfqpoint{0.520798in}{0.442177in}}{\pgfqpoint{0.719005in}{0.871884in}}%
\pgfusepath{clip}%
\pgfsetroundcap%
\pgfsetroundjoin%
\pgfsetlinewidth{0.501875pt}%
\definecolor{currentstroke}{rgb}{0.800000,0.800000,0.800000}%
\pgfsetstrokecolor{currentstroke}%
\pgfsetdash{}{0pt}%
\pgfpathmoveto{\pgfqpoint{0.880300in}{0.442177in}}%
\pgfpathlineto{\pgfqpoint{0.880300in}{1.314061in}}%
\pgfusepath{stroke}%
\end{pgfscope}%
\begin{pgfscope}%
\definecolor{textcolor}{rgb}{0.150000,0.150000,0.150000}%
\pgfsetstrokecolor{textcolor}%
\pgfsetfillcolor{textcolor}%
\pgftext[x=0.880300in,y=0.351899in,,top]{\color{textcolor}\rmfamily\fontsize{8.000000}{9.600000}\selectfont \(\displaystyle {5}\)}%
\end{pgfscope}%
\begin{pgfscope}%
\pgfpathrectangle{\pgfqpoint{0.520798in}{0.442177in}}{\pgfqpoint{0.719005in}{0.871884in}}%
\pgfusepath{clip}%
\pgfsetroundcap%
\pgfsetroundjoin%
\pgfsetlinewidth{0.501875pt}%
\definecolor{currentstroke}{rgb}{0.800000,0.800000,0.800000}%
\pgfsetstrokecolor{currentstroke}%
\pgfsetdash{}{0pt}%
\pgfpathmoveto{\pgfqpoint{1.239803in}{0.442177in}}%
\pgfpathlineto{\pgfqpoint{1.239803in}{1.314061in}}%
\pgfusepath{stroke}%
\end{pgfscope}%
\begin{pgfscope}%
\definecolor{textcolor}{rgb}{0.150000,0.150000,0.150000}%
\pgfsetstrokecolor{textcolor}%
\pgfsetfillcolor{textcolor}%
\pgftext[x=1.239803in,y=0.351899in,,top]{\color{textcolor}\rmfamily\fontsize{8.000000}{9.600000}\selectfont \(\displaystyle {10}\)}%
\end{pgfscope}%
\begin{pgfscope}%
\definecolor{textcolor}{rgb}{0.150000,0.150000,0.150000}%
\pgfsetstrokecolor{textcolor}%
\pgfsetfillcolor{textcolor}%
\pgftext[x=0.880300in,y=0.198219in,,top]{\color{textcolor}\rmfamily\fontsize{10.000000}{12.000000}\selectfont Edit distance \(\displaystyle \epsilon\)}%
\end{pgfscope}%
\begin{pgfscope}%
\pgfpathrectangle{\pgfqpoint{0.520798in}{0.442177in}}{\pgfqpoint{0.719005in}{0.871884in}}%
\pgfusepath{clip}%
\pgfsetroundcap%
\pgfsetroundjoin%
\pgfsetlinewidth{0.501875pt}%
\definecolor{currentstroke}{rgb}{0.800000,0.800000,0.800000}%
\pgfsetstrokecolor{currentstroke}%
\pgfsetdash{}{0pt}%
\pgfpathmoveto{\pgfqpoint{0.520798in}{0.442177in}}%
\pgfpathlineto{\pgfqpoint{1.239803in}{0.442177in}}%
\pgfusepath{stroke}%
\end{pgfscope}%
\begin{pgfscope}%
\definecolor{textcolor}{rgb}{0.150000,0.150000,0.150000}%
\pgfsetstrokecolor{textcolor}%
\pgfsetfillcolor{textcolor}%
\pgftext[x=0.273151in, y=0.403915in, left, base]{\color{textcolor}\rmfamily\fontsize{8.000000}{9.600000}\selectfont 0\%}%
\end{pgfscope}%
\begin{pgfscope}%
\pgfpathrectangle{\pgfqpoint{0.520798in}{0.442177in}}{\pgfqpoint{0.719005in}{0.871884in}}%
\pgfusepath{clip}%
\pgfsetroundcap%
\pgfsetroundjoin%
\pgfsetlinewidth{0.501875pt}%
\definecolor{currentstroke}{rgb}{0.800000,0.800000,0.800000}%
\pgfsetstrokecolor{currentstroke}%
\pgfsetdash{}{0pt}%
\pgfpathmoveto{\pgfqpoint{0.520798in}{0.660148in}}%
\pgfpathlineto{\pgfqpoint{1.239803in}{0.660148in}}%
\pgfusepath{stroke}%
\end{pgfscope}%
\begin{pgfscope}%
\definecolor{textcolor}{rgb}{0.150000,0.150000,0.150000}%
\pgfsetstrokecolor{textcolor}%
\pgfsetfillcolor{textcolor}%
\pgftext[x=0.214138in, y=0.621886in, left, base]{\color{textcolor}\rmfamily\fontsize{8.000000}{9.600000}\selectfont 25\%}%
\end{pgfscope}%
\begin{pgfscope}%
\pgfpathrectangle{\pgfqpoint{0.520798in}{0.442177in}}{\pgfqpoint{0.719005in}{0.871884in}}%
\pgfusepath{clip}%
\pgfsetroundcap%
\pgfsetroundjoin%
\pgfsetlinewidth{0.501875pt}%
\definecolor{currentstroke}{rgb}{0.800000,0.800000,0.800000}%
\pgfsetstrokecolor{currentstroke}%
\pgfsetdash{}{0pt}%
\pgfpathmoveto{\pgfqpoint{0.520798in}{0.878119in}}%
\pgfpathlineto{\pgfqpoint{1.239803in}{0.878119in}}%
\pgfusepath{stroke}%
\end{pgfscope}%
\begin{pgfscope}%
\definecolor{textcolor}{rgb}{0.150000,0.150000,0.150000}%
\pgfsetstrokecolor{textcolor}%
\pgfsetfillcolor{textcolor}%
\pgftext[x=0.214138in, y=0.839857in, left, base]{\color{textcolor}\rmfamily\fontsize{8.000000}{9.600000}\selectfont 50\%}%
\end{pgfscope}%
\begin{pgfscope}%
\pgfpathrectangle{\pgfqpoint{0.520798in}{0.442177in}}{\pgfqpoint{0.719005in}{0.871884in}}%
\pgfusepath{clip}%
\pgfsetroundcap%
\pgfsetroundjoin%
\pgfsetlinewidth{0.501875pt}%
\definecolor{currentstroke}{rgb}{0.800000,0.800000,0.800000}%
\pgfsetstrokecolor{currentstroke}%
\pgfsetdash{}{0pt}%
\pgfpathmoveto{\pgfqpoint{0.520798in}{1.096090in}}%
\pgfpathlineto{\pgfqpoint{1.239803in}{1.096090in}}%
\pgfusepath{stroke}%
\end{pgfscope}%
\begin{pgfscope}%
\definecolor{textcolor}{rgb}{0.150000,0.150000,0.150000}%
\pgfsetstrokecolor{textcolor}%
\pgfsetfillcolor{textcolor}%
\pgftext[x=0.214138in, y=1.057828in, left, base]{\color{textcolor}\rmfamily\fontsize{8.000000}{9.600000}\selectfont 75\%}%
\end{pgfscope}%
\begin{pgfscope}%
\pgfpathrectangle{\pgfqpoint{0.520798in}{0.442177in}}{\pgfqpoint{0.719005in}{0.871884in}}%
\pgfusepath{clip}%
\pgfsetroundcap%
\pgfsetroundjoin%
\pgfsetlinewidth{0.501875pt}%
\definecolor{currentstroke}{rgb}{0.800000,0.800000,0.800000}%
\pgfsetstrokecolor{currentstroke}%
\pgfsetdash{}{0pt}%
\pgfpathmoveto{\pgfqpoint{0.520798in}{1.314061in}}%
\pgfpathlineto{\pgfqpoint{1.239803in}{1.314061in}}%
\pgfusepath{stroke}%
\end{pgfscope}%
\begin{pgfscope}%
\definecolor{textcolor}{rgb}{0.150000,0.150000,0.150000}%
\pgfsetstrokecolor{textcolor}%
\pgfsetfillcolor{textcolor}%
\pgftext[x=0.155124in, y=1.275799in, left, base]{\color{textcolor}\rmfamily\fontsize{8.000000}{9.600000}\selectfont 100\%}%
\end{pgfscope}%
\begin{pgfscope}%
\definecolor{textcolor}{rgb}{0.150000,0.150000,0.150000}%
\pgfsetstrokecolor{textcolor}%
\pgfsetfillcolor{textcolor}%
\pgftext[x=0.099569in,y=0.878119in,,bottom,rotate=90.000000]{\color{textcolor}\rmfamily\fontsize{10.000000}{12.000000}\selectfont Cert. Acc.}%
\end{pgfscope}%
\begin{pgfscope}%
\pgfsetrectcap%
\pgfsetmiterjoin%
\pgfsetlinewidth{0.752812pt}%
\definecolor{currentstroke}{rgb}{0.700000,0.700000,0.700000}%
\pgfsetstrokecolor{currentstroke}%
\pgfsetdash{}{0pt}%
\pgfpathmoveto{\pgfqpoint{0.520798in}{0.442177in}}%
\pgfpathlineto{\pgfqpoint{0.520798in}{1.314061in}}%
\pgfusepath{stroke}%
\end{pgfscope}%
\begin{pgfscope}%
\pgfsetrectcap%
\pgfsetmiterjoin%
\pgfsetlinewidth{0.752812pt}%
\definecolor{currentstroke}{rgb}{0.700000,0.700000,0.700000}%
\pgfsetstrokecolor{currentstroke}%
\pgfsetdash{}{0pt}%
\pgfpathmoveto{\pgfqpoint{1.239803in}{0.442177in}}%
\pgfpathlineto{\pgfqpoint{1.239803in}{1.314061in}}%
\pgfusepath{stroke}%
\end{pgfscope}%
\begin{pgfscope}%
\pgfsetrectcap%
\pgfsetmiterjoin%
\pgfsetlinewidth{0.752812pt}%
\definecolor{currentstroke}{rgb}{0.700000,0.700000,0.700000}%
\pgfsetstrokecolor{currentstroke}%
\pgfsetdash{}{0pt}%
\pgfpathmoveto{\pgfqpoint{0.520798in}{0.442177in}}%
\pgfpathlineto{\pgfqpoint{1.239803in}{0.442177in}}%
\pgfusepath{stroke}%
\end{pgfscope}%
\begin{pgfscope}%
\pgfsetrectcap%
\pgfsetmiterjoin%
\pgfsetlinewidth{0.752812pt}%
\definecolor{currentstroke}{rgb}{0.700000,0.700000,0.700000}%
\pgfsetstrokecolor{currentstroke}%
\pgfsetdash{}{0pt}%
\pgfpathmoveto{\pgfqpoint{0.520798in}{1.314061in}}%
\pgfpathlineto{\pgfqpoint{1.239803in}{1.314061in}}%
\pgfusepath{stroke}%
\end{pgfscope}%
\begin{pgfscope}%
\pgfpathrectangle{\pgfqpoint{0.520798in}{0.442177in}}{\pgfqpoint{0.719005in}{0.871884in}}%
\pgfusepath{clip}%
\pgfsetroundcap%
\pgfsetroundjoin%
\pgfsetlinewidth{1.003750pt}%
\definecolor{currentstroke}{rgb}{0.003922,0.450980,0.698039}%
\pgfsetstrokecolor{currentstroke}%
\pgfsetdash{}{0pt}%
\pgfpathmoveto{\pgfqpoint{0.520798in}{1.158190in}}%
\pgfpathlineto{\pgfqpoint{0.556748in}{1.158190in}}%
\pgfpathlineto{\pgfqpoint{0.556748in}{1.090495in}}%
\pgfpathlineto{\pgfqpoint{0.628649in}{1.090495in}}%
\pgfpathlineto{\pgfqpoint{0.628649in}{0.978040in}}%
\pgfpathlineto{\pgfqpoint{0.700549in}{0.978040in}}%
\pgfpathlineto{\pgfqpoint{0.700549in}{0.703624in}}%
\pgfpathlineto{\pgfqpoint{0.772450in}{0.703624in}}%
\pgfpathlineto{\pgfqpoint{0.772450in}{0.585949in}}%
\pgfpathlineto{\pgfqpoint{0.844350in}{0.585949in}}%
\pgfpathlineto{\pgfqpoint{0.844350in}{0.569816in}}%
\pgfpathlineto{\pgfqpoint{0.916251in}{0.569816in}}%
\pgfpathlineto{\pgfqpoint{0.916251in}{0.561275in}}%
\pgfpathlineto{\pgfqpoint{0.988151in}{0.561275in}}%
\pgfpathlineto{\pgfqpoint{0.988151in}{0.547673in}}%
\pgfpathlineto{\pgfqpoint{1.060052in}{0.547673in}}%
\pgfpathlineto{\pgfqpoint{1.060052in}{0.507657in}}%
\pgfpathlineto{\pgfqpoint{1.131952in}{0.507657in}}%
\pgfpathlineto{\pgfqpoint{1.131952in}{0.500461in}}%
\pgfpathlineto{\pgfqpoint{1.203853in}{0.500461in}}%
\pgfpathlineto{\pgfqpoint{1.203853in}{0.481244in}}%
\pgfpathlineto{\pgfqpoint{1.241470in}{0.481244in}}%
\pgfusepath{stroke}%
\end{pgfscope}%
\begin{pgfscope}%
\pgfpathrectangle{\pgfqpoint{0.520798in}{0.442177in}}{\pgfqpoint{0.719005in}{0.871884in}}%
\pgfusepath{clip}%
\pgfsetbuttcap%
\pgfsetroundjoin%
\definecolor{currentfill}{rgb}{0.003922,0.450980,0.698039}%
\pgfsetfillcolor{currentfill}%
\pgfsetfillopacity{0.500000}%
\pgfsetlinewidth{0.000000pt}%
\definecolor{currentstroke}{rgb}{0.003922,0.450980,0.698039}%
\pgfsetstrokecolor{currentstroke}%
\pgfsetstrokeopacity{0.500000}%
\pgfsetdash{}{0pt}%
\pgfpathmoveto{\pgfqpoint{0.520798in}{1.173204in}}%
\pgfpathlineto{\pgfqpoint{0.520798in}{1.143176in}}%
\pgfpathlineto{\pgfqpoint{0.556748in}{1.143176in}}%
\pgfpathlineto{\pgfqpoint{0.556748in}{1.071028in}}%
\pgfpathlineto{\pgfqpoint{0.628649in}{1.071028in}}%
\pgfpathlineto{\pgfqpoint{0.628649in}{0.960594in}}%
\pgfpathlineto{\pgfqpoint{0.700549in}{0.960594in}}%
\pgfpathlineto{\pgfqpoint{0.700549in}{0.696977in}}%
\pgfpathlineto{\pgfqpoint{0.772450in}{0.696977in}}%
\pgfpathlineto{\pgfqpoint{0.772450in}{0.575272in}}%
\pgfpathlineto{\pgfqpoint{0.844350in}{0.575272in}}%
\pgfpathlineto{\pgfqpoint{0.844350in}{0.558900in}}%
\pgfpathlineto{\pgfqpoint{0.916251in}{0.558900in}}%
\pgfpathlineto{\pgfqpoint{0.916251in}{0.550809in}}%
\pgfpathlineto{\pgfqpoint{0.988151in}{0.550809in}}%
\pgfpathlineto{\pgfqpoint{0.988151in}{0.538459in}}%
\pgfpathlineto{\pgfqpoint{1.060052in}{0.538459in}}%
\pgfpathlineto{\pgfqpoint{1.060052in}{0.499955in}}%
\pgfpathlineto{\pgfqpoint{1.131952in}{0.499955in}}%
\pgfpathlineto{\pgfqpoint{1.131952in}{0.492637in}}%
\pgfpathlineto{\pgfqpoint{1.203853in}{0.492637in}}%
\pgfpathlineto{\pgfqpoint{1.203853in}{0.476098in}}%
\pgfpathlineto{\pgfqpoint{1.275753in}{0.476098in}}%
\pgfpathlineto{\pgfqpoint{1.275753in}{0.460224in}}%
\pgfpathlineto{\pgfqpoint{1.347654in}{0.460224in}}%
\pgfpathlineto{\pgfqpoint{1.347654in}{0.455551in}}%
\pgfpathlineto{\pgfqpoint{1.419554in}{0.455551in}}%
\pgfpathlineto{\pgfqpoint{1.419554in}{0.442177in}}%
\pgfpathlineto{\pgfqpoint{2.066659in}{0.442177in}}%
\pgfpathlineto{\pgfqpoint{2.066659in}{0.442177in}}%
\pgfpathlineto{\pgfqpoint{2.677813in}{0.442177in}}%
\pgfpathlineto{\pgfqpoint{2.677813in}{0.442177in}}%
\pgfpathlineto{\pgfqpoint{2.677813in}{0.442177in}}%
\pgfpathlineto{\pgfqpoint{2.066659in}{0.442177in}}%
\pgfpathlineto{\pgfqpoint{2.066659in}{0.442177in}}%
\pgfpathlineto{\pgfqpoint{1.419554in}{0.442177in}}%
\pgfpathlineto{\pgfqpoint{1.419554in}{0.462492in}}%
\pgfpathlineto{\pgfqpoint{1.347654in}{0.462492in}}%
\pgfpathlineto{\pgfqpoint{1.347654in}{0.466676in}}%
\pgfpathlineto{\pgfqpoint{1.275753in}{0.466676in}}%
\pgfpathlineto{\pgfqpoint{1.275753in}{0.486389in}}%
\pgfpathlineto{\pgfqpoint{1.203853in}{0.486389in}}%
\pgfpathlineto{\pgfqpoint{1.203853in}{0.508284in}}%
\pgfpathlineto{\pgfqpoint{1.131952in}{0.508284in}}%
\pgfpathlineto{\pgfqpoint{1.131952in}{0.515359in}}%
\pgfpathlineto{\pgfqpoint{1.060052in}{0.515359in}}%
\pgfpathlineto{\pgfqpoint{1.060052in}{0.556887in}}%
\pgfpathlineto{\pgfqpoint{0.988151in}{0.556887in}}%
\pgfpathlineto{\pgfqpoint{0.988151in}{0.571741in}}%
\pgfpathlineto{\pgfqpoint{0.916251in}{0.571741in}}%
\pgfpathlineto{\pgfqpoint{0.916251in}{0.580732in}}%
\pgfpathlineto{\pgfqpoint{0.844350in}{0.580732in}}%
\pgfpathlineto{\pgfqpoint{0.844350in}{0.596625in}}%
\pgfpathlineto{\pgfqpoint{0.772450in}{0.596625in}}%
\pgfpathlineto{\pgfqpoint{0.772450in}{0.710270in}}%
\pgfpathlineto{\pgfqpoint{0.700549in}{0.710270in}}%
\pgfpathlineto{\pgfqpoint{0.700549in}{0.995485in}}%
\pgfpathlineto{\pgfqpoint{0.628649in}{0.995485in}}%
\pgfpathlineto{\pgfqpoint{0.628649in}{1.109962in}}%
\pgfpathlineto{\pgfqpoint{0.556748in}{1.109962in}}%
\pgfpathlineto{\pgfqpoint{0.556748in}{1.173204in}}%
\pgfpathlineto{\pgfqpoint{0.520798in}{1.173204in}}%
\pgfpathlineto{\pgfqpoint{0.520798in}{1.173204in}}%
\pgfpathclose%
\pgfusepath{fill}%
\end{pgfscope}%
\begin{pgfscope}%
\pgfpathrectangle{\pgfqpoint{0.520798in}{0.442177in}}{\pgfqpoint{0.719005in}{0.871884in}}%
\pgfusepath{clip}%
\pgfsetroundcap%
\pgfsetroundjoin%
\pgfsetlinewidth{1.003750pt}%
\definecolor{currentstroke}{rgb}{0.870588,0.560784,0.019608}%
\pgfsetstrokecolor{currentstroke}%
\pgfsetdash{}{0pt}%
\pgfpathmoveto{\pgfqpoint{0.520798in}{1.158190in}}%
\pgfpathlineto{\pgfqpoint{0.556748in}{1.158190in}}%
\pgfpathlineto{\pgfqpoint{0.556748in}{1.090495in}}%
\pgfpathlineto{\pgfqpoint{0.628649in}{1.090495in}}%
\pgfpathlineto{\pgfqpoint{0.628649in}{0.978040in}}%
\pgfpathlineto{\pgfqpoint{0.700549in}{0.978040in}}%
\pgfpathlineto{\pgfqpoint{0.700549in}{0.703624in}}%
\pgfpathlineto{\pgfqpoint{0.772450in}{0.703624in}}%
\pgfpathlineto{\pgfqpoint{0.772450in}{0.585949in}}%
\pgfpathlineto{\pgfqpoint{0.844350in}{0.585949in}}%
\pgfpathlineto{\pgfqpoint{0.844350in}{0.573691in}}%
\pgfpathlineto{\pgfqpoint{0.916251in}{0.573691in}}%
\pgfpathlineto{\pgfqpoint{0.916251in}{0.563727in}}%
\pgfpathlineto{\pgfqpoint{0.988151in}{0.563727in}}%
\pgfpathlineto{\pgfqpoint{0.988151in}{0.547673in}}%
\pgfpathlineto{\pgfqpoint{1.060052in}{0.547673in}}%
\pgfpathlineto{\pgfqpoint{1.060052in}{0.507657in}}%
\pgfpathlineto{\pgfqpoint{1.131952in}{0.507657in}}%
\pgfpathlineto{\pgfqpoint{1.131952in}{0.500619in}}%
\pgfpathlineto{\pgfqpoint{1.203853in}{0.500619in}}%
\pgfpathlineto{\pgfqpoint{1.203853in}{0.481323in}}%
\pgfpathlineto{\pgfqpoint{1.241470in}{0.481323in}}%
\pgfusepath{stroke}%
\end{pgfscope}%
\begin{pgfscope}%
\pgfpathrectangle{\pgfqpoint{0.520798in}{0.442177in}}{\pgfqpoint{0.719005in}{0.871884in}}%
\pgfusepath{clip}%
\pgfsetbuttcap%
\pgfsetroundjoin%
\definecolor{currentfill}{rgb}{0.870588,0.560784,0.019608}%
\pgfsetfillcolor{currentfill}%
\pgfsetfillopacity{0.500000}%
\pgfsetlinewidth{0.000000pt}%
\definecolor{currentstroke}{rgb}{0.870588,0.560784,0.019608}%
\pgfsetstrokecolor{currentstroke}%
\pgfsetstrokeopacity{0.500000}%
\pgfsetdash{}{0pt}%
\pgfpathmoveto{\pgfqpoint{0.520798in}{1.173204in}}%
\pgfpathlineto{\pgfqpoint{0.520798in}{1.143176in}}%
\pgfpathlineto{\pgfqpoint{0.556748in}{1.143176in}}%
\pgfpathlineto{\pgfqpoint{0.556748in}{1.071028in}}%
\pgfpathlineto{\pgfqpoint{0.628649in}{1.071028in}}%
\pgfpathlineto{\pgfqpoint{0.628649in}{0.960594in}}%
\pgfpathlineto{\pgfqpoint{0.700549in}{0.960594in}}%
\pgfpathlineto{\pgfqpoint{0.700549in}{0.696977in}}%
\pgfpathlineto{\pgfqpoint{0.772450in}{0.696977in}}%
\pgfpathlineto{\pgfqpoint{0.772450in}{0.575272in}}%
\pgfpathlineto{\pgfqpoint{0.844350in}{0.575272in}}%
\pgfpathlineto{\pgfqpoint{0.844350in}{0.563222in}}%
\pgfpathlineto{\pgfqpoint{0.916251in}{0.563222in}}%
\pgfpathlineto{\pgfqpoint{0.916251in}{0.553438in}}%
\pgfpathlineto{\pgfqpoint{0.988151in}{0.553438in}}%
\pgfpathlineto{\pgfqpoint{0.988151in}{0.538459in}}%
\pgfpathlineto{\pgfqpoint{1.060052in}{0.538459in}}%
\pgfpathlineto{\pgfqpoint{1.060052in}{0.499955in}}%
\pgfpathlineto{\pgfqpoint{1.131952in}{0.499955in}}%
\pgfpathlineto{\pgfqpoint{1.131952in}{0.492873in}}%
\pgfpathlineto{\pgfqpoint{1.203853in}{0.492873in}}%
\pgfpathlineto{\pgfqpoint{1.203853in}{0.476149in}}%
\pgfpathlineto{\pgfqpoint{1.275753in}{0.476149in}}%
\pgfpathlineto{\pgfqpoint{1.275753in}{0.460579in}}%
\pgfpathlineto{\pgfqpoint{1.347654in}{0.460579in}}%
\pgfpathlineto{\pgfqpoint{1.347654in}{0.455551in}}%
\pgfpathlineto{\pgfqpoint{1.419554in}{0.455551in}}%
\pgfpathlineto{\pgfqpoint{1.419554in}{0.442177in}}%
\pgfpathlineto{\pgfqpoint{2.066659in}{0.442177in}}%
\pgfpathlineto{\pgfqpoint{2.066659in}{0.442177in}}%
\pgfpathlineto{\pgfqpoint{2.677813in}{0.442177in}}%
\pgfpathlineto{\pgfqpoint{2.677813in}{0.442177in}}%
\pgfpathlineto{\pgfqpoint{2.677813in}{0.442177in}}%
\pgfpathlineto{\pgfqpoint{2.066659in}{0.442177in}}%
\pgfpathlineto{\pgfqpoint{2.066659in}{0.442177in}}%
\pgfpathlineto{\pgfqpoint{1.419554in}{0.442177in}}%
\pgfpathlineto{\pgfqpoint{1.419554in}{0.462492in}}%
\pgfpathlineto{\pgfqpoint{1.347654in}{0.462492in}}%
\pgfpathlineto{\pgfqpoint{1.347654in}{0.467903in}}%
\pgfpathlineto{\pgfqpoint{1.275753in}{0.467903in}}%
\pgfpathlineto{\pgfqpoint{1.275753in}{0.486496in}}%
\pgfpathlineto{\pgfqpoint{1.203853in}{0.486496in}}%
\pgfpathlineto{\pgfqpoint{1.203853in}{0.508365in}}%
\pgfpathlineto{\pgfqpoint{1.131952in}{0.508365in}}%
\pgfpathlineto{\pgfqpoint{1.131952in}{0.515359in}}%
\pgfpathlineto{\pgfqpoint{1.060052in}{0.515359in}}%
\pgfpathlineto{\pgfqpoint{1.060052in}{0.556887in}}%
\pgfpathlineto{\pgfqpoint{0.988151in}{0.556887in}}%
\pgfpathlineto{\pgfqpoint{0.988151in}{0.574015in}}%
\pgfpathlineto{\pgfqpoint{0.916251in}{0.574015in}}%
\pgfpathlineto{\pgfqpoint{0.916251in}{0.584160in}}%
\pgfpathlineto{\pgfqpoint{0.844350in}{0.584160in}}%
\pgfpathlineto{\pgfqpoint{0.844350in}{0.596625in}}%
\pgfpathlineto{\pgfqpoint{0.772450in}{0.596625in}}%
\pgfpathlineto{\pgfqpoint{0.772450in}{0.710270in}}%
\pgfpathlineto{\pgfqpoint{0.700549in}{0.710270in}}%
\pgfpathlineto{\pgfqpoint{0.700549in}{0.995485in}}%
\pgfpathlineto{\pgfqpoint{0.628649in}{0.995485in}}%
\pgfpathlineto{\pgfqpoint{0.628649in}{1.109962in}}%
\pgfpathlineto{\pgfqpoint{0.556748in}{1.109962in}}%
\pgfpathlineto{\pgfqpoint{0.556748in}{1.173204in}}%
\pgfpathlineto{\pgfqpoint{0.520798in}{1.173204in}}%
\pgfpathlineto{\pgfqpoint{0.520798in}{1.173204in}}%
\pgfpathclose%
\pgfusepath{fill}%
\end{pgfscope}%
\begin{pgfscope}%
\pgfpathrectangle{\pgfqpoint{0.520798in}{0.442177in}}{\pgfqpoint{0.719005in}{0.871884in}}%
\pgfusepath{clip}%
\pgfsetroundcap%
\pgfsetroundjoin%
\pgfsetlinewidth{1.003750pt}%
\definecolor{currentstroke}{rgb}{0.007843,0.619608,0.450980}%
\pgfsetstrokecolor{currentstroke}%
\pgfsetdash{}{0pt}%
\pgfpathmoveto{\pgfqpoint{0.520798in}{1.158190in}}%
\pgfpathlineto{\pgfqpoint{0.556748in}{1.158190in}}%
\pgfpathlineto{\pgfqpoint{0.556748in}{1.090495in}}%
\pgfpathlineto{\pgfqpoint{0.628649in}{1.090495in}}%
\pgfpathlineto{\pgfqpoint{0.628649in}{0.978040in}}%
\pgfpathlineto{\pgfqpoint{0.700549in}{0.978040in}}%
\pgfpathlineto{\pgfqpoint{0.700549in}{0.703624in}}%
\pgfpathlineto{\pgfqpoint{0.772450in}{0.703624in}}%
\pgfpathlineto{\pgfqpoint{0.772450in}{0.585949in}}%
\pgfpathlineto{\pgfqpoint{0.844350in}{0.585949in}}%
\pgfpathlineto{\pgfqpoint{0.844350in}{0.573691in}}%
\pgfpathlineto{\pgfqpoint{0.916251in}{0.573691in}}%
\pgfpathlineto{\pgfqpoint{0.916251in}{0.563727in}}%
\pgfpathlineto{\pgfqpoint{0.988151in}{0.563727in}}%
\pgfpathlineto{\pgfqpoint{0.988151in}{0.547673in}}%
\pgfpathlineto{\pgfqpoint{1.060052in}{0.547673in}}%
\pgfpathlineto{\pgfqpoint{1.060052in}{0.507657in}}%
\pgfpathlineto{\pgfqpoint{1.131952in}{0.507657in}}%
\pgfpathlineto{\pgfqpoint{1.131952in}{0.500619in}}%
\pgfpathlineto{\pgfqpoint{1.203853in}{0.500619in}}%
\pgfpathlineto{\pgfqpoint{1.203853in}{0.481323in}}%
\pgfpathlineto{\pgfqpoint{1.241470in}{0.481323in}}%
\pgfusepath{stroke}%
\end{pgfscope}%
\begin{pgfscope}%
\pgfpathrectangle{\pgfqpoint{0.520798in}{0.442177in}}{\pgfqpoint{0.719005in}{0.871884in}}%
\pgfusepath{clip}%
\pgfsetbuttcap%
\pgfsetroundjoin%
\definecolor{currentfill}{rgb}{0.007843,0.619608,0.450980}%
\pgfsetfillcolor{currentfill}%
\pgfsetfillopacity{0.500000}%
\pgfsetlinewidth{0.000000pt}%
\definecolor{currentstroke}{rgb}{0.007843,0.619608,0.450980}%
\pgfsetstrokecolor{currentstroke}%
\pgfsetstrokeopacity{0.500000}%
\pgfsetdash{}{0pt}%
\pgfpathmoveto{\pgfqpoint{0.520798in}{1.173204in}}%
\pgfpathlineto{\pgfqpoint{0.520798in}{1.143176in}}%
\pgfpathlineto{\pgfqpoint{0.556748in}{1.143176in}}%
\pgfpathlineto{\pgfqpoint{0.556748in}{1.071028in}}%
\pgfpathlineto{\pgfqpoint{0.628649in}{1.071028in}}%
\pgfpathlineto{\pgfqpoint{0.628649in}{0.960594in}}%
\pgfpathlineto{\pgfqpoint{0.700549in}{0.960594in}}%
\pgfpathlineto{\pgfqpoint{0.700549in}{0.696977in}}%
\pgfpathlineto{\pgfqpoint{0.772450in}{0.696977in}}%
\pgfpathlineto{\pgfqpoint{0.772450in}{0.575272in}}%
\pgfpathlineto{\pgfqpoint{0.844350in}{0.575272in}}%
\pgfpathlineto{\pgfqpoint{0.844350in}{0.563222in}}%
\pgfpathlineto{\pgfqpoint{0.916251in}{0.563222in}}%
\pgfpathlineto{\pgfqpoint{0.916251in}{0.553438in}}%
\pgfpathlineto{\pgfqpoint{0.988151in}{0.553438in}}%
\pgfpathlineto{\pgfqpoint{0.988151in}{0.538459in}}%
\pgfpathlineto{\pgfqpoint{1.060052in}{0.538459in}}%
\pgfpathlineto{\pgfqpoint{1.060052in}{0.499955in}}%
\pgfpathlineto{\pgfqpoint{1.131952in}{0.499955in}}%
\pgfpathlineto{\pgfqpoint{1.131952in}{0.492873in}}%
\pgfpathlineto{\pgfqpoint{1.203853in}{0.492873in}}%
\pgfpathlineto{\pgfqpoint{1.203853in}{0.476149in}}%
\pgfpathlineto{\pgfqpoint{1.275753in}{0.476149in}}%
\pgfpathlineto{\pgfqpoint{1.275753in}{0.460579in}}%
\pgfpathlineto{\pgfqpoint{1.347654in}{0.460579in}}%
\pgfpathlineto{\pgfqpoint{1.347654in}{0.455551in}}%
\pgfpathlineto{\pgfqpoint{1.419554in}{0.455551in}}%
\pgfpathlineto{\pgfqpoint{1.419554in}{0.442177in}}%
\pgfpathlineto{\pgfqpoint{2.066659in}{0.442177in}}%
\pgfpathlineto{\pgfqpoint{2.066659in}{0.442177in}}%
\pgfpathlineto{\pgfqpoint{2.677813in}{0.442177in}}%
\pgfpathlineto{\pgfqpoint{2.677813in}{0.442177in}}%
\pgfpathlineto{\pgfqpoint{2.677813in}{0.442177in}}%
\pgfpathlineto{\pgfqpoint{2.066659in}{0.442177in}}%
\pgfpathlineto{\pgfqpoint{2.066659in}{0.442177in}}%
\pgfpathlineto{\pgfqpoint{1.419554in}{0.442177in}}%
\pgfpathlineto{\pgfqpoint{1.419554in}{0.462492in}}%
\pgfpathlineto{\pgfqpoint{1.347654in}{0.462492in}}%
\pgfpathlineto{\pgfqpoint{1.347654in}{0.467903in}}%
\pgfpathlineto{\pgfqpoint{1.275753in}{0.467903in}}%
\pgfpathlineto{\pgfqpoint{1.275753in}{0.486496in}}%
\pgfpathlineto{\pgfqpoint{1.203853in}{0.486496in}}%
\pgfpathlineto{\pgfqpoint{1.203853in}{0.508365in}}%
\pgfpathlineto{\pgfqpoint{1.131952in}{0.508365in}}%
\pgfpathlineto{\pgfqpoint{1.131952in}{0.515359in}}%
\pgfpathlineto{\pgfqpoint{1.060052in}{0.515359in}}%
\pgfpathlineto{\pgfqpoint{1.060052in}{0.556887in}}%
\pgfpathlineto{\pgfqpoint{0.988151in}{0.556887in}}%
\pgfpathlineto{\pgfqpoint{0.988151in}{0.574015in}}%
\pgfpathlineto{\pgfqpoint{0.916251in}{0.574015in}}%
\pgfpathlineto{\pgfqpoint{0.916251in}{0.584160in}}%
\pgfpathlineto{\pgfqpoint{0.844350in}{0.584160in}}%
\pgfpathlineto{\pgfqpoint{0.844350in}{0.596625in}}%
\pgfpathlineto{\pgfqpoint{0.772450in}{0.596625in}}%
\pgfpathlineto{\pgfqpoint{0.772450in}{0.710270in}}%
\pgfpathlineto{\pgfqpoint{0.700549in}{0.710270in}}%
\pgfpathlineto{\pgfqpoint{0.700549in}{0.995485in}}%
\pgfpathlineto{\pgfqpoint{0.628649in}{0.995485in}}%
\pgfpathlineto{\pgfqpoint{0.628649in}{1.109962in}}%
\pgfpathlineto{\pgfqpoint{0.556748in}{1.109962in}}%
\pgfpathlineto{\pgfqpoint{0.556748in}{1.173204in}}%
\pgfpathlineto{\pgfqpoint{0.520798in}{1.173204in}}%
\pgfpathlineto{\pgfqpoint{0.520798in}{1.173204in}}%
\pgfpathclose%
\pgfusepath{fill}%
\end{pgfscope}%
\begin{pgfscope}%
\definecolor{textcolor}{rgb}{0.150000,0.150000,0.150000}%
\pgfsetstrokecolor{textcolor}%
\pgfsetfillcolor{textcolor}%
\pgftext[x=0.880300in,y=1.397395in,,base]{\color{textcolor}\rmfamily\fontsize{9.000000}{10.800000}\selectfont \(\displaystyle c_X^+=1\)}%
\end{pgfscope}%
\begin{pgfscope}%
\pgfsetbuttcap%
\pgfsetmiterjoin%
\definecolor{currentfill}{rgb}{1.000000,1.000000,1.000000}%
\pgfsetfillcolor{currentfill}%
\pgfsetfillopacity{0.800000}%
\pgfsetlinewidth{1.003750pt}%
\definecolor{currentstroke}{rgb}{0.800000,0.800000,0.800000}%
\pgfsetstrokecolor{currentstroke}%
\pgfsetstrokeopacity{0.800000}%
\pgfsetdash{}{0pt}%
\pgfpathmoveto{\pgfqpoint{0.785843in}{0.638103in}}%
\pgfpathlineto{\pgfqpoint{1.191192in}{0.638103in}}%
\pgfpathlineto{\pgfqpoint{1.191192in}{1.265450in}}%
\pgfpathlineto{\pgfqpoint{0.785843in}{1.265450in}}%
\pgfpathlineto{\pgfqpoint{0.785843in}{0.638103in}}%
\pgfpathclose%
\pgfusepath{stroke,fill}%
\end{pgfscope}%
\begin{pgfscope}%
\definecolor{textcolor}{rgb}{0.150000,0.150000,0.150000}%
\pgfsetstrokecolor{textcolor}%
\pgfsetfillcolor{textcolor}%
\pgftext[x=0.912291in,y=1.113275in,left,base]{\color{textcolor}\rmfamily\fontsize{9.000000}{10.800000}\selectfont \(\displaystyle c_X^-\)}%
\end{pgfscope}%
\begin{pgfscope}%
\pgfsetroundcap%
\pgfsetroundjoin%
\pgfsetlinewidth{1.003750pt}%
\definecolor{currentstroke}{rgb}{0.003922,0.450980,0.698039}%
\pgfsetstrokecolor{currentstroke}%
\pgfsetdash{}{0pt}%
\pgfpathmoveto{\pgfqpoint{0.824732in}{1.001052in}}%
\pgfpathlineto{\pgfqpoint{0.873343in}{1.001052in}}%
\pgfpathlineto{\pgfqpoint{0.873343in}{1.001052in}}%
\pgfpathlineto{\pgfqpoint{0.970565in}{1.001052in}}%
\pgfpathlineto{\pgfqpoint{0.970565in}{1.001052in}}%
\pgfpathlineto{\pgfqpoint{1.019176in}{1.001052in}}%
\pgfusepath{stroke}%
\end{pgfscope}%
\begin{pgfscope}%
\definecolor{textcolor}{rgb}{0.150000,0.150000,0.150000}%
\pgfsetstrokecolor{textcolor}%
\pgfsetfillcolor{textcolor}%
\pgftext[x=1.096954in,y=0.967024in,left,base]{\color{textcolor}\rmfamily\fontsize{7.000000}{8.400000}\selectfont 1}%
\end{pgfscope}%
\begin{pgfscope}%
\pgfsetroundcap%
\pgfsetroundjoin%
\pgfsetlinewidth{1.003750pt}%
\definecolor{currentstroke}{rgb}{0.870588,0.560784,0.019608}%
\pgfsetstrokecolor{currentstroke}%
\pgfsetdash{}{0pt}%
\pgfpathmoveto{\pgfqpoint{0.824732in}{0.865486in}}%
\pgfpathlineto{\pgfqpoint{0.873343in}{0.865486in}}%
\pgfpathlineto{\pgfqpoint{0.873343in}{0.865486in}}%
\pgfpathlineto{\pgfqpoint{0.970565in}{0.865486in}}%
\pgfpathlineto{\pgfqpoint{0.970565in}{0.865486in}}%
\pgfpathlineto{\pgfqpoint{1.019176in}{0.865486in}}%
\pgfusepath{stroke}%
\end{pgfscope}%
\begin{pgfscope}%
\definecolor{textcolor}{rgb}{0.150000,0.150000,0.150000}%
\pgfsetstrokecolor{textcolor}%
\pgfsetfillcolor{textcolor}%
\pgftext[x=1.096954in,y=0.831458in,left,base]{\color{textcolor}\rmfamily\fontsize{7.000000}{8.400000}\selectfont 2}%
\end{pgfscope}%
\begin{pgfscope}%
\pgfsetroundcap%
\pgfsetroundjoin%
\pgfsetlinewidth{1.003750pt}%
\definecolor{currentstroke}{rgb}{0.007843,0.619608,0.450980}%
\pgfsetstrokecolor{currentstroke}%
\pgfsetdash{}{0pt}%
\pgfpathmoveto{\pgfqpoint{0.824732in}{0.729919in}}%
\pgfpathlineto{\pgfqpoint{0.873343in}{0.729919in}}%
\pgfpathlineto{\pgfqpoint{0.873343in}{0.729919in}}%
\pgfpathlineto{\pgfqpoint{0.970565in}{0.729919in}}%
\pgfpathlineto{\pgfqpoint{0.970565in}{0.729919in}}%
\pgfpathlineto{\pgfqpoint{1.019176in}{0.729919in}}%
\pgfusepath{stroke}%
\end{pgfscope}%
\begin{pgfscope}%
\definecolor{textcolor}{rgb}{0.150000,0.150000,0.150000}%
\pgfsetstrokecolor{textcolor}%
\pgfsetfillcolor{textcolor}%
\pgftext[x=1.096954in,y=0.695892in,left,base]{\color{textcolor}\rmfamily\fontsize{7.000000}{8.400000}\selectfont 4}%
\end{pgfscope}%
\end{pgfpicture}%
\makeatother%
\endgroup%

%% file: figures/experiments/graphs/sparse_smoothing/nodes_attributes/node_classification-Citeseer-GCN-hidden=32-p_adj_plus=0.0-p_adj_minus=0.0-p_att_plus=0.001-p_att_minus=0.8-multi_class_cert-B.pgf
\begingroup%
\makeatletter%
\begin{pgfpicture}%
\pgfpathrectangle{\pgfpointorigin}{\pgfqpoint{1.375000in}{1.581250in}}%
\pgfusepath{use as bounding box, clip}%
\begin{pgfscope}%
\pgfsetbuttcap%
\pgfsetmiterjoin%
\definecolor{currentfill}{rgb}{1.000000,1.000000,1.000000}%
\pgfsetfillcolor{currentfill}%
\pgfsetlinewidth{0.000000pt}%
\definecolor{currentstroke}{rgb}{1.000000,1.000000,1.000000}%
\pgfsetstrokecolor{currentstroke}%
\pgfsetdash{}{0pt}%
\pgfpathmoveto{\pgfqpoint{0.000000in}{0.000000in}}%
\pgfpathlineto{\pgfqpoint{1.375000in}{0.000000in}}%
\pgfpathlineto{\pgfqpoint{1.375000in}{1.581250in}}%
\pgfpathlineto{\pgfqpoint{0.000000in}{1.581250in}}%
\pgfpathlineto{\pgfqpoint{0.000000in}{0.000000in}}%
\pgfpathclose%
\pgfusepath{fill}%
\end{pgfscope}%
\begin{pgfscope}%
\pgfsetbuttcap%
\pgfsetmiterjoin%
\definecolor{currentfill}{rgb}{1.000000,1.000000,1.000000}%
\pgfsetfillcolor{currentfill}%
\pgfsetlinewidth{0.000000pt}%
\definecolor{currentstroke}{rgb}{0.000000,0.000000,0.000000}%
\pgfsetstrokecolor{currentstroke}%
\pgfsetstrokeopacity{0.000000}%
\pgfsetdash{}{0pt}%
\pgfpathmoveto{\pgfqpoint{0.520798in}{0.442177in}}%
\pgfpathlineto{\pgfqpoint{1.239803in}{0.442177in}}%
\pgfpathlineto{\pgfqpoint{1.239803in}{1.314061in}}%
\pgfpathlineto{\pgfqpoint{0.520798in}{1.314061in}}%
\pgfpathlineto{\pgfqpoint{0.520798in}{0.442177in}}%
\pgfpathclose%
\pgfusepath{fill}%
\end{pgfscope}%
\begin{pgfscope}%
\pgfpathrectangle{\pgfqpoint{0.520798in}{0.442177in}}{\pgfqpoint{0.719005in}{0.871884in}}%
\pgfusepath{clip}%
\pgfsetroundcap%
\pgfsetroundjoin%
\pgfsetlinewidth{0.501875pt}%
\definecolor{currentstroke}{rgb}{0.800000,0.800000,0.800000}%
\pgfsetstrokecolor{currentstroke}%
\pgfsetdash{}{0pt}%
\pgfpathmoveto{\pgfqpoint{0.520798in}{0.442177in}}%
\pgfpathlineto{\pgfqpoint{0.520798in}{1.314061in}}%
\pgfusepath{stroke}%
\end{pgfscope}%
\begin{pgfscope}%
\definecolor{textcolor}{rgb}{0.150000,0.150000,0.150000}%
\pgfsetstrokecolor{textcolor}%
\pgfsetfillcolor{textcolor}%
\pgftext[x=0.520798in,y=0.351899in,,top]{\color{textcolor}\rmfamily\fontsize{8.000000}{9.600000}\selectfont \(\displaystyle {0}\)}%
\end{pgfscope}%
\begin{pgfscope}%
\pgfpathrectangle{\pgfqpoint{0.520798in}{0.442177in}}{\pgfqpoint{0.719005in}{0.871884in}}%
\pgfusepath{clip}%
\pgfsetroundcap%
\pgfsetroundjoin%
\pgfsetlinewidth{0.501875pt}%
\definecolor{currentstroke}{rgb}{0.800000,0.800000,0.800000}%
\pgfsetstrokecolor{currentstroke}%
\pgfsetdash{}{0pt}%
\pgfpathmoveto{\pgfqpoint{0.880300in}{0.442177in}}%
\pgfpathlineto{\pgfqpoint{0.880300in}{1.314061in}}%
\pgfusepath{stroke}%
\end{pgfscope}%
\begin{pgfscope}%
\definecolor{textcolor}{rgb}{0.150000,0.150000,0.150000}%
\pgfsetstrokecolor{textcolor}%
\pgfsetfillcolor{textcolor}%
\pgftext[x=0.880300in,y=0.351899in,,top]{\color{textcolor}\rmfamily\fontsize{8.000000}{9.600000}\selectfont \(\displaystyle {5}\)}%
\end{pgfscope}%
\begin{pgfscope}%
\pgfpathrectangle{\pgfqpoint{0.520798in}{0.442177in}}{\pgfqpoint{0.719005in}{0.871884in}}%
\pgfusepath{clip}%
\pgfsetroundcap%
\pgfsetroundjoin%
\pgfsetlinewidth{0.501875pt}%
\definecolor{currentstroke}{rgb}{0.800000,0.800000,0.800000}%
\pgfsetstrokecolor{currentstroke}%
\pgfsetdash{}{0pt}%
\pgfpathmoveto{\pgfqpoint{1.239803in}{0.442177in}}%
\pgfpathlineto{\pgfqpoint{1.239803in}{1.314061in}}%
\pgfusepath{stroke}%
\end{pgfscope}%
\begin{pgfscope}%
\definecolor{textcolor}{rgb}{0.150000,0.150000,0.150000}%
\pgfsetstrokecolor{textcolor}%
\pgfsetfillcolor{textcolor}%
\pgftext[x=1.239803in,y=0.351899in,,top]{\color{textcolor}\rmfamily\fontsize{8.000000}{9.600000}\selectfont \(\displaystyle {10}\)}%
\end{pgfscope}%
\begin{pgfscope}%
\definecolor{textcolor}{rgb}{0.150000,0.150000,0.150000}%
\pgfsetstrokecolor{textcolor}%
\pgfsetfillcolor{textcolor}%
\pgftext[x=0.880300in,y=0.198219in,,top]{\color{textcolor}\rmfamily\fontsize{10.000000}{12.000000}\selectfont Edit distance \(\displaystyle \epsilon\)}%
\end{pgfscope}%
\begin{pgfscope}%
\pgfpathrectangle{\pgfqpoint{0.520798in}{0.442177in}}{\pgfqpoint{0.719005in}{0.871884in}}%
\pgfusepath{clip}%
\pgfsetroundcap%
\pgfsetroundjoin%
\pgfsetlinewidth{0.501875pt}%
\definecolor{currentstroke}{rgb}{0.800000,0.800000,0.800000}%
\pgfsetstrokecolor{currentstroke}%
\pgfsetdash{}{0pt}%
\pgfpathmoveto{\pgfqpoint{0.520798in}{0.442177in}}%
\pgfpathlineto{\pgfqpoint{1.239803in}{0.442177in}}%
\pgfusepath{stroke}%
\end{pgfscope}%
\begin{pgfscope}%
\definecolor{textcolor}{rgb}{0.150000,0.150000,0.150000}%
\pgfsetstrokecolor{textcolor}%
\pgfsetfillcolor{textcolor}%
\pgftext[x=0.273151in, y=0.403915in, left, base]{\color{textcolor}\rmfamily\fontsize{8.000000}{9.600000}\selectfont 0\%}%
\end{pgfscope}%
\begin{pgfscope}%
\pgfpathrectangle{\pgfqpoint{0.520798in}{0.442177in}}{\pgfqpoint{0.719005in}{0.871884in}}%
\pgfusepath{clip}%
\pgfsetroundcap%
\pgfsetroundjoin%
\pgfsetlinewidth{0.501875pt}%
\definecolor{currentstroke}{rgb}{0.800000,0.800000,0.800000}%
\pgfsetstrokecolor{currentstroke}%
\pgfsetdash{}{0pt}%
\pgfpathmoveto{\pgfqpoint{0.520798in}{0.660148in}}%
\pgfpathlineto{\pgfqpoint{1.239803in}{0.660148in}}%
\pgfusepath{stroke}%
\end{pgfscope}%
\begin{pgfscope}%
\definecolor{textcolor}{rgb}{0.150000,0.150000,0.150000}%
\pgfsetstrokecolor{textcolor}%
\pgfsetfillcolor{textcolor}%
\pgftext[x=0.214138in, y=0.621886in, left, base]{\color{textcolor}\rmfamily\fontsize{8.000000}{9.600000}\selectfont 25\%}%
\end{pgfscope}%
\begin{pgfscope}%
\pgfpathrectangle{\pgfqpoint{0.520798in}{0.442177in}}{\pgfqpoint{0.719005in}{0.871884in}}%
\pgfusepath{clip}%
\pgfsetroundcap%
\pgfsetroundjoin%
\pgfsetlinewidth{0.501875pt}%
\definecolor{currentstroke}{rgb}{0.800000,0.800000,0.800000}%
\pgfsetstrokecolor{currentstroke}%
\pgfsetdash{}{0pt}%
\pgfpathmoveto{\pgfqpoint{0.520798in}{0.878119in}}%
\pgfpathlineto{\pgfqpoint{1.239803in}{0.878119in}}%
\pgfusepath{stroke}%
\end{pgfscope}%
\begin{pgfscope}%
\definecolor{textcolor}{rgb}{0.150000,0.150000,0.150000}%
\pgfsetstrokecolor{textcolor}%
\pgfsetfillcolor{textcolor}%
\pgftext[x=0.214138in, y=0.839857in, left, base]{\color{textcolor}\rmfamily\fontsize{8.000000}{9.600000}\selectfont 50\%}%
\end{pgfscope}%
\begin{pgfscope}%
\pgfpathrectangle{\pgfqpoint{0.520798in}{0.442177in}}{\pgfqpoint{0.719005in}{0.871884in}}%
\pgfusepath{clip}%
\pgfsetroundcap%
\pgfsetroundjoin%
\pgfsetlinewidth{0.501875pt}%
\definecolor{currentstroke}{rgb}{0.800000,0.800000,0.800000}%
\pgfsetstrokecolor{currentstroke}%
\pgfsetdash{}{0pt}%
\pgfpathmoveto{\pgfqpoint{0.520798in}{1.096090in}}%
\pgfpathlineto{\pgfqpoint{1.239803in}{1.096090in}}%
\pgfusepath{stroke}%
\end{pgfscope}%
\begin{pgfscope}%
\definecolor{textcolor}{rgb}{0.150000,0.150000,0.150000}%
\pgfsetstrokecolor{textcolor}%
\pgfsetfillcolor{textcolor}%
\pgftext[x=0.214138in, y=1.057828in, left, base]{\color{textcolor}\rmfamily\fontsize{8.000000}{9.600000}\selectfont 75\%}%
\end{pgfscope}%
\begin{pgfscope}%
\pgfpathrectangle{\pgfqpoint{0.520798in}{0.442177in}}{\pgfqpoint{0.719005in}{0.871884in}}%
\pgfusepath{clip}%
\pgfsetroundcap%
\pgfsetroundjoin%
\pgfsetlinewidth{0.501875pt}%
\definecolor{currentstroke}{rgb}{0.800000,0.800000,0.800000}%
\pgfsetstrokecolor{currentstroke}%
\pgfsetdash{}{0pt}%
\pgfpathmoveto{\pgfqpoint{0.520798in}{1.314061in}}%
\pgfpathlineto{\pgfqpoint{1.239803in}{1.314061in}}%
\pgfusepath{stroke}%
\end{pgfscope}%
\begin{pgfscope}%
\definecolor{textcolor}{rgb}{0.150000,0.150000,0.150000}%
\pgfsetstrokecolor{textcolor}%
\pgfsetfillcolor{textcolor}%
\pgftext[x=0.155124in, y=1.275799in, left, base]{\color{textcolor}\rmfamily\fontsize{8.000000}{9.600000}\selectfont 100\%}%
\end{pgfscope}%
\begin{pgfscope}%
\definecolor{textcolor}{rgb}{0.150000,0.150000,0.150000}%
\pgfsetstrokecolor{textcolor}%
\pgfsetfillcolor{textcolor}%
\pgftext[x=0.099569in,y=0.878119in,,bottom,rotate=90.000000]{\color{textcolor}\rmfamily\fontsize{10.000000}{12.000000}\selectfont Cert. Acc.}%
\end{pgfscope}%
\begin{pgfscope}%
\pgfsetrectcap%
\pgfsetmiterjoin%
\pgfsetlinewidth{0.752812pt}%
\definecolor{currentstroke}{rgb}{0.700000,0.700000,0.700000}%
\pgfsetstrokecolor{currentstroke}%
\pgfsetdash{}{0pt}%
\pgfpathmoveto{\pgfqpoint{0.520798in}{0.442177in}}%
\pgfpathlineto{\pgfqpoint{0.520798in}{1.314061in}}%
\pgfusepath{stroke}%
\end{pgfscope}%
\begin{pgfscope}%
\pgfsetrectcap%
\pgfsetmiterjoin%
\pgfsetlinewidth{0.752812pt}%
\definecolor{currentstroke}{rgb}{0.700000,0.700000,0.700000}%
\pgfsetstrokecolor{currentstroke}%
\pgfsetdash{}{0pt}%
\pgfpathmoveto{\pgfqpoint{1.239803in}{0.442177in}}%
\pgfpathlineto{\pgfqpoint{1.239803in}{1.314061in}}%
\pgfusepath{stroke}%
\end{pgfscope}%
\begin{pgfscope}%
\pgfsetrectcap%
\pgfsetmiterjoin%
\pgfsetlinewidth{0.752812pt}%
\definecolor{currentstroke}{rgb}{0.700000,0.700000,0.700000}%
\pgfsetstrokecolor{currentstroke}%
\pgfsetdash{}{0pt}%
\pgfpathmoveto{\pgfqpoint{0.520798in}{0.442177in}}%
\pgfpathlineto{\pgfqpoint{1.239803in}{0.442177in}}%
\pgfusepath{stroke}%
\end{pgfscope}%
\begin{pgfscope}%
\pgfsetrectcap%
\pgfsetmiterjoin%
\pgfsetlinewidth{0.752812pt}%
\definecolor{currentstroke}{rgb}{0.700000,0.700000,0.700000}%
\pgfsetstrokecolor{currentstroke}%
\pgfsetdash{}{0pt}%
\pgfpathmoveto{\pgfqpoint{0.520798in}{1.314061in}}%
\pgfpathlineto{\pgfqpoint{1.239803in}{1.314061in}}%
\pgfusepath{stroke}%
\end{pgfscope}%
\begin{pgfscope}%
\pgfpathrectangle{\pgfqpoint{0.520798in}{0.442177in}}{\pgfqpoint{0.719005in}{0.871884in}}%
\pgfusepath{clip}%
\pgfsetroundcap%
\pgfsetroundjoin%
\pgfsetlinewidth{1.003750pt}%
\definecolor{currentstroke}{rgb}{0.003922,0.450980,0.698039}%
\pgfsetstrokecolor{currentstroke}%
\pgfsetdash{}{0pt}%
\pgfpathmoveto{\pgfqpoint{0.520798in}{1.059638in}}%
\pgfpathlineto{\pgfqpoint{0.556748in}{1.059638in}}%
\pgfpathlineto{\pgfqpoint{0.556748in}{0.994989in}}%
\pgfpathlineto{\pgfqpoint{0.628649in}{0.994989in}}%
\pgfpathlineto{\pgfqpoint{0.628649in}{0.902235in}}%
\pgfpathlineto{\pgfqpoint{0.700549in}{0.902235in}}%
\pgfpathlineto{\pgfqpoint{0.700549in}{0.690293in}}%
\pgfpathlineto{\pgfqpoint{0.772450in}{0.690293in}}%
\pgfpathlineto{\pgfqpoint{0.772450in}{0.595962in}}%
\pgfpathlineto{\pgfqpoint{0.844350in}{0.595962in}}%
\pgfpathlineto{\pgfqpoint{0.844350in}{0.586873in}}%
\pgfpathlineto{\pgfqpoint{0.916251in}{0.586873in}}%
\pgfpathlineto{\pgfqpoint{0.916251in}{0.580936in}}%
\pgfpathlineto{\pgfqpoint{0.988151in}{0.580936in}}%
\pgfpathlineto{\pgfqpoint{0.988151in}{0.570084in}}%
\pgfpathlineto{\pgfqpoint{1.060052in}{0.570084in}}%
\pgfpathlineto{\pgfqpoint{1.060052in}{0.532333in}}%
\pgfpathlineto{\pgfqpoint{1.131952in}{0.532333in}}%
\pgfpathlineto{\pgfqpoint{1.131952in}{0.525748in}}%
\pgfpathlineto{\pgfqpoint{1.203853in}{0.525748in}}%
\pgfpathlineto{\pgfqpoint{1.203853in}{0.511835in}}%
\pgfpathlineto{\pgfqpoint{1.241470in}{0.511835in}}%
\pgfusepath{stroke}%
\end{pgfscope}%
\begin{pgfscope}%
\pgfpathrectangle{\pgfqpoint{0.520798in}{0.442177in}}{\pgfqpoint{0.719005in}{0.871884in}}%
\pgfusepath{clip}%
\pgfsetbuttcap%
\pgfsetroundjoin%
\definecolor{currentfill}{rgb}{0.003922,0.450980,0.698039}%
\pgfsetfillcolor{currentfill}%
\pgfsetfillopacity{0.500000}%
\pgfsetlinewidth{0.000000pt}%
\definecolor{currentstroke}{rgb}{0.003922,0.450980,0.698039}%
\pgfsetstrokecolor{currentstroke}%
\pgfsetstrokeopacity{0.500000}%
\pgfsetdash{}{0pt}%
\pgfpathmoveto{\pgfqpoint{0.520798in}{1.075224in}}%
\pgfpathlineto{\pgfqpoint{0.520798in}{1.044052in}}%
\pgfpathlineto{\pgfqpoint{0.556748in}{1.044052in}}%
\pgfpathlineto{\pgfqpoint{0.556748in}{0.975845in}}%
\pgfpathlineto{\pgfqpoint{0.628649in}{0.975845in}}%
\pgfpathlineto{\pgfqpoint{0.628649in}{0.882557in}}%
\pgfpathlineto{\pgfqpoint{0.700549in}{0.882557in}}%
\pgfpathlineto{\pgfqpoint{0.700549in}{0.674081in}}%
\pgfpathlineto{\pgfqpoint{0.772450in}{0.674081in}}%
\pgfpathlineto{\pgfqpoint{0.772450in}{0.582632in}}%
\pgfpathlineto{\pgfqpoint{0.844350in}{0.582632in}}%
\pgfpathlineto{\pgfqpoint{0.844350in}{0.573880in}}%
\pgfpathlineto{\pgfqpoint{0.916251in}{0.573880in}}%
\pgfpathlineto{\pgfqpoint{0.916251in}{0.567716in}}%
\pgfpathlineto{\pgfqpoint{0.988151in}{0.567716in}}%
\pgfpathlineto{\pgfqpoint{0.988151in}{0.557587in}}%
\pgfpathlineto{\pgfqpoint{1.060052in}{0.557587in}}%
\pgfpathlineto{\pgfqpoint{1.060052in}{0.526441in}}%
\pgfpathlineto{\pgfqpoint{1.131952in}{0.526441in}}%
\pgfpathlineto{\pgfqpoint{1.131952in}{0.520234in}}%
\pgfpathlineto{\pgfqpoint{1.203853in}{0.520234in}}%
\pgfpathlineto{\pgfqpoint{1.203853in}{0.507570in}}%
\pgfpathlineto{\pgfqpoint{1.275753in}{0.507570in}}%
\pgfpathlineto{\pgfqpoint{1.275753in}{0.494265in}}%
\pgfpathlineto{\pgfqpoint{1.347654in}{0.494265in}}%
\pgfpathlineto{\pgfqpoint{1.347654in}{0.489315in}}%
\pgfpathlineto{\pgfqpoint{1.419554in}{0.489315in}}%
\pgfpathlineto{\pgfqpoint{1.419554in}{0.442177in}}%
\pgfpathlineto{\pgfqpoint{2.066659in}{0.442177in}}%
\pgfpathlineto{\pgfqpoint{2.066659in}{0.442177in}}%
\pgfpathlineto{\pgfqpoint{2.677813in}{0.442177in}}%
\pgfpathlineto{\pgfqpoint{2.677813in}{0.442177in}}%
\pgfpathlineto{\pgfqpoint{2.677813in}{0.442177in}}%
\pgfpathlineto{\pgfqpoint{2.066659in}{0.442177in}}%
\pgfpathlineto{\pgfqpoint{2.066659in}{0.442177in}}%
\pgfpathlineto{\pgfqpoint{1.419554in}{0.442177in}}%
\pgfpathlineto{\pgfqpoint{1.419554in}{0.493728in}}%
\pgfpathlineto{\pgfqpoint{1.347654in}{0.493728in}}%
\pgfpathlineto{\pgfqpoint{1.347654in}{0.501950in}}%
\pgfpathlineto{\pgfqpoint{1.275753in}{0.501950in}}%
\pgfpathlineto{\pgfqpoint{1.275753in}{0.516099in}}%
\pgfpathlineto{\pgfqpoint{1.203853in}{0.516099in}}%
\pgfpathlineto{\pgfqpoint{1.203853in}{0.531262in}}%
\pgfpathlineto{\pgfqpoint{1.131952in}{0.531262in}}%
\pgfpathlineto{\pgfqpoint{1.131952in}{0.538226in}}%
\pgfpathlineto{\pgfqpoint{1.060052in}{0.538226in}}%
\pgfpathlineto{\pgfqpoint{1.060052in}{0.582581in}}%
\pgfpathlineto{\pgfqpoint{0.988151in}{0.582581in}}%
\pgfpathlineto{\pgfqpoint{0.988151in}{0.594156in}}%
\pgfpathlineto{\pgfqpoint{0.916251in}{0.594156in}}%
\pgfpathlineto{\pgfqpoint{0.916251in}{0.599865in}}%
\pgfpathlineto{\pgfqpoint{0.844350in}{0.599865in}}%
\pgfpathlineto{\pgfqpoint{0.844350in}{0.609293in}}%
\pgfpathlineto{\pgfqpoint{0.772450in}{0.609293in}}%
\pgfpathlineto{\pgfqpoint{0.772450in}{0.706505in}}%
\pgfpathlineto{\pgfqpoint{0.700549in}{0.706505in}}%
\pgfpathlineto{\pgfqpoint{0.700549in}{0.921913in}}%
\pgfpathlineto{\pgfqpoint{0.628649in}{0.921913in}}%
\pgfpathlineto{\pgfqpoint{0.628649in}{1.014133in}}%
\pgfpathlineto{\pgfqpoint{0.556748in}{1.014133in}}%
\pgfpathlineto{\pgfqpoint{0.556748in}{1.075224in}}%
\pgfpathlineto{\pgfqpoint{0.520798in}{1.075224in}}%
\pgfpathlineto{\pgfqpoint{0.520798in}{1.075224in}}%
\pgfpathclose%
\pgfusepath{fill}%
\end{pgfscope}%
\begin{pgfscope}%
\pgfpathrectangle{\pgfqpoint{0.520798in}{0.442177in}}{\pgfqpoint{0.719005in}{0.871884in}}%
\pgfusepath{clip}%
\pgfsetroundcap%
\pgfsetroundjoin%
\pgfsetlinewidth{1.003750pt}%
\definecolor{currentstroke}{rgb}{0.870588,0.560784,0.019608}%
\pgfsetstrokecolor{currentstroke}%
\pgfsetdash{}{0pt}%
\pgfpathmoveto{\pgfqpoint{0.520798in}{1.059638in}}%
\pgfpathlineto{\pgfqpoint{0.556748in}{1.059638in}}%
\pgfpathlineto{\pgfqpoint{0.556748in}{1.010386in}}%
\pgfpathlineto{\pgfqpoint{0.628649in}{1.010386in}}%
\pgfpathlineto{\pgfqpoint{0.628649in}{0.965215in}}%
\pgfpathlineto{\pgfqpoint{0.700549in}{0.965215in}}%
\pgfpathlineto{\pgfqpoint{0.700549in}{0.926907in}}%
\pgfpathlineto{\pgfqpoint{0.772450in}{0.926907in}}%
\pgfpathlineto{\pgfqpoint{0.772450in}{0.890084in}}%
\pgfpathlineto{\pgfqpoint{0.844350in}{0.890084in}}%
\pgfpathlineto{\pgfqpoint{0.844350in}{0.857713in}}%
\pgfpathlineto{\pgfqpoint{0.916251in}{0.857713in}}%
\pgfpathlineto{\pgfqpoint{0.916251in}{0.690293in}}%
\pgfpathlineto{\pgfqpoint{0.988151in}{0.690293in}}%
\pgfpathlineto{\pgfqpoint{0.988151in}{0.679255in}}%
\pgfpathlineto{\pgfqpoint{1.060052in}{0.679255in}}%
\pgfpathlineto{\pgfqpoint{1.060052in}{0.595962in}}%
\pgfpathlineto{\pgfqpoint{1.131952in}{0.595962in}}%
\pgfpathlineto{\pgfqpoint{1.131952in}{0.589470in}}%
\pgfpathlineto{\pgfqpoint{1.203853in}{0.589470in}}%
\pgfpathlineto{\pgfqpoint{1.203853in}{0.582513in}}%
\pgfpathlineto{\pgfqpoint{1.241470in}{0.582513in}}%
\pgfusepath{stroke}%
\end{pgfscope}%
\begin{pgfscope}%
\pgfpathrectangle{\pgfqpoint{0.520798in}{0.442177in}}{\pgfqpoint{0.719005in}{0.871884in}}%
\pgfusepath{clip}%
\pgfsetbuttcap%
\pgfsetroundjoin%
\definecolor{currentfill}{rgb}{0.870588,0.560784,0.019608}%
\pgfsetfillcolor{currentfill}%
\pgfsetfillopacity{0.500000}%
\pgfsetlinewidth{0.000000pt}%
\definecolor{currentstroke}{rgb}{0.870588,0.560784,0.019608}%
\pgfsetstrokecolor{currentstroke}%
\pgfsetstrokeopacity{0.500000}%
\pgfsetdash{}{0pt}%
\pgfpathmoveto{\pgfqpoint{0.520798in}{1.075224in}}%
\pgfpathlineto{\pgfqpoint{0.520798in}{1.044052in}}%
\pgfpathlineto{\pgfqpoint{0.556748in}{1.044052in}}%
\pgfpathlineto{\pgfqpoint{0.556748in}{0.992114in}}%
\pgfpathlineto{\pgfqpoint{0.628649in}{0.992114in}}%
\pgfpathlineto{\pgfqpoint{0.628649in}{0.943626in}}%
\pgfpathlineto{\pgfqpoint{0.700549in}{0.943626in}}%
\pgfpathlineto{\pgfqpoint{0.700549in}{0.906077in}}%
\pgfpathlineto{\pgfqpoint{0.772450in}{0.906077in}}%
\pgfpathlineto{\pgfqpoint{0.772450in}{0.870734in}}%
\pgfpathlineto{\pgfqpoint{0.844350in}{0.870734in}}%
\pgfpathlineto{\pgfqpoint{0.844350in}{0.837104in}}%
\pgfpathlineto{\pgfqpoint{0.916251in}{0.837104in}}%
\pgfpathlineto{\pgfqpoint{0.916251in}{0.674081in}}%
\pgfpathlineto{\pgfqpoint{0.988151in}{0.674081in}}%
\pgfpathlineto{\pgfqpoint{0.988151in}{0.663686in}}%
\pgfpathlineto{\pgfqpoint{1.060052in}{0.663686in}}%
\pgfpathlineto{\pgfqpoint{1.060052in}{0.582632in}}%
\pgfpathlineto{\pgfqpoint{1.131952in}{0.582632in}}%
\pgfpathlineto{\pgfqpoint{1.131952in}{0.575499in}}%
\pgfpathlineto{\pgfqpoint{1.203853in}{0.575499in}}%
\pgfpathlineto{\pgfqpoint{1.203853in}{0.568841in}}%
\pgfpathlineto{\pgfqpoint{1.275753in}{0.568841in}}%
\pgfpathlineto{\pgfqpoint{1.275753in}{0.563978in}}%
\pgfpathlineto{\pgfqpoint{1.347654in}{0.563978in}}%
\pgfpathlineto{\pgfqpoint{1.347654in}{0.559518in}}%
\pgfpathlineto{\pgfqpoint{1.419554in}{0.559518in}}%
\pgfpathlineto{\pgfqpoint{1.419554in}{0.554915in}}%
\pgfpathlineto{\pgfqpoint{1.491455in}{0.554915in}}%
\pgfpathlineto{\pgfqpoint{1.491455in}{0.549254in}}%
\pgfpathlineto{\pgfqpoint{1.563355in}{0.549254in}}%
\pgfpathlineto{\pgfqpoint{1.563355in}{0.545796in}}%
\pgfpathlineto{\pgfqpoint{1.635256in}{0.545796in}}%
\pgfpathlineto{\pgfqpoint{1.635256in}{0.525749in}}%
\pgfpathlineto{\pgfqpoint{1.707156in}{0.525749in}}%
\pgfpathlineto{\pgfqpoint{1.707156in}{0.524310in}}%
\pgfpathlineto{\pgfqpoint{1.779057in}{0.524310in}}%
\pgfpathlineto{\pgfqpoint{1.779057in}{0.518997in}}%
\pgfpathlineto{\pgfqpoint{1.850957in}{0.518997in}}%
\pgfpathlineto{\pgfqpoint{1.850957in}{0.515937in}}%
\pgfpathlineto{\pgfqpoint{1.922858in}{0.515937in}}%
\pgfpathlineto{\pgfqpoint{1.922858in}{0.507163in}}%
\pgfpathlineto{\pgfqpoint{1.994758in}{0.507163in}}%
\pgfpathlineto{\pgfqpoint{1.994758in}{0.495959in}}%
\pgfpathlineto{\pgfqpoint{2.066659in}{0.495959in}}%
\pgfpathlineto{\pgfqpoint{2.066659in}{0.490623in}}%
\pgfpathlineto{\pgfqpoint{2.138559in}{0.490623in}}%
\pgfpathlineto{\pgfqpoint{2.138559in}{0.486315in}}%
\pgfpathlineto{\pgfqpoint{2.210460in}{0.486315in}}%
\pgfpathlineto{\pgfqpoint{2.210460in}{0.469881in}}%
\pgfpathlineto{\pgfqpoint{2.282360in}{0.469881in}}%
\pgfpathlineto{\pgfqpoint{2.282360in}{0.442177in}}%
\pgfpathlineto{\pgfqpoint{2.498062in}{0.442177in}}%
\pgfpathlineto{\pgfqpoint{2.498062in}{0.442177in}}%
\pgfpathlineto{\pgfqpoint{2.677813in}{0.442177in}}%
\pgfpathlineto{\pgfqpoint{2.677813in}{0.442177in}}%
\pgfpathlineto{\pgfqpoint{2.677813in}{0.442177in}}%
\pgfpathlineto{\pgfqpoint{2.498062in}{0.442177in}}%
\pgfpathlineto{\pgfqpoint{2.498062in}{0.442177in}}%
\pgfpathlineto{\pgfqpoint{2.282360in}{0.442177in}}%
\pgfpathlineto{\pgfqpoint{2.282360in}{0.477916in}}%
\pgfpathlineto{\pgfqpoint{2.210460in}{0.477916in}}%
\pgfpathlineto{\pgfqpoint{2.210460in}{0.493019in}}%
\pgfpathlineto{\pgfqpoint{2.138559in}{0.493019in}}%
\pgfpathlineto{\pgfqpoint{2.138559in}{0.496501in}}%
\pgfpathlineto{\pgfqpoint{2.066659in}{0.496501in}}%
\pgfpathlineto{\pgfqpoint{2.066659in}{0.503224in}}%
\pgfpathlineto{\pgfqpoint{1.994758in}{0.503224in}}%
\pgfpathlineto{\pgfqpoint{1.994758in}{0.514466in}}%
\pgfpathlineto{\pgfqpoint{1.922858in}{0.514466in}}%
\pgfpathlineto{\pgfqpoint{1.922858in}{0.526840in}}%
\pgfpathlineto{\pgfqpoint{1.850957in}{0.526840in}}%
\pgfpathlineto{\pgfqpoint{1.850957in}{0.530272in}}%
\pgfpathlineto{\pgfqpoint{1.779057in}{0.530272in}}%
\pgfpathlineto{\pgfqpoint{1.779057in}{0.536275in}}%
\pgfpathlineto{\pgfqpoint{1.707156in}{0.536275in}}%
\pgfpathlineto{\pgfqpoint{1.707156in}{0.537805in}}%
\pgfpathlineto{\pgfqpoint{1.635256in}{0.537805in}}%
\pgfpathlineto{\pgfqpoint{1.635256in}{0.568216in}}%
\pgfpathlineto{\pgfqpoint{1.563355in}{0.568216in}}%
\pgfpathlineto{\pgfqpoint{1.563355in}{0.572549in}}%
\pgfpathlineto{\pgfqpoint{1.491455in}{0.572549in}}%
\pgfpathlineto{\pgfqpoint{1.491455in}{0.576163in}}%
\pgfpathlineto{\pgfqpoint{1.419554in}{0.576163in}}%
\pgfpathlineto{\pgfqpoint{1.419554in}{0.581763in}}%
\pgfpathlineto{\pgfqpoint{1.347654in}{0.581763in}}%
\pgfpathlineto{\pgfqpoint{1.347654in}{0.588434in}}%
\pgfpathlineto{\pgfqpoint{1.275753in}{0.588434in}}%
\pgfpathlineto{\pgfqpoint{1.275753in}{0.596185in}}%
\pgfpathlineto{\pgfqpoint{1.203853in}{0.596185in}}%
\pgfpathlineto{\pgfqpoint{1.203853in}{0.603440in}}%
\pgfpathlineto{\pgfqpoint{1.131952in}{0.603440in}}%
\pgfpathlineto{\pgfqpoint{1.131952in}{0.609293in}}%
\pgfpathlineto{\pgfqpoint{1.060052in}{0.609293in}}%
\pgfpathlineto{\pgfqpoint{1.060052in}{0.694824in}}%
\pgfpathlineto{\pgfqpoint{0.988151in}{0.694824in}}%
\pgfpathlineto{\pgfqpoint{0.988151in}{0.706505in}}%
\pgfpathlineto{\pgfqpoint{0.916251in}{0.706505in}}%
\pgfpathlineto{\pgfqpoint{0.916251in}{0.878323in}}%
\pgfpathlineto{\pgfqpoint{0.844350in}{0.878323in}}%
\pgfpathlineto{\pgfqpoint{0.844350in}{0.909435in}}%
\pgfpathlineto{\pgfqpoint{0.772450in}{0.909435in}}%
\pgfpathlineto{\pgfqpoint{0.772450in}{0.947738in}}%
\pgfpathlineto{\pgfqpoint{0.700549in}{0.947738in}}%
\pgfpathlineto{\pgfqpoint{0.700549in}{0.986803in}}%
\pgfpathlineto{\pgfqpoint{0.628649in}{0.986803in}}%
\pgfpathlineto{\pgfqpoint{0.628649in}{1.028657in}}%
\pgfpathlineto{\pgfqpoint{0.556748in}{1.028657in}}%
\pgfpathlineto{\pgfqpoint{0.556748in}{1.075224in}}%
\pgfpathlineto{\pgfqpoint{0.520798in}{1.075224in}}%
\pgfpathlineto{\pgfqpoint{0.520798in}{1.075224in}}%
\pgfpathclose%
\pgfusepath{fill}%
\end{pgfscope}%
\begin{pgfscope}%
\pgfpathrectangle{\pgfqpoint{0.520798in}{0.442177in}}{\pgfqpoint{0.719005in}{0.871884in}}%
\pgfusepath{clip}%
\pgfsetroundcap%
\pgfsetroundjoin%
\pgfsetlinewidth{1.003750pt}%
\definecolor{currentstroke}{rgb}{0.007843,0.619608,0.450980}%
\pgfsetstrokecolor{currentstroke}%
\pgfsetdash{}{0pt}%
\pgfpathmoveto{\pgfqpoint{0.520798in}{1.059638in}}%
\pgfpathlineto{\pgfqpoint{0.556748in}{1.059638in}}%
\pgfpathlineto{\pgfqpoint{0.556748in}{1.010386in}}%
\pgfpathlineto{\pgfqpoint{0.628649in}{1.010386in}}%
\pgfpathlineto{\pgfqpoint{0.628649in}{0.965215in}}%
\pgfpathlineto{\pgfqpoint{0.700549in}{0.965215in}}%
\pgfpathlineto{\pgfqpoint{0.700549in}{0.926907in}}%
\pgfpathlineto{\pgfqpoint{0.772450in}{0.926907in}}%
\pgfpathlineto{\pgfqpoint{0.772450in}{0.890084in}}%
\pgfpathlineto{\pgfqpoint{0.844350in}{0.890084in}}%
\pgfpathlineto{\pgfqpoint{0.844350in}{0.857713in}}%
\pgfpathlineto{\pgfqpoint{0.916251in}{0.857713in}}%
\pgfpathlineto{\pgfqpoint{0.916251in}{0.829423in}}%
\pgfpathlineto{\pgfqpoint{0.988151in}{0.829423in}}%
\pgfpathlineto{\pgfqpoint{0.988151in}{0.803267in}}%
\pgfpathlineto{\pgfqpoint{1.060052in}{0.803267in}}%
\pgfpathlineto{\pgfqpoint{1.060052in}{0.779151in}}%
\pgfpathlineto{\pgfqpoint{1.131952in}{0.779151in}}%
\pgfpathlineto{\pgfqpoint{1.131952in}{0.760600in}}%
\pgfpathlineto{\pgfqpoint{1.203853in}{0.760600in}}%
\pgfpathlineto{\pgfqpoint{1.203853in}{0.742699in}}%
\pgfpathlineto{\pgfqpoint{1.241470in}{0.742699in}}%
\pgfusepath{stroke}%
\end{pgfscope}%
\begin{pgfscope}%
\pgfpathrectangle{\pgfqpoint{0.520798in}{0.442177in}}{\pgfqpoint{0.719005in}{0.871884in}}%
\pgfusepath{clip}%
\pgfsetbuttcap%
\pgfsetroundjoin%
\definecolor{currentfill}{rgb}{0.007843,0.619608,0.450980}%
\pgfsetfillcolor{currentfill}%
\pgfsetfillopacity{0.500000}%
\pgfsetlinewidth{0.000000pt}%
\definecolor{currentstroke}{rgb}{0.007843,0.619608,0.450980}%
\pgfsetstrokecolor{currentstroke}%
\pgfsetstrokeopacity{0.500000}%
\pgfsetdash{}{0pt}%
\pgfpathmoveto{\pgfqpoint{0.520798in}{1.075224in}}%
\pgfpathlineto{\pgfqpoint{0.520798in}{1.044052in}}%
\pgfpathlineto{\pgfqpoint{0.556748in}{1.044052in}}%
\pgfpathlineto{\pgfqpoint{0.556748in}{0.992114in}}%
\pgfpathlineto{\pgfqpoint{0.628649in}{0.992114in}}%
\pgfpathlineto{\pgfqpoint{0.628649in}{0.943626in}}%
\pgfpathlineto{\pgfqpoint{0.700549in}{0.943626in}}%
\pgfpathlineto{\pgfqpoint{0.700549in}{0.906077in}}%
\pgfpathlineto{\pgfqpoint{0.772450in}{0.906077in}}%
\pgfpathlineto{\pgfqpoint{0.772450in}{0.870734in}}%
\pgfpathlineto{\pgfqpoint{0.844350in}{0.870734in}}%
\pgfpathlineto{\pgfqpoint{0.844350in}{0.837104in}}%
\pgfpathlineto{\pgfqpoint{0.916251in}{0.837104in}}%
\pgfpathlineto{\pgfqpoint{0.916251in}{0.810120in}}%
\pgfpathlineto{\pgfqpoint{0.988151in}{0.810120in}}%
\pgfpathlineto{\pgfqpoint{0.988151in}{0.786258in}}%
\pgfpathlineto{\pgfqpoint{1.060052in}{0.786258in}}%
\pgfpathlineto{\pgfqpoint{1.060052in}{0.760123in}}%
\pgfpathlineto{\pgfqpoint{1.131952in}{0.760123in}}%
\pgfpathlineto{\pgfqpoint{1.131952in}{0.742695in}}%
\pgfpathlineto{\pgfqpoint{1.203853in}{0.742695in}}%
\pgfpathlineto{\pgfqpoint{1.203853in}{0.723733in}}%
\pgfpathlineto{\pgfqpoint{1.275753in}{0.723733in}}%
\pgfpathlineto{\pgfqpoint{1.275753in}{0.705938in}}%
\pgfpathlineto{\pgfqpoint{1.347654in}{0.705938in}}%
\pgfpathlineto{\pgfqpoint{1.347654in}{0.674081in}}%
\pgfpathlineto{\pgfqpoint{1.419554in}{0.674081in}}%
\pgfpathlineto{\pgfqpoint{1.419554in}{0.663686in}}%
\pgfpathlineto{\pgfqpoint{1.491455in}{0.663686in}}%
\pgfpathlineto{\pgfqpoint{1.491455in}{0.652376in}}%
\pgfpathlineto{\pgfqpoint{1.563355in}{0.652376in}}%
\pgfpathlineto{\pgfqpoint{1.563355in}{0.642547in}}%
\pgfpathlineto{\pgfqpoint{1.635256in}{0.642547in}}%
\pgfpathlineto{\pgfqpoint{1.635256in}{0.582632in}}%
\pgfpathlineto{\pgfqpoint{1.707156in}{0.582632in}}%
\pgfpathlineto{\pgfqpoint{1.707156in}{0.575499in}}%
\pgfpathlineto{\pgfqpoint{1.779057in}{0.575499in}}%
\pgfpathlineto{\pgfqpoint{1.779057in}{0.569869in}}%
\pgfpathlineto{\pgfqpoint{1.850957in}{0.569869in}}%
\pgfpathlineto{\pgfqpoint{1.850957in}{0.564530in}}%
\pgfpathlineto{\pgfqpoint{1.922858in}{0.564530in}}%
\pgfpathlineto{\pgfqpoint{1.922858in}{0.560460in}}%
\pgfpathlineto{\pgfqpoint{1.994758in}{0.560460in}}%
\pgfpathlineto{\pgfqpoint{1.994758in}{0.556537in}}%
\pgfpathlineto{\pgfqpoint{2.066659in}{0.556537in}}%
\pgfpathlineto{\pgfqpoint{2.066659in}{0.552409in}}%
\pgfpathlineto{\pgfqpoint{2.138559in}{0.552409in}}%
\pgfpathlineto{\pgfqpoint{2.138559in}{0.549428in}}%
\pgfpathlineto{\pgfqpoint{2.210460in}{0.549428in}}%
\pgfpathlineto{\pgfqpoint{2.210460in}{0.547177in}}%
\pgfpathlineto{\pgfqpoint{2.282360in}{0.547177in}}%
\pgfpathlineto{\pgfqpoint{2.282360in}{0.544505in}}%
\pgfpathlineto{\pgfqpoint{2.354261in}{0.544505in}}%
\pgfpathlineto{\pgfqpoint{2.354261in}{0.541554in}}%
\pgfpathlineto{\pgfqpoint{2.426161in}{0.541554in}}%
\pgfpathlineto{\pgfqpoint{2.426161in}{0.536658in}}%
\pgfpathlineto{\pgfqpoint{2.498062in}{0.536658in}}%
\pgfpathlineto{\pgfqpoint{2.498062in}{0.534057in}}%
\pgfpathlineto{\pgfqpoint{2.569962in}{0.534057in}}%
\pgfpathlineto{\pgfqpoint{2.569962in}{0.531316in}}%
\pgfpathlineto{\pgfqpoint{2.641863in}{0.531316in}}%
\pgfpathlineto{\pgfqpoint{2.641863in}{0.529550in}}%
\pgfpathlineto{\pgfqpoint{2.677813in}{0.529550in}}%
\pgfpathlineto{\pgfqpoint{2.677813in}{0.547174in}}%
\pgfpathlineto{\pgfqpoint{2.677813in}{0.547174in}}%
\pgfpathlineto{\pgfqpoint{2.641863in}{0.547174in}}%
\pgfpathlineto{\pgfqpoint{2.641863in}{0.549861in}}%
\pgfpathlineto{\pgfqpoint{2.569962in}{0.549861in}}%
\pgfpathlineto{\pgfqpoint{2.569962in}{0.552871in}}%
\pgfpathlineto{\pgfqpoint{2.498062in}{0.552871in}}%
\pgfpathlineto{\pgfqpoint{2.498062in}{0.555835in}}%
\pgfpathlineto{\pgfqpoint{2.426161in}{0.555835in}}%
\pgfpathlineto{\pgfqpoint{2.426161in}{0.558916in}}%
\pgfpathlineto{\pgfqpoint{2.354261in}{0.558916in}}%
\pgfpathlineto{\pgfqpoint{2.354261in}{0.560231in}}%
\pgfpathlineto{\pgfqpoint{2.282360in}{0.560231in}}%
\pgfpathlineto{\pgfqpoint{2.282360in}{0.564053in}}%
\pgfpathlineto{\pgfqpoint{2.210460in}{0.564053in}}%
\pgfpathlineto{\pgfqpoint{2.210460in}{0.570149in}}%
\pgfpathlineto{\pgfqpoint{2.138559in}{0.570149in}}%
\pgfpathlineto{\pgfqpoint{2.138559in}{0.573846in}}%
\pgfpathlineto{\pgfqpoint{2.066659in}{0.573846in}}%
\pgfpathlineto{\pgfqpoint{2.066659in}{0.577139in}}%
\pgfpathlineto{\pgfqpoint{1.994758in}{0.577139in}}%
\pgfpathlineto{\pgfqpoint{1.994758in}{0.582862in}}%
\pgfpathlineto{\pgfqpoint{1.922858in}{0.582862in}}%
\pgfpathlineto{\pgfqpoint{1.922858in}{0.589552in}}%
\pgfpathlineto{\pgfqpoint{1.850957in}{0.589552in}}%
\pgfpathlineto{\pgfqpoint{1.850957in}{0.597384in}}%
\pgfpathlineto{\pgfqpoint{1.779057in}{0.597384in}}%
\pgfpathlineto{\pgfqpoint{1.779057in}{0.603440in}}%
\pgfpathlineto{\pgfqpoint{1.707156in}{0.603440in}}%
\pgfpathlineto{\pgfqpoint{1.707156in}{0.609293in}}%
\pgfpathlineto{\pgfqpoint{1.635256in}{0.609293in}}%
\pgfpathlineto{\pgfqpoint{1.635256in}{0.671998in}}%
\pgfpathlineto{\pgfqpoint{1.563355in}{0.671998in}}%
\pgfpathlineto{\pgfqpoint{1.563355in}{0.682389in}}%
\pgfpathlineto{\pgfqpoint{1.491455in}{0.682389in}}%
\pgfpathlineto{\pgfqpoint{1.491455in}{0.694824in}}%
\pgfpathlineto{\pgfqpoint{1.419554in}{0.694824in}}%
\pgfpathlineto{\pgfqpoint{1.419554in}{0.706505in}}%
\pgfpathlineto{\pgfqpoint{1.347654in}{0.706505in}}%
\pgfpathlineto{\pgfqpoint{1.347654in}{0.741802in}}%
\pgfpathlineto{\pgfqpoint{1.275753in}{0.741802in}}%
\pgfpathlineto{\pgfqpoint{1.275753in}{0.761664in}}%
\pgfpathlineto{\pgfqpoint{1.203853in}{0.761664in}}%
\pgfpathlineto{\pgfqpoint{1.203853in}{0.778506in}}%
\pgfpathlineto{\pgfqpoint{1.131952in}{0.778506in}}%
\pgfpathlineto{\pgfqpoint{1.131952in}{0.798179in}}%
\pgfpathlineto{\pgfqpoint{1.060052in}{0.798179in}}%
\pgfpathlineto{\pgfqpoint{1.060052in}{0.820275in}}%
\pgfpathlineto{\pgfqpoint{0.988151in}{0.820275in}}%
\pgfpathlineto{\pgfqpoint{0.988151in}{0.848726in}}%
\pgfpathlineto{\pgfqpoint{0.916251in}{0.848726in}}%
\pgfpathlineto{\pgfqpoint{0.916251in}{0.878323in}}%
\pgfpathlineto{\pgfqpoint{0.844350in}{0.878323in}}%
\pgfpathlineto{\pgfqpoint{0.844350in}{0.909435in}}%
\pgfpathlineto{\pgfqpoint{0.772450in}{0.909435in}}%
\pgfpathlineto{\pgfqpoint{0.772450in}{0.947738in}}%
\pgfpathlineto{\pgfqpoint{0.700549in}{0.947738in}}%
\pgfpathlineto{\pgfqpoint{0.700549in}{0.986803in}}%
\pgfpathlineto{\pgfqpoint{0.628649in}{0.986803in}}%
\pgfpathlineto{\pgfqpoint{0.628649in}{1.028657in}}%
\pgfpathlineto{\pgfqpoint{0.556748in}{1.028657in}}%
\pgfpathlineto{\pgfqpoint{0.556748in}{1.075224in}}%
\pgfpathlineto{\pgfqpoint{0.520798in}{1.075224in}}%
\pgfpathlineto{\pgfqpoint{0.520798in}{1.075224in}}%
\pgfpathclose%
\pgfusepath{fill}%
\end{pgfscope}%
\begin{pgfscope}%
\definecolor{textcolor}{rgb}{0.150000,0.150000,0.150000}%
\pgfsetstrokecolor{textcolor}%
\pgfsetfillcolor{textcolor}%
\pgftext[x=0.880300in,y=1.397395in,,base]{\color{textcolor}\rmfamily\fontsize{9.000000}{10.800000}\selectfont \(\displaystyle c_X^-=1\)}%
\end{pgfscope}%
\begin{pgfscope}%
\pgfsetbuttcap%
\pgfsetmiterjoin%
\definecolor{currentfill}{rgb}{1.000000,1.000000,1.000000}%
\pgfsetfillcolor{currentfill}%
\pgfsetfillopacity{0.800000}%
\pgfsetlinewidth{1.003750pt}%
\definecolor{currentstroke}{rgb}{0.800000,0.800000,0.800000}%
\pgfsetstrokecolor{currentstroke}%
\pgfsetstrokeopacity{0.800000}%
\pgfsetdash{}{0pt}%
\pgfpathmoveto{\pgfqpoint{0.785843in}{0.638103in}}%
\pgfpathlineto{\pgfqpoint{1.191192in}{0.638103in}}%
\pgfpathlineto{\pgfqpoint{1.191192in}{1.265450in}}%
\pgfpathlineto{\pgfqpoint{0.785843in}{1.265450in}}%
\pgfpathlineto{\pgfqpoint{0.785843in}{0.638103in}}%
\pgfpathclose%
\pgfusepath{stroke,fill}%
\end{pgfscope}%
\begin{pgfscope}%
\definecolor{textcolor}{rgb}{0.150000,0.150000,0.150000}%
\pgfsetstrokecolor{textcolor}%
\pgfsetfillcolor{textcolor}%
\pgftext[x=0.912291in,y=1.113275in,left,base]{\color{textcolor}\rmfamily\fontsize{9.000000}{10.800000}\selectfont \(\displaystyle c_X^+\)}%
\end{pgfscope}%
\begin{pgfscope}%
\pgfsetroundcap%
\pgfsetroundjoin%
\pgfsetlinewidth{1.003750pt}%
\definecolor{currentstroke}{rgb}{0.003922,0.450980,0.698039}%
\pgfsetstrokecolor{currentstroke}%
\pgfsetdash{}{0pt}%
\pgfpathmoveto{\pgfqpoint{0.824732in}{1.001052in}}%
\pgfpathlineto{\pgfqpoint{0.873343in}{1.001052in}}%
\pgfpathlineto{\pgfqpoint{0.873343in}{1.001052in}}%
\pgfpathlineto{\pgfqpoint{0.970565in}{1.001052in}}%
\pgfpathlineto{\pgfqpoint{0.970565in}{1.001052in}}%
\pgfpathlineto{\pgfqpoint{1.019176in}{1.001052in}}%
\pgfusepath{stroke}%
\end{pgfscope}%
\begin{pgfscope}%
\definecolor{textcolor}{rgb}{0.150000,0.150000,0.150000}%
\pgfsetstrokecolor{textcolor}%
\pgfsetfillcolor{textcolor}%
\pgftext[x=1.096954in,y=0.967024in,left,base]{\color{textcolor}\rmfamily\fontsize{7.000000}{8.400000}\selectfont 1}%
\end{pgfscope}%
\begin{pgfscope}%
\pgfsetroundcap%
\pgfsetroundjoin%
\pgfsetlinewidth{1.003750pt}%
\definecolor{currentstroke}{rgb}{0.870588,0.560784,0.019608}%
\pgfsetstrokecolor{currentstroke}%
\pgfsetdash{}{0pt}%
\pgfpathmoveto{\pgfqpoint{0.824732in}{0.865486in}}%
\pgfpathlineto{\pgfqpoint{0.873343in}{0.865486in}}%
\pgfpathlineto{\pgfqpoint{0.873343in}{0.865486in}}%
\pgfpathlineto{\pgfqpoint{0.970565in}{0.865486in}}%
\pgfpathlineto{\pgfqpoint{0.970565in}{0.865486in}}%
\pgfpathlineto{\pgfqpoint{1.019176in}{0.865486in}}%
\pgfusepath{stroke}%
\end{pgfscope}%
\begin{pgfscope}%
\definecolor{textcolor}{rgb}{0.150000,0.150000,0.150000}%
\pgfsetstrokecolor{textcolor}%
\pgfsetfillcolor{textcolor}%
\pgftext[x=1.096954in,y=0.831458in,left,base]{\color{textcolor}\rmfamily\fontsize{7.000000}{8.400000}\selectfont 2}%
\end{pgfscope}%
\begin{pgfscope}%
\pgfsetroundcap%
\pgfsetroundjoin%
\pgfsetlinewidth{1.003750pt}%
\definecolor{currentstroke}{rgb}{0.007843,0.619608,0.450980}%
\pgfsetstrokecolor{currentstroke}%
\pgfsetdash{}{0pt}%
\pgfpathmoveto{\pgfqpoint{0.824732in}{0.729919in}}%
\pgfpathlineto{\pgfqpoint{0.873343in}{0.729919in}}%
\pgfpathlineto{\pgfqpoint{0.873343in}{0.729919in}}%
\pgfpathlineto{\pgfqpoint{0.970565in}{0.729919in}}%
\pgfpathlineto{\pgfqpoint{0.970565in}{0.729919in}}%
\pgfpathlineto{\pgfqpoint{1.019176in}{0.729919in}}%
\pgfusepath{stroke}%
\end{pgfscope}%
\begin{pgfscope}%
\definecolor{textcolor}{rgb}{0.150000,0.150000,0.150000}%
\pgfsetstrokecolor{textcolor}%
\pgfsetfillcolor{textcolor}%
\pgftext[x=1.096954in,y=0.695892in,left,base]{\color{textcolor}\rmfamily\fontsize{7.000000}{8.400000}\selectfont 4}%
\end{pgfscope}%
\end{pgfpicture}%
\makeatother%
\endgroup%

%% file: figures/experiments/graphs/sparse_smoothing/nodes_attributes/node_classification-Citeseer-GCN-hidden=32-p_adj_plus=0.0-p_adj_minus=0.0-p_att_plus=0.001-p_att_minus=0.8-multi_class_cert-A.pgf
\begingroup%
\makeatletter%
\begin{pgfpicture}%
\pgfpathrectangle{\pgfpointorigin}{\pgfqpoint{1.375000in}{1.581250in}}%
\pgfusepath{use as bounding box, clip}%
\begin{pgfscope}%
\pgfsetbuttcap%
\pgfsetmiterjoin%
\definecolor{currentfill}{rgb}{1.000000,1.000000,1.000000}%
\pgfsetfillcolor{currentfill}%
\pgfsetlinewidth{0.000000pt}%
\definecolor{currentstroke}{rgb}{1.000000,1.000000,1.000000}%
\pgfsetstrokecolor{currentstroke}%
\pgfsetdash{}{0pt}%
\pgfpathmoveto{\pgfqpoint{0.000000in}{0.000000in}}%
\pgfpathlineto{\pgfqpoint{1.375000in}{0.000000in}}%
\pgfpathlineto{\pgfqpoint{1.375000in}{1.581250in}}%
\pgfpathlineto{\pgfqpoint{0.000000in}{1.581250in}}%
\pgfpathlineto{\pgfqpoint{0.000000in}{0.000000in}}%
\pgfpathclose%
\pgfusepath{fill}%
\end{pgfscope}%
\begin{pgfscope}%
\pgfsetbuttcap%
\pgfsetmiterjoin%
\definecolor{currentfill}{rgb}{1.000000,1.000000,1.000000}%
\pgfsetfillcolor{currentfill}%
\pgfsetlinewidth{0.000000pt}%
\definecolor{currentstroke}{rgb}{0.000000,0.000000,0.000000}%
\pgfsetstrokecolor{currentstroke}%
\pgfsetstrokeopacity{0.000000}%
\pgfsetdash{}{0pt}%
\pgfpathmoveto{\pgfqpoint{0.520798in}{0.442177in}}%
\pgfpathlineto{\pgfqpoint{1.239803in}{0.442177in}}%
\pgfpathlineto{\pgfqpoint{1.239803in}{1.314061in}}%
\pgfpathlineto{\pgfqpoint{0.520798in}{1.314061in}}%
\pgfpathlineto{\pgfqpoint{0.520798in}{0.442177in}}%
\pgfpathclose%
\pgfusepath{fill}%
\end{pgfscope}%
\begin{pgfscope}%
\pgfpathrectangle{\pgfqpoint{0.520798in}{0.442177in}}{\pgfqpoint{0.719005in}{0.871884in}}%
\pgfusepath{clip}%
\pgfsetroundcap%
\pgfsetroundjoin%
\pgfsetlinewidth{0.501875pt}%
\definecolor{currentstroke}{rgb}{0.800000,0.800000,0.800000}%
\pgfsetstrokecolor{currentstroke}%
\pgfsetdash{}{0pt}%
\pgfpathmoveto{\pgfqpoint{0.520798in}{0.442177in}}%
\pgfpathlineto{\pgfqpoint{0.520798in}{1.314061in}}%
\pgfusepath{stroke}%
\end{pgfscope}%
\begin{pgfscope}%
\definecolor{textcolor}{rgb}{0.150000,0.150000,0.150000}%
\pgfsetstrokecolor{textcolor}%
\pgfsetfillcolor{textcolor}%
\pgftext[x=0.520798in,y=0.351899in,,top]{\color{textcolor}\rmfamily\fontsize{8.000000}{9.600000}\selectfont \(\displaystyle {0}\)}%
\end{pgfscope}%
\begin{pgfscope}%
\pgfpathrectangle{\pgfqpoint{0.520798in}{0.442177in}}{\pgfqpoint{0.719005in}{0.871884in}}%
\pgfusepath{clip}%
\pgfsetroundcap%
\pgfsetroundjoin%
\pgfsetlinewidth{0.501875pt}%
\definecolor{currentstroke}{rgb}{0.800000,0.800000,0.800000}%
\pgfsetstrokecolor{currentstroke}%
\pgfsetdash{}{0pt}%
\pgfpathmoveto{\pgfqpoint{0.880300in}{0.442177in}}%
\pgfpathlineto{\pgfqpoint{0.880300in}{1.314061in}}%
\pgfusepath{stroke}%
\end{pgfscope}%
\begin{pgfscope}%
\definecolor{textcolor}{rgb}{0.150000,0.150000,0.150000}%
\pgfsetstrokecolor{textcolor}%
\pgfsetfillcolor{textcolor}%
\pgftext[x=0.880300in,y=0.351899in,,top]{\color{textcolor}\rmfamily\fontsize{8.000000}{9.600000}\selectfont \(\displaystyle {5}\)}%
\end{pgfscope}%
\begin{pgfscope}%
\pgfpathrectangle{\pgfqpoint{0.520798in}{0.442177in}}{\pgfqpoint{0.719005in}{0.871884in}}%
\pgfusepath{clip}%
\pgfsetroundcap%
\pgfsetroundjoin%
\pgfsetlinewidth{0.501875pt}%
\definecolor{currentstroke}{rgb}{0.800000,0.800000,0.800000}%
\pgfsetstrokecolor{currentstroke}%
\pgfsetdash{}{0pt}%
\pgfpathmoveto{\pgfqpoint{1.239803in}{0.442177in}}%
\pgfpathlineto{\pgfqpoint{1.239803in}{1.314061in}}%
\pgfusepath{stroke}%
\end{pgfscope}%
\begin{pgfscope}%
\definecolor{textcolor}{rgb}{0.150000,0.150000,0.150000}%
\pgfsetstrokecolor{textcolor}%
\pgfsetfillcolor{textcolor}%
\pgftext[x=1.239803in,y=0.351899in,,top]{\color{textcolor}\rmfamily\fontsize{8.000000}{9.600000}\selectfont \(\displaystyle {10}\)}%
\end{pgfscope}%
\begin{pgfscope}%
\definecolor{textcolor}{rgb}{0.150000,0.150000,0.150000}%
\pgfsetstrokecolor{textcolor}%
\pgfsetfillcolor{textcolor}%
\pgftext[x=0.880300in,y=0.198219in,,top]{\color{textcolor}\rmfamily\fontsize{10.000000}{12.000000}\selectfont Edit distance \(\displaystyle \epsilon\)}%
\end{pgfscope}%
\begin{pgfscope}%
\pgfpathrectangle{\pgfqpoint{0.520798in}{0.442177in}}{\pgfqpoint{0.719005in}{0.871884in}}%
\pgfusepath{clip}%
\pgfsetroundcap%
\pgfsetroundjoin%
\pgfsetlinewidth{0.501875pt}%
\definecolor{currentstroke}{rgb}{0.800000,0.800000,0.800000}%
\pgfsetstrokecolor{currentstroke}%
\pgfsetdash{}{0pt}%
\pgfpathmoveto{\pgfqpoint{0.520798in}{0.442177in}}%
\pgfpathlineto{\pgfqpoint{1.239803in}{0.442177in}}%
\pgfusepath{stroke}%
\end{pgfscope}%
\begin{pgfscope}%
\definecolor{textcolor}{rgb}{0.150000,0.150000,0.150000}%
\pgfsetstrokecolor{textcolor}%
\pgfsetfillcolor{textcolor}%
\pgftext[x=0.273151in, y=0.403915in, left, base]{\color{textcolor}\rmfamily\fontsize{8.000000}{9.600000}\selectfont 0\%}%
\end{pgfscope}%
\begin{pgfscope}%
\pgfpathrectangle{\pgfqpoint{0.520798in}{0.442177in}}{\pgfqpoint{0.719005in}{0.871884in}}%
\pgfusepath{clip}%
\pgfsetroundcap%
\pgfsetroundjoin%
\pgfsetlinewidth{0.501875pt}%
\definecolor{currentstroke}{rgb}{0.800000,0.800000,0.800000}%
\pgfsetstrokecolor{currentstroke}%
\pgfsetdash{}{0pt}%
\pgfpathmoveto{\pgfqpoint{0.520798in}{0.660148in}}%
\pgfpathlineto{\pgfqpoint{1.239803in}{0.660148in}}%
\pgfusepath{stroke}%
\end{pgfscope}%
\begin{pgfscope}%
\definecolor{textcolor}{rgb}{0.150000,0.150000,0.150000}%
\pgfsetstrokecolor{textcolor}%
\pgfsetfillcolor{textcolor}%
\pgftext[x=0.214138in, y=0.621886in, left, base]{\color{textcolor}\rmfamily\fontsize{8.000000}{9.600000}\selectfont 25\%}%
\end{pgfscope}%
\begin{pgfscope}%
\pgfpathrectangle{\pgfqpoint{0.520798in}{0.442177in}}{\pgfqpoint{0.719005in}{0.871884in}}%
\pgfusepath{clip}%
\pgfsetroundcap%
\pgfsetroundjoin%
\pgfsetlinewidth{0.501875pt}%
\definecolor{currentstroke}{rgb}{0.800000,0.800000,0.800000}%
\pgfsetstrokecolor{currentstroke}%
\pgfsetdash{}{0pt}%
\pgfpathmoveto{\pgfqpoint{0.520798in}{0.878119in}}%
\pgfpathlineto{\pgfqpoint{1.239803in}{0.878119in}}%
\pgfusepath{stroke}%
\end{pgfscope}%
\begin{pgfscope}%
\definecolor{textcolor}{rgb}{0.150000,0.150000,0.150000}%
\pgfsetstrokecolor{textcolor}%
\pgfsetfillcolor{textcolor}%
\pgftext[x=0.214138in, y=0.839857in, left, base]{\color{textcolor}\rmfamily\fontsize{8.000000}{9.600000}\selectfont 50\%}%
\end{pgfscope}%
\begin{pgfscope}%
\pgfpathrectangle{\pgfqpoint{0.520798in}{0.442177in}}{\pgfqpoint{0.719005in}{0.871884in}}%
\pgfusepath{clip}%
\pgfsetroundcap%
\pgfsetroundjoin%
\pgfsetlinewidth{0.501875pt}%
\definecolor{currentstroke}{rgb}{0.800000,0.800000,0.800000}%
\pgfsetstrokecolor{currentstroke}%
\pgfsetdash{}{0pt}%
\pgfpathmoveto{\pgfqpoint{0.520798in}{1.096090in}}%
\pgfpathlineto{\pgfqpoint{1.239803in}{1.096090in}}%
\pgfusepath{stroke}%
\end{pgfscope}%
\begin{pgfscope}%
\definecolor{textcolor}{rgb}{0.150000,0.150000,0.150000}%
\pgfsetstrokecolor{textcolor}%
\pgfsetfillcolor{textcolor}%
\pgftext[x=0.214138in, y=1.057828in, left, base]{\color{textcolor}\rmfamily\fontsize{8.000000}{9.600000}\selectfont 75\%}%
\end{pgfscope}%
\begin{pgfscope}%
\pgfpathrectangle{\pgfqpoint{0.520798in}{0.442177in}}{\pgfqpoint{0.719005in}{0.871884in}}%
\pgfusepath{clip}%
\pgfsetroundcap%
\pgfsetroundjoin%
\pgfsetlinewidth{0.501875pt}%
\definecolor{currentstroke}{rgb}{0.800000,0.800000,0.800000}%
\pgfsetstrokecolor{currentstroke}%
\pgfsetdash{}{0pt}%
\pgfpathmoveto{\pgfqpoint{0.520798in}{1.314061in}}%
\pgfpathlineto{\pgfqpoint{1.239803in}{1.314061in}}%
\pgfusepath{stroke}%
\end{pgfscope}%
\begin{pgfscope}%
\definecolor{textcolor}{rgb}{0.150000,0.150000,0.150000}%
\pgfsetstrokecolor{textcolor}%
\pgfsetfillcolor{textcolor}%
\pgftext[x=0.155124in, y=1.275799in, left, base]{\color{textcolor}\rmfamily\fontsize{8.000000}{9.600000}\selectfont 100\%}%
\end{pgfscope}%
\begin{pgfscope}%
\definecolor{textcolor}{rgb}{0.150000,0.150000,0.150000}%
\pgfsetstrokecolor{textcolor}%
\pgfsetfillcolor{textcolor}%
\pgftext[x=0.099569in,y=0.878119in,,bottom,rotate=90.000000]{\color{textcolor}\rmfamily\fontsize{10.000000}{12.000000}\selectfont Cert. Acc.}%
\end{pgfscope}%
\begin{pgfscope}%
\pgfsetrectcap%
\pgfsetmiterjoin%
\pgfsetlinewidth{0.752812pt}%
\definecolor{currentstroke}{rgb}{0.700000,0.700000,0.700000}%
\pgfsetstrokecolor{currentstroke}%
\pgfsetdash{}{0pt}%
\pgfpathmoveto{\pgfqpoint{0.520798in}{0.442177in}}%
\pgfpathlineto{\pgfqpoint{0.520798in}{1.314061in}}%
\pgfusepath{stroke}%
\end{pgfscope}%
\begin{pgfscope}%
\pgfsetrectcap%
\pgfsetmiterjoin%
\pgfsetlinewidth{0.752812pt}%
\definecolor{currentstroke}{rgb}{0.700000,0.700000,0.700000}%
\pgfsetstrokecolor{currentstroke}%
\pgfsetdash{}{0pt}%
\pgfpathmoveto{\pgfqpoint{1.239803in}{0.442177in}}%
\pgfpathlineto{\pgfqpoint{1.239803in}{1.314061in}}%
\pgfusepath{stroke}%
\end{pgfscope}%
\begin{pgfscope}%
\pgfsetrectcap%
\pgfsetmiterjoin%
\pgfsetlinewidth{0.752812pt}%
\definecolor{currentstroke}{rgb}{0.700000,0.700000,0.700000}%
\pgfsetstrokecolor{currentstroke}%
\pgfsetdash{}{0pt}%
\pgfpathmoveto{\pgfqpoint{0.520798in}{0.442177in}}%
\pgfpathlineto{\pgfqpoint{1.239803in}{0.442177in}}%
\pgfusepath{stroke}%
\end{pgfscope}%
\begin{pgfscope}%
\pgfsetrectcap%
\pgfsetmiterjoin%
\pgfsetlinewidth{0.752812pt}%
\definecolor{currentstroke}{rgb}{0.700000,0.700000,0.700000}%
\pgfsetstrokecolor{currentstroke}%
\pgfsetdash{}{0pt}%
\pgfpathmoveto{\pgfqpoint{0.520798in}{1.314061in}}%
\pgfpathlineto{\pgfqpoint{1.239803in}{1.314061in}}%
\pgfusepath{stroke}%
\end{pgfscope}%
\begin{pgfscope}%
\pgfpathrectangle{\pgfqpoint{0.520798in}{0.442177in}}{\pgfqpoint{0.719005in}{0.871884in}}%
\pgfusepath{clip}%
\pgfsetroundcap%
\pgfsetroundjoin%
\pgfsetlinewidth{1.003750pt}%
\definecolor{currentstroke}{rgb}{0.003922,0.450980,0.698039}%
\pgfsetstrokecolor{currentstroke}%
\pgfsetdash{}{0pt}%
\pgfpathmoveto{\pgfqpoint{0.520798in}{1.059638in}}%
\pgfpathlineto{\pgfqpoint{0.556748in}{1.059638in}}%
\pgfpathlineto{\pgfqpoint{0.556748in}{0.994989in}}%
\pgfpathlineto{\pgfqpoint{0.628649in}{0.994989in}}%
\pgfpathlineto{\pgfqpoint{0.628649in}{0.902235in}}%
\pgfpathlineto{\pgfqpoint{0.700549in}{0.902235in}}%
\pgfpathlineto{\pgfqpoint{0.700549in}{0.690293in}}%
\pgfpathlineto{\pgfqpoint{0.772450in}{0.690293in}}%
\pgfpathlineto{\pgfqpoint{0.772450in}{0.595962in}}%
\pgfpathlineto{\pgfqpoint{0.844350in}{0.595962in}}%
\pgfpathlineto{\pgfqpoint{0.844350in}{0.586873in}}%
\pgfpathlineto{\pgfqpoint{0.916251in}{0.586873in}}%
\pgfpathlineto{\pgfqpoint{0.916251in}{0.580936in}}%
\pgfpathlineto{\pgfqpoint{0.988151in}{0.580936in}}%
\pgfpathlineto{\pgfqpoint{0.988151in}{0.570084in}}%
\pgfpathlineto{\pgfqpoint{1.060052in}{0.570084in}}%
\pgfpathlineto{\pgfqpoint{1.060052in}{0.532333in}}%
\pgfpathlineto{\pgfqpoint{1.131952in}{0.532333in}}%
\pgfpathlineto{\pgfqpoint{1.131952in}{0.525748in}}%
\pgfpathlineto{\pgfqpoint{1.203853in}{0.525748in}}%
\pgfpathlineto{\pgfqpoint{1.203853in}{0.511835in}}%
\pgfpathlineto{\pgfqpoint{1.241470in}{0.511835in}}%
\pgfusepath{stroke}%
\end{pgfscope}%
\begin{pgfscope}%
\pgfpathrectangle{\pgfqpoint{0.520798in}{0.442177in}}{\pgfqpoint{0.719005in}{0.871884in}}%
\pgfusepath{clip}%
\pgfsetbuttcap%
\pgfsetroundjoin%
\definecolor{currentfill}{rgb}{0.003922,0.450980,0.698039}%
\pgfsetfillcolor{currentfill}%
\pgfsetfillopacity{0.500000}%
\pgfsetlinewidth{0.000000pt}%
\definecolor{currentstroke}{rgb}{0.003922,0.450980,0.698039}%
\pgfsetstrokecolor{currentstroke}%
\pgfsetstrokeopacity{0.500000}%
\pgfsetdash{}{0pt}%
\pgfpathmoveto{\pgfqpoint{0.520798in}{1.075224in}}%
\pgfpathlineto{\pgfqpoint{0.520798in}{1.044052in}}%
\pgfpathlineto{\pgfqpoint{0.556748in}{1.044052in}}%
\pgfpathlineto{\pgfqpoint{0.556748in}{0.975845in}}%
\pgfpathlineto{\pgfqpoint{0.628649in}{0.975845in}}%
\pgfpathlineto{\pgfqpoint{0.628649in}{0.882557in}}%
\pgfpathlineto{\pgfqpoint{0.700549in}{0.882557in}}%
\pgfpathlineto{\pgfqpoint{0.700549in}{0.674081in}}%
\pgfpathlineto{\pgfqpoint{0.772450in}{0.674081in}}%
\pgfpathlineto{\pgfqpoint{0.772450in}{0.582632in}}%
\pgfpathlineto{\pgfqpoint{0.844350in}{0.582632in}}%
\pgfpathlineto{\pgfqpoint{0.844350in}{0.573880in}}%
\pgfpathlineto{\pgfqpoint{0.916251in}{0.573880in}}%
\pgfpathlineto{\pgfqpoint{0.916251in}{0.567716in}}%
\pgfpathlineto{\pgfqpoint{0.988151in}{0.567716in}}%
\pgfpathlineto{\pgfqpoint{0.988151in}{0.557587in}}%
\pgfpathlineto{\pgfqpoint{1.060052in}{0.557587in}}%
\pgfpathlineto{\pgfqpoint{1.060052in}{0.526441in}}%
\pgfpathlineto{\pgfqpoint{1.131952in}{0.526441in}}%
\pgfpathlineto{\pgfqpoint{1.131952in}{0.520234in}}%
\pgfpathlineto{\pgfqpoint{1.203853in}{0.520234in}}%
\pgfpathlineto{\pgfqpoint{1.203853in}{0.507570in}}%
\pgfpathlineto{\pgfqpoint{1.275753in}{0.507570in}}%
\pgfpathlineto{\pgfqpoint{1.275753in}{0.494265in}}%
\pgfpathlineto{\pgfqpoint{1.347654in}{0.494265in}}%
\pgfpathlineto{\pgfqpoint{1.347654in}{0.489315in}}%
\pgfpathlineto{\pgfqpoint{1.419554in}{0.489315in}}%
\pgfpathlineto{\pgfqpoint{1.419554in}{0.442177in}}%
\pgfpathlineto{\pgfqpoint{2.066659in}{0.442177in}}%
\pgfpathlineto{\pgfqpoint{2.066659in}{0.442177in}}%
\pgfpathlineto{\pgfqpoint{2.677813in}{0.442177in}}%
\pgfpathlineto{\pgfqpoint{2.677813in}{0.442177in}}%
\pgfpathlineto{\pgfqpoint{2.677813in}{0.442177in}}%
\pgfpathlineto{\pgfqpoint{2.066659in}{0.442177in}}%
\pgfpathlineto{\pgfqpoint{2.066659in}{0.442177in}}%
\pgfpathlineto{\pgfqpoint{1.419554in}{0.442177in}}%
\pgfpathlineto{\pgfqpoint{1.419554in}{0.493728in}}%
\pgfpathlineto{\pgfqpoint{1.347654in}{0.493728in}}%
\pgfpathlineto{\pgfqpoint{1.347654in}{0.501950in}}%
\pgfpathlineto{\pgfqpoint{1.275753in}{0.501950in}}%
\pgfpathlineto{\pgfqpoint{1.275753in}{0.516099in}}%
\pgfpathlineto{\pgfqpoint{1.203853in}{0.516099in}}%
\pgfpathlineto{\pgfqpoint{1.203853in}{0.531262in}}%
\pgfpathlineto{\pgfqpoint{1.131952in}{0.531262in}}%
\pgfpathlineto{\pgfqpoint{1.131952in}{0.538226in}}%
\pgfpathlineto{\pgfqpoint{1.060052in}{0.538226in}}%
\pgfpathlineto{\pgfqpoint{1.060052in}{0.582581in}}%
\pgfpathlineto{\pgfqpoint{0.988151in}{0.582581in}}%
\pgfpathlineto{\pgfqpoint{0.988151in}{0.594156in}}%
\pgfpathlineto{\pgfqpoint{0.916251in}{0.594156in}}%
\pgfpathlineto{\pgfqpoint{0.916251in}{0.599865in}}%
\pgfpathlineto{\pgfqpoint{0.844350in}{0.599865in}}%
\pgfpathlineto{\pgfqpoint{0.844350in}{0.609293in}}%
\pgfpathlineto{\pgfqpoint{0.772450in}{0.609293in}}%
\pgfpathlineto{\pgfqpoint{0.772450in}{0.706505in}}%
\pgfpathlineto{\pgfqpoint{0.700549in}{0.706505in}}%
\pgfpathlineto{\pgfqpoint{0.700549in}{0.921913in}}%
\pgfpathlineto{\pgfqpoint{0.628649in}{0.921913in}}%
\pgfpathlineto{\pgfqpoint{0.628649in}{1.014133in}}%
\pgfpathlineto{\pgfqpoint{0.556748in}{1.014133in}}%
\pgfpathlineto{\pgfqpoint{0.556748in}{1.075224in}}%
\pgfpathlineto{\pgfqpoint{0.520798in}{1.075224in}}%
\pgfpathlineto{\pgfqpoint{0.520798in}{1.075224in}}%
\pgfpathclose%
\pgfusepath{fill}%
\end{pgfscope}%
\begin{pgfscope}%
\pgfpathrectangle{\pgfqpoint{0.520798in}{0.442177in}}{\pgfqpoint{0.719005in}{0.871884in}}%
\pgfusepath{clip}%
\pgfsetroundcap%
\pgfsetroundjoin%
\pgfsetlinewidth{1.003750pt}%
\definecolor{currentstroke}{rgb}{0.870588,0.560784,0.019608}%
\pgfsetstrokecolor{currentstroke}%
\pgfsetdash{}{0pt}%
\pgfpathmoveto{\pgfqpoint{0.520798in}{1.059638in}}%
\pgfpathlineto{\pgfqpoint{0.556748in}{1.059638in}}%
\pgfpathlineto{\pgfqpoint{0.556748in}{0.994989in}}%
\pgfpathlineto{\pgfqpoint{0.628649in}{0.994989in}}%
\pgfpathlineto{\pgfqpoint{0.628649in}{0.902235in}}%
\pgfpathlineto{\pgfqpoint{0.700549in}{0.902235in}}%
\pgfpathlineto{\pgfqpoint{0.700549in}{0.690293in}}%
\pgfpathlineto{\pgfqpoint{0.772450in}{0.690293in}}%
\pgfpathlineto{\pgfqpoint{0.772450in}{0.595962in}}%
\pgfpathlineto{\pgfqpoint{0.844350in}{0.595962in}}%
\pgfpathlineto{\pgfqpoint{0.844350in}{0.589748in}}%
\pgfpathlineto{\pgfqpoint{0.916251in}{0.589748in}}%
\pgfpathlineto{\pgfqpoint{0.916251in}{0.583441in}}%
\pgfpathlineto{\pgfqpoint{0.988151in}{0.583441in}}%
\pgfpathlineto{\pgfqpoint{0.988151in}{0.570084in}}%
\pgfpathlineto{\pgfqpoint{1.060052in}{0.570084in}}%
\pgfpathlineto{\pgfqpoint{1.060052in}{0.532426in}}%
\pgfpathlineto{\pgfqpoint{1.131952in}{0.532426in}}%
\pgfpathlineto{\pgfqpoint{1.131952in}{0.525933in}}%
\pgfpathlineto{\pgfqpoint{1.203853in}{0.525933in}}%
\pgfpathlineto{\pgfqpoint{1.203853in}{0.512020in}}%
\pgfpathlineto{\pgfqpoint{1.241470in}{0.512020in}}%
\pgfusepath{stroke}%
\end{pgfscope}%
\begin{pgfscope}%
\pgfpathrectangle{\pgfqpoint{0.520798in}{0.442177in}}{\pgfqpoint{0.719005in}{0.871884in}}%
\pgfusepath{clip}%
\pgfsetbuttcap%
\pgfsetroundjoin%
\definecolor{currentfill}{rgb}{0.870588,0.560784,0.019608}%
\pgfsetfillcolor{currentfill}%
\pgfsetfillopacity{0.500000}%
\pgfsetlinewidth{0.000000pt}%
\definecolor{currentstroke}{rgb}{0.870588,0.560784,0.019608}%
\pgfsetstrokecolor{currentstroke}%
\pgfsetstrokeopacity{0.500000}%
\pgfsetdash{}{0pt}%
\pgfpathmoveto{\pgfqpoint{0.520798in}{1.075224in}}%
\pgfpathlineto{\pgfqpoint{0.520798in}{1.044052in}}%
\pgfpathlineto{\pgfqpoint{0.556748in}{1.044052in}}%
\pgfpathlineto{\pgfqpoint{0.556748in}{0.975845in}}%
\pgfpathlineto{\pgfqpoint{0.628649in}{0.975845in}}%
\pgfpathlineto{\pgfqpoint{0.628649in}{0.882557in}}%
\pgfpathlineto{\pgfqpoint{0.700549in}{0.882557in}}%
\pgfpathlineto{\pgfqpoint{0.700549in}{0.674081in}}%
\pgfpathlineto{\pgfqpoint{0.772450in}{0.674081in}}%
\pgfpathlineto{\pgfqpoint{0.772450in}{0.582632in}}%
\pgfpathlineto{\pgfqpoint{0.844350in}{0.582632in}}%
\pgfpathlineto{\pgfqpoint{0.844350in}{0.577073in}}%
\pgfpathlineto{\pgfqpoint{0.916251in}{0.577073in}}%
\pgfpathlineto{\pgfqpoint{0.916251in}{0.570707in}}%
\pgfpathlineto{\pgfqpoint{0.988151in}{0.570707in}}%
\pgfpathlineto{\pgfqpoint{0.988151in}{0.557587in}}%
\pgfpathlineto{\pgfqpoint{1.060052in}{0.557587in}}%
\pgfpathlineto{\pgfqpoint{1.060052in}{0.526600in}}%
\pgfpathlineto{\pgfqpoint{1.131952in}{0.526600in}}%
\pgfpathlineto{\pgfqpoint{1.131952in}{0.520552in}}%
\pgfpathlineto{\pgfqpoint{1.203853in}{0.520552in}}%
\pgfpathlineto{\pgfqpoint{1.203853in}{0.507734in}}%
\pgfpathlineto{\pgfqpoint{1.275753in}{0.507734in}}%
\pgfpathlineto{\pgfqpoint{1.275753in}{0.494565in}}%
\pgfpathlineto{\pgfqpoint{1.347654in}{0.494565in}}%
\pgfpathlineto{\pgfqpoint{1.347654in}{0.489315in}}%
\pgfpathlineto{\pgfqpoint{1.419554in}{0.489315in}}%
\pgfpathlineto{\pgfqpoint{1.419554in}{0.442177in}}%
\pgfpathlineto{\pgfqpoint{2.066659in}{0.442177in}}%
\pgfpathlineto{\pgfqpoint{2.066659in}{0.442177in}}%
\pgfpathlineto{\pgfqpoint{2.677813in}{0.442177in}}%
\pgfpathlineto{\pgfqpoint{2.677813in}{0.442177in}}%
\pgfpathlineto{\pgfqpoint{2.677813in}{0.442177in}}%
\pgfpathlineto{\pgfqpoint{2.066659in}{0.442177in}}%
\pgfpathlineto{\pgfqpoint{2.066659in}{0.442177in}}%
\pgfpathlineto{\pgfqpoint{1.419554in}{0.442177in}}%
\pgfpathlineto{\pgfqpoint{1.419554in}{0.493728in}}%
\pgfpathlineto{\pgfqpoint{1.347654in}{0.493728in}}%
\pgfpathlineto{\pgfqpoint{1.347654in}{0.503134in}}%
\pgfpathlineto{\pgfqpoint{1.275753in}{0.503134in}}%
\pgfpathlineto{\pgfqpoint{1.275753in}{0.516307in}}%
\pgfpathlineto{\pgfqpoint{1.203853in}{0.516307in}}%
\pgfpathlineto{\pgfqpoint{1.203853in}{0.531315in}}%
\pgfpathlineto{\pgfqpoint{1.131952in}{0.531315in}}%
\pgfpathlineto{\pgfqpoint{1.131952in}{0.538253in}}%
\pgfpathlineto{\pgfqpoint{1.060052in}{0.538253in}}%
\pgfpathlineto{\pgfqpoint{1.060052in}{0.582581in}}%
\pgfpathlineto{\pgfqpoint{0.988151in}{0.582581in}}%
\pgfpathlineto{\pgfqpoint{0.988151in}{0.596174in}}%
\pgfpathlineto{\pgfqpoint{0.916251in}{0.596174in}}%
\pgfpathlineto{\pgfqpoint{0.916251in}{0.602423in}}%
\pgfpathlineto{\pgfqpoint{0.844350in}{0.602423in}}%
\pgfpathlineto{\pgfqpoint{0.844350in}{0.609293in}}%
\pgfpathlineto{\pgfqpoint{0.772450in}{0.609293in}}%
\pgfpathlineto{\pgfqpoint{0.772450in}{0.706505in}}%
\pgfpathlineto{\pgfqpoint{0.700549in}{0.706505in}}%
\pgfpathlineto{\pgfqpoint{0.700549in}{0.921913in}}%
\pgfpathlineto{\pgfqpoint{0.628649in}{0.921913in}}%
\pgfpathlineto{\pgfqpoint{0.628649in}{1.014133in}}%
\pgfpathlineto{\pgfqpoint{0.556748in}{1.014133in}}%
\pgfpathlineto{\pgfqpoint{0.556748in}{1.075224in}}%
\pgfpathlineto{\pgfqpoint{0.520798in}{1.075224in}}%
\pgfpathlineto{\pgfqpoint{0.520798in}{1.075224in}}%
\pgfpathclose%
\pgfusepath{fill}%
\end{pgfscope}%
\begin{pgfscope}%
\pgfpathrectangle{\pgfqpoint{0.520798in}{0.442177in}}{\pgfqpoint{0.719005in}{0.871884in}}%
\pgfusepath{clip}%
\pgfsetroundcap%
\pgfsetroundjoin%
\pgfsetlinewidth{1.003750pt}%
\definecolor{currentstroke}{rgb}{0.007843,0.619608,0.450980}%
\pgfsetstrokecolor{currentstroke}%
\pgfsetdash{}{0pt}%
\pgfpathmoveto{\pgfqpoint{0.520798in}{1.059638in}}%
\pgfpathlineto{\pgfqpoint{0.556748in}{1.059638in}}%
\pgfpathlineto{\pgfqpoint{0.556748in}{0.994989in}}%
\pgfpathlineto{\pgfqpoint{0.628649in}{0.994989in}}%
\pgfpathlineto{\pgfqpoint{0.628649in}{0.902235in}}%
\pgfpathlineto{\pgfqpoint{0.700549in}{0.902235in}}%
\pgfpathlineto{\pgfqpoint{0.700549in}{0.690293in}}%
\pgfpathlineto{\pgfqpoint{0.772450in}{0.690293in}}%
\pgfpathlineto{\pgfqpoint{0.772450in}{0.595962in}}%
\pgfpathlineto{\pgfqpoint{0.844350in}{0.595962in}}%
\pgfpathlineto{\pgfqpoint{0.844350in}{0.589748in}}%
\pgfpathlineto{\pgfqpoint{0.916251in}{0.589748in}}%
\pgfpathlineto{\pgfqpoint{0.916251in}{0.583441in}}%
\pgfpathlineto{\pgfqpoint{0.988151in}{0.583441in}}%
\pgfpathlineto{\pgfqpoint{0.988151in}{0.570084in}}%
\pgfpathlineto{\pgfqpoint{1.060052in}{0.570084in}}%
\pgfpathlineto{\pgfqpoint{1.060052in}{0.532426in}}%
\pgfpathlineto{\pgfqpoint{1.131952in}{0.532426in}}%
\pgfpathlineto{\pgfqpoint{1.131952in}{0.525933in}}%
\pgfpathlineto{\pgfqpoint{1.203853in}{0.525933in}}%
\pgfpathlineto{\pgfqpoint{1.203853in}{0.512020in}}%
\pgfpathlineto{\pgfqpoint{1.241470in}{0.512020in}}%
\pgfusepath{stroke}%
\end{pgfscope}%
\begin{pgfscope}%
\pgfpathrectangle{\pgfqpoint{0.520798in}{0.442177in}}{\pgfqpoint{0.719005in}{0.871884in}}%
\pgfusepath{clip}%
\pgfsetbuttcap%
\pgfsetroundjoin%
\definecolor{currentfill}{rgb}{0.007843,0.619608,0.450980}%
\pgfsetfillcolor{currentfill}%
\pgfsetfillopacity{0.500000}%
\pgfsetlinewidth{0.000000pt}%
\definecolor{currentstroke}{rgb}{0.007843,0.619608,0.450980}%
\pgfsetstrokecolor{currentstroke}%
\pgfsetstrokeopacity{0.500000}%
\pgfsetdash{}{0pt}%
\pgfpathmoveto{\pgfqpoint{0.520798in}{1.075224in}}%
\pgfpathlineto{\pgfqpoint{0.520798in}{1.044052in}}%
\pgfpathlineto{\pgfqpoint{0.556748in}{1.044052in}}%
\pgfpathlineto{\pgfqpoint{0.556748in}{0.975845in}}%
\pgfpathlineto{\pgfqpoint{0.628649in}{0.975845in}}%
\pgfpathlineto{\pgfqpoint{0.628649in}{0.882557in}}%
\pgfpathlineto{\pgfqpoint{0.700549in}{0.882557in}}%
\pgfpathlineto{\pgfqpoint{0.700549in}{0.674081in}}%
\pgfpathlineto{\pgfqpoint{0.772450in}{0.674081in}}%
\pgfpathlineto{\pgfqpoint{0.772450in}{0.582632in}}%
\pgfpathlineto{\pgfqpoint{0.844350in}{0.582632in}}%
\pgfpathlineto{\pgfqpoint{0.844350in}{0.577073in}}%
\pgfpathlineto{\pgfqpoint{0.916251in}{0.577073in}}%
\pgfpathlineto{\pgfqpoint{0.916251in}{0.570707in}}%
\pgfpathlineto{\pgfqpoint{0.988151in}{0.570707in}}%
\pgfpathlineto{\pgfqpoint{0.988151in}{0.557587in}}%
\pgfpathlineto{\pgfqpoint{1.060052in}{0.557587in}}%
\pgfpathlineto{\pgfqpoint{1.060052in}{0.526600in}}%
\pgfpathlineto{\pgfqpoint{1.131952in}{0.526600in}}%
\pgfpathlineto{\pgfqpoint{1.131952in}{0.520552in}}%
\pgfpathlineto{\pgfqpoint{1.203853in}{0.520552in}}%
\pgfpathlineto{\pgfqpoint{1.203853in}{0.507734in}}%
\pgfpathlineto{\pgfqpoint{1.275753in}{0.507734in}}%
\pgfpathlineto{\pgfqpoint{1.275753in}{0.494565in}}%
\pgfpathlineto{\pgfqpoint{1.347654in}{0.494565in}}%
\pgfpathlineto{\pgfqpoint{1.347654in}{0.489315in}}%
\pgfpathlineto{\pgfqpoint{1.419554in}{0.489315in}}%
\pgfpathlineto{\pgfqpoint{1.419554in}{0.442177in}}%
\pgfpathlineto{\pgfqpoint{2.066659in}{0.442177in}}%
\pgfpathlineto{\pgfqpoint{2.066659in}{0.442177in}}%
\pgfpathlineto{\pgfqpoint{2.677813in}{0.442177in}}%
\pgfpathlineto{\pgfqpoint{2.677813in}{0.442177in}}%
\pgfpathlineto{\pgfqpoint{2.677813in}{0.442177in}}%
\pgfpathlineto{\pgfqpoint{2.066659in}{0.442177in}}%
\pgfpathlineto{\pgfqpoint{2.066659in}{0.442177in}}%
\pgfpathlineto{\pgfqpoint{1.419554in}{0.442177in}}%
\pgfpathlineto{\pgfqpoint{1.419554in}{0.493728in}}%
\pgfpathlineto{\pgfqpoint{1.347654in}{0.493728in}}%
\pgfpathlineto{\pgfqpoint{1.347654in}{0.503134in}}%
\pgfpathlineto{\pgfqpoint{1.275753in}{0.503134in}}%
\pgfpathlineto{\pgfqpoint{1.275753in}{0.516307in}}%
\pgfpathlineto{\pgfqpoint{1.203853in}{0.516307in}}%
\pgfpathlineto{\pgfqpoint{1.203853in}{0.531315in}}%
\pgfpathlineto{\pgfqpoint{1.131952in}{0.531315in}}%
\pgfpathlineto{\pgfqpoint{1.131952in}{0.538253in}}%
\pgfpathlineto{\pgfqpoint{1.060052in}{0.538253in}}%
\pgfpathlineto{\pgfqpoint{1.060052in}{0.582581in}}%
\pgfpathlineto{\pgfqpoint{0.988151in}{0.582581in}}%
\pgfpathlineto{\pgfqpoint{0.988151in}{0.596174in}}%
\pgfpathlineto{\pgfqpoint{0.916251in}{0.596174in}}%
\pgfpathlineto{\pgfqpoint{0.916251in}{0.602423in}}%
\pgfpathlineto{\pgfqpoint{0.844350in}{0.602423in}}%
\pgfpathlineto{\pgfqpoint{0.844350in}{0.609293in}}%
\pgfpathlineto{\pgfqpoint{0.772450in}{0.609293in}}%
\pgfpathlineto{\pgfqpoint{0.772450in}{0.706505in}}%
\pgfpathlineto{\pgfqpoint{0.700549in}{0.706505in}}%
\pgfpathlineto{\pgfqpoint{0.700549in}{0.921913in}}%
\pgfpathlineto{\pgfqpoint{0.628649in}{0.921913in}}%
\pgfpathlineto{\pgfqpoint{0.628649in}{1.014133in}}%
\pgfpathlineto{\pgfqpoint{0.556748in}{1.014133in}}%
\pgfpathlineto{\pgfqpoint{0.556748in}{1.075224in}}%
\pgfpathlineto{\pgfqpoint{0.520798in}{1.075224in}}%
\pgfpathlineto{\pgfqpoint{0.520798in}{1.075224in}}%
\pgfpathclose%
\pgfusepath{fill}%
\end{pgfscope}%
\begin{pgfscope}%
\definecolor{textcolor}{rgb}{0.150000,0.150000,0.150000}%
\pgfsetstrokecolor{textcolor}%
\pgfsetfillcolor{textcolor}%
\pgftext[x=0.880300in,y=1.397395in,,base]{\color{textcolor}\rmfamily\fontsize{9.000000}{10.800000}\selectfont \(\displaystyle c_X^+=1\)}%
\end{pgfscope}%
\begin{pgfscope}%
\pgfsetbuttcap%
\pgfsetmiterjoin%
\definecolor{currentfill}{rgb}{1.000000,1.000000,1.000000}%
\pgfsetfillcolor{currentfill}%
\pgfsetfillopacity{0.800000}%
\pgfsetlinewidth{1.003750pt}%
\definecolor{currentstroke}{rgb}{0.800000,0.800000,0.800000}%
\pgfsetstrokecolor{currentstroke}%
\pgfsetstrokeopacity{0.800000}%
\pgfsetdash{}{0pt}%
\pgfpathmoveto{\pgfqpoint{0.785843in}{0.638103in}}%
\pgfpathlineto{\pgfqpoint{1.191192in}{0.638103in}}%
\pgfpathlineto{\pgfqpoint{1.191192in}{1.265450in}}%
\pgfpathlineto{\pgfqpoint{0.785843in}{1.265450in}}%
\pgfpathlineto{\pgfqpoint{0.785843in}{0.638103in}}%
\pgfpathclose%
\pgfusepath{stroke,fill}%
\end{pgfscope}%
\begin{pgfscope}%
\definecolor{textcolor}{rgb}{0.150000,0.150000,0.150000}%
\pgfsetstrokecolor{textcolor}%
\pgfsetfillcolor{textcolor}%
\pgftext[x=0.912291in,y=1.113275in,left,base]{\color{textcolor}\rmfamily\fontsize{9.000000}{10.800000}\selectfont \(\displaystyle c_X^-\)}%
\end{pgfscope}%
\begin{pgfscope}%
\pgfsetroundcap%
\pgfsetroundjoin%
\pgfsetlinewidth{1.003750pt}%
\definecolor{currentstroke}{rgb}{0.003922,0.450980,0.698039}%
\pgfsetstrokecolor{currentstroke}%
\pgfsetdash{}{0pt}%
\pgfpathmoveto{\pgfqpoint{0.824732in}{1.001052in}}%
\pgfpathlineto{\pgfqpoint{0.873343in}{1.001052in}}%
\pgfpathlineto{\pgfqpoint{0.873343in}{1.001052in}}%
\pgfpathlineto{\pgfqpoint{0.970565in}{1.001052in}}%
\pgfpathlineto{\pgfqpoint{0.970565in}{1.001052in}}%
\pgfpathlineto{\pgfqpoint{1.019176in}{1.001052in}}%
\pgfusepath{stroke}%
\end{pgfscope}%
\begin{pgfscope}%
\definecolor{textcolor}{rgb}{0.150000,0.150000,0.150000}%
\pgfsetstrokecolor{textcolor}%
\pgfsetfillcolor{textcolor}%
\pgftext[x=1.096954in,y=0.967024in,left,base]{\color{textcolor}\rmfamily\fontsize{7.000000}{8.400000}\selectfont 1}%
\end{pgfscope}%
\begin{pgfscope}%
\pgfsetroundcap%
\pgfsetroundjoin%
\pgfsetlinewidth{1.003750pt}%
\definecolor{currentstroke}{rgb}{0.870588,0.560784,0.019608}%
\pgfsetstrokecolor{currentstroke}%
\pgfsetdash{}{0pt}%
\pgfpathmoveto{\pgfqpoint{0.824732in}{0.865486in}}%
\pgfpathlineto{\pgfqpoint{0.873343in}{0.865486in}}%
\pgfpathlineto{\pgfqpoint{0.873343in}{0.865486in}}%
\pgfpathlineto{\pgfqpoint{0.970565in}{0.865486in}}%
\pgfpathlineto{\pgfqpoint{0.970565in}{0.865486in}}%
\pgfpathlineto{\pgfqpoint{1.019176in}{0.865486in}}%
\pgfusepath{stroke}%
\end{pgfscope}%
\begin{pgfscope}%
\definecolor{textcolor}{rgb}{0.150000,0.150000,0.150000}%
\pgfsetstrokecolor{textcolor}%
\pgfsetfillcolor{textcolor}%
\pgftext[x=1.096954in,y=0.831458in,left,base]{\color{textcolor}\rmfamily\fontsize{7.000000}{8.400000}\selectfont 2}%
\end{pgfscope}%
\begin{pgfscope}%
\pgfsetroundcap%
\pgfsetroundjoin%
\pgfsetlinewidth{1.003750pt}%
\definecolor{currentstroke}{rgb}{0.007843,0.619608,0.450980}%
\pgfsetstrokecolor{currentstroke}%
\pgfsetdash{}{0pt}%
\pgfpathmoveto{\pgfqpoint{0.824732in}{0.729919in}}%
\pgfpathlineto{\pgfqpoint{0.873343in}{0.729919in}}%
\pgfpathlineto{\pgfqpoint{0.873343in}{0.729919in}}%
\pgfpathlineto{\pgfqpoint{0.970565in}{0.729919in}}%
\pgfpathlineto{\pgfqpoint{0.970565in}{0.729919in}}%
\pgfpathlineto{\pgfqpoint{1.019176in}{0.729919in}}%
\pgfusepath{stroke}%
\end{pgfscope}%
\begin{pgfscope}%
\definecolor{textcolor}{rgb}{0.150000,0.150000,0.150000}%
\pgfsetstrokecolor{textcolor}%
\pgfsetfillcolor{textcolor}%
\pgftext[x=1.096954in,y=0.695892in,left,base]{\color{textcolor}\rmfamily\fontsize{7.000000}{8.400000}\selectfont 4}%
\end{pgfscope}%
\end{pgfpicture}%
\makeatother%
\endgroup%

%% file: figures/experiments/graphs/sparse_smoothing/nodes_structure/node_classification-Citeseer-GCN-hidden=32-p_adj_plus=0.001-p_adj_minus=0.8-p_att_plus=0.0-p_att_minus=0.0-multi_class_cert-B.pgf
\begingroup%
\makeatletter%
\begin{pgfpicture}%
\pgfpathrectangle{\pgfpointorigin}{\pgfqpoint{1.375000in}{1.581250in}}%
\pgfusepath{use as bounding box, clip}%
\begin{pgfscope}%
\pgfsetbuttcap%
\pgfsetmiterjoin%
\definecolor{currentfill}{rgb}{1.000000,1.000000,1.000000}%
\pgfsetfillcolor{currentfill}%
\pgfsetlinewidth{0.000000pt}%
\definecolor{currentstroke}{rgb}{1.000000,1.000000,1.000000}%
\pgfsetstrokecolor{currentstroke}%
\pgfsetdash{}{0pt}%
\pgfpathmoveto{\pgfqpoint{0.000000in}{0.000000in}}%
\pgfpathlineto{\pgfqpoint{1.375000in}{0.000000in}}%
\pgfpathlineto{\pgfqpoint{1.375000in}{1.581250in}}%
\pgfpathlineto{\pgfqpoint{0.000000in}{1.581250in}}%
\pgfpathlineto{\pgfqpoint{0.000000in}{0.000000in}}%
\pgfpathclose%
\pgfusepath{fill}%
\end{pgfscope}%
\begin{pgfscope}%
\pgfsetbuttcap%
\pgfsetmiterjoin%
\definecolor{currentfill}{rgb}{1.000000,1.000000,1.000000}%
\pgfsetfillcolor{currentfill}%
\pgfsetlinewidth{0.000000pt}%
\definecolor{currentstroke}{rgb}{0.000000,0.000000,0.000000}%
\pgfsetstrokecolor{currentstroke}%
\pgfsetstrokeopacity{0.000000}%
\pgfsetdash{}{0pt}%
\pgfpathmoveto{\pgfqpoint{0.520798in}{0.442177in}}%
\pgfpathlineto{\pgfqpoint{1.239803in}{0.442177in}}%
\pgfpathlineto{\pgfqpoint{1.239803in}{1.314061in}}%
\pgfpathlineto{\pgfqpoint{0.520798in}{1.314061in}}%
\pgfpathlineto{\pgfqpoint{0.520798in}{0.442177in}}%
\pgfpathclose%
\pgfusepath{fill}%
\end{pgfscope}%
\begin{pgfscope}%
\pgfpathrectangle{\pgfqpoint{0.520798in}{0.442177in}}{\pgfqpoint{0.719005in}{0.871884in}}%
\pgfusepath{clip}%
\pgfsetroundcap%
\pgfsetroundjoin%
\pgfsetlinewidth{0.501875pt}%
\definecolor{currentstroke}{rgb}{0.800000,0.800000,0.800000}%
\pgfsetstrokecolor{currentstroke}%
\pgfsetdash{}{0pt}%
\pgfpathmoveto{\pgfqpoint{0.520798in}{0.442177in}}%
\pgfpathlineto{\pgfqpoint{0.520798in}{1.314061in}}%
\pgfusepath{stroke}%
\end{pgfscope}%
\begin{pgfscope}%
\definecolor{textcolor}{rgb}{0.150000,0.150000,0.150000}%
\pgfsetstrokecolor{textcolor}%
\pgfsetfillcolor{textcolor}%
\pgftext[x=0.520798in,y=0.351899in,,top]{\color{textcolor}\rmfamily\fontsize{8.000000}{9.600000}\selectfont \(\displaystyle {0}\)}%
\end{pgfscope}%
\begin{pgfscope}%
\pgfpathrectangle{\pgfqpoint{0.520798in}{0.442177in}}{\pgfqpoint{0.719005in}{0.871884in}}%
\pgfusepath{clip}%
\pgfsetroundcap%
\pgfsetroundjoin%
\pgfsetlinewidth{0.501875pt}%
\definecolor{currentstroke}{rgb}{0.800000,0.800000,0.800000}%
\pgfsetstrokecolor{currentstroke}%
\pgfsetdash{}{0pt}%
\pgfpathmoveto{\pgfqpoint{0.880300in}{0.442177in}}%
\pgfpathlineto{\pgfqpoint{0.880300in}{1.314061in}}%
\pgfusepath{stroke}%
\end{pgfscope}%
\begin{pgfscope}%
\definecolor{textcolor}{rgb}{0.150000,0.150000,0.150000}%
\pgfsetstrokecolor{textcolor}%
\pgfsetfillcolor{textcolor}%
\pgftext[x=0.880300in,y=0.351899in,,top]{\color{textcolor}\rmfamily\fontsize{8.000000}{9.600000}\selectfont \(\displaystyle {5}\)}%
\end{pgfscope}%
\begin{pgfscope}%
\pgfpathrectangle{\pgfqpoint{0.520798in}{0.442177in}}{\pgfqpoint{0.719005in}{0.871884in}}%
\pgfusepath{clip}%
\pgfsetroundcap%
\pgfsetroundjoin%
\pgfsetlinewidth{0.501875pt}%
\definecolor{currentstroke}{rgb}{0.800000,0.800000,0.800000}%
\pgfsetstrokecolor{currentstroke}%
\pgfsetdash{}{0pt}%
\pgfpathmoveto{\pgfqpoint{1.239803in}{0.442177in}}%
\pgfpathlineto{\pgfqpoint{1.239803in}{1.314061in}}%
\pgfusepath{stroke}%
\end{pgfscope}%
\begin{pgfscope}%
\definecolor{textcolor}{rgb}{0.150000,0.150000,0.150000}%
\pgfsetstrokecolor{textcolor}%
\pgfsetfillcolor{textcolor}%
\pgftext[x=1.239803in,y=0.351899in,,top]{\color{textcolor}\rmfamily\fontsize{8.000000}{9.600000}\selectfont \(\displaystyle {10}\)}%
\end{pgfscope}%
\begin{pgfscope}%
\definecolor{textcolor}{rgb}{0.150000,0.150000,0.150000}%
\pgfsetstrokecolor{textcolor}%
\pgfsetfillcolor{textcolor}%
\pgftext[x=0.880300in,y=0.198219in,,top]{\color{textcolor}\rmfamily\fontsize{10.000000}{12.000000}\selectfont Edit distance \(\displaystyle \epsilon\)}%
\end{pgfscope}%
\begin{pgfscope}%
\pgfpathrectangle{\pgfqpoint{0.520798in}{0.442177in}}{\pgfqpoint{0.719005in}{0.871884in}}%
\pgfusepath{clip}%
\pgfsetroundcap%
\pgfsetroundjoin%
\pgfsetlinewidth{0.501875pt}%
\definecolor{currentstroke}{rgb}{0.800000,0.800000,0.800000}%
\pgfsetstrokecolor{currentstroke}%
\pgfsetdash{}{0pt}%
\pgfpathmoveto{\pgfqpoint{0.520798in}{0.442177in}}%
\pgfpathlineto{\pgfqpoint{1.239803in}{0.442177in}}%
\pgfusepath{stroke}%
\end{pgfscope}%
\begin{pgfscope}%
\definecolor{textcolor}{rgb}{0.150000,0.150000,0.150000}%
\pgfsetstrokecolor{textcolor}%
\pgfsetfillcolor{textcolor}%
\pgftext[x=0.273151in, y=0.403915in, left, base]{\color{textcolor}\rmfamily\fontsize{8.000000}{9.600000}\selectfont 0\%}%
\end{pgfscope}%
\begin{pgfscope}%
\pgfpathrectangle{\pgfqpoint{0.520798in}{0.442177in}}{\pgfqpoint{0.719005in}{0.871884in}}%
\pgfusepath{clip}%
\pgfsetroundcap%
\pgfsetroundjoin%
\pgfsetlinewidth{0.501875pt}%
\definecolor{currentstroke}{rgb}{0.800000,0.800000,0.800000}%
\pgfsetstrokecolor{currentstroke}%
\pgfsetdash{}{0pt}%
\pgfpathmoveto{\pgfqpoint{0.520798in}{0.660148in}}%
\pgfpathlineto{\pgfqpoint{1.239803in}{0.660148in}}%
\pgfusepath{stroke}%
\end{pgfscope}%
\begin{pgfscope}%
\definecolor{textcolor}{rgb}{0.150000,0.150000,0.150000}%
\pgfsetstrokecolor{textcolor}%
\pgfsetfillcolor{textcolor}%
\pgftext[x=0.214138in, y=0.621886in, left, base]{\color{textcolor}\rmfamily\fontsize{8.000000}{9.600000}\selectfont 25\%}%
\end{pgfscope}%
\begin{pgfscope}%
\pgfpathrectangle{\pgfqpoint{0.520798in}{0.442177in}}{\pgfqpoint{0.719005in}{0.871884in}}%
\pgfusepath{clip}%
\pgfsetroundcap%
\pgfsetroundjoin%
\pgfsetlinewidth{0.501875pt}%
\definecolor{currentstroke}{rgb}{0.800000,0.800000,0.800000}%
\pgfsetstrokecolor{currentstroke}%
\pgfsetdash{}{0pt}%
\pgfpathmoveto{\pgfqpoint{0.520798in}{0.878119in}}%
\pgfpathlineto{\pgfqpoint{1.239803in}{0.878119in}}%
\pgfusepath{stroke}%
\end{pgfscope}%
\begin{pgfscope}%
\definecolor{textcolor}{rgb}{0.150000,0.150000,0.150000}%
\pgfsetstrokecolor{textcolor}%
\pgfsetfillcolor{textcolor}%
\pgftext[x=0.214138in, y=0.839857in, left, base]{\color{textcolor}\rmfamily\fontsize{8.000000}{9.600000}\selectfont 50\%}%
\end{pgfscope}%
\begin{pgfscope}%
\pgfpathrectangle{\pgfqpoint{0.520798in}{0.442177in}}{\pgfqpoint{0.719005in}{0.871884in}}%
\pgfusepath{clip}%
\pgfsetroundcap%
\pgfsetroundjoin%
\pgfsetlinewidth{0.501875pt}%
\definecolor{currentstroke}{rgb}{0.800000,0.800000,0.800000}%
\pgfsetstrokecolor{currentstroke}%
\pgfsetdash{}{0pt}%
\pgfpathmoveto{\pgfqpoint{0.520798in}{1.096090in}}%
\pgfpathlineto{\pgfqpoint{1.239803in}{1.096090in}}%
\pgfusepath{stroke}%
\end{pgfscope}%
\begin{pgfscope}%
\definecolor{textcolor}{rgb}{0.150000,0.150000,0.150000}%
\pgfsetstrokecolor{textcolor}%
\pgfsetfillcolor{textcolor}%
\pgftext[x=0.214138in, y=1.057828in, left, base]{\color{textcolor}\rmfamily\fontsize{8.000000}{9.600000}\selectfont 75\%}%
\end{pgfscope}%
\begin{pgfscope}%
\pgfpathrectangle{\pgfqpoint{0.520798in}{0.442177in}}{\pgfqpoint{0.719005in}{0.871884in}}%
\pgfusepath{clip}%
\pgfsetroundcap%
\pgfsetroundjoin%
\pgfsetlinewidth{0.501875pt}%
\definecolor{currentstroke}{rgb}{0.800000,0.800000,0.800000}%
\pgfsetstrokecolor{currentstroke}%
\pgfsetdash{}{0pt}%
\pgfpathmoveto{\pgfqpoint{0.520798in}{1.314061in}}%
\pgfpathlineto{\pgfqpoint{1.239803in}{1.314061in}}%
\pgfusepath{stroke}%
\end{pgfscope}%
\begin{pgfscope}%
\definecolor{textcolor}{rgb}{0.150000,0.150000,0.150000}%
\pgfsetstrokecolor{textcolor}%
\pgfsetfillcolor{textcolor}%
\pgftext[x=0.155124in, y=1.275799in, left, base]{\color{textcolor}\rmfamily\fontsize{8.000000}{9.600000}\selectfont 100\%}%
\end{pgfscope}%
\begin{pgfscope}%
\definecolor{textcolor}{rgb}{0.150000,0.150000,0.150000}%
\pgfsetstrokecolor{textcolor}%
\pgfsetfillcolor{textcolor}%
\pgftext[x=0.099569in,y=0.878119in,,bottom,rotate=90.000000]{\color{textcolor}\rmfamily\fontsize{10.000000}{12.000000}\selectfont Cert. Acc.}%
\end{pgfscope}%
\begin{pgfscope}%
\pgfsetrectcap%
\pgfsetmiterjoin%
\pgfsetlinewidth{0.752812pt}%
\definecolor{currentstroke}{rgb}{0.700000,0.700000,0.700000}%
\pgfsetstrokecolor{currentstroke}%
\pgfsetdash{}{0pt}%
\pgfpathmoveto{\pgfqpoint{0.520798in}{0.442177in}}%
\pgfpathlineto{\pgfqpoint{0.520798in}{1.314061in}}%
\pgfusepath{stroke}%
\end{pgfscope}%
\begin{pgfscope}%
\pgfsetrectcap%
\pgfsetmiterjoin%
\pgfsetlinewidth{0.752812pt}%
\definecolor{currentstroke}{rgb}{0.700000,0.700000,0.700000}%
\pgfsetstrokecolor{currentstroke}%
\pgfsetdash{}{0pt}%
\pgfpathmoveto{\pgfqpoint{1.239803in}{0.442177in}}%
\pgfpathlineto{\pgfqpoint{1.239803in}{1.314061in}}%
\pgfusepath{stroke}%
\end{pgfscope}%
\begin{pgfscope}%
\pgfsetrectcap%
\pgfsetmiterjoin%
\pgfsetlinewidth{0.752812pt}%
\definecolor{currentstroke}{rgb}{0.700000,0.700000,0.700000}%
\pgfsetstrokecolor{currentstroke}%
\pgfsetdash{}{0pt}%
\pgfpathmoveto{\pgfqpoint{0.520798in}{0.442177in}}%
\pgfpathlineto{\pgfqpoint{1.239803in}{0.442177in}}%
\pgfusepath{stroke}%
\end{pgfscope}%
\begin{pgfscope}%
\pgfsetrectcap%
\pgfsetmiterjoin%
\pgfsetlinewidth{0.752812pt}%
\definecolor{currentstroke}{rgb}{0.700000,0.700000,0.700000}%
\pgfsetstrokecolor{currentstroke}%
\pgfsetdash{}{0pt}%
\pgfpathmoveto{\pgfqpoint{0.520798in}{1.314061in}}%
\pgfpathlineto{\pgfqpoint{1.239803in}{1.314061in}}%
\pgfusepath{stroke}%
\end{pgfscope}%
\begin{pgfscope}%
\pgfpathrectangle{\pgfqpoint{0.520798in}{0.442177in}}{\pgfqpoint{0.719005in}{0.871884in}}%
\pgfusepath{clip}%
\pgfsetroundcap%
\pgfsetroundjoin%
\pgfsetlinewidth{1.003750pt}%
\definecolor{currentstroke}{rgb}{0.003922,0.450980,0.698039}%
\pgfsetstrokecolor{currentstroke}%
\pgfsetdash{}{0pt}%
\pgfpathmoveto{\pgfqpoint{0.520798in}{0.969018in}}%
\pgfpathlineto{\pgfqpoint{0.556748in}{0.969018in}}%
\pgfpathlineto{\pgfqpoint{0.556748in}{0.860960in}}%
\pgfpathlineto{\pgfqpoint{0.628649in}{0.860960in}}%
\pgfpathlineto{\pgfqpoint{0.628649in}{0.679441in}}%
\pgfpathlineto{\pgfqpoint{0.700549in}{0.679441in}}%
\pgfpathlineto{\pgfqpoint{0.700549in}{0.469354in}}%
\pgfpathlineto{\pgfqpoint{0.772450in}{0.469354in}}%
\pgfpathlineto{\pgfqpoint{0.772450in}{0.448113in}}%
\pgfpathlineto{\pgfqpoint{0.844350in}{0.448113in}}%
\pgfpathlineto{\pgfqpoint{0.844350in}{0.445145in}}%
\pgfpathlineto{\pgfqpoint{0.916251in}{0.445145in}}%
\pgfpathlineto{\pgfqpoint{0.916251in}{0.444310in}}%
\pgfpathlineto{\pgfqpoint{0.988151in}{0.444310in}}%
\pgfpathlineto{\pgfqpoint{0.988151in}{0.443568in}}%
\pgfpathlineto{\pgfqpoint{1.060052in}{0.443568in}}%
\pgfpathlineto{\pgfqpoint{1.060052in}{0.442548in}}%
\pgfpathlineto{\pgfqpoint{1.131952in}{0.442548in}}%
\pgfpathlineto{\pgfqpoint{1.131952in}{0.442362in}}%
\pgfpathlineto{\pgfqpoint{1.239803in}{0.442362in}}%
\pgfpathlineto{\pgfqpoint{1.239803in}{0.442270in}}%
\pgfpathlineto{\pgfqpoint{1.241470in}{0.442270in}}%
\pgfusepath{stroke}%
\end{pgfscope}%
\begin{pgfscope}%
\pgfpathrectangle{\pgfqpoint{0.520798in}{0.442177in}}{\pgfqpoint{0.719005in}{0.871884in}}%
\pgfusepath{clip}%
\pgfsetbuttcap%
\pgfsetroundjoin%
\definecolor{currentfill}{rgb}{0.003922,0.450980,0.698039}%
\pgfsetfillcolor{currentfill}%
\pgfsetfillopacity{0.500000}%
\pgfsetlinewidth{0.000000pt}%
\definecolor{currentstroke}{rgb}{0.003922,0.450980,0.698039}%
\pgfsetstrokecolor{currentstroke}%
\pgfsetstrokeopacity{0.500000}%
\pgfsetdash{}{0pt}%
\pgfpathmoveto{\pgfqpoint{0.520798in}{0.987622in}}%
\pgfpathlineto{\pgfqpoint{0.520798in}{0.950414in}}%
\pgfpathlineto{\pgfqpoint{0.556748in}{0.950414in}}%
\pgfpathlineto{\pgfqpoint{0.556748in}{0.839442in}}%
\pgfpathlineto{\pgfqpoint{0.628649in}{0.839442in}}%
\pgfpathlineto{\pgfqpoint{0.628649in}{0.664144in}}%
\pgfpathlineto{\pgfqpoint{0.700549in}{0.664144in}}%
\pgfpathlineto{\pgfqpoint{0.700549in}{0.465489in}}%
\pgfpathlineto{\pgfqpoint{0.772450in}{0.465489in}}%
\pgfpathlineto{\pgfqpoint{0.772450in}{0.447080in}}%
\pgfpathlineto{\pgfqpoint{0.844350in}{0.447080in}}%
\pgfpathlineto{\pgfqpoint{0.844350in}{0.444451in}}%
\pgfpathlineto{\pgfqpoint{0.916251in}{0.444451in}}%
\pgfpathlineto{\pgfqpoint{0.916251in}{0.443502in}}%
\pgfpathlineto{\pgfqpoint{0.988151in}{0.443502in}}%
\pgfpathlineto{\pgfqpoint{0.988151in}{0.442912in}}%
\pgfpathlineto{\pgfqpoint{1.060052in}{0.442912in}}%
\pgfpathlineto{\pgfqpoint{1.060052in}{0.442362in}}%
\pgfpathlineto{\pgfqpoint{1.131952in}{0.442362in}}%
\pgfpathlineto{\pgfqpoint{1.131952in}{0.442135in}}%
\pgfpathlineto{\pgfqpoint{1.239803in}{0.442135in}}%
\pgfpathlineto{\pgfqpoint{1.239803in}{0.442084in}}%
\pgfpathlineto{\pgfqpoint{1.383604in}{0.442084in}}%
\pgfpathlineto{\pgfqpoint{1.383604in}{0.442177in}}%
\pgfpathlineto{\pgfqpoint{2.066659in}{0.442177in}}%
\pgfpathlineto{\pgfqpoint{2.066659in}{0.442177in}}%
\pgfpathlineto{\pgfqpoint{2.677813in}{0.442177in}}%
\pgfpathlineto{\pgfqpoint{2.677813in}{0.442177in}}%
\pgfpathlineto{\pgfqpoint{2.677813in}{0.442177in}}%
\pgfpathlineto{\pgfqpoint{2.066659in}{0.442177in}}%
\pgfpathlineto{\pgfqpoint{2.066659in}{0.442177in}}%
\pgfpathlineto{\pgfqpoint{1.383604in}{0.442177in}}%
\pgfpathlineto{\pgfqpoint{1.383604in}{0.442455in}}%
\pgfpathlineto{\pgfqpoint{1.239803in}{0.442455in}}%
\pgfpathlineto{\pgfqpoint{1.239803in}{0.442590in}}%
\pgfpathlineto{\pgfqpoint{1.131952in}{0.442590in}}%
\pgfpathlineto{\pgfqpoint{1.131952in}{0.442733in}}%
\pgfpathlineto{\pgfqpoint{1.060052in}{0.442733in}}%
\pgfpathlineto{\pgfqpoint{1.060052in}{0.444224in}}%
\pgfpathlineto{\pgfqpoint{0.988151in}{0.444224in}}%
\pgfpathlineto{\pgfqpoint{0.988151in}{0.445119in}}%
\pgfpathlineto{\pgfqpoint{0.916251in}{0.445119in}}%
\pgfpathlineto{\pgfqpoint{0.916251in}{0.445839in}}%
\pgfpathlineto{\pgfqpoint{0.844350in}{0.445839in}}%
\pgfpathlineto{\pgfqpoint{0.844350in}{0.449146in}}%
\pgfpathlineto{\pgfqpoint{0.772450in}{0.449146in}}%
\pgfpathlineto{\pgfqpoint{0.772450in}{0.473218in}}%
\pgfpathlineto{\pgfqpoint{0.700549in}{0.473218in}}%
\pgfpathlineto{\pgfqpoint{0.700549in}{0.694737in}}%
\pgfpathlineto{\pgfqpoint{0.628649in}{0.694737in}}%
\pgfpathlineto{\pgfqpoint{0.628649in}{0.882478in}}%
\pgfpathlineto{\pgfqpoint{0.556748in}{0.882478in}}%
\pgfpathlineto{\pgfqpoint{0.556748in}{0.987622in}}%
\pgfpathlineto{\pgfqpoint{0.520798in}{0.987622in}}%
\pgfpathlineto{\pgfqpoint{0.520798in}{0.987622in}}%
\pgfpathclose%
\pgfusepath{fill}%
\end{pgfscope}%
\begin{pgfscope}%
\pgfpathrectangle{\pgfqpoint{0.520798in}{0.442177in}}{\pgfqpoint{0.719005in}{0.871884in}}%
\pgfusepath{clip}%
\pgfsetroundcap%
\pgfsetroundjoin%
\pgfsetlinewidth{1.003750pt}%
\definecolor{currentstroke}{rgb}{0.870588,0.560784,0.019608}%
\pgfsetstrokecolor{currentstroke}%
\pgfsetdash{}{0pt}%
\pgfpathmoveto{\pgfqpoint{0.520798in}{0.969018in}}%
\pgfpathlineto{\pgfqpoint{0.556748in}{0.969018in}}%
\pgfpathlineto{\pgfqpoint{0.556748in}{0.884890in}}%
\pgfpathlineto{\pgfqpoint{0.628649in}{0.884890in}}%
\pgfpathlineto{\pgfqpoint{0.628649in}{0.794826in}}%
\pgfpathlineto{\pgfqpoint{0.700549in}{0.794826in}}%
\pgfpathlineto{\pgfqpoint{0.700549in}{0.720160in}}%
\pgfpathlineto{\pgfqpoint{0.772450in}{0.720160in}}%
\pgfpathlineto{\pgfqpoint{0.772450in}{0.663394in}}%
\pgfpathlineto{\pgfqpoint{0.844350in}{0.663394in}}%
\pgfpathlineto{\pgfqpoint{0.844350in}{0.618780in}}%
\pgfpathlineto{\pgfqpoint{0.916251in}{0.618780in}}%
\pgfpathlineto{\pgfqpoint{0.916251in}{0.469354in}}%
\pgfpathlineto{\pgfqpoint{0.988151in}{0.469354in}}%
\pgfpathlineto{\pgfqpoint{0.988151in}{0.463510in}}%
\pgfpathlineto{\pgfqpoint{1.060052in}{0.463510in}}%
\pgfpathlineto{\pgfqpoint{1.060052in}{0.448113in}}%
\pgfpathlineto{\pgfqpoint{1.131952in}{0.448113in}}%
\pgfpathlineto{\pgfqpoint{1.131952in}{0.446907in}}%
\pgfpathlineto{\pgfqpoint{1.203853in}{0.446907in}}%
\pgfpathlineto{\pgfqpoint{1.203853in}{0.444959in}}%
\pgfpathlineto{\pgfqpoint{1.241470in}{0.444959in}}%
\pgfusepath{stroke}%
\end{pgfscope}%
\begin{pgfscope}%
\pgfpathrectangle{\pgfqpoint{0.520798in}{0.442177in}}{\pgfqpoint{0.719005in}{0.871884in}}%
\pgfusepath{clip}%
\pgfsetbuttcap%
\pgfsetroundjoin%
\definecolor{currentfill}{rgb}{0.870588,0.560784,0.019608}%
\pgfsetfillcolor{currentfill}%
\pgfsetfillopacity{0.500000}%
\pgfsetlinewidth{0.000000pt}%
\definecolor{currentstroke}{rgb}{0.870588,0.560784,0.019608}%
\pgfsetstrokecolor{currentstroke}%
\pgfsetstrokeopacity{0.500000}%
\pgfsetdash{}{0pt}%
\pgfpathmoveto{\pgfqpoint{0.520798in}{0.987622in}}%
\pgfpathlineto{\pgfqpoint{0.520798in}{0.950414in}}%
\pgfpathlineto{\pgfqpoint{0.556748in}{0.950414in}}%
\pgfpathlineto{\pgfqpoint{0.556748in}{0.863105in}}%
\pgfpathlineto{\pgfqpoint{0.628649in}{0.863105in}}%
\pgfpathlineto{\pgfqpoint{0.628649in}{0.775311in}}%
\pgfpathlineto{\pgfqpoint{0.700549in}{0.775311in}}%
\pgfpathlineto{\pgfqpoint{0.700549in}{0.702589in}}%
\pgfpathlineto{\pgfqpoint{0.772450in}{0.702589in}}%
\pgfpathlineto{\pgfqpoint{0.772450in}{0.647580in}}%
\pgfpathlineto{\pgfqpoint{0.844350in}{0.647580in}}%
\pgfpathlineto{\pgfqpoint{0.844350in}{0.603339in}}%
\pgfpathlineto{\pgfqpoint{0.916251in}{0.603339in}}%
\pgfpathlineto{\pgfqpoint{0.916251in}{0.465489in}}%
\pgfpathlineto{\pgfqpoint{0.988151in}{0.465489in}}%
\pgfpathlineto{\pgfqpoint{0.988151in}{0.460102in}}%
\pgfpathlineto{\pgfqpoint{1.060052in}{0.460102in}}%
\pgfpathlineto{\pgfqpoint{1.060052in}{0.447080in}}%
\pgfpathlineto{\pgfqpoint{1.131952in}{0.447080in}}%
\pgfpathlineto{\pgfqpoint{1.131952in}{0.445684in}}%
\pgfpathlineto{\pgfqpoint{1.203853in}{0.445684in}}%
\pgfpathlineto{\pgfqpoint{1.203853in}{0.444451in}}%
\pgfpathlineto{\pgfqpoint{1.275753in}{0.444451in}}%
\pgfpathlineto{\pgfqpoint{1.275753in}{0.443464in}}%
\pgfpathlineto{\pgfqpoint{1.347654in}{0.443464in}}%
\pgfpathlineto{\pgfqpoint{1.347654in}{0.442641in}}%
\pgfpathlineto{\pgfqpoint{1.419554in}{0.442641in}}%
\pgfpathlineto{\pgfqpoint{1.419554in}{0.442574in}}%
\pgfpathlineto{\pgfqpoint{1.491455in}{0.442574in}}%
\pgfpathlineto{\pgfqpoint{1.491455in}{0.442582in}}%
\pgfpathlineto{\pgfqpoint{1.563355in}{0.442582in}}%
\pgfpathlineto{\pgfqpoint{1.563355in}{0.442596in}}%
\pgfpathlineto{\pgfqpoint{1.635256in}{0.442596in}}%
\pgfpathlineto{\pgfqpoint{1.635256in}{0.442362in}}%
\pgfpathlineto{\pgfqpoint{1.743107in}{0.442362in}}%
\pgfpathlineto{\pgfqpoint{1.743107in}{0.442135in}}%
\pgfpathlineto{\pgfqpoint{1.922858in}{0.442135in}}%
\pgfpathlineto{\pgfqpoint{1.922858in}{0.442084in}}%
\pgfpathlineto{\pgfqpoint{2.174510in}{0.442084in}}%
\pgfpathlineto{\pgfqpoint{2.174510in}{0.442177in}}%
\pgfpathlineto{\pgfqpoint{2.498062in}{0.442177in}}%
\pgfpathlineto{\pgfqpoint{2.498062in}{0.442177in}}%
\pgfpathlineto{\pgfqpoint{2.677813in}{0.442177in}}%
\pgfpathlineto{\pgfqpoint{2.677813in}{0.442177in}}%
\pgfpathlineto{\pgfqpoint{2.677813in}{0.442177in}}%
\pgfpathlineto{\pgfqpoint{2.498062in}{0.442177in}}%
\pgfpathlineto{\pgfqpoint{2.498062in}{0.442177in}}%
\pgfpathlineto{\pgfqpoint{2.174510in}{0.442177in}}%
\pgfpathlineto{\pgfqpoint{2.174510in}{0.442455in}}%
\pgfpathlineto{\pgfqpoint{1.922858in}{0.442455in}}%
\pgfpathlineto{\pgfqpoint{1.922858in}{0.442590in}}%
\pgfpathlineto{\pgfqpoint{1.743107in}{0.442590in}}%
\pgfpathlineto{\pgfqpoint{1.743107in}{0.442733in}}%
\pgfpathlineto{\pgfqpoint{1.635256in}{0.442733in}}%
\pgfpathlineto{\pgfqpoint{1.635256in}{0.443612in}}%
\pgfpathlineto{\pgfqpoint{1.563355in}{0.443612in}}%
\pgfpathlineto{\pgfqpoint{1.563355in}{0.443812in}}%
\pgfpathlineto{\pgfqpoint{1.491455in}{0.443812in}}%
\pgfpathlineto{\pgfqpoint{1.491455in}{0.444191in}}%
\pgfpathlineto{\pgfqpoint{1.419554in}{0.444191in}}%
\pgfpathlineto{\pgfqpoint{1.419554in}{0.444496in}}%
\pgfpathlineto{\pgfqpoint{1.347654in}{0.444496in}}%
\pgfpathlineto{\pgfqpoint{1.347654in}{0.444971in}}%
\pgfpathlineto{\pgfqpoint{1.275753in}{0.444971in}}%
\pgfpathlineto{\pgfqpoint{1.275753in}{0.445467in}}%
\pgfpathlineto{\pgfqpoint{1.203853in}{0.445467in}}%
\pgfpathlineto{\pgfqpoint{1.203853in}{0.448131in}}%
\pgfpathlineto{\pgfqpoint{1.131952in}{0.448131in}}%
\pgfpathlineto{\pgfqpoint{1.131952in}{0.449146in}}%
\pgfpathlineto{\pgfqpoint{1.060052in}{0.449146in}}%
\pgfpathlineto{\pgfqpoint{1.060052in}{0.466918in}}%
\pgfpathlineto{\pgfqpoint{0.988151in}{0.466918in}}%
\pgfpathlineto{\pgfqpoint{0.988151in}{0.473218in}}%
\pgfpathlineto{\pgfqpoint{0.916251in}{0.473218in}}%
\pgfpathlineto{\pgfqpoint{0.916251in}{0.634221in}}%
\pgfpathlineto{\pgfqpoint{0.844350in}{0.634221in}}%
\pgfpathlineto{\pgfqpoint{0.844350in}{0.679209in}}%
\pgfpathlineto{\pgfqpoint{0.772450in}{0.679209in}}%
\pgfpathlineto{\pgfqpoint{0.772450in}{0.737730in}}%
\pgfpathlineto{\pgfqpoint{0.700549in}{0.737730in}}%
\pgfpathlineto{\pgfqpoint{0.700549in}{0.814341in}}%
\pgfpathlineto{\pgfqpoint{0.628649in}{0.814341in}}%
\pgfpathlineto{\pgfqpoint{0.628649in}{0.906676in}}%
\pgfpathlineto{\pgfqpoint{0.556748in}{0.906676in}}%
\pgfpathlineto{\pgfqpoint{0.556748in}{0.987622in}}%
\pgfpathlineto{\pgfqpoint{0.520798in}{0.987622in}}%
\pgfpathlineto{\pgfqpoint{0.520798in}{0.987622in}}%
\pgfpathclose%
\pgfusepath{fill}%
\end{pgfscope}%
\begin{pgfscope}%
\pgfpathrectangle{\pgfqpoint{0.520798in}{0.442177in}}{\pgfqpoint{0.719005in}{0.871884in}}%
\pgfusepath{clip}%
\pgfsetroundcap%
\pgfsetroundjoin%
\pgfsetlinewidth{1.003750pt}%
\definecolor{currentstroke}{rgb}{0.007843,0.619608,0.450980}%
\pgfsetstrokecolor{currentstroke}%
\pgfsetdash{}{0pt}%
\pgfpathmoveto{\pgfqpoint{0.520798in}{0.969018in}}%
\pgfpathlineto{\pgfqpoint{0.556748in}{0.969018in}}%
\pgfpathlineto{\pgfqpoint{0.556748in}{0.884890in}}%
\pgfpathlineto{\pgfqpoint{0.628649in}{0.884890in}}%
\pgfpathlineto{\pgfqpoint{0.628649in}{0.794826in}}%
\pgfpathlineto{\pgfqpoint{0.700549in}{0.794826in}}%
\pgfpathlineto{\pgfqpoint{0.700549in}{0.720160in}}%
\pgfpathlineto{\pgfqpoint{0.772450in}{0.720160in}}%
\pgfpathlineto{\pgfqpoint{0.772450in}{0.663394in}}%
\pgfpathlineto{\pgfqpoint{0.844350in}{0.663394in}}%
\pgfpathlineto{\pgfqpoint{0.844350in}{0.618780in}}%
\pgfpathlineto{\pgfqpoint{0.916251in}{0.618780in}}%
\pgfpathlineto{\pgfqpoint{0.916251in}{0.581307in}}%
\pgfpathlineto{\pgfqpoint{0.988151in}{0.581307in}}%
\pgfpathlineto{\pgfqpoint{0.988151in}{0.551162in}}%
\pgfpathlineto{\pgfqpoint{1.060052in}{0.551162in}}%
\pgfpathlineto{\pgfqpoint{1.060052in}{0.527696in}}%
\pgfpathlineto{\pgfqpoint{1.131952in}{0.527696in}}%
\pgfpathlineto{\pgfqpoint{1.131952in}{0.510165in}}%
\pgfpathlineto{\pgfqpoint{1.203853in}{0.510165in}}%
\pgfpathlineto{\pgfqpoint{1.203853in}{0.496623in}}%
\pgfpathlineto{\pgfqpoint{1.241470in}{0.496623in}}%
\pgfusepath{stroke}%
\end{pgfscope}%
\begin{pgfscope}%
\pgfpathrectangle{\pgfqpoint{0.520798in}{0.442177in}}{\pgfqpoint{0.719005in}{0.871884in}}%
\pgfusepath{clip}%
\pgfsetbuttcap%
\pgfsetroundjoin%
\definecolor{currentfill}{rgb}{0.007843,0.619608,0.450980}%
\pgfsetfillcolor{currentfill}%
\pgfsetfillopacity{0.500000}%
\pgfsetlinewidth{0.000000pt}%
\definecolor{currentstroke}{rgb}{0.007843,0.619608,0.450980}%
\pgfsetstrokecolor{currentstroke}%
\pgfsetstrokeopacity{0.500000}%
\pgfsetdash{}{0pt}%
\pgfpathmoveto{\pgfqpoint{0.520798in}{0.987622in}}%
\pgfpathlineto{\pgfqpoint{0.520798in}{0.950414in}}%
\pgfpathlineto{\pgfqpoint{0.556748in}{0.950414in}}%
\pgfpathlineto{\pgfqpoint{0.556748in}{0.863105in}}%
\pgfpathlineto{\pgfqpoint{0.628649in}{0.863105in}}%
\pgfpathlineto{\pgfqpoint{0.628649in}{0.775311in}}%
\pgfpathlineto{\pgfqpoint{0.700549in}{0.775311in}}%
\pgfpathlineto{\pgfqpoint{0.700549in}{0.702589in}}%
\pgfpathlineto{\pgfqpoint{0.772450in}{0.702589in}}%
\pgfpathlineto{\pgfqpoint{0.772450in}{0.647580in}}%
\pgfpathlineto{\pgfqpoint{0.844350in}{0.647580in}}%
\pgfpathlineto{\pgfqpoint{0.844350in}{0.603339in}}%
\pgfpathlineto{\pgfqpoint{0.916251in}{0.603339in}}%
\pgfpathlineto{\pgfqpoint{0.916251in}{0.566092in}}%
\pgfpathlineto{\pgfqpoint{0.988151in}{0.566092in}}%
\pgfpathlineto{\pgfqpoint{0.988151in}{0.537843in}}%
\pgfpathlineto{\pgfqpoint{1.060052in}{0.537843in}}%
\pgfpathlineto{\pgfqpoint{1.060052in}{0.516738in}}%
\pgfpathlineto{\pgfqpoint{1.131952in}{0.516738in}}%
\pgfpathlineto{\pgfqpoint{1.131952in}{0.500649in}}%
\pgfpathlineto{\pgfqpoint{1.203853in}{0.500649in}}%
\pgfpathlineto{\pgfqpoint{1.203853in}{0.490220in}}%
\pgfpathlineto{\pgfqpoint{1.275753in}{0.490220in}}%
\pgfpathlineto{\pgfqpoint{1.275753in}{0.479688in}}%
\pgfpathlineto{\pgfqpoint{1.347654in}{0.479688in}}%
\pgfpathlineto{\pgfqpoint{1.347654in}{0.465489in}}%
\pgfpathlineto{\pgfqpoint{1.419554in}{0.465489in}}%
\pgfpathlineto{\pgfqpoint{1.419554in}{0.460102in}}%
\pgfpathlineto{\pgfqpoint{1.491455in}{0.460102in}}%
\pgfpathlineto{\pgfqpoint{1.491455in}{0.456835in}}%
\pgfpathlineto{\pgfqpoint{1.563355in}{0.456835in}}%
\pgfpathlineto{\pgfqpoint{1.563355in}{0.454273in}}%
\pgfpathlineto{\pgfqpoint{1.635256in}{0.454273in}}%
\pgfpathlineto{\pgfqpoint{1.635256in}{0.447080in}}%
\pgfpathlineto{\pgfqpoint{1.707156in}{0.447080in}}%
\pgfpathlineto{\pgfqpoint{1.707156in}{0.445684in}}%
\pgfpathlineto{\pgfqpoint{1.779057in}{0.445684in}}%
\pgfpathlineto{\pgfqpoint{1.779057in}{0.445145in}}%
\pgfpathlineto{\pgfqpoint{1.850957in}{0.445145in}}%
\pgfpathlineto{\pgfqpoint{1.850957in}{0.444672in}}%
\pgfpathlineto{\pgfqpoint{1.922858in}{0.444672in}}%
\pgfpathlineto{\pgfqpoint{1.922858in}{0.443202in}}%
\pgfpathlineto{\pgfqpoint{1.994758in}{0.443202in}}%
\pgfpathlineto{\pgfqpoint{1.994758in}{0.442912in}}%
\pgfpathlineto{\pgfqpoint{2.066659in}{0.442912in}}%
\pgfpathlineto{\pgfqpoint{2.066659in}{0.442596in}}%
\pgfpathlineto{\pgfqpoint{2.138559in}{0.442596in}}%
\pgfpathlineto{\pgfqpoint{2.138559in}{0.442596in}}%
\pgfpathlineto{\pgfqpoint{2.318311in}{0.442596in}}%
\pgfpathlineto{\pgfqpoint{2.318311in}{0.442446in}}%
\pgfpathlineto{\pgfqpoint{2.569962in}{0.442446in}}%
\pgfpathlineto{\pgfqpoint{2.569962in}{0.442353in}}%
\pgfpathlineto{\pgfqpoint{2.677813in}{0.442353in}}%
\pgfpathlineto{\pgfqpoint{2.677813in}{0.443299in}}%
\pgfpathlineto{\pgfqpoint{2.677813in}{0.443299in}}%
\pgfpathlineto{\pgfqpoint{2.569962in}{0.443299in}}%
\pgfpathlineto{\pgfqpoint{2.569962in}{0.443392in}}%
\pgfpathlineto{\pgfqpoint{2.318311in}{0.443392in}}%
\pgfpathlineto{\pgfqpoint{2.318311in}{0.443612in}}%
\pgfpathlineto{\pgfqpoint{2.138559in}{0.443612in}}%
\pgfpathlineto{\pgfqpoint{2.138559in}{0.443984in}}%
\pgfpathlineto{\pgfqpoint{2.066659in}{0.443984in}}%
\pgfpathlineto{\pgfqpoint{2.066659in}{0.444224in}}%
\pgfpathlineto{\pgfqpoint{1.994758in}{0.444224in}}%
\pgfpathlineto{\pgfqpoint{1.994758in}{0.444861in}}%
\pgfpathlineto{\pgfqpoint{1.922858in}{0.444861in}}%
\pgfpathlineto{\pgfqpoint{1.922858in}{0.445618in}}%
\pgfpathlineto{\pgfqpoint{1.850957in}{0.445618in}}%
\pgfpathlineto{\pgfqpoint{1.850957in}{0.446258in}}%
\pgfpathlineto{\pgfqpoint{1.779057in}{0.446258in}}%
\pgfpathlineto{\pgfqpoint{1.779057in}{0.448131in}}%
\pgfpathlineto{\pgfqpoint{1.707156in}{0.448131in}}%
\pgfpathlineto{\pgfqpoint{1.707156in}{0.449146in}}%
\pgfpathlineto{\pgfqpoint{1.635256in}{0.449146in}}%
\pgfpathlineto{\pgfqpoint{1.635256in}{0.458278in}}%
\pgfpathlineto{\pgfqpoint{1.563355in}{0.458278in}}%
\pgfpathlineto{\pgfqpoint{1.563355in}{0.461652in}}%
\pgfpathlineto{\pgfqpoint{1.491455in}{0.461652in}}%
\pgfpathlineto{\pgfqpoint{1.491455in}{0.466918in}}%
\pgfpathlineto{\pgfqpoint{1.419554in}{0.466918in}}%
\pgfpathlineto{\pgfqpoint{1.419554in}{0.473218in}}%
\pgfpathlineto{\pgfqpoint{1.347654in}{0.473218in}}%
\pgfpathlineto{\pgfqpoint{1.347654in}{0.490370in}}%
\pgfpathlineto{\pgfqpoint{1.275753in}{0.490370in}}%
\pgfpathlineto{\pgfqpoint{1.275753in}{0.503027in}}%
\pgfpathlineto{\pgfqpoint{1.203853in}{0.503027in}}%
\pgfpathlineto{\pgfqpoint{1.203853in}{0.519681in}}%
\pgfpathlineto{\pgfqpoint{1.131952in}{0.519681in}}%
\pgfpathlineto{\pgfqpoint{1.131952in}{0.538653in}}%
\pgfpathlineto{\pgfqpoint{1.060052in}{0.538653in}}%
\pgfpathlineto{\pgfqpoint{1.060052in}{0.564481in}}%
\pgfpathlineto{\pgfqpoint{0.988151in}{0.564481in}}%
\pgfpathlineto{\pgfqpoint{0.988151in}{0.596523in}}%
\pgfpathlineto{\pgfqpoint{0.916251in}{0.596523in}}%
\pgfpathlineto{\pgfqpoint{0.916251in}{0.634221in}}%
\pgfpathlineto{\pgfqpoint{0.844350in}{0.634221in}}%
\pgfpathlineto{\pgfqpoint{0.844350in}{0.679209in}}%
\pgfpathlineto{\pgfqpoint{0.772450in}{0.679209in}}%
\pgfpathlineto{\pgfqpoint{0.772450in}{0.737730in}}%
\pgfpathlineto{\pgfqpoint{0.700549in}{0.737730in}}%
\pgfpathlineto{\pgfqpoint{0.700549in}{0.814341in}}%
\pgfpathlineto{\pgfqpoint{0.628649in}{0.814341in}}%
\pgfpathlineto{\pgfqpoint{0.628649in}{0.906676in}}%
\pgfpathlineto{\pgfqpoint{0.556748in}{0.906676in}}%
\pgfpathlineto{\pgfqpoint{0.556748in}{0.987622in}}%
\pgfpathlineto{\pgfqpoint{0.520798in}{0.987622in}}%
\pgfpathlineto{\pgfqpoint{0.520798in}{0.987622in}}%
\pgfpathclose%
\pgfusepath{fill}%
\end{pgfscope}%
\begin{pgfscope}%
\definecolor{textcolor}{rgb}{0.150000,0.150000,0.150000}%
\pgfsetstrokecolor{textcolor}%
\pgfsetfillcolor{textcolor}%
\pgftext[x=0.880300in,y=1.397395in,,base]{\color{textcolor}\rmfamily\fontsize{9.000000}{10.800000}\selectfont \(\displaystyle c_A^-=1\)}%
\end{pgfscope}%
\begin{pgfscope}%
\pgfsetbuttcap%
\pgfsetmiterjoin%
\definecolor{currentfill}{rgb}{1.000000,1.000000,1.000000}%
\pgfsetfillcolor{currentfill}%
\pgfsetfillopacity{0.800000}%
\pgfsetlinewidth{1.003750pt}%
\definecolor{currentstroke}{rgb}{0.800000,0.800000,0.800000}%
\pgfsetstrokecolor{currentstroke}%
\pgfsetstrokeopacity{0.800000}%
\pgfsetdash{}{0pt}%
\pgfpathmoveto{\pgfqpoint{0.785843in}{0.638103in}}%
\pgfpathlineto{\pgfqpoint{1.191192in}{0.638103in}}%
\pgfpathlineto{\pgfqpoint{1.191192in}{1.265450in}}%
\pgfpathlineto{\pgfqpoint{0.785843in}{1.265450in}}%
\pgfpathlineto{\pgfqpoint{0.785843in}{0.638103in}}%
\pgfpathclose%
\pgfusepath{stroke,fill}%
\end{pgfscope}%
\begin{pgfscope}%
\definecolor{textcolor}{rgb}{0.150000,0.150000,0.150000}%
\pgfsetstrokecolor{textcolor}%
\pgfsetfillcolor{textcolor}%
\pgftext[x=0.918271in,y=1.113275in,left,base]{\color{textcolor}\rmfamily\fontsize{9.000000}{10.800000}\selectfont \(\displaystyle c_A^+\)}%
\end{pgfscope}%
\begin{pgfscope}%
\pgfsetroundcap%
\pgfsetroundjoin%
\pgfsetlinewidth{1.003750pt}%
\definecolor{currentstroke}{rgb}{0.003922,0.450980,0.698039}%
\pgfsetstrokecolor{currentstroke}%
\pgfsetdash{}{0pt}%
\pgfpathmoveto{\pgfqpoint{0.824732in}{1.001052in}}%
\pgfpathlineto{\pgfqpoint{0.873343in}{1.001052in}}%
\pgfpathlineto{\pgfqpoint{0.873343in}{1.001052in}}%
\pgfpathlineto{\pgfqpoint{0.970565in}{1.001052in}}%
\pgfpathlineto{\pgfqpoint{0.970565in}{1.001052in}}%
\pgfpathlineto{\pgfqpoint{1.019176in}{1.001052in}}%
\pgfusepath{stroke}%
\end{pgfscope}%
\begin{pgfscope}%
\definecolor{textcolor}{rgb}{0.150000,0.150000,0.150000}%
\pgfsetstrokecolor{textcolor}%
\pgfsetfillcolor{textcolor}%
\pgftext[x=1.096954in,y=0.967024in,left,base]{\color{textcolor}\rmfamily\fontsize{7.000000}{8.400000}\selectfont 1}%
\end{pgfscope}%
\begin{pgfscope}%
\pgfsetroundcap%
\pgfsetroundjoin%
\pgfsetlinewidth{1.003750pt}%
\definecolor{currentstroke}{rgb}{0.870588,0.560784,0.019608}%
\pgfsetstrokecolor{currentstroke}%
\pgfsetdash{}{0pt}%
\pgfpathmoveto{\pgfqpoint{0.824732in}{0.865486in}}%
\pgfpathlineto{\pgfqpoint{0.873343in}{0.865486in}}%
\pgfpathlineto{\pgfqpoint{0.873343in}{0.865486in}}%
\pgfpathlineto{\pgfqpoint{0.970565in}{0.865486in}}%
\pgfpathlineto{\pgfqpoint{0.970565in}{0.865486in}}%
\pgfpathlineto{\pgfqpoint{1.019176in}{0.865486in}}%
\pgfusepath{stroke}%
\end{pgfscope}%
\begin{pgfscope}%
\definecolor{textcolor}{rgb}{0.150000,0.150000,0.150000}%
\pgfsetstrokecolor{textcolor}%
\pgfsetfillcolor{textcolor}%
\pgftext[x=1.096954in,y=0.831458in,left,base]{\color{textcolor}\rmfamily\fontsize{7.000000}{8.400000}\selectfont 2}%
\end{pgfscope}%
\begin{pgfscope}%
\pgfsetroundcap%
\pgfsetroundjoin%
\pgfsetlinewidth{1.003750pt}%
\definecolor{currentstroke}{rgb}{0.007843,0.619608,0.450980}%
\pgfsetstrokecolor{currentstroke}%
\pgfsetdash{}{0pt}%
\pgfpathmoveto{\pgfqpoint{0.824732in}{0.729919in}}%
\pgfpathlineto{\pgfqpoint{0.873343in}{0.729919in}}%
\pgfpathlineto{\pgfqpoint{0.873343in}{0.729919in}}%
\pgfpathlineto{\pgfqpoint{0.970565in}{0.729919in}}%
\pgfpathlineto{\pgfqpoint{0.970565in}{0.729919in}}%
\pgfpathlineto{\pgfqpoint{1.019176in}{0.729919in}}%
\pgfusepath{stroke}%
\end{pgfscope}%
\begin{pgfscope}%
\definecolor{textcolor}{rgb}{0.150000,0.150000,0.150000}%
\pgfsetstrokecolor{textcolor}%
\pgfsetfillcolor{textcolor}%
\pgftext[x=1.096954in,y=0.695892in,left,base]{\color{textcolor}\rmfamily\fontsize{7.000000}{8.400000}\selectfont 4}%
\end{pgfscope}%
\end{pgfpicture}%
\makeatother%
\endgroup%

%% file: figures/experiments/graphs/sparse_smoothing/nodes_structure/node_classification-Citeseer-GCN-hidden=32-p_adj_plus=0.001-p_adj_minus=0.8-p_att_plus=0.0-p_att_minus=0.0-multi_class_cert-A.pgf
\begingroup%
\makeatletter%
\begin{pgfpicture}%
\pgfpathrectangle{\pgfpointorigin}{\pgfqpoint{1.375000in}{1.581250in}}%
\pgfusepath{use as bounding box, clip}%
\begin{pgfscope}%
\pgfsetbuttcap%
\pgfsetmiterjoin%
\definecolor{currentfill}{rgb}{1.000000,1.000000,1.000000}%
\pgfsetfillcolor{currentfill}%
\pgfsetlinewidth{0.000000pt}%
\definecolor{currentstroke}{rgb}{1.000000,1.000000,1.000000}%
\pgfsetstrokecolor{currentstroke}%
\pgfsetdash{}{0pt}%
\pgfpathmoveto{\pgfqpoint{0.000000in}{0.000000in}}%
\pgfpathlineto{\pgfqpoint{1.375000in}{0.000000in}}%
\pgfpathlineto{\pgfqpoint{1.375000in}{1.581250in}}%
\pgfpathlineto{\pgfqpoint{0.000000in}{1.581250in}}%
\pgfpathlineto{\pgfqpoint{0.000000in}{0.000000in}}%
\pgfpathclose%
\pgfusepath{fill}%
\end{pgfscope}%
\begin{pgfscope}%
\pgfsetbuttcap%
\pgfsetmiterjoin%
\definecolor{currentfill}{rgb}{1.000000,1.000000,1.000000}%
\pgfsetfillcolor{currentfill}%
\pgfsetlinewidth{0.000000pt}%
\definecolor{currentstroke}{rgb}{0.000000,0.000000,0.000000}%
\pgfsetstrokecolor{currentstroke}%
\pgfsetstrokeopacity{0.000000}%
\pgfsetdash{}{0pt}%
\pgfpathmoveto{\pgfqpoint{0.520798in}{0.442177in}}%
\pgfpathlineto{\pgfqpoint{1.239803in}{0.442177in}}%
\pgfpathlineto{\pgfqpoint{1.239803in}{1.314061in}}%
\pgfpathlineto{\pgfqpoint{0.520798in}{1.314061in}}%
\pgfpathlineto{\pgfqpoint{0.520798in}{0.442177in}}%
\pgfpathclose%
\pgfusepath{fill}%
\end{pgfscope}%
\begin{pgfscope}%
\pgfpathrectangle{\pgfqpoint{0.520798in}{0.442177in}}{\pgfqpoint{0.719005in}{0.871884in}}%
\pgfusepath{clip}%
\pgfsetroundcap%
\pgfsetroundjoin%
\pgfsetlinewidth{0.501875pt}%
\definecolor{currentstroke}{rgb}{0.800000,0.800000,0.800000}%
\pgfsetstrokecolor{currentstroke}%
\pgfsetdash{}{0pt}%
\pgfpathmoveto{\pgfqpoint{0.520798in}{0.442177in}}%
\pgfpathlineto{\pgfqpoint{0.520798in}{1.314061in}}%
\pgfusepath{stroke}%
\end{pgfscope}%
\begin{pgfscope}%
\definecolor{textcolor}{rgb}{0.150000,0.150000,0.150000}%
\pgfsetstrokecolor{textcolor}%
\pgfsetfillcolor{textcolor}%
\pgftext[x=0.520798in,y=0.351899in,,top]{\color{textcolor}\rmfamily\fontsize{8.000000}{9.600000}\selectfont \(\displaystyle {0}\)}%
\end{pgfscope}%
\begin{pgfscope}%
\pgfpathrectangle{\pgfqpoint{0.520798in}{0.442177in}}{\pgfqpoint{0.719005in}{0.871884in}}%
\pgfusepath{clip}%
\pgfsetroundcap%
\pgfsetroundjoin%
\pgfsetlinewidth{0.501875pt}%
\definecolor{currentstroke}{rgb}{0.800000,0.800000,0.800000}%
\pgfsetstrokecolor{currentstroke}%
\pgfsetdash{}{0pt}%
\pgfpathmoveto{\pgfqpoint{0.880300in}{0.442177in}}%
\pgfpathlineto{\pgfqpoint{0.880300in}{1.314061in}}%
\pgfusepath{stroke}%
\end{pgfscope}%
\begin{pgfscope}%
\definecolor{textcolor}{rgb}{0.150000,0.150000,0.150000}%
\pgfsetstrokecolor{textcolor}%
\pgfsetfillcolor{textcolor}%
\pgftext[x=0.880300in,y=0.351899in,,top]{\color{textcolor}\rmfamily\fontsize{8.000000}{9.600000}\selectfont \(\displaystyle {5}\)}%
\end{pgfscope}%
\begin{pgfscope}%
\pgfpathrectangle{\pgfqpoint{0.520798in}{0.442177in}}{\pgfqpoint{0.719005in}{0.871884in}}%
\pgfusepath{clip}%
\pgfsetroundcap%
\pgfsetroundjoin%
\pgfsetlinewidth{0.501875pt}%
\definecolor{currentstroke}{rgb}{0.800000,0.800000,0.800000}%
\pgfsetstrokecolor{currentstroke}%
\pgfsetdash{}{0pt}%
\pgfpathmoveto{\pgfqpoint{1.239803in}{0.442177in}}%
\pgfpathlineto{\pgfqpoint{1.239803in}{1.314061in}}%
\pgfusepath{stroke}%
\end{pgfscope}%
\begin{pgfscope}%
\definecolor{textcolor}{rgb}{0.150000,0.150000,0.150000}%
\pgfsetstrokecolor{textcolor}%
\pgfsetfillcolor{textcolor}%
\pgftext[x=1.239803in,y=0.351899in,,top]{\color{textcolor}\rmfamily\fontsize{8.000000}{9.600000}\selectfont \(\displaystyle {10}\)}%
\end{pgfscope}%
\begin{pgfscope}%
\definecolor{textcolor}{rgb}{0.150000,0.150000,0.150000}%
\pgfsetstrokecolor{textcolor}%
\pgfsetfillcolor{textcolor}%
\pgftext[x=0.880300in,y=0.198219in,,top]{\color{textcolor}\rmfamily\fontsize{10.000000}{12.000000}\selectfont Edit distance \(\displaystyle \epsilon\)}%
\end{pgfscope}%
\begin{pgfscope}%
\pgfpathrectangle{\pgfqpoint{0.520798in}{0.442177in}}{\pgfqpoint{0.719005in}{0.871884in}}%
\pgfusepath{clip}%
\pgfsetroundcap%
\pgfsetroundjoin%
\pgfsetlinewidth{0.501875pt}%
\definecolor{currentstroke}{rgb}{0.800000,0.800000,0.800000}%
\pgfsetstrokecolor{currentstroke}%
\pgfsetdash{}{0pt}%
\pgfpathmoveto{\pgfqpoint{0.520798in}{0.442177in}}%
\pgfpathlineto{\pgfqpoint{1.239803in}{0.442177in}}%
\pgfusepath{stroke}%
\end{pgfscope}%
\begin{pgfscope}%
\definecolor{textcolor}{rgb}{0.150000,0.150000,0.150000}%
\pgfsetstrokecolor{textcolor}%
\pgfsetfillcolor{textcolor}%
\pgftext[x=0.273151in, y=0.403915in, left, base]{\color{textcolor}\rmfamily\fontsize{8.000000}{9.600000}\selectfont 0\%}%
\end{pgfscope}%
\begin{pgfscope}%
\pgfpathrectangle{\pgfqpoint{0.520798in}{0.442177in}}{\pgfqpoint{0.719005in}{0.871884in}}%
\pgfusepath{clip}%
\pgfsetroundcap%
\pgfsetroundjoin%
\pgfsetlinewidth{0.501875pt}%
\definecolor{currentstroke}{rgb}{0.800000,0.800000,0.800000}%
\pgfsetstrokecolor{currentstroke}%
\pgfsetdash{}{0pt}%
\pgfpathmoveto{\pgfqpoint{0.520798in}{0.660148in}}%
\pgfpathlineto{\pgfqpoint{1.239803in}{0.660148in}}%
\pgfusepath{stroke}%
\end{pgfscope}%
\begin{pgfscope}%
\definecolor{textcolor}{rgb}{0.150000,0.150000,0.150000}%
\pgfsetstrokecolor{textcolor}%
\pgfsetfillcolor{textcolor}%
\pgftext[x=0.214138in, y=0.621886in, left, base]{\color{textcolor}\rmfamily\fontsize{8.000000}{9.600000}\selectfont 25\%}%
\end{pgfscope}%
\begin{pgfscope}%
\pgfpathrectangle{\pgfqpoint{0.520798in}{0.442177in}}{\pgfqpoint{0.719005in}{0.871884in}}%
\pgfusepath{clip}%
\pgfsetroundcap%
\pgfsetroundjoin%
\pgfsetlinewidth{0.501875pt}%
\definecolor{currentstroke}{rgb}{0.800000,0.800000,0.800000}%
\pgfsetstrokecolor{currentstroke}%
\pgfsetdash{}{0pt}%
\pgfpathmoveto{\pgfqpoint{0.520798in}{0.878119in}}%
\pgfpathlineto{\pgfqpoint{1.239803in}{0.878119in}}%
\pgfusepath{stroke}%
\end{pgfscope}%
\begin{pgfscope}%
\definecolor{textcolor}{rgb}{0.150000,0.150000,0.150000}%
\pgfsetstrokecolor{textcolor}%
\pgfsetfillcolor{textcolor}%
\pgftext[x=0.214138in, y=0.839857in, left, base]{\color{textcolor}\rmfamily\fontsize{8.000000}{9.600000}\selectfont 50\%}%
\end{pgfscope}%
\begin{pgfscope}%
\pgfpathrectangle{\pgfqpoint{0.520798in}{0.442177in}}{\pgfqpoint{0.719005in}{0.871884in}}%
\pgfusepath{clip}%
\pgfsetroundcap%
\pgfsetroundjoin%
\pgfsetlinewidth{0.501875pt}%
\definecolor{currentstroke}{rgb}{0.800000,0.800000,0.800000}%
\pgfsetstrokecolor{currentstroke}%
\pgfsetdash{}{0pt}%
\pgfpathmoveto{\pgfqpoint{0.520798in}{1.096090in}}%
\pgfpathlineto{\pgfqpoint{1.239803in}{1.096090in}}%
\pgfusepath{stroke}%
\end{pgfscope}%
\begin{pgfscope}%
\definecolor{textcolor}{rgb}{0.150000,0.150000,0.150000}%
\pgfsetstrokecolor{textcolor}%
\pgfsetfillcolor{textcolor}%
\pgftext[x=0.214138in, y=1.057828in, left, base]{\color{textcolor}\rmfamily\fontsize{8.000000}{9.600000}\selectfont 75\%}%
\end{pgfscope}%
\begin{pgfscope}%
\pgfpathrectangle{\pgfqpoint{0.520798in}{0.442177in}}{\pgfqpoint{0.719005in}{0.871884in}}%
\pgfusepath{clip}%
\pgfsetroundcap%
\pgfsetroundjoin%
\pgfsetlinewidth{0.501875pt}%
\definecolor{currentstroke}{rgb}{0.800000,0.800000,0.800000}%
\pgfsetstrokecolor{currentstroke}%
\pgfsetdash{}{0pt}%
\pgfpathmoveto{\pgfqpoint{0.520798in}{1.314061in}}%
\pgfpathlineto{\pgfqpoint{1.239803in}{1.314061in}}%
\pgfusepath{stroke}%
\end{pgfscope}%
\begin{pgfscope}%
\definecolor{textcolor}{rgb}{0.150000,0.150000,0.150000}%
\pgfsetstrokecolor{textcolor}%
\pgfsetfillcolor{textcolor}%
\pgftext[x=0.155124in, y=1.275799in, left, base]{\color{textcolor}\rmfamily\fontsize{8.000000}{9.600000}\selectfont 100\%}%
\end{pgfscope}%
\begin{pgfscope}%
\definecolor{textcolor}{rgb}{0.150000,0.150000,0.150000}%
\pgfsetstrokecolor{textcolor}%
\pgfsetfillcolor{textcolor}%
\pgftext[x=0.099569in,y=0.878119in,,bottom,rotate=90.000000]{\color{textcolor}\rmfamily\fontsize{10.000000}{12.000000}\selectfont Cert. Acc.}%
\end{pgfscope}%
\begin{pgfscope}%
\pgfsetrectcap%
\pgfsetmiterjoin%
\pgfsetlinewidth{0.752812pt}%
\definecolor{currentstroke}{rgb}{0.700000,0.700000,0.700000}%
\pgfsetstrokecolor{currentstroke}%
\pgfsetdash{}{0pt}%
\pgfpathmoveto{\pgfqpoint{0.520798in}{0.442177in}}%
\pgfpathlineto{\pgfqpoint{0.520798in}{1.314061in}}%
\pgfusepath{stroke}%
\end{pgfscope}%
\begin{pgfscope}%
\pgfsetrectcap%
\pgfsetmiterjoin%
\pgfsetlinewidth{0.752812pt}%
\definecolor{currentstroke}{rgb}{0.700000,0.700000,0.700000}%
\pgfsetstrokecolor{currentstroke}%
\pgfsetdash{}{0pt}%
\pgfpathmoveto{\pgfqpoint{1.239803in}{0.442177in}}%
\pgfpathlineto{\pgfqpoint{1.239803in}{1.314061in}}%
\pgfusepath{stroke}%
\end{pgfscope}%
\begin{pgfscope}%
\pgfsetrectcap%
\pgfsetmiterjoin%
\pgfsetlinewidth{0.752812pt}%
\definecolor{currentstroke}{rgb}{0.700000,0.700000,0.700000}%
\pgfsetstrokecolor{currentstroke}%
\pgfsetdash{}{0pt}%
\pgfpathmoveto{\pgfqpoint{0.520798in}{0.442177in}}%
\pgfpathlineto{\pgfqpoint{1.239803in}{0.442177in}}%
\pgfusepath{stroke}%
\end{pgfscope}%
\begin{pgfscope}%
\pgfsetrectcap%
\pgfsetmiterjoin%
\pgfsetlinewidth{0.752812pt}%
\definecolor{currentstroke}{rgb}{0.700000,0.700000,0.700000}%
\pgfsetstrokecolor{currentstroke}%
\pgfsetdash{}{0pt}%
\pgfpathmoveto{\pgfqpoint{0.520798in}{1.314061in}}%
\pgfpathlineto{\pgfqpoint{1.239803in}{1.314061in}}%
\pgfusepath{stroke}%
\end{pgfscope}%
\begin{pgfscope}%
\pgfpathrectangle{\pgfqpoint{0.520798in}{0.442177in}}{\pgfqpoint{0.719005in}{0.871884in}}%
\pgfusepath{clip}%
\pgfsetroundcap%
\pgfsetroundjoin%
\pgfsetlinewidth{1.003750pt}%
\definecolor{currentstroke}{rgb}{0.003922,0.450980,0.698039}%
\pgfsetstrokecolor{currentstroke}%
\pgfsetdash{}{0pt}%
\pgfpathmoveto{\pgfqpoint{0.520798in}{0.969018in}}%
\pgfpathlineto{\pgfqpoint{0.556748in}{0.969018in}}%
\pgfpathlineto{\pgfqpoint{0.556748in}{0.860960in}}%
\pgfpathlineto{\pgfqpoint{0.628649in}{0.860960in}}%
\pgfpathlineto{\pgfqpoint{0.628649in}{0.679441in}}%
\pgfpathlineto{\pgfqpoint{0.700549in}{0.679441in}}%
\pgfpathlineto{\pgfqpoint{0.700549in}{0.469354in}}%
\pgfpathlineto{\pgfqpoint{0.772450in}{0.469354in}}%
\pgfpathlineto{\pgfqpoint{0.772450in}{0.448113in}}%
\pgfpathlineto{\pgfqpoint{0.844350in}{0.448113in}}%
\pgfpathlineto{\pgfqpoint{0.844350in}{0.445145in}}%
\pgfpathlineto{\pgfqpoint{0.916251in}{0.445145in}}%
\pgfpathlineto{\pgfqpoint{0.916251in}{0.444310in}}%
\pgfpathlineto{\pgfqpoint{0.988151in}{0.444310in}}%
\pgfpathlineto{\pgfqpoint{0.988151in}{0.443568in}}%
\pgfpathlineto{\pgfqpoint{1.060052in}{0.443568in}}%
\pgfpathlineto{\pgfqpoint{1.060052in}{0.442548in}}%
\pgfpathlineto{\pgfqpoint{1.131952in}{0.442548in}}%
\pgfpathlineto{\pgfqpoint{1.131952in}{0.442362in}}%
\pgfpathlineto{\pgfqpoint{1.239803in}{0.442362in}}%
\pgfpathlineto{\pgfqpoint{1.239803in}{0.442270in}}%
\pgfpathlineto{\pgfqpoint{1.241470in}{0.442270in}}%
\pgfusepath{stroke}%
\end{pgfscope}%
\begin{pgfscope}%
\pgfpathrectangle{\pgfqpoint{0.520798in}{0.442177in}}{\pgfqpoint{0.719005in}{0.871884in}}%
\pgfusepath{clip}%
\pgfsetbuttcap%
\pgfsetroundjoin%
\definecolor{currentfill}{rgb}{0.003922,0.450980,0.698039}%
\pgfsetfillcolor{currentfill}%
\pgfsetfillopacity{0.500000}%
\pgfsetlinewidth{0.000000pt}%
\definecolor{currentstroke}{rgb}{0.003922,0.450980,0.698039}%
\pgfsetstrokecolor{currentstroke}%
\pgfsetstrokeopacity{0.500000}%
\pgfsetdash{}{0pt}%
\pgfpathmoveto{\pgfqpoint{0.520798in}{0.987622in}}%
\pgfpathlineto{\pgfqpoint{0.520798in}{0.950414in}}%
\pgfpathlineto{\pgfqpoint{0.556748in}{0.950414in}}%
\pgfpathlineto{\pgfqpoint{0.556748in}{0.839442in}}%
\pgfpathlineto{\pgfqpoint{0.628649in}{0.839442in}}%
\pgfpathlineto{\pgfqpoint{0.628649in}{0.664144in}}%
\pgfpathlineto{\pgfqpoint{0.700549in}{0.664144in}}%
\pgfpathlineto{\pgfqpoint{0.700549in}{0.465489in}}%
\pgfpathlineto{\pgfqpoint{0.772450in}{0.465489in}}%
\pgfpathlineto{\pgfqpoint{0.772450in}{0.447080in}}%
\pgfpathlineto{\pgfqpoint{0.844350in}{0.447080in}}%
\pgfpathlineto{\pgfqpoint{0.844350in}{0.444451in}}%
\pgfpathlineto{\pgfqpoint{0.916251in}{0.444451in}}%
\pgfpathlineto{\pgfqpoint{0.916251in}{0.443502in}}%
\pgfpathlineto{\pgfqpoint{0.988151in}{0.443502in}}%
\pgfpathlineto{\pgfqpoint{0.988151in}{0.442912in}}%
\pgfpathlineto{\pgfqpoint{1.060052in}{0.442912in}}%
\pgfpathlineto{\pgfqpoint{1.060052in}{0.442362in}}%
\pgfpathlineto{\pgfqpoint{1.131952in}{0.442362in}}%
\pgfpathlineto{\pgfqpoint{1.131952in}{0.442135in}}%
\pgfpathlineto{\pgfqpoint{1.239803in}{0.442135in}}%
\pgfpathlineto{\pgfqpoint{1.239803in}{0.442084in}}%
\pgfpathlineto{\pgfqpoint{1.383604in}{0.442084in}}%
\pgfpathlineto{\pgfqpoint{1.383604in}{0.442177in}}%
\pgfpathlineto{\pgfqpoint{2.066659in}{0.442177in}}%
\pgfpathlineto{\pgfqpoint{2.066659in}{0.442177in}}%
\pgfpathlineto{\pgfqpoint{2.677813in}{0.442177in}}%
\pgfpathlineto{\pgfqpoint{2.677813in}{0.442177in}}%
\pgfpathlineto{\pgfqpoint{2.677813in}{0.442177in}}%
\pgfpathlineto{\pgfqpoint{2.066659in}{0.442177in}}%
\pgfpathlineto{\pgfqpoint{2.066659in}{0.442177in}}%
\pgfpathlineto{\pgfqpoint{1.383604in}{0.442177in}}%
\pgfpathlineto{\pgfqpoint{1.383604in}{0.442455in}}%
\pgfpathlineto{\pgfqpoint{1.239803in}{0.442455in}}%
\pgfpathlineto{\pgfqpoint{1.239803in}{0.442590in}}%
\pgfpathlineto{\pgfqpoint{1.131952in}{0.442590in}}%
\pgfpathlineto{\pgfqpoint{1.131952in}{0.442733in}}%
\pgfpathlineto{\pgfqpoint{1.060052in}{0.442733in}}%
\pgfpathlineto{\pgfqpoint{1.060052in}{0.444224in}}%
\pgfpathlineto{\pgfqpoint{0.988151in}{0.444224in}}%
\pgfpathlineto{\pgfqpoint{0.988151in}{0.445119in}}%
\pgfpathlineto{\pgfqpoint{0.916251in}{0.445119in}}%
\pgfpathlineto{\pgfqpoint{0.916251in}{0.445839in}}%
\pgfpathlineto{\pgfqpoint{0.844350in}{0.445839in}}%
\pgfpathlineto{\pgfqpoint{0.844350in}{0.449146in}}%
\pgfpathlineto{\pgfqpoint{0.772450in}{0.449146in}}%
\pgfpathlineto{\pgfqpoint{0.772450in}{0.473218in}}%
\pgfpathlineto{\pgfqpoint{0.700549in}{0.473218in}}%
\pgfpathlineto{\pgfqpoint{0.700549in}{0.694737in}}%
\pgfpathlineto{\pgfqpoint{0.628649in}{0.694737in}}%
\pgfpathlineto{\pgfqpoint{0.628649in}{0.882478in}}%
\pgfpathlineto{\pgfqpoint{0.556748in}{0.882478in}}%
\pgfpathlineto{\pgfqpoint{0.556748in}{0.987622in}}%
\pgfpathlineto{\pgfqpoint{0.520798in}{0.987622in}}%
\pgfpathlineto{\pgfqpoint{0.520798in}{0.987622in}}%
\pgfpathclose%
\pgfusepath{fill}%
\end{pgfscope}%
\begin{pgfscope}%
\pgfpathrectangle{\pgfqpoint{0.520798in}{0.442177in}}{\pgfqpoint{0.719005in}{0.871884in}}%
\pgfusepath{clip}%
\pgfsetroundcap%
\pgfsetroundjoin%
\pgfsetlinewidth{1.003750pt}%
\definecolor{currentstroke}{rgb}{0.870588,0.560784,0.019608}%
\pgfsetstrokecolor{currentstroke}%
\pgfsetdash{}{0pt}%
\pgfpathmoveto{\pgfqpoint{0.520798in}{0.969018in}}%
\pgfpathlineto{\pgfqpoint{0.556748in}{0.969018in}}%
\pgfpathlineto{\pgfqpoint{0.556748in}{0.860960in}}%
\pgfpathlineto{\pgfqpoint{0.628649in}{0.860960in}}%
\pgfpathlineto{\pgfqpoint{0.628649in}{0.679441in}}%
\pgfpathlineto{\pgfqpoint{0.700549in}{0.679441in}}%
\pgfpathlineto{\pgfqpoint{0.700549in}{0.469354in}}%
\pgfpathlineto{\pgfqpoint{0.772450in}{0.469354in}}%
\pgfpathlineto{\pgfqpoint{0.772450in}{0.448113in}}%
\pgfpathlineto{\pgfqpoint{0.844350in}{0.448113in}}%
\pgfpathlineto{\pgfqpoint{0.844350in}{0.445238in}}%
\pgfpathlineto{\pgfqpoint{0.916251in}{0.445238in}}%
\pgfpathlineto{\pgfqpoint{0.916251in}{0.444496in}}%
\pgfpathlineto{\pgfqpoint{0.988151in}{0.444496in}}%
\pgfpathlineto{\pgfqpoint{0.988151in}{0.443568in}}%
\pgfpathlineto{\pgfqpoint{1.060052in}{0.443568in}}%
\pgfpathlineto{\pgfqpoint{1.060052in}{0.442548in}}%
\pgfpathlineto{\pgfqpoint{1.131952in}{0.442548in}}%
\pgfpathlineto{\pgfqpoint{1.131952in}{0.442362in}}%
\pgfpathlineto{\pgfqpoint{1.239803in}{0.442362in}}%
\pgfpathlineto{\pgfqpoint{1.239803in}{0.442270in}}%
\pgfpathlineto{\pgfqpoint{1.241470in}{0.442270in}}%
\pgfusepath{stroke}%
\end{pgfscope}%
\begin{pgfscope}%
\pgfpathrectangle{\pgfqpoint{0.520798in}{0.442177in}}{\pgfqpoint{0.719005in}{0.871884in}}%
\pgfusepath{clip}%
\pgfsetbuttcap%
\pgfsetroundjoin%
\definecolor{currentfill}{rgb}{0.870588,0.560784,0.019608}%
\pgfsetfillcolor{currentfill}%
\pgfsetfillopacity{0.500000}%
\pgfsetlinewidth{0.000000pt}%
\definecolor{currentstroke}{rgb}{0.870588,0.560784,0.019608}%
\pgfsetstrokecolor{currentstroke}%
\pgfsetstrokeopacity{0.500000}%
\pgfsetdash{}{0pt}%
\pgfpathmoveto{\pgfqpoint{0.520798in}{0.987622in}}%
\pgfpathlineto{\pgfqpoint{0.520798in}{0.950414in}}%
\pgfpathlineto{\pgfqpoint{0.556748in}{0.950414in}}%
\pgfpathlineto{\pgfqpoint{0.556748in}{0.839442in}}%
\pgfpathlineto{\pgfqpoint{0.628649in}{0.839442in}}%
\pgfpathlineto{\pgfqpoint{0.628649in}{0.664144in}}%
\pgfpathlineto{\pgfqpoint{0.700549in}{0.664144in}}%
\pgfpathlineto{\pgfqpoint{0.700549in}{0.465489in}}%
\pgfpathlineto{\pgfqpoint{0.772450in}{0.465489in}}%
\pgfpathlineto{\pgfqpoint{0.772450in}{0.447080in}}%
\pgfpathlineto{\pgfqpoint{0.844350in}{0.447080in}}%
\pgfpathlineto{\pgfqpoint{0.844350in}{0.444544in}}%
\pgfpathlineto{\pgfqpoint{0.916251in}{0.444544in}}%
\pgfpathlineto{\pgfqpoint{0.916251in}{0.443480in}}%
\pgfpathlineto{\pgfqpoint{0.988151in}{0.443480in}}%
\pgfpathlineto{\pgfqpoint{0.988151in}{0.442912in}}%
\pgfpathlineto{\pgfqpoint{1.060052in}{0.442912in}}%
\pgfpathlineto{\pgfqpoint{1.060052in}{0.442362in}}%
\pgfpathlineto{\pgfqpoint{1.131952in}{0.442362in}}%
\pgfpathlineto{\pgfqpoint{1.131952in}{0.442135in}}%
\pgfpathlineto{\pgfqpoint{1.239803in}{0.442135in}}%
\pgfpathlineto{\pgfqpoint{1.239803in}{0.442084in}}%
\pgfpathlineto{\pgfqpoint{1.383604in}{0.442084in}}%
\pgfpathlineto{\pgfqpoint{1.383604in}{0.442177in}}%
\pgfpathlineto{\pgfqpoint{2.066659in}{0.442177in}}%
\pgfpathlineto{\pgfqpoint{2.066659in}{0.442177in}}%
\pgfpathlineto{\pgfqpoint{2.677813in}{0.442177in}}%
\pgfpathlineto{\pgfqpoint{2.677813in}{0.442177in}}%
\pgfpathlineto{\pgfqpoint{2.677813in}{0.442177in}}%
\pgfpathlineto{\pgfqpoint{2.066659in}{0.442177in}}%
\pgfpathlineto{\pgfqpoint{2.066659in}{0.442177in}}%
\pgfpathlineto{\pgfqpoint{1.383604in}{0.442177in}}%
\pgfpathlineto{\pgfqpoint{1.383604in}{0.442455in}}%
\pgfpathlineto{\pgfqpoint{1.239803in}{0.442455in}}%
\pgfpathlineto{\pgfqpoint{1.239803in}{0.442590in}}%
\pgfpathlineto{\pgfqpoint{1.131952in}{0.442590in}}%
\pgfpathlineto{\pgfqpoint{1.131952in}{0.442733in}}%
\pgfpathlineto{\pgfqpoint{1.060052in}{0.442733in}}%
\pgfpathlineto{\pgfqpoint{1.060052in}{0.444224in}}%
\pgfpathlineto{\pgfqpoint{0.988151in}{0.444224in}}%
\pgfpathlineto{\pgfqpoint{0.988151in}{0.445512in}}%
\pgfpathlineto{\pgfqpoint{0.916251in}{0.445512in}}%
\pgfpathlineto{\pgfqpoint{0.916251in}{0.445932in}}%
\pgfpathlineto{\pgfqpoint{0.844350in}{0.445932in}}%
\pgfpathlineto{\pgfqpoint{0.844350in}{0.449146in}}%
\pgfpathlineto{\pgfqpoint{0.772450in}{0.449146in}}%
\pgfpathlineto{\pgfqpoint{0.772450in}{0.473218in}}%
\pgfpathlineto{\pgfqpoint{0.700549in}{0.473218in}}%
\pgfpathlineto{\pgfqpoint{0.700549in}{0.694737in}}%
\pgfpathlineto{\pgfqpoint{0.628649in}{0.694737in}}%
\pgfpathlineto{\pgfqpoint{0.628649in}{0.882478in}}%
\pgfpathlineto{\pgfqpoint{0.556748in}{0.882478in}}%
\pgfpathlineto{\pgfqpoint{0.556748in}{0.987622in}}%
\pgfpathlineto{\pgfqpoint{0.520798in}{0.987622in}}%
\pgfpathlineto{\pgfqpoint{0.520798in}{0.987622in}}%
\pgfpathclose%
\pgfusepath{fill}%
\end{pgfscope}%
\begin{pgfscope}%
\pgfpathrectangle{\pgfqpoint{0.520798in}{0.442177in}}{\pgfqpoint{0.719005in}{0.871884in}}%
\pgfusepath{clip}%
\pgfsetroundcap%
\pgfsetroundjoin%
\pgfsetlinewidth{1.003750pt}%
\definecolor{currentstroke}{rgb}{0.007843,0.619608,0.450980}%
\pgfsetstrokecolor{currentstroke}%
\pgfsetdash{}{0pt}%
\pgfpathmoveto{\pgfqpoint{0.520798in}{0.969018in}}%
\pgfpathlineto{\pgfqpoint{0.556748in}{0.969018in}}%
\pgfpathlineto{\pgfqpoint{0.556748in}{0.860960in}}%
\pgfpathlineto{\pgfqpoint{0.628649in}{0.860960in}}%
\pgfpathlineto{\pgfqpoint{0.628649in}{0.679441in}}%
\pgfpathlineto{\pgfqpoint{0.700549in}{0.679441in}}%
\pgfpathlineto{\pgfqpoint{0.700549in}{0.469354in}}%
\pgfpathlineto{\pgfqpoint{0.772450in}{0.469354in}}%
\pgfpathlineto{\pgfqpoint{0.772450in}{0.448113in}}%
\pgfpathlineto{\pgfqpoint{0.844350in}{0.448113in}}%
\pgfpathlineto{\pgfqpoint{0.844350in}{0.445238in}}%
\pgfpathlineto{\pgfqpoint{0.916251in}{0.445238in}}%
\pgfpathlineto{\pgfqpoint{0.916251in}{0.444496in}}%
\pgfpathlineto{\pgfqpoint{0.988151in}{0.444496in}}%
\pgfpathlineto{\pgfqpoint{0.988151in}{0.443568in}}%
\pgfpathlineto{\pgfqpoint{1.060052in}{0.443568in}}%
\pgfpathlineto{\pgfqpoint{1.060052in}{0.442548in}}%
\pgfpathlineto{\pgfqpoint{1.131952in}{0.442548in}}%
\pgfpathlineto{\pgfqpoint{1.131952in}{0.442362in}}%
\pgfpathlineto{\pgfqpoint{1.239803in}{0.442362in}}%
\pgfpathlineto{\pgfqpoint{1.239803in}{0.442270in}}%
\pgfpathlineto{\pgfqpoint{1.241470in}{0.442270in}}%
\pgfusepath{stroke}%
\end{pgfscope}%
\begin{pgfscope}%
\pgfpathrectangle{\pgfqpoint{0.520798in}{0.442177in}}{\pgfqpoint{0.719005in}{0.871884in}}%
\pgfusepath{clip}%
\pgfsetbuttcap%
\pgfsetroundjoin%
\definecolor{currentfill}{rgb}{0.007843,0.619608,0.450980}%
\pgfsetfillcolor{currentfill}%
\pgfsetfillopacity{0.500000}%
\pgfsetlinewidth{0.000000pt}%
\definecolor{currentstroke}{rgb}{0.007843,0.619608,0.450980}%
\pgfsetstrokecolor{currentstroke}%
\pgfsetstrokeopacity{0.500000}%
\pgfsetdash{}{0pt}%
\pgfpathmoveto{\pgfqpoint{0.520798in}{0.987622in}}%
\pgfpathlineto{\pgfqpoint{0.520798in}{0.950414in}}%
\pgfpathlineto{\pgfqpoint{0.556748in}{0.950414in}}%
\pgfpathlineto{\pgfqpoint{0.556748in}{0.839442in}}%
\pgfpathlineto{\pgfqpoint{0.628649in}{0.839442in}}%
\pgfpathlineto{\pgfqpoint{0.628649in}{0.664144in}}%
\pgfpathlineto{\pgfqpoint{0.700549in}{0.664144in}}%
\pgfpathlineto{\pgfqpoint{0.700549in}{0.465489in}}%
\pgfpathlineto{\pgfqpoint{0.772450in}{0.465489in}}%
\pgfpathlineto{\pgfqpoint{0.772450in}{0.447080in}}%
\pgfpathlineto{\pgfqpoint{0.844350in}{0.447080in}}%
\pgfpathlineto{\pgfqpoint{0.844350in}{0.444544in}}%
\pgfpathlineto{\pgfqpoint{0.916251in}{0.444544in}}%
\pgfpathlineto{\pgfqpoint{0.916251in}{0.443480in}}%
\pgfpathlineto{\pgfqpoint{0.988151in}{0.443480in}}%
\pgfpathlineto{\pgfqpoint{0.988151in}{0.442912in}}%
\pgfpathlineto{\pgfqpoint{1.060052in}{0.442912in}}%
\pgfpathlineto{\pgfqpoint{1.060052in}{0.442362in}}%
\pgfpathlineto{\pgfqpoint{1.131952in}{0.442362in}}%
\pgfpathlineto{\pgfqpoint{1.131952in}{0.442135in}}%
\pgfpathlineto{\pgfqpoint{1.239803in}{0.442135in}}%
\pgfpathlineto{\pgfqpoint{1.239803in}{0.442084in}}%
\pgfpathlineto{\pgfqpoint{1.383604in}{0.442084in}}%
\pgfpathlineto{\pgfqpoint{1.383604in}{0.442177in}}%
\pgfpathlineto{\pgfqpoint{2.066659in}{0.442177in}}%
\pgfpathlineto{\pgfqpoint{2.066659in}{0.442177in}}%
\pgfpathlineto{\pgfqpoint{2.677813in}{0.442177in}}%
\pgfpathlineto{\pgfqpoint{2.677813in}{0.442177in}}%
\pgfpathlineto{\pgfqpoint{2.677813in}{0.442177in}}%
\pgfpathlineto{\pgfqpoint{2.066659in}{0.442177in}}%
\pgfpathlineto{\pgfqpoint{2.066659in}{0.442177in}}%
\pgfpathlineto{\pgfqpoint{1.383604in}{0.442177in}}%
\pgfpathlineto{\pgfqpoint{1.383604in}{0.442455in}}%
\pgfpathlineto{\pgfqpoint{1.239803in}{0.442455in}}%
\pgfpathlineto{\pgfqpoint{1.239803in}{0.442590in}}%
\pgfpathlineto{\pgfqpoint{1.131952in}{0.442590in}}%
\pgfpathlineto{\pgfqpoint{1.131952in}{0.442733in}}%
\pgfpathlineto{\pgfqpoint{1.060052in}{0.442733in}}%
\pgfpathlineto{\pgfqpoint{1.060052in}{0.444224in}}%
\pgfpathlineto{\pgfqpoint{0.988151in}{0.444224in}}%
\pgfpathlineto{\pgfqpoint{0.988151in}{0.445512in}}%
\pgfpathlineto{\pgfqpoint{0.916251in}{0.445512in}}%
\pgfpathlineto{\pgfqpoint{0.916251in}{0.445932in}}%
\pgfpathlineto{\pgfqpoint{0.844350in}{0.445932in}}%
\pgfpathlineto{\pgfqpoint{0.844350in}{0.449146in}}%
\pgfpathlineto{\pgfqpoint{0.772450in}{0.449146in}}%
\pgfpathlineto{\pgfqpoint{0.772450in}{0.473218in}}%
\pgfpathlineto{\pgfqpoint{0.700549in}{0.473218in}}%
\pgfpathlineto{\pgfqpoint{0.700549in}{0.694737in}}%
\pgfpathlineto{\pgfqpoint{0.628649in}{0.694737in}}%
\pgfpathlineto{\pgfqpoint{0.628649in}{0.882478in}}%
\pgfpathlineto{\pgfqpoint{0.556748in}{0.882478in}}%
\pgfpathlineto{\pgfqpoint{0.556748in}{0.987622in}}%
\pgfpathlineto{\pgfqpoint{0.520798in}{0.987622in}}%
\pgfpathlineto{\pgfqpoint{0.520798in}{0.987622in}}%
\pgfpathclose%
\pgfusepath{fill}%
\end{pgfscope}%
\begin{pgfscope}%
\definecolor{textcolor}{rgb}{0.150000,0.150000,0.150000}%
\pgfsetstrokecolor{textcolor}%
\pgfsetfillcolor{textcolor}%
\pgftext[x=0.880300in,y=1.397395in,,base]{\color{textcolor}\rmfamily\fontsize{9.000000}{10.800000}\selectfont \(\displaystyle c_A^+=1\)}%
\end{pgfscope}%
\begin{pgfscope}%
\pgfsetbuttcap%
\pgfsetmiterjoin%
\definecolor{currentfill}{rgb}{1.000000,1.000000,1.000000}%
\pgfsetfillcolor{currentfill}%
\pgfsetfillopacity{0.800000}%
\pgfsetlinewidth{1.003750pt}%
\definecolor{currentstroke}{rgb}{0.800000,0.800000,0.800000}%
\pgfsetstrokecolor{currentstroke}%
\pgfsetstrokeopacity{0.800000}%
\pgfsetdash{}{0pt}%
\pgfpathmoveto{\pgfqpoint{0.785843in}{0.638103in}}%
\pgfpathlineto{\pgfqpoint{1.191192in}{0.638103in}}%
\pgfpathlineto{\pgfqpoint{1.191192in}{1.265450in}}%
\pgfpathlineto{\pgfqpoint{0.785843in}{1.265450in}}%
\pgfpathlineto{\pgfqpoint{0.785843in}{0.638103in}}%
\pgfpathclose%
\pgfusepath{stroke,fill}%
\end{pgfscope}%
\begin{pgfscope}%
\definecolor{textcolor}{rgb}{0.150000,0.150000,0.150000}%
\pgfsetstrokecolor{textcolor}%
\pgfsetfillcolor{textcolor}%
\pgftext[x=0.917113in,y=1.113275in,left,base]{\color{textcolor}\rmfamily\fontsize{9.000000}{10.800000}\selectfont \(\displaystyle c_A^-\)}%
\end{pgfscope}%
\begin{pgfscope}%
\pgfsetroundcap%
\pgfsetroundjoin%
\pgfsetlinewidth{1.003750pt}%
\definecolor{currentstroke}{rgb}{0.003922,0.450980,0.698039}%
\pgfsetstrokecolor{currentstroke}%
\pgfsetdash{}{0pt}%
\pgfpathmoveto{\pgfqpoint{0.824732in}{1.001052in}}%
\pgfpathlineto{\pgfqpoint{0.873343in}{1.001052in}}%
\pgfpathlineto{\pgfqpoint{0.873343in}{1.001052in}}%
\pgfpathlineto{\pgfqpoint{0.970565in}{1.001052in}}%
\pgfpathlineto{\pgfqpoint{0.970565in}{1.001052in}}%
\pgfpathlineto{\pgfqpoint{1.019176in}{1.001052in}}%
\pgfusepath{stroke}%
\end{pgfscope}%
\begin{pgfscope}%
\definecolor{textcolor}{rgb}{0.150000,0.150000,0.150000}%
\pgfsetstrokecolor{textcolor}%
\pgfsetfillcolor{textcolor}%
\pgftext[x=1.096954in,y=0.967024in,left,base]{\color{textcolor}\rmfamily\fontsize{7.000000}{8.400000}\selectfont 1}%
\end{pgfscope}%
\begin{pgfscope}%
\pgfsetroundcap%
\pgfsetroundjoin%
\pgfsetlinewidth{1.003750pt}%
\definecolor{currentstroke}{rgb}{0.870588,0.560784,0.019608}%
\pgfsetstrokecolor{currentstroke}%
\pgfsetdash{}{0pt}%
\pgfpathmoveto{\pgfqpoint{0.824732in}{0.865486in}}%
\pgfpathlineto{\pgfqpoint{0.873343in}{0.865486in}}%
\pgfpathlineto{\pgfqpoint{0.873343in}{0.865486in}}%
\pgfpathlineto{\pgfqpoint{0.970565in}{0.865486in}}%
\pgfpathlineto{\pgfqpoint{0.970565in}{0.865486in}}%
\pgfpathlineto{\pgfqpoint{1.019176in}{0.865486in}}%
\pgfusepath{stroke}%
\end{pgfscope}%
\begin{pgfscope}%
\definecolor{textcolor}{rgb}{0.150000,0.150000,0.150000}%
\pgfsetstrokecolor{textcolor}%
\pgfsetfillcolor{textcolor}%
\pgftext[x=1.096954in,y=0.831458in,left,base]{\color{textcolor}\rmfamily\fontsize{7.000000}{8.400000}\selectfont 2}%
\end{pgfscope}%
\begin{pgfscope}%
\pgfsetroundcap%
\pgfsetroundjoin%
\pgfsetlinewidth{1.003750pt}%
\definecolor{currentstroke}{rgb}{0.007843,0.619608,0.450980}%
\pgfsetstrokecolor{currentstroke}%
\pgfsetdash{}{0pt}%
\pgfpathmoveto{\pgfqpoint{0.824732in}{0.729919in}}%
\pgfpathlineto{\pgfqpoint{0.873343in}{0.729919in}}%
\pgfpathlineto{\pgfqpoint{0.873343in}{0.729919in}}%
\pgfpathlineto{\pgfqpoint{0.970565in}{0.729919in}}%
\pgfpathlineto{\pgfqpoint{0.970565in}{0.729919in}}%
\pgfpathlineto{\pgfqpoint{1.019176in}{0.729919in}}%
\pgfusepath{stroke}%
\end{pgfscope}%
\begin{pgfscope}%
\definecolor{textcolor}{rgb}{0.150000,0.150000,0.150000}%
\pgfsetstrokecolor{textcolor}%
\pgfsetfillcolor{textcolor}%
\pgftext[x=1.096954in,y=0.695892in,left,base]{\color{textcolor}\rmfamily\fontsize{7.000000}{8.400000}\selectfont 4}%
\end{pgfscope}%
\end{pgfpicture}%
\makeatother%
\endgroup%

%% file: figures/experiments/graphs/sparse_smoothing/nodes_attributes/node_classification-Cora-GAT-hidden=8-p_adj_plus=0.0-p_adj_minus=0.0-p_att_plus=0.001-p_att_minus=0.8-multi_class_cert-B.pgf
\begingroup%
\makeatletter%
\begin{pgfpicture}%
\pgfpathrectangle{\pgfpointorigin}{\pgfqpoint{1.375000in}{1.581250in}}%
\pgfusepath{use as bounding box, clip}%
\begin{pgfscope}%
\pgfsetbuttcap%
\pgfsetmiterjoin%
\definecolor{currentfill}{rgb}{1.000000,1.000000,1.000000}%
\pgfsetfillcolor{currentfill}%
\pgfsetlinewidth{0.000000pt}%
\definecolor{currentstroke}{rgb}{1.000000,1.000000,1.000000}%
\pgfsetstrokecolor{currentstroke}%
\pgfsetdash{}{0pt}%
\pgfpathmoveto{\pgfqpoint{0.000000in}{0.000000in}}%
\pgfpathlineto{\pgfqpoint{1.375000in}{0.000000in}}%
\pgfpathlineto{\pgfqpoint{1.375000in}{1.581250in}}%
\pgfpathlineto{\pgfqpoint{0.000000in}{1.581250in}}%
\pgfpathlineto{\pgfqpoint{0.000000in}{0.000000in}}%
\pgfpathclose%
\pgfusepath{fill}%
\end{pgfscope}%
\begin{pgfscope}%
\pgfsetbuttcap%
\pgfsetmiterjoin%
\definecolor{currentfill}{rgb}{1.000000,1.000000,1.000000}%
\pgfsetfillcolor{currentfill}%
\pgfsetlinewidth{0.000000pt}%
\definecolor{currentstroke}{rgb}{0.000000,0.000000,0.000000}%
\pgfsetstrokecolor{currentstroke}%
\pgfsetstrokeopacity{0.000000}%
\pgfsetdash{}{0pt}%
\pgfpathmoveto{\pgfqpoint{0.520798in}{0.442177in}}%
\pgfpathlineto{\pgfqpoint{1.239803in}{0.442177in}}%
\pgfpathlineto{\pgfqpoint{1.239803in}{1.314061in}}%
\pgfpathlineto{\pgfqpoint{0.520798in}{1.314061in}}%
\pgfpathlineto{\pgfqpoint{0.520798in}{0.442177in}}%
\pgfpathclose%
\pgfusepath{fill}%
\end{pgfscope}%
\begin{pgfscope}%
\pgfpathrectangle{\pgfqpoint{0.520798in}{0.442177in}}{\pgfqpoint{0.719005in}{0.871884in}}%
\pgfusepath{clip}%
\pgfsetroundcap%
\pgfsetroundjoin%
\pgfsetlinewidth{0.501875pt}%
\definecolor{currentstroke}{rgb}{0.800000,0.800000,0.800000}%
\pgfsetstrokecolor{currentstroke}%
\pgfsetdash{}{0pt}%
\pgfpathmoveto{\pgfqpoint{0.520798in}{0.442177in}}%
\pgfpathlineto{\pgfqpoint{0.520798in}{1.314061in}}%
\pgfusepath{stroke}%
\end{pgfscope}%
\begin{pgfscope}%
\definecolor{textcolor}{rgb}{0.150000,0.150000,0.150000}%
\pgfsetstrokecolor{textcolor}%
\pgfsetfillcolor{textcolor}%
\pgftext[x=0.520798in,y=0.351899in,,top]{\color{textcolor}\rmfamily\fontsize{8.000000}{9.600000}\selectfont \(\displaystyle {0}\)}%
\end{pgfscope}%
\begin{pgfscope}%
\pgfpathrectangle{\pgfqpoint{0.520798in}{0.442177in}}{\pgfqpoint{0.719005in}{0.871884in}}%
\pgfusepath{clip}%
\pgfsetroundcap%
\pgfsetroundjoin%
\pgfsetlinewidth{0.501875pt}%
\definecolor{currentstroke}{rgb}{0.800000,0.800000,0.800000}%
\pgfsetstrokecolor{currentstroke}%
\pgfsetdash{}{0pt}%
\pgfpathmoveto{\pgfqpoint{0.880300in}{0.442177in}}%
\pgfpathlineto{\pgfqpoint{0.880300in}{1.314061in}}%
\pgfusepath{stroke}%
\end{pgfscope}%
\begin{pgfscope}%
\definecolor{textcolor}{rgb}{0.150000,0.150000,0.150000}%
\pgfsetstrokecolor{textcolor}%
\pgfsetfillcolor{textcolor}%
\pgftext[x=0.880300in,y=0.351899in,,top]{\color{textcolor}\rmfamily\fontsize{8.000000}{9.600000}\selectfont \(\displaystyle {5}\)}%
\end{pgfscope}%
\begin{pgfscope}%
\pgfpathrectangle{\pgfqpoint{0.520798in}{0.442177in}}{\pgfqpoint{0.719005in}{0.871884in}}%
\pgfusepath{clip}%
\pgfsetroundcap%
\pgfsetroundjoin%
\pgfsetlinewidth{0.501875pt}%
\definecolor{currentstroke}{rgb}{0.800000,0.800000,0.800000}%
\pgfsetstrokecolor{currentstroke}%
\pgfsetdash{}{0pt}%
\pgfpathmoveto{\pgfqpoint{1.239803in}{0.442177in}}%
\pgfpathlineto{\pgfqpoint{1.239803in}{1.314061in}}%
\pgfusepath{stroke}%
\end{pgfscope}%
\begin{pgfscope}%
\definecolor{textcolor}{rgb}{0.150000,0.150000,0.150000}%
\pgfsetstrokecolor{textcolor}%
\pgfsetfillcolor{textcolor}%
\pgftext[x=1.239803in,y=0.351899in,,top]{\color{textcolor}\rmfamily\fontsize{8.000000}{9.600000}\selectfont \(\displaystyle {10}\)}%
\end{pgfscope}%
\begin{pgfscope}%
\definecolor{textcolor}{rgb}{0.150000,0.150000,0.150000}%
\pgfsetstrokecolor{textcolor}%
\pgfsetfillcolor{textcolor}%
\pgftext[x=0.880300in,y=0.198219in,,top]{\color{textcolor}\rmfamily\fontsize{10.000000}{12.000000}\selectfont Edit distance \(\displaystyle \epsilon\)}%
\end{pgfscope}%
\begin{pgfscope}%
\pgfpathrectangle{\pgfqpoint{0.520798in}{0.442177in}}{\pgfqpoint{0.719005in}{0.871884in}}%
\pgfusepath{clip}%
\pgfsetroundcap%
\pgfsetroundjoin%
\pgfsetlinewidth{0.501875pt}%
\definecolor{currentstroke}{rgb}{0.800000,0.800000,0.800000}%
\pgfsetstrokecolor{currentstroke}%
\pgfsetdash{}{0pt}%
\pgfpathmoveto{\pgfqpoint{0.520798in}{0.442177in}}%
\pgfpathlineto{\pgfqpoint{1.239803in}{0.442177in}}%
\pgfusepath{stroke}%
\end{pgfscope}%
\begin{pgfscope}%
\definecolor{textcolor}{rgb}{0.150000,0.150000,0.150000}%
\pgfsetstrokecolor{textcolor}%
\pgfsetfillcolor{textcolor}%
\pgftext[x=0.273151in, y=0.403915in, left, base]{\color{textcolor}\rmfamily\fontsize{8.000000}{9.600000}\selectfont 0\%}%
\end{pgfscope}%
\begin{pgfscope}%
\pgfpathrectangle{\pgfqpoint{0.520798in}{0.442177in}}{\pgfqpoint{0.719005in}{0.871884in}}%
\pgfusepath{clip}%
\pgfsetroundcap%
\pgfsetroundjoin%
\pgfsetlinewidth{0.501875pt}%
\definecolor{currentstroke}{rgb}{0.800000,0.800000,0.800000}%
\pgfsetstrokecolor{currentstroke}%
\pgfsetdash{}{0pt}%
\pgfpathmoveto{\pgfqpoint{0.520798in}{0.660148in}}%
\pgfpathlineto{\pgfqpoint{1.239803in}{0.660148in}}%
\pgfusepath{stroke}%
\end{pgfscope}%
\begin{pgfscope}%
\definecolor{textcolor}{rgb}{0.150000,0.150000,0.150000}%
\pgfsetstrokecolor{textcolor}%
\pgfsetfillcolor{textcolor}%
\pgftext[x=0.214138in, y=0.621886in, left, base]{\color{textcolor}\rmfamily\fontsize{8.000000}{9.600000}\selectfont 25\%}%
\end{pgfscope}%
\begin{pgfscope}%
\pgfpathrectangle{\pgfqpoint{0.520798in}{0.442177in}}{\pgfqpoint{0.719005in}{0.871884in}}%
\pgfusepath{clip}%
\pgfsetroundcap%
\pgfsetroundjoin%
\pgfsetlinewidth{0.501875pt}%
\definecolor{currentstroke}{rgb}{0.800000,0.800000,0.800000}%
\pgfsetstrokecolor{currentstroke}%
\pgfsetdash{}{0pt}%
\pgfpathmoveto{\pgfqpoint{0.520798in}{0.878119in}}%
\pgfpathlineto{\pgfqpoint{1.239803in}{0.878119in}}%
\pgfusepath{stroke}%
\end{pgfscope}%
\begin{pgfscope}%
\definecolor{textcolor}{rgb}{0.150000,0.150000,0.150000}%
\pgfsetstrokecolor{textcolor}%
\pgfsetfillcolor{textcolor}%
\pgftext[x=0.214138in, y=0.839857in, left, base]{\color{textcolor}\rmfamily\fontsize{8.000000}{9.600000}\selectfont 50\%}%
\end{pgfscope}%
\begin{pgfscope}%
\pgfpathrectangle{\pgfqpoint{0.520798in}{0.442177in}}{\pgfqpoint{0.719005in}{0.871884in}}%
\pgfusepath{clip}%
\pgfsetroundcap%
\pgfsetroundjoin%
\pgfsetlinewidth{0.501875pt}%
\definecolor{currentstroke}{rgb}{0.800000,0.800000,0.800000}%
\pgfsetstrokecolor{currentstroke}%
\pgfsetdash{}{0pt}%
\pgfpathmoveto{\pgfqpoint{0.520798in}{1.096090in}}%
\pgfpathlineto{\pgfqpoint{1.239803in}{1.096090in}}%
\pgfusepath{stroke}%
\end{pgfscope}%
\begin{pgfscope}%
\definecolor{textcolor}{rgb}{0.150000,0.150000,0.150000}%
\pgfsetstrokecolor{textcolor}%
\pgfsetfillcolor{textcolor}%
\pgftext[x=0.214138in, y=1.057828in, left, base]{\color{textcolor}\rmfamily\fontsize{8.000000}{9.600000}\selectfont 75\%}%
\end{pgfscope}%
\begin{pgfscope}%
\pgfpathrectangle{\pgfqpoint{0.520798in}{0.442177in}}{\pgfqpoint{0.719005in}{0.871884in}}%
\pgfusepath{clip}%
\pgfsetroundcap%
\pgfsetroundjoin%
\pgfsetlinewidth{0.501875pt}%
\definecolor{currentstroke}{rgb}{0.800000,0.800000,0.800000}%
\pgfsetstrokecolor{currentstroke}%
\pgfsetdash{}{0pt}%
\pgfpathmoveto{\pgfqpoint{0.520798in}{1.314061in}}%
\pgfpathlineto{\pgfqpoint{1.239803in}{1.314061in}}%
\pgfusepath{stroke}%
\end{pgfscope}%
\begin{pgfscope}%
\definecolor{textcolor}{rgb}{0.150000,0.150000,0.150000}%
\pgfsetstrokecolor{textcolor}%
\pgfsetfillcolor{textcolor}%
\pgftext[x=0.155124in, y=1.275799in, left, base]{\color{textcolor}\rmfamily\fontsize{8.000000}{9.600000}\selectfont 100\%}%
\end{pgfscope}%
\begin{pgfscope}%
\definecolor{textcolor}{rgb}{0.150000,0.150000,0.150000}%
\pgfsetstrokecolor{textcolor}%
\pgfsetfillcolor{textcolor}%
\pgftext[x=0.099569in,y=0.878119in,,bottom,rotate=90.000000]{\color{textcolor}\rmfamily\fontsize{10.000000}{12.000000}\selectfont Cert. Acc.}%
\end{pgfscope}%
\begin{pgfscope}%
\pgfsetrectcap%
\pgfsetmiterjoin%
\pgfsetlinewidth{0.752812pt}%
\definecolor{currentstroke}{rgb}{0.700000,0.700000,0.700000}%
\pgfsetstrokecolor{currentstroke}%
\pgfsetdash{}{0pt}%
\pgfpathmoveto{\pgfqpoint{0.520798in}{0.442177in}}%
\pgfpathlineto{\pgfqpoint{0.520798in}{1.314061in}}%
\pgfusepath{stroke}%
\end{pgfscope}%
\begin{pgfscope}%
\pgfsetrectcap%
\pgfsetmiterjoin%
\pgfsetlinewidth{0.752812pt}%
\definecolor{currentstroke}{rgb}{0.700000,0.700000,0.700000}%
\pgfsetstrokecolor{currentstroke}%
\pgfsetdash{}{0pt}%
\pgfpathmoveto{\pgfqpoint{1.239803in}{0.442177in}}%
\pgfpathlineto{\pgfqpoint{1.239803in}{1.314061in}}%
\pgfusepath{stroke}%
\end{pgfscope}%
\begin{pgfscope}%
\pgfsetrectcap%
\pgfsetmiterjoin%
\pgfsetlinewidth{0.752812pt}%
\definecolor{currentstroke}{rgb}{0.700000,0.700000,0.700000}%
\pgfsetstrokecolor{currentstroke}%
\pgfsetdash{}{0pt}%
\pgfpathmoveto{\pgfqpoint{0.520798in}{0.442177in}}%
\pgfpathlineto{\pgfqpoint{1.239803in}{0.442177in}}%
\pgfusepath{stroke}%
\end{pgfscope}%
\begin{pgfscope}%
\pgfsetrectcap%
\pgfsetmiterjoin%
\pgfsetlinewidth{0.752812pt}%
\definecolor{currentstroke}{rgb}{0.700000,0.700000,0.700000}%
\pgfsetstrokecolor{currentstroke}%
\pgfsetdash{}{0pt}%
\pgfpathmoveto{\pgfqpoint{0.520798in}{1.314061in}}%
\pgfpathlineto{\pgfqpoint{1.239803in}{1.314061in}}%
\pgfusepath{stroke}%
\end{pgfscope}%
\begin{pgfscope}%
\pgfpathrectangle{\pgfqpoint{0.520798in}{0.442177in}}{\pgfqpoint{0.719005in}{0.871884in}}%
\pgfusepath{clip}%
\pgfsetroundcap%
\pgfsetroundjoin%
\pgfsetlinewidth{1.003750pt}%
\definecolor{currentstroke}{rgb}{0.003922,0.450980,0.698039}%
\pgfsetstrokecolor{currentstroke}%
\pgfsetdash{}{0pt}%
\pgfpathmoveto{\pgfqpoint{0.520798in}{1.159297in}}%
\pgfpathlineto{\pgfqpoint{0.556748in}{1.159297in}}%
\pgfpathlineto{\pgfqpoint{0.556748in}{1.099036in}}%
\pgfpathlineto{\pgfqpoint{0.628649in}{1.099036in}}%
\pgfpathlineto{\pgfqpoint{0.628649in}{1.003900in}}%
\pgfpathlineto{\pgfqpoint{0.700549in}{1.003900in}}%
\pgfpathlineto{\pgfqpoint{0.700549in}{0.739052in}}%
\pgfpathlineto{\pgfqpoint{0.772450in}{0.739052in}}%
\pgfpathlineto{\pgfqpoint{0.772450in}{0.604375in}}%
\pgfpathlineto{\pgfqpoint{0.844350in}{0.604375in}}%
\pgfpathlineto{\pgfqpoint{0.844350in}{0.587372in}}%
\pgfpathlineto{\pgfqpoint{0.916251in}{0.587372in}}%
\pgfpathlineto{\pgfqpoint{0.916251in}{0.577724in}}%
\pgfpathlineto{\pgfqpoint{0.988151in}{0.577724in}}%
\pgfpathlineto{\pgfqpoint{0.988151in}{0.559377in}}%
\pgfpathlineto{\pgfqpoint{1.060052in}{0.559377in}}%
\pgfpathlineto{\pgfqpoint{1.060052in}{0.506471in}}%
\pgfpathlineto{\pgfqpoint{1.131952in}{0.506471in}}%
\pgfpathlineto{\pgfqpoint{1.131952in}{0.499591in}}%
\pgfpathlineto{\pgfqpoint{1.203853in}{0.499591in}}%
\pgfpathlineto{\pgfqpoint{1.203853in}{0.474680in}}%
\pgfpathlineto{\pgfqpoint{1.241470in}{0.474680in}}%
\pgfusepath{stroke}%
\end{pgfscope}%
\begin{pgfscope}%
\pgfpathrectangle{\pgfqpoint{0.520798in}{0.442177in}}{\pgfqpoint{0.719005in}{0.871884in}}%
\pgfusepath{clip}%
\pgfsetbuttcap%
\pgfsetroundjoin%
\definecolor{currentfill}{rgb}{0.003922,0.450980,0.698039}%
\pgfsetfillcolor{currentfill}%
\pgfsetfillopacity{0.500000}%
\pgfsetlinewidth{0.000000pt}%
\definecolor{currentstroke}{rgb}{0.003922,0.450980,0.698039}%
\pgfsetstrokecolor{currentstroke}%
\pgfsetstrokeopacity{0.500000}%
\pgfsetdash{}{0pt}%
\pgfpathmoveto{\pgfqpoint{0.520798in}{1.172451in}}%
\pgfpathlineto{\pgfqpoint{0.520798in}{1.146143in}}%
\pgfpathlineto{\pgfqpoint{0.556748in}{1.146143in}}%
\pgfpathlineto{\pgfqpoint{0.556748in}{1.082811in}}%
\pgfpathlineto{\pgfqpoint{0.628649in}{1.082811in}}%
\pgfpathlineto{\pgfqpoint{0.628649in}{0.989028in}}%
\pgfpathlineto{\pgfqpoint{0.700549in}{0.989028in}}%
\pgfpathlineto{\pgfqpoint{0.700549in}{0.734887in}}%
\pgfpathlineto{\pgfqpoint{0.772450in}{0.734887in}}%
\pgfpathlineto{\pgfqpoint{0.772450in}{0.592079in}}%
\pgfpathlineto{\pgfqpoint{0.844350in}{0.592079in}}%
\pgfpathlineto{\pgfqpoint{0.844350in}{0.573918in}}%
\pgfpathlineto{\pgfqpoint{0.916251in}{0.573918in}}%
\pgfpathlineto{\pgfqpoint{0.916251in}{0.565347in}}%
\pgfpathlineto{\pgfqpoint{0.988151in}{0.565347in}}%
\pgfpathlineto{\pgfqpoint{0.988151in}{0.548400in}}%
\pgfpathlineto{\pgfqpoint{1.060052in}{0.548400in}}%
\pgfpathlineto{\pgfqpoint{1.060052in}{0.496812in}}%
\pgfpathlineto{\pgfqpoint{1.131952in}{0.496812in}}%
\pgfpathlineto{\pgfqpoint{1.131952in}{0.490452in}}%
\pgfpathlineto{\pgfqpoint{1.203853in}{0.490452in}}%
\pgfpathlineto{\pgfqpoint{1.203853in}{0.468317in}}%
\pgfpathlineto{\pgfqpoint{1.275753in}{0.468317in}}%
\pgfpathlineto{\pgfqpoint{1.275753in}{0.452427in}}%
\pgfpathlineto{\pgfqpoint{1.347654in}{0.452427in}}%
\pgfpathlineto{\pgfqpoint{1.347654in}{0.449330in}}%
\pgfpathlineto{\pgfqpoint{1.419554in}{0.449330in}}%
\pgfpathlineto{\pgfqpoint{1.419554in}{0.442177in}}%
\pgfpathlineto{\pgfqpoint{2.066659in}{0.442177in}}%
\pgfpathlineto{\pgfqpoint{2.066659in}{0.442177in}}%
\pgfpathlineto{\pgfqpoint{2.677813in}{0.442177in}}%
\pgfpathlineto{\pgfqpoint{2.677813in}{0.442177in}}%
\pgfpathlineto{\pgfqpoint{2.677813in}{0.442177in}}%
\pgfpathlineto{\pgfqpoint{2.066659in}{0.442177in}}%
\pgfpathlineto{\pgfqpoint{2.066659in}{0.442177in}}%
\pgfpathlineto{\pgfqpoint{1.419554in}{0.442177in}}%
\pgfpathlineto{\pgfqpoint{1.419554in}{0.455111in}}%
\pgfpathlineto{\pgfqpoint{1.347654in}{0.455111in}}%
\pgfpathlineto{\pgfqpoint{1.347654in}{0.459922in}}%
\pgfpathlineto{\pgfqpoint{1.275753in}{0.459922in}}%
\pgfpathlineto{\pgfqpoint{1.275753in}{0.481043in}}%
\pgfpathlineto{\pgfqpoint{1.203853in}{0.481043in}}%
\pgfpathlineto{\pgfqpoint{1.203853in}{0.508729in}}%
\pgfpathlineto{\pgfqpoint{1.131952in}{0.508729in}}%
\pgfpathlineto{\pgfqpoint{1.131952in}{0.516129in}}%
\pgfpathlineto{\pgfqpoint{1.060052in}{0.516129in}}%
\pgfpathlineto{\pgfqpoint{1.060052in}{0.570354in}}%
\pgfpathlineto{\pgfqpoint{0.988151in}{0.570354in}}%
\pgfpathlineto{\pgfqpoint{0.988151in}{0.590101in}}%
\pgfpathlineto{\pgfqpoint{0.916251in}{0.590101in}}%
\pgfpathlineto{\pgfqpoint{0.916251in}{0.600827in}}%
\pgfpathlineto{\pgfqpoint{0.844350in}{0.600827in}}%
\pgfpathlineto{\pgfqpoint{0.844350in}{0.616671in}}%
\pgfpathlineto{\pgfqpoint{0.772450in}{0.616671in}}%
\pgfpathlineto{\pgfqpoint{0.772450in}{0.743218in}}%
\pgfpathlineto{\pgfqpoint{0.700549in}{0.743218in}}%
\pgfpathlineto{\pgfqpoint{0.700549in}{1.018772in}}%
\pgfpathlineto{\pgfqpoint{0.628649in}{1.018772in}}%
\pgfpathlineto{\pgfqpoint{0.628649in}{1.115261in}}%
\pgfpathlineto{\pgfqpoint{0.556748in}{1.115261in}}%
\pgfpathlineto{\pgfqpoint{0.556748in}{1.172451in}}%
\pgfpathlineto{\pgfqpoint{0.520798in}{1.172451in}}%
\pgfpathlineto{\pgfqpoint{0.520798in}{1.172451in}}%
\pgfpathclose%
\pgfusepath{fill}%
\end{pgfscope}%
\begin{pgfscope}%
\pgfpathrectangle{\pgfqpoint{0.520798in}{0.442177in}}{\pgfqpoint{0.719005in}{0.871884in}}%
\pgfusepath{clip}%
\pgfsetroundcap%
\pgfsetroundjoin%
\pgfsetlinewidth{1.003750pt}%
\definecolor{currentstroke}{rgb}{0.870588,0.560784,0.019608}%
\pgfsetstrokecolor{currentstroke}%
\pgfsetdash{}{0pt}%
\pgfpathmoveto{\pgfqpoint{0.520798in}{1.159297in}}%
\pgfpathlineto{\pgfqpoint{0.556748in}{1.159297in}}%
\pgfpathlineto{\pgfqpoint{0.556748in}{1.111215in}}%
\pgfpathlineto{\pgfqpoint{0.628649in}{1.111215in}}%
\pgfpathlineto{\pgfqpoint{0.628649in}{1.068906in}}%
\pgfpathlineto{\pgfqpoint{0.700549in}{1.068906in}}%
\pgfpathlineto{\pgfqpoint{0.700549in}{1.027229in}}%
\pgfpathlineto{\pgfqpoint{0.772450in}{1.027229in}}%
\pgfpathlineto{\pgfqpoint{0.772450in}{0.993224in}}%
\pgfpathlineto{\pgfqpoint{0.844350in}{0.993224in}}%
\pgfpathlineto{\pgfqpoint{0.844350in}{0.958348in}}%
\pgfpathlineto{\pgfqpoint{0.916251in}{0.958348in}}%
\pgfpathlineto{\pgfqpoint{0.916251in}{0.739052in}}%
\pgfpathlineto{\pgfqpoint{0.988151in}{0.739052in}}%
\pgfpathlineto{\pgfqpoint{0.988151in}{0.721417in}}%
\pgfpathlineto{\pgfqpoint{1.060052in}{0.721417in}}%
\pgfpathlineto{\pgfqpoint{1.060052in}{0.604375in}}%
\pgfpathlineto{\pgfqpoint{1.131952in}{0.604375in}}%
\pgfpathlineto{\pgfqpoint{1.131952in}{0.594332in}}%
\pgfpathlineto{\pgfqpoint{1.203853in}{0.594332in}}%
\pgfpathlineto{\pgfqpoint{1.203853in}{0.581916in}}%
\pgfpathlineto{\pgfqpoint{1.241470in}{0.581916in}}%
\pgfusepath{stroke}%
\end{pgfscope}%
\begin{pgfscope}%
\pgfpathrectangle{\pgfqpoint{0.520798in}{0.442177in}}{\pgfqpoint{0.719005in}{0.871884in}}%
\pgfusepath{clip}%
\pgfsetbuttcap%
\pgfsetroundjoin%
\definecolor{currentfill}{rgb}{0.870588,0.560784,0.019608}%
\pgfsetfillcolor{currentfill}%
\pgfsetfillopacity{0.500000}%
\pgfsetlinewidth{0.000000pt}%
\definecolor{currentstroke}{rgb}{0.870588,0.560784,0.019608}%
\pgfsetstrokecolor{currentstroke}%
\pgfsetstrokeopacity{0.500000}%
\pgfsetdash{}{0pt}%
\pgfpathmoveto{\pgfqpoint{0.520798in}{1.172451in}}%
\pgfpathlineto{\pgfqpoint{0.520798in}{1.146143in}}%
\pgfpathlineto{\pgfqpoint{0.556748in}{1.146143in}}%
\pgfpathlineto{\pgfqpoint{0.556748in}{1.094566in}}%
\pgfpathlineto{\pgfqpoint{0.628649in}{1.094566in}}%
\pgfpathlineto{\pgfqpoint{0.628649in}{1.053196in}}%
\pgfpathlineto{\pgfqpoint{0.700549in}{1.053196in}}%
\pgfpathlineto{\pgfqpoint{0.700549in}{1.011576in}}%
\pgfpathlineto{\pgfqpoint{0.772450in}{1.011576in}}%
\pgfpathlineto{\pgfqpoint{0.772450in}{0.977493in}}%
\pgfpathlineto{\pgfqpoint{0.844350in}{0.977493in}}%
\pgfpathlineto{\pgfqpoint{0.844350in}{0.940428in}}%
\pgfpathlineto{\pgfqpoint{0.916251in}{0.940428in}}%
\pgfpathlineto{\pgfqpoint{0.916251in}{0.734887in}}%
\pgfpathlineto{\pgfqpoint{0.988151in}{0.734887in}}%
\pgfpathlineto{\pgfqpoint{0.988151in}{0.717374in}}%
\pgfpathlineto{\pgfqpoint{1.060052in}{0.717374in}}%
\pgfpathlineto{\pgfqpoint{1.060052in}{0.592079in}}%
\pgfpathlineto{\pgfqpoint{1.131952in}{0.592079in}}%
\pgfpathlineto{\pgfqpoint{1.131952in}{0.580954in}}%
\pgfpathlineto{\pgfqpoint{1.203853in}{0.580954in}}%
\pgfpathlineto{\pgfqpoint{1.203853in}{0.569140in}}%
\pgfpathlineto{\pgfqpoint{1.275753in}{0.569140in}}%
\pgfpathlineto{\pgfqpoint{1.275753in}{0.561365in}}%
\pgfpathlineto{\pgfqpoint{1.347654in}{0.561365in}}%
\pgfpathlineto{\pgfqpoint{1.347654in}{0.552513in}}%
\pgfpathlineto{\pgfqpoint{1.419554in}{0.552513in}}%
\pgfpathlineto{\pgfqpoint{1.419554in}{0.544235in}}%
\pgfpathlineto{\pgfqpoint{1.491455in}{0.544235in}}%
\pgfpathlineto{\pgfqpoint{1.491455in}{0.535910in}}%
\pgfpathlineto{\pgfqpoint{1.563355in}{0.535910in}}%
\pgfpathlineto{\pgfqpoint{1.563355in}{0.530942in}}%
\pgfpathlineto{\pgfqpoint{1.635256in}{0.530942in}}%
\pgfpathlineto{\pgfqpoint{1.635256in}{0.496475in}}%
\pgfpathlineto{\pgfqpoint{1.707156in}{0.496475in}}%
\pgfpathlineto{\pgfqpoint{1.707156in}{0.493515in}}%
\pgfpathlineto{\pgfqpoint{1.779057in}{0.493515in}}%
\pgfpathlineto{\pgfqpoint{1.779057in}{0.488932in}}%
\pgfpathlineto{\pgfqpoint{1.850957in}{0.488932in}}%
\pgfpathlineto{\pgfqpoint{1.850957in}{0.484130in}}%
\pgfpathlineto{\pgfqpoint{1.922858in}{0.484130in}}%
\pgfpathlineto{\pgfqpoint{1.922858in}{0.468061in}}%
\pgfpathlineto{\pgfqpoint{1.994758in}{0.468061in}}%
\pgfpathlineto{\pgfqpoint{1.994758in}{0.451870in}}%
\pgfpathlineto{\pgfqpoint{2.066659in}{0.451870in}}%
\pgfpathlineto{\pgfqpoint{2.066659in}{0.449987in}}%
\pgfpathlineto{\pgfqpoint{2.138559in}{0.449987in}}%
\pgfpathlineto{\pgfqpoint{2.138559in}{0.448776in}}%
\pgfpathlineto{\pgfqpoint{2.210460in}{0.448776in}}%
\pgfpathlineto{\pgfqpoint{2.210460in}{0.443890in}}%
\pgfpathlineto{\pgfqpoint{2.282360in}{0.443890in}}%
\pgfpathlineto{\pgfqpoint{2.282360in}{0.442177in}}%
\pgfpathlineto{\pgfqpoint{2.498062in}{0.442177in}}%
\pgfpathlineto{\pgfqpoint{2.498062in}{0.442177in}}%
\pgfpathlineto{\pgfqpoint{2.677813in}{0.442177in}}%
\pgfpathlineto{\pgfqpoint{2.677813in}{0.442177in}}%
\pgfpathlineto{\pgfqpoint{2.677813in}{0.442177in}}%
\pgfpathlineto{\pgfqpoint{2.498062in}{0.442177in}}%
\pgfpathlineto{\pgfqpoint{2.498062in}{0.442177in}}%
\pgfpathlineto{\pgfqpoint{2.282360in}{0.442177in}}%
\pgfpathlineto{\pgfqpoint{2.282360in}{0.446949in}}%
\pgfpathlineto{\pgfqpoint{2.210460in}{0.446949in}}%
\pgfpathlineto{\pgfqpoint{2.210460in}{0.454399in}}%
\pgfpathlineto{\pgfqpoint{2.138559in}{0.454399in}}%
\pgfpathlineto{\pgfqpoint{2.138559in}{0.456668in}}%
\pgfpathlineto{\pgfqpoint{2.066659in}{0.456668in}}%
\pgfpathlineto{\pgfqpoint{2.066659in}{0.459372in}}%
\pgfpathlineto{\pgfqpoint{1.994758in}{0.459372in}}%
\pgfpathlineto{\pgfqpoint{1.994758in}{0.480982in}}%
\pgfpathlineto{\pgfqpoint{1.922858in}{0.480982in}}%
\pgfpathlineto{\pgfqpoint{1.922858in}{0.500342in}}%
\pgfpathlineto{\pgfqpoint{1.850957in}{0.500342in}}%
\pgfpathlineto{\pgfqpoint{1.850957in}{0.505188in}}%
\pgfpathlineto{\pgfqpoint{1.779057in}{0.505188in}}%
\pgfpathlineto{\pgfqpoint{1.779057in}{0.511676in}}%
\pgfpathlineto{\pgfqpoint{1.707156in}{0.511676in}}%
\pgfpathlineto{\pgfqpoint{1.707156in}{0.515675in}}%
\pgfpathlineto{\pgfqpoint{1.635256in}{0.515675in}}%
\pgfpathlineto{\pgfqpoint{1.635256in}{0.553174in}}%
\pgfpathlineto{\pgfqpoint{1.563355in}{0.553174in}}%
\pgfpathlineto{\pgfqpoint{1.563355in}{0.559910in}}%
\pgfpathlineto{\pgfqpoint{1.491455in}{0.559910in}}%
\pgfpathlineto{\pgfqpoint{1.491455in}{0.568350in}}%
\pgfpathlineto{\pgfqpoint{1.419554in}{0.568350in}}%
\pgfpathlineto{\pgfqpoint{1.419554in}{0.576680in}}%
\pgfpathlineto{\pgfqpoint{1.347654in}{0.576680in}}%
\pgfpathlineto{\pgfqpoint{1.347654in}{0.585385in}}%
\pgfpathlineto{\pgfqpoint{1.275753in}{0.585385in}}%
\pgfpathlineto{\pgfqpoint{1.275753in}{0.594691in}}%
\pgfpathlineto{\pgfqpoint{1.203853in}{0.594691in}}%
\pgfpathlineto{\pgfqpoint{1.203853in}{0.607709in}}%
\pgfpathlineto{\pgfqpoint{1.131952in}{0.607709in}}%
\pgfpathlineto{\pgfqpoint{1.131952in}{0.616671in}}%
\pgfpathlineto{\pgfqpoint{1.060052in}{0.616671in}}%
\pgfpathlineto{\pgfqpoint{1.060052in}{0.725460in}}%
\pgfpathlineto{\pgfqpoint{0.988151in}{0.725460in}}%
\pgfpathlineto{\pgfqpoint{0.988151in}{0.743218in}}%
\pgfpathlineto{\pgfqpoint{0.916251in}{0.743218in}}%
\pgfpathlineto{\pgfqpoint{0.916251in}{0.976268in}}%
\pgfpathlineto{\pgfqpoint{0.844350in}{0.976268in}}%
\pgfpathlineto{\pgfqpoint{0.844350in}{1.008954in}}%
\pgfpathlineto{\pgfqpoint{0.772450in}{1.008954in}}%
\pgfpathlineto{\pgfqpoint{0.772450in}{1.042882in}}%
\pgfpathlineto{\pgfqpoint{0.700549in}{1.042882in}}%
\pgfpathlineto{\pgfqpoint{0.700549in}{1.084615in}}%
\pgfpathlineto{\pgfqpoint{0.628649in}{1.084615in}}%
\pgfpathlineto{\pgfqpoint{0.628649in}{1.127863in}}%
\pgfpathlineto{\pgfqpoint{0.556748in}{1.127863in}}%
\pgfpathlineto{\pgfqpoint{0.556748in}{1.172451in}}%
\pgfpathlineto{\pgfqpoint{0.520798in}{1.172451in}}%
\pgfpathlineto{\pgfqpoint{0.520798in}{1.172451in}}%
\pgfpathclose%
\pgfusepath{fill}%
\end{pgfscope}%
\begin{pgfscope}%
\pgfpathrectangle{\pgfqpoint{0.520798in}{0.442177in}}{\pgfqpoint{0.719005in}{0.871884in}}%
\pgfusepath{clip}%
\pgfsetroundcap%
\pgfsetroundjoin%
\pgfsetlinewidth{1.003750pt}%
\definecolor{currentstroke}{rgb}{0.007843,0.619608,0.450980}%
\pgfsetstrokecolor{currentstroke}%
\pgfsetdash{}{0pt}%
\pgfpathmoveto{\pgfqpoint{0.520798in}{1.159297in}}%
\pgfpathlineto{\pgfqpoint{0.556748in}{1.159297in}}%
\pgfpathlineto{\pgfqpoint{0.556748in}{1.111215in}}%
\pgfpathlineto{\pgfqpoint{0.628649in}{1.111215in}}%
\pgfpathlineto{\pgfqpoint{0.628649in}{1.068906in}}%
\pgfpathlineto{\pgfqpoint{0.700549in}{1.068906in}}%
\pgfpathlineto{\pgfqpoint{0.700549in}{1.027229in}}%
\pgfpathlineto{\pgfqpoint{0.772450in}{1.027229in}}%
\pgfpathlineto{\pgfqpoint{0.772450in}{0.993224in}}%
\pgfpathlineto{\pgfqpoint{0.844350in}{0.993224in}}%
\pgfpathlineto{\pgfqpoint{0.844350in}{0.958348in}}%
\pgfpathlineto{\pgfqpoint{0.916251in}{0.958348in}}%
\pgfpathlineto{\pgfqpoint{0.916251in}{0.924343in}}%
\pgfpathlineto{\pgfqpoint{0.988151in}{0.924343in}}%
\pgfpathlineto{\pgfqpoint{0.988151in}{0.891998in}}%
\pgfpathlineto{\pgfqpoint{1.060052in}{0.891998in}}%
\pgfpathlineto{\pgfqpoint{1.060052in}{0.863370in}}%
\pgfpathlineto{\pgfqpoint{1.131952in}{0.863370in}}%
\pgfpathlineto{\pgfqpoint{1.131952in}{0.836087in}}%
\pgfpathlineto{\pgfqpoint{1.203853in}{0.836087in}}%
\pgfpathlineto{\pgfqpoint{1.203853in}{0.812046in}}%
\pgfpathlineto{\pgfqpoint{1.241470in}{0.812046in}}%
\pgfusepath{stroke}%
\end{pgfscope}%
\begin{pgfscope}%
\pgfpathrectangle{\pgfqpoint{0.520798in}{0.442177in}}{\pgfqpoint{0.719005in}{0.871884in}}%
\pgfusepath{clip}%
\pgfsetbuttcap%
\pgfsetroundjoin%
\definecolor{currentfill}{rgb}{0.007843,0.619608,0.450980}%
\pgfsetfillcolor{currentfill}%
\pgfsetfillopacity{0.500000}%
\pgfsetlinewidth{0.000000pt}%
\definecolor{currentstroke}{rgb}{0.007843,0.619608,0.450980}%
\pgfsetstrokecolor{currentstroke}%
\pgfsetstrokeopacity{0.500000}%
\pgfsetdash{}{0pt}%
\pgfpathmoveto{\pgfqpoint{0.520798in}{1.172451in}}%
\pgfpathlineto{\pgfqpoint{0.520798in}{1.146143in}}%
\pgfpathlineto{\pgfqpoint{0.556748in}{1.146143in}}%
\pgfpathlineto{\pgfqpoint{0.556748in}{1.094566in}}%
\pgfpathlineto{\pgfqpoint{0.628649in}{1.094566in}}%
\pgfpathlineto{\pgfqpoint{0.628649in}{1.053196in}}%
\pgfpathlineto{\pgfqpoint{0.700549in}{1.053196in}}%
\pgfpathlineto{\pgfqpoint{0.700549in}{1.011576in}}%
\pgfpathlineto{\pgfqpoint{0.772450in}{1.011576in}}%
\pgfpathlineto{\pgfqpoint{0.772450in}{0.977493in}}%
\pgfpathlineto{\pgfqpoint{0.844350in}{0.977493in}}%
\pgfpathlineto{\pgfqpoint{0.844350in}{0.940428in}}%
\pgfpathlineto{\pgfqpoint{0.916251in}{0.940428in}}%
\pgfpathlineto{\pgfqpoint{0.916251in}{0.908822in}}%
\pgfpathlineto{\pgfqpoint{0.988151in}{0.908822in}}%
\pgfpathlineto{\pgfqpoint{0.988151in}{0.878190in}}%
\pgfpathlineto{\pgfqpoint{1.060052in}{0.878190in}}%
\pgfpathlineto{\pgfqpoint{1.060052in}{0.850711in}}%
\pgfpathlineto{\pgfqpoint{1.131952in}{0.850711in}}%
\pgfpathlineto{\pgfqpoint{1.131952in}{0.827498in}}%
\pgfpathlineto{\pgfqpoint{1.203853in}{0.827498in}}%
\pgfpathlineto{\pgfqpoint{1.203853in}{0.803794in}}%
\pgfpathlineto{\pgfqpoint{1.275753in}{0.803794in}}%
\pgfpathlineto{\pgfqpoint{1.275753in}{0.778909in}}%
\pgfpathlineto{\pgfqpoint{1.347654in}{0.778909in}}%
\pgfpathlineto{\pgfqpoint{1.347654in}{0.734887in}}%
\pgfpathlineto{\pgfqpoint{1.419554in}{0.734887in}}%
\pgfpathlineto{\pgfqpoint{1.419554in}{0.717374in}}%
\pgfpathlineto{\pgfqpoint{1.491455in}{0.717374in}}%
\pgfpathlineto{\pgfqpoint{1.491455in}{0.699218in}}%
\pgfpathlineto{\pgfqpoint{1.563355in}{0.699218in}}%
\pgfpathlineto{\pgfqpoint{1.563355in}{0.680973in}}%
\pgfpathlineto{\pgfqpoint{1.635256in}{0.680973in}}%
\pgfpathlineto{\pgfqpoint{1.635256in}{0.592079in}}%
\pgfpathlineto{\pgfqpoint{1.707156in}{0.592079in}}%
\pgfpathlineto{\pgfqpoint{1.707156in}{0.580954in}}%
\pgfpathlineto{\pgfqpoint{1.779057in}{0.580954in}}%
\pgfpathlineto{\pgfqpoint{1.779057in}{0.572561in}}%
\pgfpathlineto{\pgfqpoint{1.850957in}{0.572561in}}%
\pgfpathlineto{\pgfqpoint{1.850957in}{0.564983in}}%
\pgfpathlineto{\pgfqpoint{1.922858in}{0.564983in}}%
\pgfpathlineto{\pgfqpoint{1.922858in}{0.555749in}}%
\pgfpathlineto{\pgfqpoint{1.994758in}{0.555749in}}%
\pgfpathlineto{\pgfqpoint{1.994758in}{0.549759in}}%
\pgfpathlineto{\pgfqpoint{2.066659in}{0.549759in}}%
\pgfpathlineto{\pgfqpoint{2.066659in}{0.543251in}}%
\pgfpathlineto{\pgfqpoint{2.138559in}{0.543251in}}%
\pgfpathlineto{\pgfqpoint{2.138559in}{0.536710in}}%
\pgfpathlineto{\pgfqpoint{2.210460in}{0.536710in}}%
\pgfpathlineto{\pgfqpoint{2.210460in}{0.531906in}}%
\pgfpathlineto{\pgfqpoint{2.282360in}{0.531906in}}%
\pgfpathlineto{\pgfqpoint{2.282360in}{0.528818in}}%
\pgfpathlineto{\pgfqpoint{2.354261in}{0.528818in}}%
\pgfpathlineto{\pgfqpoint{2.354261in}{0.524467in}}%
\pgfpathlineto{\pgfqpoint{2.426161in}{0.524467in}}%
\pgfpathlineto{\pgfqpoint{2.426161in}{0.521011in}}%
\pgfpathlineto{\pgfqpoint{2.498062in}{0.521011in}}%
\pgfpathlineto{\pgfqpoint{2.498062in}{0.517143in}}%
\pgfpathlineto{\pgfqpoint{2.569962in}{0.517143in}}%
\pgfpathlineto{\pgfqpoint{2.569962in}{0.512957in}}%
\pgfpathlineto{\pgfqpoint{2.641863in}{0.512957in}}%
\pgfpathlineto{\pgfqpoint{2.641863in}{0.508674in}}%
\pgfpathlineto{\pgfqpoint{2.677813in}{0.508674in}}%
\pgfpathlineto{\pgfqpoint{2.677813in}{0.525303in}}%
\pgfpathlineto{\pgfqpoint{2.677813in}{0.525303in}}%
\pgfpathlineto{\pgfqpoint{2.641863in}{0.525303in}}%
\pgfpathlineto{\pgfqpoint{2.641863in}{0.530827in}}%
\pgfpathlineto{\pgfqpoint{2.569962in}{0.530827in}}%
\pgfpathlineto{\pgfqpoint{2.569962in}{0.535340in}}%
\pgfpathlineto{\pgfqpoint{2.498062in}{0.535340in}}%
\pgfpathlineto{\pgfqpoint{2.498062in}{0.539855in}}%
\pgfpathlineto{\pgfqpoint{2.426161in}{0.539855in}}%
\pgfpathlineto{\pgfqpoint{2.426161in}{0.545889in}}%
\pgfpathlineto{\pgfqpoint{2.354261in}{0.545889in}}%
\pgfpathlineto{\pgfqpoint{2.354261in}{0.550554in}}%
\pgfpathlineto{\pgfqpoint{2.282360in}{0.550554in}}%
\pgfpathlineto{\pgfqpoint{2.282360in}{0.556006in}}%
\pgfpathlineto{\pgfqpoint{2.210460in}{0.556006in}}%
\pgfpathlineto{\pgfqpoint{2.210460in}{0.561957in}}%
\pgfpathlineto{\pgfqpoint{2.138559in}{0.561957in}}%
\pgfpathlineto{\pgfqpoint{2.138559in}{0.567437in}}%
\pgfpathlineto{\pgfqpoint{2.066659in}{0.567437in}}%
\pgfpathlineto{\pgfqpoint{2.066659in}{0.575480in}}%
\pgfpathlineto{\pgfqpoint{1.994758in}{0.575480in}}%
\pgfpathlineto{\pgfqpoint{1.994758in}{0.581827in}}%
\pgfpathlineto{\pgfqpoint{1.922858in}{0.581827in}}%
\pgfpathlineto{\pgfqpoint{1.922858in}{0.589200in}}%
\pgfpathlineto{\pgfqpoint{1.850957in}{0.589200in}}%
\pgfpathlineto{\pgfqpoint{1.850957in}{0.597597in}}%
\pgfpathlineto{\pgfqpoint{1.779057in}{0.597597in}}%
\pgfpathlineto{\pgfqpoint{1.779057in}{0.607709in}}%
\pgfpathlineto{\pgfqpoint{1.707156in}{0.607709in}}%
\pgfpathlineto{\pgfqpoint{1.707156in}{0.616671in}}%
\pgfpathlineto{\pgfqpoint{1.635256in}{0.616671in}}%
\pgfpathlineto{\pgfqpoint{1.635256in}{0.696539in}}%
\pgfpathlineto{\pgfqpoint{1.563355in}{0.696539in}}%
\pgfpathlineto{\pgfqpoint{1.563355in}{0.710718in}}%
\pgfpathlineto{\pgfqpoint{1.491455in}{0.710718in}}%
\pgfpathlineto{\pgfqpoint{1.491455in}{0.725460in}}%
\pgfpathlineto{\pgfqpoint{1.419554in}{0.725460in}}%
\pgfpathlineto{\pgfqpoint{1.419554in}{0.743218in}}%
\pgfpathlineto{\pgfqpoint{1.347654in}{0.743218in}}%
\pgfpathlineto{\pgfqpoint{1.347654in}{0.793304in}}%
\pgfpathlineto{\pgfqpoint{1.275753in}{0.793304in}}%
\pgfpathlineto{\pgfqpoint{1.275753in}{0.820297in}}%
\pgfpathlineto{\pgfqpoint{1.203853in}{0.820297in}}%
\pgfpathlineto{\pgfqpoint{1.203853in}{0.844675in}}%
\pgfpathlineto{\pgfqpoint{1.131952in}{0.844675in}}%
\pgfpathlineto{\pgfqpoint{1.131952in}{0.876029in}}%
\pgfpathlineto{\pgfqpoint{1.060052in}{0.876029in}}%
\pgfpathlineto{\pgfqpoint{1.060052in}{0.905806in}}%
\pgfpathlineto{\pgfqpoint{0.988151in}{0.905806in}}%
\pgfpathlineto{\pgfqpoint{0.988151in}{0.939863in}}%
\pgfpathlineto{\pgfqpoint{0.916251in}{0.939863in}}%
\pgfpathlineto{\pgfqpoint{0.916251in}{0.976268in}}%
\pgfpathlineto{\pgfqpoint{0.844350in}{0.976268in}}%
\pgfpathlineto{\pgfqpoint{0.844350in}{1.008954in}}%
\pgfpathlineto{\pgfqpoint{0.772450in}{1.008954in}}%
\pgfpathlineto{\pgfqpoint{0.772450in}{1.042882in}}%
\pgfpathlineto{\pgfqpoint{0.700549in}{1.042882in}}%
\pgfpathlineto{\pgfqpoint{0.700549in}{1.084615in}}%
\pgfpathlineto{\pgfqpoint{0.628649in}{1.084615in}}%
\pgfpathlineto{\pgfqpoint{0.628649in}{1.127863in}}%
\pgfpathlineto{\pgfqpoint{0.556748in}{1.127863in}}%
\pgfpathlineto{\pgfqpoint{0.556748in}{1.172451in}}%
\pgfpathlineto{\pgfqpoint{0.520798in}{1.172451in}}%
\pgfpathlineto{\pgfqpoint{0.520798in}{1.172451in}}%
\pgfpathclose%
\pgfusepath{fill}%
\end{pgfscope}%
\begin{pgfscope}%
\definecolor{textcolor}{rgb}{0.150000,0.150000,0.150000}%
\pgfsetstrokecolor{textcolor}%
\pgfsetfillcolor{textcolor}%
\pgftext[x=0.880300in,y=1.397395in,,base]{\color{textcolor}\rmfamily\fontsize{9.000000}{10.800000}\selectfont \(\displaystyle c_X^-=1\)}%
\end{pgfscope}%
\begin{pgfscope}%
\pgfsetbuttcap%
\pgfsetmiterjoin%
\definecolor{currentfill}{rgb}{1.000000,1.000000,1.000000}%
\pgfsetfillcolor{currentfill}%
\pgfsetfillopacity{0.800000}%
\pgfsetlinewidth{1.003750pt}%
\definecolor{currentstroke}{rgb}{0.800000,0.800000,0.800000}%
\pgfsetstrokecolor{currentstroke}%
\pgfsetstrokeopacity{0.800000}%
\pgfsetdash{}{0pt}%
\pgfpathmoveto{\pgfqpoint{0.785843in}{0.638103in}}%
\pgfpathlineto{\pgfqpoint{1.191192in}{0.638103in}}%
\pgfpathlineto{\pgfqpoint{1.191192in}{1.265450in}}%
\pgfpathlineto{\pgfqpoint{0.785843in}{1.265450in}}%
\pgfpathlineto{\pgfqpoint{0.785843in}{0.638103in}}%
\pgfpathclose%
\pgfusepath{stroke,fill}%
\end{pgfscope}%
\begin{pgfscope}%
\definecolor{textcolor}{rgb}{0.150000,0.150000,0.150000}%
\pgfsetstrokecolor{textcolor}%
\pgfsetfillcolor{textcolor}%
\pgftext[x=0.912291in,y=1.113275in,left,base]{\color{textcolor}\rmfamily\fontsize{9.000000}{10.800000}\selectfont \(\displaystyle c_X^+\)}%
\end{pgfscope}%
\begin{pgfscope}%
\pgfsetroundcap%
\pgfsetroundjoin%
\pgfsetlinewidth{1.003750pt}%
\definecolor{currentstroke}{rgb}{0.003922,0.450980,0.698039}%
\pgfsetstrokecolor{currentstroke}%
\pgfsetdash{}{0pt}%
\pgfpathmoveto{\pgfqpoint{0.824732in}{1.001052in}}%
\pgfpathlineto{\pgfqpoint{0.873343in}{1.001052in}}%
\pgfpathlineto{\pgfqpoint{0.873343in}{1.001052in}}%
\pgfpathlineto{\pgfqpoint{0.970565in}{1.001052in}}%
\pgfpathlineto{\pgfqpoint{0.970565in}{1.001052in}}%
\pgfpathlineto{\pgfqpoint{1.019176in}{1.001052in}}%
\pgfusepath{stroke}%
\end{pgfscope}%
\begin{pgfscope}%
\definecolor{textcolor}{rgb}{0.150000,0.150000,0.150000}%
\pgfsetstrokecolor{textcolor}%
\pgfsetfillcolor{textcolor}%
\pgftext[x=1.096954in,y=0.967024in,left,base]{\color{textcolor}\rmfamily\fontsize{7.000000}{8.400000}\selectfont 1}%
\end{pgfscope}%
\begin{pgfscope}%
\pgfsetroundcap%
\pgfsetroundjoin%
\pgfsetlinewidth{1.003750pt}%
\definecolor{currentstroke}{rgb}{0.870588,0.560784,0.019608}%
\pgfsetstrokecolor{currentstroke}%
\pgfsetdash{}{0pt}%
\pgfpathmoveto{\pgfqpoint{0.824732in}{0.865486in}}%
\pgfpathlineto{\pgfqpoint{0.873343in}{0.865486in}}%
\pgfpathlineto{\pgfqpoint{0.873343in}{0.865486in}}%
\pgfpathlineto{\pgfqpoint{0.970565in}{0.865486in}}%
\pgfpathlineto{\pgfqpoint{0.970565in}{0.865486in}}%
\pgfpathlineto{\pgfqpoint{1.019176in}{0.865486in}}%
\pgfusepath{stroke}%
\end{pgfscope}%
\begin{pgfscope}%
\definecolor{textcolor}{rgb}{0.150000,0.150000,0.150000}%
\pgfsetstrokecolor{textcolor}%
\pgfsetfillcolor{textcolor}%
\pgftext[x=1.096954in,y=0.831458in,left,base]{\color{textcolor}\rmfamily\fontsize{7.000000}{8.400000}\selectfont 2}%
\end{pgfscope}%
\begin{pgfscope}%
\pgfsetroundcap%
\pgfsetroundjoin%
\pgfsetlinewidth{1.003750pt}%
\definecolor{currentstroke}{rgb}{0.007843,0.619608,0.450980}%
\pgfsetstrokecolor{currentstroke}%
\pgfsetdash{}{0pt}%
\pgfpathmoveto{\pgfqpoint{0.824732in}{0.729919in}}%
\pgfpathlineto{\pgfqpoint{0.873343in}{0.729919in}}%
\pgfpathlineto{\pgfqpoint{0.873343in}{0.729919in}}%
\pgfpathlineto{\pgfqpoint{0.970565in}{0.729919in}}%
\pgfpathlineto{\pgfqpoint{0.970565in}{0.729919in}}%
\pgfpathlineto{\pgfqpoint{1.019176in}{0.729919in}}%
\pgfusepath{stroke}%
\end{pgfscope}%
\begin{pgfscope}%
\definecolor{textcolor}{rgb}{0.150000,0.150000,0.150000}%
\pgfsetstrokecolor{textcolor}%
\pgfsetfillcolor{textcolor}%
\pgftext[x=1.096954in,y=0.695892in,left,base]{\color{textcolor}\rmfamily\fontsize{7.000000}{8.400000}\selectfont 4}%
\end{pgfscope}%
\end{pgfpicture}%
\makeatother%
\endgroup%

%% file: figures/experiments/graphs/sparse_smoothing/nodes_attributes/node_classification-Cora-GAT-hidden=8-p_adj_plus=0.0-p_adj_minus=0.0-p_att_plus=0.001-p_att_minus=0.8-multi_class_cert-A.pgf
\begingroup%
\makeatletter%
\begin{pgfpicture}%
\pgfpathrectangle{\pgfpointorigin}{\pgfqpoint{1.375000in}{1.581250in}}%
\pgfusepath{use as bounding box, clip}%
\begin{pgfscope}%
\pgfsetbuttcap%
\pgfsetmiterjoin%
\definecolor{currentfill}{rgb}{1.000000,1.000000,1.000000}%
\pgfsetfillcolor{currentfill}%
\pgfsetlinewidth{0.000000pt}%
\definecolor{currentstroke}{rgb}{1.000000,1.000000,1.000000}%
\pgfsetstrokecolor{currentstroke}%
\pgfsetdash{}{0pt}%
\pgfpathmoveto{\pgfqpoint{0.000000in}{0.000000in}}%
\pgfpathlineto{\pgfqpoint{1.375000in}{0.000000in}}%
\pgfpathlineto{\pgfqpoint{1.375000in}{1.581250in}}%
\pgfpathlineto{\pgfqpoint{0.000000in}{1.581250in}}%
\pgfpathlineto{\pgfqpoint{0.000000in}{0.000000in}}%
\pgfpathclose%
\pgfusepath{fill}%
\end{pgfscope}%
\begin{pgfscope}%
\pgfsetbuttcap%
\pgfsetmiterjoin%
\definecolor{currentfill}{rgb}{1.000000,1.000000,1.000000}%
\pgfsetfillcolor{currentfill}%
\pgfsetlinewidth{0.000000pt}%
\definecolor{currentstroke}{rgb}{0.000000,0.000000,0.000000}%
\pgfsetstrokecolor{currentstroke}%
\pgfsetstrokeopacity{0.000000}%
\pgfsetdash{}{0pt}%
\pgfpathmoveto{\pgfqpoint{0.520798in}{0.442177in}}%
\pgfpathlineto{\pgfqpoint{1.239803in}{0.442177in}}%
\pgfpathlineto{\pgfqpoint{1.239803in}{1.314061in}}%
\pgfpathlineto{\pgfqpoint{0.520798in}{1.314061in}}%
\pgfpathlineto{\pgfqpoint{0.520798in}{0.442177in}}%
\pgfpathclose%
\pgfusepath{fill}%
\end{pgfscope}%
\begin{pgfscope}%
\pgfpathrectangle{\pgfqpoint{0.520798in}{0.442177in}}{\pgfqpoint{0.719005in}{0.871884in}}%
\pgfusepath{clip}%
\pgfsetroundcap%
\pgfsetroundjoin%
\pgfsetlinewidth{0.501875pt}%
\definecolor{currentstroke}{rgb}{0.800000,0.800000,0.800000}%
\pgfsetstrokecolor{currentstroke}%
\pgfsetdash{}{0pt}%
\pgfpathmoveto{\pgfqpoint{0.520798in}{0.442177in}}%
\pgfpathlineto{\pgfqpoint{0.520798in}{1.314061in}}%
\pgfusepath{stroke}%
\end{pgfscope}%
\begin{pgfscope}%
\definecolor{textcolor}{rgb}{0.150000,0.150000,0.150000}%
\pgfsetstrokecolor{textcolor}%
\pgfsetfillcolor{textcolor}%
\pgftext[x=0.520798in,y=0.351899in,,top]{\color{textcolor}\rmfamily\fontsize{8.000000}{9.600000}\selectfont \(\displaystyle {0}\)}%
\end{pgfscope}%
\begin{pgfscope}%
\pgfpathrectangle{\pgfqpoint{0.520798in}{0.442177in}}{\pgfqpoint{0.719005in}{0.871884in}}%
\pgfusepath{clip}%
\pgfsetroundcap%
\pgfsetroundjoin%
\pgfsetlinewidth{0.501875pt}%
\definecolor{currentstroke}{rgb}{0.800000,0.800000,0.800000}%
\pgfsetstrokecolor{currentstroke}%
\pgfsetdash{}{0pt}%
\pgfpathmoveto{\pgfqpoint{0.880300in}{0.442177in}}%
\pgfpathlineto{\pgfqpoint{0.880300in}{1.314061in}}%
\pgfusepath{stroke}%
\end{pgfscope}%
\begin{pgfscope}%
\definecolor{textcolor}{rgb}{0.150000,0.150000,0.150000}%
\pgfsetstrokecolor{textcolor}%
\pgfsetfillcolor{textcolor}%
\pgftext[x=0.880300in,y=0.351899in,,top]{\color{textcolor}\rmfamily\fontsize{8.000000}{9.600000}\selectfont \(\displaystyle {5}\)}%
\end{pgfscope}%
\begin{pgfscope}%
\pgfpathrectangle{\pgfqpoint{0.520798in}{0.442177in}}{\pgfqpoint{0.719005in}{0.871884in}}%
\pgfusepath{clip}%
\pgfsetroundcap%
\pgfsetroundjoin%
\pgfsetlinewidth{0.501875pt}%
\definecolor{currentstroke}{rgb}{0.800000,0.800000,0.800000}%
\pgfsetstrokecolor{currentstroke}%
\pgfsetdash{}{0pt}%
\pgfpathmoveto{\pgfqpoint{1.239803in}{0.442177in}}%
\pgfpathlineto{\pgfqpoint{1.239803in}{1.314061in}}%
\pgfusepath{stroke}%
\end{pgfscope}%
\begin{pgfscope}%
\definecolor{textcolor}{rgb}{0.150000,0.150000,0.150000}%
\pgfsetstrokecolor{textcolor}%
\pgfsetfillcolor{textcolor}%
\pgftext[x=1.239803in,y=0.351899in,,top]{\color{textcolor}\rmfamily\fontsize{8.000000}{9.600000}\selectfont \(\displaystyle {10}\)}%
\end{pgfscope}%
\begin{pgfscope}%
\definecolor{textcolor}{rgb}{0.150000,0.150000,0.150000}%
\pgfsetstrokecolor{textcolor}%
\pgfsetfillcolor{textcolor}%
\pgftext[x=0.880300in,y=0.198219in,,top]{\color{textcolor}\rmfamily\fontsize{10.000000}{12.000000}\selectfont Edit distance \(\displaystyle \epsilon\)}%
\end{pgfscope}%
\begin{pgfscope}%
\pgfpathrectangle{\pgfqpoint{0.520798in}{0.442177in}}{\pgfqpoint{0.719005in}{0.871884in}}%
\pgfusepath{clip}%
\pgfsetroundcap%
\pgfsetroundjoin%
\pgfsetlinewidth{0.501875pt}%
\definecolor{currentstroke}{rgb}{0.800000,0.800000,0.800000}%
\pgfsetstrokecolor{currentstroke}%
\pgfsetdash{}{0pt}%
\pgfpathmoveto{\pgfqpoint{0.520798in}{0.442177in}}%
\pgfpathlineto{\pgfqpoint{1.239803in}{0.442177in}}%
\pgfusepath{stroke}%
\end{pgfscope}%
\begin{pgfscope}%
\definecolor{textcolor}{rgb}{0.150000,0.150000,0.150000}%
\pgfsetstrokecolor{textcolor}%
\pgfsetfillcolor{textcolor}%
\pgftext[x=0.273151in, y=0.403915in, left, base]{\color{textcolor}\rmfamily\fontsize{8.000000}{9.600000}\selectfont 0\%}%
\end{pgfscope}%
\begin{pgfscope}%
\pgfpathrectangle{\pgfqpoint{0.520798in}{0.442177in}}{\pgfqpoint{0.719005in}{0.871884in}}%
\pgfusepath{clip}%
\pgfsetroundcap%
\pgfsetroundjoin%
\pgfsetlinewidth{0.501875pt}%
\definecolor{currentstroke}{rgb}{0.800000,0.800000,0.800000}%
\pgfsetstrokecolor{currentstroke}%
\pgfsetdash{}{0pt}%
\pgfpathmoveto{\pgfqpoint{0.520798in}{0.660148in}}%
\pgfpathlineto{\pgfqpoint{1.239803in}{0.660148in}}%
\pgfusepath{stroke}%
\end{pgfscope}%
\begin{pgfscope}%
\definecolor{textcolor}{rgb}{0.150000,0.150000,0.150000}%
\pgfsetstrokecolor{textcolor}%
\pgfsetfillcolor{textcolor}%
\pgftext[x=0.214138in, y=0.621886in, left, base]{\color{textcolor}\rmfamily\fontsize{8.000000}{9.600000}\selectfont 25\%}%
\end{pgfscope}%
\begin{pgfscope}%
\pgfpathrectangle{\pgfqpoint{0.520798in}{0.442177in}}{\pgfqpoint{0.719005in}{0.871884in}}%
\pgfusepath{clip}%
\pgfsetroundcap%
\pgfsetroundjoin%
\pgfsetlinewidth{0.501875pt}%
\definecolor{currentstroke}{rgb}{0.800000,0.800000,0.800000}%
\pgfsetstrokecolor{currentstroke}%
\pgfsetdash{}{0pt}%
\pgfpathmoveto{\pgfqpoint{0.520798in}{0.878119in}}%
\pgfpathlineto{\pgfqpoint{1.239803in}{0.878119in}}%
\pgfusepath{stroke}%
\end{pgfscope}%
\begin{pgfscope}%
\definecolor{textcolor}{rgb}{0.150000,0.150000,0.150000}%
\pgfsetstrokecolor{textcolor}%
\pgfsetfillcolor{textcolor}%
\pgftext[x=0.214138in, y=0.839857in, left, base]{\color{textcolor}\rmfamily\fontsize{8.000000}{9.600000}\selectfont 50\%}%
\end{pgfscope}%
\begin{pgfscope}%
\pgfpathrectangle{\pgfqpoint{0.520798in}{0.442177in}}{\pgfqpoint{0.719005in}{0.871884in}}%
\pgfusepath{clip}%
\pgfsetroundcap%
\pgfsetroundjoin%
\pgfsetlinewidth{0.501875pt}%
\definecolor{currentstroke}{rgb}{0.800000,0.800000,0.800000}%
\pgfsetstrokecolor{currentstroke}%
\pgfsetdash{}{0pt}%
\pgfpathmoveto{\pgfqpoint{0.520798in}{1.096090in}}%
\pgfpathlineto{\pgfqpoint{1.239803in}{1.096090in}}%
\pgfusepath{stroke}%
\end{pgfscope}%
\begin{pgfscope}%
\definecolor{textcolor}{rgb}{0.150000,0.150000,0.150000}%
\pgfsetstrokecolor{textcolor}%
\pgfsetfillcolor{textcolor}%
\pgftext[x=0.214138in, y=1.057828in, left, base]{\color{textcolor}\rmfamily\fontsize{8.000000}{9.600000}\selectfont 75\%}%
\end{pgfscope}%
\begin{pgfscope}%
\pgfpathrectangle{\pgfqpoint{0.520798in}{0.442177in}}{\pgfqpoint{0.719005in}{0.871884in}}%
\pgfusepath{clip}%
\pgfsetroundcap%
\pgfsetroundjoin%
\pgfsetlinewidth{0.501875pt}%
\definecolor{currentstroke}{rgb}{0.800000,0.800000,0.800000}%
\pgfsetstrokecolor{currentstroke}%
\pgfsetdash{}{0pt}%
\pgfpathmoveto{\pgfqpoint{0.520798in}{1.314061in}}%
\pgfpathlineto{\pgfqpoint{1.239803in}{1.314061in}}%
\pgfusepath{stroke}%
\end{pgfscope}%
\begin{pgfscope}%
\definecolor{textcolor}{rgb}{0.150000,0.150000,0.150000}%
\pgfsetstrokecolor{textcolor}%
\pgfsetfillcolor{textcolor}%
\pgftext[x=0.155124in, y=1.275799in, left, base]{\color{textcolor}\rmfamily\fontsize{8.000000}{9.600000}\selectfont 100\%}%
\end{pgfscope}%
\begin{pgfscope}%
\definecolor{textcolor}{rgb}{0.150000,0.150000,0.150000}%
\pgfsetstrokecolor{textcolor}%
\pgfsetfillcolor{textcolor}%
\pgftext[x=0.099569in,y=0.878119in,,bottom,rotate=90.000000]{\color{textcolor}\rmfamily\fontsize{10.000000}{12.000000}\selectfont Cert. Acc.}%
\end{pgfscope}%
\begin{pgfscope}%
\pgfsetrectcap%
\pgfsetmiterjoin%
\pgfsetlinewidth{0.752812pt}%
\definecolor{currentstroke}{rgb}{0.700000,0.700000,0.700000}%
\pgfsetstrokecolor{currentstroke}%
\pgfsetdash{}{0pt}%
\pgfpathmoveto{\pgfqpoint{0.520798in}{0.442177in}}%
\pgfpathlineto{\pgfqpoint{0.520798in}{1.314061in}}%
\pgfusepath{stroke}%
\end{pgfscope}%
\begin{pgfscope}%
\pgfsetrectcap%
\pgfsetmiterjoin%
\pgfsetlinewidth{0.752812pt}%
\definecolor{currentstroke}{rgb}{0.700000,0.700000,0.700000}%
\pgfsetstrokecolor{currentstroke}%
\pgfsetdash{}{0pt}%
\pgfpathmoveto{\pgfqpoint{1.239803in}{0.442177in}}%
\pgfpathlineto{\pgfqpoint{1.239803in}{1.314061in}}%
\pgfusepath{stroke}%
\end{pgfscope}%
\begin{pgfscope}%
\pgfsetrectcap%
\pgfsetmiterjoin%
\pgfsetlinewidth{0.752812pt}%
\definecolor{currentstroke}{rgb}{0.700000,0.700000,0.700000}%
\pgfsetstrokecolor{currentstroke}%
\pgfsetdash{}{0pt}%
\pgfpathmoveto{\pgfqpoint{0.520798in}{0.442177in}}%
\pgfpathlineto{\pgfqpoint{1.239803in}{0.442177in}}%
\pgfusepath{stroke}%
\end{pgfscope}%
\begin{pgfscope}%
\pgfsetrectcap%
\pgfsetmiterjoin%
\pgfsetlinewidth{0.752812pt}%
\definecolor{currentstroke}{rgb}{0.700000,0.700000,0.700000}%
\pgfsetstrokecolor{currentstroke}%
\pgfsetdash{}{0pt}%
\pgfpathmoveto{\pgfqpoint{0.520798in}{1.314061in}}%
\pgfpathlineto{\pgfqpoint{1.239803in}{1.314061in}}%
\pgfusepath{stroke}%
\end{pgfscope}%
\begin{pgfscope}%
\pgfpathrectangle{\pgfqpoint{0.520798in}{0.442177in}}{\pgfqpoint{0.719005in}{0.871884in}}%
\pgfusepath{clip}%
\pgfsetroundcap%
\pgfsetroundjoin%
\pgfsetlinewidth{1.003750pt}%
\definecolor{currentstroke}{rgb}{0.003922,0.450980,0.698039}%
\pgfsetstrokecolor{currentstroke}%
\pgfsetdash{}{0pt}%
\pgfpathmoveto{\pgfqpoint{0.520798in}{1.159297in}}%
\pgfpathlineto{\pgfqpoint{0.556748in}{1.159297in}}%
\pgfpathlineto{\pgfqpoint{0.556748in}{1.099036in}}%
\pgfpathlineto{\pgfqpoint{0.628649in}{1.099036in}}%
\pgfpathlineto{\pgfqpoint{0.628649in}{1.003900in}}%
\pgfpathlineto{\pgfqpoint{0.700549in}{1.003900in}}%
\pgfpathlineto{\pgfqpoint{0.700549in}{0.739052in}}%
\pgfpathlineto{\pgfqpoint{0.772450in}{0.739052in}}%
\pgfpathlineto{\pgfqpoint{0.772450in}{0.604375in}}%
\pgfpathlineto{\pgfqpoint{0.844350in}{0.604375in}}%
\pgfpathlineto{\pgfqpoint{0.844350in}{0.587372in}}%
\pgfpathlineto{\pgfqpoint{0.916251in}{0.587372in}}%
\pgfpathlineto{\pgfqpoint{0.916251in}{0.577724in}}%
\pgfpathlineto{\pgfqpoint{0.988151in}{0.577724in}}%
\pgfpathlineto{\pgfqpoint{0.988151in}{0.559377in}}%
\pgfpathlineto{\pgfqpoint{1.060052in}{0.559377in}}%
\pgfpathlineto{\pgfqpoint{1.060052in}{0.506471in}}%
\pgfpathlineto{\pgfqpoint{1.131952in}{0.506471in}}%
\pgfpathlineto{\pgfqpoint{1.131952in}{0.499591in}}%
\pgfpathlineto{\pgfqpoint{1.203853in}{0.499591in}}%
\pgfpathlineto{\pgfqpoint{1.203853in}{0.474680in}}%
\pgfpathlineto{\pgfqpoint{1.241470in}{0.474680in}}%
\pgfusepath{stroke}%
\end{pgfscope}%
\begin{pgfscope}%
\pgfpathrectangle{\pgfqpoint{0.520798in}{0.442177in}}{\pgfqpoint{0.719005in}{0.871884in}}%
\pgfusepath{clip}%
\pgfsetbuttcap%
\pgfsetroundjoin%
\definecolor{currentfill}{rgb}{0.003922,0.450980,0.698039}%
\pgfsetfillcolor{currentfill}%
\pgfsetfillopacity{0.500000}%
\pgfsetlinewidth{0.000000pt}%
\definecolor{currentstroke}{rgb}{0.003922,0.450980,0.698039}%
\pgfsetstrokecolor{currentstroke}%
\pgfsetstrokeopacity{0.500000}%
\pgfsetdash{}{0pt}%
\pgfpathmoveto{\pgfqpoint{0.520798in}{1.172451in}}%
\pgfpathlineto{\pgfqpoint{0.520798in}{1.146143in}}%
\pgfpathlineto{\pgfqpoint{0.556748in}{1.146143in}}%
\pgfpathlineto{\pgfqpoint{0.556748in}{1.082811in}}%
\pgfpathlineto{\pgfqpoint{0.628649in}{1.082811in}}%
\pgfpathlineto{\pgfqpoint{0.628649in}{0.989028in}}%
\pgfpathlineto{\pgfqpoint{0.700549in}{0.989028in}}%
\pgfpathlineto{\pgfqpoint{0.700549in}{0.734887in}}%
\pgfpathlineto{\pgfqpoint{0.772450in}{0.734887in}}%
\pgfpathlineto{\pgfqpoint{0.772450in}{0.592079in}}%
\pgfpathlineto{\pgfqpoint{0.844350in}{0.592079in}}%
\pgfpathlineto{\pgfqpoint{0.844350in}{0.573918in}}%
\pgfpathlineto{\pgfqpoint{0.916251in}{0.573918in}}%
\pgfpathlineto{\pgfqpoint{0.916251in}{0.565347in}}%
\pgfpathlineto{\pgfqpoint{0.988151in}{0.565347in}}%
\pgfpathlineto{\pgfqpoint{0.988151in}{0.548400in}}%
\pgfpathlineto{\pgfqpoint{1.060052in}{0.548400in}}%
\pgfpathlineto{\pgfqpoint{1.060052in}{0.496812in}}%
\pgfpathlineto{\pgfqpoint{1.131952in}{0.496812in}}%
\pgfpathlineto{\pgfqpoint{1.131952in}{0.490452in}}%
\pgfpathlineto{\pgfqpoint{1.203853in}{0.490452in}}%
\pgfpathlineto{\pgfqpoint{1.203853in}{0.468317in}}%
\pgfpathlineto{\pgfqpoint{1.275753in}{0.468317in}}%
\pgfpathlineto{\pgfqpoint{1.275753in}{0.452427in}}%
\pgfpathlineto{\pgfqpoint{1.347654in}{0.452427in}}%
\pgfpathlineto{\pgfqpoint{1.347654in}{0.449330in}}%
\pgfpathlineto{\pgfqpoint{1.419554in}{0.449330in}}%
\pgfpathlineto{\pgfqpoint{1.419554in}{0.442177in}}%
\pgfpathlineto{\pgfqpoint{2.066659in}{0.442177in}}%
\pgfpathlineto{\pgfqpoint{2.066659in}{0.442177in}}%
\pgfpathlineto{\pgfqpoint{2.677813in}{0.442177in}}%
\pgfpathlineto{\pgfqpoint{2.677813in}{0.442177in}}%
\pgfpathlineto{\pgfqpoint{2.677813in}{0.442177in}}%
\pgfpathlineto{\pgfqpoint{2.066659in}{0.442177in}}%
\pgfpathlineto{\pgfqpoint{2.066659in}{0.442177in}}%
\pgfpathlineto{\pgfqpoint{1.419554in}{0.442177in}}%
\pgfpathlineto{\pgfqpoint{1.419554in}{0.455111in}}%
\pgfpathlineto{\pgfqpoint{1.347654in}{0.455111in}}%
\pgfpathlineto{\pgfqpoint{1.347654in}{0.459922in}}%
\pgfpathlineto{\pgfqpoint{1.275753in}{0.459922in}}%
\pgfpathlineto{\pgfqpoint{1.275753in}{0.481043in}}%
\pgfpathlineto{\pgfqpoint{1.203853in}{0.481043in}}%
\pgfpathlineto{\pgfqpoint{1.203853in}{0.508729in}}%
\pgfpathlineto{\pgfqpoint{1.131952in}{0.508729in}}%
\pgfpathlineto{\pgfqpoint{1.131952in}{0.516129in}}%
\pgfpathlineto{\pgfqpoint{1.060052in}{0.516129in}}%
\pgfpathlineto{\pgfqpoint{1.060052in}{0.570354in}}%
\pgfpathlineto{\pgfqpoint{0.988151in}{0.570354in}}%
\pgfpathlineto{\pgfqpoint{0.988151in}{0.590101in}}%
\pgfpathlineto{\pgfqpoint{0.916251in}{0.590101in}}%
\pgfpathlineto{\pgfqpoint{0.916251in}{0.600827in}}%
\pgfpathlineto{\pgfqpoint{0.844350in}{0.600827in}}%
\pgfpathlineto{\pgfqpoint{0.844350in}{0.616671in}}%
\pgfpathlineto{\pgfqpoint{0.772450in}{0.616671in}}%
\pgfpathlineto{\pgfqpoint{0.772450in}{0.743218in}}%
\pgfpathlineto{\pgfqpoint{0.700549in}{0.743218in}}%
\pgfpathlineto{\pgfqpoint{0.700549in}{1.018772in}}%
\pgfpathlineto{\pgfqpoint{0.628649in}{1.018772in}}%
\pgfpathlineto{\pgfqpoint{0.628649in}{1.115261in}}%
\pgfpathlineto{\pgfqpoint{0.556748in}{1.115261in}}%
\pgfpathlineto{\pgfqpoint{0.556748in}{1.172451in}}%
\pgfpathlineto{\pgfqpoint{0.520798in}{1.172451in}}%
\pgfpathlineto{\pgfqpoint{0.520798in}{1.172451in}}%
\pgfpathclose%
\pgfusepath{fill}%
\end{pgfscope}%
\begin{pgfscope}%
\pgfpathrectangle{\pgfqpoint{0.520798in}{0.442177in}}{\pgfqpoint{0.719005in}{0.871884in}}%
\pgfusepath{clip}%
\pgfsetroundcap%
\pgfsetroundjoin%
\pgfsetlinewidth{1.003750pt}%
\definecolor{currentstroke}{rgb}{0.870588,0.560784,0.019608}%
\pgfsetstrokecolor{currentstroke}%
\pgfsetdash{}{0pt}%
\pgfpathmoveto{\pgfqpoint{0.520798in}{1.159297in}}%
\pgfpathlineto{\pgfqpoint{0.556748in}{1.159297in}}%
\pgfpathlineto{\pgfqpoint{0.556748in}{1.099036in}}%
\pgfpathlineto{\pgfqpoint{0.628649in}{1.099036in}}%
\pgfpathlineto{\pgfqpoint{0.628649in}{1.003900in}}%
\pgfpathlineto{\pgfqpoint{0.700549in}{1.003900in}}%
\pgfpathlineto{\pgfqpoint{0.700549in}{0.739052in}}%
\pgfpathlineto{\pgfqpoint{0.772450in}{0.739052in}}%
\pgfpathlineto{\pgfqpoint{0.772450in}{0.604375in}}%
\pgfpathlineto{\pgfqpoint{0.844350in}{0.604375in}}%
\pgfpathlineto{\pgfqpoint{0.844350in}{0.590694in}}%
\pgfpathlineto{\pgfqpoint{0.916251in}{0.590694in}}%
\pgfpathlineto{\pgfqpoint{0.916251in}{0.581125in}}%
\pgfpathlineto{\pgfqpoint{0.988151in}{0.581125in}}%
\pgfpathlineto{\pgfqpoint{0.988151in}{0.559377in}}%
\pgfpathlineto{\pgfqpoint{1.060052in}{0.559377in}}%
\pgfpathlineto{\pgfqpoint{1.060052in}{0.506471in}}%
\pgfpathlineto{\pgfqpoint{1.131952in}{0.506471in}}%
\pgfpathlineto{\pgfqpoint{1.131952in}{0.499670in}}%
\pgfpathlineto{\pgfqpoint{1.203853in}{0.499670in}}%
\pgfpathlineto{\pgfqpoint{1.203853in}{0.474680in}}%
\pgfpathlineto{\pgfqpoint{1.241470in}{0.474680in}}%
\pgfusepath{stroke}%
\end{pgfscope}%
\begin{pgfscope}%
\pgfpathrectangle{\pgfqpoint{0.520798in}{0.442177in}}{\pgfqpoint{0.719005in}{0.871884in}}%
\pgfusepath{clip}%
\pgfsetbuttcap%
\pgfsetroundjoin%
\definecolor{currentfill}{rgb}{0.870588,0.560784,0.019608}%
\pgfsetfillcolor{currentfill}%
\pgfsetfillopacity{0.500000}%
\pgfsetlinewidth{0.000000pt}%
\definecolor{currentstroke}{rgb}{0.870588,0.560784,0.019608}%
\pgfsetstrokecolor{currentstroke}%
\pgfsetstrokeopacity{0.500000}%
\pgfsetdash{}{0pt}%
\pgfpathmoveto{\pgfqpoint{0.520798in}{1.172451in}}%
\pgfpathlineto{\pgfqpoint{0.520798in}{1.146143in}}%
\pgfpathlineto{\pgfqpoint{0.556748in}{1.146143in}}%
\pgfpathlineto{\pgfqpoint{0.556748in}{1.082811in}}%
\pgfpathlineto{\pgfqpoint{0.628649in}{1.082811in}}%
\pgfpathlineto{\pgfqpoint{0.628649in}{0.989028in}}%
\pgfpathlineto{\pgfqpoint{0.700549in}{0.989028in}}%
\pgfpathlineto{\pgfqpoint{0.700549in}{0.734887in}}%
\pgfpathlineto{\pgfqpoint{0.772450in}{0.734887in}}%
\pgfpathlineto{\pgfqpoint{0.772450in}{0.592079in}}%
\pgfpathlineto{\pgfqpoint{0.844350in}{0.592079in}}%
\pgfpathlineto{\pgfqpoint{0.844350in}{0.577867in}}%
\pgfpathlineto{\pgfqpoint{0.916251in}{0.577867in}}%
\pgfpathlineto{\pgfqpoint{0.916251in}{0.568507in}}%
\pgfpathlineto{\pgfqpoint{0.988151in}{0.568507in}}%
\pgfpathlineto{\pgfqpoint{0.988151in}{0.548400in}}%
\pgfpathlineto{\pgfqpoint{1.060052in}{0.548400in}}%
\pgfpathlineto{\pgfqpoint{1.060052in}{0.496812in}}%
\pgfpathlineto{\pgfqpoint{1.131952in}{0.496812in}}%
\pgfpathlineto{\pgfqpoint{1.131952in}{0.490408in}}%
\pgfpathlineto{\pgfqpoint{1.203853in}{0.490408in}}%
\pgfpathlineto{\pgfqpoint{1.203853in}{0.468317in}}%
\pgfpathlineto{\pgfqpoint{1.275753in}{0.468317in}}%
\pgfpathlineto{\pgfqpoint{1.275753in}{0.452494in}}%
\pgfpathlineto{\pgfqpoint{1.347654in}{0.452494in}}%
\pgfpathlineto{\pgfqpoint{1.347654in}{0.449330in}}%
\pgfpathlineto{\pgfqpoint{1.419554in}{0.449330in}}%
\pgfpathlineto{\pgfqpoint{1.419554in}{0.442177in}}%
\pgfpathlineto{\pgfqpoint{2.066659in}{0.442177in}}%
\pgfpathlineto{\pgfqpoint{2.066659in}{0.442177in}}%
\pgfpathlineto{\pgfqpoint{2.677813in}{0.442177in}}%
\pgfpathlineto{\pgfqpoint{2.677813in}{0.442177in}}%
\pgfpathlineto{\pgfqpoint{2.677813in}{0.442177in}}%
\pgfpathlineto{\pgfqpoint{2.066659in}{0.442177in}}%
\pgfpathlineto{\pgfqpoint{2.066659in}{0.442177in}}%
\pgfpathlineto{\pgfqpoint{1.419554in}{0.442177in}}%
\pgfpathlineto{\pgfqpoint{1.419554in}{0.455111in}}%
\pgfpathlineto{\pgfqpoint{1.347654in}{0.455111in}}%
\pgfpathlineto{\pgfqpoint{1.347654in}{0.460487in}}%
\pgfpathlineto{\pgfqpoint{1.275753in}{0.460487in}}%
\pgfpathlineto{\pgfqpoint{1.275753in}{0.481043in}}%
\pgfpathlineto{\pgfqpoint{1.203853in}{0.481043in}}%
\pgfpathlineto{\pgfqpoint{1.203853in}{0.508932in}}%
\pgfpathlineto{\pgfqpoint{1.131952in}{0.508932in}}%
\pgfpathlineto{\pgfqpoint{1.131952in}{0.516129in}}%
\pgfpathlineto{\pgfqpoint{1.060052in}{0.516129in}}%
\pgfpathlineto{\pgfqpoint{1.060052in}{0.570354in}}%
\pgfpathlineto{\pgfqpoint{0.988151in}{0.570354in}}%
\pgfpathlineto{\pgfqpoint{0.988151in}{0.593742in}}%
\pgfpathlineto{\pgfqpoint{0.916251in}{0.593742in}}%
\pgfpathlineto{\pgfqpoint{0.916251in}{0.603520in}}%
\pgfpathlineto{\pgfqpoint{0.844350in}{0.603520in}}%
\pgfpathlineto{\pgfqpoint{0.844350in}{0.616671in}}%
\pgfpathlineto{\pgfqpoint{0.772450in}{0.616671in}}%
\pgfpathlineto{\pgfqpoint{0.772450in}{0.743218in}}%
\pgfpathlineto{\pgfqpoint{0.700549in}{0.743218in}}%
\pgfpathlineto{\pgfqpoint{0.700549in}{1.018772in}}%
\pgfpathlineto{\pgfqpoint{0.628649in}{1.018772in}}%
\pgfpathlineto{\pgfqpoint{0.628649in}{1.115261in}}%
\pgfpathlineto{\pgfqpoint{0.556748in}{1.115261in}}%
\pgfpathlineto{\pgfqpoint{0.556748in}{1.172451in}}%
\pgfpathlineto{\pgfqpoint{0.520798in}{1.172451in}}%
\pgfpathlineto{\pgfqpoint{0.520798in}{1.172451in}}%
\pgfpathclose%
\pgfusepath{fill}%
\end{pgfscope}%
\begin{pgfscope}%
\pgfpathrectangle{\pgfqpoint{0.520798in}{0.442177in}}{\pgfqpoint{0.719005in}{0.871884in}}%
\pgfusepath{clip}%
\pgfsetroundcap%
\pgfsetroundjoin%
\pgfsetlinewidth{1.003750pt}%
\definecolor{currentstroke}{rgb}{0.007843,0.619608,0.450980}%
\pgfsetstrokecolor{currentstroke}%
\pgfsetdash{}{0pt}%
\pgfpathmoveto{\pgfqpoint{0.520798in}{1.159297in}}%
\pgfpathlineto{\pgfqpoint{0.556748in}{1.159297in}}%
\pgfpathlineto{\pgfqpoint{0.556748in}{1.099036in}}%
\pgfpathlineto{\pgfqpoint{0.628649in}{1.099036in}}%
\pgfpathlineto{\pgfqpoint{0.628649in}{1.003900in}}%
\pgfpathlineto{\pgfqpoint{0.700549in}{1.003900in}}%
\pgfpathlineto{\pgfqpoint{0.700549in}{0.739052in}}%
\pgfpathlineto{\pgfqpoint{0.772450in}{0.739052in}}%
\pgfpathlineto{\pgfqpoint{0.772450in}{0.604375in}}%
\pgfpathlineto{\pgfqpoint{0.844350in}{0.604375in}}%
\pgfpathlineto{\pgfqpoint{0.844350in}{0.590694in}}%
\pgfpathlineto{\pgfqpoint{0.916251in}{0.590694in}}%
\pgfpathlineto{\pgfqpoint{0.916251in}{0.581125in}}%
\pgfpathlineto{\pgfqpoint{0.988151in}{0.581125in}}%
\pgfpathlineto{\pgfqpoint{0.988151in}{0.559377in}}%
\pgfpathlineto{\pgfqpoint{1.060052in}{0.559377in}}%
\pgfpathlineto{\pgfqpoint{1.060052in}{0.506471in}}%
\pgfpathlineto{\pgfqpoint{1.131952in}{0.506471in}}%
\pgfpathlineto{\pgfqpoint{1.131952in}{0.499670in}}%
\pgfpathlineto{\pgfqpoint{1.203853in}{0.499670in}}%
\pgfpathlineto{\pgfqpoint{1.203853in}{0.474680in}}%
\pgfpathlineto{\pgfqpoint{1.241470in}{0.474680in}}%
\pgfusepath{stroke}%
\end{pgfscope}%
\begin{pgfscope}%
\pgfpathrectangle{\pgfqpoint{0.520798in}{0.442177in}}{\pgfqpoint{0.719005in}{0.871884in}}%
\pgfusepath{clip}%
\pgfsetbuttcap%
\pgfsetroundjoin%
\definecolor{currentfill}{rgb}{0.007843,0.619608,0.450980}%
\pgfsetfillcolor{currentfill}%
\pgfsetfillopacity{0.500000}%
\pgfsetlinewidth{0.000000pt}%
\definecolor{currentstroke}{rgb}{0.007843,0.619608,0.450980}%
\pgfsetstrokecolor{currentstroke}%
\pgfsetstrokeopacity{0.500000}%
\pgfsetdash{}{0pt}%
\pgfpathmoveto{\pgfqpoint{0.520798in}{1.172451in}}%
\pgfpathlineto{\pgfqpoint{0.520798in}{1.146143in}}%
\pgfpathlineto{\pgfqpoint{0.556748in}{1.146143in}}%
\pgfpathlineto{\pgfqpoint{0.556748in}{1.082811in}}%
\pgfpathlineto{\pgfqpoint{0.628649in}{1.082811in}}%
\pgfpathlineto{\pgfqpoint{0.628649in}{0.989028in}}%
\pgfpathlineto{\pgfqpoint{0.700549in}{0.989028in}}%
\pgfpathlineto{\pgfqpoint{0.700549in}{0.734887in}}%
\pgfpathlineto{\pgfqpoint{0.772450in}{0.734887in}}%
\pgfpathlineto{\pgfqpoint{0.772450in}{0.592079in}}%
\pgfpathlineto{\pgfqpoint{0.844350in}{0.592079in}}%
\pgfpathlineto{\pgfqpoint{0.844350in}{0.577867in}}%
\pgfpathlineto{\pgfqpoint{0.916251in}{0.577867in}}%
\pgfpathlineto{\pgfqpoint{0.916251in}{0.568507in}}%
\pgfpathlineto{\pgfqpoint{0.988151in}{0.568507in}}%
\pgfpathlineto{\pgfqpoint{0.988151in}{0.548400in}}%
\pgfpathlineto{\pgfqpoint{1.060052in}{0.548400in}}%
\pgfpathlineto{\pgfqpoint{1.060052in}{0.496812in}}%
\pgfpathlineto{\pgfqpoint{1.131952in}{0.496812in}}%
\pgfpathlineto{\pgfqpoint{1.131952in}{0.490408in}}%
\pgfpathlineto{\pgfqpoint{1.203853in}{0.490408in}}%
\pgfpathlineto{\pgfqpoint{1.203853in}{0.468317in}}%
\pgfpathlineto{\pgfqpoint{1.275753in}{0.468317in}}%
\pgfpathlineto{\pgfqpoint{1.275753in}{0.452494in}}%
\pgfpathlineto{\pgfqpoint{1.347654in}{0.452494in}}%
\pgfpathlineto{\pgfqpoint{1.347654in}{0.449330in}}%
\pgfpathlineto{\pgfqpoint{1.419554in}{0.449330in}}%
\pgfpathlineto{\pgfqpoint{1.419554in}{0.442177in}}%
\pgfpathlineto{\pgfqpoint{2.066659in}{0.442177in}}%
\pgfpathlineto{\pgfqpoint{2.066659in}{0.442177in}}%
\pgfpathlineto{\pgfqpoint{2.677813in}{0.442177in}}%
\pgfpathlineto{\pgfqpoint{2.677813in}{0.442177in}}%
\pgfpathlineto{\pgfqpoint{2.677813in}{0.442177in}}%
\pgfpathlineto{\pgfqpoint{2.066659in}{0.442177in}}%
\pgfpathlineto{\pgfqpoint{2.066659in}{0.442177in}}%
\pgfpathlineto{\pgfqpoint{1.419554in}{0.442177in}}%
\pgfpathlineto{\pgfqpoint{1.419554in}{0.455111in}}%
\pgfpathlineto{\pgfqpoint{1.347654in}{0.455111in}}%
\pgfpathlineto{\pgfqpoint{1.347654in}{0.460487in}}%
\pgfpathlineto{\pgfqpoint{1.275753in}{0.460487in}}%
\pgfpathlineto{\pgfqpoint{1.275753in}{0.481043in}}%
\pgfpathlineto{\pgfqpoint{1.203853in}{0.481043in}}%
\pgfpathlineto{\pgfqpoint{1.203853in}{0.508932in}}%
\pgfpathlineto{\pgfqpoint{1.131952in}{0.508932in}}%
\pgfpathlineto{\pgfqpoint{1.131952in}{0.516129in}}%
\pgfpathlineto{\pgfqpoint{1.060052in}{0.516129in}}%
\pgfpathlineto{\pgfqpoint{1.060052in}{0.570354in}}%
\pgfpathlineto{\pgfqpoint{0.988151in}{0.570354in}}%
\pgfpathlineto{\pgfqpoint{0.988151in}{0.593742in}}%
\pgfpathlineto{\pgfqpoint{0.916251in}{0.593742in}}%
\pgfpathlineto{\pgfqpoint{0.916251in}{0.603520in}}%
\pgfpathlineto{\pgfqpoint{0.844350in}{0.603520in}}%
\pgfpathlineto{\pgfqpoint{0.844350in}{0.616671in}}%
\pgfpathlineto{\pgfqpoint{0.772450in}{0.616671in}}%
\pgfpathlineto{\pgfqpoint{0.772450in}{0.743218in}}%
\pgfpathlineto{\pgfqpoint{0.700549in}{0.743218in}}%
\pgfpathlineto{\pgfqpoint{0.700549in}{1.018772in}}%
\pgfpathlineto{\pgfqpoint{0.628649in}{1.018772in}}%
\pgfpathlineto{\pgfqpoint{0.628649in}{1.115261in}}%
\pgfpathlineto{\pgfqpoint{0.556748in}{1.115261in}}%
\pgfpathlineto{\pgfqpoint{0.556748in}{1.172451in}}%
\pgfpathlineto{\pgfqpoint{0.520798in}{1.172451in}}%
\pgfpathlineto{\pgfqpoint{0.520798in}{1.172451in}}%
\pgfpathclose%
\pgfusepath{fill}%
\end{pgfscope}%
\begin{pgfscope}%
\definecolor{textcolor}{rgb}{0.150000,0.150000,0.150000}%
\pgfsetstrokecolor{textcolor}%
\pgfsetfillcolor{textcolor}%
\pgftext[x=0.880300in,y=1.397395in,,base]{\color{textcolor}\rmfamily\fontsize{9.000000}{10.800000}\selectfont \(\displaystyle c_X^+=1\)}%
\end{pgfscope}%
\begin{pgfscope}%
\pgfsetbuttcap%
\pgfsetmiterjoin%
\definecolor{currentfill}{rgb}{1.000000,1.000000,1.000000}%
\pgfsetfillcolor{currentfill}%
\pgfsetfillopacity{0.800000}%
\pgfsetlinewidth{1.003750pt}%
\definecolor{currentstroke}{rgb}{0.800000,0.800000,0.800000}%
\pgfsetstrokecolor{currentstroke}%
\pgfsetstrokeopacity{0.800000}%
\pgfsetdash{}{0pt}%
\pgfpathmoveto{\pgfqpoint{0.785843in}{0.638103in}}%
\pgfpathlineto{\pgfqpoint{1.191192in}{0.638103in}}%
\pgfpathlineto{\pgfqpoint{1.191192in}{1.265450in}}%
\pgfpathlineto{\pgfqpoint{0.785843in}{1.265450in}}%
\pgfpathlineto{\pgfqpoint{0.785843in}{0.638103in}}%
\pgfpathclose%
\pgfusepath{stroke,fill}%
\end{pgfscope}%
\begin{pgfscope}%
\definecolor{textcolor}{rgb}{0.150000,0.150000,0.150000}%
\pgfsetstrokecolor{textcolor}%
\pgfsetfillcolor{textcolor}%
\pgftext[x=0.912291in,y=1.113275in,left,base]{\color{textcolor}\rmfamily\fontsize{9.000000}{10.800000}\selectfont \(\displaystyle c_X^-\)}%
\end{pgfscope}%
\begin{pgfscope}%
\pgfsetroundcap%
\pgfsetroundjoin%
\pgfsetlinewidth{1.003750pt}%
\definecolor{currentstroke}{rgb}{0.003922,0.450980,0.698039}%
\pgfsetstrokecolor{currentstroke}%
\pgfsetdash{}{0pt}%
\pgfpathmoveto{\pgfqpoint{0.824732in}{1.001052in}}%
\pgfpathlineto{\pgfqpoint{0.873343in}{1.001052in}}%
\pgfpathlineto{\pgfqpoint{0.873343in}{1.001052in}}%
\pgfpathlineto{\pgfqpoint{0.970565in}{1.001052in}}%
\pgfpathlineto{\pgfqpoint{0.970565in}{1.001052in}}%
\pgfpathlineto{\pgfqpoint{1.019176in}{1.001052in}}%
\pgfusepath{stroke}%
\end{pgfscope}%
\begin{pgfscope}%
\definecolor{textcolor}{rgb}{0.150000,0.150000,0.150000}%
\pgfsetstrokecolor{textcolor}%
\pgfsetfillcolor{textcolor}%
\pgftext[x=1.096954in,y=0.967024in,left,base]{\color{textcolor}\rmfamily\fontsize{7.000000}{8.400000}\selectfont 1}%
\end{pgfscope}%
\begin{pgfscope}%
\pgfsetroundcap%
\pgfsetroundjoin%
\pgfsetlinewidth{1.003750pt}%
\definecolor{currentstroke}{rgb}{0.870588,0.560784,0.019608}%
\pgfsetstrokecolor{currentstroke}%
\pgfsetdash{}{0pt}%
\pgfpathmoveto{\pgfqpoint{0.824732in}{0.865486in}}%
\pgfpathlineto{\pgfqpoint{0.873343in}{0.865486in}}%
\pgfpathlineto{\pgfqpoint{0.873343in}{0.865486in}}%
\pgfpathlineto{\pgfqpoint{0.970565in}{0.865486in}}%
\pgfpathlineto{\pgfqpoint{0.970565in}{0.865486in}}%
\pgfpathlineto{\pgfqpoint{1.019176in}{0.865486in}}%
\pgfusepath{stroke}%
\end{pgfscope}%
\begin{pgfscope}%
\definecolor{textcolor}{rgb}{0.150000,0.150000,0.150000}%
\pgfsetstrokecolor{textcolor}%
\pgfsetfillcolor{textcolor}%
\pgftext[x=1.096954in,y=0.831458in,left,base]{\color{textcolor}\rmfamily\fontsize{7.000000}{8.400000}\selectfont 2}%
\end{pgfscope}%
\begin{pgfscope}%
\pgfsetroundcap%
\pgfsetroundjoin%
\pgfsetlinewidth{1.003750pt}%
\definecolor{currentstroke}{rgb}{0.007843,0.619608,0.450980}%
\pgfsetstrokecolor{currentstroke}%
\pgfsetdash{}{0pt}%
\pgfpathmoveto{\pgfqpoint{0.824732in}{0.729919in}}%
\pgfpathlineto{\pgfqpoint{0.873343in}{0.729919in}}%
\pgfpathlineto{\pgfqpoint{0.873343in}{0.729919in}}%
\pgfpathlineto{\pgfqpoint{0.970565in}{0.729919in}}%
\pgfpathlineto{\pgfqpoint{0.970565in}{0.729919in}}%
\pgfpathlineto{\pgfqpoint{1.019176in}{0.729919in}}%
\pgfusepath{stroke}%
\end{pgfscope}%
\begin{pgfscope}%
\definecolor{textcolor}{rgb}{0.150000,0.150000,0.150000}%
\pgfsetstrokecolor{textcolor}%
\pgfsetfillcolor{textcolor}%
\pgftext[x=1.096954in,y=0.695892in,left,base]{\color{textcolor}\rmfamily\fontsize{7.000000}{8.400000}\selectfont 4}%
\end{pgfscope}%
\end{pgfpicture}%
\makeatother%
\endgroup%

%% file: figures/experiments/graphs/sparse_smoothing/nodes_structure/node_classification-Cora-GAT-hidden=8-p_adj_plus=0.001-p_adj_minus=0.8-p_att_plus=0.0-p_att_minus=0.0-multi_class_cert-B.pgf
\begingroup%
\makeatletter%
\begin{pgfpicture}%
\pgfpathrectangle{\pgfpointorigin}{\pgfqpoint{1.375000in}{1.581250in}}%
\pgfusepath{use as bounding box, clip}%
\begin{pgfscope}%
\pgfsetbuttcap%
\pgfsetmiterjoin%
\definecolor{currentfill}{rgb}{1.000000,1.000000,1.000000}%
\pgfsetfillcolor{currentfill}%
\pgfsetlinewidth{0.000000pt}%
\definecolor{currentstroke}{rgb}{1.000000,1.000000,1.000000}%
\pgfsetstrokecolor{currentstroke}%
\pgfsetdash{}{0pt}%
\pgfpathmoveto{\pgfqpoint{0.000000in}{0.000000in}}%
\pgfpathlineto{\pgfqpoint{1.375000in}{0.000000in}}%
\pgfpathlineto{\pgfqpoint{1.375000in}{1.581250in}}%
\pgfpathlineto{\pgfqpoint{0.000000in}{1.581250in}}%
\pgfpathlineto{\pgfqpoint{0.000000in}{0.000000in}}%
\pgfpathclose%
\pgfusepath{fill}%
\end{pgfscope}%
\begin{pgfscope}%
\pgfsetbuttcap%
\pgfsetmiterjoin%
\definecolor{currentfill}{rgb}{1.000000,1.000000,1.000000}%
\pgfsetfillcolor{currentfill}%
\pgfsetlinewidth{0.000000pt}%
\definecolor{currentstroke}{rgb}{0.000000,0.000000,0.000000}%
\pgfsetstrokecolor{currentstroke}%
\pgfsetstrokeopacity{0.000000}%
\pgfsetdash{}{0pt}%
\pgfpathmoveto{\pgfqpoint{0.520798in}{0.442177in}}%
\pgfpathlineto{\pgfqpoint{1.239803in}{0.442177in}}%
\pgfpathlineto{\pgfqpoint{1.239803in}{1.314061in}}%
\pgfpathlineto{\pgfqpoint{0.520798in}{1.314061in}}%
\pgfpathlineto{\pgfqpoint{0.520798in}{0.442177in}}%
\pgfpathclose%
\pgfusepath{fill}%
\end{pgfscope}%
\begin{pgfscope}%
\pgfpathrectangle{\pgfqpoint{0.520798in}{0.442177in}}{\pgfqpoint{0.719005in}{0.871884in}}%
\pgfusepath{clip}%
\pgfsetroundcap%
\pgfsetroundjoin%
\pgfsetlinewidth{0.501875pt}%
\definecolor{currentstroke}{rgb}{0.800000,0.800000,0.800000}%
\pgfsetstrokecolor{currentstroke}%
\pgfsetdash{}{0pt}%
\pgfpathmoveto{\pgfqpoint{0.520798in}{0.442177in}}%
\pgfpathlineto{\pgfqpoint{0.520798in}{1.314061in}}%
\pgfusepath{stroke}%
\end{pgfscope}%
\begin{pgfscope}%
\definecolor{textcolor}{rgb}{0.150000,0.150000,0.150000}%
\pgfsetstrokecolor{textcolor}%
\pgfsetfillcolor{textcolor}%
\pgftext[x=0.520798in,y=0.351899in,,top]{\color{textcolor}\rmfamily\fontsize{8.000000}{9.600000}\selectfont \(\displaystyle {0}\)}%
\end{pgfscope}%
\begin{pgfscope}%
\pgfpathrectangle{\pgfqpoint{0.520798in}{0.442177in}}{\pgfqpoint{0.719005in}{0.871884in}}%
\pgfusepath{clip}%
\pgfsetroundcap%
\pgfsetroundjoin%
\pgfsetlinewidth{0.501875pt}%
\definecolor{currentstroke}{rgb}{0.800000,0.800000,0.800000}%
\pgfsetstrokecolor{currentstroke}%
\pgfsetdash{}{0pt}%
\pgfpathmoveto{\pgfqpoint{0.880300in}{0.442177in}}%
\pgfpathlineto{\pgfqpoint{0.880300in}{1.314061in}}%
\pgfusepath{stroke}%
\end{pgfscope}%
\begin{pgfscope}%
\definecolor{textcolor}{rgb}{0.150000,0.150000,0.150000}%
\pgfsetstrokecolor{textcolor}%
\pgfsetfillcolor{textcolor}%
\pgftext[x=0.880300in,y=0.351899in,,top]{\color{textcolor}\rmfamily\fontsize{8.000000}{9.600000}\selectfont \(\displaystyle {5}\)}%
\end{pgfscope}%
\begin{pgfscope}%
\pgfpathrectangle{\pgfqpoint{0.520798in}{0.442177in}}{\pgfqpoint{0.719005in}{0.871884in}}%
\pgfusepath{clip}%
\pgfsetroundcap%
\pgfsetroundjoin%
\pgfsetlinewidth{0.501875pt}%
\definecolor{currentstroke}{rgb}{0.800000,0.800000,0.800000}%
\pgfsetstrokecolor{currentstroke}%
\pgfsetdash{}{0pt}%
\pgfpathmoveto{\pgfqpoint{1.239803in}{0.442177in}}%
\pgfpathlineto{\pgfqpoint{1.239803in}{1.314061in}}%
\pgfusepath{stroke}%
\end{pgfscope}%
\begin{pgfscope}%
\definecolor{textcolor}{rgb}{0.150000,0.150000,0.150000}%
\pgfsetstrokecolor{textcolor}%
\pgfsetfillcolor{textcolor}%
\pgftext[x=1.239803in,y=0.351899in,,top]{\color{textcolor}\rmfamily\fontsize{8.000000}{9.600000}\selectfont \(\displaystyle {10}\)}%
\end{pgfscope}%
\begin{pgfscope}%
\definecolor{textcolor}{rgb}{0.150000,0.150000,0.150000}%
\pgfsetstrokecolor{textcolor}%
\pgfsetfillcolor{textcolor}%
\pgftext[x=0.880300in,y=0.198219in,,top]{\color{textcolor}\rmfamily\fontsize{10.000000}{12.000000}\selectfont Edit distance \(\displaystyle \epsilon\)}%
\end{pgfscope}%
\begin{pgfscope}%
\pgfpathrectangle{\pgfqpoint{0.520798in}{0.442177in}}{\pgfqpoint{0.719005in}{0.871884in}}%
\pgfusepath{clip}%
\pgfsetroundcap%
\pgfsetroundjoin%
\pgfsetlinewidth{0.501875pt}%
\definecolor{currentstroke}{rgb}{0.800000,0.800000,0.800000}%
\pgfsetstrokecolor{currentstroke}%
\pgfsetdash{}{0pt}%
\pgfpathmoveto{\pgfqpoint{0.520798in}{0.442177in}}%
\pgfpathlineto{\pgfqpoint{1.239803in}{0.442177in}}%
\pgfusepath{stroke}%
\end{pgfscope}%
\begin{pgfscope}%
\definecolor{textcolor}{rgb}{0.150000,0.150000,0.150000}%
\pgfsetstrokecolor{textcolor}%
\pgfsetfillcolor{textcolor}%
\pgftext[x=0.273151in, y=0.403915in, left, base]{\color{textcolor}\rmfamily\fontsize{8.000000}{9.600000}\selectfont 0\%}%
\end{pgfscope}%
\begin{pgfscope}%
\pgfpathrectangle{\pgfqpoint{0.520798in}{0.442177in}}{\pgfqpoint{0.719005in}{0.871884in}}%
\pgfusepath{clip}%
\pgfsetroundcap%
\pgfsetroundjoin%
\pgfsetlinewidth{0.501875pt}%
\definecolor{currentstroke}{rgb}{0.800000,0.800000,0.800000}%
\pgfsetstrokecolor{currentstroke}%
\pgfsetdash{}{0pt}%
\pgfpathmoveto{\pgfqpoint{0.520798in}{0.660148in}}%
\pgfpathlineto{\pgfqpoint{1.239803in}{0.660148in}}%
\pgfusepath{stroke}%
\end{pgfscope}%
\begin{pgfscope}%
\definecolor{textcolor}{rgb}{0.150000,0.150000,0.150000}%
\pgfsetstrokecolor{textcolor}%
\pgfsetfillcolor{textcolor}%
\pgftext[x=0.214138in, y=0.621886in, left, base]{\color{textcolor}\rmfamily\fontsize{8.000000}{9.600000}\selectfont 25\%}%
\end{pgfscope}%
\begin{pgfscope}%
\pgfpathrectangle{\pgfqpoint{0.520798in}{0.442177in}}{\pgfqpoint{0.719005in}{0.871884in}}%
\pgfusepath{clip}%
\pgfsetroundcap%
\pgfsetroundjoin%
\pgfsetlinewidth{0.501875pt}%
\definecolor{currentstroke}{rgb}{0.800000,0.800000,0.800000}%
\pgfsetstrokecolor{currentstroke}%
\pgfsetdash{}{0pt}%
\pgfpathmoveto{\pgfqpoint{0.520798in}{0.878119in}}%
\pgfpathlineto{\pgfqpoint{1.239803in}{0.878119in}}%
\pgfusepath{stroke}%
\end{pgfscope}%
\begin{pgfscope}%
\definecolor{textcolor}{rgb}{0.150000,0.150000,0.150000}%
\pgfsetstrokecolor{textcolor}%
\pgfsetfillcolor{textcolor}%
\pgftext[x=0.214138in, y=0.839857in, left, base]{\color{textcolor}\rmfamily\fontsize{8.000000}{9.600000}\selectfont 50\%}%
\end{pgfscope}%
\begin{pgfscope}%
\pgfpathrectangle{\pgfqpoint{0.520798in}{0.442177in}}{\pgfqpoint{0.719005in}{0.871884in}}%
\pgfusepath{clip}%
\pgfsetroundcap%
\pgfsetroundjoin%
\pgfsetlinewidth{0.501875pt}%
\definecolor{currentstroke}{rgb}{0.800000,0.800000,0.800000}%
\pgfsetstrokecolor{currentstroke}%
\pgfsetdash{}{0pt}%
\pgfpathmoveto{\pgfqpoint{0.520798in}{1.096090in}}%
\pgfpathlineto{\pgfqpoint{1.239803in}{1.096090in}}%
\pgfusepath{stroke}%
\end{pgfscope}%
\begin{pgfscope}%
\definecolor{textcolor}{rgb}{0.150000,0.150000,0.150000}%
\pgfsetstrokecolor{textcolor}%
\pgfsetfillcolor{textcolor}%
\pgftext[x=0.214138in, y=1.057828in, left, base]{\color{textcolor}\rmfamily\fontsize{8.000000}{9.600000}\selectfont 75\%}%
\end{pgfscope}%
\begin{pgfscope}%
\pgfpathrectangle{\pgfqpoint{0.520798in}{0.442177in}}{\pgfqpoint{0.719005in}{0.871884in}}%
\pgfusepath{clip}%
\pgfsetroundcap%
\pgfsetroundjoin%
\pgfsetlinewidth{0.501875pt}%
\definecolor{currentstroke}{rgb}{0.800000,0.800000,0.800000}%
\pgfsetstrokecolor{currentstroke}%
\pgfsetdash{}{0pt}%
\pgfpathmoveto{\pgfqpoint{0.520798in}{1.314061in}}%
\pgfpathlineto{\pgfqpoint{1.239803in}{1.314061in}}%
\pgfusepath{stroke}%
\end{pgfscope}%
\begin{pgfscope}%
\definecolor{textcolor}{rgb}{0.150000,0.150000,0.150000}%
\pgfsetstrokecolor{textcolor}%
\pgfsetfillcolor{textcolor}%
\pgftext[x=0.155124in, y=1.275799in, left, base]{\color{textcolor}\rmfamily\fontsize{8.000000}{9.600000}\selectfont 100\%}%
\end{pgfscope}%
\begin{pgfscope}%
\definecolor{textcolor}{rgb}{0.150000,0.150000,0.150000}%
\pgfsetstrokecolor{textcolor}%
\pgfsetfillcolor{textcolor}%
\pgftext[x=0.099569in,y=0.878119in,,bottom,rotate=90.000000]{\color{textcolor}\rmfamily\fontsize{10.000000}{12.000000}\selectfont Cert. Acc.}%
\end{pgfscope}%
\begin{pgfscope}%
\pgfsetrectcap%
\pgfsetmiterjoin%
\pgfsetlinewidth{0.752812pt}%
\definecolor{currentstroke}{rgb}{0.700000,0.700000,0.700000}%
\pgfsetstrokecolor{currentstroke}%
\pgfsetdash{}{0pt}%
\pgfpathmoveto{\pgfqpoint{0.520798in}{0.442177in}}%
\pgfpathlineto{\pgfqpoint{0.520798in}{1.314061in}}%
\pgfusepath{stroke}%
\end{pgfscope}%
\begin{pgfscope}%
\pgfsetrectcap%
\pgfsetmiterjoin%
\pgfsetlinewidth{0.752812pt}%
\definecolor{currentstroke}{rgb}{0.700000,0.700000,0.700000}%
\pgfsetstrokecolor{currentstroke}%
\pgfsetdash{}{0pt}%
\pgfpathmoveto{\pgfqpoint{1.239803in}{0.442177in}}%
\pgfpathlineto{\pgfqpoint{1.239803in}{1.314061in}}%
\pgfusepath{stroke}%
\end{pgfscope}%
\begin{pgfscope}%
\pgfsetrectcap%
\pgfsetmiterjoin%
\pgfsetlinewidth{0.752812pt}%
\definecolor{currentstroke}{rgb}{0.700000,0.700000,0.700000}%
\pgfsetstrokecolor{currentstroke}%
\pgfsetdash{}{0pt}%
\pgfpathmoveto{\pgfqpoint{0.520798in}{0.442177in}}%
\pgfpathlineto{\pgfqpoint{1.239803in}{0.442177in}}%
\pgfusepath{stroke}%
\end{pgfscope}%
\begin{pgfscope}%
\pgfsetrectcap%
\pgfsetmiterjoin%
\pgfsetlinewidth{0.752812pt}%
\definecolor{currentstroke}{rgb}{0.700000,0.700000,0.700000}%
\pgfsetstrokecolor{currentstroke}%
\pgfsetdash{}{0pt}%
\pgfpathmoveto{\pgfqpoint{0.520798in}{1.314061in}}%
\pgfpathlineto{\pgfqpoint{1.239803in}{1.314061in}}%
\pgfusepath{stroke}%
\end{pgfscope}%
\begin{pgfscope}%
\pgfpathrectangle{\pgfqpoint{0.520798in}{0.442177in}}{\pgfqpoint{0.719005in}{0.871884in}}%
\pgfusepath{clip}%
\pgfsetroundcap%
\pgfsetroundjoin%
\pgfsetlinewidth{1.003750pt}%
\definecolor{currentstroke}{rgb}{0.003922,0.450980,0.698039}%
\pgfsetstrokecolor{currentstroke}%
\pgfsetdash{}{0pt}%
\pgfpathmoveto{\pgfqpoint{0.520798in}{1.029997in}}%
\pgfpathlineto{\pgfqpoint{0.556748in}{1.029997in}}%
\pgfpathlineto{\pgfqpoint{0.556748in}{0.869301in}}%
\pgfpathlineto{\pgfqpoint{0.628649in}{0.869301in}}%
\pgfpathlineto{\pgfqpoint{0.628649in}{0.664715in}}%
\pgfpathlineto{\pgfqpoint{0.700549in}{0.664715in}}%
\pgfpathlineto{\pgfqpoint{0.700549in}{0.456491in}}%
\pgfpathlineto{\pgfqpoint{0.772450in}{0.456491in}}%
\pgfpathlineto{\pgfqpoint{0.772450in}{0.445103in}}%
\pgfpathlineto{\pgfqpoint{0.844350in}{0.445103in}}%
\pgfpathlineto{\pgfqpoint{0.844350in}{0.443917in}}%
\pgfpathlineto{\pgfqpoint{0.916251in}{0.443917in}}%
\pgfpathlineto{\pgfqpoint{0.916251in}{0.443363in}}%
\pgfpathlineto{\pgfqpoint{0.988151in}{0.443363in}}%
\pgfpathlineto{\pgfqpoint{0.988151in}{0.443205in}}%
\pgfpathlineto{\pgfqpoint{1.060052in}{0.443205in}}%
\pgfpathlineto{\pgfqpoint{1.060052in}{0.442572in}}%
\pgfpathlineto{\pgfqpoint{1.131952in}{0.442572in}}%
\pgfpathlineto{\pgfqpoint{1.131952in}{0.442493in}}%
\pgfpathlineto{\pgfqpoint{1.241470in}{0.442493in}}%
\pgfusepath{stroke}%
\end{pgfscope}%
\begin{pgfscope}%
\pgfpathrectangle{\pgfqpoint{0.520798in}{0.442177in}}{\pgfqpoint{0.719005in}{0.871884in}}%
\pgfusepath{clip}%
\pgfsetbuttcap%
\pgfsetroundjoin%
\definecolor{currentfill}{rgb}{0.003922,0.450980,0.698039}%
\pgfsetfillcolor{currentfill}%
\pgfsetfillopacity{0.500000}%
\pgfsetlinewidth{0.000000pt}%
\definecolor{currentstroke}{rgb}{0.003922,0.450980,0.698039}%
\pgfsetstrokecolor{currentstroke}%
\pgfsetstrokeopacity{0.500000}%
\pgfsetdash{}{0pt}%
\pgfpathmoveto{\pgfqpoint{0.520798in}{1.048669in}}%
\pgfpathlineto{\pgfqpoint{0.520798in}{1.011325in}}%
\pgfpathlineto{\pgfqpoint{0.556748in}{1.011325in}}%
\pgfpathlineto{\pgfqpoint{0.556748in}{0.850500in}}%
\pgfpathlineto{\pgfqpoint{0.628649in}{0.850500in}}%
\pgfpathlineto{\pgfqpoint{0.628649in}{0.649529in}}%
\pgfpathlineto{\pgfqpoint{0.700549in}{0.649529in}}%
\pgfpathlineto{\pgfqpoint{0.700549in}{0.452881in}}%
\pgfpathlineto{\pgfqpoint{0.772450in}{0.452881in}}%
\pgfpathlineto{\pgfqpoint{0.772450in}{0.443914in}}%
\pgfpathlineto{\pgfqpoint{0.844350in}{0.443914in}}%
\pgfpathlineto{\pgfqpoint{0.844350in}{0.443274in}}%
\pgfpathlineto{\pgfqpoint{0.916251in}{0.443274in}}%
\pgfpathlineto{\pgfqpoint{0.916251in}{0.443009in}}%
\pgfpathlineto{\pgfqpoint{0.988151in}{0.443009in}}%
\pgfpathlineto{\pgfqpoint{0.988151in}{0.442802in}}%
\pgfpathlineto{\pgfqpoint{1.060052in}{0.442802in}}%
\pgfpathlineto{\pgfqpoint{1.060052in}{0.442572in}}%
\pgfpathlineto{\pgfqpoint{1.131952in}{0.442572in}}%
\pgfpathlineto{\pgfqpoint{1.131952in}{0.442335in}}%
\pgfpathlineto{\pgfqpoint{1.311703in}{0.442335in}}%
\pgfpathlineto{\pgfqpoint{1.311703in}{0.442177in}}%
\pgfpathlineto{\pgfqpoint{2.066659in}{0.442177in}}%
\pgfpathlineto{\pgfqpoint{2.066659in}{0.442177in}}%
\pgfpathlineto{\pgfqpoint{2.677813in}{0.442177in}}%
\pgfpathlineto{\pgfqpoint{2.677813in}{0.442177in}}%
\pgfpathlineto{\pgfqpoint{2.677813in}{0.442177in}}%
\pgfpathlineto{\pgfqpoint{2.066659in}{0.442177in}}%
\pgfpathlineto{\pgfqpoint{2.066659in}{0.442177in}}%
\pgfpathlineto{\pgfqpoint{1.311703in}{0.442177in}}%
\pgfpathlineto{\pgfqpoint{1.311703in}{0.442651in}}%
\pgfpathlineto{\pgfqpoint{1.131952in}{0.442651in}}%
\pgfpathlineto{\pgfqpoint{1.131952in}{0.442572in}}%
\pgfpathlineto{\pgfqpoint{1.060052in}{0.442572in}}%
\pgfpathlineto{\pgfqpoint{1.060052in}{0.443608in}}%
\pgfpathlineto{\pgfqpoint{0.988151in}{0.443608in}}%
\pgfpathlineto{\pgfqpoint{0.988151in}{0.443717in}}%
\pgfpathlineto{\pgfqpoint{0.916251in}{0.443717in}}%
\pgfpathlineto{\pgfqpoint{0.916251in}{0.444559in}}%
\pgfpathlineto{\pgfqpoint{0.844350in}{0.444559in}}%
\pgfpathlineto{\pgfqpoint{0.844350in}{0.446292in}}%
\pgfpathlineto{\pgfqpoint{0.772450in}{0.446292in}}%
\pgfpathlineto{\pgfqpoint{0.772450in}{0.460101in}}%
\pgfpathlineto{\pgfqpoint{0.700549in}{0.460101in}}%
\pgfpathlineto{\pgfqpoint{0.700549in}{0.679901in}}%
\pgfpathlineto{\pgfqpoint{0.628649in}{0.679901in}}%
\pgfpathlineto{\pgfqpoint{0.628649in}{0.888103in}}%
\pgfpathlineto{\pgfqpoint{0.556748in}{0.888103in}}%
\pgfpathlineto{\pgfqpoint{0.556748in}{1.048669in}}%
\pgfpathlineto{\pgfqpoint{0.520798in}{1.048669in}}%
\pgfpathlineto{\pgfqpoint{0.520798in}{1.048669in}}%
\pgfpathclose%
\pgfusepath{fill}%
\end{pgfscope}%
\begin{pgfscope}%
\pgfpathrectangle{\pgfqpoint{0.520798in}{0.442177in}}{\pgfqpoint{0.719005in}{0.871884in}}%
\pgfusepath{clip}%
\pgfsetroundcap%
\pgfsetroundjoin%
\pgfsetlinewidth{1.003750pt}%
\definecolor{currentstroke}{rgb}{0.870588,0.560784,0.019608}%
\pgfsetstrokecolor{currentstroke}%
\pgfsetdash{}{0pt}%
\pgfpathmoveto{\pgfqpoint{0.520798in}{1.029997in}}%
\pgfpathlineto{\pgfqpoint{0.556748in}{1.029997in}}%
\pgfpathlineto{\pgfqpoint{0.556748in}{0.900618in}}%
\pgfpathlineto{\pgfqpoint{0.628649in}{0.900618in}}%
\pgfpathlineto{\pgfqpoint{0.628649in}{0.798602in}}%
\pgfpathlineto{\pgfqpoint{0.700549in}{0.798602in}}%
\pgfpathlineto{\pgfqpoint{0.700549in}{0.712797in}}%
\pgfpathlineto{\pgfqpoint{0.772450in}{0.712797in}}%
\pgfpathlineto{\pgfqpoint{0.772450in}{0.647000in}}%
\pgfpathlineto{\pgfqpoint{0.844350in}{0.647000in}}%
\pgfpathlineto{\pgfqpoint{0.844350in}{0.598681in}}%
\pgfpathlineto{\pgfqpoint{0.916251in}{0.598681in}}%
\pgfpathlineto{\pgfqpoint{0.916251in}{0.456491in}}%
\pgfpathlineto{\pgfqpoint{0.988151in}{0.456491in}}%
\pgfpathlineto{\pgfqpoint{0.988151in}{0.453881in}}%
\pgfpathlineto{\pgfqpoint{1.060052in}{0.453881in}}%
\pgfpathlineto{\pgfqpoint{1.060052in}{0.445103in}}%
\pgfpathlineto{\pgfqpoint{1.131952in}{0.445103in}}%
\pgfpathlineto{\pgfqpoint{1.131952in}{0.444707in}}%
\pgfpathlineto{\pgfqpoint{1.203853in}{0.444707in}}%
\pgfpathlineto{\pgfqpoint{1.203853in}{0.443679in}}%
\pgfpathlineto{\pgfqpoint{1.241470in}{0.443679in}}%
\pgfusepath{stroke}%
\end{pgfscope}%
\begin{pgfscope}%
\pgfpathrectangle{\pgfqpoint{0.520798in}{0.442177in}}{\pgfqpoint{0.719005in}{0.871884in}}%
\pgfusepath{clip}%
\pgfsetbuttcap%
\pgfsetroundjoin%
\definecolor{currentfill}{rgb}{0.870588,0.560784,0.019608}%
\pgfsetfillcolor{currentfill}%
\pgfsetfillopacity{0.500000}%
\pgfsetlinewidth{0.000000pt}%
\definecolor{currentstroke}{rgb}{0.870588,0.560784,0.019608}%
\pgfsetstrokecolor{currentstroke}%
\pgfsetstrokeopacity{0.500000}%
\pgfsetdash{}{0pt}%
\pgfpathmoveto{\pgfqpoint{0.520798in}{1.048669in}}%
\pgfpathlineto{\pgfqpoint{0.520798in}{1.011325in}}%
\pgfpathlineto{\pgfqpoint{0.556748in}{1.011325in}}%
\pgfpathlineto{\pgfqpoint{0.556748in}{0.880028in}}%
\pgfpathlineto{\pgfqpoint{0.628649in}{0.880028in}}%
\pgfpathlineto{\pgfqpoint{0.628649in}{0.779727in}}%
\pgfpathlineto{\pgfqpoint{0.700549in}{0.779727in}}%
\pgfpathlineto{\pgfqpoint{0.700549in}{0.697061in}}%
\pgfpathlineto{\pgfqpoint{0.772450in}{0.697061in}}%
\pgfpathlineto{\pgfqpoint{0.772450in}{0.631489in}}%
\pgfpathlineto{\pgfqpoint{0.844350in}{0.631489in}}%
\pgfpathlineto{\pgfqpoint{0.844350in}{0.585262in}}%
\pgfpathlineto{\pgfqpoint{0.916251in}{0.585262in}}%
\pgfpathlineto{\pgfqpoint{0.916251in}{0.452881in}}%
\pgfpathlineto{\pgfqpoint{0.988151in}{0.452881in}}%
\pgfpathlineto{\pgfqpoint{0.988151in}{0.450539in}}%
\pgfpathlineto{\pgfqpoint{1.060052in}{0.450539in}}%
\pgfpathlineto{\pgfqpoint{1.060052in}{0.443914in}}%
\pgfpathlineto{\pgfqpoint{1.131952in}{0.443914in}}%
\pgfpathlineto{\pgfqpoint{1.131952in}{0.443974in}}%
\pgfpathlineto{\pgfqpoint{1.203853in}{0.443974in}}%
\pgfpathlineto{\pgfqpoint{1.203853in}{0.443292in}}%
\pgfpathlineto{\pgfqpoint{1.275753in}{0.443292in}}%
\pgfpathlineto{\pgfqpoint{1.275753in}{0.442802in}}%
\pgfpathlineto{\pgfqpoint{1.383604in}{0.442802in}}%
\pgfpathlineto{\pgfqpoint{1.383604in}{0.442659in}}%
\pgfpathlineto{\pgfqpoint{1.491455in}{0.442659in}}%
\pgfpathlineto{\pgfqpoint{1.491455in}{0.442593in}}%
\pgfpathlineto{\pgfqpoint{1.563355in}{0.442593in}}%
\pgfpathlineto{\pgfqpoint{1.563355in}{0.442493in}}%
\pgfpathlineto{\pgfqpoint{1.635256in}{0.442493in}}%
\pgfpathlineto{\pgfqpoint{1.635256in}{0.442572in}}%
\pgfpathlineto{\pgfqpoint{1.743107in}{0.442572in}}%
\pgfpathlineto{\pgfqpoint{1.743107in}{0.442335in}}%
\pgfpathlineto{\pgfqpoint{2.066659in}{0.442335in}}%
\pgfpathlineto{\pgfqpoint{2.066659in}{0.442177in}}%
\pgfpathlineto{\pgfqpoint{2.498062in}{0.442177in}}%
\pgfpathlineto{\pgfqpoint{2.498062in}{0.442177in}}%
\pgfpathlineto{\pgfqpoint{2.677813in}{0.442177in}}%
\pgfpathlineto{\pgfqpoint{2.677813in}{0.442177in}}%
\pgfpathlineto{\pgfqpoint{2.677813in}{0.442177in}}%
\pgfpathlineto{\pgfqpoint{2.498062in}{0.442177in}}%
\pgfpathlineto{\pgfqpoint{2.498062in}{0.442177in}}%
\pgfpathlineto{\pgfqpoint{2.066659in}{0.442177in}}%
\pgfpathlineto{\pgfqpoint{2.066659in}{0.442651in}}%
\pgfpathlineto{\pgfqpoint{1.743107in}{0.442651in}}%
\pgfpathlineto{\pgfqpoint{1.743107in}{0.442572in}}%
\pgfpathlineto{\pgfqpoint{1.635256in}{0.442572in}}%
\pgfpathlineto{\pgfqpoint{1.635256in}{0.443126in}}%
\pgfpathlineto{\pgfqpoint{1.563355in}{0.443126in}}%
\pgfpathlineto{\pgfqpoint{1.563355in}{0.443184in}}%
\pgfpathlineto{\pgfqpoint{1.491455in}{0.443184in}}%
\pgfpathlineto{\pgfqpoint{1.491455in}{0.443434in}}%
\pgfpathlineto{\pgfqpoint{1.383604in}{0.443434in}}%
\pgfpathlineto{\pgfqpoint{1.383604in}{0.443608in}}%
\pgfpathlineto{\pgfqpoint{1.275753in}{0.443608in}}%
\pgfpathlineto{\pgfqpoint{1.275753in}{0.444067in}}%
\pgfpathlineto{\pgfqpoint{1.203853in}{0.444067in}}%
\pgfpathlineto{\pgfqpoint{1.203853in}{0.445441in}}%
\pgfpathlineto{\pgfqpoint{1.131952in}{0.445441in}}%
\pgfpathlineto{\pgfqpoint{1.131952in}{0.446292in}}%
\pgfpathlineto{\pgfqpoint{1.060052in}{0.446292in}}%
\pgfpathlineto{\pgfqpoint{1.060052in}{0.457223in}}%
\pgfpathlineto{\pgfqpoint{0.988151in}{0.457223in}}%
\pgfpathlineto{\pgfqpoint{0.988151in}{0.460101in}}%
\pgfpathlineto{\pgfqpoint{0.916251in}{0.460101in}}%
\pgfpathlineto{\pgfqpoint{0.916251in}{0.612100in}}%
\pgfpathlineto{\pgfqpoint{0.844350in}{0.612100in}}%
\pgfpathlineto{\pgfqpoint{0.844350in}{0.662512in}}%
\pgfpathlineto{\pgfqpoint{0.772450in}{0.662512in}}%
\pgfpathlineto{\pgfqpoint{0.772450in}{0.728534in}}%
\pgfpathlineto{\pgfqpoint{0.700549in}{0.728534in}}%
\pgfpathlineto{\pgfqpoint{0.700549in}{0.817477in}}%
\pgfpathlineto{\pgfqpoint{0.628649in}{0.817477in}}%
\pgfpathlineto{\pgfqpoint{0.628649in}{0.921208in}}%
\pgfpathlineto{\pgfqpoint{0.556748in}{0.921208in}}%
\pgfpathlineto{\pgfqpoint{0.556748in}{1.048669in}}%
\pgfpathlineto{\pgfqpoint{0.520798in}{1.048669in}}%
\pgfpathlineto{\pgfqpoint{0.520798in}{1.048669in}}%
\pgfpathclose%
\pgfusepath{fill}%
\end{pgfscope}%
\begin{pgfscope}%
\pgfpathrectangle{\pgfqpoint{0.520798in}{0.442177in}}{\pgfqpoint{0.719005in}{0.871884in}}%
\pgfusepath{clip}%
\pgfsetroundcap%
\pgfsetroundjoin%
\pgfsetlinewidth{1.003750pt}%
\definecolor{currentstroke}{rgb}{0.007843,0.619608,0.450980}%
\pgfsetstrokecolor{currentstroke}%
\pgfsetdash{}{0pt}%
\pgfpathmoveto{\pgfqpoint{0.520798in}{1.029997in}}%
\pgfpathlineto{\pgfqpoint{0.556748in}{1.029997in}}%
\pgfpathlineto{\pgfqpoint{0.556748in}{0.900618in}}%
\pgfpathlineto{\pgfqpoint{0.628649in}{0.900618in}}%
\pgfpathlineto{\pgfqpoint{0.628649in}{0.798602in}}%
\pgfpathlineto{\pgfqpoint{0.700549in}{0.798602in}}%
\pgfpathlineto{\pgfqpoint{0.700549in}{0.712797in}}%
\pgfpathlineto{\pgfqpoint{0.772450in}{0.712797in}}%
\pgfpathlineto{\pgfqpoint{0.772450in}{0.647000in}}%
\pgfpathlineto{\pgfqpoint{0.844350in}{0.647000in}}%
\pgfpathlineto{\pgfqpoint{0.844350in}{0.598681in}}%
\pgfpathlineto{\pgfqpoint{0.916251in}{0.598681in}}%
\pgfpathlineto{\pgfqpoint{0.916251in}{0.562224in}}%
\pgfpathlineto{\pgfqpoint{0.988151in}{0.562224in}}%
\pgfpathlineto{\pgfqpoint{0.988151in}{0.531777in}}%
\pgfpathlineto{\pgfqpoint{1.060052in}{0.531777in}}%
\pgfpathlineto{\pgfqpoint{1.060052in}{0.507183in}}%
\pgfpathlineto{\pgfqpoint{1.131952in}{0.507183in}}%
\pgfpathlineto{\pgfqpoint{1.131952in}{0.489626in}}%
\pgfpathlineto{\pgfqpoint{1.203853in}{0.489626in}}%
\pgfpathlineto{\pgfqpoint{1.203853in}{0.477210in}}%
\pgfpathlineto{\pgfqpoint{1.241470in}{0.477210in}}%
\pgfusepath{stroke}%
\end{pgfscope}%
\begin{pgfscope}%
\pgfpathrectangle{\pgfqpoint{0.520798in}{0.442177in}}{\pgfqpoint{0.719005in}{0.871884in}}%
\pgfusepath{clip}%
\pgfsetbuttcap%
\pgfsetroundjoin%
\definecolor{currentfill}{rgb}{0.007843,0.619608,0.450980}%
\pgfsetfillcolor{currentfill}%
\pgfsetfillopacity{0.500000}%
\pgfsetlinewidth{0.000000pt}%
\definecolor{currentstroke}{rgb}{0.007843,0.619608,0.450980}%
\pgfsetstrokecolor{currentstroke}%
\pgfsetstrokeopacity{0.500000}%
\pgfsetdash{}{0pt}%
\pgfpathmoveto{\pgfqpoint{0.520798in}{1.048669in}}%
\pgfpathlineto{\pgfqpoint{0.520798in}{1.011325in}}%
\pgfpathlineto{\pgfqpoint{0.556748in}{1.011325in}}%
\pgfpathlineto{\pgfqpoint{0.556748in}{0.880028in}}%
\pgfpathlineto{\pgfqpoint{0.628649in}{0.880028in}}%
\pgfpathlineto{\pgfqpoint{0.628649in}{0.779727in}}%
\pgfpathlineto{\pgfqpoint{0.700549in}{0.779727in}}%
\pgfpathlineto{\pgfqpoint{0.700549in}{0.697061in}}%
\pgfpathlineto{\pgfqpoint{0.772450in}{0.697061in}}%
\pgfpathlineto{\pgfqpoint{0.772450in}{0.631489in}}%
\pgfpathlineto{\pgfqpoint{0.844350in}{0.631489in}}%
\pgfpathlineto{\pgfqpoint{0.844350in}{0.585262in}}%
\pgfpathlineto{\pgfqpoint{0.916251in}{0.585262in}}%
\pgfpathlineto{\pgfqpoint{0.916251in}{0.551270in}}%
\pgfpathlineto{\pgfqpoint{0.988151in}{0.551270in}}%
\pgfpathlineto{\pgfqpoint{0.988151in}{0.522976in}}%
\pgfpathlineto{\pgfqpoint{1.060052in}{0.522976in}}%
\pgfpathlineto{\pgfqpoint{1.060052in}{0.498909in}}%
\pgfpathlineto{\pgfqpoint{1.131952in}{0.498909in}}%
\pgfpathlineto{\pgfqpoint{1.131952in}{0.482383in}}%
\pgfpathlineto{\pgfqpoint{1.203853in}{0.482383in}}%
\pgfpathlineto{\pgfqpoint{1.203853in}{0.471294in}}%
\pgfpathlineto{\pgfqpoint{1.275753in}{0.471294in}}%
\pgfpathlineto{\pgfqpoint{1.275753in}{0.464323in}}%
\pgfpathlineto{\pgfqpoint{1.347654in}{0.464323in}}%
\pgfpathlineto{\pgfqpoint{1.347654in}{0.452881in}}%
\pgfpathlineto{\pgfqpoint{1.419554in}{0.452881in}}%
\pgfpathlineto{\pgfqpoint{1.419554in}{0.450539in}}%
\pgfpathlineto{\pgfqpoint{1.491455in}{0.450539in}}%
\pgfpathlineto{\pgfqpoint{1.491455in}{0.449543in}}%
\pgfpathlineto{\pgfqpoint{1.563355in}{0.449543in}}%
\pgfpathlineto{\pgfqpoint{1.563355in}{0.447526in}}%
\pgfpathlineto{\pgfqpoint{1.635256in}{0.447526in}}%
\pgfpathlineto{\pgfqpoint{1.635256in}{0.443914in}}%
\pgfpathlineto{\pgfqpoint{1.707156in}{0.443914in}}%
\pgfpathlineto{\pgfqpoint{1.707156in}{0.443974in}}%
\pgfpathlineto{\pgfqpoint{1.779057in}{0.443974in}}%
\pgfpathlineto{\pgfqpoint{1.779057in}{0.443687in}}%
\pgfpathlineto{\pgfqpoint{1.850957in}{0.443687in}}%
\pgfpathlineto{\pgfqpoint{1.850957in}{0.443146in}}%
\pgfpathlineto{\pgfqpoint{1.922858in}{0.443146in}}%
\pgfpathlineto{\pgfqpoint{1.922858in}{0.442802in}}%
\pgfpathlineto{\pgfqpoint{1.994758in}{0.442802in}}%
\pgfpathlineto{\pgfqpoint{1.994758in}{0.442659in}}%
\pgfpathlineto{\pgfqpoint{2.066659in}{0.442659in}}%
\pgfpathlineto{\pgfqpoint{2.066659in}{0.442593in}}%
\pgfpathlineto{\pgfqpoint{2.282360in}{0.442593in}}%
\pgfpathlineto{\pgfqpoint{2.282360in}{0.442537in}}%
\pgfpathlineto{\pgfqpoint{2.498062in}{0.442537in}}%
\pgfpathlineto{\pgfqpoint{2.498062in}{0.442493in}}%
\pgfpathlineto{\pgfqpoint{2.605913in}{0.442493in}}%
\pgfpathlineto{\pgfqpoint{2.605913in}{0.442493in}}%
\pgfpathlineto{\pgfqpoint{2.677813in}{0.442493in}}%
\pgfpathlineto{\pgfqpoint{2.677813in}{0.442809in}}%
\pgfpathlineto{\pgfqpoint{2.677813in}{0.442809in}}%
\pgfpathlineto{\pgfqpoint{2.605913in}{0.442809in}}%
\pgfpathlineto{\pgfqpoint{2.605913in}{0.442809in}}%
\pgfpathlineto{\pgfqpoint{2.498062in}{0.442809in}}%
\pgfpathlineto{\pgfqpoint{2.498062in}{0.442924in}}%
\pgfpathlineto{\pgfqpoint{2.282360in}{0.442924in}}%
\pgfpathlineto{\pgfqpoint{2.282360in}{0.443184in}}%
\pgfpathlineto{\pgfqpoint{2.066659in}{0.443184in}}%
\pgfpathlineto{\pgfqpoint{2.066659in}{0.443434in}}%
\pgfpathlineto{\pgfqpoint{1.994758in}{0.443434in}}%
\pgfpathlineto{\pgfqpoint{1.994758in}{0.443608in}}%
\pgfpathlineto{\pgfqpoint{1.922858in}{0.443608in}}%
\pgfpathlineto{\pgfqpoint{1.922858in}{0.443738in}}%
\pgfpathlineto{\pgfqpoint{1.850957in}{0.443738in}}%
\pgfpathlineto{\pgfqpoint{1.850957in}{0.444462in}}%
\pgfpathlineto{\pgfqpoint{1.779057in}{0.444462in}}%
\pgfpathlineto{\pgfqpoint{1.779057in}{0.445441in}}%
\pgfpathlineto{\pgfqpoint{1.707156in}{0.445441in}}%
\pgfpathlineto{\pgfqpoint{1.707156in}{0.446292in}}%
\pgfpathlineto{\pgfqpoint{1.635256in}{0.446292in}}%
\pgfpathlineto{\pgfqpoint{1.635256in}{0.452011in}}%
\pgfpathlineto{\pgfqpoint{1.563355in}{0.452011in}}%
\pgfpathlineto{\pgfqpoint{1.563355in}{0.454897in}}%
\pgfpathlineto{\pgfqpoint{1.491455in}{0.454897in}}%
\pgfpathlineto{\pgfqpoint{1.491455in}{0.457223in}}%
\pgfpathlineto{\pgfqpoint{1.419554in}{0.457223in}}%
\pgfpathlineto{\pgfqpoint{1.419554in}{0.460101in}}%
\pgfpathlineto{\pgfqpoint{1.347654in}{0.460101in}}%
\pgfpathlineto{\pgfqpoint{1.347654in}{0.474598in}}%
\pgfpathlineto{\pgfqpoint{1.275753in}{0.474598in}}%
\pgfpathlineto{\pgfqpoint{1.275753in}{0.483126in}}%
\pgfpathlineto{\pgfqpoint{1.203853in}{0.483126in}}%
\pgfpathlineto{\pgfqpoint{1.203853in}{0.496870in}}%
\pgfpathlineto{\pgfqpoint{1.131952in}{0.496870in}}%
\pgfpathlineto{\pgfqpoint{1.131952in}{0.515456in}}%
\pgfpathlineto{\pgfqpoint{1.060052in}{0.515456in}}%
\pgfpathlineto{\pgfqpoint{1.060052in}{0.540579in}}%
\pgfpathlineto{\pgfqpoint{0.988151in}{0.540579in}}%
\pgfpathlineto{\pgfqpoint{0.988151in}{0.573178in}}%
\pgfpathlineto{\pgfqpoint{0.916251in}{0.573178in}}%
\pgfpathlineto{\pgfqpoint{0.916251in}{0.612100in}}%
\pgfpathlineto{\pgfqpoint{0.844350in}{0.612100in}}%
\pgfpathlineto{\pgfqpoint{0.844350in}{0.662512in}}%
\pgfpathlineto{\pgfqpoint{0.772450in}{0.662512in}}%
\pgfpathlineto{\pgfqpoint{0.772450in}{0.728534in}}%
\pgfpathlineto{\pgfqpoint{0.700549in}{0.728534in}}%
\pgfpathlineto{\pgfqpoint{0.700549in}{0.817477in}}%
\pgfpathlineto{\pgfqpoint{0.628649in}{0.817477in}}%
\pgfpathlineto{\pgfqpoint{0.628649in}{0.921208in}}%
\pgfpathlineto{\pgfqpoint{0.556748in}{0.921208in}}%
\pgfpathlineto{\pgfqpoint{0.556748in}{1.048669in}}%
\pgfpathlineto{\pgfqpoint{0.520798in}{1.048669in}}%
\pgfpathlineto{\pgfqpoint{0.520798in}{1.048669in}}%
\pgfpathclose%
\pgfusepath{fill}%
\end{pgfscope}%
\begin{pgfscope}%
\definecolor{textcolor}{rgb}{0.150000,0.150000,0.150000}%
\pgfsetstrokecolor{textcolor}%
\pgfsetfillcolor{textcolor}%
\pgftext[x=0.880300in,y=1.397395in,,base]{\color{textcolor}\rmfamily\fontsize{9.000000}{10.800000}\selectfont \(\displaystyle c_A^-=1\)}%
\end{pgfscope}%
\begin{pgfscope}%
\pgfsetbuttcap%
\pgfsetmiterjoin%
\definecolor{currentfill}{rgb}{1.000000,1.000000,1.000000}%
\pgfsetfillcolor{currentfill}%
\pgfsetfillopacity{0.800000}%
\pgfsetlinewidth{1.003750pt}%
\definecolor{currentstroke}{rgb}{0.800000,0.800000,0.800000}%
\pgfsetstrokecolor{currentstroke}%
\pgfsetstrokeopacity{0.800000}%
\pgfsetdash{}{0pt}%
\pgfpathmoveto{\pgfqpoint{0.785843in}{0.638103in}}%
\pgfpathlineto{\pgfqpoint{1.191192in}{0.638103in}}%
\pgfpathlineto{\pgfqpoint{1.191192in}{1.265450in}}%
\pgfpathlineto{\pgfqpoint{0.785843in}{1.265450in}}%
\pgfpathlineto{\pgfqpoint{0.785843in}{0.638103in}}%
\pgfpathclose%
\pgfusepath{stroke,fill}%
\end{pgfscope}%
\begin{pgfscope}%
\definecolor{textcolor}{rgb}{0.150000,0.150000,0.150000}%
\pgfsetstrokecolor{textcolor}%
\pgfsetfillcolor{textcolor}%
\pgftext[x=0.918271in,y=1.113275in,left,base]{\color{textcolor}\rmfamily\fontsize{9.000000}{10.800000}\selectfont \(\displaystyle c_A^+\)}%
\end{pgfscope}%
\begin{pgfscope}%
\pgfsetroundcap%
\pgfsetroundjoin%
\pgfsetlinewidth{1.003750pt}%
\definecolor{currentstroke}{rgb}{0.003922,0.450980,0.698039}%
\pgfsetstrokecolor{currentstroke}%
\pgfsetdash{}{0pt}%
\pgfpathmoveto{\pgfqpoint{0.824732in}{1.001052in}}%
\pgfpathlineto{\pgfqpoint{0.873343in}{1.001052in}}%
\pgfpathlineto{\pgfqpoint{0.873343in}{1.001052in}}%
\pgfpathlineto{\pgfqpoint{0.970565in}{1.001052in}}%
\pgfpathlineto{\pgfqpoint{0.970565in}{1.001052in}}%
\pgfpathlineto{\pgfqpoint{1.019176in}{1.001052in}}%
\pgfusepath{stroke}%
\end{pgfscope}%
\begin{pgfscope}%
\definecolor{textcolor}{rgb}{0.150000,0.150000,0.150000}%
\pgfsetstrokecolor{textcolor}%
\pgfsetfillcolor{textcolor}%
\pgftext[x=1.096954in,y=0.967024in,left,base]{\color{textcolor}\rmfamily\fontsize{7.000000}{8.400000}\selectfont 1}%
\end{pgfscope}%
\begin{pgfscope}%
\pgfsetroundcap%
\pgfsetroundjoin%
\pgfsetlinewidth{1.003750pt}%
\definecolor{currentstroke}{rgb}{0.870588,0.560784,0.019608}%
\pgfsetstrokecolor{currentstroke}%
\pgfsetdash{}{0pt}%
\pgfpathmoveto{\pgfqpoint{0.824732in}{0.865486in}}%
\pgfpathlineto{\pgfqpoint{0.873343in}{0.865486in}}%
\pgfpathlineto{\pgfqpoint{0.873343in}{0.865486in}}%
\pgfpathlineto{\pgfqpoint{0.970565in}{0.865486in}}%
\pgfpathlineto{\pgfqpoint{0.970565in}{0.865486in}}%
\pgfpathlineto{\pgfqpoint{1.019176in}{0.865486in}}%
\pgfusepath{stroke}%
\end{pgfscope}%
\begin{pgfscope}%
\definecolor{textcolor}{rgb}{0.150000,0.150000,0.150000}%
\pgfsetstrokecolor{textcolor}%
\pgfsetfillcolor{textcolor}%
\pgftext[x=1.096954in,y=0.831458in,left,base]{\color{textcolor}\rmfamily\fontsize{7.000000}{8.400000}\selectfont 2}%
\end{pgfscope}%
\begin{pgfscope}%
\pgfsetroundcap%
\pgfsetroundjoin%
\pgfsetlinewidth{1.003750pt}%
\definecolor{currentstroke}{rgb}{0.007843,0.619608,0.450980}%
\pgfsetstrokecolor{currentstroke}%
\pgfsetdash{}{0pt}%
\pgfpathmoveto{\pgfqpoint{0.824732in}{0.729919in}}%
\pgfpathlineto{\pgfqpoint{0.873343in}{0.729919in}}%
\pgfpathlineto{\pgfqpoint{0.873343in}{0.729919in}}%
\pgfpathlineto{\pgfqpoint{0.970565in}{0.729919in}}%
\pgfpathlineto{\pgfqpoint{0.970565in}{0.729919in}}%
\pgfpathlineto{\pgfqpoint{1.019176in}{0.729919in}}%
\pgfusepath{stroke}%
\end{pgfscope}%
\begin{pgfscope}%
\definecolor{textcolor}{rgb}{0.150000,0.150000,0.150000}%
\pgfsetstrokecolor{textcolor}%
\pgfsetfillcolor{textcolor}%
\pgftext[x=1.096954in,y=0.695892in,left,base]{\color{textcolor}\rmfamily\fontsize{7.000000}{8.400000}\selectfont 4}%
\end{pgfscope}%
\end{pgfpicture}%
\makeatother%
\endgroup%

%% file: figures/experiments/graphs/sparse_smoothing/nodes_structure/node_classification-Cora-GAT-hidden=8-p_adj_plus=0.001-p_adj_minus=0.8-p_att_plus=0.0-p_att_minus=0.0-multi_class_cert-A.pgf
\begingroup%
\makeatletter%
\begin{pgfpicture}%
\pgfpathrectangle{\pgfpointorigin}{\pgfqpoint{1.375000in}{1.581250in}}%
\pgfusepath{use as bounding box, clip}%
\begin{pgfscope}%
\pgfsetbuttcap%
\pgfsetmiterjoin%
\definecolor{currentfill}{rgb}{1.000000,1.000000,1.000000}%
\pgfsetfillcolor{currentfill}%
\pgfsetlinewidth{0.000000pt}%
\definecolor{currentstroke}{rgb}{1.000000,1.000000,1.000000}%
\pgfsetstrokecolor{currentstroke}%
\pgfsetdash{}{0pt}%
\pgfpathmoveto{\pgfqpoint{0.000000in}{0.000000in}}%
\pgfpathlineto{\pgfqpoint{1.375000in}{0.000000in}}%
\pgfpathlineto{\pgfqpoint{1.375000in}{1.581250in}}%
\pgfpathlineto{\pgfqpoint{0.000000in}{1.581250in}}%
\pgfpathlineto{\pgfqpoint{0.000000in}{0.000000in}}%
\pgfpathclose%
\pgfusepath{fill}%
\end{pgfscope}%
\begin{pgfscope}%
\pgfsetbuttcap%
\pgfsetmiterjoin%
\definecolor{currentfill}{rgb}{1.000000,1.000000,1.000000}%
\pgfsetfillcolor{currentfill}%
\pgfsetlinewidth{0.000000pt}%
\definecolor{currentstroke}{rgb}{0.000000,0.000000,0.000000}%
\pgfsetstrokecolor{currentstroke}%
\pgfsetstrokeopacity{0.000000}%
\pgfsetdash{}{0pt}%
\pgfpathmoveto{\pgfqpoint{0.520798in}{0.442177in}}%
\pgfpathlineto{\pgfqpoint{1.239803in}{0.442177in}}%
\pgfpathlineto{\pgfqpoint{1.239803in}{1.314061in}}%
\pgfpathlineto{\pgfqpoint{0.520798in}{1.314061in}}%
\pgfpathlineto{\pgfqpoint{0.520798in}{0.442177in}}%
\pgfpathclose%
\pgfusepath{fill}%
\end{pgfscope}%
\begin{pgfscope}%
\pgfpathrectangle{\pgfqpoint{0.520798in}{0.442177in}}{\pgfqpoint{0.719005in}{0.871884in}}%
\pgfusepath{clip}%
\pgfsetroundcap%
\pgfsetroundjoin%
\pgfsetlinewidth{0.501875pt}%
\definecolor{currentstroke}{rgb}{0.800000,0.800000,0.800000}%
\pgfsetstrokecolor{currentstroke}%
\pgfsetdash{}{0pt}%
\pgfpathmoveto{\pgfqpoint{0.520798in}{0.442177in}}%
\pgfpathlineto{\pgfqpoint{0.520798in}{1.314061in}}%
\pgfusepath{stroke}%
\end{pgfscope}%
\begin{pgfscope}%
\definecolor{textcolor}{rgb}{0.150000,0.150000,0.150000}%
\pgfsetstrokecolor{textcolor}%
\pgfsetfillcolor{textcolor}%
\pgftext[x=0.520798in,y=0.351899in,,top]{\color{textcolor}\rmfamily\fontsize{8.000000}{9.600000}\selectfont \(\displaystyle {0}\)}%
\end{pgfscope}%
\begin{pgfscope}%
\pgfpathrectangle{\pgfqpoint{0.520798in}{0.442177in}}{\pgfqpoint{0.719005in}{0.871884in}}%
\pgfusepath{clip}%
\pgfsetroundcap%
\pgfsetroundjoin%
\pgfsetlinewidth{0.501875pt}%
\definecolor{currentstroke}{rgb}{0.800000,0.800000,0.800000}%
\pgfsetstrokecolor{currentstroke}%
\pgfsetdash{}{0pt}%
\pgfpathmoveto{\pgfqpoint{0.880300in}{0.442177in}}%
\pgfpathlineto{\pgfqpoint{0.880300in}{1.314061in}}%
\pgfusepath{stroke}%
\end{pgfscope}%
\begin{pgfscope}%
\definecolor{textcolor}{rgb}{0.150000,0.150000,0.150000}%
\pgfsetstrokecolor{textcolor}%
\pgfsetfillcolor{textcolor}%
\pgftext[x=0.880300in,y=0.351899in,,top]{\color{textcolor}\rmfamily\fontsize{8.000000}{9.600000}\selectfont \(\displaystyle {5}\)}%
\end{pgfscope}%
\begin{pgfscope}%
\pgfpathrectangle{\pgfqpoint{0.520798in}{0.442177in}}{\pgfqpoint{0.719005in}{0.871884in}}%
\pgfusepath{clip}%
\pgfsetroundcap%
\pgfsetroundjoin%
\pgfsetlinewidth{0.501875pt}%
\definecolor{currentstroke}{rgb}{0.800000,0.800000,0.800000}%
\pgfsetstrokecolor{currentstroke}%
\pgfsetdash{}{0pt}%
\pgfpathmoveto{\pgfqpoint{1.239803in}{0.442177in}}%
\pgfpathlineto{\pgfqpoint{1.239803in}{1.314061in}}%
\pgfusepath{stroke}%
\end{pgfscope}%
\begin{pgfscope}%
\definecolor{textcolor}{rgb}{0.150000,0.150000,0.150000}%
\pgfsetstrokecolor{textcolor}%
\pgfsetfillcolor{textcolor}%
\pgftext[x=1.239803in,y=0.351899in,,top]{\color{textcolor}\rmfamily\fontsize{8.000000}{9.600000}\selectfont \(\displaystyle {10}\)}%
\end{pgfscope}%
\begin{pgfscope}%
\definecolor{textcolor}{rgb}{0.150000,0.150000,0.150000}%
\pgfsetstrokecolor{textcolor}%
\pgfsetfillcolor{textcolor}%
\pgftext[x=0.880300in,y=0.198219in,,top]{\color{textcolor}\rmfamily\fontsize{10.000000}{12.000000}\selectfont Edit distance \(\displaystyle \epsilon\)}%
\end{pgfscope}%
\begin{pgfscope}%
\pgfpathrectangle{\pgfqpoint{0.520798in}{0.442177in}}{\pgfqpoint{0.719005in}{0.871884in}}%
\pgfusepath{clip}%
\pgfsetroundcap%
\pgfsetroundjoin%
\pgfsetlinewidth{0.501875pt}%
\definecolor{currentstroke}{rgb}{0.800000,0.800000,0.800000}%
\pgfsetstrokecolor{currentstroke}%
\pgfsetdash{}{0pt}%
\pgfpathmoveto{\pgfqpoint{0.520798in}{0.442177in}}%
\pgfpathlineto{\pgfqpoint{1.239803in}{0.442177in}}%
\pgfusepath{stroke}%
\end{pgfscope}%
\begin{pgfscope}%
\definecolor{textcolor}{rgb}{0.150000,0.150000,0.150000}%
\pgfsetstrokecolor{textcolor}%
\pgfsetfillcolor{textcolor}%
\pgftext[x=0.273151in, y=0.403915in, left, base]{\color{textcolor}\rmfamily\fontsize{8.000000}{9.600000}\selectfont 0\%}%
\end{pgfscope}%
\begin{pgfscope}%
\pgfpathrectangle{\pgfqpoint{0.520798in}{0.442177in}}{\pgfqpoint{0.719005in}{0.871884in}}%
\pgfusepath{clip}%
\pgfsetroundcap%
\pgfsetroundjoin%
\pgfsetlinewidth{0.501875pt}%
\definecolor{currentstroke}{rgb}{0.800000,0.800000,0.800000}%
\pgfsetstrokecolor{currentstroke}%
\pgfsetdash{}{0pt}%
\pgfpathmoveto{\pgfqpoint{0.520798in}{0.660148in}}%
\pgfpathlineto{\pgfqpoint{1.239803in}{0.660148in}}%
\pgfusepath{stroke}%
\end{pgfscope}%
\begin{pgfscope}%
\definecolor{textcolor}{rgb}{0.150000,0.150000,0.150000}%
\pgfsetstrokecolor{textcolor}%
\pgfsetfillcolor{textcolor}%
\pgftext[x=0.214138in, y=0.621886in, left, base]{\color{textcolor}\rmfamily\fontsize{8.000000}{9.600000}\selectfont 25\%}%
\end{pgfscope}%
\begin{pgfscope}%
\pgfpathrectangle{\pgfqpoint{0.520798in}{0.442177in}}{\pgfqpoint{0.719005in}{0.871884in}}%
\pgfusepath{clip}%
\pgfsetroundcap%
\pgfsetroundjoin%
\pgfsetlinewidth{0.501875pt}%
\definecolor{currentstroke}{rgb}{0.800000,0.800000,0.800000}%
\pgfsetstrokecolor{currentstroke}%
\pgfsetdash{}{0pt}%
\pgfpathmoveto{\pgfqpoint{0.520798in}{0.878119in}}%
\pgfpathlineto{\pgfqpoint{1.239803in}{0.878119in}}%
\pgfusepath{stroke}%
\end{pgfscope}%
\begin{pgfscope}%
\definecolor{textcolor}{rgb}{0.150000,0.150000,0.150000}%
\pgfsetstrokecolor{textcolor}%
\pgfsetfillcolor{textcolor}%
\pgftext[x=0.214138in, y=0.839857in, left, base]{\color{textcolor}\rmfamily\fontsize{8.000000}{9.600000}\selectfont 50\%}%
\end{pgfscope}%
\begin{pgfscope}%
\pgfpathrectangle{\pgfqpoint{0.520798in}{0.442177in}}{\pgfqpoint{0.719005in}{0.871884in}}%
\pgfusepath{clip}%
\pgfsetroundcap%
\pgfsetroundjoin%
\pgfsetlinewidth{0.501875pt}%
\definecolor{currentstroke}{rgb}{0.800000,0.800000,0.800000}%
\pgfsetstrokecolor{currentstroke}%
\pgfsetdash{}{0pt}%
\pgfpathmoveto{\pgfqpoint{0.520798in}{1.096090in}}%
\pgfpathlineto{\pgfqpoint{1.239803in}{1.096090in}}%
\pgfusepath{stroke}%
\end{pgfscope}%
\begin{pgfscope}%
\definecolor{textcolor}{rgb}{0.150000,0.150000,0.150000}%
\pgfsetstrokecolor{textcolor}%
\pgfsetfillcolor{textcolor}%
\pgftext[x=0.214138in, y=1.057828in, left, base]{\color{textcolor}\rmfamily\fontsize{8.000000}{9.600000}\selectfont 75\%}%
\end{pgfscope}%
\begin{pgfscope}%
\pgfpathrectangle{\pgfqpoint{0.520798in}{0.442177in}}{\pgfqpoint{0.719005in}{0.871884in}}%
\pgfusepath{clip}%
\pgfsetroundcap%
\pgfsetroundjoin%
\pgfsetlinewidth{0.501875pt}%
\definecolor{currentstroke}{rgb}{0.800000,0.800000,0.800000}%
\pgfsetstrokecolor{currentstroke}%
\pgfsetdash{}{0pt}%
\pgfpathmoveto{\pgfqpoint{0.520798in}{1.314061in}}%
\pgfpathlineto{\pgfqpoint{1.239803in}{1.314061in}}%
\pgfusepath{stroke}%
\end{pgfscope}%
\begin{pgfscope}%
\definecolor{textcolor}{rgb}{0.150000,0.150000,0.150000}%
\pgfsetstrokecolor{textcolor}%
\pgfsetfillcolor{textcolor}%
\pgftext[x=0.155124in, y=1.275799in, left, base]{\color{textcolor}\rmfamily\fontsize{8.000000}{9.600000}\selectfont 100\%}%
\end{pgfscope}%
\begin{pgfscope}%
\definecolor{textcolor}{rgb}{0.150000,0.150000,0.150000}%
\pgfsetstrokecolor{textcolor}%
\pgfsetfillcolor{textcolor}%
\pgftext[x=0.099569in,y=0.878119in,,bottom,rotate=90.000000]{\color{textcolor}\rmfamily\fontsize{10.000000}{12.000000}\selectfont Cert. Acc.}%
\end{pgfscope}%
\begin{pgfscope}%
\pgfsetrectcap%
\pgfsetmiterjoin%
\pgfsetlinewidth{0.752812pt}%
\definecolor{currentstroke}{rgb}{0.700000,0.700000,0.700000}%
\pgfsetstrokecolor{currentstroke}%
\pgfsetdash{}{0pt}%
\pgfpathmoveto{\pgfqpoint{0.520798in}{0.442177in}}%
\pgfpathlineto{\pgfqpoint{0.520798in}{1.314061in}}%
\pgfusepath{stroke}%
\end{pgfscope}%
\begin{pgfscope}%
\pgfsetrectcap%
\pgfsetmiterjoin%
\pgfsetlinewidth{0.752812pt}%
\definecolor{currentstroke}{rgb}{0.700000,0.700000,0.700000}%
\pgfsetstrokecolor{currentstroke}%
\pgfsetdash{}{0pt}%
\pgfpathmoveto{\pgfqpoint{1.239803in}{0.442177in}}%
\pgfpathlineto{\pgfqpoint{1.239803in}{1.314061in}}%
\pgfusepath{stroke}%
\end{pgfscope}%
\begin{pgfscope}%
\pgfsetrectcap%
\pgfsetmiterjoin%
\pgfsetlinewidth{0.752812pt}%
\definecolor{currentstroke}{rgb}{0.700000,0.700000,0.700000}%
\pgfsetstrokecolor{currentstroke}%
\pgfsetdash{}{0pt}%
\pgfpathmoveto{\pgfqpoint{0.520798in}{0.442177in}}%
\pgfpathlineto{\pgfqpoint{1.239803in}{0.442177in}}%
\pgfusepath{stroke}%
\end{pgfscope}%
\begin{pgfscope}%
\pgfsetrectcap%
\pgfsetmiterjoin%
\pgfsetlinewidth{0.752812pt}%
\definecolor{currentstroke}{rgb}{0.700000,0.700000,0.700000}%
\pgfsetstrokecolor{currentstroke}%
\pgfsetdash{}{0pt}%
\pgfpathmoveto{\pgfqpoint{0.520798in}{1.314061in}}%
\pgfpathlineto{\pgfqpoint{1.239803in}{1.314061in}}%
\pgfusepath{stroke}%
\end{pgfscope}%
\begin{pgfscope}%
\pgfpathrectangle{\pgfqpoint{0.520798in}{0.442177in}}{\pgfqpoint{0.719005in}{0.871884in}}%
\pgfusepath{clip}%
\pgfsetroundcap%
\pgfsetroundjoin%
\pgfsetlinewidth{1.003750pt}%
\definecolor{currentstroke}{rgb}{0.003922,0.450980,0.698039}%
\pgfsetstrokecolor{currentstroke}%
\pgfsetdash{}{0pt}%
\pgfpathmoveto{\pgfqpoint{0.520798in}{1.029997in}}%
\pgfpathlineto{\pgfqpoint{0.556748in}{1.029997in}}%
\pgfpathlineto{\pgfqpoint{0.556748in}{0.869301in}}%
\pgfpathlineto{\pgfqpoint{0.628649in}{0.869301in}}%
\pgfpathlineto{\pgfqpoint{0.628649in}{0.664715in}}%
\pgfpathlineto{\pgfqpoint{0.700549in}{0.664715in}}%
\pgfpathlineto{\pgfqpoint{0.700549in}{0.456491in}}%
\pgfpathlineto{\pgfqpoint{0.772450in}{0.456491in}}%
\pgfpathlineto{\pgfqpoint{0.772450in}{0.445103in}}%
\pgfpathlineto{\pgfqpoint{0.844350in}{0.445103in}}%
\pgfpathlineto{\pgfqpoint{0.844350in}{0.443917in}}%
\pgfpathlineto{\pgfqpoint{0.916251in}{0.443917in}}%
\pgfpathlineto{\pgfqpoint{0.916251in}{0.443363in}}%
\pgfpathlineto{\pgfqpoint{0.988151in}{0.443363in}}%
\pgfpathlineto{\pgfqpoint{0.988151in}{0.443205in}}%
\pgfpathlineto{\pgfqpoint{1.060052in}{0.443205in}}%
\pgfpathlineto{\pgfqpoint{1.060052in}{0.442572in}}%
\pgfpathlineto{\pgfqpoint{1.131952in}{0.442572in}}%
\pgfpathlineto{\pgfqpoint{1.131952in}{0.442493in}}%
\pgfpathlineto{\pgfqpoint{1.241470in}{0.442493in}}%
\pgfusepath{stroke}%
\end{pgfscope}%
\begin{pgfscope}%
\pgfpathrectangle{\pgfqpoint{0.520798in}{0.442177in}}{\pgfqpoint{0.719005in}{0.871884in}}%
\pgfusepath{clip}%
\pgfsetbuttcap%
\pgfsetroundjoin%
\definecolor{currentfill}{rgb}{0.003922,0.450980,0.698039}%
\pgfsetfillcolor{currentfill}%
\pgfsetfillopacity{0.500000}%
\pgfsetlinewidth{0.000000pt}%
\definecolor{currentstroke}{rgb}{0.003922,0.450980,0.698039}%
\pgfsetstrokecolor{currentstroke}%
\pgfsetstrokeopacity{0.500000}%
\pgfsetdash{}{0pt}%
\pgfpathmoveto{\pgfqpoint{0.520798in}{1.048669in}}%
\pgfpathlineto{\pgfqpoint{0.520798in}{1.011325in}}%
\pgfpathlineto{\pgfqpoint{0.556748in}{1.011325in}}%
\pgfpathlineto{\pgfqpoint{0.556748in}{0.850500in}}%
\pgfpathlineto{\pgfqpoint{0.628649in}{0.850500in}}%
\pgfpathlineto{\pgfqpoint{0.628649in}{0.649529in}}%
\pgfpathlineto{\pgfqpoint{0.700549in}{0.649529in}}%
\pgfpathlineto{\pgfqpoint{0.700549in}{0.452881in}}%
\pgfpathlineto{\pgfqpoint{0.772450in}{0.452881in}}%
\pgfpathlineto{\pgfqpoint{0.772450in}{0.443914in}}%
\pgfpathlineto{\pgfqpoint{0.844350in}{0.443914in}}%
\pgfpathlineto{\pgfqpoint{0.844350in}{0.443274in}}%
\pgfpathlineto{\pgfqpoint{0.916251in}{0.443274in}}%
\pgfpathlineto{\pgfqpoint{0.916251in}{0.443009in}}%
\pgfpathlineto{\pgfqpoint{0.988151in}{0.443009in}}%
\pgfpathlineto{\pgfqpoint{0.988151in}{0.442802in}}%
\pgfpathlineto{\pgfqpoint{1.060052in}{0.442802in}}%
\pgfpathlineto{\pgfqpoint{1.060052in}{0.442572in}}%
\pgfpathlineto{\pgfqpoint{1.131952in}{0.442572in}}%
\pgfpathlineto{\pgfqpoint{1.131952in}{0.442335in}}%
\pgfpathlineto{\pgfqpoint{1.311703in}{0.442335in}}%
\pgfpathlineto{\pgfqpoint{1.311703in}{0.442177in}}%
\pgfpathlineto{\pgfqpoint{2.066659in}{0.442177in}}%
\pgfpathlineto{\pgfqpoint{2.066659in}{0.442177in}}%
\pgfpathlineto{\pgfqpoint{2.677813in}{0.442177in}}%
\pgfpathlineto{\pgfqpoint{2.677813in}{0.442177in}}%
\pgfpathlineto{\pgfqpoint{2.677813in}{0.442177in}}%
\pgfpathlineto{\pgfqpoint{2.066659in}{0.442177in}}%
\pgfpathlineto{\pgfqpoint{2.066659in}{0.442177in}}%
\pgfpathlineto{\pgfqpoint{1.311703in}{0.442177in}}%
\pgfpathlineto{\pgfqpoint{1.311703in}{0.442651in}}%
\pgfpathlineto{\pgfqpoint{1.131952in}{0.442651in}}%
\pgfpathlineto{\pgfqpoint{1.131952in}{0.442572in}}%
\pgfpathlineto{\pgfqpoint{1.060052in}{0.442572in}}%
\pgfpathlineto{\pgfqpoint{1.060052in}{0.443608in}}%
\pgfpathlineto{\pgfqpoint{0.988151in}{0.443608in}}%
\pgfpathlineto{\pgfqpoint{0.988151in}{0.443717in}}%
\pgfpathlineto{\pgfqpoint{0.916251in}{0.443717in}}%
\pgfpathlineto{\pgfqpoint{0.916251in}{0.444559in}}%
\pgfpathlineto{\pgfqpoint{0.844350in}{0.444559in}}%
\pgfpathlineto{\pgfqpoint{0.844350in}{0.446292in}}%
\pgfpathlineto{\pgfqpoint{0.772450in}{0.446292in}}%
\pgfpathlineto{\pgfqpoint{0.772450in}{0.460101in}}%
\pgfpathlineto{\pgfqpoint{0.700549in}{0.460101in}}%
\pgfpathlineto{\pgfqpoint{0.700549in}{0.679901in}}%
\pgfpathlineto{\pgfqpoint{0.628649in}{0.679901in}}%
\pgfpathlineto{\pgfqpoint{0.628649in}{0.888103in}}%
\pgfpathlineto{\pgfqpoint{0.556748in}{0.888103in}}%
\pgfpathlineto{\pgfqpoint{0.556748in}{1.048669in}}%
\pgfpathlineto{\pgfqpoint{0.520798in}{1.048669in}}%
\pgfpathlineto{\pgfqpoint{0.520798in}{1.048669in}}%
\pgfpathclose%
\pgfusepath{fill}%
\end{pgfscope}%
\begin{pgfscope}%
\pgfpathrectangle{\pgfqpoint{0.520798in}{0.442177in}}{\pgfqpoint{0.719005in}{0.871884in}}%
\pgfusepath{clip}%
\pgfsetroundcap%
\pgfsetroundjoin%
\pgfsetlinewidth{1.003750pt}%
\definecolor{currentstroke}{rgb}{0.870588,0.560784,0.019608}%
\pgfsetstrokecolor{currentstroke}%
\pgfsetdash{}{0pt}%
\pgfpathmoveto{\pgfqpoint{0.520798in}{1.029997in}}%
\pgfpathlineto{\pgfqpoint{0.556748in}{1.029997in}}%
\pgfpathlineto{\pgfqpoint{0.556748in}{0.869301in}}%
\pgfpathlineto{\pgfqpoint{0.628649in}{0.869301in}}%
\pgfpathlineto{\pgfqpoint{0.628649in}{0.664715in}}%
\pgfpathlineto{\pgfqpoint{0.700549in}{0.664715in}}%
\pgfpathlineto{\pgfqpoint{0.700549in}{0.456491in}}%
\pgfpathlineto{\pgfqpoint{0.772450in}{0.456491in}}%
\pgfpathlineto{\pgfqpoint{0.772450in}{0.445103in}}%
\pgfpathlineto{\pgfqpoint{0.844350in}{0.445103in}}%
\pgfpathlineto{\pgfqpoint{0.844350in}{0.443996in}}%
\pgfpathlineto{\pgfqpoint{0.916251in}{0.443996in}}%
\pgfpathlineto{\pgfqpoint{0.916251in}{0.443521in}}%
\pgfpathlineto{\pgfqpoint{0.988151in}{0.443521in}}%
\pgfpathlineto{\pgfqpoint{0.988151in}{0.443205in}}%
\pgfpathlineto{\pgfqpoint{1.060052in}{0.443205in}}%
\pgfpathlineto{\pgfqpoint{1.060052in}{0.442572in}}%
\pgfpathlineto{\pgfqpoint{1.131952in}{0.442572in}}%
\pgfpathlineto{\pgfqpoint{1.131952in}{0.442493in}}%
\pgfpathlineto{\pgfqpoint{1.241470in}{0.442493in}}%
\pgfusepath{stroke}%
\end{pgfscope}%
\begin{pgfscope}%
\pgfpathrectangle{\pgfqpoint{0.520798in}{0.442177in}}{\pgfqpoint{0.719005in}{0.871884in}}%
\pgfusepath{clip}%
\pgfsetbuttcap%
\pgfsetroundjoin%
\definecolor{currentfill}{rgb}{0.870588,0.560784,0.019608}%
\pgfsetfillcolor{currentfill}%
\pgfsetfillopacity{0.500000}%
\pgfsetlinewidth{0.000000pt}%
\definecolor{currentstroke}{rgb}{0.870588,0.560784,0.019608}%
\pgfsetstrokecolor{currentstroke}%
\pgfsetstrokeopacity{0.500000}%
\pgfsetdash{}{0pt}%
\pgfpathmoveto{\pgfqpoint{0.520798in}{1.048669in}}%
\pgfpathlineto{\pgfqpoint{0.520798in}{1.011325in}}%
\pgfpathlineto{\pgfqpoint{0.556748in}{1.011325in}}%
\pgfpathlineto{\pgfqpoint{0.556748in}{0.850500in}}%
\pgfpathlineto{\pgfqpoint{0.628649in}{0.850500in}}%
\pgfpathlineto{\pgfqpoint{0.628649in}{0.649529in}}%
\pgfpathlineto{\pgfqpoint{0.700549in}{0.649529in}}%
\pgfpathlineto{\pgfqpoint{0.700549in}{0.452881in}}%
\pgfpathlineto{\pgfqpoint{0.772450in}{0.452881in}}%
\pgfpathlineto{\pgfqpoint{0.772450in}{0.443914in}}%
\pgfpathlineto{\pgfqpoint{0.844350in}{0.443914in}}%
\pgfpathlineto{\pgfqpoint{0.844350in}{0.443353in}}%
\pgfpathlineto{\pgfqpoint{0.916251in}{0.443353in}}%
\pgfpathlineto{\pgfqpoint{0.916251in}{0.443205in}}%
\pgfpathlineto{\pgfqpoint{0.988151in}{0.443205in}}%
\pgfpathlineto{\pgfqpoint{0.988151in}{0.442802in}}%
\pgfpathlineto{\pgfqpoint{1.060052in}{0.442802in}}%
\pgfpathlineto{\pgfqpoint{1.060052in}{0.442572in}}%
\pgfpathlineto{\pgfqpoint{1.131952in}{0.442572in}}%
\pgfpathlineto{\pgfqpoint{1.131952in}{0.442335in}}%
\pgfpathlineto{\pgfqpoint{1.311703in}{0.442335in}}%
\pgfpathlineto{\pgfqpoint{1.311703in}{0.442177in}}%
\pgfpathlineto{\pgfqpoint{2.066659in}{0.442177in}}%
\pgfpathlineto{\pgfqpoint{2.066659in}{0.442177in}}%
\pgfpathlineto{\pgfqpoint{2.677813in}{0.442177in}}%
\pgfpathlineto{\pgfqpoint{2.677813in}{0.442177in}}%
\pgfpathlineto{\pgfqpoint{2.677813in}{0.442177in}}%
\pgfpathlineto{\pgfqpoint{2.066659in}{0.442177in}}%
\pgfpathlineto{\pgfqpoint{2.066659in}{0.442177in}}%
\pgfpathlineto{\pgfqpoint{1.311703in}{0.442177in}}%
\pgfpathlineto{\pgfqpoint{1.311703in}{0.442651in}}%
\pgfpathlineto{\pgfqpoint{1.131952in}{0.442651in}}%
\pgfpathlineto{\pgfqpoint{1.131952in}{0.442572in}}%
\pgfpathlineto{\pgfqpoint{1.060052in}{0.442572in}}%
\pgfpathlineto{\pgfqpoint{1.060052in}{0.443608in}}%
\pgfpathlineto{\pgfqpoint{0.988151in}{0.443608in}}%
\pgfpathlineto{\pgfqpoint{0.988151in}{0.443838in}}%
\pgfpathlineto{\pgfqpoint{0.916251in}{0.443838in}}%
\pgfpathlineto{\pgfqpoint{0.916251in}{0.444638in}}%
\pgfpathlineto{\pgfqpoint{0.844350in}{0.444638in}}%
\pgfpathlineto{\pgfqpoint{0.844350in}{0.446292in}}%
\pgfpathlineto{\pgfqpoint{0.772450in}{0.446292in}}%
\pgfpathlineto{\pgfqpoint{0.772450in}{0.460101in}}%
\pgfpathlineto{\pgfqpoint{0.700549in}{0.460101in}}%
\pgfpathlineto{\pgfqpoint{0.700549in}{0.679901in}}%
\pgfpathlineto{\pgfqpoint{0.628649in}{0.679901in}}%
\pgfpathlineto{\pgfqpoint{0.628649in}{0.888103in}}%
\pgfpathlineto{\pgfqpoint{0.556748in}{0.888103in}}%
\pgfpathlineto{\pgfqpoint{0.556748in}{1.048669in}}%
\pgfpathlineto{\pgfqpoint{0.520798in}{1.048669in}}%
\pgfpathlineto{\pgfqpoint{0.520798in}{1.048669in}}%
\pgfpathclose%
\pgfusepath{fill}%
\end{pgfscope}%
\begin{pgfscope}%
\pgfpathrectangle{\pgfqpoint{0.520798in}{0.442177in}}{\pgfqpoint{0.719005in}{0.871884in}}%
\pgfusepath{clip}%
\pgfsetroundcap%
\pgfsetroundjoin%
\pgfsetlinewidth{1.003750pt}%
\definecolor{currentstroke}{rgb}{0.007843,0.619608,0.450980}%
\pgfsetstrokecolor{currentstroke}%
\pgfsetdash{}{0pt}%
\pgfpathmoveto{\pgfqpoint{0.520798in}{1.029997in}}%
\pgfpathlineto{\pgfqpoint{0.556748in}{1.029997in}}%
\pgfpathlineto{\pgfqpoint{0.556748in}{0.869301in}}%
\pgfpathlineto{\pgfqpoint{0.628649in}{0.869301in}}%
\pgfpathlineto{\pgfqpoint{0.628649in}{0.664715in}}%
\pgfpathlineto{\pgfqpoint{0.700549in}{0.664715in}}%
\pgfpathlineto{\pgfqpoint{0.700549in}{0.456491in}}%
\pgfpathlineto{\pgfqpoint{0.772450in}{0.456491in}}%
\pgfpathlineto{\pgfqpoint{0.772450in}{0.445103in}}%
\pgfpathlineto{\pgfqpoint{0.844350in}{0.445103in}}%
\pgfpathlineto{\pgfqpoint{0.844350in}{0.443996in}}%
\pgfpathlineto{\pgfqpoint{0.916251in}{0.443996in}}%
\pgfpathlineto{\pgfqpoint{0.916251in}{0.443521in}}%
\pgfpathlineto{\pgfqpoint{0.988151in}{0.443521in}}%
\pgfpathlineto{\pgfqpoint{0.988151in}{0.443205in}}%
\pgfpathlineto{\pgfqpoint{1.060052in}{0.443205in}}%
\pgfpathlineto{\pgfqpoint{1.060052in}{0.442572in}}%
\pgfpathlineto{\pgfqpoint{1.131952in}{0.442572in}}%
\pgfpathlineto{\pgfqpoint{1.131952in}{0.442493in}}%
\pgfpathlineto{\pgfqpoint{1.241470in}{0.442493in}}%
\pgfusepath{stroke}%
\end{pgfscope}%
\begin{pgfscope}%
\pgfpathrectangle{\pgfqpoint{0.520798in}{0.442177in}}{\pgfqpoint{0.719005in}{0.871884in}}%
\pgfusepath{clip}%
\pgfsetbuttcap%
\pgfsetroundjoin%
\definecolor{currentfill}{rgb}{0.007843,0.619608,0.450980}%
\pgfsetfillcolor{currentfill}%
\pgfsetfillopacity{0.500000}%
\pgfsetlinewidth{0.000000pt}%
\definecolor{currentstroke}{rgb}{0.007843,0.619608,0.450980}%
\pgfsetstrokecolor{currentstroke}%
\pgfsetstrokeopacity{0.500000}%
\pgfsetdash{}{0pt}%
\pgfpathmoveto{\pgfqpoint{0.520798in}{1.048669in}}%
\pgfpathlineto{\pgfqpoint{0.520798in}{1.011325in}}%
\pgfpathlineto{\pgfqpoint{0.556748in}{1.011325in}}%
\pgfpathlineto{\pgfqpoint{0.556748in}{0.850500in}}%
\pgfpathlineto{\pgfqpoint{0.628649in}{0.850500in}}%
\pgfpathlineto{\pgfqpoint{0.628649in}{0.649529in}}%
\pgfpathlineto{\pgfqpoint{0.700549in}{0.649529in}}%
\pgfpathlineto{\pgfqpoint{0.700549in}{0.452881in}}%
\pgfpathlineto{\pgfqpoint{0.772450in}{0.452881in}}%
\pgfpathlineto{\pgfqpoint{0.772450in}{0.443914in}}%
\pgfpathlineto{\pgfqpoint{0.844350in}{0.443914in}}%
\pgfpathlineto{\pgfqpoint{0.844350in}{0.443353in}}%
\pgfpathlineto{\pgfqpoint{0.916251in}{0.443353in}}%
\pgfpathlineto{\pgfqpoint{0.916251in}{0.443205in}}%
\pgfpathlineto{\pgfqpoint{0.988151in}{0.443205in}}%
\pgfpathlineto{\pgfqpoint{0.988151in}{0.442802in}}%
\pgfpathlineto{\pgfqpoint{1.060052in}{0.442802in}}%
\pgfpathlineto{\pgfqpoint{1.060052in}{0.442572in}}%
\pgfpathlineto{\pgfqpoint{1.131952in}{0.442572in}}%
\pgfpathlineto{\pgfqpoint{1.131952in}{0.442335in}}%
\pgfpathlineto{\pgfqpoint{1.311703in}{0.442335in}}%
\pgfpathlineto{\pgfqpoint{1.311703in}{0.442177in}}%
\pgfpathlineto{\pgfqpoint{2.066659in}{0.442177in}}%
\pgfpathlineto{\pgfqpoint{2.066659in}{0.442177in}}%
\pgfpathlineto{\pgfqpoint{2.677813in}{0.442177in}}%
\pgfpathlineto{\pgfqpoint{2.677813in}{0.442177in}}%
\pgfpathlineto{\pgfqpoint{2.677813in}{0.442177in}}%
\pgfpathlineto{\pgfqpoint{2.066659in}{0.442177in}}%
\pgfpathlineto{\pgfqpoint{2.066659in}{0.442177in}}%
\pgfpathlineto{\pgfqpoint{1.311703in}{0.442177in}}%
\pgfpathlineto{\pgfqpoint{1.311703in}{0.442651in}}%
\pgfpathlineto{\pgfqpoint{1.131952in}{0.442651in}}%
\pgfpathlineto{\pgfqpoint{1.131952in}{0.442572in}}%
\pgfpathlineto{\pgfqpoint{1.060052in}{0.442572in}}%
\pgfpathlineto{\pgfqpoint{1.060052in}{0.443608in}}%
\pgfpathlineto{\pgfqpoint{0.988151in}{0.443608in}}%
\pgfpathlineto{\pgfqpoint{0.988151in}{0.443838in}}%
\pgfpathlineto{\pgfqpoint{0.916251in}{0.443838in}}%
\pgfpathlineto{\pgfqpoint{0.916251in}{0.444638in}}%
\pgfpathlineto{\pgfqpoint{0.844350in}{0.444638in}}%
\pgfpathlineto{\pgfqpoint{0.844350in}{0.446292in}}%
\pgfpathlineto{\pgfqpoint{0.772450in}{0.446292in}}%
\pgfpathlineto{\pgfqpoint{0.772450in}{0.460101in}}%
\pgfpathlineto{\pgfqpoint{0.700549in}{0.460101in}}%
\pgfpathlineto{\pgfqpoint{0.700549in}{0.679901in}}%
\pgfpathlineto{\pgfqpoint{0.628649in}{0.679901in}}%
\pgfpathlineto{\pgfqpoint{0.628649in}{0.888103in}}%
\pgfpathlineto{\pgfqpoint{0.556748in}{0.888103in}}%
\pgfpathlineto{\pgfqpoint{0.556748in}{1.048669in}}%
\pgfpathlineto{\pgfqpoint{0.520798in}{1.048669in}}%
\pgfpathlineto{\pgfqpoint{0.520798in}{1.048669in}}%
\pgfpathclose%
\pgfusepath{fill}%
\end{pgfscope}%
\begin{pgfscope}%
\definecolor{textcolor}{rgb}{0.150000,0.150000,0.150000}%
\pgfsetstrokecolor{textcolor}%
\pgfsetfillcolor{textcolor}%
\pgftext[x=0.880300in,y=1.397395in,,base]{\color{textcolor}\rmfamily\fontsize{9.000000}{10.800000}\selectfont \(\displaystyle c_A^+=1\)}%
\end{pgfscope}%
\begin{pgfscope}%
\pgfsetbuttcap%
\pgfsetmiterjoin%
\definecolor{currentfill}{rgb}{1.000000,1.000000,1.000000}%
\pgfsetfillcolor{currentfill}%
\pgfsetfillopacity{0.800000}%
\pgfsetlinewidth{1.003750pt}%
\definecolor{currentstroke}{rgb}{0.800000,0.800000,0.800000}%
\pgfsetstrokecolor{currentstroke}%
\pgfsetstrokeopacity{0.800000}%
\pgfsetdash{}{0pt}%
\pgfpathmoveto{\pgfqpoint{0.785843in}{0.638103in}}%
\pgfpathlineto{\pgfqpoint{1.191192in}{0.638103in}}%
\pgfpathlineto{\pgfqpoint{1.191192in}{1.265450in}}%
\pgfpathlineto{\pgfqpoint{0.785843in}{1.265450in}}%
\pgfpathlineto{\pgfqpoint{0.785843in}{0.638103in}}%
\pgfpathclose%
\pgfusepath{stroke,fill}%
\end{pgfscope}%
\begin{pgfscope}%
\definecolor{textcolor}{rgb}{0.150000,0.150000,0.150000}%
\pgfsetstrokecolor{textcolor}%
\pgfsetfillcolor{textcolor}%
\pgftext[x=0.917113in,y=1.113275in,left,base]{\color{textcolor}\rmfamily\fontsize{9.000000}{10.800000}\selectfont \(\displaystyle c_A^-\)}%
\end{pgfscope}%
\begin{pgfscope}%
\pgfsetroundcap%
\pgfsetroundjoin%
\pgfsetlinewidth{1.003750pt}%
\definecolor{currentstroke}{rgb}{0.003922,0.450980,0.698039}%
\pgfsetstrokecolor{currentstroke}%
\pgfsetdash{}{0pt}%
\pgfpathmoveto{\pgfqpoint{0.824732in}{1.001052in}}%
\pgfpathlineto{\pgfqpoint{0.873343in}{1.001052in}}%
\pgfpathlineto{\pgfqpoint{0.873343in}{1.001052in}}%
\pgfpathlineto{\pgfqpoint{0.970565in}{1.001052in}}%
\pgfpathlineto{\pgfqpoint{0.970565in}{1.001052in}}%
\pgfpathlineto{\pgfqpoint{1.019176in}{1.001052in}}%
\pgfusepath{stroke}%
\end{pgfscope}%
\begin{pgfscope}%
\definecolor{textcolor}{rgb}{0.150000,0.150000,0.150000}%
\pgfsetstrokecolor{textcolor}%
\pgfsetfillcolor{textcolor}%
\pgftext[x=1.096954in,y=0.967024in,left,base]{\color{textcolor}\rmfamily\fontsize{7.000000}{8.400000}\selectfont 1}%
\end{pgfscope}%
\begin{pgfscope}%
\pgfsetroundcap%
\pgfsetroundjoin%
\pgfsetlinewidth{1.003750pt}%
\definecolor{currentstroke}{rgb}{0.870588,0.560784,0.019608}%
\pgfsetstrokecolor{currentstroke}%
\pgfsetdash{}{0pt}%
\pgfpathmoveto{\pgfqpoint{0.824732in}{0.865486in}}%
\pgfpathlineto{\pgfqpoint{0.873343in}{0.865486in}}%
\pgfpathlineto{\pgfqpoint{0.873343in}{0.865486in}}%
\pgfpathlineto{\pgfqpoint{0.970565in}{0.865486in}}%
\pgfpathlineto{\pgfqpoint{0.970565in}{0.865486in}}%
\pgfpathlineto{\pgfqpoint{1.019176in}{0.865486in}}%
\pgfusepath{stroke}%
\end{pgfscope}%
\begin{pgfscope}%
\definecolor{textcolor}{rgb}{0.150000,0.150000,0.150000}%
\pgfsetstrokecolor{textcolor}%
\pgfsetfillcolor{textcolor}%
\pgftext[x=1.096954in,y=0.831458in,left,base]{\color{textcolor}\rmfamily\fontsize{7.000000}{8.400000}\selectfont 2}%
\end{pgfscope}%
\begin{pgfscope}%
\pgfsetroundcap%
\pgfsetroundjoin%
\pgfsetlinewidth{1.003750pt}%
\definecolor{currentstroke}{rgb}{0.007843,0.619608,0.450980}%
\pgfsetstrokecolor{currentstroke}%
\pgfsetdash{}{0pt}%
\pgfpathmoveto{\pgfqpoint{0.824732in}{0.729919in}}%
\pgfpathlineto{\pgfqpoint{0.873343in}{0.729919in}}%
\pgfpathlineto{\pgfqpoint{0.873343in}{0.729919in}}%
\pgfpathlineto{\pgfqpoint{0.970565in}{0.729919in}}%
\pgfpathlineto{\pgfqpoint{0.970565in}{0.729919in}}%
\pgfpathlineto{\pgfqpoint{1.019176in}{0.729919in}}%
\pgfusepath{stroke}%
\end{pgfscope}%
\begin{pgfscope}%
\definecolor{textcolor}{rgb}{0.150000,0.150000,0.150000}%
\pgfsetstrokecolor{textcolor}%
\pgfsetfillcolor{textcolor}%
\pgftext[x=1.096954in,y=0.695892in,left,base]{\color{textcolor}\rmfamily\fontsize{7.000000}{8.400000}\selectfont 4}%
\end{pgfscope}%
\end{pgfpicture}%
\makeatother%
\endgroup%

%% file: figures/experiments/graphs/sparse_smoothing/nodes_attributes/node_classification-Citeseer-GAT-hidden=8-p_adj_plus=0.0-p_adj_minus=0.0-p_att_plus=0.001-p_att_minus=0.8-multi_class_cert-B.pgf
\begingroup%
\makeatletter%
\begin{pgfpicture}%
\pgfpathrectangle{\pgfpointorigin}{\pgfqpoint{1.375000in}{1.581250in}}%
\pgfusepath{use as bounding box, clip}%
\begin{pgfscope}%
\pgfsetbuttcap%
\pgfsetmiterjoin%
\definecolor{currentfill}{rgb}{1.000000,1.000000,1.000000}%
\pgfsetfillcolor{currentfill}%
\pgfsetlinewidth{0.000000pt}%
\definecolor{currentstroke}{rgb}{1.000000,1.000000,1.000000}%
\pgfsetstrokecolor{currentstroke}%
\pgfsetdash{}{0pt}%
\pgfpathmoveto{\pgfqpoint{0.000000in}{0.000000in}}%
\pgfpathlineto{\pgfqpoint{1.375000in}{0.000000in}}%
\pgfpathlineto{\pgfqpoint{1.375000in}{1.581250in}}%
\pgfpathlineto{\pgfqpoint{0.000000in}{1.581250in}}%
\pgfpathlineto{\pgfqpoint{0.000000in}{0.000000in}}%
\pgfpathclose%
\pgfusepath{fill}%
\end{pgfscope}%
\begin{pgfscope}%
\pgfsetbuttcap%
\pgfsetmiterjoin%
\definecolor{currentfill}{rgb}{1.000000,1.000000,1.000000}%
\pgfsetfillcolor{currentfill}%
\pgfsetlinewidth{0.000000pt}%
\definecolor{currentstroke}{rgb}{0.000000,0.000000,0.000000}%
\pgfsetstrokecolor{currentstroke}%
\pgfsetstrokeopacity{0.000000}%
\pgfsetdash{}{0pt}%
\pgfpathmoveto{\pgfqpoint{0.520798in}{0.442177in}}%
\pgfpathlineto{\pgfqpoint{1.239803in}{0.442177in}}%
\pgfpathlineto{\pgfqpoint{1.239803in}{1.314061in}}%
\pgfpathlineto{\pgfqpoint{0.520798in}{1.314061in}}%
\pgfpathlineto{\pgfqpoint{0.520798in}{0.442177in}}%
\pgfpathclose%
\pgfusepath{fill}%
\end{pgfscope}%
\begin{pgfscope}%
\pgfpathrectangle{\pgfqpoint{0.520798in}{0.442177in}}{\pgfqpoint{0.719005in}{0.871884in}}%
\pgfusepath{clip}%
\pgfsetroundcap%
\pgfsetroundjoin%
\pgfsetlinewidth{0.501875pt}%
\definecolor{currentstroke}{rgb}{0.800000,0.800000,0.800000}%
\pgfsetstrokecolor{currentstroke}%
\pgfsetdash{}{0pt}%
\pgfpathmoveto{\pgfqpoint{0.520798in}{0.442177in}}%
\pgfpathlineto{\pgfqpoint{0.520798in}{1.314061in}}%
\pgfusepath{stroke}%
\end{pgfscope}%
\begin{pgfscope}%
\definecolor{textcolor}{rgb}{0.150000,0.150000,0.150000}%
\pgfsetstrokecolor{textcolor}%
\pgfsetfillcolor{textcolor}%
\pgftext[x=0.520798in,y=0.351899in,,top]{\color{textcolor}\rmfamily\fontsize{8.000000}{9.600000}\selectfont \(\displaystyle {0}\)}%
\end{pgfscope}%
\begin{pgfscope}%
\pgfpathrectangle{\pgfqpoint{0.520798in}{0.442177in}}{\pgfqpoint{0.719005in}{0.871884in}}%
\pgfusepath{clip}%
\pgfsetroundcap%
\pgfsetroundjoin%
\pgfsetlinewidth{0.501875pt}%
\definecolor{currentstroke}{rgb}{0.800000,0.800000,0.800000}%
\pgfsetstrokecolor{currentstroke}%
\pgfsetdash{}{0pt}%
\pgfpathmoveto{\pgfqpoint{0.880300in}{0.442177in}}%
\pgfpathlineto{\pgfqpoint{0.880300in}{1.314061in}}%
\pgfusepath{stroke}%
\end{pgfscope}%
\begin{pgfscope}%
\definecolor{textcolor}{rgb}{0.150000,0.150000,0.150000}%
\pgfsetstrokecolor{textcolor}%
\pgfsetfillcolor{textcolor}%
\pgftext[x=0.880300in,y=0.351899in,,top]{\color{textcolor}\rmfamily\fontsize{8.000000}{9.600000}\selectfont \(\displaystyle {5}\)}%
\end{pgfscope}%
\begin{pgfscope}%
\pgfpathrectangle{\pgfqpoint{0.520798in}{0.442177in}}{\pgfqpoint{0.719005in}{0.871884in}}%
\pgfusepath{clip}%
\pgfsetroundcap%
\pgfsetroundjoin%
\pgfsetlinewidth{0.501875pt}%
\definecolor{currentstroke}{rgb}{0.800000,0.800000,0.800000}%
\pgfsetstrokecolor{currentstroke}%
\pgfsetdash{}{0pt}%
\pgfpathmoveto{\pgfqpoint{1.239803in}{0.442177in}}%
\pgfpathlineto{\pgfqpoint{1.239803in}{1.314061in}}%
\pgfusepath{stroke}%
\end{pgfscope}%
\begin{pgfscope}%
\definecolor{textcolor}{rgb}{0.150000,0.150000,0.150000}%
\pgfsetstrokecolor{textcolor}%
\pgfsetfillcolor{textcolor}%
\pgftext[x=1.239803in,y=0.351899in,,top]{\color{textcolor}\rmfamily\fontsize{8.000000}{9.600000}\selectfont \(\displaystyle {10}\)}%
\end{pgfscope}%
\begin{pgfscope}%
\definecolor{textcolor}{rgb}{0.150000,0.150000,0.150000}%
\pgfsetstrokecolor{textcolor}%
\pgfsetfillcolor{textcolor}%
\pgftext[x=0.880300in,y=0.198219in,,top]{\color{textcolor}\rmfamily\fontsize{10.000000}{12.000000}\selectfont Edit distance \(\displaystyle \epsilon\)}%
\end{pgfscope}%
\begin{pgfscope}%
\pgfpathrectangle{\pgfqpoint{0.520798in}{0.442177in}}{\pgfqpoint{0.719005in}{0.871884in}}%
\pgfusepath{clip}%
\pgfsetroundcap%
\pgfsetroundjoin%
\pgfsetlinewidth{0.501875pt}%
\definecolor{currentstroke}{rgb}{0.800000,0.800000,0.800000}%
\pgfsetstrokecolor{currentstroke}%
\pgfsetdash{}{0pt}%
\pgfpathmoveto{\pgfqpoint{0.520798in}{0.442177in}}%
\pgfpathlineto{\pgfqpoint{1.239803in}{0.442177in}}%
\pgfusepath{stroke}%
\end{pgfscope}%
\begin{pgfscope}%
\definecolor{textcolor}{rgb}{0.150000,0.150000,0.150000}%
\pgfsetstrokecolor{textcolor}%
\pgfsetfillcolor{textcolor}%
\pgftext[x=0.273151in, y=0.403915in, left, base]{\color{textcolor}\rmfamily\fontsize{8.000000}{9.600000}\selectfont 0\%}%
\end{pgfscope}%
\begin{pgfscope}%
\pgfpathrectangle{\pgfqpoint{0.520798in}{0.442177in}}{\pgfqpoint{0.719005in}{0.871884in}}%
\pgfusepath{clip}%
\pgfsetroundcap%
\pgfsetroundjoin%
\pgfsetlinewidth{0.501875pt}%
\definecolor{currentstroke}{rgb}{0.800000,0.800000,0.800000}%
\pgfsetstrokecolor{currentstroke}%
\pgfsetdash{}{0pt}%
\pgfpathmoveto{\pgfqpoint{0.520798in}{0.660148in}}%
\pgfpathlineto{\pgfqpoint{1.239803in}{0.660148in}}%
\pgfusepath{stroke}%
\end{pgfscope}%
\begin{pgfscope}%
\definecolor{textcolor}{rgb}{0.150000,0.150000,0.150000}%
\pgfsetstrokecolor{textcolor}%
\pgfsetfillcolor{textcolor}%
\pgftext[x=0.214138in, y=0.621886in, left, base]{\color{textcolor}\rmfamily\fontsize{8.000000}{9.600000}\selectfont 25\%}%
\end{pgfscope}%
\begin{pgfscope}%
\pgfpathrectangle{\pgfqpoint{0.520798in}{0.442177in}}{\pgfqpoint{0.719005in}{0.871884in}}%
\pgfusepath{clip}%
\pgfsetroundcap%
\pgfsetroundjoin%
\pgfsetlinewidth{0.501875pt}%
\definecolor{currentstroke}{rgb}{0.800000,0.800000,0.800000}%
\pgfsetstrokecolor{currentstroke}%
\pgfsetdash{}{0pt}%
\pgfpathmoveto{\pgfqpoint{0.520798in}{0.878119in}}%
\pgfpathlineto{\pgfqpoint{1.239803in}{0.878119in}}%
\pgfusepath{stroke}%
\end{pgfscope}%
\begin{pgfscope}%
\definecolor{textcolor}{rgb}{0.150000,0.150000,0.150000}%
\pgfsetstrokecolor{textcolor}%
\pgfsetfillcolor{textcolor}%
\pgftext[x=0.214138in, y=0.839857in, left, base]{\color{textcolor}\rmfamily\fontsize{8.000000}{9.600000}\selectfont 50\%}%
\end{pgfscope}%
\begin{pgfscope}%
\pgfpathrectangle{\pgfqpoint{0.520798in}{0.442177in}}{\pgfqpoint{0.719005in}{0.871884in}}%
\pgfusepath{clip}%
\pgfsetroundcap%
\pgfsetroundjoin%
\pgfsetlinewidth{0.501875pt}%
\definecolor{currentstroke}{rgb}{0.800000,0.800000,0.800000}%
\pgfsetstrokecolor{currentstroke}%
\pgfsetdash{}{0pt}%
\pgfpathmoveto{\pgfqpoint{0.520798in}{1.096090in}}%
\pgfpathlineto{\pgfqpoint{1.239803in}{1.096090in}}%
\pgfusepath{stroke}%
\end{pgfscope}%
\begin{pgfscope}%
\definecolor{textcolor}{rgb}{0.150000,0.150000,0.150000}%
\pgfsetstrokecolor{textcolor}%
\pgfsetfillcolor{textcolor}%
\pgftext[x=0.214138in, y=1.057828in, left, base]{\color{textcolor}\rmfamily\fontsize{8.000000}{9.600000}\selectfont 75\%}%
\end{pgfscope}%
\begin{pgfscope}%
\pgfpathrectangle{\pgfqpoint{0.520798in}{0.442177in}}{\pgfqpoint{0.719005in}{0.871884in}}%
\pgfusepath{clip}%
\pgfsetroundcap%
\pgfsetroundjoin%
\pgfsetlinewidth{0.501875pt}%
\definecolor{currentstroke}{rgb}{0.800000,0.800000,0.800000}%
\pgfsetstrokecolor{currentstroke}%
\pgfsetdash{}{0pt}%
\pgfpathmoveto{\pgfqpoint{0.520798in}{1.314061in}}%
\pgfpathlineto{\pgfqpoint{1.239803in}{1.314061in}}%
\pgfusepath{stroke}%
\end{pgfscope}%
\begin{pgfscope}%
\definecolor{textcolor}{rgb}{0.150000,0.150000,0.150000}%
\pgfsetstrokecolor{textcolor}%
\pgfsetfillcolor{textcolor}%
\pgftext[x=0.155124in, y=1.275799in, left, base]{\color{textcolor}\rmfamily\fontsize{8.000000}{9.600000}\selectfont 100\%}%
\end{pgfscope}%
\begin{pgfscope}%
\definecolor{textcolor}{rgb}{0.150000,0.150000,0.150000}%
\pgfsetstrokecolor{textcolor}%
\pgfsetfillcolor{textcolor}%
\pgftext[x=0.099569in,y=0.878119in,,bottom,rotate=90.000000]{\color{textcolor}\rmfamily\fontsize{10.000000}{12.000000}\selectfont Cert. Acc.}%
\end{pgfscope}%
\begin{pgfscope}%
\pgfsetrectcap%
\pgfsetmiterjoin%
\pgfsetlinewidth{0.752812pt}%
\definecolor{currentstroke}{rgb}{0.700000,0.700000,0.700000}%
\pgfsetstrokecolor{currentstroke}%
\pgfsetdash{}{0pt}%
\pgfpathmoveto{\pgfqpoint{0.520798in}{0.442177in}}%
\pgfpathlineto{\pgfqpoint{0.520798in}{1.314061in}}%
\pgfusepath{stroke}%
\end{pgfscope}%
\begin{pgfscope}%
\pgfsetrectcap%
\pgfsetmiterjoin%
\pgfsetlinewidth{0.752812pt}%
\definecolor{currentstroke}{rgb}{0.700000,0.700000,0.700000}%
\pgfsetstrokecolor{currentstroke}%
\pgfsetdash{}{0pt}%
\pgfpathmoveto{\pgfqpoint{1.239803in}{0.442177in}}%
\pgfpathlineto{\pgfqpoint{1.239803in}{1.314061in}}%
\pgfusepath{stroke}%
\end{pgfscope}%
\begin{pgfscope}%
\pgfsetrectcap%
\pgfsetmiterjoin%
\pgfsetlinewidth{0.752812pt}%
\definecolor{currentstroke}{rgb}{0.700000,0.700000,0.700000}%
\pgfsetstrokecolor{currentstroke}%
\pgfsetdash{}{0pt}%
\pgfpathmoveto{\pgfqpoint{0.520798in}{0.442177in}}%
\pgfpathlineto{\pgfqpoint{1.239803in}{0.442177in}}%
\pgfusepath{stroke}%
\end{pgfscope}%
\begin{pgfscope}%
\pgfsetrectcap%
\pgfsetmiterjoin%
\pgfsetlinewidth{0.752812pt}%
\definecolor{currentstroke}{rgb}{0.700000,0.700000,0.700000}%
\pgfsetstrokecolor{currentstroke}%
\pgfsetdash{}{0pt}%
\pgfpathmoveto{\pgfqpoint{0.520798in}{1.314061in}}%
\pgfpathlineto{\pgfqpoint{1.239803in}{1.314061in}}%
\pgfusepath{stroke}%
\end{pgfscope}%
\begin{pgfscope}%
\pgfpathrectangle{\pgfqpoint{0.520798in}{0.442177in}}{\pgfqpoint{0.719005in}{0.871884in}}%
\pgfusepath{clip}%
\pgfsetroundcap%
\pgfsetroundjoin%
\pgfsetlinewidth{1.003750pt}%
\definecolor{currentstroke}{rgb}{0.003922,0.450980,0.698039}%
\pgfsetstrokecolor{currentstroke}%
\pgfsetdash{}{0pt}%
\pgfpathmoveto{\pgfqpoint{0.520798in}{1.064090in}}%
\pgfpathlineto{\pgfqpoint{0.556748in}{1.064090in}}%
\pgfpathlineto{\pgfqpoint{0.556748in}{1.008252in}}%
\pgfpathlineto{\pgfqpoint{0.628649in}{1.008252in}}%
\pgfpathlineto{\pgfqpoint{0.628649in}{0.921806in}}%
\pgfpathlineto{\pgfqpoint{0.700549in}{0.921806in}}%
\pgfpathlineto{\pgfqpoint{0.700549in}{0.713296in}}%
\pgfpathlineto{\pgfqpoint{0.772450in}{0.713296in}}%
\pgfpathlineto{\pgfqpoint{0.772450in}{0.612194in}}%
\pgfpathlineto{\pgfqpoint{0.844350in}{0.612194in}}%
\pgfpathlineto{\pgfqpoint{0.844350in}{0.600322in}}%
\pgfpathlineto{\pgfqpoint{0.916251in}{0.600322in}}%
\pgfpathlineto{\pgfqpoint{0.916251in}{0.592994in}}%
\pgfpathlineto{\pgfqpoint{0.988151in}{0.592994in}}%
\pgfpathlineto{\pgfqpoint{0.988151in}{0.581307in}}%
\pgfpathlineto{\pgfqpoint{1.060052in}{0.581307in}}%
\pgfpathlineto{\pgfqpoint{1.060052in}{0.536136in}}%
\pgfpathlineto{\pgfqpoint{1.131952in}{0.536136in}}%
\pgfpathlineto{\pgfqpoint{1.131952in}{0.527232in}}%
\pgfpathlineto{\pgfqpoint{1.203853in}{0.527232in}}%
\pgfpathlineto{\pgfqpoint{1.203853in}{0.509887in}}%
\pgfpathlineto{\pgfqpoint{1.241470in}{0.509887in}}%
\pgfusepath{stroke}%
\end{pgfscope}%
\begin{pgfscope}%
\pgfpathrectangle{\pgfqpoint{0.520798in}{0.442177in}}{\pgfqpoint{0.719005in}{0.871884in}}%
\pgfusepath{clip}%
\pgfsetbuttcap%
\pgfsetroundjoin%
\definecolor{currentfill}{rgb}{0.003922,0.450980,0.698039}%
\pgfsetfillcolor{currentfill}%
\pgfsetfillopacity{0.500000}%
\pgfsetlinewidth{0.000000pt}%
\definecolor{currentstroke}{rgb}{0.003922,0.450980,0.698039}%
\pgfsetstrokecolor{currentstroke}%
\pgfsetstrokeopacity{0.500000}%
\pgfsetdash{}{0pt}%
\pgfpathmoveto{\pgfqpoint{0.520798in}{1.076194in}}%
\pgfpathlineto{\pgfqpoint{0.520798in}{1.051986in}}%
\pgfpathlineto{\pgfqpoint{0.556748in}{1.051986in}}%
\pgfpathlineto{\pgfqpoint{0.556748in}{0.994752in}}%
\pgfpathlineto{\pgfqpoint{0.628649in}{0.994752in}}%
\pgfpathlineto{\pgfqpoint{0.628649in}{0.904877in}}%
\pgfpathlineto{\pgfqpoint{0.700549in}{0.904877in}}%
\pgfpathlineto{\pgfqpoint{0.700549in}{0.694707in}}%
\pgfpathlineto{\pgfqpoint{0.772450in}{0.694707in}}%
\pgfpathlineto{\pgfqpoint{0.772450in}{0.597071in}}%
\pgfpathlineto{\pgfqpoint{0.844350in}{0.597071in}}%
\pgfpathlineto{\pgfqpoint{0.844350in}{0.585768in}}%
\pgfpathlineto{\pgfqpoint{0.916251in}{0.585768in}}%
\pgfpathlineto{\pgfqpoint{0.916251in}{0.579635in}}%
\pgfpathlineto{\pgfqpoint{0.988151in}{0.579635in}}%
\pgfpathlineto{\pgfqpoint{0.988151in}{0.568385in}}%
\pgfpathlineto{\pgfqpoint{1.060052in}{0.568385in}}%
\pgfpathlineto{\pgfqpoint{1.060052in}{0.525437in}}%
\pgfpathlineto{\pgfqpoint{1.131952in}{0.525437in}}%
\pgfpathlineto{\pgfqpoint{1.131952in}{0.515930in}}%
\pgfpathlineto{\pgfqpoint{1.203853in}{0.515930in}}%
\pgfpathlineto{\pgfqpoint{1.203853in}{0.500124in}}%
\pgfpathlineto{\pgfqpoint{1.275753in}{0.500124in}}%
\pgfpathlineto{\pgfqpoint{1.275753in}{0.490095in}}%
\pgfpathlineto{\pgfqpoint{1.347654in}{0.490095in}}%
\pgfpathlineto{\pgfqpoint{1.347654in}{0.480913in}}%
\pgfpathlineto{\pgfqpoint{1.419554in}{0.480913in}}%
\pgfpathlineto{\pgfqpoint{1.419554in}{0.442177in}}%
\pgfpathlineto{\pgfqpoint{2.066659in}{0.442177in}}%
\pgfpathlineto{\pgfqpoint{2.066659in}{0.442177in}}%
\pgfpathlineto{\pgfqpoint{2.677813in}{0.442177in}}%
\pgfpathlineto{\pgfqpoint{2.677813in}{0.442177in}}%
\pgfpathlineto{\pgfqpoint{2.677813in}{0.442177in}}%
\pgfpathlineto{\pgfqpoint{2.066659in}{0.442177in}}%
\pgfpathlineto{\pgfqpoint{2.066659in}{0.442177in}}%
\pgfpathlineto{\pgfqpoint{1.419554in}{0.442177in}}%
\pgfpathlineto{\pgfqpoint{1.419554in}{0.498420in}}%
\pgfpathlineto{\pgfqpoint{1.347654in}{0.498420in}}%
\pgfpathlineto{\pgfqpoint{1.347654in}{0.503337in}}%
\pgfpathlineto{\pgfqpoint{1.275753in}{0.503337in}}%
\pgfpathlineto{\pgfqpoint{1.275753in}{0.519650in}}%
\pgfpathlineto{\pgfqpoint{1.203853in}{0.519650in}}%
\pgfpathlineto{\pgfqpoint{1.203853in}{0.538533in}}%
\pgfpathlineto{\pgfqpoint{1.131952in}{0.538533in}}%
\pgfpathlineto{\pgfqpoint{1.131952in}{0.546836in}}%
\pgfpathlineto{\pgfqpoint{1.060052in}{0.546836in}}%
\pgfpathlineto{\pgfqpoint{1.060052in}{0.594230in}}%
\pgfpathlineto{\pgfqpoint{0.988151in}{0.594230in}}%
\pgfpathlineto{\pgfqpoint{0.988151in}{0.606353in}}%
\pgfpathlineto{\pgfqpoint{0.916251in}{0.606353in}}%
\pgfpathlineto{\pgfqpoint{0.916251in}{0.614876in}}%
\pgfpathlineto{\pgfqpoint{0.844350in}{0.614876in}}%
\pgfpathlineto{\pgfqpoint{0.844350in}{0.627318in}}%
\pgfpathlineto{\pgfqpoint{0.772450in}{0.627318in}}%
\pgfpathlineto{\pgfqpoint{0.772450in}{0.731885in}}%
\pgfpathlineto{\pgfqpoint{0.700549in}{0.731885in}}%
\pgfpathlineto{\pgfqpoint{0.700549in}{0.938736in}}%
\pgfpathlineto{\pgfqpoint{0.628649in}{0.938736in}}%
\pgfpathlineto{\pgfqpoint{0.628649in}{1.021753in}}%
\pgfpathlineto{\pgfqpoint{0.556748in}{1.021753in}}%
\pgfpathlineto{\pgfqpoint{0.556748in}{1.076194in}}%
\pgfpathlineto{\pgfqpoint{0.520798in}{1.076194in}}%
\pgfpathlineto{\pgfqpoint{0.520798in}{1.076194in}}%
\pgfpathclose%
\pgfusepath{fill}%
\end{pgfscope}%
\begin{pgfscope}%
\pgfpathrectangle{\pgfqpoint{0.520798in}{0.442177in}}{\pgfqpoint{0.719005in}{0.871884in}}%
\pgfusepath{clip}%
\pgfsetroundcap%
\pgfsetroundjoin%
\pgfsetlinewidth{1.003750pt}%
\definecolor{currentstroke}{rgb}{0.870588,0.560784,0.019608}%
\pgfsetstrokecolor{currentstroke}%
\pgfsetdash{}{0pt}%
\pgfpathmoveto{\pgfqpoint{0.520798in}{1.064090in}}%
\pgfpathlineto{\pgfqpoint{0.556748in}{1.064090in}}%
\pgfpathlineto{\pgfqpoint{0.556748in}{1.020589in}}%
\pgfpathlineto{\pgfqpoint{0.628649in}{1.020589in}}%
\pgfpathlineto{\pgfqpoint{0.628649in}{0.978571in}}%
\pgfpathlineto{\pgfqpoint{0.700549in}{0.978571in}}%
\pgfpathlineto{\pgfqpoint{0.700549in}{0.942583in}}%
\pgfpathlineto{\pgfqpoint{0.772450in}{0.942583in}}%
\pgfpathlineto{\pgfqpoint{0.772450in}{0.911603in}}%
\pgfpathlineto{\pgfqpoint{0.844350in}{0.911603in}}%
\pgfpathlineto{\pgfqpoint{0.844350in}{0.882293in}}%
\pgfpathlineto{\pgfqpoint{0.916251in}{0.882293in}}%
\pgfpathlineto{\pgfqpoint{0.916251in}{0.713296in}}%
\pgfpathlineto{\pgfqpoint{0.988151in}{0.713296in}}%
\pgfpathlineto{\pgfqpoint{0.988151in}{0.702536in}}%
\pgfpathlineto{\pgfqpoint{1.060052in}{0.702536in}}%
\pgfpathlineto{\pgfqpoint{1.060052in}{0.612194in}}%
\pgfpathlineto{\pgfqpoint{1.131952in}{0.612194in}}%
\pgfpathlineto{\pgfqpoint{1.131952in}{0.603754in}}%
\pgfpathlineto{\pgfqpoint{1.203853in}{0.603754in}}%
\pgfpathlineto{\pgfqpoint{1.203853in}{0.594757in}}%
\pgfpathlineto{\pgfqpoint{1.241470in}{0.594757in}}%
\pgfusepath{stroke}%
\end{pgfscope}%
\begin{pgfscope}%
\pgfpathrectangle{\pgfqpoint{0.520798in}{0.442177in}}{\pgfqpoint{0.719005in}{0.871884in}}%
\pgfusepath{clip}%
\pgfsetbuttcap%
\pgfsetroundjoin%
\definecolor{currentfill}{rgb}{0.870588,0.560784,0.019608}%
\pgfsetfillcolor{currentfill}%
\pgfsetfillopacity{0.500000}%
\pgfsetlinewidth{0.000000pt}%
\definecolor{currentstroke}{rgb}{0.870588,0.560784,0.019608}%
\pgfsetstrokecolor{currentstroke}%
\pgfsetstrokeopacity{0.500000}%
\pgfsetdash{}{0pt}%
\pgfpathmoveto{\pgfqpoint{0.520798in}{1.076194in}}%
\pgfpathlineto{\pgfqpoint{0.520798in}{1.051986in}}%
\pgfpathlineto{\pgfqpoint{0.556748in}{1.051986in}}%
\pgfpathlineto{\pgfqpoint{0.556748in}{1.007252in}}%
\pgfpathlineto{\pgfqpoint{0.628649in}{1.007252in}}%
\pgfpathlineto{\pgfqpoint{0.628649in}{0.963425in}}%
\pgfpathlineto{\pgfqpoint{0.700549in}{0.963425in}}%
\pgfpathlineto{\pgfqpoint{0.700549in}{0.927260in}}%
\pgfpathlineto{\pgfqpoint{0.772450in}{0.927260in}}%
\pgfpathlineto{\pgfqpoint{0.772450in}{0.893372in}}%
\pgfpathlineto{\pgfqpoint{0.844350in}{0.893372in}}%
\pgfpathlineto{\pgfqpoint{0.844350in}{0.862018in}}%
\pgfpathlineto{\pgfqpoint{0.916251in}{0.862018in}}%
\pgfpathlineto{\pgfqpoint{0.916251in}{0.694707in}}%
\pgfpathlineto{\pgfqpoint{0.988151in}{0.694707in}}%
\pgfpathlineto{\pgfqpoint{0.988151in}{0.683337in}}%
\pgfpathlineto{\pgfqpoint{1.060052in}{0.683337in}}%
\pgfpathlineto{\pgfqpoint{1.060052in}{0.597071in}}%
\pgfpathlineto{\pgfqpoint{1.131952in}{0.597071in}}%
\pgfpathlineto{\pgfqpoint{1.131952in}{0.589497in}}%
\pgfpathlineto{\pgfqpoint{1.203853in}{0.589497in}}%
\pgfpathlineto{\pgfqpoint{1.203853in}{0.580847in}}%
\pgfpathlineto{\pgfqpoint{1.275753in}{0.580847in}}%
\pgfpathlineto{\pgfqpoint{1.275753in}{0.573044in}}%
\pgfpathlineto{\pgfqpoint{1.347654in}{0.573044in}}%
\pgfpathlineto{\pgfqpoint{1.347654in}{0.569428in}}%
\pgfpathlineto{\pgfqpoint{1.419554in}{0.569428in}}%
\pgfpathlineto{\pgfqpoint{1.419554in}{0.563968in}}%
\pgfpathlineto{\pgfqpoint{1.491455in}{0.563968in}}%
\pgfpathlineto{\pgfqpoint{1.491455in}{0.558460in}}%
\pgfpathlineto{\pgfqpoint{1.563355in}{0.558460in}}%
\pgfpathlineto{\pgfqpoint{1.563355in}{0.554127in}}%
\pgfpathlineto{\pgfqpoint{1.635256in}{0.554127in}}%
\pgfpathlineto{\pgfqpoint{1.635256in}{0.525153in}}%
\pgfpathlineto{\pgfqpoint{1.707156in}{0.525153in}}%
\pgfpathlineto{\pgfqpoint{1.707156in}{0.521607in}}%
\pgfpathlineto{\pgfqpoint{1.779057in}{0.521607in}}%
\pgfpathlineto{\pgfqpoint{1.779057in}{0.514356in}}%
\pgfpathlineto{\pgfqpoint{1.850957in}{0.514356in}}%
\pgfpathlineto{\pgfqpoint{1.850957in}{0.511425in}}%
\pgfpathlineto{\pgfqpoint{1.922858in}{0.511425in}}%
\pgfpathlineto{\pgfqpoint{1.922858in}{0.499652in}}%
\pgfpathlineto{\pgfqpoint{1.994758in}{0.499652in}}%
\pgfpathlineto{\pgfqpoint{1.994758in}{0.491679in}}%
\pgfpathlineto{\pgfqpoint{2.066659in}{0.491679in}}%
\pgfpathlineto{\pgfqpoint{2.066659in}{0.482718in}}%
\pgfpathlineto{\pgfqpoint{2.138559in}{0.482718in}}%
\pgfpathlineto{\pgfqpoint{2.138559in}{0.478458in}}%
\pgfpathlineto{\pgfqpoint{2.210460in}{0.478458in}}%
\pgfpathlineto{\pgfqpoint{2.210460in}{0.463744in}}%
\pgfpathlineto{\pgfqpoint{2.282360in}{0.463744in}}%
\pgfpathlineto{\pgfqpoint{2.282360in}{0.442177in}}%
\pgfpathlineto{\pgfqpoint{2.498062in}{0.442177in}}%
\pgfpathlineto{\pgfqpoint{2.498062in}{0.442177in}}%
\pgfpathlineto{\pgfqpoint{2.677813in}{0.442177in}}%
\pgfpathlineto{\pgfqpoint{2.677813in}{0.442177in}}%
\pgfpathlineto{\pgfqpoint{2.677813in}{0.442177in}}%
\pgfpathlineto{\pgfqpoint{2.498062in}{0.442177in}}%
\pgfpathlineto{\pgfqpoint{2.498062in}{0.442177in}}%
\pgfpathlineto{\pgfqpoint{2.282360in}{0.442177in}}%
\pgfpathlineto{\pgfqpoint{2.282360in}{0.482384in}}%
\pgfpathlineto{\pgfqpoint{2.210460in}{0.482384in}}%
\pgfpathlineto{\pgfqpoint{2.210460in}{0.496609in}}%
\pgfpathlineto{\pgfqpoint{2.138559in}{0.496609in}}%
\pgfpathlineto{\pgfqpoint{2.138559in}{0.499584in}}%
\pgfpathlineto{\pgfqpoint{2.066659in}{0.499584in}}%
\pgfpathlineto{\pgfqpoint{2.066659in}{0.506761in}}%
\pgfpathlineto{\pgfqpoint{1.994758in}{0.506761in}}%
\pgfpathlineto{\pgfqpoint{1.994758in}{0.518267in}}%
\pgfpathlineto{\pgfqpoint{1.922858in}{0.518267in}}%
\pgfpathlineto{\pgfqpoint{1.922858in}{0.531167in}}%
\pgfpathlineto{\pgfqpoint{1.850957in}{0.531167in}}%
\pgfpathlineto{\pgfqpoint{1.850957in}{0.536583in}}%
\pgfpathlineto{\pgfqpoint{1.779057in}{0.536583in}}%
\pgfpathlineto{\pgfqpoint{1.779057in}{0.544172in}}%
\pgfpathlineto{\pgfqpoint{1.707156in}{0.544172in}}%
\pgfpathlineto{\pgfqpoint{1.707156in}{0.546563in}}%
\pgfpathlineto{\pgfqpoint{1.635256in}{0.546563in}}%
\pgfpathlineto{\pgfqpoint{1.635256in}{0.576766in}}%
\pgfpathlineto{\pgfqpoint{1.563355in}{0.576766in}}%
\pgfpathlineto{\pgfqpoint{1.563355in}{0.582265in}}%
\pgfpathlineto{\pgfqpoint{1.491455in}{0.582265in}}%
\pgfpathlineto{\pgfqpoint{1.491455in}{0.587887in}}%
\pgfpathlineto{\pgfqpoint{1.419554in}{0.587887in}}%
\pgfpathlineto{\pgfqpoint{1.419554in}{0.595969in}}%
\pgfpathlineto{\pgfqpoint{1.347654in}{0.595969in}}%
\pgfpathlineto{\pgfqpoint{1.347654in}{0.601815in}}%
\pgfpathlineto{\pgfqpoint{1.275753in}{0.601815in}}%
\pgfpathlineto{\pgfqpoint{1.275753in}{0.608667in}}%
\pgfpathlineto{\pgfqpoint{1.203853in}{0.608667in}}%
\pgfpathlineto{\pgfqpoint{1.203853in}{0.618011in}}%
\pgfpathlineto{\pgfqpoint{1.131952in}{0.618011in}}%
\pgfpathlineto{\pgfqpoint{1.131952in}{0.627318in}}%
\pgfpathlineto{\pgfqpoint{1.060052in}{0.627318in}}%
\pgfpathlineto{\pgfqpoint{1.060052in}{0.721736in}}%
\pgfpathlineto{\pgfqpoint{0.988151in}{0.721736in}}%
\pgfpathlineto{\pgfqpoint{0.988151in}{0.731885in}}%
\pgfpathlineto{\pgfqpoint{0.916251in}{0.731885in}}%
\pgfpathlineto{\pgfqpoint{0.916251in}{0.902568in}}%
\pgfpathlineto{\pgfqpoint{0.844350in}{0.902568in}}%
\pgfpathlineto{\pgfqpoint{0.844350in}{0.929834in}}%
\pgfpathlineto{\pgfqpoint{0.772450in}{0.929834in}}%
\pgfpathlineto{\pgfqpoint{0.772450in}{0.957905in}}%
\pgfpathlineto{\pgfqpoint{0.700549in}{0.957905in}}%
\pgfpathlineto{\pgfqpoint{0.700549in}{0.993718in}}%
\pgfpathlineto{\pgfqpoint{0.628649in}{0.993718in}}%
\pgfpathlineto{\pgfqpoint{0.628649in}{1.033925in}}%
\pgfpathlineto{\pgfqpoint{0.556748in}{1.033925in}}%
\pgfpathlineto{\pgfqpoint{0.556748in}{1.076194in}}%
\pgfpathlineto{\pgfqpoint{0.520798in}{1.076194in}}%
\pgfpathlineto{\pgfqpoint{0.520798in}{1.076194in}}%
\pgfpathclose%
\pgfusepath{fill}%
\end{pgfscope}%
\begin{pgfscope}%
\pgfpathrectangle{\pgfqpoint{0.520798in}{0.442177in}}{\pgfqpoint{0.719005in}{0.871884in}}%
\pgfusepath{clip}%
\pgfsetroundcap%
\pgfsetroundjoin%
\pgfsetlinewidth{1.003750pt}%
\definecolor{currentstroke}{rgb}{0.007843,0.619608,0.450980}%
\pgfsetstrokecolor{currentstroke}%
\pgfsetdash{}{0pt}%
\pgfpathmoveto{\pgfqpoint{0.520798in}{1.064090in}}%
\pgfpathlineto{\pgfqpoint{0.556748in}{1.064090in}}%
\pgfpathlineto{\pgfqpoint{0.556748in}{1.020589in}}%
\pgfpathlineto{\pgfqpoint{0.628649in}{1.020589in}}%
\pgfpathlineto{\pgfqpoint{0.628649in}{0.978571in}}%
\pgfpathlineto{\pgfqpoint{0.700549in}{0.978571in}}%
\pgfpathlineto{\pgfqpoint{0.700549in}{0.942583in}}%
\pgfpathlineto{\pgfqpoint{0.772450in}{0.942583in}}%
\pgfpathlineto{\pgfqpoint{0.772450in}{0.911603in}}%
\pgfpathlineto{\pgfqpoint{0.844350in}{0.911603in}}%
\pgfpathlineto{\pgfqpoint{0.844350in}{0.882293in}}%
\pgfpathlineto{\pgfqpoint{0.916251in}{0.882293in}}%
\pgfpathlineto{\pgfqpoint{0.916251in}{0.854931in}}%
\pgfpathlineto{\pgfqpoint{0.988151in}{0.854931in}}%
\pgfpathlineto{\pgfqpoint{0.988151in}{0.828960in}}%
\pgfpathlineto{\pgfqpoint{1.060052in}{0.828960in}}%
\pgfpathlineto{\pgfqpoint{1.060052in}{0.805864in}}%
\pgfpathlineto{\pgfqpoint{1.131952in}{0.805864in}}%
\pgfpathlineto{\pgfqpoint{1.131952in}{0.784716in}}%
\pgfpathlineto{\pgfqpoint{1.203853in}{0.784716in}}%
\pgfpathlineto{\pgfqpoint{1.203853in}{0.766722in}}%
\pgfpathlineto{\pgfqpoint{1.241470in}{0.766722in}}%
\pgfusepath{stroke}%
\end{pgfscope}%
\begin{pgfscope}%
\pgfpathrectangle{\pgfqpoint{0.520798in}{0.442177in}}{\pgfqpoint{0.719005in}{0.871884in}}%
\pgfusepath{clip}%
\pgfsetbuttcap%
\pgfsetroundjoin%
\definecolor{currentfill}{rgb}{0.007843,0.619608,0.450980}%
\pgfsetfillcolor{currentfill}%
\pgfsetfillopacity{0.500000}%
\pgfsetlinewidth{0.000000pt}%
\definecolor{currentstroke}{rgb}{0.007843,0.619608,0.450980}%
\pgfsetstrokecolor{currentstroke}%
\pgfsetstrokeopacity{0.500000}%
\pgfsetdash{}{0pt}%
\pgfpathmoveto{\pgfqpoint{0.520798in}{1.076194in}}%
\pgfpathlineto{\pgfqpoint{0.520798in}{1.051986in}}%
\pgfpathlineto{\pgfqpoint{0.556748in}{1.051986in}}%
\pgfpathlineto{\pgfqpoint{0.556748in}{1.007252in}}%
\pgfpathlineto{\pgfqpoint{0.628649in}{1.007252in}}%
\pgfpathlineto{\pgfqpoint{0.628649in}{0.963425in}}%
\pgfpathlineto{\pgfqpoint{0.700549in}{0.963425in}}%
\pgfpathlineto{\pgfqpoint{0.700549in}{0.927260in}}%
\pgfpathlineto{\pgfqpoint{0.772450in}{0.927260in}}%
\pgfpathlineto{\pgfqpoint{0.772450in}{0.893372in}}%
\pgfpathlineto{\pgfqpoint{0.844350in}{0.893372in}}%
\pgfpathlineto{\pgfqpoint{0.844350in}{0.862018in}}%
\pgfpathlineto{\pgfqpoint{0.916251in}{0.862018in}}%
\pgfpathlineto{\pgfqpoint{0.916251in}{0.836188in}}%
\pgfpathlineto{\pgfqpoint{0.988151in}{0.836188in}}%
\pgfpathlineto{\pgfqpoint{0.988151in}{0.811123in}}%
\pgfpathlineto{\pgfqpoint{1.060052in}{0.811123in}}%
\pgfpathlineto{\pgfqpoint{1.060052in}{0.790630in}}%
\pgfpathlineto{\pgfqpoint{1.131952in}{0.790630in}}%
\pgfpathlineto{\pgfqpoint{1.131952in}{0.769224in}}%
\pgfpathlineto{\pgfqpoint{1.203853in}{0.769224in}}%
\pgfpathlineto{\pgfqpoint{1.203853in}{0.750303in}}%
\pgfpathlineto{\pgfqpoint{1.275753in}{0.750303in}}%
\pgfpathlineto{\pgfqpoint{1.275753in}{0.735638in}}%
\pgfpathlineto{\pgfqpoint{1.347654in}{0.735638in}}%
\pgfpathlineto{\pgfqpoint{1.347654in}{0.694707in}}%
\pgfpathlineto{\pgfqpoint{1.419554in}{0.694707in}}%
\pgfpathlineto{\pgfqpoint{1.419554in}{0.683337in}}%
\pgfpathlineto{\pgfqpoint{1.491455in}{0.683337in}}%
\pgfpathlineto{\pgfqpoint{1.491455in}{0.672293in}}%
\pgfpathlineto{\pgfqpoint{1.563355in}{0.672293in}}%
\pgfpathlineto{\pgfqpoint{1.563355in}{0.661498in}}%
\pgfpathlineto{\pgfqpoint{1.635256in}{0.661498in}}%
\pgfpathlineto{\pgfqpoint{1.635256in}{0.597071in}}%
\pgfpathlineto{\pgfqpoint{1.707156in}{0.597071in}}%
\pgfpathlineto{\pgfqpoint{1.707156in}{0.589497in}}%
\pgfpathlineto{\pgfqpoint{1.779057in}{0.589497in}}%
\pgfpathlineto{\pgfqpoint{1.779057in}{0.582630in}}%
\pgfpathlineto{\pgfqpoint{1.850957in}{0.582630in}}%
\pgfpathlineto{\pgfqpoint{1.850957in}{0.574162in}}%
\pgfpathlineto{\pgfqpoint{1.922858in}{0.574162in}}%
\pgfpathlineto{\pgfqpoint{1.922858in}{0.570308in}}%
\pgfpathlineto{\pgfqpoint{1.994758in}{0.570308in}}%
\pgfpathlineto{\pgfqpoint{1.994758in}{0.564714in}}%
\pgfpathlineto{\pgfqpoint{2.066659in}{0.564714in}}%
\pgfpathlineto{\pgfqpoint{2.066659in}{0.560629in}}%
\pgfpathlineto{\pgfqpoint{2.138559in}{0.560629in}}%
\pgfpathlineto{\pgfqpoint{2.138559in}{0.557329in}}%
\pgfpathlineto{\pgfqpoint{2.210460in}{0.557329in}}%
\pgfpathlineto{\pgfqpoint{2.210460in}{0.553176in}}%
\pgfpathlineto{\pgfqpoint{2.282360in}{0.553176in}}%
\pgfpathlineto{\pgfqpoint{2.282360in}{0.549181in}}%
\pgfpathlineto{\pgfqpoint{2.354261in}{0.549181in}}%
\pgfpathlineto{\pgfqpoint{2.354261in}{0.545526in}}%
\pgfpathlineto{\pgfqpoint{2.426161in}{0.545526in}}%
\pgfpathlineto{\pgfqpoint{2.426161in}{0.541576in}}%
\pgfpathlineto{\pgfqpoint{2.498062in}{0.541576in}}%
\pgfpathlineto{\pgfqpoint{2.498062in}{0.539014in}}%
\pgfpathlineto{\pgfqpoint{2.569962in}{0.539014in}}%
\pgfpathlineto{\pgfqpoint{2.569962in}{0.536144in}}%
\pgfpathlineto{\pgfqpoint{2.641863in}{0.536144in}}%
\pgfpathlineto{\pgfqpoint{2.641863in}{0.532787in}}%
\pgfpathlineto{\pgfqpoint{2.677813in}{0.532787in}}%
\pgfpathlineto{\pgfqpoint{2.677813in}{0.554511in}}%
\pgfpathlineto{\pgfqpoint{2.677813in}{0.554511in}}%
\pgfpathlineto{\pgfqpoint{2.641863in}{0.554511in}}%
\pgfpathlineto{\pgfqpoint{2.641863in}{0.557276in}}%
\pgfpathlineto{\pgfqpoint{2.569962in}{0.557276in}}%
\pgfpathlineto{\pgfqpoint{2.569962in}{0.560714in}}%
\pgfpathlineto{\pgfqpoint{2.498062in}{0.560714in}}%
\pgfpathlineto{\pgfqpoint{2.498062in}{0.563716in}}%
\pgfpathlineto{\pgfqpoint{2.426161in}{0.563716in}}%
\pgfpathlineto{\pgfqpoint{2.426161in}{0.567187in}}%
\pgfpathlineto{\pgfqpoint{2.354261in}{0.567187in}}%
\pgfpathlineto{\pgfqpoint{2.354261in}{0.571323in}}%
\pgfpathlineto{\pgfqpoint{2.282360in}{0.571323in}}%
\pgfpathlineto{\pgfqpoint{2.282360in}{0.576047in}}%
\pgfpathlineto{\pgfqpoint{2.210460in}{0.576047in}}%
\pgfpathlineto{\pgfqpoint{2.210460in}{0.579871in}}%
\pgfpathlineto{\pgfqpoint{2.138559in}{0.579871in}}%
\pgfpathlineto{\pgfqpoint{2.138559in}{0.584919in}}%
\pgfpathlineto{\pgfqpoint{2.066659in}{0.584919in}}%
\pgfpathlineto{\pgfqpoint{2.066659in}{0.590294in}}%
\pgfpathlineto{\pgfqpoint{1.994758in}{0.590294in}}%
\pgfpathlineto{\pgfqpoint{1.994758in}{0.598243in}}%
\pgfpathlineto{\pgfqpoint{1.922858in}{0.598243in}}%
\pgfpathlineto{\pgfqpoint{1.922858in}{0.603478in}}%
\pgfpathlineto{\pgfqpoint{1.850957in}{0.603478in}}%
\pgfpathlineto{\pgfqpoint{1.850957in}{0.609852in}}%
\pgfpathlineto{\pgfqpoint{1.779057in}{0.609852in}}%
\pgfpathlineto{\pgfqpoint{1.779057in}{0.618011in}}%
\pgfpathlineto{\pgfqpoint{1.707156in}{0.618011in}}%
\pgfpathlineto{\pgfqpoint{1.707156in}{0.627318in}}%
\pgfpathlineto{\pgfqpoint{1.635256in}{0.627318in}}%
\pgfpathlineto{\pgfqpoint{1.635256in}{0.695899in}}%
\pgfpathlineto{\pgfqpoint{1.563355in}{0.695899in}}%
\pgfpathlineto{\pgfqpoint{1.563355in}{0.709406in}}%
\pgfpathlineto{\pgfqpoint{1.491455in}{0.709406in}}%
\pgfpathlineto{\pgfqpoint{1.491455in}{0.721736in}}%
\pgfpathlineto{\pgfqpoint{1.419554in}{0.721736in}}%
\pgfpathlineto{\pgfqpoint{1.419554in}{0.731885in}}%
\pgfpathlineto{\pgfqpoint{1.347654in}{0.731885in}}%
\pgfpathlineto{\pgfqpoint{1.347654in}{0.766269in}}%
\pgfpathlineto{\pgfqpoint{1.275753in}{0.766269in}}%
\pgfpathlineto{\pgfqpoint{1.275753in}{0.783141in}}%
\pgfpathlineto{\pgfqpoint{1.203853in}{0.783141in}}%
\pgfpathlineto{\pgfqpoint{1.203853in}{0.800208in}}%
\pgfpathlineto{\pgfqpoint{1.131952in}{0.800208in}}%
\pgfpathlineto{\pgfqpoint{1.131952in}{0.821098in}}%
\pgfpathlineto{\pgfqpoint{1.060052in}{0.821098in}}%
\pgfpathlineto{\pgfqpoint{1.060052in}{0.846796in}}%
\pgfpathlineto{\pgfqpoint{0.988151in}{0.846796in}}%
\pgfpathlineto{\pgfqpoint{0.988151in}{0.873673in}}%
\pgfpathlineto{\pgfqpoint{0.916251in}{0.873673in}}%
\pgfpathlineto{\pgfqpoint{0.916251in}{0.902568in}}%
\pgfpathlineto{\pgfqpoint{0.844350in}{0.902568in}}%
\pgfpathlineto{\pgfqpoint{0.844350in}{0.929834in}}%
\pgfpathlineto{\pgfqpoint{0.772450in}{0.929834in}}%
\pgfpathlineto{\pgfqpoint{0.772450in}{0.957905in}}%
\pgfpathlineto{\pgfqpoint{0.700549in}{0.957905in}}%
\pgfpathlineto{\pgfqpoint{0.700549in}{0.993718in}}%
\pgfpathlineto{\pgfqpoint{0.628649in}{0.993718in}}%
\pgfpathlineto{\pgfqpoint{0.628649in}{1.033925in}}%
\pgfpathlineto{\pgfqpoint{0.556748in}{1.033925in}}%
\pgfpathlineto{\pgfqpoint{0.556748in}{1.076194in}}%
\pgfpathlineto{\pgfqpoint{0.520798in}{1.076194in}}%
\pgfpathlineto{\pgfqpoint{0.520798in}{1.076194in}}%
\pgfpathclose%
\pgfusepath{fill}%
\end{pgfscope}%
\begin{pgfscope}%
\definecolor{textcolor}{rgb}{0.150000,0.150000,0.150000}%
\pgfsetstrokecolor{textcolor}%
\pgfsetfillcolor{textcolor}%
\pgftext[x=0.880300in,y=1.397395in,,base]{\color{textcolor}\rmfamily\fontsize{9.000000}{10.800000}\selectfont \(\displaystyle c_X^-=1\)}%
\end{pgfscope}%
\begin{pgfscope}%
\pgfsetbuttcap%
\pgfsetmiterjoin%
\definecolor{currentfill}{rgb}{1.000000,1.000000,1.000000}%
\pgfsetfillcolor{currentfill}%
\pgfsetfillopacity{0.800000}%
\pgfsetlinewidth{1.003750pt}%
\definecolor{currentstroke}{rgb}{0.800000,0.800000,0.800000}%
\pgfsetstrokecolor{currentstroke}%
\pgfsetstrokeopacity{0.800000}%
\pgfsetdash{}{0pt}%
\pgfpathmoveto{\pgfqpoint{0.785843in}{0.638103in}}%
\pgfpathlineto{\pgfqpoint{1.191192in}{0.638103in}}%
\pgfpathlineto{\pgfqpoint{1.191192in}{1.265450in}}%
\pgfpathlineto{\pgfqpoint{0.785843in}{1.265450in}}%
\pgfpathlineto{\pgfqpoint{0.785843in}{0.638103in}}%
\pgfpathclose%
\pgfusepath{stroke,fill}%
\end{pgfscope}%
\begin{pgfscope}%
\definecolor{textcolor}{rgb}{0.150000,0.150000,0.150000}%
\pgfsetstrokecolor{textcolor}%
\pgfsetfillcolor{textcolor}%
\pgftext[x=0.912291in,y=1.113275in,left,base]{\color{textcolor}\rmfamily\fontsize{9.000000}{10.800000}\selectfont \(\displaystyle c_X^+\)}%
\end{pgfscope}%
\begin{pgfscope}%
\pgfsetroundcap%
\pgfsetroundjoin%
\pgfsetlinewidth{1.003750pt}%
\definecolor{currentstroke}{rgb}{0.003922,0.450980,0.698039}%
\pgfsetstrokecolor{currentstroke}%
\pgfsetdash{}{0pt}%
\pgfpathmoveto{\pgfqpoint{0.824732in}{1.001052in}}%
\pgfpathlineto{\pgfqpoint{0.873343in}{1.001052in}}%
\pgfpathlineto{\pgfqpoint{0.873343in}{1.001052in}}%
\pgfpathlineto{\pgfqpoint{0.970565in}{1.001052in}}%
\pgfpathlineto{\pgfqpoint{0.970565in}{1.001052in}}%
\pgfpathlineto{\pgfqpoint{1.019176in}{1.001052in}}%
\pgfusepath{stroke}%
\end{pgfscope}%
\begin{pgfscope}%
\definecolor{textcolor}{rgb}{0.150000,0.150000,0.150000}%
\pgfsetstrokecolor{textcolor}%
\pgfsetfillcolor{textcolor}%
\pgftext[x=1.096954in,y=0.967024in,left,base]{\color{textcolor}\rmfamily\fontsize{7.000000}{8.400000}\selectfont 1}%
\end{pgfscope}%
\begin{pgfscope}%
\pgfsetroundcap%
\pgfsetroundjoin%
\pgfsetlinewidth{1.003750pt}%
\definecolor{currentstroke}{rgb}{0.870588,0.560784,0.019608}%
\pgfsetstrokecolor{currentstroke}%
\pgfsetdash{}{0pt}%
\pgfpathmoveto{\pgfqpoint{0.824732in}{0.865486in}}%
\pgfpathlineto{\pgfqpoint{0.873343in}{0.865486in}}%
\pgfpathlineto{\pgfqpoint{0.873343in}{0.865486in}}%
\pgfpathlineto{\pgfqpoint{0.970565in}{0.865486in}}%
\pgfpathlineto{\pgfqpoint{0.970565in}{0.865486in}}%
\pgfpathlineto{\pgfqpoint{1.019176in}{0.865486in}}%
\pgfusepath{stroke}%
\end{pgfscope}%
\begin{pgfscope}%
\definecolor{textcolor}{rgb}{0.150000,0.150000,0.150000}%
\pgfsetstrokecolor{textcolor}%
\pgfsetfillcolor{textcolor}%
\pgftext[x=1.096954in,y=0.831458in,left,base]{\color{textcolor}\rmfamily\fontsize{7.000000}{8.400000}\selectfont 2}%
\end{pgfscope}%
\begin{pgfscope}%
\pgfsetroundcap%
\pgfsetroundjoin%
\pgfsetlinewidth{1.003750pt}%
\definecolor{currentstroke}{rgb}{0.007843,0.619608,0.450980}%
\pgfsetstrokecolor{currentstroke}%
\pgfsetdash{}{0pt}%
\pgfpathmoveto{\pgfqpoint{0.824732in}{0.729919in}}%
\pgfpathlineto{\pgfqpoint{0.873343in}{0.729919in}}%
\pgfpathlineto{\pgfqpoint{0.873343in}{0.729919in}}%
\pgfpathlineto{\pgfqpoint{0.970565in}{0.729919in}}%
\pgfpathlineto{\pgfqpoint{0.970565in}{0.729919in}}%
\pgfpathlineto{\pgfqpoint{1.019176in}{0.729919in}}%
\pgfusepath{stroke}%
\end{pgfscope}%
\begin{pgfscope}%
\definecolor{textcolor}{rgb}{0.150000,0.150000,0.150000}%
\pgfsetstrokecolor{textcolor}%
\pgfsetfillcolor{textcolor}%
\pgftext[x=1.096954in,y=0.695892in,left,base]{\color{textcolor}\rmfamily\fontsize{7.000000}{8.400000}\selectfont 4}%
\end{pgfscope}%
\end{pgfpicture}%
\makeatother%
\endgroup%

%% file: figures/experiments/graphs/sparse_smoothing/nodes_attributes/node_classification-Citeseer-GAT-hidden=8-p_adj_plus=0.0-p_adj_minus=0.0-p_att_plus=0.001-p_att_minus=0.8-multi_class_cert-A.pgf
\begingroup%
\makeatletter%
\begin{pgfpicture}%
\pgfpathrectangle{\pgfpointorigin}{\pgfqpoint{1.375000in}{1.581250in}}%
\pgfusepath{use as bounding box, clip}%
\begin{pgfscope}%
\pgfsetbuttcap%
\pgfsetmiterjoin%
\definecolor{currentfill}{rgb}{1.000000,1.000000,1.000000}%
\pgfsetfillcolor{currentfill}%
\pgfsetlinewidth{0.000000pt}%
\definecolor{currentstroke}{rgb}{1.000000,1.000000,1.000000}%
\pgfsetstrokecolor{currentstroke}%
\pgfsetdash{}{0pt}%
\pgfpathmoveto{\pgfqpoint{0.000000in}{0.000000in}}%
\pgfpathlineto{\pgfqpoint{1.375000in}{0.000000in}}%
\pgfpathlineto{\pgfqpoint{1.375000in}{1.581250in}}%
\pgfpathlineto{\pgfqpoint{0.000000in}{1.581250in}}%
\pgfpathlineto{\pgfqpoint{0.000000in}{0.000000in}}%
\pgfpathclose%
\pgfusepath{fill}%
\end{pgfscope}%
\begin{pgfscope}%
\pgfsetbuttcap%
\pgfsetmiterjoin%
\definecolor{currentfill}{rgb}{1.000000,1.000000,1.000000}%
\pgfsetfillcolor{currentfill}%
\pgfsetlinewidth{0.000000pt}%
\definecolor{currentstroke}{rgb}{0.000000,0.000000,0.000000}%
\pgfsetstrokecolor{currentstroke}%
\pgfsetstrokeopacity{0.000000}%
\pgfsetdash{}{0pt}%
\pgfpathmoveto{\pgfqpoint{0.520798in}{0.442177in}}%
\pgfpathlineto{\pgfqpoint{1.239803in}{0.442177in}}%
\pgfpathlineto{\pgfqpoint{1.239803in}{1.314061in}}%
\pgfpathlineto{\pgfqpoint{0.520798in}{1.314061in}}%
\pgfpathlineto{\pgfqpoint{0.520798in}{0.442177in}}%
\pgfpathclose%
\pgfusepath{fill}%
\end{pgfscope}%
\begin{pgfscope}%
\pgfpathrectangle{\pgfqpoint{0.520798in}{0.442177in}}{\pgfqpoint{0.719005in}{0.871884in}}%
\pgfusepath{clip}%
\pgfsetroundcap%
\pgfsetroundjoin%
\pgfsetlinewidth{0.501875pt}%
\definecolor{currentstroke}{rgb}{0.800000,0.800000,0.800000}%
\pgfsetstrokecolor{currentstroke}%
\pgfsetdash{}{0pt}%
\pgfpathmoveto{\pgfqpoint{0.520798in}{0.442177in}}%
\pgfpathlineto{\pgfqpoint{0.520798in}{1.314061in}}%
\pgfusepath{stroke}%
\end{pgfscope}%
\begin{pgfscope}%
\definecolor{textcolor}{rgb}{0.150000,0.150000,0.150000}%
\pgfsetstrokecolor{textcolor}%
\pgfsetfillcolor{textcolor}%
\pgftext[x=0.520798in,y=0.351899in,,top]{\color{textcolor}\rmfamily\fontsize{8.000000}{9.600000}\selectfont \(\displaystyle {0}\)}%
\end{pgfscope}%
\begin{pgfscope}%
\pgfpathrectangle{\pgfqpoint{0.520798in}{0.442177in}}{\pgfqpoint{0.719005in}{0.871884in}}%
\pgfusepath{clip}%
\pgfsetroundcap%
\pgfsetroundjoin%
\pgfsetlinewidth{0.501875pt}%
\definecolor{currentstroke}{rgb}{0.800000,0.800000,0.800000}%
\pgfsetstrokecolor{currentstroke}%
\pgfsetdash{}{0pt}%
\pgfpathmoveto{\pgfqpoint{0.880300in}{0.442177in}}%
\pgfpathlineto{\pgfqpoint{0.880300in}{1.314061in}}%
\pgfusepath{stroke}%
\end{pgfscope}%
\begin{pgfscope}%
\definecolor{textcolor}{rgb}{0.150000,0.150000,0.150000}%
\pgfsetstrokecolor{textcolor}%
\pgfsetfillcolor{textcolor}%
\pgftext[x=0.880300in,y=0.351899in,,top]{\color{textcolor}\rmfamily\fontsize{8.000000}{9.600000}\selectfont \(\displaystyle {5}\)}%
\end{pgfscope}%
\begin{pgfscope}%
\pgfpathrectangle{\pgfqpoint{0.520798in}{0.442177in}}{\pgfqpoint{0.719005in}{0.871884in}}%
\pgfusepath{clip}%
\pgfsetroundcap%
\pgfsetroundjoin%
\pgfsetlinewidth{0.501875pt}%
\definecolor{currentstroke}{rgb}{0.800000,0.800000,0.800000}%
\pgfsetstrokecolor{currentstroke}%
\pgfsetdash{}{0pt}%
\pgfpathmoveto{\pgfqpoint{1.239803in}{0.442177in}}%
\pgfpathlineto{\pgfqpoint{1.239803in}{1.314061in}}%
\pgfusepath{stroke}%
\end{pgfscope}%
\begin{pgfscope}%
\definecolor{textcolor}{rgb}{0.150000,0.150000,0.150000}%
\pgfsetstrokecolor{textcolor}%
\pgfsetfillcolor{textcolor}%
\pgftext[x=1.239803in,y=0.351899in,,top]{\color{textcolor}\rmfamily\fontsize{8.000000}{9.600000}\selectfont \(\displaystyle {10}\)}%
\end{pgfscope}%
\begin{pgfscope}%
\definecolor{textcolor}{rgb}{0.150000,0.150000,0.150000}%
\pgfsetstrokecolor{textcolor}%
\pgfsetfillcolor{textcolor}%
\pgftext[x=0.880300in,y=0.198219in,,top]{\color{textcolor}\rmfamily\fontsize{10.000000}{12.000000}\selectfont Edit distance \(\displaystyle \epsilon\)}%
\end{pgfscope}%
\begin{pgfscope}%
\pgfpathrectangle{\pgfqpoint{0.520798in}{0.442177in}}{\pgfqpoint{0.719005in}{0.871884in}}%
\pgfusepath{clip}%
\pgfsetroundcap%
\pgfsetroundjoin%
\pgfsetlinewidth{0.501875pt}%
\definecolor{currentstroke}{rgb}{0.800000,0.800000,0.800000}%
\pgfsetstrokecolor{currentstroke}%
\pgfsetdash{}{0pt}%
\pgfpathmoveto{\pgfqpoint{0.520798in}{0.442177in}}%
\pgfpathlineto{\pgfqpoint{1.239803in}{0.442177in}}%
\pgfusepath{stroke}%
\end{pgfscope}%
\begin{pgfscope}%
\definecolor{textcolor}{rgb}{0.150000,0.150000,0.150000}%
\pgfsetstrokecolor{textcolor}%
\pgfsetfillcolor{textcolor}%
\pgftext[x=0.273151in, y=0.403915in, left, base]{\color{textcolor}\rmfamily\fontsize{8.000000}{9.600000}\selectfont 0\%}%
\end{pgfscope}%
\begin{pgfscope}%
\pgfpathrectangle{\pgfqpoint{0.520798in}{0.442177in}}{\pgfqpoint{0.719005in}{0.871884in}}%
\pgfusepath{clip}%
\pgfsetroundcap%
\pgfsetroundjoin%
\pgfsetlinewidth{0.501875pt}%
\definecolor{currentstroke}{rgb}{0.800000,0.800000,0.800000}%
\pgfsetstrokecolor{currentstroke}%
\pgfsetdash{}{0pt}%
\pgfpathmoveto{\pgfqpoint{0.520798in}{0.660148in}}%
\pgfpathlineto{\pgfqpoint{1.239803in}{0.660148in}}%
\pgfusepath{stroke}%
\end{pgfscope}%
\begin{pgfscope}%
\definecolor{textcolor}{rgb}{0.150000,0.150000,0.150000}%
\pgfsetstrokecolor{textcolor}%
\pgfsetfillcolor{textcolor}%
\pgftext[x=0.214138in, y=0.621886in, left, base]{\color{textcolor}\rmfamily\fontsize{8.000000}{9.600000}\selectfont 25\%}%
\end{pgfscope}%
\begin{pgfscope}%
\pgfpathrectangle{\pgfqpoint{0.520798in}{0.442177in}}{\pgfqpoint{0.719005in}{0.871884in}}%
\pgfusepath{clip}%
\pgfsetroundcap%
\pgfsetroundjoin%
\pgfsetlinewidth{0.501875pt}%
\definecolor{currentstroke}{rgb}{0.800000,0.800000,0.800000}%
\pgfsetstrokecolor{currentstroke}%
\pgfsetdash{}{0pt}%
\pgfpathmoveto{\pgfqpoint{0.520798in}{0.878119in}}%
\pgfpathlineto{\pgfqpoint{1.239803in}{0.878119in}}%
\pgfusepath{stroke}%
\end{pgfscope}%
\begin{pgfscope}%
\definecolor{textcolor}{rgb}{0.150000,0.150000,0.150000}%
\pgfsetstrokecolor{textcolor}%
\pgfsetfillcolor{textcolor}%
\pgftext[x=0.214138in, y=0.839857in, left, base]{\color{textcolor}\rmfamily\fontsize{8.000000}{9.600000}\selectfont 50\%}%
\end{pgfscope}%
\begin{pgfscope}%
\pgfpathrectangle{\pgfqpoint{0.520798in}{0.442177in}}{\pgfqpoint{0.719005in}{0.871884in}}%
\pgfusepath{clip}%
\pgfsetroundcap%
\pgfsetroundjoin%
\pgfsetlinewidth{0.501875pt}%
\definecolor{currentstroke}{rgb}{0.800000,0.800000,0.800000}%
\pgfsetstrokecolor{currentstroke}%
\pgfsetdash{}{0pt}%
\pgfpathmoveto{\pgfqpoint{0.520798in}{1.096090in}}%
\pgfpathlineto{\pgfqpoint{1.239803in}{1.096090in}}%
\pgfusepath{stroke}%
\end{pgfscope}%
\begin{pgfscope}%
\definecolor{textcolor}{rgb}{0.150000,0.150000,0.150000}%
\pgfsetstrokecolor{textcolor}%
\pgfsetfillcolor{textcolor}%
\pgftext[x=0.214138in, y=1.057828in, left, base]{\color{textcolor}\rmfamily\fontsize{8.000000}{9.600000}\selectfont 75\%}%
\end{pgfscope}%
\begin{pgfscope}%
\pgfpathrectangle{\pgfqpoint{0.520798in}{0.442177in}}{\pgfqpoint{0.719005in}{0.871884in}}%
\pgfusepath{clip}%
\pgfsetroundcap%
\pgfsetroundjoin%
\pgfsetlinewidth{0.501875pt}%
\definecolor{currentstroke}{rgb}{0.800000,0.800000,0.800000}%
\pgfsetstrokecolor{currentstroke}%
\pgfsetdash{}{0pt}%
\pgfpathmoveto{\pgfqpoint{0.520798in}{1.314061in}}%
\pgfpathlineto{\pgfqpoint{1.239803in}{1.314061in}}%
\pgfusepath{stroke}%
\end{pgfscope}%
\begin{pgfscope}%
\definecolor{textcolor}{rgb}{0.150000,0.150000,0.150000}%
\pgfsetstrokecolor{textcolor}%
\pgfsetfillcolor{textcolor}%
\pgftext[x=0.155124in, y=1.275799in, left, base]{\color{textcolor}\rmfamily\fontsize{8.000000}{9.600000}\selectfont 100\%}%
\end{pgfscope}%
\begin{pgfscope}%
\definecolor{textcolor}{rgb}{0.150000,0.150000,0.150000}%
\pgfsetstrokecolor{textcolor}%
\pgfsetfillcolor{textcolor}%
\pgftext[x=0.099569in,y=0.878119in,,bottom,rotate=90.000000]{\color{textcolor}\rmfamily\fontsize{10.000000}{12.000000}\selectfont Cert. Acc.}%
\end{pgfscope}%
\begin{pgfscope}%
\pgfsetrectcap%
\pgfsetmiterjoin%
\pgfsetlinewidth{0.752812pt}%
\definecolor{currentstroke}{rgb}{0.700000,0.700000,0.700000}%
\pgfsetstrokecolor{currentstroke}%
\pgfsetdash{}{0pt}%
\pgfpathmoveto{\pgfqpoint{0.520798in}{0.442177in}}%
\pgfpathlineto{\pgfqpoint{0.520798in}{1.314061in}}%
\pgfusepath{stroke}%
\end{pgfscope}%
\begin{pgfscope}%
\pgfsetrectcap%
\pgfsetmiterjoin%
\pgfsetlinewidth{0.752812pt}%
\definecolor{currentstroke}{rgb}{0.700000,0.700000,0.700000}%
\pgfsetstrokecolor{currentstroke}%
\pgfsetdash{}{0pt}%
\pgfpathmoveto{\pgfqpoint{1.239803in}{0.442177in}}%
\pgfpathlineto{\pgfqpoint{1.239803in}{1.314061in}}%
\pgfusepath{stroke}%
\end{pgfscope}%
\begin{pgfscope}%
\pgfsetrectcap%
\pgfsetmiterjoin%
\pgfsetlinewidth{0.752812pt}%
\definecolor{currentstroke}{rgb}{0.700000,0.700000,0.700000}%
\pgfsetstrokecolor{currentstroke}%
\pgfsetdash{}{0pt}%
\pgfpathmoveto{\pgfqpoint{0.520798in}{0.442177in}}%
\pgfpathlineto{\pgfqpoint{1.239803in}{0.442177in}}%
\pgfusepath{stroke}%
\end{pgfscope}%
\begin{pgfscope}%
\pgfsetrectcap%
\pgfsetmiterjoin%
\pgfsetlinewidth{0.752812pt}%
\definecolor{currentstroke}{rgb}{0.700000,0.700000,0.700000}%
\pgfsetstrokecolor{currentstroke}%
\pgfsetdash{}{0pt}%
\pgfpathmoveto{\pgfqpoint{0.520798in}{1.314061in}}%
\pgfpathlineto{\pgfqpoint{1.239803in}{1.314061in}}%
\pgfusepath{stroke}%
\end{pgfscope}%
\begin{pgfscope}%
\pgfpathrectangle{\pgfqpoint{0.520798in}{0.442177in}}{\pgfqpoint{0.719005in}{0.871884in}}%
\pgfusepath{clip}%
\pgfsetroundcap%
\pgfsetroundjoin%
\pgfsetlinewidth{1.003750pt}%
\definecolor{currentstroke}{rgb}{0.003922,0.450980,0.698039}%
\pgfsetstrokecolor{currentstroke}%
\pgfsetdash{}{0pt}%
\pgfpathmoveto{\pgfqpoint{0.520798in}{1.064090in}}%
\pgfpathlineto{\pgfqpoint{0.556748in}{1.064090in}}%
\pgfpathlineto{\pgfqpoint{0.556748in}{1.008252in}}%
\pgfpathlineto{\pgfqpoint{0.628649in}{1.008252in}}%
\pgfpathlineto{\pgfqpoint{0.628649in}{0.921806in}}%
\pgfpathlineto{\pgfqpoint{0.700549in}{0.921806in}}%
\pgfpathlineto{\pgfqpoint{0.700549in}{0.713296in}}%
\pgfpathlineto{\pgfqpoint{0.772450in}{0.713296in}}%
\pgfpathlineto{\pgfqpoint{0.772450in}{0.612194in}}%
\pgfpathlineto{\pgfqpoint{0.844350in}{0.612194in}}%
\pgfpathlineto{\pgfqpoint{0.844350in}{0.600322in}}%
\pgfpathlineto{\pgfqpoint{0.916251in}{0.600322in}}%
\pgfpathlineto{\pgfqpoint{0.916251in}{0.592994in}}%
\pgfpathlineto{\pgfqpoint{0.988151in}{0.592994in}}%
\pgfpathlineto{\pgfqpoint{0.988151in}{0.581307in}}%
\pgfpathlineto{\pgfqpoint{1.060052in}{0.581307in}}%
\pgfpathlineto{\pgfqpoint{1.060052in}{0.536136in}}%
\pgfpathlineto{\pgfqpoint{1.131952in}{0.536136in}}%
\pgfpathlineto{\pgfqpoint{1.131952in}{0.527232in}}%
\pgfpathlineto{\pgfqpoint{1.203853in}{0.527232in}}%
\pgfpathlineto{\pgfqpoint{1.203853in}{0.509887in}}%
\pgfpathlineto{\pgfqpoint{1.241470in}{0.509887in}}%
\pgfusepath{stroke}%
\end{pgfscope}%
\begin{pgfscope}%
\pgfpathrectangle{\pgfqpoint{0.520798in}{0.442177in}}{\pgfqpoint{0.719005in}{0.871884in}}%
\pgfusepath{clip}%
\pgfsetbuttcap%
\pgfsetroundjoin%
\definecolor{currentfill}{rgb}{0.003922,0.450980,0.698039}%
\pgfsetfillcolor{currentfill}%
\pgfsetfillopacity{0.500000}%
\pgfsetlinewidth{0.000000pt}%
\definecolor{currentstroke}{rgb}{0.003922,0.450980,0.698039}%
\pgfsetstrokecolor{currentstroke}%
\pgfsetstrokeopacity{0.500000}%
\pgfsetdash{}{0pt}%
\pgfpathmoveto{\pgfqpoint{0.520798in}{1.076194in}}%
\pgfpathlineto{\pgfqpoint{0.520798in}{1.051986in}}%
\pgfpathlineto{\pgfqpoint{0.556748in}{1.051986in}}%
\pgfpathlineto{\pgfqpoint{0.556748in}{0.994752in}}%
\pgfpathlineto{\pgfqpoint{0.628649in}{0.994752in}}%
\pgfpathlineto{\pgfqpoint{0.628649in}{0.904877in}}%
\pgfpathlineto{\pgfqpoint{0.700549in}{0.904877in}}%
\pgfpathlineto{\pgfqpoint{0.700549in}{0.694707in}}%
\pgfpathlineto{\pgfqpoint{0.772450in}{0.694707in}}%
\pgfpathlineto{\pgfqpoint{0.772450in}{0.597071in}}%
\pgfpathlineto{\pgfqpoint{0.844350in}{0.597071in}}%
\pgfpathlineto{\pgfqpoint{0.844350in}{0.585768in}}%
\pgfpathlineto{\pgfqpoint{0.916251in}{0.585768in}}%
\pgfpathlineto{\pgfqpoint{0.916251in}{0.579635in}}%
\pgfpathlineto{\pgfqpoint{0.988151in}{0.579635in}}%
\pgfpathlineto{\pgfqpoint{0.988151in}{0.568385in}}%
\pgfpathlineto{\pgfqpoint{1.060052in}{0.568385in}}%
\pgfpathlineto{\pgfqpoint{1.060052in}{0.525437in}}%
\pgfpathlineto{\pgfqpoint{1.131952in}{0.525437in}}%
\pgfpathlineto{\pgfqpoint{1.131952in}{0.515930in}}%
\pgfpathlineto{\pgfqpoint{1.203853in}{0.515930in}}%
\pgfpathlineto{\pgfqpoint{1.203853in}{0.500124in}}%
\pgfpathlineto{\pgfqpoint{1.275753in}{0.500124in}}%
\pgfpathlineto{\pgfqpoint{1.275753in}{0.490095in}}%
\pgfpathlineto{\pgfqpoint{1.347654in}{0.490095in}}%
\pgfpathlineto{\pgfqpoint{1.347654in}{0.480913in}}%
\pgfpathlineto{\pgfqpoint{1.419554in}{0.480913in}}%
\pgfpathlineto{\pgfqpoint{1.419554in}{0.442177in}}%
\pgfpathlineto{\pgfqpoint{2.066659in}{0.442177in}}%
\pgfpathlineto{\pgfqpoint{2.066659in}{0.442177in}}%
\pgfpathlineto{\pgfqpoint{2.677813in}{0.442177in}}%
\pgfpathlineto{\pgfqpoint{2.677813in}{0.442177in}}%
\pgfpathlineto{\pgfqpoint{2.677813in}{0.442177in}}%
\pgfpathlineto{\pgfqpoint{2.066659in}{0.442177in}}%
\pgfpathlineto{\pgfqpoint{2.066659in}{0.442177in}}%
\pgfpathlineto{\pgfqpoint{1.419554in}{0.442177in}}%
\pgfpathlineto{\pgfqpoint{1.419554in}{0.498420in}}%
\pgfpathlineto{\pgfqpoint{1.347654in}{0.498420in}}%
\pgfpathlineto{\pgfqpoint{1.347654in}{0.503337in}}%
\pgfpathlineto{\pgfqpoint{1.275753in}{0.503337in}}%
\pgfpathlineto{\pgfqpoint{1.275753in}{0.519650in}}%
\pgfpathlineto{\pgfqpoint{1.203853in}{0.519650in}}%
\pgfpathlineto{\pgfqpoint{1.203853in}{0.538533in}}%
\pgfpathlineto{\pgfqpoint{1.131952in}{0.538533in}}%
\pgfpathlineto{\pgfqpoint{1.131952in}{0.546836in}}%
\pgfpathlineto{\pgfqpoint{1.060052in}{0.546836in}}%
\pgfpathlineto{\pgfqpoint{1.060052in}{0.594230in}}%
\pgfpathlineto{\pgfqpoint{0.988151in}{0.594230in}}%
\pgfpathlineto{\pgfqpoint{0.988151in}{0.606353in}}%
\pgfpathlineto{\pgfqpoint{0.916251in}{0.606353in}}%
\pgfpathlineto{\pgfqpoint{0.916251in}{0.614876in}}%
\pgfpathlineto{\pgfqpoint{0.844350in}{0.614876in}}%
\pgfpathlineto{\pgfqpoint{0.844350in}{0.627318in}}%
\pgfpathlineto{\pgfqpoint{0.772450in}{0.627318in}}%
\pgfpathlineto{\pgfqpoint{0.772450in}{0.731885in}}%
\pgfpathlineto{\pgfqpoint{0.700549in}{0.731885in}}%
\pgfpathlineto{\pgfqpoint{0.700549in}{0.938736in}}%
\pgfpathlineto{\pgfqpoint{0.628649in}{0.938736in}}%
\pgfpathlineto{\pgfqpoint{0.628649in}{1.021753in}}%
\pgfpathlineto{\pgfqpoint{0.556748in}{1.021753in}}%
\pgfpathlineto{\pgfqpoint{0.556748in}{1.076194in}}%
\pgfpathlineto{\pgfqpoint{0.520798in}{1.076194in}}%
\pgfpathlineto{\pgfqpoint{0.520798in}{1.076194in}}%
\pgfpathclose%
\pgfusepath{fill}%
\end{pgfscope}%
\begin{pgfscope}%
\pgfpathrectangle{\pgfqpoint{0.520798in}{0.442177in}}{\pgfqpoint{0.719005in}{0.871884in}}%
\pgfusepath{clip}%
\pgfsetroundcap%
\pgfsetroundjoin%
\pgfsetlinewidth{1.003750pt}%
\definecolor{currentstroke}{rgb}{0.870588,0.560784,0.019608}%
\pgfsetstrokecolor{currentstroke}%
\pgfsetdash{}{0pt}%
\pgfpathmoveto{\pgfqpoint{0.520798in}{1.064090in}}%
\pgfpathlineto{\pgfqpoint{0.556748in}{1.064090in}}%
\pgfpathlineto{\pgfqpoint{0.556748in}{1.008252in}}%
\pgfpathlineto{\pgfqpoint{0.628649in}{1.008252in}}%
\pgfpathlineto{\pgfqpoint{0.628649in}{0.921806in}}%
\pgfpathlineto{\pgfqpoint{0.700549in}{0.921806in}}%
\pgfpathlineto{\pgfqpoint{0.700549in}{0.713296in}}%
\pgfpathlineto{\pgfqpoint{0.772450in}{0.713296in}}%
\pgfpathlineto{\pgfqpoint{0.772450in}{0.612194in}}%
\pgfpathlineto{\pgfqpoint{0.844350in}{0.612194in}}%
\pgfpathlineto{\pgfqpoint{0.844350in}{0.604403in}}%
\pgfpathlineto{\pgfqpoint{0.916251in}{0.604403in}}%
\pgfpathlineto{\pgfqpoint{0.916251in}{0.597168in}}%
\pgfpathlineto{\pgfqpoint{0.988151in}{0.597168in}}%
\pgfpathlineto{\pgfqpoint{0.988151in}{0.581307in}}%
\pgfpathlineto{\pgfqpoint{1.060052in}{0.581307in}}%
\pgfpathlineto{\pgfqpoint{1.060052in}{0.536136in}}%
\pgfpathlineto{\pgfqpoint{1.131952in}{0.536136in}}%
\pgfpathlineto{\pgfqpoint{1.131952in}{0.527325in}}%
\pgfpathlineto{\pgfqpoint{1.203853in}{0.527325in}}%
\pgfpathlineto{\pgfqpoint{1.203853in}{0.510072in}}%
\pgfpathlineto{\pgfqpoint{1.241470in}{0.510072in}}%
\pgfusepath{stroke}%
\end{pgfscope}%
\begin{pgfscope}%
\pgfpathrectangle{\pgfqpoint{0.520798in}{0.442177in}}{\pgfqpoint{0.719005in}{0.871884in}}%
\pgfusepath{clip}%
\pgfsetbuttcap%
\pgfsetroundjoin%
\definecolor{currentfill}{rgb}{0.870588,0.560784,0.019608}%
\pgfsetfillcolor{currentfill}%
\pgfsetfillopacity{0.500000}%
\pgfsetlinewidth{0.000000pt}%
\definecolor{currentstroke}{rgb}{0.870588,0.560784,0.019608}%
\pgfsetstrokecolor{currentstroke}%
\pgfsetstrokeopacity{0.500000}%
\pgfsetdash{}{0pt}%
\pgfpathmoveto{\pgfqpoint{0.520798in}{1.076194in}}%
\pgfpathlineto{\pgfqpoint{0.520798in}{1.051986in}}%
\pgfpathlineto{\pgfqpoint{0.556748in}{1.051986in}}%
\pgfpathlineto{\pgfqpoint{0.556748in}{0.994752in}}%
\pgfpathlineto{\pgfqpoint{0.628649in}{0.994752in}}%
\pgfpathlineto{\pgfqpoint{0.628649in}{0.904877in}}%
\pgfpathlineto{\pgfqpoint{0.700549in}{0.904877in}}%
\pgfpathlineto{\pgfqpoint{0.700549in}{0.694707in}}%
\pgfpathlineto{\pgfqpoint{0.772450in}{0.694707in}}%
\pgfpathlineto{\pgfqpoint{0.772450in}{0.597071in}}%
\pgfpathlineto{\pgfqpoint{0.844350in}{0.597071in}}%
\pgfpathlineto{\pgfqpoint{0.844350in}{0.589363in}}%
\pgfpathlineto{\pgfqpoint{0.916251in}{0.589363in}}%
\pgfpathlineto{\pgfqpoint{0.916251in}{0.583369in}}%
\pgfpathlineto{\pgfqpoint{0.988151in}{0.583369in}}%
\pgfpathlineto{\pgfqpoint{0.988151in}{0.568385in}}%
\pgfpathlineto{\pgfqpoint{1.060052in}{0.568385in}}%
\pgfpathlineto{\pgfqpoint{1.060052in}{0.525437in}}%
\pgfpathlineto{\pgfqpoint{1.131952in}{0.525437in}}%
\pgfpathlineto{\pgfqpoint{1.131952in}{0.516061in}}%
\pgfpathlineto{\pgfqpoint{1.203853in}{0.516061in}}%
\pgfpathlineto{\pgfqpoint{1.203853in}{0.500320in}}%
\pgfpathlineto{\pgfqpoint{1.275753in}{0.500320in}}%
\pgfpathlineto{\pgfqpoint{1.275753in}{0.490662in}}%
\pgfpathlineto{\pgfqpoint{1.347654in}{0.490662in}}%
\pgfpathlineto{\pgfqpoint{1.347654in}{0.480913in}}%
\pgfpathlineto{\pgfqpoint{1.419554in}{0.480913in}}%
\pgfpathlineto{\pgfqpoint{1.419554in}{0.442177in}}%
\pgfpathlineto{\pgfqpoint{2.066659in}{0.442177in}}%
\pgfpathlineto{\pgfqpoint{2.066659in}{0.442177in}}%
\pgfpathlineto{\pgfqpoint{2.677813in}{0.442177in}}%
\pgfpathlineto{\pgfqpoint{2.677813in}{0.442177in}}%
\pgfpathlineto{\pgfqpoint{2.677813in}{0.442177in}}%
\pgfpathlineto{\pgfqpoint{2.066659in}{0.442177in}}%
\pgfpathlineto{\pgfqpoint{2.066659in}{0.442177in}}%
\pgfpathlineto{\pgfqpoint{1.419554in}{0.442177in}}%
\pgfpathlineto{\pgfqpoint{1.419554in}{0.498420in}}%
\pgfpathlineto{\pgfqpoint{1.347654in}{0.498420in}}%
\pgfpathlineto{\pgfqpoint{1.347654in}{0.504625in}}%
\pgfpathlineto{\pgfqpoint{1.275753in}{0.504625in}}%
\pgfpathlineto{\pgfqpoint{1.275753in}{0.519825in}}%
\pgfpathlineto{\pgfqpoint{1.203853in}{0.519825in}}%
\pgfpathlineto{\pgfqpoint{1.203853in}{0.538588in}}%
\pgfpathlineto{\pgfqpoint{1.131952in}{0.538588in}}%
\pgfpathlineto{\pgfqpoint{1.131952in}{0.546836in}}%
\pgfpathlineto{\pgfqpoint{1.060052in}{0.546836in}}%
\pgfpathlineto{\pgfqpoint{1.060052in}{0.594230in}}%
\pgfpathlineto{\pgfqpoint{0.988151in}{0.594230in}}%
\pgfpathlineto{\pgfqpoint{0.988151in}{0.610968in}}%
\pgfpathlineto{\pgfqpoint{0.916251in}{0.610968in}}%
\pgfpathlineto{\pgfqpoint{0.916251in}{0.619443in}}%
\pgfpathlineto{\pgfqpoint{0.844350in}{0.619443in}}%
\pgfpathlineto{\pgfqpoint{0.844350in}{0.627318in}}%
\pgfpathlineto{\pgfqpoint{0.772450in}{0.627318in}}%
\pgfpathlineto{\pgfqpoint{0.772450in}{0.731885in}}%
\pgfpathlineto{\pgfqpoint{0.700549in}{0.731885in}}%
\pgfpathlineto{\pgfqpoint{0.700549in}{0.938736in}}%
\pgfpathlineto{\pgfqpoint{0.628649in}{0.938736in}}%
\pgfpathlineto{\pgfqpoint{0.628649in}{1.021753in}}%
\pgfpathlineto{\pgfqpoint{0.556748in}{1.021753in}}%
\pgfpathlineto{\pgfqpoint{0.556748in}{1.076194in}}%
\pgfpathlineto{\pgfqpoint{0.520798in}{1.076194in}}%
\pgfpathlineto{\pgfqpoint{0.520798in}{1.076194in}}%
\pgfpathclose%
\pgfusepath{fill}%
\end{pgfscope}%
\begin{pgfscope}%
\pgfpathrectangle{\pgfqpoint{0.520798in}{0.442177in}}{\pgfqpoint{0.719005in}{0.871884in}}%
\pgfusepath{clip}%
\pgfsetroundcap%
\pgfsetroundjoin%
\pgfsetlinewidth{1.003750pt}%
\definecolor{currentstroke}{rgb}{0.007843,0.619608,0.450980}%
\pgfsetstrokecolor{currentstroke}%
\pgfsetdash{}{0pt}%
\pgfpathmoveto{\pgfqpoint{0.520798in}{1.064090in}}%
\pgfpathlineto{\pgfqpoint{0.556748in}{1.064090in}}%
\pgfpathlineto{\pgfqpoint{0.556748in}{1.008252in}}%
\pgfpathlineto{\pgfqpoint{0.628649in}{1.008252in}}%
\pgfpathlineto{\pgfqpoint{0.628649in}{0.921806in}}%
\pgfpathlineto{\pgfqpoint{0.700549in}{0.921806in}}%
\pgfpathlineto{\pgfqpoint{0.700549in}{0.713296in}}%
\pgfpathlineto{\pgfqpoint{0.772450in}{0.713296in}}%
\pgfpathlineto{\pgfqpoint{0.772450in}{0.612194in}}%
\pgfpathlineto{\pgfqpoint{0.844350in}{0.612194in}}%
\pgfpathlineto{\pgfqpoint{0.844350in}{0.604403in}}%
\pgfpathlineto{\pgfqpoint{0.916251in}{0.604403in}}%
\pgfpathlineto{\pgfqpoint{0.916251in}{0.597168in}}%
\pgfpathlineto{\pgfqpoint{0.988151in}{0.597168in}}%
\pgfpathlineto{\pgfqpoint{0.988151in}{0.581307in}}%
\pgfpathlineto{\pgfqpoint{1.060052in}{0.581307in}}%
\pgfpathlineto{\pgfqpoint{1.060052in}{0.536136in}}%
\pgfpathlineto{\pgfqpoint{1.131952in}{0.536136in}}%
\pgfpathlineto{\pgfqpoint{1.131952in}{0.527325in}}%
\pgfpathlineto{\pgfqpoint{1.203853in}{0.527325in}}%
\pgfpathlineto{\pgfqpoint{1.203853in}{0.510072in}}%
\pgfpathlineto{\pgfqpoint{1.241470in}{0.510072in}}%
\pgfusepath{stroke}%
\end{pgfscope}%
\begin{pgfscope}%
\pgfpathrectangle{\pgfqpoint{0.520798in}{0.442177in}}{\pgfqpoint{0.719005in}{0.871884in}}%
\pgfusepath{clip}%
\pgfsetbuttcap%
\pgfsetroundjoin%
\definecolor{currentfill}{rgb}{0.007843,0.619608,0.450980}%
\pgfsetfillcolor{currentfill}%
\pgfsetfillopacity{0.500000}%
\pgfsetlinewidth{0.000000pt}%
\definecolor{currentstroke}{rgb}{0.007843,0.619608,0.450980}%
\pgfsetstrokecolor{currentstroke}%
\pgfsetstrokeopacity{0.500000}%
\pgfsetdash{}{0pt}%
\pgfpathmoveto{\pgfqpoint{0.520798in}{1.076194in}}%
\pgfpathlineto{\pgfqpoint{0.520798in}{1.051986in}}%
\pgfpathlineto{\pgfqpoint{0.556748in}{1.051986in}}%
\pgfpathlineto{\pgfqpoint{0.556748in}{0.994752in}}%
\pgfpathlineto{\pgfqpoint{0.628649in}{0.994752in}}%
\pgfpathlineto{\pgfqpoint{0.628649in}{0.904877in}}%
\pgfpathlineto{\pgfqpoint{0.700549in}{0.904877in}}%
\pgfpathlineto{\pgfqpoint{0.700549in}{0.694707in}}%
\pgfpathlineto{\pgfqpoint{0.772450in}{0.694707in}}%
\pgfpathlineto{\pgfqpoint{0.772450in}{0.597071in}}%
\pgfpathlineto{\pgfqpoint{0.844350in}{0.597071in}}%
\pgfpathlineto{\pgfqpoint{0.844350in}{0.589363in}}%
\pgfpathlineto{\pgfqpoint{0.916251in}{0.589363in}}%
\pgfpathlineto{\pgfqpoint{0.916251in}{0.583369in}}%
\pgfpathlineto{\pgfqpoint{0.988151in}{0.583369in}}%
\pgfpathlineto{\pgfqpoint{0.988151in}{0.568385in}}%
\pgfpathlineto{\pgfqpoint{1.060052in}{0.568385in}}%
\pgfpathlineto{\pgfqpoint{1.060052in}{0.525437in}}%
\pgfpathlineto{\pgfqpoint{1.131952in}{0.525437in}}%
\pgfpathlineto{\pgfqpoint{1.131952in}{0.516061in}}%
\pgfpathlineto{\pgfqpoint{1.203853in}{0.516061in}}%
\pgfpathlineto{\pgfqpoint{1.203853in}{0.500320in}}%
\pgfpathlineto{\pgfqpoint{1.275753in}{0.500320in}}%
\pgfpathlineto{\pgfqpoint{1.275753in}{0.490662in}}%
\pgfpathlineto{\pgfqpoint{1.347654in}{0.490662in}}%
\pgfpathlineto{\pgfqpoint{1.347654in}{0.480913in}}%
\pgfpathlineto{\pgfqpoint{1.419554in}{0.480913in}}%
\pgfpathlineto{\pgfqpoint{1.419554in}{0.442177in}}%
\pgfpathlineto{\pgfqpoint{2.066659in}{0.442177in}}%
\pgfpathlineto{\pgfqpoint{2.066659in}{0.442177in}}%
\pgfpathlineto{\pgfqpoint{2.677813in}{0.442177in}}%
\pgfpathlineto{\pgfqpoint{2.677813in}{0.442177in}}%
\pgfpathlineto{\pgfqpoint{2.677813in}{0.442177in}}%
\pgfpathlineto{\pgfqpoint{2.066659in}{0.442177in}}%
\pgfpathlineto{\pgfqpoint{2.066659in}{0.442177in}}%
\pgfpathlineto{\pgfqpoint{1.419554in}{0.442177in}}%
\pgfpathlineto{\pgfqpoint{1.419554in}{0.498420in}}%
\pgfpathlineto{\pgfqpoint{1.347654in}{0.498420in}}%
\pgfpathlineto{\pgfqpoint{1.347654in}{0.504625in}}%
\pgfpathlineto{\pgfqpoint{1.275753in}{0.504625in}}%
\pgfpathlineto{\pgfqpoint{1.275753in}{0.519825in}}%
\pgfpathlineto{\pgfqpoint{1.203853in}{0.519825in}}%
\pgfpathlineto{\pgfqpoint{1.203853in}{0.538588in}}%
\pgfpathlineto{\pgfqpoint{1.131952in}{0.538588in}}%
\pgfpathlineto{\pgfqpoint{1.131952in}{0.546836in}}%
\pgfpathlineto{\pgfqpoint{1.060052in}{0.546836in}}%
\pgfpathlineto{\pgfqpoint{1.060052in}{0.594230in}}%
\pgfpathlineto{\pgfqpoint{0.988151in}{0.594230in}}%
\pgfpathlineto{\pgfqpoint{0.988151in}{0.610968in}}%
\pgfpathlineto{\pgfqpoint{0.916251in}{0.610968in}}%
\pgfpathlineto{\pgfqpoint{0.916251in}{0.619443in}}%
\pgfpathlineto{\pgfqpoint{0.844350in}{0.619443in}}%
\pgfpathlineto{\pgfqpoint{0.844350in}{0.627318in}}%
\pgfpathlineto{\pgfqpoint{0.772450in}{0.627318in}}%
\pgfpathlineto{\pgfqpoint{0.772450in}{0.731885in}}%
\pgfpathlineto{\pgfqpoint{0.700549in}{0.731885in}}%
\pgfpathlineto{\pgfqpoint{0.700549in}{0.938736in}}%
\pgfpathlineto{\pgfqpoint{0.628649in}{0.938736in}}%
\pgfpathlineto{\pgfqpoint{0.628649in}{1.021753in}}%
\pgfpathlineto{\pgfqpoint{0.556748in}{1.021753in}}%
\pgfpathlineto{\pgfqpoint{0.556748in}{1.076194in}}%
\pgfpathlineto{\pgfqpoint{0.520798in}{1.076194in}}%
\pgfpathlineto{\pgfqpoint{0.520798in}{1.076194in}}%
\pgfpathclose%
\pgfusepath{fill}%
\end{pgfscope}%
\begin{pgfscope}%
\definecolor{textcolor}{rgb}{0.150000,0.150000,0.150000}%
\pgfsetstrokecolor{textcolor}%
\pgfsetfillcolor{textcolor}%
\pgftext[x=0.880300in,y=1.397395in,,base]{\color{textcolor}\rmfamily\fontsize{9.000000}{10.800000}\selectfont \(\displaystyle c_X^+=1\)}%
\end{pgfscope}%
\begin{pgfscope}%
\pgfsetbuttcap%
\pgfsetmiterjoin%
\definecolor{currentfill}{rgb}{1.000000,1.000000,1.000000}%
\pgfsetfillcolor{currentfill}%
\pgfsetfillopacity{0.800000}%
\pgfsetlinewidth{1.003750pt}%
\definecolor{currentstroke}{rgb}{0.800000,0.800000,0.800000}%
\pgfsetstrokecolor{currentstroke}%
\pgfsetstrokeopacity{0.800000}%
\pgfsetdash{}{0pt}%
\pgfpathmoveto{\pgfqpoint{0.785843in}{0.638103in}}%
\pgfpathlineto{\pgfqpoint{1.191192in}{0.638103in}}%
\pgfpathlineto{\pgfqpoint{1.191192in}{1.265450in}}%
\pgfpathlineto{\pgfqpoint{0.785843in}{1.265450in}}%
\pgfpathlineto{\pgfqpoint{0.785843in}{0.638103in}}%
\pgfpathclose%
\pgfusepath{stroke,fill}%
\end{pgfscope}%
\begin{pgfscope}%
\definecolor{textcolor}{rgb}{0.150000,0.150000,0.150000}%
\pgfsetstrokecolor{textcolor}%
\pgfsetfillcolor{textcolor}%
\pgftext[x=0.912291in,y=1.113275in,left,base]{\color{textcolor}\rmfamily\fontsize{9.000000}{10.800000}\selectfont \(\displaystyle c_X^-\)}%
\end{pgfscope}%
\begin{pgfscope}%
\pgfsetroundcap%
\pgfsetroundjoin%
\pgfsetlinewidth{1.003750pt}%
\definecolor{currentstroke}{rgb}{0.003922,0.450980,0.698039}%
\pgfsetstrokecolor{currentstroke}%
\pgfsetdash{}{0pt}%
\pgfpathmoveto{\pgfqpoint{0.824732in}{1.001052in}}%
\pgfpathlineto{\pgfqpoint{0.873343in}{1.001052in}}%
\pgfpathlineto{\pgfqpoint{0.873343in}{1.001052in}}%
\pgfpathlineto{\pgfqpoint{0.970565in}{1.001052in}}%
\pgfpathlineto{\pgfqpoint{0.970565in}{1.001052in}}%
\pgfpathlineto{\pgfqpoint{1.019176in}{1.001052in}}%
\pgfusepath{stroke}%
\end{pgfscope}%
\begin{pgfscope}%
\definecolor{textcolor}{rgb}{0.150000,0.150000,0.150000}%
\pgfsetstrokecolor{textcolor}%
\pgfsetfillcolor{textcolor}%
\pgftext[x=1.096954in,y=0.967024in,left,base]{\color{textcolor}\rmfamily\fontsize{7.000000}{8.400000}\selectfont 1}%
\end{pgfscope}%
\begin{pgfscope}%
\pgfsetroundcap%
\pgfsetroundjoin%
\pgfsetlinewidth{1.003750pt}%
\definecolor{currentstroke}{rgb}{0.870588,0.560784,0.019608}%
\pgfsetstrokecolor{currentstroke}%
\pgfsetdash{}{0pt}%
\pgfpathmoveto{\pgfqpoint{0.824732in}{0.865486in}}%
\pgfpathlineto{\pgfqpoint{0.873343in}{0.865486in}}%
\pgfpathlineto{\pgfqpoint{0.873343in}{0.865486in}}%
\pgfpathlineto{\pgfqpoint{0.970565in}{0.865486in}}%
\pgfpathlineto{\pgfqpoint{0.970565in}{0.865486in}}%
\pgfpathlineto{\pgfqpoint{1.019176in}{0.865486in}}%
\pgfusepath{stroke}%
\end{pgfscope}%
\begin{pgfscope}%
\definecolor{textcolor}{rgb}{0.150000,0.150000,0.150000}%
\pgfsetstrokecolor{textcolor}%
\pgfsetfillcolor{textcolor}%
\pgftext[x=1.096954in,y=0.831458in,left,base]{\color{textcolor}\rmfamily\fontsize{7.000000}{8.400000}\selectfont 2}%
\end{pgfscope}%
\begin{pgfscope}%
\pgfsetroundcap%
\pgfsetroundjoin%
\pgfsetlinewidth{1.003750pt}%
\definecolor{currentstroke}{rgb}{0.007843,0.619608,0.450980}%
\pgfsetstrokecolor{currentstroke}%
\pgfsetdash{}{0pt}%
\pgfpathmoveto{\pgfqpoint{0.824732in}{0.729919in}}%
\pgfpathlineto{\pgfqpoint{0.873343in}{0.729919in}}%
\pgfpathlineto{\pgfqpoint{0.873343in}{0.729919in}}%
\pgfpathlineto{\pgfqpoint{0.970565in}{0.729919in}}%
\pgfpathlineto{\pgfqpoint{0.970565in}{0.729919in}}%
\pgfpathlineto{\pgfqpoint{1.019176in}{0.729919in}}%
\pgfusepath{stroke}%
\end{pgfscope}%
\begin{pgfscope}%
\definecolor{textcolor}{rgb}{0.150000,0.150000,0.150000}%
\pgfsetstrokecolor{textcolor}%
\pgfsetfillcolor{textcolor}%
\pgftext[x=1.096954in,y=0.695892in,left,base]{\color{textcolor}\rmfamily\fontsize{7.000000}{8.400000}\selectfont 4}%
\end{pgfscope}%
\end{pgfpicture}%
\makeatother%
\endgroup%

%% file: figures/experiments/graphs/sparse_smoothing/nodes_structure/node_classification-Citeseer-GAT-hidden=8-p_adj_plus=0.001-p_adj_minus=0.8-p_att_plus=0.0-p_att_minus=0.0-multi_class_cert-B.pgf
\begingroup%
\makeatletter%
\begin{pgfpicture}%
\pgfpathrectangle{\pgfpointorigin}{\pgfqpoint{1.375000in}{1.581250in}}%
\pgfusepath{use as bounding box, clip}%
\begin{pgfscope}%
\pgfsetbuttcap%
\pgfsetmiterjoin%
\definecolor{currentfill}{rgb}{1.000000,1.000000,1.000000}%
\pgfsetfillcolor{currentfill}%
\pgfsetlinewidth{0.000000pt}%
\definecolor{currentstroke}{rgb}{1.000000,1.000000,1.000000}%
\pgfsetstrokecolor{currentstroke}%
\pgfsetdash{}{0pt}%
\pgfpathmoveto{\pgfqpoint{0.000000in}{0.000000in}}%
\pgfpathlineto{\pgfqpoint{1.375000in}{0.000000in}}%
\pgfpathlineto{\pgfqpoint{1.375000in}{1.581250in}}%
\pgfpathlineto{\pgfqpoint{0.000000in}{1.581250in}}%
\pgfpathlineto{\pgfqpoint{0.000000in}{0.000000in}}%
\pgfpathclose%
\pgfusepath{fill}%
\end{pgfscope}%
\begin{pgfscope}%
\pgfsetbuttcap%
\pgfsetmiterjoin%
\definecolor{currentfill}{rgb}{1.000000,1.000000,1.000000}%
\pgfsetfillcolor{currentfill}%
\pgfsetlinewidth{0.000000pt}%
\definecolor{currentstroke}{rgb}{0.000000,0.000000,0.000000}%
\pgfsetstrokecolor{currentstroke}%
\pgfsetstrokeopacity{0.000000}%
\pgfsetdash{}{0pt}%
\pgfpathmoveto{\pgfqpoint{0.520798in}{0.442177in}}%
\pgfpathlineto{\pgfqpoint{1.239803in}{0.442177in}}%
\pgfpathlineto{\pgfqpoint{1.239803in}{1.314061in}}%
\pgfpathlineto{\pgfqpoint{0.520798in}{1.314061in}}%
\pgfpathlineto{\pgfqpoint{0.520798in}{0.442177in}}%
\pgfpathclose%
\pgfusepath{fill}%
\end{pgfscope}%
\begin{pgfscope}%
\pgfpathrectangle{\pgfqpoint{0.520798in}{0.442177in}}{\pgfqpoint{0.719005in}{0.871884in}}%
\pgfusepath{clip}%
\pgfsetroundcap%
\pgfsetroundjoin%
\pgfsetlinewidth{0.501875pt}%
\definecolor{currentstroke}{rgb}{0.800000,0.800000,0.800000}%
\pgfsetstrokecolor{currentstroke}%
\pgfsetdash{}{0pt}%
\pgfpathmoveto{\pgfqpoint{0.520798in}{0.442177in}}%
\pgfpathlineto{\pgfqpoint{0.520798in}{1.314061in}}%
\pgfusepath{stroke}%
\end{pgfscope}%
\begin{pgfscope}%
\definecolor{textcolor}{rgb}{0.150000,0.150000,0.150000}%
\pgfsetstrokecolor{textcolor}%
\pgfsetfillcolor{textcolor}%
\pgftext[x=0.520798in,y=0.351899in,,top]{\color{textcolor}\rmfamily\fontsize{8.000000}{9.600000}\selectfont \(\displaystyle {0}\)}%
\end{pgfscope}%
\begin{pgfscope}%
\pgfpathrectangle{\pgfqpoint{0.520798in}{0.442177in}}{\pgfqpoint{0.719005in}{0.871884in}}%
\pgfusepath{clip}%
\pgfsetroundcap%
\pgfsetroundjoin%
\pgfsetlinewidth{0.501875pt}%
\definecolor{currentstroke}{rgb}{0.800000,0.800000,0.800000}%
\pgfsetstrokecolor{currentstroke}%
\pgfsetdash{}{0pt}%
\pgfpathmoveto{\pgfqpoint{0.880300in}{0.442177in}}%
\pgfpathlineto{\pgfqpoint{0.880300in}{1.314061in}}%
\pgfusepath{stroke}%
\end{pgfscope}%
\begin{pgfscope}%
\definecolor{textcolor}{rgb}{0.150000,0.150000,0.150000}%
\pgfsetstrokecolor{textcolor}%
\pgfsetfillcolor{textcolor}%
\pgftext[x=0.880300in,y=0.351899in,,top]{\color{textcolor}\rmfamily\fontsize{8.000000}{9.600000}\selectfont \(\displaystyle {5}\)}%
\end{pgfscope}%
\begin{pgfscope}%
\pgfpathrectangle{\pgfqpoint{0.520798in}{0.442177in}}{\pgfqpoint{0.719005in}{0.871884in}}%
\pgfusepath{clip}%
\pgfsetroundcap%
\pgfsetroundjoin%
\pgfsetlinewidth{0.501875pt}%
\definecolor{currentstroke}{rgb}{0.800000,0.800000,0.800000}%
\pgfsetstrokecolor{currentstroke}%
\pgfsetdash{}{0pt}%
\pgfpathmoveto{\pgfqpoint{1.239803in}{0.442177in}}%
\pgfpathlineto{\pgfqpoint{1.239803in}{1.314061in}}%
\pgfusepath{stroke}%
\end{pgfscope}%
\begin{pgfscope}%
\definecolor{textcolor}{rgb}{0.150000,0.150000,0.150000}%
\pgfsetstrokecolor{textcolor}%
\pgfsetfillcolor{textcolor}%
\pgftext[x=1.239803in,y=0.351899in,,top]{\color{textcolor}\rmfamily\fontsize{8.000000}{9.600000}\selectfont \(\displaystyle {10}\)}%
\end{pgfscope}%
\begin{pgfscope}%
\definecolor{textcolor}{rgb}{0.150000,0.150000,0.150000}%
\pgfsetstrokecolor{textcolor}%
\pgfsetfillcolor{textcolor}%
\pgftext[x=0.880300in,y=0.198219in,,top]{\color{textcolor}\rmfamily\fontsize{10.000000}{12.000000}\selectfont Edit distance \(\displaystyle \epsilon\)}%
\end{pgfscope}%
\begin{pgfscope}%
\pgfpathrectangle{\pgfqpoint{0.520798in}{0.442177in}}{\pgfqpoint{0.719005in}{0.871884in}}%
\pgfusepath{clip}%
\pgfsetroundcap%
\pgfsetroundjoin%
\pgfsetlinewidth{0.501875pt}%
\definecolor{currentstroke}{rgb}{0.800000,0.800000,0.800000}%
\pgfsetstrokecolor{currentstroke}%
\pgfsetdash{}{0pt}%
\pgfpathmoveto{\pgfqpoint{0.520798in}{0.442177in}}%
\pgfpathlineto{\pgfqpoint{1.239803in}{0.442177in}}%
\pgfusepath{stroke}%
\end{pgfscope}%
\begin{pgfscope}%
\definecolor{textcolor}{rgb}{0.150000,0.150000,0.150000}%
\pgfsetstrokecolor{textcolor}%
\pgfsetfillcolor{textcolor}%
\pgftext[x=0.273151in, y=0.403915in, left, base]{\color{textcolor}\rmfamily\fontsize{8.000000}{9.600000}\selectfont 0\%}%
\end{pgfscope}%
\begin{pgfscope}%
\pgfpathrectangle{\pgfqpoint{0.520798in}{0.442177in}}{\pgfqpoint{0.719005in}{0.871884in}}%
\pgfusepath{clip}%
\pgfsetroundcap%
\pgfsetroundjoin%
\pgfsetlinewidth{0.501875pt}%
\definecolor{currentstroke}{rgb}{0.800000,0.800000,0.800000}%
\pgfsetstrokecolor{currentstroke}%
\pgfsetdash{}{0pt}%
\pgfpathmoveto{\pgfqpoint{0.520798in}{0.660148in}}%
\pgfpathlineto{\pgfqpoint{1.239803in}{0.660148in}}%
\pgfusepath{stroke}%
\end{pgfscope}%
\begin{pgfscope}%
\definecolor{textcolor}{rgb}{0.150000,0.150000,0.150000}%
\pgfsetstrokecolor{textcolor}%
\pgfsetfillcolor{textcolor}%
\pgftext[x=0.214138in, y=0.621886in, left, base]{\color{textcolor}\rmfamily\fontsize{8.000000}{9.600000}\selectfont 25\%}%
\end{pgfscope}%
\begin{pgfscope}%
\pgfpathrectangle{\pgfqpoint{0.520798in}{0.442177in}}{\pgfqpoint{0.719005in}{0.871884in}}%
\pgfusepath{clip}%
\pgfsetroundcap%
\pgfsetroundjoin%
\pgfsetlinewidth{0.501875pt}%
\definecolor{currentstroke}{rgb}{0.800000,0.800000,0.800000}%
\pgfsetstrokecolor{currentstroke}%
\pgfsetdash{}{0pt}%
\pgfpathmoveto{\pgfqpoint{0.520798in}{0.878119in}}%
\pgfpathlineto{\pgfqpoint{1.239803in}{0.878119in}}%
\pgfusepath{stroke}%
\end{pgfscope}%
\begin{pgfscope}%
\definecolor{textcolor}{rgb}{0.150000,0.150000,0.150000}%
\pgfsetstrokecolor{textcolor}%
\pgfsetfillcolor{textcolor}%
\pgftext[x=0.214138in, y=0.839857in, left, base]{\color{textcolor}\rmfamily\fontsize{8.000000}{9.600000}\selectfont 50\%}%
\end{pgfscope}%
\begin{pgfscope}%
\pgfpathrectangle{\pgfqpoint{0.520798in}{0.442177in}}{\pgfqpoint{0.719005in}{0.871884in}}%
\pgfusepath{clip}%
\pgfsetroundcap%
\pgfsetroundjoin%
\pgfsetlinewidth{0.501875pt}%
\definecolor{currentstroke}{rgb}{0.800000,0.800000,0.800000}%
\pgfsetstrokecolor{currentstroke}%
\pgfsetdash{}{0pt}%
\pgfpathmoveto{\pgfqpoint{0.520798in}{1.096090in}}%
\pgfpathlineto{\pgfqpoint{1.239803in}{1.096090in}}%
\pgfusepath{stroke}%
\end{pgfscope}%
\begin{pgfscope}%
\definecolor{textcolor}{rgb}{0.150000,0.150000,0.150000}%
\pgfsetstrokecolor{textcolor}%
\pgfsetfillcolor{textcolor}%
\pgftext[x=0.214138in, y=1.057828in, left, base]{\color{textcolor}\rmfamily\fontsize{8.000000}{9.600000}\selectfont 75\%}%
\end{pgfscope}%
\begin{pgfscope}%
\pgfpathrectangle{\pgfqpoint{0.520798in}{0.442177in}}{\pgfqpoint{0.719005in}{0.871884in}}%
\pgfusepath{clip}%
\pgfsetroundcap%
\pgfsetroundjoin%
\pgfsetlinewidth{0.501875pt}%
\definecolor{currentstroke}{rgb}{0.800000,0.800000,0.800000}%
\pgfsetstrokecolor{currentstroke}%
\pgfsetdash{}{0pt}%
\pgfpathmoveto{\pgfqpoint{0.520798in}{1.314061in}}%
\pgfpathlineto{\pgfqpoint{1.239803in}{1.314061in}}%
\pgfusepath{stroke}%
\end{pgfscope}%
\begin{pgfscope}%
\definecolor{textcolor}{rgb}{0.150000,0.150000,0.150000}%
\pgfsetstrokecolor{textcolor}%
\pgfsetfillcolor{textcolor}%
\pgftext[x=0.155124in, y=1.275799in, left, base]{\color{textcolor}\rmfamily\fontsize{8.000000}{9.600000}\selectfont 100\%}%
\end{pgfscope}%
\begin{pgfscope}%
\definecolor{textcolor}{rgb}{0.150000,0.150000,0.150000}%
\pgfsetstrokecolor{textcolor}%
\pgfsetfillcolor{textcolor}%
\pgftext[x=0.099569in,y=0.878119in,,bottom,rotate=90.000000]{\color{textcolor}\rmfamily\fontsize{10.000000}{12.000000}\selectfont Cert. Acc.}%
\end{pgfscope}%
\begin{pgfscope}%
\pgfsetrectcap%
\pgfsetmiterjoin%
\pgfsetlinewidth{0.752812pt}%
\definecolor{currentstroke}{rgb}{0.700000,0.700000,0.700000}%
\pgfsetstrokecolor{currentstroke}%
\pgfsetdash{}{0pt}%
\pgfpathmoveto{\pgfqpoint{0.520798in}{0.442177in}}%
\pgfpathlineto{\pgfqpoint{0.520798in}{1.314061in}}%
\pgfusepath{stroke}%
\end{pgfscope}%
\begin{pgfscope}%
\pgfsetrectcap%
\pgfsetmiterjoin%
\pgfsetlinewidth{0.752812pt}%
\definecolor{currentstroke}{rgb}{0.700000,0.700000,0.700000}%
\pgfsetstrokecolor{currentstroke}%
\pgfsetdash{}{0pt}%
\pgfpathmoveto{\pgfqpoint{1.239803in}{0.442177in}}%
\pgfpathlineto{\pgfqpoint{1.239803in}{1.314061in}}%
\pgfusepath{stroke}%
\end{pgfscope}%
\begin{pgfscope}%
\pgfsetrectcap%
\pgfsetmiterjoin%
\pgfsetlinewidth{0.752812pt}%
\definecolor{currentstroke}{rgb}{0.700000,0.700000,0.700000}%
\pgfsetstrokecolor{currentstroke}%
\pgfsetdash{}{0pt}%
\pgfpathmoveto{\pgfqpoint{0.520798in}{0.442177in}}%
\pgfpathlineto{\pgfqpoint{1.239803in}{0.442177in}}%
\pgfusepath{stroke}%
\end{pgfscope}%
\begin{pgfscope}%
\pgfsetrectcap%
\pgfsetmiterjoin%
\pgfsetlinewidth{0.752812pt}%
\definecolor{currentstroke}{rgb}{0.700000,0.700000,0.700000}%
\pgfsetstrokecolor{currentstroke}%
\pgfsetdash{}{0pt}%
\pgfpathmoveto{\pgfqpoint{0.520798in}{1.314061in}}%
\pgfpathlineto{\pgfqpoint{1.239803in}{1.314061in}}%
\pgfusepath{stroke}%
\end{pgfscope}%
\begin{pgfscope}%
\pgfpathrectangle{\pgfqpoint{0.520798in}{0.442177in}}{\pgfqpoint{0.719005in}{0.871884in}}%
\pgfusepath{clip}%
\pgfsetroundcap%
\pgfsetroundjoin%
\pgfsetlinewidth{1.003750pt}%
\definecolor{currentstroke}{rgb}{0.003922,0.450980,0.698039}%
\pgfsetstrokecolor{currentstroke}%
\pgfsetdash{}{0pt}%
\pgfpathmoveto{\pgfqpoint{0.520798in}{0.956218in}}%
\pgfpathlineto{\pgfqpoint{0.556748in}{0.956218in}}%
\pgfpathlineto{\pgfqpoint{0.556748in}{0.818849in}}%
\pgfpathlineto{\pgfqpoint{0.628649in}{0.818849in}}%
\pgfpathlineto{\pgfqpoint{0.628649in}{0.642339in}}%
\pgfpathlineto{\pgfqpoint{0.700549in}{0.642339in}}%
\pgfpathlineto{\pgfqpoint{0.700549in}{0.461748in}}%
\pgfpathlineto{\pgfqpoint{0.772450in}{0.461748in}}%
\pgfpathlineto{\pgfqpoint{0.772450in}{0.446165in}}%
\pgfpathlineto{\pgfqpoint{0.844350in}{0.446165in}}%
\pgfpathlineto{\pgfqpoint{0.844350in}{0.444681in}}%
\pgfpathlineto{\pgfqpoint{0.916251in}{0.444681in}}%
\pgfpathlineto{\pgfqpoint{0.916251in}{0.444403in}}%
\pgfpathlineto{\pgfqpoint{0.988151in}{0.444403in}}%
\pgfpathlineto{\pgfqpoint{0.988151in}{0.443939in}}%
\pgfpathlineto{\pgfqpoint{1.060052in}{0.443939in}}%
\pgfpathlineto{\pgfqpoint{1.060052in}{0.442455in}}%
\pgfpathlineto{\pgfqpoint{1.167902in}{0.442455in}}%
\pgfpathlineto{\pgfqpoint{1.167902in}{0.442362in}}%
\pgfpathlineto{\pgfqpoint{1.241470in}{0.442362in}}%
\pgfusepath{stroke}%
\end{pgfscope}%
\begin{pgfscope}%
\pgfpathrectangle{\pgfqpoint{0.520798in}{0.442177in}}{\pgfqpoint{0.719005in}{0.871884in}}%
\pgfusepath{clip}%
\pgfsetbuttcap%
\pgfsetroundjoin%
\definecolor{currentfill}{rgb}{0.003922,0.450980,0.698039}%
\pgfsetfillcolor{currentfill}%
\pgfsetfillopacity{0.500000}%
\pgfsetlinewidth{0.000000pt}%
\definecolor{currentstroke}{rgb}{0.003922,0.450980,0.698039}%
\pgfsetstrokecolor{currentstroke}%
\pgfsetstrokeopacity{0.500000}%
\pgfsetdash{}{0pt}%
\pgfpathmoveto{\pgfqpoint{0.520798in}{0.970311in}}%
\pgfpathlineto{\pgfqpoint{0.520798in}{0.942125in}}%
\pgfpathlineto{\pgfqpoint{0.556748in}{0.942125in}}%
\pgfpathlineto{\pgfqpoint{0.556748in}{0.803775in}}%
\pgfpathlineto{\pgfqpoint{0.628649in}{0.803775in}}%
\pgfpathlineto{\pgfqpoint{0.628649in}{0.636253in}}%
\pgfpathlineto{\pgfqpoint{0.700549in}{0.636253in}}%
\pgfpathlineto{\pgfqpoint{0.700549in}{0.456187in}}%
\pgfpathlineto{\pgfqpoint{0.772450in}{0.456187in}}%
\pgfpathlineto{\pgfqpoint{0.772450in}{0.444206in}}%
\pgfpathlineto{\pgfqpoint{0.844350in}{0.444206in}}%
\pgfpathlineto{\pgfqpoint{0.844350in}{0.442981in}}%
\pgfpathlineto{\pgfqpoint{0.916251in}{0.442981in}}%
\pgfpathlineto{\pgfqpoint{0.916251in}{0.442786in}}%
\pgfpathlineto{\pgfqpoint{0.988151in}{0.442786in}}%
\pgfpathlineto{\pgfqpoint{0.988151in}{0.442551in}}%
\pgfpathlineto{\pgfqpoint{1.060052in}{0.442551in}}%
\pgfpathlineto{\pgfqpoint{1.060052in}{0.442228in}}%
\pgfpathlineto{\pgfqpoint{1.167902in}{0.442228in}}%
\pgfpathlineto{\pgfqpoint{1.167902in}{0.442135in}}%
\pgfpathlineto{\pgfqpoint{1.311703in}{0.442135in}}%
\pgfpathlineto{\pgfqpoint{1.311703in}{0.442084in}}%
\pgfpathlineto{\pgfqpoint{1.419554in}{0.442084in}}%
\pgfpathlineto{\pgfqpoint{1.419554in}{0.442177in}}%
\pgfpathlineto{\pgfqpoint{2.066659in}{0.442177in}}%
\pgfpathlineto{\pgfqpoint{2.066659in}{0.442177in}}%
\pgfpathlineto{\pgfqpoint{2.677813in}{0.442177in}}%
\pgfpathlineto{\pgfqpoint{2.677813in}{0.442177in}}%
\pgfpathlineto{\pgfqpoint{2.677813in}{0.442177in}}%
\pgfpathlineto{\pgfqpoint{2.066659in}{0.442177in}}%
\pgfpathlineto{\pgfqpoint{2.066659in}{0.442177in}}%
\pgfpathlineto{\pgfqpoint{1.419554in}{0.442177in}}%
\pgfpathlineto{\pgfqpoint{1.419554in}{0.442455in}}%
\pgfpathlineto{\pgfqpoint{1.311703in}{0.442455in}}%
\pgfpathlineto{\pgfqpoint{1.311703in}{0.442590in}}%
\pgfpathlineto{\pgfqpoint{1.167902in}{0.442590in}}%
\pgfpathlineto{\pgfqpoint{1.167902in}{0.442682in}}%
\pgfpathlineto{\pgfqpoint{1.060052in}{0.442682in}}%
\pgfpathlineto{\pgfqpoint{1.060052in}{0.445327in}}%
\pgfpathlineto{\pgfqpoint{0.988151in}{0.445327in}}%
\pgfpathlineto{\pgfqpoint{0.988151in}{0.446020in}}%
\pgfpathlineto{\pgfqpoint{0.916251in}{0.446020in}}%
\pgfpathlineto{\pgfqpoint{0.916251in}{0.446381in}}%
\pgfpathlineto{\pgfqpoint{0.844350in}{0.446381in}}%
\pgfpathlineto{\pgfqpoint{0.844350in}{0.448124in}}%
\pgfpathlineto{\pgfqpoint{0.772450in}{0.448124in}}%
\pgfpathlineto{\pgfqpoint{0.772450in}{0.467308in}}%
\pgfpathlineto{\pgfqpoint{0.700549in}{0.467308in}}%
\pgfpathlineto{\pgfqpoint{0.700549in}{0.648426in}}%
\pgfpathlineto{\pgfqpoint{0.628649in}{0.648426in}}%
\pgfpathlineto{\pgfqpoint{0.628649in}{0.833924in}}%
\pgfpathlineto{\pgfqpoint{0.556748in}{0.833924in}}%
\pgfpathlineto{\pgfqpoint{0.556748in}{0.970311in}}%
\pgfpathlineto{\pgfqpoint{0.520798in}{0.970311in}}%
\pgfpathlineto{\pgfqpoint{0.520798in}{0.970311in}}%
\pgfpathclose%
\pgfusepath{fill}%
\end{pgfscope}%
\begin{pgfscope}%
\pgfpathrectangle{\pgfqpoint{0.520798in}{0.442177in}}{\pgfqpoint{0.719005in}{0.871884in}}%
\pgfusepath{clip}%
\pgfsetroundcap%
\pgfsetroundjoin%
\pgfsetlinewidth{1.003750pt}%
\definecolor{currentstroke}{rgb}{0.870588,0.560784,0.019608}%
\pgfsetstrokecolor{currentstroke}%
\pgfsetdash{}{0pt}%
\pgfpathmoveto{\pgfqpoint{0.520798in}{0.956218in}}%
\pgfpathlineto{\pgfqpoint{0.556748in}{0.956218in}}%
\pgfpathlineto{\pgfqpoint{0.556748in}{0.846768in}}%
\pgfpathlineto{\pgfqpoint{0.628649in}{0.846768in}}%
\pgfpathlineto{\pgfqpoint{0.628649in}{0.756519in}}%
\pgfpathlineto{\pgfqpoint{0.700549in}{0.756519in}}%
\pgfpathlineto{\pgfqpoint{0.700549in}{0.681667in}}%
\pgfpathlineto{\pgfqpoint{0.772450in}{0.681667in}}%
\pgfpathlineto{\pgfqpoint{0.772450in}{0.627684in}}%
\pgfpathlineto{\pgfqpoint{0.844350in}{0.627684in}}%
\pgfpathlineto{\pgfqpoint{0.844350in}{0.584461in}}%
\pgfpathlineto{\pgfqpoint{0.916251in}{0.584461in}}%
\pgfpathlineto{\pgfqpoint{0.916251in}{0.461748in}}%
\pgfpathlineto{\pgfqpoint{0.988151in}{0.461748in}}%
\pgfpathlineto{\pgfqpoint{0.988151in}{0.458223in}}%
\pgfpathlineto{\pgfqpoint{1.060052in}{0.458223in}}%
\pgfpathlineto{\pgfqpoint{1.060052in}{0.446165in}}%
\pgfpathlineto{\pgfqpoint{1.131952in}{0.446165in}}%
\pgfpathlineto{\pgfqpoint{1.131952in}{0.445701in}}%
\pgfpathlineto{\pgfqpoint{1.203853in}{0.445701in}}%
\pgfpathlineto{\pgfqpoint{1.203853in}{0.444588in}}%
\pgfpathlineto{\pgfqpoint{1.241470in}{0.444588in}}%
\pgfusepath{stroke}%
\end{pgfscope}%
\begin{pgfscope}%
\pgfpathrectangle{\pgfqpoint{0.520798in}{0.442177in}}{\pgfqpoint{0.719005in}{0.871884in}}%
\pgfusepath{clip}%
\pgfsetbuttcap%
\pgfsetroundjoin%
\definecolor{currentfill}{rgb}{0.870588,0.560784,0.019608}%
\pgfsetfillcolor{currentfill}%
\pgfsetfillopacity{0.500000}%
\pgfsetlinewidth{0.000000pt}%
\definecolor{currentstroke}{rgb}{0.870588,0.560784,0.019608}%
\pgfsetstrokecolor{currentstroke}%
\pgfsetstrokeopacity{0.500000}%
\pgfsetdash{}{0pt}%
\pgfpathmoveto{\pgfqpoint{0.520798in}{0.970311in}}%
\pgfpathlineto{\pgfqpoint{0.520798in}{0.942125in}}%
\pgfpathlineto{\pgfqpoint{0.556748in}{0.942125in}}%
\pgfpathlineto{\pgfqpoint{0.556748in}{0.831268in}}%
\pgfpathlineto{\pgfqpoint{0.628649in}{0.831268in}}%
\pgfpathlineto{\pgfqpoint{0.628649in}{0.748394in}}%
\pgfpathlineto{\pgfqpoint{0.700549in}{0.748394in}}%
\pgfpathlineto{\pgfqpoint{0.700549in}{0.677410in}}%
\pgfpathlineto{\pgfqpoint{0.772450in}{0.677410in}}%
\pgfpathlineto{\pgfqpoint{0.772450in}{0.619088in}}%
\pgfpathlineto{\pgfqpoint{0.844350in}{0.619088in}}%
\pgfpathlineto{\pgfqpoint{0.844350in}{0.577624in}}%
\pgfpathlineto{\pgfqpoint{0.916251in}{0.577624in}}%
\pgfpathlineto{\pgfqpoint{0.916251in}{0.456187in}}%
\pgfpathlineto{\pgfqpoint{0.988151in}{0.456187in}}%
\pgfpathlineto{\pgfqpoint{0.988151in}{0.454260in}}%
\pgfpathlineto{\pgfqpoint{1.060052in}{0.454260in}}%
\pgfpathlineto{\pgfqpoint{1.060052in}{0.444206in}}%
\pgfpathlineto{\pgfqpoint{1.131952in}{0.444206in}}%
\pgfpathlineto{\pgfqpoint{1.131952in}{0.443699in}}%
\pgfpathlineto{\pgfqpoint{1.203853in}{0.443699in}}%
\pgfpathlineto{\pgfqpoint{1.203853in}{0.442795in}}%
\pgfpathlineto{\pgfqpoint{1.275753in}{0.442795in}}%
\pgfpathlineto{\pgfqpoint{1.275753in}{0.442646in}}%
\pgfpathlineto{\pgfqpoint{1.383604in}{0.442646in}}%
\pgfpathlineto{\pgfqpoint{1.383604in}{0.442551in}}%
\pgfpathlineto{\pgfqpoint{1.491455in}{0.442551in}}%
\pgfpathlineto{\pgfqpoint{1.491455in}{0.442359in}}%
\pgfpathlineto{\pgfqpoint{1.563355in}{0.442359in}}%
\pgfpathlineto{\pgfqpoint{1.563355in}{0.442511in}}%
\pgfpathlineto{\pgfqpoint{1.635256in}{0.442511in}}%
\pgfpathlineto{\pgfqpoint{1.635256in}{0.442228in}}%
\pgfpathlineto{\pgfqpoint{1.815007in}{0.442228in}}%
\pgfpathlineto{\pgfqpoint{1.815007in}{0.442135in}}%
\pgfpathlineto{\pgfqpoint{2.030709in}{0.442135in}}%
\pgfpathlineto{\pgfqpoint{2.030709in}{0.442084in}}%
\pgfpathlineto{\pgfqpoint{2.210460in}{0.442084in}}%
\pgfpathlineto{\pgfqpoint{2.210460in}{0.442177in}}%
\pgfpathlineto{\pgfqpoint{2.498062in}{0.442177in}}%
\pgfpathlineto{\pgfqpoint{2.498062in}{0.442177in}}%
\pgfpathlineto{\pgfqpoint{2.677813in}{0.442177in}}%
\pgfpathlineto{\pgfqpoint{2.677813in}{0.442177in}}%
\pgfpathlineto{\pgfqpoint{2.677813in}{0.442177in}}%
\pgfpathlineto{\pgfqpoint{2.498062in}{0.442177in}}%
\pgfpathlineto{\pgfqpoint{2.498062in}{0.442177in}}%
\pgfpathlineto{\pgfqpoint{2.210460in}{0.442177in}}%
\pgfpathlineto{\pgfqpoint{2.210460in}{0.442455in}}%
\pgfpathlineto{\pgfqpoint{2.030709in}{0.442455in}}%
\pgfpathlineto{\pgfqpoint{2.030709in}{0.442590in}}%
\pgfpathlineto{\pgfqpoint{1.815007in}{0.442590in}}%
\pgfpathlineto{\pgfqpoint{1.815007in}{0.442682in}}%
\pgfpathlineto{\pgfqpoint{1.635256in}{0.442682in}}%
\pgfpathlineto{\pgfqpoint{1.635256in}{0.444626in}}%
\pgfpathlineto{\pgfqpoint{1.563355in}{0.444626in}}%
\pgfpathlineto{\pgfqpoint{1.563355in}{0.445148in}}%
\pgfpathlineto{\pgfqpoint{1.491455in}{0.445148in}}%
\pgfpathlineto{\pgfqpoint{1.491455in}{0.445327in}}%
\pgfpathlineto{\pgfqpoint{1.383604in}{0.445327in}}%
\pgfpathlineto{\pgfqpoint{1.383604in}{0.445603in}}%
\pgfpathlineto{\pgfqpoint{1.275753in}{0.445603in}}%
\pgfpathlineto{\pgfqpoint{1.275753in}{0.446382in}}%
\pgfpathlineto{\pgfqpoint{1.203853in}{0.446382in}}%
\pgfpathlineto{\pgfqpoint{1.203853in}{0.447704in}}%
\pgfpathlineto{\pgfqpoint{1.131952in}{0.447704in}}%
\pgfpathlineto{\pgfqpoint{1.131952in}{0.448124in}}%
\pgfpathlineto{\pgfqpoint{1.060052in}{0.448124in}}%
\pgfpathlineto{\pgfqpoint{1.060052in}{0.462187in}}%
\pgfpathlineto{\pgfqpoint{0.988151in}{0.462187in}}%
\pgfpathlineto{\pgfqpoint{0.988151in}{0.467308in}}%
\pgfpathlineto{\pgfqpoint{0.916251in}{0.467308in}}%
\pgfpathlineto{\pgfqpoint{0.916251in}{0.591298in}}%
\pgfpathlineto{\pgfqpoint{0.844350in}{0.591298in}}%
\pgfpathlineto{\pgfqpoint{0.844350in}{0.636281in}}%
\pgfpathlineto{\pgfqpoint{0.772450in}{0.636281in}}%
\pgfpathlineto{\pgfqpoint{0.772450in}{0.685923in}}%
\pgfpathlineto{\pgfqpoint{0.700549in}{0.685923in}}%
\pgfpathlineto{\pgfqpoint{0.700549in}{0.764644in}}%
\pgfpathlineto{\pgfqpoint{0.628649in}{0.764644in}}%
\pgfpathlineto{\pgfqpoint{0.628649in}{0.862268in}}%
\pgfpathlineto{\pgfqpoint{0.556748in}{0.862268in}}%
\pgfpathlineto{\pgfqpoint{0.556748in}{0.970311in}}%
\pgfpathlineto{\pgfqpoint{0.520798in}{0.970311in}}%
\pgfpathlineto{\pgfqpoint{0.520798in}{0.970311in}}%
\pgfpathclose%
\pgfusepath{fill}%
\end{pgfscope}%
\begin{pgfscope}%
\pgfpathrectangle{\pgfqpoint{0.520798in}{0.442177in}}{\pgfqpoint{0.719005in}{0.871884in}}%
\pgfusepath{clip}%
\pgfsetroundcap%
\pgfsetroundjoin%
\pgfsetlinewidth{1.003750pt}%
\definecolor{currentstroke}{rgb}{0.007843,0.619608,0.450980}%
\pgfsetstrokecolor{currentstroke}%
\pgfsetdash{}{0pt}%
\pgfpathmoveto{\pgfqpoint{0.520798in}{0.956218in}}%
\pgfpathlineto{\pgfqpoint{0.556748in}{0.956218in}}%
\pgfpathlineto{\pgfqpoint{0.556748in}{0.846768in}}%
\pgfpathlineto{\pgfqpoint{0.628649in}{0.846768in}}%
\pgfpathlineto{\pgfqpoint{0.628649in}{0.756519in}}%
\pgfpathlineto{\pgfqpoint{0.700549in}{0.756519in}}%
\pgfpathlineto{\pgfqpoint{0.700549in}{0.681667in}}%
\pgfpathlineto{\pgfqpoint{0.772450in}{0.681667in}}%
\pgfpathlineto{\pgfqpoint{0.772450in}{0.627684in}}%
\pgfpathlineto{\pgfqpoint{0.844350in}{0.627684in}}%
\pgfpathlineto{\pgfqpoint{0.844350in}{0.584461in}}%
\pgfpathlineto{\pgfqpoint{0.916251in}{0.584461in}}%
\pgfpathlineto{\pgfqpoint{0.916251in}{0.554038in}}%
\pgfpathlineto{\pgfqpoint{0.988151in}{0.554038in}}%
\pgfpathlineto{\pgfqpoint{0.988151in}{0.526675in}}%
\pgfpathlineto{\pgfqpoint{1.060052in}{0.526675in}}%
\pgfpathlineto{\pgfqpoint{1.060052in}{0.508217in}}%
\pgfpathlineto{\pgfqpoint{1.131952in}{0.508217in}}%
\pgfpathlineto{\pgfqpoint{1.131952in}{0.493470in}}%
\pgfpathlineto{\pgfqpoint{1.203853in}{0.493470in}}%
\pgfpathlineto{\pgfqpoint{1.203853in}{0.483545in}}%
\pgfpathlineto{\pgfqpoint{1.241470in}{0.483545in}}%
\pgfusepath{stroke}%
\end{pgfscope}%
\begin{pgfscope}%
\pgfpathrectangle{\pgfqpoint{0.520798in}{0.442177in}}{\pgfqpoint{0.719005in}{0.871884in}}%
\pgfusepath{clip}%
\pgfsetbuttcap%
\pgfsetroundjoin%
\definecolor{currentfill}{rgb}{0.007843,0.619608,0.450980}%
\pgfsetfillcolor{currentfill}%
\pgfsetfillopacity{0.500000}%
\pgfsetlinewidth{0.000000pt}%
\definecolor{currentstroke}{rgb}{0.007843,0.619608,0.450980}%
\pgfsetstrokecolor{currentstroke}%
\pgfsetstrokeopacity{0.500000}%
\pgfsetdash{}{0pt}%
\pgfpathmoveto{\pgfqpoint{0.520798in}{0.970311in}}%
\pgfpathlineto{\pgfqpoint{0.520798in}{0.942125in}}%
\pgfpathlineto{\pgfqpoint{0.556748in}{0.942125in}}%
\pgfpathlineto{\pgfqpoint{0.556748in}{0.831268in}}%
\pgfpathlineto{\pgfqpoint{0.628649in}{0.831268in}}%
\pgfpathlineto{\pgfqpoint{0.628649in}{0.748394in}}%
\pgfpathlineto{\pgfqpoint{0.700549in}{0.748394in}}%
\pgfpathlineto{\pgfqpoint{0.700549in}{0.677410in}}%
\pgfpathlineto{\pgfqpoint{0.772450in}{0.677410in}}%
\pgfpathlineto{\pgfqpoint{0.772450in}{0.619088in}}%
\pgfpathlineto{\pgfqpoint{0.844350in}{0.619088in}}%
\pgfpathlineto{\pgfqpoint{0.844350in}{0.577624in}}%
\pgfpathlineto{\pgfqpoint{0.916251in}{0.577624in}}%
\pgfpathlineto{\pgfqpoint{0.916251in}{0.545881in}}%
\pgfpathlineto{\pgfqpoint{0.988151in}{0.545881in}}%
\pgfpathlineto{\pgfqpoint{0.988151in}{0.518603in}}%
\pgfpathlineto{\pgfqpoint{1.060052in}{0.518603in}}%
\pgfpathlineto{\pgfqpoint{1.060052in}{0.499265in}}%
\pgfpathlineto{\pgfqpoint{1.131952in}{0.499265in}}%
\pgfpathlineto{\pgfqpoint{1.131952in}{0.485829in}}%
\pgfpathlineto{\pgfqpoint{1.203853in}{0.485829in}}%
\pgfpathlineto{\pgfqpoint{1.203853in}{0.477538in}}%
\pgfpathlineto{\pgfqpoint{1.275753in}{0.477538in}}%
\pgfpathlineto{\pgfqpoint{1.275753in}{0.467095in}}%
\pgfpathlineto{\pgfqpoint{1.347654in}{0.467095in}}%
\pgfpathlineto{\pgfqpoint{1.347654in}{0.456187in}}%
\pgfpathlineto{\pgfqpoint{1.419554in}{0.456187in}}%
\pgfpathlineto{\pgfqpoint{1.419554in}{0.454260in}}%
\pgfpathlineto{\pgfqpoint{1.491455in}{0.454260in}}%
\pgfpathlineto{\pgfqpoint{1.491455in}{0.451988in}}%
\pgfpathlineto{\pgfqpoint{1.563355in}{0.451988in}}%
\pgfpathlineto{\pgfqpoint{1.563355in}{0.449678in}}%
\pgfpathlineto{\pgfqpoint{1.635256in}{0.449678in}}%
\pgfpathlineto{\pgfqpoint{1.635256in}{0.444206in}}%
\pgfpathlineto{\pgfqpoint{1.707156in}{0.444206in}}%
\pgfpathlineto{\pgfqpoint{1.707156in}{0.443699in}}%
\pgfpathlineto{\pgfqpoint{1.779057in}{0.443699in}}%
\pgfpathlineto{\pgfqpoint{1.779057in}{0.442726in}}%
\pgfpathlineto{\pgfqpoint{1.850957in}{0.442726in}}%
\pgfpathlineto{\pgfqpoint{1.850957in}{0.442760in}}%
\pgfpathlineto{\pgfqpoint{1.922858in}{0.442760in}}%
\pgfpathlineto{\pgfqpoint{1.922858in}{0.442763in}}%
\pgfpathlineto{\pgfqpoint{1.994758in}{0.442763in}}%
\pgfpathlineto{\pgfqpoint{1.994758in}{0.442646in}}%
\pgfpathlineto{\pgfqpoint{2.066659in}{0.442646in}}%
\pgfpathlineto{\pgfqpoint{2.066659in}{0.442480in}}%
\pgfpathlineto{\pgfqpoint{2.174510in}{0.442480in}}%
\pgfpathlineto{\pgfqpoint{2.174510in}{0.442256in}}%
\pgfpathlineto{\pgfqpoint{2.282360in}{0.442256in}}%
\pgfpathlineto{\pgfqpoint{2.282360in}{0.442249in}}%
\pgfpathlineto{\pgfqpoint{2.354261in}{0.442249in}}%
\pgfpathlineto{\pgfqpoint{2.354261in}{0.442362in}}%
\pgfpathlineto{\pgfqpoint{2.426161in}{0.442362in}}%
\pgfpathlineto{\pgfqpoint{2.426161in}{0.442353in}}%
\pgfpathlineto{\pgfqpoint{2.498062in}{0.442353in}}%
\pgfpathlineto{\pgfqpoint{2.498062in}{0.442386in}}%
\pgfpathlineto{\pgfqpoint{2.605913in}{0.442386in}}%
\pgfpathlineto{\pgfqpoint{2.605913in}{0.442386in}}%
\pgfpathlineto{\pgfqpoint{2.677813in}{0.442386in}}%
\pgfpathlineto{\pgfqpoint{2.677813in}{0.443080in}}%
\pgfpathlineto{\pgfqpoint{2.677813in}{0.443080in}}%
\pgfpathlineto{\pgfqpoint{2.605913in}{0.443080in}}%
\pgfpathlineto{\pgfqpoint{2.605913in}{0.443080in}}%
\pgfpathlineto{\pgfqpoint{2.498062in}{0.443080in}}%
\pgfpathlineto{\pgfqpoint{2.498062in}{0.443299in}}%
\pgfpathlineto{\pgfqpoint{2.426161in}{0.443299in}}%
\pgfpathlineto{\pgfqpoint{2.426161in}{0.443475in}}%
\pgfpathlineto{\pgfqpoint{2.354261in}{0.443475in}}%
\pgfpathlineto{\pgfqpoint{2.354261in}{0.444331in}}%
\pgfpathlineto{\pgfqpoint{2.282360in}{0.444331in}}%
\pgfpathlineto{\pgfqpoint{2.282360in}{0.444880in}}%
\pgfpathlineto{\pgfqpoint{2.174510in}{0.444880in}}%
\pgfpathlineto{\pgfqpoint{2.174510in}{0.445584in}}%
\pgfpathlineto{\pgfqpoint{2.066659in}{0.445584in}}%
\pgfpathlineto{\pgfqpoint{2.066659in}{0.445603in}}%
\pgfpathlineto{\pgfqpoint{1.994758in}{0.445603in}}%
\pgfpathlineto{\pgfqpoint{1.994758in}{0.445672in}}%
\pgfpathlineto{\pgfqpoint{1.922858in}{0.445672in}}%
\pgfpathlineto{\pgfqpoint{1.922858in}{0.446231in}}%
\pgfpathlineto{\pgfqpoint{1.850957in}{0.446231in}}%
\pgfpathlineto{\pgfqpoint{1.850957in}{0.447193in}}%
\pgfpathlineto{\pgfqpoint{1.779057in}{0.447193in}}%
\pgfpathlineto{\pgfqpoint{1.779057in}{0.447704in}}%
\pgfpathlineto{\pgfqpoint{1.707156in}{0.447704in}}%
\pgfpathlineto{\pgfqpoint{1.707156in}{0.448124in}}%
\pgfpathlineto{\pgfqpoint{1.635256in}{0.448124in}}%
\pgfpathlineto{\pgfqpoint{1.635256in}{0.456195in}}%
\pgfpathlineto{\pgfqpoint{1.563355in}{0.456195in}}%
\pgfpathlineto{\pgfqpoint{1.563355in}{0.458151in}}%
\pgfpathlineto{\pgfqpoint{1.491455in}{0.458151in}}%
\pgfpathlineto{\pgfqpoint{1.491455in}{0.462187in}}%
\pgfpathlineto{\pgfqpoint{1.419554in}{0.462187in}}%
\pgfpathlineto{\pgfqpoint{1.419554in}{0.467308in}}%
\pgfpathlineto{\pgfqpoint{1.347654in}{0.467308in}}%
\pgfpathlineto{\pgfqpoint{1.347654in}{0.481815in}}%
\pgfpathlineto{\pgfqpoint{1.275753in}{0.481815in}}%
\pgfpathlineto{\pgfqpoint{1.275753in}{0.489552in}}%
\pgfpathlineto{\pgfqpoint{1.203853in}{0.489552in}}%
\pgfpathlineto{\pgfqpoint{1.203853in}{0.501110in}}%
\pgfpathlineto{\pgfqpoint{1.131952in}{0.501110in}}%
\pgfpathlineto{\pgfqpoint{1.131952in}{0.517170in}}%
\pgfpathlineto{\pgfqpoint{1.060052in}{0.517170in}}%
\pgfpathlineto{\pgfqpoint{1.060052in}{0.534748in}}%
\pgfpathlineto{\pgfqpoint{0.988151in}{0.534748in}}%
\pgfpathlineto{\pgfqpoint{0.988151in}{0.562195in}}%
\pgfpathlineto{\pgfqpoint{0.916251in}{0.562195in}}%
\pgfpathlineto{\pgfqpoint{0.916251in}{0.591298in}}%
\pgfpathlineto{\pgfqpoint{0.844350in}{0.591298in}}%
\pgfpathlineto{\pgfqpoint{0.844350in}{0.636281in}}%
\pgfpathlineto{\pgfqpoint{0.772450in}{0.636281in}}%
\pgfpathlineto{\pgfqpoint{0.772450in}{0.685923in}}%
\pgfpathlineto{\pgfqpoint{0.700549in}{0.685923in}}%
\pgfpathlineto{\pgfqpoint{0.700549in}{0.764644in}}%
\pgfpathlineto{\pgfqpoint{0.628649in}{0.764644in}}%
\pgfpathlineto{\pgfqpoint{0.628649in}{0.862268in}}%
\pgfpathlineto{\pgfqpoint{0.556748in}{0.862268in}}%
\pgfpathlineto{\pgfqpoint{0.556748in}{0.970311in}}%
\pgfpathlineto{\pgfqpoint{0.520798in}{0.970311in}}%
\pgfpathlineto{\pgfqpoint{0.520798in}{0.970311in}}%
\pgfpathclose%
\pgfusepath{fill}%
\end{pgfscope}%
\begin{pgfscope}%
\definecolor{textcolor}{rgb}{0.150000,0.150000,0.150000}%
\pgfsetstrokecolor{textcolor}%
\pgfsetfillcolor{textcolor}%
\pgftext[x=0.880300in,y=1.397395in,,base]{\color{textcolor}\rmfamily\fontsize{9.000000}{10.800000}\selectfont \(\displaystyle c_A^-=1\)}%
\end{pgfscope}%
\begin{pgfscope}%
\pgfsetbuttcap%
\pgfsetmiterjoin%
\definecolor{currentfill}{rgb}{1.000000,1.000000,1.000000}%
\pgfsetfillcolor{currentfill}%
\pgfsetfillopacity{0.800000}%
\pgfsetlinewidth{1.003750pt}%
\definecolor{currentstroke}{rgb}{0.800000,0.800000,0.800000}%
\pgfsetstrokecolor{currentstroke}%
\pgfsetstrokeopacity{0.800000}%
\pgfsetdash{}{0pt}%
\pgfpathmoveto{\pgfqpoint{0.785843in}{0.638103in}}%
\pgfpathlineto{\pgfqpoint{1.191192in}{0.638103in}}%
\pgfpathlineto{\pgfqpoint{1.191192in}{1.265450in}}%
\pgfpathlineto{\pgfqpoint{0.785843in}{1.265450in}}%
\pgfpathlineto{\pgfqpoint{0.785843in}{0.638103in}}%
\pgfpathclose%
\pgfusepath{stroke,fill}%
\end{pgfscope}%
\begin{pgfscope}%
\definecolor{textcolor}{rgb}{0.150000,0.150000,0.150000}%
\pgfsetstrokecolor{textcolor}%
\pgfsetfillcolor{textcolor}%
\pgftext[x=0.918271in,y=1.113275in,left,base]{\color{textcolor}\rmfamily\fontsize{9.000000}{10.800000}\selectfont \(\displaystyle c_A^+\)}%
\end{pgfscope}%
\begin{pgfscope}%
\pgfsetroundcap%
\pgfsetroundjoin%
\pgfsetlinewidth{1.003750pt}%
\definecolor{currentstroke}{rgb}{0.003922,0.450980,0.698039}%
\pgfsetstrokecolor{currentstroke}%
\pgfsetdash{}{0pt}%
\pgfpathmoveto{\pgfqpoint{0.824732in}{1.001052in}}%
\pgfpathlineto{\pgfqpoint{0.873343in}{1.001052in}}%
\pgfpathlineto{\pgfqpoint{0.873343in}{1.001052in}}%
\pgfpathlineto{\pgfqpoint{0.970565in}{1.001052in}}%
\pgfpathlineto{\pgfqpoint{0.970565in}{1.001052in}}%
\pgfpathlineto{\pgfqpoint{1.019176in}{1.001052in}}%
\pgfusepath{stroke}%
\end{pgfscope}%
\begin{pgfscope}%
\definecolor{textcolor}{rgb}{0.150000,0.150000,0.150000}%
\pgfsetstrokecolor{textcolor}%
\pgfsetfillcolor{textcolor}%
\pgftext[x=1.096954in,y=0.967024in,left,base]{\color{textcolor}\rmfamily\fontsize{7.000000}{8.400000}\selectfont 1}%
\end{pgfscope}%
\begin{pgfscope}%
\pgfsetroundcap%
\pgfsetroundjoin%
\pgfsetlinewidth{1.003750pt}%
\definecolor{currentstroke}{rgb}{0.870588,0.560784,0.019608}%
\pgfsetstrokecolor{currentstroke}%
\pgfsetdash{}{0pt}%
\pgfpathmoveto{\pgfqpoint{0.824732in}{0.865486in}}%
\pgfpathlineto{\pgfqpoint{0.873343in}{0.865486in}}%
\pgfpathlineto{\pgfqpoint{0.873343in}{0.865486in}}%
\pgfpathlineto{\pgfqpoint{0.970565in}{0.865486in}}%
\pgfpathlineto{\pgfqpoint{0.970565in}{0.865486in}}%
\pgfpathlineto{\pgfqpoint{1.019176in}{0.865486in}}%
\pgfusepath{stroke}%
\end{pgfscope}%
\begin{pgfscope}%
\definecolor{textcolor}{rgb}{0.150000,0.150000,0.150000}%
\pgfsetstrokecolor{textcolor}%
\pgfsetfillcolor{textcolor}%
\pgftext[x=1.096954in,y=0.831458in,left,base]{\color{textcolor}\rmfamily\fontsize{7.000000}{8.400000}\selectfont 2}%
\end{pgfscope}%
\begin{pgfscope}%
\pgfsetroundcap%
\pgfsetroundjoin%
\pgfsetlinewidth{1.003750pt}%
\definecolor{currentstroke}{rgb}{0.007843,0.619608,0.450980}%
\pgfsetstrokecolor{currentstroke}%
\pgfsetdash{}{0pt}%
\pgfpathmoveto{\pgfqpoint{0.824732in}{0.729919in}}%
\pgfpathlineto{\pgfqpoint{0.873343in}{0.729919in}}%
\pgfpathlineto{\pgfqpoint{0.873343in}{0.729919in}}%
\pgfpathlineto{\pgfqpoint{0.970565in}{0.729919in}}%
\pgfpathlineto{\pgfqpoint{0.970565in}{0.729919in}}%
\pgfpathlineto{\pgfqpoint{1.019176in}{0.729919in}}%
\pgfusepath{stroke}%
\end{pgfscope}%
\begin{pgfscope}%
\definecolor{textcolor}{rgb}{0.150000,0.150000,0.150000}%
\pgfsetstrokecolor{textcolor}%
\pgfsetfillcolor{textcolor}%
\pgftext[x=1.096954in,y=0.695892in,left,base]{\color{textcolor}\rmfamily\fontsize{7.000000}{8.400000}\selectfont 4}%
\end{pgfscope}%
\end{pgfpicture}%
\makeatother%
\endgroup%

%% file: figures/experiments/graphs/sparse_smoothing/nodes_structure/node_classification-Citeseer-GAT-hidden=8-p_adj_plus=0.001-p_adj_minus=0.8-p_att_plus=0.0-p_att_minus=0.0-multi_class_cert-A.pgf
\begingroup%
\makeatletter%
\begin{pgfpicture}%
\pgfpathrectangle{\pgfpointorigin}{\pgfqpoint{1.375000in}{1.581250in}}%
\pgfusepath{use as bounding box, clip}%
\begin{pgfscope}%
\pgfsetbuttcap%
\pgfsetmiterjoin%
\definecolor{currentfill}{rgb}{1.000000,1.000000,1.000000}%
\pgfsetfillcolor{currentfill}%
\pgfsetlinewidth{0.000000pt}%
\definecolor{currentstroke}{rgb}{1.000000,1.000000,1.000000}%
\pgfsetstrokecolor{currentstroke}%
\pgfsetdash{}{0pt}%
\pgfpathmoveto{\pgfqpoint{0.000000in}{0.000000in}}%
\pgfpathlineto{\pgfqpoint{1.375000in}{0.000000in}}%
\pgfpathlineto{\pgfqpoint{1.375000in}{1.581250in}}%
\pgfpathlineto{\pgfqpoint{0.000000in}{1.581250in}}%
\pgfpathlineto{\pgfqpoint{0.000000in}{0.000000in}}%
\pgfpathclose%
\pgfusepath{fill}%
\end{pgfscope}%
\begin{pgfscope}%
\pgfsetbuttcap%
\pgfsetmiterjoin%
\definecolor{currentfill}{rgb}{1.000000,1.000000,1.000000}%
\pgfsetfillcolor{currentfill}%
\pgfsetlinewidth{0.000000pt}%
\definecolor{currentstroke}{rgb}{0.000000,0.000000,0.000000}%
\pgfsetstrokecolor{currentstroke}%
\pgfsetstrokeopacity{0.000000}%
\pgfsetdash{}{0pt}%
\pgfpathmoveto{\pgfqpoint{0.520798in}{0.442177in}}%
\pgfpathlineto{\pgfqpoint{1.239803in}{0.442177in}}%
\pgfpathlineto{\pgfqpoint{1.239803in}{1.314061in}}%
\pgfpathlineto{\pgfqpoint{0.520798in}{1.314061in}}%
\pgfpathlineto{\pgfqpoint{0.520798in}{0.442177in}}%
\pgfpathclose%
\pgfusepath{fill}%
\end{pgfscope}%
\begin{pgfscope}%
\pgfpathrectangle{\pgfqpoint{0.520798in}{0.442177in}}{\pgfqpoint{0.719005in}{0.871884in}}%
\pgfusepath{clip}%
\pgfsetroundcap%
\pgfsetroundjoin%
\pgfsetlinewidth{0.501875pt}%
\definecolor{currentstroke}{rgb}{0.800000,0.800000,0.800000}%
\pgfsetstrokecolor{currentstroke}%
\pgfsetdash{}{0pt}%
\pgfpathmoveto{\pgfqpoint{0.520798in}{0.442177in}}%
\pgfpathlineto{\pgfqpoint{0.520798in}{1.314061in}}%
\pgfusepath{stroke}%
\end{pgfscope}%
\begin{pgfscope}%
\definecolor{textcolor}{rgb}{0.150000,0.150000,0.150000}%
\pgfsetstrokecolor{textcolor}%
\pgfsetfillcolor{textcolor}%
\pgftext[x=0.520798in,y=0.351899in,,top]{\color{textcolor}\rmfamily\fontsize{8.000000}{9.600000}\selectfont \(\displaystyle {0}\)}%
\end{pgfscope}%
\begin{pgfscope}%
\pgfpathrectangle{\pgfqpoint{0.520798in}{0.442177in}}{\pgfqpoint{0.719005in}{0.871884in}}%
\pgfusepath{clip}%
\pgfsetroundcap%
\pgfsetroundjoin%
\pgfsetlinewidth{0.501875pt}%
\definecolor{currentstroke}{rgb}{0.800000,0.800000,0.800000}%
\pgfsetstrokecolor{currentstroke}%
\pgfsetdash{}{0pt}%
\pgfpathmoveto{\pgfqpoint{0.880300in}{0.442177in}}%
\pgfpathlineto{\pgfqpoint{0.880300in}{1.314061in}}%
\pgfusepath{stroke}%
\end{pgfscope}%
\begin{pgfscope}%
\definecolor{textcolor}{rgb}{0.150000,0.150000,0.150000}%
\pgfsetstrokecolor{textcolor}%
\pgfsetfillcolor{textcolor}%
\pgftext[x=0.880300in,y=0.351899in,,top]{\color{textcolor}\rmfamily\fontsize{8.000000}{9.600000}\selectfont \(\displaystyle {5}\)}%
\end{pgfscope}%
\begin{pgfscope}%
\pgfpathrectangle{\pgfqpoint{0.520798in}{0.442177in}}{\pgfqpoint{0.719005in}{0.871884in}}%
\pgfusepath{clip}%
\pgfsetroundcap%
\pgfsetroundjoin%
\pgfsetlinewidth{0.501875pt}%
\definecolor{currentstroke}{rgb}{0.800000,0.800000,0.800000}%
\pgfsetstrokecolor{currentstroke}%
\pgfsetdash{}{0pt}%
\pgfpathmoveto{\pgfqpoint{1.239803in}{0.442177in}}%
\pgfpathlineto{\pgfqpoint{1.239803in}{1.314061in}}%
\pgfusepath{stroke}%
\end{pgfscope}%
\begin{pgfscope}%
\definecolor{textcolor}{rgb}{0.150000,0.150000,0.150000}%
\pgfsetstrokecolor{textcolor}%
\pgfsetfillcolor{textcolor}%
\pgftext[x=1.239803in,y=0.351899in,,top]{\color{textcolor}\rmfamily\fontsize{8.000000}{9.600000}\selectfont \(\displaystyle {10}\)}%
\end{pgfscope}%
\begin{pgfscope}%
\definecolor{textcolor}{rgb}{0.150000,0.150000,0.150000}%
\pgfsetstrokecolor{textcolor}%
\pgfsetfillcolor{textcolor}%
\pgftext[x=0.880300in,y=0.198219in,,top]{\color{textcolor}\rmfamily\fontsize{10.000000}{12.000000}\selectfont Edit distance \(\displaystyle \epsilon\)}%
\end{pgfscope}%
\begin{pgfscope}%
\pgfpathrectangle{\pgfqpoint{0.520798in}{0.442177in}}{\pgfqpoint{0.719005in}{0.871884in}}%
\pgfusepath{clip}%
\pgfsetroundcap%
\pgfsetroundjoin%
\pgfsetlinewidth{0.501875pt}%
\definecolor{currentstroke}{rgb}{0.800000,0.800000,0.800000}%
\pgfsetstrokecolor{currentstroke}%
\pgfsetdash{}{0pt}%
\pgfpathmoveto{\pgfqpoint{0.520798in}{0.442177in}}%
\pgfpathlineto{\pgfqpoint{1.239803in}{0.442177in}}%
\pgfusepath{stroke}%
\end{pgfscope}%
\begin{pgfscope}%
\definecolor{textcolor}{rgb}{0.150000,0.150000,0.150000}%
\pgfsetstrokecolor{textcolor}%
\pgfsetfillcolor{textcolor}%
\pgftext[x=0.273151in, y=0.403915in, left, base]{\color{textcolor}\rmfamily\fontsize{8.000000}{9.600000}\selectfont 0\%}%
\end{pgfscope}%
\begin{pgfscope}%
\pgfpathrectangle{\pgfqpoint{0.520798in}{0.442177in}}{\pgfqpoint{0.719005in}{0.871884in}}%
\pgfusepath{clip}%
\pgfsetroundcap%
\pgfsetroundjoin%
\pgfsetlinewidth{0.501875pt}%
\definecolor{currentstroke}{rgb}{0.800000,0.800000,0.800000}%
\pgfsetstrokecolor{currentstroke}%
\pgfsetdash{}{0pt}%
\pgfpathmoveto{\pgfqpoint{0.520798in}{0.660148in}}%
\pgfpathlineto{\pgfqpoint{1.239803in}{0.660148in}}%
\pgfusepath{stroke}%
\end{pgfscope}%
\begin{pgfscope}%
\definecolor{textcolor}{rgb}{0.150000,0.150000,0.150000}%
\pgfsetstrokecolor{textcolor}%
\pgfsetfillcolor{textcolor}%
\pgftext[x=0.214138in, y=0.621886in, left, base]{\color{textcolor}\rmfamily\fontsize{8.000000}{9.600000}\selectfont 25\%}%
\end{pgfscope}%
\begin{pgfscope}%
\pgfpathrectangle{\pgfqpoint{0.520798in}{0.442177in}}{\pgfqpoint{0.719005in}{0.871884in}}%
\pgfusepath{clip}%
\pgfsetroundcap%
\pgfsetroundjoin%
\pgfsetlinewidth{0.501875pt}%
\definecolor{currentstroke}{rgb}{0.800000,0.800000,0.800000}%
\pgfsetstrokecolor{currentstroke}%
\pgfsetdash{}{0pt}%
\pgfpathmoveto{\pgfqpoint{0.520798in}{0.878119in}}%
\pgfpathlineto{\pgfqpoint{1.239803in}{0.878119in}}%
\pgfusepath{stroke}%
\end{pgfscope}%
\begin{pgfscope}%
\definecolor{textcolor}{rgb}{0.150000,0.150000,0.150000}%
\pgfsetstrokecolor{textcolor}%
\pgfsetfillcolor{textcolor}%
\pgftext[x=0.214138in, y=0.839857in, left, base]{\color{textcolor}\rmfamily\fontsize{8.000000}{9.600000}\selectfont 50\%}%
\end{pgfscope}%
\begin{pgfscope}%
\pgfpathrectangle{\pgfqpoint{0.520798in}{0.442177in}}{\pgfqpoint{0.719005in}{0.871884in}}%
\pgfusepath{clip}%
\pgfsetroundcap%
\pgfsetroundjoin%
\pgfsetlinewidth{0.501875pt}%
\definecolor{currentstroke}{rgb}{0.800000,0.800000,0.800000}%
\pgfsetstrokecolor{currentstroke}%
\pgfsetdash{}{0pt}%
\pgfpathmoveto{\pgfqpoint{0.520798in}{1.096090in}}%
\pgfpathlineto{\pgfqpoint{1.239803in}{1.096090in}}%
\pgfusepath{stroke}%
\end{pgfscope}%
\begin{pgfscope}%
\definecolor{textcolor}{rgb}{0.150000,0.150000,0.150000}%
\pgfsetstrokecolor{textcolor}%
\pgfsetfillcolor{textcolor}%
\pgftext[x=0.214138in, y=1.057828in, left, base]{\color{textcolor}\rmfamily\fontsize{8.000000}{9.600000}\selectfont 75\%}%
\end{pgfscope}%
\begin{pgfscope}%
\pgfpathrectangle{\pgfqpoint{0.520798in}{0.442177in}}{\pgfqpoint{0.719005in}{0.871884in}}%
\pgfusepath{clip}%
\pgfsetroundcap%
\pgfsetroundjoin%
\pgfsetlinewidth{0.501875pt}%
\definecolor{currentstroke}{rgb}{0.800000,0.800000,0.800000}%
\pgfsetstrokecolor{currentstroke}%
\pgfsetdash{}{0pt}%
\pgfpathmoveto{\pgfqpoint{0.520798in}{1.314061in}}%
\pgfpathlineto{\pgfqpoint{1.239803in}{1.314061in}}%
\pgfusepath{stroke}%
\end{pgfscope}%
\begin{pgfscope}%
\definecolor{textcolor}{rgb}{0.150000,0.150000,0.150000}%
\pgfsetstrokecolor{textcolor}%
\pgfsetfillcolor{textcolor}%
\pgftext[x=0.155124in, y=1.275799in, left, base]{\color{textcolor}\rmfamily\fontsize{8.000000}{9.600000}\selectfont 100\%}%
\end{pgfscope}%
\begin{pgfscope}%
\definecolor{textcolor}{rgb}{0.150000,0.150000,0.150000}%
\pgfsetstrokecolor{textcolor}%
\pgfsetfillcolor{textcolor}%
\pgftext[x=0.099569in,y=0.878119in,,bottom,rotate=90.000000]{\color{textcolor}\rmfamily\fontsize{10.000000}{12.000000}\selectfont Cert. Acc.}%
\end{pgfscope}%
\begin{pgfscope}%
\pgfsetrectcap%
\pgfsetmiterjoin%
\pgfsetlinewidth{0.752812pt}%
\definecolor{currentstroke}{rgb}{0.700000,0.700000,0.700000}%
\pgfsetstrokecolor{currentstroke}%
\pgfsetdash{}{0pt}%
\pgfpathmoveto{\pgfqpoint{0.520798in}{0.442177in}}%
\pgfpathlineto{\pgfqpoint{0.520798in}{1.314061in}}%
\pgfusepath{stroke}%
\end{pgfscope}%
\begin{pgfscope}%
\pgfsetrectcap%
\pgfsetmiterjoin%
\pgfsetlinewidth{0.752812pt}%
\definecolor{currentstroke}{rgb}{0.700000,0.700000,0.700000}%
\pgfsetstrokecolor{currentstroke}%
\pgfsetdash{}{0pt}%
\pgfpathmoveto{\pgfqpoint{1.239803in}{0.442177in}}%
\pgfpathlineto{\pgfqpoint{1.239803in}{1.314061in}}%
\pgfusepath{stroke}%
\end{pgfscope}%
\begin{pgfscope}%
\pgfsetrectcap%
\pgfsetmiterjoin%
\pgfsetlinewidth{0.752812pt}%
\definecolor{currentstroke}{rgb}{0.700000,0.700000,0.700000}%
\pgfsetstrokecolor{currentstroke}%
\pgfsetdash{}{0pt}%
\pgfpathmoveto{\pgfqpoint{0.520798in}{0.442177in}}%
\pgfpathlineto{\pgfqpoint{1.239803in}{0.442177in}}%
\pgfusepath{stroke}%
\end{pgfscope}%
\begin{pgfscope}%
\pgfsetrectcap%
\pgfsetmiterjoin%
\pgfsetlinewidth{0.752812pt}%
\definecolor{currentstroke}{rgb}{0.700000,0.700000,0.700000}%
\pgfsetstrokecolor{currentstroke}%
\pgfsetdash{}{0pt}%
\pgfpathmoveto{\pgfqpoint{0.520798in}{1.314061in}}%
\pgfpathlineto{\pgfqpoint{1.239803in}{1.314061in}}%
\pgfusepath{stroke}%
\end{pgfscope}%
\begin{pgfscope}%
\pgfpathrectangle{\pgfqpoint{0.520798in}{0.442177in}}{\pgfqpoint{0.719005in}{0.871884in}}%
\pgfusepath{clip}%
\pgfsetroundcap%
\pgfsetroundjoin%
\pgfsetlinewidth{1.003750pt}%
\definecolor{currentstroke}{rgb}{0.003922,0.450980,0.698039}%
\pgfsetstrokecolor{currentstroke}%
\pgfsetdash{}{0pt}%
\pgfpathmoveto{\pgfqpoint{0.520798in}{0.956218in}}%
\pgfpathlineto{\pgfqpoint{0.556748in}{0.956218in}}%
\pgfpathlineto{\pgfqpoint{0.556748in}{0.818849in}}%
\pgfpathlineto{\pgfqpoint{0.628649in}{0.818849in}}%
\pgfpathlineto{\pgfqpoint{0.628649in}{0.642339in}}%
\pgfpathlineto{\pgfqpoint{0.700549in}{0.642339in}}%
\pgfpathlineto{\pgfqpoint{0.700549in}{0.461748in}}%
\pgfpathlineto{\pgfqpoint{0.772450in}{0.461748in}}%
\pgfpathlineto{\pgfqpoint{0.772450in}{0.446165in}}%
\pgfpathlineto{\pgfqpoint{0.844350in}{0.446165in}}%
\pgfpathlineto{\pgfqpoint{0.844350in}{0.444681in}}%
\pgfpathlineto{\pgfqpoint{0.916251in}{0.444681in}}%
\pgfpathlineto{\pgfqpoint{0.916251in}{0.444403in}}%
\pgfpathlineto{\pgfqpoint{0.988151in}{0.444403in}}%
\pgfpathlineto{\pgfqpoint{0.988151in}{0.443939in}}%
\pgfpathlineto{\pgfqpoint{1.060052in}{0.443939in}}%
\pgfpathlineto{\pgfqpoint{1.060052in}{0.442455in}}%
\pgfpathlineto{\pgfqpoint{1.167902in}{0.442455in}}%
\pgfpathlineto{\pgfqpoint{1.167902in}{0.442362in}}%
\pgfpathlineto{\pgfqpoint{1.241470in}{0.442362in}}%
\pgfusepath{stroke}%
\end{pgfscope}%
\begin{pgfscope}%
\pgfpathrectangle{\pgfqpoint{0.520798in}{0.442177in}}{\pgfqpoint{0.719005in}{0.871884in}}%
\pgfusepath{clip}%
\pgfsetbuttcap%
\pgfsetroundjoin%
\definecolor{currentfill}{rgb}{0.003922,0.450980,0.698039}%
\pgfsetfillcolor{currentfill}%
\pgfsetfillopacity{0.500000}%
\pgfsetlinewidth{0.000000pt}%
\definecolor{currentstroke}{rgb}{0.003922,0.450980,0.698039}%
\pgfsetstrokecolor{currentstroke}%
\pgfsetstrokeopacity{0.500000}%
\pgfsetdash{}{0pt}%
\pgfpathmoveto{\pgfqpoint{0.520798in}{0.970311in}}%
\pgfpathlineto{\pgfqpoint{0.520798in}{0.942125in}}%
\pgfpathlineto{\pgfqpoint{0.556748in}{0.942125in}}%
\pgfpathlineto{\pgfqpoint{0.556748in}{0.803775in}}%
\pgfpathlineto{\pgfqpoint{0.628649in}{0.803775in}}%
\pgfpathlineto{\pgfqpoint{0.628649in}{0.636253in}}%
\pgfpathlineto{\pgfqpoint{0.700549in}{0.636253in}}%
\pgfpathlineto{\pgfqpoint{0.700549in}{0.456187in}}%
\pgfpathlineto{\pgfqpoint{0.772450in}{0.456187in}}%
\pgfpathlineto{\pgfqpoint{0.772450in}{0.444206in}}%
\pgfpathlineto{\pgfqpoint{0.844350in}{0.444206in}}%
\pgfpathlineto{\pgfqpoint{0.844350in}{0.442981in}}%
\pgfpathlineto{\pgfqpoint{0.916251in}{0.442981in}}%
\pgfpathlineto{\pgfqpoint{0.916251in}{0.442786in}}%
\pgfpathlineto{\pgfqpoint{0.988151in}{0.442786in}}%
\pgfpathlineto{\pgfqpoint{0.988151in}{0.442551in}}%
\pgfpathlineto{\pgfqpoint{1.060052in}{0.442551in}}%
\pgfpathlineto{\pgfqpoint{1.060052in}{0.442228in}}%
\pgfpathlineto{\pgfqpoint{1.167902in}{0.442228in}}%
\pgfpathlineto{\pgfqpoint{1.167902in}{0.442135in}}%
\pgfpathlineto{\pgfqpoint{1.311703in}{0.442135in}}%
\pgfpathlineto{\pgfqpoint{1.311703in}{0.442084in}}%
\pgfpathlineto{\pgfqpoint{1.419554in}{0.442084in}}%
\pgfpathlineto{\pgfqpoint{1.419554in}{0.442177in}}%
\pgfpathlineto{\pgfqpoint{2.066659in}{0.442177in}}%
\pgfpathlineto{\pgfqpoint{2.066659in}{0.442177in}}%
\pgfpathlineto{\pgfqpoint{2.677813in}{0.442177in}}%
\pgfpathlineto{\pgfqpoint{2.677813in}{0.442177in}}%
\pgfpathlineto{\pgfqpoint{2.677813in}{0.442177in}}%
\pgfpathlineto{\pgfqpoint{2.066659in}{0.442177in}}%
\pgfpathlineto{\pgfqpoint{2.066659in}{0.442177in}}%
\pgfpathlineto{\pgfqpoint{1.419554in}{0.442177in}}%
\pgfpathlineto{\pgfqpoint{1.419554in}{0.442455in}}%
\pgfpathlineto{\pgfqpoint{1.311703in}{0.442455in}}%
\pgfpathlineto{\pgfqpoint{1.311703in}{0.442590in}}%
\pgfpathlineto{\pgfqpoint{1.167902in}{0.442590in}}%
\pgfpathlineto{\pgfqpoint{1.167902in}{0.442682in}}%
\pgfpathlineto{\pgfqpoint{1.060052in}{0.442682in}}%
\pgfpathlineto{\pgfqpoint{1.060052in}{0.445327in}}%
\pgfpathlineto{\pgfqpoint{0.988151in}{0.445327in}}%
\pgfpathlineto{\pgfqpoint{0.988151in}{0.446020in}}%
\pgfpathlineto{\pgfqpoint{0.916251in}{0.446020in}}%
\pgfpathlineto{\pgfqpoint{0.916251in}{0.446381in}}%
\pgfpathlineto{\pgfqpoint{0.844350in}{0.446381in}}%
\pgfpathlineto{\pgfqpoint{0.844350in}{0.448124in}}%
\pgfpathlineto{\pgfqpoint{0.772450in}{0.448124in}}%
\pgfpathlineto{\pgfqpoint{0.772450in}{0.467308in}}%
\pgfpathlineto{\pgfqpoint{0.700549in}{0.467308in}}%
\pgfpathlineto{\pgfqpoint{0.700549in}{0.648426in}}%
\pgfpathlineto{\pgfqpoint{0.628649in}{0.648426in}}%
\pgfpathlineto{\pgfqpoint{0.628649in}{0.833924in}}%
\pgfpathlineto{\pgfqpoint{0.556748in}{0.833924in}}%
\pgfpathlineto{\pgfqpoint{0.556748in}{0.970311in}}%
\pgfpathlineto{\pgfqpoint{0.520798in}{0.970311in}}%
\pgfpathlineto{\pgfqpoint{0.520798in}{0.970311in}}%
\pgfpathclose%
\pgfusepath{fill}%
\end{pgfscope}%
\begin{pgfscope}%
\pgfpathrectangle{\pgfqpoint{0.520798in}{0.442177in}}{\pgfqpoint{0.719005in}{0.871884in}}%
\pgfusepath{clip}%
\pgfsetroundcap%
\pgfsetroundjoin%
\pgfsetlinewidth{1.003750pt}%
\definecolor{currentstroke}{rgb}{0.870588,0.560784,0.019608}%
\pgfsetstrokecolor{currentstroke}%
\pgfsetdash{}{0pt}%
\pgfpathmoveto{\pgfqpoint{0.520798in}{0.956218in}}%
\pgfpathlineto{\pgfqpoint{0.556748in}{0.956218in}}%
\pgfpathlineto{\pgfqpoint{0.556748in}{0.818849in}}%
\pgfpathlineto{\pgfqpoint{0.628649in}{0.818849in}}%
\pgfpathlineto{\pgfqpoint{0.628649in}{0.642339in}}%
\pgfpathlineto{\pgfqpoint{0.700549in}{0.642339in}}%
\pgfpathlineto{\pgfqpoint{0.700549in}{0.461748in}}%
\pgfpathlineto{\pgfqpoint{0.772450in}{0.461748in}}%
\pgfpathlineto{\pgfqpoint{0.772450in}{0.446165in}}%
\pgfpathlineto{\pgfqpoint{0.844350in}{0.446165in}}%
\pgfpathlineto{\pgfqpoint{0.844350in}{0.444774in}}%
\pgfpathlineto{\pgfqpoint{0.916251in}{0.444774in}}%
\pgfpathlineto{\pgfqpoint{0.916251in}{0.444403in}}%
\pgfpathlineto{\pgfqpoint{0.988151in}{0.444403in}}%
\pgfpathlineto{\pgfqpoint{0.988151in}{0.443939in}}%
\pgfpathlineto{\pgfqpoint{1.060052in}{0.443939in}}%
\pgfpathlineto{\pgfqpoint{1.060052in}{0.442455in}}%
\pgfpathlineto{\pgfqpoint{1.167902in}{0.442455in}}%
\pgfpathlineto{\pgfqpoint{1.167902in}{0.442362in}}%
\pgfpathlineto{\pgfqpoint{1.241470in}{0.442362in}}%
\pgfusepath{stroke}%
\end{pgfscope}%
\begin{pgfscope}%
\pgfpathrectangle{\pgfqpoint{0.520798in}{0.442177in}}{\pgfqpoint{0.719005in}{0.871884in}}%
\pgfusepath{clip}%
\pgfsetbuttcap%
\pgfsetroundjoin%
\definecolor{currentfill}{rgb}{0.870588,0.560784,0.019608}%
\pgfsetfillcolor{currentfill}%
\pgfsetfillopacity{0.500000}%
\pgfsetlinewidth{0.000000pt}%
\definecolor{currentstroke}{rgb}{0.870588,0.560784,0.019608}%
\pgfsetstrokecolor{currentstroke}%
\pgfsetstrokeopacity{0.500000}%
\pgfsetdash{}{0pt}%
\pgfpathmoveto{\pgfqpoint{0.520798in}{0.970311in}}%
\pgfpathlineto{\pgfqpoint{0.520798in}{0.942125in}}%
\pgfpathlineto{\pgfqpoint{0.556748in}{0.942125in}}%
\pgfpathlineto{\pgfqpoint{0.556748in}{0.803775in}}%
\pgfpathlineto{\pgfqpoint{0.628649in}{0.803775in}}%
\pgfpathlineto{\pgfqpoint{0.628649in}{0.636253in}}%
\pgfpathlineto{\pgfqpoint{0.700549in}{0.636253in}}%
\pgfpathlineto{\pgfqpoint{0.700549in}{0.456187in}}%
\pgfpathlineto{\pgfqpoint{0.772450in}{0.456187in}}%
\pgfpathlineto{\pgfqpoint{0.772450in}{0.444206in}}%
\pgfpathlineto{\pgfqpoint{0.844350in}{0.444206in}}%
\pgfpathlineto{\pgfqpoint{0.844350in}{0.443151in}}%
\pgfpathlineto{\pgfqpoint{0.916251in}{0.443151in}}%
\pgfpathlineto{\pgfqpoint{0.916251in}{0.442786in}}%
\pgfpathlineto{\pgfqpoint{0.988151in}{0.442786in}}%
\pgfpathlineto{\pgfqpoint{0.988151in}{0.442551in}}%
\pgfpathlineto{\pgfqpoint{1.060052in}{0.442551in}}%
\pgfpathlineto{\pgfqpoint{1.060052in}{0.442228in}}%
\pgfpathlineto{\pgfqpoint{1.167902in}{0.442228in}}%
\pgfpathlineto{\pgfqpoint{1.167902in}{0.442135in}}%
\pgfpathlineto{\pgfqpoint{1.311703in}{0.442135in}}%
\pgfpathlineto{\pgfqpoint{1.311703in}{0.442084in}}%
\pgfpathlineto{\pgfqpoint{1.419554in}{0.442084in}}%
\pgfpathlineto{\pgfqpoint{1.419554in}{0.442177in}}%
\pgfpathlineto{\pgfqpoint{2.066659in}{0.442177in}}%
\pgfpathlineto{\pgfqpoint{2.066659in}{0.442177in}}%
\pgfpathlineto{\pgfqpoint{2.677813in}{0.442177in}}%
\pgfpathlineto{\pgfqpoint{2.677813in}{0.442177in}}%
\pgfpathlineto{\pgfqpoint{2.677813in}{0.442177in}}%
\pgfpathlineto{\pgfqpoint{2.066659in}{0.442177in}}%
\pgfpathlineto{\pgfqpoint{2.066659in}{0.442177in}}%
\pgfpathlineto{\pgfqpoint{1.419554in}{0.442177in}}%
\pgfpathlineto{\pgfqpoint{1.419554in}{0.442455in}}%
\pgfpathlineto{\pgfqpoint{1.311703in}{0.442455in}}%
\pgfpathlineto{\pgfqpoint{1.311703in}{0.442590in}}%
\pgfpathlineto{\pgfqpoint{1.167902in}{0.442590in}}%
\pgfpathlineto{\pgfqpoint{1.167902in}{0.442682in}}%
\pgfpathlineto{\pgfqpoint{1.060052in}{0.442682in}}%
\pgfpathlineto{\pgfqpoint{1.060052in}{0.445327in}}%
\pgfpathlineto{\pgfqpoint{0.988151in}{0.445327in}}%
\pgfpathlineto{\pgfqpoint{0.988151in}{0.446020in}}%
\pgfpathlineto{\pgfqpoint{0.916251in}{0.446020in}}%
\pgfpathlineto{\pgfqpoint{0.916251in}{0.446396in}}%
\pgfpathlineto{\pgfqpoint{0.844350in}{0.446396in}}%
\pgfpathlineto{\pgfqpoint{0.844350in}{0.448124in}}%
\pgfpathlineto{\pgfqpoint{0.772450in}{0.448124in}}%
\pgfpathlineto{\pgfqpoint{0.772450in}{0.467308in}}%
\pgfpathlineto{\pgfqpoint{0.700549in}{0.467308in}}%
\pgfpathlineto{\pgfqpoint{0.700549in}{0.648426in}}%
\pgfpathlineto{\pgfqpoint{0.628649in}{0.648426in}}%
\pgfpathlineto{\pgfqpoint{0.628649in}{0.833924in}}%
\pgfpathlineto{\pgfqpoint{0.556748in}{0.833924in}}%
\pgfpathlineto{\pgfqpoint{0.556748in}{0.970311in}}%
\pgfpathlineto{\pgfqpoint{0.520798in}{0.970311in}}%
\pgfpathlineto{\pgfqpoint{0.520798in}{0.970311in}}%
\pgfpathclose%
\pgfusepath{fill}%
\end{pgfscope}%
\begin{pgfscope}%
\pgfpathrectangle{\pgfqpoint{0.520798in}{0.442177in}}{\pgfqpoint{0.719005in}{0.871884in}}%
\pgfusepath{clip}%
\pgfsetroundcap%
\pgfsetroundjoin%
\pgfsetlinewidth{1.003750pt}%
\definecolor{currentstroke}{rgb}{0.007843,0.619608,0.450980}%
\pgfsetstrokecolor{currentstroke}%
\pgfsetdash{}{0pt}%
\pgfpathmoveto{\pgfqpoint{0.520798in}{0.956218in}}%
\pgfpathlineto{\pgfqpoint{0.556748in}{0.956218in}}%
\pgfpathlineto{\pgfqpoint{0.556748in}{0.818849in}}%
\pgfpathlineto{\pgfqpoint{0.628649in}{0.818849in}}%
\pgfpathlineto{\pgfqpoint{0.628649in}{0.642339in}}%
\pgfpathlineto{\pgfqpoint{0.700549in}{0.642339in}}%
\pgfpathlineto{\pgfqpoint{0.700549in}{0.461748in}}%
\pgfpathlineto{\pgfqpoint{0.772450in}{0.461748in}}%
\pgfpathlineto{\pgfqpoint{0.772450in}{0.446165in}}%
\pgfpathlineto{\pgfqpoint{0.844350in}{0.446165in}}%
\pgfpathlineto{\pgfqpoint{0.844350in}{0.444774in}}%
\pgfpathlineto{\pgfqpoint{0.916251in}{0.444774in}}%
\pgfpathlineto{\pgfqpoint{0.916251in}{0.444403in}}%
\pgfpathlineto{\pgfqpoint{0.988151in}{0.444403in}}%
\pgfpathlineto{\pgfqpoint{0.988151in}{0.443939in}}%
\pgfpathlineto{\pgfqpoint{1.060052in}{0.443939in}}%
\pgfpathlineto{\pgfqpoint{1.060052in}{0.442455in}}%
\pgfpathlineto{\pgfqpoint{1.167902in}{0.442455in}}%
\pgfpathlineto{\pgfqpoint{1.167902in}{0.442362in}}%
\pgfpathlineto{\pgfqpoint{1.241470in}{0.442362in}}%
\pgfusepath{stroke}%
\end{pgfscope}%
\begin{pgfscope}%
\pgfpathrectangle{\pgfqpoint{0.520798in}{0.442177in}}{\pgfqpoint{0.719005in}{0.871884in}}%
\pgfusepath{clip}%
\pgfsetbuttcap%
\pgfsetroundjoin%
\definecolor{currentfill}{rgb}{0.007843,0.619608,0.450980}%
\pgfsetfillcolor{currentfill}%
\pgfsetfillopacity{0.500000}%
\pgfsetlinewidth{0.000000pt}%
\definecolor{currentstroke}{rgb}{0.007843,0.619608,0.450980}%
\pgfsetstrokecolor{currentstroke}%
\pgfsetstrokeopacity{0.500000}%
\pgfsetdash{}{0pt}%
\pgfpathmoveto{\pgfqpoint{0.520798in}{0.970311in}}%
\pgfpathlineto{\pgfqpoint{0.520798in}{0.942125in}}%
\pgfpathlineto{\pgfqpoint{0.556748in}{0.942125in}}%
\pgfpathlineto{\pgfqpoint{0.556748in}{0.803775in}}%
\pgfpathlineto{\pgfqpoint{0.628649in}{0.803775in}}%
\pgfpathlineto{\pgfqpoint{0.628649in}{0.636253in}}%
\pgfpathlineto{\pgfqpoint{0.700549in}{0.636253in}}%
\pgfpathlineto{\pgfqpoint{0.700549in}{0.456187in}}%
\pgfpathlineto{\pgfqpoint{0.772450in}{0.456187in}}%
\pgfpathlineto{\pgfqpoint{0.772450in}{0.444206in}}%
\pgfpathlineto{\pgfqpoint{0.844350in}{0.444206in}}%
\pgfpathlineto{\pgfqpoint{0.844350in}{0.443151in}}%
\pgfpathlineto{\pgfqpoint{0.916251in}{0.443151in}}%
\pgfpathlineto{\pgfqpoint{0.916251in}{0.442786in}}%
\pgfpathlineto{\pgfqpoint{0.988151in}{0.442786in}}%
\pgfpathlineto{\pgfqpoint{0.988151in}{0.442551in}}%
\pgfpathlineto{\pgfqpoint{1.060052in}{0.442551in}}%
\pgfpathlineto{\pgfqpoint{1.060052in}{0.442228in}}%
\pgfpathlineto{\pgfqpoint{1.167902in}{0.442228in}}%
\pgfpathlineto{\pgfqpoint{1.167902in}{0.442135in}}%
\pgfpathlineto{\pgfqpoint{1.311703in}{0.442135in}}%
\pgfpathlineto{\pgfqpoint{1.311703in}{0.442084in}}%
\pgfpathlineto{\pgfqpoint{1.419554in}{0.442084in}}%
\pgfpathlineto{\pgfqpoint{1.419554in}{0.442177in}}%
\pgfpathlineto{\pgfqpoint{2.066659in}{0.442177in}}%
\pgfpathlineto{\pgfqpoint{2.066659in}{0.442177in}}%
\pgfpathlineto{\pgfqpoint{2.677813in}{0.442177in}}%
\pgfpathlineto{\pgfqpoint{2.677813in}{0.442177in}}%
\pgfpathlineto{\pgfqpoint{2.677813in}{0.442177in}}%
\pgfpathlineto{\pgfqpoint{2.066659in}{0.442177in}}%
\pgfpathlineto{\pgfqpoint{2.066659in}{0.442177in}}%
\pgfpathlineto{\pgfqpoint{1.419554in}{0.442177in}}%
\pgfpathlineto{\pgfqpoint{1.419554in}{0.442455in}}%
\pgfpathlineto{\pgfqpoint{1.311703in}{0.442455in}}%
\pgfpathlineto{\pgfqpoint{1.311703in}{0.442590in}}%
\pgfpathlineto{\pgfqpoint{1.167902in}{0.442590in}}%
\pgfpathlineto{\pgfqpoint{1.167902in}{0.442682in}}%
\pgfpathlineto{\pgfqpoint{1.060052in}{0.442682in}}%
\pgfpathlineto{\pgfqpoint{1.060052in}{0.445327in}}%
\pgfpathlineto{\pgfqpoint{0.988151in}{0.445327in}}%
\pgfpathlineto{\pgfqpoint{0.988151in}{0.446020in}}%
\pgfpathlineto{\pgfqpoint{0.916251in}{0.446020in}}%
\pgfpathlineto{\pgfqpoint{0.916251in}{0.446396in}}%
\pgfpathlineto{\pgfqpoint{0.844350in}{0.446396in}}%
\pgfpathlineto{\pgfqpoint{0.844350in}{0.448124in}}%
\pgfpathlineto{\pgfqpoint{0.772450in}{0.448124in}}%
\pgfpathlineto{\pgfqpoint{0.772450in}{0.467308in}}%
\pgfpathlineto{\pgfqpoint{0.700549in}{0.467308in}}%
\pgfpathlineto{\pgfqpoint{0.700549in}{0.648426in}}%
\pgfpathlineto{\pgfqpoint{0.628649in}{0.648426in}}%
\pgfpathlineto{\pgfqpoint{0.628649in}{0.833924in}}%
\pgfpathlineto{\pgfqpoint{0.556748in}{0.833924in}}%
\pgfpathlineto{\pgfqpoint{0.556748in}{0.970311in}}%
\pgfpathlineto{\pgfqpoint{0.520798in}{0.970311in}}%
\pgfpathlineto{\pgfqpoint{0.520798in}{0.970311in}}%
\pgfpathclose%
\pgfusepath{fill}%
\end{pgfscope}%
\begin{pgfscope}%
\definecolor{textcolor}{rgb}{0.150000,0.150000,0.150000}%
\pgfsetstrokecolor{textcolor}%
\pgfsetfillcolor{textcolor}%
\pgftext[x=0.880300in,y=1.397395in,,base]{\color{textcolor}\rmfamily\fontsize{9.000000}{10.800000}\selectfont \(\displaystyle c_A^+=1\)}%
\end{pgfscope}%
\begin{pgfscope}%
\pgfsetbuttcap%
\pgfsetmiterjoin%
\definecolor{currentfill}{rgb}{1.000000,1.000000,1.000000}%
\pgfsetfillcolor{currentfill}%
\pgfsetfillopacity{0.800000}%
\pgfsetlinewidth{1.003750pt}%
\definecolor{currentstroke}{rgb}{0.800000,0.800000,0.800000}%
\pgfsetstrokecolor{currentstroke}%
\pgfsetstrokeopacity{0.800000}%
\pgfsetdash{}{0pt}%
\pgfpathmoveto{\pgfqpoint{0.785843in}{0.638103in}}%
\pgfpathlineto{\pgfqpoint{1.191192in}{0.638103in}}%
\pgfpathlineto{\pgfqpoint{1.191192in}{1.265450in}}%
\pgfpathlineto{\pgfqpoint{0.785843in}{1.265450in}}%
\pgfpathlineto{\pgfqpoint{0.785843in}{0.638103in}}%
\pgfpathclose%
\pgfusepath{stroke,fill}%
\end{pgfscope}%
\begin{pgfscope}%
\definecolor{textcolor}{rgb}{0.150000,0.150000,0.150000}%
\pgfsetstrokecolor{textcolor}%
\pgfsetfillcolor{textcolor}%
\pgftext[x=0.917113in,y=1.113275in,left,base]{\color{textcolor}\rmfamily\fontsize{9.000000}{10.800000}\selectfont \(\displaystyle c_A^-\)}%
\end{pgfscope}%
\begin{pgfscope}%
\pgfsetroundcap%
\pgfsetroundjoin%
\pgfsetlinewidth{1.003750pt}%
\definecolor{currentstroke}{rgb}{0.003922,0.450980,0.698039}%
\pgfsetstrokecolor{currentstroke}%
\pgfsetdash{}{0pt}%
\pgfpathmoveto{\pgfqpoint{0.824732in}{1.001052in}}%
\pgfpathlineto{\pgfqpoint{0.873343in}{1.001052in}}%
\pgfpathlineto{\pgfqpoint{0.873343in}{1.001052in}}%
\pgfpathlineto{\pgfqpoint{0.970565in}{1.001052in}}%
\pgfpathlineto{\pgfqpoint{0.970565in}{1.001052in}}%
\pgfpathlineto{\pgfqpoint{1.019176in}{1.001052in}}%
\pgfusepath{stroke}%
\end{pgfscope}%
\begin{pgfscope}%
\definecolor{textcolor}{rgb}{0.150000,0.150000,0.150000}%
\pgfsetstrokecolor{textcolor}%
\pgfsetfillcolor{textcolor}%
\pgftext[x=1.096954in,y=0.967024in,left,base]{\color{textcolor}\rmfamily\fontsize{7.000000}{8.400000}\selectfont 1}%
\end{pgfscope}%
\begin{pgfscope}%
\pgfsetroundcap%
\pgfsetroundjoin%
\pgfsetlinewidth{1.003750pt}%
\definecolor{currentstroke}{rgb}{0.870588,0.560784,0.019608}%
\pgfsetstrokecolor{currentstroke}%
\pgfsetdash{}{0pt}%
\pgfpathmoveto{\pgfqpoint{0.824732in}{0.865486in}}%
\pgfpathlineto{\pgfqpoint{0.873343in}{0.865486in}}%
\pgfpathlineto{\pgfqpoint{0.873343in}{0.865486in}}%
\pgfpathlineto{\pgfqpoint{0.970565in}{0.865486in}}%
\pgfpathlineto{\pgfqpoint{0.970565in}{0.865486in}}%
\pgfpathlineto{\pgfqpoint{1.019176in}{0.865486in}}%
\pgfusepath{stroke}%
\end{pgfscope}%
\begin{pgfscope}%
\definecolor{textcolor}{rgb}{0.150000,0.150000,0.150000}%
\pgfsetstrokecolor{textcolor}%
\pgfsetfillcolor{textcolor}%
\pgftext[x=1.096954in,y=0.831458in,left,base]{\color{textcolor}\rmfamily\fontsize{7.000000}{8.400000}\selectfont 2}%
\end{pgfscope}%
\begin{pgfscope}%
\pgfsetroundcap%
\pgfsetroundjoin%
\pgfsetlinewidth{1.003750pt}%
\definecolor{currentstroke}{rgb}{0.007843,0.619608,0.450980}%
\pgfsetstrokecolor{currentstroke}%
\pgfsetdash{}{0pt}%
\pgfpathmoveto{\pgfqpoint{0.824732in}{0.729919in}}%
\pgfpathlineto{\pgfqpoint{0.873343in}{0.729919in}}%
\pgfpathlineto{\pgfqpoint{0.873343in}{0.729919in}}%
\pgfpathlineto{\pgfqpoint{0.970565in}{0.729919in}}%
\pgfpathlineto{\pgfqpoint{0.970565in}{0.729919in}}%
\pgfpathlineto{\pgfqpoint{1.019176in}{0.729919in}}%
\pgfusepath{stroke}%
\end{pgfscope}%
\begin{pgfscope}%
\definecolor{textcolor}{rgb}{0.150000,0.150000,0.150000}%
\pgfsetstrokecolor{textcolor}%
\pgfsetfillcolor{textcolor}%
\pgftext[x=1.096954in,y=0.695892in,left,base]{\color{textcolor}\rmfamily\fontsize{7.000000}{8.400000}\selectfont 4}%
\end{pgfscope}%
\end{pgfpicture}%
\makeatother%
\endgroup%

%% file: figures/experiments/graphs/sparse_smoothing/nodes_attributes/node_classification-Cora-APPNP-hidden=32-p_adj_plus=0.0-p_adj_minus=0.0-p_att_plus=0.001-p_att_minus=0.8-multi_class_cert-B.pgf
\begingroup%
\makeatletter%
\begin{pgfpicture}%
\pgfpathrectangle{\pgfpointorigin}{\pgfqpoint{1.375000in}{1.581250in}}%
\pgfusepath{use as bounding box, clip}%
\begin{pgfscope}%
\pgfsetbuttcap%
\pgfsetmiterjoin%
\definecolor{currentfill}{rgb}{1.000000,1.000000,1.000000}%
\pgfsetfillcolor{currentfill}%
\pgfsetlinewidth{0.000000pt}%
\definecolor{currentstroke}{rgb}{1.000000,1.000000,1.000000}%
\pgfsetstrokecolor{currentstroke}%
\pgfsetdash{}{0pt}%
\pgfpathmoveto{\pgfqpoint{0.000000in}{0.000000in}}%
\pgfpathlineto{\pgfqpoint{1.375000in}{0.000000in}}%
\pgfpathlineto{\pgfqpoint{1.375000in}{1.581250in}}%
\pgfpathlineto{\pgfqpoint{0.000000in}{1.581250in}}%
\pgfpathlineto{\pgfqpoint{0.000000in}{0.000000in}}%
\pgfpathclose%
\pgfusepath{fill}%
\end{pgfscope}%
\begin{pgfscope}%
\pgfsetbuttcap%
\pgfsetmiterjoin%
\definecolor{currentfill}{rgb}{1.000000,1.000000,1.000000}%
\pgfsetfillcolor{currentfill}%
\pgfsetlinewidth{0.000000pt}%
\definecolor{currentstroke}{rgb}{0.000000,0.000000,0.000000}%
\pgfsetstrokecolor{currentstroke}%
\pgfsetstrokeopacity{0.000000}%
\pgfsetdash{}{0pt}%
\pgfpathmoveto{\pgfqpoint{0.520798in}{0.442177in}}%
\pgfpathlineto{\pgfqpoint{1.239803in}{0.442177in}}%
\pgfpathlineto{\pgfqpoint{1.239803in}{1.314061in}}%
\pgfpathlineto{\pgfqpoint{0.520798in}{1.314061in}}%
\pgfpathlineto{\pgfqpoint{0.520798in}{0.442177in}}%
\pgfpathclose%
\pgfusepath{fill}%
\end{pgfscope}%
\begin{pgfscope}%
\pgfpathrectangle{\pgfqpoint{0.520798in}{0.442177in}}{\pgfqpoint{0.719005in}{0.871884in}}%
\pgfusepath{clip}%
\pgfsetroundcap%
\pgfsetroundjoin%
\pgfsetlinewidth{0.501875pt}%
\definecolor{currentstroke}{rgb}{0.800000,0.800000,0.800000}%
\pgfsetstrokecolor{currentstroke}%
\pgfsetdash{}{0pt}%
\pgfpathmoveto{\pgfqpoint{0.520798in}{0.442177in}}%
\pgfpathlineto{\pgfqpoint{0.520798in}{1.314061in}}%
\pgfusepath{stroke}%
\end{pgfscope}%
\begin{pgfscope}%
\definecolor{textcolor}{rgb}{0.150000,0.150000,0.150000}%
\pgfsetstrokecolor{textcolor}%
\pgfsetfillcolor{textcolor}%
\pgftext[x=0.520798in,y=0.351899in,,top]{\color{textcolor}\rmfamily\fontsize{8.000000}{9.600000}\selectfont \(\displaystyle {0}\)}%
\end{pgfscope}%
\begin{pgfscope}%
\pgfpathrectangle{\pgfqpoint{0.520798in}{0.442177in}}{\pgfqpoint{0.719005in}{0.871884in}}%
\pgfusepath{clip}%
\pgfsetroundcap%
\pgfsetroundjoin%
\pgfsetlinewidth{0.501875pt}%
\definecolor{currentstroke}{rgb}{0.800000,0.800000,0.800000}%
\pgfsetstrokecolor{currentstroke}%
\pgfsetdash{}{0pt}%
\pgfpathmoveto{\pgfqpoint{0.880300in}{0.442177in}}%
\pgfpathlineto{\pgfqpoint{0.880300in}{1.314061in}}%
\pgfusepath{stroke}%
\end{pgfscope}%
\begin{pgfscope}%
\definecolor{textcolor}{rgb}{0.150000,0.150000,0.150000}%
\pgfsetstrokecolor{textcolor}%
\pgfsetfillcolor{textcolor}%
\pgftext[x=0.880300in,y=0.351899in,,top]{\color{textcolor}\rmfamily\fontsize{8.000000}{9.600000}\selectfont \(\displaystyle {5}\)}%
\end{pgfscope}%
\begin{pgfscope}%
\pgfpathrectangle{\pgfqpoint{0.520798in}{0.442177in}}{\pgfqpoint{0.719005in}{0.871884in}}%
\pgfusepath{clip}%
\pgfsetroundcap%
\pgfsetroundjoin%
\pgfsetlinewidth{0.501875pt}%
\definecolor{currentstroke}{rgb}{0.800000,0.800000,0.800000}%
\pgfsetstrokecolor{currentstroke}%
\pgfsetdash{}{0pt}%
\pgfpathmoveto{\pgfqpoint{1.239803in}{0.442177in}}%
\pgfpathlineto{\pgfqpoint{1.239803in}{1.314061in}}%
\pgfusepath{stroke}%
\end{pgfscope}%
\begin{pgfscope}%
\definecolor{textcolor}{rgb}{0.150000,0.150000,0.150000}%
\pgfsetstrokecolor{textcolor}%
\pgfsetfillcolor{textcolor}%
\pgftext[x=1.239803in,y=0.351899in,,top]{\color{textcolor}\rmfamily\fontsize{8.000000}{9.600000}\selectfont \(\displaystyle {10}\)}%
\end{pgfscope}%
\begin{pgfscope}%
\definecolor{textcolor}{rgb}{0.150000,0.150000,0.150000}%
\pgfsetstrokecolor{textcolor}%
\pgfsetfillcolor{textcolor}%
\pgftext[x=0.880300in,y=0.198219in,,top]{\color{textcolor}\rmfamily\fontsize{10.000000}{12.000000}\selectfont Edit distance \(\displaystyle \epsilon\)}%
\end{pgfscope}%
\begin{pgfscope}%
\pgfpathrectangle{\pgfqpoint{0.520798in}{0.442177in}}{\pgfqpoint{0.719005in}{0.871884in}}%
\pgfusepath{clip}%
\pgfsetroundcap%
\pgfsetroundjoin%
\pgfsetlinewidth{0.501875pt}%
\definecolor{currentstroke}{rgb}{0.800000,0.800000,0.800000}%
\pgfsetstrokecolor{currentstroke}%
\pgfsetdash{}{0pt}%
\pgfpathmoveto{\pgfqpoint{0.520798in}{0.442177in}}%
\pgfpathlineto{\pgfqpoint{1.239803in}{0.442177in}}%
\pgfusepath{stroke}%
\end{pgfscope}%
\begin{pgfscope}%
\definecolor{textcolor}{rgb}{0.150000,0.150000,0.150000}%
\pgfsetstrokecolor{textcolor}%
\pgfsetfillcolor{textcolor}%
\pgftext[x=0.273151in, y=0.403915in, left, base]{\color{textcolor}\rmfamily\fontsize{8.000000}{9.600000}\selectfont 0\%}%
\end{pgfscope}%
\begin{pgfscope}%
\pgfpathrectangle{\pgfqpoint{0.520798in}{0.442177in}}{\pgfqpoint{0.719005in}{0.871884in}}%
\pgfusepath{clip}%
\pgfsetroundcap%
\pgfsetroundjoin%
\pgfsetlinewidth{0.501875pt}%
\definecolor{currentstroke}{rgb}{0.800000,0.800000,0.800000}%
\pgfsetstrokecolor{currentstroke}%
\pgfsetdash{}{0pt}%
\pgfpathmoveto{\pgfqpoint{0.520798in}{0.660148in}}%
\pgfpathlineto{\pgfqpoint{1.239803in}{0.660148in}}%
\pgfusepath{stroke}%
\end{pgfscope}%
\begin{pgfscope}%
\definecolor{textcolor}{rgb}{0.150000,0.150000,0.150000}%
\pgfsetstrokecolor{textcolor}%
\pgfsetfillcolor{textcolor}%
\pgftext[x=0.214138in, y=0.621886in, left, base]{\color{textcolor}\rmfamily\fontsize{8.000000}{9.600000}\selectfont 25\%}%
\end{pgfscope}%
\begin{pgfscope}%
\pgfpathrectangle{\pgfqpoint{0.520798in}{0.442177in}}{\pgfqpoint{0.719005in}{0.871884in}}%
\pgfusepath{clip}%
\pgfsetroundcap%
\pgfsetroundjoin%
\pgfsetlinewidth{0.501875pt}%
\definecolor{currentstroke}{rgb}{0.800000,0.800000,0.800000}%
\pgfsetstrokecolor{currentstroke}%
\pgfsetdash{}{0pt}%
\pgfpathmoveto{\pgfqpoint{0.520798in}{0.878119in}}%
\pgfpathlineto{\pgfqpoint{1.239803in}{0.878119in}}%
\pgfusepath{stroke}%
\end{pgfscope}%
\begin{pgfscope}%
\definecolor{textcolor}{rgb}{0.150000,0.150000,0.150000}%
\pgfsetstrokecolor{textcolor}%
\pgfsetfillcolor{textcolor}%
\pgftext[x=0.214138in, y=0.839857in, left, base]{\color{textcolor}\rmfamily\fontsize{8.000000}{9.600000}\selectfont 50\%}%
\end{pgfscope}%
\begin{pgfscope}%
\pgfpathrectangle{\pgfqpoint{0.520798in}{0.442177in}}{\pgfqpoint{0.719005in}{0.871884in}}%
\pgfusepath{clip}%
\pgfsetroundcap%
\pgfsetroundjoin%
\pgfsetlinewidth{0.501875pt}%
\definecolor{currentstroke}{rgb}{0.800000,0.800000,0.800000}%
\pgfsetstrokecolor{currentstroke}%
\pgfsetdash{}{0pt}%
\pgfpathmoveto{\pgfqpoint{0.520798in}{1.096090in}}%
\pgfpathlineto{\pgfqpoint{1.239803in}{1.096090in}}%
\pgfusepath{stroke}%
\end{pgfscope}%
\begin{pgfscope}%
\definecolor{textcolor}{rgb}{0.150000,0.150000,0.150000}%
\pgfsetstrokecolor{textcolor}%
\pgfsetfillcolor{textcolor}%
\pgftext[x=0.214138in, y=1.057828in, left, base]{\color{textcolor}\rmfamily\fontsize{8.000000}{9.600000}\selectfont 75\%}%
\end{pgfscope}%
\begin{pgfscope}%
\pgfpathrectangle{\pgfqpoint{0.520798in}{0.442177in}}{\pgfqpoint{0.719005in}{0.871884in}}%
\pgfusepath{clip}%
\pgfsetroundcap%
\pgfsetroundjoin%
\pgfsetlinewidth{0.501875pt}%
\definecolor{currentstroke}{rgb}{0.800000,0.800000,0.800000}%
\pgfsetstrokecolor{currentstroke}%
\pgfsetdash{}{0pt}%
\pgfpathmoveto{\pgfqpoint{0.520798in}{1.314061in}}%
\pgfpathlineto{\pgfqpoint{1.239803in}{1.314061in}}%
\pgfusepath{stroke}%
\end{pgfscope}%
\begin{pgfscope}%
\definecolor{textcolor}{rgb}{0.150000,0.150000,0.150000}%
\pgfsetstrokecolor{textcolor}%
\pgfsetfillcolor{textcolor}%
\pgftext[x=0.155124in, y=1.275799in, left, base]{\color{textcolor}\rmfamily\fontsize{8.000000}{9.600000}\selectfont 100\%}%
\end{pgfscope}%
\begin{pgfscope}%
\definecolor{textcolor}{rgb}{0.150000,0.150000,0.150000}%
\pgfsetstrokecolor{textcolor}%
\pgfsetfillcolor{textcolor}%
\pgftext[x=0.099569in,y=0.878119in,,bottom,rotate=90.000000]{\color{textcolor}\rmfamily\fontsize{10.000000}{12.000000}\selectfont Cert. Acc.}%
\end{pgfscope}%
\begin{pgfscope}%
\pgfsetrectcap%
\pgfsetmiterjoin%
\pgfsetlinewidth{0.752812pt}%
\definecolor{currentstroke}{rgb}{0.700000,0.700000,0.700000}%
\pgfsetstrokecolor{currentstroke}%
\pgfsetdash{}{0pt}%
\pgfpathmoveto{\pgfqpoint{0.520798in}{0.442177in}}%
\pgfpathlineto{\pgfqpoint{0.520798in}{1.314061in}}%
\pgfusepath{stroke}%
\end{pgfscope}%
\begin{pgfscope}%
\pgfsetrectcap%
\pgfsetmiterjoin%
\pgfsetlinewidth{0.752812pt}%
\definecolor{currentstroke}{rgb}{0.700000,0.700000,0.700000}%
\pgfsetstrokecolor{currentstroke}%
\pgfsetdash{}{0pt}%
\pgfpathmoveto{\pgfqpoint{1.239803in}{0.442177in}}%
\pgfpathlineto{\pgfqpoint{1.239803in}{1.314061in}}%
\pgfusepath{stroke}%
\end{pgfscope}%
\begin{pgfscope}%
\pgfsetrectcap%
\pgfsetmiterjoin%
\pgfsetlinewidth{0.752812pt}%
\definecolor{currentstroke}{rgb}{0.700000,0.700000,0.700000}%
\pgfsetstrokecolor{currentstroke}%
\pgfsetdash{}{0pt}%
\pgfpathmoveto{\pgfqpoint{0.520798in}{0.442177in}}%
\pgfpathlineto{\pgfqpoint{1.239803in}{0.442177in}}%
\pgfusepath{stroke}%
\end{pgfscope}%
\begin{pgfscope}%
\pgfsetrectcap%
\pgfsetmiterjoin%
\pgfsetlinewidth{0.752812pt}%
\definecolor{currentstroke}{rgb}{0.700000,0.700000,0.700000}%
\pgfsetstrokecolor{currentstroke}%
\pgfsetdash{}{0pt}%
\pgfpathmoveto{\pgfqpoint{0.520798in}{1.314061in}}%
\pgfpathlineto{\pgfqpoint{1.239803in}{1.314061in}}%
\pgfusepath{stroke}%
\end{pgfscope}%
\begin{pgfscope}%
\pgfpathrectangle{\pgfqpoint{0.520798in}{0.442177in}}{\pgfqpoint{0.719005in}{0.871884in}}%
\pgfusepath{clip}%
\pgfsetroundcap%
\pgfsetroundjoin%
\pgfsetlinewidth{1.003750pt}%
\definecolor{currentstroke}{rgb}{0.003922,0.450980,0.698039}%
\pgfsetstrokecolor{currentstroke}%
\pgfsetdash{}{0pt}%
\pgfpathmoveto{\pgfqpoint{0.520798in}{1.165703in}}%
\pgfpathlineto{\pgfqpoint{0.556748in}{1.165703in}}%
\pgfpathlineto{\pgfqpoint{0.556748in}{1.104255in}}%
\pgfpathlineto{\pgfqpoint{0.628649in}{1.104255in}}%
\pgfpathlineto{\pgfqpoint{0.628649in}{1.001765in}}%
\pgfpathlineto{\pgfqpoint{0.700549in}{1.001765in}}%
\pgfpathlineto{\pgfqpoint{0.700549in}{0.709792in}}%
\pgfpathlineto{\pgfqpoint{0.772450in}{0.709792in}}%
\pgfpathlineto{\pgfqpoint{0.772450in}{0.580097in}}%
\pgfpathlineto{\pgfqpoint{0.844350in}{0.580097in}}%
\pgfpathlineto{\pgfqpoint{0.844350in}{0.564913in}}%
\pgfpathlineto{\pgfqpoint{0.916251in}{0.564913in}}%
\pgfpathlineto{\pgfqpoint{0.916251in}{0.553683in}}%
\pgfpathlineto{\pgfqpoint{0.988151in}{0.553683in}}%
\pgfpathlineto{\pgfqpoint{0.988151in}{0.539844in}}%
\pgfpathlineto{\pgfqpoint{1.060052in}{0.539844in}}%
\pgfpathlineto{\pgfqpoint{1.060052in}{0.492869in}}%
\pgfpathlineto{\pgfqpoint{1.131952in}{0.492869in}}%
\pgfpathlineto{\pgfqpoint{1.131952in}{0.486305in}}%
\pgfpathlineto{\pgfqpoint{1.203853in}{0.486305in}}%
\pgfpathlineto{\pgfqpoint{1.203853in}{0.466376in}}%
\pgfpathlineto{\pgfqpoint{1.241470in}{0.466376in}}%
\pgfusepath{stroke}%
\end{pgfscope}%
\begin{pgfscope}%
\pgfpathrectangle{\pgfqpoint{0.520798in}{0.442177in}}{\pgfqpoint{0.719005in}{0.871884in}}%
\pgfusepath{clip}%
\pgfsetbuttcap%
\pgfsetroundjoin%
\definecolor{currentfill}{rgb}{0.003922,0.450980,0.698039}%
\pgfsetfillcolor{currentfill}%
\pgfsetfillopacity{0.500000}%
\pgfsetlinewidth{0.000000pt}%
\definecolor{currentstroke}{rgb}{0.003922,0.450980,0.698039}%
\pgfsetstrokecolor{currentstroke}%
\pgfsetstrokeopacity{0.500000}%
\pgfsetdash{}{0pt}%
\pgfpathmoveto{\pgfqpoint{0.520798in}{1.178455in}}%
\pgfpathlineto{\pgfqpoint{0.520798in}{1.152950in}}%
\pgfpathlineto{\pgfqpoint{0.556748in}{1.152950in}}%
\pgfpathlineto{\pgfqpoint{0.556748in}{1.088241in}}%
\pgfpathlineto{\pgfqpoint{0.628649in}{1.088241in}}%
\pgfpathlineto{\pgfqpoint{0.628649in}{0.984178in}}%
\pgfpathlineto{\pgfqpoint{0.700549in}{0.984178in}}%
\pgfpathlineto{\pgfqpoint{0.700549in}{0.703216in}}%
\pgfpathlineto{\pgfqpoint{0.772450in}{0.703216in}}%
\pgfpathlineto{\pgfqpoint{0.772450in}{0.570999in}}%
\pgfpathlineto{\pgfqpoint{0.844350in}{0.570999in}}%
\pgfpathlineto{\pgfqpoint{0.844350in}{0.554989in}}%
\pgfpathlineto{\pgfqpoint{0.916251in}{0.554989in}}%
\pgfpathlineto{\pgfqpoint{0.916251in}{0.542660in}}%
\pgfpathlineto{\pgfqpoint{0.988151in}{0.542660in}}%
\pgfpathlineto{\pgfqpoint{0.988151in}{0.529201in}}%
\pgfpathlineto{\pgfqpoint{1.060052in}{0.529201in}}%
\pgfpathlineto{\pgfqpoint{1.060052in}{0.484179in}}%
\pgfpathlineto{\pgfqpoint{1.131952in}{0.484179in}}%
\pgfpathlineto{\pgfqpoint{1.131952in}{0.477227in}}%
\pgfpathlineto{\pgfqpoint{1.203853in}{0.477227in}}%
\pgfpathlineto{\pgfqpoint{1.203853in}{0.460660in}}%
\pgfpathlineto{\pgfqpoint{1.275753in}{0.460660in}}%
\pgfpathlineto{\pgfqpoint{1.275753in}{0.452951in}}%
\pgfpathlineto{\pgfqpoint{1.347654in}{0.452951in}}%
\pgfpathlineto{\pgfqpoint{1.347654in}{0.448808in}}%
\pgfpathlineto{\pgfqpoint{1.419554in}{0.448808in}}%
\pgfpathlineto{\pgfqpoint{1.419554in}{0.442177in}}%
\pgfpathlineto{\pgfqpoint{2.066659in}{0.442177in}}%
\pgfpathlineto{\pgfqpoint{2.066659in}{0.442177in}}%
\pgfpathlineto{\pgfqpoint{2.677813in}{0.442177in}}%
\pgfpathlineto{\pgfqpoint{2.677813in}{0.442177in}}%
\pgfpathlineto{\pgfqpoint{2.677813in}{0.442177in}}%
\pgfpathlineto{\pgfqpoint{2.066659in}{0.442177in}}%
\pgfpathlineto{\pgfqpoint{2.066659in}{0.442177in}}%
\pgfpathlineto{\pgfqpoint{1.419554in}{0.442177in}}%
\pgfpathlineto{\pgfqpoint{1.419554in}{0.454209in}}%
\pgfpathlineto{\pgfqpoint{1.347654in}{0.454209in}}%
\pgfpathlineto{\pgfqpoint{1.347654in}{0.457183in}}%
\pgfpathlineto{\pgfqpoint{1.275753in}{0.457183in}}%
\pgfpathlineto{\pgfqpoint{1.275753in}{0.472092in}}%
\pgfpathlineto{\pgfqpoint{1.203853in}{0.472092in}}%
\pgfpathlineto{\pgfqpoint{1.203853in}{0.495382in}}%
\pgfpathlineto{\pgfqpoint{1.131952in}{0.495382in}}%
\pgfpathlineto{\pgfqpoint{1.131952in}{0.501558in}}%
\pgfpathlineto{\pgfqpoint{1.060052in}{0.501558in}}%
\pgfpathlineto{\pgfqpoint{1.060052in}{0.550486in}}%
\pgfpathlineto{\pgfqpoint{0.988151in}{0.550486in}}%
\pgfpathlineto{\pgfqpoint{0.988151in}{0.564707in}}%
\pgfpathlineto{\pgfqpoint{0.916251in}{0.564707in}}%
\pgfpathlineto{\pgfqpoint{0.916251in}{0.574836in}}%
\pgfpathlineto{\pgfqpoint{0.844350in}{0.574836in}}%
\pgfpathlineto{\pgfqpoint{0.844350in}{0.589194in}}%
\pgfpathlineto{\pgfqpoint{0.772450in}{0.589194in}}%
\pgfpathlineto{\pgfqpoint{0.772450in}{0.716368in}}%
\pgfpathlineto{\pgfqpoint{0.700549in}{0.716368in}}%
\pgfpathlineto{\pgfqpoint{0.700549in}{1.019351in}}%
\pgfpathlineto{\pgfqpoint{0.628649in}{1.019351in}}%
\pgfpathlineto{\pgfqpoint{0.628649in}{1.120270in}}%
\pgfpathlineto{\pgfqpoint{0.556748in}{1.120270in}}%
\pgfpathlineto{\pgfqpoint{0.556748in}{1.178455in}}%
\pgfpathlineto{\pgfqpoint{0.520798in}{1.178455in}}%
\pgfpathlineto{\pgfqpoint{0.520798in}{1.178455in}}%
\pgfpathclose%
\pgfusepath{fill}%
\end{pgfscope}%
\begin{pgfscope}%
\pgfpathrectangle{\pgfqpoint{0.520798in}{0.442177in}}{\pgfqpoint{0.719005in}{0.871884in}}%
\pgfusepath{clip}%
\pgfsetroundcap%
\pgfsetroundjoin%
\pgfsetlinewidth{1.003750pt}%
\definecolor{currentstroke}{rgb}{0.870588,0.560784,0.019608}%
\pgfsetstrokecolor{currentstroke}%
\pgfsetdash{}{0pt}%
\pgfpathmoveto{\pgfqpoint{0.520798in}{1.165703in}}%
\pgfpathlineto{\pgfqpoint{0.556748in}{1.165703in}}%
\pgfpathlineto{\pgfqpoint{0.556748in}{1.116355in}}%
\pgfpathlineto{\pgfqpoint{0.628649in}{1.116355in}}%
\pgfpathlineto{\pgfqpoint{0.628649in}{1.071990in}}%
\pgfpathlineto{\pgfqpoint{0.700549in}{1.071990in}}%
\pgfpathlineto{\pgfqpoint{0.700549in}{1.029443in}}%
\pgfpathlineto{\pgfqpoint{0.772450in}{1.029443in}}%
\pgfpathlineto{\pgfqpoint{0.772450in}{0.988558in}}%
\pgfpathlineto{\pgfqpoint{0.844350in}{0.988558in}}%
\pgfpathlineto{\pgfqpoint{0.844350in}{0.947514in}}%
\pgfpathlineto{\pgfqpoint{0.916251in}{0.947514in}}%
\pgfpathlineto{\pgfqpoint{0.916251in}{0.709792in}}%
\pgfpathlineto{\pgfqpoint{0.988151in}{0.709792in}}%
\pgfpathlineto{\pgfqpoint{0.988151in}{0.693501in}}%
\pgfpathlineto{\pgfqpoint{1.060052in}{0.693501in}}%
\pgfpathlineto{\pgfqpoint{1.060052in}{0.580097in}}%
\pgfpathlineto{\pgfqpoint{1.131952in}{0.580097in}}%
\pgfpathlineto{\pgfqpoint{1.131952in}{0.570844in}}%
\pgfpathlineto{\pgfqpoint{1.203853in}{0.570844in}}%
\pgfpathlineto{\pgfqpoint{1.203853in}{0.559456in}}%
\pgfpathlineto{\pgfqpoint{1.241470in}{0.559456in}}%
\pgfusepath{stroke}%
\end{pgfscope}%
\begin{pgfscope}%
\pgfpathrectangle{\pgfqpoint{0.520798in}{0.442177in}}{\pgfqpoint{0.719005in}{0.871884in}}%
\pgfusepath{clip}%
\pgfsetbuttcap%
\pgfsetroundjoin%
\definecolor{currentfill}{rgb}{0.870588,0.560784,0.019608}%
\pgfsetfillcolor{currentfill}%
\pgfsetfillopacity{0.500000}%
\pgfsetlinewidth{0.000000pt}%
\definecolor{currentstroke}{rgb}{0.870588,0.560784,0.019608}%
\pgfsetstrokecolor{currentstroke}%
\pgfsetstrokeopacity{0.500000}%
\pgfsetdash{}{0pt}%
\pgfpathmoveto{\pgfqpoint{0.520798in}{1.178455in}}%
\pgfpathlineto{\pgfqpoint{0.520798in}{1.152950in}}%
\pgfpathlineto{\pgfqpoint{0.556748in}{1.152950in}}%
\pgfpathlineto{\pgfqpoint{0.556748in}{1.100834in}}%
\pgfpathlineto{\pgfqpoint{0.628649in}{1.100834in}}%
\pgfpathlineto{\pgfqpoint{0.628649in}{1.054462in}}%
\pgfpathlineto{\pgfqpoint{0.700549in}{1.054462in}}%
\pgfpathlineto{\pgfqpoint{0.700549in}{1.011533in}}%
\pgfpathlineto{\pgfqpoint{0.772450in}{1.011533in}}%
\pgfpathlineto{\pgfqpoint{0.772450in}{0.971389in}}%
\pgfpathlineto{\pgfqpoint{0.844350in}{0.971389in}}%
\pgfpathlineto{\pgfqpoint{0.844350in}{0.932061in}}%
\pgfpathlineto{\pgfqpoint{0.916251in}{0.932061in}}%
\pgfpathlineto{\pgfqpoint{0.916251in}{0.703216in}}%
\pgfpathlineto{\pgfqpoint{0.988151in}{0.703216in}}%
\pgfpathlineto{\pgfqpoint{0.988151in}{0.687039in}}%
\pgfpathlineto{\pgfqpoint{1.060052in}{0.687039in}}%
\pgfpathlineto{\pgfqpoint{1.060052in}{0.570999in}}%
\pgfpathlineto{\pgfqpoint{1.131952in}{0.570999in}}%
\pgfpathlineto{\pgfqpoint{1.131952in}{0.561461in}}%
\pgfpathlineto{\pgfqpoint{1.203853in}{0.561461in}}%
\pgfpathlineto{\pgfqpoint{1.203853in}{0.548077in}}%
\pgfpathlineto{\pgfqpoint{1.275753in}{0.548077in}}%
\pgfpathlineto{\pgfqpoint{1.275753in}{0.540632in}}%
\pgfpathlineto{\pgfqpoint{1.347654in}{0.540632in}}%
\pgfpathlineto{\pgfqpoint{1.347654in}{0.531497in}}%
\pgfpathlineto{\pgfqpoint{1.419554in}{0.531497in}}%
\pgfpathlineto{\pgfqpoint{1.419554in}{0.523922in}}%
\pgfpathlineto{\pgfqpoint{1.491455in}{0.523922in}}%
\pgfpathlineto{\pgfqpoint{1.491455in}{0.517574in}}%
\pgfpathlineto{\pgfqpoint{1.563355in}{0.517574in}}%
\pgfpathlineto{\pgfqpoint{1.563355in}{0.510894in}}%
\pgfpathlineto{\pgfqpoint{1.635256in}{0.510894in}}%
\pgfpathlineto{\pgfqpoint{1.635256in}{0.484077in}}%
\pgfpathlineto{\pgfqpoint{1.707156in}{0.484077in}}%
\pgfpathlineto{\pgfqpoint{1.707156in}{0.480767in}}%
\pgfpathlineto{\pgfqpoint{1.779057in}{0.480767in}}%
\pgfpathlineto{\pgfqpoint{1.779057in}{0.474706in}}%
\pgfpathlineto{\pgfqpoint{1.850957in}{0.474706in}}%
\pgfpathlineto{\pgfqpoint{1.850957in}{0.472148in}}%
\pgfpathlineto{\pgfqpoint{1.922858in}{0.472148in}}%
\pgfpathlineto{\pgfqpoint{1.922858in}{0.460399in}}%
\pgfpathlineto{\pgfqpoint{1.994758in}{0.460399in}}%
\pgfpathlineto{\pgfqpoint{1.994758in}{0.452521in}}%
\pgfpathlineto{\pgfqpoint{2.066659in}{0.452521in}}%
\pgfpathlineto{\pgfqpoint{2.066659in}{0.450281in}}%
\pgfpathlineto{\pgfqpoint{2.138559in}{0.450281in}}%
\pgfpathlineto{\pgfqpoint{2.138559in}{0.448413in}}%
\pgfpathlineto{\pgfqpoint{2.210460in}{0.448413in}}%
\pgfpathlineto{\pgfqpoint{2.210460in}{0.444979in}}%
\pgfpathlineto{\pgfqpoint{2.282360in}{0.444979in}}%
\pgfpathlineto{\pgfqpoint{2.282360in}{0.442177in}}%
\pgfpathlineto{\pgfqpoint{2.498062in}{0.442177in}}%
\pgfpathlineto{\pgfqpoint{2.498062in}{0.442177in}}%
\pgfpathlineto{\pgfqpoint{2.677813in}{0.442177in}}%
\pgfpathlineto{\pgfqpoint{2.677813in}{0.442177in}}%
\pgfpathlineto{\pgfqpoint{2.677813in}{0.442177in}}%
\pgfpathlineto{\pgfqpoint{2.498062in}{0.442177in}}%
\pgfpathlineto{\pgfqpoint{2.498062in}{0.442177in}}%
\pgfpathlineto{\pgfqpoint{2.282360in}{0.442177in}}%
\pgfpathlineto{\pgfqpoint{2.282360in}{0.447125in}}%
\pgfpathlineto{\pgfqpoint{2.210460in}{0.447125in}}%
\pgfpathlineto{\pgfqpoint{2.210460in}{0.453814in}}%
\pgfpathlineto{\pgfqpoint{2.138559in}{0.453814in}}%
\pgfpathlineto{\pgfqpoint{2.138559in}{0.455267in}}%
\pgfpathlineto{\pgfqpoint{2.066659in}{0.455267in}}%
\pgfpathlineto{\pgfqpoint{2.066659in}{0.456664in}}%
\pgfpathlineto{\pgfqpoint{1.994758in}{0.456664in}}%
\pgfpathlineto{\pgfqpoint{1.994758in}{0.472037in}}%
\pgfpathlineto{\pgfqpoint{1.922858in}{0.472037in}}%
\pgfpathlineto{\pgfqpoint{1.922858in}{0.487650in}}%
\pgfpathlineto{\pgfqpoint{1.850957in}{0.487650in}}%
\pgfpathlineto{\pgfqpoint{1.850957in}{0.492526in}}%
\pgfpathlineto{\pgfqpoint{1.779057in}{0.492526in}}%
\pgfpathlineto{\pgfqpoint{1.779057in}{0.498011in}}%
\pgfpathlineto{\pgfqpoint{1.707156in}{0.498011in}}%
\pgfpathlineto{\pgfqpoint{1.707156in}{0.501028in}}%
\pgfpathlineto{\pgfqpoint{1.635256in}{0.501028in}}%
\pgfpathlineto{\pgfqpoint{1.635256in}{0.534314in}}%
\pgfpathlineto{\pgfqpoint{1.563355in}{0.534314in}}%
\pgfpathlineto{\pgfqpoint{1.563355in}{0.539970in}}%
\pgfpathlineto{\pgfqpoint{1.491455in}{0.539970in}}%
\pgfpathlineto{\pgfqpoint{1.491455in}{0.545801in}}%
\pgfpathlineto{\pgfqpoint{1.419554in}{0.545801in}}%
\pgfpathlineto{\pgfqpoint{1.419554in}{0.553093in}}%
\pgfpathlineto{\pgfqpoint{1.347654in}{0.553093in}}%
\pgfpathlineto{\pgfqpoint{1.347654in}{0.562780in}}%
\pgfpathlineto{\pgfqpoint{1.275753in}{0.562780in}}%
\pgfpathlineto{\pgfqpoint{1.275753in}{0.570836in}}%
\pgfpathlineto{\pgfqpoint{1.203853in}{0.570836in}}%
\pgfpathlineto{\pgfqpoint{1.203853in}{0.580227in}}%
\pgfpathlineto{\pgfqpoint{1.131952in}{0.580227in}}%
\pgfpathlineto{\pgfqpoint{1.131952in}{0.589194in}}%
\pgfpathlineto{\pgfqpoint{1.060052in}{0.589194in}}%
\pgfpathlineto{\pgfqpoint{1.060052in}{0.699963in}}%
\pgfpathlineto{\pgfqpoint{0.988151in}{0.699963in}}%
\pgfpathlineto{\pgfqpoint{0.988151in}{0.716368in}}%
\pgfpathlineto{\pgfqpoint{0.916251in}{0.716368in}}%
\pgfpathlineto{\pgfqpoint{0.916251in}{0.962966in}}%
\pgfpathlineto{\pgfqpoint{0.844350in}{0.962966in}}%
\pgfpathlineto{\pgfqpoint{0.844350in}{1.005727in}}%
\pgfpathlineto{\pgfqpoint{0.772450in}{1.005727in}}%
\pgfpathlineto{\pgfqpoint{0.772450in}{1.047354in}}%
\pgfpathlineto{\pgfqpoint{0.700549in}{1.047354in}}%
\pgfpathlineto{\pgfqpoint{0.700549in}{1.089518in}}%
\pgfpathlineto{\pgfqpoint{0.628649in}{1.089518in}}%
\pgfpathlineto{\pgfqpoint{0.628649in}{1.131876in}}%
\pgfpathlineto{\pgfqpoint{0.556748in}{1.131876in}}%
\pgfpathlineto{\pgfqpoint{0.556748in}{1.178455in}}%
\pgfpathlineto{\pgfqpoint{0.520798in}{1.178455in}}%
\pgfpathlineto{\pgfqpoint{0.520798in}{1.178455in}}%
\pgfpathclose%
\pgfusepath{fill}%
\end{pgfscope}%
\begin{pgfscope}%
\pgfpathrectangle{\pgfqpoint{0.520798in}{0.442177in}}{\pgfqpoint{0.719005in}{0.871884in}}%
\pgfusepath{clip}%
\pgfsetroundcap%
\pgfsetroundjoin%
\pgfsetlinewidth{1.003750pt}%
\definecolor{currentstroke}{rgb}{0.007843,0.619608,0.450980}%
\pgfsetstrokecolor{currentstroke}%
\pgfsetdash{}{0pt}%
\pgfpathmoveto{\pgfqpoint{0.520798in}{1.165703in}}%
\pgfpathlineto{\pgfqpoint{0.556748in}{1.165703in}}%
\pgfpathlineto{\pgfqpoint{0.556748in}{1.116355in}}%
\pgfpathlineto{\pgfqpoint{0.628649in}{1.116355in}}%
\pgfpathlineto{\pgfqpoint{0.628649in}{1.071990in}}%
\pgfpathlineto{\pgfqpoint{0.700549in}{1.071990in}}%
\pgfpathlineto{\pgfqpoint{0.700549in}{1.029443in}}%
\pgfpathlineto{\pgfqpoint{0.772450in}{1.029443in}}%
\pgfpathlineto{\pgfqpoint{0.772450in}{0.988558in}}%
\pgfpathlineto{\pgfqpoint{0.844350in}{0.988558in}}%
\pgfpathlineto{\pgfqpoint{0.844350in}{0.947514in}}%
\pgfpathlineto{\pgfqpoint{0.916251in}{0.947514in}}%
\pgfpathlineto{\pgfqpoint{0.916251in}{0.912480in}}%
\pgfpathlineto{\pgfqpoint{0.988151in}{0.912480in}}%
\pgfpathlineto{\pgfqpoint{0.988151in}{0.879898in}}%
\pgfpathlineto{\pgfqpoint{1.060052in}{0.879898in}}%
\pgfpathlineto{\pgfqpoint{1.060052in}{0.845181in}}%
\pgfpathlineto{\pgfqpoint{1.131952in}{0.845181in}}%
\pgfpathlineto{\pgfqpoint{1.131952in}{0.815209in}}%
\pgfpathlineto{\pgfqpoint{1.203853in}{0.815209in}}%
\pgfpathlineto{\pgfqpoint{1.203853in}{0.788321in}}%
\pgfpathlineto{\pgfqpoint{1.241470in}{0.788321in}}%
\pgfusepath{stroke}%
\end{pgfscope}%
\begin{pgfscope}%
\pgfpathrectangle{\pgfqpoint{0.520798in}{0.442177in}}{\pgfqpoint{0.719005in}{0.871884in}}%
\pgfusepath{clip}%
\pgfsetbuttcap%
\pgfsetroundjoin%
\definecolor{currentfill}{rgb}{0.007843,0.619608,0.450980}%
\pgfsetfillcolor{currentfill}%
\pgfsetfillopacity{0.500000}%
\pgfsetlinewidth{0.000000pt}%
\definecolor{currentstroke}{rgb}{0.007843,0.619608,0.450980}%
\pgfsetstrokecolor{currentstroke}%
\pgfsetstrokeopacity{0.500000}%
\pgfsetdash{}{0pt}%
\pgfpathmoveto{\pgfqpoint{0.520798in}{1.178455in}}%
\pgfpathlineto{\pgfqpoint{0.520798in}{1.152950in}}%
\pgfpathlineto{\pgfqpoint{0.556748in}{1.152950in}}%
\pgfpathlineto{\pgfqpoint{0.556748in}{1.100834in}}%
\pgfpathlineto{\pgfqpoint{0.628649in}{1.100834in}}%
\pgfpathlineto{\pgfqpoint{0.628649in}{1.054462in}}%
\pgfpathlineto{\pgfqpoint{0.700549in}{1.054462in}}%
\pgfpathlineto{\pgfqpoint{0.700549in}{1.011533in}}%
\pgfpathlineto{\pgfqpoint{0.772450in}{1.011533in}}%
\pgfpathlineto{\pgfqpoint{0.772450in}{0.971389in}}%
\pgfpathlineto{\pgfqpoint{0.844350in}{0.971389in}}%
\pgfpathlineto{\pgfqpoint{0.844350in}{0.932061in}}%
\pgfpathlineto{\pgfqpoint{0.916251in}{0.932061in}}%
\pgfpathlineto{\pgfqpoint{0.916251in}{0.901595in}}%
\pgfpathlineto{\pgfqpoint{0.988151in}{0.901595in}}%
\pgfpathlineto{\pgfqpoint{0.988151in}{0.871707in}}%
\pgfpathlineto{\pgfqpoint{1.060052in}{0.871707in}}%
\pgfpathlineto{\pgfqpoint{1.060052in}{0.837669in}}%
\pgfpathlineto{\pgfqpoint{1.131952in}{0.837669in}}%
\pgfpathlineto{\pgfqpoint{1.131952in}{0.806298in}}%
\pgfpathlineto{\pgfqpoint{1.203853in}{0.806298in}}%
\pgfpathlineto{\pgfqpoint{1.203853in}{0.781419in}}%
\pgfpathlineto{\pgfqpoint{1.275753in}{0.781419in}}%
\pgfpathlineto{\pgfqpoint{1.275753in}{0.759225in}}%
\pgfpathlineto{\pgfqpoint{1.347654in}{0.759225in}}%
\pgfpathlineto{\pgfqpoint{1.347654in}{0.703216in}}%
\pgfpathlineto{\pgfqpoint{1.419554in}{0.703216in}}%
\pgfpathlineto{\pgfqpoint{1.419554in}{0.687039in}}%
\pgfpathlineto{\pgfqpoint{1.491455in}{0.687039in}}%
\pgfpathlineto{\pgfqpoint{1.491455in}{0.670900in}}%
\pgfpathlineto{\pgfqpoint{1.563355in}{0.670900in}}%
\pgfpathlineto{\pgfqpoint{1.563355in}{0.652459in}}%
\pgfpathlineto{\pgfqpoint{1.635256in}{0.652459in}}%
\pgfpathlineto{\pgfqpoint{1.635256in}{0.570999in}}%
\pgfpathlineto{\pgfqpoint{1.707156in}{0.570999in}}%
\pgfpathlineto{\pgfqpoint{1.707156in}{0.561453in}}%
\pgfpathlineto{\pgfqpoint{1.779057in}{0.561453in}}%
\pgfpathlineto{\pgfqpoint{1.779057in}{0.551869in}}%
\pgfpathlineto{\pgfqpoint{1.850957in}{0.551869in}}%
\pgfpathlineto{\pgfqpoint{1.850957in}{0.543978in}}%
\pgfpathlineto{\pgfqpoint{1.922858in}{0.543978in}}%
\pgfpathlineto{\pgfqpoint{1.922858in}{0.536453in}}%
\pgfpathlineto{\pgfqpoint{1.994758in}{0.536453in}}%
\pgfpathlineto{\pgfqpoint{1.994758in}{0.528082in}}%
\pgfpathlineto{\pgfqpoint{2.066659in}{0.528082in}}%
\pgfpathlineto{\pgfqpoint{2.066659in}{0.522336in}}%
\pgfpathlineto{\pgfqpoint{2.138559in}{0.522336in}}%
\pgfpathlineto{\pgfqpoint{2.138559in}{0.516814in}}%
\pgfpathlineto{\pgfqpoint{2.210460in}{0.516814in}}%
\pgfpathlineto{\pgfqpoint{2.210460in}{0.512486in}}%
\pgfpathlineto{\pgfqpoint{2.282360in}{0.512486in}}%
\pgfpathlineto{\pgfqpoint{2.282360in}{0.508256in}}%
\pgfpathlineto{\pgfqpoint{2.354261in}{0.508256in}}%
\pgfpathlineto{\pgfqpoint{2.354261in}{0.504793in}}%
\pgfpathlineto{\pgfqpoint{2.426161in}{0.504793in}}%
\pgfpathlineto{\pgfqpoint{2.426161in}{0.502902in}}%
\pgfpathlineto{\pgfqpoint{2.498062in}{0.502902in}}%
\pgfpathlineto{\pgfqpoint{2.498062in}{0.499364in}}%
\pgfpathlineto{\pgfqpoint{2.569962in}{0.499364in}}%
\pgfpathlineto{\pgfqpoint{2.569962in}{0.495878in}}%
\pgfpathlineto{\pgfqpoint{2.641863in}{0.495878in}}%
\pgfpathlineto{\pgfqpoint{2.641863in}{0.491397in}}%
\pgfpathlineto{\pgfqpoint{2.677813in}{0.491397in}}%
\pgfpathlineto{\pgfqpoint{2.677813in}{0.509683in}}%
\pgfpathlineto{\pgfqpoint{2.677813in}{0.509683in}}%
\pgfpathlineto{\pgfqpoint{2.641863in}{0.509683in}}%
\pgfpathlineto{\pgfqpoint{2.641863in}{0.513742in}}%
\pgfpathlineto{\pgfqpoint{2.569962in}{0.513742in}}%
\pgfpathlineto{\pgfqpoint{2.569962in}{0.518007in}}%
\pgfpathlineto{\pgfqpoint{2.498062in}{0.518007in}}%
\pgfpathlineto{\pgfqpoint{2.498062in}{0.522693in}}%
\pgfpathlineto{\pgfqpoint{2.426161in}{0.522693in}}%
\pgfpathlineto{\pgfqpoint{2.426161in}{0.526021in}}%
\pgfpathlineto{\pgfqpoint{2.354261in}{0.526021in}}%
\pgfpathlineto{\pgfqpoint{2.354261in}{0.529517in}}%
\pgfpathlineto{\pgfqpoint{2.282360in}{0.529517in}}%
\pgfpathlineto{\pgfqpoint{2.282360in}{0.534303in}}%
\pgfpathlineto{\pgfqpoint{2.210460in}{0.534303in}}%
\pgfpathlineto{\pgfqpoint{2.210460in}{0.539307in}}%
\pgfpathlineto{\pgfqpoint{2.138559in}{0.539307in}}%
\pgfpathlineto{\pgfqpoint{2.138559in}{0.545489in}}%
\pgfpathlineto{\pgfqpoint{2.066659in}{0.545489in}}%
\pgfpathlineto{\pgfqpoint{2.066659in}{0.551131in}}%
\pgfpathlineto{\pgfqpoint{1.994758in}{0.551131in}}%
\pgfpathlineto{\pgfqpoint{1.994758in}{0.558260in}}%
\pgfpathlineto{\pgfqpoint{1.922858in}{0.558260in}}%
\pgfpathlineto{\pgfqpoint{1.922858in}{0.566868in}}%
\pgfpathlineto{\pgfqpoint{1.850957in}{0.566868in}}%
\pgfpathlineto{\pgfqpoint{1.850957in}{0.573844in}}%
\pgfpathlineto{\pgfqpoint{1.779057in}{0.573844in}}%
\pgfpathlineto{\pgfqpoint{1.779057in}{0.580077in}}%
\pgfpathlineto{\pgfqpoint{1.707156in}{0.580077in}}%
\pgfpathlineto{\pgfqpoint{1.707156in}{0.589194in}}%
\pgfpathlineto{\pgfqpoint{1.635256in}{0.589194in}}%
\pgfpathlineto{\pgfqpoint{1.635256in}{0.666215in}}%
\pgfpathlineto{\pgfqpoint{1.563355in}{0.666215in}}%
\pgfpathlineto{\pgfqpoint{1.563355in}{0.683520in}}%
\pgfpathlineto{\pgfqpoint{1.491455in}{0.683520in}}%
\pgfpathlineto{\pgfqpoint{1.491455in}{0.699963in}}%
\pgfpathlineto{\pgfqpoint{1.419554in}{0.699963in}}%
\pgfpathlineto{\pgfqpoint{1.419554in}{0.716368in}}%
\pgfpathlineto{\pgfqpoint{1.347654in}{0.716368in}}%
\pgfpathlineto{\pgfqpoint{1.347654in}{0.769018in}}%
\pgfpathlineto{\pgfqpoint{1.275753in}{0.769018in}}%
\pgfpathlineto{\pgfqpoint{1.275753in}{0.795222in}}%
\pgfpathlineto{\pgfqpoint{1.203853in}{0.795222in}}%
\pgfpathlineto{\pgfqpoint{1.203853in}{0.824120in}}%
\pgfpathlineto{\pgfqpoint{1.131952in}{0.824120in}}%
\pgfpathlineto{\pgfqpoint{1.131952in}{0.852694in}}%
\pgfpathlineto{\pgfqpoint{1.060052in}{0.852694in}}%
\pgfpathlineto{\pgfqpoint{1.060052in}{0.888090in}}%
\pgfpathlineto{\pgfqpoint{0.988151in}{0.888090in}}%
\pgfpathlineto{\pgfqpoint{0.988151in}{0.923366in}}%
\pgfpathlineto{\pgfqpoint{0.916251in}{0.923366in}}%
\pgfpathlineto{\pgfqpoint{0.916251in}{0.962966in}}%
\pgfpathlineto{\pgfqpoint{0.844350in}{0.962966in}}%
\pgfpathlineto{\pgfqpoint{0.844350in}{1.005727in}}%
\pgfpathlineto{\pgfqpoint{0.772450in}{1.005727in}}%
\pgfpathlineto{\pgfqpoint{0.772450in}{1.047354in}}%
\pgfpathlineto{\pgfqpoint{0.700549in}{1.047354in}}%
\pgfpathlineto{\pgfqpoint{0.700549in}{1.089518in}}%
\pgfpathlineto{\pgfqpoint{0.628649in}{1.089518in}}%
\pgfpathlineto{\pgfqpoint{0.628649in}{1.131876in}}%
\pgfpathlineto{\pgfqpoint{0.556748in}{1.131876in}}%
\pgfpathlineto{\pgfqpoint{0.556748in}{1.178455in}}%
\pgfpathlineto{\pgfqpoint{0.520798in}{1.178455in}}%
\pgfpathlineto{\pgfqpoint{0.520798in}{1.178455in}}%
\pgfpathclose%
\pgfusepath{fill}%
\end{pgfscope}%
\begin{pgfscope}%
\definecolor{textcolor}{rgb}{0.150000,0.150000,0.150000}%
\pgfsetstrokecolor{textcolor}%
\pgfsetfillcolor{textcolor}%
\pgftext[x=0.880300in,y=1.397395in,,base]{\color{textcolor}\rmfamily\fontsize{9.000000}{10.800000}\selectfont \(\displaystyle c_X^-=1\)}%
\end{pgfscope}%
\begin{pgfscope}%
\pgfsetbuttcap%
\pgfsetmiterjoin%
\definecolor{currentfill}{rgb}{1.000000,1.000000,1.000000}%
\pgfsetfillcolor{currentfill}%
\pgfsetfillopacity{0.800000}%
\pgfsetlinewidth{1.003750pt}%
\definecolor{currentstroke}{rgb}{0.800000,0.800000,0.800000}%
\pgfsetstrokecolor{currentstroke}%
\pgfsetstrokeopacity{0.800000}%
\pgfsetdash{}{0pt}%
\pgfpathmoveto{\pgfqpoint{0.785843in}{0.638103in}}%
\pgfpathlineto{\pgfqpoint{1.191192in}{0.638103in}}%
\pgfpathlineto{\pgfqpoint{1.191192in}{1.265450in}}%
\pgfpathlineto{\pgfqpoint{0.785843in}{1.265450in}}%
\pgfpathlineto{\pgfqpoint{0.785843in}{0.638103in}}%
\pgfpathclose%
\pgfusepath{stroke,fill}%
\end{pgfscope}%
\begin{pgfscope}%
\definecolor{textcolor}{rgb}{0.150000,0.150000,0.150000}%
\pgfsetstrokecolor{textcolor}%
\pgfsetfillcolor{textcolor}%
\pgftext[x=0.912291in,y=1.113275in,left,base]{\color{textcolor}\rmfamily\fontsize{9.000000}{10.800000}\selectfont \(\displaystyle c_X^+\)}%
\end{pgfscope}%
\begin{pgfscope}%
\pgfsetroundcap%
\pgfsetroundjoin%
\pgfsetlinewidth{1.003750pt}%
\definecolor{currentstroke}{rgb}{0.003922,0.450980,0.698039}%
\pgfsetstrokecolor{currentstroke}%
\pgfsetdash{}{0pt}%
\pgfpathmoveto{\pgfqpoint{0.824732in}{1.001052in}}%
\pgfpathlineto{\pgfqpoint{0.873343in}{1.001052in}}%
\pgfpathlineto{\pgfqpoint{0.873343in}{1.001052in}}%
\pgfpathlineto{\pgfqpoint{0.970565in}{1.001052in}}%
\pgfpathlineto{\pgfqpoint{0.970565in}{1.001052in}}%
\pgfpathlineto{\pgfqpoint{1.019176in}{1.001052in}}%
\pgfusepath{stroke}%
\end{pgfscope}%
\begin{pgfscope}%
\definecolor{textcolor}{rgb}{0.150000,0.150000,0.150000}%
\pgfsetstrokecolor{textcolor}%
\pgfsetfillcolor{textcolor}%
\pgftext[x=1.096954in,y=0.967024in,left,base]{\color{textcolor}\rmfamily\fontsize{7.000000}{8.400000}\selectfont 1}%
\end{pgfscope}%
\begin{pgfscope}%
\pgfsetroundcap%
\pgfsetroundjoin%
\pgfsetlinewidth{1.003750pt}%
\definecolor{currentstroke}{rgb}{0.870588,0.560784,0.019608}%
\pgfsetstrokecolor{currentstroke}%
\pgfsetdash{}{0pt}%
\pgfpathmoveto{\pgfqpoint{0.824732in}{0.865486in}}%
\pgfpathlineto{\pgfqpoint{0.873343in}{0.865486in}}%
\pgfpathlineto{\pgfqpoint{0.873343in}{0.865486in}}%
\pgfpathlineto{\pgfqpoint{0.970565in}{0.865486in}}%
\pgfpathlineto{\pgfqpoint{0.970565in}{0.865486in}}%
\pgfpathlineto{\pgfqpoint{1.019176in}{0.865486in}}%
\pgfusepath{stroke}%
\end{pgfscope}%
\begin{pgfscope}%
\definecolor{textcolor}{rgb}{0.150000,0.150000,0.150000}%
\pgfsetstrokecolor{textcolor}%
\pgfsetfillcolor{textcolor}%
\pgftext[x=1.096954in,y=0.831458in,left,base]{\color{textcolor}\rmfamily\fontsize{7.000000}{8.400000}\selectfont 2}%
\end{pgfscope}%
\begin{pgfscope}%
\pgfsetroundcap%
\pgfsetroundjoin%
\pgfsetlinewidth{1.003750pt}%
\definecolor{currentstroke}{rgb}{0.007843,0.619608,0.450980}%
\pgfsetstrokecolor{currentstroke}%
\pgfsetdash{}{0pt}%
\pgfpathmoveto{\pgfqpoint{0.824732in}{0.729919in}}%
\pgfpathlineto{\pgfqpoint{0.873343in}{0.729919in}}%
\pgfpathlineto{\pgfqpoint{0.873343in}{0.729919in}}%
\pgfpathlineto{\pgfqpoint{0.970565in}{0.729919in}}%
\pgfpathlineto{\pgfqpoint{0.970565in}{0.729919in}}%
\pgfpathlineto{\pgfqpoint{1.019176in}{0.729919in}}%
\pgfusepath{stroke}%
\end{pgfscope}%
\begin{pgfscope}%
\definecolor{textcolor}{rgb}{0.150000,0.150000,0.150000}%
\pgfsetstrokecolor{textcolor}%
\pgfsetfillcolor{textcolor}%
\pgftext[x=1.096954in,y=0.695892in,left,base]{\color{textcolor}\rmfamily\fontsize{7.000000}{8.400000}\selectfont 4}%
\end{pgfscope}%
\end{pgfpicture}%
\makeatother%
\endgroup%

%% file: figures/experiments/graphs/sparse_smoothing/nodes_attributes/node_classification-Cora-APPNP-hidden=32-p_adj_plus=0.0-p_adj_minus=0.0-p_att_plus=0.001-p_att_minus=0.8-multi_class_cert-A.pgf
\begingroup%
\makeatletter%
\begin{pgfpicture}%
\pgfpathrectangle{\pgfpointorigin}{\pgfqpoint{1.375000in}{1.581250in}}%
\pgfusepath{use as bounding box, clip}%
\begin{pgfscope}%
\pgfsetbuttcap%
\pgfsetmiterjoin%
\definecolor{currentfill}{rgb}{1.000000,1.000000,1.000000}%
\pgfsetfillcolor{currentfill}%
\pgfsetlinewidth{0.000000pt}%
\definecolor{currentstroke}{rgb}{1.000000,1.000000,1.000000}%
\pgfsetstrokecolor{currentstroke}%
\pgfsetdash{}{0pt}%
\pgfpathmoveto{\pgfqpoint{0.000000in}{0.000000in}}%
\pgfpathlineto{\pgfqpoint{1.375000in}{0.000000in}}%
\pgfpathlineto{\pgfqpoint{1.375000in}{1.581250in}}%
\pgfpathlineto{\pgfqpoint{0.000000in}{1.581250in}}%
\pgfpathlineto{\pgfqpoint{0.000000in}{0.000000in}}%
\pgfpathclose%
\pgfusepath{fill}%
\end{pgfscope}%
\begin{pgfscope}%
\pgfsetbuttcap%
\pgfsetmiterjoin%
\definecolor{currentfill}{rgb}{1.000000,1.000000,1.000000}%
\pgfsetfillcolor{currentfill}%
\pgfsetlinewidth{0.000000pt}%
\definecolor{currentstroke}{rgb}{0.000000,0.000000,0.000000}%
\pgfsetstrokecolor{currentstroke}%
\pgfsetstrokeopacity{0.000000}%
\pgfsetdash{}{0pt}%
\pgfpathmoveto{\pgfqpoint{0.520798in}{0.442177in}}%
\pgfpathlineto{\pgfqpoint{1.239803in}{0.442177in}}%
\pgfpathlineto{\pgfqpoint{1.239803in}{1.314061in}}%
\pgfpathlineto{\pgfqpoint{0.520798in}{1.314061in}}%
\pgfpathlineto{\pgfqpoint{0.520798in}{0.442177in}}%
\pgfpathclose%
\pgfusepath{fill}%
\end{pgfscope}%
\begin{pgfscope}%
\pgfpathrectangle{\pgfqpoint{0.520798in}{0.442177in}}{\pgfqpoint{0.719005in}{0.871884in}}%
\pgfusepath{clip}%
\pgfsetroundcap%
\pgfsetroundjoin%
\pgfsetlinewidth{0.501875pt}%
\definecolor{currentstroke}{rgb}{0.800000,0.800000,0.800000}%
\pgfsetstrokecolor{currentstroke}%
\pgfsetdash{}{0pt}%
\pgfpathmoveto{\pgfqpoint{0.520798in}{0.442177in}}%
\pgfpathlineto{\pgfqpoint{0.520798in}{1.314061in}}%
\pgfusepath{stroke}%
\end{pgfscope}%
\begin{pgfscope}%
\definecolor{textcolor}{rgb}{0.150000,0.150000,0.150000}%
\pgfsetstrokecolor{textcolor}%
\pgfsetfillcolor{textcolor}%
\pgftext[x=0.520798in,y=0.351899in,,top]{\color{textcolor}\rmfamily\fontsize{8.000000}{9.600000}\selectfont \(\displaystyle {0}\)}%
\end{pgfscope}%
\begin{pgfscope}%
\pgfpathrectangle{\pgfqpoint{0.520798in}{0.442177in}}{\pgfqpoint{0.719005in}{0.871884in}}%
\pgfusepath{clip}%
\pgfsetroundcap%
\pgfsetroundjoin%
\pgfsetlinewidth{0.501875pt}%
\definecolor{currentstroke}{rgb}{0.800000,0.800000,0.800000}%
\pgfsetstrokecolor{currentstroke}%
\pgfsetdash{}{0pt}%
\pgfpathmoveto{\pgfqpoint{0.880300in}{0.442177in}}%
\pgfpathlineto{\pgfqpoint{0.880300in}{1.314061in}}%
\pgfusepath{stroke}%
\end{pgfscope}%
\begin{pgfscope}%
\definecolor{textcolor}{rgb}{0.150000,0.150000,0.150000}%
\pgfsetstrokecolor{textcolor}%
\pgfsetfillcolor{textcolor}%
\pgftext[x=0.880300in,y=0.351899in,,top]{\color{textcolor}\rmfamily\fontsize{8.000000}{9.600000}\selectfont \(\displaystyle {5}\)}%
\end{pgfscope}%
\begin{pgfscope}%
\pgfpathrectangle{\pgfqpoint{0.520798in}{0.442177in}}{\pgfqpoint{0.719005in}{0.871884in}}%
\pgfusepath{clip}%
\pgfsetroundcap%
\pgfsetroundjoin%
\pgfsetlinewidth{0.501875pt}%
\definecolor{currentstroke}{rgb}{0.800000,0.800000,0.800000}%
\pgfsetstrokecolor{currentstroke}%
\pgfsetdash{}{0pt}%
\pgfpathmoveto{\pgfqpoint{1.239803in}{0.442177in}}%
\pgfpathlineto{\pgfqpoint{1.239803in}{1.314061in}}%
\pgfusepath{stroke}%
\end{pgfscope}%
\begin{pgfscope}%
\definecolor{textcolor}{rgb}{0.150000,0.150000,0.150000}%
\pgfsetstrokecolor{textcolor}%
\pgfsetfillcolor{textcolor}%
\pgftext[x=1.239803in,y=0.351899in,,top]{\color{textcolor}\rmfamily\fontsize{8.000000}{9.600000}\selectfont \(\displaystyle {10}\)}%
\end{pgfscope}%
\begin{pgfscope}%
\definecolor{textcolor}{rgb}{0.150000,0.150000,0.150000}%
\pgfsetstrokecolor{textcolor}%
\pgfsetfillcolor{textcolor}%
\pgftext[x=0.880300in,y=0.198219in,,top]{\color{textcolor}\rmfamily\fontsize{10.000000}{12.000000}\selectfont Edit distance \(\displaystyle \epsilon\)}%
\end{pgfscope}%
\begin{pgfscope}%
\pgfpathrectangle{\pgfqpoint{0.520798in}{0.442177in}}{\pgfqpoint{0.719005in}{0.871884in}}%
\pgfusepath{clip}%
\pgfsetroundcap%
\pgfsetroundjoin%
\pgfsetlinewidth{0.501875pt}%
\definecolor{currentstroke}{rgb}{0.800000,0.800000,0.800000}%
\pgfsetstrokecolor{currentstroke}%
\pgfsetdash{}{0pt}%
\pgfpathmoveto{\pgfqpoint{0.520798in}{0.442177in}}%
\pgfpathlineto{\pgfqpoint{1.239803in}{0.442177in}}%
\pgfusepath{stroke}%
\end{pgfscope}%
\begin{pgfscope}%
\definecolor{textcolor}{rgb}{0.150000,0.150000,0.150000}%
\pgfsetstrokecolor{textcolor}%
\pgfsetfillcolor{textcolor}%
\pgftext[x=0.273151in, y=0.403915in, left, base]{\color{textcolor}\rmfamily\fontsize{8.000000}{9.600000}\selectfont 0\%}%
\end{pgfscope}%
\begin{pgfscope}%
\pgfpathrectangle{\pgfqpoint{0.520798in}{0.442177in}}{\pgfqpoint{0.719005in}{0.871884in}}%
\pgfusepath{clip}%
\pgfsetroundcap%
\pgfsetroundjoin%
\pgfsetlinewidth{0.501875pt}%
\definecolor{currentstroke}{rgb}{0.800000,0.800000,0.800000}%
\pgfsetstrokecolor{currentstroke}%
\pgfsetdash{}{0pt}%
\pgfpathmoveto{\pgfqpoint{0.520798in}{0.660148in}}%
\pgfpathlineto{\pgfqpoint{1.239803in}{0.660148in}}%
\pgfusepath{stroke}%
\end{pgfscope}%
\begin{pgfscope}%
\definecolor{textcolor}{rgb}{0.150000,0.150000,0.150000}%
\pgfsetstrokecolor{textcolor}%
\pgfsetfillcolor{textcolor}%
\pgftext[x=0.214138in, y=0.621886in, left, base]{\color{textcolor}\rmfamily\fontsize{8.000000}{9.600000}\selectfont 25\%}%
\end{pgfscope}%
\begin{pgfscope}%
\pgfpathrectangle{\pgfqpoint{0.520798in}{0.442177in}}{\pgfqpoint{0.719005in}{0.871884in}}%
\pgfusepath{clip}%
\pgfsetroundcap%
\pgfsetroundjoin%
\pgfsetlinewidth{0.501875pt}%
\definecolor{currentstroke}{rgb}{0.800000,0.800000,0.800000}%
\pgfsetstrokecolor{currentstroke}%
\pgfsetdash{}{0pt}%
\pgfpathmoveto{\pgfqpoint{0.520798in}{0.878119in}}%
\pgfpathlineto{\pgfqpoint{1.239803in}{0.878119in}}%
\pgfusepath{stroke}%
\end{pgfscope}%
\begin{pgfscope}%
\definecolor{textcolor}{rgb}{0.150000,0.150000,0.150000}%
\pgfsetstrokecolor{textcolor}%
\pgfsetfillcolor{textcolor}%
\pgftext[x=0.214138in, y=0.839857in, left, base]{\color{textcolor}\rmfamily\fontsize{8.000000}{9.600000}\selectfont 50\%}%
\end{pgfscope}%
\begin{pgfscope}%
\pgfpathrectangle{\pgfqpoint{0.520798in}{0.442177in}}{\pgfqpoint{0.719005in}{0.871884in}}%
\pgfusepath{clip}%
\pgfsetroundcap%
\pgfsetroundjoin%
\pgfsetlinewidth{0.501875pt}%
\definecolor{currentstroke}{rgb}{0.800000,0.800000,0.800000}%
\pgfsetstrokecolor{currentstroke}%
\pgfsetdash{}{0pt}%
\pgfpathmoveto{\pgfqpoint{0.520798in}{1.096090in}}%
\pgfpathlineto{\pgfqpoint{1.239803in}{1.096090in}}%
\pgfusepath{stroke}%
\end{pgfscope}%
\begin{pgfscope}%
\definecolor{textcolor}{rgb}{0.150000,0.150000,0.150000}%
\pgfsetstrokecolor{textcolor}%
\pgfsetfillcolor{textcolor}%
\pgftext[x=0.214138in, y=1.057828in, left, base]{\color{textcolor}\rmfamily\fontsize{8.000000}{9.600000}\selectfont 75\%}%
\end{pgfscope}%
\begin{pgfscope}%
\pgfpathrectangle{\pgfqpoint{0.520798in}{0.442177in}}{\pgfqpoint{0.719005in}{0.871884in}}%
\pgfusepath{clip}%
\pgfsetroundcap%
\pgfsetroundjoin%
\pgfsetlinewidth{0.501875pt}%
\definecolor{currentstroke}{rgb}{0.800000,0.800000,0.800000}%
\pgfsetstrokecolor{currentstroke}%
\pgfsetdash{}{0pt}%
\pgfpathmoveto{\pgfqpoint{0.520798in}{1.314061in}}%
\pgfpathlineto{\pgfqpoint{1.239803in}{1.314061in}}%
\pgfusepath{stroke}%
\end{pgfscope}%
\begin{pgfscope}%
\definecolor{textcolor}{rgb}{0.150000,0.150000,0.150000}%
\pgfsetstrokecolor{textcolor}%
\pgfsetfillcolor{textcolor}%
\pgftext[x=0.155124in, y=1.275799in, left, base]{\color{textcolor}\rmfamily\fontsize{8.000000}{9.600000}\selectfont 100\%}%
\end{pgfscope}%
\begin{pgfscope}%
\definecolor{textcolor}{rgb}{0.150000,0.150000,0.150000}%
\pgfsetstrokecolor{textcolor}%
\pgfsetfillcolor{textcolor}%
\pgftext[x=0.099569in,y=0.878119in,,bottom,rotate=90.000000]{\color{textcolor}\rmfamily\fontsize{10.000000}{12.000000}\selectfont Cert. Acc.}%
\end{pgfscope}%
\begin{pgfscope}%
\pgfsetrectcap%
\pgfsetmiterjoin%
\pgfsetlinewidth{0.752812pt}%
\definecolor{currentstroke}{rgb}{0.700000,0.700000,0.700000}%
\pgfsetstrokecolor{currentstroke}%
\pgfsetdash{}{0pt}%
\pgfpathmoveto{\pgfqpoint{0.520798in}{0.442177in}}%
\pgfpathlineto{\pgfqpoint{0.520798in}{1.314061in}}%
\pgfusepath{stroke}%
\end{pgfscope}%
\begin{pgfscope}%
\pgfsetrectcap%
\pgfsetmiterjoin%
\pgfsetlinewidth{0.752812pt}%
\definecolor{currentstroke}{rgb}{0.700000,0.700000,0.700000}%
\pgfsetstrokecolor{currentstroke}%
\pgfsetdash{}{0pt}%
\pgfpathmoveto{\pgfqpoint{1.239803in}{0.442177in}}%
\pgfpathlineto{\pgfqpoint{1.239803in}{1.314061in}}%
\pgfusepath{stroke}%
\end{pgfscope}%
\begin{pgfscope}%
\pgfsetrectcap%
\pgfsetmiterjoin%
\pgfsetlinewidth{0.752812pt}%
\definecolor{currentstroke}{rgb}{0.700000,0.700000,0.700000}%
\pgfsetstrokecolor{currentstroke}%
\pgfsetdash{}{0pt}%
\pgfpathmoveto{\pgfqpoint{0.520798in}{0.442177in}}%
\pgfpathlineto{\pgfqpoint{1.239803in}{0.442177in}}%
\pgfusepath{stroke}%
\end{pgfscope}%
\begin{pgfscope}%
\pgfsetrectcap%
\pgfsetmiterjoin%
\pgfsetlinewidth{0.752812pt}%
\definecolor{currentstroke}{rgb}{0.700000,0.700000,0.700000}%
\pgfsetstrokecolor{currentstroke}%
\pgfsetdash{}{0pt}%
\pgfpathmoveto{\pgfqpoint{0.520798in}{1.314061in}}%
\pgfpathlineto{\pgfqpoint{1.239803in}{1.314061in}}%
\pgfusepath{stroke}%
\end{pgfscope}%
\begin{pgfscope}%
\pgfpathrectangle{\pgfqpoint{0.520798in}{0.442177in}}{\pgfqpoint{0.719005in}{0.871884in}}%
\pgfusepath{clip}%
\pgfsetroundcap%
\pgfsetroundjoin%
\pgfsetlinewidth{1.003750pt}%
\definecolor{currentstroke}{rgb}{0.003922,0.450980,0.698039}%
\pgfsetstrokecolor{currentstroke}%
\pgfsetdash{}{0pt}%
\pgfpathmoveto{\pgfqpoint{0.520798in}{1.165703in}}%
\pgfpathlineto{\pgfqpoint{0.556748in}{1.165703in}}%
\pgfpathlineto{\pgfqpoint{0.556748in}{1.104255in}}%
\pgfpathlineto{\pgfqpoint{0.628649in}{1.104255in}}%
\pgfpathlineto{\pgfqpoint{0.628649in}{1.001765in}}%
\pgfpathlineto{\pgfqpoint{0.700549in}{1.001765in}}%
\pgfpathlineto{\pgfqpoint{0.700549in}{0.709792in}}%
\pgfpathlineto{\pgfqpoint{0.772450in}{0.709792in}}%
\pgfpathlineto{\pgfqpoint{0.772450in}{0.580097in}}%
\pgfpathlineto{\pgfqpoint{0.844350in}{0.580097in}}%
\pgfpathlineto{\pgfqpoint{0.844350in}{0.564913in}}%
\pgfpathlineto{\pgfqpoint{0.916251in}{0.564913in}}%
\pgfpathlineto{\pgfqpoint{0.916251in}{0.553683in}}%
\pgfpathlineto{\pgfqpoint{0.988151in}{0.553683in}}%
\pgfpathlineto{\pgfqpoint{0.988151in}{0.539844in}}%
\pgfpathlineto{\pgfqpoint{1.060052in}{0.539844in}}%
\pgfpathlineto{\pgfqpoint{1.060052in}{0.492869in}}%
\pgfpathlineto{\pgfqpoint{1.131952in}{0.492869in}}%
\pgfpathlineto{\pgfqpoint{1.131952in}{0.486305in}}%
\pgfpathlineto{\pgfqpoint{1.203853in}{0.486305in}}%
\pgfpathlineto{\pgfqpoint{1.203853in}{0.466376in}}%
\pgfpathlineto{\pgfqpoint{1.241470in}{0.466376in}}%
\pgfusepath{stroke}%
\end{pgfscope}%
\begin{pgfscope}%
\pgfpathrectangle{\pgfqpoint{0.520798in}{0.442177in}}{\pgfqpoint{0.719005in}{0.871884in}}%
\pgfusepath{clip}%
\pgfsetbuttcap%
\pgfsetroundjoin%
\definecolor{currentfill}{rgb}{0.003922,0.450980,0.698039}%
\pgfsetfillcolor{currentfill}%
\pgfsetfillopacity{0.500000}%
\pgfsetlinewidth{0.000000pt}%
\definecolor{currentstroke}{rgb}{0.003922,0.450980,0.698039}%
\pgfsetstrokecolor{currentstroke}%
\pgfsetstrokeopacity{0.500000}%
\pgfsetdash{}{0pt}%
\pgfpathmoveto{\pgfqpoint{0.520798in}{1.178455in}}%
\pgfpathlineto{\pgfqpoint{0.520798in}{1.152950in}}%
\pgfpathlineto{\pgfqpoint{0.556748in}{1.152950in}}%
\pgfpathlineto{\pgfqpoint{0.556748in}{1.088241in}}%
\pgfpathlineto{\pgfqpoint{0.628649in}{1.088241in}}%
\pgfpathlineto{\pgfqpoint{0.628649in}{0.984178in}}%
\pgfpathlineto{\pgfqpoint{0.700549in}{0.984178in}}%
\pgfpathlineto{\pgfqpoint{0.700549in}{0.703216in}}%
\pgfpathlineto{\pgfqpoint{0.772450in}{0.703216in}}%
\pgfpathlineto{\pgfqpoint{0.772450in}{0.570999in}}%
\pgfpathlineto{\pgfqpoint{0.844350in}{0.570999in}}%
\pgfpathlineto{\pgfqpoint{0.844350in}{0.554989in}}%
\pgfpathlineto{\pgfqpoint{0.916251in}{0.554989in}}%
\pgfpathlineto{\pgfqpoint{0.916251in}{0.542660in}}%
\pgfpathlineto{\pgfqpoint{0.988151in}{0.542660in}}%
\pgfpathlineto{\pgfqpoint{0.988151in}{0.529201in}}%
\pgfpathlineto{\pgfqpoint{1.060052in}{0.529201in}}%
\pgfpathlineto{\pgfqpoint{1.060052in}{0.484179in}}%
\pgfpathlineto{\pgfqpoint{1.131952in}{0.484179in}}%
\pgfpathlineto{\pgfqpoint{1.131952in}{0.477227in}}%
\pgfpathlineto{\pgfqpoint{1.203853in}{0.477227in}}%
\pgfpathlineto{\pgfqpoint{1.203853in}{0.460660in}}%
\pgfpathlineto{\pgfqpoint{1.275753in}{0.460660in}}%
\pgfpathlineto{\pgfqpoint{1.275753in}{0.452951in}}%
\pgfpathlineto{\pgfqpoint{1.347654in}{0.452951in}}%
\pgfpathlineto{\pgfqpoint{1.347654in}{0.448808in}}%
\pgfpathlineto{\pgfqpoint{1.419554in}{0.448808in}}%
\pgfpathlineto{\pgfqpoint{1.419554in}{0.442177in}}%
\pgfpathlineto{\pgfqpoint{2.066659in}{0.442177in}}%
\pgfpathlineto{\pgfqpoint{2.066659in}{0.442177in}}%
\pgfpathlineto{\pgfqpoint{2.677813in}{0.442177in}}%
\pgfpathlineto{\pgfqpoint{2.677813in}{0.442177in}}%
\pgfpathlineto{\pgfqpoint{2.677813in}{0.442177in}}%
\pgfpathlineto{\pgfqpoint{2.066659in}{0.442177in}}%
\pgfpathlineto{\pgfqpoint{2.066659in}{0.442177in}}%
\pgfpathlineto{\pgfqpoint{1.419554in}{0.442177in}}%
\pgfpathlineto{\pgfqpoint{1.419554in}{0.454209in}}%
\pgfpathlineto{\pgfqpoint{1.347654in}{0.454209in}}%
\pgfpathlineto{\pgfqpoint{1.347654in}{0.457183in}}%
\pgfpathlineto{\pgfqpoint{1.275753in}{0.457183in}}%
\pgfpathlineto{\pgfqpoint{1.275753in}{0.472092in}}%
\pgfpathlineto{\pgfqpoint{1.203853in}{0.472092in}}%
\pgfpathlineto{\pgfqpoint{1.203853in}{0.495382in}}%
\pgfpathlineto{\pgfqpoint{1.131952in}{0.495382in}}%
\pgfpathlineto{\pgfqpoint{1.131952in}{0.501558in}}%
\pgfpathlineto{\pgfqpoint{1.060052in}{0.501558in}}%
\pgfpathlineto{\pgfqpoint{1.060052in}{0.550486in}}%
\pgfpathlineto{\pgfqpoint{0.988151in}{0.550486in}}%
\pgfpathlineto{\pgfqpoint{0.988151in}{0.564707in}}%
\pgfpathlineto{\pgfqpoint{0.916251in}{0.564707in}}%
\pgfpathlineto{\pgfqpoint{0.916251in}{0.574836in}}%
\pgfpathlineto{\pgfqpoint{0.844350in}{0.574836in}}%
\pgfpathlineto{\pgfqpoint{0.844350in}{0.589194in}}%
\pgfpathlineto{\pgfqpoint{0.772450in}{0.589194in}}%
\pgfpathlineto{\pgfqpoint{0.772450in}{0.716368in}}%
\pgfpathlineto{\pgfqpoint{0.700549in}{0.716368in}}%
\pgfpathlineto{\pgfqpoint{0.700549in}{1.019351in}}%
\pgfpathlineto{\pgfqpoint{0.628649in}{1.019351in}}%
\pgfpathlineto{\pgfqpoint{0.628649in}{1.120270in}}%
\pgfpathlineto{\pgfqpoint{0.556748in}{1.120270in}}%
\pgfpathlineto{\pgfqpoint{0.556748in}{1.178455in}}%
\pgfpathlineto{\pgfqpoint{0.520798in}{1.178455in}}%
\pgfpathlineto{\pgfqpoint{0.520798in}{1.178455in}}%
\pgfpathclose%
\pgfusepath{fill}%
\end{pgfscope}%
\begin{pgfscope}%
\pgfpathrectangle{\pgfqpoint{0.520798in}{0.442177in}}{\pgfqpoint{0.719005in}{0.871884in}}%
\pgfusepath{clip}%
\pgfsetroundcap%
\pgfsetroundjoin%
\pgfsetlinewidth{1.003750pt}%
\definecolor{currentstroke}{rgb}{0.870588,0.560784,0.019608}%
\pgfsetstrokecolor{currentstroke}%
\pgfsetdash{}{0pt}%
\pgfpathmoveto{\pgfqpoint{0.520798in}{1.165703in}}%
\pgfpathlineto{\pgfqpoint{0.556748in}{1.165703in}}%
\pgfpathlineto{\pgfqpoint{0.556748in}{1.104255in}}%
\pgfpathlineto{\pgfqpoint{0.628649in}{1.104255in}}%
\pgfpathlineto{\pgfqpoint{0.628649in}{1.001765in}}%
\pgfpathlineto{\pgfqpoint{0.700549in}{1.001765in}}%
\pgfpathlineto{\pgfqpoint{0.700549in}{0.709792in}}%
\pgfpathlineto{\pgfqpoint{0.772450in}{0.709792in}}%
\pgfpathlineto{\pgfqpoint{0.772450in}{0.580097in}}%
\pgfpathlineto{\pgfqpoint{0.844350in}{0.580097in}}%
\pgfpathlineto{\pgfqpoint{0.844350in}{0.567918in}}%
\pgfpathlineto{\pgfqpoint{0.916251in}{0.567918in}}%
\pgfpathlineto{\pgfqpoint{0.916251in}{0.556767in}}%
\pgfpathlineto{\pgfqpoint{0.988151in}{0.556767in}}%
\pgfpathlineto{\pgfqpoint{0.988151in}{0.539844in}}%
\pgfpathlineto{\pgfqpoint{1.060052in}{0.539844in}}%
\pgfpathlineto{\pgfqpoint{1.060052in}{0.492948in}}%
\pgfpathlineto{\pgfqpoint{1.131952in}{0.492948in}}%
\pgfpathlineto{\pgfqpoint{1.131952in}{0.486305in}}%
\pgfpathlineto{\pgfqpoint{1.203853in}{0.486305in}}%
\pgfpathlineto{\pgfqpoint{1.203853in}{0.466376in}}%
\pgfpathlineto{\pgfqpoint{1.241470in}{0.466376in}}%
\pgfusepath{stroke}%
\end{pgfscope}%
\begin{pgfscope}%
\pgfpathrectangle{\pgfqpoint{0.520798in}{0.442177in}}{\pgfqpoint{0.719005in}{0.871884in}}%
\pgfusepath{clip}%
\pgfsetbuttcap%
\pgfsetroundjoin%
\definecolor{currentfill}{rgb}{0.870588,0.560784,0.019608}%
\pgfsetfillcolor{currentfill}%
\pgfsetfillopacity{0.500000}%
\pgfsetlinewidth{0.000000pt}%
\definecolor{currentstroke}{rgb}{0.870588,0.560784,0.019608}%
\pgfsetstrokecolor{currentstroke}%
\pgfsetstrokeopacity{0.500000}%
\pgfsetdash{}{0pt}%
\pgfpathmoveto{\pgfqpoint{0.520798in}{1.178455in}}%
\pgfpathlineto{\pgfqpoint{0.520798in}{1.152950in}}%
\pgfpathlineto{\pgfqpoint{0.556748in}{1.152950in}}%
\pgfpathlineto{\pgfqpoint{0.556748in}{1.088241in}}%
\pgfpathlineto{\pgfqpoint{0.628649in}{1.088241in}}%
\pgfpathlineto{\pgfqpoint{0.628649in}{0.984178in}}%
\pgfpathlineto{\pgfqpoint{0.700549in}{0.984178in}}%
\pgfpathlineto{\pgfqpoint{0.700549in}{0.703216in}}%
\pgfpathlineto{\pgfqpoint{0.772450in}{0.703216in}}%
\pgfpathlineto{\pgfqpoint{0.772450in}{0.570999in}}%
\pgfpathlineto{\pgfqpoint{0.844350in}{0.570999in}}%
\pgfpathlineto{\pgfqpoint{0.844350in}{0.557571in}}%
\pgfpathlineto{\pgfqpoint{0.916251in}{0.557571in}}%
\pgfpathlineto{\pgfqpoint{0.916251in}{0.546525in}}%
\pgfpathlineto{\pgfqpoint{0.988151in}{0.546525in}}%
\pgfpathlineto{\pgfqpoint{0.988151in}{0.529201in}}%
\pgfpathlineto{\pgfqpoint{1.060052in}{0.529201in}}%
\pgfpathlineto{\pgfqpoint{1.060052in}{0.484139in}}%
\pgfpathlineto{\pgfqpoint{1.131952in}{0.484139in}}%
\pgfpathlineto{\pgfqpoint{1.131952in}{0.477227in}}%
\pgfpathlineto{\pgfqpoint{1.203853in}{0.477227in}}%
\pgfpathlineto{\pgfqpoint{1.203853in}{0.460660in}}%
\pgfpathlineto{\pgfqpoint{1.275753in}{0.460660in}}%
\pgfpathlineto{\pgfqpoint{1.275753in}{0.453534in}}%
\pgfpathlineto{\pgfqpoint{1.347654in}{0.453534in}}%
\pgfpathlineto{\pgfqpoint{1.347654in}{0.448808in}}%
\pgfpathlineto{\pgfqpoint{1.419554in}{0.448808in}}%
\pgfpathlineto{\pgfqpoint{1.419554in}{0.442177in}}%
\pgfpathlineto{\pgfqpoint{2.066659in}{0.442177in}}%
\pgfpathlineto{\pgfqpoint{2.066659in}{0.442177in}}%
\pgfpathlineto{\pgfqpoint{2.677813in}{0.442177in}}%
\pgfpathlineto{\pgfqpoint{2.677813in}{0.442177in}}%
\pgfpathlineto{\pgfqpoint{2.677813in}{0.442177in}}%
\pgfpathlineto{\pgfqpoint{2.066659in}{0.442177in}}%
\pgfpathlineto{\pgfqpoint{2.066659in}{0.442177in}}%
\pgfpathlineto{\pgfqpoint{1.419554in}{0.442177in}}%
\pgfpathlineto{\pgfqpoint{1.419554in}{0.454209in}}%
\pgfpathlineto{\pgfqpoint{1.347654in}{0.454209in}}%
\pgfpathlineto{\pgfqpoint{1.347654in}{0.458024in}}%
\pgfpathlineto{\pgfqpoint{1.275753in}{0.458024in}}%
\pgfpathlineto{\pgfqpoint{1.275753in}{0.472092in}}%
\pgfpathlineto{\pgfqpoint{1.203853in}{0.472092in}}%
\pgfpathlineto{\pgfqpoint{1.203853in}{0.495382in}}%
\pgfpathlineto{\pgfqpoint{1.131952in}{0.495382in}}%
\pgfpathlineto{\pgfqpoint{1.131952in}{0.501756in}}%
\pgfpathlineto{\pgfqpoint{1.060052in}{0.501756in}}%
\pgfpathlineto{\pgfqpoint{1.060052in}{0.550486in}}%
\pgfpathlineto{\pgfqpoint{0.988151in}{0.550486in}}%
\pgfpathlineto{\pgfqpoint{0.988151in}{0.567010in}}%
\pgfpathlineto{\pgfqpoint{0.916251in}{0.567010in}}%
\pgfpathlineto{\pgfqpoint{0.916251in}{0.578265in}}%
\pgfpathlineto{\pgfqpoint{0.844350in}{0.578265in}}%
\pgfpathlineto{\pgfqpoint{0.844350in}{0.589194in}}%
\pgfpathlineto{\pgfqpoint{0.772450in}{0.589194in}}%
\pgfpathlineto{\pgfqpoint{0.772450in}{0.716368in}}%
\pgfpathlineto{\pgfqpoint{0.700549in}{0.716368in}}%
\pgfpathlineto{\pgfqpoint{0.700549in}{1.019351in}}%
\pgfpathlineto{\pgfqpoint{0.628649in}{1.019351in}}%
\pgfpathlineto{\pgfqpoint{0.628649in}{1.120270in}}%
\pgfpathlineto{\pgfqpoint{0.556748in}{1.120270in}}%
\pgfpathlineto{\pgfqpoint{0.556748in}{1.178455in}}%
\pgfpathlineto{\pgfqpoint{0.520798in}{1.178455in}}%
\pgfpathlineto{\pgfqpoint{0.520798in}{1.178455in}}%
\pgfpathclose%
\pgfusepath{fill}%
\end{pgfscope}%
\begin{pgfscope}%
\pgfpathrectangle{\pgfqpoint{0.520798in}{0.442177in}}{\pgfqpoint{0.719005in}{0.871884in}}%
\pgfusepath{clip}%
\pgfsetroundcap%
\pgfsetroundjoin%
\pgfsetlinewidth{1.003750pt}%
\definecolor{currentstroke}{rgb}{0.007843,0.619608,0.450980}%
\pgfsetstrokecolor{currentstroke}%
\pgfsetdash{}{0pt}%
\pgfpathmoveto{\pgfqpoint{0.520798in}{1.165703in}}%
\pgfpathlineto{\pgfqpoint{0.556748in}{1.165703in}}%
\pgfpathlineto{\pgfqpoint{0.556748in}{1.104255in}}%
\pgfpathlineto{\pgfqpoint{0.628649in}{1.104255in}}%
\pgfpathlineto{\pgfqpoint{0.628649in}{1.001765in}}%
\pgfpathlineto{\pgfqpoint{0.700549in}{1.001765in}}%
\pgfpathlineto{\pgfqpoint{0.700549in}{0.709792in}}%
\pgfpathlineto{\pgfqpoint{0.772450in}{0.709792in}}%
\pgfpathlineto{\pgfqpoint{0.772450in}{0.580097in}}%
\pgfpathlineto{\pgfqpoint{0.844350in}{0.580097in}}%
\pgfpathlineto{\pgfqpoint{0.844350in}{0.567918in}}%
\pgfpathlineto{\pgfqpoint{0.916251in}{0.567918in}}%
\pgfpathlineto{\pgfqpoint{0.916251in}{0.556767in}}%
\pgfpathlineto{\pgfqpoint{0.988151in}{0.556767in}}%
\pgfpathlineto{\pgfqpoint{0.988151in}{0.539844in}}%
\pgfpathlineto{\pgfqpoint{1.060052in}{0.539844in}}%
\pgfpathlineto{\pgfqpoint{1.060052in}{0.492948in}}%
\pgfpathlineto{\pgfqpoint{1.131952in}{0.492948in}}%
\pgfpathlineto{\pgfqpoint{1.131952in}{0.486305in}}%
\pgfpathlineto{\pgfqpoint{1.203853in}{0.486305in}}%
\pgfpathlineto{\pgfqpoint{1.203853in}{0.466376in}}%
\pgfpathlineto{\pgfqpoint{1.241470in}{0.466376in}}%
\pgfusepath{stroke}%
\end{pgfscope}%
\begin{pgfscope}%
\pgfpathrectangle{\pgfqpoint{0.520798in}{0.442177in}}{\pgfqpoint{0.719005in}{0.871884in}}%
\pgfusepath{clip}%
\pgfsetbuttcap%
\pgfsetroundjoin%
\definecolor{currentfill}{rgb}{0.007843,0.619608,0.450980}%
\pgfsetfillcolor{currentfill}%
\pgfsetfillopacity{0.500000}%
\pgfsetlinewidth{0.000000pt}%
\definecolor{currentstroke}{rgb}{0.007843,0.619608,0.450980}%
\pgfsetstrokecolor{currentstroke}%
\pgfsetstrokeopacity{0.500000}%
\pgfsetdash{}{0pt}%
\pgfpathmoveto{\pgfqpoint{0.520798in}{1.178455in}}%
\pgfpathlineto{\pgfqpoint{0.520798in}{1.152950in}}%
\pgfpathlineto{\pgfqpoint{0.556748in}{1.152950in}}%
\pgfpathlineto{\pgfqpoint{0.556748in}{1.088241in}}%
\pgfpathlineto{\pgfqpoint{0.628649in}{1.088241in}}%
\pgfpathlineto{\pgfqpoint{0.628649in}{0.984178in}}%
\pgfpathlineto{\pgfqpoint{0.700549in}{0.984178in}}%
\pgfpathlineto{\pgfqpoint{0.700549in}{0.703216in}}%
\pgfpathlineto{\pgfqpoint{0.772450in}{0.703216in}}%
\pgfpathlineto{\pgfqpoint{0.772450in}{0.570999in}}%
\pgfpathlineto{\pgfqpoint{0.844350in}{0.570999in}}%
\pgfpathlineto{\pgfqpoint{0.844350in}{0.557571in}}%
\pgfpathlineto{\pgfqpoint{0.916251in}{0.557571in}}%
\pgfpathlineto{\pgfqpoint{0.916251in}{0.546525in}}%
\pgfpathlineto{\pgfqpoint{0.988151in}{0.546525in}}%
\pgfpathlineto{\pgfqpoint{0.988151in}{0.529201in}}%
\pgfpathlineto{\pgfqpoint{1.060052in}{0.529201in}}%
\pgfpathlineto{\pgfqpoint{1.060052in}{0.484139in}}%
\pgfpathlineto{\pgfqpoint{1.131952in}{0.484139in}}%
\pgfpathlineto{\pgfqpoint{1.131952in}{0.477227in}}%
\pgfpathlineto{\pgfqpoint{1.203853in}{0.477227in}}%
\pgfpathlineto{\pgfqpoint{1.203853in}{0.460660in}}%
\pgfpathlineto{\pgfqpoint{1.275753in}{0.460660in}}%
\pgfpathlineto{\pgfqpoint{1.275753in}{0.453534in}}%
\pgfpathlineto{\pgfqpoint{1.347654in}{0.453534in}}%
\pgfpathlineto{\pgfqpoint{1.347654in}{0.448808in}}%
\pgfpathlineto{\pgfqpoint{1.419554in}{0.448808in}}%
\pgfpathlineto{\pgfqpoint{1.419554in}{0.442177in}}%
\pgfpathlineto{\pgfqpoint{2.066659in}{0.442177in}}%
\pgfpathlineto{\pgfqpoint{2.066659in}{0.442177in}}%
\pgfpathlineto{\pgfqpoint{2.677813in}{0.442177in}}%
\pgfpathlineto{\pgfqpoint{2.677813in}{0.442177in}}%
\pgfpathlineto{\pgfqpoint{2.677813in}{0.442177in}}%
\pgfpathlineto{\pgfqpoint{2.066659in}{0.442177in}}%
\pgfpathlineto{\pgfqpoint{2.066659in}{0.442177in}}%
\pgfpathlineto{\pgfqpoint{1.419554in}{0.442177in}}%
\pgfpathlineto{\pgfqpoint{1.419554in}{0.454209in}}%
\pgfpathlineto{\pgfqpoint{1.347654in}{0.454209in}}%
\pgfpathlineto{\pgfqpoint{1.347654in}{0.458024in}}%
\pgfpathlineto{\pgfqpoint{1.275753in}{0.458024in}}%
\pgfpathlineto{\pgfqpoint{1.275753in}{0.472092in}}%
\pgfpathlineto{\pgfqpoint{1.203853in}{0.472092in}}%
\pgfpathlineto{\pgfqpoint{1.203853in}{0.495382in}}%
\pgfpathlineto{\pgfqpoint{1.131952in}{0.495382in}}%
\pgfpathlineto{\pgfqpoint{1.131952in}{0.501756in}}%
\pgfpathlineto{\pgfqpoint{1.060052in}{0.501756in}}%
\pgfpathlineto{\pgfqpoint{1.060052in}{0.550486in}}%
\pgfpathlineto{\pgfqpoint{0.988151in}{0.550486in}}%
\pgfpathlineto{\pgfqpoint{0.988151in}{0.567010in}}%
\pgfpathlineto{\pgfqpoint{0.916251in}{0.567010in}}%
\pgfpathlineto{\pgfqpoint{0.916251in}{0.578265in}}%
\pgfpathlineto{\pgfqpoint{0.844350in}{0.578265in}}%
\pgfpathlineto{\pgfqpoint{0.844350in}{0.589194in}}%
\pgfpathlineto{\pgfqpoint{0.772450in}{0.589194in}}%
\pgfpathlineto{\pgfqpoint{0.772450in}{0.716368in}}%
\pgfpathlineto{\pgfqpoint{0.700549in}{0.716368in}}%
\pgfpathlineto{\pgfqpoint{0.700549in}{1.019351in}}%
\pgfpathlineto{\pgfqpoint{0.628649in}{1.019351in}}%
\pgfpathlineto{\pgfqpoint{0.628649in}{1.120270in}}%
\pgfpathlineto{\pgfqpoint{0.556748in}{1.120270in}}%
\pgfpathlineto{\pgfqpoint{0.556748in}{1.178455in}}%
\pgfpathlineto{\pgfqpoint{0.520798in}{1.178455in}}%
\pgfpathlineto{\pgfqpoint{0.520798in}{1.178455in}}%
\pgfpathclose%
\pgfusepath{fill}%
\end{pgfscope}%
\begin{pgfscope}%
\definecolor{textcolor}{rgb}{0.150000,0.150000,0.150000}%
\pgfsetstrokecolor{textcolor}%
\pgfsetfillcolor{textcolor}%
\pgftext[x=0.880300in,y=1.397395in,,base]{\color{textcolor}\rmfamily\fontsize{9.000000}{10.800000}\selectfont \(\displaystyle c_X^+=1\)}%
\end{pgfscope}%
\begin{pgfscope}%
\pgfsetbuttcap%
\pgfsetmiterjoin%
\definecolor{currentfill}{rgb}{1.000000,1.000000,1.000000}%
\pgfsetfillcolor{currentfill}%
\pgfsetfillopacity{0.800000}%
\pgfsetlinewidth{1.003750pt}%
\definecolor{currentstroke}{rgb}{0.800000,0.800000,0.800000}%
\pgfsetstrokecolor{currentstroke}%
\pgfsetstrokeopacity{0.800000}%
\pgfsetdash{}{0pt}%
\pgfpathmoveto{\pgfqpoint{0.785843in}{0.638103in}}%
\pgfpathlineto{\pgfqpoint{1.191192in}{0.638103in}}%
\pgfpathlineto{\pgfqpoint{1.191192in}{1.265450in}}%
\pgfpathlineto{\pgfqpoint{0.785843in}{1.265450in}}%
\pgfpathlineto{\pgfqpoint{0.785843in}{0.638103in}}%
\pgfpathclose%
\pgfusepath{stroke,fill}%
\end{pgfscope}%
\begin{pgfscope}%
\definecolor{textcolor}{rgb}{0.150000,0.150000,0.150000}%
\pgfsetstrokecolor{textcolor}%
\pgfsetfillcolor{textcolor}%
\pgftext[x=0.912291in,y=1.113275in,left,base]{\color{textcolor}\rmfamily\fontsize{9.000000}{10.800000}\selectfont \(\displaystyle c_X^-\)}%
\end{pgfscope}%
\begin{pgfscope}%
\pgfsetroundcap%
\pgfsetroundjoin%
\pgfsetlinewidth{1.003750pt}%
\definecolor{currentstroke}{rgb}{0.003922,0.450980,0.698039}%
\pgfsetstrokecolor{currentstroke}%
\pgfsetdash{}{0pt}%
\pgfpathmoveto{\pgfqpoint{0.824732in}{1.001052in}}%
\pgfpathlineto{\pgfqpoint{0.873343in}{1.001052in}}%
\pgfpathlineto{\pgfqpoint{0.873343in}{1.001052in}}%
\pgfpathlineto{\pgfqpoint{0.970565in}{1.001052in}}%
\pgfpathlineto{\pgfqpoint{0.970565in}{1.001052in}}%
\pgfpathlineto{\pgfqpoint{1.019176in}{1.001052in}}%
\pgfusepath{stroke}%
\end{pgfscope}%
\begin{pgfscope}%
\definecolor{textcolor}{rgb}{0.150000,0.150000,0.150000}%
\pgfsetstrokecolor{textcolor}%
\pgfsetfillcolor{textcolor}%
\pgftext[x=1.096954in,y=0.967024in,left,base]{\color{textcolor}\rmfamily\fontsize{7.000000}{8.400000}\selectfont 1}%
\end{pgfscope}%
\begin{pgfscope}%
\pgfsetroundcap%
\pgfsetroundjoin%
\pgfsetlinewidth{1.003750pt}%
\definecolor{currentstroke}{rgb}{0.870588,0.560784,0.019608}%
\pgfsetstrokecolor{currentstroke}%
\pgfsetdash{}{0pt}%
\pgfpathmoveto{\pgfqpoint{0.824732in}{0.865486in}}%
\pgfpathlineto{\pgfqpoint{0.873343in}{0.865486in}}%
\pgfpathlineto{\pgfqpoint{0.873343in}{0.865486in}}%
\pgfpathlineto{\pgfqpoint{0.970565in}{0.865486in}}%
\pgfpathlineto{\pgfqpoint{0.970565in}{0.865486in}}%
\pgfpathlineto{\pgfqpoint{1.019176in}{0.865486in}}%
\pgfusepath{stroke}%
\end{pgfscope}%
\begin{pgfscope}%
\definecolor{textcolor}{rgb}{0.150000,0.150000,0.150000}%
\pgfsetstrokecolor{textcolor}%
\pgfsetfillcolor{textcolor}%
\pgftext[x=1.096954in,y=0.831458in,left,base]{\color{textcolor}\rmfamily\fontsize{7.000000}{8.400000}\selectfont 2}%
\end{pgfscope}%
\begin{pgfscope}%
\pgfsetroundcap%
\pgfsetroundjoin%
\pgfsetlinewidth{1.003750pt}%
\definecolor{currentstroke}{rgb}{0.007843,0.619608,0.450980}%
\pgfsetstrokecolor{currentstroke}%
\pgfsetdash{}{0pt}%
\pgfpathmoveto{\pgfqpoint{0.824732in}{0.729919in}}%
\pgfpathlineto{\pgfqpoint{0.873343in}{0.729919in}}%
\pgfpathlineto{\pgfqpoint{0.873343in}{0.729919in}}%
\pgfpathlineto{\pgfqpoint{0.970565in}{0.729919in}}%
\pgfpathlineto{\pgfqpoint{0.970565in}{0.729919in}}%
\pgfpathlineto{\pgfqpoint{1.019176in}{0.729919in}}%
\pgfusepath{stroke}%
\end{pgfscope}%
\begin{pgfscope}%
\definecolor{textcolor}{rgb}{0.150000,0.150000,0.150000}%
\pgfsetstrokecolor{textcolor}%
\pgfsetfillcolor{textcolor}%
\pgftext[x=1.096954in,y=0.695892in,left,base]{\color{textcolor}\rmfamily\fontsize{7.000000}{8.400000}\selectfont 4}%
\end{pgfscope}%
\end{pgfpicture}%
\makeatother%
\endgroup%

%% file: figures/experiments/graphs/sparse_smoothing/nodes_structure/node_classification-Cora-APPNP-hidden=32-p_adj_plus=0.001-p_adj_minus=0.8-p_att_plus=0.0-p_att_minus=0.0-multi_class_cert-B.pgf
\begingroup%
\makeatletter%
\begin{pgfpicture}%
\pgfpathrectangle{\pgfpointorigin}{\pgfqpoint{1.375000in}{1.581250in}}%
\pgfusepath{use as bounding box, clip}%
\begin{pgfscope}%
\pgfsetbuttcap%
\pgfsetmiterjoin%
\definecolor{currentfill}{rgb}{1.000000,1.000000,1.000000}%
\pgfsetfillcolor{currentfill}%
\pgfsetlinewidth{0.000000pt}%
\definecolor{currentstroke}{rgb}{1.000000,1.000000,1.000000}%
\pgfsetstrokecolor{currentstroke}%
\pgfsetdash{}{0pt}%
\pgfpathmoveto{\pgfqpoint{0.000000in}{0.000000in}}%
\pgfpathlineto{\pgfqpoint{1.375000in}{0.000000in}}%
\pgfpathlineto{\pgfqpoint{1.375000in}{1.581250in}}%
\pgfpathlineto{\pgfqpoint{0.000000in}{1.581250in}}%
\pgfpathlineto{\pgfqpoint{0.000000in}{0.000000in}}%
\pgfpathclose%
\pgfusepath{fill}%
\end{pgfscope}%
\begin{pgfscope}%
\pgfsetbuttcap%
\pgfsetmiterjoin%
\definecolor{currentfill}{rgb}{1.000000,1.000000,1.000000}%
\pgfsetfillcolor{currentfill}%
\pgfsetlinewidth{0.000000pt}%
\definecolor{currentstroke}{rgb}{0.000000,0.000000,0.000000}%
\pgfsetstrokecolor{currentstroke}%
\pgfsetstrokeopacity{0.000000}%
\pgfsetdash{}{0pt}%
\pgfpathmoveto{\pgfqpoint{0.520798in}{0.442177in}}%
\pgfpathlineto{\pgfqpoint{1.239803in}{0.442177in}}%
\pgfpathlineto{\pgfqpoint{1.239803in}{1.314061in}}%
\pgfpathlineto{\pgfqpoint{0.520798in}{1.314061in}}%
\pgfpathlineto{\pgfqpoint{0.520798in}{0.442177in}}%
\pgfpathclose%
\pgfusepath{fill}%
\end{pgfscope}%
\begin{pgfscope}%
\pgfpathrectangle{\pgfqpoint{0.520798in}{0.442177in}}{\pgfqpoint{0.719005in}{0.871884in}}%
\pgfusepath{clip}%
\pgfsetroundcap%
\pgfsetroundjoin%
\pgfsetlinewidth{0.501875pt}%
\definecolor{currentstroke}{rgb}{0.800000,0.800000,0.800000}%
\pgfsetstrokecolor{currentstroke}%
\pgfsetdash{}{0pt}%
\pgfpathmoveto{\pgfqpoint{0.520798in}{0.442177in}}%
\pgfpathlineto{\pgfqpoint{0.520798in}{1.314061in}}%
\pgfusepath{stroke}%
\end{pgfscope}%
\begin{pgfscope}%
\definecolor{textcolor}{rgb}{0.150000,0.150000,0.150000}%
\pgfsetstrokecolor{textcolor}%
\pgfsetfillcolor{textcolor}%
\pgftext[x=0.520798in,y=0.351899in,,top]{\color{textcolor}\rmfamily\fontsize{8.000000}{9.600000}\selectfont \(\displaystyle {0}\)}%
\end{pgfscope}%
\begin{pgfscope}%
\pgfpathrectangle{\pgfqpoint{0.520798in}{0.442177in}}{\pgfqpoint{0.719005in}{0.871884in}}%
\pgfusepath{clip}%
\pgfsetroundcap%
\pgfsetroundjoin%
\pgfsetlinewidth{0.501875pt}%
\definecolor{currentstroke}{rgb}{0.800000,0.800000,0.800000}%
\pgfsetstrokecolor{currentstroke}%
\pgfsetdash{}{0pt}%
\pgfpathmoveto{\pgfqpoint{0.880300in}{0.442177in}}%
\pgfpathlineto{\pgfqpoint{0.880300in}{1.314061in}}%
\pgfusepath{stroke}%
\end{pgfscope}%
\begin{pgfscope}%
\definecolor{textcolor}{rgb}{0.150000,0.150000,0.150000}%
\pgfsetstrokecolor{textcolor}%
\pgfsetfillcolor{textcolor}%
\pgftext[x=0.880300in,y=0.351899in,,top]{\color{textcolor}\rmfamily\fontsize{8.000000}{9.600000}\selectfont \(\displaystyle {5}\)}%
\end{pgfscope}%
\begin{pgfscope}%
\pgfpathrectangle{\pgfqpoint{0.520798in}{0.442177in}}{\pgfqpoint{0.719005in}{0.871884in}}%
\pgfusepath{clip}%
\pgfsetroundcap%
\pgfsetroundjoin%
\pgfsetlinewidth{0.501875pt}%
\definecolor{currentstroke}{rgb}{0.800000,0.800000,0.800000}%
\pgfsetstrokecolor{currentstroke}%
\pgfsetdash{}{0pt}%
\pgfpathmoveto{\pgfqpoint{1.239803in}{0.442177in}}%
\pgfpathlineto{\pgfqpoint{1.239803in}{1.314061in}}%
\pgfusepath{stroke}%
\end{pgfscope}%
\begin{pgfscope}%
\definecolor{textcolor}{rgb}{0.150000,0.150000,0.150000}%
\pgfsetstrokecolor{textcolor}%
\pgfsetfillcolor{textcolor}%
\pgftext[x=1.239803in,y=0.351899in,,top]{\color{textcolor}\rmfamily\fontsize{8.000000}{9.600000}\selectfont \(\displaystyle {10}\)}%
\end{pgfscope}%
\begin{pgfscope}%
\definecolor{textcolor}{rgb}{0.150000,0.150000,0.150000}%
\pgfsetstrokecolor{textcolor}%
\pgfsetfillcolor{textcolor}%
\pgftext[x=0.880300in,y=0.198219in,,top]{\color{textcolor}\rmfamily\fontsize{10.000000}{12.000000}\selectfont Edit distance \(\displaystyle \epsilon\)}%
\end{pgfscope}%
\begin{pgfscope}%
\pgfpathrectangle{\pgfqpoint{0.520798in}{0.442177in}}{\pgfqpoint{0.719005in}{0.871884in}}%
\pgfusepath{clip}%
\pgfsetroundcap%
\pgfsetroundjoin%
\pgfsetlinewidth{0.501875pt}%
\definecolor{currentstroke}{rgb}{0.800000,0.800000,0.800000}%
\pgfsetstrokecolor{currentstroke}%
\pgfsetdash{}{0pt}%
\pgfpathmoveto{\pgfqpoint{0.520798in}{0.442177in}}%
\pgfpathlineto{\pgfqpoint{1.239803in}{0.442177in}}%
\pgfusepath{stroke}%
\end{pgfscope}%
\begin{pgfscope}%
\definecolor{textcolor}{rgb}{0.150000,0.150000,0.150000}%
\pgfsetstrokecolor{textcolor}%
\pgfsetfillcolor{textcolor}%
\pgftext[x=0.273151in, y=0.403915in, left, base]{\color{textcolor}\rmfamily\fontsize{8.000000}{9.600000}\selectfont 0\%}%
\end{pgfscope}%
\begin{pgfscope}%
\pgfpathrectangle{\pgfqpoint{0.520798in}{0.442177in}}{\pgfqpoint{0.719005in}{0.871884in}}%
\pgfusepath{clip}%
\pgfsetroundcap%
\pgfsetroundjoin%
\pgfsetlinewidth{0.501875pt}%
\definecolor{currentstroke}{rgb}{0.800000,0.800000,0.800000}%
\pgfsetstrokecolor{currentstroke}%
\pgfsetdash{}{0pt}%
\pgfpathmoveto{\pgfqpoint{0.520798in}{0.660148in}}%
\pgfpathlineto{\pgfqpoint{1.239803in}{0.660148in}}%
\pgfusepath{stroke}%
\end{pgfscope}%
\begin{pgfscope}%
\definecolor{textcolor}{rgb}{0.150000,0.150000,0.150000}%
\pgfsetstrokecolor{textcolor}%
\pgfsetfillcolor{textcolor}%
\pgftext[x=0.214138in, y=0.621886in, left, base]{\color{textcolor}\rmfamily\fontsize{8.000000}{9.600000}\selectfont 25\%}%
\end{pgfscope}%
\begin{pgfscope}%
\pgfpathrectangle{\pgfqpoint{0.520798in}{0.442177in}}{\pgfqpoint{0.719005in}{0.871884in}}%
\pgfusepath{clip}%
\pgfsetroundcap%
\pgfsetroundjoin%
\pgfsetlinewidth{0.501875pt}%
\definecolor{currentstroke}{rgb}{0.800000,0.800000,0.800000}%
\pgfsetstrokecolor{currentstroke}%
\pgfsetdash{}{0pt}%
\pgfpathmoveto{\pgfqpoint{0.520798in}{0.878119in}}%
\pgfpathlineto{\pgfqpoint{1.239803in}{0.878119in}}%
\pgfusepath{stroke}%
\end{pgfscope}%
\begin{pgfscope}%
\definecolor{textcolor}{rgb}{0.150000,0.150000,0.150000}%
\pgfsetstrokecolor{textcolor}%
\pgfsetfillcolor{textcolor}%
\pgftext[x=0.214138in, y=0.839857in, left, base]{\color{textcolor}\rmfamily\fontsize{8.000000}{9.600000}\selectfont 50\%}%
\end{pgfscope}%
\begin{pgfscope}%
\pgfpathrectangle{\pgfqpoint{0.520798in}{0.442177in}}{\pgfqpoint{0.719005in}{0.871884in}}%
\pgfusepath{clip}%
\pgfsetroundcap%
\pgfsetroundjoin%
\pgfsetlinewidth{0.501875pt}%
\definecolor{currentstroke}{rgb}{0.800000,0.800000,0.800000}%
\pgfsetstrokecolor{currentstroke}%
\pgfsetdash{}{0pt}%
\pgfpathmoveto{\pgfqpoint{0.520798in}{1.096090in}}%
\pgfpathlineto{\pgfqpoint{1.239803in}{1.096090in}}%
\pgfusepath{stroke}%
\end{pgfscope}%
\begin{pgfscope}%
\definecolor{textcolor}{rgb}{0.150000,0.150000,0.150000}%
\pgfsetstrokecolor{textcolor}%
\pgfsetfillcolor{textcolor}%
\pgftext[x=0.214138in, y=1.057828in, left, base]{\color{textcolor}\rmfamily\fontsize{8.000000}{9.600000}\selectfont 75\%}%
\end{pgfscope}%
\begin{pgfscope}%
\pgfpathrectangle{\pgfqpoint{0.520798in}{0.442177in}}{\pgfqpoint{0.719005in}{0.871884in}}%
\pgfusepath{clip}%
\pgfsetroundcap%
\pgfsetroundjoin%
\pgfsetlinewidth{0.501875pt}%
\definecolor{currentstroke}{rgb}{0.800000,0.800000,0.800000}%
\pgfsetstrokecolor{currentstroke}%
\pgfsetdash{}{0pt}%
\pgfpathmoveto{\pgfqpoint{0.520798in}{1.314061in}}%
\pgfpathlineto{\pgfqpoint{1.239803in}{1.314061in}}%
\pgfusepath{stroke}%
\end{pgfscope}%
\begin{pgfscope}%
\definecolor{textcolor}{rgb}{0.150000,0.150000,0.150000}%
\pgfsetstrokecolor{textcolor}%
\pgfsetfillcolor{textcolor}%
\pgftext[x=0.155124in, y=1.275799in, left, base]{\color{textcolor}\rmfamily\fontsize{8.000000}{9.600000}\selectfont 100\%}%
\end{pgfscope}%
\begin{pgfscope}%
\definecolor{textcolor}{rgb}{0.150000,0.150000,0.150000}%
\pgfsetstrokecolor{textcolor}%
\pgfsetfillcolor{textcolor}%
\pgftext[x=0.099569in,y=0.878119in,,bottom,rotate=90.000000]{\color{textcolor}\rmfamily\fontsize{10.000000}{12.000000}\selectfont Cert. Acc.}%
\end{pgfscope}%
\begin{pgfscope}%
\pgfsetrectcap%
\pgfsetmiterjoin%
\pgfsetlinewidth{0.752812pt}%
\definecolor{currentstroke}{rgb}{0.700000,0.700000,0.700000}%
\pgfsetstrokecolor{currentstroke}%
\pgfsetdash{}{0pt}%
\pgfpathmoveto{\pgfqpoint{0.520798in}{0.442177in}}%
\pgfpathlineto{\pgfqpoint{0.520798in}{1.314061in}}%
\pgfusepath{stroke}%
\end{pgfscope}%
\begin{pgfscope}%
\pgfsetrectcap%
\pgfsetmiterjoin%
\pgfsetlinewidth{0.752812pt}%
\definecolor{currentstroke}{rgb}{0.700000,0.700000,0.700000}%
\pgfsetstrokecolor{currentstroke}%
\pgfsetdash{}{0pt}%
\pgfpathmoveto{\pgfqpoint{1.239803in}{0.442177in}}%
\pgfpathlineto{\pgfqpoint{1.239803in}{1.314061in}}%
\pgfusepath{stroke}%
\end{pgfscope}%
\begin{pgfscope}%
\pgfsetrectcap%
\pgfsetmiterjoin%
\pgfsetlinewidth{0.752812pt}%
\definecolor{currentstroke}{rgb}{0.700000,0.700000,0.700000}%
\pgfsetstrokecolor{currentstroke}%
\pgfsetdash{}{0pt}%
\pgfpathmoveto{\pgfqpoint{0.520798in}{0.442177in}}%
\pgfpathlineto{\pgfqpoint{1.239803in}{0.442177in}}%
\pgfusepath{stroke}%
\end{pgfscope}%
\begin{pgfscope}%
\pgfsetrectcap%
\pgfsetmiterjoin%
\pgfsetlinewidth{0.752812pt}%
\definecolor{currentstroke}{rgb}{0.700000,0.700000,0.700000}%
\pgfsetstrokecolor{currentstroke}%
\pgfsetdash{}{0pt}%
\pgfpathmoveto{\pgfqpoint{0.520798in}{1.314061in}}%
\pgfpathlineto{\pgfqpoint{1.239803in}{1.314061in}}%
\pgfusepath{stroke}%
\end{pgfscope}%
\begin{pgfscope}%
\pgfpathrectangle{\pgfqpoint{0.520798in}{0.442177in}}{\pgfqpoint{0.719005in}{0.871884in}}%
\pgfusepath{clip}%
\pgfsetroundcap%
\pgfsetroundjoin%
\pgfsetlinewidth{1.003750pt}%
\definecolor{currentstroke}{rgb}{0.003922,0.450980,0.698039}%
\pgfsetstrokecolor{currentstroke}%
\pgfsetdash{}{0pt}%
\pgfpathmoveto{\pgfqpoint{0.520798in}{1.024224in}}%
\pgfpathlineto{\pgfqpoint{0.556748in}{1.024224in}}%
\pgfpathlineto{\pgfqpoint{0.556748in}{0.950519in}}%
\pgfpathlineto{\pgfqpoint{0.628649in}{0.950519in}}%
\pgfpathlineto{\pgfqpoint{0.628649in}{0.843520in}}%
\pgfpathlineto{\pgfqpoint{0.700549in}{0.843520in}}%
\pgfpathlineto{\pgfqpoint{0.700549in}{0.596309in}}%
\pgfpathlineto{\pgfqpoint{0.772450in}{0.596309in}}%
\pgfpathlineto{\pgfqpoint{0.772450in}{0.504336in}}%
\pgfpathlineto{\pgfqpoint{0.844350in}{0.504336in}}%
\pgfpathlineto{\pgfqpoint{0.844350in}{0.490733in}}%
\pgfpathlineto{\pgfqpoint{0.916251in}{0.490733in}}%
\pgfpathlineto{\pgfqpoint{0.916251in}{0.484486in}}%
\pgfpathlineto{\pgfqpoint{0.988151in}{0.484486in}}%
\pgfpathlineto{\pgfqpoint{0.988151in}{0.474838in}}%
\pgfpathlineto{\pgfqpoint{1.060052in}{0.474838in}}%
\pgfpathlineto{\pgfqpoint{1.060052in}{0.449848in}}%
\pgfpathlineto{\pgfqpoint{1.131952in}{0.449848in}}%
\pgfpathlineto{\pgfqpoint{1.131952in}{0.447396in}}%
\pgfpathlineto{\pgfqpoint{1.203853in}{0.447396in}}%
\pgfpathlineto{\pgfqpoint{1.203853in}{0.444391in}}%
\pgfpathlineto{\pgfqpoint{1.241470in}{0.444391in}}%
\pgfusepath{stroke}%
\end{pgfscope}%
\begin{pgfscope}%
\pgfpathrectangle{\pgfqpoint{0.520798in}{0.442177in}}{\pgfqpoint{0.719005in}{0.871884in}}%
\pgfusepath{clip}%
\pgfsetbuttcap%
\pgfsetroundjoin%
\definecolor{currentfill}{rgb}{0.003922,0.450980,0.698039}%
\pgfsetfillcolor{currentfill}%
\pgfsetfillopacity{0.500000}%
\pgfsetlinewidth{0.000000pt}%
\definecolor{currentstroke}{rgb}{0.003922,0.450980,0.698039}%
\pgfsetstrokecolor{currentstroke}%
\pgfsetstrokeopacity{0.500000}%
\pgfsetdash{}{0pt}%
\pgfpathmoveto{\pgfqpoint{0.520798in}{1.033671in}}%
\pgfpathlineto{\pgfqpoint{0.520798in}{1.014777in}}%
\pgfpathlineto{\pgfqpoint{0.556748in}{1.014777in}}%
\pgfpathlineto{\pgfqpoint{0.556748in}{0.940742in}}%
\pgfpathlineto{\pgfqpoint{0.628649in}{0.940742in}}%
\pgfpathlineto{\pgfqpoint{0.628649in}{0.838275in}}%
\pgfpathlineto{\pgfqpoint{0.700549in}{0.838275in}}%
\pgfpathlineto{\pgfqpoint{0.700549in}{0.594011in}}%
\pgfpathlineto{\pgfqpoint{0.772450in}{0.594011in}}%
\pgfpathlineto{\pgfqpoint{0.772450in}{0.500531in}}%
\pgfpathlineto{\pgfqpoint{0.844350in}{0.500531in}}%
\pgfpathlineto{\pgfqpoint{0.844350in}{0.488576in}}%
\pgfpathlineto{\pgfqpoint{0.916251in}{0.488576in}}%
\pgfpathlineto{\pgfqpoint{0.916251in}{0.482048in}}%
\pgfpathlineto{\pgfqpoint{0.988151in}{0.482048in}}%
\pgfpathlineto{\pgfqpoint{0.988151in}{0.472137in}}%
\pgfpathlineto{\pgfqpoint{1.060052in}{0.472137in}}%
\pgfpathlineto{\pgfqpoint{1.060052in}{0.446738in}}%
\pgfpathlineto{\pgfqpoint{1.131952in}{0.446738in}}%
\pgfpathlineto{\pgfqpoint{1.131952in}{0.445569in}}%
\pgfpathlineto{\pgfqpoint{1.203853in}{0.445569in}}%
\pgfpathlineto{\pgfqpoint{1.203853in}{0.443616in}}%
\pgfpathlineto{\pgfqpoint{1.275753in}{0.443616in}}%
\pgfpathlineto{\pgfqpoint{1.275753in}{0.443199in}}%
\pgfpathlineto{\pgfqpoint{1.347654in}{0.443199in}}%
\pgfpathlineto{\pgfqpoint{1.347654in}{0.443009in}}%
\pgfpathlineto{\pgfqpoint{1.419554in}{0.443009in}}%
\pgfpathlineto{\pgfqpoint{1.419554in}{0.442177in}}%
\pgfpathlineto{\pgfqpoint{2.066659in}{0.442177in}}%
\pgfpathlineto{\pgfqpoint{2.066659in}{0.442177in}}%
\pgfpathlineto{\pgfqpoint{2.677813in}{0.442177in}}%
\pgfpathlineto{\pgfqpoint{2.677813in}{0.442177in}}%
\pgfpathlineto{\pgfqpoint{2.677813in}{0.442177in}}%
\pgfpathlineto{\pgfqpoint{2.066659in}{0.442177in}}%
\pgfpathlineto{\pgfqpoint{2.066659in}{0.442177in}}%
\pgfpathlineto{\pgfqpoint{1.419554in}{0.442177in}}%
\pgfpathlineto{\pgfqpoint{1.419554in}{0.443717in}}%
\pgfpathlineto{\pgfqpoint{1.347654in}{0.443717in}}%
\pgfpathlineto{\pgfqpoint{1.347654in}{0.444318in}}%
\pgfpathlineto{\pgfqpoint{1.275753in}{0.444318in}}%
\pgfpathlineto{\pgfqpoint{1.275753in}{0.445166in}}%
\pgfpathlineto{\pgfqpoint{1.203853in}{0.445166in}}%
\pgfpathlineto{\pgfqpoint{1.203853in}{0.449224in}}%
\pgfpathlineto{\pgfqpoint{1.131952in}{0.449224in}}%
\pgfpathlineto{\pgfqpoint{1.131952in}{0.452957in}}%
\pgfpathlineto{\pgfqpoint{1.060052in}{0.452957in}}%
\pgfpathlineto{\pgfqpoint{1.060052in}{0.477538in}}%
\pgfpathlineto{\pgfqpoint{0.988151in}{0.477538in}}%
\pgfpathlineto{\pgfqpoint{0.988151in}{0.486923in}}%
\pgfpathlineto{\pgfqpoint{0.916251in}{0.486923in}}%
\pgfpathlineto{\pgfqpoint{0.916251in}{0.492891in}}%
\pgfpathlineto{\pgfqpoint{0.844350in}{0.492891in}}%
\pgfpathlineto{\pgfqpoint{0.844350in}{0.508140in}}%
\pgfpathlineto{\pgfqpoint{0.772450in}{0.508140in}}%
\pgfpathlineto{\pgfqpoint{0.772450in}{0.598606in}}%
\pgfpathlineto{\pgfqpoint{0.700549in}{0.598606in}}%
\pgfpathlineto{\pgfqpoint{0.700549in}{0.848766in}}%
\pgfpathlineto{\pgfqpoint{0.628649in}{0.848766in}}%
\pgfpathlineto{\pgfqpoint{0.628649in}{0.960297in}}%
\pgfpathlineto{\pgfqpoint{0.556748in}{0.960297in}}%
\pgfpathlineto{\pgfqpoint{0.556748in}{1.033671in}}%
\pgfpathlineto{\pgfqpoint{0.520798in}{1.033671in}}%
\pgfpathlineto{\pgfqpoint{0.520798in}{1.033671in}}%
\pgfpathclose%
\pgfusepath{fill}%
\end{pgfscope}%
\begin{pgfscope}%
\pgfpathrectangle{\pgfqpoint{0.520798in}{0.442177in}}{\pgfqpoint{0.719005in}{0.871884in}}%
\pgfusepath{clip}%
\pgfsetroundcap%
\pgfsetroundjoin%
\pgfsetlinewidth{1.003750pt}%
\definecolor{currentstroke}{rgb}{0.870588,0.560784,0.019608}%
\pgfsetstrokecolor{currentstroke}%
\pgfsetdash{}{0pt}%
\pgfpathmoveto{\pgfqpoint{0.520798in}{1.024224in}}%
\pgfpathlineto{\pgfqpoint{0.556748in}{1.024224in}}%
\pgfpathlineto{\pgfqpoint{0.556748in}{0.965703in}}%
\pgfpathlineto{\pgfqpoint{0.628649in}{0.965703in}}%
\pgfpathlineto{\pgfqpoint{0.628649in}{0.916830in}}%
\pgfpathlineto{\pgfqpoint{0.700549in}{0.916830in}}%
\pgfpathlineto{\pgfqpoint{0.700549in}{0.872860in}}%
\pgfpathlineto{\pgfqpoint{0.772450in}{0.872860in}}%
\pgfpathlineto{\pgfqpoint{0.772450in}{0.831974in}}%
\pgfpathlineto{\pgfqpoint{0.844350in}{0.831974in}}%
\pgfpathlineto{\pgfqpoint{0.844350in}{0.793936in}}%
\pgfpathlineto{\pgfqpoint{0.916251in}{0.793936in}}%
\pgfpathlineto{\pgfqpoint{0.916251in}{0.596309in}}%
\pgfpathlineto{\pgfqpoint{0.988151in}{0.596309in}}%
\pgfpathlineto{\pgfqpoint{0.988151in}{0.582785in}}%
\pgfpathlineto{\pgfqpoint{1.060052in}{0.582785in}}%
\pgfpathlineto{\pgfqpoint{1.060052in}{0.504336in}}%
\pgfpathlineto{\pgfqpoint{1.131952in}{0.504336in}}%
\pgfpathlineto{\pgfqpoint{1.131952in}{0.497851in}}%
\pgfpathlineto{\pgfqpoint{1.203853in}{0.497851in}}%
\pgfpathlineto{\pgfqpoint{1.203853in}{0.488519in}}%
\pgfpathlineto{\pgfqpoint{1.241470in}{0.488519in}}%
\pgfusepath{stroke}%
\end{pgfscope}%
\begin{pgfscope}%
\pgfpathrectangle{\pgfqpoint{0.520798in}{0.442177in}}{\pgfqpoint{0.719005in}{0.871884in}}%
\pgfusepath{clip}%
\pgfsetbuttcap%
\pgfsetroundjoin%
\definecolor{currentfill}{rgb}{0.870588,0.560784,0.019608}%
\pgfsetfillcolor{currentfill}%
\pgfsetfillopacity{0.500000}%
\pgfsetlinewidth{0.000000pt}%
\definecolor{currentstroke}{rgb}{0.870588,0.560784,0.019608}%
\pgfsetstrokecolor{currentstroke}%
\pgfsetstrokeopacity{0.500000}%
\pgfsetdash{}{0pt}%
\pgfpathmoveto{\pgfqpoint{0.520798in}{1.033671in}}%
\pgfpathlineto{\pgfqpoint{0.520798in}{1.014777in}}%
\pgfpathlineto{\pgfqpoint{0.556748in}{1.014777in}}%
\pgfpathlineto{\pgfqpoint{0.556748in}{0.957481in}}%
\pgfpathlineto{\pgfqpoint{0.628649in}{0.957481in}}%
\pgfpathlineto{\pgfqpoint{0.628649in}{0.908215in}}%
\pgfpathlineto{\pgfqpoint{0.700549in}{0.908215in}}%
\pgfpathlineto{\pgfqpoint{0.700549in}{0.864962in}}%
\pgfpathlineto{\pgfqpoint{0.772450in}{0.864962in}}%
\pgfpathlineto{\pgfqpoint{0.772450in}{0.827442in}}%
\pgfpathlineto{\pgfqpoint{0.844350in}{0.827442in}}%
\pgfpathlineto{\pgfqpoint{0.844350in}{0.791378in}}%
\pgfpathlineto{\pgfqpoint{0.916251in}{0.791378in}}%
\pgfpathlineto{\pgfqpoint{0.916251in}{0.594011in}}%
\pgfpathlineto{\pgfqpoint{0.988151in}{0.594011in}}%
\pgfpathlineto{\pgfqpoint{0.988151in}{0.579558in}}%
\pgfpathlineto{\pgfqpoint{1.060052in}{0.579558in}}%
\pgfpathlineto{\pgfqpoint{1.060052in}{0.500531in}}%
\pgfpathlineto{\pgfqpoint{1.131952in}{0.500531in}}%
\pgfpathlineto{\pgfqpoint{1.131952in}{0.494427in}}%
\pgfpathlineto{\pgfqpoint{1.203853in}{0.494427in}}%
\pgfpathlineto{\pgfqpoint{1.203853in}{0.486466in}}%
\pgfpathlineto{\pgfqpoint{1.275753in}{0.486466in}}%
\pgfpathlineto{\pgfqpoint{1.275753in}{0.482658in}}%
\pgfpathlineto{\pgfqpoint{1.347654in}{0.482658in}}%
\pgfpathlineto{\pgfqpoint{1.347654in}{0.475557in}}%
\pgfpathlineto{\pgfqpoint{1.419554in}{0.475557in}}%
\pgfpathlineto{\pgfqpoint{1.419554in}{0.472160in}}%
\pgfpathlineto{\pgfqpoint{1.491455in}{0.472160in}}%
\pgfpathlineto{\pgfqpoint{1.491455in}{0.467234in}}%
\pgfpathlineto{\pgfqpoint{1.563355in}{0.467234in}}%
\pgfpathlineto{\pgfqpoint{1.563355in}{0.463255in}}%
\pgfpathlineto{\pgfqpoint{1.635256in}{0.463255in}}%
\pgfpathlineto{\pgfqpoint{1.635256in}{0.446806in}}%
\pgfpathlineto{\pgfqpoint{1.707156in}{0.446806in}}%
\pgfpathlineto{\pgfqpoint{1.707156in}{0.446150in}}%
\pgfpathlineto{\pgfqpoint{1.779057in}{0.446150in}}%
\pgfpathlineto{\pgfqpoint{1.779057in}{0.445353in}}%
\pgfpathlineto{\pgfqpoint{1.850957in}{0.445353in}}%
\pgfpathlineto{\pgfqpoint{1.850957in}{0.444262in}}%
\pgfpathlineto{\pgfqpoint{1.922858in}{0.444262in}}%
\pgfpathlineto{\pgfqpoint{1.922858in}{0.443537in}}%
\pgfpathlineto{\pgfqpoint{1.994758in}{0.443537in}}%
\pgfpathlineto{\pgfqpoint{1.994758in}{0.443047in}}%
\pgfpathlineto{\pgfqpoint{2.066659in}{0.443047in}}%
\pgfpathlineto{\pgfqpoint{2.066659in}{0.442932in}}%
\pgfpathlineto{\pgfqpoint{2.174510in}{0.442932in}}%
\pgfpathlineto{\pgfqpoint{2.174510in}{0.442537in}}%
\pgfpathlineto{\pgfqpoint{2.282360in}{0.442537in}}%
\pgfpathlineto{\pgfqpoint{2.282360in}{0.442177in}}%
\pgfpathlineto{\pgfqpoint{2.498062in}{0.442177in}}%
\pgfpathlineto{\pgfqpoint{2.498062in}{0.442177in}}%
\pgfpathlineto{\pgfqpoint{2.677813in}{0.442177in}}%
\pgfpathlineto{\pgfqpoint{2.677813in}{0.442177in}}%
\pgfpathlineto{\pgfqpoint{2.677813in}{0.442177in}}%
\pgfpathlineto{\pgfqpoint{2.498062in}{0.442177in}}%
\pgfpathlineto{\pgfqpoint{2.498062in}{0.442177in}}%
\pgfpathlineto{\pgfqpoint{2.282360in}{0.442177in}}%
\pgfpathlineto{\pgfqpoint{2.282360in}{0.442924in}}%
\pgfpathlineto{\pgfqpoint{2.174510in}{0.442924in}}%
\pgfpathlineto{\pgfqpoint{2.174510in}{0.443320in}}%
\pgfpathlineto{\pgfqpoint{2.066659in}{0.443320in}}%
\pgfpathlineto{\pgfqpoint{2.066659in}{0.443996in}}%
\pgfpathlineto{\pgfqpoint{1.994758in}{0.443996in}}%
\pgfpathlineto{\pgfqpoint{1.994758in}{0.445087in}}%
\pgfpathlineto{\pgfqpoint{1.922858in}{0.445087in}}%
\pgfpathlineto{\pgfqpoint{1.922858in}{0.446576in}}%
\pgfpathlineto{\pgfqpoint{1.850957in}{0.446576in}}%
\pgfpathlineto{\pgfqpoint{1.850957in}{0.448648in}}%
\pgfpathlineto{\pgfqpoint{1.779057in}{0.448648in}}%
\pgfpathlineto{\pgfqpoint{1.779057in}{0.452122in}}%
\pgfpathlineto{\pgfqpoint{1.707156in}{0.452122in}}%
\pgfpathlineto{\pgfqpoint{1.707156in}{0.452732in}}%
\pgfpathlineto{\pgfqpoint{1.635256in}{0.452732in}}%
\pgfpathlineto{\pgfqpoint{1.635256in}{0.468548in}}%
\pgfpathlineto{\pgfqpoint{1.563355in}{0.468548in}}%
\pgfpathlineto{\pgfqpoint{1.563355in}{0.472794in}}%
\pgfpathlineto{\pgfqpoint{1.491455in}{0.472794in}}%
\pgfpathlineto{\pgfqpoint{1.491455in}{0.476566in}}%
\pgfpathlineto{\pgfqpoint{1.419554in}{0.476566in}}%
\pgfpathlineto{\pgfqpoint{1.419554in}{0.481711in}}%
\pgfpathlineto{\pgfqpoint{1.347654in}{0.481711in}}%
\pgfpathlineto{\pgfqpoint{1.347654in}{0.485998in}}%
\pgfpathlineto{\pgfqpoint{1.275753in}{0.485998in}}%
\pgfpathlineto{\pgfqpoint{1.275753in}{0.490572in}}%
\pgfpathlineto{\pgfqpoint{1.203853in}{0.490572in}}%
\pgfpathlineto{\pgfqpoint{1.203853in}{0.501274in}}%
\pgfpathlineto{\pgfqpoint{1.131952in}{0.501274in}}%
\pgfpathlineto{\pgfqpoint{1.131952in}{0.508140in}}%
\pgfpathlineto{\pgfqpoint{1.060052in}{0.508140in}}%
\pgfpathlineto{\pgfqpoint{1.060052in}{0.586013in}}%
\pgfpathlineto{\pgfqpoint{0.988151in}{0.586013in}}%
\pgfpathlineto{\pgfqpoint{0.988151in}{0.598606in}}%
\pgfpathlineto{\pgfqpoint{0.916251in}{0.598606in}}%
\pgfpathlineto{\pgfqpoint{0.916251in}{0.796493in}}%
\pgfpathlineto{\pgfqpoint{0.844350in}{0.796493in}}%
\pgfpathlineto{\pgfqpoint{0.844350in}{0.836506in}}%
\pgfpathlineto{\pgfqpoint{0.772450in}{0.836506in}}%
\pgfpathlineto{\pgfqpoint{0.772450in}{0.880758in}}%
\pgfpathlineto{\pgfqpoint{0.700549in}{0.880758in}}%
\pgfpathlineto{\pgfqpoint{0.700549in}{0.925444in}}%
\pgfpathlineto{\pgfqpoint{0.628649in}{0.925444in}}%
\pgfpathlineto{\pgfqpoint{0.628649in}{0.973925in}}%
\pgfpathlineto{\pgfqpoint{0.556748in}{0.973925in}}%
\pgfpathlineto{\pgfqpoint{0.556748in}{1.033671in}}%
\pgfpathlineto{\pgfqpoint{0.520798in}{1.033671in}}%
\pgfpathlineto{\pgfqpoint{0.520798in}{1.033671in}}%
\pgfpathclose%
\pgfusepath{fill}%
\end{pgfscope}%
\begin{pgfscope}%
\pgfpathrectangle{\pgfqpoint{0.520798in}{0.442177in}}{\pgfqpoint{0.719005in}{0.871884in}}%
\pgfusepath{clip}%
\pgfsetroundcap%
\pgfsetroundjoin%
\pgfsetlinewidth{1.003750pt}%
\definecolor{currentstroke}{rgb}{0.007843,0.619608,0.450980}%
\pgfsetstrokecolor{currentstroke}%
\pgfsetdash{}{0pt}%
\pgfpathmoveto{\pgfqpoint{0.520798in}{1.024224in}}%
\pgfpathlineto{\pgfqpoint{0.556748in}{1.024224in}}%
\pgfpathlineto{\pgfqpoint{0.556748in}{0.965703in}}%
\pgfpathlineto{\pgfqpoint{0.628649in}{0.965703in}}%
\pgfpathlineto{\pgfqpoint{0.628649in}{0.916830in}}%
\pgfpathlineto{\pgfqpoint{0.700549in}{0.916830in}}%
\pgfpathlineto{\pgfqpoint{0.700549in}{0.872860in}}%
\pgfpathlineto{\pgfqpoint{0.772450in}{0.872860in}}%
\pgfpathlineto{\pgfqpoint{0.772450in}{0.831974in}}%
\pgfpathlineto{\pgfqpoint{0.844350in}{0.831974in}}%
\pgfpathlineto{\pgfqpoint{0.844350in}{0.793936in}}%
\pgfpathlineto{\pgfqpoint{0.916251in}{0.793936in}}%
\pgfpathlineto{\pgfqpoint{0.916251in}{0.761275in}}%
\pgfpathlineto{\pgfqpoint{0.988151in}{0.761275in}}%
\pgfpathlineto{\pgfqpoint{0.988151in}{0.732647in}}%
\pgfpathlineto{\pgfqpoint{1.060052in}{0.732647in}}%
\pgfpathlineto{\pgfqpoint{1.060052in}{0.704177in}}%
\pgfpathlineto{\pgfqpoint{1.131952in}{0.704177in}}%
\pgfpathlineto{\pgfqpoint{1.131952in}{0.677605in}}%
\pgfpathlineto{\pgfqpoint{1.203853in}{0.677605in}}%
\pgfpathlineto{\pgfqpoint{1.203853in}{0.656016in}}%
\pgfpathlineto{\pgfqpoint{1.241470in}{0.656016in}}%
\pgfusepath{stroke}%
\end{pgfscope}%
\begin{pgfscope}%
\pgfpathrectangle{\pgfqpoint{0.520798in}{0.442177in}}{\pgfqpoint{0.719005in}{0.871884in}}%
\pgfusepath{clip}%
\pgfsetbuttcap%
\pgfsetroundjoin%
\definecolor{currentfill}{rgb}{0.007843,0.619608,0.450980}%
\pgfsetfillcolor{currentfill}%
\pgfsetfillopacity{0.500000}%
\pgfsetlinewidth{0.000000pt}%
\definecolor{currentstroke}{rgb}{0.007843,0.619608,0.450980}%
\pgfsetstrokecolor{currentstroke}%
\pgfsetstrokeopacity{0.500000}%
\pgfsetdash{}{0pt}%
\pgfpathmoveto{\pgfqpoint{0.520798in}{1.033671in}}%
\pgfpathlineto{\pgfqpoint{0.520798in}{1.014777in}}%
\pgfpathlineto{\pgfqpoint{0.556748in}{1.014777in}}%
\pgfpathlineto{\pgfqpoint{0.556748in}{0.957481in}}%
\pgfpathlineto{\pgfqpoint{0.628649in}{0.957481in}}%
\pgfpathlineto{\pgfqpoint{0.628649in}{0.908215in}}%
\pgfpathlineto{\pgfqpoint{0.700549in}{0.908215in}}%
\pgfpathlineto{\pgfqpoint{0.700549in}{0.864962in}}%
\pgfpathlineto{\pgfqpoint{0.772450in}{0.864962in}}%
\pgfpathlineto{\pgfqpoint{0.772450in}{0.827442in}}%
\pgfpathlineto{\pgfqpoint{0.844350in}{0.827442in}}%
\pgfpathlineto{\pgfqpoint{0.844350in}{0.791378in}}%
\pgfpathlineto{\pgfqpoint{0.916251in}{0.791378in}}%
\pgfpathlineto{\pgfqpoint{0.916251in}{0.758688in}}%
\pgfpathlineto{\pgfqpoint{0.988151in}{0.758688in}}%
\pgfpathlineto{\pgfqpoint{0.988151in}{0.727641in}}%
\pgfpathlineto{\pgfqpoint{1.060052in}{0.727641in}}%
\pgfpathlineto{\pgfqpoint{1.060052in}{0.700171in}}%
\pgfpathlineto{\pgfqpoint{1.131952in}{0.700171in}}%
\pgfpathlineto{\pgfqpoint{1.131952in}{0.675417in}}%
\pgfpathlineto{\pgfqpoint{1.203853in}{0.675417in}}%
\pgfpathlineto{\pgfqpoint{1.203853in}{0.652742in}}%
\pgfpathlineto{\pgfqpoint{1.275753in}{0.652742in}}%
\pgfpathlineto{\pgfqpoint{1.275753in}{0.631331in}}%
\pgfpathlineto{\pgfqpoint{1.347654in}{0.631331in}}%
\pgfpathlineto{\pgfqpoint{1.347654in}{0.594011in}}%
\pgfpathlineto{\pgfqpoint{1.419554in}{0.594011in}}%
\pgfpathlineto{\pgfqpoint{1.419554in}{0.579558in}}%
\pgfpathlineto{\pgfqpoint{1.491455in}{0.579558in}}%
\pgfpathlineto{\pgfqpoint{1.491455in}{0.566487in}}%
\pgfpathlineto{\pgfqpoint{1.563355in}{0.566487in}}%
\pgfpathlineto{\pgfqpoint{1.563355in}{0.555238in}}%
\pgfpathlineto{\pgfqpoint{1.635256in}{0.555238in}}%
\pgfpathlineto{\pgfqpoint{1.635256in}{0.500531in}}%
\pgfpathlineto{\pgfqpoint{1.707156in}{0.500531in}}%
\pgfpathlineto{\pgfqpoint{1.707156in}{0.494427in}}%
\pgfpathlineto{\pgfqpoint{1.779057in}{0.494427in}}%
\pgfpathlineto{\pgfqpoint{1.779057in}{0.489042in}}%
\pgfpathlineto{\pgfqpoint{1.850957in}{0.489042in}}%
\pgfpathlineto{\pgfqpoint{1.850957in}{0.485733in}}%
\pgfpathlineto{\pgfqpoint{1.922858in}{0.485733in}}%
\pgfpathlineto{\pgfqpoint{1.922858in}{0.480869in}}%
\pgfpathlineto{\pgfqpoint{1.994758in}{0.480869in}}%
\pgfpathlineto{\pgfqpoint{1.994758in}{0.477186in}}%
\pgfpathlineto{\pgfqpoint{2.066659in}{0.477186in}}%
\pgfpathlineto{\pgfqpoint{2.066659in}{0.472675in}}%
\pgfpathlineto{\pgfqpoint{2.138559in}{0.472675in}}%
\pgfpathlineto{\pgfqpoint{2.138559in}{0.468374in}}%
\pgfpathlineto{\pgfqpoint{2.210460in}{0.468374in}}%
\pgfpathlineto{\pgfqpoint{2.210460in}{0.465164in}}%
\pgfpathlineto{\pgfqpoint{2.282360in}{0.465164in}}%
\pgfpathlineto{\pgfqpoint{2.282360in}{0.462231in}}%
\pgfpathlineto{\pgfqpoint{2.354261in}{0.462231in}}%
\pgfpathlineto{\pgfqpoint{2.354261in}{0.459796in}}%
\pgfpathlineto{\pgfqpoint{2.426161in}{0.459796in}}%
\pgfpathlineto{\pgfqpoint{2.426161in}{0.457722in}}%
\pgfpathlineto{\pgfqpoint{2.498062in}{0.457722in}}%
\pgfpathlineto{\pgfqpoint{2.498062in}{0.456344in}}%
\pgfpathlineto{\pgfqpoint{2.569962in}{0.456344in}}%
\pgfpathlineto{\pgfqpoint{2.569962in}{0.453521in}}%
\pgfpathlineto{\pgfqpoint{2.641863in}{0.453521in}}%
\pgfpathlineto{\pgfqpoint{2.641863in}{0.452521in}}%
\pgfpathlineto{\pgfqpoint{2.677813in}{0.452521in}}%
\pgfpathlineto{\pgfqpoint{2.677813in}{0.456664in}}%
\pgfpathlineto{\pgfqpoint{2.677813in}{0.456664in}}%
\pgfpathlineto{\pgfqpoint{2.641863in}{0.456664in}}%
\pgfpathlineto{\pgfqpoint{2.641863in}{0.459618in}}%
\pgfpathlineto{\pgfqpoint{2.569962in}{0.459618in}}%
\pgfpathlineto{\pgfqpoint{2.569962in}{0.460750in}}%
\pgfpathlineto{\pgfqpoint{2.498062in}{0.460750in}}%
\pgfpathlineto{\pgfqpoint{2.498062in}{0.462535in}}%
\pgfpathlineto{\pgfqpoint{2.426161in}{0.462535in}}%
\pgfpathlineto{\pgfqpoint{2.426161in}{0.463625in}}%
\pgfpathlineto{\pgfqpoint{2.354261in}{0.463625in}}%
\pgfpathlineto{\pgfqpoint{2.354261in}{0.466092in}}%
\pgfpathlineto{\pgfqpoint{2.282360in}{0.466092in}}%
\pgfpathlineto{\pgfqpoint{2.282360in}{0.469802in}}%
\pgfpathlineto{\pgfqpoint{2.210460in}{0.469802in}}%
\pgfpathlineto{\pgfqpoint{2.210460in}{0.473077in}}%
\pgfpathlineto{\pgfqpoint{2.138559in}{0.473077in}}%
\pgfpathlineto{\pgfqpoint{2.138559in}{0.477159in}}%
\pgfpathlineto{\pgfqpoint{2.066659in}{0.477159in}}%
\pgfpathlineto{\pgfqpoint{2.066659in}{0.480556in}}%
\pgfpathlineto{\pgfqpoint{1.994758in}{0.480556in}}%
\pgfpathlineto{\pgfqpoint{1.994758in}{0.485414in}}%
\pgfpathlineto{\pgfqpoint{1.922858in}{0.485414in}}%
\pgfpathlineto{\pgfqpoint{1.922858in}{0.488775in}}%
\pgfpathlineto{\pgfqpoint{1.850957in}{0.488775in}}%
\pgfpathlineto{\pgfqpoint{1.850957in}{0.495114in}}%
\pgfpathlineto{\pgfqpoint{1.779057in}{0.495114in}}%
\pgfpathlineto{\pgfqpoint{1.779057in}{0.501274in}}%
\pgfpathlineto{\pgfqpoint{1.707156in}{0.501274in}}%
\pgfpathlineto{\pgfqpoint{1.707156in}{0.508140in}}%
\pgfpathlineto{\pgfqpoint{1.635256in}{0.508140in}}%
\pgfpathlineto{\pgfqpoint{1.635256in}{0.564149in}}%
\pgfpathlineto{\pgfqpoint{1.563355in}{0.564149in}}%
\pgfpathlineto{\pgfqpoint{1.563355in}{0.574727in}}%
\pgfpathlineto{\pgfqpoint{1.491455in}{0.574727in}}%
\pgfpathlineto{\pgfqpoint{1.491455in}{0.586013in}}%
\pgfpathlineto{\pgfqpoint{1.419554in}{0.586013in}}%
\pgfpathlineto{\pgfqpoint{1.419554in}{0.598606in}}%
\pgfpathlineto{\pgfqpoint{1.347654in}{0.598606in}}%
\pgfpathlineto{\pgfqpoint{1.347654in}{0.639420in}}%
\pgfpathlineto{\pgfqpoint{1.275753in}{0.639420in}}%
\pgfpathlineto{\pgfqpoint{1.275753in}{0.659290in}}%
\pgfpathlineto{\pgfqpoint{1.203853in}{0.659290in}}%
\pgfpathlineto{\pgfqpoint{1.203853in}{0.679794in}}%
\pgfpathlineto{\pgfqpoint{1.131952in}{0.679794in}}%
\pgfpathlineto{\pgfqpoint{1.131952in}{0.708183in}}%
\pgfpathlineto{\pgfqpoint{1.060052in}{0.708183in}}%
\pgfpathlineto{\pgfqpoint{1.060052in}{0.737652in}}%
\pgfpathlineto{\pgfqpoint{0.988151in}{0.737652in}}%
\pgfpathlineto{\pgfqpoint{0.988151in}{0.763862in}}%
\pgfpathlineto{\pgfqpoint{0.916251in}{0.763862in}}%
\pgfpathlineto{\pgfqpoint{0.916251in}{0.796493in}}%
\pgfpathlineto{\pgfqpoint{0.844350in}{0.796493in}}%
\pgfpathlineto{\pgfqpoint{0.844350in}{0.836506in}}%
\pgfpathlineto{\pgfqpoint{0.772450in}{0.836506in}}%
\pgfpathlineto{\pgfqpoint{0.772450in}{0.880758in}}%
\pgfpathlineto{\pgfqpoint{0.700549in}{0.880758in}}%
\pgfpathlineto{\pgfqpoint{0.700549in}{0.925444in}}%
\pgfpathlineto{\pgfqpoint{0.628649in}{0.925444in}}%
\pgfpathlineto{\pgfqpoint{0.628649in}{0.973925in}}%
\pgfpathlineto{\pgfqpoint{0.556748in}{0.973925in}}%
\pgfpathlineto{\pgfqpoint{0.556748in}{1.033671in}}%
\pgfpathlineto{\pgfqpoint{0.520798in}{1.033671in}}%
\pgfpathlineto{\pgfqpoint{0.520798in}{1.033671in}}%
\pgfpathclose%
\pgfusepath{fill}%
\end{pgfscope}%
\begin{pgfscope}%
\definecolor{textcolor}{rgb}{0.150000,0.150000,0.150000}%
\pgfsetstrokecolor{textcolor}%
\pgfsetfillcolor{textcolor}%
\pgftext[x=0.880300in,y=1.397395in,,base]{\color{textcolor}\rmfamily\fontsize{9.000000}{10.800000}\selectfont \(\displaystyle c_A^-=1\)}%
\end{pgfscope}%
\begin{pgfscope}%
\pgfsetbuttcap%
\pgfsetmiterjoin%
\definecolor{currentfill}{rgb}{1.000000,1.000000,1.000000}%
\pgfsetfillcolor{currentfill}%
\pgfsetfillopacity{0.800000}%
\pgfsetlinewidth{1.003750pt}%
\definecolor{currentstroke}{rgb}{0.800000,0.800000,0.800000}%
\pgfsetstrokecolor{currentstroke}%
\pgfsetstrokeopacity{0.800000}%
\pgfsetdash{}{0pt}%
\pgfpathmoveto{\pgfqpoint{0.785843in}{0.638103in}}%
\pgfpathlineto{\pgfqpoint{1.191192in}{0.638103in}}%
\pgfpathlineto{\pgfqpoint{1.191192in}{1.265450in}}%
\pgfpathlineto{\pgfqpoint{0.785843in}{1.265450in}}%
\pgfpathlineto{\pgfqpoint{0.785843in}{0.638103in}}%
\pgfpathclose%
\pgfusepath{stroke,fill}%
\end{pgfscope}%
\begin{pgfscope}%
\definecolor{textcolor}{rgb}{0.150000,0.150000,0.150000}%
\pgfsetstrokecolor{textcolor}%
\pgfsetfillcolor{textcolor}%
\pgftext[x=0.918271in,y=1.113275in,left,base]{\color{textcolor}\rmfamily\fontsize{9.000000}{10.800000}\selectfont \(\displaystyle c_A^+\)}%
\end{pgfscope}%
\begin{pgfscope}%
\pgfsetroundcap%
\pgfsetroundjoin%
\pgfsetlinewidth{1.003750pt}%
\definecolor{currentstroke}{rgb}{0.003922,0.450980,0.698039}%
\pgfsetstrokecolor{currentstroke}%
\pgfsetdash{}{0pt}%
\pgfpathmoveto{\pgfqpoint{0.824732in}{1.001052in}}%
\pgfpathlineto{\pgfqpoint{0.873343in}{1.001052in}}%
\pgfpathlineto{\pgfqpoint{0.873343in}{1.001052in}}%
\pgfpathlineto{\pgfqpoint{0.970565in}{1.001052in}}%
\pgfpathlineto{\pgfqpoint{0.970565in}{1.001052in}}%
\pgfpathlineto{\pgfqpoint{1.019176in}{1.001052in}}%
\pgfusepath{stroke}%
\end{pgfscope}%
\begin{pgfscope}%
\definecolor{textcolor}{rgb}{0.150000,0.150000,0.150000}%
\pgfsetstrokecolor{textcolor}%
\pgfsetfillcolor{textcolor}%
\pgftext[x=1.096954in,y=0.967024in,left,base]{\color{textcolor}\rmfamily\fontsize{7.000000}{8.400000}\selectfont 1}%
\end{pgfscope}%
\begin{pgfscope}%
\pgfsetroundcap%
\pgfsetroundjoin%
\pgfsetlinewidth{1.003750pt}%
\definecolor{currentstroke}{rgb}{0.870588,0.560784,0.019608}%
\pgfsetstrokecolor{currentstroke}%
\pgfsetdash{}{0pt}%
\pgfpathmoveto{\pgfqpoint{0.824732in}{0.865486in}}%
\pgfpathlineto{\pgfqpoint{0.873343in}{0.865486in}}%
\pgfpathlineto{\pgfqpoint{0.873343in}{0.865486in}}%
\pgfpathlineto{\pgfqpoint{0.970565in}{0.865486in}}%
\pgfpathlineto{\pgfqpoint{0.970565in}{0.865486in}}%
\pgfpathlineto{\pgfqpoint{1.019176in}{0.865486in}}%
\pgfusepath{stroke}%
\end{pgfscope}%
\begin{pgfscope}%
\definecolor{textcolor}{rgb}{0.150000,0.150000,0.150000}%
\pgfsetstrokecolor{textcolor}%
\pgfsetfillcolor{textcolor}%
\pgftext[x=1.096954in,y=0.831458in,left,base]{\color{textcolor}\rmfamily\fontsize{7.000000}{8.400000}\selectfont 2}%
\end{pgfscope}%
\begin{pgfscope}%
\pgfsetroundcap%
\pgfsetroundjoin%
\pgfsetlinewidth{1.003750pt}%
\definecolor{currentstroke}{rgb}{0.007843,0.619608,0.450980}%
\pgfsetstrokecolor{currentstroke}%
\pgfsetdash{}{0pt}%
\pgfpathmoveto{\pgfqpoint{0.824732in}{0.729919in}}%
\pgfpathlineto{\pgfqpoint{0.873343in}{0.729919in}}%
\pgfpathlineto{\pgfqpoint{0.873343in}{0.729919in}}%
\pgfpathlineto{\pgfqpoint{0.970565in}{0.729919in}}%
\pgfpathlineto{\pgfqpoint{0.970565in}{0.729919in}}%
\pgfpathlineto{\pgfqpoint{1.019176in}{0.729919in}}%
\pgfusepath{stroke}%
\end{pgfscope}%
\begin{pgfscope}%
\definecolor{textcolor}{rgb}{0.150000,0.150000,0.150000}%
\pgfsetstrokecolor{textcolor}%
\pgfsetfillcolor{textcolor}%
\pgftext[x=1.096954in,y=0.695892in,left,base]{\color{textcolor}\rmfamily\fontsize{7.000000}{8.400000}\selectfont 4}%
\end{pgfscope}%
\end{pgfpicture}%
\makeatother%
\endgroup%

%% file: figures/experiments/graphs/sparse_smoothing/nodes_structure/node_classification-Cora-APPNP-hidden=32-p_adj_plus=0.001-p_adj_minus=0.8-p_att_plus=0.0-p_att_minus=0.0-multi_class_cert-A.pgf
\begingroup%
\makeatletter%
\begin{pgfpicture}%
\pgfpathrectangle{\pgfpointorigin}{\pgfqpoint{1.375000in}{1.581250in}}%
\pgfusepath{use as bounding box, clip}%
\begin{pgfscope}%
\pgfsetbuttcap%
\pgfsetmiterjoin%
\definecolor{currentfill}{rgb}{1.000000,1.000000,1.000000}%
\pgfsetfillcolor{currentfill}%
\pgfsetlinewidth{0.000000pt}%
\definecolor{currentstroke}{rgb}{1.000000,1.000000,1.000000}%
\pgfsetstrokecolor{currentstroke}%
\pgfsetdash{}{0pt}%
\pgfpathmoveto{\pgfqpoint{0.000000in}{0.000000in}}%
\pgfpathlineto{\pgfqpoint{1.375000in}{0.000000in}}%
\pgfpathlineto{\pgfqpoint{1.375000in}{1.581250in}}%
\pgfpathlineto{\pgfqpoint{0.000000in}{1.581250in}}%
\pgfpathlineto{\pgfqpoint{0.000000in}{0.000000in}}%
\pgfpathclose%
\pgfusepath{fill}%
\end{pgfscope}%
\begin{pgfscope}%
\pgfsetbuttcap%
\pgfsetmiterjoin%
\definecolor{currentfill}{rgb}{1.000000,1.000000,1.000000}%
\pgfsetfillcolor{currentfill}%
\pgfsetlinewidth{0.000000pt}%
\definecolor{currentstroke}{rgb}{0.000000,0.000000,0.000000}%
\pgfsetstrokecolor{currentstroke}%
\pgfsetstrokeopacity{0.000000}%
\pgfsetdash{}{0pt}%
\pgfpathmoveto{\pgfqpoint{0.520798in}{0.442177in}}%
\pgfpathlineto{\pgfqpoint{1.239803in}{0.442177in}}%
\pgfpathlineto{\pgfqpoint{1.239803in}{1.314061in}}%
\pgfpathlineto{\pgfqpoint{0.520798in}{1.314061in}}%
\pgfpathlineto{\pgfqpoint{0.520798in}{0.442177in}}%
\pgfpathclose%
\pgfusepath{fill}%
\end{pgfscope}%
\begin{pgfscope}%
\pgfpathrectangle{\pgfqpoint{0.520798in}{0.442177in}}{\pgfqpoint{0.719005in}{0.871884in}}%
\pgfusepath{clip}%
\pgfsetroundcap%
\pgfsetroundjoin%
\pgfsetlinewidth{0.501875pt}%
\definecolor{currentstroke}{rgb}{0.800000,0.800000,0.800000}%
\pgfsetstrokecolor{currentstroke}%
\pgfsetdash{}{0pt}%
\pgfpathmoveto{\pgfqpoint{0.520798in}{0.442177in}}%
\pgfpathlineto{\pgfqpoint{0.520798in}{1.314061in}}%
\pgfusepath{stroke}%
\end{pgfscope}%
\begin{pgfscope}%
\definecolor{textcolor}{rgb}{0.150000,0.150000,0.150000}%
\pgfsetstrokecolor{textcolor}%
\pgfsetfillcolor{textcolor}%
\pgftext[x=0.520798in,y=0.351899in,,top]{\color{textcolor}\rmfamily\fontsize{8.000000}{9.600000}\selectfont \(\displaystyle {0}\)}%
\end{pgfscope}%
\begin{pgfscope}%
\pgfpathrectangle{\pgfqpoint{0.520798in}{0.442177in}}{\pgfqpoint{0.719005in}{0.871884in}}%
\pgfusepath{clip}%
\pgfsetroundcap%
\pgfsetroundjoin%
\pgfsetlinewidth{0.501875pt}%
\definecolor{currentstroke}{rgb}{0.800000,0.800000,0.800000}%
\pgfsetstrokecolor{currentstroke}%
\pgfsetdash{}{0pt}%
\pgfpathmoveto{\pgfqpoint{0.880300in}{0.442177in}}%
\pgfpathlineto{\pgfqpoint{0.880300in}{1.314061in}}%
\pgfusepath{stroke}%
\end{pgfscope}%
\begin{pgfscope}%
\definecolor{textcolor}{rgb}{0.150000,0.150000,0.150000}%
\pgfsetstrokecolor{textcolor}%
\pgfsetfillcolor{textcolor}%
\pgftext[x=0.880300in,y=0.351899in,,top]{\color{textcolor}\rmfamily\fontsize{8.000000}{9.600000}\selectfont \(\displaystyle {5}\)}%
\end{pgfscope}%
\begin{pgfscope}%
\pgfpathrectangle{\pgfqpoint{0.520798in}{0.442177in}}{\pgfqpoint{0.719005in}{0.871884in}}%
\pgfusepath{clip}%
\pgfsetroundcap%
\pgfsetroundjoin%
\pgfsetlinewidth{0.501875pt}%
\definecolor{currentstroke}{rgb}{0.800000,0.800000,0.800000}%
\pgfsetstrokecolor{currentstroke}%
\pgfsetdash{}{0pt}%
\pgfpathmoveto{\pgfqpoint{1.239803in}{0.442177in}}%
\pgfpathlineto{\pgfqpoint{1.239803in}{1.314061in}}%
\pgfusepath{stroke}%
\end{pgfscope}%
\begin{pgfscope}%
\definecolor{textcolor}{rgb}{0.150000,0.150000,0.150000}%
\pgfsetstrokecolor{textcolor}%
\pgfsetfillcolor{textcolor}%
\pgftext[x=1.239803in,y=0.351899in,,top]{\color{textcolor}\rmfamily\fontsize{8.000000}{9.600000}\selectfont \(\displaystyle {10}\)}%
\end{pgfscope}%
\begin{pgfscope}%
\definecolor{textcolor}{rgb}{0.150000,0.150000,0.150000}%
\pgfsetstrokecolor{textcolor}%
\pgfsetfillcolor{textcolor}%
\pgftext[x=0.880300in,y=0.198219in,,top]{\color{textcolor}\rmfamily\fontsize{10.000000}{12.000000}\selectfont Edit distance \(\displaystyle \epsilon\)}%
\end{pgfscope}%
\begin{pgfscope}%
\pgfpathrectangle{\pgfqpoint{0.520798in}{0.442177in}}{\pgfqpoint{0.719005in}{0.871884in}}%
\pgfusepath{clip}%
\pgfsetroundcap%
\pgfsetroundjoin%
\pgfsetlinewidth{0.501875pt}%
\definecolor{currentstroke}{rgb}{0.800000,0.800000,0.800000}%
\pgfsetstrokecolor{currentstroke}%
\pgfsetdash{}{0pt}%
\pgfpathmoveto{\pgfqpoint{0.520798in}{0.442177in}}%
\pgfpathlineto{\pgfqpoint{1.239803in}{0.442177in}}%
\pgfusepath{stroke}%
\end{pgfscope}%
\begin{pgfscope}%
\definecolor{textcolor}{rgb}{0.150000,0.150000,0.150000}%
\pgfsetstrokecolor{textcolor}%
\pgfsetfillcolor{textcolor}%
\pgftext[x=0.273151in, y=0.403915in, left, base]{\color{textcolor}\rmfamily\fontsize{8.000000}{9.600000}\selectfont 0\%}%
\end{pgfscope}%
\begin{pgfscope}%
\pgfpathrectangle{\pgfqpoint{0.520798in}{0.442177in}}{\pgfqpoint{0.719005in}{0.871884in}}%
\pgfusepath{clip}%
\pgfsetroundcap%
\pgfsetroundjoin%
\pgfsetlinewidth{0.501875pt}%
\definecolor{currentstroke}{rgb}{0.800000,0.800000,0.800000}%
\pgfsetstrokecolor{currentstroke}%
\pgfsetdash{}{0pt}%
\pgfpathmoveto{\pgfqpoint{0.520798in}{0.660148in}}%
\pgfpathlineto{\pgfqpoint{1.239803in}{0.660148in}}%
\pgfusepath{stroke}%
\end{pgfscope}%
\begin{pgfscope}%
\definecolor{textcolor}{rgb}{0.150000,0.150000,0.150000}%
\pgfsetstrokecolor{textcolor}%
\pgfsetfillcolor{textcolor}%
\pgftext[x=0.214138in, y=0.621886in, left, base]{\color{textcolor}\rmfamily\fontsize{8.000000}{9.600000}\selectfont 25\%}%
\end{pgfscope}%
\begin{pgfscope}%
\pgfpathrectangle{\pgfqpoint{0.520798in}{0.442177in}}{\pgfqpoint{0.719005in}{0.871884in}}%
\pgfusepath{clip}%
\pgfsetroundcap%
\pgfsetroundjoin%
\pgfsetlinewidth{0.501875pt}%
\definecolor{currentstroke}{rgb}{0.800000,0.800000,0.800000}%
\pgfsetstrokecolor{currentstroke}%
\pgfsetdash{}{0pt}%
\pgfpathmoveto{\pgfqpoint{0.520798in}{0.878119in}}%
\pgfpathlineto{\pgfqpoint{1.239803in}{0.878119in}}%
\pgfusepath{stroke}%
\end{pgfscope}%
\begin{pgfscope}%
\definecolor{textcolor}{rgb}{0.150000,0.150000,0.150000}%
\pgfsetstrokecolor{textcolor}%
\pgfsetfillcolor{textcolor}%
\pgftext[x=0.214138in, y=0.839857in, left, base]{\color{textcolor}\rmfamily\fontsize{8.000000}{9.600000}\selectfont 50\%}%
\end{pgfscope}%
\begin{pgfscope}%
\pgfpathrectangle{\pgfqpoint{0.520798in}{0.442177in}}{\pgfqpoint{0.719005in}{0.871884in}}%
\pgfusepath{clip}%
\pgfsetroundcap%
\pgfsetroundjoin%
\pgfsetlinewidth{0.501875pt}%
\definecolor{currentstroke}{rgb}{0.800000,0.800000,0.800000}%
\pgfsetstrokecolor{currentstroke}%
\pgfsetdash{}{0pt}%
\pgfpathmoveto{\pgfqpoint{0.520798in}{1.096090in}}%
\pgfpathlineto{\pgfqpoint{1.239803in}{1.096090in}}%
\pgfusepath{stroke}%
\end{pgfscope}%
\begin{pgfscope}%
\definecolor{textcolor}{rgb}{0.150000,0.150000,0.150000}%
\pgfsetstrokecolor{textcolor}%
\pgfsetfillcolor{textcolor}%
\pgftext[x=0.214138in, y=1.057828in, left, base]{\color{textcolor}\rmfamily\fontsize{8.000000}{9.600000}\selectfont 75\%}%
\end{pgfscope}%
\begin{pgfscope}%
\pgfpathrectangle{\pgfqpoint{0.520798in}{0.442177in}}{\pgfqpoint{0.719005in}{0.871884in}}%
\pgfusepath{clip}%
\pgfsetroundcap%
\pgfsetroundjoin%
\pgfsetlinewidth{0.501875pt}%
\definecolor{currentstroke}{rgb}{0.800000,0.800000,0.800000}%
\pgfsetstrokecolor{currentstroke}%
\pgfsetdash{}{0pt}%
\pgfpathmoveto{\pgfqpoint{0.520798in}{1.314061in}}%
\pgfpathlineto{\pgfqpoint{1.239803in}{1.314061in}}%
\pgfusepath{stroke}%
\end{pgfscope}%
\begin{pgfscope}%
\definecolor{textcolor}{rgb}{0.150000,0.150000,0.150000}%
\pgfsetstrokecolor{textcolor}%
\pgfsetfillcolor{textcolor}%
\pgftext[x=0.155124in, y=1.275799in, left, base]{\color{textcolor}\rmfamily\fontsize{8.000000}{9.600000}\selectfont 100\%}%
\end{pgfscope}%
\begin{pgfscope}%
\definecolor{textcolor}{rgb}{0.150000,0.150000,0.150000}%
\pgfsetstrokecolor{textcolor}%
\pgfsetfillcolor{textcolor}%
\pgftext[x=0.099569in,y=0.878119in,,bottom,rotate=90.000000]{\color{textcolor}\rmfamily\fontsize{10.000000}{12.000000}\selectfont Cert. Acc.}%
\end{pgfscope}%
\begin{pgfscope}%
\pgfsetrectcap%
\pgfsetmiterjoin%
\pgfsetlinewidth{0.752812pt}%
\definecolor{currentstroke}{rgb}{0.700000,0.700000,0.700000}%
\pgfsetstrokecolor{currentstroke}%
\pgfsetdash{}{0pt}%
\pgfpathmoveto{\pgfqpoint{0.520798in}{0.442177in}}%
\pgfpathlineto{\pgfqpoint{0.520798in}{1.314061in}}%
\pgfusepath{stroke}%
\end{pgfscope}%
\begin{pgfscope}%
\pgfsetrectcap%
\pgfsetmiterjoin%
\pgfsetlinewidth{0.752812pt}%
\definecolor{currentstroke}{rgb}{0.700000,0.700000,0.700000}%
\pgfsetstrokecolor{currentstroke}%
\pgfsetdash{}{0pt}%
\pgfpathmoveto{\pgfqpoint{1.239803in}{0.442177in}}%
\pgfpathlineto{\pgfqpoint{1.239803in}{1.314061in}}%
\pgfusepath{stroke}%
\end{pgfscope}%
\begin{pgfscope}%
\pgfsetrectcap%
\pgfsetmiterjoin%
\pgfsetlinewidth{0.752812pt}%
\definecolor{currentstroke}{rgb}{0.700000,0.700000,0.700000}%
\pgfsetstrokecolor{currentstroke}%
\pgfsetdash{}{0pt}%
\pgfpathmoveto{\pgfqpoint{0.520798in}{0.442177in}}%
\pgfpathlineto{\pgfqpoint{1.239803in}{0.442177in}}%
\pgfusepath{stroke}%
\end{pgfscope}%
\begin{pgfscope}%
\pgfsetrectcap%
\pgfsetmiterjoin%
\pgfsetlinewidth{0.752812pt}%
\definecolor{currentstroke}{rgb}{0.700000,0.700000,0.700000}%
\pgfsetstrokecolor{currentstroke}%
\pgfsetdash{}{0pt}%
\pgfpathmoveto{\pgfqpoint{0.520798in}{1.314061in}}%
\pgfpathlineto{\pgfqpoint{1.239803in}{1.314061in}}%
\pgfusepath{stroke}%
\end{pgfscope}%
\begin{pgfscope}%
\pgfpathrectangle{\pgfqpoint{0.520798in}{0.442177in}}{\pgfqpoint{0.719005in}{0.871884in}}%
\pgfusepath{clip}%
\pgfsetroundcap%
\pgfsetroundjoin%
\pgfsetlinewidth{1.003750pt}%
\definecolor{currentstroke}{rgb}{0.003922,0.450980,0.698039}%
\pgfsetstrokecolor{currentstroke}%
\pgfsetdash{}{0pt}%
\pgfpathmoveto{\pgfqpoint{0.520798in}{1.024224in}}%
\pgfpathlineto{\pgfqpoint{0.556748in}{1.024224in}}%
\pgfpathlineto{\pgfqpoint{0.556748in}{0.950519in}}%
\pgfpathlineto{\pgfqpoint{0.628649in}{0.950519in}}%
\pgfpathlineto{\pgfqpoint{0.628649in}{0.843520in}}%
\pgfpathlineto{\pgfqpoint{0.700549in}{0.843520in}}%
\pgfpathlineto{\pgfqpoint{0.700549in}{0.596309in}}%
\pgfpathlineto{\pgfqpoint{0.772450in}{0.596309in}}%
\pgfpathlineto{\pgfqpoint{0.772450in}{0.504336in}}%
\pgfpathlineto{\pgfqpoint{0.844350in}{0.504336in}}%
\pgfpathlineto{\pgfqpoint{0.844350in}{0.490733in}}%
\pgfpathlineto{\pgfqpoint{0.916251in}{0.490733in}}%
\pgfpathlineto{\pgfqpoint{0.916251in}{0.484486in}}%
\pgfpathlineto{\pgfqpoint{0.988151in}{0.484486in}}%
\pgfpathlineto{\pgfqpoint{0.988151in}{0.474838in}}%
\pgfpathlineto{\pgfqpoint{1.060052in}{0.474838in}}%
\pgfpathlineto{\pgfqpoint{1.060052in}{0.449848in}}%
\pgfpathlineto{\pgfqpoint{1.131952in}{0.449848in}}%
\pgfpathlineto{\pgfqpoint{1.131952in}{0.447396in}}%
\pgfpathlineto{\pgfqpoint{1.203853in}{0.447396in}}%
\pgfpathlineto{\pgfqpoint{1.203853in}{0.444391in}}%
\pgfpathlineto{\pgfqpoint{1.241470in}{0.444391in}}%
\pgfusepath{stroke}%
\end{pgfscope}%
\begin{pgfscope}%
\pgfpathrectangle{\pgfqpoint{0.520798in}{0.442177in}}{\pgfqpoint{0.719005in}{0.871884in}}%
\pgfusepath{clip}%
\pgfsetbuttcap%
\pgfsetroundjoin%
\definecolor{currentfill}{rgb}{0.003922,0.450980,0.698039}%
\pgfsetfillcolor{currentfill}%
\pgfsetfillopacity{0.500000}%
\pgfsetlinewidth{0.000000pt}%
\definecolor{currentstroke}{rgb}{0.003922,0.450980,0.698039}%
\pgfsetstrokecolor{currentstroke}%
\pgfsetstrokeopacity{0.500000}%
\pgfsetdash{}{0pt}%
\pgfpathmoveto{\pgfqpoint{0.520798in}{1.033671in}}%
\pgfpathlineto{\pgfqpoint{0.520798in}{1.014777in}}%
\pgfpathlineto{\pgfqpoint{0.556748in}{1.014777in}}%
\pgfpathlineto{\pgfqpoint{0.556748in}{0.940742in}}%
\pgfpathlineto{\pgfqpoint{0.628649in}{0.940742in}}%
\pgfpathlineto{\pgfqpoint{0.628649in}{0.838275in}}%
\pgfpathlineto{\pgfqpoint{0.700549in}{0.838275in}}%
\pgfpathlineto{\pgfqpoint{0.700549in}{0.594011in}}%
\pgfpathlineto{\pgfqpoint{0.772450in}{0.594011in}}%
\pgfpathlineto{\pgfqpoint{0.772450in}{0.500531in}}%
\pgfpathlineto{\pgfqpoint{0.844350in}{0.500531in}}%
\pgfpathlineto{\pgfqpoint{0.844350in}{0.488576in}}%
\pgfpathlineto{\pgfqpoint{0.916251in}{0.488576in}}%
\pgfpathlineto{\pgfqpoint{0.916251in}{0.482048in}}%
\pgfpathlineto{\pgfqpoint{0.988151in}{0.482048in}}%
\pgfpathlineto{\pgfqpoint{0.988151in}{0.472137in}}%
\pgfpathlineto{\pgfqpoint{1.060052in}{0.472137in}}%
\pgfpathlineto{\pgfqpoint{1.060052in}{0.446738in}}%
\pgfpathlineto{\pgfqpoint{1.131952in}{0.446738in}}%
\pgfpathlineto{\pgfqpoint{1.131952in}{0.445569in}}%
\pgfpathlineto{\pgfqpoint{1.203853in}{0.445569in}}%
\pgfpathlineto{\pgfqpoint{1.203853in}{0.443616in}}%
\pgfpathlineto{\pgfqpoint{1.275753in}{0.443616in}}%
\pgfpathlineto{\pgfqpoint{1.275753in}{0.443199in}}%
\pgfpathlineto{\pgfqpoint{1.347654in}{0.443199in}}%
\pgfpathlineto{\pgfqpoint{1.347654in}{0.443009in}}%
\pgfpathlineto{\pgfqpoint{1.419554in}{0.443009in}}%
\pgfpathlineto{\pgfqpoint{1.419554in}{0.442177in}}%
\pgfpathlineto{\pgfqpoint{2.066659in}{0.442177in}}%
\pgfpathlineto{\pgfqpoint{2.066659in}{0.442177in}}%
\pgfpathlineto{\pgfqpoint{2.677813in}{0.442177in}}%
\pgfpathlineto{\pgfqpoint{2.677813in}{0.442177in}}%
\pgfpathlineto{\pgfqpoint{2.677813in}{0.442177in}}%
\pgfpathlineto{\pgfqpoint{2.066659in}{0.442177in}}%
\pgfpathlineto{\pgfqpoint{2.066659in}{0.442177in}}%
\pgfpathlineto{\pgfqpoint{1.419554in}{0.442177in}}%
\pgfpathlineto{\pgfqpoint{1.419554in}{0.443717in}}%
\pgfpathlineto{\pgfqpoint{1.347654in}{0.443717in}}%
\pgfpathlineto{\pgfqpoint{1.347654in}{0.444318in}}%
\pgfpathlineto{\pgfqpoint{1.275753in}{0.444318in}}%
\pgfpathlineto{\pgfqpoint{1.275753in}{0.445166in}}%
\pgfpathlineto{\pgfqpoint{1.203853in}{0.445166in}}%
\pgfpathlineto{\pgfqpoint{1.203853in}{0.449224in}}%
\pgfpathlineto{\pgfqpoint{1.131952in}{0.449224in}}%
\pgfpathlineto{\pgfqpoint{1.131952in}{0.452957in}}%
\pgfpathlineto{\pgfqpoint{1.060052in}{0.452957in}}%
\pgfpathlineto{\pgfqpoint{1.060052in}{0.477538in}}%
\pgfpathlineto{\pgfqpoint{0.988151in}{0.477538in}}%
\pgfpathlineto{\pgfqpoint{0.988151in}{0.486923in}}%
\pgfpathlineto{\pgfqpoint{0.916251in}{0.486923in}}%
\pgfpathlineto{\pgfqpoint{0.916251in}{0.492891in}}%
\pgfpathlineto{\pgfqpoint{0.844350in}{0.492891in}}%
\pgfpathlineto{\pgfqpoint{0.844350in}{0.508140in}}%
\pgfpathlineto{\pgfqpoint{0.772450in}{0.508140in}}%
\pgfpathlineto{\pgfqpoint{0.772450in}{0.598606in}}%
\pgfpathlineto{\pgfqpoint{0.700549in}{0.598606in}}%
\pgfpathlineto{\pgfqpoint{0.700549in}{0.848766in}}%
\pgfpathlineto{\pgfqpoint{0.628649in}{0.848766in}}%
\pgfpathlineto{\pgfqpoint{0.628649in}{0.960297in}}%
\pgfpathlineto{\pgfqpoint{0.556748in}{0.960297in}}%
\pgfpathlineto{\pgfqpoint{0.556748in}{1.033671in}}%
\pgfpathlineto{\pgfqpoint{0.520798in}{1.033671in}}%
\pgfpathlineto{\pgfqpoint{0.520798in}{1.033671in}}%
\pgfpathclose%
\pgfusepath{fill}%
\end{pgfscope}%
\begin{pgfscope}%
\pgfpathrectangle{\pgfqpoint{0.520798in}{0.442177in}}{\pgfqpoint{0.719005in}{0.871884in}}%
\pgfusepath{clip}%
\pgfsetroundcap%
\pgfsetroundjoin%
\pgfsetlinewidth{1.003750pt}%
\definecolor{currentstroke}{rgb}{0.870588,0.560784,0.019608}%
\pgfsetstrokecolor{currentstroke}%
\pgfsetdash{}{0pt}%
\pgfpathmoveto{\pgfqpoint{0.520798in}{1.024224in}}%
\pgfpathlineto{\pgfqpoint{0.556748in}{1.024224in}}%
\pgfpathlineto{\pgfqpoint{0.556748in}{0.950519in}}%
\pgfpathlineto{\pgfqpoint{0.628649in}{0.950519in}}%
\pgfpathlineto{\pgfqpoint{0.628649in}{0.843520in}}%
\pgfpathlineto{\pgfqpoint{0.700549in}{0.843520in}}%
\pgfpathlineto{\pgfqpoint{0.700549in}{0.596309in}}%
\pgfpathlineto{\pgfqpoint{0.772450in}{0.596309in}}%
\pgfpathlineto{\pgfqpoint{0.772450in}{0.504336in}}%
\pgfpathlineto{\pgfqpoint{0.844350in}{0.504336in}}%
\pgfpathlineto{\pgfqpoint{0.844350in}{0.491999in}}%
\pgfpathlineto{\pgfqpoint{0.916251in}{0.491999in}}%
\pgfpathlineto{\pgfqpoint{0.916251in}{0.485830in}}%
\pgfpathlineto{\pgfqpoint{0.988151in}{0.485830in}}%
\pgfpathlineto{\pgfqpoint{0.988151in}{0.474838in}}%
\pgfpathlineto{\pgfqpoint{1.060052in}{0.474838in}}%
\pgfpathlineto{\pgfqpoint{1.060052in}{0.449848in}}%
\pgfpathlineto{\pgfqpoint{1.131952in}{0.449848in}}%
\pgfpathlineto{\pgfqpoint{1.131952in}{0.447396in}}%
\pgfpathlineto{\pgfqpoint{1.203853in}{0.447396in}}%
\pgfpathlineto{\pgfqpoint{1.203853in}{0.444707in}}%
\pgfpathlineto{\pgfqpoint{1.241470in}{0.444707in}}%
\pgfusepath{stroke}%
\end{pgfscope}%
\begin{pgfscope}%
\pgfpathrectangle{\pgfqpoint{0.520798in}{0.442177in}}{\pgfqpoint{0.719005in}{0.871884in}}%
\pgfusepath{clip}%
\pgfsetbuttcap%
\pgfsetroundjoin%
\definecolor{currentfill}{rgb}{0.870588,0.560784,0.019608}%
\pgfsetfillcolor{currentfill}%
\pgfsetfillopacity{0.500000}%
\pgfsetlinewidth{0.000000pt}%
\definecolor{currentstroke}{rgb}{0.870588,0.560784,0.019608}%
\pgfsetstrokecolor{currentstroke}%
\pgfsetstrokeopacity{0.500000}%
\pgfsetdash{}{0pt}%
\pgfpathmoveto{\pgfqpoint{0.520798in}{1.033671in}}%
\pgfpathlineto{\pgfqpoint{0.520798in}{1.014777in}}%
\pgfpathlineto{\pgfqpoint{0.556748in}{1.014777in}}%
\pgfpathlineto{\pgfqpoint{0.556748in}{0.940742in}}%
\pgfpathlineto{\pgfqpoint{0.628649in}{0.940742in}}%
\pgfpathlineto{\pgfqpoint{0.628649in}{0.838275in}}%
\pgfpathlineto{\pgfqpoint{0.700549in}{0.838275in}}%
\pgfpathlineto{\pgfqpoint{0.700549in}{0.594011in}}%
\pgfpathlineto{\pgfqpoint{0.772450in}{0.594011in}}%
\pgfpathlineto{\pgfqpoint{0.772450in}{0.500531in}}%
\pgfpathlineto{\pgfqpoint{0.844350in}{0.500531in}}%
\pgfpathlineto{\pgfqpoint{0.844350in}{0.489613in}}%
\pgfpathlineto{\pgfqpoint{0.916251in}{0.489613in}}%
\pgfpathlineto{\pgfqpoint{0.916251in}{0.483450in}}%
\pgfpathlineto{\pgfqpoint{0.988151in}{0.483450in}}%
\pgfpathlineto{\pgfqpoint{0.988151in}{0.472137in}}%
\pgfpathlineto{\pgfqpoint{1.060052in}{0.472137in}}%
\pgfpathlineto{\pgfqpoint{1.060052in}{0.446738in}}%
\pgfpathlineto{\pgfqpoint{1.131952in}{0.446738in}}%
\pgfpathlineto{\pgfqpoint{1.131952in}{0.445569in}}%
\pgfpathlineto{\pgfqpoint{1.203853in}{0.445569in}}%
\pgfpathlineto{\pgfqpoint{1.203853in}{0.444171in}}%
\pgfpathlineto{\pgfqpoint{1.275753in}{0.444171in}}%
\pgfpathlineto{\pgfqpoint{1.275753in}{0.443313in}}%
\pgfpathlineto{\pgfqpoint{1.347654in}{0.443313in}}%
\pgfpathlineto{\pgfqpoint{1.347654in}{0.443009in}}%
\pgfpathlineto{\pgfqpoint{1.419554in}{0.443009in}}%
\pgfpathlineto{\pgfqpoint{1.419554in}{0.442177in}}%
\pgfpathlineto{\pgfqpoint{2.066659in}{0.442177in}}%
\pgfpathlineto{\pgfqpoint{2.066659in}{0.442177in}}%
\pgfpathlineto{\pgfqpoint{2.677813in}{0.442177in}}%
\pgfpathlineto{\pgfqpoint{2.677813in}{0.442177in}}%
\pgfpathlineto{\pgfqpoint{2.677813in}{0.442177in}}%
\pgfpathlineto{\pgfqpoint{2.066659in}{0.442177in}}%
\pgfpathlineto{\pgfqpoint{2.066659in}{0.442177in}}%
\pgfpathlineto{\pgfqpoint{1.419554in}{0.442177in}}%
\pgfpathlineto{\pgfqpoint{1.419554in}{0.443717in}}%
\pgfpathlineto{\pgfqpoint{1.347654in}{0.443717in}}%
\pgfpathlineto{\pgfqpoint{1.347654in}{0.444362in}}%
\pgfpathlineto{\pgfqpoint{1.275753in}{0.444362in}}%
\pgfpathlineto{\pgfqpoint{1.275753in}{0.445244in}}%
\pgfpathlineto{\pgfqpoint{1.203853in}{0.445244in}}%
\pgfpathlineto{\pgfqpoint{1.203853in}{0.449224in}}%
\pgfpathlineto{\pgfqpoint{1.131952in}{0.449224in}}%
\pgfpathlineto{\pgfqpoint{1.131952in}{0.452957in}}%
\pgfpathlineto{\pgfqpoint{1.060052in}{0.452957in}}%
\pgfpathlineto{\pgfqpoint{1.060052in}{0.477538in}}%
\pgfpathlineto{\pgfqpoint{0.988151in}{0.477538in}}%
\pgfpathlineto{\pgfqpoint{0.988151in}{0.488211in}}%
\pgfpathlineto{\pgfqpoint{0.916251in}{0.488211in}}%
\pgfpathlineto{\pgfqpoint{0.916251in}{0.494384in}}%
\pgfpathlineto{\pgfqpoint{0.844350in}{0.494384in}}%
\pgfpathlineto{\pgfqpoint{0.844350in}{0.508140in}}%
\pgfpathlineto{\pgfqpoint{0.772450in}{0.508140in}}%
\pgfpathlineto{\pgfqpoint{0.772450in}{0.598606in}}%
\pgfpathlineto{\pgfqpoint{0.700549in}{0.598606in}}%
\pgfpathlineto{\pgfqpoint{0.700549in}{0.848766in}}%
\pgfpathlineto{\pgfqpoint{0.628649in}{0.848766in}}%
\pgfpathlineto{\pgfqpoint{0.628649in}{0.960297in}}%
\pgfpathlineto{\pgfqpoint{0.556748in}{0.960297in}}%
\pgfpathlineto{\pgfqpoint{0.556748in}{1.033671in}}%
\pgfpathlineto{\pgfqpoint{0.520798in}{1.033671in}}%
\pgfpathlineto{\pgfqpoint{0.520798in}{1.033671in}}%
\pgfpathclose%
\pgfusepath{fill}%
\end{pgfscope}%
\begin{pgfscope}%
\pgfpathrectangle{\pgfqpoint{0.520798in}{0.442177in}}{\pgfqpoint{0.719005in}{0.871884in}}%
\pgfusepath{clip}%
\pgfsetroundcap%
\pgfsetroundjoin%
\pgfsetlinewidth{1.003750pt}%
\definecolor{currentstroke}{rgb}{0.007843,0.619608,0.450980}%
\pgfsetstrokecolor{currentstroke}%
\pgfsetdash{}{0pt}%
\pgfpathmoveto{\pgfqpoint{0.520798in}{1.024224in}}%
\pgfpathlineto{\pgfqpoint{0.556748in}{1.024224in}}%
\pgfpathlineto{\pgfqpoint{0.556748in}{0.950519in}}%
\pgfpathlineto{\pgfqpoint{0.628649in}{0.950519in}}%
\pgfpathlineto{\pgfqpoint{0.628649in}{0.843520in}}%
\pgfpathlineto{\pgfqpoint{0.700549in}{0.843520in}}%
\pgfpathlineto{\pgfqpoint{0.700549in}{0.596309in}}%
\pgfpathlineto{\pgfqpoint{0.772450in}{0.596309in}}%
\pgfpathlineto{\pgfqpoint{0.772450in}{0.504336in}}%
\pgfpathlineto{\pgfqpoint{0.844350in}{0.504336in}}%
\pgfpathlineto{\pgfqpoint{0.844350in}{0.491999in}}%
\pgfpathlineto{\pgfqpoint{0.916251in}{0.491999in}}%
\pgfpathlineto{\pgfqpoint{0.916251in}{0.485830in}}%
\pgfpathlineto{\pgfqpoint{0.988151in}{0.485830in}}%
\pgfpathlineto{\pgfqpoint{0.988151in}{0.474838in}}%
\pgfpathlineto{\pgfqpoint{1.060052in}{0.474838in}}%
\pgfpathlineto{\pgfqpoint{1.060052in}{0.449848in}}%
\pgfpathlineto{\pgfqpoint{1.131952in}{0.449848in}}%
\pgfpathlineto{\pgfqpoint{1.131952in}{0.447396in}}%
\pgfpathlineto{\pgfqpoint{1.203853in}{0.447396in}}%
\pgfpathlineto{\pgfqpoint{1.203853in}{0.444707in}}%
\pgfpathlineto{\pgfqpoint{1.241470in}{0.444707in}}%
\pgfusepath{stroke}%
\end{pgfscope}%
\begin{pgfscope}%
\pgfpathrectangle{\pgfqpoint{0.520798in}{0.442177in}}{\pgfqpoint{0.719005in}{0.871884in}}%
\pgfusepath{clip}%
\pgfsetbuttcap%
\pgfsetroundjoin%
\definecolor{currentfill}{rgb}{0.007843,0.619608,0.450980}%
\pgfsetfillcolor{currentfill}%
\pgfsetfillopacity{0.500000}%
\pgfsetlinewidth{0.000000pt}%
\definecolor{currentstroke}{rgb}{0.007843,0.619608,0.450980}%
\pgfsetstrokecolor{currentstroke}%
\pgfsetstrokeopacity{0.500000}%
\pgfsetdash{}{0pt}%
\pgfpathmoveto{\pgfqpoint{0.520798in}{1.033671in}}%
\pgfpathlineto{\pgfqpoint{0.520798in}{1.014777in}}%
\pgfpathlineto{\pgfqpoint{0.556748in}{1.014777in}}%
\pgfpathlineto{\pgfqpoint{0.556748in}{0.940742in}}%
\pgfpathlineto{\pgfqpoint{0.628649in}{0.940742in}}%
\pgfpathlineto{\pgfqpoint{0.628649in}{0.838275in}}%
\pgfpathlineto{\pgfqpoint{0.700549in}{0.838275in}}%
\pgfpathlineto{\pgfqpoint{0.700549in}{0.594011in}}%
\pgfpathlineto{\pgfqpoint{0.772450in}{0.594011in}}%
\pgfpathlineto{\pgfqpoint{0.772450in}{0.500531in}}%
\pgfpathlineto{\pgfqpoint{0.844350in}{0.500531in}}%
\pgfpathlineto{\pgfqpoint{0.844350in}{0.489613in}}%
\pgfpathlineto{\pgfqpoint{0.916251in}{0.489613in}}%
\pgfpathlineto{\pgfqpoint{0.916251in}{0.483450in}}%
\pgfpathlineto{\pgfqpoint{0.988151in}{0.483450in}}%
\pgfpathlineto{\pgfqpoint{0.988151in}{0.472137in}}%
\pgfpathlineto{\pgfqpoint{1.060052in}{0.472137in}}%
\pgfpathlineto{\pgfqpoint{1.060052in}{0.446738in}}%
\pgfpathlineto{\pgfqpoint{1.131952in}{0.446738in}}%
\pgfpathlineto{\pgfqpoint{1.131952in}{0.445569in}}%
\pgfpathlineto{\pgfqpoint{1.203853in}{0.445569in}}%
\pgfpathlineto{\pgfqpoint{1.203853in}{0.444171in}}%
\pgfpathlineto{\pgfqpoint{1.275753in}{0.444171in}}%
\pgfpathlineto{\pgfqpoint{1.275753in}{0.443313in}}%
\pgfpathlineto{\pgfqpoint{1.347654in}{0.443313in}}%
\pgfpathlineto{\pgfqpoint{1.347654in}{0.443009in}}%
\pgfpathlineto{\pgfqpoint{1.419554in}{0.443009in}}%
\pgfpathlineto{\pgfqpoint{1.419554in}{0.442177in}}%
\pgfpathlineto{\pgfqpoint{2.066659in}{0.442177in}}%
\pgfpathlineto{\pgfqpoint{2.066659in}{0.442177in}}%
\pgfpathlineto{\pgfqpoint{2.677813in}{0.442177in}}%
\pgfpathlineto{\pgfqpoint{2.677813in}{0.442177in}}%
\pgfpathlineto{\pgfqpoint{2.677813in}{0.442177in}}%
\pgfpathlineto{\pgfqpoint{2.066659in}{0.442177in}}%
\pgfpathlineto{\pgfqpoint{2.066659in}{0.442177in}}%
\pgfpathlineto{\pgfqpoint{1.419554in}{0.442177in}}%
\pgfpathlineto{\pgfqpoint{1.419554in}{0.443717in}}%
\pgfpathlineto{\pgfqpoint{1.347654in}{0.443717in}}%
\pgfpathlineto{\pgfqpoint{1.347654in}{0.444362in}}%
\pgfpathlineto{\pgfqpoint{1.275753in}{0.444362in}}%
\pgfpathlineto{\pgfqpoint{1.275753in}{0.445244in}}%
\pgfpathlineto{\pgfqpoint{1.203853in}{0.445244in}}%
\pgfpathlineto{\pgfqpoint{1.203853in}{0.449224in}}%
\pgfpathlineto{\pgfqpoint{1.131952in}{0.449224in}}%
\pgfpathlineto{\pgfqpoint{1.131952in}{0.452957in}}%
\pgfpathlineto{\pgfqpoint{1.060052in}{0.452957in}}%
\pgfpathlineto{\pgfqpoint{1.060052in}{0.477538in}}%
\pgfpathlineto{\pgfqpoint{0.988151in}{0.477538in}}%
\pgfpathlineto{\pgfqpoint{0.988151in}{0.488211in}}%
\pgfpathlineto{\pgfqpoint{0.916251in}{0.488211in}}%
\pgfpathlineto{\pgfqpoint{0.916251in}{0.494384in}}%
\pgfpathlineto{\pgfqpoint{0.844350in}{0.494384in}}%
\pgfpathlineto{\pgfqpoint{0.844350in}{0.508140in}}%
\pgfpathlineto{\pgfqpoint{0.772450in}{0.508140in}}%
\pgfpathlineto{\pgfqpoint{0.772450in}{0.598606in}}%
\pgfpathlineto{\pgfqpoint{0.700549in}{0.598606in}}%
\pgfpathlineto{\pgfqpoint{0.700549in}{0.848766in}}%
\pgfpathlineto{\pgfqpoint{0.628649in}{0.848766in}}%
\pgfpathlineto{\pgfqpoint{0.628649in}{0.960297in}}%
\pgfpathlineto{\pgfqpoint{0.556748in}{0.960297in}}%
\pgfpathlineto{\pgfqpoint{0.556748in}{1.033671in}}%
\pgfpathlineto{\pgfqpoint{0.520798in}{1.033671in}}%
\pgfpathlineto{\pgfqpoint{0.520798in}{1.033671in}}%
\pgfpathclose%
\pgfusepath{fill}%
\end{pgfscope}%
\begin{pgfscope}%
\definecolor{textcolor}{rgb}{0.150000,0.150000,0.150000}%
\pgfsetstrokecolor{textcolor}%
\pgfsetfillcolor{textcolor}%
\pgftext[x=0.880300in,y=1.397395in,,base]{\color{textcolor}\rmfamily\fontsize{9.000000}{10.800000}\selectfont \(\displaystyle c_A^+=1\)}%
\end{pgfscope}%
\begin{pgfscope}%
\pgfsetbuttcap%
\pgfsetmiterjoin%
\definecolor{currentfill}{rgb}{1.000000,1.000000,1.000000}%
\pgfsetfillcolor{currentfill}%
\pgfsetfillopacity{0.800000}%
\pgfsetlinewidth{1.003750pt}%
\definecolor{currentstroke}{rgb}{0.800000,0.800000,0.800000}%
\pgfsetstrokecolor{currentstroke}%
\pgfsetstrokeopacity{0.800000}%
\pgfsetdash{}{0pt}%
\pgfpathmoveto{\pgfqpoint{0.785843in}{0.638103in}}%
\pgfpathlineto{\pgfqpoint{1.191192in}{0.638103in}}%
\pgfpathlineto{\pgfqpoint{1.191192in}{1.265450in}}%
\pgfpathlineto{\pgfqpoint{0.785843in}{1.265450in}}%
\pgfpathlineto{\pgfqpoint{0.785843in}{0.638103in}}%
\pgfpathclose%
\pgfusepath{stroke,fill}%
\end{pgfscope}%
\begin{pgfscope}%
\definecolor{textcolor}{rgb}{0.150000,0.150000,0.150000}%
\pgfsetstrokecolor{textcolor}%
\pgfsetfillcolor{textcolor}%
\pgftext[x=0.917113in,y=1.113275in,left,base]{\color{textcolor}\rmfamily\fontsize{9.000000}{10.800000}\selectfont \(\displaystyle c_A^-\)}%
\end{pgfscope}%
\begin{pgfscope}%
\pgfsetroundcap%
\pgfsetroundjoin%
\pgfsetlinewidth{1.003750pt}%
\definecolor{currentstroke}{rgb}{0.003922,0.450980,0.698039}%
\pgfsetstrokecolor{currentstroke}%
\pgfsetdash{}{0pt}%
\pgfpathmoveto{\pgfqpoint{0.824732in}{1.001052in}}%
\pgfpathlineto{\pgfqpoint{0.873343in}{1.001052in}}%
\pgfpathlineto{\pgfqpoint{0.873343in}{1.001052in}}%
\pgfpathlineto{\pgfqpoint{0.970565in}{1.001052in}}%
\pgfpathlineto{\pgfqpoint{0.970565in}{1.001052in}}%
\pgfpathlineto{\pgfqpoint{1.019176in}{1.001052in}}%
\pgfusepath{stroke}%
\end{pgfscope}%
\begin{pgfscope}%
\definecolor{textcolor}{rgb}{0.150000,0.150000,0.150000}%
\pgfsetstrokecolor{textcolor}%
\pgfsetfillcolor{textcolor}%
\pgftext[x=1.096954in,y=0.967024in,left,base]{\color{textcolor}\rmfamily\fontsize{7.000000}{8.400000}\selectfont 1}%
\end{pgfscope}%
\begin{pgfscope}%
\pgfsetroundcap%
\pgfsetroundjoin%
\pgfsetlinewidth{1.003750pt}%
\definecolor{currentstroke}{rgb}{0.870588,0.560784,0.019608}%
\pgfsetstrokecolor{currentstroke}%
\pgfsetdash{}{0pt}%
\pgfpathmoveto{\pgfqpoint{0.824732in}{0.865486in}}%
\pgfpathlineto{\pgfqpoint{0.873343in}{0.865486in}}%
\pgfpathlineto{\pgfqpoint{0.873343in}{0.865486in}}%
\pgfpathlineto{\pgfqpoint{0.970565in}{0.865486in}}%
\pgfpathlineto{\pgfqpoint{0.970565in}{0.865486in}}%
\pgfpathlineto{\pgfqpoint{1.019176in}{0.865486in}}%
\pgfusepath{stroke}%
\end{pgfscope}%
\begin{pgfscope}%
\definecolor{textcolor}{rgb}{0.150000,0.150000,0.150000}%
\pgfsetstrokecolor{textcolor}%
\pgfsetfillcolor{textcolor}%
\pgftext[x=1.096954in,y=0.831458in,left,base]{\color{textcolor}\rmfamily\fontsize{7.000000}{8.400000}\selectfont 2}%
\end{pgfscope}%
\begin{pgfscope}%
\pgfsetroundcap%
\pgfsetroundjoin%
\pgfsetlinewidth{1.003750pt}%
\definecolor{currentstroke}{rgb}{0.007843,0.619608,0.450980}%
\pgfsetstrokecolor{currentstroke}%
\pgfsetdash{}{0pt}%
\pgfpathmoveto{\pgfqpoint{0.824732in}{0.729919in}}%
\pgfpathlineto{\pgfqpoint{0.873343in}{0.729919in}}%
\pgfpathlineto{\pgfqpoint{0.873343in}{0.729919in}}%
\pgfpathlineto{\pgfqpoint{0.970565in}{0.729919in}}%
\pgfpathlineto{\pgfqpoint{0.970565in}{0.729919in}}%
\pgfpathlineto{\pgfqpoint{1.019176in}{0.729919in}}%
\pgfusepath{stroke}%
\end{pgfscope}%
\begin{pgfscope}%
\definecolor{textcolor}{rgb}{0.150000,0.150000,0.150000}%
\pgfsetstrokecolor{textcolor}%
\pgfsetfillcolor{textcolor}%
\pgftext[x=1.096954in,y=0.695892in,left,base]{\color{textcolor}\rmfamily\fontsize{7.000000}{8.400000}\selectfont 4}%
\end{pgfscope}%
\end{pgfpicture}%
\makeatother%
\endgroup%

%% file: figures/experiments/graphs/sparse_smoothing/nodes_attributes/node_classification-Citeseer-APPNP-hidden=32-p_adj_plus=0.0-p_adj_minus=0.0-p_att_plus=0.001-p_att_minus=0.8-multi_class_cert-B.pgf
\begingroup%
\makeatletter%
\begin{pgfpicture}%
\pgfpathrectangle{\pgfpointorigin}{\pgfqpoint{1.375000in}{1.581250in}}%
\pgfusepath{use as bounding box, clip}%
\begin{pgfscope}%
\pgfsetbuttcap%
\pgfsetmiterjoin%
\definecolor{currentfill}{rgb}{1.000000,1.000000,1.000000}%
\pgfsetfillcolor{currentfill}%
\pgfsetlinewidth{0.000000pt}%
\definecolor{currentstroke}{rgb}{1.000000,1.000000,1.000000}%
\pgfsetstrokecolor{currentstroke}%
\pgfsetdash{}{0pt}%
\pgfpathmoveto{\pgfqpoint{0.000000in}{0.000000in}}%
\pgfpathlineto{\pgfqpoint{1.375000in}{0.000000in}}%
\pgfpathlineto{\pgfqpoint{1.375000in}{1.581250in}}%
\pgfpathlineto{\pgfqpoint{0.000000in}{1.581250in}}%
\pgfpathlineto{\pgfqpoint{0.000000in}{0.000000in}}%
\pgfpathclose%
\pgfusepath{fill}%
\end{pgfscope}%
\begin{pgfscope}%
\pgfsetbuttcap%
\pgfsetmiterjoin%
\definecolor{currentfill}{rgb}{1.000000,1.000000,1.000000}%
\pgfsetfillcolor{currentfill}%
\pgfsetlinewidth{0.000000pt}%
\definecolor{currentstroke}{rgb}{0.000000,0.000000,0.000000}%
\pgfsetstrokecolor{currentstroke}%
\pgfsetstrokeopacity{0.000000}%
\pgfsetdash{}{0pt}%
\pgfpathmoveto{\pgfqpoint{0.520798in}{0.442177in}}%
\pgfpathlineto{\pgfqpoint{1.239803in}{0.442177in}}%
\pgfpathlineto{\pgfqpoint{1.239803in}{1.314061in}}%
\pgfpathlineto{\pgfqpoint{0.520798in}{1.314061in}}%
\pgfpathlineto{\pgfqpoint{0.520798in}{0.442177in}}%
\pgfpathclose%
\pgfusepath{fill}%
\end{pgfscope}%
\begin{pgfscope}%
\pgfpathrectangle{\pgfqpoint{0.520798in}{0.442177in}}{\pgfqpoint{0.719005in}{0.871884in}}%
\pgfusepath{clip}%
\pgfsetroundcap%
\pgfsetroundjoin%
\pgfsetlinewidth{0.501875pt}%
\definecolor{currentstroke}{rgb}{0.800000,0.800000,0.800000}%
\pgfsetstrokecolor{currentstroke}%
\pgfsetdash{}{0pt}%
\pgfpathmoveto{\pgfqpoint{0.520798in}{0.442177in}}%
\pgfpathlineto{\pgfqpoint{0.520798in}{1.314061in}}%
\pgfusepath{stroke}%
\end{pgfscope}%
\begin{pgfscope}%
\definecolor{textcolor}{rgb}{0.150000,0.150000,0.150000}%
\pgfsetstrokecolor{textcolor}%
\pgfsetfillcolor{textcolor}%
\pgftext[x=0.520798in,y=0.351899in,,top]{\color{textcolor}\rmfamily\fontsize{8.000000}{9.600000}\selectfont \(\displaystyle {0}\)}%
\end{pgfscope}%
\begin{pgfscope}%
\pgfpathrectangle{\pgfqpoint{0.520798in}{0.442177in}}{\pgfqpoint{0.719005in}{0.871884in}}%
\pgfusepath{clip}%
\pgfsetroundcap%
\pgfsetroundjoin%
\pgfsetlinewidth{0.501875pt}%
\definecolor{currentstroke}{rgb}{0.800000,0.800000,0.800000}%
\pgfsetstrokecolor{currentstroke}%
\pgfsetdash{}{0pt}%
\pgfpathmoveto{\pgfqpoint{0.880300in}{0.442177in}}%
\pgfpathlineto{\pgfqpoint{0.880300in}{1.314061in}}%
\pgfusepath{stroke}%
\end{pgfscope}%
\begin{pgfscope}%
\definecolor{textcolor}{rgb}{0.150000,0.150000,0.150000}%
\pgfsetstrokecolor{textcolor}%
\pgfsetfillcolor{textcolor}%
\pgftext[x=0.880300in,y=0.351899in,,top]{\color{textcolor}\rmfamily\fontsize{8.000000}{9.600000}\selectfont \(\displaystyle {5}\)}%
\end{pgfscope}%
\begin{pgfscope}%
\pgfpathrectangle{\pgfqpoint{0.520798in}{0.442177in}}{\pgfqpoint{0.719005in}{0.871884in}}%
\pgfusepath{clip}%
\pgfsetroundcap%
\pgfsetroundjoin%
\pgfsetlinewidth{0.501875pt}%
\definecolor{currentstroke}{rgb}{0.800000,0.800000,0.800000}%
\pgfsetstrokecolor{currentstroke}%
\pgfsetdash{}{0pt}%
\pgfpathmoveto{\pgfqpoint{1.239803in}{0.442177in}}%
\pgfpathlineto{\pgfqpoint{1.239803in}{1.314061in}}%
\pgfusepath{stroke}%
\end{pgfscope}%
\begin{pgfscope}%
\definecolor{textcolor}{rgb}{0.150000,0.150000,0.150000}%
\pgfsetstrokecolor{textcolor}%
\pgfsetfillcolor{textcolor}%
\pgftext[x=1.239803in,y=0.351899in,,top]{\color{textcolor}\rmfamily\fontsize{8.000000}{9.600000}\selectfont \(\displaystyle {10}\)}%
\end{pgfscope}%
\begin{pgfscope}%
\definecolor{textcolor}{rgb}{0.150000,0.150000,0.150000}%
\pgfsetstrokecolor{textcolor}%
\pgfsetfillcolor{textcolor}%
\pgftext[x=0.880300in,y=0.198219in,,top]{\color{textcolor}\rmfamily\fontsize{10.000000}{12.000000}\selectfont Edit distance \(\displaystyle \epsilon\)}%
\end{pgfscope}%
\begin{pgfscope}%
\pgfpathrectangle{\pgfqpoint{0.520798in}{0.442177in}}{\pgfqpoint{0.719005in}{0.871884in}}%
\pgfusepath{clip}%
\pgfsetroundcap%
\pgfsetroundjoin%
\pgfsetlinewidth{0.501875pt}%
\definecolor{currentstroke}{rgb}{0.800000,0.800000,0.800000}%
\pgfsetstrokecolor{currentstroke}%
\pgfsetdash{}{0pt}%
\pgfpathmoveto{\pgfqpoint{0.520798in}{0.442177in}}%
\pgfpathlineto{\pgfqpoint{1.239803in}{0.442177in}}%
\pgfusepath{stroke}%
\end{pgfscope}%
\begin{pgfscope}%
\definecolor{textcolor}{rgb}{0.150000,0.150000,0.150000}%
\pgfsetstrokecolor{textcolor}%
\pgfsetfillcolor{textcolor}%
\pgftext[x=0.273151in, y=0.403915in, left, base]{\color{textcolor}\rmfamily\fontsize{8.000000}{9.600000}\selectfont 0\%}%
\end{pgfscope}%
\begin{pgfscope}%
\pgfpathrectangle{\pgfqpoint{0.520798in}{0.442177in}}{\pgfqpoint{0.719005in}{0.871884in}}%
\pgfusepath{clip}%
\pgfsetroundcap%
\pgfsetroundjoin%
\pgfsetlinewidth{0.501875pt}%
\definecolor{currentstroke}{rgb}{0.800000,0.800000,0.800000}%
\pgfsetstrokecolor{currentstroke}%
\pgfsetdash{}{0pt}%
\pgfpathmoveto{\pgfqpoint{0.520798in}{0.660148in}}%
\pgfpathlineto{\pgfqpoint{1.239803in}{0.660148in}}%
\pgfusepath{stroke}%
\end{pgfscope}%
\begin{pgfscope}%
\definecolor{textcolor}{rgb}{0.150000,0.150000,0.150000}%
\pgfsetstrokecolor{textcolor}%
\pgfsetfillcolor{textcolor}%
\pgftext[x=0.214138in, y=0.621886in, left, base]{\color{textcolor}\rmfamily\fontsize{8.000000}{9.600000}\selectfont 25\%}%
\end{pgfscope}%
\begin{pgfscope}%
\pgfpathrectangle{\pgfqpoint{0.520798in}{0.442177in}}{\pgfqpoint{0.719005in}{0.871884in}}%
\pgfusepath{clip}%
\pgfsetroundcap%
\pgfsetroundjoin%
\pgfsetlinewidth{0.501875pt}%
\definecolor{currentstroke}{rgb}{0.800000,0.800000,0.800000}%
\pgfsetstrokecolor{currentstroke}%
\pgfsetdash{}{0pt}%
\pgfpathmoveto{\pgfqpoint{0.520798in}{0.878119in}}%
\pgfpathlineto{\pgfqpoint{1.239803in}{0.878119in}}%
\pgfusepath{stroke}%
\end{pgfscope}%
\begin{pgfscope}%
\definecolor{textcolor}{rgb}{0.150000,0.150000,0.150000}%
\pgfsetstrokecolor{textcolor}%
\pgfsetfillcolor{textcolor}%
\pgftext[x=0.214138in, y=0.839857in, left, base]{\color{textcolor}\rmfamily\fontsize{8.000000}{9.600000}\selectfont 50\%}%
\end{pgfscope}%
\begin{pgfscope}%
\pgfpathrectangle{\pgfqpoint{0.520798in}{0.442177in}}{\pgfqpoint{0.719005in}{0.871884in}}%
\pgfusepath{clip}%
\pgfsetroundcap%
\pgfsetroundjoin%
\pgfsetlinewidth{0.501875pt}%
\definecolor{currentstroke}{rgb}{0.800000,0.800000,0.800000}%
\pgfsetstrokecolor{currentstroke}%
\pgfsetdash{}{0pt}%
\pgfpathmoveto{\pgfqpoint{0.520798in}{1.096090in}}%
\pgfpathlineto{\pgfqpoint{1.239803in}{1.096090in}}%
\pgfusepath{stroke}%
\end{pgfscope}%
\begin{pgfscope}%
\definecolor{textcolor}{rgb}{0.150000,0.150000,0.150000}%
\pgfsetstrokecolor{textcolor}%
\pgfsetfillcolor{textcolor}%
\pgftext[x=0.214138in, y=1.057828in, left, base]{\color{textcolor}\rmfamily\fontsize{8.000000}{9.600000}\selectfont 75\%}%
\end{pgfscope}%
\begin{pgfscope}%
\pgfpathrectangle{\pgfqpoint{0.520798in}{0.442177in}}{\pgfqpoint{0.719005in}{0.871884in}}%
\pgfusepath{clip}%
\pgfsetroundcap%
\pgfsetroundjoin%
\pgfsetlinewidth{0.501875pt}%
\definecolor{currentstroke}{rgb}{0.800000,0.800000,0.800000}%
\pgfsetstrokecolor{currentstroke}%
\pgfsetdash{}{0pt}%
\pgfpathmoveto{\pgfqpoint{0.520798in}{1.314061in}}%
\pgfpathlineto{\pgfqpoint{1.239803in}{1.314061in}}%
\pgfusepath{stroke}%
\end{pgfscope}%
\begin{pgfscope}%
\definecolor{textcolor}{rgb}{0.150000,0.150000,0.150000}%
\pgfsetstrokecolor{textcolor}%
\pgfsetfillcolor{textcolor}%
\pgftext[x=0.155124in, y=1.275799in, left, base]{\color{textcolor}\rmfamily\fontsize{8.000000}{9.600000}\selectfont 100\%}%
\end{pgfscope}%
\begin{pgfscope}%
\definecolor{textcolor}{rgb}{0.150000,0.150000,0.150000}%
\pgfsetstrokecolor{textcolor}%
\pgfsetfillcolor{textcolor}%
\pgftext[x=0.099569in,y=0.878119in,,bottom,rotate=90.000000]{\color{textcolor}\rmfamily\fontsize{10.000000}{12.000000}\selectfont Cert. Acc.}%
\end{pgfscope}%
\begin{pgfscope}%
\pgfsetrectcap%
\pgfsetmiterjoin%
\pgfsetlinewidth{0.752812pt}%
\definecolor{currentstroke}{rgb}{0.700000,0.700000,0.700000}%
\pgfsetstrokecolor{currentstroke}%
\pgfsetdash{}{0pt}%
\pgfpathmoveto{\pgfqpoint{0.520798in}{0.442177in}}%
\pgfpathlineto{\pgfqpoint{0.520798in}{1.314061in}}%
\pgfusepath{stroke}%
\end{pgfscope}%
\begin{pgfscope}%
\pgfsetrectcap%
\pgfsetmiterjoin%
\pgfsetlinewidth{0.752812pt}%
\definecolor{currentstroke}{rgb}{0.700000,0.700000,0.700000}%
\pgfsetstrokecolor{currentstroke}%
\pgfsetdash{}{0pt}%
\pgfpathmoveto{\pgfqpoint{1.239803in}{0.442177in}}%
\pgfpathlineto{\pgfqpoint{1.239803in}{1.314061in}}%
\pgfusepath{stroke}%
\end{pgfscope}%
\begin{pgfscope}%
\pgfsetrectcap%
\pgfsetmiterjoin%
\pgfsetlinewidth{0.752812pt}%
\definecolor{currentstroke}{rgb}{0.700000,0.700000,0.700000}%
\pgfsetstrokecolor{currentstroke}%
\pgfsetdash{}{0pt}%
\pgfpathmoveto{\pgfqpoint{0.520798in}{0.442177in}}%
\pgfpathlineto{\pgfqpoint{1.239803in}{0.442177in}}%
\pgfusepath{stroke}%
\end{pgfscope}%
\begin{pgfscope}%
\pgfsetrectcap%
\pgfsetmiterjoin%
\pgfsetlinewidth{0.752812pt}%
\definecolor{currentstroke}{rgb}{0.700000,0.700000,0.700000}%
\pgfsetstrokecolor{currentstroke}%
\pgfsetdash{}{0pt}%
\pgfpathmoveto{\pgfqpoint{0.520798in}{1.314061in}}%
\pgfpathlineto{\pgfqpoint{1.239803in}{1.314061in}}%
\pgfusepath{stroke}%
\end{pgfscope}%
\begin{pgfscope}%
\pgfpathrectangle{\pgfqpoint{0.520798in}{0.442177in}}{\pgfqpoint{0.719005in}{0.871884in}}%
\pgfusepath{clip}%
\pgfsetroundcap%
\pgfsetroundjoin%
\pgfsetlinewidth{1.003750pt}%
\definecolor{currentstroke}{rgb}{0.003922,0.450980,0.698039}%
\pgfsetstrokecolor{currentstroke}%
\pgfsetdash{}{0pt}%
\pgfpathmoveto{\pgfqpoint{0.520798in}{1.076241in}}%
\pgfpathlineto{\pgfqpoint{0.556748in}{1.076241in}}%
\pgfpathlineto{\pgfqpoint{0.556748in}{1.022815in}}%
\pgfpathlineto{\pgfqpoint{0.628649in}{1.022815in}}%
\pgfpathlineto{\pgfqpoint{0.628649in}{0.941284in}}%
\pgfpathlineto{\pgfqpoint{0.700549in}{0.941284in}}%
\pgfpathlineto{\pgfqpoint{0.700549in}{0.732403in}}%
\pgfpathlineto{\pgfqpoint{0.772450in}{0.732403in}}%
\pgfpathlineto{\pgfqpoint{0.772450in}{0.617667in}}%
\pgfpathlineto{\pgfqpoint{0.844350in}{0.617667in}}%
\pgfpathlineto{\pgfqpoint{0.844350in}{0.606907in}}%
\pgfpathlineto{\pgfqpoint{0.916251in}{0.606907in}}%
\pgfpathlineto{\pgfqpoint{0.916251in}{0.599209in}}%
\pgfpathlineto{\pgfqpoint{0.988151in}{0.599209in}}%
\pgfpathlineto{\pgfqpoint{0.988151in}{0.586223in}}%
\pgfpathlineto{\pgfqpoint{1.060052in}{0.586223in}}%
\pgfpathlineto{\pgfqpoint{1.060052in}{0.537157in}}%
\pgfpathlineto{\pgfqpoint{1.131952in}{0.537157in}}%
\pgfpathlineto{\pgfqpoint{1.131952in}{0.528067in}}%
\pgfpathlineto{\pgfqpoint{1.203853in}{0.528067in}}%
\pgfpathlineto{\pgfqpoint{1.203853in}{0.511464in}}%
\pgfpathlineto{\pgfqpoint{1.241470in}{0.511464in}}%
\pgfusepath{stroke}%
\end{pgfscope}%
\begin{pgfscope}%
\pgfpathrectangle{\pgfqpoint{0.520798in}{0.442177in}}{\pgfqpoint{0.719005in}{0.871884in}}%
\pgfusepath{clip}%
\pgfsetbuttcap%
\pgfsetroundjoin%
\definecolor{currentfill}{rgb}{0.003922,0.450980,0.698039}%
\pgfsetfillcolor{currentfill}%
\pgfsetfillopacity{0.500000}%
\pgfsetlinewidth{0.000000pt}%
\definecolor{currentstroke}{rgb}{0.003922,0.450980,0.698039}%
\pgfsetstrokecolor{currentstroke}%
\pgfsetstrokeopacity{0.500000}%
\pgfsetdash{}{0pt}%
\pgfpathmoveto{\pgfqpoint{0.520798in}{1.084698in}}%
\pgfpathlineto{\pgfqpoint{0.520798in}{1.067783in}}%
\pgfpathlineto{\pgfqpoint{0.556748in}{1.067783in}}%
\pgfpathlineto{\pgfqpoint{0.556748in}{1.009211in}}%
\pgfpathlineto{\pgfqpoint{0.628649in}{1.009211in}}%
\pgfpathlineto{\pgfqpoint{0.628649in}{0.925079in}}%
\pgfpathlineto{\pgfqpoint{0.700549in}{0.925079in}}%
\pgfpathlineto{\pgfqpoint{0.700549in}{0.717807in}}%
\pgfpathlineto{\pgfqpoint{0.772450in}{0.717807in}}%
\pgfpathlineto{\pgfqpoint{0.772450in}{0.603032in}}%
\pgfpathlineto{\pgfqpoint{0.844350in}{0.603032in}}%
\pgfpathlineto{\pgfqpoint{0.844350in}{0.592293in}}%
\pgfpathlineto{\pgfqpoint{0.916251in}{0.592293in}}%
\pgfpathlineto{\pgfqpoint{0.916251in}{0.584199in}}%
\pgfpathlineto{\pgfqpoint{0.988151in}{0.584199in}}%
\pgfpathlineto{\pgfqpoint{0.988151in}{0.572922in}}%
\pgfpathlineto{\pgfqpoint{1.060052in}{0.572922in}}%
\pgfpathlineto{\pgfqpoint{1.060052in}{0.528709in}}%
\pgfpathlineto{\pgfqpoint{1.131952in}{0.528709in}}%
\pgfpathlineto{\pgfqpoint{1.131952in}{0.519275in}}%
\pgfpathlineto{\pgfqpoint{1.203853in}{0.519275in}}%
\pgfpathlineto{\pgfqpoint{1.203853in}{0.505189in}}%
\pgfpathlineto{\pgfqpoint{1.275753in}{0.505189in}}%
\pgfpathlineto{\pgfqpoint{1.275753in}{0.486922in}}%
\pgfpathlineto{\pgfqpoint{1.347654in}{0.486922in}}%
\pgfpathlineto{\pgfqpoint{1.347654in}{0.480412in}}%
\pgfpathlineto{\pgfqpoint{1.419554in}{0.480412in}}%
\pgfpathlineto{\pgfqpoint{1.419554in}{0.442177in}}%
\pgfpathlineto{\pgfqpoint{2.066659in}{0.442177in}}%
\pgfpathlineto{\pgfqpoint{2.066659in}{0.442177in}}%
\pgfpathlineto{\pgfqpoint{2.677813in}{0.442177in}}%
\pgfpathlineto{\pgfqpoint{2.677813in}{0.442177in}}%
\pgfpathlineto{\pgfqpoint{2.677813in}{0.442177in}}%
\pgfpathlineto{\pgfqpoint{2.066659in}{0.442177in}}%
\pgfpathlineto{\pgfqpoint{2.066659in}{0.442177in}}%
\pgfpathlineto{\pgfqpoint{1.419554in}{0.442177in}}%
\pgfpathlineto{\pgfqpoint{1.419554in}{0.494841in}}%
\pgfpathlineto{\pgfqpoint{1.347654in}{0.494841in}}%
\pgfpathlineto{\pgfqpoint{1.347654in}{0.502614in}}%
\pgfpathlineto{\pgfqpoint{1.275753in}{0.502614in}}%
\pgfpathlineto{\pgfqpoint{1.275753in}{0.517738in}}%
\pgfpathlineto{\pgfqpoint{1.203853in}{0.517738in}}%
\pgfpathlineto{\pgfqpoint{1.203853in}{0.536858in}}%
\pgfpathlineto{\pgfqpoint{1.131952in}{0.536858in}}%
\pgfpathlineto{\pgfqpoint{1.131952in}{0.545604in}}%
\pgfpathlineto{\pgfqpoint{1.060052in}{0.545604in}}%
\pgfpathlineto{\pgfqpoint{1.060052in}{0.599525in}}%
\pgfpathlineto{\pgfqpoint{0.988151in}{0.599525in}}%
\pgfpathlineto{\pgfqpoint{0.988151in}{0.614218in}}%
\pgfpathlineto{\pgfqpoint{0.916251in}{0.614218in}}%
\pgfpathlineto{\pgfqpoint{0.916251in}{0.621521in}}%
\pgfpathlineto{\pgfqpoint{0.844350in}{0.621521in}}%
\pgfpathlineto{\pgfqpoint{0.844350in}{0.632302in}}%
\pgfpathlineto{\pgfqpoint{0.772450in}{0.632302in}}%
\pgfpathlineto{\pgfqpoint{0.772450in}{0.746999in}}%
\pgfpathlineto{\pgfqpoint{0.700549in}{0.746999in}}%
\pgfpathlineto{\pgfqpoint{0.700549in}{0.957489in}}%
\pgfpathlineto{\pgfqpoint{0.628649in}{0.957489in}}%
\pgfpathlineto{\pgfqpoint{0.628649in}{1.036418in}}%
\pgfpathlineto{\pgfqpoint{0.556748in}{1.036418in}}%
\pgfpathlineto{\pgfqpoint{0.556748in}{1.084698in}}%
\pgfpathlineto{\pgfqpoint{0.520798in}{1.084698in}}%
\pgfpathlineto{\pgfqpoint{0.520798in}{1.084698in}}%
\pgfpathclose%
\pgfusepath{fill}%
\end{pgfscope}%
\begin{pgfscope}%
\pgfpathrectangle{\pgfqpoint{0.520798in}{0.442177in}}{\pgfqpoint{0.719005in}{0.871884in}}%
\pgfusepath{clip}%
\pgfsetroundcap%
\pgfsetroundjoin%
\pgfsetlinewidth{1.003750pt}%
\definecolor{currentstroke}{rgb}{0.870588,0.560784,0.019608}%
\pgfsetstrokecolor{currentstroke}%
\pgfsetdash{}{0pt}%
\pgfpathmoveto{\pgfqpoint{0.520798in}{1.076241in}}%
\pgfpathlineto{\pgfqpoint{0.556748in}{1.076241in}}%
\pgfpathlineto{\pgfqpoint{0.556748in}{1.034502in}}%
\pgfpathlineto{\pgfqpoint{0.628649in}{1.034502in}}%
\pgfpathlineto{\pgfqpoint{0.628649in}{0.995267in}}%
\pgfpathlineto{\pgfqpoint{0.700549in}{0.995267in}}%
\pgfpathlineto{\pgfqpoint{0.700549in}{0.961412in}}%
\pgfpathlineto{\pgfqpoint{0.772450in}{0.961412in}}%
\pgfpathlineto{\pgfqpoint{0.772450in}{0.933864in}}%
\pgfpathlineto{\pgfqpoint{0.844350in}{0.933864in}}%
\pgfpathlineto{\pgfqpoint{0.844350in}{0.906038in}}%
\pgfpathlineto{\pgfqpoint{0.916251in}{0.906038in}}%
\pgfpathlineto{\pgfqpoint{0.916251in}{0.732403in}}%
\pgfpathlineto{\pgfqpoint{0.988151in}{0.732403in}}%
\pgfpathlineto{\pgfqpoint{0.988151in}{0.718768in}}%
\pgfpathlineto{\pgfqpoint{1.060052in}{0.718768in}}%
\pgfpathlineto{\pgfqpoint{1.060052in}{0.617667in}}%
\pgfpathlineto{\pgfqpoint{1.131952in}{0.617667in}}%
\pgfpathlineto{\pgfqpoint{1.131952in}{0.610154in}}%
\pgfpathlineto{\pgfqpoint{1.203853in}{0.610154in}}%
\pgfpathlineto{\pgfqpoint{1.203853in}{0.601064in}}%
\pgfpathlineto{\pgfqpoint{1.241470in}{0.601064in}}%
\pgfusepath{stroke}%
\end{pgfscope}%
\begin{pgfscope}%
\pgfpathrectangle{\pgfqpoint{0.520798in}{0.442177in}}{\pgfqpoint{0.719005in}{0.871884in}}%
\pgfusepath{clip}%
\pgfsetbuttcap%
\pgfsetroundjoin%
\definecolor{currentfill}{rgb}{0.870588,0.560784,0.019608}%
\pgfsetfillcolor{currentfill}%
\pgfsetfillopacity{0.500000}%
\pgfsetlinewidth{0.000000pt}%
\definecolor{currentstroke}{rgb}{0.870588,0.560784,0.019608}%
\pgfsetstrokecolor{currentstroke}%
\pgfsetstrokeopacity{0.500000}%
\pgfsetdash{}{0pt}%
\pgfpathmoveto{\pgfqpoint{0.520798in}{1.084698in}}%
\pgfpathlineto{\pgfqpoint{0.520798in}{1.067783in}}%
\pgfpathlineto{\pgfqpoint{0.556748in}{1.067783in}}%
\pgfpathlineto{\pgfqpoint{0.556748in}{1.020687in}}%
\pgfpathlineto{\pgfqpoint{0.628649in}{1.020687in}}%
\pgfpathlineto{\pgfqpoint{0.628649in}{0.979272in}}%
\pgfpathlineto{\pgfqpoint{0.700549in}{0.979272in}}%
\pgfpathlineto{\pgfqpoint{0.700549in}{0.946454in}}%
\pgfpathlineto{\pgfqpoint{0.772450in}{0.946454in}}%
\pgfpathlineto{\pgfqpoint{0.772450in}{0.917054in}}%
\pgfpathlineto{\pgfqpoint{0.844350in}{0.917054in}}%
\pgfpathlineto{\pgfqpoint{0.844350in}{0.888534in}}%
\pgfpathlineto{\pgfqpoint{0.916251in}{0.888534in}}%
\pgfpathlineto{\pgfqpoint{0.916251in}{0.717807in}}%
\pgfpathlineto{\pgfqpoint{0.988151in}{0.717807in}}%
\pgfpathlineto{\pgfqpoint{0.988151in}{0.702452in}}%
\pgfpathlineto{\pgfqpoint{1.060052in}{0.702452in}}%
\pgfpathlineto{\pgfqpoint{1.060052in}{0.603032in}}%
\pgfpathlineto{\pgfqpoint{1.131952in}{0.603032in}}%
\pgfpathlineto{\pgfqpoint{1.131952in}{0.595039in}}%
\pgfpathlineto{\pgfqpoint{1.203853in}{0.595039in}}%
\pgfpathlineto{\pgfqpoint{1.203853in}{0.586633in}}%
\pgfpathlineto{\pgfqpoint{1.275753in}{0.586633in}}%
\pgfpathlineto{\pgfqpoint{1.275753in}{0.580247in}}%
\pgfpathlineto{\pgfqpoint{1.347654in}{0.580247in}}%
\pgfpathlineto{\pgfqpoint{1.347654in}{0.573946in}}%
\pgfpathlineto{\pgfqpoint{1.419554in}{0.573946in}}%
\pgfpathlineto{\pgfqpoint{1.419554in}{0.568846in}}%
\pgfpathlineto{\pgfqpoint{1.491455in}{0.568846in}}%
\pgfpathlineto{\pgfqpoint{1.491455in}{0.563131in}}%
\pgfpathlineto{\pgfqpoint{1.563355in}{0.563131in}}%
\pgfpathlineto{\pgfqpoint{1.563355in}{0.559200in}}%
\pgfpathlineto{\pgfqpoint{1.635256in}{0.559200in}}%
\pgfpathlineto{\pgfqpoint{1.635256in}{0.528124in}}%
\pgfpathlineto{\pgfqpoint{1.707156in}{0.528124in}}%
\pgfpathlineto{\pgfqpoint{1.707156in}{0.524241in}}%
\pgfpathlineto{\pgfqpoint{1.779057in}{0.524241in}}%
\pgfpathlineto{\pgfqpoint{1.779057in}{0.516763in}}%
\pgfpathlineto{\pgfqpoint{1.850957in}{0.516763in}}%
\pgfpathlineto{\pgfqpoint{1.850957in}{0.512653in}}%
\pgfpathlineto{\pgfqpoint{1.922858in}{0.512653in}}%
\pgfpathlineto{\pgfqpoint{1.922858in}{0.504897in}}%
\pgfpathlineto{\pgfqpoint{1.994758in}{0.504897in}}%
\pgfpathlineto{\pgfqpoint{1.994758in}{0.489966in}}%
\pgfpathlineto{\pgfqpoint{2.066659in}{0.489966in}}%
\pgfpathlineto{\pgfqpoint{2.066659in}{0.481476in}}%
\pgfpathlineto{\pgfqpoint{2.138559in}{0.481476in}}%
\pgfpathlineto{\pgfqpoint{2.138559in}{0.479217in}}%
\pgfpathlineto{\pgfqpoint{2.210460in}{0.479217in}}%
\pgfpathlineto{\pgfqpoint{2.210460in}{0.463427in}}%
\pgfpathlineto{\pgfqpoint{2.282360in}{0.463427in}}%
\pgfpathlineto{\pgfqpoint{2.282360in}{0.442177in}}%
\pgfpathlineto{\pgfqpoint{2.498062in}{0.442177in}}%
\pgfpathlineto{\pgfqpoint{2.498062in}{0.442177in}}%
\pgfpathlineto{\pgfqpoint{2.677813in}{0.442177in}}%
\pgfpathlineto{\pgfqpoint{2.677813in}{0.442177in}}%
\pgfpathlineto{\pgfqpoint{2.677813in}{0.442177in}}%
\pgfpathlineto{\pgfqpoint{2.498062in}{0.442177in}}%
\pgfpathlineto{\pgfqpoint{2.498062in}{0.442177in}}%
\pgfpathlineto{\pgfqpoint{2.282360in}{0.442177in}}%
\pgfpathlineto{\pgfqpoint{2.282360in}{0.473611in}}%
\pgfpathlineto{\pgfqpoint{2.210460in}{0.473611in}}%
\pgfpathlineto{\pgfqpoint{2.210460in}{0.490841in}}%
\pgfpathlineto{\pgfqpoint{2.138559in}{0.490841in}}%
\pgfpathlineto{\pgfqpoint{2.138559in}{0.496558in}}%
\pgfpathlineto{\pgfqpoint{2.066659in}{0.496558in}}%
\pgfpathlineto{\pgfqpoint{2.066659in}{0.505135in}}%
\pgfpathlineto{\pgfqpoint{1.994758in}{0.505135in}}%
\pgfpathlineto{\pgfqpoint{1.994758in}{0.516732in}}%
\pgfpathlineto{\pgfqpoint{1.922858in}{0.516732in}}%
\pgfpathlineto{\pgfqpoint{1.922858in}{0.529752in}}%
\pgfpathlineto{\pgfqpoint{1.850957in}{0.529752in}}%
\pgfpathlineto{\pgfqpoint{1.850957in}{0.534547in}}%
\pgfpathlineto{\pgfqpoint{1.779057in}{0.534547in}}%
\pgfpathlineto{\pgfqpoint{1.779057in}{0.542838in}}%
\pgfpathlineto{\pgfqpoint{1.707156in}{0.542838in}}%
\pgfpathlineto{\pgfqpoint{1.707156in}{0.545447in}}%
\pgfpathlineto{\pgfqpoint{1.635256in}{0.545447in}}%
\pgfpathlineto{\pgfqpoint{1.635256in}{0.583936in}}%
\pgfpathlineto{\pgfqpoint{1.563355in}{0.583936in}}%
\pgfpathlineto{\pgfqpoint{1.563355in}{0.589281in}}%
\pgfpathlineto{\pgfqpoint{1.491455in}{0.589281in}}%
\pgfpathlineto{\pgfqpoint{1.491455in}{0.593769in}}%
\pgfpathlineto{\pgfqpoint{1.419554in}{0.593769in}}%
\pgfpathlineto{\pgfqpoint{1.419554in}{0.601283in}}%
\pgfpathlineto{\pgfqpoint{1.347654in}{0.601283in}}%
\pgfpathlineto{\pgfqpoint{1.347654in}{0.608524in}}%
\pgfpathlineto{\pgfqpoint{1.275753in}{0.608524in}}%
\pgfpathlineto{\pgfqpoint{1.275753in}{0.615495in}}%
\pgfpathlineto{\pgfqpoint{1.203853in}{0.615495in}}%
\pgfpathlineto{\pgfqpoint{1.203853in}{0.625268in}}%
\pgfpathlineto{\pgfqpoint{1.131952in}{0.625268in}}%
\pgfpathlineto{\pgfqpoint{1.131952in}{0.632302in}}%
\pgfpathlineto{\pgfqpoint{1.060052in}{0.632302in}}%
\pgfpathlineto{\pgfqpoint{1.060052in}{0.735085in}}%
\pgfpathlineto{\pgfqpoint{0.988151in}{0.735085in}}%
\pgfpathlineto{\pgfqpoint{0.988151in}{0.746999in}}%
\pgfpathlineto{\pgfqpoint{0.916251in}{0.746999in}}%
\pgfpathlineto{\pgfqpoint{0.916251in}{0.923542in}}%
\pgfpathlineto{\pgfqpoint{0.844350in}{0.923542in}}%
\pgfpathlineto{\pgfqpoint{0.844350in}{0.950674in}}%
\pgfpathlineto{\pgfqpoint{0.772450in}{0.950674in}}%
\pgfpathlineto{\pgfqpoint{0.772450in}{0.976370in}}%
\pgfpathlineto{\pgfqpoint{0.700549in}{0.976370in}}%
\pgfpathlineto{\pgfqpoint{0.700549in}{1.011261in}}%
\pgfpathlineto{\pgfqpoint{0.628649in}{1.011261in}}%
\pgfpathlineto{\pgfqpoint{0.628649in}{1.048317in}}%
\pgfpathlineto{\pgfqpoint{0.556748in}{1.048317in}}%
\pgfpathlineto{\pgfqpoint{0.556748in}{1.084698in}}%
\pgfpathlineto{\pgfqpoint{0.520798in}{1.084698in}}%
\pgfpathlineto{\pgfqpoint{0.520798in}{1.084698in}}%
\pgfpathclose%
\pgfusepath{fill}%
\end{pgfscope}%
\begin{pgfscope}%
\pgfpathrectangle{\pgfqpoint{0.520798in}{0.442177in}}{\pgfqpoint{0.719005in}{0.871884in}}%
\pgfusepath{clip}%
\pgfsetroundcap%
\pgfsetroundjoin%
\pgfsetlinewidth{1.003750pt}%
\definecolor{currentstroke}{rgb}{0.007843,0.619608,0.450980}%
\pgfsetstrokecolor{currentstroke}%
\pgfsetdash{}{0pt}%
\pgfpathmoveto{\pgfqpoint{0.520798in}{1.076241in}}%
\pgfpathlineto{\pgfqpoint{0.556748in}{1.076241in}}%
\pgfpathlineto{\pgfqpoint{0.556748in}{1.034502in}}%
\pgfpathlineto{\pgfqpoint{0.628649in}{1.034502in}}%
\pgfpathlineto{\pgfqpoint{0.628649in}{0.995267in}}%
\pgfpathlineto{\pgfqpoint{0.700549in}{0.995267in}}%
\pgfpathlineto{\pgfqpoint{0.700549in}{0.961412in}}%
\pgfpathlineto{\pgfqpoint{0.772450in}{0.961412in}}%
\pgfpathlineto{\pgfqpoint{0.772450in}{0.933864in}}%
\pgfpathlineto{\pgfqpoint{0.844350in}{0.933864in}}%
\pgfpathlineto{\pgfqpoint{0.844350in}{0.906038in}}%
\pgfpathlineto{\pgfqpoint{0.916251in}{0.906038in}}%
\pgfpathlineto{\pgfqpoint{0.916251in}{0.877748in}}%
\pgfpathlineto{\pgfqpoint{0.988151in}{0.877748in}}%
\pgfpathlineto{\pgfqpoint{0.988151in}{0.853818in}}%
\pgfpathlineto{\pgfqpoint{1.060052in}{0.853818in}}%
\pgfpathlineto{\pgfqpoint{1.060052in}{0.831000in}}%
\pgfpathlineto{\pgfqpoint{1.131952in}{0.831000in}}%
\pgfpathlineto{\pgfqpoint{1.131952in}{0.811893in}}%
\pgfpathlineto{\pgfqpoint{1.203853in}{0.811893in}}%
\pgfpathlineto{\pgfqpoint{1.203853in}{0.791116in}}%
\pgfpathlineto{\pgfqpoint{1.241470in}{0.791116in}}%
\pgfusepath{stroke}%
\end{pgfscope}%
\begin{pgfscope}%
\pgfpathrectangle{\pgfqpoint{0.520798in}{0.442177in}}{\pgfqpoint{0.719005in}{0.871884in}}%
\pgfusepath{clip}%
\pgfsetbuttcap%
\pgfsetroundjoin%
\definecolor{currentfill}{rgb}{0.007843,0.619608,0.450980}%
\pgfsetfillcolor{currentfill}%
\pgfsetfillopacity{0.500000}%
\pgfsetlinewidth{0.000000pt}%
\definecolor{currentstroke}{rgb}{0.007843,0.619608,0.450980}%
\pgfsetstrokecolor{currentstroke}%
\pgfsetstrokeopacity{0.500000}%
\pgfsetdash{}{0pt}%
\pgfpathmoveto{\pgfqpoint{0.520798in}{1.084698in}}%
\pgfpathlineto{\pgfqpoint{0.520798in}{1.067783in}}%
\pgfpathlineto{\pgfqpoint{0.556748in}{1.067783in}}%
\pgfpathlineto{\pgfqpoint{0.556748in}{1.020687in}}%
\pgfpathlineto{\pgfqpoint{0.628649in}{1.020687in}}%
\pgfpathlineto{\pgfqpoint{0.628649in}{0.979272in}}%
\pgfpathlineto{\pgfqpoint{0.700549in}{0.979272in}}%
\pgfpathlineto{\pgfqpoint{0.700549in}{0.946454in}}%
\pgfpathlineto{\pgfqpoint{0.772450in}{0.946454in}}%
\pgfpathlineto{\pgfqpoint{0.772450in}{0.917054in}}%
\pgfpathlineto{\pgfqpoint{0.844350in}{0.917054in}}%
\pgfpathlineto{\pgfqpoint{0.844350in}{0.888534in}}%
\pgfpathlineto{\pgfqpoint{0.916251in}{0.888534in}}%
\pgfpathlineto{\pgfqpoint{0.916251in}{0.861668in}}%
\pgfpathlineto{\pgfqpoint{0.988151in}{0.861668in}}%
\pgfpathlineto{\pgfqpoint{0.988151in}{0.838454in}}%
\pgfpathlineto{\pgfqpoint{1.060052in}{0.838454in}}%
\pgfpathlineto{\pgfqpoint{1.060052in}{0.815220in}}%
\pgfpathlineto{\pgfqpoint{1.131952in}{0.815220in}}%
\pgfpathlineto{\pgfqpoint{1.131952in}{0.795851in}}%
\pgfpathlineto{\pgfqpoint{1.203853in}{0.795851in}}%
\pgfpathlineto{\pgfqpoint{1.203853in}{0.776306in}}%
\pgfpathlineto{\pgfqpoint{1.275753in}{0.776306in}}%
\pgfpathlineto{\pgfqpoint{1.275753in}{0.754266in}}%
\pgfpathlineto{\pgfqpoint{1.347654in}{0.754266in}}%
\pgfpathlineto{\pgfqpoint{1.347654in}{0.717807in}}%
\pgfpathlineto{\pgfqpoint{1.419554in}{0.717807in}}%
\pgfpathlineto{\pgfqpoint{1.419554in}{0.702452in}}%
\pgfpathlineto{\pgfqpoint{1.491455in}{0.702452in}}%
\pgfpathlineto{\pgfqpoint{1.491455in}{0.690710in}}%
\pgfpathlineto{\pgfqpoint{1.563355in}{0.690710in}}%
\pgfpathlineto{\pgfqpoint{1.563355in}{0.679530in}}%
\pgfpathlineto{\pgfqpoint{1.635256in}{0.679530in}}%
\pgfpathlineto{\pgfqpoint{1.635256in}{0.603032in}}%
\pgfpathlineto{\pgfqpoint{1.707156in}{0.603032in}}%
\pgfpathlineto{\pgfqpoint{1.707156in}{0.595039in}}%
\pgfpathlineto{\pgfqpoint{1.779057in}{0.595039in}}%
\pgfpathlineto{\pgfqpoint{1.779057in}{0.587815in}}%
\pgfpathlineto{\pgfqpoint{1.850957in}{0.587815in}}%
\pgfpathlineto{\pgfqpoint{1.850957in}{0.581541in}}%
\pgfpathlineto{\pgfqpoint{1.922858in}{0.581541in}}%
\pgfpathlineto{\pgfqpoint{1.922858in}{0.575564in}}%
\pgfpathlineto{\pgfqpoint{1.994758in}{0.575564in}}%
\pgfpathlineto{\pgfqpoint{1.994758in}{0.570676in}}%
\pgfpathlineto{\pgfqpoint{2.066659in}{0.570676in}}%
\pgfpathlineto{\pgfqpoint{2.066659in}{0.565836in}}%
\pgfpathlineto{\pgfqpoint{2.138559in}{0.565836in}}%
\pgfpathlineto{\pgfqpoint{2.138559in}{0.562277in}}%
\pgfpathlineto{\pgfqpoint{2.210460in}{0.562277in}}%
\pgfpathlineto{\pgfqpoint{2.210460in}{0.557011in}}%
\pgfpathlineto{\pgfqpoint{2.282360in}{0.557011in}}%
\pgfpathlineto{\pgfqpoint{2.282360in}{0.555271in}}%
\pgfpathlineto{\pgfqpoint{2.354261in}{0.555271in}}%
\pgfpathlineto{\pgfqpoint{2.354261in}{0.551008in}}%
\pgfpathlineto{\pgfqpoint{2.426161in}{0.551008in}}%
\pgfpathlineto{\pgfqpoint{2.426161in}{0.546809in}}%
\pgfpathlineto{\pgfqpoint{2.498062in}{0.546809in}}%
\pgfpathlineto{\pgfqpoint{2.498062in}{0.542291in}}%
\pgfpathlineto{\pgfqpoint{2.569962in}{0.542291in}}%
\pgfpathlineto{\pgfqpoint{2.569962in}{0.540062in}}%
\pgfpathlineto{\pgfqpoint{2.641863in}{0.540062in}}%
\pgfpathlineto{\pgfqpoint{2.641863in}{0.536234in}}%
\pgfpathlineto{\pgfqpoint{2.677813in}{0.536234in}}%
\pgfpathlineto{\pgfqpoint{2.677813in}{0.557186in}}%
\pgfpathlineto{\pgfqpoint{2.677813in}{0.557186in}}%
\pgfpathlineto{\pgfqpoint{2.641863in}{0.557186in}}%
\pgfpathlineto{\pgfqpoint{2.641863in}{0.561335in}}%
\pgfpathlineto{\pgfqpoint{2.569962in}{0.561335in}}%
\pgfpathlineto{\pgfqpoint{2.569962in}{0.564486in}}%
\pgfpathlineto{\pgfqpoint{2.498062in}{0.564486in}}%
\pgfpathlineto{\pgfqpoint{2.498062in}{0.568872in}}%
\pgfpathlineto{\pgfqpoint{2.426161in}{0.568872in}}%
\pgfpathlineto{\pgfqpoint{2.426161in}{0.571723in}}%
\pgfpathlineto{\pgfqpoint{2.354261in}{0.571723in}}%
\pgfpathlineto{\pgfqpoint{2.354261in}{0.577663in}}%
\pgfpathlineto{\pgfqpoint{2.282360in}{0.577663in}}%
\pgfpathlineto{\pgfqpoint{2.282360in}{0.581488in}}%
\pgfpathlineto{\pgfqpoint{2.210460in}{0.581488in}}%
\pgfpathlineto{\pgfqpoint{2.210460in}{0.587352in}}%
\pgfpathlineto{\pgfqpoint{2.138559in}{0.587352in}}%
\pgfpathlineto{\pgfqpoint{2.138559in}{0.593068in}}%
\pgfpathlineto{\pgfqpoint{2.066659in}{0.593068in}}%
\pgfpathlineto{\pgfqpoint{2.066659in}{0.597689in}}%
\pgfpathlineto{\pgfqpoint{1.994758in}{0.597689in}}%
\pgfpathlineto{\pgfqpoint{1.994758in}{0.603746in}}%
\pgfpathlineto{\pgfqpoint{1.922858in}{0.603746in}}%
\pgfpathlineto{\pgfqpoint{1.922858in}{0.609827in}}%
\pgfpathlineto{\pgfqpoint{1.850957in}{0.609827in}}%
\pgfpathlineto{\pgfqpoint{1.850957in}{0.617467in}}%
\pgfpathlineto{\pgfqpoint{1.779057in}{0.617467in}}%
\pgfpathlineto{\pgfqpoint{1.779057in}{0.625268in}}%
\pgfpathlineto{\pgfqpoint{1.707156in}{0.625268in}}%
\pgfpathlineto{\pgfqpoint{1.707156in}{0.632302in}}%
\pgfpathlineto{\pgfqpoint{1.635256in}{0.632302in}}%
\pgfpathlineto{\pgfqpoint{1.635256in}{0.707734in}}%
\pgfpathlineto{\pgfqpoint{1.563355in}{0.707734in}}%
\pgfpathlineto{\pgfqpoint{1.563355in}{0.721412in}}%
\pgfpathlineto{\pgfqpoint{1.491455in}{0.721412in}}%
\pgfpathlineto{\pgfqpoint{1.491455in}{0.735085in}}%
\pgfpathlineto{\pgfqpoint{1.419554in}{0.735085in}}%
\pgfpathlineto{\pgfqpoint{1.419554in}{0.746999in}}%
\pgfpathlineto{\pgfqpoint{1.347654in}{0.746999in}}%
\pgfpathlineto{\pgfqpoint{1.347654in}{0.784186in}}%
\pgfpathlineto{\pgfqpoint{1.275753in}{0.784186in}}%
\pgfpathlineto{\pgfqpoint{1.275753in}{0.805927in}}%
\pgfpathlineto{\pgfqpoint{1.203853in}{0.805927in}}%
\pgfpathlineto{\pgfqpoint{1.203853in}{0.827935in}}%
\pgfpathlineto{\pgfqpoint{1.131952in}{0.827935in}}%
\pgfpathlineto{\pgfqpoint{1.131952in}{0.846781in}}%
\pgfpathlineto{\pgfqpoint{1.060052in}{0.846781in}}%
\pgfpathlineto{\pgfqpoint{1.060052in}{0.869181in}}%
\pgfpathlineto{\pgfqpoint{0.988151in}{0.869181in}}%
\pgfpathlineto{\pgfqpoint{0.988151in}{0.893828in}}%
\pgfpathlineto{\pgfqpoint{0.916251in}{0.893828in}}%
\pgfpathlineto{\pgfqpoint{0.916251in}{0.923542in}}%
\pgfpathlineto{\pgfqpoint{0.844350in}{0.923542in}}%
\pgfpathlineto{\pgfqpoint{0.844350in}{0.950674in}}%
\pgfpathlineto{\pgfqpoint{0.772450in}{0.950674in}}%
\pgfpathlineto{\pgfqpoint{0.772450in}{0.976370in}}%
\pgfpathlineto{\pgfqpoint{0.700549in}{0.976370in}}%
\pgfpathlineto{\pgfqpoint{0.700549in}{1.011261in}}%
\pgfpathlineto{\pgfqpoint{0.628649in}{1.011261in}}%
\pgfpathlineto{\pgfqpoint{0.628649in}{1.048317in}}%
\pgfpathlineto{\pgfqpoint{0.556748in}{1.048317in}}%
\pgfpathlineto{\pgfqpoint{0.556748in}{1.084698in}}%
\pgfpathlineto{\pgfqpoint{0.520798in}{1.084698in}}%
\pgfpathlineto{\pgfqpoint{0.520798in}{1.084698in}}%
\pgfpathclose%
\pgfusepath{fill}%
\end{pgfscope}%
\begin{pgfscope}%
\definecolor{textcolor}{rgb}{0.150000,0.150000,0.150000}%
\pgfsetstrokecolor{textcolor}%
\pgfsetfillcolor{textcolor}%
\pgftext[x=0.880300in,y=1.397395in,,base]{\color{textcolor}\rmfamily\fontsize{9.000000}{10.800000}\selectfont \(\displaystyle c_X^-=1\)}%
\end{pgfscope}%
\begin{pgfscope}%
\pgfsetbuttcap%
\pgfsetmiterjoin%
\definecolor{currentfill}{rgb}{1.000000,1.000000,1.000000}%
\pgfsetfillcolor{currentfill}%
\pgfsetfillopacity{0.800000}%
\pgfsetlinewidth{1.003750pt}%
\definecolor{currentstroke}{rgb}{0.800000,0.800000,0.800000}%
\pgfsetstrokecolor{currentstroke}%
\pgfsetstrokeopacity{0.800000}%
\pgfsetdash{}{0pt}%
\pgfpathmoveto{\pgfqpoint{0.785843in}{0.638103in}}%
\pgfpathlineto{\pgfqpoint{1.191192in}{0.638103in}}%
\pgfpathlineto{\pgfqpoint{1.191192in}{1.265450in}}%
\pgfpathlineto{\pgfqpoint{0.785843in}{1.265450in}}%
\pgfpathlineto{\pgfqpoint{0.785843in}{0.638103in}}%
\pgfpathclose%
\pgfusepath{stroke,fill}%
\end{pgfscope}%
\begin{pgfscope}%
\definecolor{textcolor}{rgb}{0.150000,0.150000,0.150000}%
\pgfsetstrokecolor{textcolor}%
\pgfsetfillcolor{textcolor}%
\pgftext[x=0.912291in,y=1.113275in,left,base]{\color{textcolor}\rmfamily\fontsize{9.000000}{10.800000}\selectfont \(\displaystyle c_X^+\)}%
\end{pgfscope}%
\begin{pgfscope}%
\pgfsetroundcap%
\pgfsetroundjoin%
\pgfsetlinewidth{1.003750pt}%
\definecolor{currentstroke}{rgb}{0.003922,0.450980,0.698039}%
\pgfsetstrokecolor{currentstroke}%
\pgfsetdash{}{0pt}%
\pgfpathmoveto{\pgfqpoint{0.824732in}{1.001052in}}%
\pgfpathlineto{\pgfqpoint{0.873343in}{1.001052in}}%
\pgfpathlineto{\pgfqpoint{0.873343in}{1.001052in}}%
\pgfpathlineto{\pgfqpoint{0.970565in}{1.001052in}}%
\pgfpathlineto{\pgfqpoint{0.970565in}{1.001052in}}%
\pgfpathlineto{\pgfqpoint{1.019176in}{1.001052in}}%
\pgfusepath{stroke}%
\end{pgfscope}%
\begin{pgfscope}%
\definecolor{textcolor}{rgb}{0.150000,0.150000,0.150000}%
\pgfsetstrokecolor{textcolor}%
\pgfsetfillcolor{textcolor}%
\pgftext[x=1.096954in,y=0.967024in,left,base]{\color{textcolor}\rmfamily\fontsize{7.000000}{8.400000}\selectfont 1}%
\end{pgfscope}%
\begin{pgfscope}%
\pgfsetroundcap%
\pgfsetroundjoin%
\pgfsetlinewidth{1.003750pt}%
\definecolor{currentstroke}{rgb}{0.870588,0.560784,0.019608}%
\pgfsetstrokecolor{currentstroke}%
\pgfsetdash{}{0pt}%
\pgfpathmoveto{\pgfqpoint{0.824732in}{0.865486in}}%
\pgfpathlineto{\pgfqpoint{0.873343in}{0.865486in}}%
\pgfpathlineto{\pgfqpoint{0.873343in}{0.865486in}}%
\pgfpathlineto{\pgfqpoint{0.970565in}{0.865486in}}%
\pgfpathlineto{\pgfqpoint{0.970565in}{0.865486in}}%
\pgfpathlineto{\pgfqpoint{1.019176in}{0.865486in}}%
\pgfusepath{stroke}%
\end{pgfscope}%
\begin{pgfscope}%
\definecolor{textcolor}{rgb}{0.150000,0.150000,0.150000}%
\pgfsetstrokecolor{textcolor}%
\pgfsetfillcolor{textcolor}%
\pgftext[x=1.096954in,y=0.831458in,left,base]{\color{textcolor}\rmfamily\fontsize{7.000000}{8.400000}\selectfont 2}%
\end{pgfscope}%
\begin{pgfscope}%
\pgfsetroundcap%
\pgfsetroundjoin%
\pgfsetlinewidth{1.003750pt}%
\definecolor{currentstroke}{rgb}{0.007843,0.619608,0.450980}%
\pgfsetstrokecolor{currentstroke}%
\pgfsetdash{}{0pt}%
\pgfpathmoveto{\pgfqpoint{0.824732in}{0.729919in}}%
\pgfpathlineto{\pgfqpoint{0.873343in}{0.729919in}}%
\pgfpathlineto{\pgfqpoint{0.873343in}{0.729919in}}%
\pgfpathlineto{\pgfqpoint{0.970565in}{0.729919in}}%
\pgfpathlineto{\pgfqpoint{0.970565in}{0.729919in}}%
\pgfpathlineto{\pgfqpoint{1.019176in}{0.729919in}}%
\pgfusepath{stroke}%
\end{pgfscope}%
\begin{pgfscope}%
\definecolor{textcolor}{rgb}{0.150000,0.150000,0.150000}%
\pgfsetstrokecolor{textcolor}%
\pgfsetfillcolor{textcolor}%
\pgftext[x=1.096954in,y=0.695892in,left,base]{\color{textcolor}\rmfamily\fontsize{7.000000}{8.400000}\selectfont 4}%
\end{pgfscope}%
\end{pgfpicture}%
\makeatother%
\endgroup%

%% file: figures/experiments/graphs/sparse_smoothing/nodes_attributes/node_classification-Citeseer-APPNP-hidden=32-p_adj_plus=0.0-p_adj_minus=0.0-p_att_plus=0.001-p_att_minus=0.8-multi_class_cert-A.pgf
\begingroup%
\makeatletter%
\begin{pgfpicture}%
\pgfpathrectangle{\pgfpointorigin}{\pgfqpoint{1.375000in}{1.581250in}}%
\pgfusepath{use as bounding box, clip}%
\begin{pgfscope}%
\pgfsetbuttcap%
\pgfsetmiterjoin%
\definecolor{currentfill}{rgb}{1.000000,1.000000,1.000000}%
\pgfsetfillcolor{currentfill}%
\pgfsetlinewidth{0.000000pt}%
\definecolor{currentstroke}{rgb}{1.000000,1.000000,1.000000}%
\pgfsetstrokecolor{currentstroke}%
\pgfsetdash{}{0pt}%
\pgfpathmoveto{\pgfqpoint{0.000000in}{0.000000in}}%
\pgfpathlineto{\pgfqpoint{1.375000in}{0.000000in}}%
\pgfpathlineto{\pgfqpoint{1.375000in}{1.581250in}}%
\pgfpathlineto{\pgfqpoint{0.000000in}{1.581250in}}%
\pgfpathlineto{\pgfqpoint{0.000000in}{0.000000in}}%
\pgfpathclose%
\pgfusepath{fill}%
\end{pgfscope}%
\begin{pgfscope}%
\pgfsetbuttcap%
\pgfsetmiterjoin%
\definecolor{currentfill}{rgb}{1.000000,1.000000,1.000000}%
\pgfsetfillcolor{currentfill}%
\pgfsetlinewidth{0.000000pt}%
\definecolor{currentstroke}{rgb}{0.000000,0.000000,0.000000}%
\pgfsetstrokecolor{currentstroke}%
\pgfsetstrokeopacity{0.000000}%
\pgfsetdash{}{0pt}%
\pgfpathmoveto{\pgfqpoint{0.520798in}{0.442177in}}%
\pgfpathlineto{\pgfqpoint{1.239803in}{0.442177in}}%
\pgfpathlineto{\pgfqpoint{1.239803in}{1.314061in}}%
\pgfpathlineto{\pgfqpoint{0.520798in}{1.314061in}}%
\pgfpathlineto{\pgfqpoint{0.520798in}{0.442177in}}%
\pgfpathclose%
\pgfusepath{fill}%
\end{pgfscope}%
\begin{pgfscope}%
\pgfpathrectangle{\pgfqpoint{0.520798in}{0.442177in}}{\pgfqpoint{0.719005in}{0.871884in}}%
\pgfusepath{clip}%
\pgfsetroundcap%
\pgfsetroundjoin%
\pgfsetlinewidth{0.501875pt}%
\definecolor{currentstroke}{rgb}{0.800000,0.800000,0.800000}%
\pgfsetstrokecolor{currentstroke}%
\pgfsetdash{}{0pt}%
\pgfpathmoveto{\pgfqpoint{0.520798in}{0.442177in}}%
\pgfpathlineto{\pgfqpoint{0.520798in}{1.314061in}}%
\pgfusepath{stroke}%
\end{pgfscope}%
\begin{pgfscope}%
\definecolor{textcolor}{rgb}{0.150000,0.150000,0.150000}%
\pgfsetstrokecolor{textcolor}%
\pgfsetfillcolor{textcolor}%
\pgftext[x=0.520798in,y=0.351899in,,top]{\color{textcolor}\rmfamily\fontsize{8.000000}{9.600000}\selectfont \(\displaystyle {0}\)}%
\end{pgfscope}%
\begin{pgfscope}%
\pgfpathrectangle{\pgfqpoint{0.520798in}{0.442177in}}{\pgfqpoint{0.719005in}{0.871884in}}%
\pgfusepath{clip}%
\pgfsetroundcap%
\pgfsetroundjoin%
\pgfsetlinewidth{0.501875pt}%
\definecolor{currentstroke}{rgb}{0.800000,0.800000,0.800000}%
\pgfsetstrokecolor{currentstroke}%
\pgfsetdash{}{0pt}%
\pgfpathmoveto{\pgfqpoint{0.880300in}{0.442177in}}%
\pgfpathlineto{\pgfqpoint{0.880300in}{1.314061in}}%
\pgfusepath{stroke}%
\end{pgfscope}%
\begin{pgfscope}%
\definecolor{textcolor}{rgb}{0.150000,0.150000,0.150000}%
\pgfsetstrokecolor{textcolor}%
\pgfsetfillcolor{textcolor}%
\pgftext[x=0.880300in,y=0.351899in,,top]{\color{textcolor}\rmfamily\fontsize{8.000000}{9.600000}\selectfont \(\displaystyle {5}\)}%
\end{pgfscope}%
\begin{pgfscope}%
\pgfpathrectangle{\pgfqpoint{0.520798in}{0.442177in}}{\pgfqpoint{0.719005in}{0.871884in}}%
\pgfusepath{clip}%
\pgfsetroundcap%
\pgfsetroundjoin%
\pgfsetlinewidth{0.501875pt}%
\definecolor{currentstroke}{rgb}{0.800000,0.800000,0.800000}%
\pgfsetstrokecolor{currentstroke}%
\pgfsetdash{}{0pt}%
\pgfpathmoveto{\pgfqpoint{1.239803in}{0.442177in}}%
\pgfpathlineto{\pgfqpoint{1.239803in}{1.314061in}}%
\pgfusepath{stroke}%
\end{pgfscope}%
\begin{pgfscope}%
\definecolor{textcolor}{rgb}{0.150000,0.150000,0.150000}%
\pgfsetstrokecolor{textcolor}%
\pgfsetfillcolor{textcolor}%
\pgftext[x=1.239803in,y=0.351899in,,top]{\color{textcolor}\rmfamily\fontsize{8.000000}{9.600000}\selectfont \(\displaystyle {10}\)}%
\end{pgfscope}%
\begin{pgfscope}%
\definecolor{textcolor}{rgb}{0.150000,0.150000,0.150000}%
\pgfsetstrokecolor{textcolor}%
\pgfsetfillcolor{textcolor}%
\pgftext[x=0.880300in,y=0.198219in,,top]{\color{textcolor}\rmfamily\fontsize{10.000000}{12.000000}\selectfont Edit distance \(\displaystyle \epsilon\)}%
\end{pgfscope}%
\begin{pgfscope}%
\pgfpathrectangle{\pgfqpoint{0.520798in}{0.442177in}}{\pgfqpoint{0.719005in}{0.871884in}}%
\pgfusepath{clip}%
\pgfsetroundcap%
\pgfsetroundjoin%
\pgfsetlinewidth{0.501875pt}%
\definecolor{currentstroke}{rgb}{0.800000,0.800000,0.800000}%
\pgfsetstrokecolor{currentstroke}%
\pgfsetdash{}{0pt}%
\pgfpathmoveto{\pgfqpoint{0.520798in}{0.442177in}}%
\pgfpathlineto{\pgfqpoint{1.239803in}{0.442177in}}%
\pgfusepath{stroke}%
\end{pgfscope}%
\begin{pgfscope}%
\definecolor{textcolor}{rgb}{0.150000,0.150000,0.150000}%
\pgfsetstrokecolor{textcolor}%
\pgfsetfillcolor{textcolor}%
\pgftext[x=0.273151in, y=0.403915in, left, base]{\color{textcolor}\rmfamily\fontsize{8.000000}{9.600000}\selectfont 0\%}%
\end{pgfscope}%
\begin{pgfscope}%
\pgfpathrectangle{\pgfqpoint{0.520798in}{0.442177in}}{\pgfqpoint{0.719005in}{0.871884in}}%
\pgfusepath{clip}%
\pgfsetroundcap%
\pgfsetroundjoin%
\pgfsetlinewidth{0.501875pt}%
\definecolor{currentstroke}{rgb}{0.800000,0.800000,0.800000}%
\pgfsetstrokecolor{currentstroke}%
\pgfsetdash{}{0pt}%
\pgfpathmoveto{\pgfqpoint{0.520798in}{0.660148in}}%
\pgfpathlineto{\pgfqpoint{1.239803in}{0.660148in}}%
\pgfusepath{stroke}%
\end{pgfscope}%
\begin{pgfscope}%
\definecolor{textcolor}{rgb}{0.150000,0.150000,0.150000}%
\pgfsetstrokecolor{textcolor}%
\pgfsetfillcolor{textcolor}%
\pgftext[x=0.214138in, y=0.621886in, left, base]{\color{textcolor}\rmfamily\fontsize{8.000000}{9.600000}\selectfont 25\%}%
\end{pgfscope}%
\begin{pgfscope}%
\pgfpathrectangle{\pgfqpoint{0.520798in}{0.442177in}}{\pgfqpoint{0.719005in}{0.871884in}}%
\pgfusepath{clip}%
\pgfsetroundcap%
\pgfsetroundjoin%
\pgfsetlinewidth{0.501875pt}%
\definecolor{currentstroke}{rgb}{0.800000,0.800000,0.800000}%
\pgfsetstrokecolor{currentstroke}%
\pgfsetdash{}{0pt}%
\pgfpathmoveto{\pgfqpoint{0.520798in}{0.878119in}}%
\pgfpathlineto{\pgfqpoint{1.239803in}{0.878119in}}%
\pgfusepath{stroke}%
\end{pgfscope}%
\begin{pgfscope}%
\definecolor{textcolor}{rgb}{0.150000,0.150000,0.150000}%
\pgfsetstrokecolor{textcolor}%
\pgfsetfillcolor{textcolor}%
\pgftext[x=0.214138in, y=0.839857in, left, base]{\color{textcolor}\rmfamily\fontsize{8.000000}{9.600000}\selectfont 50\%}%
\end{pgfscope}%
\begin{pgfscope}%
\pgfpathrectangle{\pgfqpoint{0.520798in}{0.442177in}}{\pgfqpoint{0.719005in}{0.871884in}}%
\pgfusepath{clip}%
\pgfsetroundcap%
\pgfsetroundjoin%
\pgfsetlinewidth{0.501875pt}%
\definecolor{currentstroke}{rgb}{0.800000,0.800000,0.800000}%
\pgfsetstrokecolor{currentstroke}%
\pgfsetdash{}{0pt}%
\pgfpathmoveto{\pgfqpoint{0.520798in}{1.096090in}}%
\pgfpathlineto{\pgfqpoint{1.239803in}{1.096090in}}%
\pgfusepath{stroke}%
\end{pgfscope}%
\begin{pgfscope}%
\definecolor{textcolor}{rgb}{0.150000,0.150000,0.150000}%
\pgfsetstrokecolor{textcolor}%
\pgfsetfillcolor{textcolor}%
\pgftext[x=0.214138in, y=1.057828in, left, base]{\color{textcolor}\rmfamily\fontsize{8.000000}{9.600000}\selectfont 75\%}%
\end{pgfscope}%
\begin{pgfscope}%
\pgfpathrectangle{\pgfqpoint{0.520798in}{0.442177in}}{\pgfqpoint{0.719005in}{0.871884in}}%
\pgfusepath{clip}%
\pgfsetroundcap%
\pgfsetroundjoin%
\pgfsetlinewidth{0.501875pt}%
\definecolor{currentstroke}{rgb}{0.800000,0.800000,0.800000}%
\pgfsetstrokecolor{currentstroke}%
\pgfsetdash{}{0pt}%
\pgfpathmoveto{\pgfqpoint{0.520798in}{1.314061in}}%
\pgfpathlineto{\pgfqpoint{1.239803in}{1.314061in}}%
\pgfusepath{stroke}%
\end{pgfscope}%
\begin{pgfscope}%
\definecolor{textcolor}{rgb}{0.150000,0.150000,0.150000}%
\pgfsetstrokecolor{textcolor}%
\pgfsetfillcolor{textcolor}%
\pgftext[x=0.155124in, y=1.275799in, left, base]{\color{textcolor}\rmfamily\fontsize{8.000000}{9.600000}\selectfont 100\%}%
\end{pgfscope}%
\begin{pgfscope}%
\definecolor{textcolor}{rgb}{0.150000,0.150000,0.150000}%
\pgfsetstrokecolor{textcolor}%
\pgfsetfillcolor{textcolor}%
\pgftext[x=0.099569in,y=0.878119in,,bottom,rotate=90.000000]{\color{textcolor}\rmfamily\fontsize{10.000000}{12.000000}\selectfont Cert. Acc.}%
\end{pgfscope}%
\begin{pgfscope}%
\pgfsetrectcap%
\pgfsetmiterjoin%
\pgfsetlinewidth{0.752812pt}%
\definecolor{currentstroke}{rgb}{0.700000,0.700000,0.700000}%
\pgfsetstrokecolor{currentstroke}%
\pgfsetdash{}{0pt}%
\pgfpathmoveto{\pgfqpoint{0.520798in}{0.442177in}}%
\pgfpathlineto{\pgfqpoint{0.520798in}{1.314061in}}%
\pgfusepath{stroke}%
\end{pgfscope}%
\begin{pgfscope}%
\pgfsetrectcap%
\pgfsetmiterjoin%
\pgfsetlinewidth{0.752812pt}%
\definecolor{currentstroke}{rgb}{0.700000,0.700000,0.700000}%
\pgfsetstrokecolor{currentstroke}%
\pgfsetdash{}{0pt}%
\pgfpathmoveto{\pgfqpoint{1.239803in}{0.442177in}}%
\pgfpathlineto{\pgfqpoint{1.239803in}{1.314061in}}%
\pgfusepath{stroke}%
\end{pgfscope}%
\begin{pgfscope}%
\pgfsetrectcap%
\pgfsetmiterjoin%
\pgfsetlinewidth{0.752812pt}%
\definecolor{currentstroke}{rgb}{0.700000,0.700000,0.700000}%
\pgfsetstrokecolor{currentstroke}%
\pgfsetdash{}{0pt}%
\pgfpathmoveto{\pgfqpoint{0.520798in}{0.442177in}}%
\pgfpathlineto{\pgfqpoint{1.239803in}{0.442177in}}%
\pgfusepath{stroke}%
\end{pgfscope}%
\begin{pgfscope}%
\pgfsetrectcap%
\pgfsetmiterjoin%
\pgfsetlinewidth{0.752812pt}%
\definecolor{currentstroke}{rgb}{0.700000,0.700000,0.700000}%
\pgfsetstrokecolor{currentstroke}%
\pgfsetdash{}{0pt}%
\pgfpathmoveto{\pgfqpoint{0.520798in}{1.314061in}}%
\pgfpathlineto{\pgfqpoint{1.239803in}{1.314061in}}%
\pgfusepath{stroke}%
\end{pgfscope}%
\begin{pgfscope}%
\pgfpathrectangle{\pgfqpoint{0.520798in}{0.442177in}}{\pgfqpoint{0.719005in}{0.871884in}}%
\pgfusepath{clip}%
\pgfsetroundcap%
\pgfsetroundjoin%
\pgfsetlinewidth{1.003750pt}%
\definecolor{currentstroke}{rgb}{0.003922,0.450980,0.698039}%
\pgfsetstrokecolor{currentstroke}%
\pgfsetdash{}{0pt}%
\pgfpathmoveto{\pgfqpoint{0.520798in}{1.076241in}}%
\pgfpathlineto{\pgfqpoint{0.556748in}{1.076241in}}%
\pgfpathlineto{\pgfqpoint{0.556748in}{1.022815in}}%
\pgfpathlineto{\pgfqpoint{0.628649in}{1.022815in}}%
\pgfpathlineto{\pgfqpoint{0.628649in}{0.941284in}}%
\pgfpathlineto{\pgfqpoint{0.700549in}{0.941284in}}%
\pgfpathlineto{\pgfqpoint{0.700549in}{0.732403in}}%
\pgfpathlineto{\pgfqpoint{0.772450in}{0.732403in}}%
\pgfpathlineto{\pgfqpoint{0.772450in}{0.617667in}}%
\pgfpathlineto{\pgfqpoint{0.844350in}{0.617667in}}%
\pgfpathlineto{\pgfqpoint{0.844350in}{0.606907in}}%
\pgfpathlineto{\pgfqpoint{0.916251in}{0.606907in}}%
\pgfpathlineto{\pgfqpoint{0.916251in}{0.599209in}}%
\pgfpathlineto{\pgfqpoint{0.988151in}{0.599209in}}%
\pgfpathlineto{\pgfqpoint{0.988151in}{0.586223in}}%
\pgfpathlineto{\pgfqpoint{1.060052in}{0.586223in}}%
\pgfpathlineto{\pgfqpoint{1.060052in}{0.537157in}}%
\pgfpathlineto{\pgfqpoint{1.131952in}{0.537157in}}%
\pgfpathlineto{\pgfqpoint{1.131952in}{0.528067in}}%
\pgfpathlineto{\pgfqpoint{1.203853in}{0.528067in}}%
\pgfpathlineto{\pgfqpoint{1.203853in}{0.511464in}}%
\pgfpathlineto{\pgfqpoint{1.241470in}{0.511464in}}%
\pgfusepath{stroke}%
\end{pgfscope}%
\begin{pgfscope}%
\pgfpathrectangle{\pgfqpoint{0.520798in}{0.442177in}}{\pgfqpoint{0.719005in}{0.871884in}}%
\pgfusepath{clip}%
\pgfsetbuttcap%
\pgfsetroundjoin%
\definecolor{currentfill}{rgb}{0.003922,0.450980,0.698039}%
\pgfsetfillcolor{currentfill}%
\pgfsetfillopacity{0.500000}%
\pgfsetlinewidth{0.000000pt}%
\definecolor{currentstroke}{rgb}{0.003922,0.450980,0.698039}%
\pgfsetstrokecolor{currentstroke}%
\pgfsetstrokeopacity{0.500000}%
\pgfsetdash{}{0pt}%
\pgfpathmoveto{\pgfqpoint{0.520798in}{1.084698in}}%
\pgfpathlineto{\pgfqpoint{0.520798in}{1.067783in}}%
\pgfpathlineto{\pgfqpoint{0.556748in}{1.067783in}}%
\pgfpathlineto{\pgfqpoint{0.556748in}{1.009211in}}%
\pgfpathlineto{\pgfqpoint{0.628649in}{1.009211in}}%
\pgfpathlineto{\pgfqpoint{0.628649in}{0.925079in}}%
\pgfpathlineto{\pgfqpoint{0.700549in}{0.925079in}}%
\pgfpathlineto{\pgfqpoint{0.700549in}{0.717807in}}%
\pgfpathlineto{\pgfqpoint{0.772450in}{0.717807in}}%
\pgfpathlineto{\pgfqpoint{0.772450in}{0.603032in}}%
\pgfpathlineto{\pgfqpoint{0.844350in}{0.603032in}}%
\pgfpathlineto{\pgfqpoint{0.844350in}{0.592293in}}%
\pgfpathlineto{\pgfqpoint{0.916251in}{0.592293in}}%
\pgfpathlineto{\pgfqpoint{0.916251in}{0.584199in}}%
\pgfpathlineto{\pgfqpoint{0.988151in}{0.584199in}}%
\pgfpathlineto{\pgfqpoint{0.988151in}{0.572922in}}%
\pgfpathlineto{\pgfqpoint{1.060052in}{0.572922in}}%
\pgfpathlineto{\pgfqpoint{1.060052in}{0.528709in}}%
\pgfpathlineto{\pgfqpoint{1.131952in}{0.528709in}}%
\pgfpathlineto{\pgfqpoint{1.131952in}{0.519275in}}%
\pgfpathlineto{\pgfqpoint{1.203853in}{0.519275in}}%
\pgfpathlineto{\pgfqpoint{1.203853in}{0.505189in}}%
\pgfpathlineto{\pgfqpoint{1.275753in}{0.505189in}}%
\pgfpathlineto{\pgfqpoint{1.275753in}{0.486922in}}%
\pgfpathlineto{\pgfqpoint{1.347654in}{0.486922in}}%
\pgfpathlineto{\pgfqpoint{1.347654in}{0.480412in}}%
\pgfpathlineto{\pgfqpoint{1.419554in}{0.480412in}}%
\pgfpathlineto{\pgfqpoint{1.419554in}{0.442177in}}%
\pgfpathlineto{\pgfqpoint{2.066659in}{0.442177in}}%
\pgfpathlineto{\pgfqpoint{2.066659in}{0.442177in}}%
\pgfpathlineto{\pgfqpoint{2.677813in}{0.442177in}}%
\pgfpathlineto{\pgfqpoint{2.677813in}{0.442177in}}%
\pgfpathlineto{\pgfqpoint{2.677813in}{0.442177in}}%
\pgfpathlineto{\pgfqpoint{2.066659in}{0.442177in}}%
\pgfpathlineto{\pgfqpoint{2.066659in}{0.442177in}}%
\pgfpathlineto{\pgfqpoint{1.419554in}{0.442177in}}%
\pgfpathlineto{\pgfqpoint{1.419554in}{0.494841in}}%
\pgfpathlineto{\pgfqpoint{1.347654in}{0.494841in}}%
\pgfpathlineto{\pgfqpoint{1.347654in}{0.502614in}}%
\pgfpathlineto{\pgfqpoint{1.275753in}{0.502614in}}%
\pgfpathlineto{\pgfqpoint{1.275753in}{0.517738in}}%
\pgfpathlineto{\pgfqpoint{1.203853in}{0.517738in}}%
\pgfpathlineto{\pgfqpoint{1.203853in}{0.536858in}}%
\pgfpathlineto{\pgfqpoint{1.131952in}{0.536858in}}%
\pgfpathlineto{\pgfqpoint{1.131952in}{0.545604in}}%
\pgfpathlineto{\pgfqpoint{1.060052in}{0.545604in}}%
\pgfpathlineto{\pgfqpoint{1.060052in}{0.599525in}}%
\pgfpathlineto{\pgfqpoint{0.988151in}{0.599525in}}%
\pgfpathlineto{\pgfqpoint{0.988151in}{0.614218in}}%
\pgfpathlineto{\pgfqpoint{0.916251in}{0.614218in}}%
\pgfpathlineto{\pgfqpoint{0.916251in}{0.621521in}}%
\pgfpathlineto{\pgfqpoint{0.844350in}{0.621521in}}%
\pgfpathlineto{\pgfqpoint{0.844350in}{0.632302in}}%
\pgfpathlineto{\pgfqpoint{0.772450in}{0.632302in}}%
\pgfpathlineto{\pgfqpoint{0.772450in}{0.746999in}}%
\pgfpathlineto{\pgfqpoint{0.700549in}{0.746999in}}%
\pgfpathlineto{\pgfqpoint{0.700549in}{0.957489in}}%
\pgfpathlineto{\pgfqpoint{0.628649in}{0.957489in}}%
\pgfpathlineto{\pgfqpoint{0.628649in}{1.036418in}}%
\pgfpathlineto{\pgfqpoint{0.556748in}{1.036418in}}%
\pgfpathlineto{\pgfqpoint{0.556748in}{1.084698in}}%
\pgfpathlineto{\pgfqpoint{0.520798in}{1.084698in}}%
\pgfpathlineto{\pgfqpoint{0.520798in}{1.084698in}}%
\pgfpathclose%
\pgfusepath{fill}%
\end{pgfscope}%
\begin{pgfscope}%
\pgfpathrectangle{\pgfqpoint{0.520798in}{0.442177in}}{\pgfqpoint{0.719005in}{0.871884in}}%
\pgfusepath{clip}%
\pgfsetroundcap%
\pgfsetroundjoin%
\pgfsetlinewidth{1.003750pt}%
\definecolor{currentstroke}{rgb}{0.870588,0.560784,0.019608}%
\pgfsetstrokecolor{currentstroke}%
\pgfsetdash{}{0pt}%
\pgfpathmoveto{\pgfqpoint{0.520798in}{1.076241in}}%
\pgfpathlineto{\pgfqpoint{0.556748in}{1.076241in}}%
\pgfpathlineto{\pgfqpoint{0.556748in}{1.022815in}}%
\pgfpathlineto{\pgfqpoint{0.628649in}{1.022815in}}%
\pgfpathlineto{\pgfqpoint{0.628649in}{0.941284in}}%
\pgfpathlineto{\pgfqpoint{0.700549in}{0.941284in}}%
\pgfpathlineto{\pgfqpoint{0.700549in}{0.732403in}}%
\pgfpathlineto{\pgfqpoint{0.772450in}{0.732403in}}%
\pgfpathlineto{\pgfqpoint{0.772450in}{0.617667in}}%
\pgfpathlineto{\pgfqpoint{0.844350in}{0.617667in}}%
\pgfpathlineto{\pgfqpoint{0.844350in}{0.610617in}}%
\pgfpathlineto{\pgfqpoint{0.916251in}{0.610617in}}%
\pgfpathlineto{\pgfqpoint{0.916251in}{0.603475in}}%
\pgfpathlineto{\pgfqpoint{0.988151in}{0.603475in}}%
\pgfpathlineto{\pgfqpoint{0.988151in}{0.586223in}}%
\pgfpathlineto{\pgfqpoint{1.060052in}{0.586223in}}%
\pgfpathlineto{\pgfqpoint{1.060052in}{0.537157in}}%
\pgfpathlineto{\pgfqpoint{1.131952in}{0.537157in}}%
\pgfpathlineto{\pgfqpoint{1.131952in}{0.528067in}}%
\pgfpathlineto{\pgfqpoint{1.203853in}{0.528067in}}%
\pgfpathlineto{\pgfqpoint{1.203853in}{0.511557in}}%
\pgfpathlineto{\pgfqpoint{1.241470in}{0.511557in}}%
\pgfusepath{stroke}%
\end{pgfscope}%
\begin{pgfscope}%
\pgfpathrectangle{\pgfqpoint{0.520798in}{0.442177in}}{\pgfqpoint{0.719005in}{0.871884in}}%
\pgfusepath{clip}%
\pgfsetbuttcap%
\pgfsetroundjoin%
\definecolor{currentfill}{rgb}{0.870588,0.560784,0.019608}%
\pgfsetfillcolor{currentfill}%
\pgfsetfillopacity{0.500000}%
\pgfsetlinewidth{0.000000pt}%
\definecolor{currentstroke}{rgb}{0.870588,0.560784,0.019608}%
\pgfsetstrokecolor{currentstroke}%
\pgfsetstrokeopacity{0.500000}%
\pgfsetdash{}{0pt}%
\pgfpathmoveto{\pgfqpoint{0.520798in}{1.084698in}}%
\pgfpathlineto{\pgfqpoint{0.520798in}{1.067783in}}%
\pgfpathlineto{\pgfqpoint{0.556748in}{1.067783in}}%
\pgfpathlineto{\pgfqpoint{0.556748in}{1.009211in}}%
\pgfpathlineto{\pgfqpoint{0.628649in}{1.009211in}}%
\pgfpathlineto{\pgfqpoint{0.628649in}{0.925079in}}%
\pgfpathlineto{\pgfqpoint{0.700549in}{0.925079in}}%
\pgfpathlineto{\pgfqpoint{0.700549in}{0.717807in}}%
\pgfpathlineto{\pgfqpoint{0.772450in}{0.717807in}}%
\pgfpathlineto{\pgfqpoint{0.772450in}{0.603032in}}%
\pgfpathlineto{\pgfqpoint{0.844350in}{0.603032in}}%
\pgfpathlineto{\pgfqpoint{0.844350in}{0.596419in}}%
\pgfpathlineto{\pgfqpoint{0.916251in}{0.596419in}}%
\pgfpathlineto{\pgfqpoint{0.916251in}{0.587874in}}%
\pgfpathlineto{\pgfqpoint{0.988151in}{0.587874in}}%
\pgfpathlineto{\pgfqpoint{0.988151in}{0.572922in}}%
\pgfpathlineto{\pgfqpoint{1.060052in}{0.572922in}}%
\pgfpathlineto{\pgfqpoint{1.060052in}{0.528709in}}%
\pgfpathlineto{\pgfqpoint{1.131952in}{0.528709in}}%
\pgfpathlineto{\pgfqpoint{1.131952in}{0.519275in}}%
\pgfpathlineto{\pgfqpoint{1.203853in}{0.519275in}}%
\pgfpathlineto{\pgfqpoint{1.203853in}{0.505303in}}%
\pgfpathlineto{\pgfqpoint{1.275753in}{0.505303in}}%
\pgfpathlineto{\pgfqpoint{1.275753in}{0.487009in}}%
\pgfpathlineto{\pgfqpoint{1.347654in}{0.487009in}}%
\pgfpathlineto{\pgfqpoint{1.347654in}{0.480412in}}%
\pgfpathlineto{\pgfqpoint{1.419554in}{0.480412in}}%
\pgfpathlineto{\pgfqpoint{1.419554in}{0.442177in}}%
\pgfpathlineto{\pgfqpoint{2.066659in}{0.442177in}}%
\pgfpathlineto{\pgfqpoint{2.066659in}{0.442177in}}%
\pgfpathlineto{\pgfqpoint{2.677813in}{0.442177in}}%
\pgfpathlineto{\pgfqpoint{2.677813in}{0.442177in}}%
\pgfpathlineto{\pgfqpoint{2.677813in}{0.442177in}}%
\pgfpathlineto{\pgfqpoint{2.066659in}{0.442177in}}%
\pgfpathlineto{\pgfqpoint{2.066659in}{0.442177in}}%
\pgfpathlineto{\pgfqpoint{1.419554in}{0.442177in}}%
\pgfpathlineto{\pgfqpoint{1.419554in}{0.494841in}}%
\pgfpathlineto{\pgfqpoint{1.347654in}{0.494841in}}%
\pgfpathlineto{\pgfqpoint{1.347654in}{0.503270in}}%
\pgfpathlineto{\pgfqpoint{1.275753in}{0.503270in}}%
\pgfpathlineto{\pgfqpoint{1.275753in}{0.517810in}}%
\pgfpathlineto{\pgfqpoint{1.203853in}{0.517810in}}%
\pgfpathlineto{\pgfqpoint{1.203853in}{0.536858in}}%
\pgfpathlineto{\pgfqpoint{1.131952in}{0.536858in}}%
\pgfpathlineto{\pgfqpoint{1.131952in}{0.545604in}}%
\pgfpathlineto{\pgfqpoint{1.060052in}{0.545604in}}%
\pgfpathlineto{\pgfqpoint{1.060052in}{0.599525in}}%
\pgfpathlineto{\pgfqpoint{0.988151in}{0.599525in}}%
\pgfpathlineto{\pgfqpoint{0.988151in}{0.619077in}}%
\pgfpathlineto{\pgfqpoint{0.916251in}{0.619077in}}%
\pgfpathlineto{\pgfqpoint{0.916251in}{0.624816in}}%
\pgfpathlineto{\pgfqpoint{0.844350in}{0.624816in}}%
\pgfpathlineto{\pgfqpoint{0.844350in}{0.632302in}}%
\pgfpathlineto{\pgfqpoint{0.772450in}{0.632302in}}%
\pgfpathlineto{\pgfqpoint{0.772450in}{0.746999in}}%
\pgfpathlineto{\pgfqpoint{0.700549in}{0.746999in}}%
\pgfpathlineto{\pgfqpoint{0.700549in}{0.957489in}}%
\pgfpathlineto{\pgfqpoint{0.628649in}{0.957489in}}%
\pgfpathlineto{\pgfqpoint{0.628649in}{1.036418in}}%
\pgfpathlineto{\pgfqpoint{0.556748in}{1.036418in}}%
\pgfpathlineto{\pgfqpoint{0.556748in}{1.084698in}}%
\pgfpathlineto{\pgfqpoint{0.520798in}{1.084698in}}%
\pgfpathlineto{\pgfqpoint{0.520798in}{1.084698in}}%
\pgfpathclose%
\pgfusepath{fill}%
\end{pgfscope}%
\begin{pgfscope}%
\pgfpathrectangle{\pgfqpoint{0.520798in}{0.442177in}}{\pgfqpoint{0.719005in}{0.871884in}}%
\pgfusepath{clip}%
\pgfsetroundcap%
\pgfsetroundjoin%
\pgfsetlinewidth{1.003750pt}%
\definecolor{currentstroke}{rgb}{0.007843,0.619608,0.450980}%
\pgfsetstrokecolor{currentstroke}%
\pgfsetdash{}{0pt}%
\pgfpathmoveto{\pgfqpoint{0.520798in}{1.076241in}}%
\pgfpathlineto{\pgfqpoint{0.556748in}{1.076241in}}%
\pgfpathlineto{\pgfqpoint{0.556748in}{1.022815in}}%
\pgfpathlineto{\pgfqpoint{0.628649in}{1.022815in}}%
\pgfpathlineto{\pgfqpoint{0.628649in}{0.941284in}}%
\pgfpathlineto{\pgfqpoint{0.700549in}{0.941284in}}%
\pgfpathlineto{\pgfqpoint{0.700549in}{0.732403in}}%
\pgfpathlineto{\pgfqpoint{0.772450in}{0.732403in}}%
\pgfpathlineto{\pgfqpoint{0.772450in}{0.617667in}}%
\pgfpathlineto{\pgfqpoint{0.844350in}{0.617667in}}%
\pgfpathlineto{\pgfqpoint{0.844350in}{0.610617in}}%
\pgfpathlineto{\pgfqpoint{0.916251in}{0.610617in}}%
\pgfpathlineto{\pgfqpoint{0.916251in}{0.603475in}}%
\pgfpathlineto{\pgfqpoint{0.988151in}{0.603475in}}%
\pgfpathlineto{\pgfqpoint{0.988151in}{0.586223in}}%
\pgfpathlineto{\pgfqpoint{1.060052in}{0.586223in}}%
\pgfpathlineto{\pgfqpoint{1.060052in}{0.537157in}}%
\pgfpathlineto{\pgfqpoint{1.131952in}{0.537157in}}%
\pgfpathlineto{\pgfqpoint{1.131952in}{0.528067in}}%
\pgfpathlineto{\pgfqpoint{1.203853in}{0.528067in}}%
\pgfpathlineto{\pgfqpoint{1.203853in}{0.511557in}}%
\pgfpathlineto{\pgfqpoint{1.241470in}{0.511557in}}%
\pgfusepath{stroke}%
\end{pgfscope}%
\begin{pgfscope}%
\pgfpathrectangle{\pgfqpoint{0.520798in}{0.442177in}}{\pgfqpoint{0.719005in}{0.871884in}}%
\pgfusepath{clip}%
\pgfsetbuttcap%
\pgfsetroundjoin%
\definecolor{currentfill}{rgb}{0.007843,0.619608,0.450980}%
\pgfsetfillcolor{currentfill}%
\pgfsetfillopacity{0.500000}%
\pgfsetlinewidth{0.000000pt}%
\definecolor{currentstroke}{rgb}{0.007843,0.619608,0.450980}%
\pgfsetstrokecolor{currentstroke}%
\pgfsetstrokeopacity{0.500000}%
\pgfsetdash{}{0pt}%
\pgfpathmoveto{\pgfqpoint{0.520798in}{1.084698in}}%
\pgfpathlineto{\pgfqpoint{0.520798in}{1.067783in}}%
\pgfpathlineto{\pgfqpoint{0.556748in}{1.067783in}}%
\pgfpathlineto{\pgfqpoint{0.556748in}{1.009211in}}%
\pgfpathlineto{\pgfqpoint{0.628649in}{1.009211in}}%
\pgfpathlineto{\pgfqpoint{0.628649in}{0.925079in}}%
\pgfpathlineto{\pgfqpoint{0.700549in}{0.925079in}}%
\pgfpathlineto{\pgfqpoint{0.700549in}{0.717807in}}%
\pgfpathlineto{\pgfqpoint{0.772450in}{0.717807in}}%
\pgfpathlineto{\pgfqpoint{0.772450in}{0.603032in}}%
\pgfpathlineto{\pgfqpoint{0.844350in}{0.603032in}}%
\pgfpathlineto{\pgfqpoint{0.844350in}{0.596419in}}%
\pgfpathlineto{\pgfqpoint{0.916251in}{0.596419in}}%
\pgfpathlineto{\pgfqpoint{0.916251in}{0.587874in}}%
\pgfpathlineto{\pgfqpoint{0.988151in}{0.587874in}}%
\pgfpathlineto{\pgfqpoint{0.988151in}{0.572922in}}%
\pgfpathlineto{\pgfqpoint{1.060052in}{0.572922in}}%
\pgfpathlineto{\pgfqpoint{1.060052in}{0.528709in}}%
\pgfpathlineto{\pgfqpoint{1.131952in}{0.528709in}}%
\pgfpathlineto{\pgfqpoint{1.131952in}{0.519275in}}%
\pgfpathlineto{\pgfqpoint{1.203853in}{0.519275in}}%
\pgfpathlineto{\pgfqpoint{1.203853in}{0.505303in}}%
\pgfpathlineto{\pgfqpoint{1.275753in}{0.505303in}}%
\pgfpathlineto{\pgfqpoint{1.275753in}{0.487009in}}%
\pgfpathlineto{\pgfqpoint{1.347654in}{0.487009in}}%
\pgfpathlineto{\pgfqpoint{1.347654in}{0.480412in}}%
\pgfpathlineto{\pgfqpoint{1.419554in}{0.480412in}}%
\pgfpathlineto{\pgfqpoint{1.419554in}{0.442177in}}%
\pgfpathlineto{\pgfqpoint{2.066659in}{0.442177in}}%
\pgfpathlineto{\pgfqpoint{2.066659in}{0.442177in}}%
\pgfpathlineto{\pgfqpoint{2.677813in}{0.442177in}}%
\pgfpathlineto{\pgfqpoint{2.677813in}{0.442177in}}%
\pgfpathlineto{\pgfqpoint{2.677813in}{0.442177in}}%
\pgfpathlineto{\pgfqpoint{2.066659in}{0.442177in}}%
\pgfpathlineto{\pgfqpoint{2.066659in}{0.442177in}}%
\pgfpathlineto{\pgfqpoint{1.419554in}{0.442177in}}%
\pgfpathlineto{\pgfqpoint{1.419554in}{0.494841in}}%
\pgfpathlineto{\pgfqpoint{1.347654in}{0.494841in}}%
\pgfpathlineto{\pgfqpoint{1.347654in}{0.503270in}}%
\pgfpathlineto{\pgfqpoint{1.275753in}{0.503270in}}%
\pgfpathlineto{\pgfqpoint{1.275753in}{0.517810in}}%
\pgfpathlineto{\pgfqpoint{1.203853in}{0.517810in}}%
\pgfpathlineto{\pgfqpoint{1.203853in}{0.536858in}}%
\pgfpathlineto{\pgfqpoint{1.131952in}{0.536858in}}%
\pgfpathlineto{\pgfqpoint{1.131952in}{0.545604in}}%
\pgfpathlineto{\pgfqpoint{1.060052in}{0.545604in}}%
\pgfpathlineto{\pgfqpoint{1.060052in}{0.599525in}}%
\pgfpathlineto{\pgfqpoint{0.988151in}{0.599525in}}%
\pgfpathlineto{\pgfqpoint{0.988151in}{0.619077in}}%
\pgfpathlineto{\pgfqpoint{0.916251in}{0.619077in}}%
\pgfpathlineto{\pgfqpoint{0.916251in}{0.624816in}}%
\pgfpathlineto{\pgfqpoint{0.844350in}{0.624816in}}%
\pgfpathlineto{\pgfqpoint{0.844350in}{0.632302in}}%
\pgfpathlineto{\pgfqpoint{0.772450in}{0.632302in}}%
\pgfpathlineto{\pgfqpoint{0.772450in}{0.746999in}}%
\pgfpathlineto{\pgfqpoint{0.700549in}{0.746999in}}%
\pgfpathlineto{\pgfqpoint{0.700549in}{0.957489in}}%
\pgfpathlineto{\pgfqpoint{0.628649in}{0.957489in}}%
\pgfpathlineto{\pgfqpoint{0.628649in}{1.036418in}}%
\pgfpathlineto{\pgfqpoint{0.556748in}{1.036418in}}%
\pgfpathlineto{\pgfqpoint{0.556748in}{1.084698in}}%
\pgfpathlineto{\pgfqpoint{0.520798in}{1.084698in}}%
\pgfpathlineto{\pgfqpoint{0.520798in}{1.084698in}}%
\pgfpathclose%
\pgfusepath{fill}%
\end{pgfscope}%
\begin{pgfscope}%
\definecolor{textcolor}{rgb}{0.150000,0.150000,0.150000}%
\pgfsetstrokecolor{textcolor}%
\pgfsetfillcolor{textcolor}%
\pgftext[x=0.880300in,y=1.397395in,,base]{\color{textcolor}\rmfamily\fontsize{9.000000}{10.800000}\selectfont \(\displaystyle c_X^+=1\)}%
\end{pgfscope}%
\begin{pgfscope}%
\pgfsetbuttcap%
\pgfsetmiterjoin%
\definecolor{currentfill}{rgb}{1.000000,1.000000,1.000000}%
\pgfsetfillcolor{currentfill}%
\pgfsetfillopacity{0.800000}%
\pgfsetlinewidth{1.003750pt}%
\definecolor{currentstroke}{rgb}{0.800000,0.800000,0.800000}%
\pgfsetstrokecolor{currentstroke}%
\pgfsetstrokeopacity{0.800000}%
\pgfsetdash{}{0pt}%
\pgfpathmoveto{\pgfqpoint{0.785843in}{0.638103in}}%
\pgfpathlineto{\pgfqpoint{1.191192in}{0.638103in}}%
\pgfpathlineto{\pgfqpoint{1.191192in}{1.265450in}}%
\pgfpathlineto{\pgfqpoint{0.785843in}{1.265450in}}%
\pgfpathlineto{\pgfqpoint{0.785843in}{0.638103in}}%
\pgfpathclose%
\pgfusepath{stroke,fill}%
\end{pgfscope}%
\begin{pgfscope}%
\definecolor{textcolor}{rgb}{0.150000,0.150000,0.150000}%
\pgfsetstrokecolor{textcolor}%
\pgfsetfillcolor{textcolor}%
\pgftext[x=0.912291in,y=1.113275in,left,base]{\color{textcolor}\rmfamily\fontsize{9.000000}{10.800000}\selectfont \(\displaystyle c_X^-\)}%
\end{pgfscope}%
\begin{pgfscope}%
\pgfsetroundcap%
\pgfsetroundjoin%
\pgfsetlinewidth{1.003750pt}%
\definecolor{currentstroke}{rgb}{0.003922,0.450980,0.698039}%
\pgfsetstrokecolor{currentstroke}%
\pgfsetdash{}{0pt}%
\pgfpathmoveto{\pgfqpoint{0.824732in}{1.001052in}}%
\pgfpathlineto{\pgfqpoint{0.873343in}{1.001052in}}%
\pgfpathlineto{\pgfqpoint{0.873343in}{1.001052in}}%
\pgfpathlineto{\pgfqpoint{0.970565in}{1.001052in}}%
\pgfpathlineto{\pgfqpoint{0.970565in}{1.001052in}}%
\pgfpathlineto{\pgfqpoint{1.019176in}{1.001052in}}%
\pgfusepath{stroke}%
\end{pgfscope}%
\begin{pgfscope}%
\definecolor{textcolor}{rgb}{0.150000,0.150000,0.150000}%
\pgfsetstrokecolor{textcolor}%
\pgfsetfillcolor{textcolor}%
\pgftext[x=1.096954in,y=0.967024in,left,base]{\color{textcolor}\rmfamily\fontsize{7.000000}{8.400000}\selectfont 1}%
\end{pgfscope}%
\begin{pgfscope}%
\pgfsetroundcap%
\pgfsetroundjoin%
\pgfsetlinewidth{1.003750pt}%
\definecolor{currentstroke}{rgb}{0.870588,0.560784,0.019608}%
\pgfsetstrokecolor{currentstroke}%
\pgfsetdash{}{0pt}%
\pgfpathmoveto{\pgfqpoint{0.824732in}{0.865486in}}%
\pgfpathlineto{\pgfqpoint{0.873343in}{0.865486in}}%
\pgfpathlineto{\pgfqpoint{0.873343in}{0.865486in}}%
\pgfpathlineto{\pgfqpoint{0.970565in}{0.865486in}}%
\pgfpathlineto{\pgfqpoint{0.970565in}{0.865486in}}%
\pgfpathlineto{\pgfqpoint{1.019176in}{0.865486in}}%
\pgfusepath{stroke}%
\end{pgfscope}%
\begin{pgfscope}%
\definecolor{textcolor}{rgb}{0.150000,0.150000,0.150000}%
\pgfsetstrokecolor{textcolor}%
\pgfsetfillcolor{textcolor}%
\pgftext[x=1.096954in,y=0.831458in,left,base]{\color{textcolor}\rmfamily\fontsize{7.000000}{8.400000}\selectfont 2}%
\end{pgfscope}%
\begin{pgfscope}%
\pgfsetroundcap%
\pgfsetroundjoin%
\pgfsetlinewidth{1.003750pt}%
\definecolor{currentstroke}{rgb}{0.007843,0.619608,0.450980}%
\pgfsetstrokecolor{currentstroke}%
\pgfsetdash{}{0pt}%
\pgfpathmoveto{\pgfqpoint{0.824732in}{0.729919in}}%
\pgfpathlineto{\pgfqpoint{0.873343in}{0.729919in}}%
\pgfpathlineto{\pgfqpoint{0.873343in}{0.729919in}}%
\pgfpathlineto{\pgfqpoint{0.970565in}{0.729919in}}%
\pgfpathlineto{\pgfqpoint{0.970565in}{0.729919in}}%
\pgfpathlineto{\pgfqpoint{1.019176in}{0.729919in}}%
\pgfusepath{stroke}%
\end{pgfscope}%
\begin{pgfscope}%
\definecolor{textcolor}{rgb}{0.150000,0.150000,0.150000}%
\pgfsetstrokecolor{textcolor}%
\pgfsetfillcolor{textcolor}%
\pgftext[x=1.096954in,y=0.695892in,left,base]{\color{textcolor}\rmfamily\fontsize{7.000000}{8.400000}\selectfont 4}%
\end{pgfscope}%
\end{pgfpicture}%
\makeatother%
\endgroup%

%% file: figures/experiments/graphs/sparse_smoothing/nodes_structure/node_classification-Citeseer-APPNP-hidden=32-p_adj_plus=0.001-p_adj_minus=0.8-p_att_plus=0.0-p_att_minus=0.0-multi_class_cert-B.pgf
\begingroup%
\makeatletter%
\begin{pgfpicture}%
\pgfpathrectangle{\pgfpointorigin}{\pgfqpoint{1.375000in}{1.581250in}}%
\pgfusepath{use as bounding box, clip}%
\begin{pgfscope}%
\pgfsetbuttcap%
\pgfsetmiterjoin%
\definecolor{currentfill}{rgb}{1.000000,1.000000,1.000000}%
\pgfsetfillcolor{currentfill}%
\pgfsetlinewidth{0.000000pt}%
\definecolor{currentstroke}{rgb}{1.000000,1.000000,1.000000}%
\pgfsetstrokecolor{currentstroke}%
\pgfsetdash{}{0pt}%
\pgfpathmoveto{\pgfqpoint{0.000000in}{0.000000in}}%
\pgfpathlineto{\pgfqpoint{1.375000in}{0.000000in}}%
\pgfpathlineto{\pgfqpoint{1.375000in}{1.581250in}}%
\pgfpathlineto{\pgfqpoint{0.000000in}{1.581250in}}%
\pgfpathlineto{\pgfqpoint{0.000000in}{0.000000in}}%
\pgfpathclose%
\pgfusepath{fill}%
\end{pgfscope}%
\begin{pgfscope}%
\pgfsetbuttcap%
\pgfsetmiterjoin%
\definecolor{currentfill}{rgb}{1.000000,1.000000,1.000000}%
\pgfsetfillcolor{currentfill}%
\pgfsetlinewidth{0.000000pt}%
\definecolor{currentstroke}{rgb}{0.000000,0.000000,0.000000}%
\pgfsetstrokecolor{currentstroke}%
\pgfsetstrokeopacity{0.000000}%
\pgfsetdash{}{0pt}%
\pgfpathmoveto{\pgfqpoint{0.520798in}{0.442177in}}%
\pgfpathlineto{\pgfqpoint{1.239803in}{0.442177in}}%
\pgfpathlineto{\pgfqpoint{1.239803in}{1.314061in}}%
\pgfpathlineto{\pgfqpoint{0.520798in}{1.314061in}}%
\pgfpathlineto{\pgfqpoint{0.520798in}{0.442177in}}%
\pgfpathclose%
\pgfusepath{fill}%
\end{pgfscope}%
\begin{pgfscope}%
\pgfpathrectangle{\pgfqpoint{0.520798in}{0.442177in}}{\pgfqpoint{0.719005in}{0.871884in}}%
\pgfusepath{clip}%
\pgfsetroundcap%
\pgfsetroundjoin%
\pgfsetlinewidth{0.501875pt}%
\definecolor{currentstroke}{rgb}{0.800000,0.800000,0.800000}%
\pgfsetstrokecolor{currentstroke}%
\pgfsetdash{}{0pt}%
\pgfpathmoveto{\pgfqpoint{0.520798in}{0.442177in}}%
\pgfpathlineto{\pgfqpoint{0.520798in}{1.314061in}}%
\pgfusepath{stroke}%
\end{pgfscope}%
\begin{pgfscope}%
\definecolor{textcolor}{rgb}{0.150000,0.150000,0.150000}%
\pgfsetstrokecolor{textcolor}%
\pgfsetfillcolor{textcolor}%
\pgftext[x=0.520798in,y=0.351899in,,top]{\color{textcolor}\rmfamily\fontsize{8.000000}{9.600000}\selectfont \(\displaystyle {0}\)}%
\end{pgfscope}%
\begin{pgfscope}%
\pgfpathrectangle{\pgfqpoint{0.520798in}{0.442177in}}{\pgfqpoint{0.719005in}{0.871884in}}%
\pgfusepath{clip}%
\pgfsetroundcap%
\pgfsetroundjoin%
\pgfsetlinewidth{0.501875pt}%
\definecolor{currentstroke}{rgb}{0.800000,0.800000,0.800000}%
\pgfsetstrokecolor{currentstroke}%
\pgfsetdash{}{0pt}%
\pgfpathmoveto{\pgfqpoint{0.880300in}{0.442177in}}%
\pgfpathlineto{\pgfqpoint{0.880300in}{1.314061in}}%
\pgfusepath{stroke}%
\end{pgfscope}%
\begin{pgfscope}%
\definecolor{textcolor}{rgb}{0.150000,0.150000,0.150000}%
\pgfsetstrokecolor{textcolor}%
\pgfsetfillcolor{textcolor}%
\pgftext[x=0.880300in,y=0.351899in,,top]{\color{textcolor}\rmfamily\fontsize{8.000000}{9.600000}\selectfont \(\displaystyle {5}\)}%
\end{pgfscope}%
\begin{pgfscope}%
\pgfpathrectangle{\pgfqpoint{0.520798in}{0.442177in}}{\pgfqpoint{0.719005in}{0.871884in}}%
\pgfusepath{clip}%
\pgfsetroundcap%
\pgfsetroundjoin%
\pgfsetlinewidth{0.501875pt}%
\definecolor{currentstroke}{rgb}{0.800000,0.800000,0.800000}%
\pgfsetstrokecolor{currentstroke}%
\pgfsetdash{}{0pt}%
\pgfpathmoveto{\pgfqpoint{1.239803in}{0.442177in}}%
\pgfpathlineto{\pgfqpoint{1.239803in}{1.314061in}}%
\pgfusepath{stroke}%
\end{pgfscope}%
\begin{pgfscope}%
\definecolor{textcolor}{rgb}{0.150000,0.150000,0.150000}%
\pgfsetstrokecolor{textcolor}%
\pgfsetfillcolor{textcolor}%
\pgftext[x=1.239803in,y=0.351899in,,top]{\color{textcolor}\rmfamily\fontsize{8.000000}{9.600000}\selectfont \(\displaystyle {10}\)}%
\end{pgfscope}%
\begin{pgfscope}%
\definecolor{textcolor}{rgb}{0.150000,0.150000,0.150000}%
\pgfsetstrokecolor{textcolor}%
\pgfsetfillcolor{textcolor}%
\pgftext[x=0.880300in,y=0.198219in,,top]{\color{textcolor}\rmfamily\fontsize{10.000000}{12.000000}\selectfont Edit distance \(\displaystyle \epsilon\)}%
\end{pgfscope}%
\begin{pgfscope}%
\pgfpathrectangle{\pgfqpoint{0.520798in}{0.442177in}}{\pgfqpoint{0.719005in}{0.871884in}}%
\pgfusepath{clip}%
\pgfsetroundcap%
\pgfsetroundjoin%
\pgfsetlinewidth{0.501875pt}%
\definecolor{currentstroke}{rgb}{0.800000,0.800000,0.800000}%
\pgfsetstrokecolor{currentstroke}%
\pgfsetdash{}{0pt}%
\pgfpathmoveto{\pgfqpoint{0.520798in}{0.442177in}}%
\pgfpathlineto{\pgfqpoint{1.239803in}{0.442177in}}%
\pgfusepath{stroke}%
\end{pgfscope}%
\begin{pgfscope}%
\definecolor{textcolor}{rgb}{0.150000,0.150000,0.150000}%
\pgfsetstrokecolor{textcolor}%
\pgfsetfillcolor{textcolor}%
\pgftext[x=0.273151in, y=0.403915in, left, base]{\color{textcolor}\rmfamily\fontsize{8.000000}{9.600000}\selectfont 0\%}%
\end{pgfscope}%
\begin{pgfscope}%
\pgfpathrectangle{\pgfqpoint{0.520798in}{0.442177in}}{\pgfqpoint{0.719005in}{0.871884in}}%
\pgfusepath{clip}%
\pgfsetroundcap%
\pgfsetroundjoin%
\pgfsetlinewidth{0.501875pt}%
\definecolor{currentstroke}{rgb}{0.800000,0.800000,0.800000}%
\pgfsetstrokecolor{currentstroke}%
\pgfsetdash{}{0pt}%
\pgfpathmoveto{\pgfqpoint{0.520798in}{0.660148in}}%
\pgfpathlineto{\pgfqpoint{1.239803in}{0.660148in}}%
\pgfusepath{stroke}%
\end{pgfscope}%
\begin{pgfscope}%
\definecolor{textcolor}{rgb}{0.150000,0.150000,0.150000}%
\pgfsetstrokecolor{textcolor}%
\pgfsetfillcolor{textcolor}%
\pgftext[x=0.214138in, y=0.621886in, left, base]{\color{textcolor}\rmfamily\fontsize{8.000000}{9.600000}\selectfont 25\%}%
\end{pgfscope}%
\begin{pgfscope}%
\pgfpathrectangle{\pgfqpoint{0.520798in}{0.442177in}}{\pgfqpoint{0.719005in}{0.871884in}}%
\pgfusepath{clip}%
\pgfsetroundcap%
\pgfsetroundjoin%
\pgfsetlinewidth{0.501875pt}%
\definecolor{currentstroke}{rgb}{0.800000,0.800000,0.800000}%
\pgfsetstrokecolor{currentstroke}%
\pgfsetdash{}{0pt}%
\pgfpathmoveto{\pgfqpoint{0.520798in}{0.878119in}}%
\pgfpathlineto{\pgfqpoint{1.239803in}{0.878119in}}%
\pgfusepath{stroke}%
\end{pgfscope}%
\begin{pgfscope}%
\definecolor{textcolor}{rgb}{0.150000,0.150000,0.150000}%
\pgfsetstrokecolor{textcolor}%
\pgfsetfillcolor{textcolor}%
\pgftext[x=0.214138in, y=0.839857in, left, base]{\color{textcolor}\rmfamily\fontsize{8.000000}{9.600000}\selectfont 50\%}%
\end{pgfscope}%
\begin{pgfscope}%
\pgfpathrectangle{\pgfqpoint{0.520798in}{0.442177in}}{\pgfqpoint{0.719005in}{0.871884in}}%
\pgfusepath{clip}%
\pgfsetroundcap%
\pgfsetroundjoin%
\pgfsetlinewidth{0.501875pt}%
\definecolor{currentstroke}{rgb}{0.800000,0.800000,0.800000}%
\pgfsetstrokecolor{currentstroke}%
\pgfsetdash{}{0pt}%
\pgfpathmoveto{\pgfqpoint{0.520798in}{1.096090in}}%
\pgfpathlineto{\pgfqpoint{1.239803in}{1.096090in}}%
\pgfusepath{stroke}%
\end{pgfscope}%
\begin{pgfscope}%
\definecolor{textcolor}{rgb}{0.150000,0.150000,0.150000}%
\pgfsetstrokecolor{textcolor}%
\pgfsetfillcolor{textcolor}%
\pgftext[x=0.214138in, y=1.057828in, left, base]{\color{textcolor}\rmfamily\fontsize{8.000000}{9.600000}\selectfont 75\%}%
\end{pgfscope}%
\begin{pgfscope}%
\pgfpathrectangle{\pgfqpoint{0.520798in}{0.442177in}}{\pgfqpoint{0.719005in}{0.871884in}}%
\pgfusepath{clip}%
\pgfsetroundcap%
\pgfsetroundjoin%
\pgfsetlinewidth{0.501875pt}%
\definecolor{currentstroke}{rgb}{0.800000,0.800000,0.800000}%
\pgfsetstrokecolor{currentstroke}%
\pgfsetdash{}{0pt}%
\pgfpathmoveto{\pgfqpoint{0.520798in}{1.314061in}}%
\pgfpathlineto{\pgfqpoint{1.239803in}{1.314061in}}%
\pgfusepath{stroke}%
\end{pgfscope}%
\begin{pgfscope}%
\definecolor{textcolor}{rgb}{0.150000,0.150000,0.150000}%
\pgfsetstrokecolor{textcolor}%
\pgfsetfillcolor{textcolor}%
\pgftext[x=0.155124in, y=1.275799in, left, base]{\color{textcolor}\rmfamily\fontsize{8.000000}{9.600000}\selectfont 100\%}%
\end{pgfscope}%
\begin{pgfscope}%
\definecolor{textcolor}{rgb}{0.150000,0.150000,0.150000}%
\pgfsetstrokecolor{textcolor}%
\pgfsetfillcolor{textcolor}%
\pgftext[x=0.099569in,y=0.878119in,,bottom,rotate=90.000000]{\color{textcolor}\rmfamily\fontsize{10.000000}{12.000000}\selectfont Cert. Acc.}%
\end{pgfscope}%
\begin{pgfscope}%
\pgfsetrectcap%
\pgfsetmiterjoin%
\pgfsetlinewidth{0.752812pt}%
\definecolor{currentstroke}{rgb}{0.700000,0.700000,0.700000}%
\pgfsetstrokecolor{currentstroke}%
\pgfsetdash{}{0pt}%
\pgfpathmoveto{\pgfqpoint{0.520798in}{0.442177in}}%
\pgfpathlineto{\pgfqpoint{0.520798in}{1.314061in}}%
\pgfusepath{stroke}%
\end{pgfscope}%
\begin{pgfscope}%
\pgfsetrectcap%
\pgfsetmiterjoin%
\pgfsetlinewidth{0.752812pt}%
\definecolor{currentstroke}{rgb}{0.700000,0.700000,0.700000}%
\pgfsetstrokecolor{currentstroke}%
\pgfsetdash{}{0pt}%
\pgfpathmoveto{\pgfqpoint{1.239803in}{0.442177in}}%
\pgfpathlineto{\pgfqpoint{1.239803in}{1.314061in}}%
\pgfusepath{stroke}%
\end{pgfscope}%
\begin{pgfscope}%
\pgfsetrectcap%
\pgfsetmiterjoin%
\pgfsetlinewidth{0.752812pt}%
\definecolor{currentstroke}{rgb}{0.700000,0.700000,0.700000}%
\pgfsetstrokecolor{currentstroke}%
\pgfsetdash{}{0pt}%
\pgfpathmoveto{\pgfqpoint{0.520798in}{0.442177in}}%
\pgfpathlineto{\pgfqpoint{1.239803in}{0.442177in}}%
\pgfusepath{stroke}%
\end{pgfscope}%
\begin{pgfscope}%
\pgfsetrectcap%
\pgfsetmiterjoin%
\pgfsetlinewidth{0.752812pt}%
\definecolor{currentstroke}{rgb}{0.700000,0.700000,0.700000}%
\pgfsetstrokecolor{currentstroke}%
\pgfsetdash{}{0pt}%
\pgfpathmoveto{\pgfqpoint{0.520798in}{1.314061in}}%
\pgfpathlineto{\pgfqpoint{1.239803in}{1.314061in}}%
\pgfusepath{stroke}%
\end{pgfscope}%
\begin{pgfscope}%
\pgfpathrectangle{\pgfqpoint{0.520798in}{0.442177in}}{\pgfqpoint{0.719005in}{0.871884in}}%
\pgfusepath{clip}%
\pgfsetroundcap%
\pgfsetroundjoin%
\pgfsetlinewidth{1.003750pt}%
\definecolor{currentstroke}{rgb}{0.003922,0.450980,0.698039}%
\pgfsetstrokecolor{currentstroke}%
\pgfsetdash{}{0pt}%
\pgfpathmoveto{\pgfqpoint{0.520798in}{0.980890in}}%
\pgfpathlineto{\pgfqpoint{0.556748in}{0.980890in}}%
\pgfpathlineto{\pgfqpoint{0.556748in}{0.907522in}}%
\pgfpathlineto{\pgfqpoint{0.628649in}{0.907522in}}%
\pgfpathlineto{\pgfqpoint{0.628649in}{0.790281in}}%
\pgfpathlineto{\pgfqpoint{0.700549in}{0.790281in}}%
\pgfpathlineto{\pgfqpoint{0.700549in}{0.545504in}}%
\pgfpathlineto{\pgfqpoint{0.772450in}{0.545504in}}%
\pgfpathlineto{\pgfqpoint{0.772450in}{0.480113in}}%
\pgfpathlineto{\pgfqpoint{0.844350in}{0.480113in}}%
\pgfpathlineto{\pgfqpoint{0.844350in}{0.470559in}}%
\pgfpathlineto{\pgfqpoint{0.916251in}{0.470559in}}%
\pgfpathlineto{\pgfqpoint{0.916251in}{0.466757in}}%
\pgfpathlineto{\pgfqpoint{0.988151in}{0.466757in}}%
\pgfpathlineto{\pgfqpoint{0.988151in}{0.461377in}}%
\pgfpathlineto{\pgfqpoint{1.060052in}{0.461377in}}%
\pgfpathlineto{\pgfqpoint{1.060052in}{0.446907in}}%
\pgfpathlineto{\pgfqpoint{1.131952in}{0.446907in}}%
\pgfpathlineto{\pgfqpoint{1.131952in}{0.445794in}}%
\pgfpathlineto{\pgfqpoint{1.203853in}{0.445794in}}%
\pgfpathlineto{\pgfqpoint{1.203853in}{0.443939in}}%
\pgfpathlineto{\pgfqpoint{1.241470in}{0.443939in}}%
\pgfusepath{stroke}%
\end{pgfscope}%
\begin{pgfscope}%
\pgfpathrectangle{\pgfqpoint{0.520798in}{0.442177in}}{\pgfqpoint{0.719005in}{0.871884in}}%
\pgfusepath{clip}%
\pgfsetbuttcap%
\pgfsetroundjoin%
\definecolor{currentfill}{rgb}{0.003922,0.450980,0.698039}%
\pgfsetfillcolor{currentfill}%
\pgfsetfillopacity{0.500000}%
\pgfsetlinewidth{0.000000pt}%
\definecolor{currentstroke}{rgb}{0.003922,0.450980,0.698039}%
\pgfsetstrokecolor{currentstroke}%
\pgfsetstrokeopacity{0.500000}%
\pgfsetdash{}{0pt}%
\pgfpathmoveto{\pgfqpoint{0.520798in}{0.995273in}}%
\pgfpathlineto{\pgfqpoint{0.520798in}{0.966507in}}%
\pgfpathlineto{\pgfqpoint{0.556748in}{0.966507in}}%
\pgfpathlineto{\pgfqpoint{0.556748in}{0.888332in}}%
\pgfpathlineto{\pgfqpoint{0.628649in}{0.888332in}}%
\pgfpathlineto{\pgfqpoint{0.628649in}{0.774020in}}%
\pgfpathlineto{\pgfqpoint{0.700549in}{0.774020in}}%
\pgfpathlineto{\pgfqpoint{0.700549in}{0.536089in}}%
\pgfpathlineto{\pgfqpoint{0.772450in}{0.536089in}}%
\pgfpathlineto{\pgfqpoint{0.772450in}{0.473678in}}%
\pgfpathlineto{\pgfqpoint{0.844350in}{0.473678in}}%
\pgfpathlineto{\pgfqpoint{0.844350in}{0.465927in}}%
\pgfpathlineto{\pgfqpoint{0.916251in}{0.465927in}}%
\pgfpathlineto{\pgfqpoint{0.916251in}{0.462318in}}%
\pgfpathlineto{\pgfqpoint{0.988151in}{0.462318in}}%
\pgfpathlineto{\pgfqpoint{0.988151in}{0.457141in}}%
\pgfpathlineto{\pgfqpoint{1.060052in}{0.457141in}}%
\pgfpathlineto{\pgfqpoint{1.060052in}{0.445550in}}%
\pgfpathlineto{\pgfqpoint{1.131952in}{0.445550in}}%
\pgfpathlineto{\pgfqpoint{1.131952in}{0.444231in}}%
\pgfpathlineto{\pgfqpoint{1.203853in}{0.444231in}}%
\pgfpathlineto{\pgfqpoint{1.203853in}{0.442826in}}%
\pgfpathlineto{\pgfqpoint{1.275753in}{0.442826in}}%
\pgfpathlineto{\pgfqpoint{1.275753in}{0.442817in}}%
\pgfpathlineto{\pgfqpoint{1.347654in}{0.442817in}}%
\pgfpathlineto{\pgfqpoint{1.347654in}{0.442557in}}%
\pgfpathlineto{\pgfqpoint{1.419554in}{0.442557in}}%
\pgfpathlineto{\pgfqpoint{1.419554in}{0.442177in}}%
\pgfpathlineto{\pgfqpoint{2.066659in}{0.442177in}}%
\pgfpathlineto{\pgfqpoint{2.066659in}{0.442177in}}%
\pgfpathlineto{\pgfqpoint{2.677813in}{0.442177in}}%
\pgfpathlineto{\pgfqpoint{2.677813in}{0.442177in}}%
\pgfpathlineto{\pgfqpoint{2.677813in}{0.442177in}}%
\pgfpathlineto{\pgfqpoint{2.066659in}{0.442177in}}%
\pgfpathlineto{\pgfqpoint{2.066659in}{0.442177in}}%
\pgfpathlineto{\pgfqpoint{1.419554in}{0.442177in}}%
\pgfpathlineto{\pgfqpoint{1.419554in}{0.443466in}}%
\pgfpathlineto{\pgfqpoint{1.347654in}{0.443466in}}%
\pgfpathlineto{\pgfqpoint{1.347654in}{0.443763in}}%
\pgfpathlineto{\pgfqpoint{1.275753in}{0.443763in}}%
\pgfpathlineto{\pgfqpoint{1.275753in}{0.445052in}}%
\pgfpathlineto{\pgfqpoint{1.203853in}{0.445052in}}%
\pgfpathlineto{\pgfqpoint{1.203853in}{0.447357in}}%
\pgfpathlineto{\pgfqpoint{1.131952in}{0.447357in}}%
\pgfpathlineto{\pgfqpoint{1.131952in}{0.448264in}}%
\pgfpathlineto{\pgfqpoint{1.060052in}{0.448264in}}%
\pgfpathlineto{\pgfqpoint{1.060052in}{0.465613in}}%
\pgfpathlineto{\pgfqpoint{0.988151in}{0.465613in}}%
\pgfpathlineto{\pgfqpoint{0.988151in}{0.471195in}}%
\pgfpathlineto{\pgfqpoint{0.916251in}{0.471195in}}%
\pgfpathlineto{\pgfqpoint{0.916251in}{0.475192in}}%
\pgfpathlineto{\pgfqpoint{0.844350in}{0.475192in}}%
\pgfpathlineto{\pgfqpoint{0.844350in}{0.486549in}}%
\pgfpathlineto{\pgfqpoint{0.772450in}{0.486549in}}%
\pgfpathlineto{\pgfqpoint{0.772450in}{0.554920in}}%
\pgfpathlineto{\pgfqpoint{0.700549in}{0.554920in}}%
\pgfpathlineto{\pgfqpoint{0.700549in}{0.806543in}}%
\pgfpathlineto{\pgfqpoint{0.628649in}{0.806543in}}%
\pgfpathlineto{\pgfqpoint{0.628649in}{0.926712in}}%
\pgfpathlineto{\pgfqpoint{0.556748in}{0.926712in}}%
\pgfpathlineto{\pgfqpoint{0.556748in}{0.995273in}}%
\pgfpathlineto{\pgfqpoint{0.520798in}{0.995273in}}%
\pgfpathlineto{\pgfqpoint{0.520798in}{0.995273in}}%
\pgfpathclose%
\pgfusepath{fill}%
\end{pgfscope}%
\begin{pgfscope}%
\pgfpathrectangle{\pgfqpoint{0.520798in}{0.442177in}}{\pgfqpoint{0.719005in}{0.871884in}}%
\pgfusepath{clip}%
\pgfsetroundcap%
\pgfsetroundjoin%
\pgfsetlinewidth{1.003750pt}%
\definecolor{currentstroke}{rgb}{0.870588,0.560784,0.019608}%
\pgfsetstrokecolor{currentstroke}%
\pgfsetdash{}{0pt}%
\pgfpathmoveto{\pgfqpoint{0.520798in}{0.980890in}}%
\pgfpathlineto{\pgfqpoint{0.556748in}{0.980890in}}%
\pgfpathlineto{\pgfqpoint{0.556748in}{0.923290in}}%
\pgfpathlineto{\pgfqpoint{0.628649in}{0.923290in}}%
\pgfpathlineto{\pgfqpoint{0.628649in}{0.871626in}}%
\pgfpathlineto{\pgfqpoint{0.700549in}{0.871626in}}%
\pgfpathlineto{\pgfqpoint{0.700549in}{0.823765in}}%
\pgfpathlineto{\pgfqpoint{0.772450in}{0.823765in}}%
\pgfpathlineto{\pgfqpoint{0.772450in}{0.777018in}}%
\pgfpathlineto{\pgfqpoint{0.844350in}{0.777018in}}%
\pgfpathlineto{\pgfqpoint{0.844350in}{0.738247in}}%
\pgfpathlineto{\pgfqpoint{0.916251in}{0.738247in}}%
\pgfpathlineto{\pgfqpoint{0.916251in}{0.545504in}}%
\pgfpathlineto{\pgfqpoint{0.988151in}{0.545504in}}%
\pgfpathlineto{\pgfqpoint{0.988151in}{0.533725in}}%
\pgfpathlineto{\pgfqpoint{1.060052in}{0.533725in}}%
\pgfpathlineto{\pgfqpoint{1.060052in}{0.480113in}}%
\pgfpathlineto{\pgfqpoint{1.131952in}{0.480113in}}%
\pgfpathlineto{\pgfqpoint{1.131952in}{0.476217in}}%
\pgfpathlineto{\pgfqpoint{1.203853in}{0.476217in}}%
\pgfpathlineto{\pgfqpoint{1.203853in}{0.469168in}}%
\pgfpathlineto{\pgfqpoint{1.241470in}{0.469168in}}%
\pgfusepath{stroke}%
\end{pgfscope}%
\begin{pgfscope}%
\pgfpathrectangle{\pgfqpoint{0.520798in}{0.442177in}}{\pgfqpoint{0.719005in}{0.871884in}}%
\pgfusepath{clip}%
\pgfsetbuttcap%
\pgfsetroundjoin%
\definecolor{currentfill}{rgb}{0.870588,0.560784,0.019608}%
\pgfsetfillcolor{currentfill}%
\pgfsetfillopacity{0.500000}%
\pgfsetlinewidth{0.000000pt}%
\definecolor{currentstroke}{rgb}{0.870588,0.560784,0.019608}%
\pgfsetstrokecolor{currentstroke}%
\pgfsetstrokeopacity{0.500000}%
\pgfsetdash{}{0pt}%
\pgfpathmoveto{\pgfqpoint{0.520798in}{0.995273in}}%
\pgfpathlineto{\pgfqpoint{0.520798in}{0.966507in}}%
\pgfpathlineto{\pgfqpoint{0.556748in}{0.966507in}}%
\pgfpathlineto{\pgfqpoint{0.556748in}{0.906785in}}%
\pgfpathlineto{\pgfqpoint{0.628649in}{0.906785in}}%
\pgfpathlineto{\pgfqpoint{0.628649in}{0.854728in}}%
\pgfpathlineto{\pgfqpoint{0.700549in}{0.854728in}}%
\pgfpathlineto{\pgfqpoint{0.700549in}{0.807544in}}%
\pgfpathlineto{\pgfqpoint{0.772450in}{0.807544in}}%
\pgfpathlineto{\pgfqpoint{0.772450in}{0.760974in}}%
\pgfpathlineto{\pgfqpoint{0.844350in}{0.760974in}}%
\pgfpathlineto{\pgfqpoint{0.844350in}{0.723335in}}%
\pgfpathlineto{\pgfqpoint{0.916251in}{0.723335in}}%
\pgfpathlineto{\pgfqpoint{0.916251in}{0.536089in}}%
\pgfpathlineto{\pgfqpoint{0.988151in}{0.536089in}}%
\pgfpathlineto{\pgfqpoint{0.988151in}{0.524854in}}%
\pgfpathlineto{\pgfqpoint{1.060052in}{0.524854in}}%
\pgfpathlineto{\pgfqpoint{1.060052in}{0.473678in}}%
\pgfpathlineto{\pgfqpoint{1.131952in}{0.473678in}}%
\pgfpathlineto{\pgfqpoint{1.131952in}{0.470812in}}%
\pgfpathlineto{\pgfqpoint{1.203853in}{0.470812in}}%
\pgfpathlineto{\pgfqpoint{1.203853in}{0.464527in}}%
\pgfpathlineto{\pgfqpoint{1.275753in}{0.464527in}}%
\pgfpathlineto{\pgfqpoint{1.275753in}{0.462511in}}%
\pgfpathlineto{\pgfqpoint{1.347654in}{0.462511in}}%
\pgfpathlineto{\pgfqpoint{1.347654in}{0.460218in}}%
\pgfpathlineto{\pgfqpoint{1.419554in}{0.460218in}}%
\pgfpathlineto{\pgfqpoint{1.419554in}{0.456643in}}%
\pgfpathlineto{\pgfqpoint{1.491455in}{0.456643in}}%
\pgfpathlineto{\pgfqpoint{1.491455in}{0.455335in}}%
\pgfpathlineto{\pgfqpoint{1.563355in}{0.455335in}}%
\pgfpathlineto{\pgfqpoint{1.563355in}{0.454077in}}%
\pgfpathlineto{\pgfqpoint{1.635256in}{0.454077in}}%
\pgfpathlineto{\pgfqpoint{1.635256in}{0.445550in}}%
\pgfpathlineto{\pgfqpoint{1.707156in}{0.445550in}}%
\pgfpathlineto{\pgfqpoint{1.707156in}{0.445205in}}%
\pgfpathlineto{\pgfqpoint{1.779057in}{0.445205in}}%
\pgfpathlineto{\pgfqpoint{1.779057in}{0.444096in}}%
\pgfpathlineto{\pgfqpoint{1.850957in}{0.444096in}}%
\pgfpathlineto{\pgfqpoint{1.850957in}{0.443653in}}%
\pgfpathlineto{\pgfqpoint{1.922858in}{0.443653in}}%
\pgfpathlineto{\pgfqpoint{1.922858in}{0.442633in}}%
\pgfpathlineto{\pgfqpoint{1.994758in}{0.442633in}}%
\pgfpathlineto{\pgfqpoint{1.994758in}{0.442677in}}%
\pgfpathlineto{\pgfqpoint{2.066659in}{0.442677in}}%
\pgfpathlineto{\pgfqpoint{2.066659in}{0.442557in}}%
\pgfpathlineto{\pgfqpoint{2.138559in}{0.442557in}}%
\pgfpathlineto{\pgfqpoint{2.138559in}{0.442548in}}%
\pgfpathlineto{\pgfqpoint{2.210460in}{0.442548in}}%
\pgfpathlineto{\pgfqpoint{2.210460in}{0.442228in}}%
\pgfpathlineto{\pgfqpoint{2.282360in}{0.442228in}}%
\pgfpathlineto{\pgfqpoint{2.282360in}{0.442177in}}%
\pgfpathlineto{\pgfqpoint{2.498062in}{0.442177in}}%
\pgfpathlineto{\pgfqpoint{2.498062in}{0.442177in}}%
\pgfpathlineto{\pgfqpoint{2.677813in}{0.442177in}}%
\pgfpathlineto{\pgfqpoint{2.677813in}{0.442177in}}%
\pgfpathlineto{\pgfqpoint{2.677813in}{0.442177in}}%
\pgfpathlineto{\pgfqpoint{2.498062in}{0.442177in}}%
\pgfpathlineto{\pgfqpoint{2.498062in}{0.442177in}}%
\pgfpathlineto{\pgfqpoint{2.282360in}{0.442177in}}%
\pgfpathlineto{\pgfqpoint{2.282360in}{0.442682in}}%
\pgfpathlineto{\pgfqpoint{2.210460in}{0.442682in}}%
\pgfpathlineto{\pgfqpoint{2.210460in}{0.443290in}}%
\pgfpathlineto{\pgfqpoint{2.138559in}{0.443290in}}%
\pgfpathlineto{\pgfqpoint{2.138559in}{0.443466in}}%
\pgfpathlineto{\pgfqpoint{2.066659in}{0.443466in}}%
\pgfpathlineto{\pgfqpoint{2.066659in}{0.444273in}}%
\pgfpathlineto{\pgfqpoint{1.994758in}{0.444273in}}%
\pgfpathlineto{\pgfqpoint{1.994758in}{0.444874in}}%
\pgfpathlineto{\pgfqpoint{1.922858in}{0.444874in}}%
\pgfpathlineto{\pgfqpoint{1.922858in}{0.445895in}}%
\pgfpathlineto{\pgfqpoint{1.850957in}{0.445895in}}%
\pgfpathlineto{\pgfqpoint{1.850957in}{0.447121in}}%
\pgfpathlineto{\pgfqpoint{1.779057in}{0.447121in}}%
\pgfpathlineto{\pgfqpoint{1.779057in}{0.447867in}}%
\pgfpathlineto{\pgfqpoint{1.707156in}{0.447867in}}%
\pgfpathlineto{\pgfqpoint{1.707156in}{0.448264in}}%
\pgfpathlineto{\pgfqpoint{1.635256in}{0.448264in}}%
\pgfpathlineto{\pgfqpoint{1.635256in}{0.460329in}}%
\pgfpathlineto{\pgfqpoint{1.563355in}{0.460329in}}%
\pgfpathlineto{\pgfqpoint{1.563355in}{0.462039in}}%
\pgfpathlineto{\pgfqpoint{1.491455in}{0.462039in}}%
\pgfpathlineto{\pgfqpoint{1.491455in}{0.465925in}}%
\pgfpathlineto{\pgfqpoint{1.419554in}{0.465925in}}%
\pgfpathlineto{\pgfqpoint{1.419554in}{0.468658in}}%
\pgfpathlineto{\pgfqpoint{1.347654in}{0.468658in}}%
\pgfpathlineto{\pgfqpoint{1.347654in}{0.470631in}}%
\pgfpathlineto{\pgfqpoint{1.275753in}{0.470631in}}%
\pgfpathlineto{\pgfqpoint{1.275753in}{0.473810in}}%
\pgfpathlineto{\pgfqpoint{1.203853in}{0.473810in}}%
\pgfpathlineto{\pgfqpoint{1.203853in}{0.481623in}}%
\pgfpathlineto{\pgfqpoint{1.131952in}{0.481623in}}%
\pgfpathlineto{\pgfqpoint{1.131952in}{0.486549in}}%
\pgfpathlineto{\pgfqpoint{1.060052in}{0.486549in}}%
\pgfpathlineto{\pgfqpoint{1.060052in}{0.542595in}}%
\pgfpathlineto{\pgfqpoint{0.988151in}{0.542595in}}%
\pgfpathlineto{\pgfqpoint{0.988151in}{0.554920in}}%
\pgfpathlineto{\pgfqpoint{0.916251in}{0.554920in}}%
\pgfpathlineto{\pgfqpoint{0.916251in}{0.753158in}}%
\pgfpathlineto{\pgfqpoint{0.844350in}{0.753158in}}%
\pgfpathlineto{\pgfqpoint{0.844350in}{0.793062in}}%
\pgfpathlineto{\pgfqpoint{0.772450in}{0.793062in}}%
\pgfpathlineto{\pgfqpoint{0.772450in}{0.839986in}}%
\pgfpathlineto{\pgfqpoint{0.700549in}{0.839986in}}%
\pgfpathlineto{\pgfqpoint{0.700549in}{0.888524in}}%
\pgfpathlineto{\pgfqpoint{0.628649in}{0.888524in}}%
\pgfpathlineto{\pgfqpoint{0.628649in}{0.939796in}}%
\pgfpathlineto{\pgfqpoint{0.556748in}{0.939796in}}%
\pgfpathlineto{\pgfqpoint{0.556748in}{0.995273in}}%
\pgfpathlineto{\pgfqpoint{0.520798in}{0.995273in}}%
\pgfpathlineto{\pgfqpoint{0.520798in}{0.995273in}}%
\pgfpathclose%
\pgfusepath{fill}%
\end{pgfscope}%
\begin{pgfscope}%
\pgfpathrectangle{\pgfqpoint{0.520798in}{0.442177in}}{\pgfqpoint{0.719005in}{0.871884in}}%
\pgfusepath{clip}%
\pgfsetroundcap%
\pgfsetroundjoin%
\pgfsetlinewidth{1.003750pt}%
\definecolor{currentstroke}{rgb}{0.007843,0.619608,0.450980}%
\pgfsetstrokecolor{currentstroke}%
\pgfsetdash{}{0pt}%
\pgfpathmoveto{\pgfqpoint{0.520798in}{0.980890in}}%
\pgfpathlineto{\pgfqpoint{0.556748in}{0.980890in}}%
\pgfpathlineto{\pgfqpoint{0.556748in}{0.923290in}}%
\pgfpathlineto{\pgfqpoint{0.628649in}{0.923290in}}%
\pgfpathlineto{\pgfqpoint{0.628649in}{0.871626in}}%
\pgfpathlineto{\pgfqpoint{0.700549in}{0.871626in}}%
\pgfpathlineto{\pgfqpoint{0.700549in}{0.823765in}}%
\pgfpathlineto{\pgfqpoint{0.772450in}{0.823765in}}%
\pgfpathlineto{\pgfqpoint{0.772450in}{0.777018in}}%
\pgfpathlineto{\pgfqpoint{0.844350in}{0.777018in}}%
\pgfpathlineto{\pgfqpoint{0.844350in}{0.738247in}}%
\pgfpathlineto{\pgfqpoint{0.916251in}{0.738247in}}%
\pgfpathlineto{\pgfqpoint{0.916251in}{0.703557in}}%
\pgfpathlineto{\pgfqpoint{0.988151in}{0.703557in}}%
\pgfpathlineto{\pgfqpoint{0.988151in}{0.673690in}}%
\pgfpathlineto{\pgfqpoint{1.060052in}{0.673690in}}%
\pgfpathlineto{\pgfqpoint{1.060052in}{0.647348in}}%
\pgfpathlineto{\pgfqpoint{1.131952in}{0.647348in}}%
\pgfpathlineto{\pgfqpoint{1.131952in}{0.625736in}}%
\pgfpathlineto{\pgfqpoint{1.203853in}{0.625736in}}%
\pgfpathlineto{\pgfqpoint{1.203853in}{0.603104in}}%
\pgfpathlineto{\pgfqpoint{1.241470in}{0.603104in}}%
\pgfusepath{stroke}%
\end{pgfscope}%
\begin{pgfscope}%
\pgfpathrectangle{\pgfqpoint{0.520798in}{0.442177in}}{\pgfqpoint{0.719005in}{0.871884in}}%
\pgfusepath{clip}%
\pgfsetbuttcap%
\pgfsetroundjoin%
\definecolor{currentfill}{rgb}{0.007843,0.619608,0.450980}%
\pgfsetfillcolor{currentfill}%
\pgfsetfillopacity{0.500000}%
\pgfsetlinewidth{0.000000pt}%
\definecolor{currentstroke}{rgb}{0.007843,0.619608,0.450980}%
\pgfsetstrokecolor{currentstroke}%
\pgfsetstrokeopacity{0.500000}%
\pgfsetdash{}{0pt}%
\pgfpathmoveto{\pgfqpoint{0.520798in}{0.995273in}}%
\pgfpathlineto{\pgfqpoint{0.520798in}{0.966507in}}%
\pgfpathlineto{\pgfqpoint{0.556748in}{0.966507in}}%
\pgfpathlineto{\pgfqpoint{0.556748in}{0.906785in}}%
\pgfpathlineto{\pgfqpoint{0.628649in}{0.906785in}}%
\pgfpathlineto{\pgfqpoint{0.628649in}{0.854728in}}%
\pgfpathlineto{\pgfqpoint{0.700549in}{0.854728in}}%
\pgfpathlineto{\pgfqpoint{0.700549in}{0.807544in}}%
\pgfpathlineto{\pgfqpoint{0.772450in}{0.807544in}}%
\pgfpathlineto{\pgfqpoint{0.772450in}{0.760974in}}%
\pgfpathlineto{\pgfqpoint{0.844350in}{0.760974in}}%
\pgfpathlineto{\pgfqpoint{0.844350in}{0.723335in}}%
\pgfpathlineto{\pgfqpoint{0.916251in}{0.723335in}}%
\pgfpathlineto{\pgfqpoint{0.916251in}{0.689913in}}%
\pgfpathlineto{\pgfqpoint{0.988151in}{0.689913in}}%
\pgfpathlineto{\pgfqpoint{0.988151in}{0.660363in}}%
\pgfpathlineto{\pgfqpoint{1.060052in}{0.660363in}}%
\pgfpathlineto{\pgfqpoint{1.060052in}{0.634881in}}%
\pgfpathlineto{\pgfqpoint{1.131952in}{0.634881in}}%
\pgfpathlineto{\pgfqpoint{1.131952in}{0.613741in}}%
\pgfpathlineto{\pgfqpoint{1.203853in}{0.613741in}}%
\pgfpathlineto{\pgfqpoint{1.203853in}{0.589469in}}%
\pgfpathlineto{\pgfqpoint{1.275753in}{0.589469in}}%
\pgfpathlineto{\pgfqpoint{1.275753in}{0.568645in}}%
\pgfpathlineto{\pgfqpoint{1.347654in}{0.568645in}}%
\pgfpathlineto{\pgfqpoint{1.347654in}{0.536089in}}%
\pgfpathlineto{\pgfqpoint{1.419554in}{0.536089in}}%
\pgfpathlineto{\pgfqpoint{1.419554in}{0.524854in}}%
\pgfpathlineto{\pgfqpoint{1.491455in}{0.524854in}}%
\pgfpathlineto{\pgfqpoint{1.491455in}{0.515215in}}%
\pgfpathlineto{\pgfqpoint{1.563355in}{0.515215in}}%
\pgfpathlineto{\pgfqpoint{1.563355in}{0.506617in}}%
\pgfpathlineto{\pgfqpoint{1.635256in}{0.506617in}}%
\pgfpathlineto{\pgfqpoint{1.635256in}{0.473678in}}%
\pgfpathlineto{\pgfqpoint{1.707156in}{0.473678in}}%
\pgfpathlineto{\pgfqpoint{1.707156in}{0.470812in}}%
\pgfpathlineto{\pgfqpoint{1.779057in}{0.470812in}}%
\pgfpathlineto{\pgfqpoint{1.779057in}{0.466795in}}%
\pgfpathlineto{\pgfqpoint{1.850957in}{0.466795in}}%
\pgfpathlineto{\pgfqpoint{1.850957in}{0.464416in}}%
\pgfpathlineto{\pgfqpoint{1.922858in}{0.464416in}}%
\pgfpathlineto{\pgfqpoint{1.922858in}{0.461891in}}%
\pgfpathlineto{\pgfqpoint{1.994758in}{0.461891in}}%
\pgfpathlineto{\pgfqpoint{1.994758in}{0.458735in}}%
\pgfpathlineto{\pgfqpoint{2.066659in}{0.458735in}}%
\pgfpathlineto{\pgfqpoint{2.066659in}{0.457095in}}%
\pgfpathlineto{\pgfqpoint{2.138559in}{0.457095in}}%
\pgfpathlineto{\pgfqpoint{2.138559in}{0.456340in}}%
\pgfpathlineto{\pgfqpoint{2.210460in}{0.456340in}}%
\pgfpathlineto{\pgfqpoint{2.210460in}{0.454459in}}%
\pgfpathlineto{\pgfqpoint{2.282360in}{0.454459in}}%
\pgfpathlineto{\pgfqpoint{2.282360in}{0.452609in}}%
\pgfpathlineto{\pgfqpoint{2.354261in}{0.452609in}}%
\pgfpathlineto{\pgfqpoint{2.354261in}{0.450707in}}%
\pgfpathlineto{\pgfqpoint{2.426161in}{0.450707in}}%
\pgfpathlineto{\pgfqpoint{2.426161in}{0.449424in}}%
\pgfpathlineto{\pgfqpoint{2.498062in}{0.449424in}}%
\pgfpathlineto{\pgfqpoint{2.498062in}{0.449118in}}%
\pgfpathlineto{\pgfqpoint{2.569962in}{0.449118in}}%
\pgfpathlineto{\pgfqpoint{2.569962in}{0.448282in}}%
\pgfpathlineto{\pgfqpoint{2.641863in}{0.448282in}}%
\pgfpathlineto{\pgfqpoint{2.641863in}{0.447825in}}%
\pgfpathlineto{\pgfqpoint{2.677813in}{0.447825in}}%
\pgfpathlineto{\pgfqpoint{2.677813in}{0.452112in}}%
\pgfpathlineto{\pgfqpoint{2.677813in}{0.452112in}}%
\pgfpathlineto{\pgfqpoint{2.641863in}{0.452112in}}%
\pgfpathlineto{\pgfqpoint{2.641863in}{0.452953in}}%
\pgfpathlineto{\pgfqpoint{2.569962in}{0.452953in}}%
\pgfpathlineto{\pgfqpoint{2.569962in}{0.454714in}}%
\pgfpathlineto{\pgfqpoint{2.498062in}{0.454714in}}%
\pgfpathlineto{\pgfqpoint{2.498062in}{0.455892in}}%
\pgfpathlineto{\pgfqpoint{2.426161in}{0.455892in}}%
\pgfpathlineto{\pgfqpoint{2.426161in}{0.457949in}}%
\pgfpathlineto{\pgfqpoint{2.354261in}{0.457949in}}%
\pgfpathlineto{\pgfqpoint{2.354261in}{0.459385in}}%
\pgfpathlineto{\pgfqpoint{2.282360in}{0.459385in}}%
\pgfpathlineto{\pgfqpoint{2.282360in}{0.460874in}}%
\pgfpathlineto{\pgfqpoint{2.210460in}{0.460874in}}%
\pgfpathlineto{\pgfqpoint{2.210460in}{0.462889in}}%
\pgfpathlineto{\pgfqpoint{2.138559in}{0.462889in}}%
\pgfpathlineto{\pgfqpoint{2.138559in}{0.465287in}}%
\pgfpathlineto{\pgfqpoint{2.066659in}{0.465287in}}%
\pgfpathlineto{\pgfqpoint{2.066659in}{0.468285in}}%
\pgfpathlineto{\pgfqpoint{1.994758in}{0.468285in}}%
\pgfpathlineto{\pgfqpoint{1.994758in}{0.471251in}}%
\pgfpathlineto{\pgfqpoint{1.922858in}{0.471251in}}%
\pgfpathlineto{\pgfqpoint{1.922858in}{0.473549in}}%
\pgfpathlineto{\pgfqpoint{1.850957in}{0.473549in}}%
\pgfpathlineto{\pgfqpoint{1.850957in}{0.477848in}}%
\pgfpathlineto{\pgfqpoint{1.779057in}{0.477848in}}%
\pgfpathlineto{\pgfqpoint{1.779057in}{0.481623in}}%
\pgfpathlineto{\pgfqpoint{1.707156in}{0.481623in}}%
\pgfpathlineto{\pgfqpoint{1.707156in}{0.486549in}}%
\pgfpathlineto{\pgfqpoint{1.635256in}{0.486549in}}%
\pgfpathlineto{\pgfqpoint{1.635256in}{0.520762in}}%
\pgfpathlineto{\pgfqpoint{1.563355in}{0.520762in}}%
\pgfpathlineto{\pgfqpoint{1.563355in}{0.533498in}}%
\pgfpathlineto{\pgfqpoint{1.491455in}{0.533498in}}%
\pgfpathlineto{\pgfqpoint{1.491455in}{0.542595in}}%
\pgfpathlineto{\pgfqpoint{1.419554in}{0.542595in}}%
\pgfpathlineto{\pgfqpoint{1.419554in}{0.554920in}}%
\pgfpathlineto{\pgfqpoint{1.347654in}{0.554920in}}%
\pgfpathlineto{\pgfqpoint{1.347654in}{0.594711in}}%
\pgfpathlineto{\pgfqpoint{1.275753in}{0.594711in}}%
\pgfpathlineto{\pgfqpoint{1.275753in}{0.616740in}}%
\pgfpathlineto{\pgfqpoint{1.203853in}{0.616740in}}%
\pgfpathlineto{\pgfqpoint{1.203853in}{0.637731in}}%
\pgfpathlineto{\pgfqpoint{1.131952in}{0.637731in}}%
\pgfpathlineto{\pgfqpoint{1.131952in}{0.659815in}}%
\pgfpathlineto{\pgfqpoint{1.060052in}{0.659815in}}%
\pgfpathlineto{\pgfqpoint{1.060052in}{0.687017in}}%
\pgfpathlineto{\pgfqpoint{0.988151in}{0.687017in}}%
\pgfpathlineto{\pgfqpoint{0.988151in}{0.717200in}}%
\pgfpathlineto{\pgfqpoint{0.916251in}{0.717200in}}%
\pgfpathlineto{\pgfqpoint{0.916251in}{0.753158in}}%
\pgfpathlineto{\pgfqpoint{0.844350in}{0.753158in}}%
\pgfpathlineto{\pgfqpoint{0.844350in}{0.793062in}}%
\pgfpathlineto{\pgfqpoint{0.772450in}{0.793062in}}%
\pgfpathlineto{\pgfqpoint{0.772450in}{0.839986in}}%
\pgfpathlineto{\pgfqpoint{0.700549in}{0.839986in}}%
\pgfpathlineto{\pgfqpoint{0.700549in}{0.888524in}}%
\pgfpathlineto{\pgfqpoint{0.628649in}{0.888524in}}%
\pgfpathlineto{\pgfqpoint{0.628649in}{0.939796in}}%
\pgfpathlineto{\pgfqpoint{0.556748in}{0.939796in}}%
\pgfpathlineto{\pgfqpoint{0.556748in}{0.995273in}}%
\pgfpathlineto{\pgfqpoint{0.520798in}{0.995273in}}%
\pgfpathlineto{\pgfqpoint{0.520798in}{0.995273in}}%
\pgfpathclose%
\pgfusepath{fill}%
\end{pgfscope}%
\begin{pgfscope}%
\definecolor{textcolor}{rgb}{0.150000,0.150000,0.150000}%
\pgfsetstrokecolor{textcolor}%
\pgfsetfillcolor{textcolor}%
\pgftext[x=0.880300in,y=1.397395in,,base]{\color{textcolor}\rmfamily\fontsize{9.000000}{10.800000}\selectfont \(\displaystyle c_A^-=1\)}%
\end{pgfscope}%
\begin{pgfscope}%
\pgfsetbuttcap%
\pgfsetmiterjoin%
\definecolor{currentfill}{rgb}{1.000000,1.000000,1.000000}%
\pgfsetfillcolor{currentfill}%
\pgfsetfillopacity{0.800000}%
\pgfsetlinewidth{1.003750pt}%
\definecolor{currentstroke}{rgb}{0.800000,0.800000,0.800000}%
\pgfsetstrokecolor{currentstroke}%
\pgfsetstrokeopacity{0.800000}%
\pgfsetdash{}{0pt}%
\pgfpathmoveto{\pgfqpoint{0.785843in}{0.638103in}}%
\pgfpathlineto{\pgfqpoint{1.191192in}{0.638103in}}%
\pgfpathlineto{\pgfqpoint{1.191192in}{1.265450in}}%
\pgfpathlineto{\pgfqpoint{0.785843in}{1.265450in}}%
\pgfpathlineto{\pgfqpoint{0.785843in}{0.638103in}}%
\pgfpathclose%
\pgfusepath{stroke,fill}%
\end{pgfscope}%
\begin{pgfscope}%
\definecolor{textcolor}{rgb}{0.150000,0.150000,0.150000}%
\pgfsetstrokecolor{textcolor}%
\pgfsetfillcolor{textcolor}%
\pgftext[x=0.918271in,y=1.113275in,left,base]{\color{textcolor}\rmfamily\fontsize{9.000000}{10.800000}\selectfont \(\displaystyle c_A^+\)}%
\end{pgfscope}%
\begin{pgfscope}%
\pgfsetroundcap%
\pgfsetroundjoin%
\pgfsetlinewidth{1.003750pt}%
\definecolor{currentstroke}{rgb}{0.003922,0.450980,0.698039}%
\pgfsetstrokecolor{currentstroke}%
\pgfsetdash{}{0pt}%
\pgfpathmoveto{\pgfqpoint{0.824732in}{1.001052in}}%
\pgfpathlineto{\pgfqpoint{0.873343in}{1.001052in}}%
\pgfpathlineto{\pgfqpoint{0.873343in}{1.001052in}}%
\pgfpathlineto{\pgfqpoint{0.970565in}{1.001052in}}%
\pgfpathlineto{\pgfqpoint{0.970565in}{1.001052in}}%
\pgfpathlineto{\pgfqpoint{1.019176in}{1.001052in}}%
\pgfusepath{stroke}%
\end{pgfscope}%
\begin{pgfscope}%
\definecolor{textcolor}{rgb}{0.150000,0.150000,0.150000}%
\pgfsetstrokecolor{textcolor}%
\pgfsetfillcolor{textcolor}%
\pgftext[x=1.096954in,y=0.967024in,left,base]{\color{textcolor}\rmfamily\fontsize{7.000000}{8.400000}\selectfont 1}%
\end{pgfscope}%
\begin{pgfscope}%
\pgfsetroundcap%
\pgfsetroundjoin%
\pgfsetlinewidth{1.003750pt}%
\definecolor{currentstroke}{rgb}{0.870588,0.560784,0.019608}%
\pgfsetstrokecolor{currentstroke}%
\pgfsetdash{}{0pt}%
\pgfpathmoveto{\pgfqpoint{0.824732in}{0.865486in}}%
\pgfpathlineto{\pgfqpoint{0.873343in}{0.865486in}}%
\pgfpathlineto{\pgfqpoint{0.873343in}{0.865486in}}%
\pgfpathlineto{\pgfqpoint{0.970565in}{0.865486in}}%
\pgfpathlineto{\pgfqpoint{0.970565in}{0.865486in}}%
\pgfpathlineto{\pgfqpoint{1.019176in}{0.865486in}}%
\pgfusepath{stroke}%
\end{pgfscope}%
\begin{pgfscope}%
\definecolor{textcolor}{rgb}{0.150000,0.150000,0.150000}%
\pgfsetstrokecolor{textcolor}%
\pgfsetfillcolor{textcolor}%
\pgftext[x=1.096954in,y=0.831458in,left,base]{\color{textcolor}\rmfamily\fontsize{7.000000}{8.400000}\selectfont 2}%
\end{pgfscope}%
\begin{pgfscope}%
\pgfsetroundcap%
\pgfsetroundjoin%
\pgfsetlinewidth{1.003750pt}%
\definecolor{currentstroke}{rgb}{0.007843,0.619608,0.450980}%
\pgfsetstrokecolor{currentstroke}%
\pgfsetdash{}{0pt}%
\pgfpathmoveto{\pgfqpoint{0.824732in}{0.729919in}}%
\pgfpathlineto{\pgfqpoint{0.873343in}{0.729919in}}%
\pgfpathlineto{\pgfqpoint{0.873343in}{0.729919in}}%
\pgfpathlineto{\pgfqpoint{0.970565in}{0.729919in}}%
\pgfpathlineto{\pgfqpoint{0.970565in}{0.729919in}}%
\pgfpathlineto{\pgfqpoint{1.019176in}{0.729919in}}%
\pgfusepath{stroke}%
\end{pgfscope}%
\begin{pgfscope}%
\definecolor{textcolor}{rgb}{0.150000,0.150000,0.150000}%
\pgfsetstrokecolor{textcolor}%
\pgfsetfillcolor{textcolor}%
\pgftext[x=1.096954in,y=0.695892in,left,base]{\color{textcolor}\rmfamily\fontsize{7.000000}{8.400000}\selectfont 4}%
\end{pgfscope}%
\end{pgfpicture}%
\makeatother%
\endgroup%

%% file: figures/experiments/graphs/sparse_smoothing/nodes_structure/node_classification-Citeseer-APPNP-hidden=32-p_adj_plus=0.001-p_adj_minus=0.8-p_att_plus=0.0-p_att_minus=0.0-multi_class_cert-A.pgf
\begingroup%
\makeatletter%
\begin{pgfpicture}%
\pgfpathrectangle{\pgfpointorigin}{\pgfqpoint{1.375000in}{1.581250in}}%
\pgfusepath{use as bounding box, clip}%
\begin{pgfscope}%
\pgfsetbuttcap%
\pgfsetmiterjoin%
\definecolor{currentfill}{rgb}{1.000000,1.000000,1.000000}%
\pgfsetfillcolor{currentfill}%
\pgfsetlinewidth{0.000000pt}%
\definecolor{currentstroke}{rgb}{1.000000,1.000000,1.000000}%
\pgfsetstrokecolor{currentstroke}%
\pgfsetdash{}{0pt}%
\pgfpathmoveto{\pgfqpoint{0.000000in}{0.000000in}}%
\pgfpathlineto{\pgfqpoint{1.375000in}{0.000000in}}%
\pgfpathlineto{\pgfqpoint{1.375000in}{1.581250in}}%
\pgfpathlineto{\pgfqpoint{0.000000in}{1.581250in}}%
\pgfpathlineto{\pgfqpoint{0.000000in}{0.000000in}}%
\pgfpathclose%
\pgfusepath{fill}%
\end{pgfscope}%
\begin{pgfscope}%
\pgfsetbuttcap%
\pgfsetmiterjoin%
\definecolor{currentfill}{rgb}{1.000000,1.000000,1.000000}%
\pgfsetfillcolor{currentfill}%
\pgfsetlinewidth{0.000000pt}%
\definecolor{currentstroke}{rgb}{0.000000,0.000000,0.000000}%
\pgfsetstrokecolor{currentstroke}%
\pgfsetstrokeopacity{0.000000}%
\pgfsetdash{}{0pt}%
\pgfpathmoveto{\pgfqpoint{0.520798in}{0.442177in}}%
\pgfpathlineto{\pgfqpoint{1.239803in}{0.442177in}}%
\pgfpathlineto{\pgfqpoint{1.239803in}{1.314061in}}%
\pgfpathlineto{\pgfqpoint{0.520798in}{1.314061in}}%
\pgfpathlineto{\pgfqpoint{0.520798in}{0.442177in}}%
\pgfpathclose%
\pgfusepath{fill}%
\end{pgfscope}%
\begin{pgfscope}%
\pgfpathrectangle{\pgfqpoint{0.520798in}{0.442177in}}{\pgfqpoint{0.719005in}{0.871884in}}%
\pgfusepath{clip}%
\pgfsetroundcap%
\pgfsetroundjoin%
\pgfsetlinewidth{0.501875pt}%
\definecolor{currentstroke}{rgb}{0.800000,0.800000,0.800000}%
\pgfsetstrokecolor{currentstroke}%
\pgfsetdash{}{0pt}%
\pgfpathmoveto{\pgfqpoint{0.520798in}{0.442177in}}%
\pgfpathlineto{\pgfqpoint{0.520798in}{1.314061in}}%
\pgfusepath{stroke}%
\end{pgfscope}%
\begin{pgfscope}%
\definecolor{textcolor}{rgb}{0.150000,0.150000,0.150000}%
\pgfsetstrokecolor{textcolor}%
\pgfsetfillcolor{textcolor}%
\pgftext[x=0.520798in,y=0.351899in,,top]{\color{textcolor}\rmfamily\fontsize{8.000000}{9.600000}\selectfont \(\displaystyle {0}\)}%
\end{pgfscope}%
\begin{pgfscope}%
\pgfpathrectangle{\pgfqpoint{0.520798in}{0.442177in}}{\pgfqpoint{0.719005in}{0.871884in}}%
\pgfusepath{clip}%
\pgfsetroundcap%
\pgfsetroundjoin%
\pgfsetlinewidth{0.501875pt}%
\definecolor{currentstroke}{rgb}{0.800000,0.800000,0.800000}%
\pgfsetstrokecolor{currentstroke}%
\pgfsetdash{}{0pt}%
\pgfpathmoveto{\pgfqpoint{0.880300in}{0.442177in}}%
\pgfpathlineto{\pgfqpoint{0.880300in}{1.314061in}}%
\pgfusepath{stroke}%
\end{pgfscope}%
\begin{pgfscope}%
\definecolor{textcolor}{rgb}{0.150000,0.150000,0.150000}%
\pgfsetstrokecolor{textcolor}%
\pgfsetfillcolor{textcolor}%
\pgftext[x=0.880300in,y=0.351899in,,top]{\color{textcolor}\rmfamily\fontsize{8.000000}{9.600000}\selectfont \(\displaystyle {5}\)}%
\end{pgfscope}%
\begin{pgfscope}%
\pgfpathrectangle{\pgfqpoint{0.520798in}{0.442177in}}{\pgfqpoint{0.719005in}{0.871884in}}%
\pgfusepath{clip}%
\pgfsetroundcap%
\pgfsetroundjoin%
\pgfsetlinewidth{0.501875pt}%
\definecolor{currentstroke}{rgb}{0.800000,0.800000,0.800000}%
\pgfsetstrokecolor{currentstroke}%
\pgfsetdash{}{0pt}%
\pgfpathmoveto{\pgfqpoint{1.239803in}{0.442177in}}%
\pgfpathlineto{\pgfqpoint{1.239803in}{1.314061in}}%
\pgfusepath{stroke}%
\end{pgfscope}%
\begin{pgfscope}%
\definecolor{textcolor}{rgb}{0.150000,0.150000,0.150000}%
\pgfsetstrokecolor{textcolor}%
\pgfsetfillcolor{textcolor}%
\pgftext[x=1.239803in,y=0.351899in,,top]{\color{textcolor}\rmfamily\fontsize{8.000000}{9.600000}\selectfont \(\displaystyle {10}\)}%
\end{pgfscope}%
\begin{pgfscope}%
\definecolor{textcolor}{rgb}{0.150000,0.150000,0.150000}%
\pgfsetstrokecolor{textcolor}%
\pgfsetfillcolor{textcolor}%
\pgftext[x=0.880300in,y=0.198219in,,top]{\color{textcolor}\rmfamily\fontsize{10.000000}{12.000000}\selectfont Edit distance \(\displaystyle \epsilon\)}%
\end{pgfscope}%
\begin{pgfscope}%
\pgfpathrectangle{\pgfqpoint{0.520798in}{0.442177in}}{\pgfqpoint{0.719005in}{0.871884in}}%
\pgfusepath{clip}%
\pgfsetroundcap%
\pgfsetroundjoin%
\pgfsetlinewidth{0.501875pt}%
\definecolor{currentstroke}{rgb}{0.800000,0.800000,0.800000}%
\pgfsetstrokecolor{currentstroke}%
\pgfsetdash{}{0pt}%
\pgfpathmoveto{\pgfqpoint{0.520798in}{0.442177in}}%
\pgfpathlineto{\pgfqpoint{1.239803in}{0.442177in}}%
\pgfusepath{stroke}%
\end{pgfscope}%
\begin{pgfscope}%
\definecolor{textcolor}{rgb}{0.150000,0.150000,0.150000}%
\pgfsetstrokecolor{textcolor}%
\pgfsetfillcolor{textcolor}%
\pgftext[x=0.273151in, y=0.403915in, left, base]{\color{textcolor}\rmfamily\fontsize{8.000000}{9.600000}\selectfont 0\%}%
\end{pgfscope}%
\begin{pgfscope}%
\pgfpathrectangle{\pgfqpoint{0.520798in}{0.442177in}}{\pgfqpoint{0.719005in}{0.871884in}}%
\pgfusepath{clip}%
\pgfsetroundcap%
\pgfsetroundjoin%
\pgfsetlinewidth{0.501875pt}%
\definecolor{currentstroke}{rgb}{0.800000,0.800000,0.800000}%
\pgfsetstrokecolor{currentstroke}%
\pgfsetdash{}{0pt}%
\pgfpathmoveto{\pgfqpoint{0.520798in}{0.660148in}}%
\pgfpathlineto{\pgfqpoint{1.239803in}{0.660148in}}%
\pgfusepath{stroke}%
\end{pgfscope}%
\begin{pgfscope}%
\definecolor{textcolor}{rgb}{0.150000,0.150000,0.150000}%
\pgfsetstrokecolor{textcolor}%
\pgfsetfillcolor{textcolor}%
\pgftext[x=0.214138in, y=0.621886in, left, base]{\color{textcolor}\rmfamily\fontsize{8.000000}{9.600000}\selectfont 25\%}%
\end{pgfscope}%
\begin{pgfscope}%
\pgfpathrectangle{\pgfqpoint{0.520798in}{0.442177in}}{\pgfqpoint{0.719005in}{0.871884in}}%
\pgfusepath{clip}%
\pgfsetroundcap%
\pgfsetroundjoin%
\pgfsetlinewidth{0.501875pt}%
\definecolor{currentstroke}{rgb}{0.800000,0.800000,0.800000}%
\pgfsetstrokecolor{currentstroke}%
\pgfsetdash{}{0pt}%
\pgfpathmoveto{\pgfqpoint{0.520798in}{0.878119in}}%
\pgfpathlineto{\pgfqpoint{1.239803in}{0.878119in}}%
\pgfusepath{stroke}%
\end{pgfscope}%
\begin{pgfscope}%
\definecolor{textcolor}{rgb}{0.150000,0.150000,0.150000}%
\pgfsetstrokecolor{textcolor}%
\pgfsetfillcolor{textcolor}%
\pgftext[x=0.214138in, y=0.839857in, left, base]{\color{textcolor}\rmfamily\fontsize{8.000000}{9.600000}\selectfont 50\%}%
\end{pgfscope}%
\begin{pgfscope}%
\pgfpathrectangle{\pgfqpoint{0.520798in}{0.442177in}}{\pgfqpoint{0.719005in}{0.871884in}}%
\pgfusepath{clip}%
\pgfsetroundcap%
\pgfsetroundjoin%
\pgfsetlinewidth{0.501875pt}%
\definecolor{currentstroke}{rgb}{0.800000,0.800000,0.800000}%
\pgfsetstrokecolor{currentstroke}%
\pgfsetdash{}{0pt}%
\pgfpathmoveto{\pgfqpoint{0.520798in}{1.096090in}}%
\pgfpathlineto{\pgfqpoint{1.239803in}{1.096090in}}%
\pgfusepath{stroke}%
\end{pgfscope}%
\begin{pgfscope}%
\definecolor{textcolor}{rgb}{0.150000,0.150000,0.150000}%
\pgfsetstrokecolor{textcolor}%
\pgfsetfillcolor{textcolor}%
\pgftext[x=0.214138in, y=1.057828in, left, base]{\color{textcolor}\rmfamily\fontsize{8.000000}{9.600000}\selectfont 75\%}%
\end{pgfscope}%
\begin{pgfscope}%
\pgfpathrectangle{\pgfqpoint{0.520798in}{0.442177in}}{\pgfqpoint{0.719005in}{0.871884in}}%
\pgfusepath{clip}%
\pgfsetroundcap%
\pgfsetroundjoin%
\pgfsetlinewidth{0.501875pt}%
\definecolor{currentstroke}{rgb}{0.800000,0.800000,0.800000}%
\pgfsetstrokecolor{currentstroke}%
\pgfsetdash{}{0pt}%
\pgfpathmoveto{\pgfqpoint{0.520798in}{1.314061in}}%
\pgfpathlineto{\pgfqpoint{1.239803in}{1.314061in}}%
\pgfusepath{stroke}%
\end{pgfscope}%
\begin{pgfscope}%
\definecolor{textcolor}{rgb}{0.150000,0.150000,0.150000}%
\pgfsetstrokecolor{textcolor}%
\pgfsetfillcolor{textcolor}%
\pgftext[x=0.155124in, y=1.275799in, left, base]{\color{textcolor}\rmfamily\fontsize{8.000000}{9.600000}\selectfont 100\%}%
\end{pgfscope}%
\begin{pgfscope}%
\definecolor{textcolor}{rgb}{0.150000,0.150000,0.150000}%
\pgfsetstrokecolor{textcolor}%
\pgfsetfillcolor{textcolor}%
\pgftext[x=0.099569in,y=0.878119in,,bottom,rotate=90.000000]{\color{textcolor}\rmfamily\fontsize{10.000000}{12.000000}\selectfont Cert. Acc.}%
\end{pgfscope}%
\begin{pgfscope}%
\pgfsetrectcap%
\pgfsetmiterjoin%
\pgfsetlinewidth{0.752812pt}%
\definecolor{currentstroke}{rgb}{0.700000,0.700000,0.700000}%
\pgfsetstrokecolor{currentstroke}%
\pgfsetdash{}{0pt}%
\pgfpathmoveto{\pgfqpoint{0.520798in}{0.442177in}}%
\pgfpathlineto{\pgfqpoint{0.520798in}{1.314061in}}%
\pgfusepath{stroke}%
\end{pgfscope}%
\begin{pgfscope}%
\pgfsetrectcap%
\pgfsetmiterjoin%
\pgfsetlinewidth{0.752812pt}%
\definecolor{currentstroke}{rgb}{0.700000,0.700000,0.700000}%
\pgfsetstrokecolor{currentstroke}%
\pgfsetdash{}{0pt}%
\pgfpathmoveto{\pgfqpoint{1.239803in}{0.442177in}}%
\pgfpathlineto{\pgfqpoint{1.239803in}{1.314061in}}%
\pgfusepath{stroke}%
\end{pgfscope}%
\begin{pgfscope}%
\pgfsetrectcap%
\pgfsetmiterjoin%
\pgfsetlinewidth{0.752812pt}%
\definecolor{currentstroke}{rgb}{0.700000,0.700000,0.700000}%
\pgfsetstrokecolor{currentstroke}%
\pgfsetdash{}{0pt}%
\pgfpathmoveto{\pgfqpoint{0.520798in}{0.442177in}}%
\pgfpathlineto{\pgfqpoint{1.239803in}{0.442177in}}%
\pgfusepath{stroke}%
\end{pgfscope}%
\begin{pgfscope}%
\pgfsetrectcap%
\pgfsetmiterjoin%
\pgfsetlinewidth{0.752812pt}%
\definecolor{currentstroke}{rgb}{0.700000,0.700000,0.700000}%
\pgfsetstrokecolor{currentstroke}%
\pgfsetdash{}{0pt}%
\pgfpathmoveto{\pgfqpoint{0.520798in}{1.314061in}}%
\pgfpathlineto{\pgfqpoint{1.239803in}{1.314061in}}%
\pgfusepath{stroke}%
\end{pgfscope}%
\begin{pgfscope}%
\pgfpathrectangle{\pgfqpoint{0.520798in}{0.442177in}}{\pgfqpoint{0.719005in}{0.871884in}}%
\pgfusepath{clip}%
\pgfsetroundcap%
\pgfsetroundjoin%
\pgfsetlinewidth{1.003750pt}%
\definecolor{currentstroke}{rgb}{0.003922,0.450980,0.698039}%
\pgfsetstrokecolor{currentstroke}%
\pgfsetdash{}{0pt}%
\pgfpathmoveto{\pgfqpoint{0.520798in}{0.980890in}}%
\pgfpathlineto{\pgfqpoint{0.556748in}{0.980890in}}%
\pgfpathlineto{\pgfqpoint{0.556748in}{0.907522in}}%
\pgfpathlineto{\pgfqpoint{0.628649in}{0.907522in}}%
\pgfpathlineto{\pgfqpoint{0.628649in}{0.790281in}}%
\pgfpathlineto{\pgfqpoint{0.700549in}{0.790281in}}%
\pgfpathlineto{\pgfqpoint{0.700549in}{0.545504in}}%
\pgfpathlineto{\pgfqpoint{0.772450in}{0.545504in}}%
\pgfpathlineto{\pgfqpoint{0.772450in}{0.480113in}}%
\pgfpathlineto{\pgfqpoint{0.844350in}{0.480113in}}%
\pgfpathlineto{\pgfqpoint{0.844350in}{0.470559in}}%
\pgfpathlineto{\pgfqpoint{0.916251in}{0.470559in}}%
\pgfpathlineto{\pgfqpoint{0.916251in}{0.466757in}}%
\pgfpathlineto{\pgfqpoint{0.988151in}{0.466757in}}%
\pgfpathlineto{\pgfqpoint{0.988151in}{0.461377in}}%
\pgfpathlineto{\pgfqpoint{1.060052in}{0.461377in}}%
\pgfpathlineto{\pgfqpoint{1.060052in}{0.446907in}}%
\pgfpathlineto{\pgfqpoint{1.131952in}{0.446907in}}%
\pgfpathlineto{\pgfqpoint{1.131952in}{0.445794in}}%
\pgfpathlineto{\pgfqpoint{1.203853in}{0.445794in}}%
\pgfpathlineto{\pgfqpoint{1.203853in}{0.443939in}}%
\pgfpathlineto{\pgfqpoint{1.241470in}{0.443939in}}%
\pgfusepath{stroke}%
\end{pgfscope}%
\begin{pgfscope}%
\pgfpathrectangle{\pgfqpoint{0.520798in}{0.442177in}}{\pgfqpoint{0.719005in}{0.871884in}}%
\pgfusepath{clip}%
\pgfsetbuttcap%
\pgfsetroundjoin%
\definecolor{currentfill}{rgb}{0.003922,0.450980,0.698039}%
\pgfsetfillcolor{currentfill}%
\pgfsetfillopacity{0.500000}%
\pgfsetlinewidth{0.000000pt}%
\definecolor{currentstroke}{rgb}{0.003922,0.450980,0.698039}%
\pgfsetstrokecolor{currentstroke}%
\pgfsetstrokeopacity{0.500000}%
\pgfsetdash{}{0pt}%
\pgfpathmoveto{\pgfqpoint{0.520798in}{0.995273in}}%
\pgfpathlineto{\pgfqpoint{0.520798in}{0.966507in}}%
\pgfpathlineto{\pgfqpoint{0.556748in}{0.966507in}}%
\pgfpathlineto{\pgfqpoint{0.556748in}{0.888332in}}%
\pgfpathlineto{\pgfqpoint{0.628649in}{0.888332in}}%
\pgfpathlineto{\pgfqpoint{0.628649in}{0.774020in}}%
\pgfpathlineto{\pgfqpoint{0.700549in}{0.774020in}}%
\pgfpathlineto{\pgfqpoint{0.700549in}{0.536089in}}%
\pgfpathlineto{\pgfqpoint{0.772450in}{0.536089in}}%
\pgfpathlineto{\pgfqpoint{0.772450in}{0.473678in}}%
\pgfpathlineto{\pgfqpoint{0.844350in}{0.473678in}}%
\pgfpathlineto{\pgfqpoint{0.844350in}{0.465927in}}%
\pgfpathlineto{\pgfqpoint{0.916251in}{0.465927in}}%
\pgfpathlineto{\pgfqpoint{0.916251in}{0.462318in}}%
\pgfpathlineto{\pgfqpoint{0.988151in}{0.462318in}}%
\pgfpathlineto{\pgfqpoint{0.988151in}{0.457141in}}%
\pgfpathlineto{\pgfqpoint{1.060052in}{0.457141in}}%
\pgfpathlineto{\pgfqpoint{1.060052in}{0.445550in}}%
\pgfpathlineto{\pgfqpoint{1.131952in}{0.445550in}}%
\pgfpathlineto{\pgfqpoint{1.131952in}{0.444231in}}%
\pgfpathlineto{\pgfqpoint{1.203853in}{0.444231in}}%
\pgfpathlineto{\pgfqpoint{1.203853in}{0.442826in}}%
\pgfpathlineto{\pgfqpoint{1.275753in}{0.442826in}}%
\pgfpathlineto{\pgfqpoint{1.275753in}{0.442817in}}%
\pgfpathlineto{\pgfqpoint{1.347654in}{0.442817in}}%
\pgfpathlineto{\pgfqpoint{1.347654in}{0.442557in}}%
\pgfpathlineto{\pgfqpoint{1.419554in}{0.442557in}}%
\pgfpathlineto{\pgfqpoint{1.419554in}{0.442177in}}%
\pgfpathlineto{\pgfqpoint{2.066659in}{0.442177in}}%
\pgfpathlineto{\pgfqpoint{2.066659in}{0.442177in}}%
\pgfpathlineto{\pgfqpoint{2.677813in}{0.442177in}}%
\pgfpathlineto{\pgfqpoint{2.677813in}{0.442177in}}%
\pgfpathlineto{\pgfqpoint{2.677813in}{0.442177in}}%
\pgfpathlineto{\pgfqpoint{2.066659in}{0.442177in}}%
\pgfpathlineto{\pgfqpoint{2.066659in}{0.442177in}}%
\pgfpathlineto{\pgfqpoint{1.419554in}{0.442177in}}%
\pgfpathlineto{\pgfqpoint{1.419554in}{0.443466in}}%
\pgfpathlineto{\pgfqpoint{1.347654in}{0.443466in}}%
\pgfpathlineto{\pgfqpoint{1.347654in}{0.443763in}}%
\pgfpathlineto{\pgfqpoint{1.275753in}{0.443763in}}%
\pgfpathlineto{\pgfqpoint{1.275753in}{0.445052in}}%
\pgfpathlineto{\pgfqpoint{1.203853in}{0.445052in}}%
\pgfpathlineto{\pgfqpoint{1.203853in}{0.447357in}}%
\pgfpathlineto{\pgfqpoint{1.131952in}{0.447357in}}%
\pgfpathlineto{\pgfqpoint{1.131952in}{0.448264in}}%
\pgfpathlineto{\pgfqpoint{1.060052in}{0.448264in}}%
\pgfpathlineto{\pgfqpoint{1.060052in}{0.465613in}}%
\pgfpathlineto{\pgfqpoint{0.988151in}{0.465613in}}%
\pgfpathlineto{\pgfqpoint{0.988151in}{0.471195in}}%
\pgfpathlineto{\pgfqpoint{0.916251in}{0.471195in}}%
\pgfpathlineto{\pgfqpoint{0.916251in}{0.475192in}}%
\pgfpathlineto{\pgfqpoint{0.844350in}{0.475192in}}%
\pgfpathlineto{\pgfqpoint{0.844350in}{0.486549in}}%
\pgfpathlineto{\pgfqpoint{0.772450in}{0.486549in}}%
\pgfpathlineto{\pgfqpoint{0.772450in}{0.554920in}}%
\pgfpathlineto{\pgfqpoint{0.700549in}{0.554920in}}%
\pgfpathlineto{\pgfqpoint{0.700549in}{0.806543in}}%
\pgfpathlineto{\pgfqpoint{0.628649in}{0.806543in}}%
\pgfpathlineto{\pgfqpoint{0.628649in}{0.926712in}}%
\pgfpathlineto{\pgfqpoint{0.556748in}{0.926712in}}%
\pgfpathlineto{\pgfqpoint{0.556748in}{0.995273in}}%
\pgfpathlineto{\pgfqpoint{0.520798in}{0.995273in}}%
\pgfpathlineto{\pgfqpoint{0.520798in}{0.995273in}}%
\pgfpathclose%
\pgfusepath{fill}%
\end{pgfscope}%
\begin{pgfscope}%
\pgfpathrectangle{\pgfqpoint{0.520798in}{0.442177in}}{\pgfqpoint{0.719005in}{0.871884in}}%
\pgfusepath{clip}%
\pgfsetroundcap%
\pgfsetroundjoin%
\pgfsetlinewidth{1.003750pt}%
\definecolor{currentstroke}{rgb}{0.870588,0.560784,0.019608}%
\pgfsetstrokecolor{currentstroke}%
\pgfsetdash{}{0pt}%
\pgfpathmoveto{\pgfqpoint{0.520798in}{0.980890in}}%
\pgfpathlineto{\pgfqpoint{0.556748in}{0.980890in}}%
\pgfpathlineto{\pgfqpoint{0.556748in}{0.907522in}}%
\pgfpathlineto{\pgfqpoint{0.628649in}{0.907522in}}%
\pgfpathlineto{\pgfqpoint{0.628649in}{0.790281in}}%
\pgfpathlineto{\pgfqpoint{0.700549in}{0.790281in}}%
\pgfpathlineto{\pgfqpoint{0.700549in}{0.545504in}}%
\pgfpathlineto{\pgfqpoint{0.772450in}{0.545504in}}%
\pgfpathlineto{\pgfqpoint{0.772450in}{0.480113in}}%
\pgfpathlineto{\pgfqpoint{0.844350in}{0.480113in}}%
\pgfpathlineto{\pgfqpoint{0.844350in}{0.471116in}}%
\pgfpathlineto{\pgfqpoint{0.916251in}{0.471116in}}%
\pgfpathlineto{\pgfqpoint{0.916251in}{0.467128in}}%
\pgfpathlineto{\pgfqpoint{0.988151in}{0.467128in}}%
\pgfpathlineto{\pgfqpoint{0.988151in}{0.461377in}}%
\pgfpathlineto{\pgfqpoint{1.060052in}{0.461377in}}%
\pgfpathlineto{\pgfqpoint{1.060052in}{0.446907in}}%
\pgfpathlineto{\pgfqpoint{1.131952in}{0.446907in}}%
\pgfpathlineto{\pgfqpoint{1.131952in}{0.445794in}}%
\pgfpathlineto{\pgfqpoint{1.203853in}{0.445794in}}%
\pgfpathlineto{\pgfqpoint{1.203853in}{0.444125in}}%
\pgfpathlineto{\pgfqpoint{1.241470in}{0.444125in}}%
\pgfusepath{stroke}%
\end{pgfscope}%
\begin{pgfscope}%
\pgfpathrectangle{\pgfqpoint{0.520798in}{0.442177in}}{\pgfqpoint{0.719005in}{0.871884in}}%
\pgfusepath{clip}%
\pgfsetbuttcap%
\pgfsetroundjoin%
\definecolor{currentfill}{rgb}{0.870588,0.560784,0.019608}%
\pgfsetfillcolor{currentfill}%
\pgfsetfillopacity{0.500000}%
\pgfsetlinewidth{0.000000pt}%
\definecolor{currentstroke}{rgb}{0.870588,0.560784,0.019608}%
\pgfsetstrokecolor{currentstroke}%
\pgfsetstrokeopacity{0.500000}%
\pgfsetdash{}{0pt}%
\pgfpathmoveto{\pgfqpoint{0.520798in}{0.995273in}}%
\pgfpathlineto{\pgfqpoint{0.520798in}{0.966507in}}%
\pgfpathlineto{\pgfqpoint{0.556748in}{0.966507in}}%
\pgfpathlineto{\pgfqpoint{0.556748in}{0.888332in}}%
\pgfpathlineto{\pgfqpoint{0.628649in}{0.888332in}}%
\pgfpathlineto{\pgfqpoint{0.628649in}{0.774020in}}%
\pgfpathlineto{\pgfqpoint{0.700549in}{0.774020in}}%
\pgfpathlineto{\pgfqpoint{0.700549in}{0.536089in}}%
\pgfpathlineto{\pgfqpoint{0.772450in}{0.536089in}}%
\pgfpathlineto{\pgfqpoint{0.772450in}{0.473678in}}%
\pgfpathlineto{\pgfqpoint{0.844350in}{0.473678in}}%
\pgfpathlineto{\pgfqpoint{0.844350in}{0.466399in}}%
\pgfpathlineto{\pgfqpoint{0.916251in}{0.466399in}}%
\pgfpathlineto{\pgfqpoint{0.916251in}{0.462599in}}%
\pgfpathlineto{\pgfqpoint{0.988151in}{0.462599in}}%
\pgfpathlineto{\pgfqpoint{0.988151in}{0.457141in}}%
\pgfpathlineto{\pgfqpoint{1.060052in}{0.457141in}}%
\pgfpathlineto{\pgfqpoint{1.060052in}{0.445550in}}%
\pgfpathlineto{\pgfqpoint{1.131952in}{0.445550in}}%
\pgfpathlineto{\pgfqpoint{1.131952in}{0.444231in}}%
\pgfpathlineto{\pgfqpoint{1.203853in}{0.444231in}}%
\pgfpathlineto{\pgfqpoint{1.203853in}{0.442937in}}%
\pgfpathlineto{\pgfqpoint{1.275753in}{0.442937in}}%
\pgfpathlineto{\pgfqpoint{1.275753in}{0.442826in}}%
\pgfpathlineto{\pgfqpoint{1.347654in}{0.442826in}}%
\pgfpathlineto{\pgfqpoint{1.347654in}{0.442557in}}%
\pgfpathlineto{\pgfqpoint{1.419554in}{0.442557in}}%
\pgfpathlineto{\pgfqpoint{1.419554in}{0.442177in}}%
\pgfpathlineto{\pgfqpoint{2.066659in}{0.442177in}}%
\pgfpathlineto{\pgfqpoint{2.066659in}{0.442177in}}%
\pgfpathlineto{\pgfqpoint{2.677813in}{0.442177in}}%
\pgfpathlineto{\pgfqpoint{2.677813in}{0.442177in}}%
\pgfpathlineto{\pgfqpoint{2.677813in}{0.442177in}}%
\pgfpathlineto{\pgfqpoint{2.066659in}{0.442177in}}%
\pgfpathlineto{\pgfqpoint{2.066659in}{0.442177in}}%
\pgfpathlineto{\pgfqpoint{1.419554in}{0.442177in}}%
\pgfpathlineto{\pgfqpoint{1.419554in}{0.443466in}}%
\pgfpathlineto{\pgfqpoint{1.347654in}{0.443466in}}%
\pgfpathlineto{\pgfqpoint{1.347654in}{0.443939in}}%
\pgfpathlineto{\pgfqpoint{1.275753in}{0.443939in}}%
\pgfpathlineto{\pgfqpoint{1.275753in}{0.445312in}}%
\pgfpathlineto{\pgfqpoint{1.203853in}{0.445312in}}%
\pgfpathlineto{\pgfqpoint{1.203853in}{0.447357in}}%
\pgfpathlineto{\pgfqpoint{1.131952in}{0.447357in}}%
\pgfpathlineto{\pgfqpoint{1.131952in}{0.448264in}}%
\pgfpathlineto{\pgfqpoint{1.060052in}{0.448264in}}%
\pgfpathlineto{\pgfqpoint{1.060052in}{0.465613in}}%
\pgfpathlineto{\pgfqpoint{0.988151in}{0.465613in}}%
\pgfpathlineto{\pgfqpoint{0.988151in}{0.471656in}}%
\pgfpathlineto{\pgfqpoint{0.916251in}{0.471656in}}%
\pgfpathlineto{\pgfqpoint{0.916251in}{0.475833in}}%
\pgfpathlineto{\pgfqpoint{0.844350in}{0.475833in}}%
\pgfpathlineto{\pgfqpoint{0.844350in}{0.486549in}}%
\pgfpathlineto{\pgfqpoint{0.772450in}{0.486549in}}%
\pgfpathlineto{\pgfqpoint{0.772450in}{0.554920in}}%
\pgfpathlineto{\pgfqpoint{0.700549in}{0.554920in}}%
\pgfpathlineto{\pgfqpoint{0.700549in}{0.806543in}}%
\pgfpathlineto{\pgfqpoint{0.628649in}{0.806543in}}%
\pgfpathlineto{\pgfqpoint{0.628649in}{0.926712in}}%
\pgfpathlineto{\pgfqpoint{0.556748in}{0.926712in}}%
\pgfpathlineto{\pgfqpoint{0.556748in}{0.995273in}}%
\pgfpathlineto{\pgfqpoint{0.520798in}{0.995273in}}%
\pgfpathlineto{\pgfqpoint{0.520798in}{0.995273in}}%
\pgfpathclose%
\pgfusepath{fill}%
\end{pgfscope}%
\begin{pgfscope}%
\pgfpathrectangle{\pgfqpoint{0.520798in}{0.442177in}}{\pgfqpoint{0.719005in}{0.871884in}}%
\pgfusepath{clip}%
\pgfsetroundcap%
\pgfsetroundjoin%
\pgfsetlinewidth{1.003750pt}%
\definecolor{currentstroke}{rgb}{0.007843,0.619608,0.450980}%
\pgfsetstrokecolor{currentstroke}%
\pgfsetdash{}{0pt}%
\pgfpathmoveto{\pgfqpoint{0.520798in}{0.980890in}}%
\pgfpathlineto{\pgfqpoint{0.556748in}{0.980890in}}%
\pgfpathlineto{\pgfqpoint{0.556748in}{0.907522in}}%
\pgfpathlineto{\pgfqpoint{0.628649in}{0.907522in}}%
\pgfpathlineto{\pgfqpoint{0.628649in}{0.790281in}}%
\pgfpathlineto{\pgfqpoint{0.700549in}{0.790281in}}%
\pgfpathlineto{\pgfqpoint{0.700549in}{0.545504in}}%
\pgfpathlineto{\pgfqpoint{0.772450in}{0.545504in}}%
\pgfpathlineto{\pgfqpoint{0.772450in}{0.480113in}}%
\pgfpathlineto{\pgfqpoint{0.844350in}{0.480113in}}%
\pgfpathlineto{\pgfqpoint{0.844350in}{0.471116in}}%
\pgfpathlineto{\pgfqpoint{0.916251in}{0.471116in}}%
\pgfpathlineto{\pgfqpoint{0.916251in}{0.467128in}}%
\pgfpathlineto{\pgfqpoint{0.988151in}{0.467128in}}%
\pgfpathlineto{\pgfqpoint{0.988151in}{0.461377in}}%
\pgfpathlineto{\pgfqpoint{1.060052in}{0.461377in}}%
\pgfpathlineto{\pgfqpoint{1.060052in}{0.446907in}}%
\pgfpathlineto{\pgfqpoint{1.131952in}{0.446907in}}%
\pgfpathlineto{\pgfqpoint{1.131952in}{0.445794in}}%
\pgfpathlineto{\pgfqpoint{1.203853in}{0.445794in}}%
\pgfpathlineto{\pgfqpoint{1.203853in}{0.444125in}}%
\pgfpathlineto{\pgfqpoint{1.241470in}{0.444125in}}%
\pgfusepath{stroke}%
\end{pgfscope}%
\begin{pgfscope}%
\pgfpathrectangle{\pgfqpoint{0.520798in}{0.442177in}}{\pgfqpoint{0.719005in}{0.871884in}}%
\pgfusepath{clip}%
\pgfsetbuttcap%
\pgfsetroundjoin%
\definecolor{currentfill}{rgb}{0.007843,0.619608,0.450980}%
\pgfsetfillcolor{currentfill}%
\pgfsetfillopacity{0.500000}%
\pgfsetlinewidth{0.000000pt}%
\definecolor{currentstroke}{rgb}{0.007843,0.619608,0.450980}%
\pgfsetstrokecolor{currentstroke}%
\pgfsetstrokeopacity{0.500000}%
\pgfsetdash{}{0pt}%
\pgfpathmoveto{\pgfqpoint{0.520798in}{0.995273in}}%
\pgfpathlineto{\pgfqpoint{0.520798in}{0.966507in}}%
\pgfpathlineto{\pgfqpoint{0.556748in}{0.966507in}}%
\pgfpathlineto{\pgfqpoint{0.556748in}{0.888332in}}%
\pgfpathlineto{\pgfqpoint{0.628649in}{0.888332in}}%
\pgfpathlineto{\pgfqpoint{0.628649in}{0.774020in}}%
\pgfpathlineto{\pgfqpoint{0.700549in}{0.774020in}}%
\pgfpathlineto{\pgfqpoint{0.700549in}{0.536089in}}%
\pgfpathlineto{\pgfqpoint{0.772450in}{0.536089in}}%
\pgfpathlineto{\pgfqpoint{0.772450in}{0.473678in}}%
\pgfpathlineto{\pgfqpoint{0.844350in}{0.473678in}}%
\pgfpathlineto{\pgfqpoint{0.844350in}{0.466399in}}%
\pgfpathlineto{\pgfqpoint{0.916251in}{0.466399in}}%
\pgfpathlineto{\pgfqpoint{0.916251in}{0.462599in}}%
\pgfpathlineto{\pgfqpoint{0.988151in}{0.462599in}}%
\pgfpathlineto{\pgfqpoint{0.988151in}{0.457141in}}%
\pgfpathlineto{\pgfqpoint{1.060052in}{0.457141in}}%
\pgfpathlineto{\pgfqpoint{1.060052in}{0.445550in}}%
\pgfpathlineto{\pgfqpoint{1.131952in}{0.445550in}}%
\pgfpathlineto{\pgfqpoint{1.131952in}{0.444231in}}%
\pgfpathlineto{\pgfqpoint{1.203853in}{0.444231in}}%
\pgfpathlineto{\pgfqpoint{1.203853in}{0.442937in}}%
\pgfpathlineto{\pgfqpoint{1.275753in}{0.442937in}}%
\pgfpathlineto{\pgfqpoint{1.275753in}{0.442826in}}%
\pgfpathlineto{\pgfqpoint{1.347654in}{0.442826in}}%
\pgfpathlineto{\pgfqpoint{1.347654in}{0.442557in}}%
\pgfpathlineto{\pgfqpoint{1.419554in}{0.442557in}}%
\pgfpathlineto{\pgfqpoint{1.419554in}{0.442177in}}%
\pgfpathlineto{\pgfqpoint{2.066659in}{0.442177in}}%
\pgfpathlineto{\pgfqpoint{2.066659in}{0.442177in}}%
\pgfpathlineto{\pgfqpoint{2.677813in}{0.442177in}}%
\pgfpathlineto{\pgfqpoint{2.677813in}{0.442177in}}%
\pgfpathlineto{\pgfqpoint{2.677813in}{0.442177in}}%
\pgfpathlineto{\pgfqpoint{2.066659in}{0.442177in}}%
\pgfpathlineto{\pgfqpoint{2.066659in}{0.442177in}}%
\pgfpathlineto{\pgfqpoint{1.419554in}{0.442177in}}%
\pgfpathlineto{\pgfqpoint{1.419554in}{0.443466in}}%
\pgfpathlineto{\pgfqpoint{1.347654in}{0.443466in}}%
\pgfpathlineto{\pgfqpoint{1.347654in}{0.443939in}}%
\pgfpathlineto{\pgfqpoint{1.275753in}{0.443939in}}%
\pgfpathlineto{\pgfqpoint{1.275753in}{0.445312in}}%
\pgfpathlineto{\pgfqpoint{1.203853in}{0.445312in}}%
\pgfpathlineto{\pgfqpoint{1.203853in}{0.447357in}}%
\pgfpathlineto{\pgfqpoint{1.131952in}{0.447357in}}%
\pgfpathlineto{\pgfqpoint{1.131952in}{0.448264in}}%
\pgfpathlineto{\pgfqpoint{1.060052in}{0.448264in}}%
\pgfpathlineto{\pgfqpoint{1.060052in}{0.465613in}}%
\pgfpathlineto{\pgfqpoint{0.988151in}{0.465613in}}%
\pgfpathlineto{\pgfqpoint{0.988151in}{0.471656in}}%
\pgfpathlineto{\pgfqpoint{0.916251in}{0.471656in}}%
\pgfpathlineto{\pgfqpoint{0.916251in}{0.475833in}}%
\pgfpathlineto{\pgfqpoint{0.844350in}{0.475833in}}%
\pgfpathlineto{\pgfqpoint{0.844350in}{0.486549in}}%
\pgfpathlineto{\pgfqpoint{0.772450in}{0.486549in}}%
\pgfpathlineto{\pgfqpoint{0.772450in}{0.554920in}}%
\pgfpathlineto{\pgfqpoint{0.700549in}{0.554920in}}%
\pgfpathlineto{\pgfqpoint{0.700549in}{0.806543in}}%
\pgfpathlineto{\pgfqpoint{0.628649in}{0.806543in}}%
\pgfpathlineto{\pgfqpoint{0.628649in}{0.926712in}}%
\pgfpathlineto{\pgfqpoint{0.556748in}{0.926712in}}%
\pgfpathlineto{\pgfqpoint{0.556748in}{0.995273in}}%
\pgfpathlineto{\pgfqpoint{0.520798in}{0.995273in}}%
\pgfpathlineto{\pgfqpoint{0.520798in}{0.995273in}}%
\pgfpathclose%
\pgfusepath{fill}%
\end{pgfscope}%
\begin{pgfscope}%
\definecolor{textcolor}{rgb}{0.150000,0.150000,0.150000}%
\pgfsetstrokecolor{textcolor}%
\pgfsetfillcolor{textcolor}%
\pgftext[x=0.880300in,y=1.397395in,,base]{\color{textcolor}\rmfamily\fontsize{9.000000}{10.800000}\selectfont \(\displaystyle c_A^+=1\)}%
\end{pgfscope}%
\begin{pgfscope}%
\pgfsetbuttcap%
\pgfsetmiterjoin%
\definecolor{currentfill}{rgb}{1.000000,1.000000,1.000000}%
\pgfsetfillcolor{currentfill}%
\pgfsetfillopacity{0.800000}%
\pgfsetlinewidth{1.003750pt}%
\definecolor{currentstroke}{rgb}{0.800000,0.800000,0.800000}%
\pgfsetstrokecolor{currentstroke}%
\pgfsetstrokeopacity{0.800000}%
\pgfsetdash{}{0pt}%
\pgfpathmoveto{\pgfqpoint{0.785843in}{0.638103in}}%
\pgfpathlineto{\pgfqpoint{1.191192in}{0.638103in}}%
\pgfpathlineto{\pgfqpoint{1.191192in}{1.265450in}}%
\pgfpathlineto{\pgfqpoint{0.785843in}{1.265450in}}%
\pgfpathlineto{\pgfqpoint{0.785843in}{0.638103in}}%
\pgfpathclose%
\pgfusepath{stroke,fill}%
\end{pgfscope}%
\begin{pgfscope}%
\definecolor{textcolor}{rgb}{0.150000,0.150000,0.150000}%
\pgfsetstrokecolor{textcolor}%
\pgfsetfillcolor{textcolor}%
\pgftext[x=0.917113in,y=1.113275in,left,base]{\color{textcolor}\rmfamily\fontsize{9.000000}{10.800000}\selectfont \(\displaystyle c_A^-\)}%
\end{pgfscope}%
\begin{pgfscope}%
\pgfsetroundcap%
\pgfsetroundjoin%
\pgfsetlinewidth{1.003750pt}%
\definecolor{currentstroke}{rgb}{0.003922,0.450980,0.698039}%
\pgfsetstrokecolor{currentstroke}%
\pgfsetdash{}{0pt}%
\pgfpathmoveto{\pgfqpoint{0.824732in}{1.001052in}}%
\pgfpathlineto{\pgfqpoint{0.873343in}{1.001052in}}%
\pgfpathlineto{\pgfqpoint{0.873343in}{1.001052in}}%
\pgfpathlineto{\pgfqpoint{0.970565in}{1.001052in}}%
\pgfpathlineto{\pgfqpoint{0.970565in}{1.001052in}}%
\pgfpathlineto{\pgfqpoint{1.019176in}{1.001052in}}%
\pgfusepath{stroke}%
\end{pgfscope}%
\begin{pgfscope}%
\definecolor{textcolor}{rgb}{0.150000,0.150000,0.150000}%
\pgfsetstrokecolor{textcolor}%
\pgfsetfillcolor{textcolor}%
\pgftext[x=1.096954in,y=0.967024in,left,base]{\color{textcolor}\rmfamily\fontsize{7.000000}{8.400000}\selectfont 1}%
\end{pgfscope}%
\begin{pgfscope}%
\pgfsetroundcap%
\pgfsetroundjoin%
\pgfsetlinewidth{1.003750pt}%
\definecolor{currentstroke}{rgb}{0.870588,0.560784,0.019608}%
\pgfsetstrokecolor{currentstroke}%
\pgfsetdash{}{0pt}%
\pgfpathmoveto{\pgfqpoint{0.824732in}{0.865486in}}%
\pgfpathlineto{\pgfqpoint{0.873343in}{0.865486in}}%
\pgfpathlineto{\pgfqpoint{0.873343in}{0.865486in}}%
\pgfpathlineto{\pgfqpoint{0.970565in}{0.865486in}}%
\pgfpathlineto{\pgfqpoint{0.970565in}{0.865486in}}%
\pgfpathlineto{\pgfqpoint{1.019176in}{0.865486in}}%
\pgfusepath{stroke}%
\end{pgfscope}%
\begin{pgfscope}%
\definecolor{textcolor}{rgb}{0.150000,0.150000,0.150000}%
\pgfsetstrokecolor{textcolor}%
\pgfsetfillcolor{textcolor}%
\pgftext[x=1.096954in,y=0.831458in,left,base]{\color{textcolor}\rmfamily\fontsize{7.000000}{8.400000}\selectfont 2}%
\end{pgfscope}%
\begin{pgfscope}%
\pgfsetroundcap%
\pgfsetroundjoin%
\pgfsetlinewidth{1.003750pt}%
\definecolor{currentstroke}{rgb}{0.007843,0.619608,0.450980}%
\pgfsetstrokecolor{currentstroke}%
\pgfsetdash{}{0pt}%
\pgfpathmoveto{\pgfqpoint{0.824732in}{0.729919in}}%
\pgfpathlineto{\pgfqpoint{0.873343in}{0.729919in}}%
\pgfpathlineto{\pgfqpoint{0.873343in}{0.729919in}}%
\pgfpathlineto{\pgfqpoint{0.970565in}{0.729919in}}%
\pgfpathlineto{\pgfqpoint{0.970565in}{0.729919in}}%
\pgfpathlineto{\pgfqpoint{1.019176in}{0.729919in}}%
\pgfusepath{stroke}%
\end{pgfscope}%
\begin{pgfscope}%
\definecolor{textcolor}{rgb}{0.150000,0.150000,0.150000}%
\pgfsetstrokecolor{textcolor}%
\pgfsetfillcolor{textcolor}%
\pgftext[x=1.096954in,y=0.695892in,left,base]{\color{textcolor}\rmfamily\fontsize{7.000000}{8.400000}\selectfont 4}%
\end{pgfscope}%
\end{pgfpicture}%
\makeatother%
\endgroup%

%% file: figures/experiments/attacks/pointnet_attack.pgf
\begingroup%
\makeatletter%
\begin{pgfpicture}%
\pgfpathrectangle{\pgfpointorigin}{\pgfqpoint{2.750000in}{1.699593in}}%
\pgfusepath{use as bounding box, clip}%
\begin{pgfscope}%
\pgfsetbuttcap%
\pgfsetmiterjoin%
\definecolor{currentfill}{rgb}{1.000000,1.000000,1.000000}%
\pgfsetfillcolor{currentfill}%
\pgfsetlinewidth{0.000000pt}%
\definecolor{currentstroke}{rgb}{1.000000,1.000000,1.000000}%
\pgfsetstrokecolor{currentstroke}%
\pgfsetdash{}{0pt}%
\pgfpathmoveto{\pgfqpoint{0.000000in}{0.000000in}}%
\pgfpathlineto{\pgfqpoint{2.750000in}{0.000000in}}%
\pgfpathlineto{\pgfqpoint{2.750000in}{1.699593in}}%
\pgfpathlineto{\pgfqpoint{0.000000in}{1.699593in}}%
\pgfpathlineto{\pgfqpoint{0.000000in}{0.000000in}}%
\pgfpathclose%
\pgfusepath{fill}%
\end{pgfscope}%
\begin{pgfscope}%
\pgfsetbuttcap%
\pgfsetmiterjoin%
\definecolor{currentfill}{rgb}{1.000000,1.000000,1.000000}%
\pgfsetfillcolor{currentfill}%
\pgfsetlinewidth{0.000000pt}%
\definecolor{currentstroke}{rgb}{0.000000,0.000000,0.000000}%
\pgfsetstrokecolor{currentstroke}%
\pgfsetstrokeopacity{0.000000}%
\pgfsetdash{}{0pt}%
\pgfpathmoveto{\pgfqpoint{0.618786in}{0.442177in}}%
\pgfpathlineto{\pgfqpoint{2.673611in}{0.442177in}}%
\pgfpathlineto{\pgfqpoint{2.673611in}{1.585085in}}%
\pgfpathlineto{\pgfqpoint{0.618786in}{1.585085in}}%
\pgfpathlineto{\pgfqpoint{0.618786in}{0.442177in}}%
\pgfpathclose%
\pgfusepath{fill}%
\end{pgfscope}%
\begin{pgfscope}%
\pgfpathrectangle{\pgfqpoint{0.618786in}{0.442177in}}{\pgfqpoint{2.054825in}{1.142908in}}%
\pgfusepath{clip}%
\pgfsetroundcap%
\pgfsetroundjoin%
\pgfsetlinewidth{0.501875pt}%
\definecolor{currentstroke}{rgb}{0.800000,0.800000,0.800000}%
\pgfsetstrokecolor{currentstroke}%
\pgfsetdash{}{0pt}%
\pgfpathmoveto{\pgfqpoint{0.618786in}{0.442177in}}%
\pgfpathlineto{\pgfqpoint{0.618786in}{1.585085in}}%
\pgfusepath{stroke}%
\end{pgfscope}%
\begin{pgfscope}%
\definecolor{textcolor}{rgb}{0.150000,0.150000,0.150000}%
\pgfsetstrokecolor{textcolor}%
\pgfsetfillcolor{textcolor}%
\pgftext[x=0.618786in,y=0.351899in,,top]{\color{textcolor}\rmfamily\fontsize{8.000000}{9.600000}\selectfont \(\displaystyle {0.0}\)}%
\end{pgfscope}%
\begin{pgfscope}%
\pgfpathrectangle{\pgfqpoint{0.618786in}{0.442177in}}{\pgfqpoint{2.054825in}{1.142908in}}%
\pgfusepath{clip}%
\pgfsetroundcap%
\pgfsetroundjoin%
\pgfsetlinewidth{0.501875pt}%
\definecolor{currentstroke}{rgb}{0.800000,0.800000,0.800000}%
\pgfsetstrokecolor{currentstroke}%
\pgfsetdash{}{0pt}%
\pgfpathmoveto{\pgfqpoint{1.205879in}{0.442177in}}%
\pgfpathlineto{\pgfqpoint{1.205879in}{1.585085in}}%
\pgfusepath{stroke}%
\end{pgfscope}%
\begin{pgfscope}%
\definecolor{textcolor}{rgb}{0.150000,0.150000,0.150000}%
\pgfsetstrokecolor{textcolor}%
\pgfsetfillcolor{textcolor}%
\pgftext[x=1.205879in,y=0.351899in,,top]{\color{textcolor}\rmfamily\fontsize{8.000000}{9.600000}\selectfont \(\displaystyle {0.2}\)}%
\end{pgfscope}%
\begin{pgfscope}%
\pgfpathrectangle{\pgfqpoint{0.618786in}{0.442177in}}{\pgfqpoint{2.054825in}{1.142908in}}%
\pgfusepath{clip}%
\pgfsetroundcap%
\pgfsetroundjoin%
\pgfsetlinewidth{0.501875pt}%
\definecolor{currentstroke}{rgb}{0.800000,0.800000,0.800000}%
\pgfsetstrokecolor{currentstroke}%
\pgfsetdash{}{0pt}%
\pgfpathmoveto{\pgfqpoint{1.792972in}{0.442177in}}%
\pgfpathlineto{\pgfqpoint{1.792972in}{1.585085in}}%
\pgfusepath{stroke}%
\end{pgfscope}%
\begin{pgfscope}%
\definecolor{textcolor}{rgb}{0.150000,0.150000,0.150000}%
\pgfsetstrokecolor{textcolor}%
\pgfsetfillcolor{textcolor}%
\pgftext[x=1.792972in,y=0.351899in,,top]{\color{textcolor}\rmfamily\fontsize{8.000000}{9.600000}\selectfont \(\displaystyle {0.4}\)}%
\end{pgfscope}%
\begin{pgfscope}%
\pgfpathrectangle{\pgfqpoint{0.618786in}{0.442177in}}{\pgfqpoint{2.054825in}{1.142908in}}%
\pgfusepath{clip}%
\pgfsetroundcap%
\pgfsetroundjoin%
\pgfsetlinewidth{0.501875pt}%
\definecolor{currentstroke}{rgb}{0.800000,0.800000,0.800000}%
\pgfsetstrokecolor{currentstroke}%
\pgfsetdash{}{0pt}%
\pgfpathmoveto{\pgfqpoint{2.380065in}{0.442177in}}%
\pgfpathlineto{\pgfqpoint{2.380065in}{1.585085in}}%
\pgfusepath{stroke}%
\end{pgfscope}%
\begin{pgfscope}%
\definecolor{textcolor}{rgb}{0.150000,0.150000,0.150000}%
\pgfsetstrokecolor{textcolor}%
\pgfsetfillcolor{textcolor}%
\pgftext[x=2.380065in,y=0.351899in,,top]{\color{textcolor}\rmfamily\fontsize{8.000000}{9.600000}\selectfont \(\displaystyle {0.6}\)}%
\end{pgfscope}%
\begin{pgfscope}%
\definecolor{textcolor}{rgb}{0.150000,0.150000,0.150000}%
\pgfsetstrokecolor{textcolor}%
\pgfsetfillcolor{textcolor}%
\pgftext[x=1.646199in,y=0.198219in,,top]{\color{textcolor}\rmfamily\fontsize{10.000000}{12.000000}\selectfont Correspondence distance \(\displaystyle \epsilon\)}%
\end{pgfscope}%
\begin{pgfscope}%
\pgfpathrectangle{\pgfqpoint{0.618786in}{0.442177in}}{\pgfqpoint{2.054825in}{1.142908in}}%
\pgfusepath{clip}%
\pgfsetroundcap%
\pgfsetroundjoin%
\pgfsetlinewidth{0.501875pt}%
\definecolor{currentstroke}{rgb}{0.800000,0.800000,0.800000}%
\pgfsetstrokecolor{currentstroke}%
\pgfsetdash{}{0pt}%
\pgfpathmoveto{\pgfqpoint{0.618786in}{0.442177in}}%
\pgfpathlineto{\pgfqpoint{2.673611in}{0.442177in}}%
\pgfusepath{stroke}%
\end{pgfscope}%
\begin{pgfscope}%
\definecolor{textcolor}{rgb}{0.150000,0.150000,0.150000}%
\pgfsetstrokecolor{textcolor}%
\pgfsetfillcolor{textcolor}%
\pgftext[x=0.371139in, y=0.403915in, left, base]{\color{textcolor}\rmfamily\fontsize{8.000000}{9.600000}\selectfont 0\%}%
\end{pgfscope}%
\begin{pgfscope}%
\pgfpathrectangle{\pgfqpoint{0.618786in}{0.442177in}}{\pgfqpoint{2.054825in}{1.142908in}}%
\pgfusepath{clip}%
\pgfsetroundcap%
\pgfsetroundjoin%
\pgfsetlinewidth{0.501875pt}%
\definecolor{currentstroke}{rgb}{0.800000,0.800000,0.800000}%
\pgfsetstrokecolor{currentstroke}%
\pgfsetdash{}{0pt}%
\pgfpathmoveto{\pgfqpoint{0.618786in}{0.670758in}}%
\pgfpathlineto{\pgfqpoint{2.673611in}{0.670758in}}%
\pgfusepath{stroke}%
\end{pgfscope}%
\begin{pgfscope}%
\definecolor{textcolor}{rgb}{0.150000,0.150000,0.150000}%
\pgfsetstrokecolor{textcolor}%
\pgfsetfillcolor{textcolor}%
\pgftext[x=0.312126in, y=0.632496in, left, base]{\color{textcolor}\rmfamily\fontsize{8.000000}{9.600000}\selectfont 20\%}%
\end{pgfscope}%
\begin{pgfscope}%
\pgfpathrectangle{\pgfqpoint{0.618786in}{0.442177in}}{\pgfqpoint{2.054825in}{1.142908in}}%
\pgfusepath{clip}%
\pgfsetroundcap%
\pgfsetroundjoin%
\pgfsetlinewidth{0.501875pt}%
\definecolor{currentstroke}{rgb}{0.800000,0.800000,0.800000}%
\pgfsetstrokecolor{currentstroke}%
\pgfsetdash{}{0pt}%
\pgfpathmoveto{\pgfqpoint{0.618786in}{0.899340in}}%
\pgfpathlineto{\pgfqpoint{2.673611in}{0.899340in}}%
\pgfusepath{stroke}%
\end{pgfscope}%
\begin{pgfscope}%
\definecolor{textcolor}{rgb}{0.150000,0.150000,0.150000}%
\pgfsetstrokecolor{textcolor}%
\pgfsetfillcolor{textcolor}%
\pgftext[x=0.312126in, y=0.861078in, left, base]{\color{textcolor}\rmfamily\fontsize{8.000000}{9.600000}\selectfont 40\%}%
\end{pgfscope}%
\begin{pgfscope}%
\pgfpathrectangle{\pgfqpoint{0.618786in}{0.442177in}}{\pgfqpoint{2.054825in}{1.142908in}}%
\pgfusepath{clip}%
\pgfsetroundcap%
\pgfsetroundjoin%
\pgfsetlinewidth{0.501875pt}%
\definecolor{currentstroke}{rgb}{0.800000,0.800000,0.800000}%
\pgfsetstrokecolor{currentstroke}%
\pgfsetdash{}{0pt}%
\pgfpathmoveto{\pgfqpoint{0.618786in}{1.127922in}}%
\pgfpathlineto{\pgfqpoint{2.673611in}{1.127922in}}%
\pgfusepath{stroke}%
\end{pgfscope}%
\begin{pgfscope}%
\definecolor{textcolor}{rgb}{0.150000,0.150000,0.150000}%
\pgfsetstrokecolor{textcolor}%
\pgfsetfillcolor{textcolor}%
\pgftext[x=0.312126in, y=1.089660in, left, base]{\color{textcolor}\rmfamily\fontsize{8.000000}{9.600000}\selectfont 60\%}%
\end{pgfscope}%
\begin{pgfscope}%
\pgfpathrectangle{\pgfqpoint{0.618786in}{0.442177in}}{\pgfqpoint{2.054825in}{1.142908in}}%
\pgfusepath{clip}%
\pgfsetroundcap%
\pgfsetroundjoin%
\pgfsetlinewidth{0.501875pt}%
\definecolor{currentstroke}{rgb}{0.800000,0.800000,0.800000}%
\pgfsetstrokecolor{currentstroke}%
\pgfsetdash{}{0pt}%
\pgfpathmoveto{\pgfqpoint{0.618786in}{1.356504in}}%
\pgfpathlineto{\pgfqpoint{2.673611in}{1.356504in}}%
\pgfusepath{stroke}%
\end{pgfscope}%
\begin{pgfscope}%
\definecolor{textcolor}{rgb}{0.150000,0.150000,0.150000}%
\pgfsetstrokecolor{textcolor}%
\pgfsetfillcolor{textcolor}%
\pgftext[x=0.312126in, y=1.318241in, left, base]{\color{textcolor}\rmfamily\fontsize{8.000000}{9.600000}\selectfont 80\%}%
\end{pgfscope}%
\begin{pgfscope}%
\pgfpathrectangle{\pgfqpoint{0.618786in}{0.442177in}}{\pgfqpoint{2.054825in}{1.142908in}}%
\pgfusepath{clip}%
\pgfsetroundcap%
\pgfsetroundjoin%
\pgfsetlinewidth{0.501875pt}%
\definecolor{currentstroke}{rgb}{0.800000,0.800000,0.800000}%
\pgfsetstrokecolor{currentstroke}%
\pgfsetdash{}{0pt}%
\pgfpathmoveto{\pgfqpoint{0.618786in}{1.585085in}}%
\pgfpathlineto{\pgfqpoint{2.673611in}{1.585085in}}%
\pgfusepath{stroke}%
\end{pgfscope}%
\begin{pgfscope}%
\definecolor{textcolor}{rgb}{0.150000,0.150000,0.150000}%
\pgfsetstrokecolor{textcolor}%
\pgfsetfillcolor{textcolor}%
\pgftext[x=0.253113in, y=1.546823in, left, base]{\color{textcolor}\rmfamily\fontsize{8.000000}{9.600000}\selectfont 100\%}%
\end{pgfscope}%
\begin{pgfscope}%
\definecolor{textcolor}{rgb}{0.150000,0.150000,0.150000}%
\pgfsetstrokecolor{textcolor}%
\pgfsetfillcolor{textcolor}%
\pgftext[x=0.197557in,y=1.013631in,,bottom,rotate=90.000000]{\color{textcolor}\rmfamily\fontsize{10.000000}{12.000000}\selectfont Accuracy}%
\end{pgfscope}%
\begin{pgfscope}%
\pgfsetrectcap%
\pgfsetmiterjoin%
\pgfsetlinewidth{0.752812pt}%
\definecolor{currentstroke}{rgb}{0.700000,0.700000,0.700000}%
\pgfsetstrokecolor{currentstroke}%
\pgfsetdash{}{0pt}%
\pgfpathmoveto{\pgfqpoint{0.618786in}{0.442177in}}%
\pgfpathlineto{\pgfqpoint{0.618786in}{1.585085in}}%
\pgfusepath{stroke}%
\end{pgfscope}%
\begin{pgfscope}%
\pgfsetrectcap%
\pgfsetmiterjoin%
\pgfsetlinewidth{0.752812pt}%
\definecolor{currentstroke}{rgb}{0.700000,0.700000,0.700000}%
\pgfsetstrokecolor{currentstroke}%
\pgfsetdash{}{0pt}%
\pgfpathmoveto{\pgfqpoint{2.673611in}{0.442177in}}%
\pgfpathlineto{\pgfqpoint{2.673611in}{1.585085in}}%
\pgfusepath{stroke}%
\end{pgfscope}%
\begin{pgfscope}%
\pgfsetrectcap%
\pgfsetmiterjoin%
\pgfsetlinewidth{0.752812pt}%
\definecolor{currentstroke}{rgb}{0.700000,0.700000,0.700000}%
\pgfsetstrokecolor{currentstroke}%
\pgfsetdash{}{0pt}%
\pgfpathmoveto{\pgfqpoint{0.618786in}{0.442177in}}%
\pgfpathlineto{\pgfqpoint{2.673611in}{0.442177in}}%
\pgfusepath{stroke}%
\end{pgfscope}%
\begin{pgfscope}%
\pgfsetrectcap%
\pgfsetmiterjoin%
\pgfsetlinewidth{0.752812pt}%
\definecolor{currentstroke}{rgb}{0.700000,0.700000,0.700000}%
\pgfsetstrokecolor{currentstroke}%
\pgfsetdash{}{0pt}%
\pgfpathmoveto{\pgfqpoint{0.618786in}{1.585085in}}%
\pgfpathlineto{\pgfqpoint{2.673611in}{1.585085in}}%
\pgfusepath{stroke}%
\end{pgfscope}%
\begin{pgfscope}%
\pgfpathrectangle{\pgfqpoint{0.618786in}{0.442177in}}{\pgfqpoint{2.054825in}{1.142908in}}%
\pgfusepath{clip}%
\pgfsetroundcap%
\pgfsetroundjoin%
\pgfsetlinewidth{1.003750pt}%
\definecolor{currentstroke}{rgb}{0.003922,0.450980,0.698039}%
\pgfsetstrokecolor{currentstroke}%
\pgfsetdash{}{0pt}%
\pgfpathmoveto{\pgfqpoint{0.618786in}{1.449863in}}%
\pgfpathlineto{\pgfqpoint{0.648437in}{1.430876in}}%
\pgfpathlineto{\pgfqpoint{0.678088in}{1.401238in}}%
\pgfpathlineto{\pgfqpoint{0.707740in}{1.385030in}}%
\pgfpathlineto{\pgfqpoint{0.737391in}{1.363265in}}%
\pgfpathlineto{\pgfqpoint{0.767042in}{1.335479in}}%
\pgfpathlineto{\pgfqpoint{0.796693in}{1.310009in}}%
\pgfpathlineto{\pgfqpoint{0.826344in}{1.286855in}}%
\pgfpathlineto{\pgfqpoint{0.855995in}{1.261848in}}%
\pgfpathlineto{\pgfqpoint{0.885646in}{1.234525in}}%
\pgfpathlineto{\pgfqpoint{0.915298in}{1.209518in}}%
\pgfpathlineto{\pgfqpoint{0.944949in}{1.186364in}}%
\pgfpathlineto{\pgfqpoint{0.974600in}{1.166914in}}%
\pgfpathlineto{\pgfqpoint{1.004251in}{1.139592in}}%
\pgfpathlineto{\pgfqpoint{1.033902in}{1.119216in}}%
\pgfpathlineto{\pgfqpoint{1.063553in}{1.098840in}}%
\pgfpathlineto{\pgfqpoint{1.093205in}{1.077538in}}%
\pgfpathlineto{\pgfqpoint{1.122856in}{1.058088in}}%
\pgfpathlineto{\pgfqpoint{1.152507in}{1.038638in}}%
\pgfpathlineto{\pgfqpoint{1.182158in}{1.025208in}}%
\pgfpathlineto{\pgfqpoint{1.211809in}{1.007611in}}%
\pgfpathlineto{\pgfqpoint{1.241460in}{0.989087in}}%
\pgfpathlineto{\pgfqpoint{1.271111in}{0.975194in}}%
\pgfpathlineto{\pgfqpoint{1.300763in}{0.960376in}}%
\pgfpathlineto{\pgfqpoint{1.330414in}{0.944630in}}%
\pgfpathlineto{\pgfqpoint{1.360065in}{0.931664in}}%
\pgfpathlineto{\pgfqpoint{1.389716in}{0.923328in}}%
\pgfpathlineto{\pgfqpoint{1.419367in}{0.908046in}}%
\pgfpathlineto{\pgfqpoint{1.449018in}{0.895543in}}%
\pgfpathlineto{\pgfqpoint{1.478670in}{0.886281in}}%
\pgfpathlineto{\pgfqpoint{1.508321in}{0.876093in}}%
\pgfpathlineto{\pgfqpoint{1.537972in}{0.868220in}}%
\pgfpathlineto{\pgfqpoint{1.567623in}{0.861274in}}%
\pgfpathlineto{\pgfqpoint{1.597274in}{0.852012in}}%
\pgfpathlineto{\pgfqpoint{1.626925in}{0.846918in}}%
\pgfpathlineto{\pgfqpoint{1.656576in}{0.838583in}}%
\pgfpathlineto{\pgfqpoint{1.686228in}{0.829784in}}%
\pgfpathlineto{\pgfqpoint{1.715879in}{0.822838in}}%
\pgfpathlineto{\pgfqpoint{1.745530in}{0.815891in}}%
\pgfpathlineto{\pgfqpoint{1.775181in}{0.807556in}}%
\pgfpathlineto{\pgfqpoint{1.804832in}{0.798757in}}%
\pgfpathlineto{\pgfqpoint{1.834483in}{0.790421in}}%
\pgfpathlineto{\pgfqpoint{1.864135in}{0.783012in}}%
\pgfpathlineto{\pgfqpoint{1.893786in}{0.775602in}}%
\pgfpathlineto{\pgfqpoint{1.923437in}{0.770045in}}%
\pgfpathlineto{\pgfqpoint{1.953088in}{0.763562in}}%
\pgfpathlineto{\pgfqpoint{1.982739in}{0.757542in}}%
\pgfpathlineto{\pgfqpoint{2.012390in}{0.750595in}}%
\pgfpathlineto{\pgfqpoint{2.042041in}{0.746891in}}%
\pgfpathlineto{\pgfqpoint{2.071693in}{0.739018in}}%
\pgfpathlineto{\pgfqpoint{2.101344in}{0.735313in}}%
\pgfpathlineto{\pgfqpoint{2.130995in}{0.730682in}}%
\pgfpathlineto{\pgfqpoint{2.160646in}{0.722810in}}%
\pgfpathlineto{\pgfqpoint{2.190297in}{0.717716in}}%
\pgfpathlineto{\pgfqpoint{2.219948in}{0.711233in}}%
\pgfpathlineto{\pgfqpoint{2.249600in}{0.705676in}}%
\pgfpathlineto{\pgfqpoint{2.279251in}{0.699655in}}%
\pgfpathlineto{\pgfqpoint{2.308902in}{0.692246in}}%
\pgfpathlineto{\pgfqpoint{2.338553in}{0.686226in}}%
\pgfpathlineto{\pgfqpoint{2.368204in}{0.679742in}}%
\pgfpathlineto{\pgfqpoint{2.397855in}{0.676038in}}%
\pgfpathlineto{\pgfqpoint{2.427506in}{0.669554in}}%
\pgfpathlineto{\pgfqpoint{2.457158in}{0.665387in}}%
\pgfpathlineto{\pgfqpoint{2.486809in}{0.660293in}}%
\pgfpathlineto{\pgfqpoint{2.516460in}{0.657051in}}%
\pgfpathlineto{\pgfqpoint{2.546111in}{0.652883in}}%
\pgfpathlineto{\pgfqpoint{2.575762in}{0.648715in}}%
\pgfpathlineto{\pgfqpoint{2.605413in}{0.645011in}}%
\pgfpathlineto{\pgfqpoint{2.635065in}{0.641769in}}%
\pgfpathlineto{\pgfqpoint{2.664716in}{0.637601in}}%
\pgfpathlineto{\pgfqpoint{2.675278in}{0.635952in}}%
\pgfusepath{stroke}%
\end{pgfscope}%
\end{pgfpicture}%
\makeatother%
\endgroup%

%% file: figures/experiments/attacks/gcn_attack.pgf
\begingroup%
\makeatletter%
\begin{pgfpicture}%
\pgfpathrectangle{\pgfpointorigin}{\pgfqpoint{2.750000in}{1.699593in}}%
\pgfusepath{use as bounding box, clip}%
\begin{pgfscope}%
\pgfsetbuttcap%
\pgfsetmiterjoin%
\definecolor{currentfill}{rgb}{1.000000,1.000000,1.000000}%
\pgfsetfillcolor{currentfill}%
\pgfsetlinewidth{0.000000pt}%
\definecolor{currentstroke}{rgb}{1.000000,1.000000,1.000000}%
\pgfsetstrokecolor{currentstroke}%
\pgfsetdash{}{0pt}%
\pgfpathmoveto{\pgfqpoint{0.000000in}{0.000000in}}%
\pgfpathlineto{\pgfqpoint{2.750000in}{0.000000in}}%
\pgfpathlineto{\pgfqpoint{2.750000in}{1.699593in}}%
\pgfpathlineto{\pgfqpoint{0.000000in}{1.699593in}}%
\pgfpathlineto{\pgfqpoint{0.000000in}{0.000000in}}%
\pgfpathclose%
\pgfusepath{fill}%
\end{pgfscope}%
\begin{pgfscope}%
\pgfsetbuttcap%
\pgfsetmiterjoin%
\definecolor{currentfill}{rgb}{1.000000,1.000000,1.000000}%
\pgfsetfillcolor{currentfill}%
\pgfsetlinewidth{0.000000pt}%
\definecolor{currentstroke}{rgb}{0.000000,0.000000,0.000000}%
\pgfsetstrokecolor{currentstroke}%
\pgfsetstrokeopacity{0.000000}%
\pgfsetdash{}{0pt}%
\pgfpathmoveto{\pgfqpoint{0.618786in}{0.442177in}}%
\pgfpathlineto{\pgfqpoint{2.614803in}{0.442177in}}%
\pgfpathlineto{\pgfqpoint{2.614803in}{1.585085in}}%
\pgfpathlineto{\pgfqpoint{0.618786in}{1.585085in}}%
\pgfpathlineto{\pgfqpoint{0.618786in}{0.442177in}}%
\pgfpathclose%
\pgfusepath{fill}%
\end{pgfscope}%
\begin{pgfscope}%
\pgfpathrectangle{\pgfqpoint{0.618786in}{0.442177in}}{\pgfqpoint{1.996017in}{1.142908in}}%
\pgfusepath{clip}%
\pgfsetroundcap%
\pgfsetroundjoin%
\pgfsetlinewidth{0.501875pt}%
\definecolor{currentstroke}{rgb}{0.800000,0.800000,0.800000}%
\pgfsetstrokecolor{currentstroke}%
\pgfsetdash{}{0pt}%
\pgfpathmoveto{\pgfqpoint{0.618786in}{0.442177in}}%
\pgfpathlineto{\pgfqpoint{0.618786in}{1.585085in}}%
\pgfusepath{stroke}%
\end{pgfscope}%
\begin{pgfscope}%
\definecolor{textcolor}{rgb}{0.150000,0.150000,0.150000}%
\pgfsetstrokecolor{textcolor}%
\pgfsetfillcolor{textcolor}%
\pgftext[x=0.618786in,y=0.351899in,,top]{\color{textcolor}\rmfamily\fontsize{8.000000}{9.600000}\selectfont \(\displaystyle {0}\)}%
\end{pgfscope}%
\begin{pgfscope}%
\pgfpathrectangle{\pgfqpoint{0.618786in}{0.442177in}}{\pgfqpoint{1.996017in}{1.142908in}}%
\pgfusepath{clip}%
\pgfsetroundcap%
\pgfsetroundjoin%
\pgfsetlinewidth{0.501875pt}%
\definecolor{currentstroke}{rgb}{0.800000,0.800000,0.800000}%
\pgfsetstrokecolor{currentstroke}%
\pgfsetdash{}{0pt}%
\pgfpathmoveto{\pgfqpoint{1.284125in}{0.442177in}}%
\pgfpathlineto{\pgfqpoint{1.284125in}{1.585085in}}%
\pgfusepath{stroke}%
\end{pgfscope}%
\begin{pgfscope}%
\definecolor{textcolor}{rgb}{0.150000,0.150000,0.150000}%
\pgfsetstrokecolor{textcolor}%
\pgfsetfillcolor{textcolor}%
\pgftext[x=1.284125in,y=0.351899in,,top]{\color{textcolor}\rmfamily\fontsize{8.000000}{9.600000}\selectfont \(\displaystyle {10}\)}%
\end{pgfscope}%
\begin{pgfscope}%
\pgfpathrectangle{\pgfqpoint{0.618786in}{0.442177in}}{\pgfqpoint{1.996017in}{1.142908in}}%
\pgfusepath{clip}%
\pgfsetroundcap%
\pgfsetroundjoin%
\pgfsetlinewidth{0.501875pt}%
\definecolor{currentstroke}{rgb}{0.800000,0.800000,0.800000}%
\pgfsetstrokecolor{currentstroke}%
\pgfsetdash{}{0pt}%
\pgfpathmoveto{\pgfqpoint{1.949464in}{0.442177in}}%
\pgfpathlineto{\pgfqpoint{1.949464in}{1.585085in}}%
\pgfusepath{stroke}%
\end{pgfscope}%
\begin{pgfscope}%
\definecolor{textcolor}{rgb}{0.150000,0.150000,0.150000}%
\pgfsetstrokecolor{textcolor}%
\pgfsetfillcolor{textcolor}%
\pgftext[x=1.949464in,y=0.351899in,,top]{\color{textcolor}\rmfamily\fontsize{8.000000}{9.600000}\selectfont \(\displaystyle {20}\)}%
\end{pgfscope}%
\begin{pgfscope}%
\pgfpathrectangle{\pgfqpoint{0.618786in}{0.442177in}}{\pgfqpoint{1.996017in}{1.142908in}}%
\pgfusepath{clip}%
\pgfsetroundcap%
\pgfsetroundjoin%
\pgfsetlinewidth{0.501875pt}%
\definecolor{currentstroke}{rgb}{0.800000,0.800000,0.800000}%
\pgfsetstrokecolor{currentstroke}%
\pgfsetdash{}{0pt}%
\pgfpathmoveto{\pgfqpoint{2.614803in}{0.442177in}}%
\pgfpathlineto{\pgfqpoint{2.614803in}{1.585085in}}%
\pgfusepath{stroke}%
\end{pgfscope}%
\begin{pgfscope}%
\definecolor{textcolor}{rgb}{0.150000,0.150000,0.150000}%
\pgfsetstrokecolor{textcolor}%
\pgfsetfillcolor{textcolor}%
\pgftext[x=2.614803in,y=0.351899in,,top]{\color{textcolor}\rmfamily\fontsize{8.000000}{9.600000}\selectfont \(\displaystyle {30}\)}%
\end{pgfscope}%
\begin{pgfscope}%
\definecolor{textcolor}{rgb}{0.150000,0.150000,0.150000}%
\pgfsetstrokecolor{textcolor}%
\pgfsetfillcolor{textcolor}%
\pgftext[x=1.616794in,y=0.198219in,,top]{\color{textcolor}\rmfamily\fontsize{10.000000}{12.000000}\selectfont Edit distance \(\displaystyle \epsilon\)}%
\end{pgfscope}%
\begin{pgfscope}%
\pgfpathrectangle{\pgfqpoint{0.618786in}{0.442177in}}{\pgfqpoint{1.996017in}{1.142908in}}%
\pgfusepath{clip}%
\pgfsetroundcap%
\pgfsetroundjoin%
\pgfsetlinewidth{0.501875pt}%
\definecolor{currentstroke}{rgb}{0.800000,0.800000,0.800000}%
\pgfsetstrokecolor{currentstroke}%
\pgfsetdash{}{0pt}%
\pgfpathmoveto{\pgfqpoint{0.618786in}{0.442177in}}%
\pgfpathlineto{\pgfqpoint{2.614803in}{0.442177in}}%
\pgfusepath{stroke}%
\end{pgfscope}%
\begin{pgfscope}%
\definecolor{textcolor}{rgb}{0.150000,0.150000,0.150000}%
\pgfsetstrokecolor{textcolor}%
\pgfsetfillcolor{textcolor}%
\pgftext[x=0.371139in, y=0.403915in, left, base]{\color{textcolor}\rmfamily\fontsize{8.000000}{9.600000}\selectfont 0\%}%
\end{pgfscope}%
\begin{pgfscope}%
\pgfpathrectangle{\pgfqpoint{0.618786in}{0.442177in}}{\pgfqpoint{1.996017in}{1.142908in}}%
\pgfusepath{clip}%
\pgfsetroundcap%
\pgfsetroundjoin%
\pgfsetlinewidth{0.501875pt}%
\definecolor{currentstroke}{rgb}{0.800000,0.800000,0.800000}%
\pgfsetstrokecolor{currentstroke}%
\pgfsetdash{}{0pt}%
\pgfpathmoveto{\pgfqpoint{0.618786in}{0.670758in}}%
\pgfpathlineto{\pgfqpoint{2.614803in}{0.670758in}}%
\pgfusepath{stroke}%
\end{pgfscope}%
\begin{pgfscope}%
\definecolor{textcolor}{rgb}{0.150000,0.150000,0.150000}%
\pgfsetstrokecolor{textcolor}%
\pgfsetfillcolor{textcolor}%
\pgftext[x=0.312126in, y=0.632496in, left, base]{\color{textcolor}\rmfamily\fontsize{8.000000}{9.600000}\selectfont 20\%}%
\end{pgfscope}%
\begin{pgfscope}%
\pgfpathrectangle{\pgfqpoint{0.618786in}{0.442177in}}{\pgfqpoint{1.996017in}{1.142908in}}%
\pgfusepath{clip}%
\pgfsetroundcap%
\pgfsetroundjoin%
\pgfsetlinewidth{0.501875pt}%
\definecolor{currentstroke}{rgb}{0.800000,0.800000,0.800000}%
\pgfsetstrokecolor{currentstroke}%
\pgfsetdash{}{0pt}%
\pgfpathmoveto{\pgfqpoint{0.618786in}{0.899340in}}%
\pgfpathlineto{\pgfqpoint{2.614803in}{0.899340in}}%
\pgfusepath{stroke}%
\end{pgfscope}%
\begin{pgfscope}%
\definecolor{textcolor}{rgb}{0.150000,0.150000,0.150000}%
\pgfsetstrokecolor{textcolor}%
\pgfsetfillcolor{textcolor}%
\pgftext[x=0.312126in, y=0.861078in, left, base]{\color{textcolor}\rmfamily\fontsize{8.000000}{9.600000}\selectfont 40\%}%
\end{pgfscope}%
\begin{pgfscope}%
\pgfpathrectangle{\pgfqpoint{0.618786in}{0.442177in}}{\pgfqpoint{1.996017in}{1.142908in}}%
\pgfusepath{clip}%
\pgfsetroundcap%
\pgfsetroundjoin%
\pgfsetlinewidth{0.501875pt}%
\definecolor{currentstroke}{rgb}{0.800000,0.800000,0.800000}%
\pgfsetstrokecolor{currentstroke}%
\pgfsetdash{}{0pt}%
\pgfpathmoveto{\pgfqpoint{0.618786in}{1.127922in}}%
\pgfpathlineto{\pgfqpoint{2.614803in}{1.127922in}}%
\pgfusepath{stroke}%
\end{pgfscope}%
\begin{pgfscope}%
\definecolor{textcolor}{rgb}{0.150000,0.150000,0.150000}%
\pgfsetstrokecolor{textcolor}%
\pgfsetfillcolor{textcolor}%
\pgftext[x=0.312126in, y=1.089660in, left, base]{\color{textcolor}\rmfamily\fontsize{8.000000}{9.600000}\selectfont 60\%}%
\end{pgfscope}%
\begin{pgfscope}%
\pgfpathrectangle{\pgfqpoint{0.618786in}{0.442177in}}{\pgfqpoint{1.996017in}{1.142908in}}%
\pgfusepath{clip}%
\pgfsetroundcap%
\pgfsetroundjoin%
\pgfsetlinewidth{0.501875pt}%
\definecolor{currentstroke}{rgb}{0.800000,0.800000,0.800000}%
\pgfsetstrokecolor{currentstroke}%
\pgfsetdash{}{0pt}%
\pgfpathmoveto{\pgfqpoint{0.618786in}{1.356504in}}%
\pgfpathlineto{\pgfqpoint{2.614803in}{1.356504in}}%
\pgfusepath{stroke}%
\end{pgfscope}%
\begin{pgfscope}%
\definecolor{textcolor}{rgb}{0.150000,0.150000,0.150000}%
\pgfsetstrokecolor{textcolor}%
\pgfsetfillcolor{textcolor}%
\pgftext[x=0.312126in, y=1.318241in, left, base]{\color{textcolor}\rmfamily\fontsize{8.000000}{9.600000}\selectfont 80\%}%
\end{pgfscope}%
\begin{pgfscope}%
\pgfpathrectangle{\pgfqpoint{0.618786in}{0.442177in}}{\pgfqpoint{1.996017in}{1.142908in}}%
\pgfusepath{clip}%
\pgfsetroundcap%
\pgfsetroundjoin%
\pgfsetlinewidth{0.501875pt}%
\definecolor{currentstroke}{rgb}{0.800000,0.800000,0.800000}%
\pgfsetstrokecolor{currentstroke}%
\pgfsetdash{}{0pt}%
\pgfpathmoveto{\pgfqpoint{0.618786in}{1.585085in}}%
\pgfpathlineto{\pgfqpoint{2.614803in}{1.585085in}}%
\pgfusepath{stroke}%
\end{pgfscope}%
\begin{pgfscope}%
\definecolor{textcolor}{rgb}{0.150000,0.150000,0.150000}%
\pgfsetstrokecolor{textcolor}%
\pgfsetfillcolor{textcolor}%
\pgftext[x=0.253113in, y=1.546823in, left, base]{\color{textcolor}\rmfamily\fontsize{8.000000}{9.600000}\selectfont 100\%}%
\end{pgfscope}%
\begin{pgfscope}%
\definecolor{textcolor}{rgb}{0.150000,0.150000,0.150000}%
\pgfsetstrokecolor{textcolor}%
\pgfsetfillcolor{textcolor}%
\pgftext[x=0.197557in,y=1.013631in,,bottom,rotate=90.000000]{\color{textcolor}\rmfamily\fontsize{10.000000}{12.000000}\selectfont Accuracy}%
\end{pgfscope}%
\begin{pgfscope}%
\pgfsetrectcap%
\pgfsetmiterjoin%
\pgfsetlinewidth{0.752812pt}%
\definecolor{currentstroke}{rgb}{0.700000,0.700000,0.700000}%
\pgfsetstrokecolor{currentstroke}%
\pgfsetdash{}{0pt}%
\pgfpathmoveto{\pgfqpoint{0.618786in}{0.442177in}}%
\pgfpathlineto{\pgfqpoint{0.618786in}{1.585085in}}%
\pgfusepath{stroke}%
\end{pgfscope}%
\begin{pgfscope}%
\pgfsetrectcap%
\pgfsetmiterjoin%
\pgfsetlinewidth{0.752812pt}%
\definecolor{currentstroke}{rgb}{0.700000,0.700000,0.700000}%
\pgfsetstrokecolor{currentstroke}%
\pgfsetdash{}{0pt}%
\pgfpathmoveto{\pgfqpoint{2.614803in}{0.442177in}}%
\pgfpathlineto{\pgfqpoint{2.614803in}{1.585085in}}%
\pgfusepath{stroke}%
\end{pgfscope}%
\begin{pgfscope}%
\pgfsetrectcap%
\pgfsetmiterjoin%
\pgfsetlinewidth{0.752812pt}%
\definecolor{currentstroke}{rgb}{0.700000,0.700000,0.700000}%
\pgfsetstrokecolor{currentstroke}%
\pgfsetdash{}{0pt}%
\pgfpathmoveto{\pgfqpoint{0.618786in}{0.442177in}}%
\pgfpathlineto{\pgfqpoint{2.614803in}{0.442177in}}%
\pgfusepath{stroke}%
\end{pgfscope}%
\begin{pgfscope}%
\pgfsetrectcap%
\pgfsetmiterjoin%
\pgfsetlinewidth{0.752812pt}%
\definecolor{currentstroke}{rgb}{0.700000,0.700000,0.700000}%
\pgfsetstrokecolor{currentstroke}%
\pgfsetdash{}{0pt}%
\pgfpathmoveto{\pgfqpoint{0.618786in}{1.585085in}}%
\pgfpathlineto{\pgfqpoint{2.614803in}{1.585085in}}%
\pgfusepath{stroke}%
\end{pgfscope}%
\begin{pgfscope}%
\pgfpathrectangle{\pgfqpoint{0.618786in}{0.442177in}}{\pgfqpoint{1.996017in}{1.142908in}}%
\pgfusepath{clip}%
\pgfsetroundcap%
\pgfsetroundjoin%
\pgfsetlinewidth{1.003750pt}%
\definecolor{currentstroke}{rgb}{0.003922,0.450980,0.698039}%
\pgfsetstrokecolor{currentstroke}%
\pgfsetdash{}{0pt}%
\pgfpathmoveto{\pgfqpoint{0.618786in}{1.286530in}}%
\pgfpathlineto{\pgfqpoint{0.685320in}{1.207226in}}%
\pgfpathlineto{\pgfqpoint{0.751854in}{1.120147in}}%
\pgfpathlineto{\pgfqpoint{0.818388in}{1.041880in}}%
\pgfpathlineto{\pgfqpoint{0.884922in}{0.969314in}}%
\pgfpathlineto{\pgfqpoint{0.951456in}{0.904523in}}%
\pgfpathlineto{\pgfqpoint{1.017989in}{0.853209in}}%
\pgfpathlineto{\pgfqpoint{1.084523in}{0.801895in}}%
\pgfpathlineto{\pgfqpoint{1.151057in}{0.758356in}}%
\pgfpathlineto{\pgfqpoint{1.217591in}{0.716889in}}%
\pgfpathlineto{\pgfqpoint{1.284125in}{0.684753in}}%
\pgfpathlineto{\pgfqpoint{1.350659in}{0.650025in}}%
\pgfpathlineto{\pgfqpoint{1.417193in}{0.627219in}}%
\pgfpathlineto{\pgfqpoint{1.483727in}{0.602339in}}%
\pgfpathlineto{\pgfqpoint{1.550261in}{0.576942in}}%
\pgfpathlineto{\pgfqpoint{1.616794in}{0.559318in}}%
\pgfpathlineto{\pgfqpoint{1.683328in}{0.547397in}}%
\pgfpathlineto{\pgfqpoint{1.749862in}{0.537030in}}%
\pgfpathlineto{\pgfqpoint{1.816396in}{0.526664in}}%
\pgfpathlineto{\pgfqpoint{1.882930in}{0.515779in}}%
\pgfpathlineto{\pgfqpoint{1.949464in}{0.506449in}}%
\pgfpathlineto{\pgfqpoint{2.015998in}{0.498156in}}%
\pgfpathlineto{\pgfqpoint{2.082532in}{0.489344in}}%
\pgfpathlineto{\pgfqpoint{2.149066in}{0.485198in}}%
\pgfpathlineto{\pgfqpoint{2.215600in}{0.477423in}}%
\pgfpathlineto{\pgfqpoint{2.282133in}{0.471203in}}%
\pgfpathlineto{\pgfqpoint{2.348667in}{0.468611in}}%
\pgfpathlineto{\pgfqpoint{2.415201in}{0.467575in}}%
\pgfpathlineto{\pgfqpoint{2.481735in}{0.463428in}}%
\pgfpathlineto{\pgfqpoint{2.548269in}{0.462392in}}%
\pgfpathlineto{\pgfqpoint{2.614803in}{0.461873in}}%
\pgfpathlineto{\pgfqpoint{2.616470in}{0.461847in}}%
\pgfusepath{stroke}%
\end{pgfscope}%
\end{pgfpicture}%
\makeatother%
\endgroup%

%% file: appendix/experimental_setup.tex
\section{Full experimental setup}
\label{appendix:experimental_setup}

\subsection{Point cloud classification}
For our experiments on point cloud classification, we  replicate the  experimental setup from~\cite{Schuchardt2022},
where randomized smoothing was also applied to point cloud classification.
We just use different base models, which are not rotation and translation invariant.
Each experiment (including training) is repeated $5$ times using different random seeds. We report the mean and standard deviation in certificate strength across all seeds.

\subsubsection{Data}
All experiments are performed on ModelNet\cite{Wu2015} which consists of $9843$ training samples and $2468$ test samples from $40$ classes.
We apply the same preprocessing steps as in~\cite{Qi2017} to transform the CAD models into point clouds consisting of $1024$ points, i.e.\ random sampling on the surface of the objects and normalization into the unit sphere.
We use $20\%$ of the training data for validation. All certificates are evaluated on the test data.

\subsubsection{Models}
We use two different models, PointNet~\cite{Qi2017} and DGCNN~\cite{Wang2019}.

For PointNet, we use  three linear layers ($64, 128, 1024$ neurons) before pooling and three layers ($512, 256$, $40$ neurons) after pooling.
Before feeding input point clouds into the model, we multiply them with a matrix predicted by a T-Net, i.e.\ another PointNet with three linear layers ($64, 128, 1024$ neurons) before pooling and three linear layers ($512, 256, 3 \cdot 3$ neurons) after pooling.
We use batch normalization ($\epsilon = 1e-5, \mathrm{momentum}=0.1$) for all layers, except the output layer.
We use dropout ($p=0.4$) for the second-to-last layer.

For DGCNN, we use four dynamic graph convolution layers ($64, 64, 64, 128$ neurons) with k-nearest neighbor graph construction ($k=20$).
The outputs of all four layers are concatenated, passed through a linear layer ($1024$ neurons), then max-pooled across the node dimension and finally passed through three more linear layers ($512, 256, 40$ neurons). We use batch normalization for all layers ($\epsilon = 1e-5, \mathrm{momentum}=0.1$).
We use dropout ($p=0.4$) for the second-to-last and third-to-last layer.
Before feeding input point clouds into the model, we multiply them with a $3 \times 3$ matrix predicted by the same T-Net architecture we described for PointNet.

\subsubsection{Training}
We use the same training parameters as in~\cite{Schuchardt2022}.
We train for $200$ (PointNet) or $400$ (DGCNN) epochs using Adam ($\beta_1 = 0.9, \beta_2 = 0.99$, $\epsilon=1e-8$, $\mathrm{weight\_decay} = 1e^-4$) with a learning rate of $0.001$ and exponential weight decay with factor $0.7$ every $20$ (PointNet) or $40$ (DGCNN) epochs.
We use a batch size of $128$ for PointNet and $32$ for DGCNN.
We add the T-Net loss proposed by~\cite{Qi2017} to the cross entropy loss with a weight of $0.0001$.
The data is randomly augmented via scaling by a factor uniformly sampled from $[0.8, 1.25]$. We additionally add isotropic Gaussian noise with the same standard deviation we use for randomized smoothing ($0.05$, $0.1$, $0.15$, $0.2$ or $0.25$).

\subsubsection{Certification parameters}
We perform randomized smoothing with isotropic Gaussian noise ($\sigma \in \{0.05$, $0.1$, $0.15$, $0.2$ or $0.25\}$) and majority voting~\cite{Cohen2019}.
We use $1{,}000$ Monte Carlo samples for prediction and $10{,}000$ Monte Carlo samples for certification.
We compute the Clopper-Pearson confidence intervals for  class probabilities with $\alpha=0.001$,  i.e.\ all guarantees hold with high probability $99.9\%$.

\subsubsection{Computational resources}
All experiments involving PointNet were performed on a Xeon E5-2630 v4 CPU @ $\SI{2.20}{\giga\hertz}$ with a NVIDA GTX 1080TI GPU.
The average time for training a model was $\SI{21.1}{\minute}$.
The average time for certifying the robustness of a single model on the entire test set for $50$ different budgets $\epsilon$ was $\SI{81.7}{\minute}$.

All experiments involving DGCNN were performed on an AMD EPYC 7543 CPU @ 2.80GHz  with a NVIDA A100 ($\SI{40}{\giga\byte}$) GPU.
The average time for training a model was $\SI{161.9}{\minute}$.
The average time for certifying the robustness of a single model on the entire test set for $50$ different budgets $\epsilon$ was $\SI{601.7}{\minute}$.

\subsection{Molecular force prediction}
Like with point cloud classification, we train and certify $5$ different models using different random seeds on a common benchmark dataset.
We report the mean and standard deviation in certificate strength across all seeds.

\subsubsection{Data}
We use the original (not the revised) MD17~\cite{Chmiela2017} collection of datasets.
It consists of $8$ different datasets, each consisting of a large number (between $133\,770$ and $993\,237$) of spatial configurations  of a specific molecule.
For each dataset, we use $1000$ randomly chosen configurations for training, $1000$ for validation and $1000$ for evaluating the robustness guarantees.
We use the same data split for all experiments.

\subsubsection{Models}
We use DimeNet++~\cite{Gasteiger2020}, SchNet~\cite{Schutt2018} and SphereNet~\cite{Liu2022} as our base model. All models predict atomic energies. Force are calculated via automatic differentiation w.r.t.\ the input coordinates.

For DimeNet++, we use the default model parameters for MD17 force predictions specified in~\cite{Gasteiger2020}.. There are $4$ layers with $128$ hidden channels. The triplet embedding size is $64$. The basis embedding size is $8$. The output embedding size is $256$. The number of basis functions is $8$ (bilinear), $7$ (spherical), $6$ (radial). The cutoff radius for graph construction is \SI{5}{\angstrom}. The number of residual layers before the skip connection is $1$. The number of residual layers after the skip connection is $2$.

For SchNet, we use $128$ hidden channels and $128$ filters. We set the number of interaction blocks to $6$ and use $50$ Gaussians. The cutoff distance is \SI{10}{\angstrom}, with a maximum number of $32$ neighbors. We use addition as our global readout function.

For SphereNet, we use  $4$ layers with $128$ hidden channels. The triplet embedding size is $64$. The basis embedding size is $8$ for distance, angle and torsion. The output embedding size is $256$. The number of basis functions is $8$ (bilinear), $7$ (spherical), $6$ (radial). The cutoff radius for graph construction is \SI{5}{\angstrom}. 
We use swish activation functions. The number of residual layers before the skip connection is $1$. The number of residual layers after the skip connection is $2$.

\subsubsection{Training parameters}
To train DimeNet++ and Schnet, we use ADAM with a learning rate of $0.001$. We use linear warmup with exponential decay as our learning rate scheduler ($1000$ warmup steps, $400{,}000$ decay steps, decay rate of $0.01$) and train until convergence with a patience of $50$ epochs and a convergence threshold of $10{^-4}$.

To train SphereNet, we use ADAM with a learning rate of $0.001$. We use ``reduce on plateau` as our learning rate scheduler (decay factor of $0.8$, patience of $80$ epochs, convergence threshold of $10^-4$, cooldown of $10$ epochs) and train until convergence with a patience of $50$ epochs and a convergence threshold of $10^{-4}$.

We do not add randomized smoothing noise during the training process.

\subsubsection{Certification parameters}
We perform randomized smoothing with isotropic Gaussian noise ($\sigma \in \{\SI{1}{\femto\meter}, \SI{10}{\femto\meter}, \SI{100}{\femto\meter} \}$) and center smoothing~\cite{Kumar2021} with default parameters $\alpha_1=0.005, \alpha_2=0.005, \Delta = 0.05$,  i.e.\ all guarantees hold with high probability $99\%$.
We use $10{,}000$ samples for prediction, $10{,}000$ samples to test for abstention and $10{,}000$ samples for certification.
In our experiments the smoothed model never abstained.

\subsubsection{Computational resources}
All experiments for molecular force prediction  were performed on a Xeon E5-2630 v4 CPU @ $\SI{2.20}{\giga\hertz}$ with a NVIDA GTX 1080TI GPU.
We trained $4$ models simultaneously.
The average time for training a DimeNet++ model was  $\SI{28.9}{\hour}$.
The average time for certifying the robustness of a single DimeNet++ model on $1000$ molecule configurations for $1000$ different budgets $\epsilon$ was $\SI{10.49}{\hour}$.

\subsection{Node and graph classification}
In~\cref{section:graph_edit_certificates} we generalized six approaches for  proving the robustness of graph neural networks with respect to graph edit distances.
While the approaches differ, we mostly use the same datasets and models.
As before, we perform each experiment (including training) with $5$ different random seeds
and report the mean and standard deviation in certificate strength across all seeds.

\subsubsection{Data}
We use two standard node classification datasets, Cora-ML~\cite{McCallum2000,Bojchevski2018} ($2{,}708$ nodes, $10{,}556$ edges, $1{,}433$ features, $7$ classes) and
Citeseer~\cite{Sen2008}($3{,}327$ nodes, $9{,}104$ edges, $3{,}703$ features, $6$ classes), and two graph classification sets,
PROTEINS ($1{,}113$ graphs, $3$ features, $2$ classes) and MUTAG ($188$ graphs, $7$ features, $2$ classes), which are part of the TUDataset~\cite{Morris2020}.
These datasets were also used in the papers that proposed the original certification procedures.
With Cora-ML and Citeseer, we follow the procedure from~\cite{Bojchevski2020} and use $20$ nodes per class for training, $20$ nodes per class for validation and the remainder for evaluating the certificates.
With PROTEINS and MUTAG, we use $30\%$ of graphs for training, $20\%$ for validation and the remainder for evaluating the certificates.

\subsubsection{Models}
The main model we use for node classification is a standard $2$-layer graph convolutional network with $\mathrm{ReLU}$ nonlinearities and $32$ hidden features.
We insert self-loops and perform degree normalization via $\mD^{-1 \mathbin{/} 2} \mA \mD^{-1 \mathbin{/} 2}$. 

For node classification, we use the architecture described in~\cite{Jin2020}, i.e.\ a single layer GCN with $64$ neurons, followed by a linear layer and max-pooling.
We also insert self-loops, but perform degree normalization via $\mD^{-1} \mA$.

With sparsity-aware randomized smoothing, we additionally use graph attention networks~\cite{Velickovic2018} and APPNP \citep{Bojchevski2018}. We implement all GNNs with two-layers and 32 hidden dimensions, except GAT for which we implement 8 hidden dimensions and 8 attention heads. For APPNP we further set k\_hops=$10$ and teleport probability $\alpha=0.15$.

For the experiments with certificates for the \(\pi\)-PPNP architecture we set teleport probability $\alpha= 0.15$ and use a hidden dimension of 64.

\subsubsection{Training parameters}
We train models for a maximum of $1{,}000$ (\(\pi\)-PPNP), $3{,}000$ (other node classifiers) or $200$ (graph classifiers) epochs with Adam ($lr=1e-3$, $\beta_1 = 0.9, \beta_2 = 0.99$, $\epsilon=1e-8$) and cross entropy loss.
For node classification we additionally use weight decay $ 5e-4$ and early stopping after $50$ epochs.
For node-classification we perform full-batch gradient descent. For graph classification we use minibatches of size $20$.
For the sparsity-ware randomized smoothing experiments we use a dropout of \(0.5\) on the hidden node representations after the first graph convolution for GCN and GAT, and additionally on the attention coefficients for GAT.  For randomized smoothing we additionally add sparsity-aware noise to the training data. We use the same flip probabilities as for certification.

\subsubsection{Certification parameters}
For interval bound propagation and the convex outer adversarial polytope method,  we set the nodewise local budgets $\rho_1,\dots,\rho_N$ to $1\%$ of input features, just like in~\cite{Zuegner2019cert}.

For our  generalization of the graph classification certificate from, we set the local budget $\rho_n$ of node $n$ with degree $d_n$ to 
$\min(0, d_n -\max_m d_m + 3)$, just like in~\cite{Jin2020}.
We additionally enforce symmetry of the adjacency matrix via Lagrange dualization (see~\cref{appendix:linearization_and_dualization}) and perform $200$ alternating optimization steps using the default parameters from~\cite{Jin2020}.

We perform randomized smoothing with sparse noise ($p_\mX^+=0.001$, $p_\mX^-=0.8$, $p_\mA^+=0$, $p_\mA^-=0$, or $p_\mX^+=0$, $p_\mX^-=0$, $p_\mA^+=0.001$, $p_\mA^-=0.8$) and majority voting~\cite{Bojchevski2020}.
For node classification we use $1{,}000$ Monte-Carlo samples for prediction and $1{,}000{,}000$ Monte-Carlo samples for certification.
For graph classification we use $1{,}000$ Monte-Carlo samples for prediction and $10{,}000$ Monte-Carlo samples for certification.
We compute the Clopper-Pearson confidence intervals for  class probabilities with $\alpha=0.01$,  i.e.\ all guarantees hold with high probability $99\%$.

\subsubsection{Computational resources}

All experiments  were performed on a Xeon E5-2630 v4 CPU @ $\SI{2.20}{\giga\hertz}$ with a NVIDA GTX 1080TI GPU.
With interval bound propagation, the average time of training and certifying a model for $100$ budgets $\epsilon$ was $\SI{1.2}{\min}$ on Cora-ML and $\SI{3.0}{\min}$ on Citeseer.
With the convex outer adversarial polytope method, the average time of training and certifying a model for $100$ budgets $\epsilon$ was $\SI{9.7}{\min}$ on Cora-ML and $\SI{14.4}{\min}$ on Citeseer. 
With the linearization and dualization method for graph classifiers, the average time of the average time of training and certifying a model for $100$ budgets $\epsilon$ on the entire test set  was $\SI{1.2}{\hour}$ on MUTAG and $\SI{48.1}{\hour}$ on PROTEINS. 
With the policy iteration method for \(\pi\)-PPNP, the average time for training and certifying a model for $10$ attack strengths was $\SI{126.1}{\min}$.
With sparsity-aware randomized smoothing, the average time for training and certifying a model for $1000$ budgets  $\epsilon$ was $\SI{267.7}{\min}$.

\subsection{Third-party assets}
Since we extend various existing methods for proving the robustness of graph neural networks, we naturally built upon their reference implementations. 
For interval bound propagation and convex outer adversarial polytopes, we extend the reference implementation from~\cite{Zuegner2019cert}
(\href{https://github.com/danielzuegner/robust-gcn}{https://github.com/danielzuegner/robust-gcn}).
For the linearization and dualization method, we extend the reference implementation from~\cite{Jin2020}
(\href{https://github.com/RobustGraph/RoboGraph}{https://github.com/RobustGraph/RoboGraph}).
For the policy iteration method, we extend the reference implementation from~\cite{Bojchevski2019}
(\href{https://github.com/abojchevski/graph_cert}{https://github.com/abojchevski/graph\_cert}).
For sparsity-aware randomized smoothing on graphs, we use the reference implementation from~\cite{Bojchevski2020}
(\href{https://github.com/abojchevski/sparse\_smoothing}{https://github.com/abojchevski/sparse\_smoothing}).

For equivariance-preserving randomized smoothing on point clouds, we use the training and sampling procedure from~\cite{Schuchardt2022}
(\href{https://github.com/jan-schuchardt/invariance-smoothing}{https://github.com/jan-schuchardt/invariance-smoothing}).

All of the above are available under MIT license.

To train models for molecular force prediction, we use a pre-release version of code from~\cite{Wollschlaeger2023molecules}
(\href{https://github.com/wollschl/uncertainty_for_molecules}{https://github.com/wollschl/uncertainty\_for\_molecules}), which we will include in our reference implementation.

%% file: appendix/action_induced_distance.tex
\section{Proof of Proposition 1}
\label{appendix:action_induced_distance}
In the following, we show that the action-induced distance is the only function $\dlift : \sX \times \sR \rightarrow \sR_+$ that fulfills all three desiderata defined in~\cref{section:input_distance}. These were
\begin{itemize}
    \item $\forall x,x' \in \sX, g \in \sG : \dlift(x, g \act x') = \dlift(x, x')$,
    \item $\forall x, x': \dlift(x,x') \leq \din(x,x')$,
    \item $\dlift(x, x') = \max_{\gamma \in \sD} \gamma(x, x')$,
\end{itemize}
where $\sD$ is the set of functions from $\smash{\sX \times \sX}$ to $\sR_+$ that fulfill the first two desiderata

\begin{customproposition}{1}
A function $\smash{\dlift : \sX \times \sX \rightarrow \sR_+}$ that fulfills all three desiderata for any original distance function  $\smash{\din : \sX \times \sX \rightarrow \sR_+}$ exists and is uniquely defined: $\smash{\dlift(x,x') = \min_{g \in \sG} \din(x, g \act x')}$.
\end{customproposition}
\begin{proof}
Consider a specific pair $x, x' \in \sX'$.
The first desideratum states that we must have $\dlift(x, x') = \dlift(x, g \act x')$ for all $g \in \sG$.
The second desideratum states that we must have $\dlift(x, g\act x') \leq \din(x, g\act x')$ for all $g \in \sG$.
Thus, by the transitive property, fulfilling both desiderata simultaneously for our specific $x$ and $x'$ is equivalent to
$\forall g \in \sG : \dlift(x,x') \leq \din(x,g \act x')$.
This is equivalent to $\dlift(x,x') \leq \min_{g \in \sG} \din(x, g \act x')$.
The left side of the equality is naturally maximized to fulfill the third desideratum when we have a strict equality, i.e.\ $\dlift(x,x') = \min_{g \in \sG} \din(x,g \act x')$.
\end{proof}

%% file: appendix/combining_measures_schemes.tex
\section{Combining smoothing measures and smoothing schemes}\label{appendix:combing_measures_schemes}
Most randomized smoothing literature can be divided into works that focus on investigating properties of different smoothing measures (e.g.~\cite{Bojchevski2020,Cohen2019,Fischer2020,Alfarra2022,Murarev2022,Lee2019,Yang2020} and works that propose new smoothing schemes (e.g.~\cite{Cohen2019, Kumar2020,Chiang2020,Kumar2021}.
Measure-focused works usually only consider classification tasks, while scheme-focused works usually only consider Gaussian measures.
But, as we shall detail in the following, any of the smoothing schemes discussed in~\cref{section:equivariance_smoothing} can be used with arbitrary smoothing measures to enable certification for various input distances $\din$ and output distances $\dout$.

\subsection{Smoothing measures}
Measure-focused works usually consider a base classifier $h : \sX \rightarrow \sY$ with $\sY = \{1,\dots,K\}$.
They define a family of smoothing measures $(\mu_x)_{x \in \sX}$ to construct the smoothed classifier $f(x) = \max_{k \in \sY} \Pr_{z \sim \mu_x}\left[h(z)=k\right]$.
To certify the predictions of such a smoothed classifier, they derive lower and upper bounds on the probability of classifying a perturbed input $x' \in \sX$ as a specific class $k \in \{1,\dots,k\}$, i.e.\ $\underline{p_{h,x',k}} \leq \Pr_{z \sim \mu_{x'}}\left[h(z) = k\right] \leq \overline{p_{h,x',k}}$.
These bounds are usually obtained by finding the least / most robust model that has the same clean prediction probability as base classifier $h$, i.e.
\begin{align*}
    \underline{p_{h,x',k}} &= \min_{\tilde{h} : \sX \rightarrow \sY} \Pr_{z \sim \mu_{x'}}\left[\tilde{h}(z)=k\right] \ \text{ s.t. } \ \Pr_{z \sim \mu_x}\left[\tilde{h}(z)=k\right] = \Pr_{z \sim \mu_x}\left[h(z)=k\right],
    \\
    \overline{p_{h,x',k}} &= \max_{\tilde{h} : \sX \rightarrow \sY} \Pr_{z \sim \mu_{x'}}\left[\tilde{h}(z)=k\right] \ \text{ s.t. } \ \Pr_{z \sim \mu_x}\left[\tilde{h}(z)=k\right] = \Pr_{z \sim \mu_x}\left[h(z)=k\right]
\end{align*}
As long as $\mu_x$ and $\mu_{x'}$ have a density, these problems\footnote{or continuous relaxations thereof, if any constant likelihood region has non-zero measure} can always be solved exactly via the Neyman-Pearson lemma~\cite{Neyman1933}.

For appropriately chosen smoothing measures $(\mu_x)_{x \in \sX}$, one can then identify that the perturbed probability bounds
$\underline{p_{h,x',k}}$ and $\overline{p_{h,x',k}}$ only depend on a certain distance function $\din: \sX \times \sX \rightarrow \sR_+$.
For example:
\begin{itemize}
    \item For $\sX = \sR^N$, and $\mu_\vx = \mathcal{N}(\vx,\sigma \cdot \eye)$, the bounds $\underline{p_{h,x',k}}$ and $\overline{p_{h,x',k}}$ only depend on $\din(\vx,\vx') = ||\vx - \vx'||_2$ and are monotonically decreasing / increasing~\cite{Cohen2019}.
    \item For $\sX = \sR^N$, and $\mu_\vx = \mathrm{Laplace}(\vx,\sigma \cdot \eye)$, the bounds $\underline{p_{h,x',k}}$ and $\overline{p_{h,x',k}}$ only depend on $\din(\vx,\vx') = ||\vx - \vx'||_1$ and are monotonically decreasing / increasing~\cite{Yang2020}.
    \item For $\sX = \{0,1\}$, and $\mu_\vx$ being the measure associated with i.i.d.\ flipping of input bits, the bounds $\underline{p_{h,x',k}}$ and $\overline{p_{h,x',k}}$ only depend on $\din(\vx,\vx') = ||\vx - \vx'||_0$ and are monotonically decreasing / increasing~\cite{Lee2019}.
\end{itemize}
For an overview of various additive smoothing schemes and their associated input distances, see~\cite{Yang2020}.
Note that works on non-additive smoothing measures, like ablation smoothing~\cite{Levine2020ablation} or derandomized smoothing~\cite{Levine2020derandomized}, also provide bounds $\underline{p_{h,x',k}}$ and $\overline{p_{h,x',k}}$ as a function of some input distance $\din$, such as the number of perturbed pixels or the size of an adversarial image patch.

\subsection{Smoothing schemes}\label{appendix:combing_measures_schemes_schemes}
Scheme-focused works include majority voting~\cite{Cohen2019}, expected value smoothing~\cite{Kumar2020}, median smoothing~\cite{Chiang2020} and center smoothing~\cite{Kumar2021}.
These works only consider Gaussian smoothing, i.e. $\sX = \sR^N$ and $\mu_\vx = \mathcal{N}(\vx, \sigma \cdot \eye)$.
Each one proposes a different prediction procedure and certification procedure.
The prediction procedures provide a sampling-based approximations of the smoothed prediction $\xi(\mu \cat h^{-1})$ and does not depend on the choice of smoothing measure.
The certification procedures can be adapted to arbitrary smoothing measures as follows.

\textbf{Majority voting.}
Majority voting assumes a base classifier $h : \sX \rightarrow \sY$ with $\sY = \{1,\dots,K\}$.
It  constructs a smoothed classifier $f(x) = \max_{k \in \sY} \Pr_{z \sim \mu_x}\left[h(z)=k\right]$.
The prediction $f(x) = k^*$ of this smoothed classifier is robust if $\underline{p_{h,x',k^*}} > \max_{k \neq k^*} \underline{p_{h,x',k}}$ or if  $\underline{p_{h,x',k^*}} > 0.5$.
This certification procedure can be adapted to any other smoothing measure by inserting the corresponding lower and upper bounds.
It can be generalized to multi-output classification tasks with $\sY = \{1,\dots,K\}^M$ (e.g.\ segmentation) by applying the certification procedure independently to each output dimension.

\textbf{Expected value and median smoothing.}
Median smoothing and expected value smoothing assume a base regression model $h : \sX \rightarrow \sR$.
They construct a smoothed model $f(\vx)$ via the expected value or median of pushforward measure $\mu_x \cat h^{-1}$.
To certify robustness, they require lower and upper bounds on the cumulative distribution function $\Pr_{z  \sim \mu_x'}\indicator \left[ h(z) \leq a_t  \right]$
for a finite range of thresholds $-\infty \leq a_t \leq \dots \leq a_T \leq \infty$.
Since each of the indicator functions $\indicator \left[ h(z) \leq a_t  \right]$ can be thought of as a binary classifier $g_t : \sX \rightarrow \{0,1\}$, these lower and upper bounds are given by $\underline{p_{g_t,x',1}}$ and $\overline{p_{g_t,x',1}}$.
These certification procedures can thus be adapted to any other smoothing measure by inserting the corresponding lower and upper bounds.
They can be generalized to multi-output regression tasks with $\sY = \sR^M$ by applying the certification procedure independently to each output dimension.
Instead of proving that the smoothed prediction $f(\vx)$ remains constant, these smoothing schemes guarantee that the output remains in a certified hyperrectangle
$\sH = \{\vy \in \sR^M \mid l_m \leq \evx_m \leq u_m\}$. 
One can then prove that $\dout(f(\vx), f(\vx')) \leq \delta$ by verifying that $\{\vy \in \sR^M \mid \dout(f(\vx), \vy) \leq \delta\} \subseteq \sH$, i.e.\ the hyperrectangle contains a $\dout$-ball of radius $\delta$.

\textbf{Center smoothing.}
Center smoothing is compatible with any model $h : \sX \rightarrow \sY$, as long as the output space $\sY$ fulfills a relaxed triangle inequality.
It constructs a smoothed model $f(\vx)$ via the center of the smallest ball that has at least $50\%$ probability under pushforward measure $\mu_\vx \cat h^{-1}$.
To certify robustness, it requires bounds on the cumulative distribution function
$\Pr_{z  \sim \mu_x'}\indicator \left[ \dout(f(\vx), h(\vz) \leq a_t  \right]$.
Thus, like with expected value and median smoothing, this certification procedure can be adapted to any other smoothing measure by inserting the corresponding bounds $\underline{p_{g_t,x',1}}$ and $\overline{p_{g_t,x',1}}$.
This smoothing scheme directly provides an upper bound on $\dout(f(\vx), f(\vx'))$.

Informally speaking, any of these schemes can be adapted to non-Gaussian measures by replacing any occurence of
$\Phi\left(\Phi^{-1}(\dots) - \dots\right)$ and $\Phi\left(\Phi^{-1}(\dots) + \dots\right)$, which are the formulae for $\underline{p_{h,x',k}}$ and $\overline{p_{h,x',k}}$ of Gaussian measures~\cite{Cohen2019}, with the bounds of another smoothing measure.
For example, one can combine the bounds from derandomized smoothing with the center smoothing certification procedure to prove robustness of  generative image reconstruction models to patch-based attacks.

%% file: appendix/equivariance_smoothing.tex
\section{Equivariance-preserving randomized smoothing}

\label{appendix:equivariance_smoothing}
\begin{figure}[H]
    \centering
    \includegraphics[width=\linewidth]{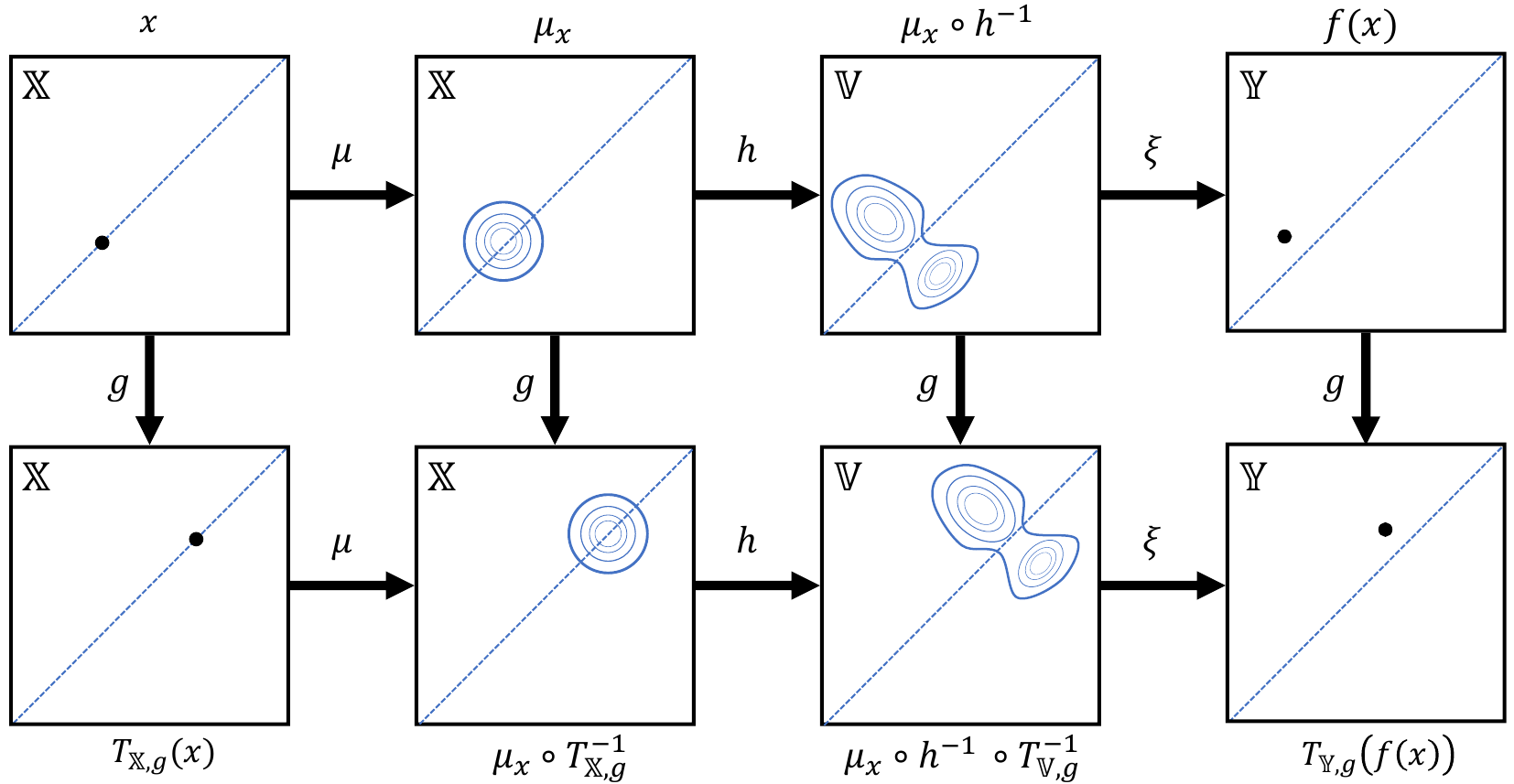}
    \caption{Example of equivariance-preserving randomized smoothing for $\sX = \sV = \sY = \sR^2$
    and diagonal translation group $\sG = (\sR,+)$, which acts via
    $T_{\cdot, g}(\vx) =  \vx + g \cdot \mathbf{1}$.
    We construct our smoothed model using
    Gaussian measures $\mu_\vx = \mathcal{N}(\vx, \sigma \cdot \eye)$,
    base model $h : \sX \rightarrow \sV$
    and expected value smoothing scheme $\xi: \Delta(\sV, \sF) \rightarrow \sY$,
    which maps from measures on intermediate space $(\sV, \sF)$ to output space $\sY$.
    Measures are visualized via isocontours of their densities.
    Because the family of measures $(\mu_\vx)_{\vx \in \sX}$ is equivariant (left cycle),
    the base model $h$ is equivariant (central cycle), 
    and the smoothing scheme $\xi$ is equivariant (right cycle),
    the smoothed model $f(\vx) = \xi(\mu_\vx \circ h^{-1})$
    is also equivariant to diagonal translation.} 
    \label{fig:explainy_smoothing}
\end{figure}

In the following, we provide more formal justifications for our randomized smoothing framework, that we introduced in section~\cref{section:equivariance_smoothing} and that is visualized in~\cref{fig:explainy_smoothing}.
We first verify the correctness of our sufficient condition for equivariance-preserving randomized smoothing. We then
consider the different equivariances preserved by different smoothing schemes and measures.
We provide a tabular overview of different schemes, measures and their properties in~\cref{table:measures,table:schemes}.

\textbf{Concerning notation for pushforward measures.} 
Consider measurable spaces $(\sX,\sD)$ and $(\sV,\sF)$, and a measurable function $h : \sX \rightarrow \sV$.
We write $h^{-1}$ to refer to the preimage of function $h$, not its inverse.
That is, for any $\sA \subseteq \sV$ we have $h^{-1}(\sA) = \{x \in \sX \mid h(x) \in \sA\}$.
Because $h$ is measurable, we have $h^{-1}(\sA) \in \sD$ whenever $\sA \in \sF$.
Thus,  the expression $\mu \cat h^{-1}(\sA)$ refers to the measure of all elements of $\sX$ that get mapped into $\sA \in \sF$,
i.e.\ it is the pushforward measure of $\mu$.

\subsection{Proof of Proposition 3}
\begin{customproposition}{3}
    Assume two measurable spaces $\smash{(\sX, \sD)}$, $\smash{(\sV, \sF)}$, an output space $\sY$ and a measurable base model $\smash{h : \sX \rightarrow \sV}$ that is equivariant with respect to the action of group $\sG$.
    Further assume that $\sG$ acts on $\sX$ and $\sV$ via measurable functions.
    Let $\smash{\xi : \Delta(\sV, \sF) \rightarrow \sY}$ be a smoothing scheme that maps from the set of probability measures $\Delta(\sV, \sF)$ on intermediate space $\smash{(\sV, \sF)}$ to the output space.
    Define $T_{\sX, g}(\cdot)$ to be the group action on set $\sX$ for a fixed $g$, i.e.\ $\smash{T_{\sX, g}(x) = g \act_\sX x}$. 
    Then, the smoothed model $f(x) = \xi(\mu_x \cat h^{-1})$ is equivariant with respect to the action of group $\sG$ if both
    \begin{itemize}
        \item the family of measures $\smash{(\mu_x)_{x \in \sX}}$ is equivariant, i.e.\ $\smash{\forall x \in \sX, g \in \sG: \mu_{g \act x} = \mu_x \cat T_{\sX,g}^{-1}}$, 
        \item and smoothing scheme $\xi$ is equivariant, i.e.\ $\smash{\forall \nu \in \Delta(\sV, \sF), g \in \sG : \xi (\nu \cat T_{\sV, g}^{-1} ) = g \act \xi(\nu)}$.
    \end{itemize}
\end{customproposition}
\begin{proof}
By definition of $f$, verifying that $\forall x \in \sX, \forall g \in \sG : f(g \act x) = g \act f(x)$ is equivalent to verifying that
$\xi(\mu_{g \act x} \cat h^{-1}) = g \act \xi(\mu_{g} \cat h^{-1})$. We can do so by first using the equivariance of the family of measures $(\mu_x)_{x \in \sX}$, then the equivariance of base model $h$ and then the equivariance of smoothing scheme $\xi$:
\begin{align*}
    \xi(\mu_{g \act x} \cat h^{-1}) &= \xi\left(\left(\mu_{x} \cat T_{\sX,g}^{-1}\right)  \cat h^{-1}\right)
    \\
    & =
    \xi\left(\mu_{x} \cat \left( T_{\sX,g}^{-1}  \cat h^{-1}\right) \right)
    \\
    & =
    \xi\left(\mu_{x} \cat \left( h \cat T_{\sX,g} \right)^{-1} \right)
    \\
    & =
    \xi\left(\mu_{x} \cat \left( T_{\sV,g} \cat h  \right)^{-1} \right)
    \\
    & =
    \xi\left(\mu_{x} \cat \left( h^{-1} \cat T_{\sV,g}^{-1} \right) \right)
    \\
    & =
    \xi\left(\left(\mu_{x} \cat  h^{-1} \right) \cat T_{\sV,g}^{-1}  \right)
    =
    g \act \xi(\mu_x \cat h^{-1}).
\end{align*}
In the second and sixth equality we used the associativity of function composition.
In the third and fifth equality we used that the preimage of a composition of functions is equivalent to the composition of the individual preimages in reverse order.
\end{proof}

\label{appendix:proof_equivariance_smoothing}

\subsection{Equivariance-preserving schemes and measures}
In the following, $\Delta(\sX,\sD)$ refers to the set of all probability measures on measurable space $(\sX,\sD)$.

\label{appendix:proof_scheme_measure_equivariances}
\subsubsection{Componentwise smoothing schemes.} 
For this type of smoothing scheme, we assume that the intermediate space and output space consist of $M$ distinct components, i.e. $\sV = \sA^M$ and $\sY = \sB^M$ for some sets $\sA$, $\sB$. We write $\mathrm{proj}_m : \sA^M \rightarrow \sA$ for the function that maps any vector $\va \in \sA^M$ to its $m$th component $a_m$.
Thus, for any measure $\nu \in \Delta(\sA^M,\sF^M)$ the function $\nu \cat \mathrm{proj}_m^{-1}$ is the $m$th marginal measure, which is a measure on the single-dimensional space $(\sA,\sF)$.
Componentwise smoothing means generating the $m$th prediction based on some quantity of the $m$th marginal measure.

\begin{proposition}\label{proposition:componentwise_schemes}
    Assume an $M$-dimensional measurable product space $(\sA^M, \sF^M)$ and an output space $\sB^M$.
    Further assume that there is a function $\kappa : \Delta(\sA, \sF) \rightarrow \sB$ such that
    $\forall \nu \in \Delta(\sA^M, \sF^M) : \xi(\nu)_m = \kappa(\nu \cat \mathrm{proj}_m^{-1})$.
    If a group $\sG$ acts on $\sA^N$ and $\sB^N$ via the same permutation, then the smoothing scheme $\xi$ is equivariant.
\end{proposition}
\begin{proof}
    Consider an arbitrary $g \in \sG$ with corresponding actions $T_{\sV, g} : \sA^M \rightarrow \sA^M$ with $T_{\sV, g}(\vv)_m = \evv_{\pi^{-1}(m)}$
    and 
    $T_{\sY, g} : \sA^M \rightarrow \sA^M$ with $T_{\sY, g}(\vy)_m = \evy_{\pi^{-1}(m)}$
    , where
    $\pi : \{1,\dots,M\} \rightarrow \{1,\dots,M\}$ is a permutation.
    For any $m$, we can use the definition $\xi$, the associativity of function composition and the fact that the preimage of a composition is equivalent to the composition of preimages in reverse order to show that
    \begin{align*}
    \xi \left(\nu \cat T_{\sV, g}^{-1} \right)_m
    &=
    \kappa \left((\nu \cat T_{\sV, g}^{-1}) \cat \mathrm{proj}_m^{-1} \right)
    \\
    &
    =
    \kappa \left(\nu \cat (T_{\sV, g}^{-1} \cat \mathrm{proj}_m^{-1} )\right)
    \\
    &=
    \kappa \left(\nu \cat (\mathrm{proj_m} \cat T_{\sV, g})^{-1}\right)
    \\
    &=
    \kappa \left(\nu \cat \mathrm{proj^{-1}_{\pi^{-1}(m)}}\right)
    =
    \xi \left(\nu \right)_{\pi^{-1}(m)}
    = (g \act \xi \left(\nu \right))_m.
    \end{align*}
    For the fourth equality we  used that selecting the $m$th element from a sequence permuted by $\pi$ is equivalent to selecting the element $\pi^{-1}(m)$ of the unpermuted sequence.
    The second to last equality is just the definition of our smoothing scheme $\xi$.
\end{proof}
In practice, componentwise smoothing schemes are not evaluated exactly but approximated using $S$ Monte Carlo samples $\vv^{(1)},\dots,\vv^{(S)} \in \sV$.
The $m$th component of smoothed prediction $y_n$ is generated based on some quantity of $\evv_m^{(1)},\dots,\evv_m^{(S)} \in \sV$, such as their mean~\cite{Kumar2020}. 
We can thus use the same argument as above to show that the Monte Carlo approximations are permutation equivariant:
Permuting the components of the samples before applying componentwise functions is equivalent to first applying componentwise functions and then performing a permutation.

\subsubsection{Expected value smoothing scheme.} 
For this smoothing scheme, we assume that the intermediate and output space coincide are real-valued, i.e.\ $\sV = \sY = \sR^M$
We then make smoothed predictions via the expected value of the model's output distribution over $\sY$.

\begin{proposition}\label{proposition:expected_value_scheme}
    Let $\mathcal{B}(\sR^M)$ be the corresponding Borel $\sigma$-algebra of $\sR^M$.
    Define expected value smoothing scheme $\xi : \Delta(\sR^M, \mathcal{B}(\sR^M)) \rightarrow \sR^M$ with 
    \begin{equation*}
        \xi(\nu) = \int_{\sR^M} \vy \ d \nu(\vy)
    \end{equation*}
    If group $\sG$ acts on $\sR^M$ via affine transformations, then smoothing scheme $\xi$ is equivariant.
\end{proposition}
\begin{proof}
    Consider an arbitrary $g \in \sG$ with corresponding action $T_{\sY,g}$. By change of variables and linearity of integration, we have
    \begin{align*}
        \xi\left(\nu \cat T_{\sY,g}^{-1}\right)
        &= \int_{\sR^M} \vy \ d \left(\nu \cat T_{\sY,g}^{-1}\right)(\vy)
        \\
        &= \int_{\sR^M} T_{\sY,g}(\vy) \ d  \nu(y)
        \\
        &= T_{\sY,g}\left(\int_{\sR^M} \vy \ d  \nu(\vy)\right)
        \\
        &= g \act \xi(\nu).
    \end{align*}
\end{proof}
In practice, expected value smoothing is not evaluated exactly but approximated using the average of $S$ Monte Carlo samples $\vv^{(1)},\dots,\vv^{(S)} \in \sR^M$~\cite{Kumar2020}.
We can thus use the same argument as above to show that the Monte Carlo approximation is equivariant to affine transformations:
Affinely transforming the samples before computing the average is equivalent to computing the average and then applying the affine transformation.

\subsubsection{Median smoothing scheme}
For this smoothing scheme, we assume that the intermediate and output space coincide and are real-valued, i.e.\ $\sV = \sY = \sR^M$.
We then make smoothed predictions via the elementwise median of the model's output distribution over $\sY$.
As before, we write $\mathrm{proj}_m : \sR^M \rightarrow \sR$ for the function that maps any vector $\vy \in \sR^M$ to its $m$th component $\evy_m$.
Thus, $\nu \cat \mathrm{proj}_m^{-1}$ is the $m$th marginal measure of $\nu$.
\begin{proposition}\label{proposition:median_scheme}
    Let $\mathcal{B}(\sR^M)$ be the  Borel $\sigma$-algebra of $\sR^M$.
    Let $F_m(\nu, x)$ be the cumulative distribution function for marginal measure $\nu \cat \mathrm{proj}_m^{-1}$, i.e.\
    \begin{equation*}
        F_m(\nu, x) = \left(\nu \cat \mathrm{proj}_m^{-1}\right)((-\infty,x]).
    \end{equation*}
    Let $F_m^{-}(\nu, p)$ and $F_m^{+}(\nu, p)$ be the corresponding lower and upper quantile functions, i.e.\
    \begin{align*}
        F_m^{-}(\nu, p) &= \inf \{x \mid  F_m(\nu, x) \geq p \}
        \\
        F_m^{+}(\nu, p) &= \sup \{x \mid  F_m(\nu, x) \leq p \}
    \end{align*}
    Define median smoothing scheme $\xi : \Delta(\sR^M, \mathcal{B}(\sR^M)) \rightarrow \sR^M$ with 
    \begin{equation*}
        \xi(\nu)_m = \frac{1}{2} \cdot 
        \left[
            F^{-}_m(\nu, \nicefrac{1}{2}) + F^{+}_m(\nu, \nicefrac{1}{2})
        \right].
    \end{equation*}
    If group $\sG$ acts on $\sY$ via elementwise linear transformations, then smoothing scheme $\xi$ is equivariant.
\end{proposition}
\begin{proof}
    Consider an arbitrary $g \in \sG$.
    Let $T_{\sY,g}$ be the corresponding group action with $T_{\sY,g}(\vy)_m = w_m \cdot \evy_m + c_m$ for some $w_m, c_m \in \sR$.
    Assume w.l.o.g.\ that $\forall m : w_m > 0$.
    
    For the cumulative distribution functions, we have
    \begin{align*}
        F_m(\nu \cat T_{\sY,g}^{-1}, x) = F_m(\nu , (x-c_m) \mathbin{/} w_m).
    \end{align*}
    For the corresponding lower and upper quantile functions, we thus have
    \begin{align*}
        F_m^{-}(\nu \cat T_{\sY,g}^{-1} , p) &= w_m \cdot F_m^{-}(\nu, p) + c_m
        \\
        F_m^{+}(\nu \cat T_{\sY,g}^{-1} , p) &= w_m \cdot F_m^{+}(\nu, p) + c_m.
    \end{align*}
    For the smoothing scheme, we thus have
    \begin{equation*}
        \xi(\nu \cat T_{\sY,g}^{-1})_m = w_m \cdot \xi(\nu)_m + c_m = (g \act \xi(\nu))_m.
    \end{equation*}
\end{proof}
In practice, median smoothing is not evaluated exactly but approximated using the elementwise median of $S$ Monte Carlo samples $\vv^{(1)},\dots,\vv^{(S)} \in \sV$~\cite{Kumar2020}.
We can thus use the same argument as above to show that the Monte Carlo approximation is equivariant to elementwise linear transformations:
Linearly transforming the samples elementwise before computing the sample median is equivalent to computing the sample median and then applying the elementwise linear transformations.

\subsubsection{Center smoothing scheme}
For this type of smoothing scheme, we assume that the intermediate and output space are identical, i.e.\ $\sV = \sY$.
Let $\sB(y,r) = \{y' \in \sY \mid \dout(y,y') \leq r\}$ be the $\dout$ ball of radius $r$ around $y \in \sY$.
Center smoothing makes its predictions using the center of the $\dout$ ball with the smallest radius among all $\dout$ balls with a measure of at least $\frac{1}{2}$.

\begin{proposition}\label{proposition:center_scheme}
    Consider a measurable space $(\sY, \sF)$, a function $\dout : \sY \times \sY \rightarrow \sR_+$, and assume that
    $\forall y \in \sY, r \geq 0 : \sB(y,r) \in \sF$.
    Define center smoothing scheme $\xi : \Delta(\sY, \sF) \rightarrow \sY$ with 
    $\xi(\nu) = \argmin_y r \text{ s.t. } \nu\left(\sB(y,r)\right) \geq \frac{1}{2}$.
    If $\sG$ acts isometrically on $\sY$, i.e.\ $\forall y, y' \in \sY, g \in \sG : \dout(g \act y, g \act y') = \dout(y, y')$, then smoothing scheme $\xi$ is equivariant.
\end{proposition}
\begin{proof}
    Consider an arbitrary $g \in \sG$ with corresponding action $T_{\sY, g}$.
    By definition of the center smoothing scheme, $\xi(\nu \cat T_{\sY, g}^{-1})$ is
    \begin{equation*}
        \left(\argmin_y r \text{ s.t. } \left(\nu \cat T_{\sY, g}^{-1}\right)\left(\sB(y,r)\right) \geq \frac{1}{2} \right)
        =
        \left(\argmin_y r \text{ s.t. } \nu\left(T_{\sY, g}^{-1}(\sB(y,r)) \geq \frac{1}{2}) \right)\right).
    \end{equation*}
    By definition of the preimage and action $T_{\sY, g}$, we have
    $T_{\sY, g}^{-1}(\sB(y,r))
    = \{y' \mid g \act y' \in \sB(y,r)\}
    = \{y' \mid \dout(y, g \act y') \leq r\}
    = \{y' \mid \dout(g^{-1} \act y, y') \leq r\} 
    = \sB\left(g^{-1} \act y, r\right),
    $
    where the second to last  equality holds because $\sG$ acts isometrically. This shows that
    \begin{equation*}
        \xi(\nu \cat T_{\sY, g}^{-1}) = \left(\argmin_y r \text{ s.t. } \nu\left(\sB(g^{-1} \act y,r)\right) \geq \frac{1}{2} \right).
    \end{equation*}
    The optimum of this problem is $g \act \xi(\nu)$, because $g^{-1} \act (g \act \xi(\nu)) = \xi(\nu)$ and $\xi(\nu)$ is the optimum of the original  problem without group action.
\end{proof}
In practice, center smoothing is  approximated using $S$ Monte Carlo samples $\vy^{(1)},\dots,\vy^{(S)} \in \sY$ by
selecting the sample with the smallest median distance $\dout$ to all other samples~\cite{Kumar2021}.
We can use a similar argument to the one above to show that the Monte Carlo approximation is isometry equivariant:
Isometries, by definition, do not change the pairwise distances and therefore do not change which sample has the smallest median distance to all other samples.

\subsubsection{Product measures}
For our discussion of product measures, note that the $\sigma$-algebra $\sD^N$ of a measurable product space $(\sA^N,\sD^N)$ is not the $n$-fold Cartesian product of the $\sigma$-algebra $\sD$. Instead, it is defined as
$\left\{ \left(\bigtimes_{n=1}^N \sS_n \right) \mid \sS_1 \in \sD,\dots, \sS_N \in \sD \right\}$, i.e.\ the set containing all Cartesian products of measurable sets.

\begin{proposition}\label{proposition:independent_noise}
    Assume that $(\mu_\vx)_{\vx \in \sA^N}$ is a family of product measures on the N-dimensional measurable product space $(\sA^N, \sD^N)$, i.e.\ there is a family of measures $(\kappa_a)_{a \in \sA}$ on $(\sA, \sD)$ such that 
    $\forall \vx \in \sA^N, \forall \left(\bigtimes_{n=1}^N \sS_n \right) \in \sD^N :
        \mu_\vx\left(\bigtimes_{n=1}^N \sS_n \right) = \prod_{n=1}^N \kappa_{\evx_n}(\sS_n).
    $
    If group $\sG$ acts on $\sA^N$ via permutation, then the family of measures $(\mu_\vx)_{\vx \in \sX^N}$ is equivariant.
\end{proposition}
\begin{proof}
    Consider an arbitrary $g \in \sG$ with action $T_{\sX, g} : \sA^N \rightarrow \sA^N$ with $T_{\sV, g}(\vx)_n = \evx_{\pi^{-1}(n)}$
    , where
    $\pi : \{1,\dots,N\} \rightarrow \{1,\dots,N\}$ is a permutation.
    For any $\left(\bigtimes_{n=1}^N \sS_n \right) \in \sD^N$, we have
    \begin{align*}
        \mu_{g \act \vx} \left(\bigtimes_{n=1}^N \sS_n \right)
        = 
        \prod_{n=1}^N \kappa_{(g \act \vx)_n}(\sS_n)
        =
        \prod_{n=1}^N \kappa_{\evx_{\pi^{-1}(n)}}(\sS_n)
        =
        \prod_{n=1}^N \kappa_{\evx_{n}}(\sS_{\pi(n)})
        = 
        \mu_{\vx} \left(\bigtimes_{n=1}^N \sS_{\pi(n)} \right).
    \end{align*}
    For the second to last equality, we have just changed the iteration order of the product from $(1,\dots,N)$ to $(\pi(1),\dots,\pi(N))$.
    Finally, it follows from the definition of our group action and preimages that 
    \begin{align*}
        \left(\bigtimes_{n=1}^N \sS_{\pi(n)} \right)
        & = 
        \left\{\vs \in \sA^N \mid \evs_1 \in \sS_{\pi(1)}, \dots, \evs_N \in \sS_{\pi(N)} \right\}
        \\
        &=
        \left\{\vs  \in \sA^N \mid \evs_{\pi^{-1}(1)} \in \sS_{1}, \dots, \evs_{\pi^{-1}(N)} \in \sS_{\pi(N)} \right\}
        \\
        &=
        \left\{\vs  \in \sA^N  \mid T_{\sX, g}(\vs) \in \left(\bigtimes_{n=1}^N \sS_n \right) \right\}
        \\
        &= 
        T_{\sX, g}^{-1} \left(\bigtimes_{n=1}^N \sS_n \right),
    \end{align*}
    and thus $\forall \vx \in \sA^N, g \in \sG: \mu_{g \act \vx} = \mu_\vx \cat T_{\sX,g}^{-1}$.
\end{proof}

\subsubsection{Isotropic Gaussian measures}

\begin{proposition}\label{proposition:gaussian_noise}
    Consider the measurable space $(\sR^{N \times D}, \mathcal{B}(\sR^{N \times D}))$, where $\mathcal{B}(\sR^{N \times D})$ is the Borel $\sigma$-algebra on $\sR^{N \times D}$.
    Let $(\mu_{\mX, \sigma})_{\mX \in \sR^{N \times D}}$ be the family of isotropic Gaussian measures with standard deviation $\sigma$ on
    $(\sR^{N \times D}, \mathcal{B}(\sR^{N \times D}))$.
    If a group $\sG$ acts isometrically on $\sR^{N \times D}$ with respect to Frobenius norm $||\cdot||_2$, then the family of measures is equivariant.
\end{proposition}
\begin{proof}
        By definition of the Gaussian measure, we have for any $\sA \in \mathcal{B}(\sR^{N \times D})$
    \begin{align*}
        \mu_{\mX, \sigma}(\sA)
        & =
        \int_{\sA} \prod_{n=1}^N \prod_{d=1}^D \frac{1}{\sqrt{2 \pi \sigma^2} }
        \exp \left(
        -\frac{1}{2\sigma^2} (Z_{n,d} - X_{n,d}^2)
        \right) \ d \mZ
        \\
        & = 
        \int_{\sA}  \frac{1}{\left(\sqrt{2 \pi \sigma^2} \right)^{N \cdot D}}
        \exp \left(
        -\frac{1}{2\sigma^2} ||\mZ - \mX||_2^2
        \right) \ d \mZ.
    \end{align*}
    By change of variables, we have
    \begin{align*}
        \left(\mu_{\mX, \sigma} \cat T_{\sX, g}^{-1}\right) (\sA)
        & =
        \int_{T_{\sX, g}^{-1} (\sA)}  \frac{1}{\left(\sqrt{2 \pi \sigma^2} \right)^{N \cdot D}}
        \exp \left(
        -\frac{1}{2\sigma^2} ||\mZ - \mX||_2^2
        \right) \ d \mZ
        \\
        & = 
        \int_{\sA} \left\lvert\det\left(J_{T^{-1}_{\sX,g}}\right) \right\rvert \frac{1}{\left(\sqrt{2 \pi} \sigma\right)^{N \cdot D}}
        \exp \left(
        -\frac{1}{2\sigma^2} || T_{\sX, g}^{-1}(\mZ) - \mX||_2^2
        \right) \ d \mZ
        \\
        & = 
        \int_{\sA} 1 \cdot  \frac{1}{\left(\sqrt{2 \pi} \sigma\right)^{N \cdot D}}
        \exp \left(
        -\frac{1}{2\sigma^2} || \mZ - T_{\sX, g}(\mX)||_2^2
        \right) \ d \mZ
        \\
        & = 
        \mu_{g \act \mX, \sigma}  (\sA).
    \end{align*}
    The third equality follows from $T_{\sX, g}$ being an isometry with respect to the Frobenius norm $||\cdot||_2$.
\end{proof}

\subsubsection{Transformation-specific measures}
In transformation-specific smoothing~\cite{Chu2022,Fischer2020,Li2021,Alfarra2022,Murarev2022},
the clean input $x \in \sX$ is transformed via randomly sampled functions from a parametric family $(\psi_\theta)_{\theta \in \Theta}$ with $\psi_\theta : \sX \rightarrow \sX$ and 
 measurable parameter space $(\Theta, \sG)$.
Let $\gamma : \sG \rightarrow \sR_+$ be the corresponding parameter distribution.
Then, transformation-specific smoothing induces a smoothing measure
\begin{equation}\label{eq:transformation_measure}
    \mu_x(\sA) = \gamma\left(
        \left\{
            \theta \in \Theta \mid f_\theta(x) \in \sA
        \right\}
    \right) 
\end{equation}
on input space $(\sX,\sD)$. \footnote{Assuming $\forall \sA \in \sD : \left\{
            \theta \mid f_\theta(x) \in \sA
        \right\} \in \sG $, i.e.\ all preimages are in the parameter $\sigma$-field $\sG$.}
In the following, we show that this induced measure inherits its equivariances from the family of transformations $(\psi_\theta)_{\theta \in \Theta}$.
\begin{proposition}\label{proposition:transformation_noise}
    Consider a measurable space $(\sX, \sD)$
    and a parametric family of transformations $(\psi_\theta)_{\theta \in \Theta}$ with $\psi_\theta : \sX \rightarrow \sX$ and 
 measurable parameter space $(\Theta, \sG)$.
    Let  $\gamma : \sG \rightarrow \sR_+$ be a measure on parameter space $(\Theta, \sG)$
    and consider the family of induced smoothing measures $(\mu_x)_{x \in \sX}$, as defined in~\cref{eq:transformation_measure}.
    If all transformations $\psi_\theta$ are equivariant to the actions of group $\sG$, then the family
    of measures is equivariant.
\end{proposition}
\begin{proof}
    Consider an arbitrary group element $g \in \sG$ with corresponding group action $T_{\sX,g}$. 
    By definition of $\mu_x$ and the equivariance of all $\psi_\theta$, we have for all $x \in \sX$ that
    \begin{align*}
        \mu_{g \act x}(\sA) & = \gamma\left(
        \left\{
            \theta \in \Theta \mid f_\theta(g \act x ) \in \sA
        \right\}
    \right) 
    \\
    & =
    \gamma\left(
        \left\{
            \theta \in \Theta \mid g \act f_\theta(x ) \in \sA
        \right\}
    \right) 
    \\
    & =
    \gamma\left(
        \left\{
            \theta \in \Theta \mid f_\theta(x ) \in  T_{\sX,g}^{-1}(\sA)
        \right\}
    \right) 
    = \left(\mu_{x} \cat T_{\sX,g}^{-1}\right) (\sA).
    \end{align*}
    The second-to-last equality holds because, by definition of the pre-image,
    $T_{\sX,g}^{-1}(\sA) = \{x \in \sX \mid g \act x \in \sA\}$ and thus
    $g \act f_\theta(x ) \in \sA \iff  f_\theta(x ) \in T_{\sX,g}^{-1}(\sA)$.
\end{proof}

\subsubsection{Sparsity-aware measures}

\begin{proposition}\label{proposition:sparsity_noise}
    Consider the measurable space $(\sX, \sD)$, where
     $\sX = \{0,1\}^{N \times D} \times \{0,1\}^{N \times N}$ is the set of all binary attributed graphs with $N$ nodes and $D$ features 
    and $\sD = \mathcal{P}(\sX)$ is its powerset.
    Let $(\mu_{\mX, \mA})_{(\mX,\mA) \in \sX}$ be the family of sparsity-aware aware measures on $(\sX,\sD)$,
    as defined by the sparsity-aware probability mass function with fixed flip probabilities 
    $p_+^\mX, p_-^\mX, p_+^\mA, p_-^\mA \in [0,1]$ (see~\cref{eq:sparse-smoothing-distribution}).
    If group $\sG$ acts on $\sX$ via graph isomorphisms, then the family of measures is equivariant.
\end{proposition}
\begin{proof}
    Group $\sG$ acting on $\sX$ via graph isomorphisms means that 
    for every $(\mX,\mA) \in \sX$ and $g \in \sG$  there is some permutation matrix $\mP \in \{0,1\}^{N \times N}$
    such that $g \act (\mX,\mA) = (\mP \mX, \mP \mA \mP^T)$.
    That is, the entries of the attribute and adjacency matrix are permuted.
    The sparsity-aware measures are product measures, i.e.\ sparsity-aware smoothing perturbs each element of the attribute and adjacency matrix independently.
    Families of product measures are equivariant with respect to the action of groups acting via permutation, see~\cref{proposition:independent_noise}.
\end{proof}

\begin{table}[h!]
\caption{Suitable measures for different input domains $\mathbb{X}$, input distances $d_\mathrm{in}$, and group actions.}
\label{table:measures}
    \begin{center}
    \centerline{
    \begin{tabular}{ | c  c  c || c| } 
     \hline
     $\mathbb{X}$ & $d_\mathrm{in}$ & Group action & Smoothing measure \\
     \hline
     \hline     
     $\vrule width 0pt height 0.35cm \mathbb{R}^{N \times D}$ &  $\ell_1, \ell_\infty$ & \makecell{Permutation,\\Translation} & $\mathrm{Uniform}(-\sigma , \sigma)$  \\
     \hline
     $\vrule width 0pt height 0.35cm \mathbb{R}^{N \times D}$ &  $\ell_1, \ell_\infty$ & \makecell{Permutation,\\Translation} & $\mathrm{Laplace}(0, \sigma)$  \\
     \hline
     $\vrule width 0pt height 0.35cm \mathbb{R}^{N \times D}$ &  $\ell_2$ & \makecell{Euclidean isometries} & $\mathcal{N}(0, \sigma)$  \\
     \hline
     $\mathbb{Z}^{N \times D}$ &  $\ell_0$ & Permutation & \makecell{Discrete smoothing\\(see Lee et al.\ [80])} \\
     \hline
     \vrule width 0pt height 0.6cm \makecell{$\{0,1\}^{N \times D} \times \{0,1\}^{N \times N}$\\(Attributed graphs)} &  \makecell{Edit cost\\(see~\cref{eq:ged_base_distance})} & Graph isomorphism & \makecell{Sparsity-aware smoothing\\(see Appendix E.7)} \\
     \hline
    \end{tabular}
    }
\end{center}
\end{table}

\clearpage

\begin{table}[ht]
    \caption{Suitable smoothing schemes for different output domains $\mathbb{Y}$, output distances $d_\mathrm{out}$, and group actions (see discussion in Section 5.1).
    (*)For expected value and median smoothing, robustness can be certified as long as the $\dout$ ball $\{y' \mid \dout(f(x), y') \leq \epsilon\}$ is contained within a hyperrectangle (see~\cref{appendix:combing_measures_schemes_schemes}).
    }
    \label{table:schemes}
    \begin{center}
    \begin{tabular}{ | c  c c || c| } 
     \hline
     $\mathbb{Y}$ & $d_\mathrm{out}$ & Group action & Smoothing scheme \\
     \hline     
     \hline
     $\vrule width 0pt height 0.35cm \{1,\dots,K\}^N$ &  $\ell_0$ & Permutation & Elementwise majority  voting \\
     \hline
     $\vrule width 0pt height 0.35cm \sR^N$ &  Any\textsuperscript{*} & \makecell{Permutation, \\ Affine} & Expected value \\
     \hline
     $\vrule width 0pt height 0.35cm \mathbb{R}^{N}$ &  Any\textsuperscript{*} & \makecell{Permutation, \\ Elementwise linear} & Elementwise median \\
     \hline
     Any &  Any & \makecell{Isometry w.r.t.\ $d_\mathrm{out}$\\(e.g.\ Euclidean isometries for $\ell_2$)} &  Center
     \\
     \hline
    \end{tabular}
\end{center}
\end{table}

\clearpage

%% file: appendix/graph_edit_distance.tex
\section{Graph edit distance certificates}
\label{appendix:graph_edit_distance}

In the following, we show how the different existing approaches for proving the robustness of graph neural networks operating on $\{0,1\}^{N \times D} \times \{0,1\}^{N \times N}$ w.r.t\ to distance $\din$
\begin{equation}
    c_\mX^+ \cdot ||(\mX' - \mX)_{+}||_0 +  c_\mX^- \cdot ||(\mX' - \mX)_{-}||_0
    + c_\mA^+ \cdot ||(\mA' - \mA)_{+}||_0 +  c_\mA^- \cdot ||(\mA' - \mA)_{-}||_0, 
\end{equation}
with costs $c_\mX^+, c_\mX^-, c_\mA^+, c_\mA^- \in \{\infty, 1\}$ can be generalized to non-uniform costs, i.e.\ $c_\mX^+, c_\mX^-, c_\mA^+, c_\mA^- \in \{\infty\} \cup \sR_+$.
Combined with~\cref{proposition:certification_reduction}, this yields robustness guarantees w.r.t.\ to the graph edit distance, where the edit operations are insertion and deletion of edges and/or node attributes. 
Note that we also consider local budgets $\rho_1,\dots,\rho_N$, which we introduce in~\cref{appendix:local_robustness}.

Before making the different generalizations, we discuss how to solve Knapsack problems with local constraints, which will be used in many of the subsequent derivations.

Please note that providing an in-depth explanation and motivation for each of the certification procedures would be out of scope.
Instead, we strongly recommend first  reading the original papers. Our discussions are to be understood as specifying changes that have to be made relative to the original certification procedures.

\subsection{Knapsack problems with local constraints}
\label{appendix:knapsack}
A recurring problem in proving the robustness of graph neural networks is selecting a set of edges or attributes that have the largest effect on a classifier's prediction while complying with global and node-wise constraints on the cost of perturbations.
This is a form of Knapsack problem.
\begin{definition}\label{definition:local_knapsack}
A knapsack problem with local constraints is a binary integer program of the form
\begin{equation}
    \begin{split}\label{eq:local_knapsack}
    &\max_{\mQ \in \{0,1\}^{N \times M}} \sum_{n=1}^N \sum_{m=1}^M \emV_{n,m} \cdot \emQ_{n,m} \\
    \text{s.t. } &
    \sum_{n=1}^N \sum_{m=1}^M \emC_{n,m} \cdot \emQ_{n,m} \leq \epsilon, 
    \quad
    \forall n :  \sum_{m=1}^M \emC_{n,m} \cdot \emQ_{n,m} \leq \rho_n,
\end{split}
\end{equation}
with value matrix $\mV \in \sR^{N \times M}$, cost matrix $\mC \in \sR_+^{N \times M}$, global budget $\epsilon \in \sR_+$ and local budgets
$\rho_1, \dots, \rho_N \in \sR_+$.
\end{definition}
Matrix $\mQ$ indicates which entry of an attribute or adjacency matrix should be adversarially perturbed.

When there are only two distinct costs, i.e. we are only concerned with attribute or adjacency perturbations, the above problem can be solved exactly via dynamic programming.
Alternatively, an upper bound can be found via linear relaxation, i.e.\ optimization over $[0,1]^{N \times M}$.

\subsubsection{Dynamic programming}
In the following, we assume that there is an index set $\sI \subseteq \{1,\dots, N\} \times \{1,\dots,M\}$ with complement $\overline{\sI}$ such that 
$\forall (n, m) \in \sI : \mC_{n,m} = c^+$ and $\forall (n, m) \in \overline{\sI} : \mC_{n,m} = c^-$, i.e.\ we only have two distinct costs.
In this case, the problem from~\cref{eq:local_knapsack} can be solved in a two-step procedure, which generalizes the certification procedure for uniform costs  discussed in Section 3 of~\cite{Jin2020}.

The first step is to find optimal solutions per row $r$, while ignoring the global budget $\epsilon$.
We precompute $N$ dictionaries\footnote{or sparse matrices} $\alpha_1, \dots, \alpha_N : \sN \times \sN \rightarrow \sR_+$ and $N$ dictionaries $\beta_1, \dots, \beta_N : N \times N \rightarrow \mathcal{P}(\{1,\dots, M\})$, where $\mathcal{P}$ is the powerset. Entry $\alpha_r(i,j)$ is the optimal value of~\cref{eq:local_knapsack} if only row $\mQ_r$ is non-zero, and exactly $i$ indices in $\sI$ and $j$ indices in $\overline{\sI}$ are non-zero, i.e.\ 
\begin{equation*}
    \begin{split}
    \alpha_r(i,j) = 
    &\max_{\mQ \in \{0,1\}^{N \times M}} \sum_{n=1}^N \sum_{m=1}^M \emV_{n,m} \cdot \emQ_{n,m} \\
    & \text{s.t. } 
    \sum_{n,m \in \sI}   \mQ_{n,m} = i,
    \quad
    \sum_{n,m \in \overline{\sI}}   \mQ_{n,m} = j,
    \quad
    \mQ_{-r} = \mZero.
\end{split}
\end{equation*}

Entry $\beta_r(i,j)$ is the corresponding set of non-zero indices of the optimal $\mQ_r$. 
Because all entries of value matrix $\mV$ and cost matrix $\mC$ are non-negative, the optimal solution is to just select the indices with the largest value $\emV_{n,m}$.
This pre-processing step is summarized in~\cref{algo:knapsack_precompute}.
Its complexity is in $\mathcal{O}(N \cdot ((\max_n \rho_n)  \mathbin{/} c^+) \cdot ((\max_n \rho_n) \mathbin{/} c^-))$, i.e.\ it scales linearly with the number of rows and the maximum number of non-zero values in a row of  $\mQ$.

The second step is to combine these local optimal solutions while complying with the global budget $\epsilon$ and local budgets $\rho_1, \dots, \rho_N$.
This can be achieved via dynamic programming.
We create dictionaries $s_1,\dots,s_N$, with $s_r(\gamma)$ being the optimal value of~\cref{eq:local_knapsack} when  only allocating budget to the first $r$ rows and incurring an overall cost of $\gamma$:
\begin{equation*}
    \begin{split}
    s_r(\gamma) = 
    &\max_{\mQ \in \{0,1\}^{N \times M}} \sum_{n=1}^N \sum_{m=1}^M \emV_{n,m} \cdot \emQ_{n,m} \\
    & \text{s.t. } 
    \sum_{n,m \in \sI}  c^+ \cdot \mQ_{n,m} 
    +
    \sum_{n,m \in \overline{\sI}}  c^- \cdot \mQ_{n,m}  = \gamma,
    \quad
    \mQ_{r:} = \mZero.
\end{split}
\end{equation*}
The first dictionary, $s_1$, can be generated from the precomputed optimal values $\alpha_1$.
After that, each dictionary $s_r$ can be generated from $s_{r-1}$, while ensuring that we remain within local budget $\rho_n$ and global budget $\epsilon$.
The optimal value is given by $\max_\gamma s_N(\gamma)$.
In addition, we generate dictionaries $t_1,\dots,t_N$, which aggregate the row-wise optima stored in $\beta_1,\dots,\beta_N$ such that $t_N(\gamma^*)$ with $\gamma^* = \mathrm{argmax}_\gamma s_N(\gamma)$ gives us the non-zero entries of the optimal $\mQ$. 
This procedure is summarized in~\cref{algo:knapsack_global}.
Its complexity is in $\mathcal{O}(N \cdot  (\epsilon  \mathbin{/} c^+) \cdot (\epsilon \mathbin{/} c^-) \cdot ((\max_n \rho_n)  \mathbin{/} c^+) \cdot ((\max_n \rho_n) \mathbin{/} c^-))$, i.e.\ it scales linearly with the number of rows, the maximum number of non-zero values in $\mQ$, and the maximum number of non-zero values in a row of $\mQ$.
Note that graph neural networks are generally brittle to small adversarial perturbations, so the algorithm only needs to be executed for small $\epsilon$.


\begin{algorithm}
\caption{Precomputation of local solutions}
\label{algo:knapsack_precompute}
\begin{algorithmic}
\For{$n = 1,\dots,N$}
    \State $\alpha_n, \beta_n \gets \mathrm{dict}(), \mathrm{dict}()$
    \State $\mathrm{best\_idx} \gets \mathrm{argsort\_desc}(\mV_n)$
    \State $\mathrm{best\_add\_idx} \gets \mathrm{best\_idx} \setminus \overline{\sI}$
    \State $\mathrm{best\_del\_idx} \gets \mathrm{best\_idx} \setminus \sI$
    \State $\mathrm{max\_adds} \gets \lfloor \rho_n \mathbin{/} c^+ \rfloor$
    \For{$i=0,\dots,\mathrm{max\_adds}$}
        \State $\mathrm{max\_dels} \gets \lfloor (\rho_n - c^+ \cdot i) \mathbin{/} c^- \rfloor$
        \For{$j=0,\dots,\mathrm{max\_dels}$}
            \State {$\alpha_n(i,j) \gets \mathrm{sum}(\{\mV_{n,k} \mid k \in \mathrm{best\_add\_idx}[: \! i]\})$}
            \State {$\alpha_n(i,j) \gets \alpha_n(i, j) + \mathrm{sum}(\{\mV_{n,k} \mid k \in \mathrm{best\_del\_idx}[: \! j]\})$}
            \State $\beta_n(i,j) \gets \mathrm{set}(\mathrm{best\_add\_idx}[: \! i]) \cup \mathrm{set}(\mathrm{best\_del\_idx}[: \! j])$ 
        \EndFor
    \EndFor
\EndFor
\State \Return $\alpha_1,\dots, \alpha_N$,  $\beta_1, \dots, \beta_N$
\end{algorithmic}
\end{algorithm}

\clearpage 
\begin{algorithm}[H]
\caption{Construction of the global solution from local solutions}
\label{algo:knapsack_global}
\begin{algorithmic}
\State $s_0, t_0 \gets \mathrm{dict}(), \mathrm{dict}()$
\State $s_0(0) \gets 0$
\State $t_0(0) \gets [ \ ]$
\For{$n = 1,\dots,N$}
    \State $s_n, t_n \gets \mathrm{dict}(), \mathrm{dict}()$
    \For{$ \mathrm{prev\_cost} \in \mathrm{keys}(s_{n-1})$}
    \State $b \gets \min(\rho_n, \epsilon - \mathrm{prev\_cost})$
    \State $\mathrm{max\_adds} \gets \lfloor b \mathbin{/} c^+ \rfloor$
    \For{$i=0,\dots,\mathrm{max\_adds}$}
        \State $\mathrm{max\_dels} \gets \lfloor (b - c^+ \cdot i) \mathbin{/} c^- \rfloor$
        \For{$j=0,\dots,\mathrm{max\_dels}$}
            \State $\gamma \gets \mathrm{prev\_cost} + c^+ \cdot i + c^- \cdot j$
            \State $v \gets s_{n-1}(\mathrm{prev\_cost}) + \alpha_n(i, j)$
            \If{$\gamma \notin \mathrm{keys}(s_n) \text{ \textbf{or} } v > s_n(\gamma)$}
                \State $s_n(\gamma) \gets v$
                \State $t_n(\gamma) \gets \mathrm{concatenate}(t_{n-1}(\mathrm{prev\_cost}), \beta(i,j))$
            \EndIf
        \EndFor
    \EndFor
    \EndFor
\EndFor
\State $\gamma^* \gets \mathrm{argmax}_{\gamma \in \mathrm{keys}(s_N)} s_N(\gamma)$
\State \Return $s_N(\gamma^*), t_N(\gamma^*)$
\end{algorithmic}
\end{algorithm}


\subsubsection{Linear relaxation}
The above dynamic programming approach can be generalized to arbitrary cost matrices $\mC \in \sR_+^{N \times M}$. However, it may not scale because it requires iterating over all possible combinations of costs admitted by global budget $\epsilon$ and local budgets $\rho_1, \dots, \rho_N$.
A more efficient alternative is upper-bounding the optimal value of the knapsack problem in~\cref{eq:local_knapsack} by relaxing the binary matrix $\mQ \in \{0,1\}^{N \times M}$ to real values $\mQ \in [0,1]^{N \times M}$:
\begin{equation}
    \begin{split}\label{eq:local_knapsack_relaxed}
    &\max_{\mQ \in \sR_{N \times M}} \sum_{n=1}^N \sum_{m=1}^M \emV_{n,m} \cdot \emQ_{n,m} \\
    \text{s.t. } &
    \sum_{n=1}^N \sum_{m=1}^M \emC_{n,m} \cdot \emQ_{n,m} \leq \epsilon, 
    \quad
    \forall n :  \sum_{m=1}^M \emC_{n,m} \cdot \emQ_{n,m} \leq \rho_n, \\
    & \forall n,m : 0 \leq \emQ_{n,m}, \quad \forall n,m : \emQ_{n,m} \leq 1.
\end{split}
\end{equation}
Intuitively, it is best to set those $Q_{n,m}$ to $1$ that have the largest value-to-cost ratio, i.e.\ the largest $\emV_{n,m} \mathbin{/} \emC_{n,m}$. To comply with the local and global budgets, one should first greedily select the $(n, m)$ with the largest ratio in each row $n$, until the local budgets $\rho_n$ are exhausted.
One should then aggregate all these row-wise optimal indices and select those with the largest ratio, until the global budget $\epsilon$ is exhausted.
This procedure is summarized in~\cref{algo:knapsack_relaxed}.

\clearpage

\begin{algorithm}
\caption{Optimal value of the linearly relaxed knapsack problem with local constraints (\cref{eq:local_knapsack_relaxed}}
\begin{algorithmic}\label{algo:knapsack_relaxed}
\State $\mQ^* \gets \mZero$ \Comment{Initialize global allocations with $N \times M$ zeros.}
\State $\mL \gets \mZero$ \Comment{Initialize row-wise allocations with $N \times M$ zeros.}
\State $\mR \gets \mV \oslash \mC$ \Comment{Elementwise division. Division by $0$ is defined as $\infty$.}
\For{$n = 1,\dots,N$}
    \State $\mathrm{best\_cols} \gets \mathrm{argsort\_desc}(\mR_n)$
    \State $b \gets \rho_n$
    \For{$m \in \mathrm{best\_cols}$}
        \State $\emL_{n,m} \gets b \mathbin{/} \emC_{n,m}$
        \State $\emL_{n, m} \gets \min(1, \emL_{n,m}))$ \Comment{Clip, to avoid allocating values larger than $1$.}
        \State $b \gets b - \emC_{n,m} \cdot \emL_{n,m}$
    \EndFor
\EndFor
\State $\mathrm{best\_idx} \gets \mathrm{argsort\_desc}(\mR)$
\State $b \gets \epsilon$
\For{$(n,m) \in \mathrm{best\_idx}$}
    \State $\emQ^*_{n,m} \gets b \mathbin{/} \emC_{n,m}$
    \State $\emQ^*_{n,m} \gets \min(\emL_{n,m}, \emQ^*_{n,m})$ \Comment{Clip, to avoid violating local budget constraints.}
    \State $b \gets b - \emC_{n,m} \cdot \emQ^*_{n,m}$
\EndFor
\State \Return $\sum_{n,m} \emV_{n,m} \cdot \emQ^*_{n,m}$
\end{algorithmic}
\end{algorithm}

\begin{proposition}\label{proposition:correctess_knapsack}
    \cref{algo:knapsack_relaxed} yields the optimal value of the linearly relaxed knapsack problem with local constraints, as defined in~\cref{eq:local_knapsack_relaxed}.
\end{proposition}
\begin{proof}
    To simplify the proof, we make certain assumptions about budgets $\epsilon$ and $\rho_1, \dots, \rho_N$.
    If $\epsilon = 0$, then the value of $\mQ$ computed by~\cref{algo:knapsack_relaxed} is obviously correct, because the only feasible solution that does not violate the global budget is $\mQ^* = \mZero$.
    If $\rho_n = 0$ for some $n$, then the value of $\mQ_n$ computed by~\cref{algo:knapsack_relaxed} is obviously correct, because the only feasible solution that does not violate the local budget is $\mQ^*_n = \mZero$.
    We thus assume w.l.o.g.\ that $\epsilon > 0$ and $\forall n: \rho_n > 0$.
    Furthermore, we assume that the budgets do not exceed the maximum overall and row-wise cost, i.e. $\epsilon \leq \sum_{n=1}^N \sum_{m=1}^M \emC_{n,m}$ and $\forall n: \sum_{m=1}^M \emC_{n,m} \leq \rho_n$, since any excess budget will not have any effect on the optimal value of~\cref{eq:local_knapsack_relaxed}.

In the following, we prove the correctness of~\cref{algo:knapsack_relaxed} by verifying that the constructed solution fulfills the Karush-Kuhn-Tucket conditions.
We define the Lagrangian 
\begin{equation*}    \begin{split}
    &L(\mQ, \lambda, \vkappa, \mT, \mU) = 
    \Bigg(
     - \sum_{n=1}^N \sum_{m=1}^M \emV_{n,m} \cdot \emQ_{n,m} 
    + \lambda 
    \left(\sum_{n=1}^N \sum_{m=1}^M \emC_{n,m} \cdot \emQ_{n,m} - \epsilon \right) 
    \\
    &
    +
    \sum_{n=1}^N 
    \kappa_n 
    \left(\sum_{m=1}^M \emC_{n,m} \cdot \emQ_{n,m} - \rho_n\right)
    - \sum_{n=1}^N \sum_{m=1}^M
    \emT_{n,m}  \cdot 
     \emQ_{n,m}
    + \sum_{n=1}^N \sum_{m=1}^M
    \emU_{n,m} \cdot 
     \left(\emQ_{n,m} - 1\right)
     \Bigg) .
     \end{split}
\end{equation*}
We introduce a negative sign in the objective, because~\cref{eq:local_knapsack_relaxed} is a maximization problem.
Variable $\lambda \in \sR_+$ corresponds to the global budget constraint, variable $\vkappa \in \sR_+^{N}$ correspond to the local budget constraints and variables $\mT, \mU \in \sR_+^{N \times M}$ correspond to the constraint that $\mQ$ should be in $[0,1]^{N \times M}$.

We claim that $\mQ$, $\lambda$, $\vkappa$, $\mT$ and $\mU$ have the following optimal values:
\begin{itemize}
    \item The optimal value of $\mQ$ is the $\mQ^*$ computed by~\cref{algo:knapsack_relaxed}.
    \item The optimal value of $\lambda$ is $\lambda^* = \min_{n,m} \emV_{n,m} \mathbin{/} \emC_{n,m} \text{ s.t. } \emQ^*_{n,m} > 0$, with $\mQ^*$ computed as in~\cref{algo:knapsack_relaxed}. That is, $\lambda^*$ is the smallest value-to-cost ratio for which some global budget is allocated.
    \item Define $o_n = \min_{m} \emV_{n,m} \mathbin{/} \emC_{n,m} \text{ s.t. } \emL_{n,m} > 0$ with $\mL$ computed as in~\cref{algo:knapsack_relaxed}.
    That is, $o_n$ is the smallest value-to-cost ratio in row $n$ for which some local budget is allocated.
    The optimal value of $\kappa_n$ is $\kappa_n^* = \max(0, o_n - \lambda^*)$.
    \item The optimal value of $\emU_{n,m}$ is $\emU_{n,m}^* = \max(\emV_{n,m} - \emC_{n,m} \cdot (\lambda^* + \kappa^*_n), 0)$.
    \item The optimal value of $\emT_{n,m}$ is $\emT_{n,m}^* = - \emV_{n,m}  + \emC_{n,m} \cdot(\lambda^* + \kappa^*_n) + \emU_{n,m}$, which is equivalent to
    $\max(0, -\emV_{n,m} + \emC_{n,m} \cdot (\lambda^* + \kappa^*_n))$.
\end{itemize}

\textbf{Stationarity. }
By the definition of $\mT^*$, we trivially have
\begin{equation*}    \nabla_{\emQ_{n,m}} L(\mQ^*, \lambda^*, \vkappa^*, \mT^*, \mU^*) = - \emV_{n,m}  + \emC_{n,m} \cdot(\lambda^* + \kappa_n^*) - \emT^*_{n,m} + \emU^*_{n,m} = 0.
\end{equation*}

\textbf{Primal feasibility.} The clipping in~\cref{algo:knapsack_relaxed} ensures that $\forall n,m : 0 \leq \emQ^*_{n,m} \leq 1$ and that global budget $\epsilon$ and local budgets $\rho_1, \dots, \rho_N$ are never exceeded.

\textbf{Dual feasibility.} Because all values $\mV$ and all costs $\mC$ are non-negative, the optimal values $\lambda^*$, $\vkappa^*$  are non-negative. The optimal values $\mT^*$ and $\mU^*$ are non-negative by definition.

\textbf{Complementary slackness.} 
We first verify complementary slackness for variable $\lambda$. 
Because we assume that the global budget $\epsilon$ does not exceed the maximum overall cost $\sum_{n=1}^N \sum_{m=1}^M \emC_{n,m}$ and all values $\mV$ are non-negative,~\cref{algo:knapsack_relaxed} will always allocate the entire global budget, i.e.
\begin{equation*}
    \lambda^* \left(\sum_{n=1}^N \sum_{m=1}^M \emC_{n,m} \cdot \emQ^*_{n,m} - \epsilon \right) = \lambda^* \cdot 0 = 0.
\end{equation*}

Next, we verify complementary slackness for variable $\vkappa$.
We can make a case distinction, based on the value of $\kappa_n^* = \max(0, o_n - \lambda^*)$. If $o_n \leq \lambda^*$, we have
\begin{equation*}
    \kappa_n \cdot 
    \left(\sum_{m=1}^M \emC_{n,m} \cdot \emQ^*_{n,m} - \rho_n\right) = 0 \cdot 
    \left(\sum_{m=1}^M \emC_{n,m} \cdot \emQ^*_{n,m} - \rho_n\right) = 0.
\end{equation*}
If $o_n > \lambda^*$, then -- by definition -- the smallest value-to-cost ratio for which budget is allocated in row $n$ is larger than the smallest value-to-cost ratio for which budget is  allocated overall. Combined with our assumption that the local budget $\rho_n$ does not exceed the overall cost $\sum_{m=1}^M \emC_{n,m}$ in row $n$, this means that the entire local budget $\rho_n$ is used up, i.e.
\begin{equation*}
    \kappa^*_n \cdot 
    \left(\sum_{m=1}^M \emC_{n,m} \cdot \emQ^*_{n,m} - \rho_n\right) = \kappa^*_n \cdot 
    0 = 0.
\end{equation*}

Next, we verify complementary slackness for variable $\mT$.
If $\emQ^*_{n,m} = 0$, we obviously have $\emT^*_{n,m} \cdot (-\emQ^*_{n,m}) = 0$.
For  $\emQ^*_{n,m} > 0$, recall that $\emT^*_{n,m} = \max(0, -\emV_{n,m} + \emC_{n,m} \cdot (\lambda^* + \kappa^*_n))$.
Because we defined $\kappa_n^* = \max(0, o_n - \lambda^*)$, we have $\lambda^* + \kappa^*_n = \max(\lambda^*, o_n)$.
We can thus verify $\emV_{n,m} \geq \emC_{n,m} \cdot \lambda^*$ and $\emV_{n,m} \geq \emC_{n,m} \cdot o_n$ to show that $\max(0, -\emV_{n,m} + \emC_{n,m} \cdot (\lambda^* + \kappa^*_n)) = 0$.

We defined $\lambda^*$ to be the smallest value-to-cost ratio for which $\emQ^*_{n,m} > 0$.
We thus know that $\emV_{n,m} \geq \emC_{n,m} \cdot \lambda^*$ whenever $\emQ^*_{n,m} > 0$.

We defined $o_n$ to be the largest value-to-cost ratio for which $\emL_{n,m} > 0$. Due to the clipping in~\cref{algo:knapsack_relaxed}, we know that $\emL_{n,m} \geq \emQ^*_{n,m}$.
We thus know that, if $\emQ^*_{n,m} > 0$, then $\emL_{n,m} > 0$ and thus $\emV_{n,m} \geq \emC_{n,m} \cdot o_n$.

Combining these two results confirms that $\emT^*_{n,m} = 0$ if $\emQ^*_{n,m} > 0 $ and thus $\emT^*_{n,m} \cdot (-\emQ^*_{n,m}) = 0$.

Finally, we verify complementary slackness for variable $\mU$ using a similar argument.
If $\emQ^*_{n,m} = 1$, we trivially have  $\emU^*_{n,m} \cdot (\emQ^*_{n,m} - 1) = 0$.
For $\emQ^*_{n,m} < 1$, recall that $\emU_{n,m}^* = \max(\emV_{n,m} - \emC_{n,m} \cdot (\lambda + \kappa_n), 0)$.
We can again use that $\lambda^* + \kappa^*_n = \max(\lambda^*, o_n)$. There are two only two potential causes for $\emQ^*_{n,m} < 1$. The first one is that all of the local budget $\rho_n$ was used up in column $m$ of row $n$ or columns with higher value-to-cost ratio. In this case, we have $V_{n,m} \leq C_{n,m} \cdot  o_n$.
The other one is that all of the global budget $\epsilon$ was used up in entry $(n,m)$ or entries with higher value-to-cost ratio. In this case, we have $\emV_{n,m} \leq C_{n,m} \cdot \lambda^*$.
Overall, this implies that $\emU^*_{n,m} \cdot (\emQ^*_{n,m} - 1) = 0 \cdot (\emQ^*_{n,m} - 1) = 0$.

\end{proof}

\subsection{Interval bound propagation for attribute perturbations}\label{appendix:certs_ibp}
\citet{Liu2020graphcert} propose to prove the robustness of $L$-layer graph convolutional networks~\cite{Kipf2017} to attribute perturbations with uniform cost
($c_\mX^+ = c_\mX^- = 1, c_\mA^+ = c_\mA^- = \infty$) via interval bound propagation.
In the following, we generalize their guarantees to arbitrary attribute perturbation costs $c_\mX^+, c_\mX^- \in \sR_+$.

Graph convolutional networks applied to attributes $\mX \in \{0,1\}^{N \times D}$ and adjacency $\mA \in \{0,1\}^{N \times N}$ are defined as
\begin{equation}\label{eq:definition_gcn}
    \begin{split}
    &\mH^{(0)} = \mX \\
    & \mZ^{(l)} = \tilde{\mA} \mH^{(l)} \mW^{(l)} + \vone (\vb^{(l)})^T \quad \text{ for } l=0,\dots,L-1 \\
    & \mH^{(l)} = \sigma^{(l)}(\mZ^{(l-1)})  \, \, \qquad  \qquad \quad \text{ for } l=1,\dots,  L,
    \end{split}
\end{equation}
where $\mW^{(0)},\dots,\mW^{(L-1)}$ are weight matrices, $\vb^{(0)},\dots,\vb^{(L-1)}$ are bias vectors, $\sigma^{(1)},\dots,\sigma^{(L)}$ are  activation functions (here assumed to be $\mathrm{ReLU}$ in the first $L-1$ layers and $\mathrm{softmax}$ in the last layer), and $\tilde{\mA}$ is the adjacency matrix after additional preprocessing steps, such as degree normalization.

Given a perturbation set $\sB$, interval bound propagation proves the robustness of a prediction $y_n = \argmax_{k} \emH^{(L)}_{n,k} = \argmax_{k} \emZ^{(L-1)}_{n,k}$ by computing elementwise lower and upper bounds
$\mR^{(l)}$ and $\mS^{(l)}$ on the pre-activation values $\mZ^{(l)}$ and elementwise lower and upper bounds 
$\hat{\mR}^{(l)}$ and $\hat{\mS}^{(l)}$ on the post-activation values $\mH^{(l)}$ via interval arithmetic.
If $\forall k \neq y_n : \emR^{(L-1)}_{n,y_n} > \emS^{(L-1)}_{n,k}$, then the prediction $y_n$ for node $n$ is provably robust.

In our case, the perturbation set is
\begin{equation}\label{eq:perturbation_set_attributes}
    \begin{split}
     \sB = \big\{(\mX', \mA') 
     \big| 
     & \mX \in \{0,1\}^{N \times D}, \mA \in \{0,1\}^{N \times N}, \\
     & \mA' = \mA,
      \\
      &
      c_\mX^+ \cdot ||(\mX' - \mX)_+||_0 + c_\mX^- \cdot ||(\mX' - \mX)_-||_0 \leq \epsilon,
      \\
      & 
      \forall n: c_\mX^+ \cdot ||(\mX_n' - \mX_n)_+||_0 + c_\mX^- \cdot ||(\mX_n' - \mX_n)_-||_0 \leq \rho_n
      \big\},
     \end{split}
\end{equation}
with global budget $\epsilon$ and local budgets $\rho_n$.

We propose to compute the lower and upper bounds $\mR^{(0)}$ and $\mS^{(0)}$ for the first pre-activation values by solving a knapsack problem with local constraints (see~\cref{definition:local_knapsack}) via the methods discussed in~\cref{appendix:knapsack}.
We define the cost matrix $\mC \in \sR_+^{N \times D}$ to be  
\begin{equation*}
    \emC_{n,d} = \begin{cases}
        c_\mX^+ & \text{if } \emX_{n,d} = 0 \\
        c_\mX^- & \text{if } \emX_{n,d} = 1.
    \end{cases}
\end{equation*}

To obtain an upper bound $\emS^{(0)}_{i,j}$, we define the value matrix $\mV   \in \sR_+^{N \times D}$ to be 
\begin{equation*}
    \emV_{n,d} = 
    \begin{cases}
        \tilde{\emA}_{i,n} \cdot \max(\emW^{(0)}_{d,j}, 0) &  \text{if } \emX_{n,d} = 0 \\
        \tilde{\emA}_{i,n} \cdot \max(-\emW^{(0)}_{d,j}, 0) &  \text{if } \emX_{n,d} = 1.
    \end{cases}
\end{equation*}
That is, setting an attribute $\emX_{n,d}$ from $0$ to $1$ makes a positive contribution to upper bound $\mS^{(0)}_{i,j}$ if the corresponding entry of the weight matrix is positive and there is an edge between nodes $n$ and $i$.
Setting an attribute $\emX_{n,d}$ from $1$ to $0$ makes a positive contribution, if the corresponding entry of the weight matrix is negative and there is an edge between nodes $n$ and $i$.
We then solve the knapsack problem and add its optimal value to the original, unperturbed pre-activation value $\emZ^{(0)}_{i,j}$. 

To obtain a lower bound $\emR^{(0)}_{i,j}$, we analogously define the value matrix $\mV   \in \sR_+^{N \times D}$ to be 
\begin{equation*}
    \emV_{n,d} = 
    \begin{cases}
        \tilde{\emA}_{i,n} \cdot \max(-\emW^{(0)}_{d,j}, 0) &  \text{if } \emX_{n,d} = 0 \\
        \tilde{\emA}_{i,n} \cdot \max(\emW^{(0)}_{d,j}, 0) &  \text{if } \emX_{n,d} = 1,
    \end{cases}
\end{equation*}
multiply the optimal value of the knapsack problem with $-1$ and  then add it to the original, unperturbed pre-activation value $\emZ^{(0)}_{i,j}$. 

After we have obtained the lower and upper bounds for the first layer's pre-activation values, we can use the same procedure as~\citet{Liu2020graphcert} to bound the subsequent post- and pre-activation values:
\begin{align*}
    \hat{\mR}^{(l)} &= \mathrm{ReLU}(\mR^{(l-1)})    \qquad \quad \qquad \qquad \qquad  \qquad \qquad \text{ for } l=1,\dots, L-1, \\
    \hat{\mS}^{(l)} &= \mathrm{ReLU}(\mS^{(l-1)}) s  \qquad \quad \qquad \qquad \qquad   \  \ \quad \qquad \text{ for } l=1,\dots, L-1, \\
    \mR^{(l)} &= (\tilde{\mA} \hat{\mR}^{(l)}) \mW^{(l)}_+ + (\tilde{\mA} \hat{\mS}^{(l)}) \mW^{(l)}_- + \vone (\vb^{(l)})^T \qquad \text{ for } l=1,\dots, L-1, \\
    \mS^{(l)} &= (\tilde{\mA} \hat{\mR}^{(l)}) \mW^{(l)}_- + (\tilde{\mA} \hat{\mS}^{(l)}) \mW^{(l)}_+ + \vone (\vb^{(l)})^T \qquad \text{ for } l=1,\dots, L-1. \\
\end{align*}
with $\mW_+ = \max(\mZero, \mW)$ and $\mW_- = \min(\mZero, \mW)$. 
If $\forall k \neq y_n : \emR^{(L-1)}_{n,y_n} > \emS^{(L-1)}_{n,k}$, then the prediction $y_n$ for node $n$ is provably robust.
In our experiments, we use linear programming and not the exact dynamic programming solver for the knapsack problems (see~\cref{appendix:knapsack}).

\subsection{Convex outer adversarial polytopes for attribute perturbations}\label{appendix:graph_edit_distance_convex}
\citet{Zuegner2019cert} also propose robustness guarantees for $L$-layer graph convolutional networks (see~\cref{eq:definition_gcn} under attribute perturbations with uniform cost
($c_\mX^+ = c_\mX^- = 1, c_\mA^+ = c_\mA^- = \infty$).
However, their method solves a a convex optimization problem and yields robustness guarantees that are always at least as strong as interval bound propagation. 
Again, we want to generalize their guarantees to arbitrary attribute perturbation costs $c_\mX^+, c_\mX^- \in \sR_+$.

Let $\mZ^{(L-1)}(\mX,\mA)$ be the pre-activation values in the last layer, given attribute matrix $\mX \in \{0,1\}^{N \times D}$ and adjacency matrix $\mA \in \{0,1\}^{N \times N}$.

A prediction $y_n = \argmax_{k} \mZ^{(L-1)}(\mX,\mA)_{n,k}$ is robust to all perturbed $\mX', \mA'$ from a perturbation set $\sB$ if
$\forall k \neq y_n, \forall (\mX', \mA') \in \sB : \mZ^{(L-1)}(\mX',\mA')_{n,y_n} > \mZ^{(L-1)}(\mX',\mA')_{n,k}$.
This is equivalent to showing that, for all $k \neq y_n$:
\begin{equation}\label{eq:attribute_margin_bound}
    \min_{(\mX', \mA') \in \sB} \mZ^{(L-1)}(\mX',\mA')_{n,y_n} - \mZ^{(L-1)}(\mX',\mA')_{n,k} > 0.
\end{equation}

Solving these $k$ optimization problems is generally intractable due to the $\mathrm{ReLU}$ nonlinearities and the discrete set of perturbed graphs $\sB$. \citet{Zuegner2019cert} thus propose to make two relaxations. The first relaxation is to relax the integrality constraints on the perturbation set $\sB$. In our case, this leads to the relaxed  set
\begin{equation}\label{eq:perturbation_set_attributes_relaxed}
    \begin{split}
     \sB = \big\{(\mX', \mA') 
     \big| 
     & \mX \in [0,1]^{N \times D}, \mA \in [0,1]^{N \times N}, \\
     & \mA' = \mA,
      \\
      &
      c_\mX^+ \cdot ||(\mX' - \mX)_+||_1 + c_\mX^- \cdot ||(\mX' - \mX)_-||_1 \leq \epsilon,
      \\
      & 
      \forall n: c_\mX^+ \cdot ||(\mX_n' - \mX_n)_+||_1 + c_\mX^- \cdot ||(\mX_n' - \mX_n)_-||_1 \leq \rho_n
      \big\}.
     \end{split}
\end{equation}
Note that we now use $\ell_1$ instead of $\ell_0$ norms.

The second relaxation is to relax the strict functional dependency between pre- and post-activation values 
$\mH^{(l)} = \mathrm{ReLU}(\mZ^{(l)})$ to its convex hull:
\begin{equation*}
    \left(\emZ^{(l)}_{n,d}, \emH^{(l)}_{n,d}\right) \in \mathrm{hull}\left(\left\{a, \mathrm{ReLU}(a) \in \sR \mid \emR^{(l)}_{n,d} \leq a \leq  \emS^{(l)}_{n,d} \right\}\right),
\end{equation*}
where $\mR^{(l)}$ and $\mS^{(l)}$ are elementwise lower and upper bounds on the pre-activation values $\mZ^{(l)}$.
We propose to use the modified interval bound propagation method introduced in~\cref{appendix:certs_ibp} to obtain these bounds.

These two relaxations leave us with an optimization problem that is almost identical to the one discussed in~\cite{Zuegner2019cert}.
We just weight insertions and deletions with constants $c_\mX^+$ and $c_\mX^+$ instead of uniform weight $1$ and have different values for the elementwise pre-activation bounds. We can thus go through the same derivations as in the proof of Theorem 4.3 in~\cite{Zuegner2019cert} while tracking the constants $c_\mX^+$ and $c_\mX^+$ to derive the dual of our relaxed optimization problem, which lower-bounds~\cref{eq:attribute_margin_bound}. We highlight the difference to the result for uniform costs in red.
The dual problem for local budgets $\vrho$ and global budget $\epsilon$ is
\begin{equation}\label{eq:daniels_dual}
    \begin{split}
    &\max_{\mOmega^{(1)},\dots,\mOmega^{(L-1)}, \vkappa, \lambda} g_{\vrho,\epsilon}(\mX, \mOmega, \vkappa, \lambda) \\
    \text{s.t. } 
    &
    \mOmega^{(l)} \in [0,1]^{N \times h^{(l)}} \text{ for } l=  L-1,\dots, \textcolor{red}{1},
    \\
    &
    \vkappa \in \sR_+^N, \lambda \in \sR_+,
    \end{split}
\end{equation}
where $h^{(l)}$ is the number of neurons in layer $l$ and
\begin{equation*}
    \begin{split}
    g_{\vrho,\epsilon}(\mX, \mOmega, \kappa, \lambda) = 
    &
    \sum_{l=
    \textcolor{red}{1}}^{L-1} \sum_{(n,j) \in \mathcal{I}^{(l)}} \frac{\emS^{(l)}_{n,j} \emR^{(l)}_{n,j}}{\emS^{(l)}_{n,j} - \emR^{(l)}_{n,j}} \left[\hat{\emPhi}^{(l)}_{n,j}\right]_+
    - \sum_{l=\textcolor{red}{0}}^{L-1} \vone^T \mPhi^{(l+1)} \vb^{(l)}
    \\
    &
    - \mathrm{Tr}\left[\mX^T \hat{\mPhi}^{(\textcolor{red}{0})}\right]
    - ||\mPsi||_1 -  \sum_{n=1}^N \rho_n \cdot \kappa_n - \epsilon \cdot \lambda,
    \end{split}
\end{equation*}
where $\mathcal{I}^{(l)}$ is the set of unstable neurons in layer $l$, i.e. $\mathcal{I}^{(l)} = \{(n,j) \mid \emR^{(l)}_{n,j} \leq 0 < \emS^{(l)}_{n,j}\}$, and
\begin{align*}
    \emPhi^{(L)}_j & = \begin{cases}
        - 1 & \text{if } j = y^* \\
        1 & \text{if } j = k \\
        0 & \text{otherwise}
    \end{cases}
    \\
    \hat{\mPhi}^{(l)} & = \tilde{\mA}^T \mPhi^{(l+1)} (\mW^{(l)})^T \quad \text{ for } \qquad \qquad \qquad \, \qquad \qquad \qquad \qquad  l= L-1,\dots, \textcolor{red}{0} \\
    \emPhi^{(l)}_{n,j} & = \begin{cases}
        0 & \text{if }   \emS^{(l)}_{n,j} \leq 0 \\
        \hat{\emPhi}_{n,j} & \text{if } \emR^{(l)}_{n,j} > 0 \\
        \frac{\emS^{(l)}_{n,j}}{\emS^{(l)}_{n,j} - \emR^{(l)}_{n,j}} \left[\hat{\emPhi}^{(l)}_{n,j}\right]_+ 
        \textcolor{red}{+} \ \emOmega^{(l)}_{n,j} \left[ \hat{\emPhi}^{(l)}_{n,j} \right]_-
        & \text{if } (n,j) \in \mathcal{I}^{(l)}
    \end{cases} \qquad \text{ for } l = L-1,\dots, \textcolor{red}{1}
    \\
    \emPsi_{n,d} & = \begin{cases}
        \max\left(\emDelta_{n,d} - \textcolor{red}{c_\mX^+} \cdot (\kappa_n + \lambda), 0\right) & \text{if } \emX_{n,d} = 0 \\
        \max\left(\emDelta_{n,d} - \textcolor{red}{c_\mX^-} \cdot (\kappa_n + \lambda), 0\right) & \text{if } \emX_{n,d} = 1
    \end{cases} \\
    \emDelta_{n,d} & = \left[\hat{\mPhi}^{(\textcolor{red}{0})}\right]_+ \cdot ( 1  - \emX_{n,d})
    \textcolor{red}{-} \left[\hat{\mPhi}^{(\textcolor{red}{0})}\right]_- \cdot \emX_{n,d}.
\end{align*}
The indexing changes are necessary because we consider $L$ and not $(L-1)$-layer networks.
The sign changes are necessary because we define the clipping of negative values differently, i.e.\ $\mX_- = \min(\mX, 0)$ and not $\mX_- = -\min(\mX, 0)$.
Aside from that, the only difference are the costs $c_\mX^+$ and $c_\mX^-$ in the definition of $\mPsi$.

Because this is a dual problem, any choice of $\mOmega$, $\vkappa$ and $\lambda$ yields a lower bound on~\cref{eq:attribute_margin_bound}.
But using optimal parameters yields a tighter bound.
\begin{proposition}
    Define value matrix $\mV = \mDelta$ and cost matrix $\mC = c_\mX^+ \cdot \left[ 1 - \mX \right]_+ + c_\mX^- \cdot \left[\mX \right]_+$.
    Then,  
    \begin{itemize}
        \item The optimal value of $\lambda$ is $\lambda^* = \min_{n,m} \emV_{n,m} \mathbin{/} \emC_{n,m} \text{ s.t. } \emQ^*_{n,m} > 0$, with $\mQ^*$ computed as in~\cref{algo:knapsack_relaxed}. That is, $\lambda^*$ is the smallest value-to-cost ratio for which some global budget is allocated.
        \item Define $o_n = \min_{m} \emV_{n,m} \mathbin{/} \emC_{n,m} \text{ s.t. } \emL_{n,m} > 0$ with $\mL$ computed as in~\cref{algo:knapsack_relaxed}.
        That is, $o_n$ is the smallest value-to-cost ratio in row $n$ for which some local budget is allocated.
        The optimal value of $\kappa_n$ is $\kappa_n^* = \max(0, o_n - \lambda^*)$.
    \end{itemize}
\end{proposition}
\begin{proof}
    For any fixed $\mOmega^{(1)},\dots,\mOmega^{(L-1)}$, 
    we can go through the same derivations as in the proofs of Theorems 4.3 and 4.4 in~\cite{Zuegner2019cert} -- while keeping track of constants $c_\mX^+, c_\mX^-$ -- to show that the above optimization problem is up to an additive constant equivalent to
    \begin{align*}
        \min_{\lambda \in \sR_+, \vkappa \in \sR_+^{N}, \mU \in \sR_+^{N \times D}} & \sum_{n,d} \emU_{n,d} + \sum_{n=1}^N \rho_n \cdot \kappa_n + \epsilon \cdot \lambda\\
        \text{s.t. } &
        \emU_{n,d} \geq \emV_{n,d} - \emC_{n,d} \cdot \kappa_n - \emC_{n,d} \cdot  \lambda
    \end{align*}
    with $\mV = \mDelta$ and $\mC = c_\mX^+ \cdot \left[ 1 - \mX \right]_+ + c_\mX^- \cdot \left[\mX \right]_+$.

    By standard construction, the dual of the above linear program is
    \begin{equation*}
    \begin{split}
    &\max_{\mQ \in \sR_{N \times M}} \sum_{n=1}^N \sum_{m=1}^M \emV_{n,m} \cdot \emQ_{n,m} \\
    \text{s.t. } &
    \sum_{n=1}^N \sum_{m=1}^M \emC_{n,m} \cdot \emQ_{n,m} \leq \epsilon, 
    \quad
    \forall n :  \sum_{m=1}^M \emC_{n,m} \cdot \emQ_{n,m} \leq \rho_n, \\
    & \forall n,m : 0 \leq \emQ_{n,m}, \quad \forall n,m : \emQ_{n,m} \leq 1.
    \end{split}
\end{equation*}
    This is exactly the linearly relaxed knapsack problem with local constraints from~\cref{eq:local_knapsack_relaxed}.
    During our proof of~\cref{proposition:correctess_knapsack} via Karush-Kuhn-Tucket conditions, we have already shown that $\lambda^*$ and $\vkappa^*$ are the optimal values of the dual (here: primal) variables.
\end{proof}
Since we have optimal values for variables $\lambda$ and $\vkappa$ of the optimization problem in~\cref{eq:daniels_dual}, we only need to choose values for $\mOmega^{(1)},\dots,\mOmega^{(L-1)}$. They can be either be optimized via gradient ascent or set to some constant value.
In our experiments, we choose
\begin{equation*}
    \emOmega^{(l)}_{n,j} =  \frac{\emS^{(l)}_{n,j}}{\emS^{(l)}_{n,j} - \emS^{(l)}_{n,j}},
\end{equation*}
as suggested in~\cite{Zuegner2019cert}.
As discussed in~\cite{Zuegner2019cert}, the efficiency of the certification procedure can be further improved by slicing the attribute and adjacency matrix to only contain nodes that influence the hidden representation of node $n$ in each layer $l$.

\subsection{Bilinear programming for adjacency perturbations}\label{appendix:bilinear_programming}
\citet{Zuegner2020} derive guarantees for the robustness of graph convolutional networks to edge deletions ($c_\mX^+ = c_\mX^- = c_\mA^+ = \infty, c_\mA^- = 1$).
Generalizing these guarantees to arbitrary costs $c_\mA^- \in \sR_+$ does not require any additional derivations.
Multiplying the cost by a factor $k$ is equivalent to dividing global budget $\epsilon$ and local budgets $\rho_1,\dots,\rho_n$ by factor $k$.

\subsection{Linearization and dualization for adjacency perturbations}
\citet{Jin2020} derive robustness guarantees for graph classifiers consisting of a $1$-layer graph convolutional network followed by a linear layer and mean pooling  under adjacency perturbations with uniform cost ($c_\mX^+ = c_\mX^- = \infty, c_\mA^+ = c_\mA^- = 1$). 
In the following, we want to generalize their guarantees to arbitrary $c_\mA^+$ and $c_\mA^-$.

In fact, the authors propose two different approaches.
The first one involves linearization of the neural network and the formulation of a Lagrange dual problem to enforce symmetry of the perturbed adjacency matrix.
The second one combines Fenchel biconjugation with convex outer adversarial polytopes to derive a lower bound that is then optimized via conjugate gradients.
While the second objective is sound, its optimization via conjugate gradient may be numerically unstable, which is why an entirely different solver is used in practice.%
\footnote{See~\href{https://github.com/RobustGraph/RoboGraph/blob/master/robograph/attack/cvx_env_solver.py}{https://github.com/RobustGraph/RoboGraph/blob/master/robograph/attack/cvx\_env\_solver.py}}
In order to not mispresent the contents of~\cite{Jin2020}, we focus on the first approach, which offers similarly strong guarantees (see Fig.4 in~\cite{Jin2020}).

The considered graph classification architecture is
\begin{equation*}
    F(\mX, \mA) = \sum_{n=1}^N  \mathrm{ReLU} \left(\mD^{-1} \tilde{\mA} \mX \mW\right) \mU \mathbin{/} N,
\end{equation*}
where $\tilde{\mA} = \mA$ is the adjacency matrix after introducing self-loops, i.e.\ setting all diagonal entries to $1$,
$\mD \in \sR^{N \times N}$ with $\mD_{i,i} = \vone^T  \tilde{\mA}_i $ is the diagonal degree matrix of $\tilde{\mA}$, $\mW \in \sR^{D \times H}$ are the graph convolution weights and $\mU \in \sR^{H \times K}$ are the linear layer weights for $K$ classes.

As before, robustness for a specific prediction $y = \argmax_k F(\mX, \mA)_k$ under a perturbation set $\sB$ can be proven by showing that the classification margin is positive, i.e.
\begin{equation*}
    \forall k \neq y : \min_{(\mX',\mA') \in \sB} F(\mX',\mA')_y - F(\mX',\mA')_k > 0.
\end{equation*}
For this specific architecture, this is equivalent to showing that, for all $k \neq y$,
\begin{equation*}
    \begin{split}
        &\min_{(\mX,\mA) \in \sB} \sum_{n=1}^N f_{n,k}(\mX', \mA') > 0 \\
        \text{with } & f_{n,k}(\mX', \mA') = \left(\vone^T  \tilde{A'}_n \right)^{-1}  \mathrm{ReLU} \left( (\tilde{\mA'}_n)^T \mX' \mW\right)  (\mU_{:,y} - \mU_{:, k}) \mathbin{/} N.
    \end{split}
\end{equation*}
Note that we could move the degree of node $n$ out of the nonlinearity because it is only a scalar factor.
Our perturbation set is
\begin{equation}\label{eq:perturbation_set_adjacency}
    \begin{split}
     \sB = \big\{(\mX', \mA') 
     \big| 
     & \mX \in \{0,1\}^{N \times D}, \mA \in \{0,1\}^{N \times N}, \\
     & \mX' = \mX,
      \\
      &
      c_\mA^+ \cdot ||(\mX' - \mX)_+||_0 + c_\mA^- \cdot ||(\mX' - \mX)_-||_0 \leq \epsilon,
      \\
      & 
      \forall n: c_\mA^+ \cdot ||(\mX_n' - \mX_n)_+||_0 + c_\mA^- \cdot ||(\mX_n' - \mX_n)_-||_0 \leq \rho_n
      \big\}.
     \end{split}
\end{equation}

\textbf{Linearization. }
Unless one wants to find the worst-case perturbation for each $f_{n,k}$ via brute-forcing (which is a viable approach for very small budgets and discussed in~\cite{Jin2020}), 
the first step of the certification procedure is to find $N$ linear models that lower-bound the nodewise functions $f_{n,k}$ for all possible perturbed inputs. That is, we want to find vectors $\vq^{(1)},\dots, \vq^{(N)} \in \sR^{N}$ and scalars $b^{(1)},\dots,b^{(N)} \in \sR^{K}$ such that
\begin{align}
    \forall_{(\mX', \mA') \in \sB} : 
    & f_{n,k}(\mX',\mA') \geq \underline{f_{n,k}}(\mX',\mA') \\
    \label{eq:linearized model}
    & \text{with } \underline{f_{n,k}}(\mX',\mA') = \left(\vone^T  \tilde{A'}_n \right)^{-1}  \left( (\tilde{\mA'}_n)^T \vq^{(n)} + b^{(n)} \right).
\end{align}
Like in~\cite{Jin2020}, we use the linear bounds from CROWN~\cite{Zhang2018}.\footnote{These linear bounds are referred to as doubleL in~\cite{Jin2020}.}
Computing these bounds requires elementwise lower and upper bounds $\vr^{(1)},\dots,\vr^{(N)}$  and $\vs^{(1)},\dots,\vs^{(N)}$  on the pre-activation values $\vz^{(n)} = ( (\tilde{\mA'}_n)^T \mX' \mW )$ with $\vz^{(n)} \in \sR^H$.

For our new perturbation set, we  propose to compute these elementwise lower and upper bounds by solving knapsack problems with local constraints via the algorithms discussed in~\cref{appendix:knapsack}.
To bound the the pre-activation values $\vz^{(n)}$ of the $n$th node, we define the cost matrix $\mC \in \sR_+^{1 \times N}$ as  
\begin{equation*}
    \emC_{1,m} = \begin{cases}
        \infty & \text{if } m = n \\
        c_\mA^+ & \text{if } m \neq n \land \emA_{n,m} = 0 \\
        c_\mA^- & \text{if } m \neq n \land \emA_{n,m} = 1.
    \end{cases}
\end{equation*}
The infinite cost for diagonal elements ensures that they are not adversarially perturbed. Afterall, there is no benefit to attacking elements that are anyway overwritten by self-loops.
The cost matrix has shape $1 \times N$, because in every row $n$, there are only $N$ edges that can be perturbed.

To obtain an upper bound $\evs^{(n)}_{h}$, we define the value matrix $\mV   \in \sR_+^{1 \times N}$ to be 
\begin{equation*}
    \emV_{1,m} = 
    \begin{cases}
        \max\left(\mX_m^T \mW_{:,h}, 0\right) &  \text{if } \emA_{n,m} = 0 \\
        \max\left(-\mX_m^T \mW_{:,h}, 0\right) &  \text{if } \emA_{n,m} = 1.
    \end{cases}
\end{equation*}
We then solve the knapsack problem, which tells us how much the pre-activation value can change under $\sB$, and then add this optimal value to the original, unperturbed pre-activation value $\evz^{(n)}_h$. 

To obtain a lower bound $\evr^{(N)}_{h}$, we analogously define the value matrix $\mV   \in \sR_+^{1 \times N}$ to be 
\begin{equation*}
    \emV_{1,m} = 
    \begin{cases}
        \max\left(-\mX_m^T \mW_{:,h}, 0\right) &  \text{if } \emA_{n,m} = 0 \\
        \max\left(\mX_m^T \mW_{:,h}, 0\right) &  \text{if } \emA_{n,m} = 1.
    \end{cases}
\end{equation*}
multiply the optimal value of the knapsack problem with $-1$ and  then add it to the original, unperturbed pre-activation value $\evz^{(N)}_h$. 

\textbf{Nodewise guarantees. }
The next step is to use the constructed linear lower bounds $\underline{f_{1,k}},\dots, \underline{f_{N,k}}$
to lower-bound the value of  functions
$f_{1,k},\dots, f_{N,k}$ 
for each possible number of perturbations admitted by local budgets $\rho_1,\dots, \rho_N$.
Note that we now need to take the degree normalization into account.

For our new perturbation set, we propose to use~\cref{algo:knapsack_precompute} to perform these precomputations.
We set the cost parameters to $c^+ = c_\mA^+$ and $c^- = c_\mA^-$.
To prevent perturbations of the diagonal entries that are overwritten with self-loops, we set the index set parameters
$\sI$ and $\overline{\sI}$ to
$\sI = \{(1,m) \mid \mA_{n,m} = 0\} \cup \{1,n\}$ and $\overline{\sI} = \{(1,m) \mid \mA_{n,m} = 1\} \cup \{1,n\}$.
We set the global and local budget parameter of the algorithm to $\epsilon$ and $\rho_n$, respectively.
Because~\cref{algo:knapsack_precompute} is defined to solve a maximization and not a minimization problem, we define the value matrix $\mV \in \sR^{1 \times N}$ to be
\begin{equation*}
    \emV_{1,m} = 
    \begin{cases}
        - \evq^{(n)}_m &  \text{if } \emA_{n,m} = 0 \\
        \evq^{(n)}_m &  \text{if } \emA_{n,m} = 1.
    \end{cases}
\end{equation*}
That is, setting $\emA_{n,m}$ from $0$ to $1$ has a large value if  $\evq^{(n)}_m$ is a negative number.
Setting $\emA_{n,m}$ from $1$ to $0$ has a large value if $\evq^{(n)}$ if a large positive number.
\cref{algo:knapsack_precompute} yields a dictionary $\alpha_n$, where $\alpha_n(i,j)$ indicates how much the value of $\underline{f_{n,k}}(\mX, \mA)$ changes when optimally inserting $i$ and deleting $j$ edges.
It further yields a dictionary $\beta_n$, where $\beta_n(i,j)$ contains the optimal set of edges to perturb in row $n$ of the adjacency matrix.
The last thing we need to do is to add them to the unperturbed values $\underline{f_{n,k}}(\mX,\mA)$ while accounting for the fact that perturbing edges also influences the degree and thus the degree normalization.
We propose to do so via~\cref{algo:knapsack_degree_normalization}
\begin{algorithm}
\caption{Computation of  $-1 \cdot \min_{\mX',\mA' \in \sB} \underline{f_{n,k}}$ from the precomputed worst-case changes without degree normalization stored in dictionary $\alpha_n$.}
\label{algo:knapsack_degree_normalization}
\begin{algorithmic}
\State $\mathrm{max\_adds} \gets \lfloor \rho_n \mathbin{/} c^+ \rfloor$
\For{$i=0,\dots,\mathrm{max\_adds}$}
    \State $\mathrm{max\_dels} \gets \lfloor (\rho_n - c^+ \cdot i) \mathbin{/} c^- \rfloor$
    \For{$j=0,\dots,\mathrm{max\_dels}$}
        \State $\mathrm{old\_degree} \gets \vone^T  \tilde{A}_n$
        \State $\mathrm{new\_degree} \gets \mathrm{old\_degree} + i - j$
        \State $\mathrm{new\_degree} \gets \max(1, \min(\mathrm{new\_degree},N))$ \Comment{Between $1$ and $N$, due to self-loop.}
        \State $\alpha_n(i,j) \gets \alpha_n(i,j) - \mathrm{old\_degree} \cdot \underline{f_{n,k}}(\mX,\mA) $
        \State $\alpha_n(i,j) \gets \alpha_n(i,j) \mathbin{/} \mathrm{new\_degree}$
    \EndFor
\EndFor
\State \Return $\alpha_n$
\end{algorithmic}
\end{algorithm}

Note that we compute the negative value of the lower bound, because all discussed algorithms are designed for maximization problems.

\textbf{Combining the nodewise guarantees. }
Now we have, for each node $n$ and every possible number of insertions $i$ and deletions $j$, the negative of a lower bound on $f_{n,k}$, which is stored in  $\alpha_n(i,j)$.
The last step is to combine these nodewise guarantees while complying with global budget $\epsilon$.
To this end, we can just reuse~\cref{algo:knapsack_global} with $c^+ = c_\mA^+$ and $c^- = c_\mA^-$.
It yields a sequence $t$ of length $N$, whose $n$th element is the set of the worst indices to perturb in row $n$ of the adjacency matrix.
It further yields a lower bound on the negative classification margin $-1 \cdot \left(F(\mX',\mA')_y - F(\mX',\mA')_k\right)$ for a specific $k$.
If after performing the above procedure for all $k \neq y$ all classification margins are positive, then the prediction is provably robust.

\textbf{Enforcing symmetry via dualization.} 
An important aspect of the discussion in~\cite{Jin2020} is that one may want to introduce the additional constraint $\mA' = \mA'^T$ when proving robustness for symmetric graphs.
The authors show that 
\begin{align*}
    &\min_{(mX,\mA') \in \sB} \sum_{n=1}^N \underline{f_{n,k}}(\mX',\mA') \quad \text{s.t. } \mA' = \mA'^T \\
    \geq
    & \max_{\mLambda \in \sR^{N \times N}} \min_{(mX,\mA') \in \sB} \sum_{n=1}^N \underline{f_{n,k}}(\mX',\mA')
    + \mathrm{Tr}((\mLambda^T - \mLambda) \mA'),
\end{align*}
with dual variable $\mLambda \in \sR^{N \times N}$.
Note that $\mathrm{Tr}((\mLambda^T - \mLambda) \mA) = \sum_{n=1}^N \sum_{m=1}^M (\mLambda^T - \mLambda)^T_{n,m} \cdot \emA'_{n,m}$.

The inner optimization problem can  be solved exactly via the ``nodewise guarantees'' and ``combining the nodewise guarantees'' steps above, after replacing each parameter vector $\vq^{(n)}$ of the linearized models $\underline{f_{n,k}}$ defined in~\cref{eq:linearized model} with
$\vq^{(n)} + \left(\vone^T \tilde{A'}_n \right) \cdot (\mLambda - \mLambda^T)_{n}$.

Because the inner optimization problem is solved exactly, dual variable $\mLambda$ can be optimized via gradient ascent on $\mathrm{Tr}((\mLambda^T - \mLambda) \mA')$ (Danskin's theorem).

\input{appendix/ged_tmp}

%% file: appendix/ged_tmp.tex
\subsection{Policy iteration for adjacency perturbations}\label{appendix:graph_edit_distance_policy_iteration}
\citet{Bojchevski2019} derive robustness certificates for models where the predictions are a linear function of the (personalized) PageRank.  Specifically, they consider the following architecture called~\(\pi\)-PPNP:
\[
    \bm{Y} = \text{softmax}(\bm{\Pi}\bm{H}),
    \quad
    \bm{H}_{v,:} = f_\theta(\bm{X}_{v,:}),
    \quad
    \bm{\Pi} = (1-\alpha)(\bm{I}_N - \alpha \bm{D}^{-1}\bm{A})^{-1}
\]
where \(\bm{X}\) the feature matrix of the graph, \(f\) is a neural network with parameter \(\theta\), \(\bm{H}_{v,:}\) the prediction for node \(v\)  and \(\bm{\Pi}\) the personalized PageRank matrix with teleport probability \(\alpha\) (i.e.\ \(\bm{\Pi}_{v,:}\) is the personalized PageRank of node \(v\)). Further note that \(\bm{I}_N\) is the identity matrix and \(\bm{D}\) is the degree matrix \(\bm{D}_{ii} = \sum_{j} = \bm{A}_{ij}\) of a graph \(G = (\mathcal{V},\mathcal{E})\) with nodes \(\mathcal{V}\) and edges \(\mathcal{E}\).

Their certificates are against adjacency perturbations with uniform costs ($c_{\bm{A}}^+ = c_{\bm{A}}^- = 1$) and we will generalize their certificate to arbitrary costs for edge insertion and deletion. Let the following set denote all admissible perturbed graphs under costs $c_{\bm{A}}^+$ for edge insertion and $c_{\bm{A}}^-$ for edge deletion:
\begin{equation}
    \begin{aligned}
        \mathcal{Q}_\mathcal{F} = \{ (\mathcal{V}, \tilde{\mathcal{E}}:= \mathcal{E}_f \cup \mathcal{F}_+)\mid & \mathcal{F}_+ \in\mathcal{P}(\mathcal{F}), \\
        & c_{\bm{A}}^+\cdot|\tilde{\mathcal{E}}\setminus\mathcal{E}| + c_{\bm{A}}^-\cdot|\mathcal{E}\setminus\tilde{\mathcal{E}}| \leq \epsilon, \\
        & c_{\bm{A}}^+\cdot|\tilde{\mathcal{E}}^v\setminus\mathcal{E}^v| + c_{\bm{A}}^-\cdot|\mathcal{E}^v\setminus\tilde{\mathcal{E}}^v| \leq \rho_v, \forall v \}
    \end{aligned}
\end{equation}
where \(\mathcal{E}_f\) is a set of fixed edges that cannot be modified, \(\mathcal{F}\subseteq \mathcal{V}\times \mathcal{V}\) the set of fragile edges that can be modified. Here, \(\mathcal{F}_+\subseteq\mathcal{F}\) denotes the set of edges that are included in the perturbed graph (analogously  \(\mathcal{F}_-\subseteq\mathcal{F}\) the set of edges that are not included anymore in the perturbed graph). Here, \(\epsilon\) denotes the global budget and \(\rho_v\) the local budget for node \(v\). 

We assume a fixed graph \(G\), set of fragile edges \(\mathcal{F}\), global budget \(B\) and local budgets \(b_v\).  Note that since \(\pi\)-PPNP separates prediction from propagation we can further assume fixed model logits \(\bm{H}\). 
Following \citet{Bojchevski2019} we further define the worst-case margin between classes \(y_t\) and \(c\) under any perturbed graph \(\tilde{G}\in\mathcal{Q}_\mathcal{F}\) (Problem 1 of \citet{Bojchevski2019}):
\begin{equation}\label{eq:pippnpProblem1}
    m^*_{y_t,c}(t) 
    = \min_{\tilde{G}\in\mathcal{Q}_\mathcal{F}} m_{y_t,c}(t)
    = \min_{\tilde{G}\in\mathcal{Q}_\mathcal{F}} \bm{\pi}_{\tilde{G}}(\bm{e}_t)^T(\bm{H}_{:,y_t} - \bm{H}_{:,c})
\end{equation}
where \(\bm{e}\) is the unit vector and \(\bm{\pi}_{\tilde{G}}(\bm{e}_t)\) the personalized PageRank vector of target node \(t\) under perturbed graph \(\tilde{G}\). 
We want to show that a specific target node \(t\) is certifiably robust w.r.t. the logits \(\bm{H}\) and the set \(\mathcal{Q}_\mathcal{F}\).
This is the case if \(m^*_{y_t,*}(t) = \min_{c\neq y_t}m^*_{y_t,c}(t) > 0\).
In the following we will derive guarantees for this under local and global adversarial budgets, respectively.

\textbf{Local constraints only.}
\citet{Bojchevski2019} phrase the problem of \autoref{eq:pippnpProblem1} as a more general average cost infinite horizon Markov decision process and present a policy iteration algorithm that solves it in polynomial time. Following their derivations we also define \(\bm{r} = -(\bm{H}_{:,y_t} - \bm{H}_{:,c})\), where \(\bm{r}_v\) denotes the rewards a random walker gets for visiting node \(v\) \citep{Bojchevski2019}. The following adapted policy iteration computes the worst-case graph \(\tilde{G}\in\mathcal{Q}_\mathcal{F}\) under arbitrary costs:

\begin{algorithm}
    \caption{Policy Iteration with local budgets under arbitrary costs}
    \label{algo:pippnp-PI}
\begin{algorithmic}[1]
    \Require Graph \(G=(\mathcal{V},\mathcal{E})\), reward \(\bm{r}\), set of fixed edges \(\mathcal{E}_f\), fragile edges \(\mathcal{F}\), local budgets \(b_v\) and costs $c_\mA^+$, for edge insertion and $c_\mA^-$ for edge deletion
    \State Initialization: arbitrary \(\mathcal{W}_0\subseteq \mathcal{F}\), \(\bm{A}^G\) corresponding to \(G\)
    \While {$\mathcal{W}_k \neq \mathcal{W}_{k-1}$}
        \State Solve $(\bm{I}_N-\alpha\bm{D}^{-1}\bm{A})\bm{x}=\bm{r}$ for $\bm{x}$ where $\bm{A}_{ij} = 1 - \bm{A}^G_{ij}$ if $(i,j)\in\mathcal{W}_k$
        \State $l_{ij}\gets (1-2\bm{A}^G_{ij})(\bm{x}_j - \frac{\bm{x}_i - \bm{r}_i}{\alpha})$ for all $(i,j)\in\mathcal{F}$
        \For {$v\in\mathcal{V}$}
            \State $\bm{m} \gets \arg\max_{\bm{m}} \sum_{(v,j)\in F^+\cup F^-} m_{j} l_{v,j} $ 
            s.t. $ \sum_{(v,j)\in F^-} c_{\bm{A}}^- m_{j} + \sum_{(v,j)\in F^+} c_{\bm{A}}^+ m_{j} \leq \rho_v$, 
            $m_{j} \in \{0,1\}$
            \State $\mathcal{L}_v \gets \{ (v,j)\in\mathcal{F} \mid \bm{m}_{j}=1\}$
        \EndFor
        \State \(\mathcal{W}_k\gets \bigcup_v \mathcal{L}_v\)
        \State $k\gets k+1$
    \EndWhile
    \Return $\mathcal{W}_k$
\end{algorithmic}
\end{algorithm}

Line 6 of the policy iteration requires us to find those fragile edges with the largest score $l$ under the local budget $\rho_v$. This is an instance of the Knapsack problem, which we discussed earlier.
In practice we solve this using dynamic programming, specifically we call \cref{algo:knapsack_global} for row-vector $\bm{V}_{1,j} \gets l_{v,j}$, $\epsilon\gets\infty$ and $\rho_1\gets \rho_v$.
Finally, note that \citet{Bojchevski2019} show the correctness of the policy iteration for arbitrary sets of admissible perturbed graphs \(\mathcal{Q}_\mathcal{F}\), thus the correctness of  \cref{algo:pippnp-PI} follows from their proof. In particular, the additional costs for insertion and deletion does not change the fact that we can model the problem as a Markov decision process.

\textbf{Local and global constraints.}
Lastly, for local and global constraints we can use the same auxiliary graph as introduced by \citet{Bojchevski2019}.
In particular, the additional constraints are only additional enrichments of the Linear Program resulting from the auxiliary graph, yielding a quadratically constrained linear program.
To account for additional costs, we have to replace constraint (4f) of their QCLP with the following constraints that also considers additional costs:
\[
    \sum_{(i,j)\in\mathcal{F}} c_{\bm{A}}^-\cdot[(i,j)\in\mathcal{E}] \beta^0_{i,j} + c_{\bm{A}}^+\cdot[(i,j)\notin{\mathcal{E}}] \beta^1_{i,j} \leq \epsilon
\]

\subsection{Sparsity-aware randomized smoothing for attribute and adjacency perturbations}
\label{appendix:ged_randomized_smoothing}

\citet{Bojchevski2020} present robustness certificates for sparse data based on randomized smoothing. Their main idea is to introduce a sparsity-aware smoothing distribution that preserves the sparsity of the underlying data distribution. Such sparsity-aware certificates for discrete data are currently state-of-the-art for certifying robustness of GNNs against structure and attribute perturbations.

\citet{Bojchevski2020} derive robustness guarantees only under uniform costs ($c_{\bm{A}}^+ = c_{\bm{A}}^- = c_{\bm{X}}^+ = c_{\bm{X}}^- = 1$). Here we make use of our findings (\cref{appendix:equivariance_smoothing}) and generalize their certificate to graph edit distances under arbitrary costs for insertion and deletion.
Specifically for sparse graphs we model adversaries that perturb nodes by adding ones (flip \(0\rightarrow 1\)) to or deleting ones (flip \(1\rightarrow 0\)) of the adjacency or feature matrix. Since adversaries can perturb both the features \(\bm{X}\) and edges \(\bm{A}\) of a graph \(G=(\bm{A},\bm{X})\) we consider the threat model
\(
    \mathcal{B}_{\epsilon}(G) = \{\tilde{G} \mid \delta(G,\tilde{G}) \leq \epsilon\} 
\)
with
\begin{equation}
    \begin{aligned}  
        \delta(G,\tilde{G}) = &
            c_{\bm{A}}^+\sum_i\sum_j \mathbb{I}(\tilde{\bm{A}}_{ij} = \bm{A}_{ij}+1) +
            c_{\bm{A}}^-\sum_i\sum_j \mathbb{I}(\tilde{\bm{A}}_{ij} = \bm{A}_{ij}-1) \\ + &
            c_{\bm{X}}^+\sum_i\sum_j \mathbb{I}(\tilde{\bm{X}}_{ij} = \bm{X}_{ij}+1) +
            c_{\bm{X}}^-\sum_i\sum_j \mathbb{I}(\tilde{\bm{X}}_{ij} = \bm{X}_{ij}-1) 
    \end{aligned}
\end{equation}
where \(\mathbb{I}\) is an indicator function and $c_{\bm{A}}^+, c_{\bm{A}}^-, c_{\bm{X}}^+, c_{\bm{X}}^-$ the corresponding costs for addition and deletion.

\textbf{Sparsity-aware smoothing distribution.}
\citet{Bojchevski2020} propose a family of smoothing distributions that preserves sparsity of the underlying data. Applied to graphs, the distribution of randomly perturbed graphs $(\tilde{\mX}, \tilde{\mA})$ given clean graph $(\mX,\mA)$ is defined by the probability mass function
$Q : \{0,1\}^{N \times D} \times \{0,1\}^{N \times N} \rightarrow [0, 1]$ with
\begin{equation}\label{eq:sparse-smoothing-distribution}
    Q(\tilde{\mX}, \tilde{\mA}) = \prod_{i,j} q_\mX(\tilde{\mX}_{i,j}) \prod_{i,j} q_\mA(\tilde{\mA}_{i,j})
\end{equation}
and elementwise functions $q_\mX, q_\mA : \{0,1\} \rightarrow [0,1]$ with
\begin{align*}
    q_\mX(\tilde{\mX}_{i,j}) & =
    \begin{cases}
        (p_{\bm{X}}^+)^{1-\bm{X}_{ij}}(p_{\bm{X}}^-)^{\bm{X}_{ij}} & \text{if } \tilde{\mX}_{i,j} \neq \mX_{i,j} \\
        1 - (p_{\bm{X}}^+)^{1-\bm{X}_{ij}}(p_{\bm{X}}^-)^{\bm{X}_{ij}}  & \text{otherwise}
    \end{cases}
    \\
    q_\mA(\tilde{\mA}_{i,j}) & =
    \begin{cases}
        (p_{\bm{A}}^+)^{1-\bm{A}_{ij}}(p_{\bm{A}}^-)^{\bm{A}_{ij}} & \text{if } \tilde{\mA}_{i,j} \neq \mA_{i,j} \\
        1 - (p_{\bm{A}}^+)^{1-\bm{A}_{ij}}(p_{\bm{A}}^-)^{\bm{A}_{ij}}  & \text{otherwise}
    \end{cases}
\end{align*}
where parameters $p_{\bm{X}}^+, p_{\bm{X}}^-, p_{\bm{A}}^+, p_{\bm{A}}^- \in [0,1]$ specify the probability of randomly inserting or deleting attributes and edges, respectively.
Note that by using different probabilities to flip \(0 \rightarrow 1\) with probability \(p^+\) and \(1 \rightarrow 0\) with probability \(p^-\), the smoothing distribution allows to preserve sparsity of the data especially for~$p^+ \ll p^-$.

\textbf{Graph edit distance certificates.}
The certificate of \citet{Bojchevski2020} guarantees robustness against the threat model \( \mathcal{B}_{r_{\bm{A}}^+,r_{\bm{A}}^-,r_{\bm{X}}^+,r_{\bm{X}}^-}(G) \), which bounds the perturbations individually by having separate radii for addition $r^+$ and deletion $r^-$.
To certify robustness under graph edit distance with arbitrary costs and given budget \(\epsilon\), we have to certify robustness with respect to all balls \( \mathcal{B}_{r_{\bm{A}}^+,r_{\bm{A}}^-,r_{\bm{X}}^+,r_{\bm{X}}^-}(G) \)~with
\[
    c_{\bm{A}}^+\cdot r_{\bm{A}}^+ + c_{\bm{A}}^-\cdot r_{\bm{A}}^- +
    c_{\bm{X}}^+\cdot r_{\bm{X}}^+ + c_{\bm{X}}^-\cdot r_{\bm{X}}^-
    \leq\epsilon.
\]
Finally note that in practice we do not have to consider all balls since if we can certify one radius the classifier is also robust for smaller radii \citep{Bojchevski2020}. Therefore the number of combinations one has to consider reduces significantly in practice.

%% file: appendix/local_robustness.tex
\section{Local budgets and local robustness}
\label{appendix:local_robustness}

As discussed in~\cref{section:defining_robustness}, 
the domain $\sX$ may be composed of $N$ distinct elements, i.e.\ $\sX = \sA^N$ for some set $\sA$.
Similarly, the task may require making $M$ distinct predictions, i.e.\ the co-domain is $\sY  = \sB^M$ for some set $\sB$. 
In certain tasks, like node classification, it is common to enforce local distance constraints on each of the $N$ elements of a perturbed input $x'$ and investigate the robustness of some subset of prediction elements $\sT \subseteq \{1,\dots,M\}$.
In the following, we discuss how to generalize our definition of robustness (see~\cref{definition:equivariant_robustness}) to enforce such local budget constraints and quantify such local robustness.

For this discussion, recall that a model is $(\sG, \din, \dout, \epsilon, \delta)$-equivariant-robust if
\begin{equation*}
\left(
    \max_{x' \in \sX} \max_{g \in \sG} \dout(f(x), g^{-1} \act f(g \act x')) \text{ s.t. }  \din(x,  x') \leq \epsilon
\right)
\leq \delta.
\end{equation*}

\textbf{Local budgets. }
Local budget constraints can be easily accounted for by replacing the original optimization domain
$\{g \act x' \mid g \in \sG, x' \in \sX, \din(x,x') \leq \epsilon\}$
in the above definition  with
\begin{equation*}
    \left\{g \act x' \mid g \in \sG, x' \in \sA^N, \din(x,x') \leq \epsilon, \forall n : \dlocal(x_n,x'_n) \leq \rho_n \right\},
\end{equation*}
with global distance 
$\din : \sA^N \rightarrow \sR_+$, global budget $\epsilon$, local distance $\dlocal : \sA \rightarrow \sR_+$, and local budgets $\rho_1, \dots, \rho_N \in \sR_+$.

\textbf{Local robustness. }
Quantifying the robustness of some subset of prediction indices $\sT \subseteq \{1,\dots,M\}$ requires a function $\dout : \sB^{|\sT|}  \times \sB^{|\sT|}  \rightarrow \sR_+$.
We need to be careful about where we introduce the indexing to pass the original $M$-dimensional predictions into $\dout$.
It is not correct to measure output distance using $\dout(f(x)_\sT, g^{-1} \act f(g \act x')_\sT)$, i.e.\ it is not correct to index before reverting the effect of group action $g \in \sG$.
Afterall, group $\sG$ may act differently on $\sB^M$ and $\sB^{|T|}$. For example, rotating a point cloud around its center of mass and then subsampling it is not the same as rotating around the center of mass of the subsampled point cloud.
Therefore, $\dout(f(x)_\sT, g^{-1} \act f(g \act x')_\sT)$ may be very large, even if $f$ is perfectly equivariant and not affected by the small perturbation $(x' - x)$.
Instead, we need to measure output distance using
\begin{equation*}
    \dout(f(x)_\sT, (g^{-1} \act f(g \act x'))_\sT).
\end{equation*}

Combining local budgets and local robustness leads to the following definition:
\begin{definition}\label{definition:equivariant_robustness_local}
Consider a ground truth function $y : \sX \rightarrow \sY$ with $\sX = \sA^N$ and $\sY = \sB^N$ for some sets $\sA, \sB$.
Further consider input distance function $\din : \sA^N \times \sA^N \rightarrow \sR_+$, local distance function $\dlocal : \sA \times \sA \rightarrow \sR_+$, a set of output indices $\sT \subseteq \{1,\dots,M\}$ and an output distance function $\dout : \sB^{|T|} \times \sB^{|T|} \rightarrow \sR_+$.
Assume that $y$ is equivariant with respect to the action of group $\sG$.
Then, a prediction $f(x)$ for clean input $x \in \sX$ is $(\sG, \din, \dout, \epsilon, \vrho, \delta, \sT)$-equivariant-robust if
\begin{align*}\label{eq:equivariant_robustness}
    &\max_{x' \in \sX} \max_{g \in \sG} \dout\left(f(x)_\sT, (g^{-1} \act f(g \act x'))_\sT\right) 
    \\ \text{ s.t. } & \din(x,  x') \leq \epsilon , \quad \forall n : \dlocal(x_n,x'_n) \leq \rho_n,
\end{align*}
is less than or equal to $\delta$.
\end{definition}

Similar to what we discussed in~\cref{section:certification}, using an equivariant model $f$ means that
$\dout\left(f(x)_\sT, (g^{-1} \act f(g \act x'))_\sT\right) = \dout\left(f(x)_\sT, f(x')_\sT\right)$ and we recover the traditional notion of adversarial robustness with local constraints for the subset of predictions $\sT$.

%% file: appendix/non_compact_sets.tex
\section{Definition of robustness for non-compact sets}\label{appendix:non_compact}
For our discussion in~\cref{section:defining_robustness}, we assumed compactness of all optimization domains, so that minima and maxima always exist.
In particular, we assumed that the set 
$\{x' \in \sX \mid \din(x,x') \leq \epsilon\}$ is compact for all $\epsilon \in \sR_+$ and thus contains all its limit points.
If this is not the case, the minimum $\min_{g \in \sG} \din(x, g \act x')$ may not exist for certain groups $\sG$ and perturbed inputs $x' \in \sX$.
In this case, the action-induced distance needs to be defined as
$    \dlift(x,x') = \inf_{g \in \sG} \din(x, g \act x')$.

Furthermore, our discussion on perturbation models in~\cref{section:input_distance} may no longer hold.
We may have
$\{x' \mid \dlift(x, x') \leq \epsilon\} \supsetneq \{g \act x' \mid g \in \sG,  \din(x,x') \leq \epsilon\}$, since the l.h.s.\ set also contains perturbed inputs $x' \in \sX$ that can be mapped arbitrarily close to a $\din$-ball of radius $\epsilon$, whereas the r.h.s.\ set only contains perturbed inputs that can be mapped into the interior of the ball (via the inverse group element $g^{-1}$).
To ensure equality, we also need to include these limit points.
This leads us to the following definition:
\begin{definition}\label{definition:equivariant_robustness_compact}
Assume that ground truth function $y : \sX \rightarrow \sY$ is equivariant with respect to the action of group $\sG$.
Further define the set of limit points
\begin{equation*}
    \sL = \{g \act x' \mid \inf_{h \in \sG} \din(x, h \act x') = \epsilon \, \land \, g \in \mathrm{arg\,inf}_{g \in \sG} \din(x, g \act x')\}.
\end{equation*}
Then, a prediction $f(x)$ for clean input $x \in \sX$ is $(\sG, \din, \dout, \epsilon, \delta)$-equivariant-robust if
\begin{equation*}
(
    \max_{x' \in \sX} \max_{g \in \sG} \dout(f(x), g^{-1} \act f(g \act x')) \ \text{ s.t. }  \ \din(x,  x') \leq \epsilon \lor x' \in \sL 
)
\leq \delta.
\end{equation*}\end{definition}
One could also constrain $x'$  to be in the closure of $\{x' \in \sX \mid \din(x,x') \leq \epsilon\}$.
This would however correspond to a potentially stronger notion of robustness, since the closure may also contain limit points that are not in $\sL$. The model may thus need to be robust to a larger set of perturbed inputs.

Note that all our discussions in~\cref{section:certification} also apply to this notion of robustness.

%% file: appendix/non_isometric_actions.tex
\section{Definition of robustness for non-isometric group actions}\label{appendix:non_isometric}
For our discussion in~\cref{section:defining_robustness}, we assumed that group $\sG$ acts isometrically on input space $\sX$, i.e.\ $\forall x, x' \in \sX, \forall g \in \sG: \din(g \act x, g \act x') = \din(x, x')$.
If this is not the case, one could also try to define an action-induced distance as follows:
\begin{equation*}
    \dlift(x,x') = \min_{g,h \in \sG} \din(h \act x, g \act x'),
\end{equation*}
i.e.\ try to minimize the distance between clean input $x$ and perturbed input $x'$ by transforming both via group actions.
However, we evidently have $\min_{g,h \in \sG} \din(h \act x, g \act x') \leq \min_{g \in \sG} \din(x, g \act x')$.
We can thus distinguish two cases

\textbf{Case 1.} Both action-induced distances are identical for all $x, x' \in \sX$ (which is always the case when $\sG$ acts isometrically).
In this case, introducing the second group element is redundant.

\textbf{Case 2.}
There is a pair $x, x' \in \sX$ such that $\min_{g,h \in \sG} \din(h \act x, g \act x') < \min_{g \in \sG} \din(x, g \act x')$.
In this case,  Desideratum $3$ from~\cref{section:defining_robustness} would be violated, i.e. the action-induced distance $\dlift$ would not optimally preserve the original distance $\din$.

Nevertheless, one could conceivably define robustness for equivariant tasks with non-isometric group actions as follows:
\begin{equation*}
(
    \max_{x' \in \sX} \max_{g,h \in \sG} \dout(h^{-1} \act f(h \act x), g^{-1} \act f(g \act x')) \ \text{ s.t. }  \ \din(x,  x') \leq \epsilon
)
\leq \delta.
\end{equation*}
All our discussions in~\cref{section:certification} also apply to this notion of robustness.

However, this notion would be qualitatively different from~\cref{definition:equivariant_robustness}:
Rather than requiring robustness robustness for a specific clean prediction $f(x)$, one would require robustness for the entire set of predictions $\{f(h \act x) \mid h \in \sG\}$.
As such, it is arguably not as good of a generalization of the classic notion of adversarial robustness.
If one opts to use this notion of robustness for certification, one should at least state it explicitly.
Afterall, being explicit about the underlying semantics of tasks and certificates is a main focus of this work.

Note that typically considered group actions (translation, rotation, reflection, permutation, etc.) act isometrically on the typically considered input spaces (e.g.\ Euclidean spaces), meaning these considerations should usually not be relevant in practice.

%% file: appendix/transformation_specific.tex
\section{Relation to transformation-specific robustness}\label{appendix:transformation_specific}
As discussed in~\cref{section:related_work}, works on transformation-specific robustness are concerned with robustness to  unnoticeable parametric transformations, such as small rotations or translations.
More formally, consider a parametric function $\psi_\theta : \sX \rightarrow \sX$ with parameter $\theta \in \Theta$.
A model is considered robust to transformation-specific attacks if
$\max_{\theta \in \Theta} \dout(f(x), f(\psi_\theta(x)) \leq \delta$ for some small $\delta \in \sR_+$.
Usually, $\dout$ is defined as the $0$--$1$-loss $\indicator[y \neq y']$~\cite{Kanbak2018,Engstrom2019,Balunovic2019,Zhao2020iso,Fischer2020,Ruoss2021,Mohapatra2021,Li2021,Alfarra2022,Murarev2022}.
The underlying assumption is that the transformations $\psi_\theta$ do not change the ground truth label of an input.

The key differences to our proposed notion of robustness are that
(1) transformation-specific robustness does not consider that there are transformations for which predictions explicitly need to change and 
(2) works on transformation-specific usually do not consider unstructured perturbations like camera noise which are only constrained by a distance function $\din$.

In the following, we demonstrate that transformation-specific robustness can nevertheless be framed as a special case of our proposed notion of robustness --- or relaxations thereof.
We can distinguish three cases, depending on the structure of $\{\psi_\theta \mid \theta \in \Theta\}$.

\textbf{Case 1.}
In the first case, the family of transformations $\{\psi_\theta \mid \theta \in \Theta\}$ forms a group $\sG$ with composition as the group operator ($\psi_\theta \compose \psi_{\theta'} = \psi_\theta \cat \psi_{\theta'}$) and application of the transformation as the group action ($\psi_\theta \act x = \psi_\theta(x)$).\footnote{Such a family of transformations is referred to as ``resolvable'' in transformation-specific robustness literature.}
In this case, the above definition of transformation-specific robustness can be reformulated as $\max_{g \in \sG} \dout(f(x), f(g \act x)) \leq \delta$.
Assuming that our distance function fulfills $\din(x,x') = 0 \iff x = x'$, this is equivalent to 
\begin{equation}\label{eq:transformation_specific_1}
    \max_{x' \in \sX} \max_{g \in \sG} \dout(f(x), f(g \act x')) \leq \delta \quad \text{ s.t. } \din(x,x') \leq 0.
\end{equation}
We observe that this is a special of our notion of robustness (see~\cref{definition:equivariant_robustness}), where the adversary has a budget of $\epsilon=0$ (i.e. can only apply group actions) and the task is group invariant.

\textbf{Case 2.}
In the second case, the set of transformations is restricted to a subset $\sH$ of $\sG$ that is not a proper group.
For example, one may restrict an adversary to small translations instead of arbitrary translations (see, e.g., \cite{Engstrom2019}).
In this case, transformation-specific robustness can be framed as
\begin{equation*}
    \max_{x' \in \sX} \max_{g \in \sH} \dout(f(x), f(g \act x')) \leq \delta \quad \text{ s.t. } \din(x,x') \leq 0,
\end{equation*}
which is a relaxation of~\cref{eq:transformation_specific_1} since $\sH \subset \sG$.

\textbf{Case 3.}
In the third case, the family $\{\psi_\theta \mid \theta \in \Theta\}$ does not have a group structure or $\psi_\theta(x)$ is not a proper group action (e.g.\ due to interpolation artifacts after rotation of an image).
In this case, we can choose an arbitrary $\epsilon \in \sR$ and define an arbitrary $\din(x,x')$ such that $\{x' \mid \din(x,x') \leq \epsilon\} = \{\psi_\theta(x) \mid \theta \in \Theta\}$. That is, the distance between $x$ and $x'$ is smaller than $\epsilon$ if $x'$ can be obtained via a transformation of $x$.
With this choice of $\din$, transformation-specific robustness is equivalent to
\begin{equation*}
        \max_{x' \in \sX} \dout(f(x), f(x')) \leq \delta \quad \text{ s.t. } \din(x,x') \leq \epsilon.
\end{equation*}
This is an instance of classic adversarial robustness, which is a special case of our proposed notion of robustness (no equivariance).